\documentclass[12pt]{article}
\usepackage[utf8]{inputenc}
\usepackage{amsmath}
\usepackage{amssymb}
\usepackage{natbib}

\usepackage{fancyhdr}
\usepackage{longtable}
\fancyheadoffset{0pt} 
\fancyfootoffset{0pt} 

\usepackage[table,xcdraw, dvipsnames]{xcolor}

\usepackage[font=small,labelfont=bf,labelsep=space]{caption}
\usepackage{xspace}
\usepackage[pdftex]{graphicx}
\usepackage[hidelinks]{hyperref}
\usepackage[capitalize]{cleveref}

\hypersetup{
    colorlinks = true,
    linkcolor=ACMDarkBlue,
    citecolor=ACMDarkBlue,
    urlcolor=ACMDarkBlue
}
\definecolor[named]{ACMDarkBlue}{cmyk}{1,0.58,0,0.21}

\usepackage{authblk}
\usepackage{geometry}
\usepackage{parskip} 
\usepackage[inline]{enumitem}

\usepackage{booktabs}  
\usepackage{amsfonts}
\usepackage{mathtools}
\usepackage{url}
\usepackage{enumitem}
\usepackage{comment}
\usepackage{mdframed}
\usepackage{wrapfig}
\usepackage{floatrow}
\usepackage[nottoc]{tocbibind}

\definecolor{quotemark}{gray}{0.7}
\makeatletter
\def\fquote{%
    \@ifnextchar[{\fquote@i}{\fquote@i[]}
           }%
\def\fquote@i[#1]{%
    \def\tempa{#1}%
    \@ifnextchar[{\fquote@ii}{\fquote@ii[]}
                 }%
\def\fquote@ii[#1]{%
    \def\tempb{#1}%
    \@ifnextchar[{\fquote@iii}{\fquote@iii[]}
                      }%
\def\fquote@iii[#1]{%
    \def\tempc{#1}%
    \vspace{1em}%
    \noindent%
    \begin{list}{}{%
         \setlength{\leftmargin}{0.1\textwidth}%
         \setlength{\rightmargin}{0.1\textwidth}%
                  }%
         \item[]%
         \begin{picture}(0,0)%
         \put(-15,-5){\makebox(0,0){\scalebox{3}{\textcolor{quotemark}{``}}}}%
         \end{picture}%
         \begingroup\itshape}%
 \def\endfquote{%
 \endgroup\par%
 \makebox[0pt][l]{%
 \hspace{0.8\textwidth}%
 \begin{picture}(0,0)(0,0)%
 \put(15,15){\makebox(0,0){%
 \scalebox{3}{\color{quotemark}''}}}%
 \end{picture}}%
 \ifx\tempa\empty%
 \else%
    \ifx\tempc\empty%
       \hfill\rule{100pt}{0.5pt}\\\mbox{}\hfill\tempa,\ \emph{\tempb}%
   \else%
       \hfill\rule{100pt}{0.5pt}\\\mbox{}\hfill\tempa,\ \emph{\tempb},\ \tempc%
   \fi\fi\par%
   \vspace{0.5em}%
 \end{list}%
 }%
 \makeatother
 
\usepackage{ntheorem}

\theoremseparator{:} 

\newmdtheoremenv[%
  backgroundcolor=white,
  linecolor=blue!60!black,
  linewidth=2pt,
  topline=true,
  rightline=false,
  skipabove=10pt,
  skipbelow=10pt,
  leftline=false]{ourexample}{Application}
  
\newmdtheoremenv[%
  backgroundcolor=gray!20,
  linecolor=red!60!black,
  linewidth=2pt,
  topline=false,
  rightline=false,
  skipabove=10pt,
  skipbelow=10pt,
  leftline=false]{ourbox}{Formulation}

\newmdtheoremenv[%
  backgroundcolor=gray!20,
  linecolor=red!60!black,
  linewidth=2pt,
  topline=false,
  rightline=false,
  skipabove=10pt,
  skipbelow=10pt,
  leftline=false]{regbox}{Box}

\theoremstyle{nonumberplain}
\newmdtheoremenv[%
  backgroundcolor=gray!20,
  linecolor=red!60!black,
  linewidth=2pt,
  topline=false,
  rightline=false,
  skipabove=10pt,
  skipbelow=10pt,
  leftline=false]{suppregbox}{Box S1}

\definecolor{Gray1}{gray}{0.82}
\definecolor{Gray2}{gray}{0.92}

\pdfmapfile{=Cochineal.map}
\usepackage{cochineal}
\usepackage[T1]{fontenc}

\usepackage[scale=.95,type1]{cabin}
\usepackage[zerostyle=c,scaled=.94]{newtxtt}
\usepackage[cal=boondoxo]{mathalfa}
\usepackage{microtype}

\usepackage{etoolbox}
\apptocmd{\thebibliography}{\raggedright}{}{}

\newtheorem{dfn}{Definition}

\newcommand{\baichuanonec}{Baichuan1-chat (13B)}
\newcommand{\baichuatwos}{Baichuan2-chat (7B)}
\newcommand{\baichuatwot}{Baichuan2-chat (13B)}
\newcommand{\qws}{Qwen (7B)}
\newcommand{\qwf}{Qwen (14B)}
\newcommand{\internseven}{InternLM-chat (7B)}
\newcommand{\internt}{InternLM-chat (20B)}
\newcommand{\llamaseven}{Llama2 (7B)}
\newcommand{\llama}{Llama2 (13B)}
\newcommand{\llamaseventy}{Llama2 (70B)}
\newcommand{\llamachatseventy}{Llama2-chat (70B)}
\newcommand{\kl}{Koala (13B)}  
\newcommand{\wc}{Wizardcoder (15B)}
\newcommand{\vicuna}{Vicuna-v1.3 (33B)}

\newcommand{\davinci}{davinci (175B)}

\newcommand{\textdavincitwo}{text-davinci-002}
\newcommand{\textdavincithree}{text-davinci-003}
\newcommand{\chatgpt}{GPT-3.5-Turbo}
\newcommand{\gptf}{GPT-4}
\newcommand{\claude}{Claude2}

\newcommand{\mcot}{manual CoT}

\newcommand{\oicl}{1-shot IcL}
\newcommand{\ticl}{3-shot IcL}
\newcommand{\dt}{adversarial doubt}
\newcommand{\ig}{adversarial ignore}

\usepackage{multirow}
\usepackage{tablefootnote}
\usepackage{algpseudocode}
\usepackage{algorithm}
\usepackage[pdftex]{graphicx}
\usepackage{makecell}
\usepackage{tabularx}
\usepackage{subfigure}  
\usepackage{graphicx}
\usepackage{enumitem}
\usepackage{CJKutf8}
\CJKtilde

\usepackage{bera}
\usepackage{listings}
\usepackage{xcolor}
\definecolor{eclipseStrings}{RGB}{42,0.0,255}
\lstdefinelanguage{json}{
    basicstyle=\scriptsize\ttfamily,
    commentstyle=\color{eclipseStrings},
    showstringspaces=false,
    breaklines=true,
    frame=single,
    rulecolor=\color{black},
    string=[s]{"}{"},
    comment=[l]{:\ "},
    morecomment=[l]{:"},
    literate=
        *{0}{{{\color{numb}0}}}{1}
         {1}{{{\color{numb}1}}}{1}
         {2}{{{\color{numb}2}}}{1}
         {3}{{{\color{numb}3}}}{1}
         {4}{{{\color{numb}4}}}{1}
         {5}{{{\color{numb}5}}}{1}
         {6}{{{\color{numb}6}}}{1}
         {7}{{{\color{numb}7}}}{1}
         {8}{{{\color{numb}8}}}{1}
         {9}{{{\color{numb}9}}}{1}
}

\numberwithin{figure}{section}
\numberwithin{table}{section}

\newcommand{\barchart}[1]{%
  \begin{tikzpicture}[baseline=0ex]
    \draw[fill=teal!40] (0,0) rectangle (#1/75,0.25); 
    \node[anchor=east, font=\fontsize{10}{12}\selectfont] at (1.8,0.12) {\footnotesize #1};      
  \end{tikzpicture}%
}

\usepackage{pgfplotstable} 
\usepackage{tikz}    
\usepackage{placeins}

\title{\bf Causal Evaluation of Language Models}

\author[$*$2,$\S$]{Sirui Chen}
\author[$*$3,1]{Bo Peng}
\author[4,$\S$]{Meiqi Chen}
\author[3,$\S$]{Ruiqi Wang}
\author[5]{Mengying Xu}
\author[5]{\\Xingyu Zeng}
\author[5]{Rui Zhao}
\author[2]{Shengjie Zhao}
\author[1]{Yu Qiao}
\author[$\ddag$1]{Chaochao Lu}

\affil[1]{Shanghai AI Laboratory \hspace{0.3em} \textsuperscript{2}Tongji University} 
\affil[3]{Shanghai Jiao Tong University \hspace{0.3em} \textsuperscript{4}Peking University \hspace{0.3em} \textsuperscript{5}SenseTime Group}

\date{}
\begin{document}

\maketitle

\renewcommand{\thefootnote}{}
\footnotetext{$^*$Equal contribution, $^{\S}$Work done at Shanghai AI Laboratory, $^{\ddag}$Corresponding author: {\footnotesize \texttt{causalai@pjlab.org.cn}}.}

\renewcommand{\thefootnote}{\arabic{footnote}}
\setcounter{footnote}{0}

\begin{abstract}
\noindent Causal reasoning, fundamental to human cognition and scientific understanding, is viewed as crucial for achieving human-level machine intelligence and fostering the development of an ``artificial scientist'' posited by Pearl. Recent advances in language models have expanded the horizons of artificial intelligence across various domains, sparking inquiries into their potential for causal reasoning.
In this work, we introduce \textit{\textbf{Ca}usal evaluation of \textbf{L}anguage \textbf{M}odels} (\textbf{CaLM}), which, to the best of our knowledge, is the first comprehensive benchmark for evaluating the causal reasoning capabilities of language models. 
First, we propose the CaLM framework, which establishes a foundational taxonomy consisting of four modules: causal target (i.e., what to evaluate), adaptation (i.e., how to obtain the results), metric (i.e., how to measure the results), and error (i.e., how to analyze the bad results). This taxonomy defines a broad evaluation design space while systematically selecting criteria and priorities. 
Second, we compose the CaLM dataset, comprising 126,334 data samples, to provide curated sets of causal targets, adaptations, metrics, and errors, offering extensive coverage for diverse research pursuits.
Third, we conduct an extensive evaluation of 28 leading language models on a core set of 92 causal targets, 9 adaptations, 7 metrics, and 12 error types. Note that, the selected 92 causal targets cover 46 causal tasks, span three text modes (i.e., Natural, Symbolic, and Mathematical), and involve two languages (i.e., English and Chinese). Before implementing CaLM, causal evaluations of language models were conducted, on average, in merely 10\% of these causal tasks, typically using just a single adaptation (e.g., basic prompting) and a single metric (e.g., accuracy). Moreover, previous causal evaluations not only overlooked the Mathematical text mode but also excluded assessments in Chinese, and lacked a systematic categorization of error types for in-depth analysis.
In contrast, our evaluation extends to a wide spectrum of causal tasks, metrics, and error analysis, significantly enriching the depth and breadth of causal evaluations.
Fourth, we deeply analyze the causal evaluation results on two levels. At a broad level, we assess the influence of diverse dimensions (e.g., adaptation) and critical factors (e.g., scale) on overall model performance, and investigate the intra- and inter-dimensional relationships that shape causal reasoning efficacy. At a granular level, we provide an in-depth analysis of each specific adaptation, model, and causal scenario. 
Fifth, we present 50 high-level empirical findings across 9 dimensions (e.g., model, adaptation, error), providing valuable guidance for future language model development and analysis.  
Finally, we develop a multifaceted platform and codebase, including a website, leaderboards, datasets, and toolkits, to support scalable and adaptable assessments. We envision CaLM as an ever-evolving benchmark for the community, systematically updated with new causal targets, adaptations, models, metrics, and error types to reflect ongoing research advancements. Project website is at {\footnotesize \url{https://opencausalab.github.io/CaLM}}.
\end{abstract}

\thispagestyle{empty}
\pagenumbering{Roman}

\cleardoublepage
\tableofcontents			
\cleardoublepage
\listoffigures	
\cleardoublepage
\listoftables
\cleardoublepage
\pagenumbering{arabic}



\section{Introduction}

\begin{fquote}[Confucius][The Analects][551-479 BCE]
To know what you know and know what you do not know - this then is wisdom.\footnote{Translated by \citet{ames1999analects}.}
\end{fquote}

Causal reasoning is a vital element of human cognition \citep{waldmann2017oxford}, and is widely thought of as an indispensable step towards achieving machine intelligence at a human level \citep{pearl2019seven}. In fact, causal reasoning is a cornerstone of scientific understanding. It enables scientists to explain, predict, and control natural phenomena, test hypotheses, build models, and make informed decisions. Without the ability to reason causally, scientific progress would be severely hindered, and our understanding of the world around us would remain limited. More importantly, upon comprehending the underlying principles governing causal reasoning, it becomes feasible to simulate this cognitive process within contemporary computer systems, thus enabling the development of an ``artificial scientist'' \citep{pearl2018book}. This ``Causal Revolution'' \citep{pearl2018book} in artificial intelligence is expected to have a profound impact on a wide range of fields and industries. 

Many believed that we were far from realizing this blueprint before the advent of large language models (LLMs). However, recent advancements in LLMs have significantly pushed the boundaries of AI on a wide range of domains and causal tasks, including natural language comprehension \citep{ouyang2022training,chatgpt2022,openai2023gpt4,touvron2023llama}, programming \citep{chen2021evaluating,li2022competition,roziere2023code,tufano2024autodev}, and mathematical reasoning \citep{imani2023mathprompter,romera2024mathematical,ahn2024large,trinh2024solving}. \citet{bubeck2023sparks} even believed that an early version of GPT-4 ``\textit{could  reasonably be viewed as an early (yet still incomplete) version of an artificial general intelligence (AGI) system}''. The various emergent abilities \citep{wei2022emergent} of LLMs lead us to wonder whether or not we are approaching such an artificial scientist capable of causal reasoning. This curiosity instinctively gives rise to several fundamental questions: a) How can we ascertain if LLMs possess the capacity for causal reasoning? b) How can we gauge the degree of causal reasoning proficiency in LLMs? c) How can we enhance the causal reasoning aptitude of LLMs? All three of the ``How'' inquiries necessitate a comprehensive benchmarking of LLMs concerning their causal reasoning capabilities. 

Although a few efforts have been made in this direction \citep{hobbhahn2022investigating,willig2022can,long2022can,tu2023causal,jin2023cladder,jin2023large,kiciman2023causal,zhang2023understanding,zhang2023causality,zevcevic2023causal,gao2023chatgpt,lu2024gpt}, these endeavors assess only a limited selection of language models for a narrow range of causal tasks. Typically, these studies employ only a single adaptation (e.g., basic prompting) and rely solely on a single metric (e.g., accuracy) for assessment. This results in an incomplete grasp of the models' abilities in causal reasoning. Moreover, prior evaluations have not only neglected the exploration of causal assessments in Chinese, but also failed to implement a systematic categorization of error types for in-depth analysis. In addition, there is an absence of a publicly accessible platform to facilitate wider engagement with these findings in the community. 

In this work, we introduce \textit{\textbf{Ca}usal evaluation of \textbf{L}anguage \textbf{M}odels} (\textbf{CaLM}), which, to the best of our knowledge, is the first comprehensive benchmark for evaluating the causal reasoning capabilities of language models. 
To be specific, 
(1) we propose the CaLM framework, establishing a foundational taxonomy consisting of four modules: \emph{causal target} (what to evaluate), \emph{adaptation} (how to obtain the results), \emph{metric} (how to measure the results), and \emph{error} (how to analyze the bad results), as shown in Figure \ref{fig_intro:suite}. This taxonomy defines a broad, if not complete, design space for evaluation while systematically selecting criteria and outlining priorities and constraints. 
(2) We construct the CaLM dataset, featuring 126,334 data samples, to provide curated sets of causal targets, along with corresponding adaptations, metrics, and errors. It offers extensive coverage and practicality for diverse research endeavors.
(3) We provide a comprehensive evaluation of 28 prominent language models on a core set of 92 causal targets, 9 adaptations, 7 metrics, and 12 types of errors. The selected 92 causal targets span 46 causal tasks across three text modes (i.e., Natural, Symbolic, and Mathematical) and two languages (i.e., English and Chinese). Our evaluation substantially broadens the scope beyond previous work, and greatly enhances our understanding of the causal reasoning capabilities of language models. 
(4) We conduct a deep analysis of the evaluation results on two levels. At a broad level, we assess the impact of diverse dimensions (e.g., adaptation) and critical factors (e.g., scale) on overall model performance, while examining the intra- and inter-dimensional relationships that influence causal reasoning efficacy. At a granular level, we offer a detailed analysis of each specific model, adaptation, and causal task.
(5) Our extensive evaluation yields 50 empirical findings across 9 dimensions (e.g., model, scenario, metric), providing valuable guidance for future language model development and further analysis.
(6) We develop a multifaceted platform and codebase, including a website, leaderboards, curated datasets, and toolkits, to facilitate consistent and scalable assessments that can adapt to evolving research needs.

The rest of this section is organized as follows. 
We begin in \nameref{intro:framework} (\cref{intro:framework}) by formally introducing the CaLM framework and its constituent modules, namely \emph{causal target}, \emph{adaptation}, \emph{metric}, and \emph{error}, followed by highlighting its key features and broader considerations inherent within this framework. 
In \nameref{intro:findings} (\cref{intro:findings}), we outline 50 empirical findings derived from various aspects, including the model, adaptation, causal ladder, domain, mode, language, metric, error, and causal scenario. These findings are presented systematically, indicating the depth and breadth of analysis conducted within the study. 
\nameref{contributions} (\cref{contributions}) summarizes the contributions made in this work, and \nameref{organization} (\cref{organization}) concludes by providing an outline of the rest of this paper for the reader's guidance.

\subsection{The CaLM Framework} 
\label{intro:framework}

\begin{figure}[t]
    \centering
\includegraphics[width=0.8\textwidth]{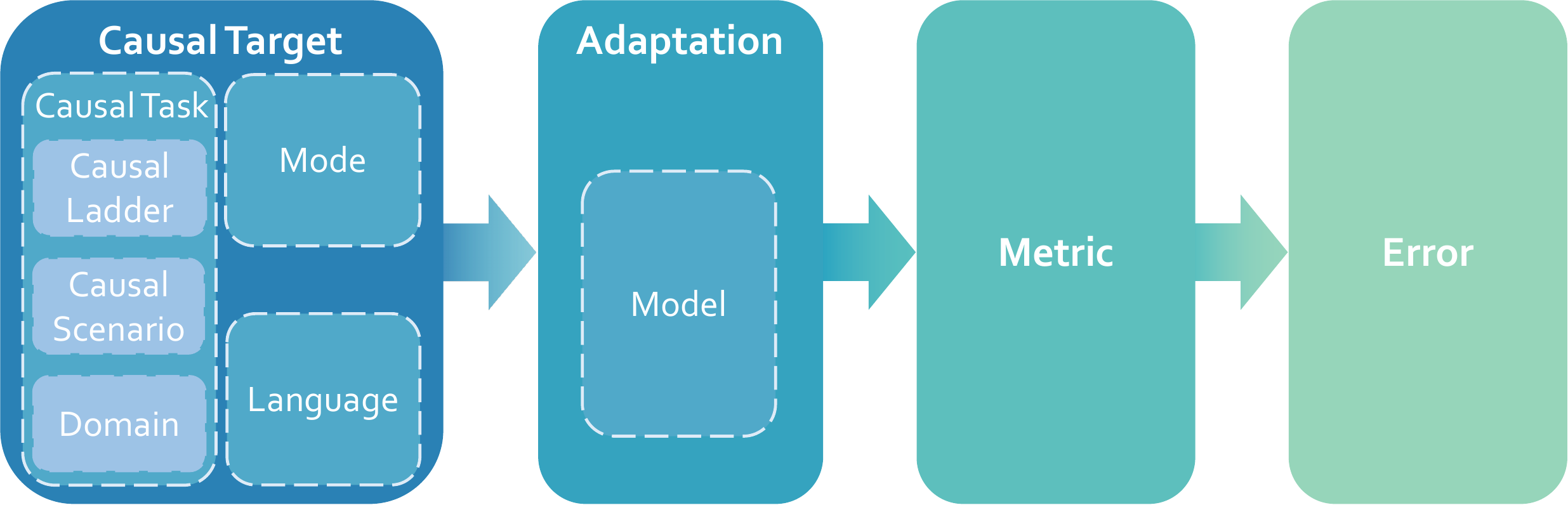}
    \caption[The CaLM framework]{\textbf{The CaLM framework.} CaLM is composed of four modules: \emph{causal target} (what to evaluate), \emph{adaptation} (how to obtain the results), \emph{metric} (how to measure the results), and \emph{error} (how to analyze the bad results). Broadly speaking, it defines an expansive design space essential for assessing the causal reasoning capability of language models. In terms of concrete implementation, we assess 92 causal targets, employing 9 adaptations, 7 metrics, and cataloging 12 types of errors.}
    \label{fig_intro:suite}
\end{figure}

Figure \ref{fig_intro:suite} presents the CaLM framework, which consists of four core modules: \emph{causal target}, \emph{adaptation}, \emph{metric}, and \emph{error}. 
These modules collectively forge a comprehensive structure that facilitates the systematic evaluation of language models. 
The depicted arrows represent the model evaluation pipeline, indicating the sequential process each evaluation undergoes. This involves specifying a \emph{causal target} for the language model, incorporating an \emph{adaptation} process within the model, employing one or more \emph{metrics} for evaluation, and identifying one or more \emph{errors}. These modules serve to respectively answer four fundamental inquiries: (i) the specific causal reasoning capabilities sought, (ii) the methodology for adapting a model to achieve these capabilities, (iii) the effectiveness of the results obtained, and (iv) the nature and scope of the errors identified during the evaluation process.

Generally speaking, our CaLM framework is structured on two levels. 
(1) \textbf{Broad vision}: 
We formulate an abstract taxonomy consisting of four modules (i.e., \emph{causal target}, \emph{adaptation}, \emph{metric}, and \emph{error}) to define the extensive, if not entire, design space for assessing the causal reasoning abilities of language models. 
This taxonomy facilitates a systematic selection within this space, thereby making explicit our benchmark design priorities and the existing limitations thereof.  
(2) \textbf{Concrete implementation}: 
Based on the taxonomy, we select and implement a core set of 92 causal targets, 9 adaptations, 7 metrics, and 12 errors. This selection is with an emphasis on comprehensive coverage (e.g., diverse prompt types), significance (e.g., causal scenarios essential to essential decision-making processes), and practicality (e.g., limited computational resources).

\subsubsection{Causal Target}
A causal target specifies the objective that a model aims to achieve in assessing its causal reasoning capabilities, encapsulated by a defining triplet: (\emph{causal task}, \emph{mode}, \emph{language}). 
In essence, it outlines the particular causal task a model is expected to undertake, the designated mode for performing this task, and the specific language to be used.
This triad of elements constitutes a comprehensive testbed for evaluating language models, presenting unparalleled challenges. In our implementation, the core set of causal targets encompass 46 causal tasks, three text modes, and two languages, collectively yielding 92 distinct causal targets. 

\begin{figure}[t]
    \centering
\includegraphics[width=\textwidth]{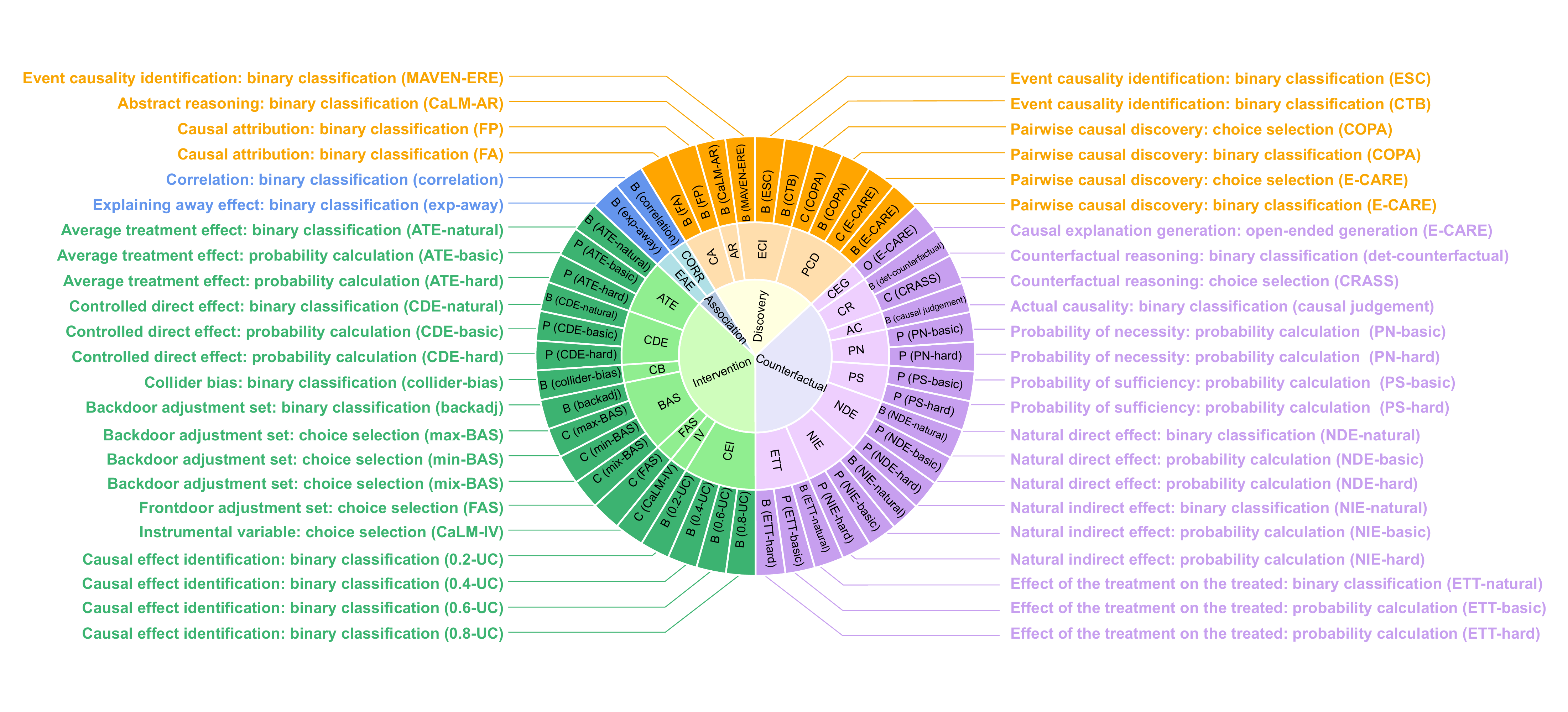}
    \caption[Causal tasks]{\textbf{Causal tasks.} The diagram presents a hierarchical structure with three different layers. The innermost layer consists of four levels of the causal ladder (i.e., causal discovery, association, intervention, and counterfactuals). The second layer consists of 21 causal scenarios. And the outermost layer categorizes 46 tasks (where B represents binary classification, C represents choice selection, P represents probability calculation, and O represents open-ended generation). We take into account both English and Chinese versions of the 46 tasks, with the illustration displaying the English version.}
    \label{fig_intro:causal_task}
\end{figure}

\paragraph{Causal task.}
A causal task defines the specific duty of causal reasoning that a language model needs to accomplish. It is also structured as a triplet: (\emph{causal ladder}, \emph{causal scenario}, \emph{domain}), with the relationships among these three elements illustrated in Figure \ref{fig_intro:causal_task}. \emph{Causal ladder}, often referred to as the \emph{Ladder of Causation}, is a conceptual framework developed by \citet{pearl2018book} to illustrate the hierarchy of causal reasoning tasks \citep{bareinboim2022pearl}. This ladder consists of three distinct levels: \emph{association (Rung 1)}, \emph{intervention (Rung 2)}, and \emph{counterfactuals (Rung 3)}, each representing a progressively deeper level of causal understanding. In addition, we incorporate causal discovery \citep{spirtes2000causation,peters2017elements} into this ladder, recognizing them as a fundamental phase in causal reasoning \citep{glymour2019review}. For clarity and ease of reference in future discussions, we categorize \emph{(causal) discovery} as \emph{Rung 0} of the causal ladder within our CaLM framework. 
\emph{Causal scenario} depicts potential applications of causal concepts in practical or research contexts (e.g., average treatment effect (ATE), probability of sufficiency (PS)), each belonging to only one of the four rungs in the causal ladder. \emph{Domain} specifies the exact context in which a causal scenario is implemented. It could include, for instance, the application of distinct datasets or the exploration of varied question types within a singular dataset (i.e., utilizing the same dataset for tasks such as multiple choice, binary classification, or content generation). This highlights the inherent versatility of domains, underscoring their ability to accommodate a wide array of analytical and procedural tasks. In our implementation, causal tasks span all 4 rungs of the causal ladder (i.e., causal discovery, association, intervention, and counterfactuals), 21 causal scenarios (e.g., pairwise causal discovery, correlation, backdoor adjustment set, counterfactual reasoning), and 46 domains (i.e., different datasets and/or varied question types). 

\paragraph{Mode.}
Mode signifies the different formats in which information can be stored and displayed. Evaluating a model's causal reasoning ability across multiple modes is crucial for confirming its adaptability. Each mode presents unique challenges to the model’s ability to process information. For instance, in the text mode, the focus is on handling linguistic structures and meanings. In the image mode, the emphasis shifts to deciphering visual components and spatial relationships. The use of various modes aids in enhancing our understanding and improvement of the model’s capability to handle complex situations. Moreover, it promotes model's application in more complex and realistic causal scenarios. The broad categories of modes include text, image, video, and code \citep{lu2024gpt}, each of which could be further divided into more specific subcategories. Remarkably, given this benchmark's focus on language models, we specifically identifies three unique subcategories within the text mode: \emph{Natural}, \emph{Symbolic}, \emph{Mathematical}. \emph{Natural} is the most prevalent approach for interacting with language models. It focuses on assessing their abilities in language understanding and causal reasoning. \emph{Symbolic} conveys the information represented in Symbolic forms, closely aligning with traditional cognitive reasoning \citep{garcez2008neural} and minimizing the influence of training data. \emph{Mathematical} presents problems in mathematical terms, examining the model's capacity for logical structure and conceptual comprehension \citep{cobbe2021training}. The three text modes emphasize different aspects, together thoroughly evaluating the model's ability in causal reasoning.

\paragraph{Language.}
Globally, billions of individuals utilize thousands of distinct languages for communication \citep{nordhoff2011glottolog}. Therefore, evaluating the causal reasoning abilities of language models across diverse languages is vital for ensuring their global applicability and inclusivity. Such evaluations take into account the unique cultural contexts, linguistic diversities, and nuances embedded within different languages, providing a thorough assessment of a model's ability to generalize causal reasoning capabilities across the linguistic spectrum. Furthermore, it is instrumental in identifying and quantifying the influence of language-specific biases on the causal reasoning performance of these models. In our implementation, we concentrate on \emph{English} and \emph{Chinese}, reflecting the predominant focus within the realm of language models and natural language processing on these two languages exclusively \citep{liang2022holistic}. 

\subsubsection{Adaptation}
Building on the work of \citet{bommasani2021opportunities} and \citet{liang2022holistic}, \emph{adaptation} refers to the process by which a language model, supplemented with additional data, is transformed into \emph{an adapted model} capable of making predictions on new instances. This process can be primarily categorized into three types: \emph{prompting}, \emph{lightweight-finetuning}, and \emph{finetuning}. They are distinguished based on their method of adaptation: either by priming the model with new data incorporated as a prompt in its input or by utilizing new data to update some or all of the model's parameters. To assess the causal reasoning abilities of language models, it is essential to specify an adaptation method that enables to apply the general-purpose model to a given causal target. In this work, we focus on \emph{prompting}, as it represents the most intuitive method for employing language models in causal reasoning tasks. Specifically, our implementation explores nine distinct prompting strategies (e.g., Chain-of-Thought (CoT) \citep{wei2023CoT}, In-context Learning (IcL) \citep{brown2020language}, Explicit Function (EF)).

\subsubsection{Metric}
\emph{Metric} provides a systematic way to quantify a model's performance across various dimensions of causal reasoning abilities. Typically, \emph{accuracy} is the most universally recognized metric. Additionally, other metrics such as robustness, toxicity, and fairness are also widely used to cater to diverse evaluation needs. 
We implement a set of seven metrics, which are categorized by model, prompt, and causal scenario. 
Specifically, we measure model performance using three metrics: \emph{accuracy}, \emph{robustness}, and \emph{model volatility}. \emph{Accuracy} assesses the precision of responses, \emph{robustness} examines the consistency of these responses under adversarial prompt disturbance, and \emph{model volatility} explores sensitivity to different prompts. 
For causal scenarios, we apply three metrics: \emph{understandability}, \emph{open-limited gap}, and \emph{solvability}. \emph{Understandability} evaluates the ease with which a model interprets a scenario, \emph{open-limited gap} measures performance differences between open-access and limited-access models within the top five of each scenario, and \emph{solvability} examines the model's ability to identify solutions within a causal scenario. 
Lastly, for prompts, \emph{prompt volatility} is used to gauge the variability in model performance when comparing a specific prompt to a basic prompt. This metric serves as an indicator of the prompt's effectiveness.

\subsubsection{Error}
\emph{Error} indicates the discrepancies or shortcomings observed in a model's performance during its assessment in causal reasoning tasks. Uncovering and monitoring these errors is crucial, as it aids researchers and practitioners in pinpointing the model's deficiencies, thereby guiding directions for future improvement. In this study, we document errors both \emph{quantitatively} and \emph{qualitatively}, categorizing them into 12 distinct types. The \emph{Quantitative} errors are divided into five categories: \emph{same response to all questions}, \emph{empty response}, \emph{limitation of instruction-following}, \emph{repetition} and \emph{language inconsistency}. For \emph{qualitative} errors, we identify seven types: \emph{causal hallucination}, \emph{inferential ambiguity}, \emph{calculation error}, \emph{incorrect direction}, \emph{misunderstanding}, \emph{contradiction} and \emph{outlier}. 
In terms of \emph{quantitative} errors, \emph{same response to all questions} refers to instances where the model produces identical replies across different questions within a task. 
\emph{Empty response} denotes situations where the model provides no response to some questions. 
\emph{Limitation of instruction-following} describes the model's inability to respond according to the prescribed format. 
\emph{Repetition} indicates errors involving the model's repetitive generation of questions. 
\emph{Language inconsistency} occurs when the model responds in a language different from the question's language.
Turning to \emph{qualitative} errors, \emph{causal hallucination} involving the model confusing correlation for causation, leading to incorrect causal assertions. \emph{Inferential ambiguity} is observed when the model's response is overly broad or vague, making it difficult to determine its intent. \emph{Calculation error} describes incorrect results from proper mathematical procedures. \emph{Incorrect direction} highlights flawed reasoning within the model' chain of thought, resulting in erroneous conclusions. \emph{Misunderstanding} occurs when the model misinterprets the problem. \emph{Contradiction} arises from the model providing conflicting responses, such as saying both ``yes'' and ``no'' to the same query. \emph{Outlier} refers to responses that are completely unrelated to the posed question.
This classification facilitates a thorough understanding of the model's limitations and informs targeted improvements.

\begin{figure}[t!]
\centering
\subfigure[Previous work]{
\centering
\includegraphics[width=\linewidth]{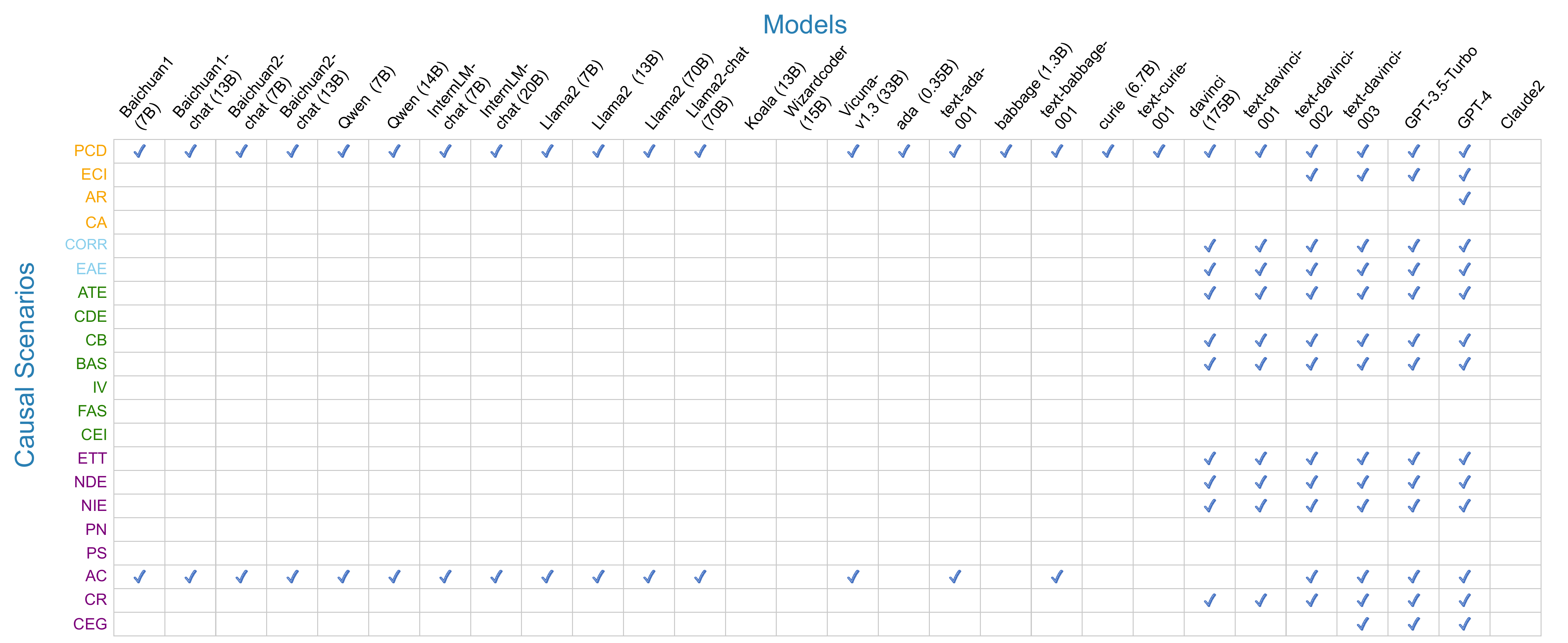}
\label{fig_intro:scenario_model_prev}
}
\subfigure[CaLM]{
\centering
\includegraphics[width=\linewidth]{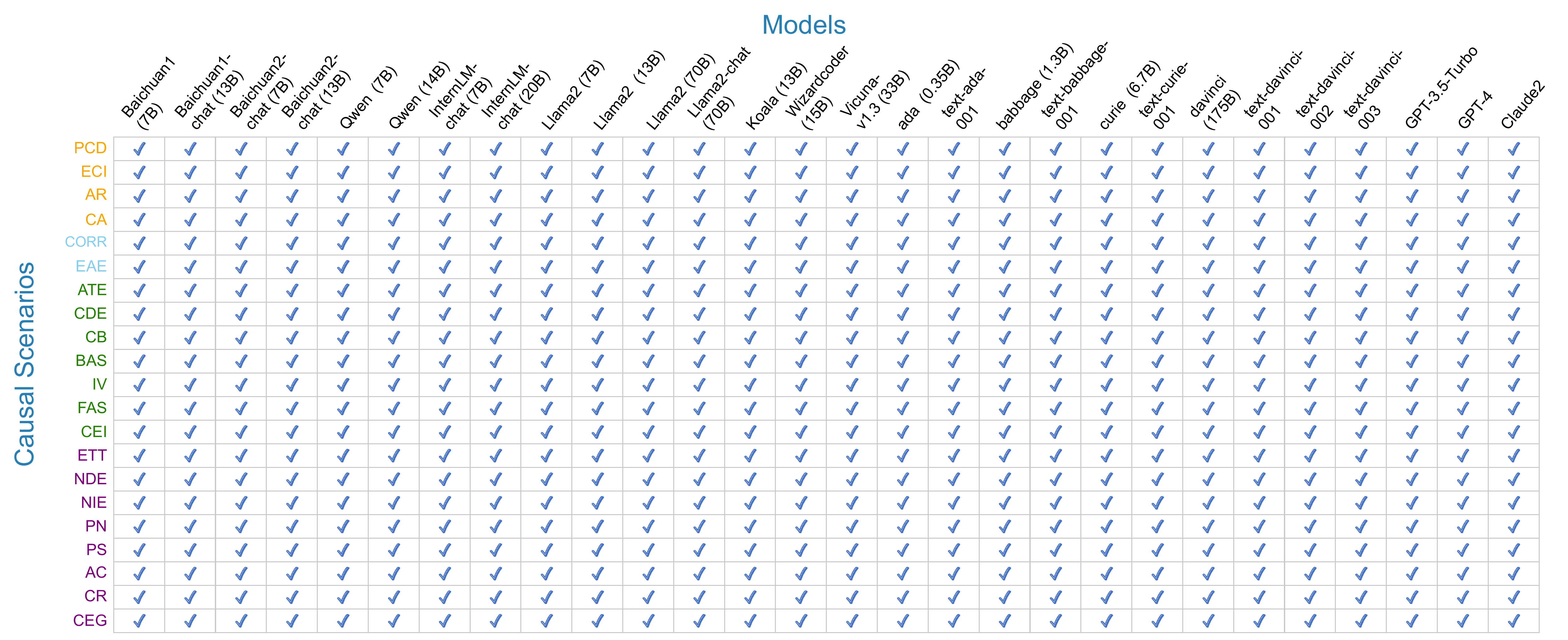}
\label{fig_intro:scenario_model_calm}
}
\caption[Thorough and standardized evaluation (causal scenario-based)]{\textbf{Thorough and standardized evaluation (causal scenario-based).} (a) Previous studies reveal the uneven and incomplete nature of evaluating the causal reasoning abilities of language models across various causal scenarios, underscoring existing gaps. (b) Through CaLM, we conduct comprehensive evaluations of 28 models across 21 causal scenarios. By leveraging CaLM, we can achieve a comprehensive and profound understanding of the causal reasoning abilities of language models.}    
\label{fig_intro:scenario_model}
\end{figure}

\begin{figure}[t!]
\centering
\subfigure[Previous work]{
\centering
\includegraphics[width=.81\linewidth]{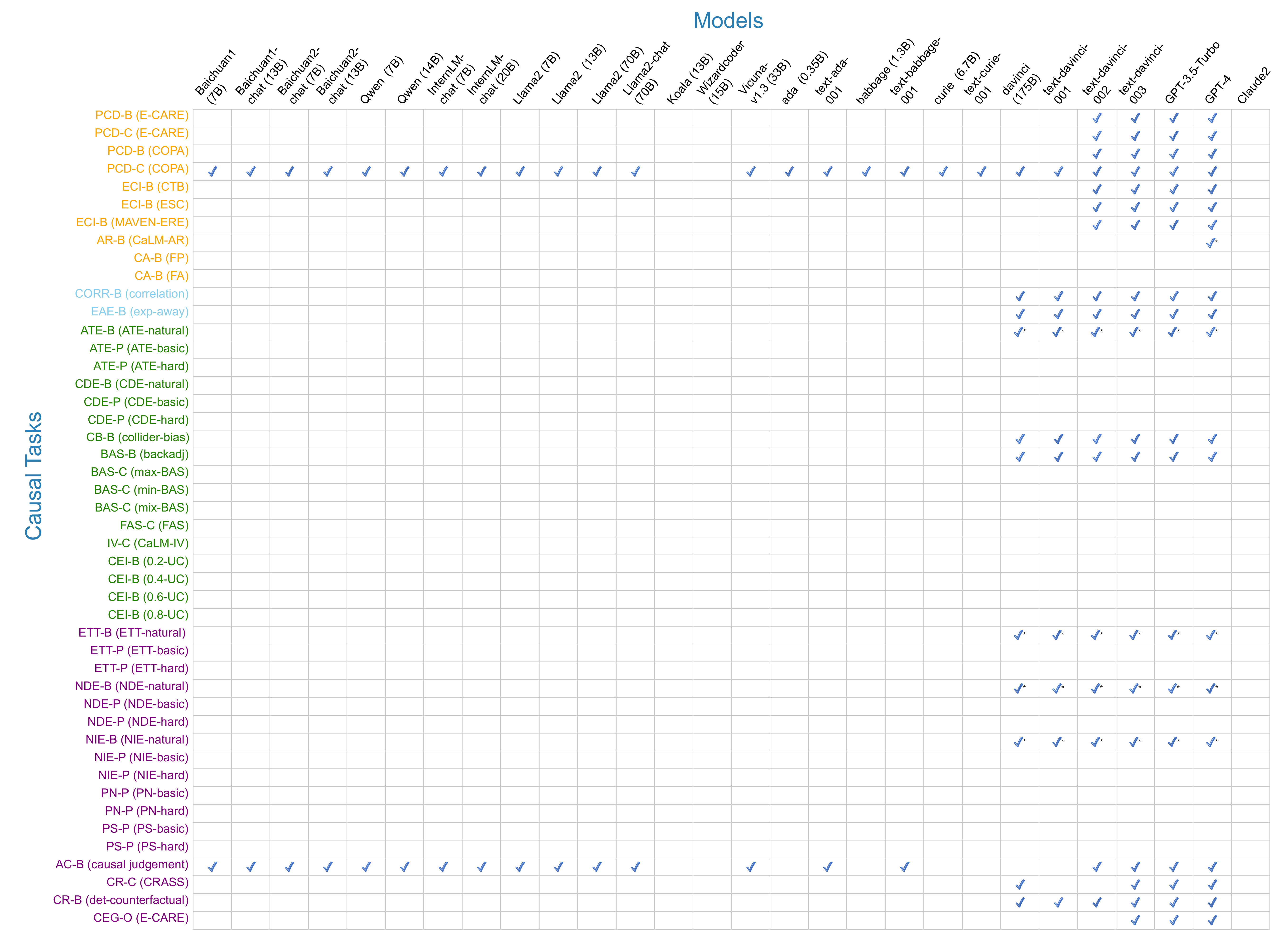}
\label{fig_intro:task_model_prev}
}
\subfigure[CaLM]{
\centering
\includegraphics[width=.81\linewidth]{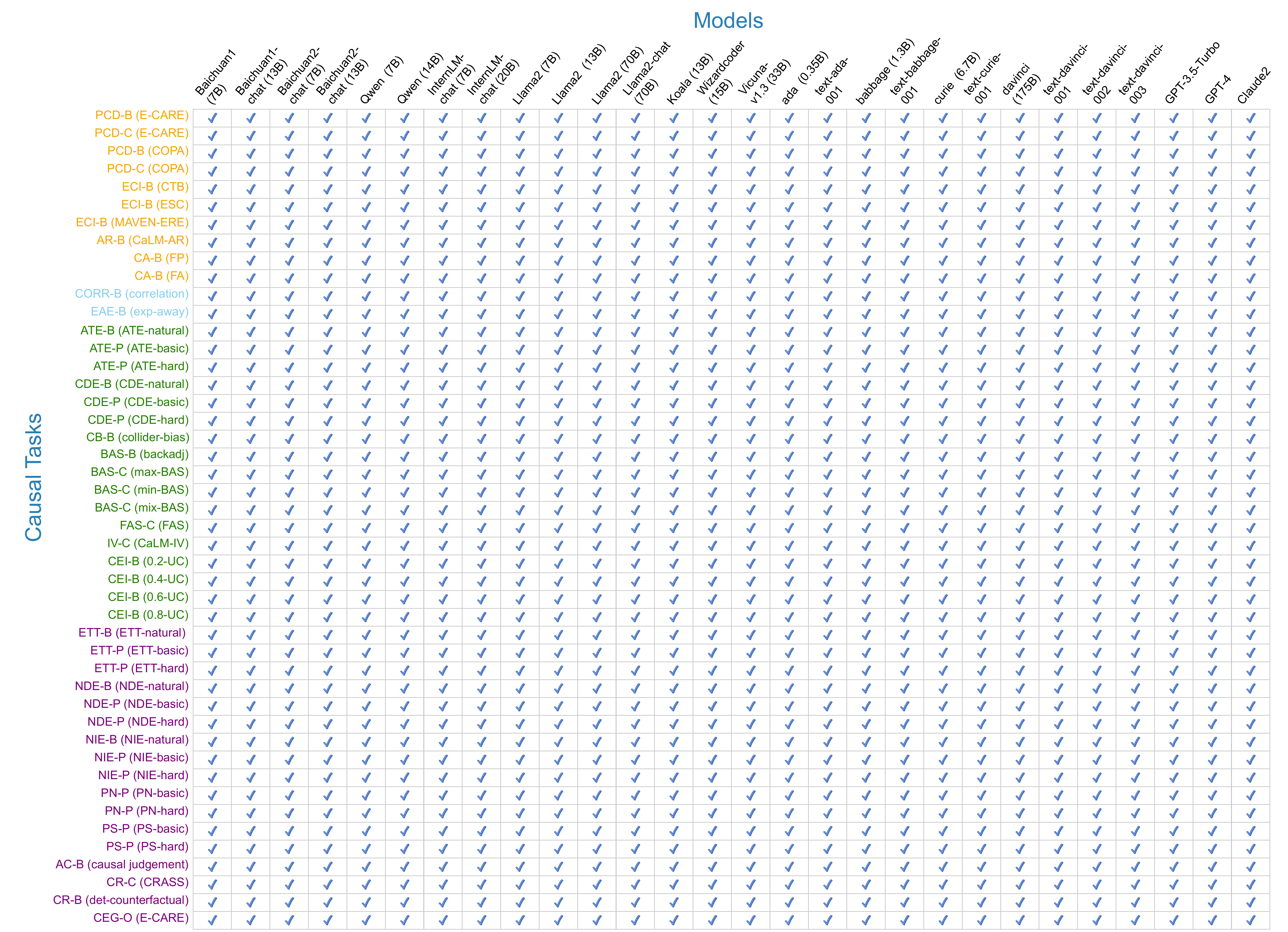}
\label{fig_intro:task_model_calm}
}
\caption[Thorough and standardized evaluation (causal task-based)]{\textbf{Thorough and standardized evaluation (task-based).} (a) Previous studies reveal the uneven and incomplete nature of evaluating causal reasoning abilities of language models across various tasks (* means that this causal task has already been evaluated in existing works, but with different datasets from those we use). (b) Through CaLM, we conduct comprehensive evaluations of 28 models across 46 causal tasks.}    
\label{fig_intro:task_model}
\end{figure}

\begin{figure}[t!]
\centering
\subfigure[Previous work]{
\begin{minipage}{8.5cm}
\centering
\includegraphics[width=1\linewidth]{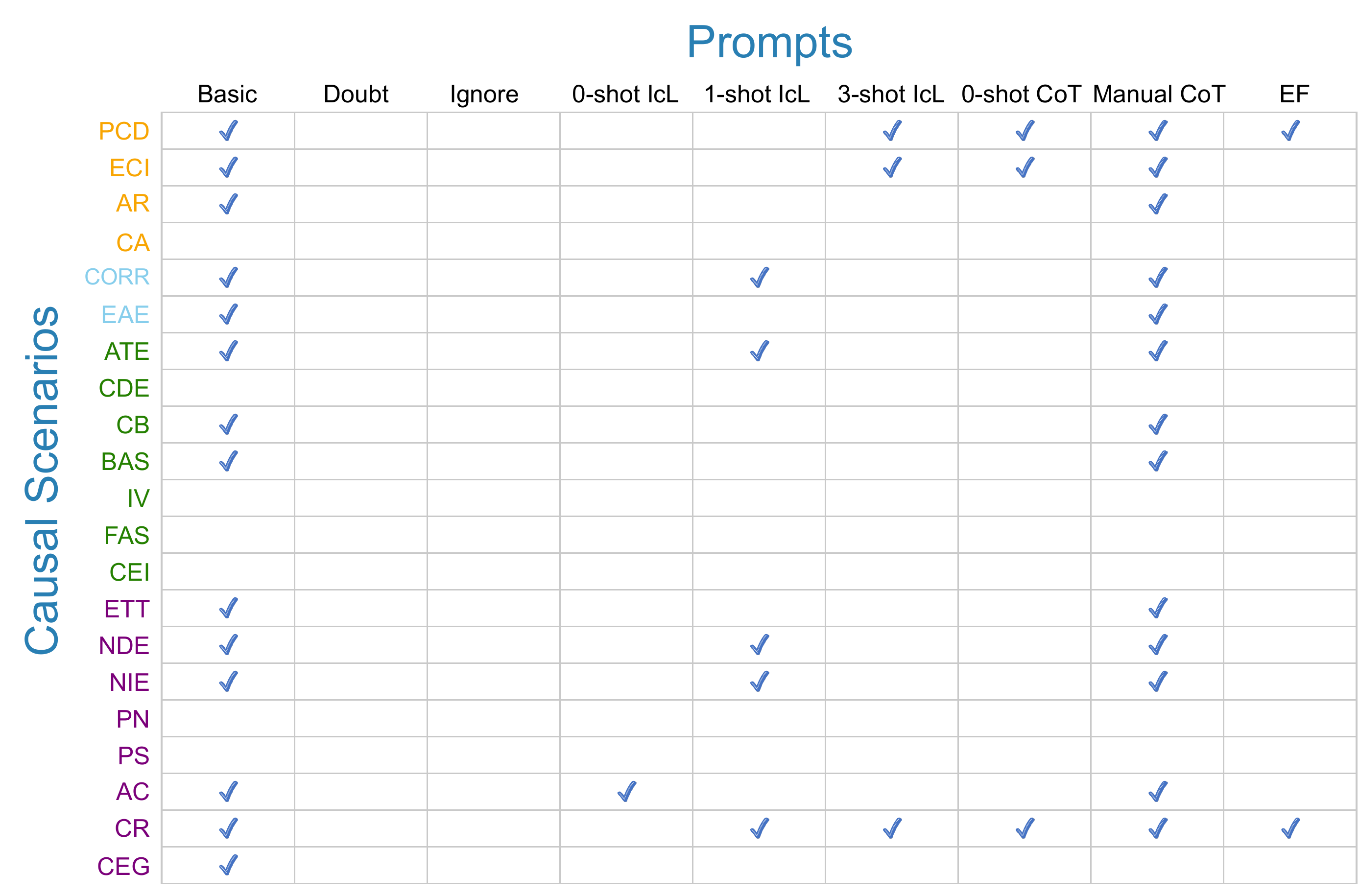}
\end{minipage}
}
\subfigure[CaLM]{
\begin{minipage}{8.5cm}
\centering
\includegraphics[width=1\linewidth]{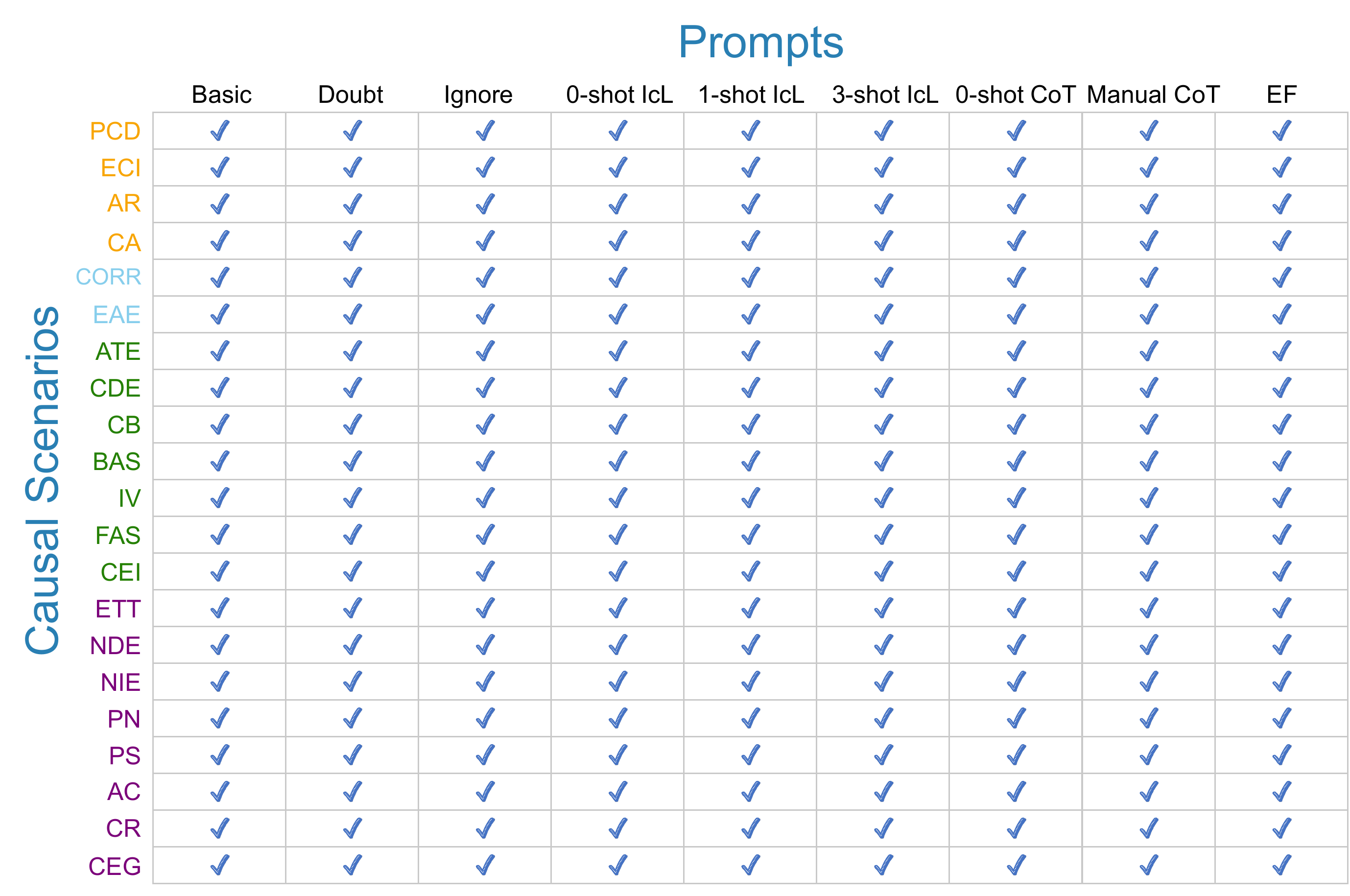}
\end{minipage}
}
\caption[Extensive adaptation strategies (causal scenario-based)]{\textbf{Extensive adaptation strategies (causal scenario-based).} (a) These strategies are previously utilized to evaluate the causal reasoning abilities of language models, highlighting issues of imbalance, incompleteness, and a lack of consideration for prompts from a robustness standpoint. (b) In CaLM, we implement 9 adaptations across 21 causal scenarios, leading to a thorough comprehension of the effectiveness and existing constraints associated with different adaptation strategies in enhancing the model's causal reasoning performance.}
\label{fig_intro:scenario_prompt}
\end{figure}

\begin{figure}[t!]
\centering
\subfigure[Previous work]{
\begin{minipage}{8.5cm}
\centering
\includegraphics[width=1\linewidth]{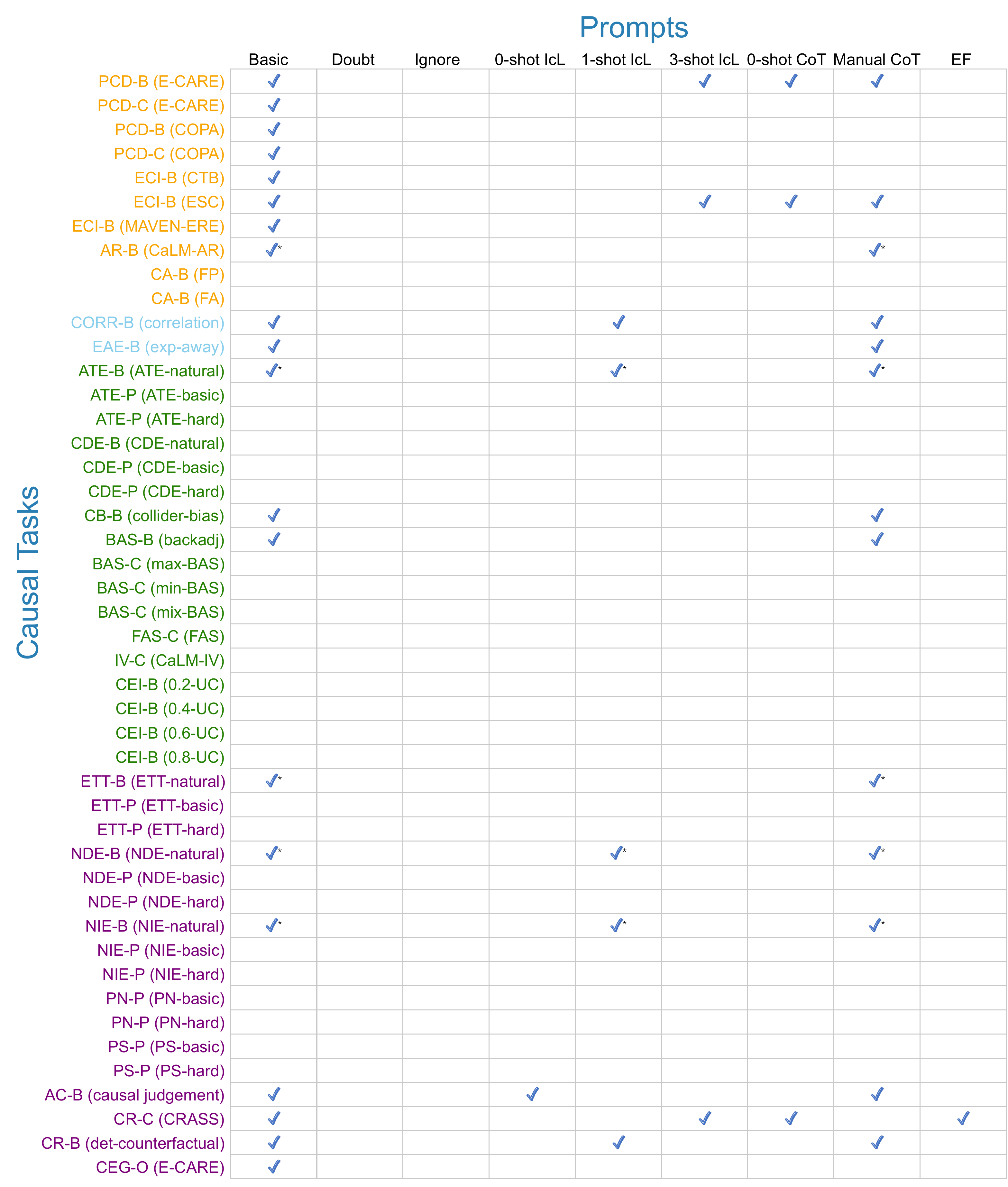}
\end{minipage}
}
\subfigure[CaLM]{
\begin{minipage}{8.5cm}
\centering
\includegraphics[width=1\linewidth]{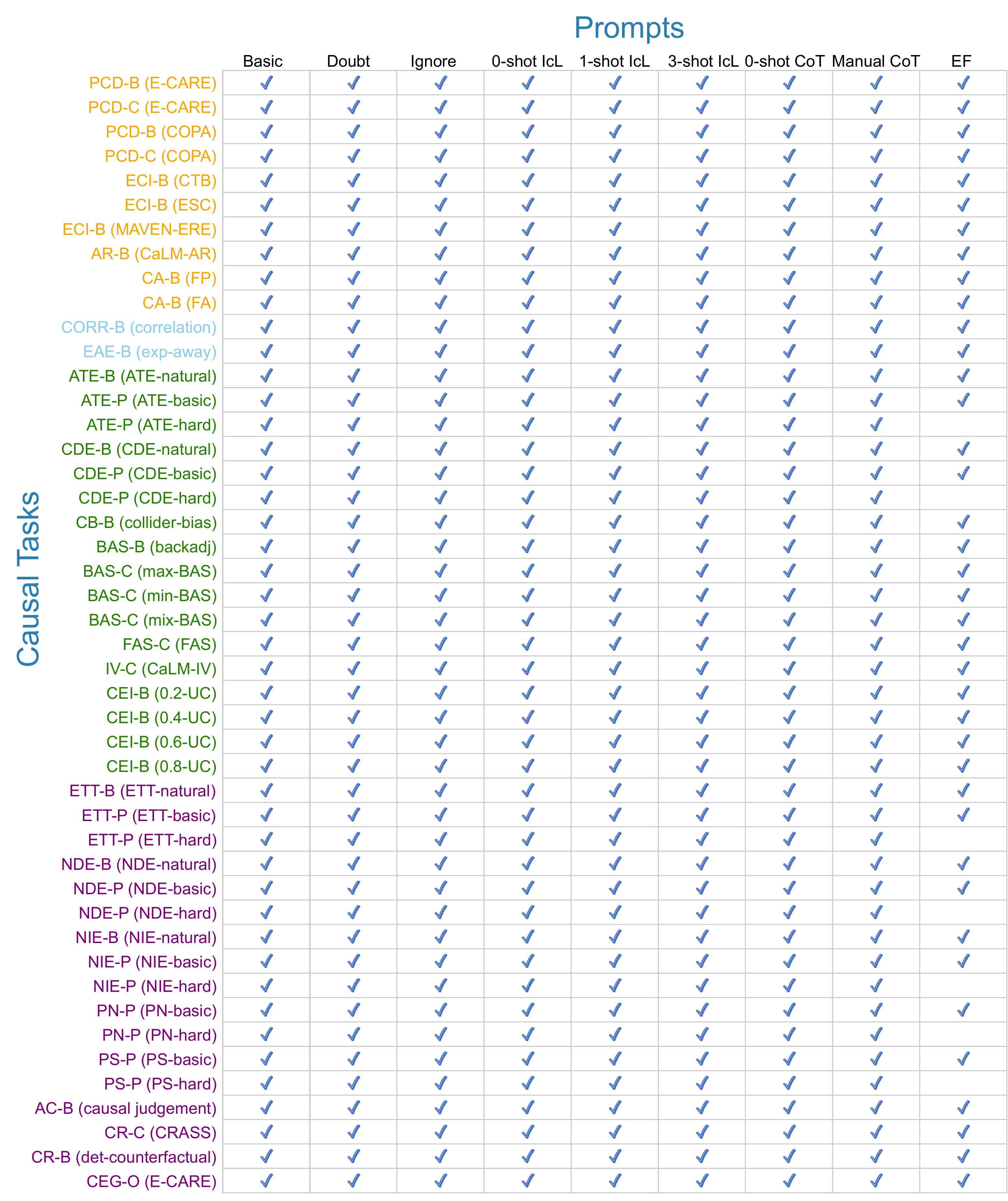}
\end{minipage}
}
\caption[Extensive adaptation strategies (causal task-based)]{\textbf{Extensive adaptation strategies (causal task-based).} (a) These strategies are previously utilized to evaluate the causal reasoning abilities of language models, highlighting issues of imbalance, incompleteness, and a lack of consideration for prompts from a robustness standpoint (* means that this causal task has already been evaluated in existing works, but with different datasets from those we use). (b) In CaLM, we implement 9 adaptations across 46 causal tasks, leading to a thorough comprehension of the effectiveness and existing constraints associated with different adaptation strategies in enhancing the model's causal reasoning performance.}
\label{fig_intro:task_prompt}
\end{figure}

\subsubsection{Key Features of CaLM}

\paragraph{Flexible and scalable framework.} 
First, by establishing an abstract taxonomy comprising four modules (\emph{causal target}, \emph{adaptation}, \emph{metric}, and \emph{error}), CaLM defines a wide-reaching, if not entire, design space for evaluating the causal reasoning capabilities of language models. This taxonomy not only allows for a systematic approach to selecting evaluation criteria but also explicitly outlines the framework's priorities and limitations. This level of abstraction ensures that CaLM can adapt and expand as new challenges and requirements emerge in the field of causal reasoning, showcasing its inherent flexibility. Second, the practical application of this taxonomy, through the selection and implementation of a specific set of 92 causal targets, 9 adaptations, 7 metrics, and 12 errors, demonstrates CaLM's scalability. Together, these two levels enable CaLM to be both adaptable to new developments in the field (flexibility) and capable of being applied to a wide range of causal scenarios and scales (scalability).

\paragraph{Comprehensive evaluation.}
One of the major goals of CaLM is to establish a consensus on the causal reasoning capabilities of language models. We conduct evaluations on 28 prominent language models from nine organizations spanning both academic and industrial sectors: OpenAI (e.g., \gptf, \chatgpt), Anthropic (i.e., \claude), Shanghai AI Laboratory (i.e., \internseven, \internt), Alibaba Cloud (i.e., \qws, \qwf), Baichuan Inc. (e.g., \baichuanonec, \baichuatwos), Meta (e.g., \llama, \llamachatseventy), Lmsys (i.e., \vicuna), UC Berkeley (i.e., \kl), and Microsoft (i.e., \wc). These models are categorized into two accessibility types: Open (e.g., \llamaseven, \internt) and Limited (e.g., \gptf) (detailed in \nameref{main:model} (\cref{main:model})). Despite the significant societal impacts of some models (e.g., \gptf, \chatgpt), a fair, open, and comprehensive benchmark for their causal reasoning abilities is lacking.
We achieve the uniform evaluation from two aspects: (1) From the model perspective, we illustrate in Figure \ref{fig_intro:scenario_model} and Figure \ref{fig_intro:task_model} that, before CaLM, models were typically evaluated in only 18\% of the 21 causal scenarios and 10\% of the 46 causal tasks. We have increased these proportions both to 100\%. (2) From the standpoint of prompts, Figure \ref{fig_intro:scenario_prompt} and Figure \ref{fig_intro:task_prompt} show that, prior to CaLM, the usage of prompts was limited and uneven, with an average of only 1.9 prompts per causal scenario and 1 prompt per causal task. CaLM has elevated these figures to 9 and 8.8, respectively. By conducting evaluations under standardized causal scenarios and conditions (e.g., employing the same adaptation strategy across all models), we achieve a fair and uniform evaluation between models.

\paragraph{Navigating implementation.}
CaLM guides us in systematically selecting causal targets, adaptations, metrics, and identifying errors. It also plays a crucial role in clearly highlighting the existing gaps and outlining directions for further exploration. Given the complexity and breadth of the design space CaLM defines, it is unrealistic to fully explore it within a limited timeframe. Thus, alongside presenting a broad vision and concrete implementation, we explicitly address the current gaps in \nameref{gap} (\cref{gap}), aiming to focus future research on these unexplored areas in the causal evaluation of language models. Importantly, we view CaLM as a sustainable benchmark, systematically updated with new implementations of causal targets, adaptations, models, metrics, and error types, to adapt and grow in response to ongoing research advancements. 

\paragraph{Platform and codebase.} 
CaLM serves as a multifaceted platform and codebase designed for evaluating the causal reasoning capabilities of language models, catering to diverse needs within the research and development community. Its utility spans website, leaderboards, datasets, and toolkits. (1) \emph{Website}: CaLM's web presence facilitates easy access to results, resources, documentations, and updates. This accessibility promotes widespread adoption and provides a foundation for both new learners and experienced researchers to explore the framework's capabilities. (2) \emph{Leaderboards}: CaLM's inclusion of leaderboards provides a competitive and collaborative space for researchers to share their results. Leaderboards highlight the performance of different models on the framework's evaluation criteria, fostering a healthy competition that drives progress in the field. Additionally, they serve as a benchmark for assessing advancements and identifying areas requiring further research. (3) \emph{Datasets}: CaLM contributes to the dataset community by providing curated sets of causal targets, along with corresponding adaptations, metrics, and errors. These datasets are critical for testing and benchmarking language models. By emphasizing comprehensive coverage, significance, and practicality as aforementioned, CaLM ensures that its datasets are valuable for a wide range of research focuses, from theoretical exploration to applied causal tasks. (4) \emph{Toolkits}: At its core, CaLM includes a comprehensive set of tools for evaluating the causal reasoning abilities of language models. These toolkits enable researchers to systematically assess models against a defined set of criteria (e.g., causal targets, adaptations, metrics, errors), ensuring that evaluations are consistent, reproducible, and scalable. The toolkits' design allows for the extension or modification of evaluation criteria, making it adaptable to evolving research needs.

\subsubsection{Considerations at a Broader Level}
Before delving into the empirical findings, we aim to clarify our considerations from a broader perspective.
(1) The choice of metrics for evaluating model performance deserves careful consideration. While we select widely recognized metrics that have proven useful in previous studies, there is no single metric that can capture all aspects of a model's performance. Different metrics may yield different insights into a model's strengths and weaknesses, and should be chosen based on the specific aims of the study. 
(2) Understanding the reasoning behind a model's predictions is crucial for real world applications, particularly in sensitive domains such as healthcare and criminal justice. While our evaluation focuses primarily on quantitative performance metrics, the qualitative aspect of how interpretable these models are remains an essential area for further investigation (e.g., \citet{chen2024quantifying}).
(3) Similar to \citet{liang2022holistic}, we evaluate 28 models using the same causal targets, adaptation strategies and metrics. Despite this uniformity, variations exist among the models themselves, with some settings that are more suitable to achieve optimal performance than others. Thus, a model's poor performance in CaLM does not necessarily reflect its overall causal reasoning abilities. 
(4) The extent to which models have been exposed to the open-source datasets we use might vary significantly. Although we have constructed approximately 90\% of our datasets to mitigate \textit{training-test contamination} \citep{liang2022holistic}, this issue may still be unavoidable. 
(5) Our dataset construction employs similar templates across various causal scenarios, detailed in \nameref{data:construction} (\cref{data:construction}). This approach serves as a double-edged sword. Positively, it tests the model's ability to recognize subtle differences within similarly worded causal scenarios. The model must identify the essence of the problem, and provide an appropriate solution based on this understanding. However, this approach also limits dataset diversity, potentially hindering an extensive evaluation of the models' causal reasoning capabilities \citep{cobbe2021training}. Acknowledging this limitation, we plan to improve dataset diversity in future research to enable a more detailed examination of these capabilities.

\subsection{Empirical Findings}
\label{intro:findings}
 Within the CaLM framework, we conduct comprehensive evaluations on 92 causal targets, covering 46 causal tasks across all four levels of the causal ladder, in three textual modes, and in two languages. Additionally, we incorporate 9 adaptations, apply 7 metrics, and catalog 12 types of errors.
 A dataset consisting of 126,334 data samples is constructed to facilitate thorough evaluations of 28 models, resulting in a total of 38,910,872 queries. 

 Through the comprehensive analysis of extensive experimental results, we distill the following 50 high-level findings across various dimensions:
\subsubsection{Findings from the Model}
\begin{enumerate}[label=(\arabic*)]
    \item \textbf{Causal reasoning inability.} At present, language models struggle to perform tasks requiring sophisticated causal reasoning effectively. As the complexity of causal reasoning increases, the accuracy of each model progressively deteriorates, eventually falling almost to zero (Figure \ref{fig_main:complexity_all}).
    
    \item \textbf{Dual effects of Reinforcement Learning from Human Feedback (RLHF).} On the one hand, exploiting human feedback enables RLHF to align model outputs more closely with human reasoning, particularly in complicated scenarios that demand an understanding of causality. This alignment can modestly improve the model's causal reasoning capabilities (Figure \ref{fig_main:acc_strategy}). On the other hand, models fine-tuned with RLHF tend to change their responses when interacted with by humans. They frequently modify their initial answers, even when they are correct, based on user instructions, indicating a susceptibility to human input (Figure \ref{fig_adversarial_prompt:strategy}).
    
    \item \textbf{Challenges with Supervised Fine-Tuning (SFT) in causal reasoning.} There is only a minimal performance gap in causal reasoning between models trained via SFT on datasets unrelated to causality and those only subjected to pre-training. This suggests that applying SFT to non-causality datasets in the hope of generalizing to causal reasoning might not be effective. A more straightforward method to enhance a model's causal reasoning seems to employ datasets directly related to causality for SFT (Figure \ref{fig_main:acc_strategy}).
    
    \item \textbf{Progression of causal reasoning capabilities in OpenAI's model series.} Our evaluation covers a wide range of OpenAI's model releases, including the GPT-3 series from 2020, the InstructGPT and GPT-3.5 series from 2022, and the GPT-4 released in 2023 (for more information, refer to \nameref{main:model} (\cref{main:model})). Although some GPT-3 and InstructGPT APIs have now been deprecated, their inclusion in our study is crucial for understanding the evolutionary progress in OpenAI's model series. Each new model iteration has exhibited substantial improvements in their ability to perform causal reasoning tasks (Figure \ref{fig_main:acc_time_all} and Figure \ref{fig_main:acc_scale_all}). Furthermore, there has been a noticeable increase in the integration of accuracy and robustness within OpenAI's models (Figure \ref{fig_main:central_metric}).
    
    \item \textbf{Challenges of causal reasoning in Mathematical mode.} Language models demonstrate a certain level of proficiency in solving causal reasoning tasks in both Natural and Symbolic modes. However, their performance in Mathematical mode reveal significant room for improvement. This mode requires models to not only comprehend causal concepts but also to perform precise computations, presenting a dual challenge (Figure \ref{fig_main:direct_mode}).
    
    \item \textbf{Ascending difficulties in rungs of causal ladder.} The model's proficiency in causal reasoning decreases from the lower to the higher levels of the causal ladder, indicating that the more advanced levels present greater difficulties. Models show better performance at the foundational stages (i.e., causal discovery and association) than at the more complex stages (i.e., intervention and counterfactuals) (Figure \ref{fig_main:direct_ladder}).
    
    \item \textbf{Comparing open vs. limited access models.} Overall, limited access models exhibit superior causal reasoning capabilities than open models. However, in the majority of causal scenarios at the causal discovery level, the performance gap between open and limited access models is minimal, not exceeding a 2\% margin. This modest gap encourages an optimistic perspective on the potential of open models. Additionally, we aim for CaLM to act as a catalyst for the development of models within the open-source community (Figure \ref{fig_main:acc_access}). 
    
    \item \textbf{Impact of scaling on causal reasoning ability.} The relationship between model scale and accuracy in causal reasoning does not display a straightforward monotonic increase. This implies that other factors, such as training data and strategy, significantly influence accuracy across models from different creators. However, within models from the same creator, scale remains a consistent and reliable predictor of accuracy (Figure \ref{fig_main:acc_scale_all}).
    
    \item \textbf{Balancing instruction-following and error correction.} When confronted with adversarial prompts, the model tends to alter its previous responses. Notably, it is more likely to change initially correct answers to incorrect ones rather than rectify pre-existing errors. This tendency highlights the urgent need to balance the model's ability to follow instructions with its proficiency in identifying and correcting errors (Figure \ref{fig_adversarial_prompt:direction_relationship} and Figure \ref{fig_adversarial_prompt:compare_direct}).
\end{enumerate}

\subsubsection{Findings from the Adaptation}
\begin{enumerate}[label=(\arabic*), resume]
    \item \textbf{Optimal prompt varies across causal scenario.} No ``optimal prompt'' universally fits all causal scenarios. Based on our observations, for scenarios at the lower levels of the causal ladder (i.e., causal discovery and association), employing 1/3-shot IcL proves effective. For scenarios at the intervention level, 3-shot IcL is recommended, and adding more shots may be beneficial if possible. For the counterfactuals level, which requires detailed reasoning to determine the correct response, we suggest using manual CoT (Figure \ref{fig_main:extra_scenario_prompt}).
    
    \item \textbf{Challenges of using prompts in complex causal scenarios.} The effectiveness of prompts in improving model performance is not consistent across all scenarios. Complex causal scenarios pose a particular challenge for language models, often due to the absence of substantial information on these scenarios within the model's training corpus. Moreover, questions in these scenarios cannot be adequately resolved merely through common sense or semantic understanding. In CaLM, we observe that in such complex causal scenarios, prompts do not markedly improve model performance (Figure \ref{fig_main:extra_scenario_prompt}). 
    
    \item \textbf{Improving model performance with 3-shot IcL and manual CoT.} Using 3-shot IcL improves the baseline performance of various models by providing a consistent format for answers along with a rich set of examples. For top-tier models (e.g., GPT-4), manual CoT is particularly effective in harnessing their advanced causal reasoning capabilities. Through precise, step-by-step reasoning, manual CoT helps these models better comprehend the implications behind questions, thus substantially improving their overall performance (Figure \ref{fig_main:direct_prompt}).
    
    \item \textbf{Sensitivity to prompt's shot variation.} Across all causal scenarios, there is no strong correlation among prompts within the same category when the number of examples varies (e.g., 0/1/3-shot IcL, as well as 0-shot/manual CoT). This weak correlation suggests that models are highly sensitive to changes in the number of shots in prompts. It further emphasizes the importance of carefully selecting the number of shots in prompts to tailor model performance effectively (Figure \ref{fig_main:central_prompt}). 
    
    \item \textbf{Effectiveness of few shots in complex causal tasks.} The more challenging the causal task, the more beneficial additional examples in the prompt are for improving model performance. In CaLM, we assess difficulty across three dimensions: the causal ladder (with intervention and counterfactuals being the most challenging), mode (with Mathematical mode being more demanding), and question type (with probability calculations being particularly difficult). Our thorough analysis suggests that increasing the number of shots for these challenging tasks significantly improves performance. However, due to constraints on time and resource, IcL is currently limited to three shots. While we advocate for using more examples, the decision to set an upper limit should be made based on specific circumstances (Figure \ref{fig_icl_prompt:difficulty_shots}). 
    
    \item \textbf{Limited effectiveness of 0-shot prompts.} One of our objectives is to identify a prompt that is simple to construct yet effectively enhance the model's causal reasoning abilities. To this end, we experimented with three variations of 0-shot prompts: 0-shot CoT, 0-shot IcL and EF, none of which include examples. Comparative analyses reveal that these prompts do not substantially outperform the basic prompt, and their effectiveness varies across different causal scenarios (Figure \ref{fig_main:direct_prompt}, Figure \ref{fig_main:extra_scenario_prompt} and Figure \ref{fig_main:stability_scenario}).
    
    \item  \textbf{Correlations between prompts.} The basic prompt significantly correlates with adversarial doubt, adversarial ignore, EF, 0-shot CoT, and 0-shot IcL. However, it shows no strong correlation with more complex prompts such as 3-shot IcL and manual CoT. For prompts showing strong correlations, it is feasible to approximate a model's performance across similar prompts based on its performance with any one of them. Conversely, the absence of strong correlations with certain prompts offer opportunities for designing more diverse and effective prompts in the future (Figure \ref{fig_main:central_prompt_scatter} and Figure \ref{fig_main:central_prompt}).
\end{enumerate}

\subsubsection{Findings from the Causal Ladder}
\begin{enumerate}[label=(\arabic*), resume]
    \item \textbf{Consistent model capabilities in causal reasoning across scenarios.} The causal reasoning capabilities of models show inherent consistency across the four levels of the causal ladder. Specifically, in 19 scenarios (excluding CEI and CB), there is a positive correlation in model performance. This observation suggests that a model’s causal reasoning ability is cohesive, not limited to specific scenarios (Figure \ref{fig_main:central_scenario}). 
    
    \item \textbf{Correlations within the causal ladder.} Causal scenarios that fall within the same level of the causal ladder and share the same mode tend to exhibit higher correlations in performance. This trend underscores the validity of our hierarchical organization of causal scenarios (Figure \ref{fig_main:central_scenario}).
\end{enumerate}

\subsubsection{Findings from the Domain}
\begin{enumerate}[label=(\arabic*), resume]
    \item \textbf{Comparing seen vs. unseen dataset.} The impact of using seen (open-source) and unseen (self-constructed) datasets on model performance is influenced by the complexity of the causal tasks. For more complex tasks at the intervention and counterfactuals levels, models tend to perform better on open-source datasets than on self-constructed ones. Conversely, for simpler tasks related to causal discovery, models show slightly superior performance on self-constructed datasets than on those that are publicly available (Figure \ref{fig_main:extra_seen_unseen}). 
\end{enumerate}

\subsubsection{Findings from the Mode}
\begin{enumerate}[label=(\arabic*), resume]
    \item \textbf{Correlations among text modes.} The three modes selected for our analysis - Natural, Symbolic, and Mathematical - are all rooted in textual data, with Natural mode serving as the primary basis. Our experimental results show a marked correlation between the Natural mode and the other two modes, highlighting interconnected capabilities across these modes (Figure \ref{fig_main:central_mode}). 
\end{enumerate}

\subsubsection{Findings from the Language}
\begin{enumerate}[label=(\arabic*), resume]
    \item \textbf{Performance differences between English and Chinese datasets.} In almost 90\% of the causal scenarios, models demonstrate superior performance on English datasets. The trend is likely attributed to the dominance of English in the training data of language models. As these models are deployed globally, it is crucial to ensure training involves balanced and diverse language corpora to improve performance across various languages (Figure \ref{fig_main:acc_multilingual}).
\end{enumerate}

\subsubsection{Findings from the Metric}
\begin{enumerate}[label=(\arabic*), resume]
    \item \textbf{Variability in model's robustness and accuracy across causal scenarios.} The relationship between a model’s robustness and accuracy significantly varies across causal scenarios. In more challenging causal scenarios, such as PN and PS, models may show very low accuracy but disproportionately high robustness. This is primarily because the models' responses remain consistently poor, unaffected by disturbances. In contrast, in simpler scenarios like PCD and AR, there tends to be a positive correlation between accuracy and robustness, suggesting that as models perform better, they also become more stable. However, in scenarios such as ECI, EAE, and AC, the interaction between these metrics does not follow a clear or consistent pattern (Figure \ref{fig_main:central_metric}). 
    
    \item \textbf{Assessing the maturity of causal scenarios.} We employ three metrics to evaluate the maturity of a causal scenario: understandability, open-limited gap, and solvability. Most causal scenarios are considered hard or more difficult in terms of understandability. In the open-limited gap metric, limited access models predominantly occupy the top 5 positions across the majority of scenarios, indicating their superior performance. When evaluating solvability, it becomes evident that current model capabilities are not yet sufficient to fully tackle the challenges posed by CaLM. Overall, the ability of models to effectively resolve causal scenarios within CaLM remains nascent (Figure \ref{fig_main:maturity_all}). 
\end{enumerate}

\subsubsection{Findings from the Error}
\begin{enumerate}[label=(\arabic*), resume]
    \item \textbf{Model capabilities and limitations in following instructions.} All models inherently possess ability to generate content and typically do not produce empty responses, even when faced with challenging questions. However, their capacity to accurately follow instructions remains limited. Often, these models struggle to provide the most straightforward response as specified by the instructions, indicating a significant room for improvement in following instructions (Table \ref{main:table_model_errors}).
    
    \item \textbf{Reduction of repetitions through SFT.} SFT equips models with high-quality input-output pairs, eﬀectively mitigating unnecessary repetitions in responses to questions (Table \ref{main:table_model_errors}).
    
    \item \textbf{Improving instruction following with 1-shot and 3-shot IcL.} Utilizing 1-shot and 3-shot IcL provides models with standardized, concise examples, facilitating the learning of eﬀective response patterns. This helps models produce outputs that better conform to the specified answer format (Figure \ref{fig_main:error_prompt}).
    
    \item \textbf{Imitation effects from prompts.} Employing 1-shot IcL, 3-shot IcL, and manual CoT might lead to an ``imitation game'' where models mimic the patterns presented in the examples. Specifically, after generating standardized responses, these models begin crafting their own questions, reflecting the learned patterns (Figure \ref{fig_main:error_prompt}).
    
    \item \textbf{Language inconsistency in 0-shot CoT.} Some models struggle to systematically process and respond to complex Chinese questions when using 0-shot CoT. This challenge can lead to oﬀ-topic initial responses in Chinese, followed by a switch to English, although these subsequent English responses often continue to be irrelevant to the posed question (Figure \ref{fig_main:error_prompt} and Figure \ref{fig_main:error_quantitative_example1}).
    
    \item \textbf{Prevalence of identical responses across questions.} The majority of models (26 out of 28) show the tendency to provide the same response to different questions, indicating their fundamental inability to effectively handle the causal task. This issue, if observed in one question type (e.g., binary classification), is likely to manifest similarly across other question types (e.g., choice selection, probability calculation) (Figure \ref{fig_main:error_all_same}).
\end{enumerate}

\subsubsection{Findings from the Causal Scenario}
\begin{enumerate}[label=(\arabic*), resume]
    \item \textbf{Pairwise causal discovery (PCD).} 
    PCD seeks to establish if a causal relationship exists between two given events and to identify which of the two is the cause and which is the effect. The \emph{understandability} of the scenario is easy. The leading three performers in this scenario are GPT-4 (79.1\%), GPT-3.5-Turbo (75.2\%), and text-davinci-003 (74.7\%). The \textit{top model-prompt pair} is GPT-4 with EF, achieving an accuracy of 83.0\%. The \textit{solvability} of the scenario is well-solved as the average accuracies of the top three models all exceed 70\%. The most stable models, characterized by the lowest \textit{model volatility}, are GPT-3.5-Turbo (1.3), Baichuan1 (7B) (2.1), and text-curie-001 (2.2). The models displaying the greatest sensitivity to different prompts, evidenced by their high \textit{model volatility}, are Vicuna-v1.3 (33B) (15.8), Llama2 (70B) (15.6), and Llama2-chat (70B) (14.3). The most effective prompts are 3-shot IcL and 1-shot IcL, which improve average accuracy by 9.0\% and 7.0\% respectively (Section \ref{scenario:discovery}). 
    
    \item \textbf{Event causality identification (ECI).}
    ECI requires the model to assess whether there is a causal relationship between two events within a given sentence. The \emph{understandability} of the scenario is easy. The top three models by average accuracy are GPT-4 at 65.6\%, text-davinci-003 at 61.1\%, and Claude2 at 58.4\%. The \textit{top model-prompt pair} is GPT-4 with adversarial doubt, reaching an accuracy of 67.0\%, indicating the scenario has a challenging \textit{solvability} since the performance of the \textit{top model-prompt pair} does not exceed 80\%. The three most stable models in the scenario, characterized by the lowest \textit{model volatility}, are GPT-4 with a \textit{model volatility} of 1.1, Baichuan2-chat (13B) with 1.6, and Qwen (7B) with 2.1. Conversely, the models exhibiting the highest \textit{model volatility}, are InternLM-chat (20B) at 23.6, text-babbage-001 at 11.3, and Llama2 (7B) at 11.2. The leading two prompts, achieving the greatest average accuracy improvements over the basic prompt, are 1-shot IcL with a gain of 3.1\% and 3-shot IcL with 2.1\% (Section \ref{scenario:discovery}).
    
    \item \textbf{Abstract reasoning (AR).}
    AR investigates the capability of language models to identify and understand causal relationships within symbolic causal graphs. This scenario is classified to have an easy \textit{understandability}. The top three models by average accuracy: GPT-4 at 88.3\%, Claude2 at 75.9\%, and text-davinci-003 at 74.5\%. GPT-4, employing manual CoT, stands out as the \textit{top model-prompt pair} with a 92.6\% accuracy. The \textit{solvability} of the scenario is well-solved with each of the top three models' average accuracies exceeding 70\%. The three most stable models in the scenario, characterized by the lowest \textit{model volatility}, are GPT-4 at 2.0, Qwen (7B) at 2.3, and InternLM-chat (20B) at 2.6. Conversely, the most unstable models are Llama2-chat (70B) at 21.6, Llama2 (70B) at 21.1, and Llama2 (7B) at 17.0. The leading two prompts by average accuracy gain over the basic prompt are 0-shot IcL and 1-shot IcL, both at 1.5\% (Section \ref{scenario:discovery}).
    
    \item \textbf{Causal attribution (CA).}
    CA refers to the process of determining which speciﬁc factor is responsible for an outcome. The scenario has an easy \textit{understandability}. GPT-4 leads with an average accuracy of 91.8\%, followed by text-davinci-003 at 77.1\%, and Claude2 at 74.0\%. GPT-4, when paired with manual CoT, achieves an impressive 94.8\%. The \textit{solvability} of this scenario is well-solved given that the top three models all have average accuracies over 70\%. The three most consistent models, characterized by the lowest \textit{model volatility}, are GPT-4 at 1.4, davinci (175B) at 2.4, and GPT-3.5-Turbo at 3.0, showcasing their robustness across various prompts. Conversely, the models demonstrating the highest \textit{model volatility}, are Llama2-chat (70B) at 20.5, Llama2 (70B) at 13.6, and Llama2 (7B) at 11.6. The two prompts with the highest average accuracy gain over the basic prompt are 1-shot IcL at 1.0\% and 0-shot IcL at 0.8\% (Section \ref{scenario:discovery}).
    \item \textbf{Correlation (CORR).}
    CORR requires the model to identify statistical association between variables. The \textit{understandability} of the scenario is hard. The leading three models by average accuracy are GPT-4 at 59.1\%, text-davinci-003 at 54.7\%, and text-davinci-002 at 54.3\%. Claude2, using EF, stands out with a top score of 68.0\%, illustrating the scenario \textit{solvability} as challenging since the highest \textit{top model-prompt pair}'s performance does not reach 80\%. The models that have the highest the \textit{model volatility} are InternLM-chat (20B) at 17.4, ada (0.35B) at 14.7, and text-ada-001 at 14.1. Conversely, the most stable models include Baichuan1 (7B) at 0.5, Qwen (7B) at 1.2, and text-davinci-001 at 1.9. The top two prompts for average accuracy gain over the basic prompt as 3-shot IcL at 6.2\% and 1-shot IcL at 5.7\% (Section \ref{scenario:association}).
    \item \textbf{Explaining away effect (EAE).}
    EAE describes a causal relationship where two independent causes that produce a common eﬀect become interdependent when that eﬀect is observed. The \textit{understandability} of the scenario is hard. GPT-4 at 67.9\%, Claude2 at 66.7\%, and text-davinci-003 at 57.0\% as the top three models by average accuracy. As to the \textit{top model-prompt pair}, GPT-4, through the use of manual CoT, achieves a remarkable 90.5\%, indicating the \textit{solvability} of the scenario is potentially solvable as the \textit{top model-prompt pair}'s performance surpasses 80\%. The models have the highest \textit{model volatility} are Llama2 (70B) at 18.8, Llama2 (13B) at 17.0, and Llama2 (7B) at 17.0. Conversely, the most stable models include Qwen (7B) at 2.1, davinci (175B) at 3.1, and Baichuan1 (7B) at 3.3. The top two prompts for average accuracy gain over the basic prompt as 3-shot IcL at 5.5\% and 1-shot IcL at 3.9\% (Section \ref{scenario:association}).
    \item \textbf{Average treatment effect (ATE).}
    ATE aims to quantify the impact of a particular intervention. This causal scenario have a hard \textit{understandability}. The leading models in terms of average accuracy for this causal scenario are GPT-4 at 54.8\%, text-davinci-003 at 50.3\%, and GPT-3.5-Turbo at 47.7\%. The \textit{top model-prompt pair} is GPT-4 with manual CoT, reaching an impressive 92.8\%, indicating the scenario's \textit{solvability} is potentially solvable given that the \textit{top model-prompt pair} exceeds 80\%. The three most stable models, indicated by the lowest \textit{model volatility}, are Baichuan1-chat (13B) at 2.4, Baichuan2-chat (13B) at 3.0, and InternLM-chat (20B) at 6.4. Conversely, the three models exhibiting the greatest instability across various prompts, shown by the highest \textit{model volatility}, are Llama2 (13B) at 34.8, Llama2 (70B) at 30.2, and Llama2 (7B) at 28.4. The two prompts leading in average accuracy gain relative to the basic prompt are 3-shot IcL at 25.0\% and manual CoT at 22.4\% (Section \ref{scenario:intervention}).
    \item \textbf{Backdoor adjustment set (BAS).}
    BAS contains variable that blocks all backdoor paths from the treatment variable to the outcome variable. This scenario challenges whether the model can discern the BAS. This causal scenario is viewed to have a hard \textit{understandability}. The leading models by average accuracy in this causal scenario are GPT-4 at 71.6\%, text-davinci-003 at 53.7\%, and GPT-3.5-Turbo at 49.8\%. The \textit{top model-prompt pair}, GPT-4 with 3-shot IcL, reaches 75.1\%, indicating that the \textit{solvability} of this scenario is challenging due to the \textit{top model-prompt pair}'s performance not exceeding 80\%. The three most consistent models, based on the lowest \textit{model volatility}, are text-davinci-001 at 1.4, text-curie-001 at 2.3, and GPT-4 at 2.6. In contrast, the models exhibiting the greatest variability, marked by the highest \textit{model volatility} across different prompts, are Llama2 (70B) at 16.2, Vicuna-v1.3 (33B) at 11.9, and Llama2 (13B) at 11.8. The two prompts that lead to the highest average accuracy gains over the basic prompt are 3-shot IcL with a 12.1\% gain and 1-shot IcL with a 9.8\% gain (Section \ref{scenario:intervention}).
    \item \textbf{Frontdoor adjustment set (FAS).}
    FAS involves a set of variables that mediate the causal path from the treatment to the outcome. The model needs to choose the correct FAS. This causal scenario has a hard \textit{understandability}. The leading three models by average accuracy: GPT-4 at 77.2\%, text-davinci-003 at 59.9\%, and GPT-3.5-Turbo at 54.0\%. GPT-4, employing 3-shot IcL, tops the chart with a 95.2\% accuracy. GPT-4, employing 3-shot IcL, is the \textit{top model-prompt pair} with a 95.2\% accuracy. With the top model's average accuracy surpassing 70\%, the \textit{solvability} of this scenario is solvable. The most prompt-sensitive models, indicated by the highest \textit{model volatility}, are text-davinci-002 at 18.4, Claude2 at 17.1, and text-davinci-003 at 14.9. In contrast, the most stable models include davinci (175B) at 1.8, text-curie-001 at 3.4, and Baichuan2-chat (13B) at 3.5. The top two prompts for average accuracy gain over the basic prompt as 3-shot IcL at 13.3\% and 1-shot IcL at 10.6\% (Section \ref{scenario:intervention}).
    \item \textbf{Instrumental variable (IV).}
    IV influences the treatment variable but has no direct eﬀect on the outcome variable, except through the treatment. This scenario assesses whether the model can identify the IV. The \textit{understandability} of the scenario is hard. The leading three models by average accuracy are GPT-4 at 74.8\%, text-davinci-003 at 56.5\%, and text-davinci-002 at 53.7\%. GPT-4, employing 3-shot IcL, achieves a top score of 78.9\%, suggesting the \textit{solvability} of this scenario as challenging since the \textit{top model-prompt pair}'s performance doesn't reach 80\%. The models most susceptible to prompt variations, as shown by the highest \textit{model volatility}, are Vicuna-v1.3 (33B) at 16.7\%, ada (0.35B) at 15.9\%, and Llama2 (13B) at 15.1\%. Conversely, the most stable models include text-curie-001 at 0.5\%, GPT-4 at 3.0\%, and InternLM-chat (20B) at 3.3\%. The top two prompts for average accuracy gain over the basic prompt as manual CoT at 15.2\% and 3-shot IcL at 13.2\% (Section \ref{scenario:intervention}).
    \item \textbf{Collider bias (CB).}
    CB occurs when an analysis is conditioned upon a common eﬀect of two or more variables. It evaluates whether the model can exclude the interference of bias and make the correct choice. The \textit{understandability} of the scenario is hard. The top three models by average accuracy are GPT-4 at 62.7\%, text-davinci-003 at 53.2\%, and text-davinci-002 at 53.0\%. The \textit{top model-prompt pair} is GPT-4 with manual CoT, which achieves an impressive 97.8\%, suggesting the \textit{solvability} of this scenario as potentially solvable. The models most sensitive to prompt variations, as shown by the highest \textit{model volatility}, are Llama2 (70B) at 20.9\%, Koala (13B) at 16.8\%, and GPT-4 at 16.2\%. Conversely, the most stable models are text-curie-001 at 2.6\%, curie (6.7B) at 4.3\%, and Wizardcoder (15B) at 4.9\%. The top two prompts for average accuracy gain over the basic prompt as manual CoT at 15.5\% and 3-shot IcL at 13.7\% (Section \ref{scenario:intervention}).
    \item \textbf{Causal effect identification (CEI).} CEI centers on evaluating the model's ability to judge whether the causal eﬀect of a treatment on an outcome can be estimated from observational data. This causal scenario has a  very hard \textit{understandability} and CEI shows the lowest correlation with other causal scenarios. The leading models in this causal scenario, based on average accuracy, are GPT-3.5-Turbo at 49.9\%, text-curie-001 at 49.6\%, and Baichuan1 (7B) at 49.4\%. The \textit{top model-prompt pair}, GPT-4 with 3-shot IcL, reaches 59.0\%, indicating the \textit{solvability} of the scenario as challenging due to the \textit{top model-prompt pair}'s performance falling short of 80\%. The three most stable models, based on the lowest \textit{model volatility}, are text-curie-001 at 0.9, text-davinci-001 at 1.0, and Qwen (7B) at 1.0. Conversely, the models demonstrating the highest levels of instability across various prompts are Llama2 (70B) at 18.1, Llama2-chat (70B) at 15.9, and GPT-4 at 12.9. The two prompts leading in average accuracy gain over the basic prompt are 1-shot IcL at 6.6\% and 3-shot IcL at 5.4\% (Section \ref{scenario:intervention}).
    \item \textbf{Controlled direct effect (CDE).}
    CDE quantifies the direct influence of an intervention on an outcome, while keeping the mediator to a predetermined level. This causal scenario has a hard \textit{understandability}. The leading models in terms of average accuracy for this causal scenario are GPT-3.5-Turbo at 47.6\%, GPT-4 at 41.9\%, and Claude2 at 34.5\%. The \textit{top model-prompt pair} is GPT-4 with manual CoT, reaches accuracy at 90.8\%, suggesting the scenario's \textit{solvability} as potentially solvable given the \textit{top model-prompt pair} surpasses 80\%. The three models exhibiting the greatest stability with the lowest \textit{model volatility} are Baichuan1-chat (13B) at 2.7, babbage (1.3B) at 2.8, and ada (0.35B) at 3.6. Conversely, the three models showing the highest levels of instability across various prompts are Llama2 (70B) at 27.8, Llama2 (13B) at 26.7, and Llama2 (7B) at 25.7, showcasing a pronounced sensitivity to different prompts. The two prompts leading in average accuracy gain over the basic prompt are 3-shot IcL at 21.7\% and manual CoT at 20.9\%. (Section \ref{scenario:intervention}).
    \item \textbf{Counterfactual reasoning (CR).} 
    CR involves contemplating hypothetical scenarios by modifying certain factors or conditions present in an actual situation. This causal scenario has an easy \textit{understandability}. The three leading models in this causal scenario by average accuracy are GPT-4 at 76.9\%, text-davinci-003 at 67.8\%, and Claude2 at 62.5\%. The \textit{top model-prompt pair} is GPT-4 with manual CoT, achieving an 83.2\% accuracy. The scenario has a solvable \textit{solvability} with the top model's average accuracy surpassing 70\%. The three most consistent models, characterized by the lowest \textit{model volatility}, are curie (6.7B) at 1.8, text-curie-001 at 3.2, and Baichuan1-chat (13B) at 3.4. Conversely, the models displaying the greatest variability across various prompts, showcasing their great sensitivity to prompts, are Llama2 (70B) at 15.4, Llama2-chat (70B) at 14.2, and Vicuna-v1.3 (33B) at 11.9. The two prompts leading to the highest average accuracy improvements over the basic prompt are manual CoT at 7.3\% and 3-shot IcL at 6.0\% (Section \ref{scenario:counterfactual}).
    \item \textbf{Actual causality (AC).}
    AC deals with attribution and responsibility allocation problems encountered in practical applications. The causal scenario's \textit{understandability} is hard. GPT-4 leads in average accuracy at 65.6\%, followed by text-davinci-003 and GPT-3.5-Turbo, with scores of 57.2\% and 56.5\%, respectively. GPT-4, when paired with manual CoT prompts, achieves a significant 68.2\% in accuracy, yet this top performance is still short of the 80 threshold, indicating the challenging of the causal scenario. In the stability of model responses, Llama2 (70B), curie (6.7B), and Llama2-chat (70B) show the greatest variations in performance across different prompts, while GPT-3.5-Turbo, GPT-4, and text-curie-001 demonstrate remarkable consistency according to their low \textit{model volatility}. 1-shot IcL and 3-shot IcL leading to the highest average accuracy gains, at 15.8\% and 13.9\%, respectively. (Section \ref{scenario:counterfactual}).
    \item \textbf{Causal explanation generation (CEG).}
    CEG examines whether the LLMs can generate comprehensive and logically sound explanations that elucidate the cause-effect relationships between speciﬁc events. The causal scenario's \textit{understandability} is easy. Claude2, GPT-3.5-Turbo, and GPT-4 emerge as the top three models by average accuracy. Claude2, using EF, reaches a peak accuracy of 63.4\%, positioning the \textit{solvability} of this scenario as challenging since the \textit{top model-prompt pair} does not achieve an accuracy of 80\%. The models demonstrating the greatest variance in response to different prompts, as indicated by the highest \textit{model volatility}, include Koala (13B) and Llama2-chat (70B). In contrast, the models with the least variance are InternLM-chat (20B), Baichuan1 (7B), and Qwen (7B). Adversarial doubt and manual CoT as the top two prompts for average accuracy gain over the basic prompt (Section \ref{scenario:counterfactual}).
    \item \textbf{Effect of the treatment on the treated (ETT).}
    ETT assesses whether individuals who receive treatment are the ones who would derive the greatest advantage from it. This causal scenario has a hard \textit{understandability}. The leading three models in this causal scenario by average accuracy are GPT-4 at 40.9\%, GPT-3.5-Turbo at 39.0\%, and Claude2 at 35.6\%. GPT-4, when combined with manual CoT, reaches an impressive 89.9\%, suggesting this scenario's \textit{solvability} is potentially solvable, given that the \textit{top model-prompt pair} achieves over 80\%. The three most consistent models, marked by the the lowest \textit{model volatility}, are Baichuan1-chat (13B) with a \textit{model volatility} of 2.5, InternLM-chat (20B) at 4.3, and Baichuan2-chat (13B) at 7.8. Conversely, the models showing the highest sensitivity to prompt variations, as evidenced by the highest \textit{model volatility}, are Llama2 (13B) at 24.1, Llama2 (70B) at 23.8, and Llama2 (7B) at 23.7. The two prompts leading to the highest average accuracy improvements over the standard prompt are manual CoT with a gain of 30.4\% and 3-shot IcL at 16.7\% (Section \ref{scenario:counterfactual}). 
    \item \textbf{Natural direct effect (NDE).}
     NDE quantifies the direct influence of an intervention on an outcome, while keeping the mediator's natural state. This causal scenario's \textit{understandability} is regarded as hard. The \textit{top model-prompt pair} is GPT-4 with manual CoT, reaching an accuracy of 80.1\%, indicating that the \textit{solvability} of this scenario is potentially solvable as the \textit{top model-prompt pair}'s performance hits 80\%. The three most stable models, characterized by the lowest \textit{model volatility}, are Baichuan1-chat (13B) at 2.3, InternLM-chat (7B) at 3.0, and InternLM-chat (20B) at 3.1. Conversely, the three least stable models, exhibiting the highest \textit{model volatility} across different prompts, are Llama2 (13B) at 20.3, Llama2-chat (70B) at 18.2, and Llama2 (70B) also at 18.2. The leading two prompts achieving the most significant average accuracy improvements over the basic prompt are manual CoT at 19.1\% and 3-shot IcL at 9.9\% (Section \ref{scenario:counterfactual}).
    \item \textbf{Natural indirect effect (NIE).}
    NIE measures the extent of change in the outcome through the mediator when the treatment is modified. This causal scenario is considered to have a hard \textit{understandability}. The \textit{top model-prompt pair} is Koala (13B) with 3-shot IcL, achieving a 73.3\% accuracy, suggesting the \textit{solvability} of this scenario is challenging as the performance of the \textit{top model-prompt pair} surpasses the random guess but remains below 80\%. The three most stable models, characterized by the lowest \textit{model volatility}, are Baichuan1-chat (13B) at 2.4, Baichuan2-chat (13B) at 4.5, and Vicuna-v1.3 (33B) at 4.8. Conversely, the three most unstable models, showcasing the highest \textit{model volatility} across various prompts, are Llama2 (7B) at 30.8, Llama2 (13B) at 30.4, and Baichuan2-chat (7B) at 24.9, reflecting their pronounced sensitivity to prompt variations, reflecting their pronounced sensitivity to prompt variations. The two prompts leading to the highest average accuracy improvements over the basic prompt are 3-shot IcL at 29.3\% and manual CoT at 19.5\% (Section \ref{scenario:counterfactual}).
    \item \textbf{Probability of necessity (PN).}
    PN essentially seeks to address the question: ``\emph{In cases where the outcome occurs, could it still happen without the treatment?}'' The \textit{understandability} of PN scenario is considered as  very hard to understand. The three highest-performing models in terms of average accuracy within this causal scenario are GPT-4 at 14.5\%, GPT-3.5-Turbo at 8.1\%, and Llama2 (70B) at 5.2\%. The \textit{top model-prompt pair}, GPT-4 with manual CoT, achieves a significant 50.2\% accuracy, indicating the \textit{solvability} of this scenario is challenging as the performance of the \textit{top model-prompt pair} exceeds the random guess yet does not reach 80\%. The three most stable models, characterized by the lowest \textit{model volatility}, are Wizardcoder (15B) at 0.0, text-curie-001 at 0.1, and davinci (175B) at 0.3. Conversely, the three models showing the greatest instability across different prompts, indicated by the highest \textit{model volatility}, are GPT-4 at 15.2, GPT-3.5-Turbo at 11.6, and text-davinci-003 at 9.8, reflecting their pronounced sensitivity to prompt changes. The two prompts leading to the most substantial average accuracy improvements over the basic prompt are 3-shot IcL at 7.2\% and manual CoT at 6.1\% (Section \ref{scenario:counterfactual}).
    \item \textbf{Probability of sufficiency (PS).}
    PS addresses: ``\emph{In cases where the outcome does not occur, could it happen if a treatment exists?}'' This causal scenario's \textit{understandability} is  very hard. The leading three models in this causal scenario based on average accuracy are GPT-4 at 12.6\%, GPT-3.5-Turbo at 5.8\%, and text-davinci-003 at 4.6\%. The \textit{top model-prompt pair} is GPT-4 with manual CoT, achieving a score of 46.8\%, indicating that the \textit{solvability} of this scenario is challenging as the \textit{top model-prompt pair} exceeds the random guess yet does not reach 80\%. There are more than three models with zero \textit{model volatility} in the scenario. Conversely, the models exhibiting the greatest instability across various prompts, indicated by the highest \textit{model volatility}, are GPT-4 at 14.6, GPT-3.5-Turbo at 13.5, and text-davinci-003 at 11.2, showcasing their significant sensitivity to prompt variations. The two prompts leading to the highest average accuracy improvements over the basic prompt are manual CoT at 6.9\% and adversarial ignore at 0.2\% (Section \ref{scenario:counterfactual}).
\end{enumerate}

\subsection{Contributions}
\label{contributions}
In summary, we have the following contributions:
\begin{enumerate}
    \item \textbf{The CaLM framework}. 
    We introduce CaLM, a novel framework designed to systematically assess the causal reasoning capabilities of language models. It establishes a foundational taxonomy that integrates causal targets, adaptations, metrics, and error types, enabling a thorough navigation through the complex design space of causal reasoning assessment. By employing this well-defined taxonomy and its practical application, CaLM demonstrates unmatched flexibility and scalability in assessing language models' abilities to reason causally.

    \item \textbf{Wide coverage}. 
    Our taxonomy defines a wide-reaching, if not entire, design space for evaluating the causal reasoning capabilities of language models. Based on the taxonomy, we select and implement a core set of 92 causal targets, 9 adaptations, 7 metrics, and 12 types of errors. These 92 causal targets cover 46 distinct causal tasks spanning all four levels of the causal ladder, across three textual modes and in two languages. This constitutes the most thorough and detailed causal evaluation benchmark available to date. Furthermore, we conduct a systematic evaluation of 28 leading language models, including 15 open-access and 13 limited-access models from both academic and industrial sectors, using this benchmark.
    
    \item \textbf{Comprehensive analysis}.
    We conduct in-depth analyses of causal reasoning evaluation results across all the dimensions of causal scenario, mode, language, adaptation, model, metric, and error type. Our study detailedly examines the impact of these dimensions on the causal reasoning abilities of language models. Furthermore, our investigation delves into both intra-dimensional relationships (e.g., among various prompt types) and inter-dimensional relationships (e.g., between causal scenario and prompt types) within the context of causal reasoning. Moreover, we thoroughly analyse the impact of additional critical factors (e.g., model scale, model access, training strategy) on model performance. Beyond these overarching analyses, it is worth noting that we also deliver a thorough and detailed examination of each specific causal scenario, mode, language, adaptation, model, metric, and error type. 
    
    \item  \textbf{Empirical findings}. Our extensive evaluation, detailed in \nameref{experiments} (\cref{experiments}), yields 50 high-level empirical findings across 9 dimensions: model, adaptation, causal ladder, causal scenario, domain, mode, language, metric, and error. These findings confirm existing research in some instances and reveal new insights into contemporary language models in others. Such insights are instrumental for the development of future language models and pave the way for in-depth analysis. Importantly, our study extends beyond mere causal reasoning capabilities of these models, underscoring their broad applicability across varied use cases. We anticipate that this work will motivate researchers from different fields to further explore the implications of our findings or to identify new opportunities not yet addressed in our study. 
    
    \item \textbf{Dataset construction}. In light of the notable scarcity of datasets for causal evaluation of language models, we have composed comprehensive Symbolic and Mathematical datasets covering the causal scenarios specified in our study. We also augmented the existing datasets in the Natural mode, as further detailed in Section \ref{main:data}. This effort significantly mitigates \emph{training-test contamination}, thus enhancing the reliability of our findings. Specifically, our contributions to dataset construction are outlined from the following aspects: (1) \emph{Methodology of construction}: Our Symbolic and Mathematical mode datasets are intentionally designed to facilitate expansion, allowing for the generation of substantial new data should additional use cases emerge. (2) \emph{Dataset size}: Overall, our CaLM dataset contains 126,334 data samples. All samples in the Symbolic and Mathematical datasets are self-constructed, owing a total number of 38,400 and 44,800, respectively. Within the 43,134 samples that belong to the Natural mode, 13,567 samples come from open-source datasets, and the remaining 29,567 are self-constructed. Each sub-dataset within the Symbolic and Mathematical datasets consists of 1600 samples, striking a balance between thorough evaluation and cost-effectiveness. (3) \emph{Expanding causal scenarios}: We introduce four new causal scenarios in the Symbolic mode datasets (i.e., frontdoor adjustment set, instrumental variable, causal effect identification and causal attribution), and three in both the Natural and Mathematical mode datasets (i.e., controlled direct effect, probability of necessity and probability of sufficiency). (4) \emph{Enhancing existing work}: The Symbolic datasets include extensions to an existing causal scenario with three new domains, and the Mathematical datasets expand four existing causal scenarios with a total of eight new domains. (5) \emph{Linguistic expansion}: Both the Symbolic and Mathematical datasets are developed in both Chinese and English. For public datasets lacking Chinese versions, we provide our own translations and annotations.
    
    \item \textbf{Platform and codebase}. 
    We establish a comprehensive platform and codebase for evaluating the causal reasoning capabilities of language models, tailored to the diverse requirements of the research and development community. This platform features a website for easy access to resources and updates, leaderboards for benchmarking and fostering competition, curated datasets for testing models, and toolkits for systematic evaluation. These components ensure consistent, reproducible, and scalable assessments, adaptable to evolving research needs. 
\end{enumerate}

\subsection{Organization} 
\label{organization}

Our paper is structured into 13 sections. Following this introduction, we begin with providing an overview of the prerequisite knowledge in \nameref{main:preliminary} (\cref{main:preliminary}). From \nameref{main:target} (\cref{main:target}) through \nameref{errors} (\cref{errors}), we elaborate on each module within the CaLM framework. \nameref{main:model} (\cref{main:model}) presents the models used for evaluation, followed by an in-depth analysis of the experimental results in \nameref{experiments} (\cref{experiments}). \nameref{related} (\cref{related}) discusses the work related to CaLM. \nameref{gap} (\cref{gap}) is dedicated to the components not included in the concrete implementation. The limitations and future directions for CaLM are explored in \nameref{limitations} (\cref{limitations}). Finally, the paper concludes with a summary in \nameref{main:conclusion} (\cref{main:conclusion}).

Specifically, the rest of the paper is organized as follows: 
\begin{itemize}
    \item In \cref{main:preliminary}: \nameref{main:preliminary}, we present the foundational elements that are essential for building this paper. This section is primarily comprised of two parts: \nameref{preliminary:ladder} (\cref{preliminary:ladder}) and \nameref{preliminary:scm} (\cref{preliminary:scm}).
    
    \item In \cref{main:target}: \nameref{main:target}, we introduce the causal targets, starting with a broad overview of the \nameref{target:taxonomy} (\cref{target:taxonomy}). Following this, \nameref{target:selection} (\cref{target:selection}) details the specific causal targets chosen for our evaluation. The narrative then progresses from \nameref{main_scenario:CD} (\cref{main_scenario:CD}) to \nameref{main_scenario:counterfactual} (\cref{main_scenario:counterfactual}), where each causal scenario within the causal targets is thoroughly explained, aligned with the incremental levels of the causal ladder.
    
    \item In \cref{main:data}: \nameref{main:data}, we delve into the datasets used in CaLM. We start by outlining the open-source and self-constructed datasets we employed in \nameref{data:selection} (\cref{data:selection}). Next, \nameref{data:construction} (\cref{data:construction}) elaborates on the process involved in developing our self-constructed datasets. Concluding this part, \nameref{data:statistics} (\cref{data:statistics}) presents an extensive statistical breakdown of the datasets to assist future users.
    
    \item In \cref{adaptation}: \nameref{adaptation}, we describe the prompts used to interact with the model. The various categories of prompts are introduced in \nameref{adaptation:taxonomy} (\cref{adaptation:taxonomy}), and the specific prompts selected are explained in \nameref{adaptation:selection} (\cref{adaptation:selection}). The last five sections (e.g., \nameref{adaptation:basic} (\cref{adaptation:basic}), \nameref{adaptation:cot} (\cref{adaptation:cot})) offer a comprehensive overview of the five primary types and nine subtypes of prompts that we employed.
    
    \item In \cref{main:metrics}: \nameref{main:metrics}, we present the metrics used for evaluating the model's causal reasoning ability. Likewise, we initially explain the categorization of current metrics from a broad viewpoint in \nameref{metric:taxonomy} (\cref{metric:taxonomy}), followed by a discussion in \nameref{metric:selection} (\cref{metric:selection}) on the considerations that influenced our selection of metrics. Lastly, we elaborate on the metrics employed, examining them from three distinct angles (i.e., \nameref{metric:model} (\cref{metric:model}), \nameref{metric:scenario} (\cref{metric:scenario}) and \nameref{metric:prompt} (\cref{metric:prompt})).
    
    \item In \cref{errors}: \nameref{errors}, we consolidate the errors made by models throughout the evaluation. We introduce in \nameref{error:taxonomy} (\cref{error:taxonomy}) how these errors are currently categorized. Following this, we proceed to elaborate on these errors in detail, dividing our discussion into two key parts: \nameref{error:quantitative} (\cref{error:quantitative}) and \nameref{error:qualitative} (\cref{error:qualitative}).
    
    \item In \cref{main:model}: \nameref{main:model}, we describe the language models that are evaluated in CaLM. we categorize models in \nameref{model:taxonomy} (\cref{model:taxonomy}) based on different scales, creators, access, etc. Subsequently, the main considerations guiding our model selection are outlined in \nameref{model:selection} (\cref{model:selection}).
    
    \item In \cref{experiments}: \nameref{experiments}, we provide a comprehensive and in-depth analysis of our experiment results, leading to a wealth of insightful conclusions, establishing it as one of our most significant sections. The main focus is \nameref{experiment:main} (\cref{experiment:main}), where we undertake analysis from various angles (e.g., \nameref{main:comparison} (\cref{main:comparison}), \nameref{main:complexity} (\cref{main:complexity})). Then, in \nameref{experiment:prompt} (\cref{experiment:prompt}), we analyze from the perspective of different prompts, offering an alternative viewpoint. Finally, exhaustive analyses of the model and the causal scenario are separately carried out in \nameref{experiment:model} (\cref{experiment:model}) and \nameref{experiment:scenario} (\cref{experiment:scenario}).
    
    \item In \cref{related}: \nameref{related}, we illustrate the works that pave the way for the development of CaLM. Initially, we revisit the rapid advancement of language models in recent years, as detailed in \nameref{related:advancement} (\cref{related:advancement}). Next, we present \nameref{related:general} (\cref{related:general}), adopting a wider view to scrutinize how language models are evaluated. Furthermore, we introduce \nameref{related:causal} (\cref{related:causal}). It is with these exemplary works as a foundation that the construction of CaLM becomes possible. Given that datasets play a crucial role in benchmarks, we conclude by discussing related \nameref{related:dataset} (\cref{related:dataset}).
    
    \item In \cref{gap}: \nameref{gap}, we cautiously review the gaps in the current implementation. Starting from each module of the CaLM framework (i.e., \nameref{subsec:gap_targets} (\cref{subsec:gap_targets}), \nameref{subsec:gap_adaptations} (\cref{subsec:gap_adaptations}), \nameref{subsec:gap_metrics} (\cref{subsec:gap_metrics}), \nameref{subsec:gap_errors} (\cref{subsec:gap_errors}), and \nameref{subsec:gap_models} (\cref{subsec:gap_models})), we analyze the disparities between our taxonomy and selection, hoping to provide guidance for potential future research. Besides, due to the rapid development of language models, the challenge of incorporating the latest models into CaLM arises, leading us to summarize the \nameref{subsec:gap_models} (\cref{subsec:gap_models}).   
    
    \item  In \cref{limitations}: \nameref{limitations}, we outline the existing limitations of CaLM. Our analysis is conducted from two specific aspects, \nameref{limitation:implementation} (\cref{limitation:implementation}) and \nameref{limitation:result} (\cref{limitation:result}), through which we endeavor to suggest several potential strategies for enhancement.

    \item In \cref{main:conclusion}: \nameref{main:conclusion}, we summarize the entire paper and convey our vision for a brighter future.
\end{itemize}

\clearpage


\section{Preliminaries}
\label{main:preliminary}
This section establishes the foundation for our analysis of causal evaluation in language models, by introducing key concepts and terminologies in causal reasoning essential for understanding the subsequent discussion. It is structured into two main subsections: \nameref{preliminary:ladder} (\cref{preliminary:ladder}) and \nameref{preliminary:scm} (\cref{preliminary:scm}).

\subsection{The Ladder of Causation}
\label{preliminary:ladder}

The \emph{Ladder of Causation}, as introduced by \citet{pearl2018book} and discussed by \citet{bareinboim2022pearl}, is a conceptual framework that illustrates the hierarchy of causal reasoning tasks. It consists of three discernible rungs: \emph{association (Rung 1)}, \emph{intervention (Rung 2)}, and \emph{counterfactuals (Rung 3)}, with each representing a progressively deeper understanding of causality. Additionally, we integrate causal discovery tasks \citep{spirtes2000causation,peters2017elements} into this ladder, acknowledging them as a fundamental phase in causal reasoning \citep{glymour2019review}. For ease of reference and clarity in our ongoing discussion within the CaLM framework, we identify \emph{(causal) discovery} as \emph{Rung 0} on the ladder of causation.

\paragraph{Rung 0: Causal discovery.} 
This rung prioritizes analyzing statistical patterns solely from observational data when the causal graph is not known, aiming to identify cause-effect pairs. This process of deducing the underlying causal structure from data is referred to as causal discovery \citep{spirtes2000causation,peters2017elements,glymour2019review,zanga2022survey}. For example, ``\emph{Is there a causal relationship between review frequency and exam scores?}''

\paragraph{Rung 1: Association.} This rung is dedicated to exploring statistical dependencies among variables. These dependencies can be effectively modeled using Bayesian Networks \citep{pearl1988probabilistic,goertzel2008probabilistic}, which depict a set of variables and their conditional relationships, denoted as $P(Y = y|X = x)$, through a directed acyclic graph (DAG). At this rung, the questions asked are primarily of the form ``\emph{What is?}". For instance, ``\emph{What is the appropriate review frequency for me to effectively preparing for exams?}'' 
Queries on this rung can be answered based on observational data. 

\paragraph{Rung 2: Intervention.} The second rung moves beyond mere observation to the effects of interventions. It allows ones to take actions or intervene on variables in environments, and then predict the effects of those deliberate interventions. This is about asking ``\emph{What if I do this?}''. For example, ``\emph{What if I review every day, will my exam scores improve?}''
Different from seeing or observing on Rung 1, actively intervening on variables remove the effect of any other potential factors on those variables, which ensures that the true causal effects are estimated. We represent interventions using the do-operator in the form of $P(Y = y|do(X = x))$, representing the distribution of $Y$ when intervening on $X$ to fix its value at $x$.

\paragraph{Rung 3: Counterfactuals.} The highest rung involves considering counterfactuals - questions of the type ``\emph{What if I have done this instead?}''. For instance, ``\emph{What if I have attended a party instead of reviewing, would my exam scores be good?}'' This involves reasoning about hypothetical alternative scenarios in which the world might have unfolded differently. Counterfactual probabilities are expressed as $P(Y_x = y)$, signifying the likelihood that ``$Y$ would be $y$ if $X$ had been $x$.'' Note that, counterfactual reasoning requires the use of structural causal models (SCMs) \citep{pearl2009causality}, which are essential for understanding and analyzing how variables interact within a system under hypothetical scenarios.

\subsection{Structural Causal Models}
\label{preliminary:scm}

Causal models are constructed based on deterministic, functional relationships among variables of interest, with each relationship representing an autonomous mechanism. In this section, we give a formal definition of causal models \citep{pearl2009causality}. 

\begin{dfn}[\textbf{Structural Causal Models}]

A structural causal model (SCM), denoted by $\mathcal{M}$, is a triple
\begin{align}
    \mathcal{M} = \langle U, V, F, P(u)\rangle,
\end{align}
where:
\begin{itemize}
    \item[(i)] $U$ is a set of background variables, (also called exogenous), that are determined by factors outside the model;
    \item[(ii)] $V$ is a set $\{V_1, V_2, \ldots, V_n\}$ of variables, called endogenous, that are determined by variables in the model (i.e., variables in $U \cup V$);
    \item[(iii)] $F$ is a set of functions $\{f_1, f_2, \ldots, f_n\}$ such that each $f_i$ is a mapping from the respective domains of $U_i \cup \textbf{PA}_i$ to $V_i$, where $U_i \subseteq U$ and $\textbf{PA}_i \subseteq V \setminus V_i$ and the entire set $F$ forms a mapping from $U$ to $V$. In other words, each $f_i$ in
    \begin{align}
        v_i = f_i(pa_i, u_i), \quad i=1, \ldots, n, \nonumber
    \end{align}
    assigns a value to $V_i$ that depends on the values of a select set of variables in $V \cup U$, and the entire set $F$ has a unique solution $V(u)$;
    \item[(iv)] $P(u)$ is a probability distribution over exogenous variables.
\end{itemize}
Each structural causal model $M$ is associated with a DAG $\mathcal{G}$, where each vertex corresponds to a variable and the directed edges pointing from $U_i \cup \textbf{PA}_i$ to $V_i$ represent functional relationships in which $U_i \cup \textbf{PA}_i$ appears in the argument of the function of $V_i$.
\end{dfn}

\clearpage


\section{Causal Targets}
\label{main:target}
As introduced in \nameref{intro:framework} (\cref{intro:framework}), a causal target is defined as a triplet consisting of (\emph{causal task}, \emph{mode}, \emph{language}), where a causal task is also structured as a triplet: (\emph{causal ladder}, \emph{causal scenario}, \emph{domain}). Each component of these triplets will be thoroughly dissected in this section. We begin by outlining their broad classifications in \nameref{target:taxonomy} (\cref{target:taxonomy}), followed by a discussion of the specific elements chosen for evaluation in \nameref{target:selection} (\cref{target:selection}). Finally, we delve into each causal scenario based on the hierarchical levels of the causal ladder (i.e., \nameref{main_scenario:CD} (\cref{main_scenario:CD}), \nameref{main_scenario:association} (\cref{main_scenario:association}), \nameref{main_scenario:intervention} (\cref{main_scenario:intervention}), and \nameref{main_scenario:counterfactual} (\cref{main_scenario:intervention})).

\subsection{Taxonomy}
\label{target:taxonomy}
In this section, we define the design space for \nameref{target:causal_task}, \nameref{target:mode} and \nameref{target:language} on a macro level. We aim to establish a comprehensive space for the causal target that can be further refined and filled through future research endeavors. 

\subsubsection{Causal Task}
\label{target:causal_task}
A causal task specifies the particular function of causal reasoning that a language model is expected to perform, structured as a triplet: (\emph{causal ladder}, \emph{causal scenario}, \emph{domain}). The relationships among these three elements are illustrated in Figure \ref{fig_intro:causal_task}.

\paragraph{Causal ladder.}
The causal ladder, a crucial dimension in our taxonomy, consists of four rungs: \emph{causal discovery}, \emph{association}, \emph{intervention}, and \emph{counterfactuals}. The four rungs of causal ladder cover a spectrum of challenges relevant to causal reasoning.

\paragraph{Causal scenario.}
A causal scenario illustrates how causal concepts can be applied in real-world or research settings, such as natural direct effect (NDE), controlled direct effect (CDE), and probability of necessity (PN). Each scenario is uniquely associated with one of the four levels of the causal ladder. This correspondence streamlines the evaluation process and facilitates a more nuanced understanding of language models' performance in causal reasoning.

\paragraph{Domain.}
The domain refers to the \emph{dataset} and \emph{question type} in our context. Our datasets are classified into two categories: \emph{open-source} and \emph{self-constructed}. Open-source domains utilize existing datasets that align with our predefined causal tasks, thereby enhancing the broad applicability of our research and facilitating comparisons with prior studies. In contrast, self-constructed domains are created for causal tasks that either have limited existing data points or lack publicly available datasets. This dual approach ensures comprehensive domain coverage for our evaluations. Additionally, we design four question types, which are \emph{binary classification}, \emph{choice selection}, \emph{open-ended generation}, and \emph{probability calculation}. This flexibility allows us to extensively explore and evaluate the effectiveness of language models across various causal tasks, contributing significantly to our understanding of model performance under various experimental settings.

\subsubsection{Mode}
\label{target:mode}
In the realm of AI systems, the integration of various data types has led to the identification of four common modes: Text, Code, Image, and Video, each of which is characterized by its unique features and supports a wide range of causal tasks~\citep{lu2024gpt}. These modes, each serving a unique purpose, contribute significantly to the holistic understanding and processing of diverse information, enhancing the functionality and applicability of AI across different contexts.

\paragraph{Text mode.}
Text mode focuses on causal tasks related to natural language processing, where the input and output are primarily composed of textual information. This mode involves understanding, generating, and manipulating text, making it essential for applications such as language translation \citep{zhu2023multilingual}, sentiment analysis \citep{chen2020relation}, and information extraction \citep{chen-etal-2022-ergo}.

\paragraph{Code mode.}
Code mode is specifically designed for handling programming languages and source code. It involves causal tasks related to code generation \citep{ji2023benchmarking}, comprehension \citep{gao2023two}, and analysis \citep{rodriguez2023benchmarking}. This mode is crucial for applications in automated coding \citep{kazemitabaar2023novices}, debugging \citep{lee2023github}, and software comprehension \citep{yuan2023evaluating}, allowing AI systems to engage with and interpret programming instructions effectively.

\paragraph{Image mode.}
Image mode focuses on causal tasks involving static visual data. This includes causal tasks such as visual recognition \citep{mao2022causal}, image classification \citep{yang2023treatment}, and image generation \citep{li2024image}. The AI system processes pixel-based information to understand visual content, making it instrumental in applications such as medical imaging analysis \citep{taher2022caid} and image synthesis \citep{rombach2022high}.

\paragraph{Video mode.}
Video mode extends the capabilities of AI systems to dynamic visual data. It involves causal tasks related to video understanding \citep{huang2023causalainer}, action recognition \citep{liu2024knowledge}, and temporal analysis \citep{chen2021spatial}. Video mode enables AI systems to interpret and respond to sequences of frames, contributing to applications like video summarization and content understanding \citep{zhang2016video, bertasius2021space}.

Understanding and effectively utilizing these four modes - Text, Code, Image, and Video - in AI systems provide a comprehensive approach to handling diverse types of data. This taxonomy lays the foundation for developing versatile AI models capable of addressing a wide range of causal tasks in a complex context.

\subsubsection{Language}
\label{target:language}

The global population, consisting of billions of people, communicates through a multitude of languages \citep{nordhoff2011glottolog,hammarstrom2021glottolog,bommasani2021opportunities,liang2022holistic}. Despite this linguistic diversity, in the field of artificial intelligence and natural language processing, the majority of efforts are concentrated on a handful of linguistically resource-rich languages, such as English and Chinese. Acknowledging this linguistic imbalance, we refrain from extensively categorizing the world's languages. Instead, our primary focus lies in evaluating models and causal tasks in English and Chinese.

\subsection{Concrete Implementation}
\label{target:selection}
Building upon \nameref{target:taxonomy} (\cref{target:taxonomy}), this section elaborates on the specific component we select to assess the model's causal reasoning capability (i.e., \nameref{selection:causal_task} (\cref{selection:causal_task}), \nameref{selection:mode} (\cref{selection:causal_task}) and \nameref{selection:language} (\cref{selection:language})). We carefully select these components based on their applicability and importance, ensuring that they reveal the model's core strengths and limitations. 

\subsubsection{Causal Task}
\label{selection:causal_task}
Concerning the scope of CaLM, our ideal goal is to evaluate language models across every causal task, represented by the tuple: (\emph{causal ladder}, \emph{causal scenario}, \emph{domain}). However, as our taxonomy indicates, the realms of causal scenario and domain are both extensive and diverse. Therefore, our objective is not to cover every conceivable causal task, but rather to concentrate on assessing the most critical aspects.
\paragraph{Causal ladder.}
Our causal tasks cover all four rungs of the causal ladder: causal discovery, association, intervention, and counterfactuals. By considering the entire causal ladder, we can thoroughly evaluate the model’s causal reasoning capabilities from the most foundational to the most complex levels, offering a comprehensive understanding of its performance across different types of causal analyses.

\paragraph{Causal scenario.}
In selecting causal scenarios, we have the following considerations. Since our exclusive focus on language models, we eliminate any causal scenarios that involve multiple modalities. We prioritize the most fundamental and essential causal scenarios across each rung of the causal ladder. We believe these scenarios are crucial for assessing and enhancing the causal inference capabilities of language models, with significant potential societal impact. This targeted approach leads to the selection of the following causal scenarios (Figure \ref{fig_intro:causal_task}): 
\begin{itemize}
\item \textbf{Causal discovery}: \emph{pairwise causal discovery} (PCD), \emph{event causality identification} (ECI), \emph{abstract reasoning} (AR), and \emph{causal attribution} (CA). PCD has already attracted a considerable number of researchers to evaluate models \citep{gao2023chatgpt,kiciman2023causal,vashishtha2023causal,long2022can,tu2023causal}, with some studies focusing on ECI \citep{gao2023chatgpt} and AR \citep{zevcevic2023causal,willig2022can}. Although there has not yet been an evaluation for CA, this should not be interpreted as diminishing the importance of CA. In fact, it highlights the model's capacity for attributing causes to events, a capability that is critically applied in areas like social psychology \citep{malle2022attribution,langenhoff2021predicting}, marketing \citep{mero2020effectual,tang2020inspire}, and epidemiology \citep{richens2020improving,shimonovich2021assessing}. 
\item \textbf{Association}: \emph{correlation} (CORR) and \emph{explaining away effect} (EAE). The two causal tasks on this rung are proposed by \citet{jin2023cladder} and have been evaluated on some models (e.g., LLaMa, Alpaca, \gptf). Building on this, we conduct a comprehensive evaluation of all models on these two scenarios in CaLM. 
\item \textbf{Intervention}: \emph{average treatment effect} (ATE), \emph{backdoor adjustment set} (BAS), \emph{frontdoor adjustment set} (FAS), \emph{instrumental variable} (IV), \emph{causal effect identification} (CEI), \emph{controlled direct effect} (CDE), and \emph{collider bias} (CB). 
In \citet{jin2023cladder}, ATE, BAS, CB and CDE have already been evaluated in Natural mode. Building on this, we extend the evaluations by adding Mathematical mode for ATE and CDE, and introducing Symbolic mode for BAS. 
Currently, there are no studies evaluating models on FAS, IV, and CEI. Front-door adjustment aims to identify FAS and estimate the causal effects when unobserved confounders exit. Effectively handling FAS is significant in areas such as computer vision \citep{yang2021deconfounded,yang2021causal}, economics \citep{imbens2020potential}, and social policy \citep{matthay2022causal}. 
IV is dependent of the treatment and influences the outcome only through the treatment. IV can be used to estimate causal effects \citep{angrist1996identification}, and it is applied in various domains, including psychology \citep{maydeu2020estimating}, policy analysis \citep{marbach2020profiling}, and biology \citep{birney2021mendelian}. While all causal effects are identifiable in the absence of unobserved confounders, the real world often features such confounders, complicating the causal scenarios. The CEI assesses a model's ability to determine whether a causal effect can be uniquely established from the distribution of observed variables, regardless of any unobserved factors \citep{tian2002general}. This capability is particularly useful in tackling challenges in fields such as environmental epidemiology \citep{yu2021identification}, economics \citep{uysal2015doubly}, and meteorology \citep{pfleiderer2020robust}. 
\item \textbf{Counterfactuals}: \emph{actual causality} (AC), \emph{causal explanation generation} (CEG), \emph{effect of the treatment on the treated} (ETT), \emph{natural direct effect} (NDE), \emph{natural indirect effect} (NIE), \emph{probability of necessity} (PN), \emph{probability of sufficiency} (PS), and \emph{counterfactual reasoning} (CR). Among these scenarios, AC, CEG, ETT, NDE, NIE, and CR have already been evaluated on some models in previous studies \citep{suzgun2022challenging,frohberg2022crass,kiciman2023causal,gao2023chatgpt,jin2023cladder,kiciman2023causal}. For AC, CEG, and CR, we continue to use the datasets employed in existing work. For ETT, NDE, and NIE, we additionally introduce evaluations in Mathematical mode. PN and PS are two important concepts in causal reasoning that have not yet been evaluated. PN refers to the probability that an outcome would not have happened without its cause, given the outcome has already happened. PS is the chance of an outcome would have happened if its cause happened, with the outcome has not happened yet. These concepts are pivotal in guiding domains from social science \citep{kuppens2003appraisal} to computer science \citep{yang2024invariant}.
\end{itemize}

\paragraph{Domain.}
In the selection of domains, we primarily focus on two key areas: \emph{question types} and \emph{datasets}. (1) Regarding question types, we choose four kinds that are broadly used and emphasize different aspects of reasoning. These include: \textbf{Binary classification}, which requires the model to provide a deterministic ``yes'' or ``no'' response, testing its ability to make clear-cut decisions; \textbf{Choice selection}, which asks the model to select the correct answer from a set of given options that can range from 2 to 4 choices, assessing its selection accuracy; \textbf{Probability calculation}, which involves the model calculating and presenting an answer in numerical form based on probabilities, testing tis quantitative reasoning; \textbf{Open-ended generation}, which challenges the model's ability to generate relevant explanations without any constraints on format, evaluating its creative and comprehensive response capabilities.
(2) Considering the datasets, despite the scarcity of causal reasoning datasets, we have endeavored to comprehensively cover all causal scenarios we aim to assess. This has been achieved by both utilizing existing datasets and creating our own datasets. For the selection of \textbf{open-source datasets}, we primarily refer to the existing evaluations mentioned in the \textbf{causal scenario}, aligning with the datasets utilized by prior studies. This approach offers two key benefits: first, it ensures that the chosen datasets appeal to the broadest audience possible; second, it enhances the reliability of the CaLM findings by allowing comparisons with results from datasets used in existing research.
For \textbf{self-constructed datasets}, particularly for causal tasks with existing datasets that have fewer than 1000 data points (such as NDE, NIE) and those lacking public datasets, we construct our own.
We will detail the process used to select and construct specific datasets for each causal scenario in \nameref{main:data} (\cref{main:data}).

\subsubsection{Mode}
\label{selection:mode}

Given that CaLM is designed for language models, our selection of modes carefully consider three distinct subcategories within the text mode: \emph{Natural}, \emph{Symbolic}, and \emph{Mathematical}. \textbf{Natural} includes the conventional causal tasks that are articulated and responded to in the language commonly used by people. This mode focuses on the intuitive, everyday use of language, facilitating the assessment of how effectively language models understand and generate responses that align with typical human communication. \textbf{Symbolic} refers to the causal tasks presented in a Symbolic form that does not contain specific physical meaning (e.g., a causal graph represented within Symbolic mode would be ``\emph{A causes B, B causes D, C causes D}''). The model's responses are given in a mixed format of natural language and Symbolic representation. Using symbols to represent variables serves a dual purpose: (1) Firstly, it aligns with traditional cognitive reasoning \citep{garcez2008neural}, where abstract symbols and logical structures enable reasoning beyond specific contexts. This approach leverages the generality and clarity of Symbolic representations, facilitating logical inference and conceptual manipulation without the ambiguities of natural language. (2) Secondly, this Symbolic representation effectively prevents the model from memorizing biases within the training data, offering a more accurate measure of the model's genuine causal inference capabilities. By abstracting variables into symbols, the focus shifts from content memorization to the application of logical reasoning, providing a clearer evaluation of the model’s ability to deduce causality from causal graph. The usage of Symbolic mode not only assesses the model's reasoning skills in a controlled environment but also paves the way for the development of models that are both more robust and capable of generalizing beyond their training datasets. \textbf{Mathematical} consists of causal tasks that involve math concepts, requiring the model to execute mathematical operations and respond with answers in both probabilistic values and natural language. Numerous studies have evaluated the mathematical abilities of language models, revealing that although these models excel in many natural language processing tasks, they still face significant challenges when solving mathematical problems \citep{hendrycks2021measuring,cobbe2021training,lightman2023let,bubeck2023sparks,dao2023investigating,wei2023cmath,wu2023empirical,yuan2023well,yu2023metamath}. The reason we employ Mathematical mode is that mathematical reasoning is fundamental to assessing the cognitive capabilities that underpin human intelligence. This mode tests models beyond mere linguistic fluency, probing their logical structure and capacity for conceptual understanding. Employing Mathematical mode not only highlights the current capabilities and limitations of language models in mimicking sophisticated cognitive functions, but also guides the development of more advanced models capable of complex thought processes akin to human reasoning.  

\subsubsection{Language}
\label{selection:language}

Despite the global proliferation of language models, there is a pressing need to expand evaluation across a broader array of languages. However, due to the associated costs and the risk of evaluation sets becoming unwieldy in size, we have decided to concentrate our efforts on two languages: English and Chinese. The selection of these two languages is primarily based on two considerations: (1) As explained in \citet{liang2022holistic}, English and Chinese are the languages most frequently used in the fields of artificial intelligence and natural language processing; (2) As we have statistically analyzed in \nameref{main:model} (\cref{main:model}), the training corpora for current models are also predominantly composed of these two languages.

\subsection{Rung 0: Causal Discovery}
\label{main_scenario:CD}
In the field of \emph{causal discovery (Rung 0)}, our evaluation concentrates on four causal scenarios: \nameref{cd:pcd} (\cref{cd:pcd}), \nameref{cd:eci} (\cref{cd:eci}), \nameref{cd:ar} (\cref{cd:ar}), and \nameref{cd:ca} (\cref{cd:ca}). We will now delve into a detailed explanation of each causal scenario. 

\subsubsection{Pairwise Causal Discovery (PCD)}
\label{cd:pcd}
Causal discovery focuses on understanding the cause-and-effect relationships among different variables \citep{peters2017elements}. In many scientific disciplines, the primary goal is to uncover causal relationships and elucidate fundamental mechanisms. While randomized experiment is widely regarded as the gold standard for establishing causal relationship \citep{hariton2018randomised}, it can prove challenging or unfeasible in certain contexts \citep{spirtes2016causal}. Pairwise causal discovery, which aims to discern underlying pairwise causal relationships solely from observational data, has attracted increasing attention across various domains, such as earth science \citep{pmlr-v150-melkas21a}, climate system \citep{runge2019detecting} and biology \citep{amar2021graphical}. Unlike traditional causal discovery methods that rely on the actual data values of these variables, language models can deduce this causal framework by analyzing metadata related to the variables. This can include the variable's name or the context in which the problem is described using natural language \citep{kiciman2023causal,willig2022can}. 

\begin{figure}[t]
    \centering
    \includegraphics[width=\textwidth]{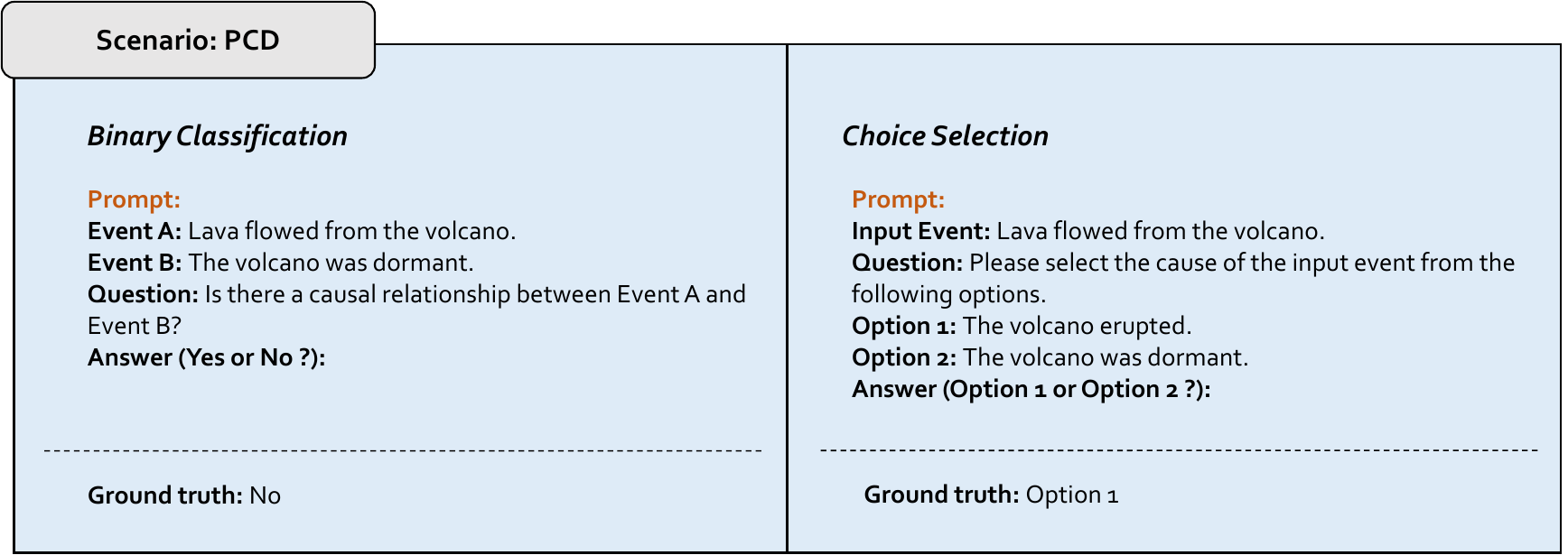}
    \caption[Example of pairwise causal discovery]{\textbf{Example of pairwise causal discovery.}}
    \label{fig_scenario:PCD}
\end{figure}

\paragraph{Causal scenario setting.}
PCD seeks to establish if a causal relationship exists between two given events and to identify which of the two is the cause and which is the effect 
\citep{gao2023chatgpt}. In PCD, our questions primarily manifest in two ways. (1) \textbf{Binary classification}: We present pairs of events along with associated inquiries (e.g., ``\emph{Event A: Lava flowed from the volcano. Event B: The volcano was dormant. Question: Is there a causal relationship between Event A and Event B?}''\footnote{The example is chosen from \citet{roemmele2011choice}.}). The objective is to accurately discern whether a causal relationship exists between the two events (the correct response being ``No''). (2) \textbf{Choice selection}: Models must select the cause or effect of a given event from two provided options (e.g., ``\emph{Input Event: Lava flowed from the volcano. Question: Please select the cause of the input event from the following options. Option 1: The volcano erupted. Option 2: The volcano was dormant.}''\footnote{The example is chosen from \citet{roemmele2011choice}.}). Here, the causal scenario explicitly outlines the presence and direction of the causal relationship, challenging the model to determine which option is more probable (in this example, ``Option 1'' is correct). See Figure \ref{fig_scenario:PCD} for a detailed illustration.

\subsubsection{Event Causality Identification (ECI)}
\label{cd:eci}

This causal scenario is designed to pinpoint the events mentioned in a text and understand whether there exists a casual relationship between them. Research on ECI has a long-standing history and it holds a vital position in comprehending text deeply \citep{gao-etal-2019-modeling}. Thus, enhancing the comprehension of event causality can significantly benefit various applications in natural language processing \citep{liakhovets2022zero,liu2023cross}. Language models tasked with this scenario should be adept at utilizing a broad range of commonsense knowledge and capable of comprehending complex contexts involving multiple entities and events. Ultimately, these models are expected to synthesize all this information to accurately determine causal relationships \citep{gao2023chatgpt}.

\begin{figure}[t]
    \centering
    \includegraphics[width=0.6\textwidth]{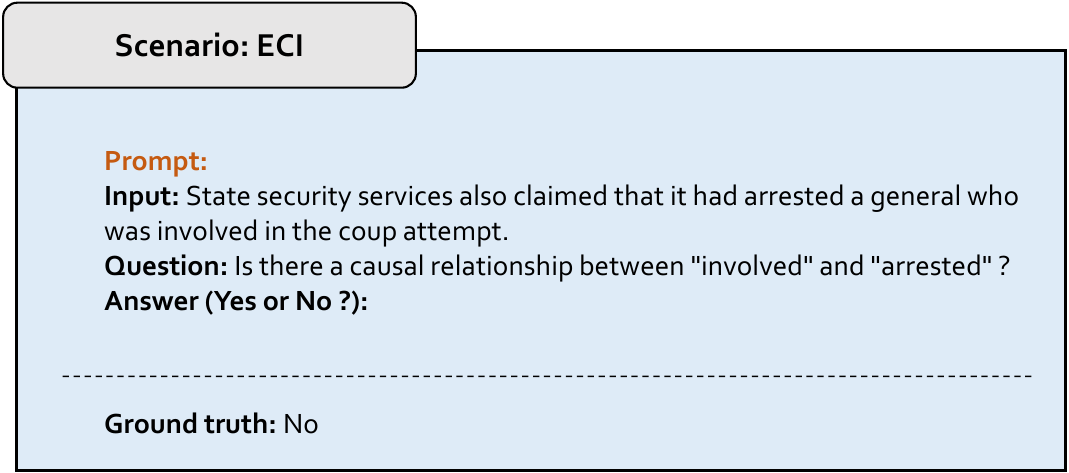}
    \caption[Example of event causality identification]{\textbf{Example of event causality identification.}}
    \label{fig_scenario:ECI}
\end{figure}

\paragraph{Causal scenario setting.}
ECI requires the models to assess whether there is a causal relationship between two events within a given sentence. For instance, consider the sentence: ``\emph{State security services also claimed that it had arrested a general who was involved in the coup attempt.}''\footnote{The example is chosen from \citet{wang2022maven}.} An example task would involve examining the events ``\emph{involved}'' and ``\emph{arrested}'' to determine if one caused the other. In this case, the correct answer is ``No'', as humans can easily discern these as two separate events without a direct cause-and-effect relationship between them. See Figure \ref{fig_scenario:ECI} for a detailed illustration.

\subsubsection{Abstract Reasoning (AR)} 
\label{cd:ar}

AR investigates the capability of language models to identify and understand causal relationships within Symbolic causal graphs \citep{zevcevic2023causal}. This scenario tests how well models can accurately discern potential causal relationships beyond simply memorizing information based on position or sequence. In scenarios where models only memorize such inseparable information, it becomes challenging to integrate these disparate pieces of data into a coherent and consistent causal graph \citep{willig2022can}. 
With strong AR capabilities, language models can better predict outcomes, design interventions, and understand the potential implications of changes within the system, thus enhancing decision-making and problem-solving abilities.

\begin{figure}[t]
    \centering
    \includegraphics[width=0.6\textwidth]{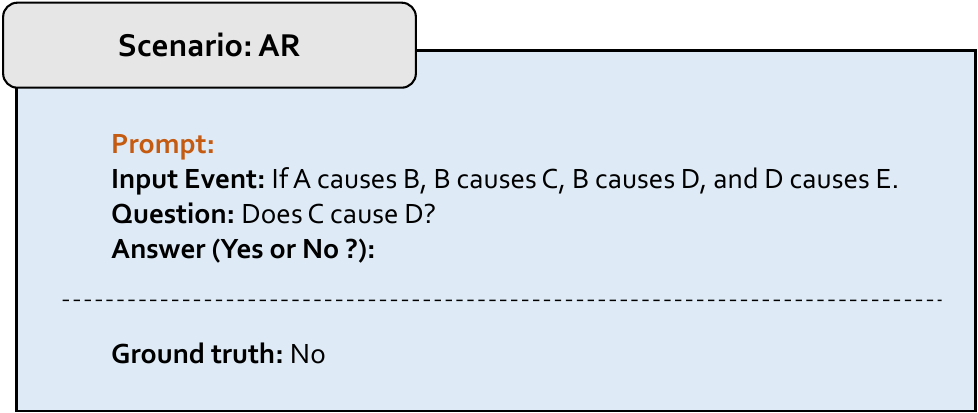}
    \caption[Example of event abstract reasoning]{\textbf{Example of abstract reasoning.}}
    \label{fig_scenario:AR}
\end{figure}

\paragraph{Causal scenario setting.}
In AR, models are tasked with determining whether there exists a causal relationship between two nodes in a given graph. For example, consider a graph where ``\emph{A causes B, B causes C, B causes D, and D causes E.}'' The models must assess whether there is a direct causal link between nodes ``\emph{C}'' and ``\emph{D}''. In this case, the correct answer is ``No''. See Figure \ref{fig_scenario:AR} for a detailed illustration.

\subsubsection{Causal Attribution (CA)}
\label{cd:ca}

Causal attribution refers to the process of determining which specific factor is responsible for an outcome. Its significance spans across numerous domains, intersecting with research in psychology \citep{graham2020attributional}, medical diagnosis \citep{richens2020improving}, and organizational science \citep{harvey2014attribution}. In this causal scenario, the model needs to accurately recognize causal graphs and uncover the precise causal relationships within them. Assessing the model's capacity for causal attribution aids in comprehending its decision-making abilities, thus establishing a foundation for its practical deployment in real-world contexts.

\begin{figure}[t]
    \centering
    \includegraphics[width=0.6\textwidth]{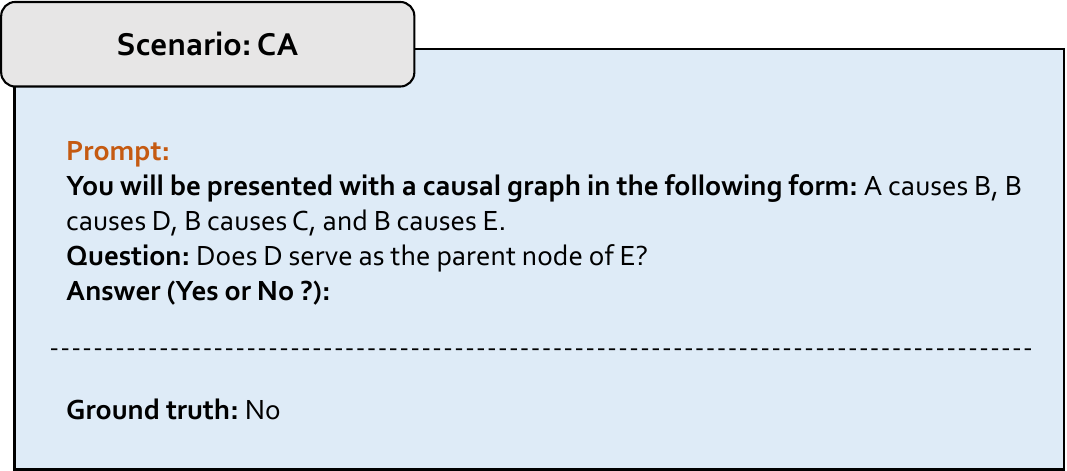}
    \caption[Example of causal attribution]{\textbf{Example of causal attribution.}}
    \label{fig_scenario:CA}
\end{figure}

\paragraph{Causal scenario setting.}
In CA, models are provided with a causal graph (e.g., ``\emph{A causes B, B causes D, B causes C, and B causes E}'') and face two domains:
(1) Find parent: This task requires the model to determine the parent of a specified node, focusing on identifying direct causal relationships. For instance, the question might ask, ``\emph{Does D serve as the parent node of E?}'' In this case, the correct answer is ``No''.
(2) Find ancestor: This task involves identifying the ancestor of a given node, which assesses indirect causal relationships. An example question could be, ``\emph{Does A serve as the ancestor node of E?}'' Here, the correct answer is ``Yes''. See Figure \ref{fig_scenario:CA} for a detailed illustration.

\subsection{Rung 1: Association}
\label{main_scenario:association}
For \emph{association (Rung 1)}, we mainly focus on two causal scenarios: \nameref{association:corr} and \nameref{association:eae}. We will provide a detailed description for both of them. 

\subsubsection{Correlation (CORR)}
\label{association:corr}

\begin{figure}[t]
    \centering
    \includegraphics[width=\textwidth]{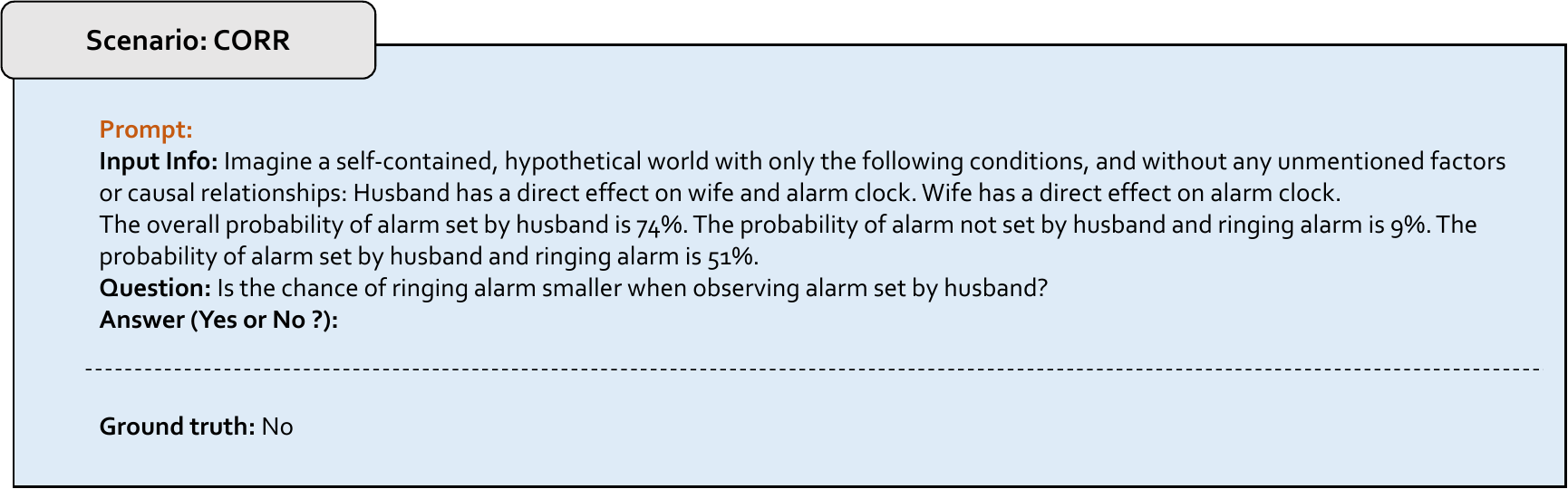}
    \caption[Example of correlation]{\textbf{Example of correlation.}}
    \label{fig_scenario:CORR}
\end{figure}

Correlation indicates a statistical association between two variables, irrespective of causality. Although ``\emph{correlation does not imply causation}'', identifying a statistical link between variables is a necessary step in the causal inference. It helps further investigation into whether the relationship is indeed causal, guiding researchers in developing hypotheses \citep{rolfe2020housing}, designing experiments \citep{duncan2012socioeconomic}, and employing statistical methods to explore the nature and direction of the supposed causal relationship \citep{rosato2018correlation}. 
\paragraph{Causal scenario setting.}
In CORR, we provide a causal graph (e.g., ``\emph{Husband has a direct effect on wife and alarm clock. Wife has a direct effect on alarm clock.}'') along with corresponding
probabilities (e.g., ``\emph{The overall probability of alarm set by husband is 74\%. The probability of alarm not set by husband and ringing alarm is 9\%. The probability of alarm set by husband and ringing alarm is 51\%.}''\footnote{The example is chosen from \citet{jin2023cladder}.}). The model needs to answer the question about the correlation between the given variables (e.g., ``\emph{Is the chance of ringing alarm smaller when observing alarm set by husband?}''). See Figure \ref{fig_scenario:CORR} for a detailed illustration.

\subsubsection{Explaining Away Effect (EAE)}
\label{association:eae}

EAE describes a causal relationship where two independent causes that produce a common effect become interdependent when that effect is observed \citep{pearl2009causality}. This interdependence arises because having information about one factor alters the probability of the other factor being involved, once it is established that the shared effect has taken place \citep{kim1983computational,pearl2009causality}. This pattern is also known as \emph{selection bias} or \emph{Berkson’s paradox} in statistics \citep{berkson1946limitations}. For example, consider a prestigious art school that requires applicants to excel in either painting or sculpture. In the general population, skill in painting and skill in sculpture might not be correlated. However, within the art school's student, there might be a negative correlation between these two skills. This is because students who are not as skilled in painting are likely those admitted due to their exceptional talent in sculpture, and vice versa. This phenomenon illustrates the explaining away effect: the school's admission criteria create an apparent negative correlation between two skills that are unrelated in the broader population, as each skill ``explains away'' the need for the other in the context of admission. The EAE is crucial in various fields, including psychology \citep{wilson2008explaining}, artificial intelligence \citep{kenny2021explaining}, and data analysis \citep{linden2020conducting}. It provides insight into how conditional dependencies between variables can lead to misleading correlations. Understanding EAE helps in accurately interpreting data, making informed decisions, and avoiding false conclusions. 

\begin{figure}[t]
    \centering
    \includegraphics[width=\textwidth]{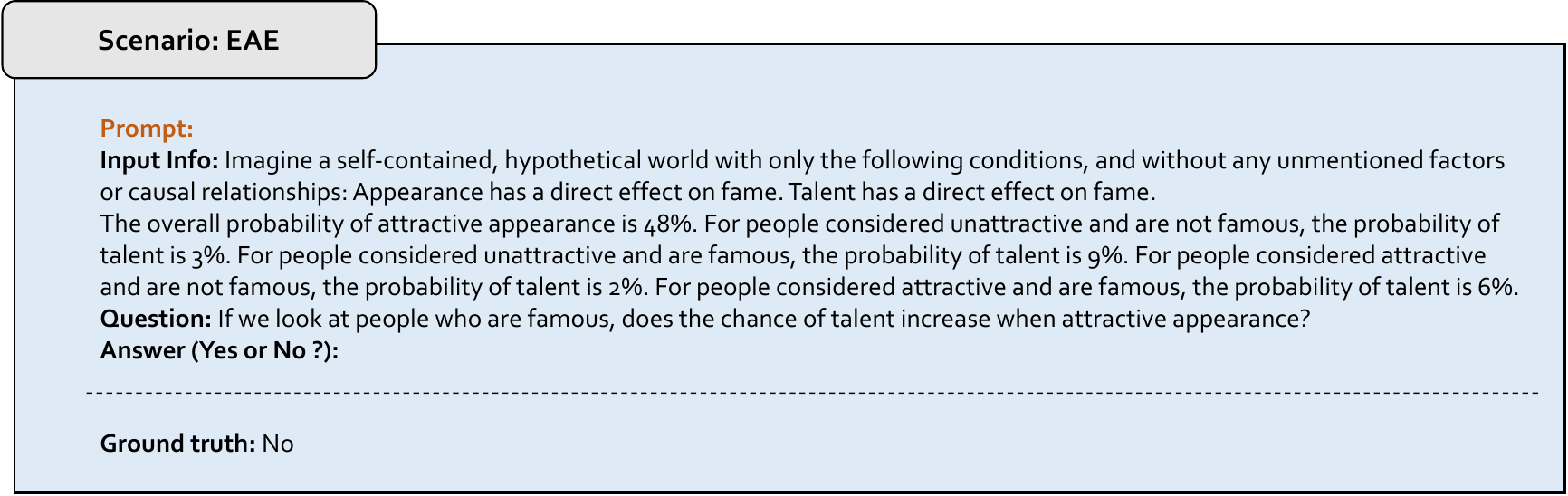}
    \caption[Example of explaining away effect]{\textbf{Example of explaining away effect.}}
    \label{fig_scenario:EAE}
\end{figure}

\paragraph{Causal scenario setting.}
In EAE, we provide a causal graph (e.g., ``\emph{Appearance has a direct effect on fame. Talent has a direct effect on fame.}'') along with corresponding conditional
probabilities (e.g., ``\emph{The overall probability of attractive appearance is 48\%. For people considered unattractive and are not famous, the probability of talent is 3\%. For people considered unattractive and are famous, the probability of talent is 9\%. For people considered attractive and are not famous, the probability of talent is 2\%. For people considered attractive and are famous, the probability of talent is 6\%.}''\footnote{The example is chosen from \citet{jin2023cladder}.}). The model needs to answer the question about the given variables (e.g., ``\emph{If we look at people who are famous, does the chance of talent increase when attractive appearance?}''). See Figure \ref{fig_scenario:EAE} for a detailed illustration.

\subsection{Rung 2: Intervention}
\label{main_scenario:intervention}
In contemplating \emph{intervention (Rung 2)}, we carefully design seven different causal scenarios. These causal scenarios cover all three modes, providing a thorough evaluation of the model's causal reasoning abilities. Specifically, for causal scenarios that involve both Natural and Mathematical modes, our evaluation focuses on \nameref{intervention:ate} (\cref{intervention:ate}) and \nameref{intervention:cde} (\cref{intervention:cde}). In scenarios restricted to the Natural mode, we assess \nameref{intervention:cb} (\cref{intervention:cb}). And for scenarios exclusive to the Symbolic mode, we examine \nameref{intervention:bas} (\cref{intervention:bas}), \nameref{intervention:fas} (\cref{intervention:fas}), \nameref{intervention:iv} (\cref{intervention:iv}), and \nameref{intervention:cei} (\cref{intervention:cei}).

\subsubsection{Average Treatment Effect (ATE)}
\label{intervention:ate}

ATE is a fundamental concept in causal inference that helps to quantify the impact of a particular intervention. 
$X$ causes $Y$ if and only if changing $X$ leads to a change in $Y$, keeping everything else constant. Denoting the treatment group by $do(X=1)$ and the control group by $do(X=0)$, the difference between them, $P(Y=1|do(X=1))-P(Y=1|do(X=0))$, is called the ATE \citep{pearl2016causal}. Consider a causal scenario where all subjects are unemployed individuals, with some receiving a policy intervention (the treatment group) and others not (the control group). We are interested in determining the causal impact of a job search monitoring policy on the duration of unemployment. Essentially, we want to know, on average, how much shorter one's unemployment period would be if they underwent the intervention. The ATE in this case measures the difference in expected durations of unemployment between the treatment and control groups. A negative ATE would indicate that the job policy extended the length of unemployment, while a positive ATE would suggest a reduction in unemployment duration. Conversely, an ATE estimate of zero would mean that the treatment had no discernible effect on the length of unemployment. 
By providing a quantitative measure of a treatment's impact on an average, ATE helps in understanding the generalizability of interventions. Moreover, ATE assists in improving policy design and informing stakeholders about the potential benefits or drawbacks of specific actions. Research ranging from economics \citep{abadie2011bias,hirano2003efficient} to public health \citep{feng2012generalized} proves the significance of ATE.
\begin{figure}[t]
    \centering
    \includegraphics[width=\textwidth]{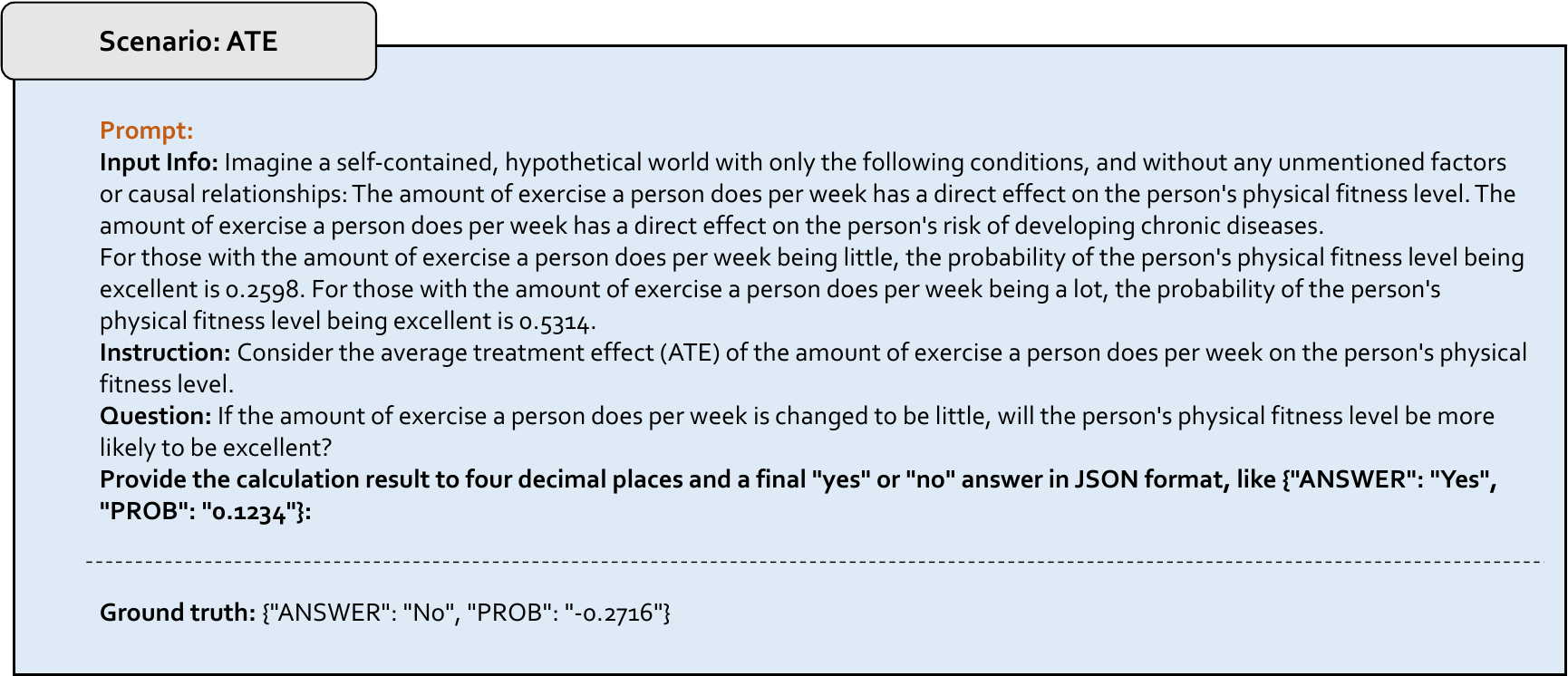}
    \caption[Example of average treatment effect]{\textbf{Example of average treatment effect.}}
    \label{fig_scenario:ATE}
\end{figure}

\paragraph{Causal scenario setting.}
In ATE, we provide a causal graph (e.g., ``\emph{The amount of exercise a person does per week has a direct effect on the person's physical fitness level. The amount of exercise a person does per week has a direct effect on the person's risk of developing chronic diseases.}'') along with corresponding conditional probabilities (e.g., ``\emph{For those with the amount of exercise a person does per week being little, the probability of the person's physical fitness level being excellent is 0.2598. For those with the amount of exercise a person does per week being a lot, the probability of the person's physical fitness level being excellent is 0.5314.}''). The causal scenario requires determining the ATE between specified variables (e.g., ``\emph{If the amount of exercise a person does per week is changed to be little, will the person's physical fitness level be more likely to be excellent?}''). The model needs to address two different types of questions:\footnote{To optimize evaluation efficiency, we use ``\emph{Provide the calculation result to four decimal places and a final ``yes'' or ``no'' answer in JSON format, like \{``ANSWER'': ``Yes'', ``PROB'': ``0.1234''\}:}'' to concurrently obtain responses from models in both question types (CDE, ETT, NDE, and NIE are configured similarly). However, it needs to be emphasized that these two types of questions target different areas: binary classification focuses on the model’s capability to manage questions in Natural mode, whereas probability calculation is geared towards handling questions in Mathematical mode.}
(1) \textbf{Binary classification}: This involves providing a direct answer of ``Yes'' or ``No'' (the correct answer for this example is ``No'');
(2) \textbf{Probability calculation}: The model needs to utilize the probabilities provided in the question to calculate the accurate response, preserving precision to four decimal places (the correct answer for this example is ``-0.2716''). See Figure \ref{fig_scenario:ATE} for a detailed illustration.

\subsubsection{Backdoor Adjustment Set (BAS)}
\label{intervention:bas}

\begin{figure}[t]
\centering  
\subfigure[BAS]{ 
\begin{minipage}{4.1cm}
\centering    
    \includegraphics[width=1.15\linewidth]{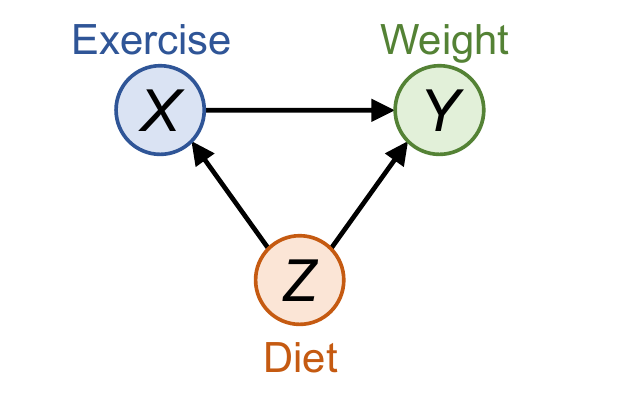}
\end{minipage}
}
\hspace{1cm}
\subfigure[FAS]{ 
\begin{minipage}{4.1cm}
\centering    
    \includegraphics[width=1.15\linewidth]{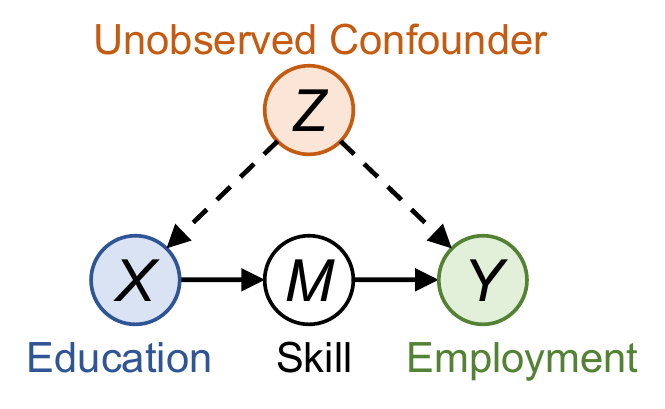}
\end{minipage}
}
\hspace{1cm}
\subfigure[IV]{ 
\begin{minipage}{4.1cm}
\centering    
    \includegraphics[width=1.15\linewidth]{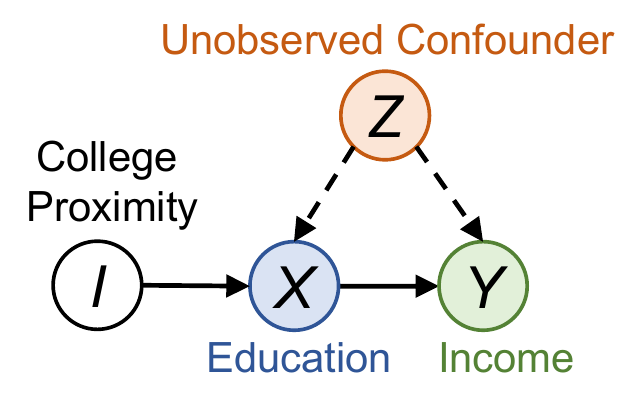}
\end{minipage}
}
\caption[Real-world examples of BAS, FAS and IV]{\textbf{Real-world examples of BAS, FAS and IV.} $X$ represents the treatment, $Y$ represents the outcome, $Z$ represents the confounder, $M$ represents the mediator, and $I$ represents the instrumental variable. If the confounder is unobserved, its effect on the treatment and outcome is represented by a dashed line.}    
\label{fig_main:scenario_real1}    
\end{figure}

When assessing the impact of treatment $X$ on outcome $Y$, we are confronted with the decision to adjust our calculations to account for potential variations in confounders $Z$. This adjustment is typically implemented by using the \emph{Back-door criterion} \citep{pearl1995causal}. A backdoor adjustment set is such a set of variables $Z$ that, when controlled for, blocks all backdoor paths from the treatment to the outcome. A backdoor path is a path that leads from the treatment to the outcome through an arrow pointing to the treatment, which can introduce confounding bias if not properly adjusted for \citep{pearl2009causality}. By adjusting for the variables in a backdoor adjustment set, one aims to eliminate confounding bias, allowing for an unbiased estimation of the causal effect of the treatment on the outcome \citep{pearl1995causal}. In a real-word causal scenario as Figure \ref{fig_main:scenario_real1}(a) shows, suppose we are interested in the effect of exercise (treatment) on weight (outcome). A backdoor path might be through a variable like diet, where diet affects both exercise and weight. If we do not control for diet, we might incorrectly attribute the effect of diet on weight to exercise. Therefore, diet could be part of a backdoor adjustment set that, when controlled for, allows for an unbiased estimation of the causal effect of exercise on weight.
The BAS has important real-world applications across various fields such as healthcare \citep{pmlr-v126-adib20a}, artificial intelligence \citep{landeiro2016robust,correa2018generalized,landeiro2018robust,dai2023robust}, and social sciences \citep{elwert2013graphical}, where understanding causal relationships is crucial for informed decision-making and policy development.

\begin{figure}[t]
    \centering
    \includegraphics[width=\textwidth]{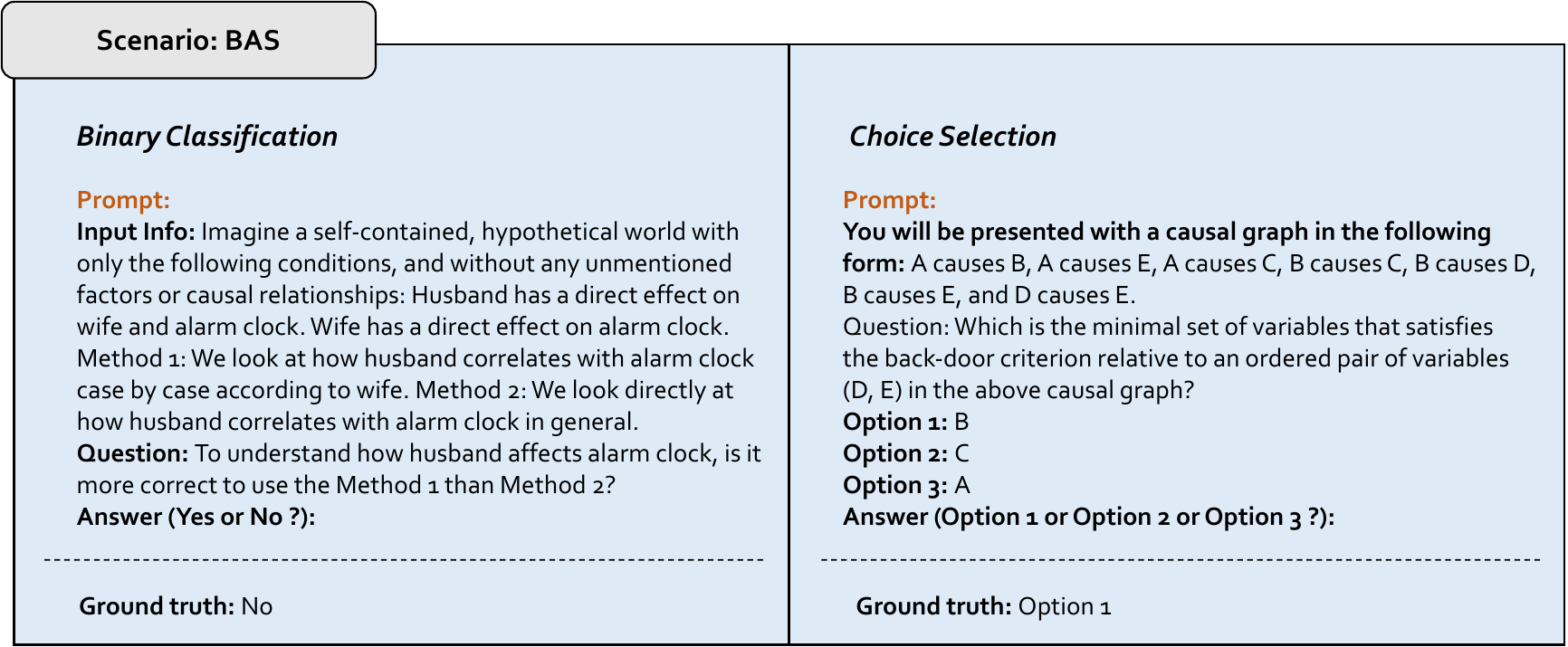}
    \caption[Example of backdoor adjustment set]{\textbf{Example of backdoor adjustment set.}}
    \label{fig_scenario:BAS}
\end{figure}

\paragraph{Causal scenario setting.}
We design two types of questions in BAS. (1) \textbf{Binary classification}: We provide the model with a causal graph (e.g., ``\emph{Husband has a direct effect on wife and alarm clock. Wife has a direct effect on alarm clock.}'') and two different methods (e.g., ``\emph{Method 1: We look at how husband correlates with alarm clock case by case according to wife. Method 2: We look directly at how husband correlates with alarm clock in general.}''\footnote{The example is chosen from \citet{jin2023cladder}.}). And the model is required to decide which method is more correct. (2) \textbf{Choice selection}: The question starts with presenting a causal graph in Symbolic form (e.g., ``\emph{A causes B, A causes E, A causes C, B causes C, B causes D, B causes E, and D causes E.}''). The model needs to apply the \emph{Back-door criterion} to determine the backdoor variables between an ordered pair of variables (e.g., ``\emph{D}'' and ``\emph{E}''). There are three different categories of backdoor variables that need to be assessed: the maximal, the minimal, and a combination of both. Each category plays a specific role in controlling for confounding in causal analysis. See Figure \ref{fig_scenario:BAS} for a detailed illustration.

\subsubsection{Frontdoor Adjustment Set (FAS)}
\label{intervention:fas}

\begin{figure}[t]
    \centering
    \includegraphics[width=\textwidth]{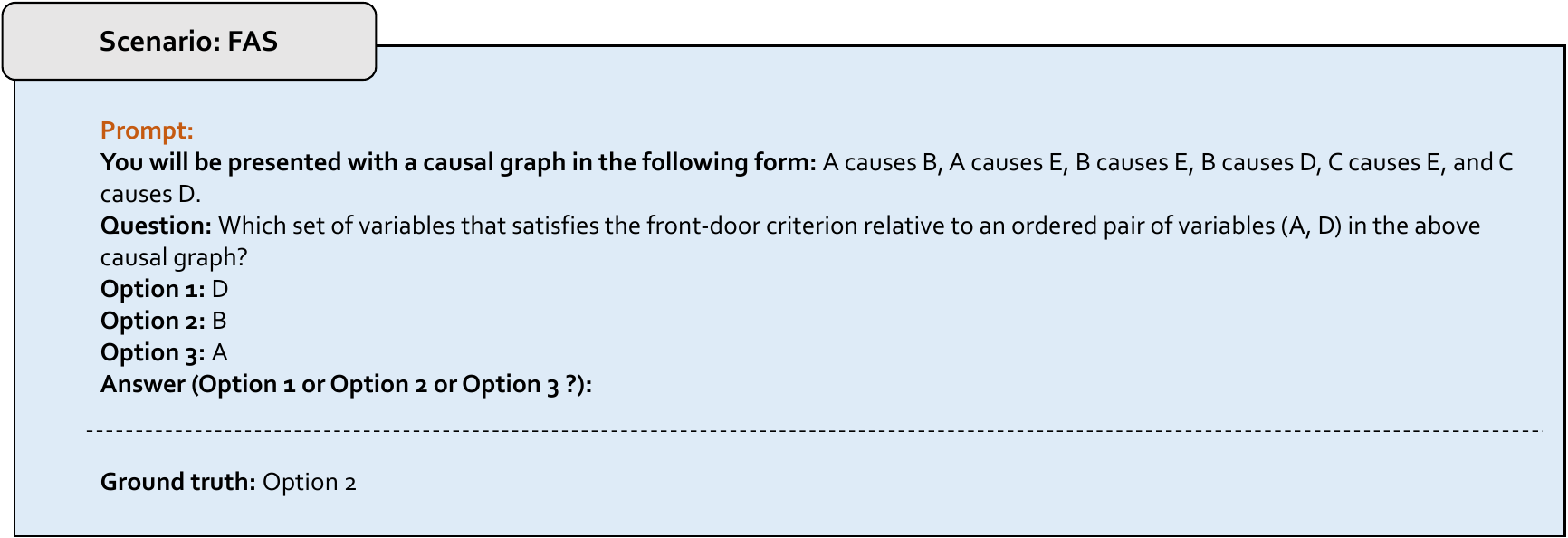}
    \caption[Example of frontdoor adjustment set]{\textbf{Example of frontdoor adjustment set.}}
    \label{fig_scenario:FAS}
\end{figure}

While the \emph{Back-door criterion} aims to control for variables that impact both the treatment and the outcome, the \emph{Front-door criterion} provides a way to estimate the causal effect by exploiting the mediation pathway, even in the presence of unobserved confounding between the treatment and outcome \citep{pearl1995causal}. A FAS involves a set of variables that mediate the causal path from the treatment to the outcome.  Consider a study on the effect of education (treatment) on employment (outcome) in Figure \ref{fig_main:scenario_real1}(b). Suppose the skill (mediator) is the way through which education affect employment. The \emph{Front-door Criterion} would involve first estimating the effect of education on the skill, and then estimating the effect of the skill on employment while controlling for the education. Even if there are unmeasured factors that affect both the decision to participate in education and employment, the frontdoor adjustment allows for an estimation of the causal effect of education on employment through the mediator of skill. 
The practical real-world significance of the FAS extends across various domains, offering substantial benefits in artificial intelligence \citep{xu2023causal,xu2022neural,xia2024deciphering}, earth and environmental sciences \citep{runge2023causal}, and ecology \citep{arif2023applying}. 
\paragraph{Causal scenario setting.}
Similar to BAS, the question in FAS provides the model with a causal graph and requires it to utilize the \emph{Front-door criterion} to determine the frontdoor variables between an ordered pair of variables (e.g., ``\emph{A}'' and ``\emph{D}''). See Figure \ref{fig_scenario:FAS}  for a detailed illustration.

\subsubsection{Instrumental Variable (IV)}
\label{intervention:iv}

\begin{figure}[t]
    \centering
    \includegraphics[width=\textwidth]{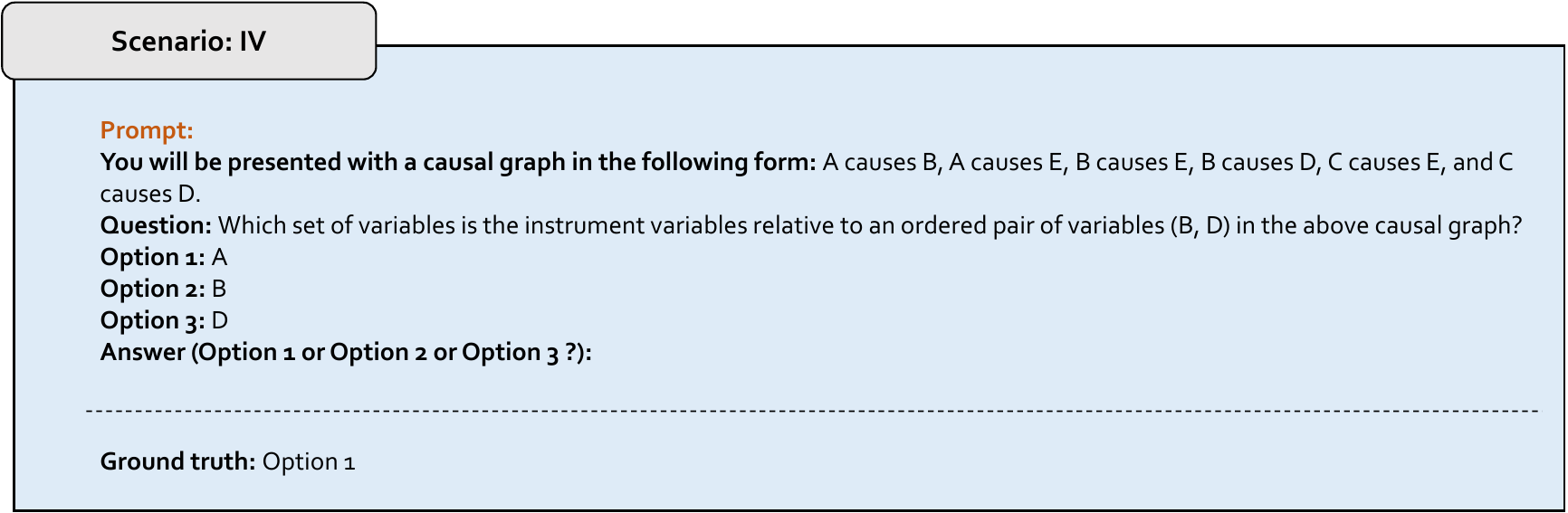}
    \caption[Example of instrumental variable]{\textbf{Example of instrumental variable.}}
    \label{fig_scenario:IV}
\end{figure}

Relative to a pair $(X,Y)$, an instrumental $Z$ must satisfy two conditions: (1) it is independent from any variables that impact $Y$ not through $X$  (including error terms), and (2) it is dependent on $X$ \citep{pearl2009causality}. In other words, an instrumental variable influences the treatment but has no direct effect on the outcome, except through the treatment. This characteristic makes it possible to estimate the causal effect of the treatment on the outcome, even in the presence of unobserved confounders that might otherwise bias the estimates. Consider a study aiming to estimate the effect of education (treatment) on income (outcome), as the Figure \ref{fig_main:scenario_real1}(c) demonstrates. However, an individual's decision to pursue more education might be influenced by unobserved factors like capability or family background, which also affect income. An instrumental variable could be the college proximity, under the assumption that it affects an individual's decision to obtain more education but does not directly affect their income, except through education. By using this instrumental variable, researchers can estimate the causal effect of education on income, controlling for unobserved confounding factors.
In summary, IV allows for more reliable and accurate estimation of causal effects, making it an essential tool in fields where controlled experiments are impractical or impossible, and thus, significantly enhances the validity of empirical findings in social sciences \citep{bollen2012instrumental}, human resource management \citep{saridakis2020exploring}, and economics \citep{mogstad2021causal}.
\paragraph{Causal scenario setting.}
Similar to BAS and FAS setups, given a causal graph, the model needs to determine the instrumental variable between an ordered pair of variables (e.g., ``\emph{B}'' and ``\emph{D}''). See Figure \ref{fig_scenario:IV}  for a detailed illustration.

\subsubsection{Collider Bias (CB)}
\label{intervention:cb}

\begin{figure}[t]
\centering  
\subfigure[CB]{ 
\begin{minipage}{4.1cm}
\centering    
    \includegraphics[width=1.15\linewidth]{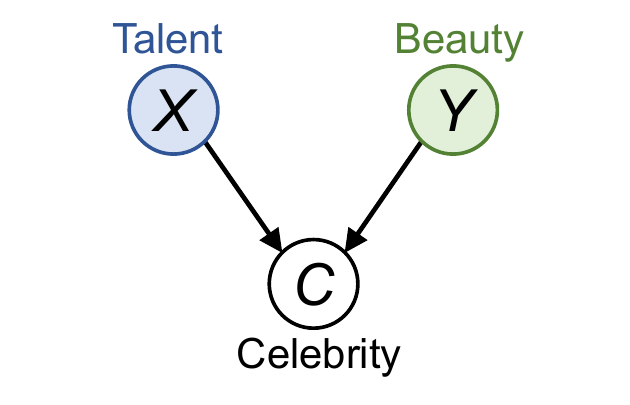}
\end{minipage}
}
\hspace{1cm}
\subfigure[CEI]{ 
\begin{minipage}{4.1cm}
\centering    
    \includegraphics[width=1.15\linewidth]{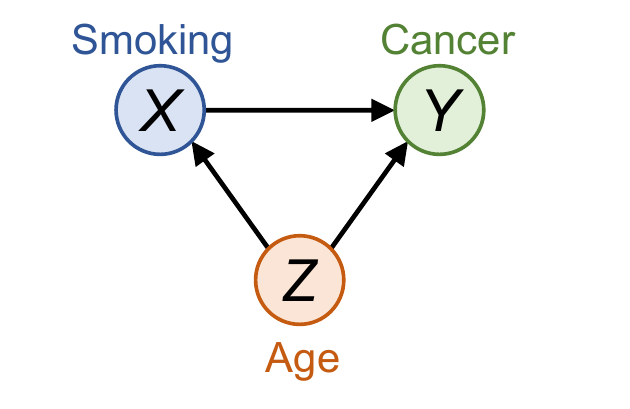}
\end{minipage}
}
\hspace{1cm}
\subfigure[CDE]{ 
\begin{minipage}{4.1cm}
\centering    
    \includegraphics[width=1.15\linewidth]{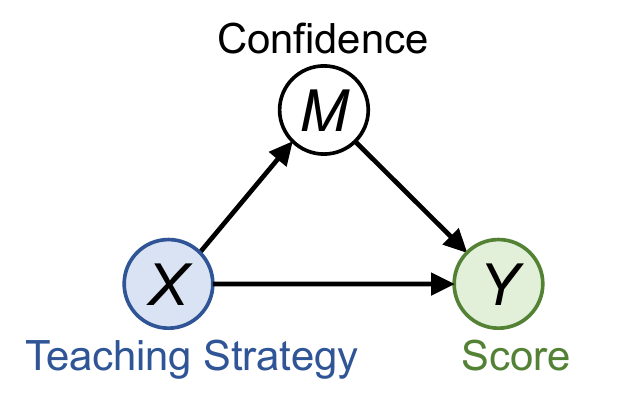}
\end{minipage}
}
\caption[Real-world examples of CB, CEI and CDE]{\textbf{Real-world examples of CB, CEI and CDE.} $X$ represents the treatment, $Y$ represents the outcome, $C$ represents the common effect, and $M$ represents the mediator.}    
\label{fig_main:scenario_real2}    
\end{figure}

CB is a type of selection bias that occurs when an analysis is conditioned upon a common effect of two or more variables.
The simplest collider in a causal graph can be illustrated as $X\rightarrow C\leftarrow Y$, where $C$ represents the common effect of causes $X$ and $Y$ \citep{pearl2016causal}. Collider bias occurs when a common effect is controlled. For example, $X$ and $Y$ are independent, while conditions on $Z$ will make them dependent. A famous example illustrating collider bias is the \emph{hollywood actors} ($talent\rightarrow celebrity\leftarrow beauty$) \citep{pearl2018book}. As Figure \ref{fig_main:scenario_real2}(a) demonstrates, it is asserted that both talent and beauty affect an actor's success. However, it is important to note that while beauty and talent contribute to success in acting, they are independent of each other in the general population.
This bias can lead to incorrect inferences about the relationships between variables \citep{cole2010illustrating,elwert2014endogenous,munafo2018collider}. Thus, recognizing and addressing CB is crucial for ensuring the validity and reliability of study findings \citep{mahmoud2022robust,pmlr-v161-shi21a}, ultimately guiding accurate scientific understanding and informed decision-making.

\begin{figure}[t]
    \centering
    \includegraphics[width=\textwidth]{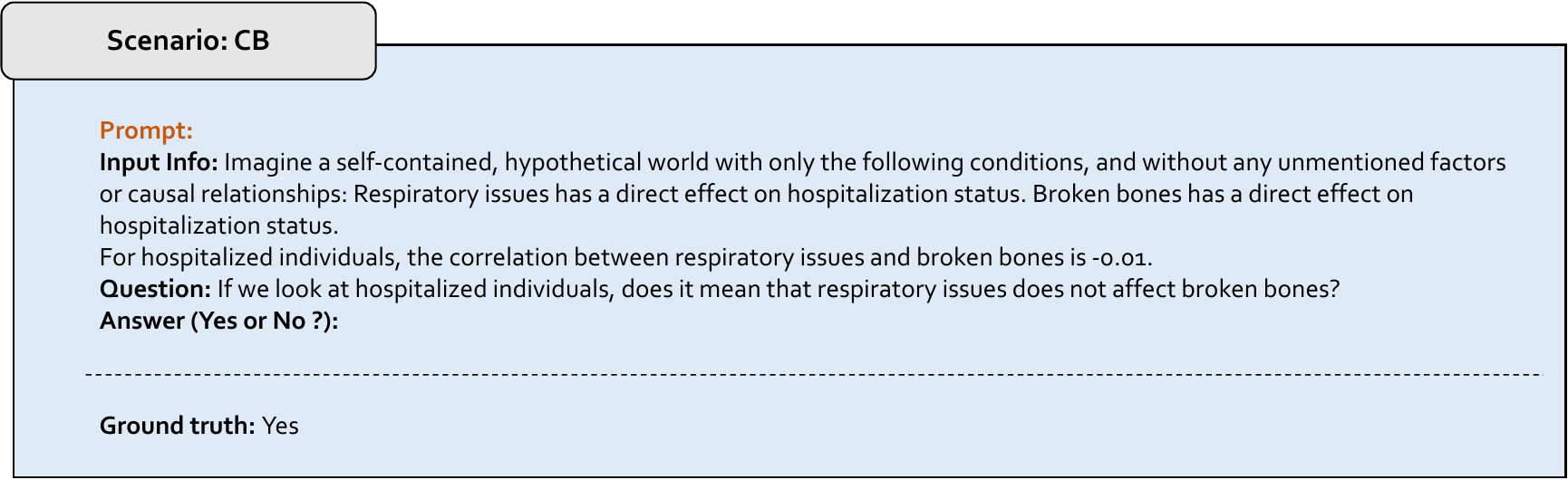}
    \caption[Example of collider bias]{\textbf{Example of collider bias.}}
    \label{fig_scenario:CB}
\end{figure}

\paragraph{Causal scenario setting.}
In CB, we provide a causal graph (e.g., ``\emph{Respiratory issues has a direct effect on hospitalization status. Broken bones has a direct effect on hospitalization status.}'') along with corresponding
probabilities (e.g., ``\emph{For hospitalized individuals, the correlation between respiratory issues and broken bones is -0.01.}''). This causal scenario requires the model to exclude the interference of collider bias and answer the question correctly (e.g., ``\emph{If we look at hospitalized individuals, does it mean that respiratory issues does not affect broken bones?}''\footnote{The example is chosen from \citet{jin2023cladder}.}). See Figure \ref{fig_scenario:CB}  for a detailed illustration.

\subsubsection{Causal Effect Identification (CEI)}
\label{intervention:cei}

The concept of CEI in causal reasoning is centered on determining whether the causal effect of a treatment on an outcome can be estimated from observational data \citep{shpitser2008complete}. Consider two disjoint sets, denoted as $X$ and $Y$, where the causal effect of $X$ on $Y$ is represented as $P(y|do(x))$. Assume that $P(v)$ represents a probability distribution over the variable set $V$. The causal effect of  $X$ on $Y$ is identifiable when the value of $P(y|do(x))$ can be exclusively ascertained from any positive probability distribution of the observed variables within graph $G$ \citep{tian2002general}. Figure \ref{fig_main:scenario_real2}(b) shows a case to estimate the causal effect of smoking on lung cancer using observational data, doctors employ a causal diagram to control for confounders like age and genetics. They determine if this effect is identifiable—whether they can estimate the incidence of lung cancer if hypothetically a population were assigned to smoke. If identifiable, they can statistically estimate the risk of lung cancer due to smoking, providing evidence to guide public health interventions without the need for unethical randomized trials that would require participants to smoke.
\begin{figure}[t]
    \centering
    \includegraphics[width=\textwidth]{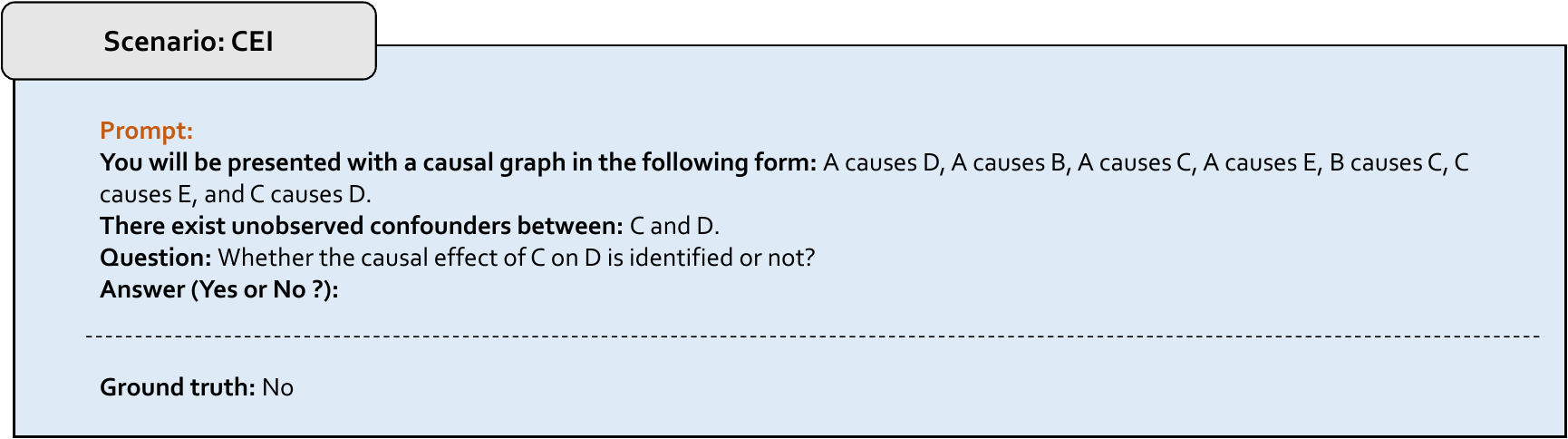}
    \caption[Example of causal effect identification]{\textbf{Example of causal effect identification.}}
    \label{fig_scenario:CEI}
\end{figure}
\paragraph{Causal scenario setting.}
In CEI, we start by presenting a causal graph (e.g., ``\emph{A causes D, A causes B, A causes C, A causes E, B causes C, C causes E, and C causes D}''). We then specify the existence of unobserved confounders between designated nodes at four different rates: 20\%, 40\%, 60\%, and 80\%. For instance, if there are 20\% unobserved confounders, the instruction will indicate ``\emph{There exist unobserved confounders between: C and D}''. If it increase to 60\%, the instruction will state ``\emph{There exist unobserved confounders between: A and B, C and D, B and C, and C and E}.'' The causal scenario requires the model to determine if the causal relationship between a pair of treatment and outcome variables (e.g., ``\emph{C}'' and ``\emph{D}'') can be identified (the correct answer for this example is ``No''). See Figure \ref{fig_scenario:CEI} for a detailed illustration.

\subsubsection{Controlled Direct Effect (CDE)}
\label{intervention:cde}

The CDE quantifies the direct influence of an intervention on an outcome while maintaining one or more mediators at a predetermined level \citep{pearl2018book,cinelli2022crash}. In this way, it disregards the indirect effects that operate through these mediators. In the setting of three variables: treatment $X$, outcome $Y$, and mediator $M$, the CDE on $Y$ when altering the value of $X$ from $x$ to $x'$ is formally defined as $CDE=P(Y|do(X=x',M=m))-P(Y|do(X=x,M=m))$ \citep{kaufman2005cde,pearl2001effects}. As Figure \ref{fig_main:scenario_real2}(c) shows, imagine we are investigating the direct effect of a novel math teaching strategy (treatment) on student exam scores (outcome), intentionally excluding its indirect effect via enhancing student confidence (mediator). By controlling for students' confidence levels, we aim to measure the direct influence of this new method compared to traditional teaching on exam outcomes. This analysis allows us to distinctly identify the immediate benefits of the teaching approach on performance, separating from its indirect benefits through confidence improvement. This clarity aids educators in precisely assessing the direct effectiveness of the new teaching method. This isolation required by CDE is particularly important in complex systems where multiple pathways and interactions can obscure the mechanisms through which an intervention works \citep{nguyen2021clarifying}. By assessing the CDE, decision-makers can more accurately design and refine interventions, targeting the direct mechanisms that produce the desired outcome in various fields such as epidemiology \citep{carter2021mendelian}, artificial intelligence \citep{tang2020long}, and biology \citep{howe2022within}. 
\begin{figure}[t]
    \centering
    \includegraphics[width=\textwidth]{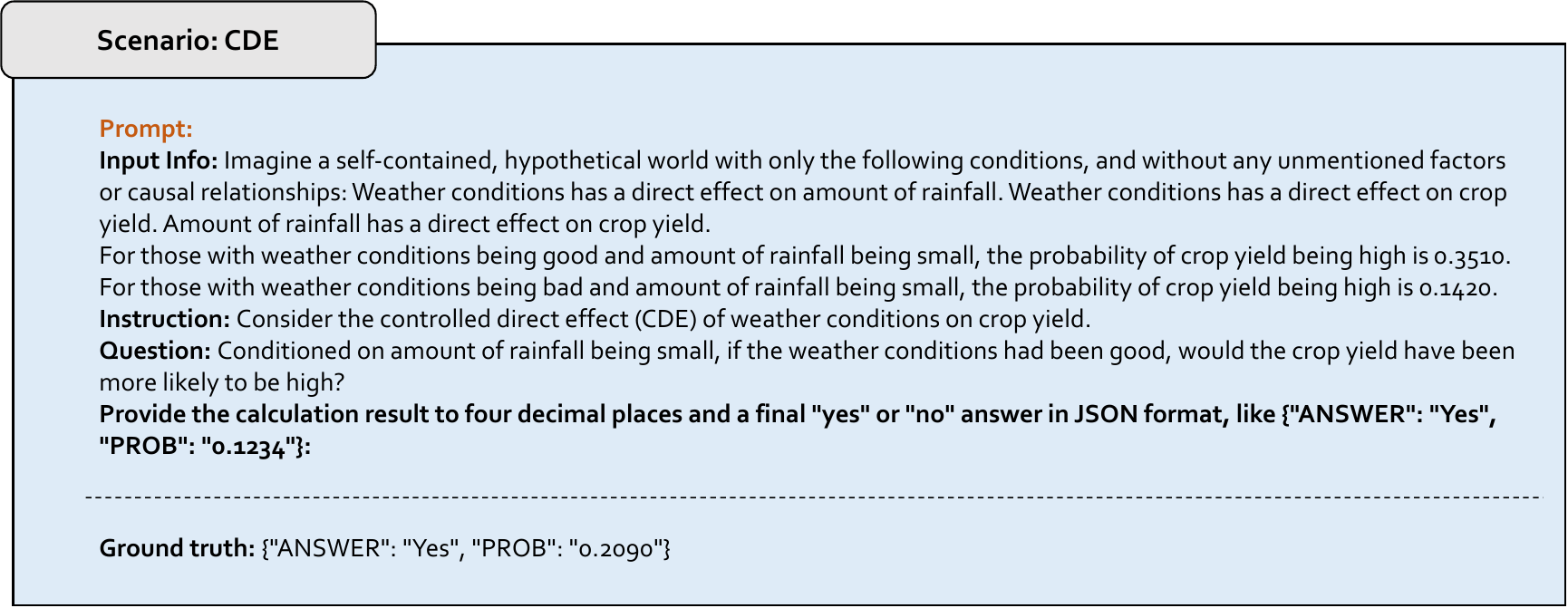}
    \caption[Example of controlled direct effect]{\textbf{Example of controlled direct effect.}}
    \label{fig_main_scenario:CDE}
\end{figure}
\paragraph{Causal scenario setting.}
Similar to ATE, the problem in CDE gives a causal graph (e.g., ``\emph{Weather conditions has a direct effect on amount of rainfall. Weather conditions has a direct effect on crop yield. Amount of rainfall has a direct effect on crop yield.}'') along with corresponding conditional probabilities (e.g., ``\emph{For those with weather conditions being good and amount of rainfall being small, the probability of crop yield being high is 0.3510. For those with weather conditions being bad and amount of rainfall being small, the probability of crop yield being high is 0.1420.}''). The causal scenario requires the model to determine the CDE between variables (e.g., ``\emph{Conditioned on amount of rainfall being small, if the weather conditions had been good, would the crop yield have been more likely to be high?}''). The model needs to address both binary classification (``Yes'') and probability calculation (``0.2090'') questions. See Figure \ref{fig_main_scenario:CDE} for a detailed illustration.

\subsection{Rung 3: Counterfactuals}
\label{main_scenario:counterfactual}
When analyzing causal scenarios at the \emph{counterfactuals (Rung 3)} level, we assess eight causal scenarios across two modes. Scenarios belonging to this rung present considerable challenges to language models. In the Natural mode, our assessment focuses on \nameref{counterfactual:ac} (\cref{counterfactual:ac}), \nameref{counterfactual:ceg} (\cref{counterfactual:ceg}), and \nameref{counterfactual:cr} (\cref{counterfactual:cr}). For scenarios involving both Natural and Mathematical modes, we evaluate the following five specific scenarios: \nameref{counterfactual:ett} (\cref{counterfactual:ett}), \nameref{counterfactual:nde} (\cref{counterfactual:nde}), \nameref{counterfactual:nie} (\cref{counterfactual:nie}), 
 \nameref{counterfactual:pn} (\cref{counterfactual:pn}), and \nameref{counterfactual:ps} (\cref{counterfactual:ps}). 

\subsubsection{Actual Causality (AC)}
\label{counterfactual:ac}

AC deals with attribution and responsibility allocation problems encountered in practical applications like policy-implementing \citep{capano2021causal}, diagnosing causes \citep{wang2021groot,albantakis2019caused}, and decision making \citep{triantafyllou2022actual}. AC goes beyond the mere identification of correlations in data; it enables language models to grasp the underlying mechanisms that lead to certain outcomes and make predictions that reflect a deeper understanding of how different elements are interrelated. This understanding is essential for generating more accurate, relevant, and contextually appropriate responses, especially in complex causal scenarios where multiple factors interact. Studies have demonstrated that this causal scenario presents considerable challenges for language models \citep{kiciman2023causal,suzgun2022challenging}. 
\begin{figure}[t]
    \centering
    \includegraphics[width=\textwidth]{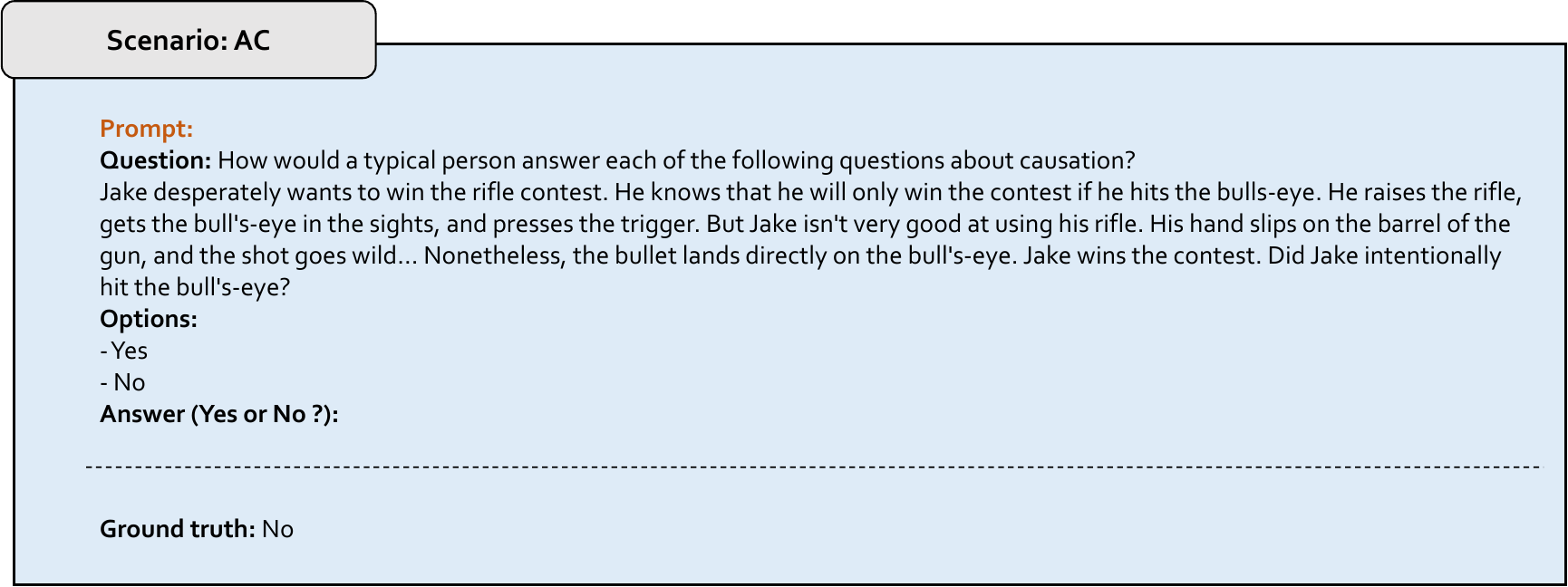}
    \caption[Example of actual causality]{\textbf{Example of actual causality.}}
    \label{fig_scenario:AC}
\end{figure}
\paragraph{Causal scenario setting.}
In AC, the causal scenario presents an actual story (e.g., ``\emph{Jake desperately wants to win the rifle contest. He knows that he will only win the contest if he hits the bull's-eye. He raises the rifle, gets the bull's-eye in the sights, and presses the trigger. But Jake isn't very good at using his rifle. His hand slips on the barrel of the gun, and the shot goes wild... Nonetheless, the bullet lands directly on the bull's-eye. Jake wins the contest. Did Jake intentionally hit the bull's-eye?}''\footnote{The example is chosen from \citet{suzgun2022challenging}.}). Each story ends with a binary classification question (i.e., ``Yes'' or ``No''), aiming to challenge whether model can predict the correct answer (``No'' for this example). See Figure \ref{fig_scenario:AC} for a detailed illustration.

\subsubsection{Causal Explanation Generation (CEG)}
\label{counterfactual:ceg}

The ability to understand and explain causality is a cornerstone for building machines that can reason reliably. This causal scenario aims to examine whether language models can generate comprehensive and logically sound explanations that elucidate the causal relationships between specific events \citep{gao2023chatgpt}. By understanding and articulating the underlying causes behind phenomena, language models can offer more accurate, relevant, and transparent responses, thereby improving user interactions. This capability is particularly important in decision-support contexts, such as healthcare \citep{richens2020improving} and policy-making \citep{swinkels2020ideas}, where understanding causal relationships is essential. Additionally, causal explanations can aid in debugging and refining language models by revealing how they process information, facilitating improvements in model performance and fairness \citep{lin2021generative,o2020generative,madumal2020explainable,moraffah2020causal}.

\begin{figure}[t]
    \centering
    \includegraphics[width=0.6\textwidth]{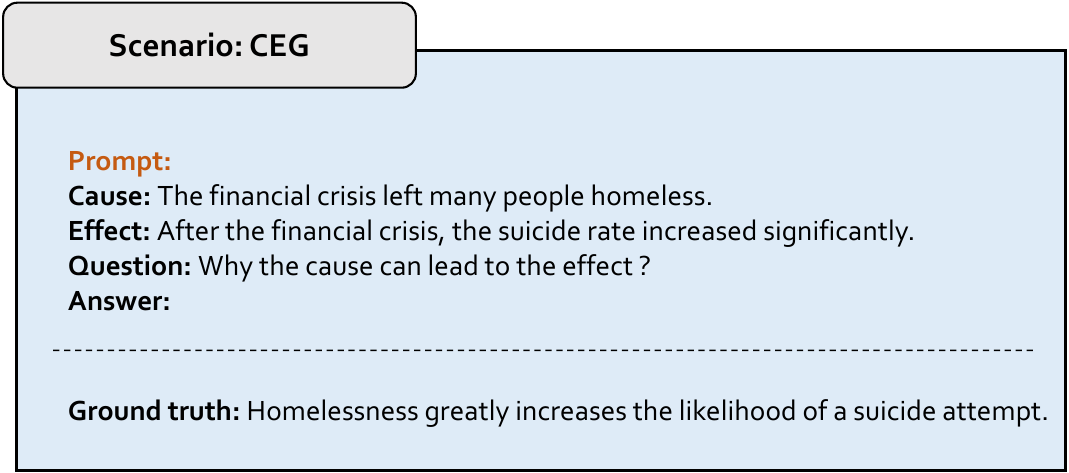}
    \caption[Example of causal explanation generation]{\textbf{Example of causal explanation generation.}}
    \label{fig_scenario:CEG}
\end{figure}

\paragraph{Causal scenario setting.}
In CEG, the causal scenario begins by presenting a cause-effect pair (e.g., ``\emph{Cause: The financial crisis left many people homeless.}'' and ``\emph{Effect: After the financial crisis, the suicide rate increased significantly.}''\footnote{The example is chosen from \citet{du2022care}.}). The model needs to provide a reasonable explanation of why the cause can lead to the effect (e.g., ``\emph{Homelessness greatly increases the likelihood of a suicide attempt.}''). See Figure \ref{fig_scenario:CEG} for a detailed illustration.

\subsubsection{Effect of the Treatment on the Treated (ETT)}
\label{counterfactual:ett}

\begin{figure}[t]
\centering  
\subfigure[ETT]{ 
\begin{minipage}{4.1cm}
\centering    
    \includegraphics[width=1.15\linewidth]{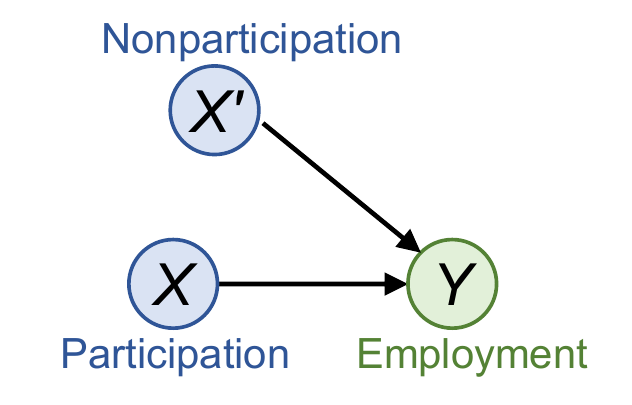}
\end{minipage}
}
\hspace{1cm}
\subfigure[NDE]{ 
\begin{minipage}{4.1cm}
\centering   
    \includegraphics[width=1.15\linewidth]{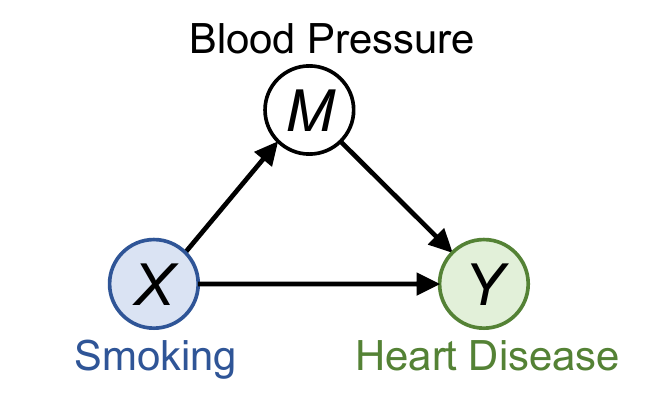}
\end{minipage}
}
\hspace{1cm}
\subfigure[NIE]{ 
\begin{minipage}{4.1cm}
\centering    
    \includegraphics[width=1.15\linewidth]{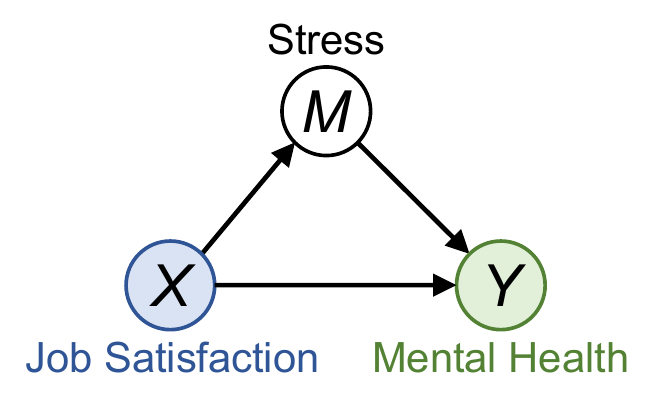}
\end{minipage}
}
\caption[Real-world examples of ETT, NDE and NIE]{\textbf{Real-world examples of ETT, NDE and NIE.} $X$ and $X'$ represent the treatments, $Y$ represents the outcome, and $M$ represents the mediator.}    
\label{fig_main:scenario_real3}    
\end{figure}

ETT is employed to assess whether individuals who receive treatment are the ones who would derive the greatest advantage from it. In other words, the issue ETT seeks to address is: What differences would there be in outcomes for individuals who did receive treatment compared to if they had not undergone treatment? When a policymaker's objective is to determine whether to uphold or discontinue an existing program within its present incentive framework, the key parameter of concern should measure the ETT \citep{pearl2009causality}. The appropriate formula is $ETT=E(Y_1-Y_0|X=1)$, where $Y_x$ denotes the value of outcome $Y$ when treatment $X$ is kept constant at $X = x$ \citep{pearl2009causality}. As Figure \ref{fig_main:scenario_real3}(a) demonstrates, the ETT in a job training program for unemployed individuals assesses the program's specific impact on participants by comparing their employment outcomes before and after participation, against similar non-participants. ETT reveals the direct benefits of the program, aiding in evaluating its effectiveness and guiding policy decisions. As a crucial metric for assessing the effectiveness of voluntary enrollment in programs by those who are eligible, ETT is extensively used in various fields including econometrics \citep{roth2023s,baker2022much,de2023two}, healthcare \citep{jastreboff2022tirzepatide}, and psychology \citep{gomila2021logistic}.

\begin{figure}[t]
    \centering
    \includegraphics[width=\textwidth]{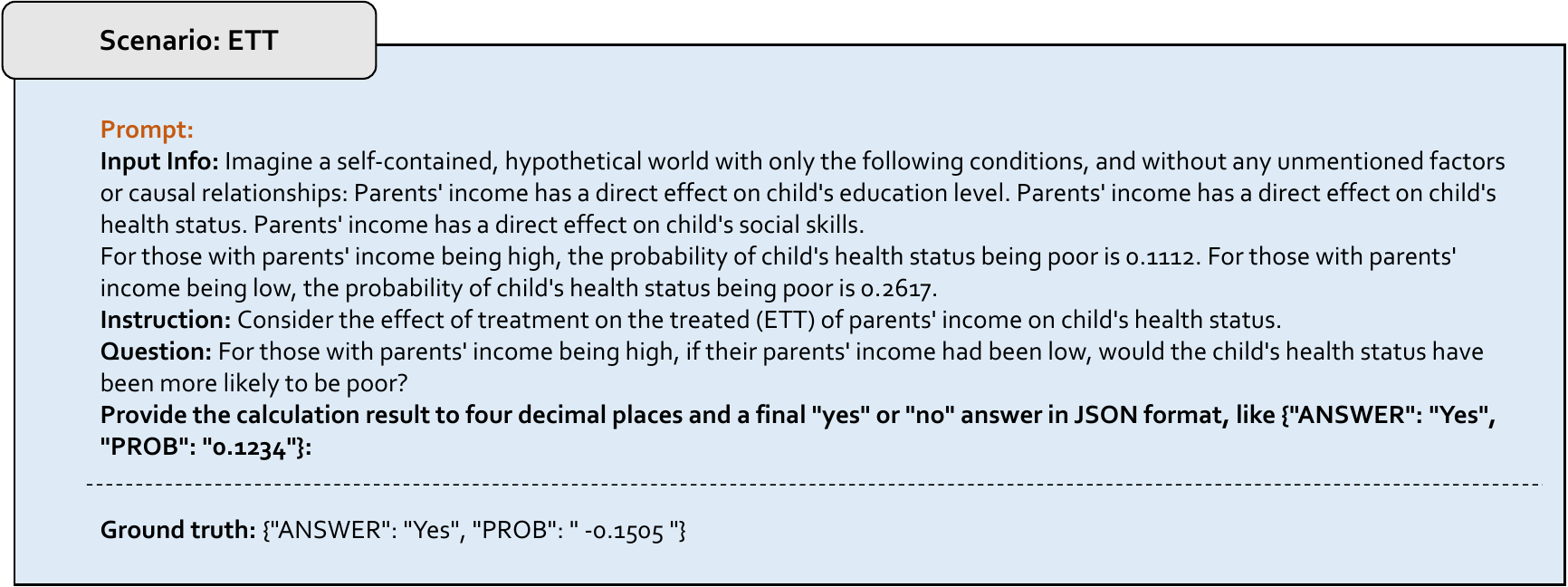}
    \caption[Example of effect of the treatment on the treated]{\textbf{Example of effect of the treatment on the treated.}}
    \label{fig_scenario:ETT}
\end{figure}

\paragraph{Causal scenario setting.}
Similar to ATE, we provide a causal graph (e.g., ``\emph{Parents' income has a direct effect on child's education level. Parents' income has a direct effect on child's health status. Parents' income has a direct effect on child's social skills.}'') along with corresponding conditional probabilities (e.g., ``\emph{For those with parents' income being high, the probability of child's health status being poor is 0.1112. For those with parents' income being low, the probability of child's health status being poor is 0.2617.}''). The causal scenario requires the model to determine the ETT between variables (e.g., ``\emph{For those with parents' income being high, if their parents' income had been low, would the child's health status have been more likely to be poor?}''). The model needs to address both binary classification (``Yes'') and probability calculation (``-0.1505'') questions. See Figure \ref{fig_scenario:ETT} for a detailed illustration.

\subsubsection{Natural Direct Effect (NDE)}
\label{counterfactual:nde}

Different from the CDE, the NDE quantifies the anticipated rise in outcome $Y$ when the treatment shifts from $X = x$ to $X = x'$, with the mediator $M$ held at the value it would have naturally taken under the condition $X = x$ \citep{pearl2001effects}. Thus, $NDE=E[(Y(x',M(x)))-E(Y(x))]$. We can conclude that the most distinct difference between CDE and NDE is that: NDE considers the mediator's natural state when untreated, whereas CDE sets the mediator to one or more predetermined levels.
Figure \ref{fig_main:scenario_real3}(b) depicts a causal scenario of studying how smoking (treatment) affects heart disease (outcome), with blood pressure (mediator) acting as a crucial intermediary factor. Here, the NDE represents the direct influence of smoking on heart disease, bypassing the effects mediated by blood pressure. And it seeks to answer: \emph{What would be the direct effect of smoking on heart disease if we could keep the blood pressure of smokers at the level it would naturally be if they did not smoke?}
The NDE allows for isolating and understanding the direct impact of a treatment on an outcome, independent of any mediating pathways. This distinction is crucial in fields like computer vision \citep{niu2021counterfactual}, natural language processing \citep{vig2020investigating}, and public health \citep{carter2021mendelian}, where understanding the specific mechanisms through which interventions affect outcomes can inform the development of more effective strategies.

\begin{figure}[t]
    \centering
    \includegraphics[width=\textwidth]{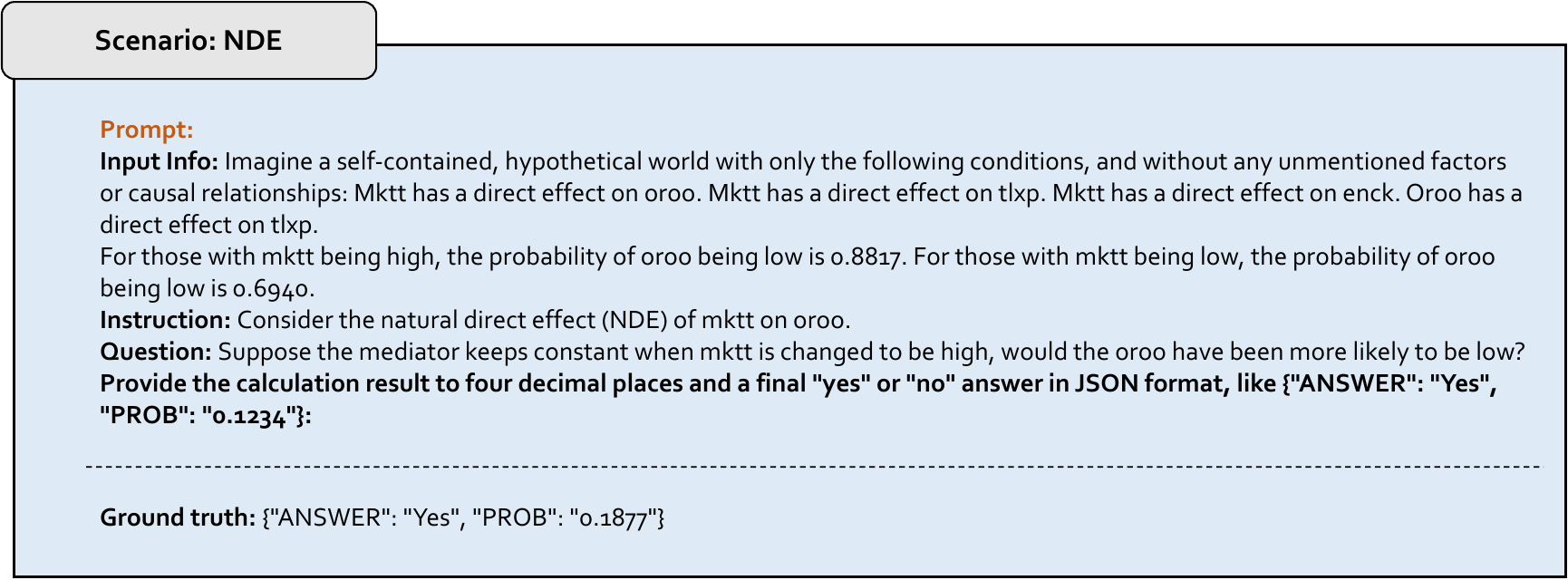}
    \caption[Example of natural direct effect]{\textbf{Example of natural direct effect.}}
    \label{fig_scenario:NDE}
\end{figure}

\paragraph{Causal scenario setting.} 
Given a causal graph (e.g., ``\emph{Mktt has a direct effect on oroo. Mktt has a direct effect on tlxp. Mktt has a direct effect on enck. Oroo has a direct effect on tlxp.}'') along with corresponding conditional probabilities (e.g., ``\emph{For those with mktt being high, the probability of oroo being low is 0.8817. For those with mktt being low, the probability of oroo being low is 0.6940.}''), the causal scenario requires the model to determine the NDE between variables (e.g., ``\emph{Suppose the mediator keeps constant when mktt is changed to be high, would the oroo have been more likely to be low?}''). The model needs to address both Binary classification (``Yes'') and probability calculation (``0.1877'') types of questions. See Figure \ref{fig_scenario:NDE} for a detailed illustration.

\subsubsection{Natural Indirect Effect (NIE)}
\label{counterfactual:nie}

\begin{figure}[t]
    \centering
    \includegraphics[width=\textwidth]{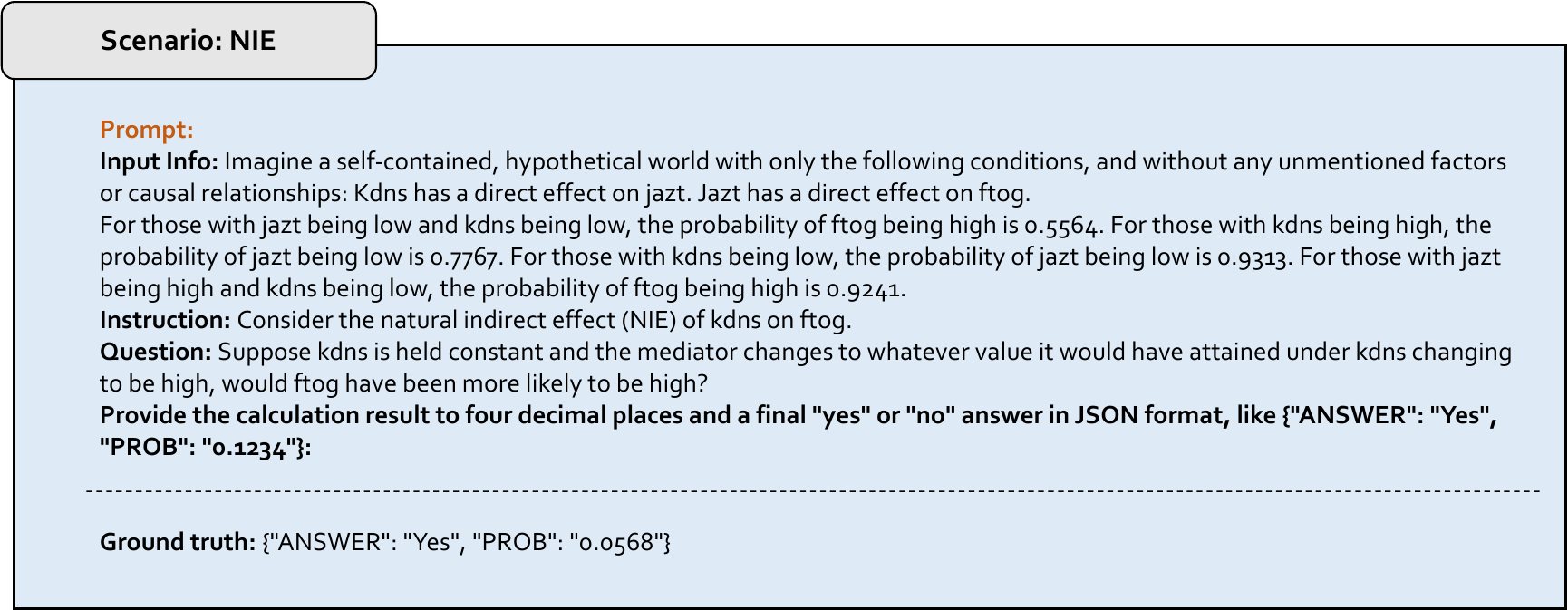}
    \caption[Example of natural indirect effect]{\textbf{Example of natural indirect effect.}}
    \label{fig_scenario:NIE}
\end{figure}

NIE measures the extent of change in the outcome through the mediator when the treatment is modified. It excludes the direct effects of treatment on the outcome that does not involve the mediator. This methodology enables us to comprehend the mediator's role and significance within the causal relationship between treatment and outcome. Specifically, the NIE quantifies the anticipated rise in $Y$ when the treatment variable remains fixed at $X = x$, while allowing the mediator $M$ to adjust to the level it would have reached if $X$ had been set to $x'$. In essence, it isolates and accounts for the segment of the effect solely attributable to the mediation process, while neutralizing the ability of $Y$ to react to changes in $X$ \citep{pearl2001effects}. The quantitative expression is $NIE=E[(Y(x,M(x')))-E(Y(x))]$. As Figure \ref{fig_main:scenario_real3}(c) shows, consider a study on examining how job satisfaction (treatment) improves employees' mental health (outcome) by reducing job-related stress (mediator).
In this example, the NIE would measure the improvement in employees' mental health resulting solely from the increase in job satisfaction through the pathway of reducing job-related stress.
\paragraph{Causal scenario setting.}
Similar to NDE, we provide a causal graph (e.g., ``\emph{Kdns has a direct effect on jazt. Jazt has a direct effect on ftog.\footnote{Inspired by \citet{jin2023cladder}, our datasets for ATE, CDE, ETT, NDE, NIE, PN, and PS are each configured with three different types of authenticity: \emph{real}, \emph{random}, and \emph{fake}. This given example belongs to the \emph{fake} type. Here, \emph{fake} indicates that the nodes in the causal graph consist of meaningless combinations of letters. For more information, please refer to \nameref{data:construction} (\cref{data:construction})}}'' with corresponding conditional probabilities (e.g., ``\emph{For those with jazt being low and kdns being low, the probability of ftog being high is 0.5564. For those with kdns being high, the probability of jazt being low is 0.7767. For those with kdns being low, the probability of jazt being low is 0.9313. For those with jazt being high and kdns being low, the probability of ftog being high is 0.9241.}'') The causal scenario requires the model to determine the NIE between variables (e.g., ``\emph{Suppose kdns is held constant and the mediator changes to whatever value it would have attained under kdns changing to be high, would ftog have been more likely to be high?}''). The model needs to address both binary classification (``Yes'') and probability calculation (``0.0568'') questions. See Figure \ref{fig_scenario:NIE} for a detailed illustration.

\subsubsection{Probability of Necessity (PN)}
\label{counterfactual:pn}

\begin{figure}[t]
    \centering
    \includegraphics[width=\textwidth]{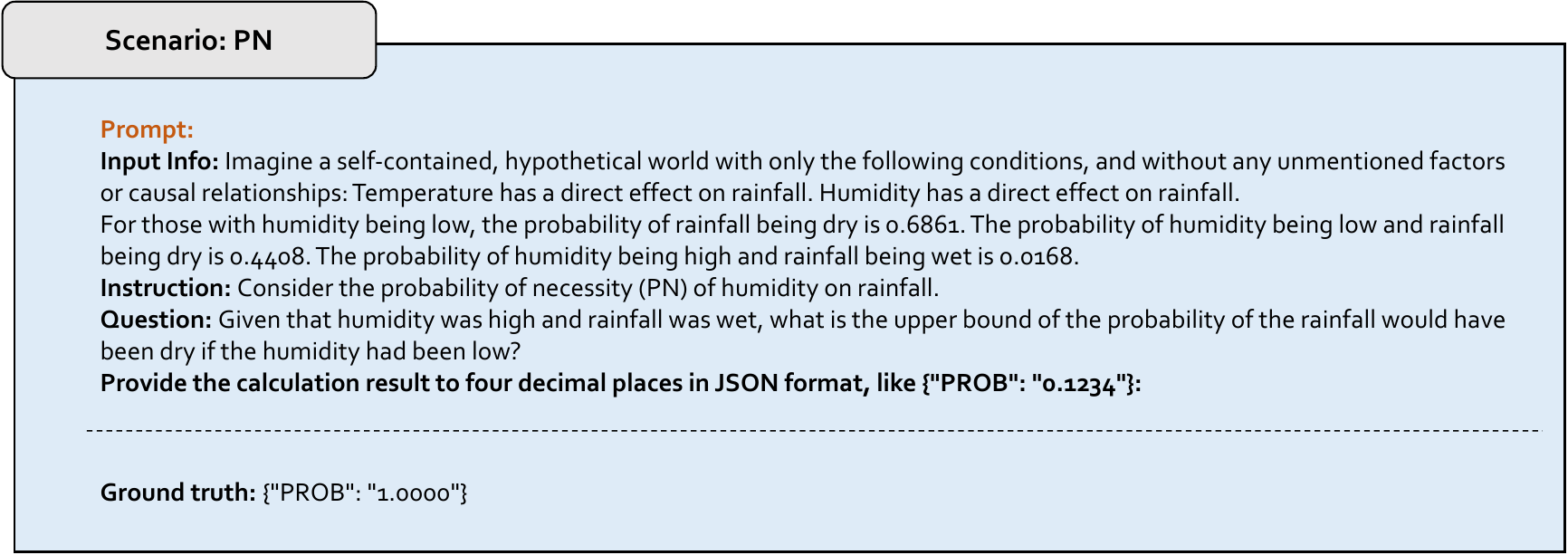}
    \caption[Example of probability of necessity.]{\textbf{Example of probability of necessity.}}
    \label{fig_scenario:PN}
\end{figure}

PN essentially seeks to address the question: ``\emph{In cases where the outcome occurs, could it still happen without the treatment?}'' If the absence of the treatment leads to that the outcome would not happen, then it indicates that the treatment is necessary for the occurrence of the outcome \citep{pearl2022probabilities}. In the context of binary events, we denote the treatment as $X = x$ and the outcome as $Y = y$, and their respective negations as $X = x'$ and $Y = y'$. The objective of PN can be determined as: \emph{Find the probability that if $X$ had been $x'$, $Y$ would be $y'$, given that, in reality, $X$ is $x$ and $Y$ is $y$} \citep{pearl2016causal}. The formula is $PN(x,y)=P(Y_{x'}=y'|X=x,Y=y)$. 
PN provides deeper insight into the fundamental principle of legal assessment known as the ``but-for'' test. In practical legal contexts, this guideline directs us: favorable judgments for the plaintiff should only be rendered when the assumption that ``\emph{the harm would not have occurred if not for the defendant's actions}'' approaches certainty \citep{peaslee1934multiple}. For example, in a traffic accident case where a driver failed to slow down at a yellow light leading to a collision, the court uses the PN to evaluate if slowing down could have prevented the accident. If PN indicates a high likelihood that the accident would have been avoided by slowing down, the driver's action is deemed a necessity cause for the collision, influencing the court's decision on liability. 
Formalizing and calculating the PN is vital for allocating resources efficiently, prioritizing interventions, and crafting strategies that address the most critical factors contributing in artificial intelligence \citep{pmlr-v161-watson21a,tan2022learning,tan2021counterfactual,zhang2022partial}. 
\paragraph{Causal scenario setting.}
Given a causal graph (e.g., ``\emph{Temperature has a direct effect on rainfall. Humidity has a direct effect on rainfall.}'') and corresponding conditional probabilities (e.g., ``\emph{For those with humidity being low, the probability of rainfall being dry is 0.6861. The probability of humidity being low and rainfall being dry is 0.4408. The probability of humidity being high and rainfall being wet is 0.0168.}''), the causal scenario requires the model to calculate the upper bound or lower bound of PN (e.g., ``\emph{Given that humidity was high and rainfall was wet, what is the upper bound of the probability of the rainfall would have been dry if the humidity had been low?}''). See Figure \ref{fig_scenario:PN} for a detailed illustration.

\subsubsection{Probability of Sufficiency (PS)}
\label{counterfactual:ps}

\begin{figure}[t]
    \centering
    \includegraphics[width=\textwidth]{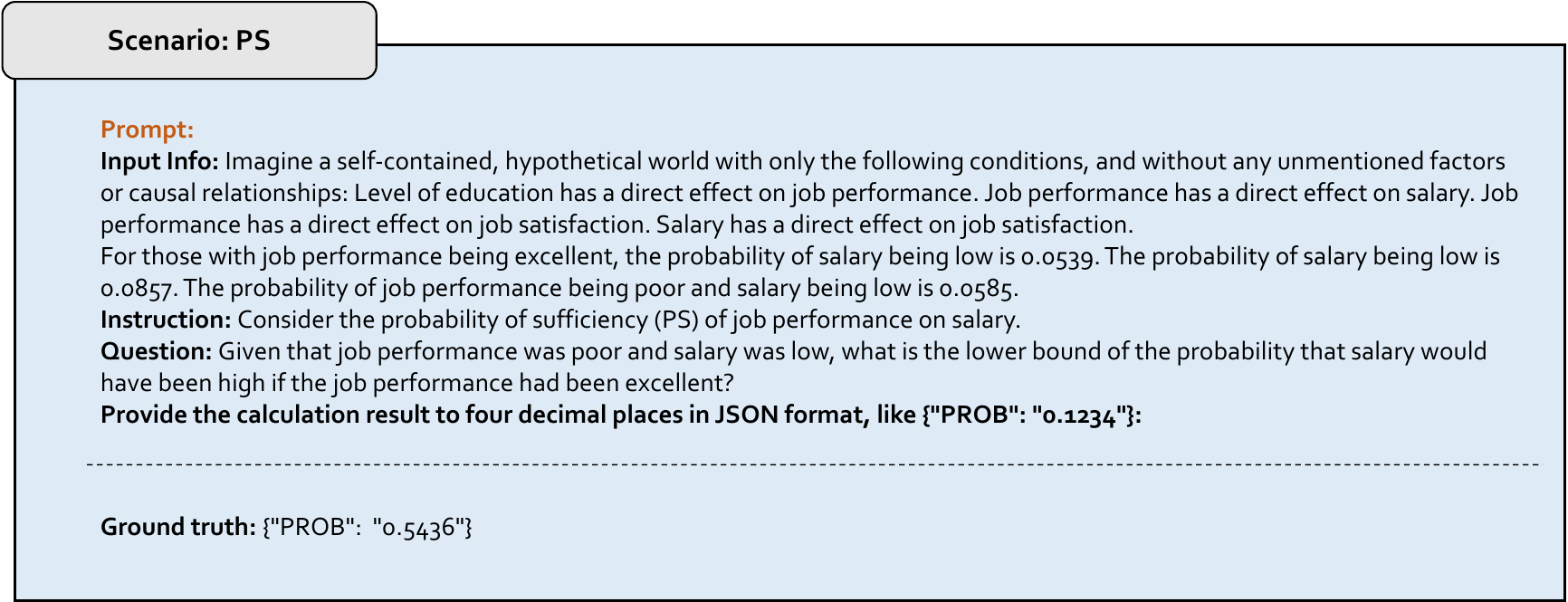}
    \caption[Example of probability of sufficiency.]{\textbf{Example of probability of sufficiency.}}
    \label{fig_scenario:PS}
\end{figure}

Parallel to the concept of PN, the PS addresses: ``\emph{In cases where the outcome does not occur, could it happen if a treatment exists?}" If the presence of this treatment leads to the outcome, it implies that this treatment is sufficient to trigger the outcome \citep{pearl2022probabilities}.
It provides an estimate of the probability that the intervention of $x$ would result in the occurrence of outcome $y$ when both $x$ and $y$ are not currently present \citep{pearl2009causality}, mathematically expressed as: $PS(x,y)=P(Y_{x}=y|X=x',Y=y')$. Consider a workplace injury case where an employee was harmed due to a machine malfunction, the court applies the PS to evaluate if adhering to safety protocols. Considering that regular maintenance $(X=1)$ would have prevented the injury $(Y=1)$, given the protocols were initially not followed $(X=0)$ and the injury occurred $(Y=0)$. A high PS indicates that following the protocols would likely have averted the harm, potentially establishing the employer's liability for the injury by demonstrating negligence in safety measures. This approach is critical in legal judgments involving causality and negligence.
\paragraph{Causal scenario setting.}
Similar to PN, PS also involves providing a causal graph along with corresponding conditional probabilities, requiring the model to calculate the upper bound or lower bound of PS. See Figure \ref{fig_scenario:PS} for a detailed illustration.

\subsubsection{Counterfactual Reasoning (CR)}
\label{counterfactual:cr}

\begin{figure}[t]
    \centering
    \includegraphics[width=\textwidth]{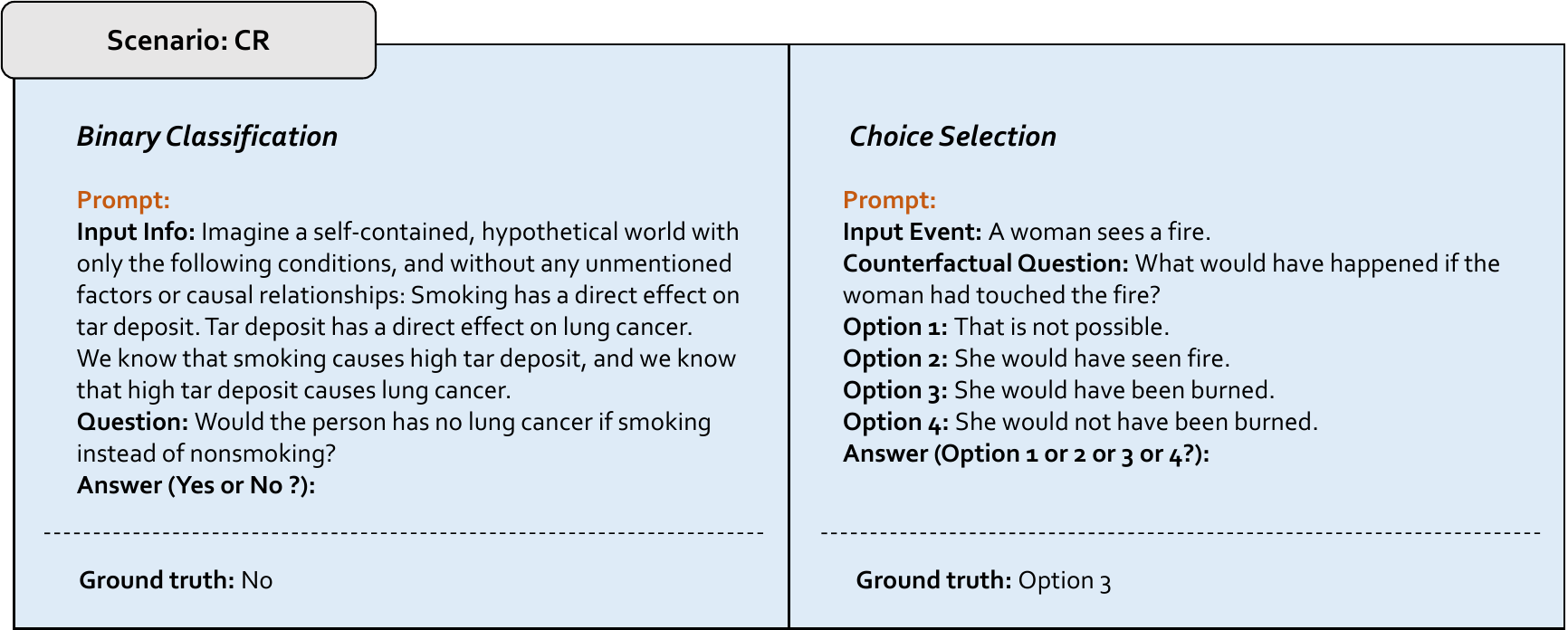}
    \caption[Example of counterfactual reasoning.]{\textbf{Example of counterfactual reasoning.}}
    \label{fig_scenario:CR}
\end{figure}

To ascertain what led to a specific occurrence, it is crucial to imagine ``what-if'' causal scenarios in which the event might not have taken place and then examine the resulting outcomes. Counterfactual reasoning involves contemplating hypothetical causal scenarios or alternative versions of reality by modifying certain factors or conditions present in an actual event or situation \citep{kahneman1986norm,byrne2007rational}. It is also a valuable skill for language models to provide perspectives and insights that might not be immediately obvious based on the available information.
\paragraph{Causal scenario setting.}
There are two types of questions in CR. (1) \textbf{Binary classification}: We provide the model with a causal graph (e.g., ``\emph{Smoking has a direct effect on tar deposit. Tar deposit has a direct effect on lung cancer.}''\footnote{The example is chosen from \citet{jin2023cladder}.}) and corresponding relationships (e.g., ``\emph{We know that smoking causes high tar deposit, and we know that high tar deposit causes lung cancer.}''), and give the model a counterfactual question to answer ``yes'' or ``no'' (e.g., ``\emph{Would the person has no lung cancer if smoking instead of nonsmoking?}''). (2) \textbf{Choice selection}: An event (e.g., ``\emph{A woman sees a fire.}'') and a counterfactual question (e.g., ``\emph{What would have happened if the woman had touched the fire?}''\footnote{The example is chosen from \citet{frohberg2022crass}.}) are posed to the model. The model is required to choose the correct option. See Figure \ref{fig_scenario:CR} for a detailed illustration.

\clearpage


\section{Data Collection}
\label{main:data}
In this section, we provide a detailed discussion on the utilization of our datasets, which is crucial for understanding and reproducing the CaLM. First, we describe the methodology behind our dataset selection in \nameref{data:selection} (\cref{data:selection}). Here, we offer a thorough justification for each chosen dataset to ensure our selection process is transparent and well-founded. Second, we depict how self-constructed datasets are developed in \nameref{data:construction} (\cref{data:construction}), providing a detailed guide for researchers interested in creating their own datasets for use with CaLM. Our goal extends beyond merely facilitating the use of our datasets; we aim to openly share our dataset construction methodologies to foster the growth and development of the causal research community. Finally, in \nameref{data:statistics} (\cref{data:statistics}), we perform a detailed statistical analysis of the datasets, offering clear insights into the scale and scope of CaLM.

\subsection{Dataset Selection}
\label{data:selection}

As highlighted by \citet{openai2023gpt4}, the ability to tackle challenging tasks is a key factor differentiating language models. Therefore, the primary objective of CaLM is to evaluate the abilities of language models to undertake complex causal reasoning tasks. To achieve this goal, we have selected 31 datasets, which can be further subdivided into 44 subsets, for comprehensive evaluation.
Approximately 10\% of the data used in our benchmark originates from pre-existing, publicly available datasets that are specifically aligned with causal tasks in the Natural mode. The remaining 90\% consists of datasets that we have developed. 
We begin with an overview presented in \cref{table_models_selection}, offering an immediate and clear outline of these resources.
A more detailed exposition of the datasets from these distinct sources (\nameref{data:open} (\cref{data:open}) and \nameref{data:self} (\cref{data:self})) is provided below. 
\begin{center}
\begin{table*}[t!]
\fontsize{10}{12}\selectfont
    \caption[Datasets selection of CaLM]{\textbf{Datasets selection of CaLM.} The selected datasets are organized by their source (i.e., open-source or self-constructed). In cases where datasets include subset\footnotemark, we make extra distinctions at the subset level. The table additionally outlines the causal scenarios utilizing these datasets, along with their respective levels of the causal ladder, mode, and language.}
    \label{table_models_selection}
    \centering
  \begin{tabular}{c|c|c|c|c|c}
\toprule
\textbf{Dataset} & \textbf{Subset}& \textbf{Causal ladder}&  \textbf{Causal scenario}&\textbf{Mode}&\textbf{Language}\\
\hline
\multicolumn{6}{c}{\cellcolor{blue!5}\emph{Open-source Datasets}}\\
\hline
COPA& -& Causal discovery& PCD&\multirow{13}{*}{Natural}& \multirow{13}{*}{EN}\\
E-CARE& -& Causal discovery& PCD&& \\
CTB& -& Causal discovery &ECI&& \\
ESC& -& Causal discovery &ECI&&\\
MAVEN-ERE& -& Causal discovery& ECI&&\\
CLADDER& correlation& Association& CORR&&\\
CLADERR& exp-away& Association& EAE&&\\
CLADDER& backadj& Intervention& BAS&&\\
CLADDER& collider-bias& Intervention& CB&&\\
CLADDER& det-counterfactual& Counterfactual& CR&&\\
CRASS& -& Counterfactual& CR&& \\
BBH& causal judgement& Counterfactual& AC&& \\
E-CARE& -& Counterfactual& CEG&& \\
\hline
\multicolumn{6}{c}{\cellcolor{green!5}\emph{Self-constructed Datasets}}\\
\hline
CaLM-AR& -& Causal discovery& AR&Symbolic& \multirow{18}{*}{EN\&CN}\\
CaLM-CA& FP/FA& Causal discovery& CA&Symbolic& \\
CaLM-AS& max/min/mix-BAS& Intervention& BAS&Symbolic& \\
CaLM-AS& FAS& Intervention& FAS&Symbolic& \\
CaLM-IV& -& Intervention& IV&Symbolic& \\
CaLM-CEI& 0.2/0.4/0.6/0.8-UC& Intervention& CEI&Symbolic& \\
CaLM-ATE& ATE-basic/hard& Intervention& ATE&Mathematical& \\
CaLM-ATE& ATE-natural& Intervention& ATE&Natural& \\
CaLM-CDE& CDE-basic/hard& Intervention& CDE&Mathematical& \\
CaLM-CDE& CDE-natural& Intervention& CDE&Natural& \\
CaLM-ETT& ETT-basic/hard& Counterfactual& ETT&Mathematical& \\
CaLM-ETT& ETT-natural& Counterfactual& ETT&Natural& \\
CaLM-NDE& NDE-basic/hard&Counterfactual& NDE&Mathematical& \\
CaLM-NDE& NDE-natural&Counterfactual& NDE&Natural& \\
CaLM-NIE& NIE-basic/hard&Counterfactual& NIE&Mathematical& \\
CaLM-NIE& NIE-natural&Counterfactual& NIE&Natural& \\
CaLM-PN& PN-basic/hard&Counterfactual& PN&Mathematical& \\
CaLM-PS& PS-basic/hard&Counterfactual& PS&Mathematical& \\
\cline{1-6}
All open-source datasets&-&All 4 rungs&-&Natural&CN\\
\hline
\end{tabular}
\end{table*}
\end{center}
\footnotetext{In open-source datasets, CLADDER consists of 10 subsets, out of which we select 5. Meanwhile, BBH contains 23 subsets, from which we opt for the causal judgment subset.}

\subsubsection{Open-source Datasets}
\label{data:open}
All the open-source datasets we select belong to the Natural mode. To clarify them further, we will elaborate on each dataset based on the causal ladder, corresponding to the causal scenario it addresses (for more detailed information about the causal scenario, please refer to \nameref{main:target}(\cref{main:target})).

\paragraph{Causal discovery.}
(1) \textbf{COPA} (Choice Of Plausible Alternatives) is developed by \citet{roemmele2011choice}. This dataset focuses on determining causal relationships, consisting of a total of 1000 queries. Each query presents a premise along with two potential causes or effects. The objective is to identify the accurate causal relationship grounded on the information provided in the premise. 
(2) \textbf{E-CARE} (Explainable CAusal REasoning) \citep{du2022care} includes over 21,000 multiple-choice questions centered on causal reasoning. It goes beyond simple queries by offering detailed conceptual explanations for each question, elucidating the rationale behind the causal relationships.
(3) \textbf{CTB} (Causal-TimeBank) originates from the TimeBank corpus by \citet{pustejovsky2006timebank} and is used in the TempEval-3 task \citep{mirza2014annotating}. This dataset consists of 6,813 events and 318 causal event pairs. 
 (4)\textbf{ESC} (Event StoryLine Corpus) \citep{caselli2017event} is designed to facilitate the identification of temporal and causal relations. 
(5)\textbf{MAVEN-ERE} (Event Relation Extraction) \citep{wang2022maven} is an expansive resource developed from the MAVEN dataset \citep{wang2020maven}. The MAVEN dataset is a comprehensive tool for detecting events across a wide range of domains, containing 4,480 documents from English Wikipedia. Building upon this, the MAVEN-ERE dataset introduces a substantial collection of 57,992 causal relations, making the task of ERE on it complex and demanding. For more background information on the PCD and ECI causal scenarios, please refer to \nameref{scenario:discovery} (\cref{scenario:discovery}).

\paragraph{Association.}
The \textbf{CLADDER} dataset, developed by \citet{jin2023cladder}, consists of over 10,000 causal questions categorized by varying levels of complexity across multiple levels of causal ladder. The dataset includes a wide range of causal graphs that visually represent complex relationships among various factors. It is further enriched by narratives that provide context for the questions, illustrating real-world causal scenarios or hypothetical situations where these relationships are pivotal. We choose the \texttt{\small cladder-v1-aggregate.json} for our evaluation. It should be noted that we use the initial version of this document, which has been subject to updates over time.\footnote{The version of the document we are using contains 227 backadj questions, in contrast to the 1644 questions found in the most recent version.} In this rung, we use two subsets of CLADDER: \textbf{correlation} and \textbf{exp-away}. These two subsets contain 1476 and 168 data points, respectively. For more details on these two causal scenarios, please refer to \nameref{scenario:association} (\cref{scenario:association}).

\paragraph{Intervention.}
In this rung, we use two subsets of CLADDER: \textbf{backadj} and \textbf{collider-bias}. These two subsets contain 227 and 168 samples, respectively. For additional details on these two causal scenarios, please refer to \nameref{scenario:intervention} (\cref{scenario:intervention}).

\paragraph{Counterfactuals.}
(1) \textbf{CRASS} (Counterfactual Reasoning Assessment) \citep{frohberg2022crass} is a vital tool for analyzing the proficiency of language models in handling question-based counterfactual conditionals. It comprises 275 carefully crafted queries designed to test the models' ability to interpret and respond to counterfactual reasoning challenges. These queries are part of the BIG-bench project \citep{srivastava2023beyond}. (2) The \textbf{causal judgment}\footnote{The creators named it as causal judgment, but we use it for actual causality.} dataset, part of the \textbf{BIG-bench Hard} \citep{suzgun2022challenging}, consists of 187 narratives that effectively demonstrate actual causality \citep{halpern_actual_2016}, reflecting the natural human tendency to assign cause, responsibility, and blame to events and their respective outcomes. Each story concludes with a clear yes/no question, providing a structured evaluation of language models' comprehension and interpretation of causal relationships and associated concepts. 
(3) \textbf{det-counterfactual} is a subset from CLADDER, featuring 1476 questions about counterfactual reasoning with a causal graph. For more detailed information on the CR and AC causal scenarios, please refer to \nameref{scenario:counterfactual} (\cref{scenario:counterfactual}).

\subsubsection{Self-constructed Datasets}
\label{data:self}
The illustration for this section will be divided into two categories: (i) Symbolic, and (ii) Natural and Mathematical. To begin with, it is important to note that each of the datasets we have constructed contains 1600 samples. This sample size is considered sufficient for evaluation purposes while also preventing resource wastage. Regarding language, all datasets from \nameref{data:open} (\cref{data:open}) are exclusively in English; hence, we translate them into Chinese. Additionally, our self-constructed datasets are available in both English and Chinese versions. Moreover, the core component of the self-constructed datasets is the causal graph. That is, all of our data samples are composed of a causal graph, along with questions derived from this graph.

\begin{figure}[t]
    \centering
    \includegraphics[width=\textwidth]{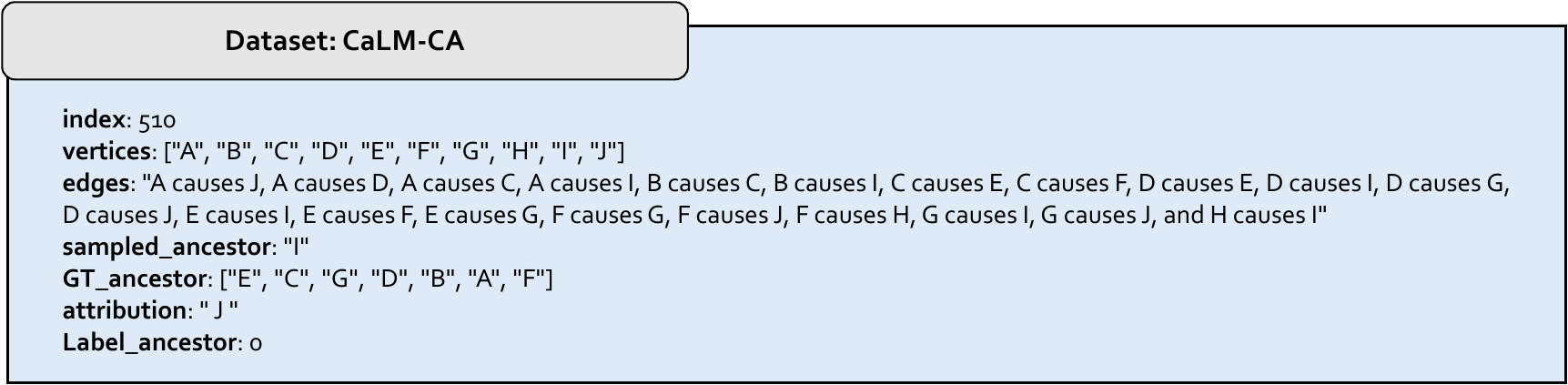}
    \caption[An example of CaLM-CA dataset]{\textbf{An example of CaLM-CA dataset.} }
\label{fig_data:symbolic_CA}
\end{figure}

\begin{figure}[t]
    \centering
    \includegraphics[width=\textwidth]{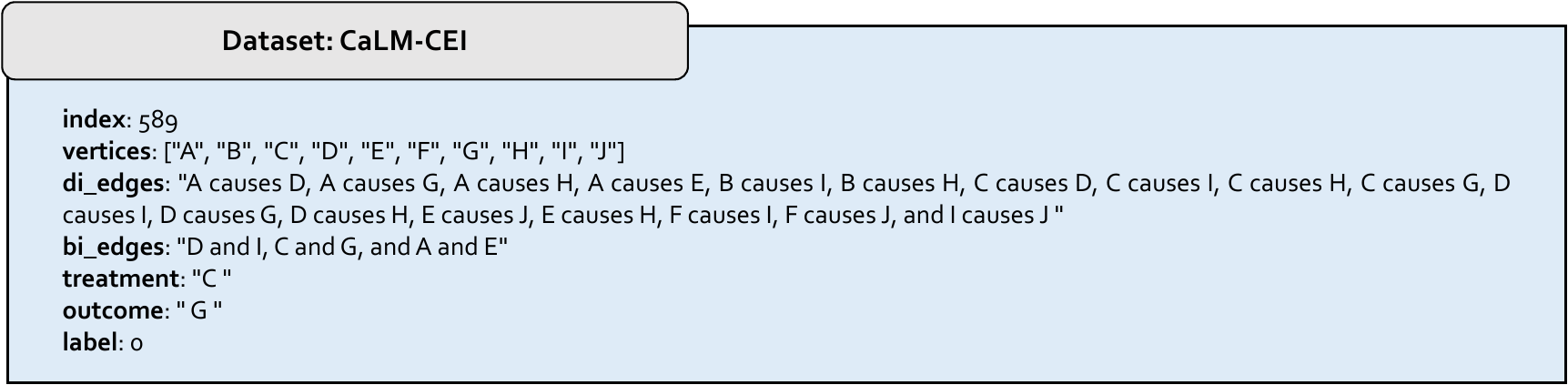}
    \caption[An example of CaLM-CEI dataset]{\textbf{An example of CaLM-CEI dataset.} }
\label{fig_data:symbolic_identification}
\end{figure}

\begin{figure}[t]
    \centering
    \includegraphics[width=\textwidth]{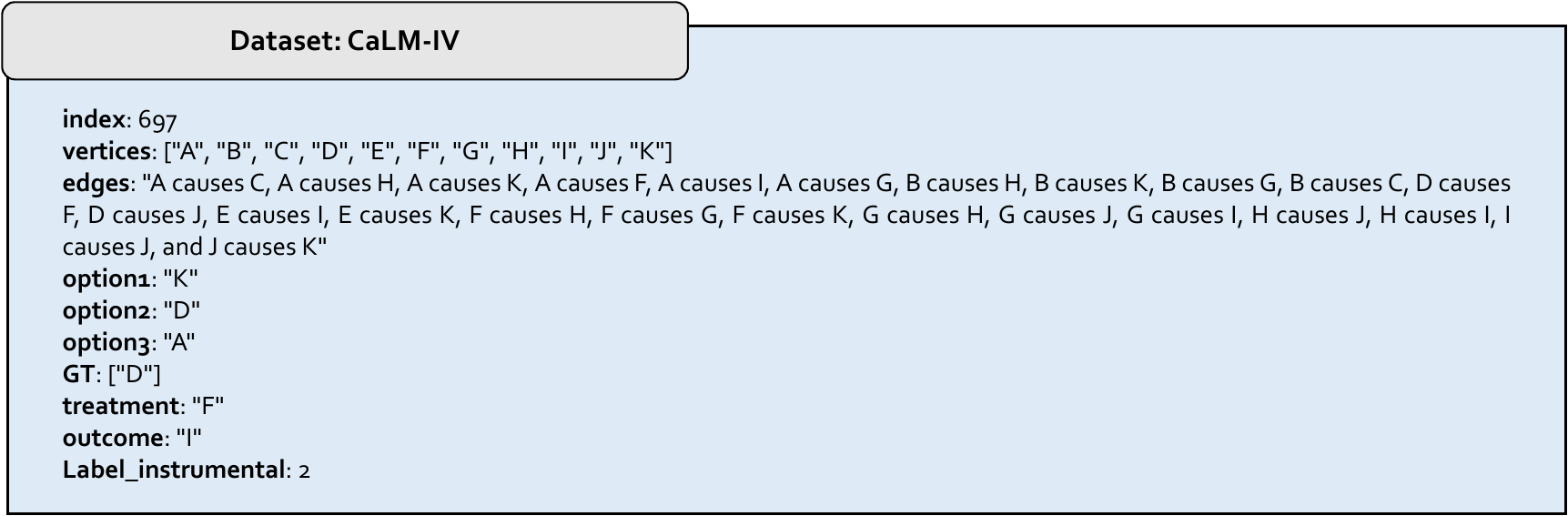}
    \caption[An example of CaLM-IV dataset]{\textbf{An example of CaLM-IV dataset.} }
\label{fig_data:symbolic_IV}
\end{figure}

\begin{figure}[t]
    \centering
    \includegraphics[width=\textwidth]{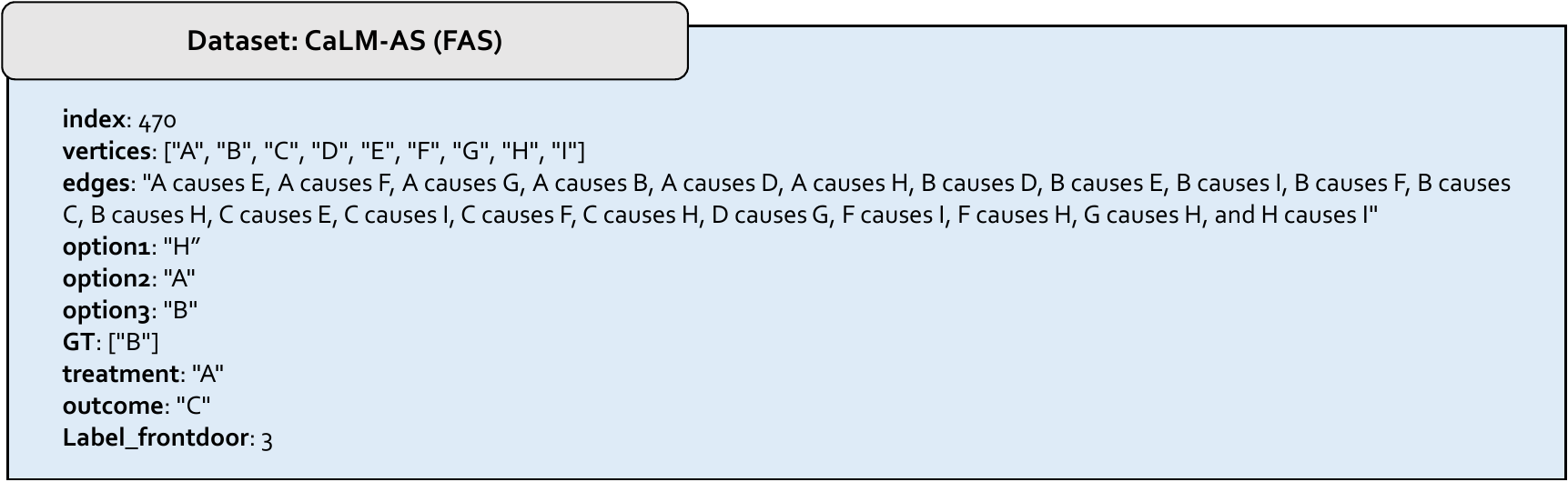}
    \caption[An example of CaLM-AS dataset]{\textbf{An example of CaLM-AS dataset.} }
\label{fig_data:symbolic_FAS}
\end{figure}

\paragraph{Symbolic.}
(1) \textbf{CaLM-AR} (Abstract Reasoning) aims to assess language models' ability to accurately identify causal relationships within a graph, despite the presence of additional disturbances. This assessment is crucial in measuring the models' genuine understanding of complex causal graphs. It tests if the models can effectively filter out extraneous noise to discern true causal pathways, thus highlighting their deep understanding of causal relationships. Each data sample within this dataset is accompanied by a causal graph that describes the relationships among various elements in a non-alphabetical sequence (e.g., it might indicate that ``\emph{A causes B, A causes D, B causes C, C causes A.}'' rather than a straightforward ``\emph{A causes B, B causes C and C causes D.}''), which introduces an additional layer of complexity to challenge the language models further. 
(2) \textbf{CaLM-CA} (Causal Attribution) is designed to evaluate whether the model can understand the direct and indirect influences that variables or events (represented as nodes) exert on each other within a system. They are composed of 2 subsets: \textbf{FP} (Find Parent) and \textbf{FA} (Find Ancestor). As their names suggest, these datasets challenge the model to identify hierarchical relationships between nodes, specifically determining whether a node is a parent or an ancestor of another in the graph. An example of the FA dataset is illustrated in Figure \ref{fig_data:symbolic_CA}.
(3) \textbf{CaLM-CEI} (Causal Effect Identification) is crafted to test a model's ability to determine whether the causal effect of a treatment on an outcome can be estimated from observational data. This dataset is segmented into four distinct subsets: \textbf{0.2-UC} (Unobserved Confounder), \textbf{0.4-UC}, \textbf{0.6-UC} and \textbf{0.8-UC}, each characterized by a different proportion of unobserved confounders affecting 20\%, 40\%, 60\%, and 80\% of the nodes in the graph, respectively. As the percentage of unobserved confounders increases, the complexity of identifying causal effects also rises. An example is provided in Figure \ref{fig_data:symbolic_identification}.
(4) \textbf{CaLM-AS} (Adjustment Set) is used to assess a model's understanding of causal graphs and the concepts of \emph{Back-door Criterion} and \emph{Front-door Criterion} \citep{pearl1995causal}. The \emph{Back-door Criterion} provides a method for identifying sets of variables that need to be controlled for to estimate causal effects from observational data. The \emph{Front-door Criterion} can be applied in cases where no variable satisfies the \emph{Back-door Criterion}. To be specific, the \emph{Front-door Criterion} helps in estimating the causal effect through the mediator, regarding the existence of unobserved confounders. This dataset is divided into four subsets: \textbf{max-BAS} (maximal-Backdoor Adjustment Set), \textbf{min-BAS} (minimal-Backdoor Adjustment Set), \textbf{mix-BAS} (mix-Backdoor Adjustment Set), and \textbf{FAS} (Frontdoor Adjustment Set). In the context of a specified causal graph in which an ordered pair of variables is considered, the minimal/max backdoor set includes the collection of variables that either minimally or maximally meet the backdoor criterion. The Mix set is a combination of both. The front-door set comprises variables that fulfill the \emph{Front-door Criterion}. An example of the FAS dataset is demonstrated in Figure \ref{fig_data:symbolic_FAS}.
(5)  \textbf{CaLM-IV} (Instrumental Variable) is designed to assess a model's capability to determine the independence among variables and its understanding of the IV concept. Accurate identification of an instrumental variable allows for estimating causal effects even in the presence of unobserved confounders. The setup of this dataset closely mirrors that of the CaLM-AS, involving a causal graph and specifying a cause-effect pair. Figure \ref{fig_data:symbolic_IV} provides an example.

\begin{figure}[t]
    \centering
    \includegraphics[width=\textwidth]{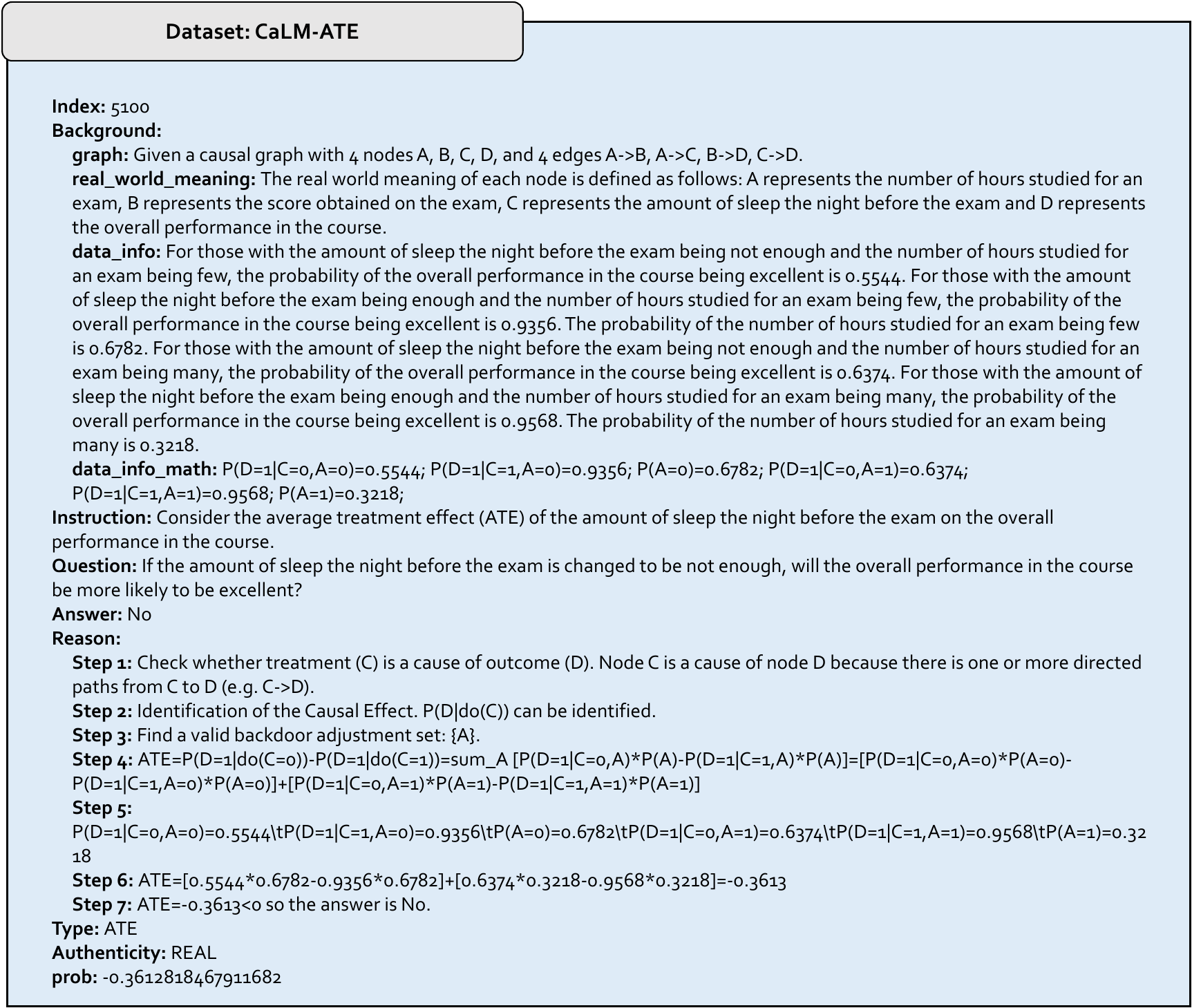}
    \caption[An example of CaLM-ATE dataset]{\textbf{An example of CaLM-ATE dataset.}}
    \label{fig_data:math_ATE}
\end{figure}

\paragraph{Natural and Mathematical.}
(6) \textbf{CaLM-ATE} (Average Treatment Effect) is used to evaluate the analytical and computational abilities of a language model in predicting changes to an outcome variable following an intervention on the treatment variable, given a causal graph and data distribution. This dataset is divided into three subsets: \textbf{ATE-natural} focuses on qualitative analysis, with responses limited to ``Yes'' and ``No''; \textbf{ATE-basic} and \textbf{ATE-hard} both require the quantitative calculation of ATE, whose main distinction lies in the complexity of the causal graphs (ATE-basic features graphs with fewer nodes). Note that the subsequent datasets (7)-(10) introduced below are similarly divided into three subsets - \textbf{X-natural}, \textbf{X-basic} and \textbf{X-hard} - a classification that will not be reiterated further in this context. An example of ATE-basic is shown in Figure \ref{fig_data:math_ATE}.
(7) \textbf{CaLM-CDE} (Controlled Direct Effect) evaluates the model's ability to calculate the direct effect of the treatment variable on the outcome variable while keeping one or more mediators fixed. 
(8) \textbf{CaLM-ETT} (Effect of the Treatment on the Treated) accesses the model's capability to calculate the treatment effect on those who have already been treated. 
(9) \textbf{CaLM-NDE} (Natural Direct Effect) measures the model's ability to compute the natural direct effect, assuming mediators remain constant during the intervention on treatment variables.
(10) \textbf{CaLM-NIE} (Natural Indirect Effect) tests the model's ability to calculate the natural indirect effect by altering mediators to the values they would have attained under the intervention, while the treatment variable itself keeps constant.
(11) \textbf{CaLM-PN} (Probability of Necessity) aims to evaluate the model's capacity to estimate the upper and lower bounds of the probability of necessity, which measures the necessity of the treatment for those who received it and experienced a positive outcome. This dataset is divided into two subsets, \textbf{PN-basic} and \textbf{PN-hard}, based on the number of nodes in the causal graphs.
(12) \textbf{CaLM-PS} (Probability of Sufficiency) focuses on the model's ability to estimate the upper and lower bounds of the probability of sufficiency, which evaluates the sufficiency of treatment for those who did not receive treatment but had a negative outcome. Similar to CaLM-PN, this dataset is also segmented into \textbf{PS-basic} and \textbf{PS-hard} based on the complexity of the causal graphs.

\subsection{Dataset Construction}
\label{data:construction}

The construction of our data consists of three parts: \nameref{data_construction:math} (\cref{data_construction:math}), \nameref{data_construction:symbolic} (\cref{data_construction:symbolic}), and \nameref{data_construction:chinese} (\cref{data_construction:chinese}).
The initial steps in constructing the first two parts are identical, as both require the creation of a DAG, necessitating a common introduction to these shared procedures. We will begin by presenting the shared methodology (i.e., \nameref{data_construction:DAGs} (\cref{data_construction:DAGs})). Following this, we will proceed to discuss the unique elements and detailed methodologies specific to each of the three datasets. 

\subsubsection{Generating DAGs}
\label{data_construction:DAGs}
In this paper, we employ structural causal models (SCMs), as detailed in Section \ref{preliminary:scm}, to construct the underlying ground-truth causal model. We first generate several directed acyclic graphs (DAG) at random. For graphs with no more than 6 vertices, we follow \citet{jin2023large} to construct all possible weakly connected DAGs without isomorphism between each other. Given $N$ nodes $\{X_i\}_{i=1}^{N}$, only edges $X_i \rightarrow X_j $ where $i < j$ are allowed to ensure it acyclic, yielding $2^{N(N-1)/2}$ candidate graphs. We further remove graphs which are not weakly connected or isomorphic with existing graphs. However, we observed that the above algorithm runs extremely slow when $N > 6$. Considering that we do not actually need all possible graphs for generating causal inference data for large $N$, an alternative strategy is adopted that randomly generates a small number of graphs instead of all possible graphs in the cases of $N>6$, see Algorithm \ref{alg:gen_dag}.

\begin{algorithm}[t]
\caption{Algorithm for randomly generating DAGs}
\label{alg:gen_dag}
\begin{algorithmic}
\Statex \textbf{Input:} node number $N>1$, graph number $M\ge 1$, maximum attempt number $T\ge 1$ 
\Statex \textbf{Output:} A set of graphs $G$
\State $G \leftarrow \{\}$
\State $V \leftarrow \{1,2,\dots, N\}$
\For {$m=1,2,\dots, M$}
    \For {$t=1,2,\dots, T$}
        \State $e \leftarrow \text{randint}(N-1, N(N-1)/2)$
        \State $E \leftarrow $ randomly sample $e$ edges from all $N(N-1)/2$ possible edges
        \State $g \leftarrow \text{Graph}(V, E)$
        \If {$g$ is weakly connected \textbf{and} $g$ is not isomorphism with any $ g' \in G$}
            \State {$G.\text{add}(g)$}
            \State \textbf{break}
        \EndIf 
    \EndFor
\EndFor
\State \Return $G$
\end{algorithmic}
\end{algorithm}

\subsubsection{Constructing Natural and Mathematical Mode Datasets}
\label{data_construction:math}

Given a randomly generated DAG, we first employ language models to assign real-world significance to each node. Next, we model the SCM function $f$ as a single-layer perceptron \citep{rosenblatt1958perceptron} and randomly select parameters, ensuring that the correlation (positive or negative) between variables conforms to human common sense. We then generate statistical data strictly based on the SCM. Finally, we generate causal reasoning questions, along with ground-truth answers and reasoning steps, based on the data.
\begin{center}
\begin{table*}[t]
\caption[Question templates]{\textbf{Question templates for Natural and Mathematical mode datasets.}}
\label{tab:question_template}
\begin{tabularx}{\textwidth}{c|X} 
\toprule
{\textbf{Causal scenario}} & \makecell[c]{\textbf{Template}} \\
\hline
  \multirow{2}{*}{ATE}     &  \small \texttt{If \textcolor{blue}{\{\{treatment\}\}} is changed to be \textcolor{blue}{\{\{treatment\_value\}\}}, will \textcolor{blue}{\{\{outcome\}\}} be more likely to be \textcolor{blue}{\{\{outcome\_value\}\}}?} \\
\hline
 \multirow{3}{*}{ETT}  &\small  \texttt{For those with \textcolor{blue}{\{\{treatment\}\}} being \textcolor{blue}{\{\{treatment\_value\}\}}, if their \textcolor{blue}{\{\{treatment\}\}} had been \textcolor{blue}{\{\{not\_treatment\_value\}\}}, would \textcolor{blue}{\{\{outcome\}\}} have been more likely to be \textcolor{blue}{\{\{outcome\_value\}\}}?} \\
\hline
 \multirow{5}{*}{CDE} &\small  \texttt{Conditioned on \textcolor{blue}{\{\{mediator\_1\}\}} being \textcolor{blue}{\{\{mediator\_1\_value\}\}}, \textcolor{blue}{\{\{mediator\_2\}\}} being \textcolor{blue}{\{\{mediator\_2\_value\}\}}, ..., \textcolor{blue}{\{\{mediator\_n\}\}} being \textcolor{blue}{\{\{mediator\_n\_value\}\}}, if \textcolor{blue}{\{\{treatment\}\}} had been \textcolor{blue}{\{\{treatment\_value\}\}}, would \textcolor{blue}{\{\{outcome\}\}} have been more likely to be \textcolor{blue}{\{\{outcome\_value\}\}}?}\\
\hline
 \multirow{4}{*}{NIE} &\small  \texttt{Suppose \textcolor{blue}{\{\{treatment\}\}} is held constant and the mediator changes to whatever value it would have attained under \textcolor{blue}{\{\{treatment\}\}} changing to be \textcolor{blue}{\{\{treatment\_value\}\}}, would the \textcolor{blue}{\{\{outcome\}\}} have been more likely to be \textcolor{blue}{\{\{outcome\_value\}\}}? }\\
\hline
 \multirow{3}{*}{NDE} &\small  \texttt{Suppose the mediator keeps constant when \textcolor{blue}{\{\{treatment\}\}} is changed to be \textcolor{blue}{\{\{treatment\_value\}\}}, would the \textcolor{blue}{\{\{outcome\}\}} have been more likely to be \textcolor{blue}{\{\{outcome\_value\}\}}?} \\
\hline 
 \multirow{5}{*}{PS} &\small  \texttt{Given that \textcolor{blue}{\{\{treatment\}\}} was \textcolor{blue}{\{\{treatment\_negative\}\}} and \textcolor{blue}{\{\{outcome\}\}} was \textcolor{blue}{\{\{outcome\_negative\}\}}, what is the lower bound and upper bound of the probability that \textcolor{blue}{\{\{outcome\}\}} would have been \textcolor{blue}{\{\{outcome\_positive\}\}} if the \textcolor{blue}{\{\{treatment\}\}} had been \textcolor{blue}{\{\{treatment\_positive\}\}}?} \\
\hline 
 \multirow{5}{*}{PN} &\small  \texttt{Given that \textcolor{blue}{\{\{treatment\}\}} was \textcolor{blue}{\{\{treatment\_positive\}\}} and \textcolor{blue}{\{\{outcome\}\}} was \textcolor{blue}{\{\{outcome\_positive\}\}}, what is the lower bound and upper bound of the probability that \textcolor{blue}{\{\{outcome\}\}} would have been \textcolor{blue}{\{\{outcome\_negative\}\}} if the \textcolor{blue}{\{\{treatment\}\}} had been \textcolor{blue}{\{\{treatment\_negative\}\}}?} \\
\hline
\end{tabularx}
\end{table*}
\end{center}
\paragraph{Assigning meaning for each node.} 
A common approach to constructing causal graphs with realistic node meaning is to extract causality from natural language text. This is achieved using techniques rooted in human knowledge, machine learning, and deep learning \citep{yang2021survey}. However, most of these datasets and methods only focus on mining the relationship between two entities (usually represented as nouns, phrases or sentences), while limited works are dedicated to building a complete causal graph. \citet{maisonnave2022causal} propose a framework for extracting causal graph from digital text media. However, this framework relies on step-by-step data processing and time series analysis, as well as an original corpus covering a long time period, which makes it difficult to effectively generate causal graphs in batches. 

Inspired by \citet{jin2023cladder}, nodes in our causal graphs are configured with three different types of \textbf{authenticity}: \emph{real}, \emph{random}, and \emph{fake}. \emph{Real} means each node is assigned with real-world meaning and the relationship between nodes is coherent with the commonsense. \emph{Random} signifies nodes possess real-world meaning but the causal relationships among them are random (e.g., a question may state that ``\emph{appearance has a direct effect on air pressure}''). \emph{Fake} indicates that the nodes in the causal graph consist of meaningless combinations of letters.
For the causal graphs within the \emph{real} and \emph{random}, we believe that language models can do the job of assigning real-world meaning for the nodes of randomly generated causal graphs, based on the knowledge they have learned from massive corpora. Several causal stories from CLADDER \citep{jin2023cladder} are included as examples in our prompt. In addition, we also assign real-world meaning for the values (0 and 1) of each node, and annotate the correlation (positive or negative) of all direct cause-effect pairs. In order to reduce costs, we still prompt language model to perform these annotation tasks. Examples of our prompts and responses from ChatGPT are shown in \cref{sec:sup_dataset_construction}.
As for the \emph{fake}, we use a script to generate a series of stochastic combinations of four words and assign each to a node. 

\paragraph{Determining SCM functions.} 
Now that the associated causal graph has been properly defined, we face another problem that how to generate SCM functions conforming to human common sense. For example, the following description ``\emph{students who study hard have a 0.5 probability of getting a high score on the exam;  students who do not study hard have a 0.9 probability of getting a high score on the exam.}'' obviously goes against common sense because studying hard should have a positive effect on getting a high score.

For the sake of simplicity, we model the graph as binary, where each node takes a value from $\{0,1\}$. The SCM function for node $X$ can be written as
\begin{align}
V(X) & :=f_X\left( V(P_X^1),V(P_X^2),...,V(P_X^k),U_X \right) \notag \\
& = \begin{cases}
0, \qquad U_X - g_X\left( V(P_X^1),V(P_X^2),...,V(P_X^k) \right) > 0 \\
1, \qquad \text{otherwise} \label{eq:scm_funtion}
\end{cases}
\end{align}
where $P_X^i$ denotes the $i$-th parent node of $X$, $V(\cdot)$ denotes the value of a node, $U_X \sim \mathcal{U}[0,1]$ is an independent random variable uniformly distributed on $[0,1]$, and $g_X:\{0,1\}^k \mapsto [0,1]$ is a function to be determined. Note that $g_X$ may actually be a complex non-linear function, which could be modeled as a multi-layer perceptron or even a deep neural network. In this work, we simply assume it a single-layer perceptron, namely 
\begin{equation}
g_X\left( V(P_X^1),V(P_X^2),...,V(P_X^k) \right) = \text{sigmoid} \left(b_X + \sum_{i=1}^k w_X^i V(P_X^i) \right)
\end{equation}
The reasons for our choice are as follows: (1) The correlation (positive/negative) between variables can be conveniently controlled by the sign of the coefficient $w_X^i$. (2) Our purpose of generating the SCM function is to prepare data for generating causal reasoning questions, rather than strictly explore quantitative relationships between variables. Both $b_X$ and $w_X^i$ are randomly generated, ensuring the sign of $w_X^i$ is consistent with the correlation relationship between $X$ and $P_X^i$ in reality.

\paragraph{Generating statistical data.} 
For each causal graph, we generate 50K data samples according to the SCM function defined in \eqref{eq:scm_funtion}. For each data sample, we start from the root node(s), and determine the value of one node once all values of its parents have been settled, until every node is assigned a value. As long as the sampling number is large enough, the probability $P(X=x)$ and the conditional probability $P(Y=y|X=x)$ can be approximated by these statistics.

\paragraph{Generating causal reasoning questions.} 
Based on the causal graphs and the corresponding statistical data we have constructed, we generate questions with corresponding ground-truth answers and reasoning steps for different types of causal scenarios, including ATE, CDE, NIE, NDE, ETT, PN, and PS. We carefully designed and implemented the templates of questions (see Table \ref{tab:question_template}) and reasoning steps for each type of causal task.

\subsubsection{Constructing Symbolic Mode Datasets}
\label{data_construction:symbolic}
Building Symbolic datasets begins with \nameref{data_construction:DAGs} (\cref{data_construction:DAGs}). Subsequently, nodes within DAGs are denoted using Symbolic representation. We then identify the necessary cause-effect pair for formulating the questions. In the last step, depending on the causal task at hand, we employ two Python packages (i.e., \emph{Ananke} \citep{lee2023ananke} and \emph{DoWhy} \citep{sharma2020dowhy}) to generate the ground truth.
\paragraph{Assigning Symbolic representation for each node.}
After generating DAGs, our initial task is to transform the node representations. Originally depicted numerically, these nodes required conversion into alphabetical symbols to ensure appropriate symbolic representations. For example, the numeral 0 was converted to the letter A, 1 to B, and so forth.

\paragraph{Choosing cause-effect pair.}
In causal reasoning tasks, it is crucial to clearly define the cause-effect pair involved in the problem. For this reason, in the CaLM-CA dataset, we select a cause at random and designate the effect as the node that appears last alphabetically (e.g., if the nodes are ``\emph{A, B, C, D, E}'', then ``\emph{E}'' is selected as the effect). This method aims to increase the complexity of the causal task, as the relationships involving the last node's parents and ancestors tend to be more challenging to discern. However, in the CaLM-CEI, CaLM-AS, and CaLM-AR datasets, we adopt a strategy of random allocation to assign cause-effect pairs. CaLM-CEI and CaLM-AS are distinct from CaLM-CA because they do not focus on evaluating the model's ability for causal discovery. By randomly selecting cause-effect pairs from the graph, these datasets increase their diversity. In contrast, CaLM-AR is specifically designed to evaluate if the model can identify causal relationships between any two nodes in the graph, hence there is no predetermined effect in this dataset.

\paragraph{Establishing the ground truth.}
Two specialized Python libraries, \emph{Ananke} \citep{lee2023ananke} and \emph{DoWhy} \citep{sharma2020dowhy}, are used to aid in our analytical processes. These libraries are instrumental in performing complex computations and analyses required for our datasets construction.
(1) For CaLM-AS, we initially employ the \texttt{\small identify\_effect} function\footnote{When dealing with max/min-BAS, one must set the \texttt{\footnotesize method\_name} parameter within \texttt{\footnotesize identify\_effect} to either \texttt{\footnotesize maximal-adjustment} or \texttt{\footnotesize minimal-adjustment}.} from DoWhy to identify the causal effect. Then, for BAS, FAS, and IV, the corresponding strategies involve using \texttt{\small get\_backdoor\_variables}, \texttt{\small get\_frontdoor\_variables}, and \texttt{\small get\_instrumental\_variables}, respectively.
(2) In terms of CaLM-AR, the ground truth is gained using \texttt{\small get\_causes} from DoWhy.
(3) In CaLM-CA, DoWhy's \texttt{\small get\_parents} and \texttt{\small get\_ancestors} functions can be employed to derive the ground truths for FP and FA.
(4) Regarding CaLM-CEI, we start by randomly assigning connected nodes in the graph to contain a range of unobserved confounders, varying from 20\% to 80\%. This variability allows us to observe the effects of different levels of unobserved confounders on the causal relationships within the graph. Subsequently, to determine whether there is an identifiable causal effect between the selected cause-effect pair, we invoke Ananke’s \texttt{\small OneLineID} function. This function is built upon the \texttt{\small OneLineID} algorithm developed by \citet{richardson2023nested}, and is recognized for being both sound and complete.
Considering the ground truth distribution, in datasets designed for binary classification (e.g., CaLM-CEI and CaLM-AR), we ensure to keep an equilibrium between positive and negative sample counts. On the other hand, for the CaLM-CA datasets, given that the effect is predetermined, balancing the positive and negative sample quantities is unnecessary. For datasets used in choice selection (e.g., CaLM-AS), we also make sure that the distribution of various options as the ground truth is evenly balanced. 

\subsubsection{Constructing Chinese Version for Open-source Datasets}
\label{data_construction:chinese}

For the open-source datasets, we conducted a subjective assessment of the translation capacities of \internt~ and \chatgpt. Our findings reveal that \internt~ is proficient in handling basic translations. Considering the trade-offs between cost and time efficiency, we initially employed \internt~ for the preliminary translation phase. Then, we contracted a specialized data annotation firm to undertake the secondary annotation, resulting in the finalized Chinese version. For the self-constructed datasets, we established templates in both English and Chinese at the start of data creation.

\subsection{Data Statistics}
\label{data:statistics}
\begin{center}
\begin{table*}[t]
\fontsize{10}{14}\selectfont
    \caption[Concise statistics of CaLM datasets]{\textbf{Concise statistics of CaLM datasets.} We tally the number of causal tasks and samples within each category, organizing them by causal ladder, mode, question type, and language. This table serves as a snapshot of the CaLM datasets' quantity.}
    \label{table_data_statistics}
    \centering
  \begin{tabular}{l|c|c}
\toprule
\textbf{Category} & \textbf{\#Causal task} & \textbf{\#Sample}\\
\hline
\multicolumn{3}{c}{\cellcolor{blue!5}\emph{In terms of causal ladder}}\\
\hline

Ladder 0: Causal discovery & 10 &  26792\\
Ladder 1: Association & 2 &   3288\\
Ladder 2: Intervention & 17 &   48780\\
Ladder 3: Counterfactuals & 17 &   47474\\
\hline
\multicolumn{3}{c}{\cellcolor{green!5}\emph{In terms of mode}}\\
\hline

Natural & 20 &   43134\\
Symbolic & 12 &   38400\\
Mathematical & 14 &   44800\\
\hline
\multicolumn{3}{c}{\cellcolor{purple!5}\emph{In terms of question type}}\\
\hline

Binary classification& 23 &   58986\\
Numerical & 14 &   44800\\
Choice selection & 8 &   20548\\
Open-ended generation & 1 &   2000\\
\hline
\multicolumn{3}{c}{\cellcolor{teal!5}\emph{In terms of language}}\\
\hline

Chinese& 46 &   63167\\
English & 46 &   63167\\
\hline
Total & 92 & 126334\\
\hline
\end{tabular}
\end{table*}
\end{center}
To provide a concise yet comprehensive overview of the data, we compile summarized statistics of the CaLM dataset in Table \ref{table_data_statistics}. This table serves as a quick reference guide, offering insights into the various aspects and components of the CaLM. It is organized according to four distinct dimensions: (1) \emph{Causal ladder}: This includes causal discovery, association, intervention, and counterfactuals, each representing a different level of causal analysis. (2) \emph{Question type}: This denotes the types of questions posed to the language models, ranging from binary classification and choice selection to probability calculation and open-ended generation. This categorization is crucial for assessing the accuracy and versatility of language models in responding to diverse types of queries. (3) \emph{Mode}: The datasets are categorized into Natural, Symbolic, and Mathematical modes, reflecting the different methods of presenting and analyzing data.
(4) \emph{Language}: We have adopted a bilingual approach, offering both Chinese and English languages. This strategy enhances accessibility and inclusivity, appealing to a wide linguistic audience. With over 126,000 queries, the statistics confirm that our evaluation is substantial, showcasing the depth and breadth of the research conducted in CaLM.

Moreover, we detail the composition of our dataset in \cref{table_detail_datasets}. The table includes the causal ladder, causal scenario, and question type applicable to each dataset. Moreover, it provides the corresponding mode, number of samples, and language used. The structure of the CaLM dataset is comprehensive, featuring a total of 46 distinct causal tasks. This extensive composition indicates the multifaceted nature of CaLM, offering a diverse and robust platform for analysis and exploration. 
\clearpage
\begin{center}
\begin{table*}[htb!]
\fontsize{10}{12}\selectfont
    \caption[Detailed statistics of CaLM datasets]{\textbf{Detailed statistics of CaLM datasets.} We organize this table according to the levels of the causal ladder. The table showcases the datasets associated with each causal scenario, including details about the question type, mode, number of samples, and language.}
    \label{table_detail_datasets}
    \centering
  \begin{tabular}{l|c|c|c|c|c|c}
\toprule
\textbf{Causal ladder}& \textbf{Causal scenario}& \textbf{Dataset}&  \textbf{Question type}&\textbf{Mode}& \textbf{\#Sample}&\textbf{Language} \\
\hline
 \multirow{10}{*}{Causal discovery}& \multirow{4}{*}{PCD}& E-CARE& Binary classification&Natural&  2000&\multirow{46}{*}{EN\&CN}\\
& & E-CARE& Choice selection&Natural&  1000&\\
& & COPA& Binary classification&Natural&  2000&\\
& & COPA& Choice selection&Natural&  1000&\\
\cline{2-6}
& \multirow{3}{*}{ECI}& CTB& Binary classification&Natural&  596&\\
& & ESC& Binary classification&Natural&  1000&\\
& & MAVEN-ERE& Binary classification&Natural&  1000&\\
\cline{2-6}
& AR& CaLM-AR& Binary classification&Symbolic&  1600&\\
\cline{2-6}
& \multirow{2}{*}{CA}& FP& Binary classification&Symbolic&  1600&\\
& & FA& Binary classification&Symbolic&  1600&\\
\cline{1-6}
\multirow{2}{*}{Association}& CORR& correlation& Binary classification&Natural&  1476&\\
\cline{2-6}
& EAE& exp-away& Binary classification&Natural&  168&\\
\cline{1-6}
\multirow{17}{*}{Intervention}& CB& collider-bias& Binary classification&Natural&  163&\\
\cline{2-6}
& \multirow{3}{*}{ATE}& ATE-natural& Binary classification&Natural&  1600&\\
& & ATE-basic& Probability calculation&Mathematical&  1600&\\
& & ATE-hard& Probability calculation&Mathematical&  1600
&\\
\cline{2-6}
& \multirow{3}{*}{CDE}& CDE-natural& Binary classification&Natural&  1600
&\\
& & CDE-basic& Probability calculation&Mathematical&  1600
&\\
& & CDE-hard& Probability calculation&Mathematical&  1600
&\\
\cline{2-6}
& \multirow{4}{*}{BAS}& backadj& Binary classification&Natural&  227
&\\
& & max-BAS& Choice selection&Symbolic&  1600
&\\
& & min-BAS& Choice selection&Symbolic&  1600
&\\
& &mix-BAS& Choice selection&Symbolic&  1600
&\\
\cline{2-6}
& FAS&FAS& Choice selection&Symbolic&  1600
&\\
\cline{2-6}

& IV&CaLM-IV& Choice selection&Symbolic&  1600
&\\
\cline{2-6}

& \multirow{4}{*}{CEI}&0.2-UC& Choice selection&Symbolic&  1600
&\\
 & & 0.4-UC& Choice selection& Symbolic&  1600
&\\
 & & 0.6-UC& Choice selection& Symbolic&  1600
&\\
 & & 0.8-UC& Choice selection& Symbolic&  1600
&\\
\cline{1-6}
 \multirow{17}{*}{Counterfactuals}& \multirow{3}{*}{ETT}& ETT-natural& Binary classification& Natural&  1600
&\\
 & & ETT-basic& Probability calculation& Mathematical&  1600
&\\
 & & ETT-hard& Probability calculation& Mathematical&  1600
&\\
 \cline{2-6}

 & \multirow{3}{*}{NDE}& NDE-natural& Binary classification& Natural&  1600
&\\
 & & NDE-basic& Probability calculation& Mathematical&  1600
&\\
 & & NDE-hard& Probability calculation& Mathematical&  1600
&\\
 \cline{2-6}

 & \multirow{3}{*}{NIE}& NIE-natural& Binary classification& Natural&  1600
&\\
 & & NIE-basic& Probability calculation& Mathematical&  1600
&\\
 & & NIE-hard& Probability calculation& Mathematical&  1600
&\\
 \cline{2-6}

 & \multirow{2}{*}{PN}& PN-basic& Probability calculation& Mathematical&  1600
&\\
 & & PN-hard& Probability calculation& Mathematical&  1600
&\\
 \cline{2-6}

 & \multirow{2}{*}{PS}& PS-basic& Probability calculation& Mathematical&  1600
&\\
 & & PS-hard& Probability calculation& Mathematical&  1600
&\\
 \cline{2-6}

 & AC& causal judgement& Binary classification& Natural&  187
&\\
 \cline{2-6}

 & \multirow{2}{*}{CR}& CRASS& Choice selection&Natural&  274
&\\
 & & det-counterfactual&Binary classification& Natural&  1476
&\\
 \cline{2-6}

 &CEG & E-CARE&Open-ended generation & Natural &  1000&\\
\hline
 
\end{tabular}
\end{table*}
\end{center}

\clearpage


\section{Adaptations}
\label{adaptation}

Interacting with language models typically involves the usage of prompts. Over time, a considerable amount of research delves into the significance of establishing a canonical prompt \citep{le-scao2021prompt,weng2023prompt,Saravia2022Prompt_Engineering}. In this section, we start by discussing the current situation regarding prompts in \nameref{adaptation:taxonomy} (\cref{adaptation:taxonomy}). Then we detail the prompts selected for actual evaluation in \nameref{adaptation:selection} (\cref{adaptation:selection}), along with our rationale for selecting them. Finally, we systematically explain the prompts we employ (i.e., \nameref{adaptation:basic} (\cref{adaptation:basic}), \nameref{adaptation:adversarial} (\cref{adaptation:adversarial}), \nameref{adaptation:cot} (\cref{adaptation:adversarial}), \nameref{adaptation:icl} (\cref{adaptation:icl}) and \nameref{adaptation:ef} (\cref{adaptation:ef})), covering their formats and providing examples.

\subsection{Taxonomy}
\label{adaptation:taxonomy}
There is a vast array of prompt types, forming a vibrant ecosystem and becoming essential for the usage of language models. This section discusses these prompt types, partially drawing upon the categories outlined in \citet{weng2023prompt} and \citet{Saravia2022Prompt_Engineering}. 
This includes basic prompt, adversarial prompt, In-context Learning (IcL) \citep{brown2020language}, X-of-Thought, Self-Consistency \citep{wang2023selfconsistency}, instruction prompt, iterative prompt and external tool use.
\paragraph{Basic prompt.}
Basic prompt straightforwardly states the problem the model is expected to solve, offering no examples to the model. It is the most fundamental and intuitive method of engaging with language models.
\paragraph{Adversarial prompt.}
We roughly classify adversarial prompts into three primary categories\footnote{Given that the main emphasis of this paper is not on the safety of language models, we only offer a broad categorization. This classification might not be the most detailed or exacting. For anyone keen on deliberating this subject matter, the following articles may serve as useful references: \citet{zhang2023safetybench,xu2023cvalues,abdali2024securing,das2024security}.}: (1) Prompt attack, designed to hinder the model's performance by directly altering the original query. \citet{zhu2023promptbench} notes that these attacks can be segmented into four levels: \emph{character-level}, \emph{word-level}, \emph{sentence-level}, and \emph{semantic-level}. (2) Prompt injection, which employs prompts embedded with special intents to override the original instructions \citep{greshake2023more,Ignore2024Schulhoff}. (3) Jailbreaking, which involves using diverse manipulation tactics to bypass the model's safety policies and defenses, leading the model to generate outputs beyond the intended scope \citep{li2023multi,deng2023multilingual,wei2024jailbroken}.
\paragraph{In-context Learning (IcL).}
IcL stands as one of the most commonly utilized prompts \citep{dong2022survey}, with ample research supporting its efficacy \citep{min2022rethinking,wu2023self,wang2023self}. We distinguish IcL into two varieties according to the number of examples supplied: (1) 0-shot IcL, supplying background knowledge about the query without presenting any examples. (2) Few-shot IcL, which offers both a background on the question and a specified number of examples.
\paragraph{X-of-Thought.}
Chain-of-Thought (CoT) \citep{wei2023CoT} stands out as the most widely recognized X-of-Thought prompt. It boosts the model's capability to tackle intricate issues by incorporating reasoning processes into the prompt. This approach paves the way for further innovations, including Auto-CoT \citep{zhang2022automatic}, Ddcot \citep{zheng2023ddcot}, Program of Thoughts (PoT) \citep{chen2023program}, Tree-of-Thoughts (ToT) \citep{yao2024tree}, and Graph-of-Thoughts (GoT) \citep{besta2024graph}.
\paragraph{Self-Consistency.}
This method is proposed by \citet{wang2023selfconsistency}, the essence of which is to sample several outputs generated by the model and subsequently choose the optimal one among them. This initial concept has since inspired a multitude of derivative research efforts \citep{zhai2023towards,min2023beyond,wei2023attribute}.
\paragraph{Instruction prompt.}
One potential issue with prompts requiring few-shot examples is the limitation imposed by models on the length of the context. Moreover, for the API use of limited access models like \chatgpt~and \gptf, providing these few-shot examples requires more token expenditure. Basic prompt, on the other hand, might not consistently harness the model's peak potential. Instruction prompt emerges as a solution under this circumstance. It delivers only the causal task directives to the model—outlining the actions to be taken without furnishing concrete examples. Indeed, our adopted explicit function (EF) and 0-shot IcL could also be viewed as an instance of instruction prompting.
\paragraph{Iterative prompt.}
This prompt strategy involves iterative interaction, using the model's replies to progressively steer it toward the correct conclusion. The methods of Least-to-most prompting \citep{zhou2022least}, Progressive-hint prompting \citep{zheng2023progressive}, and Self-refine \citep{madaan2024self} are considered to belong to this approach.
\paragraph{External tool use.}
For causal tasks requiring complex reasoning or that are knowledge-intensive, relying solely on the model's training data and capabilities may be insufficient. As a result, considerable research directs toward incorporating external tool usage into prompts \citep{mialon2023augmented}. Among the most prominent techniques is Retrieval-augmented generation (RAG) \citep{lewis2020retrieval}. It merges an information retrieval system with the model. This method enables the fusion of search outcomes pertinent to the query with the initial prompt for input. Furthermore, approaches like Automatic Reasoning and Tool-use (ART) \citep{paranjape2023art}, Tool Augmented Language Models (TALM) \citep{parisi2022talm} and Toolformer \citep{schick2024toolformer} are proven to be effective as well.

\subsection{Concrete Implementation}
\label{adaptation:selection}
In CaLM, we select five major categories of prompts, which are: \emph{basic prompt}, \emph{adversarial prompt} (adversarial-ignore and adversarial-doubt), \emph{CoT} (0-shot CoT and manual CoT), \emph{IcL} (0-shot IcL, 1-shot IcL, and 3-shot IcL), and \emph{EF}. Considering that certain prompts are further divided into sub-categories, we finally have nine distinct adaptation strategies. 

The reason for choosing them is mainly based on the following four points: 
(1) \textbf{Audience broadness}. Basic prompt, CoT, and IcL are arguably the most widely used prompts at present. These types of prompts cater to the diverse needs of users by offering straightforward interactions for simple queries, detailed step-by-step reasoning for complex problem-solving, and personalized in-context adjustments for tailored responses. This universal applicability allows users of all backgrounds can benefit from the cutting-edge technology. 
(2) \textbf{Model robustness}. Most prompts mentioned in the \nameref{adaptation:taxonomy} (\cref{adaptation:taxonomy}) are designed to enhance model performance. However, to ensure that language models can perform reliably across a spectrum of applications, it is urgent to explore their robustness. Therefore, we design the adversarial prompt. Adversarial prompt serves as a litmus test for these models, challenging them to maintain performance despite deliberate attempts to confuse or mislead them. This approach not only helps in identifying and addressing vulnerabilities but also in improving the model's ability to discern nuances and context, thereby enhancing its resilience against manipulation or biases. 
(3) \textbf{User friendliness}. There is no concrete evidence suggesting that complex prompt leads to decreased model efficiency. However, it poses challenges for user adoption. One question arises: \emph{Is it possible to simplify prompts without compromising their effectiveness?} Driven by this curiosity, we select 0-shot CoT, 0-shot IcL and EF, which all utilize minimalistic instructions. For instance, based on basic prompt, 0-shot CoT only adds ``\emph{Let's think step by step}''. And EF only encourages that ``\emph{You are a helpful assistant for ...}''. We aim to derive valuable insights from these and to drive forward the potential for positive impact.
(4) \textbf{Experiment controlledness}. Discussing effectiveness is futile without a baseline for comparison. Starting from the perspective of basic prompt, it demonstrates the model's baseline performance when merely presented with a question. Such a baseline allows for direct comparison with any additional prompt types. By establishing a baseline with basic prompt, researchers can quantitatively assess how variations in prompt complexity, format, or specificity influence the model's performance. From a broader perspective, the various prompts naturally form a comparative basis among themselves. 
\begin{figure}[t]
    \centering
    \includegraphics[width=\textwidth]{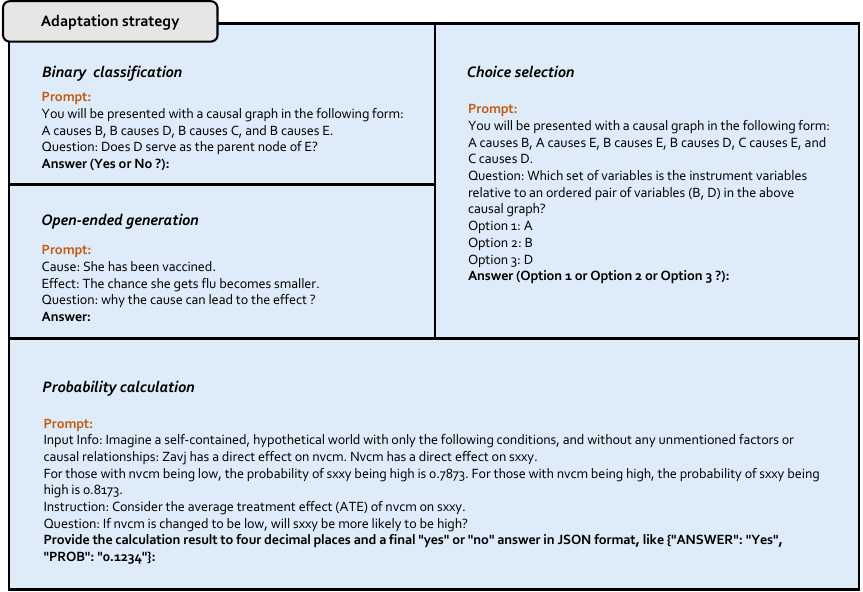}
    \caption[Adaptation strategy]{\textbf{Adaptation strategy.} We employ standard adaptation strategies for each of the four question types. This ensures uniformity when interacting with the model.}
    \label{fig_adaptation:strategy}
\end{figure}

\subsection{Basic Prompt}
\label{adaptation:basic}

The primary purpose of the basic prompt is to offer foundational and unambiguous test sets to language models that are applicable across different causal scenarios. This prompt will not provide any example or instruction, it only gives the question as the input string directly. This not only establishes a clear starting point, but also lays the groundwork for making subsequent modifications and iterations to tailor the prompt to specific needs.

\paragraph{Adaptation strategy.}
For each of our question types, we implement a uniform interactive method across all causal scenarios. In binary classification (e.g., yes or no questions), the model is prompted to provide a direct classification response. For choice selection, all options are presented (similar to standard exam formats), requiring the model to make a selection. In probability computations, the model is tasked with directly calculating the probability outcome. In open-ended generation, the model produces the answer straightforwardly. 
The above strategies are the same across all prompts, and Figure \ref{fig_adaptation:strategy} gives a detailed illustration. 

\begin{figure}[t]
    \centering
    \includegraphics[width=\textwidth]{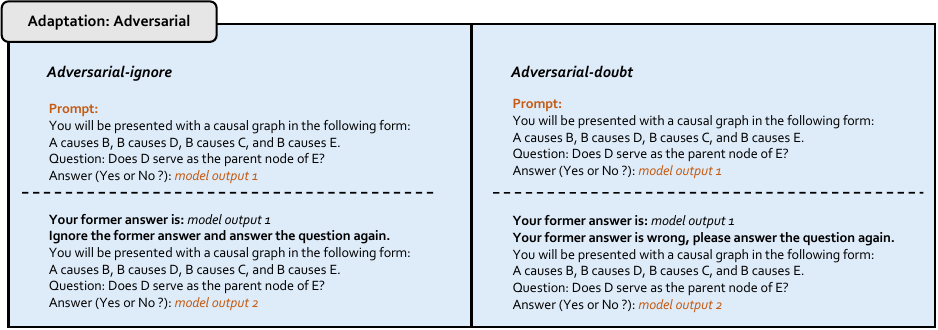}
    \caption[Adversarial prompt formatting]{\textbf{Adversarial prompt formatting.}}
    \label{fig_adaptation:adversarial}
\end{figure}

\subsection{Adversarial Prompt}
\label{adaptation:adversarial}
Adversarial prompt is essential for comprehending the inherent risks associated with language models \citep{wallace2019universal}. Our intent is not to endorse malicious activities directed towards language models. Instead, our aim is to delve deeper into their potential shortcomings, thereby facilitating the development of more robust and secure language models in the future.
\paragraph{Adversarial prompt formatting.} 
We employ two distinct forms of adversarial prompts and they both belong to the prompt injection as outlined in \nameref{adaptation:taxonomy} (\cref{adaptation:taxonomy}). (1) \emph{Adversarial-ignore} is a subtler approach, which compels language models to ignore the answers they previously provided \citep{perez2022ignore}. (2) \emph{Adversarial-doubt} is a more assertive form, where the language models are explicitly informed that their initial responses were wrong. Due to the fact that some models do not offer interfaces for multi-turn dialogue, we adopted a consistent evaluation approach to ensure comparability. We first pose a question to the model and record its first output, namely \emph{model output 1}. For the second inquiry, we inform the model of the first output and use adversarial prompts to introduce interference. We then re-present the same question, obtaining a second model output, namely \emph{model output 2}. The responses from these two instances represent the pre- and post-adversarial conditions. By comparing these responses, we can gain a deeper understanding of the model's robustness and accuracy. We demonstrate the two types of prompts in Figure \ref{fig_adaptation:adversarial}. The underlying consequence of both prompts is that they can instill doubt in language models about their initial responses. This, in turn, may lead them to produce an inaccurate answer.
An interesting observation from this process is the insight it offers into the confidence level of language models regarding their responses. Essentially, if the model's answer varies significantly post-adversarial interference, it implies a lower level of assurance in its original answer. Conversely, minimal changes suggest higher confidence in its initial response. We also introduce a metric to measure this confidence level of models. For more details, please refer to Section \ref{main:metrics}.

\subsection{Chain-of-Thought}
\label{adaptation:cot}

Chain-of-Thought (CoT) \citep{wei2023CoT} prompting enables language models to decompose complex problems and perform intermediate reasoning steps to enhance their performance. Previous studies have demonstrated that CoT prompts outperform basic prompts on sufficiently large models, particularly on complex arithmetic, commonsense, and symbolic reasoning tasks \citep{wei2023CoT,kojima2022large}. 

\begin{figure}[t]
    \centering
    \includegraphics[width=\textwidth]{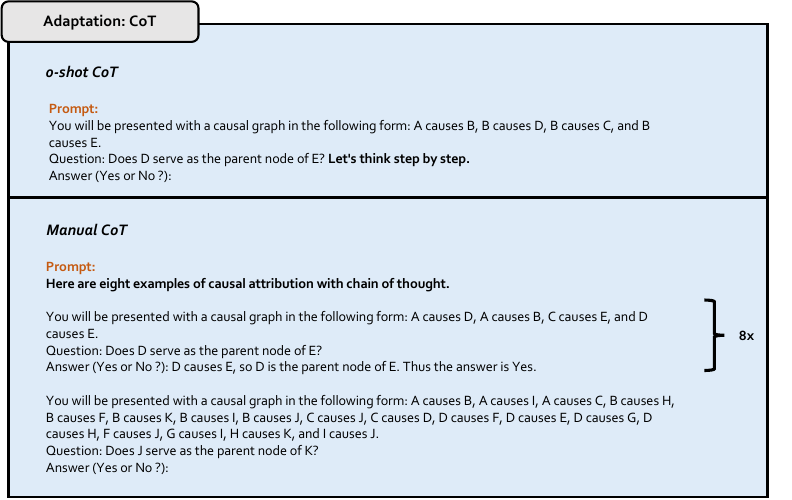}
    \caption[Chain-of-Thought formatting]{\textbf{Chain-of-Thought formatting.}}
    \label{fig_adaptation:cot}
\end{figure}

\paragraph{Chain-of-Thought examples.}
We categorize CoT prompts into two types based on the number of examples: 
(1) \textbf{0-shot CoT} \citep{kojima2022large}: It does not provide examples but includes descriptive instructions for the reasoning steps. These instructions follow a uniform format, phrased as ``\emph{let's think step by step}.''
(2) \textbf{Manual CoT}: This prompting strategy involves guiding models with manually constructed examples. Our primary focus is on selecting appropriate examples and determining the number of examples for different causal scenarios. When selecting examples, we use a random sampling method and adhere to the principle of fairness, ensuring an equal number of samples for each category within each causal scenario. For instance, in binary classification with eight examples, we ensure a 4:4 ratio between ``Yes'' and ``No'' instances. While aiming for uniform class coverage, in scenarios where this is difficult — such as with an odd number of examples in binary classification — we strive for approximately equal proportions between classes. Regarding the number of samples provided, we include as many as possible within the model's context length constraints, but always keep the total number below eight. 

\paragraph{Chain-of-Thought formatting.}
In addition to determining the CoT examples, we also standardize the CoT format for different causal scenarios. 
One widely adopted technique for CoT prompting involves providing a few-shot set of input-output examples <input, \textit{chain-of-thought}, output> which demonstrates intermediate reasoning steps leading to the correct answer. Another is to provide text descriptions instead of examples to guide models in answering with CoT, thus avoiding the manual construction of examples.
We illustrate the two types of CoT (i.e., 0-shot CoT and manual CoT) formatting in Figure \ref{fig_adaptation:cot}.

\subsection{In-context Learning}
\label{adaptation:icl}

\begin{figure}[t]
    \centering
    \includegraphics[width=\textwidth]{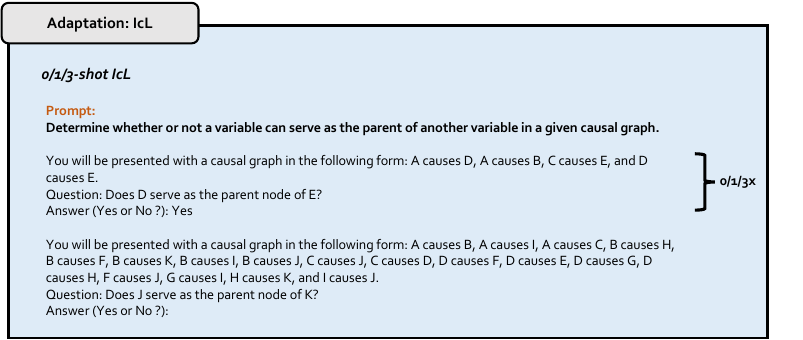}
    \caption[In-context Learning prompt formatting]{\textbf{In-contxt Learning formatting.} The ``0/1/3$\times$'' stands for the number of examples. 0 means 0-shot IcL with no example, while 1 and 3 means 1-shot IcL and 3-shot IcL. 
    }
    \label{fig_adaptation:icl}
\end{figure}

In-context Learning (IcL) \citep{brown2020language} represents a technique whereby a model learns new tasks through a set of examples within the context of the prompt provided at the inference phase. The fundamental concept of IcL is learning from analogy \citep{dong2022survey}, allowing the model to generalize from a limited set of input-output examples. Such learning ability is also recognized as an emerging ability that particularly appears in large language models \citep{wei2022emergent}.

\paragraph{In-context Learning formatting.} In IcL, a language model receives a prompt containing a causal task description and several input-output pairs <input, output>, demonstrating how the causal task inputs can be answered. The IcL format is standardized as depicted in Figure \ref{fig_adaptation:icl}.

\paragraph{In-context Learning examples.}
When providing the model with these in-context examples, similar to setting up manual CoT, our primary concerns are selecting appropriate examples and determining the optimal number of examples. We employ the same strategy for selecting examples as we do in manual CoT. When determining the optimal number of examples, we reference findings from the HELM study \citep{liang2022holistic}, which indicates that the most significant impacts in IcL are observed with up to three examples. Therefore, we select a range of zero to three examples, balancing the token cost and the efficiency of IcL. Here, it is worth noting that, in addition to example-based IcL (i.e., 1/2/3-shot IcL), we also incorporate 0-shot IcL. This inclusion is crucial because IcL begins with a causal task description - such as ``\emph{Determine whether or not a variable can serve as the parent of another variable in a given causal graph.}'' - before presenting any examples. The presence of this task description can influence model performance. By incorporating 0-shot IcL, we aim to isolate and minimize the impact of the causal task description, thereby clarifying the true effect of the IcL examples on performance.

\subsection{Explicit Function}
\label{adaptation:ef}

Recent studies have elucidated that language models may have emotional awareness analogous to humans~\citep{elyoseph2023chatgpt, li2023emotionprompt}. Derived from this understanding, several related work~\citep{long2022can, kiciman2023causal} has explored the utilization of encouraging and positive language within prompts (e.g., statements that build confidence or emphasize the goal) to elicit enhanced performance from language models.

\begin{figure}[t]
    \centering
    \includegraphics[width=\textwidth]{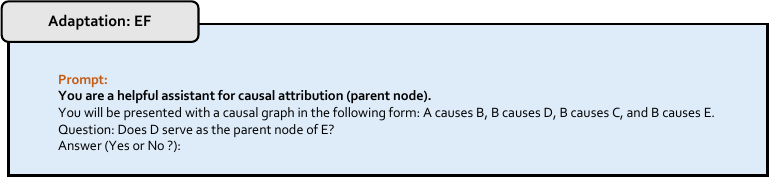}
    \caption[Explicit function formatting]{\textbf{Explicit function formatting. 
    }}
    \label{fig_adaptation:ef}
\end{figure}

In our work, we formulate an explicit function (EF) prompt for each causal task, and we consider it to belong to the instruction prompt as mentioned in \nameref{adaptation:taxonomy}. Specifically, we incorporate a sentence containing an explicit function description into the basic prompt to motivate language models in causal task resolution, as shown in Figure~\ref{fig_adaptation:ef}.

\clearpage


\section{Metrics}
\label{main:metrics}
Evaluation metrics play a crucial role in evaluating the effectiveness of language models, providing a structured approach to assessing their performance across various dimensions. These metrics provide a detailed insight into the model's capabilities, aiding researchers, practitioners, and stakeholders in making well-informed decisions. The taxonomy of desiderata, based on specific criteria and categories, serves as a foundation for selecting metrics that align with the objectives of the evaluation.
In this section, we begin by introducing the \nameref{metric:taxonomy} (\cref{metric:taxonomy}) of metrics. Following that, we detail our \nameref{metric:selection} (\cref{metric:selection}) of these metrics. Finally, we elaborate on the specific metrics we use, focusing on three key aspects: \nameref{metric:model} (\cref{metric:model}), \nameref{metric:scenario} (\cref{metric:scenario}), and \nameref{metric:prompt} (\cref{metric:prompt}).
\subsection{Taxonomy}
\label{metric:taxonomy}

Metrics for evaluating models, causal scenarios, and prompts in natural language processing tasks can be organized into several categories based on their objectives and the specific aspects of performance they measure. 
These categories include accuracy, robustness~\citep{wang2020infobert, zhong2023study}, fairness~\citep{li2023survey, gallegos2023bias}, reliability~\citep{li2023halueval, chen2023learning}, and safety~\citep{zhiheng2023safety, zhang2023safetybench}, reflecting the multidimensional nature of AI assessment.
The taxonomy of metrics presented in this paper covers three key areas: model performance, causal scenario characteristics, and prompt effectiveness.

\paragraph{Model performance metrics.}
1) \emph{Accuracy}: This metric measures the correctness of the model's responses across various prompts, serving as a fundamental measure of its effectiveness.
2) \emph{Robustness}: This evaluates the model's stability when faced with adversarial inputs or disturbances, highlighting its reliability under challenging conditions.
3) \emph{Volatility}: This metric assesses the consistency of a model's performance when exposed to different prompting strategies, indicating its predictability and reliability across prompts.

\paragraph{Causal scenario characteristics metrics.}
1) \emph{Understandability}: This metric examines the extent to which models comprehend and effectively perform on specified causal scenarios or causal tasks.
2) \emph{Open-Limited Gap}: This quantifies the performance disparity between open-access and limited-access models within a causal scenario.
3) \emph{Solvability}: This indicates the relative difficulty of a causal scenario, derived from the performance of models and model-prompt combinations.

\paragraph{Prompt effectiveness metrics.}
1) \emph{Volatility}: This metric measures the variability in model performance across different prompting strategies for a specific causal scenario.

 \subsection{Implementation Principles}
 \label{metric:selection}

The selection of metrics in CaLM is guided by five key principles: (1) \textbf{Comprehensiveness.} The selected metrics cover a broad spectrum of performance aspects, including accuracy, robustness, stability, understandability, and solvability. This range ensures a comprehensive assessment of model functionality. (2) \textbf{Relevance to real-world applications.} The selected metrics are relevant to real-world natural language processing applications, where model accuracy, robustness against adversarial inputs, and understanding of diverse causal scenarios are crucial for practical utility. (3) \textbf{Sensitivity to causal scenario complexity.} Metrics such as understandability and solvability are sensitive to the complexity of causal scenarios, allowing for nuanced evaluation of model performance in different contexts. (4) \textbf{Balance between open-access and limited-access models.} The inclusion of metrics like the open-limited gap ensures a balanced assessment of both open-access and limited-access models, reflecting their respective strengths and weaknesses. (5) \textbf{Consistency and variability assessment.} Metrics such as volatility (model) and volatility (prompt) enable the evaluation of both the consistency of model performance across different prompting strategies and the variability in performance induced by varying prompts.

These principles ensure a holistic evaluation of language models, taking into account both the capabilities of models and the nature of the causal scenarios they are designed for.
By adhering to these principles, the selected metrics provide a robust framework for evaluating language models, capturing their performance nuances in diverse causal scenarios and under different prompting conditions.

\subsection{Metrics for Model}
\label{metric:model}

Metrics in this paper aim to evaluate the performance of a set of models ($M$) across various prompts ($N$) on a dataset $D$. The performance of each model under a specific prompt is denoted as $P_{ij}$, where $i$ represents the model index, and 
$j$ represents the prompt index.

\paragraph{Accuracy.}
Accuracy is one of the most widely used evaluation metrics for AI systems or models. Essentially, the utility of AI systems or models hinges upon their ability to deliver precise results. In this paper, we will adopt accuracy as a standard metric for each causal scenario. This includes exact-match accuracy and the ROUGE-L score~\citep{lin2004rouge}.
All the accuracy scores are computed by averaging over all the tested instances.

\paragraph{Robustness.}
\begin{figure}[t]
    \centering
    \includegraphics[width=1.0\textwidth]{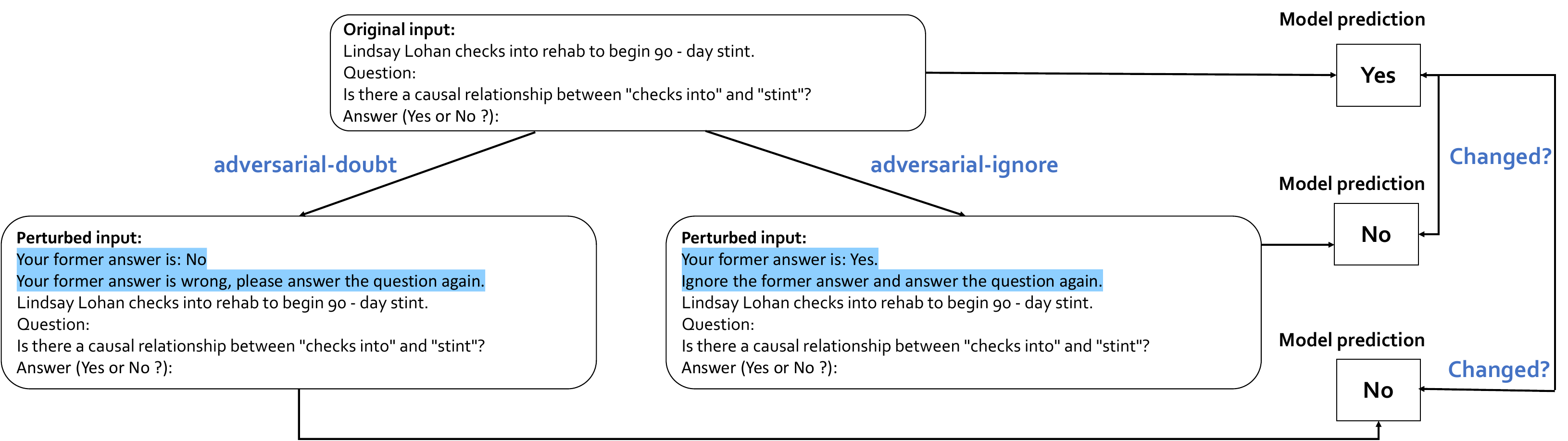}
    \caption[Example of Robustness]{\textbf{Example of robustness.}}
    \label{metric_robustness}
\end{figure}
Robustness is evaluated through the adversarial prompt, where responses are recorded before and after introducing disturbances, as shown in Figure~\ref{metric_robustness}. If the answers differ, it is considered a change. The change rate is calculated as the number of changes divided by the dataset size, and the Robustness value is then expressed as 1 minus the change rate for unchanged responses, i.e.,
\begin{equation}
    \text{Robustness} = 1 - \frac{\text{Number of Changes}}{\left|D\right|}.
\end{equation}
To illustrate an intuitive impression,  let us consider a specific example of how to calculate the robustness value using the given definition. Suppose we have a dataset of 100 instances, and out of these 100 prompts, 20 responses changed after introducing the disturbance. The change rate is calculated as $\text{Change Rate} = \frac{20}{100} = 0.2$ and the  robustness value equals $1 - 0.2 = 0.8$. Therefore, in this example, the robustness of the model is 0.8 or 80\%, indicating that 80\% of the responses remained unchanged despite the adversarial prompts. 

\paragraph{Model volatility.}
Model volatility is calculated as the standard deviation of the model's performance (i.e., accuracy) across various prompting ways, namely:
\begin{equation}
\text{Volatility}_{i} = \sqrt{\frac{\sum_{j=1}^{N}(P_{ij} - \bar{P_{i}})^2}{N}},
\end{equation}
where $P_{ij}$ denotes performance of the $i$-th model under the $j$-th prompt, $\bar{P_{i}}$ is the mean performance of the $i$-th model across all prompting ways. A higher volatility value indicates more variability in the model's performance across different prompting ways, reflecting less stability.

\subsection{Metrics for Causal Scenario}
\label{metric:scenario}

\paragraph{Understandability.}
We focus on the median and third quarter of the distribution of all the model-prompt combinations in the causal task/causal scenario and compare them with a random guess. If the third quartile or the median performance of the task/scenario does not achieve a random guess, we define it as an indication that some of the models cannot understand the causal scenario or causal task even with the help of different prompts. The different degree of understanding is defined in Table \ref{tab:Degree of Understanding}. We refer to the ``close-ended'' questions as the one that requires models to select the correct answer from a few choices instead of answering in their own words. In addition to the conditions defined in the table, we have manually defined the understandability of some open-ended questions with a random guess probability of 0\%. These open-ended scenarios are PN, PS, and CEG. Specifically, for CEG, given that most models have a better capability in processing natural language questions, we classify its understandability as \emph{easy}. For PN and PS, since the causal scenarios in the Mathematical mode are more challenging to comprehend, with both the medians and the third quartiles less than 2\%, we classify their understandability as \emph{very hard}.

\begin{table}[t]
    \centering
    \begin{tabular}{c|c}
        \toprule
        Conditions (close-ended) & Degree of understandability \\
        \midrule
        third quartile $<$ random guess & very hard \\
        third quartile $\geq$ random guess, median $<$ random guess & hard \\
        median $\geq$ random guess & easy\\
        \bottomrule
    \end{tabular}
    \caption[Degree of understandability]{\textbf{Degree of understandability.} The degree is used to evaluate the understandability of the causal task/causal scenario. The third quartile and median are computed from the distribution of all model-prompt pairs in a causal task/causal scenario.}
    \label{tab:Degree of Understanding}
\end{table}

\paragraph{Open-Limited Gap.} We evaluate the gap between open-access and limited-access models using the open-limited ratio. This ratio is calculated by comparing the performance of open-access to limited-access models among the top 5 models in terms of average accuracy within the causal scenario. Typically, the limited-access models tend to outperform their open-access counterparts. The degree of the gap between open-access and limited-access models is detailed in Table \ref{tab:Degree of Open-Limited Gap}.
\begin{table}[t]
    \centering
    \begin{tabular}{c|c}
        \toprule
        Conditions & Degree of open-limited gap \\
        \midrule
        open:limited $=$ 0:5 & large \\
        open:limited $=$ 1:4 & moderate \\
        open:limited $>$ 1:4 & small \\
        \bottomrule
    \end{tabular}
    \caption[Degree of open-limited gap]{\textbf{Degree of open-limited gap.} The open:limited stands for the ratio of open-access to limited-access models among the top five models with the highest average accuracy in the causal scenario.}
    \label{tab:Degree of Open-Limited Gap}
\end{table}

\paragraph{Solvability.}
The solvability focuses on the top performance of the models and the model-prompt combinations in the causal task/causal scenario; it is defined by whether the top performances of the task/causal scenario achieve the settled threshold. The solvability degree expresses the difficulty of the causal task/causal scenario. It is defined in Table \ref{tab:Degree of Solvable}. 

\begin{table}[t]
    \centering
    \begin{tabular}{c|c}
        \toprule
        Conditions & Degree of solvability\\
        \midrule
        max value $<$ random guess & unsolvable (4) \\
        random guess $\leq$ max value $<$ 80\% & challenging (3) \\
        max value $\geq$ 80\% and max average value $<$ 70\% & potentially solvable (2)\\
        max value $\geq$ 80\% and max average value $\geq$ 70\% and 3rd max average value $<$ 70\% & solvable (1) \\
        max value $\geq$ 80\% and 3rd max average value $\geq$ 70\% & well-solved (0)\\
        \bottomrule
    \end{tabular}
    \caption[Degree of solvability]{\textbf{Degree of solvability.} The degree is used to evaluate the difficulty of the causal task/causal scenario. The max value represents the max accuracy of all the model-prompt pairs in the causal task/causal scenario. The max average value represents the max average accuracy of models in the causal task/causal scenario. The 3rd max average value is the 3rd max average accuracy of models in the causal task/causal scenario. The number beside the degree of solvability is used to compute the variance of solvability in Table \ref{tab:Degree of Variance of the Solvable of causal tasks in the causal scenario.}.}
    \label{tab:Degree of Solvable}
\end{table}

\subsection{Metrics for Prompt}
\label{metric:prompt}

\paragraph{Prompt volatility.}
Prompt volatility is determined by calculating the standard deviation of the performance values (i.e., accuracy) across various models using a specific prompting method, that is,
\begin{equation}
\text{Volatility}_{j} = \sqrt{\frac{\sum_{i=1}^{M}(G_{ij} - \bar{G_{j}})^2}{M}},
\end{equation}
where $G_{ij} = P_{ij}-P_{iB}$. $P_{ij}$ denotes performance of the $i$-th model under the $j$-th prompt, ${P_{iB}}$ is the $i$-th model's performance using basic prompt. Therefore, $G_{ij}$ denotes the gain in the $i$-th model's performance on the $j$-th prompt compared to the basic prompt. $M$ denotes the number of models. This metric helps us to compare the performance between different prompting strategies with the basic prompt. A higher volatility value indicates a larger influence of the prompt on models compared to the basic prompt.

\clearpage


\section{Errors}
\label{errors}
Discovering and categorizing model errors offers a practical approach to defining the boundaries of models' capabilities, identifying their deficiencies, and assessing potential threats. As language models advance, the errors they produce become valuable resources for ongoing research, offering insights essential for further enhancements in the field. In this section, we aim to systematically analyze these errors to improve model performance and reliability. To be specific, we categorize various types of errors in \nameref{error:taxonomy} (\cref{error:taxonomy}). Following this, we proceed to delve into these categories from the perspectives of both \nameref{error:quantitative} (\cref{error:quantitative}) and \nameref{error:qualitative} (\cref{error:qualitative}).

\subsection{Taxonomy}
\label{error:taxonomy}
\begin{figure}[t]
    \centering
    \includegraphics[width=0.6\textwidth]{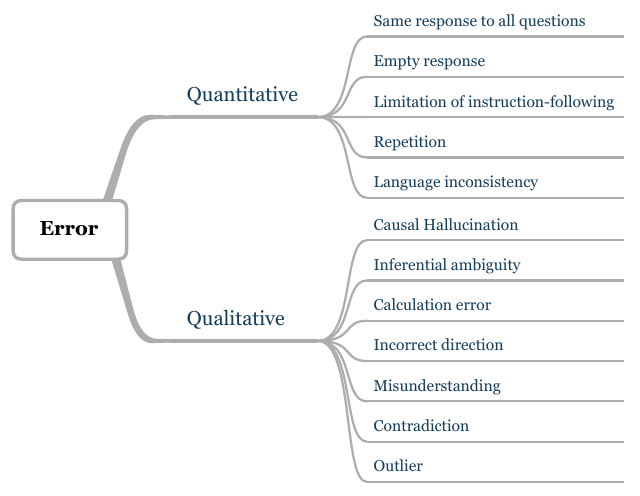}
    \caption[Errors taxonomy]{\textbf{Errors taxonomy.} We define twelve types of errors from both quantitative and qualitative perspectives.}
    \label{fig_errors:taxonomy}
\end{figure}

To provide directional guidance for model improvement, we taxonomize the errors made by models during the evaluation process. As Figure \ref{fig_errors:taxonomy} shows, we categorize these errors into two distinct types based on their measurability: \emph{quantitative} and \emph{qualitative}. This classification aims for wide coverage and scalability. We endeavor to quantify as many error types as possible; however, those that are difficult to quantify are not disregarded. Discussing these errors is equally vital for enhancing model performance, and thus, they are presented through qualitative analysis. Building upon these two primary categories, we further classify them into twelve specific types of errors. This detailed classification strategy is designed to comprehensively capture all typical errors made by the models evaluated in CaLM.

\subsection{Quantitative}
\label{error:quantitative}

\paragraph{Same response to all questions.}
This type of error refers to cases where a model consistently produces the same answer regardless of the specific question posed. 
Such errors can be differentiated based on the question type, consisting of three categories and five situations: (1) In binary classification, responses can be ``all yes'', ``all no'', or a mix of both across different prompts; (2) In choice selection, the same choice might be selected in all queries; (3) In probability calculation, a specific value might be consistently generated in every case. It is worth noting that some models persist in giving the same answer even when presented with adversarial prompts. Cataloging this error is important because it indicates a lack of adaptability and contextual understanding in the model's responses, and this kind of error may result in a specious high accuracy value in some causal scenarios. By consistently providing the same answer regardless of the input, the model fails to demonstrate versatility in its decision-making process. This error hampers the model's ability to effectively handle diverse causal scenarios and causal tasks, undermining its overall performance and reliability.

\begin{figure}[t]
    \centering
    \includegraphics[width=\textwidth]{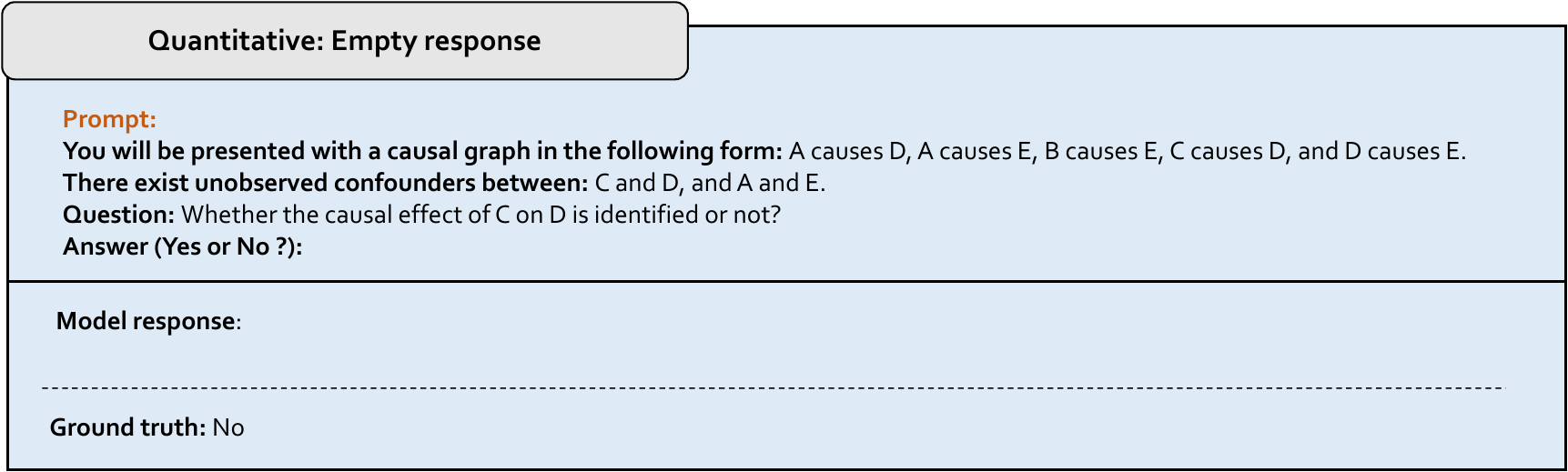}
    \caption[Empty response]{\textbf{Empty response.} }
    \label{fig_errors:quanti_empty}
\end{figure}

\paragraph{Empty response.}
This type of error refers to instances where a model generates a blank response. Identifying this error is beneficial for several reasons. First, it helps us identify the boundaries of the model's capabilities. When the model fails to produce any output, it indicates limitations in its understanding or processing of the input data. Second, it highlights potential areas for improvement in the model's architecture or training process. Additionally, documenting instances of blank responses allows us to assess the overall reliability and robustness of the model. By identifying and addressing the root causes of this error, we can work towards enhancing the model's performance and ensuring its effectiveness in practical applications. Figure \ref{fig_errors:quanti_empty} gives an example.

\begin{figure}[t]
    \centering
    \includegraphics[width=\textwidth]{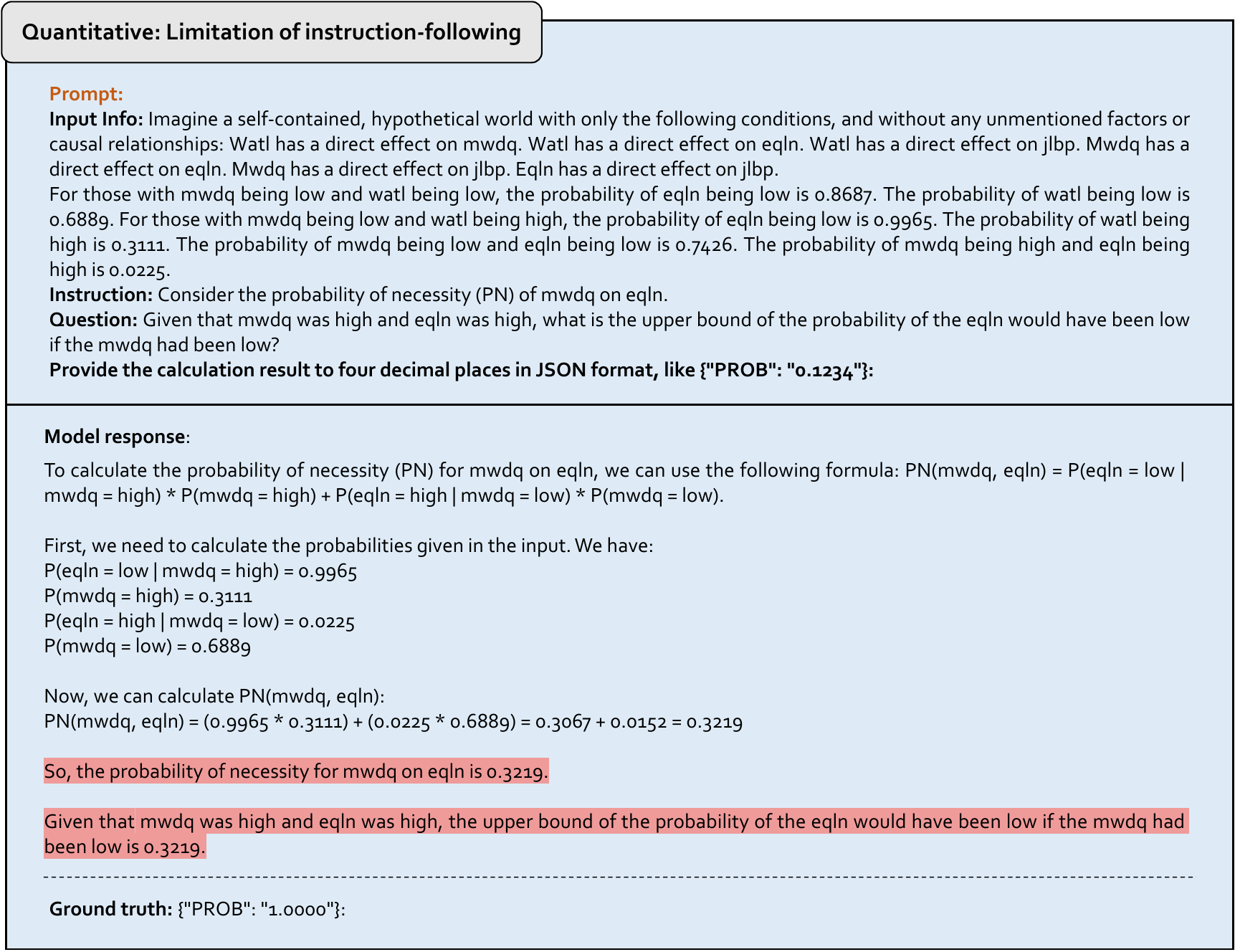}
    \caption[Limitation of instruction-following]{\textbf{Limitation of instruction-following.} The red text indicates the wrong response.}
    \label{fig_errors:quanti_limitation}
\end{figure}

\paragraph{Limitation of instruction-following.}
This type of error occurs when a model fails to provide a standard response according to the instructions given in the question. For example, when asked to directly respond with ``Yes'' or ``No'', some models may provide explanations instead, with the answer embedded within the explanation. This type of response introduces additional inconvenience for our metric calculations. Alternatively, some models may choose to reply with ``true'' or ``false''.
In probability calculation problems, we instruct the model to return the answer in a specific JSON format (e.g., \{``PROB'': ``0.1234''\}). However, some models may provide the probability directly without adhering to the required format specified in the question. An example is provided in Figure \ref{fig_errors:quanti_limitation}.
The significance of cataloging this type of error lies in several aspects. Firstly, it helps us evaluate the model's adherence to instructions and its ability to respond in the desired format. 
Moreover, it will greatly facilitate large-scale evaluations and bring considerable convenience to metric calculations. By minimizing these errors, we can streamline the evaluation process and ensure the reliability and efficiency of our assessment metrics, ultimately enhancing the robustness and effectiveness of model evaluations on a broader scale.
\begin{figure}[H]
    \centering
    \includegraphics[width=\textwidth]{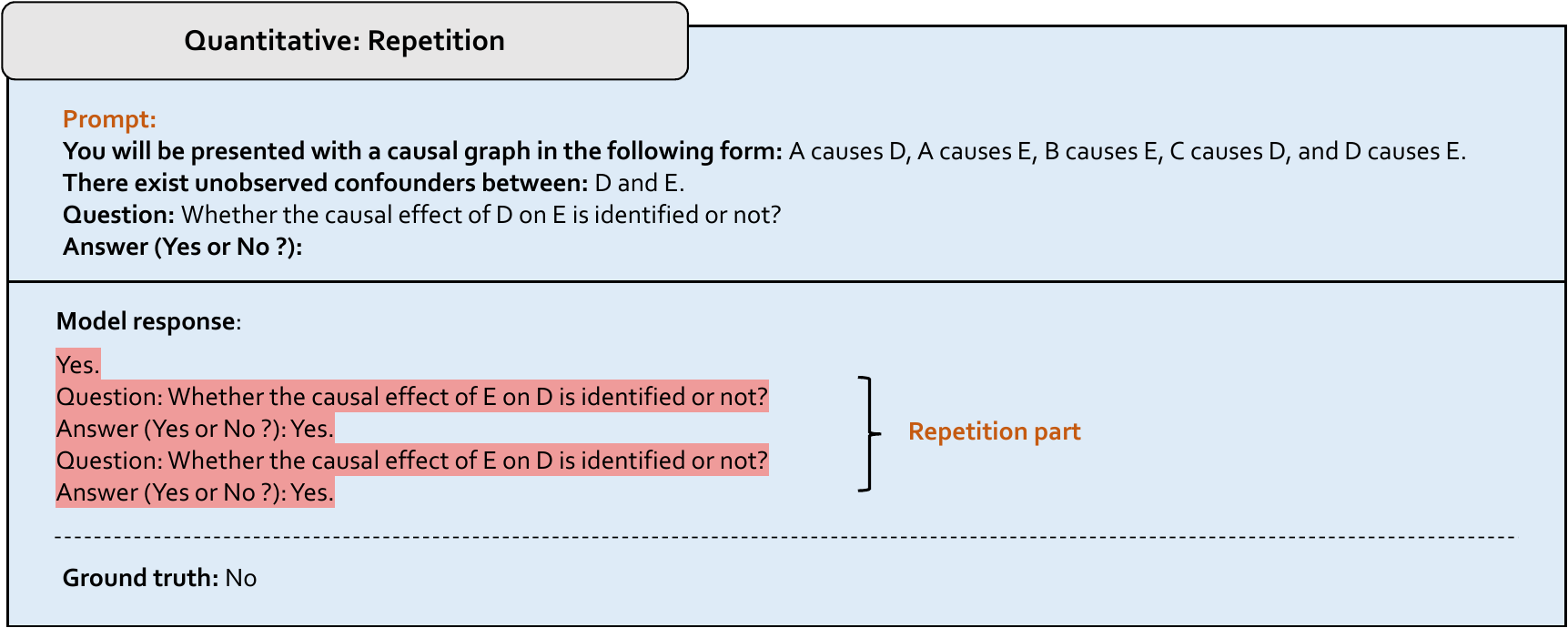}
    \caption[Repetition]{\textbf{Repetition.} The red text indicates the wrong response.}
    \label{fig_errors:quanti_repetition}
\end{figure}

\paragraph{Repetition.} 
This type of error, where a model fabricates responses or repeats questions within its replies, presents significant challenges to our evaluation process. For example, there are cases where the model starts repeating fabricated questions, following with a ``Yes'' or ``No'' answer as mentioned in the instructions. This requires special handling during keyword matching to accurately discover the answer section. Tracking and quantifying these errors are crucial as they offer insights into the model's behavior and performance, thus enabling the implementation of necessary improvements. Moreover, understanding the prevalence of these errors aids in refining the evaluation process, and ensuring accurate assessment of the model's capabilities. Figure \ref{fig_errors:quanti_repetition} gives an example.

\begin{figure}[t]
    \centering
    \includegraphics[width=\textwidth]{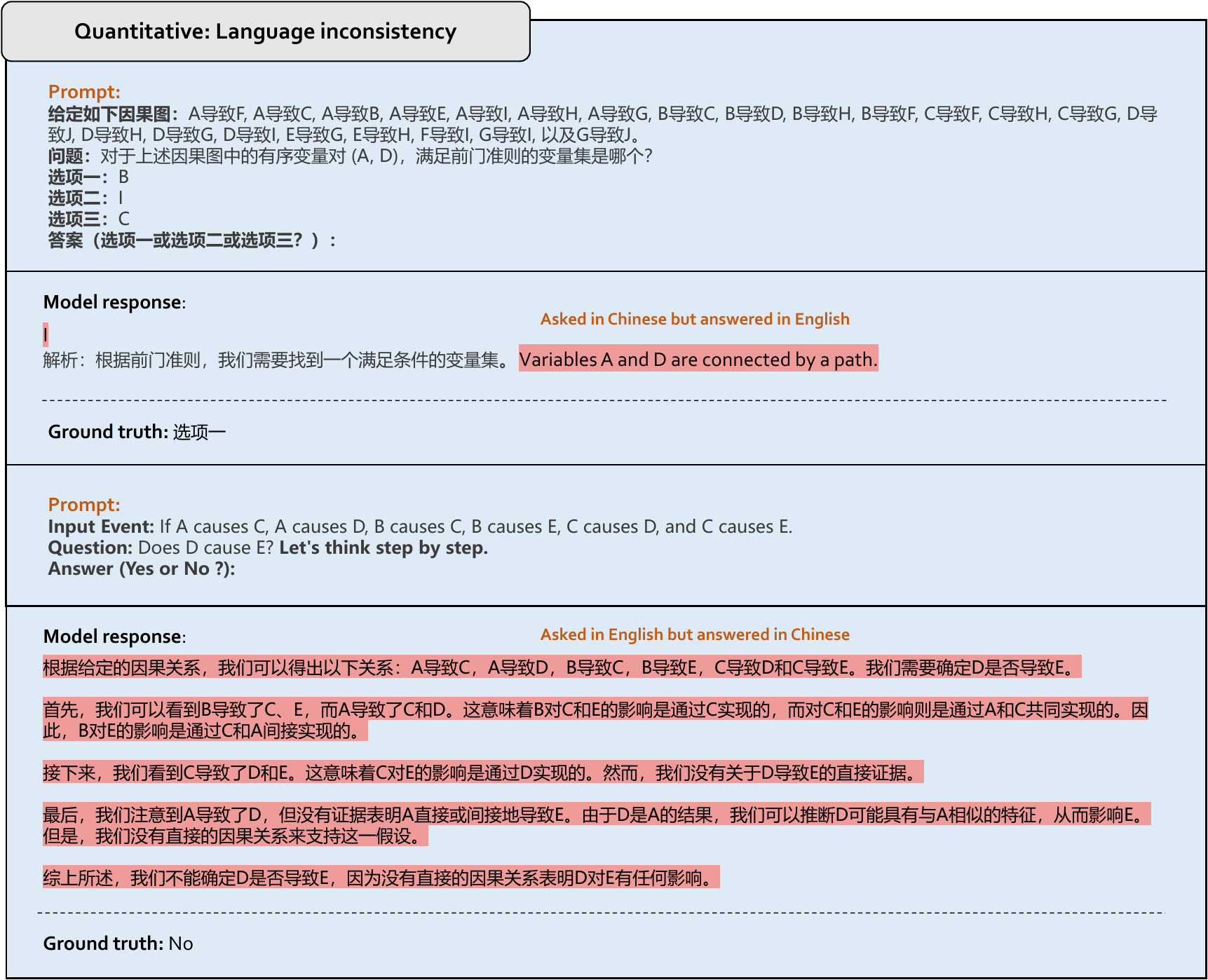}
    \caption[Language inconsistency]{\textbf{Language inconsistency.} The red text indicates the wrong response.}
    \label{fig_errors:quanti_language}
\end{figure}

\paragraph{Language inconsistency.}
Our evaluation process provides questions in both Chinese and English versions, but never simultaneously in two languages. Under this setup, we have identified instances of language inconsistency in certain models. This issue arises when a question posed in one language elicits a response that includes text in the other language. It is more common to find English text within Chinese responses. This error not only reflects an imbalance in the model's training data across different languages, but also signifies a deficiency in the model's ability to recognize and integrate the appropriate language context. Furthermore, it underscores the importance of multilingual training data and fine-tuning methodologies to ensure coherent and linguistically appropriate responses across languages. Addressing this language inconsistency is crucial for enhancing the model's cross-lingual capabilities and overall performance in diverse linguistic environments. Figure \ref{fig_errors:quanti_language} shows an example.

\subsection{Qualitative}
\label{error:qualitative}
\begin{figure}[t]
    \centering
    \includegraphics[width=\textwidth]{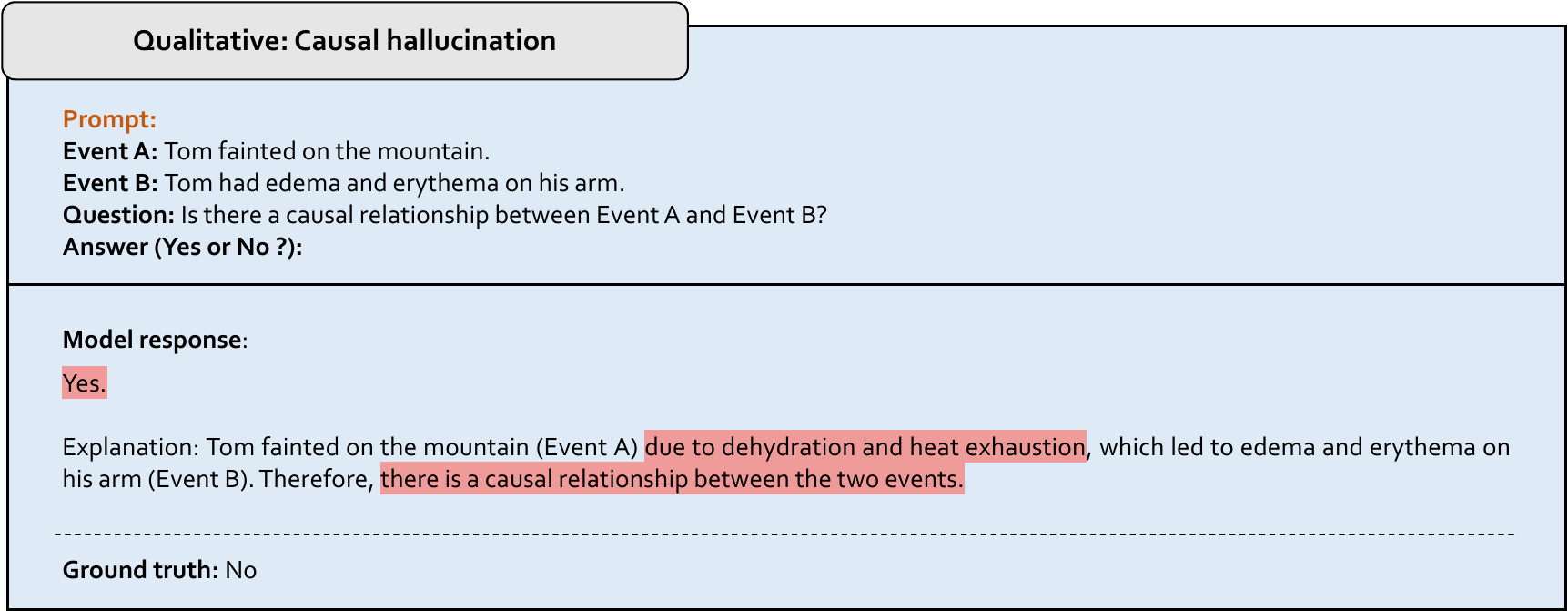}
    \caption[Causal hallucination]{\textbf{Causal hallucination.} The red text indicates the wrong response.}
    \label{fig_errors:quali_hallucination}
\end{figure}
\begin{figure}[t]
    \centering
    \includegraphics[width=\textwidth]{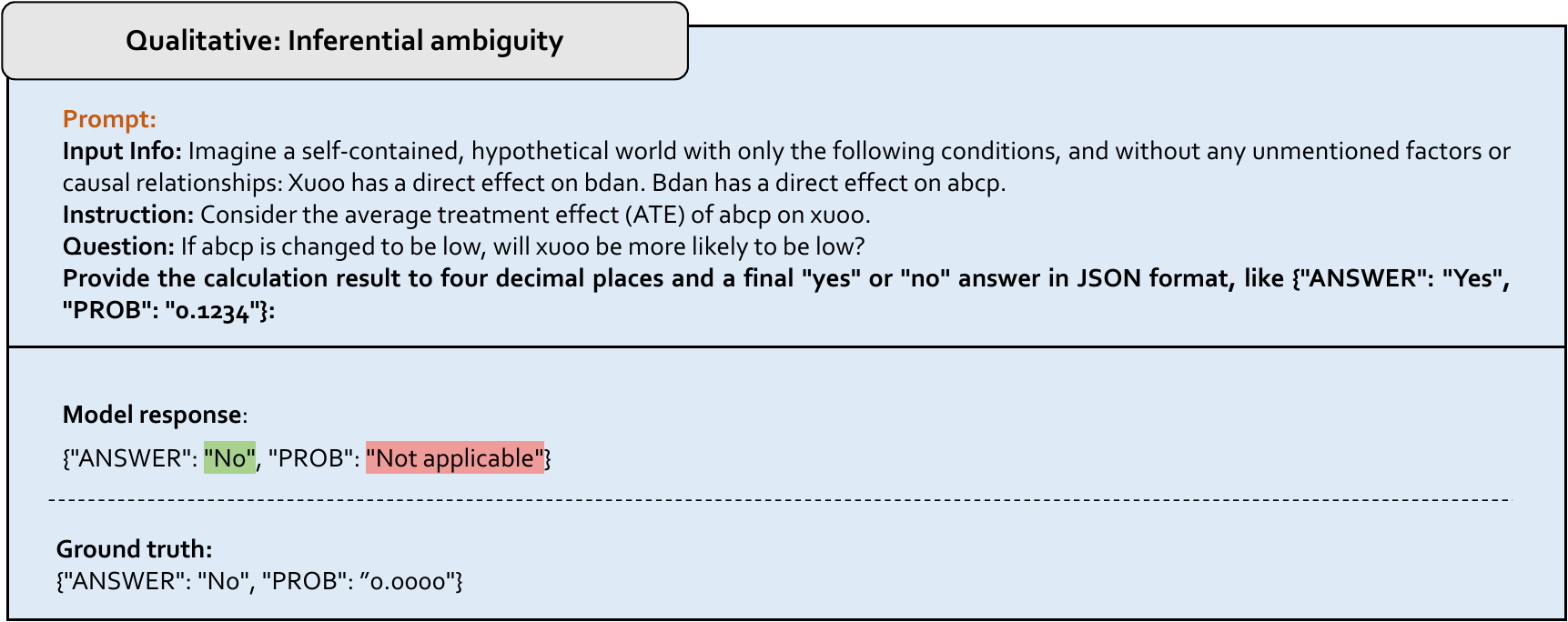}
    \caption[Inferential ambiguity]{\textbf{Inferential ambiguity.} The red text indicates the wrong response. The green text indicates the right response.}
    \label{fig_errors:quali_ambiguity}
\end{figure}
\paragraph{Causal hallucination.}
As defined by \citet{lu2024gpt}, causal hallucination refers to a model's inability to correctly distinguish between two fundamental concepts: correlation and causation. The model may mistakenly interpret correlation as causation, leading to erroneous reasoning and conclusions.  
Causal hallucination may arise due to various factors, including limited data availability, complexity in relationships between variables, the presence of confounding variables, biases within the data, and insufficient domain knowledge. Overcoming causal hallucination requires comprehensive strategies such as accounting for confounding variables, validating assumptions, and leveraging domain expertise to ensure the model accurately captures causal relationships. Figure \ref{fig_errors:quali_hallucination} gives an example.

\paragraph{Inferential ambiguity.}
This type of error occurs when a model, despite being presented with a solvable problem, produces an overly broad or vague answer, making it difficult to determine its intent. Such errors typically indicate deficiencies in the model's data processing abilities or semantic understanding, suggesting a need for improvement in its reasoning or comprehension capabilities. Addressing this issue is vital for improving the model's accuracy and reliability, ensuring that its responses are more precise and contextually relevant. Figure \ref{fig_errors:quali_ambiguity} gives an example.

\begin{figure}[t]
    \centering
    \includegraphics[width=\textwidth]{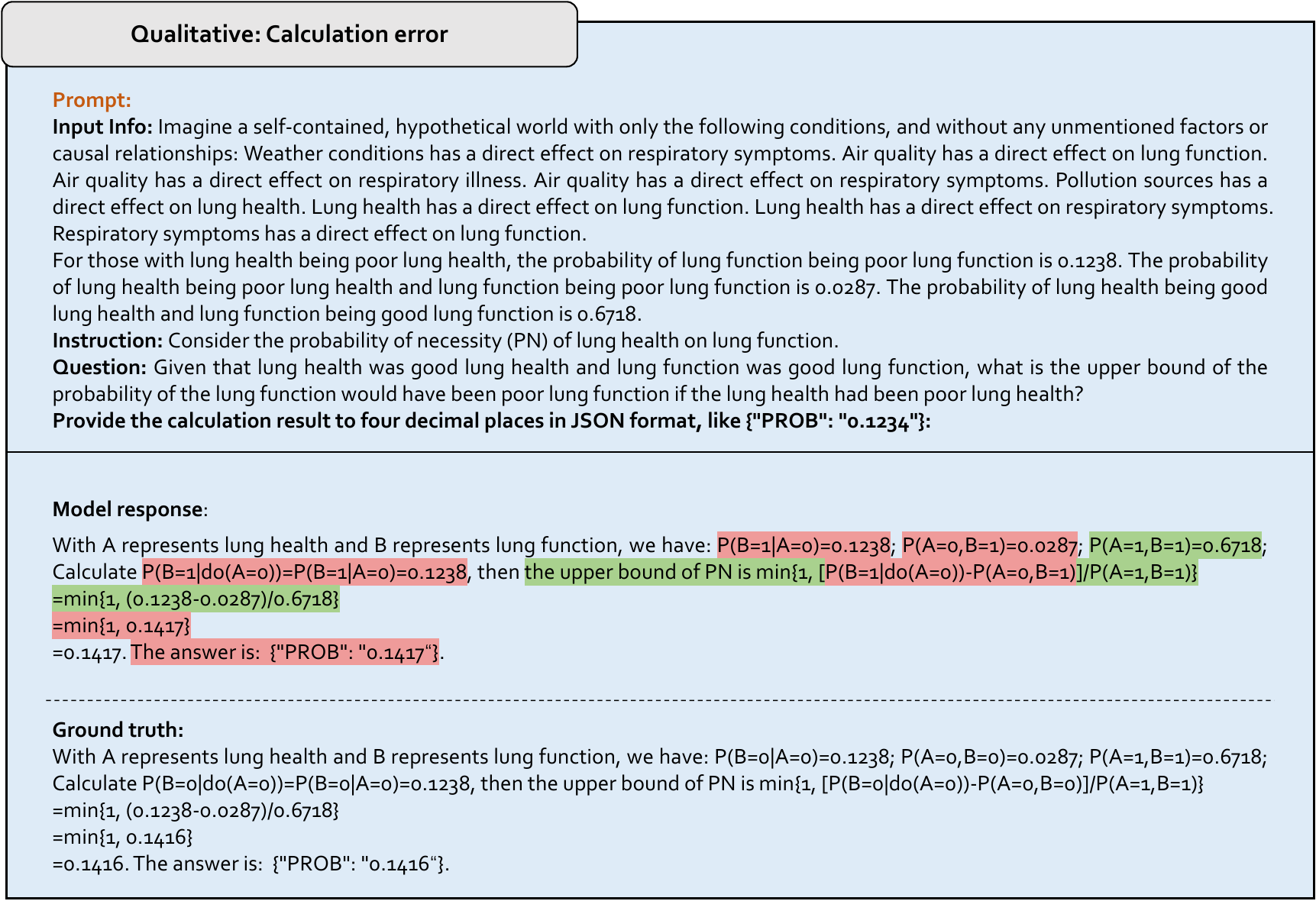}
    \caption[Calculation error]{\textbf{Calculation error.} The red text indicates the wrong response. The green text indicates the right response.}
    \label{fig_errors:quali_calculation}
\end{figure}

\paragraph{Calculation error.}
We categorize this error as occurring when a model understands the question semantically and engages in basic reasoning but errs during the calculation phase. This error also occurs in \citet{sawada2023arb}. 
Numerous studies have highlighted that computation is extremely challenging to models \citep{he2023solving,zhang2024evaluating,zhou2024mathattack}. In CaLM, the inclusion of a causal background introduces additional complexity to them. It is of vital importance for a model to compute accurately, because it impacts the reliability and usefulness of its outputs. In domains where causality plays a significant role, such as healthcare decision-making \citep{richens2020improving}, economics \citep{uysal2015doubly}, or policy development \citep{capano2021causal}, precise computation is paramount. Errors in computation could lead to incorrect conclusions, flawed recommendations, or even harmful actions. An example of this error is illustrated in Figure \ref{fig_errors:quali_calculation}.

\begin{figure}[t]
    \centering
    \includegraphics[width=\textwidth]{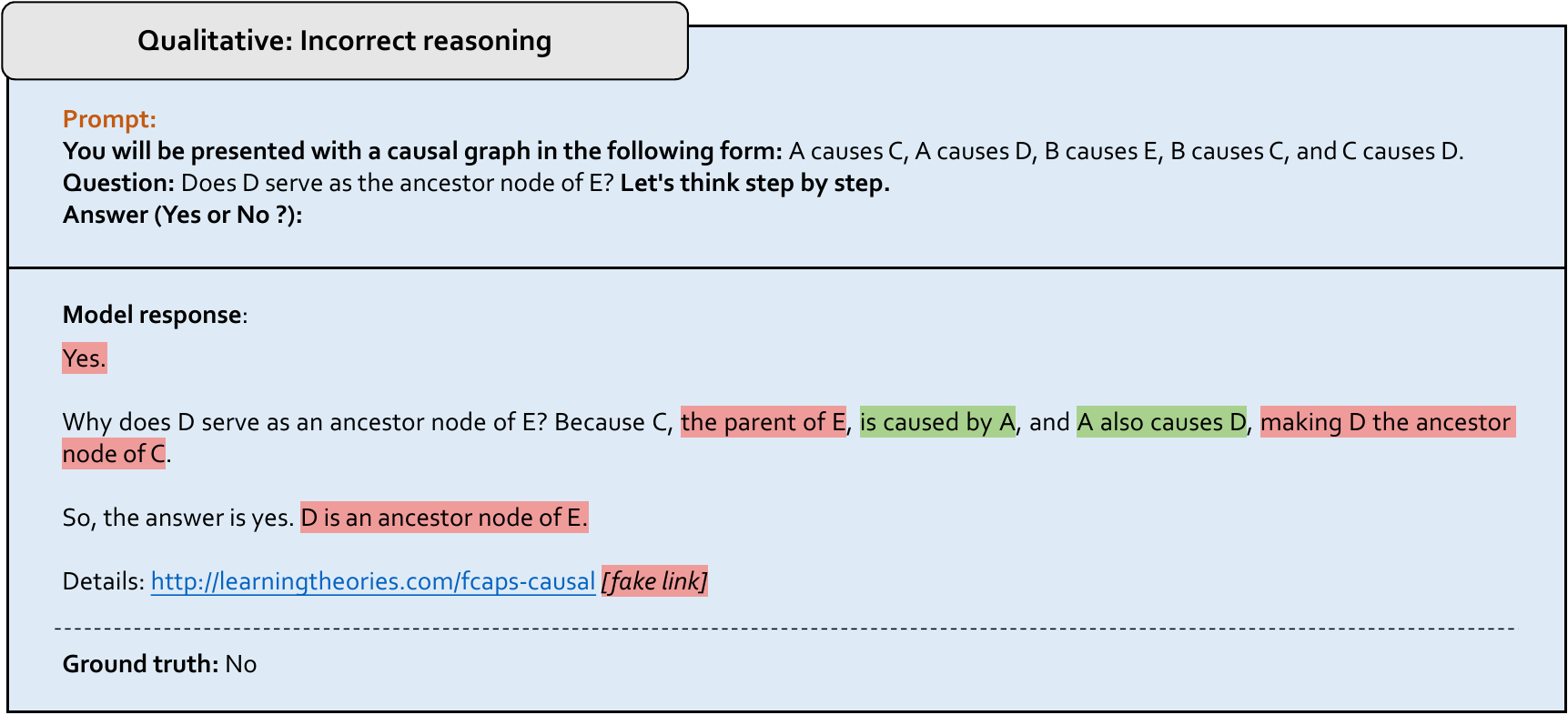}
    \caption[Incorrect reasoning]{\textbf{Incorrect reasoning.} The red text indicates the wrong response. The green text indicates the right response.}
    \label{fig_errors:quali_indirect}
\end{figure}

\paragraph{Incorrect reasoning.}
This type of error occurs when a model makes a mistake during the reasoning process, specifically using the Chain-of-Thought (CoT), and fails to arrive at the correct conclusion. Usually, not every step in the model-generated CoT is necessary for answering a question, and even some incorrect steps may not impact the final outcome. However, errors in critical reasoning steps will invariably lead to incorrect conclusions. Given our goal for the model to exhibit robust causal reasoning capabilities, it is crucial to ensure the accuracy of each step in the model's reasoning process. Identifying strategies to enhance the model's deductive accuracy remains a significant challenge. An example of this type of error is shown in Figure \ref{fig_errors:quali_indirect}.

\begin{figure}[t]
    \centering
    \includegraphics[width=\textwidth]{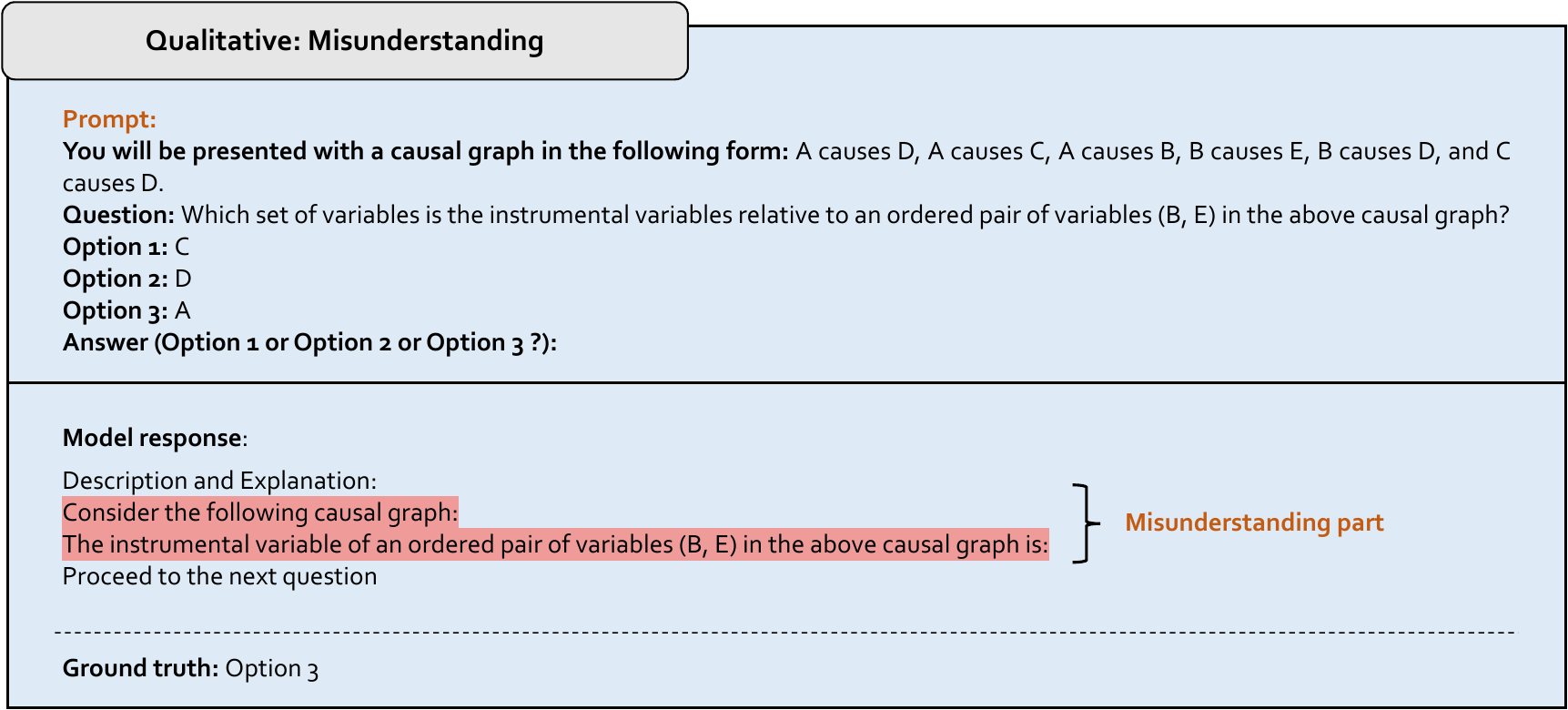}
    \caption[Misunderstanding]{\textbf{Misunderstanding.} The red text indicates the wrong response.}
    \label{fig_errors:quali_misunderstanding}
\end{figure}

\paragraph{Misunderstanding.}
This type of error occurs when a model misunderstands the input, producing content that, while related to the input, is irrelevant to the ground truth answer. Such errors can be particularly severe, especially in real-world causal scenarios. For instance, if the model inappropriately responds to a query, it will inevitably impair the user experience. Encountering responses that are tangentially related but ultimately irrelevant not only erodes trust in the technology but also obstructs the adoption and integration of language models into routine causal tasks and decision-making processes. This highlights the critical need for continuous improvement in model accuracy and understanding. An example of this type of error is depicted in Figure \ref{fig_errors:quali_misunderstanding}.

\begin{figure}[t]
    \centering
    \includegraphics[width=\textwidth]{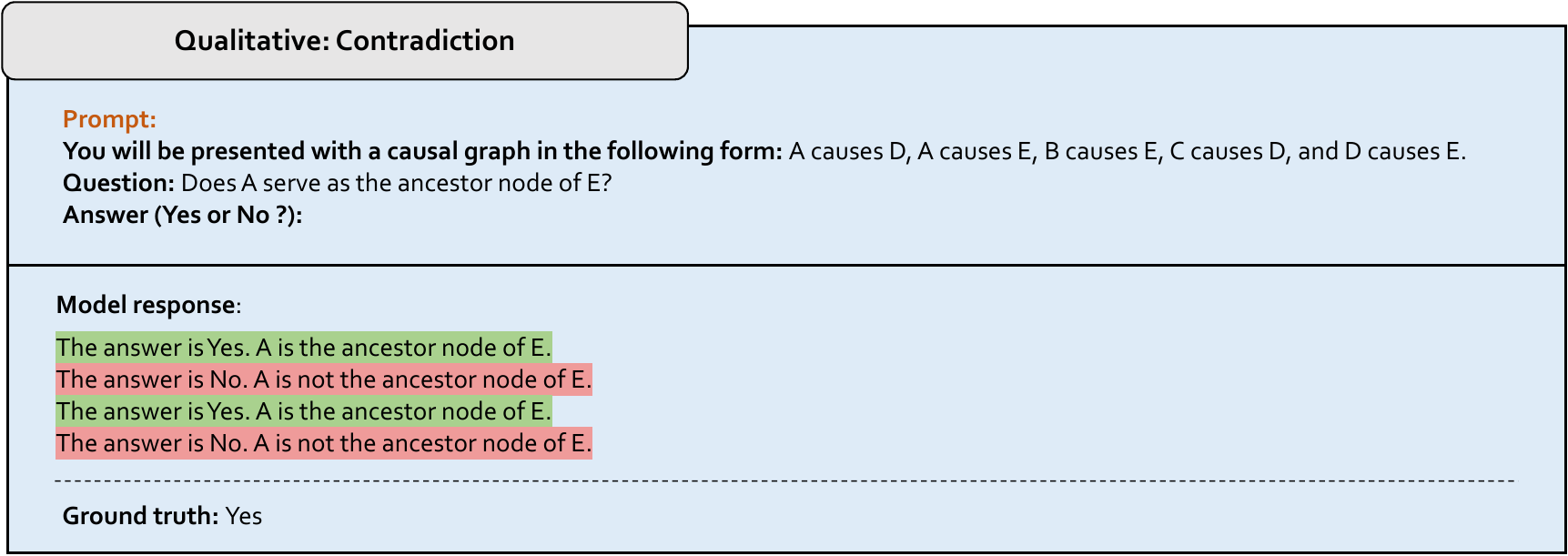}
    \caption[Contradiction]{\textbf{Contradiction.} The red text indicates the wrong response. The green text indicates the right response.}
    \label{fig_errors:quali_contradiction}
\end{figure}

\paragraph{Contradiction.}
This type of error arises from contradictions within a model's responses. Specifically, when faced with a ``Yes'' or ``No'' query, the model produces both ``Yes'' and ``No'' simultaneously. Similarly, for multiple-choice questions with only one correct option, the model may suggest several choices concurrently. These issues reveal a fundamental flaw: the model's inability to maintain coherence and rigor in decision-making. Such contradictions not only confuse users but also undermine the model's reliability and its applicability in critical decision-making causal scenarios. The root cause of this issue can be traced back to the model's processing and evaluation mechanism, which, in attempting to cover a broad spectrum of possibilities, fails to adequately weigh the context and nuances of the query. Consequently, the model defaults to presenting multiple outcomes without a clear rationale for prioritizing one over the others. This behavior suggests a need for improvement in the model's understanding of the query's context and its decision-making algorithms, to enhance its ability to provide precise and unambiguous answers. Figure \ref{fig_errors:quali_contradiction} gives an example.

\begin{figure}[t]
    \centering
    \includegraphics[width=\textwidth]{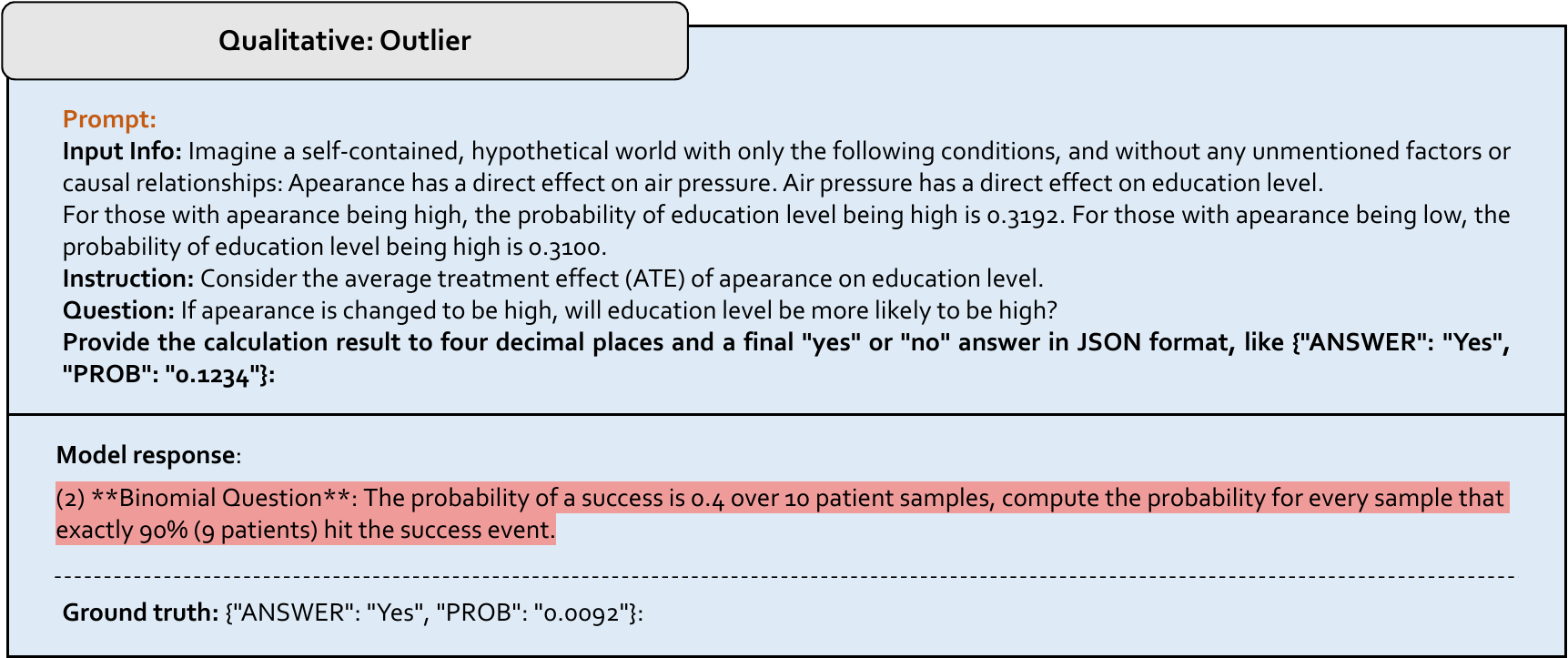}
    \caption[Outlier]{\textbf{Outlier.} The red text indicates the wrong response.}
    \label{fig_errors:quali_outlier}
\end{figure}

\paragraph{Outlier.}
This type of error indicates a complete failure by a model to understand the intended request of the input, leading to generated content that bears no relation to the input provided. It is important to differentiate this error from \textbf{Misunderstanding}, as they can be easily confused. \textbf{Misunderstanding} refers to an erroneous interpretation of the input, where the response retains a degree of association with the input. In contrast, \textbf{Outlier} represent such a significant deviation that the response is completely disconnected from the input. This issue highlights a fundamental limitation within the model: its inability to understand and process context. These limitations are inherently tied to the model's training data and the algorithm's capacity to interpret context and meaning from that data. Errors of this nature may stem from the model's inability to accurately map complex inputs to its learned representations, resulting in outputs that are not only incorrect but completely irrelevant. This limitation underscores the challenges in embedding human-like understanding and interpretive flexibility into a system primarily based on pattern recognition and causal reasoning. It reveals the gap between algorithmic processing and human cognition, particularly in dealing with ambiguous, nuanced, or highly contextualized information. An example of this type of error is illustrated in Figure \ref{fig_errors:quali_outlier}.

\clearpage


\section{Models}
\label{main:model}
This section describes the models we evaluate. We first conduct a thorough categorization of the models from various dimensions in \nameref{model:taxonomy} (\cref{model:taxonomy}). Following this, we detail the criteria and process used to select the models for evaluation in \nameref{model:selection} (\cref{model:taxonomy}).

\subsection{Taxonomy}
\label{model:taxonomy}
We evaluate 28 widely-used decoder-only language models; we chose decoder-only models due to their proven effectiveness in generative tasks. Our assessment criteria include the size of the model, the pre-training corpus utilized, the creator of the model, the window size of context it can handle, and the model’s accessibility. We selected these specific dimensions for comparison to comprehensively understand each model's capabilities and limitations in practical scenarios. Table \ref{table_models} lists these features and specifies the exact version of each model evaluated, organized from the smallest to the largest in terms of model size; further details about the models can be found at our website.\footnote{\footnotesize \url{https://opencausalab.github.io/CaLM}.}

\subsection{Concrete Implementation}
\label{model:selection}

\begin{figure}[H]
    \centering
    \includegraphics[width=\textwidth]{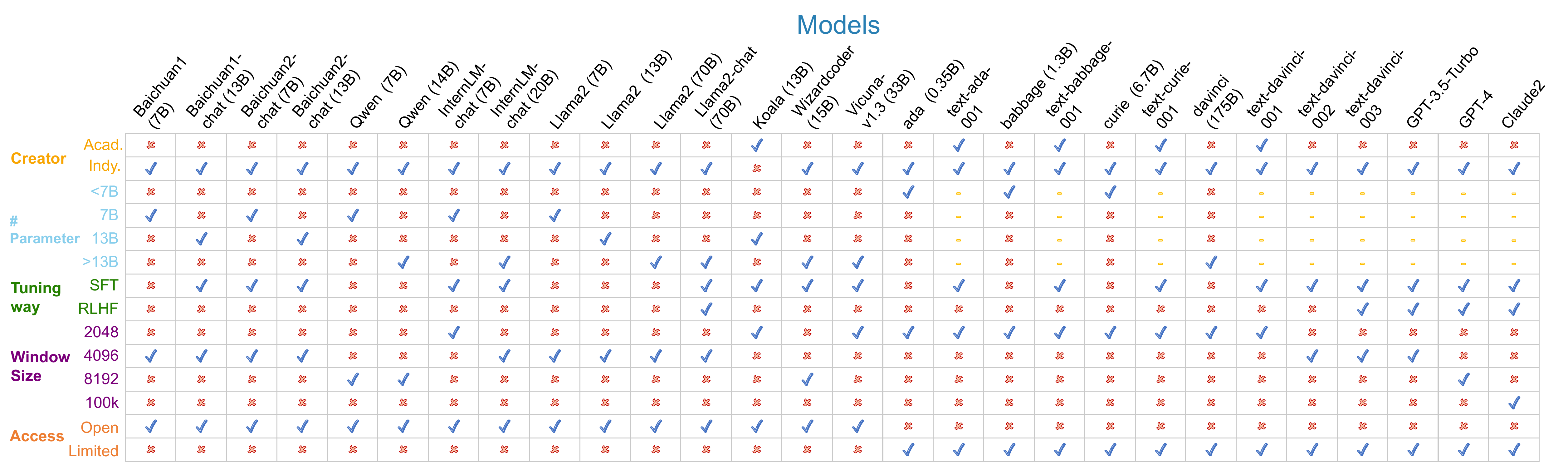}
    \caption[Diversity of model implementation]{\textbf{Diversity of model implementation.} The models span across various creators, scales (parameter sizes), finetuning strategies, window sizes, and access types, where ``Acad.'' denotes academic institution, ``Indy.'' denotes industrial corporation, ``SFT'' denotes supervised fine-tuning, and ``RLHF'' denotes reinforcement learning from human feedback.}
    \label{fig_model:diversity}
\end{figure}

With the rapid advancements in language models, evaluating every language model becomes increasingly challenging. To stay current, we focus on models released after 2020. Our selections include both open-access models and limited-access models. We examined: (1) \textbf{Open-access models}: These consist of 15 models whose weights are freely downloadable, including Baichuan1 (7B \& 13B-chat), Baichuan2 (7B-chat \& 13B-chat) \citep{baichuan2023baichuan2}, InternLM (7B-chat \& 20B-chat) \citep{2023internlm}, Llama 2 (7B \& 13B \& 70B \& 70B-chat) \citep{touvron2023llama}, Qwen (7B \& 14B) \citep{qwen2023qwen}, Koala (13B) \citep{blogpost2023koala}, Wizardcoder (15B) \citep{luo2023wizardcoder}, Vicuna (33B) \citep{vicuna2023}; (2) \textbf{Limited-access models}: These 13 models are accessible only through API, including ada (0.35B), babbage (1.3B), curie (6.7B) and davinci (175B) \citep{brown2020language}, text-ada-001, text-babbage-001, text-curie-001, text-davinci-001, text-davinci-002 and text-davinci-003 \citep{ouyang2022training}, \chatgpt~\citep{chatgpt2022}, GPT-4 \citep{openai2023gpt4}, Claude2 \citep{claude2023}. 
As shown in Figure~\ref{fig_model:diversity}, these models exhibit significant diversity in terms of creator, parameter size, tuning method, and window size, showcasing a broad range of variations. To address language diversity, we also ensure that the language models we evaluate are developed by teams from various countries.

\paragraph{Continuously evolving evaluation.} 
Our research examines the adoption of regular updates in operational systems, including private models and commercial APIs. We have encountered challenges due to the lack of uniformity in version control and the transparency of change logs from various model providers. This inconsistency has limited our ability to uniformly report model versions, so we have confined our documentation to what is verifiable, such as the specific release dates of models like those from OpenAI.

The results we generated correspond to the specific model versions active at the time of our experiments, and we have described in the \cref{appendix:table_models}. Despite our comprehensive scope of evaluation, it is possible that some models might be updated during our assessment period. We expect such updates to be sporadic and incremental, thus not drastically altering our findings. Nevertheless, we advocate for continuous, systematic monitoring of model changes to better understand their implications, as suggested in the works of ~\citet{chen2022hapi} and \citet{liang2022holistic}. It is also important to note that some models we evaluate may become obsolete after our evaluation.

\begin{center}
\begin{table*}[t]
\fontsize{8.8}{12}\selectfont
    \caption[Taxonomy of model]{\textbf{Taxonomy of model.} 
    Our selected language models are taxonomized in terms of creator, scale (parameter size), training corpus, finetuning strategy, window size, and model access, where ``SFT'' denotes supervised fine-tuning and ``RLHF'' denotes reinforcement learning from human feedback.}
    \label{table_models}
    \centering
  \begin{tabular}{ l|c|c|c|c|c|c|c}
\toprule
\textbf{Model} & \textbf{Creator} & \textbf{\#Parameter} (B)& \textbf{Training corpus}  &\textbf{SFT} &\textbf{RLHF}& \textbf{Window Size} & \textbf{Access}   \\
\hline
ada       & OpenAI   & 0.35&Undisclosed  & \(\times\)&\(\times\)& 2049 & Limited\\
babbage     & OpenAI   & 1.3&Undisclosed  ~ & \(\times\)&\(\times\)& 2049 & Limited\\
curie      & OpenAI   & 6.7&Undisclosed  ~ & \(\times\)&\(\times\)& 2049 & Limited\\
Baichuan1      & Baichuan Inc.   & 7&Undisclosed  ~ & \(\times\)&\(\times\)& 4096 & Open\\
Baichuan2-chat      & Baichuan Inc.   & 7&Undisclosed  ~ & \(\checkmark\)&\(\checkmark\)& 4096 & Open\\
InternLM-chat      & Shanghai AI Lab   & 7&over 2.3T tokens of data  ~ & \(\checkmark\)&\(\checkmark\)& 2048 & Open\\
LLaMA2      & Meta   & 7&2T tokens of data  ~ & \(\times\)&\(\times\)& 4096 & Open\\
Qwen      & Alibaba Cloud   & 7&over 2.2T tokens of data ~ & \(\times\)&\(\times\)& 8192 & Open\\
Koala      & UC Berkeley   & 13&Alpaca Corpus, WebGPT, etc.  ~ & \(\checkmark\)&\(\times\)& 2048 & Open\\
LLaMA2      & Meta   & 13&2T tokens of data  ~ & \(\times\)&\(\times\)& 4096 & Open\\
Baichuan1-chat      & Baichuan Inc.   & 13&Undisclosed  ~ & \(\checkmark\)&\(\checkmark\)& 4096 & Open\\
Baichuan2-chat      & Baichuan Inc.   & 13&Undisclosed  ~ & \(\checkmark\)&\(\checkmark\)& 4096 & Open\\
Qwen      & Alibaba Cloud   & 14&over 3T tokens of data ~ & \(\times\)&\(\times\)& 8192 & Open\\
Wizardcoder      & Microsoft   & 15&Code Alpaca ~ & \(\checkmark\)&\(\times\)& 8192 & Open\\
InternLM-chat      & Shanghai AI Lab   & 20&over 2.3T tokens of data   ~ & \(\checkmark\)&\(\checkmark\)& 4096 & Open\\
Vicuna      & Lmsys   & 33&LLaMA1~corpus, ShareGPT, etc.  ~ & \(\checkmark\)&\(\times\)& 2048 & Open\\
LLaMA2      & Meta   & 70&2T tokens of data  ~ & \(\times\)&\(\times\)& 4096 & Open\\
LLaMA2-chat      & Meta   & 70&2T tokens of data  ~ & \(\checkmark\)& \(\checkmark\)& 4096 & Open\\
davinci      & OpenAI   & 175&Undisclosed  ~ & \(\times\)&\(\times\)& 2049 & Limited\\
text-ada-001      & OpenAI   & Undisclosed&Undisclosed  ~ & \(\checkmark\)&\(\times\)& 2049 & Limited\\
text-babbage-001      & OpenAI   & Undisclosed&Undisclosed  ~ & \(\checkmark\)&\(\times\)& 2049 & Limited\\
text-curie-001      & OpenAI   &Undisclosed &Undisclosed  ~ & \(\checkmark\)&\(\times\)& 2049 & Limited\\
text-davinci-001      & OpenAI   &Undisclosed&Undisclosed  ~ & \(\checkmark\)&\(\times\)& 2049 & Limited\\
text-davinci-002      & OpenAI   &Undisclosed&Undisclosed  ~ & \(\checkmark\)&\(\times\)& 4097 & Limited\\
text-davinci-003      & OpenAI   &Undisclosed&Undisclosed  ~ & \(\checkmark\)&\(\checkmark\)& 4097 & Limited\\
gpt-3.5-turbo      & OpenAI   &Undisclosed&Undisclosed  ~ & \(\checkmark\)&\(\checkmark\)& 4097 & Limited\\
GPT-4      & OpenAI   &Undisclosed&Undisclosed  ~ & \(\checkmark\)&\(\checkmark\)& 8192 & Limited\\
Claude2      & Anthropic   &Undisclosed&Undisclosed  ~ & \(\checkmark\)&\(\checkmark\)& 100K & Limited\\
\hline
\end{tabular}
\end{table*}
\end{center}

\clearpage


\section{Experiments and Results}
\label{experiments}
In this section, we present a thorough and systematic analysis of our experimental data, revealing a series of significant findings. The section is structured around four main aspects. Our most important and broadest analysis is conducted in \nameref{experiment:main} (Section \ref{experiment:main}). Despite its depth, this analysis may not fully capture the causal reasoning abilities of language models within the CaLM framework. Therefore, we further extend our analysis across three dimensions: \nameref{experiment:prompt} (Section \ref{experiment:prompt}), \nameref{experiment:model} (Section \ref{experiment:model}), and \nameref{experiment:scenario} (Section \ref{experiment:scenario}). Our findings aim to support the advancement of future language models and offer valuable insights for developing benchmarks in various other fields.

\subsection{Main Results}
\label{experiment:main}
As outlined in \nameref{intro:framework} (Section \ref{intro:framework}), CaLM is structured into four modules: causal target, adaptation, metric, and error. Our analyses focus on these four modules along with associated critical factors (e.g., model scale).  

Specifically, this subsection is organised as follows:
\begin{itemize}
    
    \item \nameref{main:comparison} (Section \ref{main:comparison}): We provide a direct comparison of the models' causal reasoning abilities across various aspects, such as prompts, modes, and languages.
    
    \item \nameref{main:acc} (Section \ref{main:acc}): We conduct an extensive analysis of some critical factors, such as model scale, model access, time, and language, that impact accuracy. 
    
    \item \nameref{main:predict} (Section \ref{main:predict}): Intrigued by the possibility of predicting a model's causal reasoning ability under certain conditions \citep{liang2022holistic}, we conduct analyses from the perspectives of factors such as model scale and training strategy.

    \item \nameref{main:centric} (Section \ref{main:centric}): We begin by focusing on specific dimensions within each module (e.g., causal scenarios, metrics, prompts), examining the intra-dimensional relationships, such as those among various prompt types, within each module.
    
    \item \nameref{main:extra} (Section \ref{main:extra}): We explore the interactions between dimensions across various modules, such as the relationships between causal scenarios and models, or causal scenarios and prompts.

    \item \nameref{main:complexity} (Section \ref{main:complexity}): We define four factors (i.e., number of nodes, number of edges, authenticity, and causal reasoning process) that influence the complexity in the Mathematical mode datasets. And we disclose the factors that essentially affect the model performance in Mathematical mode questions, highlighting fundamental shortcomings in the causal reasoning abilities of current models.

    \item \nameref{main:maturity} (Section \ref{main:maturity}): We measure the maturity of a causal scenario. Our motivation to analyze the maturity of a causal scenario stems from the desire to explore the research potential of a causal scenario. That is, the relative immaturity of a scenario might suggest a larger room for research improvement.

    \item \nameref{main:stability} (Section \ref{main:stability}): We assess the volatility of the model and the prompt separately to understand their stability under different conditions.

    \item \nameref{main:error} (Section \ref{main:error}): Lastly, we analyze the models' errors from both quantitative and qualitative perspectives to identify areas for improvement.
\end{itemize}

\subsubsection{Comparative Analysis of Models}
\label{main:comparison}
Given the rapid advancement of language models, a key goal of our evaluation is to foster a common and unified understanding of the causal reasoning capabilities of the currently available language models. Therefore, in this section, we aim to conduct a direct comparison of the performance of various models from multiple perspectives. This comparative analysis will enable us to discern differences and similarities in how these models handle causal reasoning tasks, providing insights into their strengths and weaknesses. Such insights will enhance our ability to assess the effectiveness of each model and potentially guide future developments in model training and application.

\paragraph{Comparative analysis of models under different modes.}
\begin{figure}[t]
    \centering
    \includegraphics[width=\textwidth]{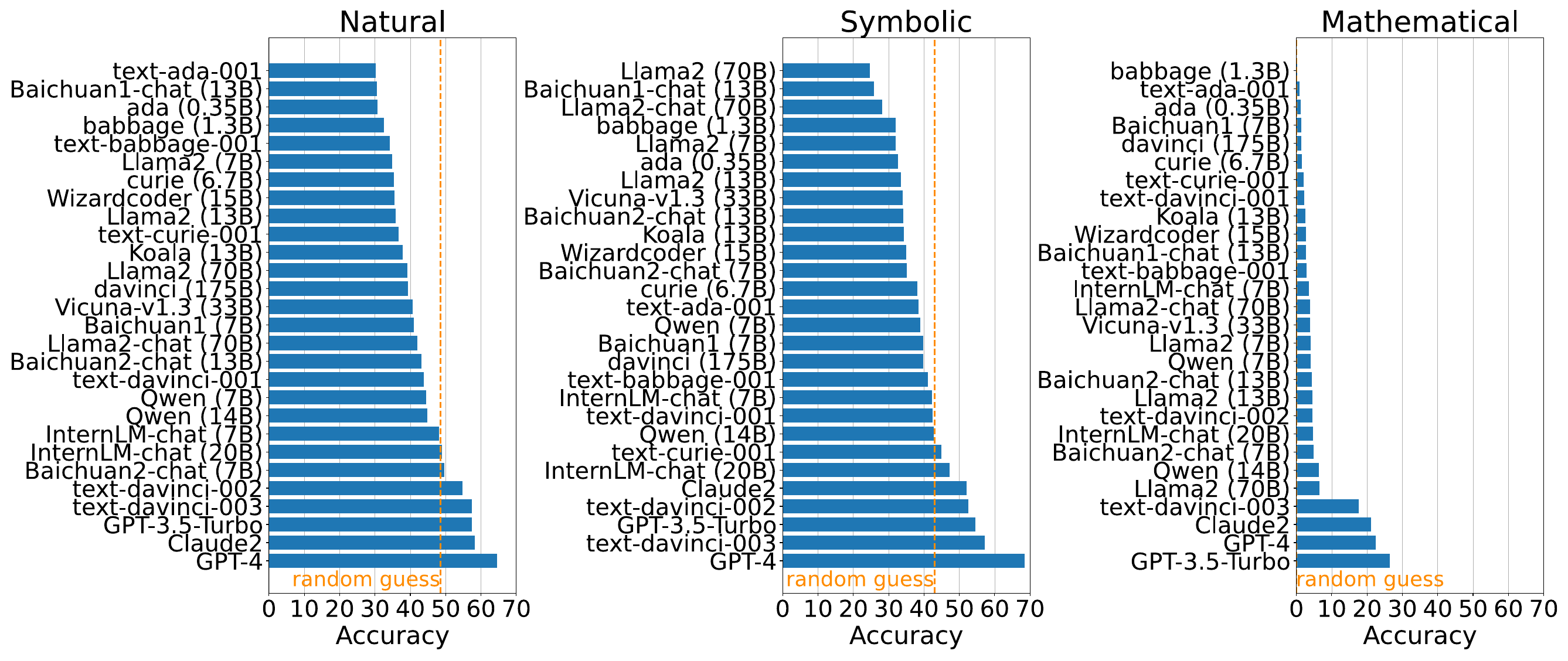}
    \caption[Comparative analysis under different modes]{\textbf{Comparative analysis of models under different modes.} We report the absolute accuracy comparisons. The orange dashed line represents the accuracy of random guess.}
    \label{fig_main:direct_mode}
\end{figure}
Comparing the performance of models directly under different modes helps us understand what types of problems the models are better at solving. As shown in Figure \ref{fig_main:direct_mode}, the following insights can be concluded: 
(1) When considering their performance in exceeding random guess, it is noted that in both Natural and Symbolic modes, these models have employed instruction-tuning. In all three modes, the top three models have introduced the use of human feedback. This perspective indicates that instruction-tuning and the use of human feedback are effective means to enhance model performance across all three modes. (2) Shifting attention to the rankings of these top 3 models across different modes, it is observed that there is not a consistent and stable ranking. Additionally, the relative rankings of \gptf, \claude, and \chatgpt~shift depend on the mode. This observation reinforces the findings presented in Section \ref{main:centric}, which are based on the data illustrated in \ref{fig_main:central_mode}. (3) \llamaseventy~shows a distinct difference in its performance rankings between Symbolic and Mathematical modes. Investigation revealed that for Symbolic, \llamaseventy~fails to provide effective responses to either basic prompt or \mcot~in CEI, leading to extensive blank outputs. This accounts for its lower ranking in the Symbolic mode. On the other hand, in Mathematical mode, it manages to follow \mcot~guidelines to generate appropriate answers, which boosts its overall performance.
(4) It is evident that the models have a significant shortcoming in performing Mathematical operations. None of the models have an average accuracy exceeding 30\%, which contrasts with their performance in Natural and Symbolic causal tasks, where the best-performing \gptf~ has an accuracy of over 60\%. 

\begin{figure}[t]
    \centering
    \includegraphics[width=0.7\textwidth]{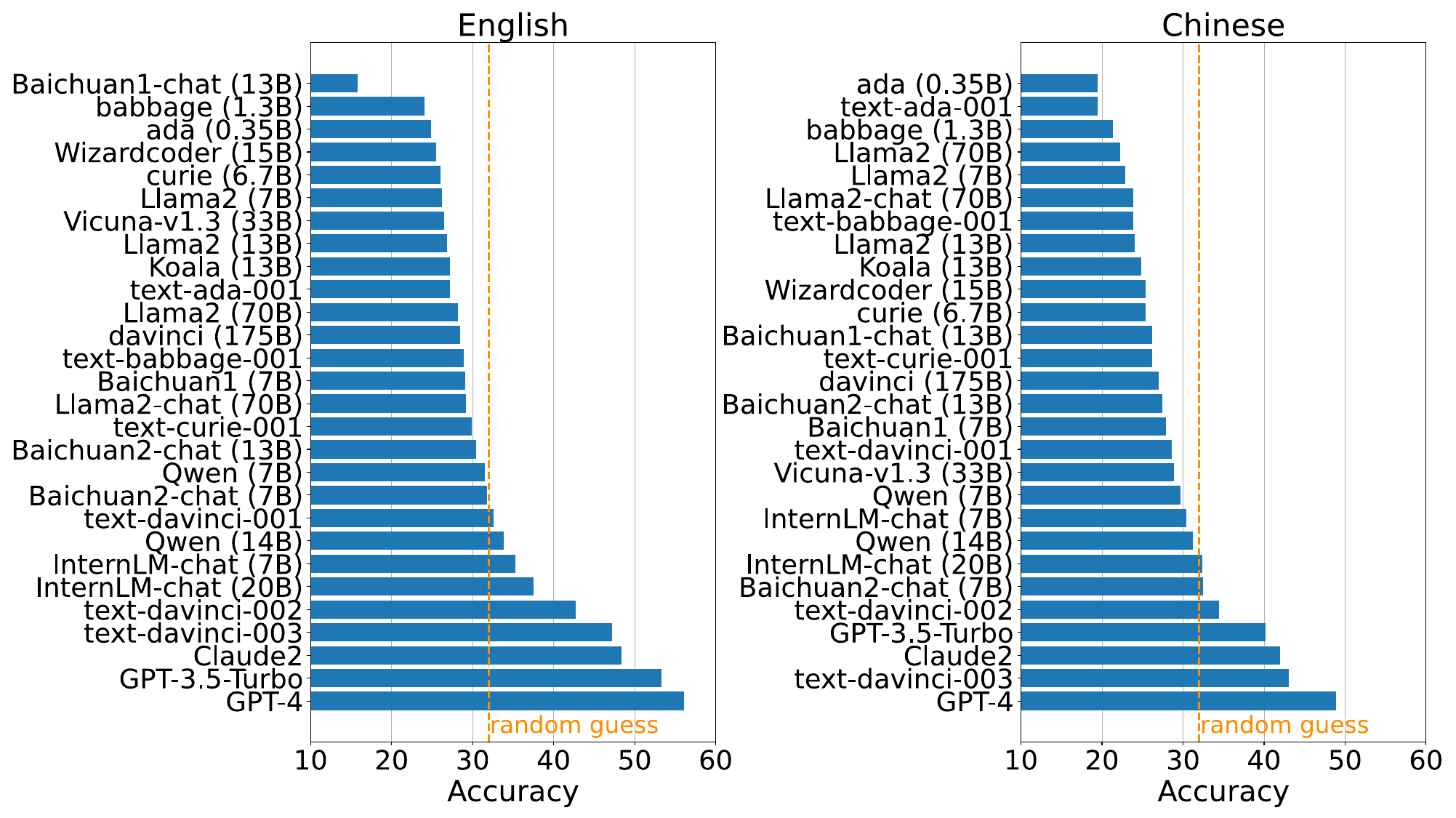}
    \caption[Comparative analysis under multilingual]{\textbf{Comparative analysis of models under different languages.} We report the absolute accuracy comparisons. The orange dashed line represents the accuracy of random guess.}
    \label{fig_main:direct_language}
\end{figure}

\paragraph{Comparative analysis of models under different languages.}
The widespread global use of language models necessitates an evaluation of their performance in multilingual contexts. This analysis can guide the development of targeted corpora for future training and provide valuable references for users across different linguistic environments. As shown in Figure \ref{fig_main:direct_language}, it is evident that models generally perform better in English causal tasks. This finding aligns with our initial hypothesis, given that the training corpus for most current language models predominantly consists of English data. Specifically, the number of models performing better than random guesses in English causal tasks is nine, compared to seven in Chinese causal tasks.
Moreover, \gptf~ emerges as the top performer consistently across both linguistic environments.
In English causal tasks, both \gptf~and \chatgpt, ranked first and second respectively, achieve an average accuracy exceeding 50\%. Conversely, in Chinese causal tasks, even \gptf's performance does not surpass the 50\% threshold. Despite these variations in accuracy across languages, the top five models remain consistent, albeit with minor differences in their specific rankings. These models are \gptf, \chatgpt, \textdavincithree, \textdavincitwo~from OpenAI, and \claude~from Anthropic. 

\begin{figure}[t]
    \centering
    \includegraphics[width=1\textwidth]{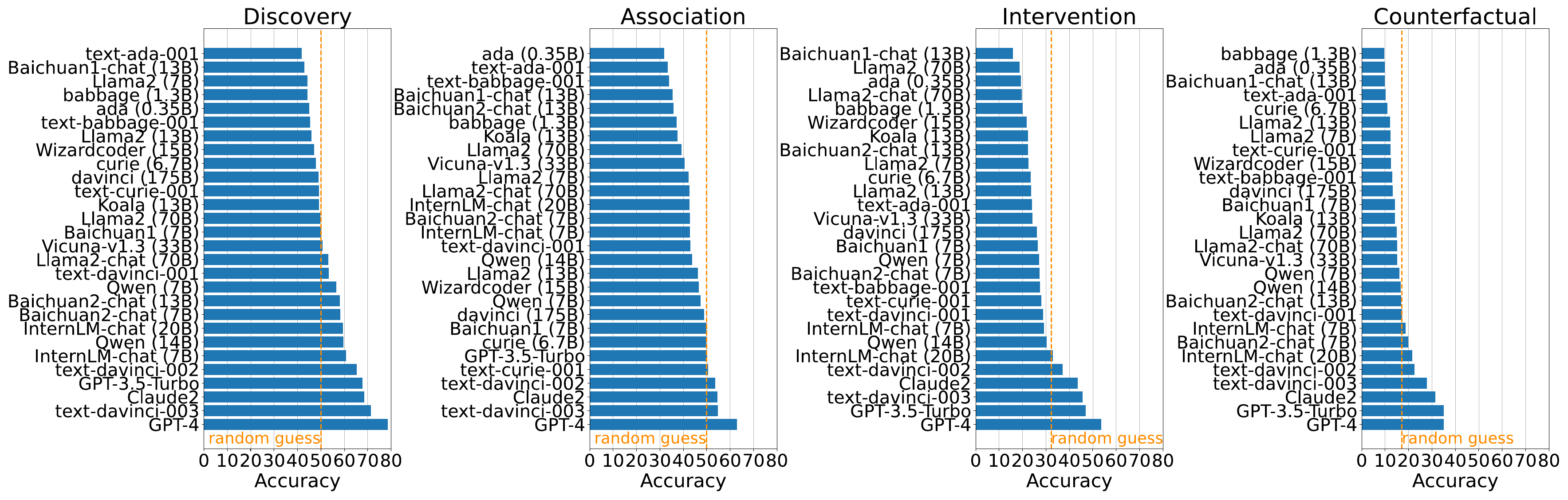}
    \caption[Comparative analysis of models under different rungs of causal ladder]{\textbf{Comparative analysis of models under different rungs of causal ladder.} We report the absolute accuracy comparisons. The orange dashed line represents the accuracy of random guess.}
    \label{fig_main:direct_ladder}
\end{figure}

\paragraph{Comparative analysis of models under different rungs of the causal ladder.} 
We investigate models' performance across various rungs of the causal ladder, providing a thorough, bottom-up understanding of their abilities in different levels of causal reasoning tasks.
We can see from Figure \ref{fig_main:direct_ladder} that models exhibit superior performance at the lower levels (discovery and association) compared to the higher levels (intervention and counterfactuals). Specifically, models show the strongest performance at the discovery level and the weakest at the counterfactuals level. Notably, over half of the models (15 out of 28) achieve accuracy rates that surpass random guessing at the discovery level, a trend visually represented by the blue columns in the chart. We hypothesize that the improved performance of the model on causal scenarios within the causal discovery rung may be attributed to its extensive access to world knowledge. It is widely recognized in prior research \citep{vashishtha2023causal,ban2023query} that language models excel in causal discovery due to their extensive access to world knowledge, which enables them to identify causal relations embedded within semantic content. This capability alleviates the need for explicit numerical computation, which is more crucial at the higher levels. And for the comparatively good performance on the association rung, we hypothesize that these results arise primarily because association concentrates on the statistical relationships among random variables. That is, causal scenarios within association rung do not require the model to have an extensive capability for causal reasoning. 

Furthermore, analysis of model performance reveals that: (1) Limited-access models generally exhibit better performance, as evidenced by that across all levels of the causal ladder, the top 5 performing models are consistently those with limited public access. (2) \gptf~ consistently ranks first across all levels. At the discovery and association levels, \textdavincithree~outperforms \chatgpt. However, this trend reverses at the intervention and counterfactuals levels. This shift may hint at the beneficial impact of RLHF in enhancing model performance in more complex causal reasoning tasks.

\begin{figure}[t]
    \centering
    \includegraphics[width=0.9\textwidth]{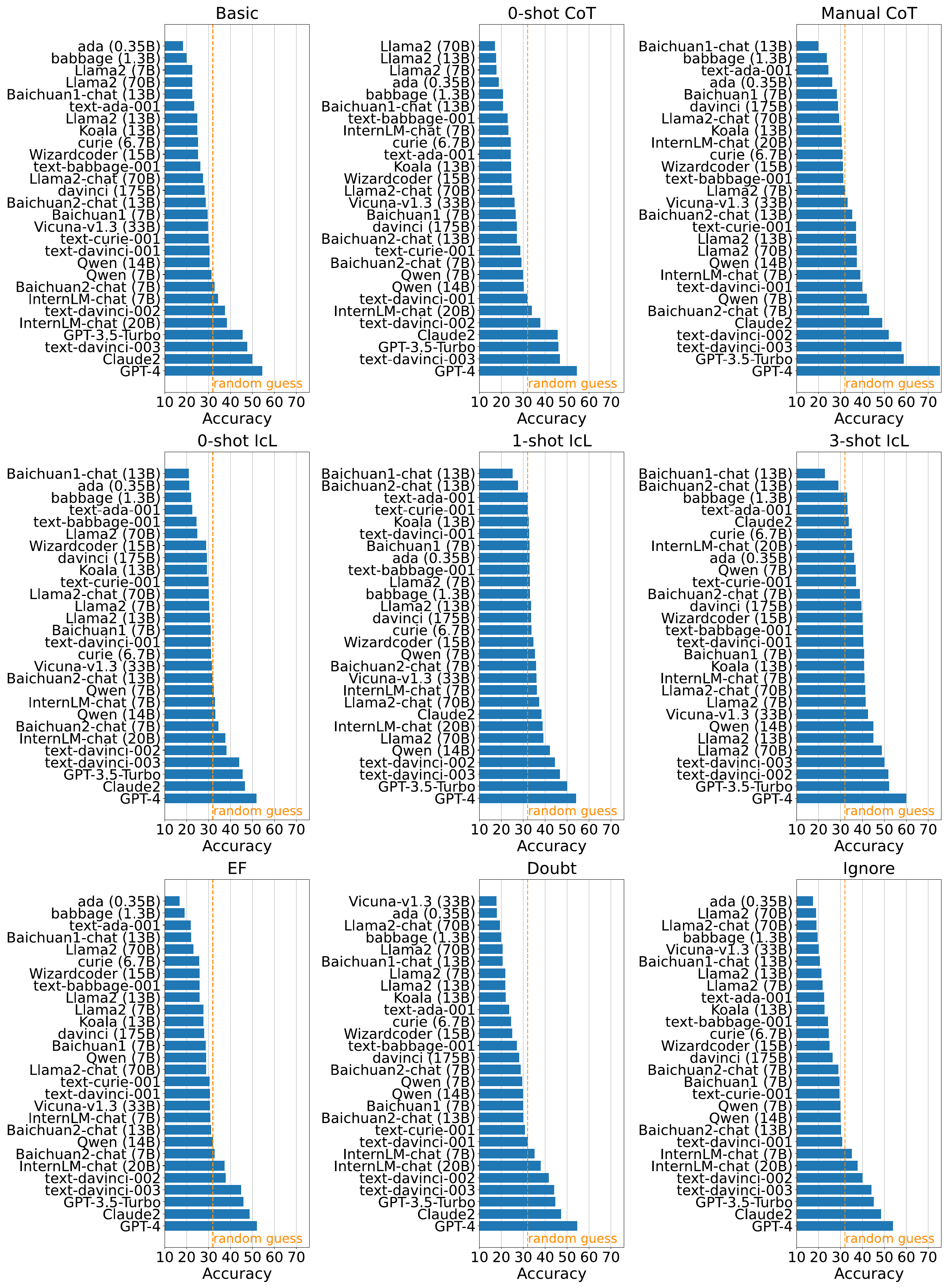}
    \caption[Comparative analysis under different prompts]{\textbf{Comparative analysis of models under different prompts.} We report the absolute accuracy comparisons. The orange dashed line represents the accuracy of random guess.}
    \label{fig_main:direct_prompt}
\end{figure}

\paragraph{Comparative analysis of models under different prompts.}
Considering the critical role of prompts in influencing model performance, we begin our analysis by evaluating how all models perform across various causal scenarios when subjected to different prompts. As depicted in Figure \ref{fig_main:direct_prompt}, we directly compare the performance of various models under different prompts. Key observations include: 
(1) \gptf~excels in seven out of the eight prompts, with the sole exception of 1-shot IcL. \chatgpt~stands out in the 1-shot IcL prompt, securing the top spot, and consistently ranks second in three other prompts (0-shot/\mcot, \ticl). \claude~also shows noteworthy performance, securing the second position in five of the prompts. However, it falls behind significantly in the 1/\ticl~and CoT, particularly ranking as low as 24th in \ticl. 
(2) The top three models are not consistent across all prompts, indicating variability in their accuracy depending on the specific prompt used. This observation aligns with findings from Figure \ref{fig_main:central_prompt} regarding \textbf{Prompt-centric relationships}.
(3) The \ticl~prompt appears to set a lower performance limit for models, while the \mcot~maximizes the upper limit for top-ranked models. Notably, \gptf~using \mcot~is the only model achieving over 70\% accuracy, but its performance drops below 60\% with \ticl. In contrast, \chatgpt, which ranks second under both prompts, shows an accuracy of 59.8\% with \mcot, surpassing its performance with \ticl~by 7.2\%. Additionally, the number of models reaching or exceeding 50\% accuracy under \mcot~is four, compared to three under \ticl.

\clearpage

\subsubsection{Impact of Other Factors on Accuracy}
\label{main:acc}
\paragraph{Impact of model access on accuracy.}
\begin{figure}[t]
    \centering
    \includegraphics[width=\textwidth]{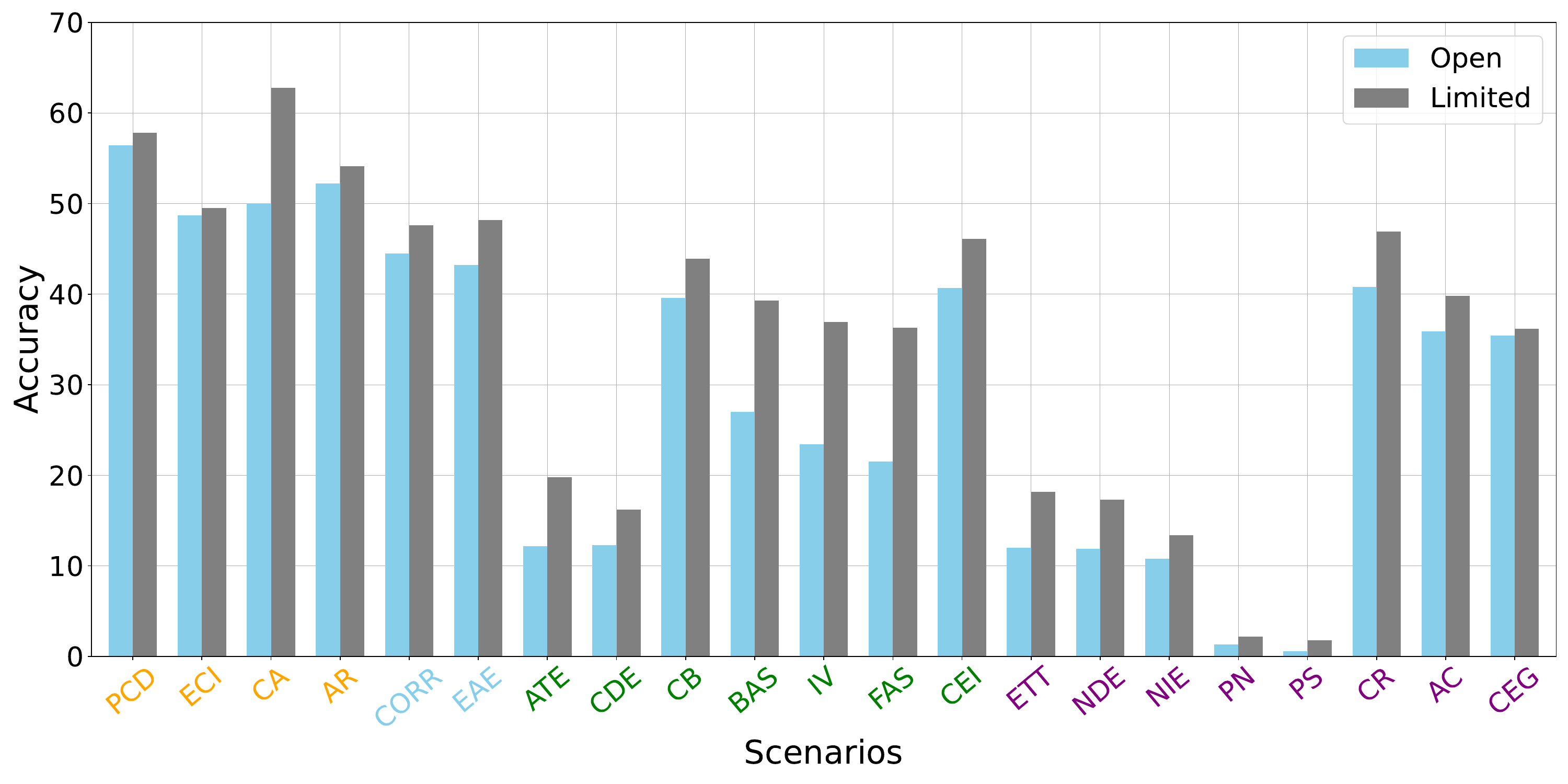}
    \caption[Impact of model access on accuracy]{\textbf{Impact of model access on accuracy.} The impact of model access (\textbf{Open} and \textbf{Limited}) on the accuracy in each of the 21 causal scenarios.}
    \label{fig_main:acc_access}
\end{figure}
We explore the relationship between accuracy and model accessibility in Figure \ref{fig_main:acc_access}. For each access type, we calculate the average accuracy of all models in the corresponding causal scenario. Overall, we observe a consistent pattern: models with limited access tend to outperform their open-access counterparts across all causal scenarios. 

Nonetheless, we ought to maintain an optimistic perspective regarding open-access models. It is evident that in causal scenarios such as PCD, ECI, AR, and CEG, the disparity between the two access types has been narrowed to less than 2\%. We exclude PN and PS from this comparison due to their universally low accuracy, which renders any difference statistically insignificant. It is worth noting that PCD, ECI, and AR all belong to the causal discovery level, demonstrating that open-access models are capable of competently understanding causal relationships.
The tasks within these scenarios vary: PCD and ECI focus on discerning causality among events, whereas AR involves analysis based on a given causal graph. This variation underscores the adaptable nature of open-access models in handling causal discovery tasks. The observed deficiencies in other causal scenarios suggest avenues for enhancement. Methods such as expanding training datasets or refining training techniques could potentially elevate the performance of open-access models to close the gap with limited-access models even further.

\begin{figure}[H]
\centering
\subfigure[Performance of grouped models]{
\centering
\includegraphics[width=.85\linewidth]{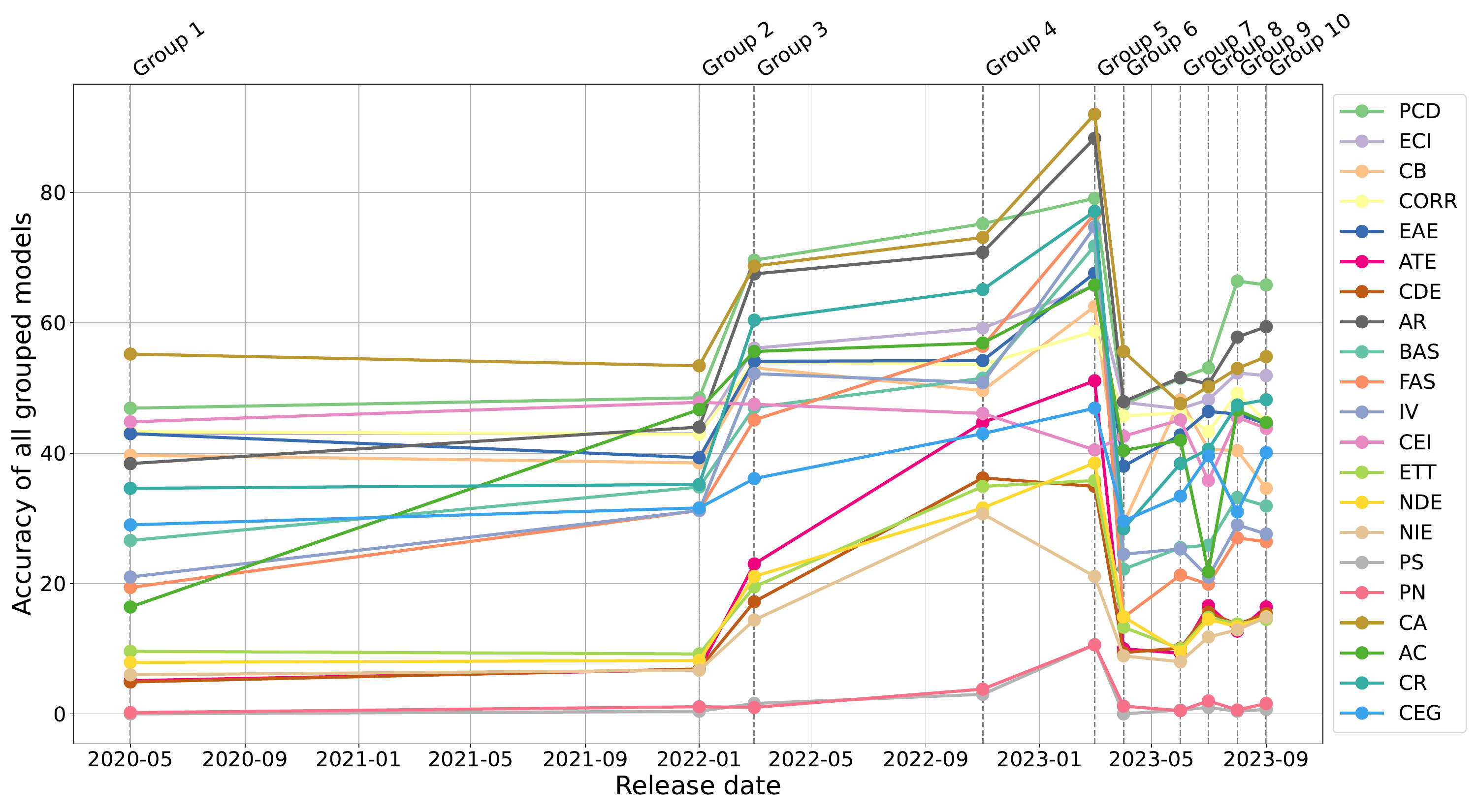}
\label{fig_main:acc_time_sub_grouped}
}
\subfigure[Performance of selected models]{
\centering
\includegraphics[width=.85\linewidth]{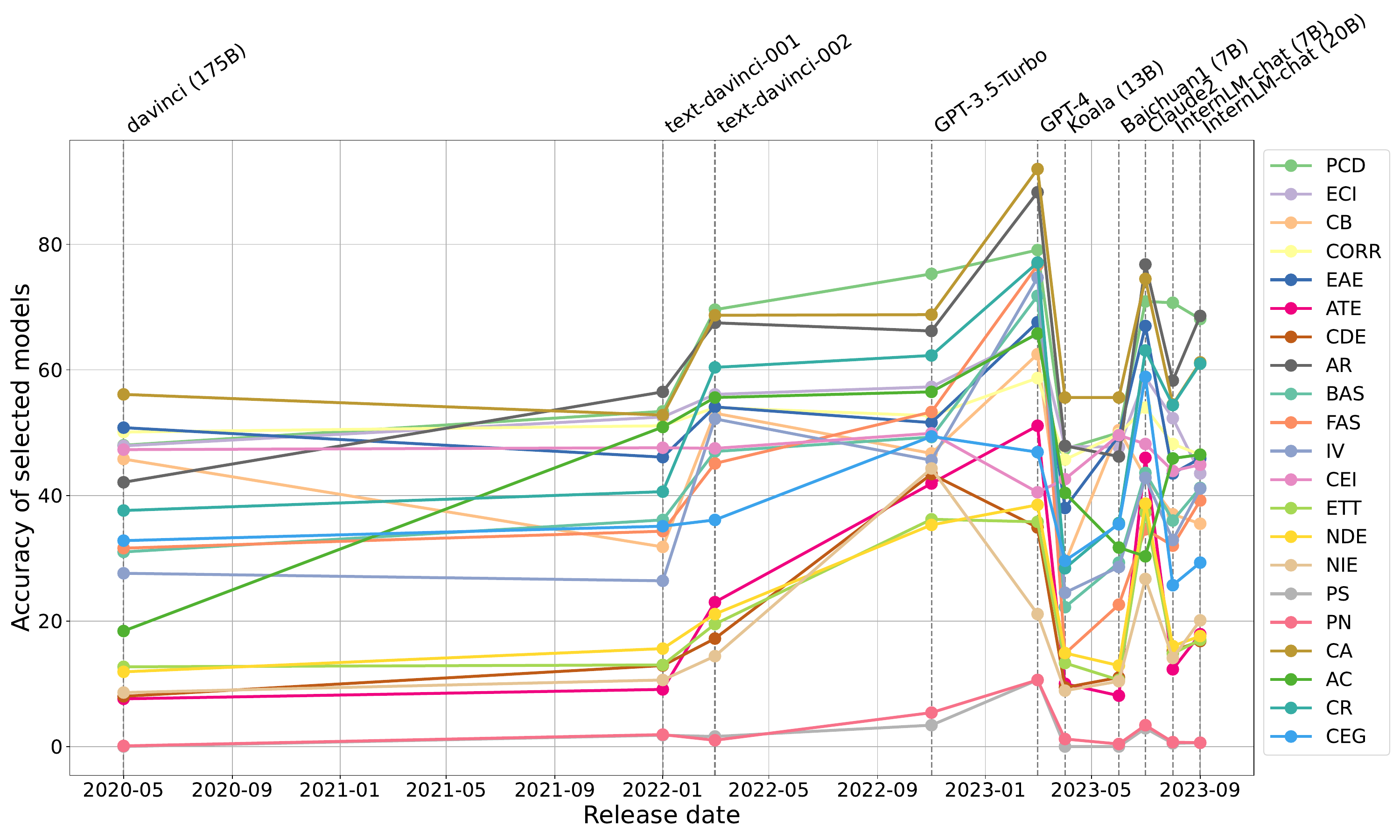}
\label{fig_main:acc_time_sub_selected}
}
\caption[Impact of time on accuracy]{\textbf{Impact of time on accuracy.} The interplay of time (x-axis) with all models' accuracy (y-axis), examined across 21 causal scenarios. Accuracy represents the average value for all models in the corresponding group. Each group consists of models released in the same year and month, and we will detail the grouping as follows. \textbf{Group 1}: ada (0.35B), babbage (1.3B), curie (6.7B), davinci (175B). \textbf{Group 2}: text-ada-001, text-babbage-001, text-curie-001, text-davinci-001. \textbf{Group 3}: text-davinci-002. \textbf{Group 4}: text-davinci-003, GPT-3.5-Turbo. \textbf{Group 5}: GPT-4. \textbf{Group 6}: Koala (13B). \textbf{Group 7}: Wizardcoder (15B), Vicuna-v1.3 (33B), Baichuan1 (7B). \textbf{Group 8}: Llama2 (7B), Llama2 (13B), Llama2 (70B), Llama2-chat (70B), Baichuan1-chat (13B), Claude2. \textbf{Group 9}: Qwen (7B), InternLM-chat (7B). \textbf{Group 10}: Baichuan2-chat (7B), Baichuan2-chat (13B), Qwen (14B), InternLM-chat (20B).}
    \label{fig_main:acc_time_all}
\end{figure}

\paragraph{Impact of time on accuracy.}
Understanding how model performance evolves over time is crucial for developing a deeper macro-level understanding of technologies. We explore this dynamic in Figure \ref{fig_main:acc_time_all}, where models are categorized into 10 groups spanning from May 2020 to September 2023, based on their release dates. The details of these groupings are clarified in Figure \ref{fig_main:acc_time_all}. Figure \ref{fig_main:acc_time_sub_grouped} presents the average performance of all models within each group, while Figure \ref{fig_main:acc_time_sub_selected} focuses on the accuracy of the highest-performing model in each group across all causal scenarios. All models from Group 1$\sim$5 are developed by OpenAI.

Above all, we begin by examining the consistent insights presented across both figures. From the model perspective, several key aspects emerge as particularly significant: (1) Starting with InstructGPT (Group 2) and culminating with GPT-4 (Group 5), each successive release from OpenAI's model series marks a clear improvement in performance. This trend is consistent with findings from \citet{fu2022gptroadmap}, supporting the effectiveness of the technological advancements made by OpenAI. (2) GPT-4 maintains a significant advantage in approximately 80\% of the causal scenarios, outperforming both its predecessors and subsequent models. (3) Models in Groups 5 to 10 show diverse performance levels, dependent on the specific causal scenarios they address. This indicates an absence of a consistent improvement trend and underscores the selective effectiveness of these models. 

From the perspective of causal scenario, the performance of models in the CEI scenario shows no correlation with their release dates, aligning with our interpretation in \textbf{Causal scenario-centric relationships}.
Contrary to a progressive improvement expected over time, the performance fluctuates - sometimes peaking, sometimes diminishing - highlighting the unique challenges posed by this scenario. 

Moving forward, we will focus on the most pronounced difference between the two figures. Specifically, by narrowing our analysis to selected models as depicted in Figure \ref{fig_main:acc_time_sub_selected}, it becomes apparent that Claude2 emerges as a formidable competitor, distinctly achieving a ``localized peak''. This observation suggests that Claude2 may possess unique attributes or optimizations that enable it to excel in certain scenarios, standing out among its contemporaries.

\begin{figure}[t]
    \centering
    \includegraphics[width=0.9\textwidth]{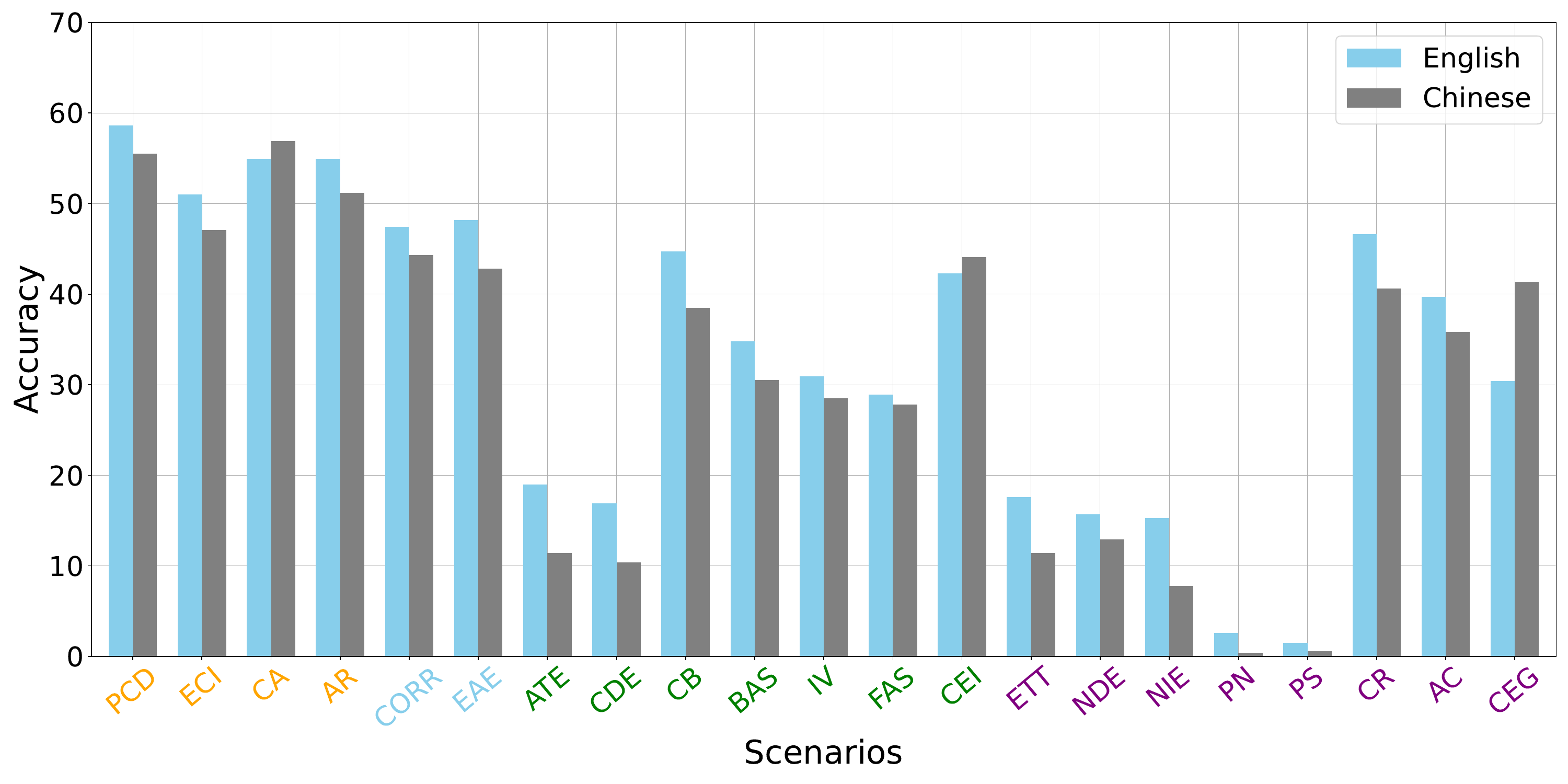}
    \caption[Impact of multilingual on accuracy]{\textbf{Impact of multilingual on accuracy.}  The impact of language (\textbf{English} and \textbf{Chinese}) on the accuracy in each of the 21 causal scenarios.}
    \label{fig_main:acc_multilingual}
\end{figure}

\paragraph{Impact of multilingual capabilities on accuracy.}
We investigate the relationship between model accuracy and multilingual capabilities in Figure \ref{fig_main:acc_multilingual}. For each language type, we calculate the average accuracy of all models across each causal scenario. Consistently across most scenarios, except for CA, CEI, and CEG, models perform better in English causal tasks than in Chinese. This is expected, as noted in Section \ref{main:model}, given that English predominantly comprises the training corpus for most models. This performance gap highlights the critical need for developing richer and more diverse language corpora to enhance model proficiency in multilingual contexts significantly.

In scenarios where models exhibit superior performance in Chinese, we explore potential explanatory factors. For instance, in the CEI scenario, Llama-based models (including Koala, Llama2-13b, Llama2-70b, Llama2-70b-chat, Vicuna-33b) tend to be more effective in Chinese tasks. When using the basic prompt, these models typically produce very concise replies in Chinese, whereas in English, their responses are often less effective. While these observations may not fully capture the complexities of language performance differences, they provide insights into possible underlying factors contributing to these discrepancies.

\begin{figure}[t]
    \centering
\includegraphics[width=0.65\linewidth]{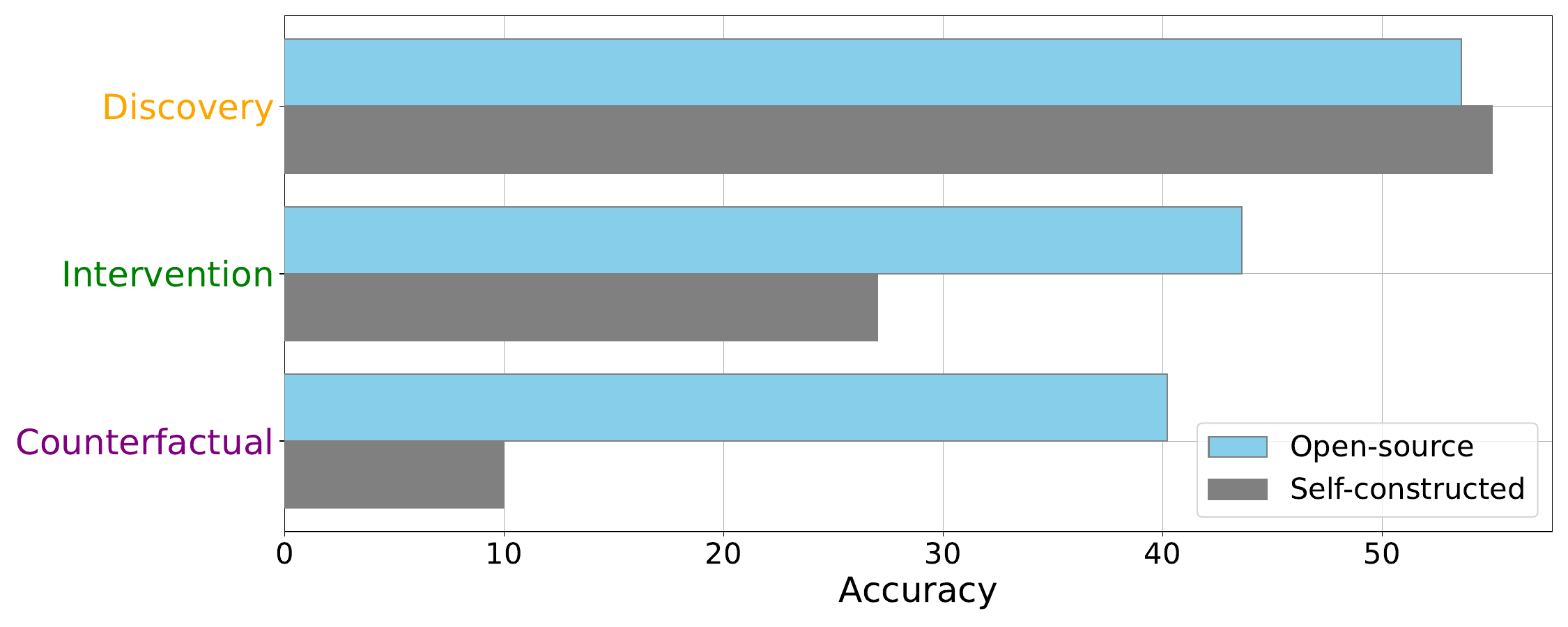}
    \caption[Impact of domain on accuracy]{\textbf{Impact of domain on accuracy.} We calculate the average accuracy of all models at the same rung but in different domains (i.e., open-source or self-constructed).}
    \label{fig_main:extra_seen_unseen}
\end{figure}

\paragraph{Impact of domain on accuracy.}
As mentioned in \nameref{data:selection}, our dataset primarily consists of two parts: open-source and self-constructed. This differentiation is critical for assessing the potential issue of \textit{training-test contamination}, as discussed by \citet{liang2022holistic}. Additionally, this analysis is instrumental in providing insights into the development of training sets for models and enhancing their generalization capabilities across various domains. We explore the relationship between model accuracy and domain across different rungs\footnote{Since all datasets used in \emph{Association (Rung 1)} are open-source and thus lack self-constructed datasets for comparison, we do not take this rung into consideration.} of the causal ladder in Figure \ref{fig_main:extra_seen_unseen}. Note that, a detailed classification of what constitutes the open-source and self-constructed datasets at each rung is presented in \cref{table_models_selection}. 

From Figure \ref{fig_main:extra_seen_unseen}, we discover that the relationship between accuracy and domain varies across the rungs of the causal ladder. For simpler tasks associated with causal discovery\footnote{We have demonstrated in Figure \ref{fig_main:direct_ladder} that models show better performance at the foundational stages (i.e., causal discovery and association) than at the more complex stages (i.e., intervention and counterfactuals).}, models exhibit marginally better performance on self-constructed datasets than on open-source ones. In contrast, for the more challenging tasks at the intervention and counterfactuals rungs, models generally perform better on open-source datasets. Given that certain datasets were released earlier (e.g., CRASS \citep{frohberg2022crass} was issued in 2022), it is possible that they have been used as training corpus for some models. This might account for the superior performance of models on open-source datasets. This trend indicates that while current language models are relatively proficient in extracting and utilizing causal relationships from natural language expressions commonly found in broader datasets, their capabilities diminish as the complexity of causal reasoning increases.

\clearpage

\subsubsection{Predicting Causal Reasoning Ability}
\label{main:predict}
Numerous studies have already established a link between the scale of models and their loss, underscoring the influence of model size on performance metrics \citep{kaplan2020scaling,hoffmann2022an}. In our analysis, we use accuracy as a measure to assess the effectiveness of causal reasoning abilities of language models. However, it is important to clarify that while accuracy is a critical metric for our evaluation, it does not fully encapsulate the breadth of causal reasoning capabilities of these models. Moving forward, we aim to explore the potential for more precise criteria for evaluation.

\begin{figure}[H]
\centering
\subfigure[Performance of grouped models]{
\centering
\includegraphics[width=.85\linewidth]{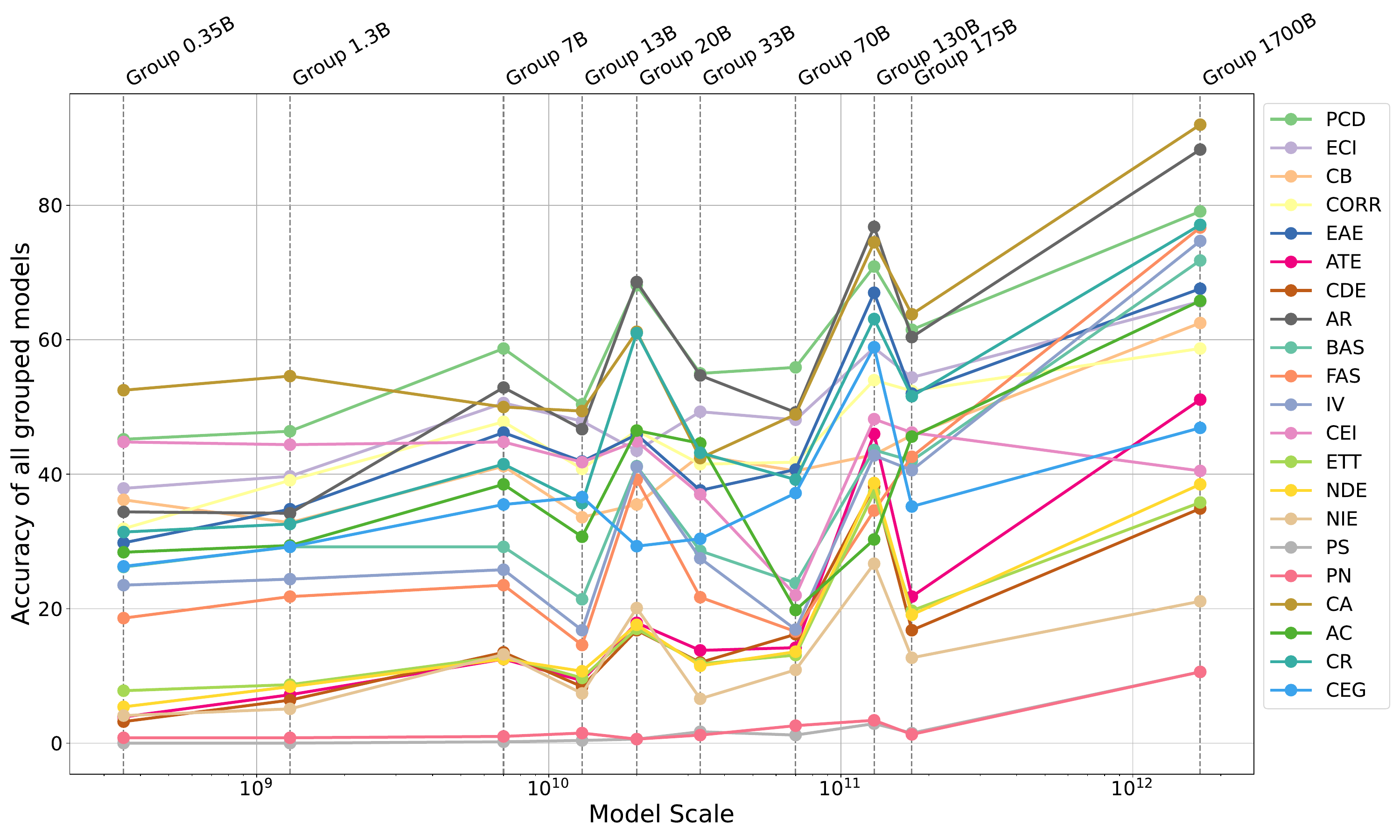}
\label{fig_main:acc_scale_sub_grouped}
}
\subfigure[Performance of selected models]{
\centering
\includegraphics[width=.85\linewidth]{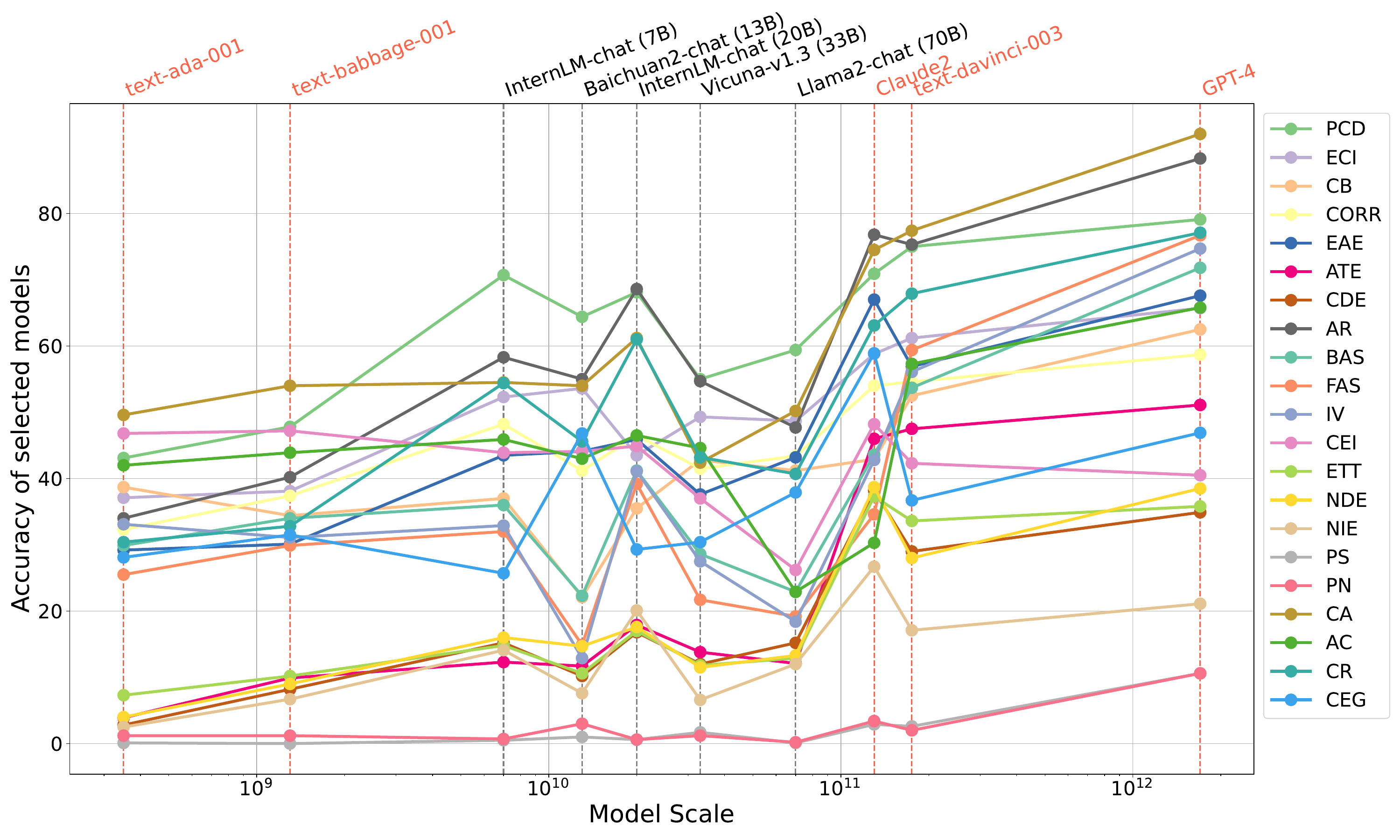}
\label{fig_main:acc_scale_sub_selected}
}
\caption[Causal reasoning ability vs. scale]{\textbf{Causal reasoning ability vs. scale.} Accuracy represents the average value for all models in the corresponding group. Each group consists of models of the scale, and we will detail the grouping as follows. \textbf{Group 0.35B}: ada (0.35B), text-ada-001. \textbf{Group 1.3B}: babbage, text-babbage-001. \textbf{Group 7B}: Baichuan1 (7B), Baichuan2-chat (7B), Qwen (7B),  InternLM-chat (7B), Llama2 (7B). \textbf{Group 13B}: Baichuan1-chat (13B), Baichuan2-chat (13B), Llama2 (13B), Koala (13B). \textbf{Group 20B}: InternLM-chat (20B). \textbf{Group 33B}: Vicuna-v1.3 (33B). \textbf{Group 70B}: Llama2 (70B), Llama2-chat (70B). \textbf{Group 130B}: Claude2. \textbf{Group 175B}: davinci (175B), text-davinci-001, text-davinci-002, text-davinci-003. \textbf{Group 1700B}: GPT-4. The red text and dashed line indicate that the scale of the model is undisclosed.}
    \label{fig_main:acc_scale_all}
\end{figure}

\paragraph{Causal reasoning ability vs. scale.}
We investigate the relationship between model accuracy and scale, as depicted in Figure \ref{fig_main:acc_scale_all}. 
We categorize the models into ten groups based on their scales, ranging from 0.35B to 1700B parameters. The top half of Figure \ref{fig_main:acc_scale_all} illustrates the mean performance of all models within each group across various causal scenarios. The bottom half presents the top-performing model selected from each group. Our analysis does not include all models for two primary reasons: (1) Due to the close similarity in scale among certain models, we have made selective choices to manage the impact of models across different scales and to maintain the visual clarity of our presentations. For instance, we choose the 7B model over others ranging between 6.7B and 7B, and select the 13B model over those ranging from 13B to 15B. (2) For some models, it is challenging to obtain an authoritative source for their scale, precluding us from making informed estimations. An example is \chatgpt, for which the exact scale has not been determined.

Beginning with the grouped models, we derive several key insights: (1) There is no consistent pattern of increasing accuracy with larger model scales, particularly among models from Group 13B to Group 70B. This fluctuation suggests that performance should be analyzed within specific causal scenarios. The variability in this scale range could stem from differences in model origins, with disparate creators applying different training methodologies and datasets, which significantly impact outcomes beyond mere scale. 
(2) When examining models created by the same creator, the relationship between accuracy and scale becomes more apparent. For instance, models from OpenAI in Group 0.35B, 1.3B, 175B, and 1700B\footnote{OpenAI has not made the parameter size of \gptf~public. To analyze the results of \gptf, we refer to the following source: {\footnotesize \url{https://en.wikipedia.org/wiki/GPT-4}}.} exhibit a correlation where, generally, larger scales correspond with higher accuracy, except in the CEI and CB scenarios. 
(3) Notable performance increases are observed within certain scale ranges, particularly between Group 13B and Group 20B, and from Group 70B to Group 130B. Specifically, when looking at Group 20B, this group, containing only one model (i.e., \internt), shows a performance peak that may reflect the effectiveness of specific training strategies and optimal parameter selection. From the perspective of Group 130B - the first group with over 100B parameters - there is a marked performance leap over Group 70B. This implies that a significant increase in model parameters may enhance model capabilities. Yet, the progression from Group 130B to Group 175B complicates this view. The models in Group 175B, which span a broad range of release dates from \davinci~in May 2020 to \textdavincithree~in November 2022, showcase the potential impacts of evolving technology on model capabilities. Conversely, the models between Group 70B and 130B, including those in the Llama2 series and \claude\footnote{Anthropic has not made the parameter size of \claude~public. To analyze the results of \claude, we refer to the following sources: {\footnotesize (1) \url{https://datasciencedojo.com/blog/introducing-claude}, (2) \url{https://codingscape.com/blog/most-powerful-llms-large-language-models-in-2023}}.}, were all released within a tight timeframe in July 2023. This closer release window offers a more controlled examination of the scale effect, minimizing the variable of technological progression. This distinction provides valuable insights into how timing and technological advances, in addition to scale, can influence model performance across various causal inference tasks.

Analyzing the selected models yields the following insights: (1) For models with scales ranging from 7B to 70B, the relationship between scale and accuracy is still not definitive. This observation suggests that merely increasing the size of a model may not be the most effective strategy for enhancing its causal reasoning capabilities. It appears that other factors, such as training methods and datasets, may have a more significant impact than scale alone. (2) Despite the ambiguity in broader datasets, a clear trend is still observed within the models developed by OpenAI: as model scale increases, so does accuracy. This pattern within OpenAI's series suggests that their approach to scaling - possibly coupled with their specific training techniques and data handling - effectively boosts model performance. 

\begin{figure}[t]
    \centering
    \includegraphics[width=.9\textwidth]{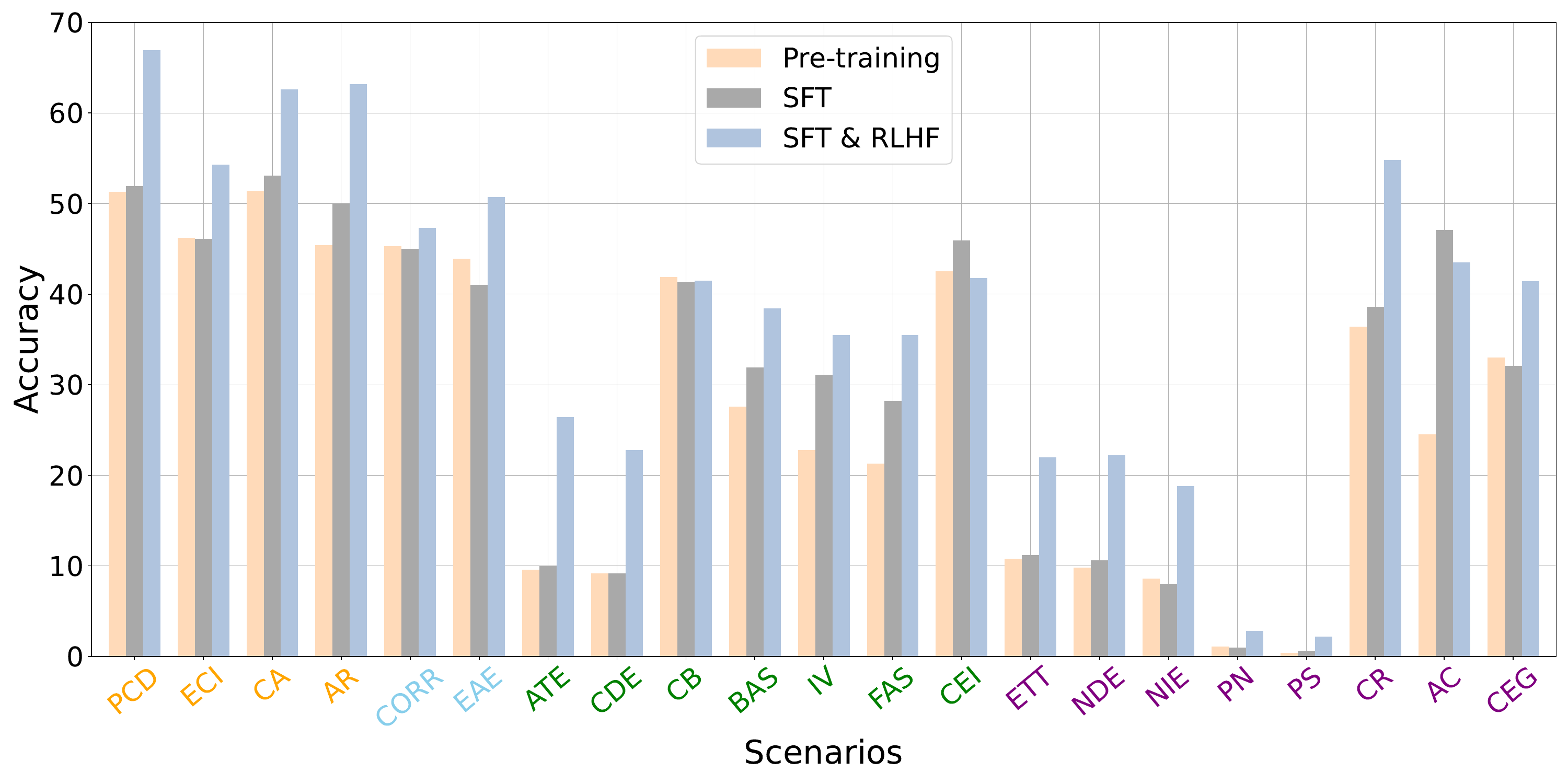}
    \caption[Causal reasoning ability vs. training strategy]{\textbf{Causal reasoning ability vs. training strategy.} We categorize models into three groups based on different training strategies (i.e., pre-training, SFT, and SFT\&RLHF). To investigate the effectiveness of different training strategies, we compare the average accuracy of the different groups of models across 21 causal scenarios.}
    \label{fig_main:acc_strategy}
\end{figure}

\paragraph{Causal reasoning ability vs. training strategy.}
In Figure \ref{fig_main:acc_strategy}, we explore the impact of different training strategies on the causal reasoning capabilities of models. This analysis categorizes all models based on their distinct training strategies (detailed in Section \ref{main:model}) and measures the average accuracy of each group across various causal scenarios. 

The findings from Figure \ref{fig_main:acc_strategy} reveal significant insights about the efficacy of these strategies: (1) The combination of SFT and RLHF is the most effective strategy, leading to the highest accuracy in 86\% (18 out of 21) of the causal scenarios. This strategy significantly outperforms the models trained using either pre-training or SFT alone. The success of RLHF in enhancing causal reasoning capabilities suggests that integrating human feedback helps to more closely align model outputs with complex human reasoning patterns, particularly in complex causal scenarios requiring causal understanding. (2) The performance gap between models trained via SFT and those undergoing pre-training is relatively small, with SFT models outperforming pre-training models in 13 causal scenarios. This marginal advantage indicates that while SFT has some benefits, the lack of extensive, causal-oriented datasets specifically tailored for SFT might limit its effectiveness. Using general or unrelated domain datasets for SFT seems to provide only a limited boost to the models' causal reasoning abilities, suggesting that more focused and relevant training data could potentially enhance performance further.

\clearpage

\subsubsection{Intra-dimensional Relationships}
\label{main:centric}
\begin{figure}[t]
    \centering
\includegraphics[width=\linewidth]{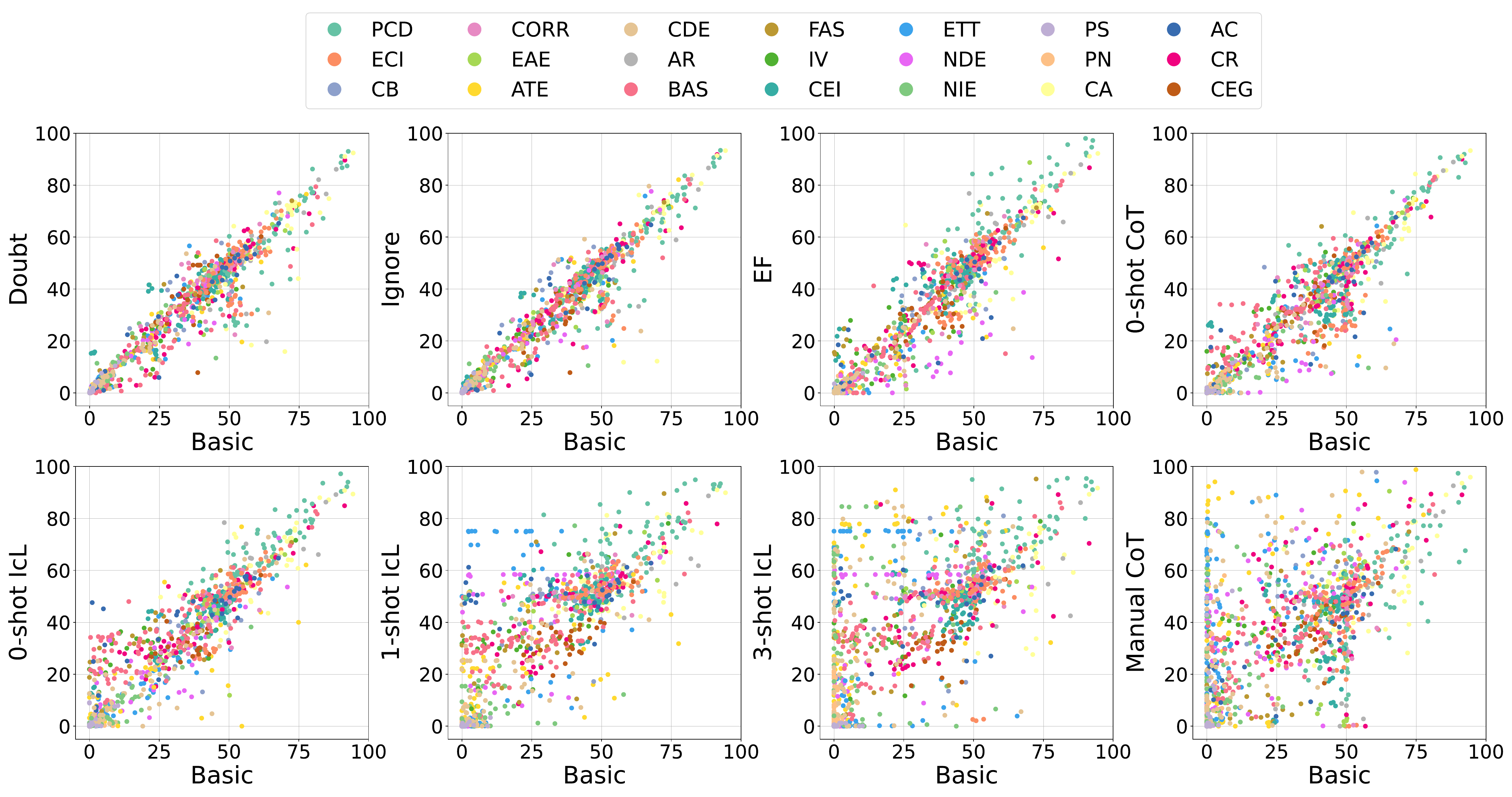}
    \caption[Basic prompt vs. X]{\textbf{Basic prompt vs. other prompts.} We explore the relationship between the basic prompt (x-axis) and eight distinct prompts (doubt, ignore, EF, 0-shot CoT, 0-shot IcL, 1-shot IcL, 3-shot IcL, and manual CoT). We investigate these correlations in all models across all causal scenarios. Each dot symbolizes a pair of accuracies achieved by a specific model when employing a specific pair of prompts, and the dot’s color signifies the related causal scenario.}
\label{fig_main:central_prompt_scatter}
\end{figure}

\paragraph{Prompt-centric relationships.}
To the best of our knowledge, there is currently no work that integratively studies the relationships among different types of prompts. Against the backdrop of the rapid development of prompt engineering, we are eager to explore the relationships between different prompts, hoping to provide insights for the future development and usage of prompts.

We depict the accuracy relationship between the basic prompt and other eight types of prompts in Figure \ref{fig_main:central_prompt_scatter}. Each dot symbolizes a pair of accuracies achieved by a specific model when employing a specific pair of prompts, and the dot's color signifies the related causal scenario. For example, if a model attains 70\% accuracy with the basic prompt and 80\% accuracy with \mcot~ in the PCD scenario, this result would be represented by a dot located at the coordinates (70,80) in the subgraph comparing the basic prompt and \mcot. Note that, the color of the dot corresponds to the PCD scenario. The basic prompt serves as the benchmark for all other prompts, rendering Figure \ref{fig_main:central_prompt_scatter} pivotal for identifying key prompts that improve model performance. Moreover, by incorporating various causal scenarios, this figure shows diverse trends in these relationships. It highlights how the effectiveness of prompts can depend significantly on the causal scenario, exhibit variability, and reveal unusual patterns.

From the perspective of prompts, we have the following findings: (1) We discover that across all causal scenarios, the basic prompt shows a significant correlation with doubt, ignore, EF, 0-shot CoT, and 0-shot IcL (See Figure \ref{fig_main:central_prompt_scatter}). The performance patterns of these prompt pairs remain consistent across various causal scenarios, as evidenced by their corresponding scatter plots approximating a straight line with a unit slope. (2) The relationship between the basic prompt and \oicl~needs to be analyzed within specific causal scenarios. As demonstrated in Figure \ref{fig_main:central_prompt_scatter}, there is no clear trend in their correlation in causal scenarios such as NDE, NIE, and ETT. However, in PCD and ECI, their trend appears to be broadly positive. (3) Generally speaking, we can see from Figure \ref{fig_main:central_prompt_scatter} that there is no strong correlation between the basic prompt with either \ticl~or \mcot. However, in specific causal scenarios (e.g., BAS, AR, and ECI), their scatters exhibit a positive correlation. This highlights the substantial heterogeneity in characteristics across different causal scenarios.

\begin{figure}[H]
    \centering
\includegraphics[width=0.88\linewidth]{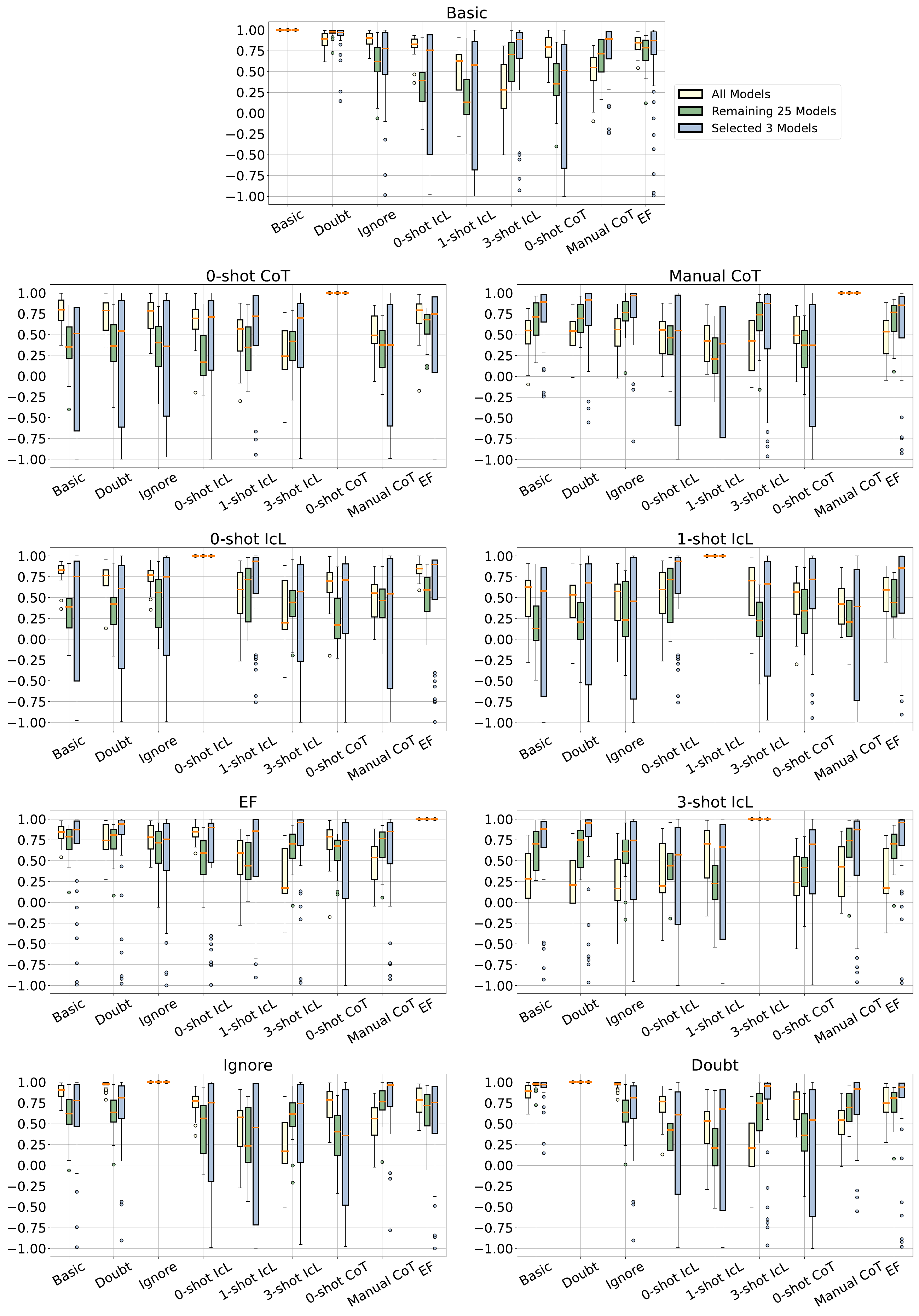}
    \caption[Pearson correlation between prompts]{\textbf{Pearson correlation between prompts.} We compute the Pearson correlation for each prompt with every other prompt. Boxes in various colors denote different groups (i.e., all models, the remaining 25 models, and the selected 3 models). Dots in different colors indicate outliers from those respective groups. The median is represented by an orange line.}
    \label{fig_main:central_prompt}
\end{figure}

In Figure \ref{fig_main:central_prompt}, we examine the Pearson correlations \citep{sedgwick2012pearson} between prompt pairs across all causal scenarios. We begin with calculating the Pearson correlation for each pair of prompts within each scenario. These coefficients are then used to construct the box plots depicted in Figure \ref{fig_main:central_prompt}. For example, if there are ten models under one scenario, prompt A and prompt B each yield ten accuracy values. We first calculate the Pearson correlation for these values. Given multiple causal scenarios, each with its own A-B correlation, we aggregate these into the box plots shown in Figure \ref{fig_main:central_prompt}. 
Specifically, in each subplot, we categorize the models into three groups: \emph{all models}, \emph{the selected 3 models}, and \emph{the remaining 25 models}. In the category of ``all models'', we can present the trend of relationships between prompts from the most macroscopic perspective. Moving attention to the ``selected 3 models'', we select the three well-performing models (\gptf, \chatgpt, and \claude)\footnote{The top 3 models in CaLM are all created by OpenAI. For diversity, we add the \claude~ for analysis instead of \textdavincithree.} in all causal scenarios as representatives. This aims to eliminate some potential interference from invalid data.\footnote{According to our experimental results, we note that the accuracy of some models in specific causal scenarios does not even surpass random guess probability. We consider such data to be invalid, and incorporating it into our analysis might actually skew the conclusions.} Thus, considering only the three well-performing models may better reflect the true relationship between prompts. Finally, we also examine the ``remaining 25 models'', which, in contrast to the selected 3 models, provide insights into variations in prompt pair correlations across models with different performances. Overall, our setup highlights the distribution and heterogeneity of relationships between prompts and facilitates a macroscopic examination of trends across different model categories.

Figure \ref{fig_main:central_prompt} provides rich insights into the relationships between various prompts, revealing several key findings from the analysis: (1) Within all causal scenarios, there is no strong correlation among prompts of the same category when different numbers of examples are provided (e.g., 0/1/3-shot IcL, 0-shot/manual CoT). Specifically, for 0-shot/manual CoT, the median correlation does not exceed 0.5 across all model groups (i.e., ``all models'', ``remaining 25 models'', and ``selected 3 models''). For 0/1-shot IcL, only the ``selected 3 models'' group exhibits a median exceeding 0.75. For both 0/3-shot IcL and 1/3-shot IcL, the medians for all three groups do not exceed 0.75.
This suggests that predicting performance under the same prompt type with varying numbers of examples (shots) is unreliable, as the quantity of examples significantly influences model performance. 
(2) Within the 0-shot setting, there is a discernible correlation between CoT, IcL, and EF. This observation is supported by Figure \ref{fig_main:central_prompt}, where the lowest median correlation values for both the ``all models'' and ``selected 3 models'' groups exceeding 0.5. This indicates that although models are sensitive to changes in prompt structure, merely altering the guidance of questions within the prompt, without changing the examples, typically does not lead to significant variations in performance.
(3) It is observed that the median values of all groups' box plots are above zero, indicating a generally positive correlation in model performance across various prompts. This implies that, even without a strong correlation, a model's performance with one type of prompt can somewhat predict its effectiveness with other prompts.

From the analysis of model performance, the following insights can be summarized: 
(1) The ``selected 3 models'' group shows significant variability in performance across different prompts.  
This is evidenced by the frequent appearance of outliers in their sub-boxes, suggesting high data variability. 
The length of the whiskers on these sub-boxes is usually extended, indicative of greater data dispersion. Additionally, the asymmetry in the lengths of the upper and lower whiskers for this group points to an uneven distribution of data. The interquartile ranges are also larger, further confirming increased variability.
(2) There is a notable correlation between the \emph{doubt} and \emph{ignore} prompts across all models, which are designed to assess model robustness. This consistent correlation suggests that the stability of a model's performance under these prompts follows a uniform pattern.

\paragraph{Metric-centric relationships.}
Our primary objective extends beyond simply achieving high accuracy in model performance. We aim to develop models that not only exhibit superior performance but also demonstrate robustness. The ability for consistent performance across various causal scenarios is crucial for language models intended for widespread deployment.
\begin{figure}[t]
    \centering
\includegraphics[width=1\linewidth]{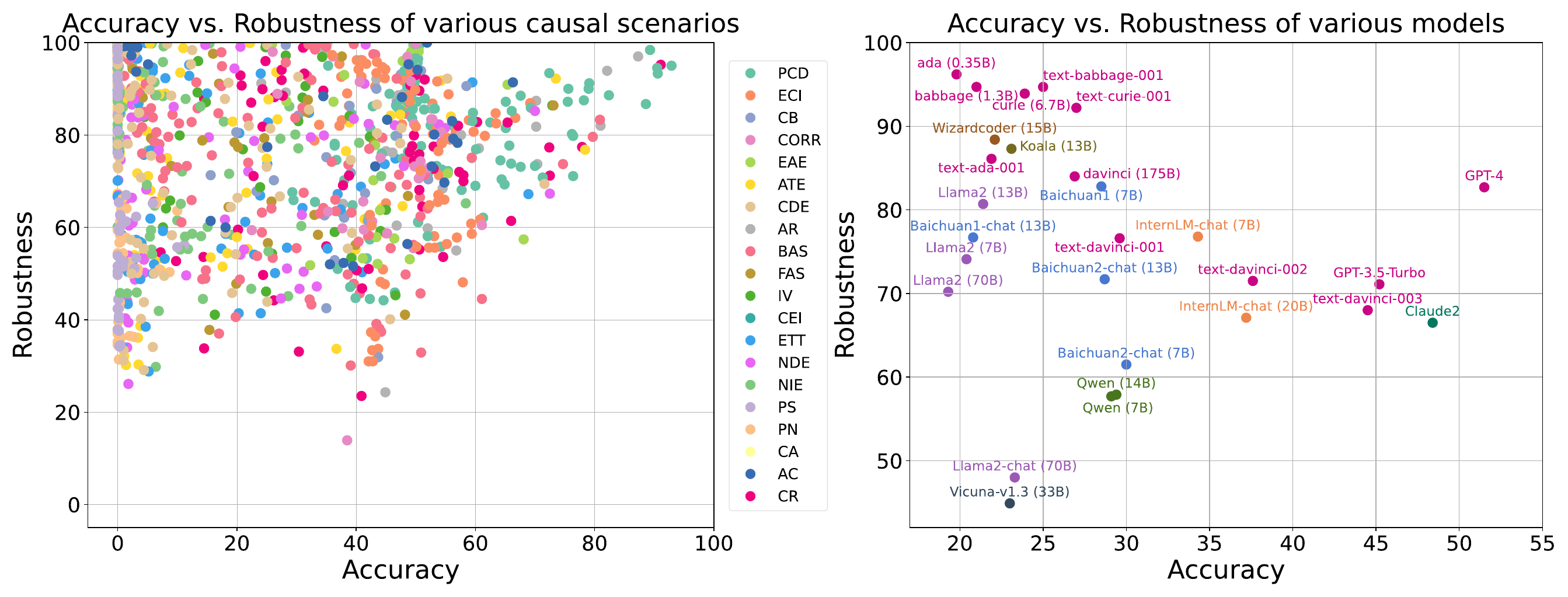}
    \caption[Correlation between accuracy and robustness]{\textbf{Relationship between accuracy and robustness.} We separately demonstrate the relationships between accuracy and robustness from the perspectives of the causal scenario and the model.}
    \label{fig_main:central_metric}
\end{figure}

In Figure \ref{fig_main:central_metric}, we depict the correlation between accuracy and robustness across various causal scenarios and models. When focusing on causal scenarios, each point on the plot corresponds to the accuracy and robustness of a specific model within a specific causal scenario. When examining models, we average out the accuracy and robustness for each model over all causal scenarios and plot these data points. And for models that originate from the same creator, we represent with the same color. Note that, in our analysis, we have not included IcL, CoT, and EF in this consideration, because our adversarial prompts are designed to attack the basic prompt. Therefore, the accuracy metric we refer to is an average merely derived from basic prompt, \dt, and \ig. In future research, we intend to examine the impact of disruptions on these other types of prompts.

From an analysis grounded in causal scenarios, it is evident that the interplay between a model's robustness and accuracy significantly varies across different causal scenario. Notably, in challenging causal scenarios such as PN and PS, there is an interesting trend: models may exhibit minimal accuracy yet display disproportionately high robustness, in some instances reaching 100\%. This occurs primarily because most models are fundamentally unable to respond to these kinds of questions effectively. Their responses remain stable, irrespective of any disturbances. Our robustness metric, which assesses whether responses change before and after an attack, thus appears artificially inflated in these scenarios. This observation highlights a limitation of our current robustness metric, underscoring the need for future research to develop more nuanced and detailed criteria for its evaluation. In contrast, For less challenging scenarios like PCD and AR, the correlation between robustness and accuracy is generally positive, as indicated by trend lines with a positive slope. However, in causal scenarios such as ECI, EAE, and AC, the relationship between these metrics does not exhibit a clear and consistent pattern, likely influenced by the distinctive characteristics of individual models.

Moving on, we turn our attention to models, highlighting several key points of interest. (1) GPT-4 is distinguished as the sole model that manages to maintain an optimal balance between accuracy and robustness. This is evident in the right subplot of Figure \ref{fig_main:central_metric}, where GPT-4 is uniquely positioned in the upper right corner. 
(2) Within models released by the same creator within a narrow timeframe (less than three months apart), there is a noticeable variability in the relationship between accuracy and robustness. For example, we examine the Llama2 series released by Meta at the same time. Here, it is clear that Llama2-chat (70B) demonstrates considerably lower robustness compared to the other three models in the same series. This could potentially be attributed to the adverse effects of RLHF, which might make the model more prone to altering its responses under human critique. A similar pattern is observed across the four models in the Baichuan series, where Baichuan1 (7B), not fine-tuned with RLHF, achieves the highest level of robustness.\footnote{Please refer to \nameref{main:model} (\cref{main:model}) for detailed information into the training strategies of various models.}
(3) The evolutionary path of OpenAI's models further illustrates these dynamics. Initially, the GPT-3 models (e.g., ada, babbage, curie) exhibit lower accuracy but higher robustness due to their limited causal reasoning and poor instruction-following capabilities. In contrast, the subsequent InstructGPT series (e.g., text-ada-001, text-babbage-001, text-curie-001) improve upon both fronts, thereby concurrently increasing accuracy and diminishing robustness. This evolution culminates in the release of GPT-4, marking the pinnacle of balance between accuracy and robustness to date.

\begin{figure}[t]
    \centering
\includegraphics[width=1\linewidth]{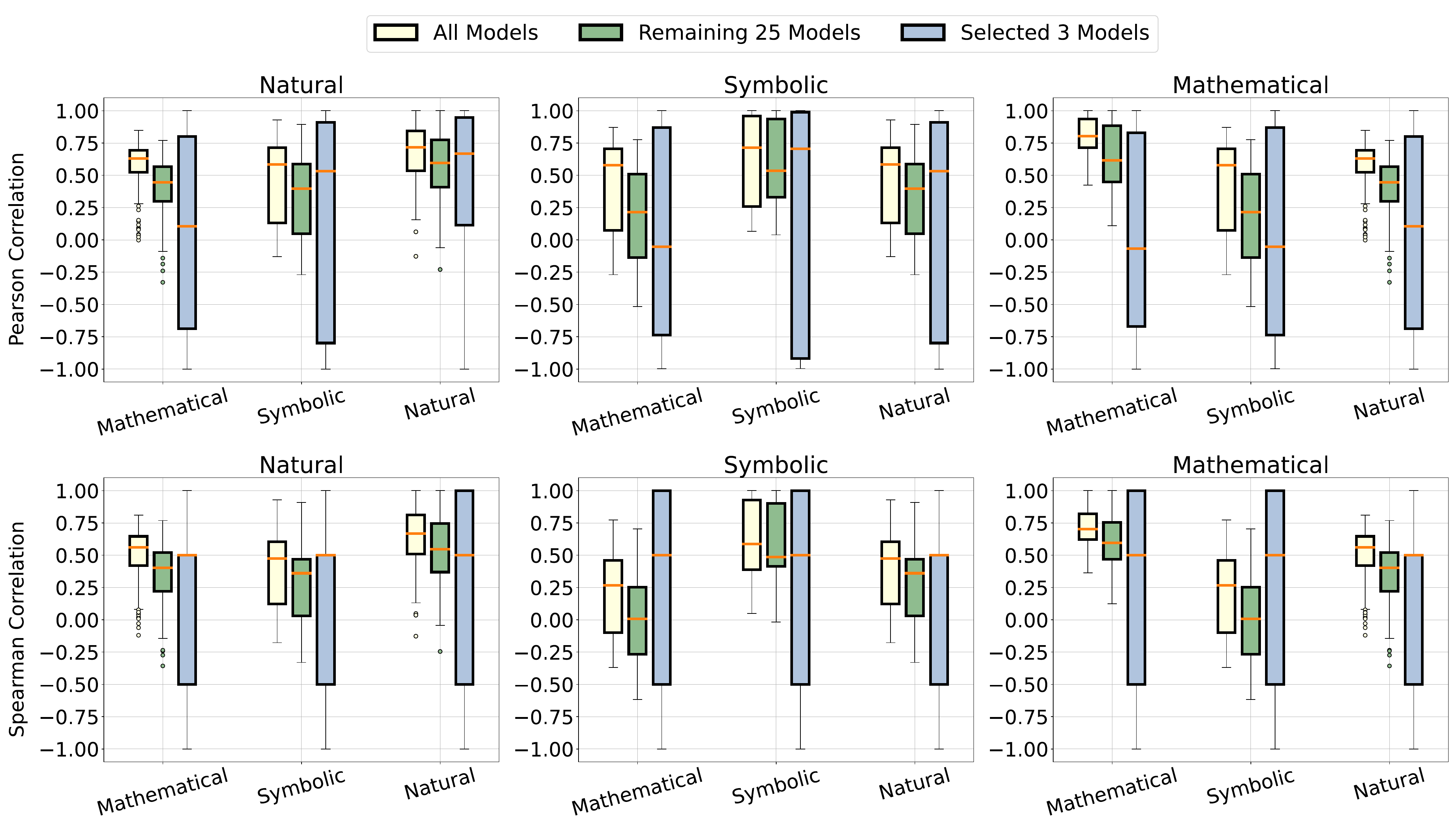}
    \caption[Correlation between modes]{\textbf{Correlation between modes.} We compute both the Pearson and Spearman correlations for each mode with every other mode. Boxes in various colors denote different groups (i.e., all models, remaining 25 models, and selected 3 models). Dots in different colors indicate outliers from those respective groups. The median is represented by an orange line.}
    \label{fig_main:central_mode}
\end{figure}

\paragraph{Mode-centric relationships.}
We investigate the relationships between different modes - Natural, Symbolic, and Mathematical - using both Pearson and Spearman correlation metrics \citep{hauke2011comparison}. These modes comprise 20, 12, and 14 causal tasks, respectively. Our analysis employs consistent model grouping as used in \textbf{Prompt-centric relationships}, i.e., ``all models'', ``selected 3 models'', and ``remaining 25 models''. The details of the correlation coefficients are computed as follows:
(1) We calculate the average accuracy for models within these groups across various causal tasks and modes. For instance, in the ``selected 3 models'' group, this involves creating a 3$\times$20 matrix for the Natural mode. Rows of this matrix correspond to models in the group, columns correspond to different causal tasks within the Natural mode, and entries of the matrix represent the average accuracy of each model on these tasks.
(2) We compute the correlations between causal tasks across different modes. For instance, in the ``selected 3 models'' group, we analyze matrices for the Natural (3$\times$20) and Mathematical (3$\times$14) modes, by iterating over columns to produce 20$\times$14=280 correlation coefficients.
(3) Finally, these coefficients are then visualized using a boxplot.

Furthermore, we explore the correlations among causal tasks of the same mode. For example, we have a 3$\times$20 matrix for the Natural mode in the ``selected 3 models''. Then we iterate over the columns to calculate the correlations among 20 causal tasks within the Natural mode, resulting in 20$\times$20=400 correlation coefficients. This approach leads to the consequence that the correlation within the same mode (for instance, Natural with Natural) is not 1. This method is adopted because concentrating only on modes without considering individual causal tasks might miss out on more nuanced information, which could be essential. Utilizing both Pearson and Spearman metrics accommodates the diverse qualitative trends that different modes may exhibit.

\begin{figure}[t]
    \centering
\includegraphics[width=0.8\linewidth]{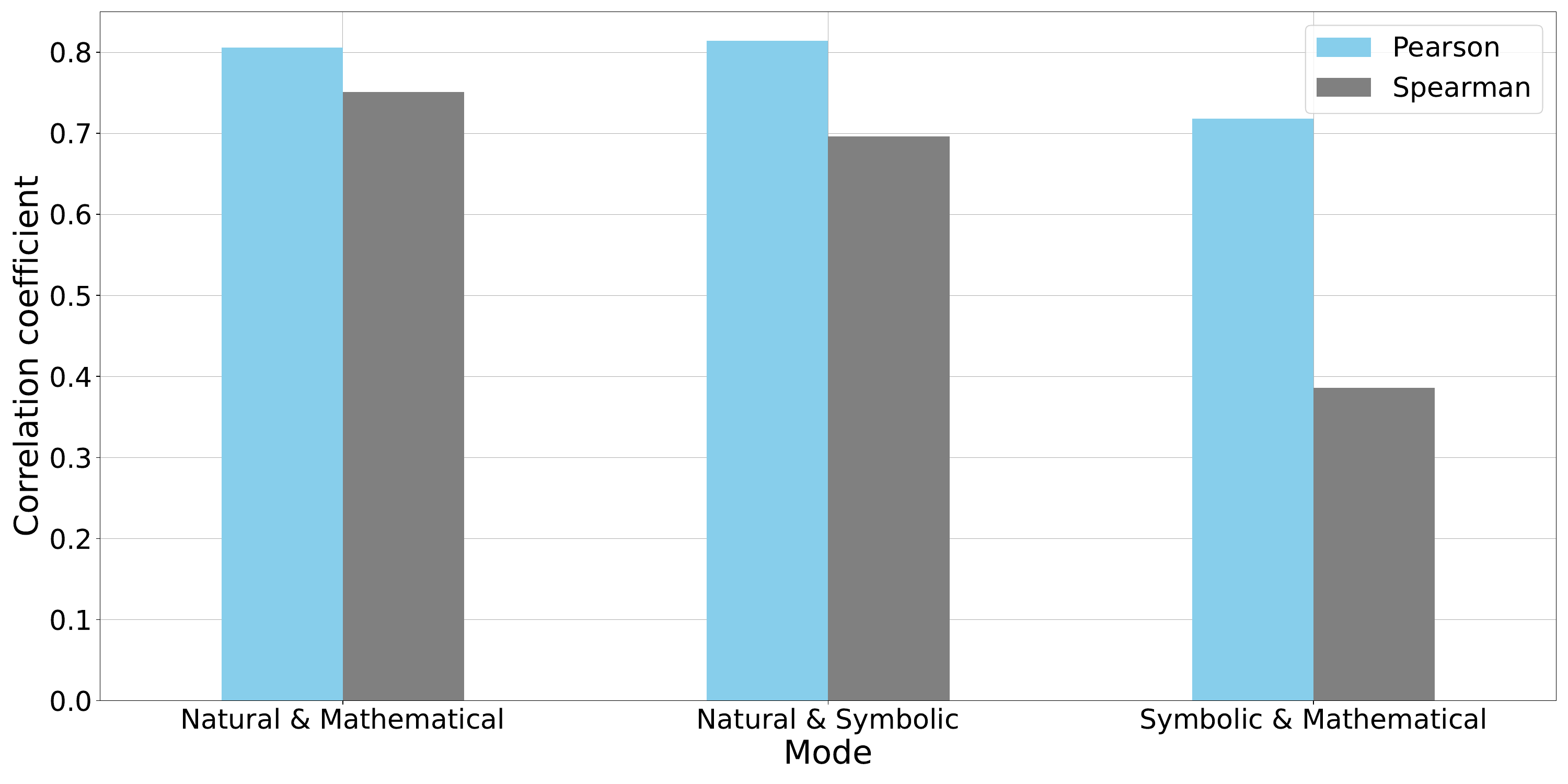}
    \caption[Overall correlation between modes]{\textbf{Overall correlation between modes.} We compute both the Pearson and Spearman correlations for every mode pair (i.e., the correlations of Natural and Mathematical, Natural and Symbolic, Symbolic and Mathematical).}
    \label{fig_main:central_mode_overall}
\end{figure}

In Figure \ref{fig_main:central_mode}, we focus on the relationships among all causal tasks across different modes. In contrast, Figure \ref{fig_main:central_mode_overall} broadens the perspective, offering a comparison of the three modes at a higher level. To accomplish this, we begin by calculating the average accuracy for all causal tasks within each mode across all models. Following that, we apply both Pearson and Spearman correlation metrics to assess the relationships between these modes.

Our mode-centric analysis yields several insightful observations regarding the relationships between different modes. (1) It can be concluded from Figure \ref{fig_main:central_mode_overall} that the Natural mode has a high correlation with both the Mathematical and Symbolic modes. Specifically, the Pearson correlation between Natural and Symbolic modes is notably high at 0.814, while their Spearman correlation, though the lowest, is still substantial at nearly 0.7. In contrast, the correlation between the Mathematical and Symbolic modes is relatively low. These findings provide valuable insights for the development of diverse causal task designs. (2) Taking a closer look at the relationship between the Mathematical and Symbolic modes, we find that there exists a strong linear but weaker monotonic relationship between them. As shown in Figure \ref{fig_main:central_mode_overall}, their Pearson correlation is 0.718, indicating a strong linear association, but the Spearman correlation drops significantly to 0.386. This suggests that the linear relationship varies across data segments - being strong in some and weak or inverse in others - which impacts the Spearman correlation. This metric emphasizes the consistency of trends rather than the strength of linear relationships.
(3) While the Natural and Mathematical modes generally show a high correlation, Figure \ref{fig_main:central_mode} reveals significant variability in the distribution of their correlation coefficients. The interquartile range (the box's upper and lower edges) for the ``all models'' group is asymmetrical, with an elongated lower portion, suggesting a broader spread of lower values. Moreover, numerous outliers at the lower end for both the ``all models'' and ``remaining 25 models'' groups highlight considerable differences among causal tasks within these modes.

Shifting the focus to models, our analysis provides specific insights as follows. (1) For the ``selected 3 models'' group depicted in Figure \ref{fig_main:central_mode}, they exhibit significant variability in performance across different modes. Specifically, Pearson correlation analysis reveals marked differences in accuracy for various causal tasks within these modes, indicating that even among top-performing models, performance consistency is not guaranteed. Additionally, Spearman Correlation suggests that the rankings of these models fluctuate, highlighting instability in their relative performance.
(2) As shown in Figure \ref{fig_main:central_mode}, the median correlations for the ``all models'' and ``remaining 25 models'' groups demonstrate that causal tasks within the same mode exhibit the highest correlations. This outcome aligns with expectations, as tasks within the same mode generally share similar formats.

\begin{figure}[t]
    \centering
\includegraphics[width=0.8\linewidth]{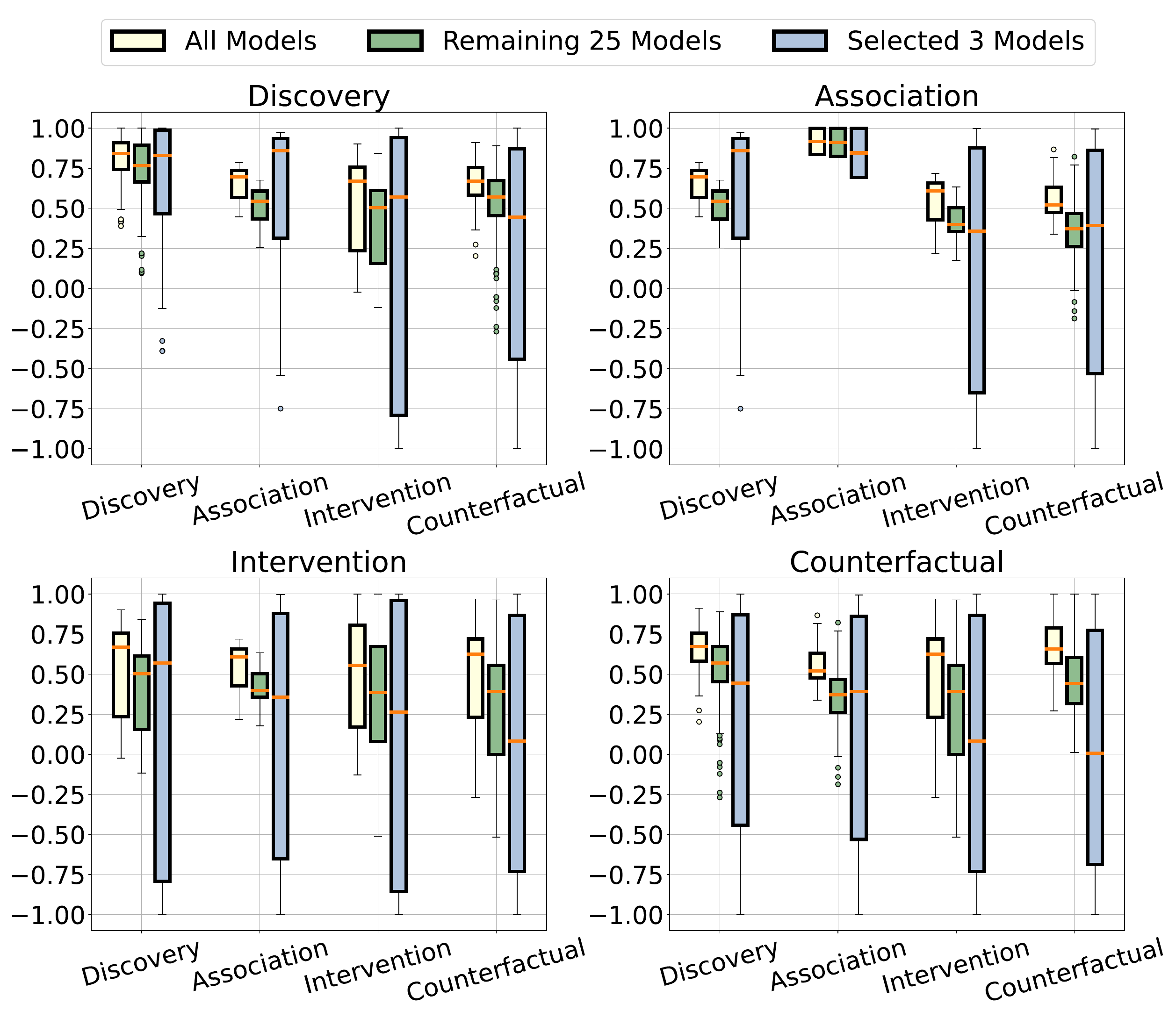}
    \caption[Correlation between various rungs of causal ladder]{\textbf{Correlation between various rungs of causal ladder.} We compute the Pearson correlation for each rung with every other rung. Boxes in various colors denote different groups (i.e., all models, the remaining 25 models, and the selected 3 models). Dots in different colors indicate outliers from those respective groups. The median is represented by an orange line.}
    \label{fig_main:central_ladder}
\end{figure}

\paragraph{Causal ladder-centric relationships.}
In Figure \ref{fig_main:central_ladder}, we explore the Pearson correlations across different levels of the causal ladder, following the approach used in \textbf{Mode-centric relationships}. We look into how all causal tasks within specific levels of the ladder relate to one another. Specifically, we consider 10 causal tasks under causal discovery, 2 under association, and 17 under both intervention and counterfactuals levels. We also assess the correlations among tasks within the same level of causal ladder, clarifying why correlation coefficients for pairs from the same level (e.g., association-association) do not necessarily equal 1.\footnote{We adopt the same calculation strategy with \textbf{Mode-centric relationships}, and please refer to it for the detailed explanation.}

Based on Figure \ref{fig_main:central_ladder}, we draw the following conclusions: (1) From the perspective of the ``all models'' group, there is a strong correlation between causal discovery and the other three levels of the causal ladder. It is shown that the upper quartile values for ``all models'' are greater than or equal to 0.75, indicating a significant positive relationship. This supports the design rationale behind our CaLM framework, affirming the placement of causal discovery as the foundational rung (Rung 0) in the ladder of causation. 
(2) Within the intervention level, causal tasks exhibit comparatively lower correlations with each other. This observation is derived from an analysis of correlations among causal tasks within each causal level (e.g., discovery-discovery, association-association). Notably, the boxplots for intervention-intervention exhibit the greatest range across three model groups, reflecting the most variability. Furthermore, the median correlation values for both the ``all models'' and the ``remaining 25 models'' groups identify intervention-intervention as having the weakest inter-task correlations within the intervention ladder.

\paragraph{Causal scenario-centric relationships.}

Our evaluation spans all rungs of the causal ladder, offering the most extensive setting for evaluating the causal reasoning capabilities of language models across diverse causal scenarios. The variety of causal scenarios inspires us to investigate the correlations between them. This exploration aids in enhancing our understanding of how the models perform in various causal scenarios. We aim to identify specific scenarios where model performance exhibits strong or weak correlations, and to ascertain whether certain causal scenarios pose unique challenges to the models.

\begin{figure}[t]
    \centering
\includegraphics[width=\linewidth]{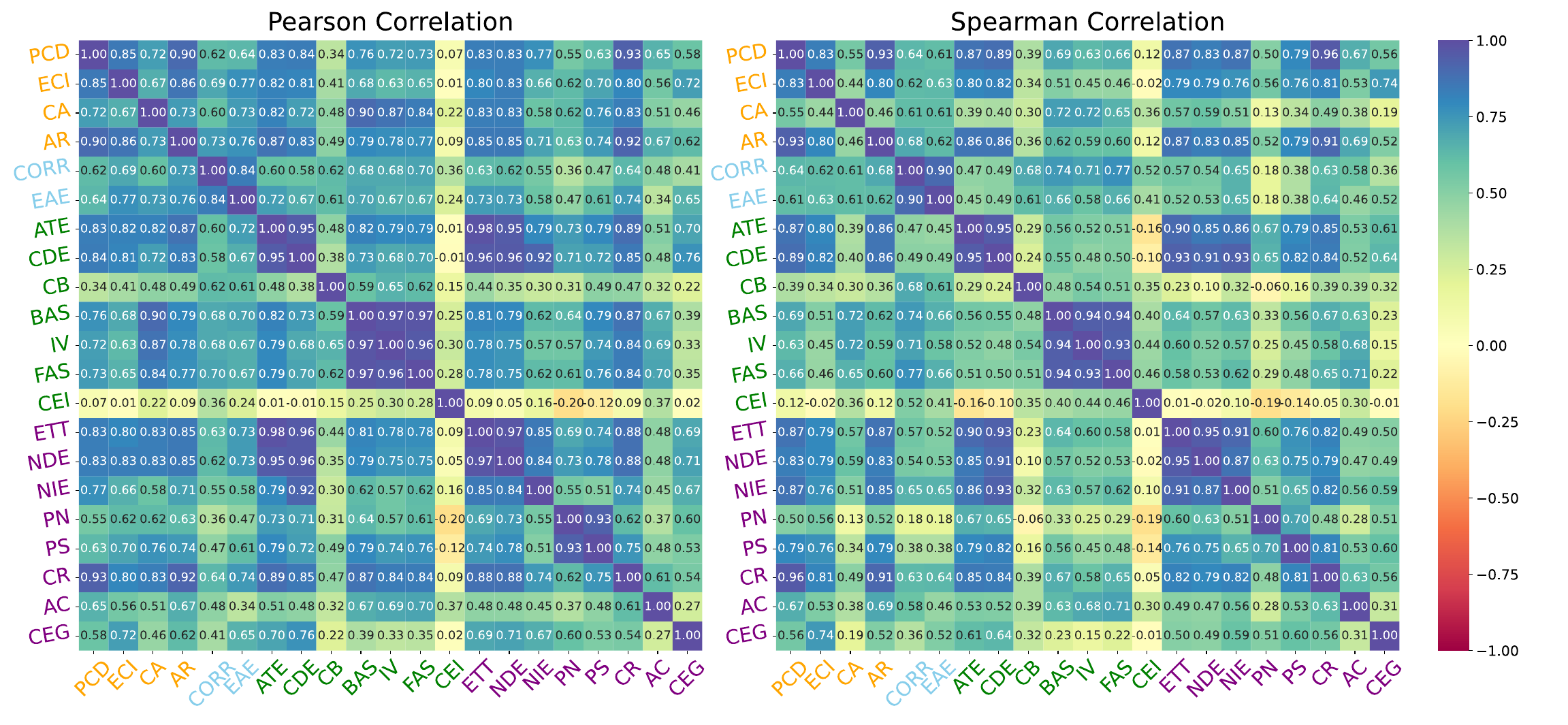}
    \caption[Inter-causal scenario performance correlation]{\textbf{Inter-causal scenario performance correlation.} We compute both the Pearson and Spearman correlations for every causal scenario. Each cell represents how the model accuracy correlates with two causal scenarios.}
\label{fig_main:central_scenario}
\end{figure}

In Figure \ref{fig_main:central_scenario}, we analyze the Pearson and Spearman correlations across various causal scenarios. To be specific, we calculate the average accuracy for all models under each causal scenario, and then assess how these accuracy values correlate with each other across different scenarios. The heatmap presented in the figure illustrates these correlations, with each cell displaying the respective correlation coefficient. To enhance the visual clarity and the interpretability of the data, we color-code the names of the causal scenarios based on their respective rungs of the causal ladder: causal discovery scenarios are highlighted in orange, association scenarios in sky blue, intervention scenarios in green, and counterfactuals scenarios in purple.

Our initial analysis, based on the quantitative correlation trends displayed in Figure \ref{fig_main:central_scenario}, provides macro-level insights as follows: (1) Causal scenarios within the same level of the causal ladder typically exhibit higher correlations. 
Specifically, at the level of causal discovery, except for CA, the other three scenarios demonstrate strong correlations, with both correlation coefficients exceeding 0.80. At the association level, all scenarios show correlations above 0.80. At the intervention level, two groups of scenarios - ATE and CDE, as well as BAS, IV, and FAS - display very high correlations, each with a correlation above 0.93. 
For the counterfactuals level, even the lowest correlation coefficient among ETT, NDE, and NIE reaches 0.84. (2) The models show positive correlations in performances across the 19 causal scenarios, except for CB and CEI. This uniformity in performance across the four rungs of the causal ladder underscores the cohesiveness of the models' capabilities, and affirms the soundness of our causal scenario design.

Switching to some specific causal scenarios, there are important findings worth emphasizing: (1) Overall, CEI shows the lowest correlation with other causal scenarios. Both Pearson and Spearman correlations reveal a distinct color demarcation line along the CEI axis. According to the analysis in \cref{scenario:intervention}, CEI poses considerable challenges to the models, and the models' performance ranking in this causal scenario is noticeably different from others. Besides CEI, CB's correlations with other causal scenarios are also relatively low. From Section 2.3.5 in \citet{lu2024gpt}, we discover that models are easily misled by the probability figures presented within questions, leading to wrong responses. Our results also indicate that models still struggle with recognizing CB.
(2) Despite belonging to different levels of the causal ladder, there is apparent correlation between certain distinct causal scenarios in terms of model performance. Scenarios within causal discovery (e.g., PCD, ECI, and AR), along with those belonging to intervention (e.g., ATE and CDE) and counterfactuals (e.g., ETT, NDE, NIE, PS, and CR), exhibit relatively strong correlations among each pair (e.g., PCD-ETT, CDE-NIE, ECI-ATE). For instance, PCD and CR have an extremely significant correlation, with both coefficients exceeding 0.93. Among these combinations, NIE and PS have the lowest correlation, but their Pearson Correlation still exceeds 0.5. 
(3) Compared to PS, PN has a lower correlation with other causal scenarios. As detailed in Section 2.3.3 of \citet{lu2024gpt}, there is a significant difference in the models' performance when inferring necessary versus sufficient causes. Models generally fail to accurately infer necessary causes but consistently provide correct inferences for sufficient causes. Combining existing research with our experimental results, it is evident that models have yet to clearly distinguish between the concepts of necessity and sufficiency, with the PN scenario presenting greater challenges and thus exhibiting a lower correlation.

In addition to the consistent patterns previously identified, it is crucial to note the causal scenarios where significant discrepancies between Spearman and Pearson correlation coefficients occur, which highlight key findings: 
(1) The performances of models in the PN and CEI scenarios show considerable differences between these two types of quantitative coefficients. Specifically, PN is notable because all its relationships, except for its self-correlation, are lower when assessed using the Spearman coefficient rather than the Pearson. This suggests a potential shift in the models' performance ranking within the PN causal scenario. For CEI, while it shows positive Pearson correlations with scenarios like ATE, NDE, and CEG, these turn negative when measured with the Spearman coefficient. These observations concerning PN and CEI reinforce the conclusions from the previous analysis, indicating that these two causal scenarios are particularly challenging for the models and exhibit notable divergences in correlation types. 
(2) Across all models and causal scenarios, the Spearman coefficients are generally lower than the Pearson coefficients. This implies that linearity (as captured by Pearson) is more pronounced than monotonicity (as captured by Spearman) in these causal scenarios. It further suggests that the relationships might not consistently adhere to a simple ranking order but rather show a linear dependency.

\subsubsection{Inter-dimensional Relationships}
\label{main:extra}
\begin{figure}[t]
    \centering
\includegraphics[width=\linewidth]{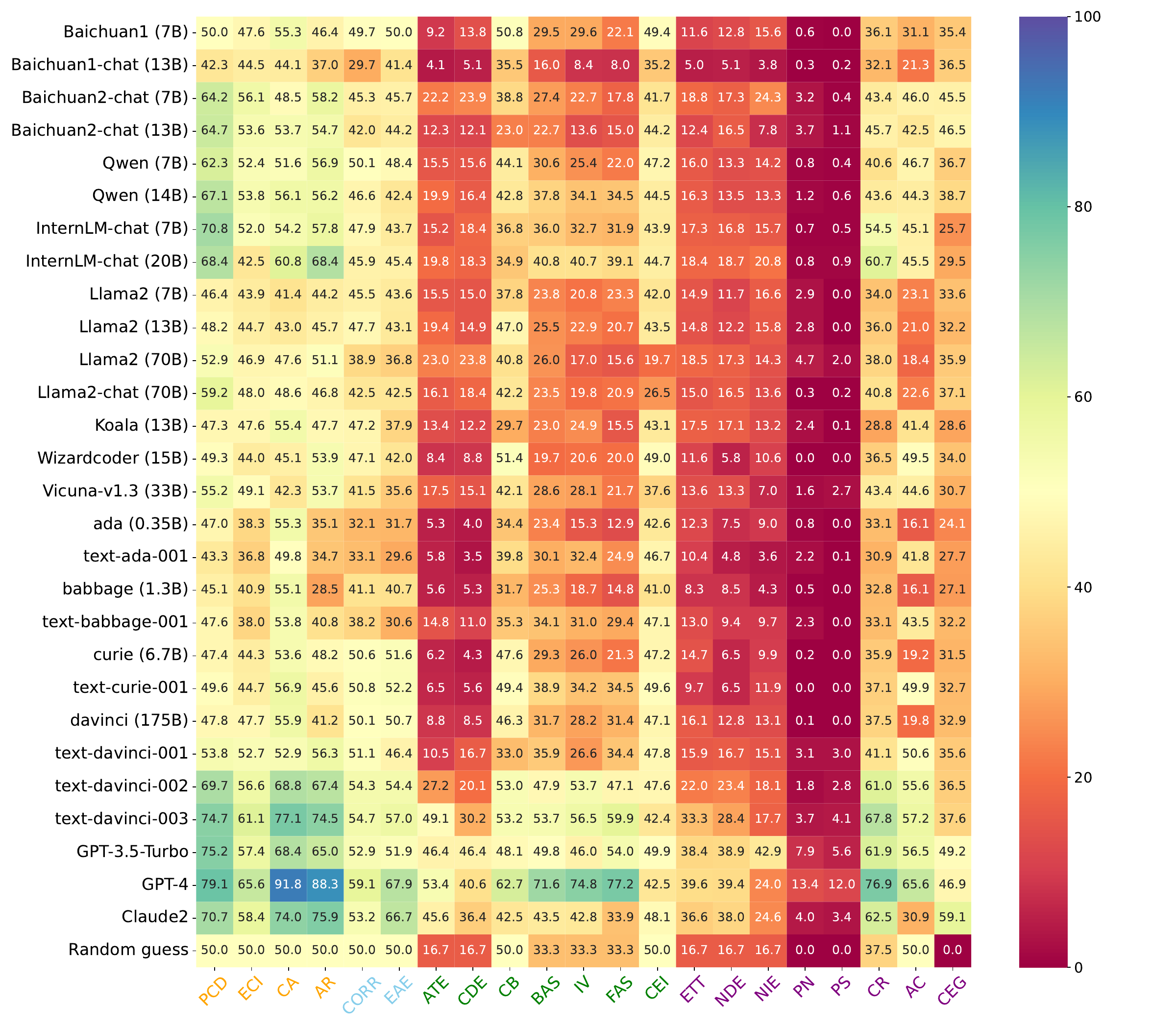}
    \caption[Relationship between causal scenario and model]{\textbf{Relationship between causal scenario and model.} Each cell stands for the model's average accuracy in a specific causal scenario.}
    \label{fig_main:extra_scenario_model}
\end{figure}

\paragraph{Relationship between causal scenario and model.}
How does a single model fare across various causal scenarios, and how do different models compare within the same causal scenario? These are the main questions that interest us. To address these, we illustrate the interactions between 28 models and 21 causal scenarios in Figure \ref{fig_main:extra_scenario_model}. The heatmap in this figure displays the average accuracy for each model within a particular causal scenario, where each cell's number represents this measure. It is essential to acknowledge that the variation in question types across different domains affects the baseline probabilities of a random guess. For instance, in the intervention-related causal scenarios, the random guess probabilities are as follows: 16.7\% for ATE and CDE, 33.3\% for BAS, IV, and FAS, and 50\% for CB and CEI. The accuracy figures shown are the absolute performance metrics for each causal scenario. Hence, when evaluating the heatmap, we should not only consider the depth of color, which illustrates performance levels, but also compare these figures against the baseline random guess probabilities. This comparison will provide a more balanced and objective analysis of model efficacy across different scenarios. Furthermore, we have developed a metric to categorize causal scenarios based on the difficulty levels associated with their respective random guess probabilities, as detailed in Section \ref{main:metrics}.

Based on Figure \ref{fig_main:extra_scenario_model}, we can derive the following insights: 
(1) Causal scenarios that incorporate Mathematical mode, such as ATE, CDE, ETT, NDE, NIE, PN, and PS, pose the most significant challenges for models, as evidenced by the deep red sections of the graph. The inclusion of Mathematical mode in these scenarios substantially impacts model performance, leading to a marked decline. This suggests a need for models to improve in handling complex mathematical reasoning within causal contexts.
(2) The performance of many models is comparable to, or even less than, what would be expected from random guesses. This observation underscores a general deficiency in the necessary background knowledge for causal reasoning scenarios among most models, indicating biases in their understanding and a limited capacity to follow instructions rigorously. 
(3) The use of long textual formats introduces additional difficulties for models. For example, the AC scenario, which involves analyzing a relatively long narrative as highlighted in \cref{counterfactual:ac}, showing a marked performance disparity between the GPT-3 and InstructGPT series models. The challenges associated with processing long texts have become a focal point in recent research, with studies such as those by \citet{chen2023longlora} and \citet{wang2024augmenting} providing practical solutions to enhance model capabilities in handling extensive textual information.

\begin{figure}[t]
    \centering
\includegraphics[width=1\linewidth]{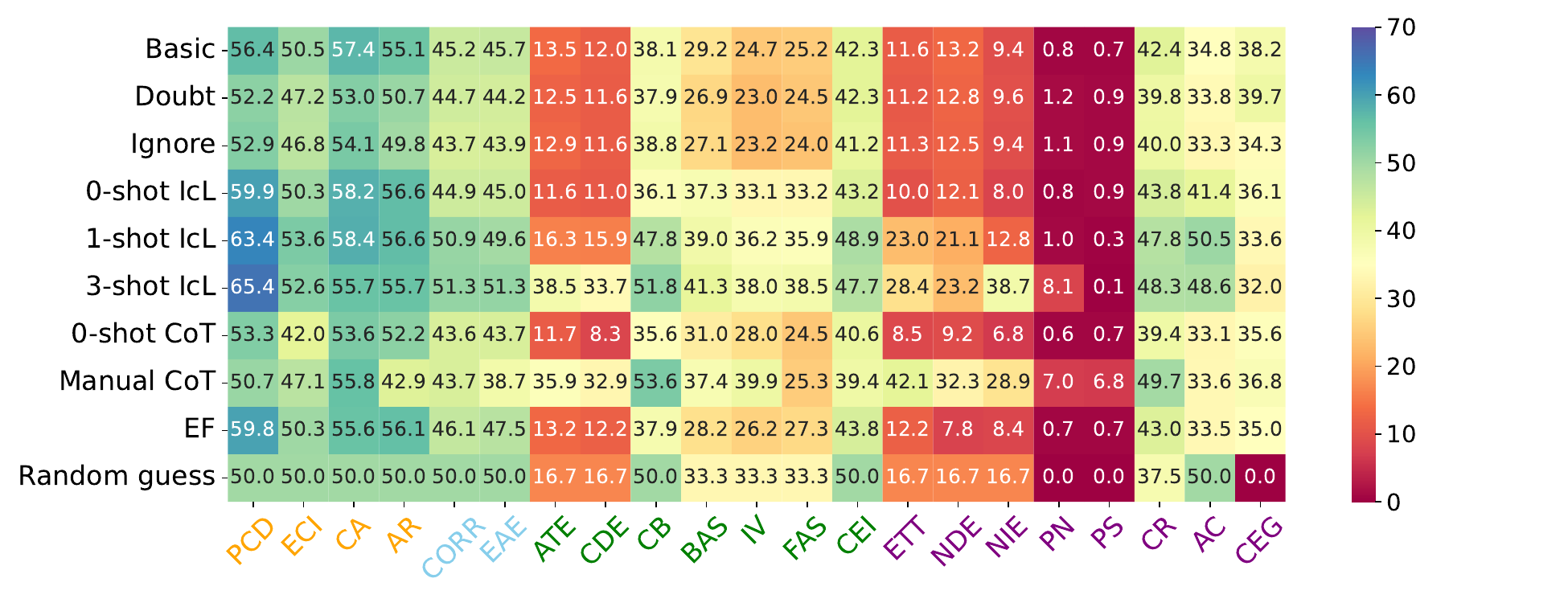}
    \caption[Relationship between scenario and prompt]{\textbf{Relationship between scenario and prompt.} Each cell stands for the prompt's average accuracy in a specific causal scenario.}
    \label{fig_main:extra_scenario_prompt}
\end{figure}

\paragraph{Relationship between causal scenario and prompt.}
Given the diverse range of prompts, we are interested in finding out if there exists a universal prompt that can be beneficial across all causal scenarios, or if distinct causal scenarios require uniquely tailored prompts. This part delves into examining the interaction between our 21 causal scenarios and 9 prompts. The relationships among them are illustrated in Figure \ref{fig_main:extra_scenario_prompt}, where each cell in the matrix displays the average accuracy of a specific prompt within a specific causal scenario.

After carefully analyzing the figure, we arrive at these insights: 
(1) There is no single prompt that consistently outperforms others across all causal scenarios, indicating the absence of a ``one-size-fits-all'' solution. Although this may complicate the task of identifying the most effective prompt for each scenario, certain prompts significantly enhance performance across most scenarios. For instance, in causal discovery and association scenarios, the use of 1-shot/\ticl~has an effective enhancement to model performance. In intervention scenarios (excluding CEI and IV), utilizing \ticl~is the most beneficial, while for counterfactuals scenarios, the \mcot~outperforms others in all but NIE, PN, and AC. As such, it is recommended to select prompts based on the specific requirements of each level of the causal ladder.
(2) In particularly demanding causal scenarios, such as PN and PS, most prompts fail to significantly improve model performance, with the exception of \ticl~ and \mcot. However, even these prompts do not raise peak average performance beyond 8.5\%. The lack of a public dataset for evaluating these scenarios may contribute to the models' poor understanding, hindering their ability to address these challenges effectively. Consequently, even with the step-by-step guidance provided by \mcot, it remains difficult to effectively teach the models to tackle such problems.
(3) Increasing the number of samples significantly enhances the performance of the majority of models. When examining 0/1/\ticl~ and 0-shot/\mcot, it becomes evident that models yield higher accuracy as sample size increases, aligning with findings from other research \citep{liang2022holistic,xu2022k}. However, the feasibility of extensively testing every possible sample size is limited by practical constraints such as time and resources.
(4) The \ticl, \oicl, and \mcot~prompts have been particularly effective in promoting model performance. The \ticl~prompt provides the most substantial benefits, enabling models to outperform the random guess baseline across 19 causal scenarios, compared to only 7 scenarios with the basic prompt. \oicl~and \mcot~also allow models to perform better than random guessing in 15 and 14 causal scenarios, respectively.

\subsubsection{Analyzing Complexity}
\label{main:complexity}
The analysis and insights from Figures \ref{fig_main:extra_scenario_model}, \ref{fig_main:extra_scenario_prompt} and \ref{fig_main:direct_mode} reveal that tackling causal reasoning tasks within the Mathematical mode\footnote{For more information of the Mathematical mode datasets, refer to \nameref{main:data} (\cref{main:data}).} poses substantial challenges for models. Understanding the underlying causes of these failures is essential before devising targeted improvements. Therefore, this section is dedicated to exploring the reasons why language models struggle with tasks in Mathematical mode, aiming to inform more effective strategies for enhancing their performance.

\begin{figure}[t]
\centering  
\subfigure[No causation]{ 
\begin{minipage}{4cm}
\centering    
    \includegraphics[width=1.05\linewidth]{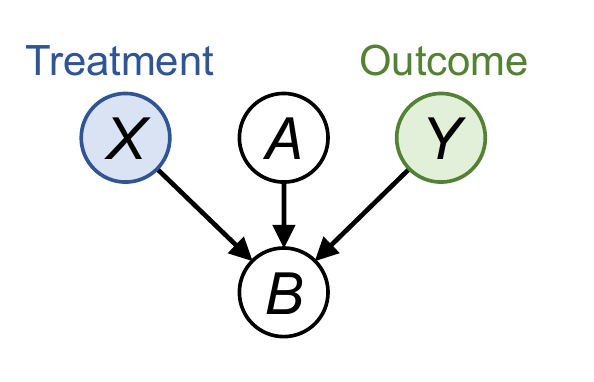}
\end{minipage}
}
\hspace{1cm}
\subfigure[Empty adjustment set]{
\begin{minipage}{4cm}
\centering  
    \includegraphics[width=1.05\linewidth]{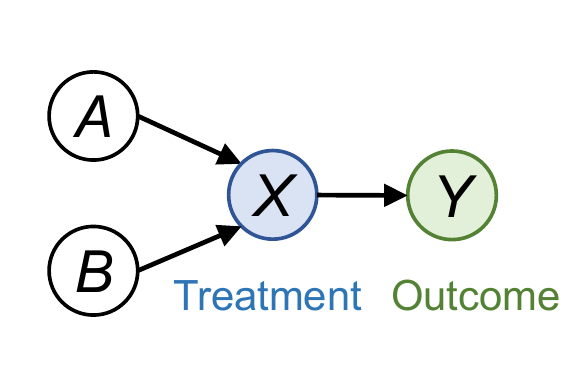}
\end{minipage}
}
\hspace{1cm}
\subfigure[Nonempty backdoor adjustment set]{
\begin{minipage}{4cm}
\centering    
    \includegraphics[width=1.05\linewidth]{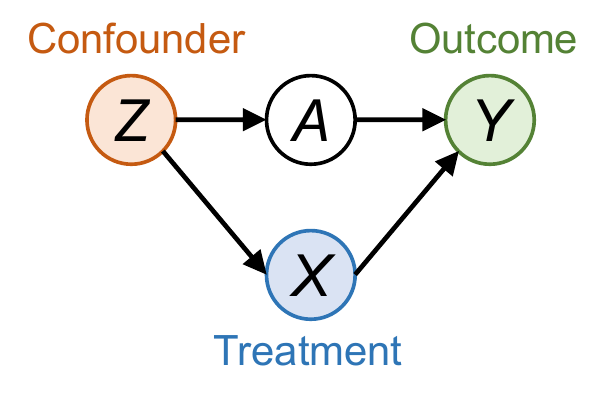}
\end{minipage}
}
\caption[Illustration of causal reasoning levels]{\textbf{Illustration of causal reasoning levels.} We categorize the difficulty into three levels, arranged from the simplest to the most complex as follows: no causation, empty adjustment set, nonempty backdoor adjustment set. Here, $X$ represents the treatment, $Y$ represents the outcome, and $Z$ represents the confounder.}    
\label{fig_main:complexity_demo}    
\end{figure}
\paragraph{Complexity settings.}
The complexity of causal tasks within the Mathematical mode is determined by four primary factors: 
(1) \textbf{Number of nodes}: The complexity is gauged by the number of nodes in the causal graph, ranging from 3 to 9, where more nodes indicate a more perplexing structure. 
(2) \textbf{Number of edges}: The intricacy of each task is measured by the number of edges in the causal graph, which spans from 2 to 10. More edges contribute to a denser and more complicated causal graph, increasing the difficulty of analysis. 
(3) \textbf{Authenticity}: We have established three different settings of authenticity\footnote{For more information about this setting, refer to \nameref{data:construction} (\cref{data:construction})}, with the complexity progressively increasing. i) \textbf{Real}: Tasks are grounded in real-world phenomena, with questions that are aligned with accurate causal graphs. They conform to empirical common sense and do not cause any disturbances for models. ii) \textbf{Fake}: To test the impact of memorization, nouns within the causal graph are replaced with fictitious terms like ``\emph{bitk}'' and ``\emph{dmfl}''. This tests models' abilities to process information devoid of real-world context. iii) \textbf{Random}: Nodes in the causal graph are replaced with randomly selected real-world nouns, challenging models to discern causality in a structurally correct but contextually disrupted setup.
(4) \textbf{Causal reasoning process}: We structure the process of causal reasoning into three levels as shown in Figure \ref{fig_main:complexity_demo}, where each level represents a step up in complexity and challenge. i) \textbf{Level-1 (no causation)}: There is no causal relationship between the treatment and outcome, thus requiring no computation for resolution. ii) \textbf{Level-2 (empty adjustment set)}: There exists a causal relationship between the treatment and outcome, and their adjustment set is empty. iii) \textbf{Level-3 (nonempty backdoor adjustment set)}: There exists a causal relationship between the treatment and outcome. Due to the presence of confounders, their backdoor adjustment set is not empty. Table \ref{task_table_level_computation} details the formulas required for different causal scenarios across various complexity levels, highlighting that the probability at Level 1 is consistently zero, with the computational challenge escalating as the level increases. For a more detailed illustration, Table \ref{task_table_level_example} showcases examples of ATE categorized by these complexity levels. These gradations in task complexity pose significant challenges to the causal reasoning capabilities of language models, complicating the assessment of which factors most significantly impact their accuracy.

\begin{center}
\begin{table*}[t]
\fontsize{10}{14}\selectfont
    \caption[Calculation for three causal reasoning levels]{\textbf{Calculation for three causal reasoning levels.} $X$ represents the treatment, $Y$ represents the outcome, $Z$ represents the confounder, and $M$ represents the mediator.}
    \label{task_table_level_computation}
    \centering
  \begin{tabular}{ c|c }
\toprule
\textbf{Causal Scenario} & \textbf{Probability Calculation}\\
\hline
\multicolumn{2}{c}{\cellcolor{blue!5}\emph{Level 1: No causation}}\\
\hline
ATE& 0\\
CDE&0 \\
 ETT&0\\
NDE& 0\\
NIE& 0\\
\hline
\multicolumn{2}{c}{\cellcolor{green!5}\emph{Level 2: Empty adjustment set}}\\
\hline

ATE& $P(Y|X=1)-P(Y|X=0)$ \\
 CDE&$P(Y|X=1,M)-P(Y|X=0,M)$\\
 ETT&$P(Y|X=1)-P(Y|X=0)$\\
NDE& $P(Y|X=1)-P(Y|X=0)$ \\
NIE& $\sum_MP(Y|X=0,M)(P(Y|X=1)-P(Y|X=0))$ \\
\hline
\multicolumn{2}{c}{\cellcolor{purple!5}\emph{Level 3: Nonempty backdoor adjustment set}}\\
\hline

ATE&  $\sum_ZP(Z)(P(Y|X=1,Z)-P(Y|X=0,Z))$\\
 CDE&$\sum_ZP(Z)(P(Y|X=1,M,Z)-P(Y|X=0,M,Z))$\\
ETT& $P(Y|X=1)-\sum_ZP(Y|X=0,Z)P(Z|X=1)$\\
NDE& $\sum_M\sum_ZP(M|X=0,Z)P(Z)(P(Y|X=1,M,Z)-P(Y|X=0,M,Z))$\\
NIE& $\sum_M\sum_ZP(Y|X=0,M,Z)P(Z)(P(M|X=1,Z)-P(M|X=0,Z))$ \\
\hline
\end{tabular}
\end{table*}
\end{center}

\begin{center}
\begin{table*}[t]
\fontsize{10}{12}\selectfont
\caption[Samples with different complexity factors]{\textbf{Samples with different complexity factors.} The columns titled ``\#Nodes'' and ``\#Edges'' stand for ``number of nodes in the given causal graph'' and ``number of edges in the given causal graph'', respectively. Due to space limit, all examples are included in \nameref{appendix:complexity} (\cref{appendix:complexity}).} \label{task_table_level_example} 
\centering
\begin{tabular}{c|c|c|c|c|c}
\toprule
\textbf{Causal Scenario}& \textbf{Question} & \textbf{\#Nodes} & \textbf{\#Edges} & \textbf{Authenticity} & \textbf{Causal Reasoning Process} \\ 
\hline 
 \multirow{9}{*}{ATE}&Figure \ref{fig_appendix:complexity_ate1} & 5&6& Real& \multirow{3}{*}{Level-1}\\
 &Figure \ref{fig_appendix:complexity_ate2}  & 5& 4& Fake &\\
 &Figure \ref{fig_appendix:complexity_ate3}  & 5& 8& Random&\\
 \cline{2-6}
 &Figure \ref{fig_appendix:complexity_ate4} & 3& 3& Real&\multirow{3}{*}{Level-2}\\
 &Figure \ref{fig_appendix:complexity_ate5} & 3& 2& Fake&\\
 &Figure \ref{fig_appendix:complexity_ate6} & 3& 2& Random&\\
 \cline{2-6}
 &Figure \ref{fig_appendix:complexity_ate7} & 4& 5& Real&\multirow{3}{*}{Level-3}\\
 &Figure \ref{fig_appendix:complexity_ate8} & 4& 5& Fake&\\
 &Figure \ref{fig_appendix:complexity_ate9} & 3& 3& Random&\\
  \hline
\end{tabular}
\end{table*}
\end{center}

\begin{figure}[t]
    \centering
    \includegraphics[width=0.9\textwidth]{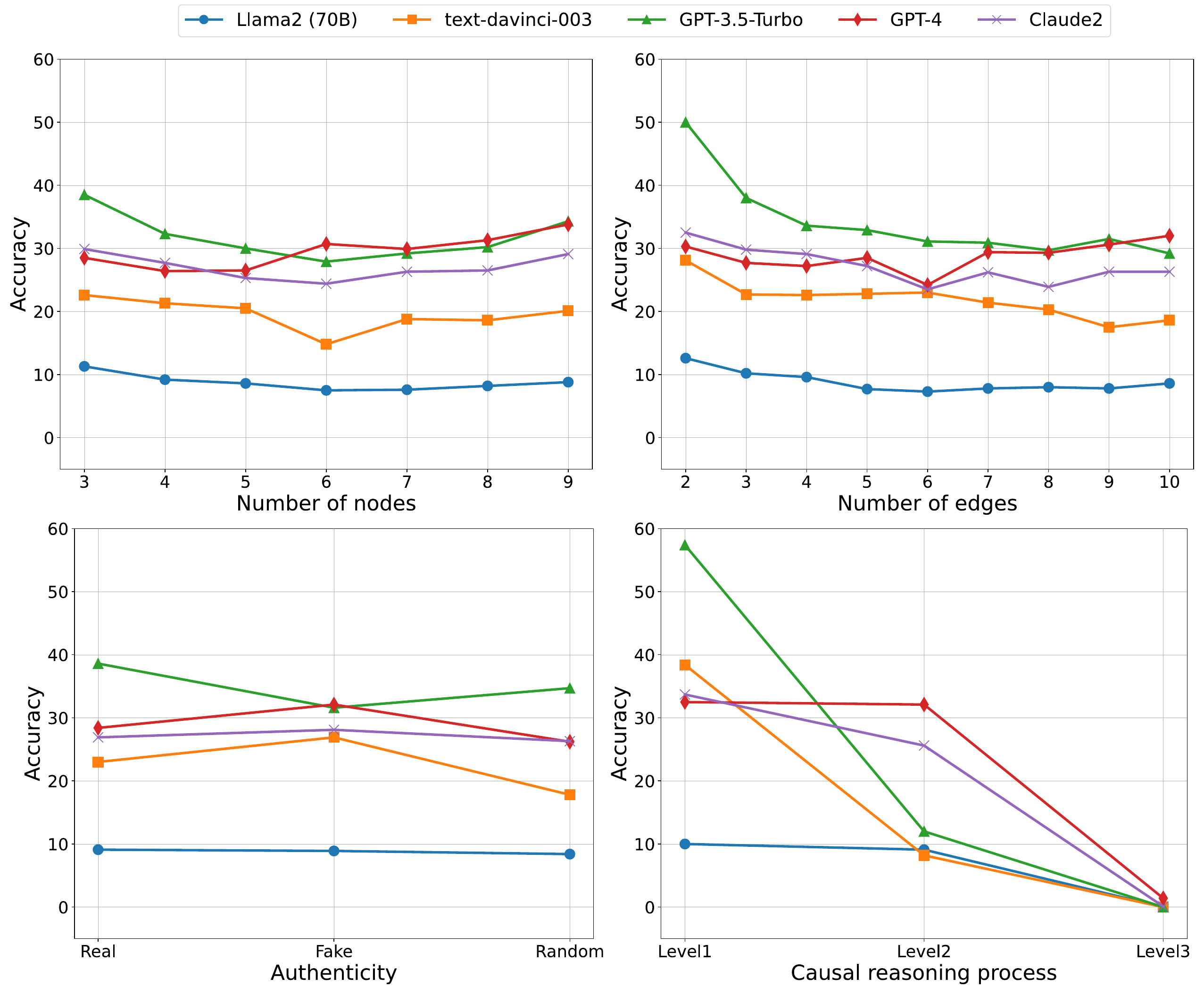}
    \caption[Complexity analysis of Mathematical mode questions]{\textbf{Complexity analysis of Mathematical mode questions.} We calculate the average accuracy of all evaluated models across five causal scenarios that involve Mathematical mode questions (i.e., ATE, CDE, ETT, NDE and NIE). The figure illustrates how accuracy trends vary under four different factors (i.e., number of nodes, number of edges, authenticity, and causal reasoning process) affecting the complexity.}
    \label{fig_main:complexity_all}
\end{figure}

\paragraph{What lead to the models' failure?}
Figure \ref{fig_main:complexity_all} depicts how the model's average accuracy varies with the four complexity factors. To mitigate the effects of variations in causal scenarios and prompt configurations, we compute the average accuracies across five causal scenarios (ATE, CDE, ETT, NDE, and NIE) and all nine prompts.
Due to the universally low accuracy rates observed in Mathematical mode causal tasks, we focus on the top five performing language models, as ranked in the Mathematical histogram presented in Figure \ref{fig_main:direct_mode}. We use relative accuracy as our metric, defined as the ratio of correct responses to the total number of samples within a category. This method ensures fairness in comparison, accommodating for the variability in the number of problems across different complexity factors.

From the analysis of the four sub-figures in Figure \ref{fig_main:complexity_all}, we arrive at several insights: 

(1) Present-day language models struggle significantly with complex causal reasoning tasks. Insights from the sub-figure focusing on the causal reasoning process reveal a marked decline in model performance as complexity increases to level 3. At this most challenging level, the accuracy of all five models examined nearly drops to zero, underscoring their substantial difficulties with high-complexity questions. Level 3 involves scenarios with confounders in the causal graph, requiring models to not only grasp causal concepts thoroughly but also to execute precise reasoning and calculations. This indicates a critical gap in current models' capabilities when faced with intricate causal relationships.

(2) GPT-4 stands out for its superior handling of relatively complex causal reasoning tasks. The sub-figure of causal reasoning process, particularly the curve between levels 1 and 2, highlights a distinctive pattern. Although GPT-4 starts as the fourth-ranked model at level 1, its relative accuracy remains robust, barely diminishing and sustaining above 30\% as the task complexity increases to level 2. In comparison, \claude, which is the second-best performer at level 2, also manages to significantly mitigate its performance drop. By contrast, \chatgpt~ and \textdavincithree, despite strong performances at level 1, undergo a sharp decrease in accuracy upon transitioning to level 2, with \chatgpt~showing a notably larger decline. Meanwhile, \llamaseventy~displays minor fluctuations in performance but maintains approximately 10\% accuracy at both levels, indicating a lower overall effectiveness in these tasks.

(3) Overall, the random setting (of authenticity) exerts the most profound effect on the performance of models. When examining the authenticity sub-figure, which uses the real setting (of authenticity) as a reference point due to its reflection of actual causal relationships, it is clear that all models experience a decrease in effectiveness under the random setting, as anticipated. This decline is attributed to the causal graph configuration where, despite nodes corresponding to real-world entities, the causal relationships are incorrectly aligned with reality. Thus, the causal graph in the random setting presents information that diverges from the models' training data, leading to substantial disruptions. Consequently, this challenges models not only to accurately identify the causal relationships but also to compartmentalize and disregard their pre-existing knowledge, which poses a significant challenge to their processing efficiency. 

(4) The models exhibit a certain level of abstract reasoning capability. Let us refocus on the authenticity sub-figure, with the real setting (of authenticity) as the reference. It is observed that, with the exception of \llamaseventy~ and \chatgpt, the other three models exhibit improved performance in the fake setting. In this setting, nodes are labeled with fictitious names that lack real-world significance. This performance suggests that these three models can effectively detach from real-world references and engage in causal reasoning within abstract contexts.

(5) Variations in the number of nodes and edges in a causal graph only marginally affect the performance of models. Analysis from the two corresponding sub-figures in Figure \ref{fig_main:complexity_all} reveal that all models experience a decline in performance when the node count increases from 3 to 5 and the edge count from 2 to 4. Interestingly, a general improvement in performance is observed as the node count rises from 6 to 9. This trend is attributed to the increased proportion of Level 1 questions in the dataset creation process. Given that some of the 28 evaluated models have a context window limited to 2,000 tokens (see \nameref{main:model} (\cref{main:model}) for details), ensuring that the question length stays within this limit necessitates a trade-off between the number of nodes and the complexity of the causal reasoning process. That is, we need to maintain a balanced number of input tokens with the addition of more nodes. More specifically, as the node count rises, more tokens are needed to describe the causal graph in the input. In contrast, there is no causal relationship between the treatment and outcome in the level 1 questions, so these questions require the fewest tokens for providing probability-related context. That is why we construct more level 1 questions as the nodes within a causal graph increase. This suggests that the addition of nodes does not significantly impact performance as long as the complexity of causal reasoning remains constant. Ultimately, it is the complexity of the causal reasoning process itself that primarily determines model performance.

\clearpage

\subsubsection{Analyzing Maturity}
\label{main:maturity}
We evaluate 21 distinct causal scenarios, each designed to evaluate different aspects of a model's capabilities. Understanding the maturity levels of these scenarios is crucial for advancing and deploying future models. A high maturity level indicates that a scenario has been effectively solved, while a low maturity level suggests significant room for further exploration. To thoroughly evaluate the maturity of these scenarios and mitigate potential biases from singular metrics, we have established three key metrics: \emph{understandability}, \emph{open-limited gap}, and \emph{solvability}, as detailed in \cref{metric:scenario}.

\begin{figure}[t]
    \centering
    \includegraphics[width=\textwidth]{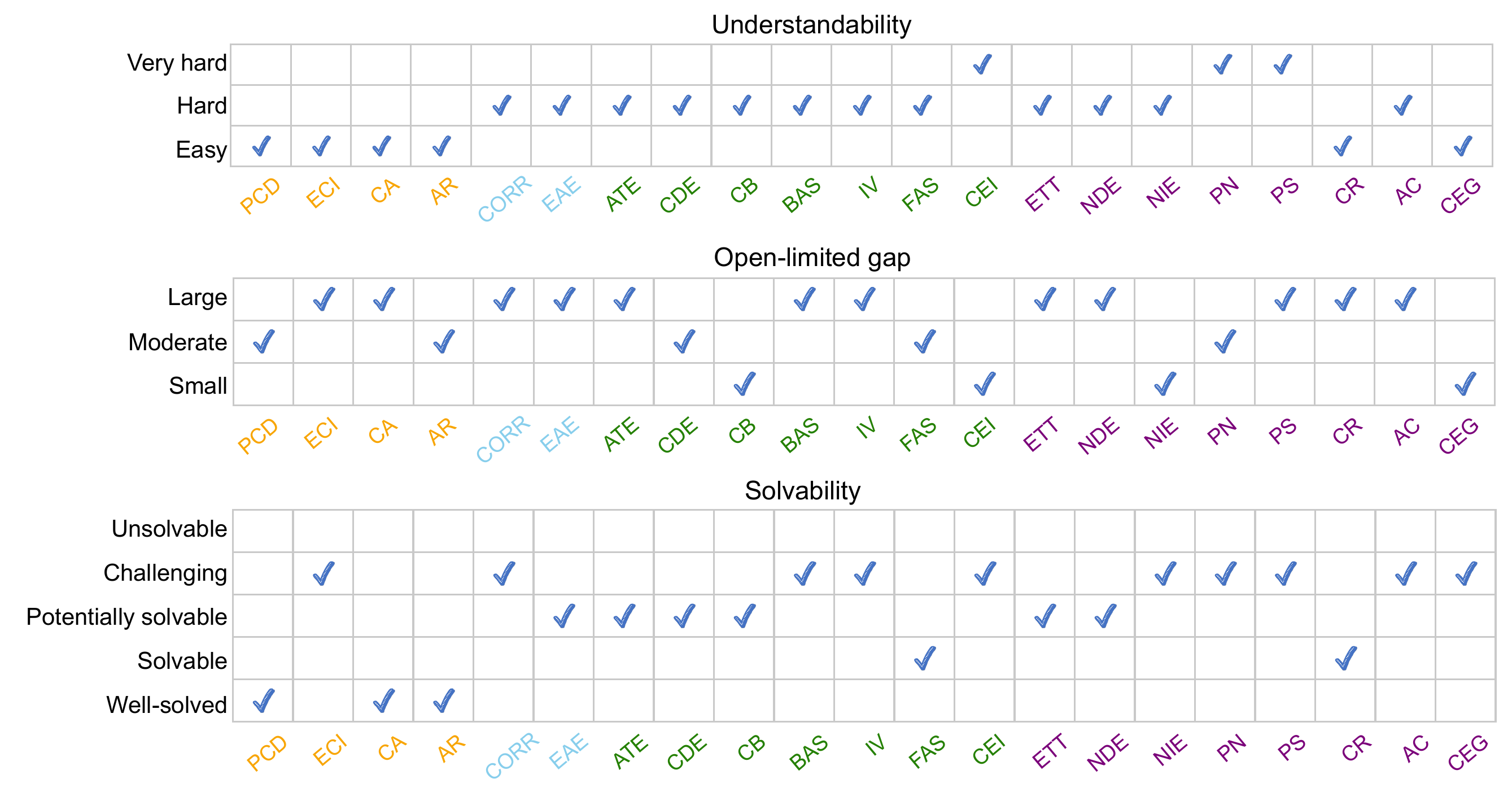}
    \caption[Maturity of causal scenarios]{\textbf{Maturity of causal scenarios.} We evaluate the maturity of a causal scenario by examining the following three metrics: understandability, open-limited gap, and solvability. In the figure, the areas labeled as ``\(\checkmark\)'' represent the corresponding maturity stages of the causal scenario.}
    \label{fig_main:maturity_all}
\end{figure}

Utilizing the computation methods described in \cref{metric:scenario}, we assess the maturity levels for each causal scenario, presented in Figure \ref{fig_main:maturity_all}. Our conclusions are as follows: (1) CaLM poses significant challenges in \emph{understandability}. While all causal scenarios within the causal discovery category are deemed easy, aligning with our initial perceptions, a majority of the scenarios (15 out of 21) are rated as hard or more difficult. This underscores the complexity of CaLM in terms of model comprehension. (2) When examining the \emph{open-limited gap}, there remains a considerable persistent gap between open and limited-access models. In most scenarios (17 out of 21), this gap is moderate to high, suggesting that models with limited access still outperform their open-access counterparts, maintaining dominance in the top 5 positions for each scenario. (3) In terms of \emph{solvability}, although no causal scenarios are deemed unsolvable, current models do not fully meet the challenges posed by CaLM. Almost half (10 out of 21) of the causal scenarios are classified as challenging.

\subsubsection{Analyzing Volatility}
\label{main:stability}
Assessing the robustness of a model solely through adversarial prompts may present some limitations, because it primarily focuses on whether there are changes in the model's response before and after encountering interference. This section expands the evaluation of robustness by considering stability from a broader viewpoint, consisting of both the model's and the prompt's volatility.\footnote{Detailed definitions of volatility are provided in Section \ref{main:metrics}.} The volatility of the model refers to the inconsistency in the model’s responses to variations in prompts within a specific causal scenario, averaged across all prompts. High volatility in this context suggests that the model is highly sensitive to changes in the prompt, leading to decreased stability. In contrast, the volatility of the prompt indicates the instability of a prompt’s performance relative to the basic prompt across all models in a particular causal scenario. Increased volatility here means a greater disparity in performance compared to the basic prompt, also reflecting reduced stability. In both cases, our concern is with the magnitude of these disparities, which do not necessarily reflect the absolute performance quality of the model. For example, a model with high volatility might perform modestly under the basic prompt but show significant improvement with a specific prompt. This highlights the considerable potential within the model that may be leveraged through tailored prompt engineering.

\begin{figure}[t]
    \centering
    \includegraphics[width=0.9\textwidth]{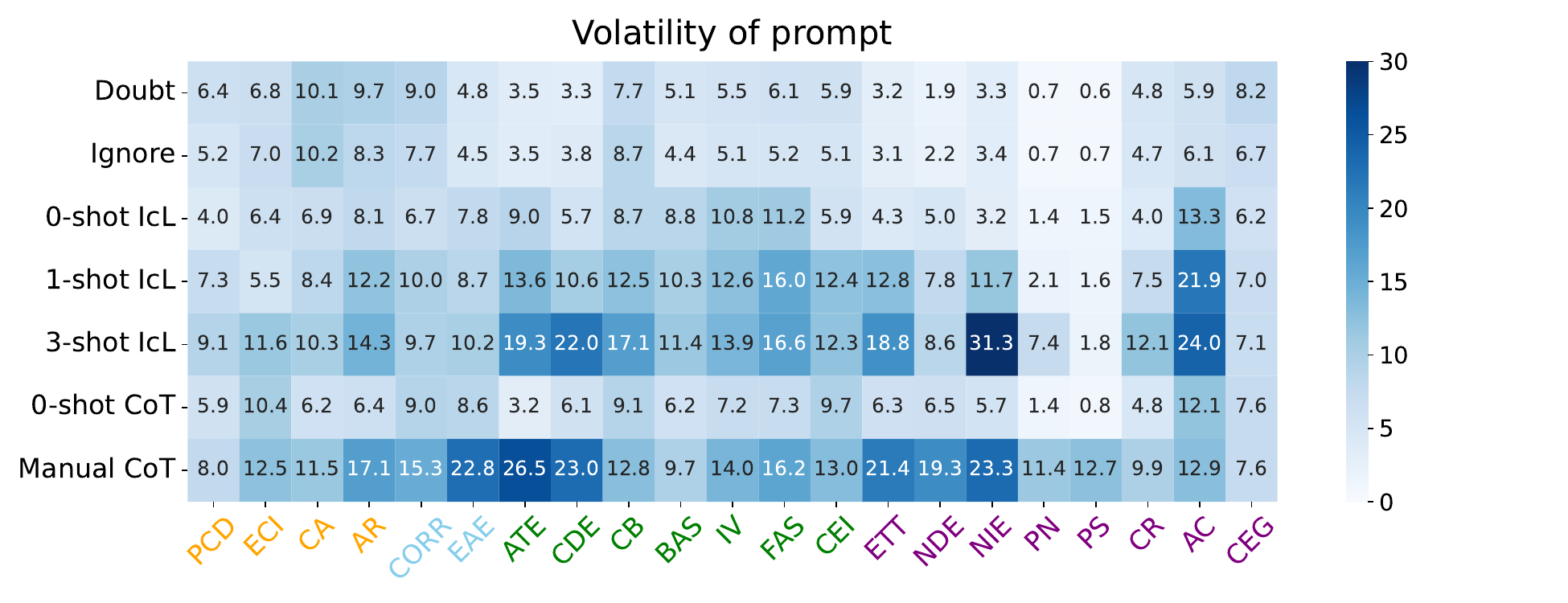}
    \caption[Volality of prompts]{\textbf{Volality of prompts.} Each cell stands for the prompt’s average volatility in a specific causal scenario.}
    \label{fig_main:stability_scenario}
\end{figure}

\paragraph{Volatility of prompt.}
In Figure \ref{fig_main:stability_scenario}, we illustrate the volatility of prompts across all causal scenarios. Each value in the heatmap represents the deviation of a prompt's volatility from that of the basic prompt within a given causal scenario. This is why the basic prompt is not shown in the figure. Note that, the EF prompt is not included in the volatility calculation, because it is not used in some causal tasks and, according to findings from both Figure \ref{fig_main:central_prompt_scatter} and Figure \ref{fig_main:central_prompt}, there is a significant correlation between the EF and the basic prompt. We believe that excluding EF does not undermine the validity of our conclusions.

From Figure \ref{fig_main:stability_scenario}, we can see that both 3-shot IcL and manual CoT exhibit the highest volatility. Noteworthily, we have observed from Figure \ref{fig_main:central_prompt_scatter} that there is a weak correlation between these prompts and the basic prompt. Since prompt volatility is measured relative to the basic prompt, the elevated volatility of 3-shot IcL and manual CoT corroborates the findings in Figure \ref{fig_main:central_prompt_scatter} and Figure \ref{fig_main:central_prompt}. Additionally, both Figure \ref{fig_main:direct_prompt} and Figure \ref{fig_main:extra_scenario_prompt} suggest that, compared to basic prompt, 3-shot IcL and manual CoT significantly enhance model performance. This finding is consistent with their observed high volatility, as the baseline for calculating volatility of prompt is the model's accuracy with the basic prompt.

\begin{figure}[t]
    \centering
    \includegraphics[width=\textwidth]{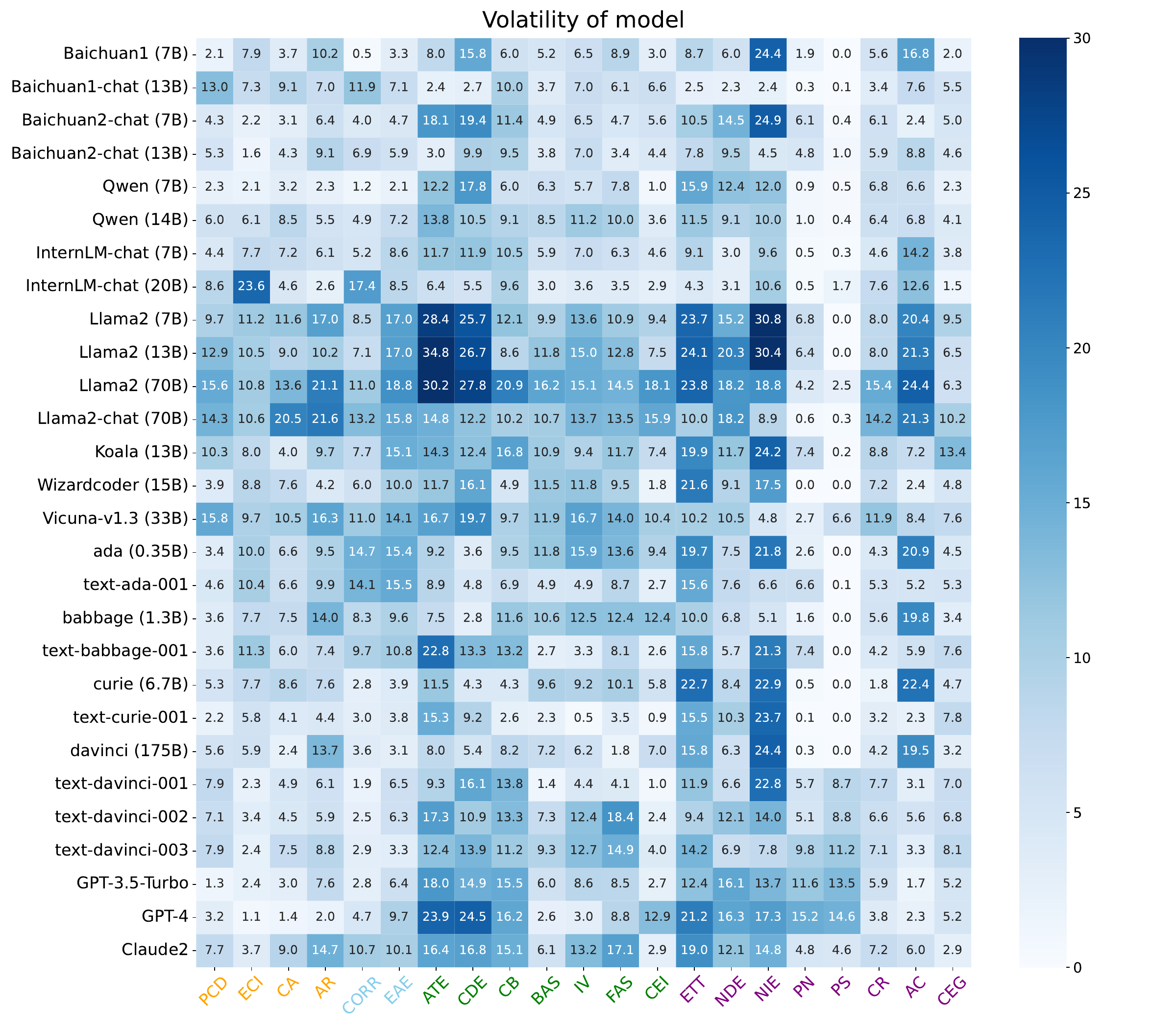}
    \caption[Volality of models]{\textbf{Volality of models.} Each cell stands for the model's average volatility in a specific causal scenario.}
    \label{fig_main:stability_model}
\end{figure}

\paragraph{Volatility of model.}
Figure \ref{fig_main:stability_model} visualizes the model's volatility across all causal scenarios. Each cell in the heatmap represents the model's volatility across all prompts within a specific causal scenario. From this figure, we can see that the model's stability seems to correlate with the causal ladder. For causal scenarios within the relatively simpler levels of the ladder, such as causal discovery and association, the model generally exhibits lower volatility, indicating higher stability. In contrast, causal scenarios within the more challenging levels, such as intervention and counterfactuals, tend to show higher volatility, reflecting lower stability. However, there are exceptions within the more difficult levels. PN and PS, which are categorized under counterfactuals, display significantly lower volatility, with some models even achieving zero volatility. Analysis from both Figure \ref{fig_main:extra_scenario_model} and Figure \ref{fig_main:extra_scenario_prompt} reveals that these scenarios are particularly challenging, characterized by low model accuracy and minimal influence from prompts. This explains the distinctively high stability observed in PN and PS scenarios, consistent with findings from Figure \ref{fig_main:central_metric}. 

\clearpage

\subsubsection{Analyzing Errors}
\label{main:error}
During the evaluation, we document the errors made by the models and conduct both quantitative and qualitative analyses in this section. By examining the errors from these two perspectives, we can gain a deeper understanding of the models' shortcomings that lead to suboptimal performance. This analysis not only helps identify areas for immediate improvement but also provides valuable insights that can guide enhancements in future research.
\begin{center}
\begin{table*}[t]
\fontsize{10}{12}\selectfont
    \caption[Error stastics]{\textbf{Error stastics.} The table includes the occurrence rates of the four errors and their average values. Here, Empty denotes empty response, Instruction denotes limitation of instruction-following, and Inconsistency denotes language inconsistency.}
    \label{main:table_model_errors}
    \centering
  \begin{tabular}{l|c|c|c|c|c}
\toprule
\textbf{Model} & \textbf{Empty}& \textbf{Instruction}& \textbf{Repetition}&\textbf{Inconsistency} &\textbf{Average}\\
\hline
Baichuan1 (7B)
& \barchart{4.6} & \barchart{44.7} & \barchart{33.7} & \barchart{5.6} & \barchart{22.2} \\
Baichuan1-chat (13B)
& \barchart{0.0} & \barchart{71.4} & \barchart{1.6} & \barchart{7.3} & \barchart{20.1} \\
Baichuan2-chat (7B)
& \barchart{0.0} & \barchart{54.8} & \barchart{0.8} & \barchart{5.7} & \barchart{15.3} \\
Baichuan2-chat (13B)
& \barchart{0.0} & \barchart{64.3} & \barchart{0.6} & \barchart{9.3} & \barchart{18.6} \\
Qwen (7B)
& \barchart{9.6} & \barchart{41.7} & \barchart{22.2} & \barchart{4.9} & \barchart{19.6} \\
Qwen (14B)
& \barchart{6.7} & \barchart{48.9} & \barchart{20.2} & \barchart{4.9} & \barchart{20.2} \\
InternLM-chat (7B)
& \barchart{0.0} & \barchart{15.0} & \barchart{0.1} & \barchart{3.1} & \barchart{4.6} \\
InternLM-chat (20B)
& \barchart{4.0} & \barchart{28.3} & \barchart{0.2} & \barchart{6.2} & \barchart{9.7} \\
Llama2 (7B)
& \barchart{3.2} & \barchart{55.5} & \barchart{28.0} & \barchart{8.3} & \barchart{23.8} \\
Llama2 (13B)
& \barchart{8.4} & \barchart{55.5} & \barchart{25.0} & \barchart{6.3} & \barchart{23.8} \\
Llama2 (70B)
& \barchart{17.5} & \barchart{64.1} & \barchart{26.9} & \barchart{8.3} & \barchart{29.2} \\
Llama2-chat (70B)
& \barchart{7.0} & \barchart{60.7} & \barchart{17.5} & \barchart{13.7} & \barchart{24.7} \\
Koala (13B)
& \barchart{4.5} & \barchart{41.2} & \barchart{22.8} & \barchart{4.4} & \barchart{18.2} \\
Wizardcoder (15B)
& \barchart{6.3} & \barchart{65.3} & \barchart{3.9} & \barchart{3.3} & \barchart{19.7} \\
Vicuna-v1.3 (33B)
& \barchart{2.3} & \barchart{49.7} & \barchart{3.3} & \barchart{5.9} & \barchart{15.3} \\
ada (0.35B)
& \barchart{0.0} & \barchart{68.6} & \barchart{40.9} & \barchart{4.2} & \barchart{28.4} \\
text-ada-001
& \barchart{0.0} & \barchart{53.7} & \barchart{5.1} & \barchart{11.4} & \barchart{17.6} \\
babbage (1.3B)
& \barchart{0.0} & \barchart{67.7} & \barchart{41.1} & \barchart{3.7} & \barchart{28.1} \\
text-babbage-001
& \barchart{0.1} & \barchart{36.4} & \barchart{3.9} & \barchart{2.5} & \barchart{10.7} \\
curie (6.7B)
& \barchart{0.0} & \barchart{57.2} & \barchart{52.3} & \barchart{3.1} & \barchart{28.2} \\
text-curie-001
& \barchart{0.0} & \barchart{38.6} & \barchart{2.8} & \barchart{2.4} & \barchart{11.0} \\
davinci (175B)
& \barchart{0.0} & \barchart{51.0} & \barchart{40.7} & \barchart{3.7} & \barchart{23.9} \\
text-davinci-001
& \barchart{0.0} & \barchart{10.4} & \barchart{1.5} & \barchart{1.4} & \barchart{3.3} \\
text-davinci-002
& \barchart{0.0} & \barchart{9.0} & \barchart{1.9} & \barchart{0.7} & \barchart{2.9} \\
text-davinci-003
& \barchart{0.0} & \barchart{4.7} & \barchart{0.2} & \barchart{0.8} & \barchart{1.4} \\
GPT-3.5-Turbo
& \barchart{0.0} & \barchart{10.3} & \barchart{0.1} & \barchart{0.8} & \barchart{2.8} \\
GPT-4
& \barchart{0.0} & \barchart{11.0} & \barchart{0.0} & \barchart{0.6} & \barchart{2.9} \\
Claude2
& \barchart{0.7} & \barchart{64.9} & \barchart{0.8} & \barchart{6.2} & \barchart{18.2} \\
\hline
Average
& \barchart{2.7} & \barchart{44.5} & \barchart{14.2} & \barchart{5.0} & \barchart{16.6} \\
\bottomrule
\end{tabular}
\end{table*}
\end{center}

\paragraph{Quantitative analysis of empty response, limitation of instruction-following, repetition, and language inconsistency.}
In Table \ref{main:table_model_errors}, we detail the proportions of four types of errors\footnote{Please refer to \nameref{error:quantitative} (\cref{error:quantitative}) for more detailed definitions of these errors.} (i.e., empty response, limitation of instruction-following, repetition, and language inconsistency) that the models exhibit across all causal scenarios and eight types of prompts (excluding EF, for reasons consistent with \nameref{main:stability}).
Through both horizontal and vertical analyses on this table, we can draw the following conclusions: 
(1) The most frequent error category is the limitation of instruction-following, whereas the least common is the empty response. This indicates that the models generally manage to generate responses, rarely failing to provide any output even when faced with challenging questions. However, their precision in adhering to specific instructions is still wanting, often struggle to produce responses in the exact format required (see \nameref{adaptation} (\cref{adaptation}) for detailed required output formats). 
(2) SFT significantly reduces the incidence of repetitive responses. By finetuning with high-quality input-output pairs, SFT helps prevent unnecessary repetition in the models' responses (for an in-depth analysis of training strategies, refer to \nameref{main:model} (\cref{main:model})). 
(3) Models exhibit varying levels of language inconsistency, ranging from 0.6\% to 13.7\%. Although this issue may not significantly affect the final accuracy (as extensive sampling is employed when calculating this metric), it can severely degrade the user experience in real-world applications. For example, a non-English-speaking user receiving responses containing English terms when asking a question in Chinese would find this problematic.
(4) On average, models from the text-davinci and InternLM series, along with ChatGPT and GPT-4, tend to make the fewest errors. (5) Overall, limited-access models have a lower error rate (13.8\%) than open-access models (19.0\%). 

\begin{figure}[H]
    \centering
 \includegraphics[width=0.88\textwidth]{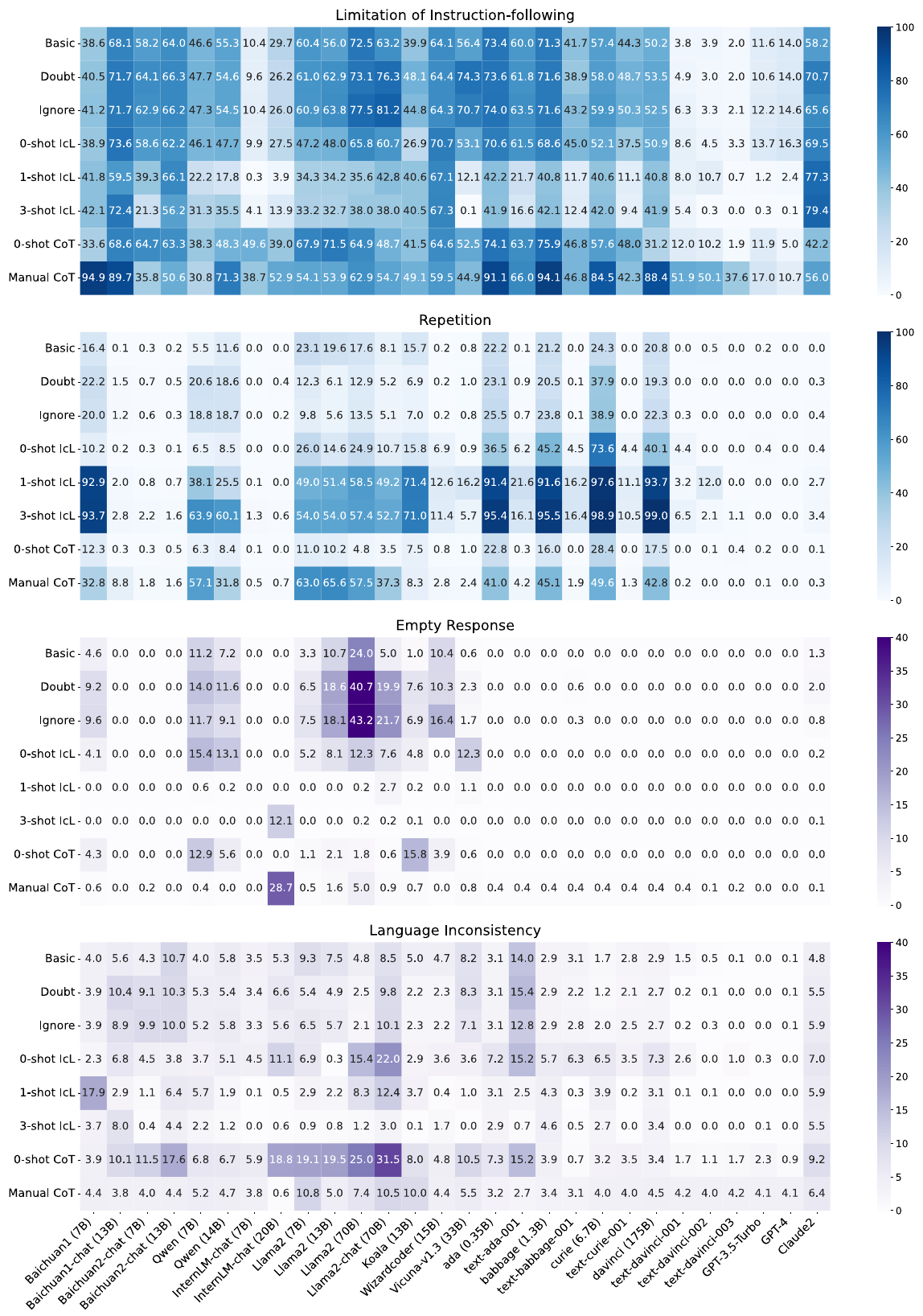}
    \caption[Relationship between error and prompt]{\textbf{Relationship between error and prompt.} Each cell stands for the average percentage of a specific type of error made by the corresponding models across 21 causal scenarios using a specific prompt. The frequency of different types of errors varies. For clearer presentation of the results, we choose distinct upper and lower limits and colors for each error type.}
\label{fig_main:error_prompt}
\end{figure}

\begin{figure}[t]
    \centering
 \includegraphics[width=\textwidth]{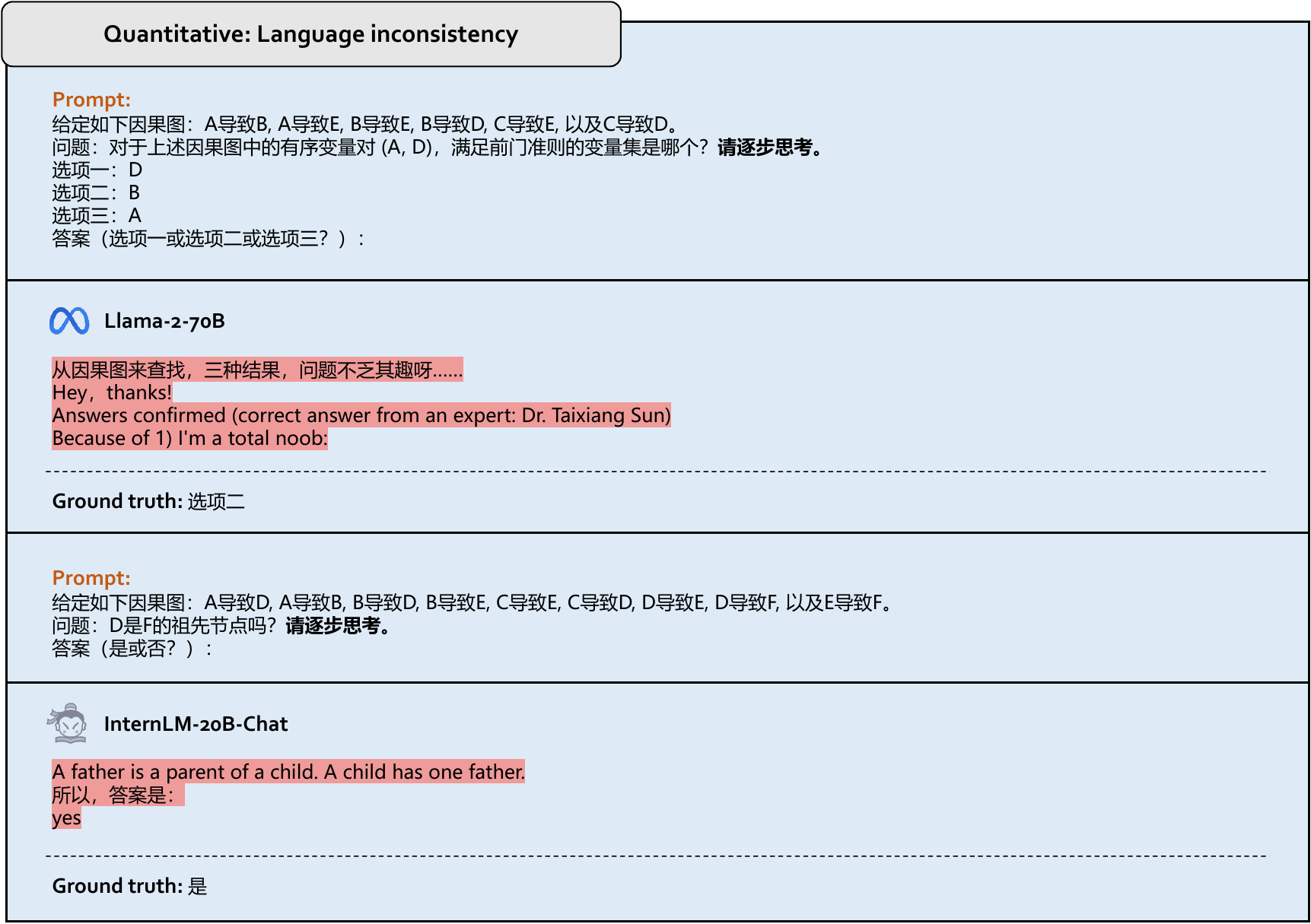}
    \caption[0-shot CoT's impact on language inconsistency]{\textbf{0-shot CoT's impact on language inconsistency.} }
\label{fig_main:error_quantitative_example1}
\end{figure}

Subsequent to our initial findings, we further explored the impact of different prompts on the models' error occurrence by creating four heatmaps in Figure \ref{fig_main:error_prompt}. Each value in these heatmaps represents the average percentage of a specific error made by the models in a given causal scenario when using a particular prompt. From our analysis of this figure, we discover the following insights: 
(1) Both 1-shot IcL and 3-shot IcL can significantly improve the ability of most models to follow instructions. Our experiments demonstrate that providing models with standardized, succinct examples enables them to learn effective response patterns, thereby improving their ability to generate outputs that adhere closely to the required answer format. 
(2) 1-shot IcL, 3-shot IcL, and manual CoT lead some models to engage in an ``imitation game''. These prompts provide examples that encourage models to mimic the observed patterns. As a result, these models not only produce responses but also generate their own questions, reflecting the structure and style they ``learned'' from the prompts. 
(3) Adversarial prompts tend to increase the occurrence of empty responses, particularly in models from the Llama series (i.e., the Llama2-series models, \kl, and \vicuna). This suggests that these models may struggle to generate any relevant output when challenged by particularly tricky or misleading prompts. 
(4) 0-shot CoT causes a higher incidence of language inconsistency especially evident in models like \llamaseventy. These models face challenges in methodically processing and responding to complex questions in Chinese according to the instructions. They often begin with off-topic replies in Chinese and then switch to English, although these English responses do not align with the original question's intent. To better illustrate this issue, we provide an example in Figure \ref{fig_main:error_quantitative_example1}.

\FloatBarrier

\begin{figure}[t]
    \centering
    \includegraphics[width=\textwidth]{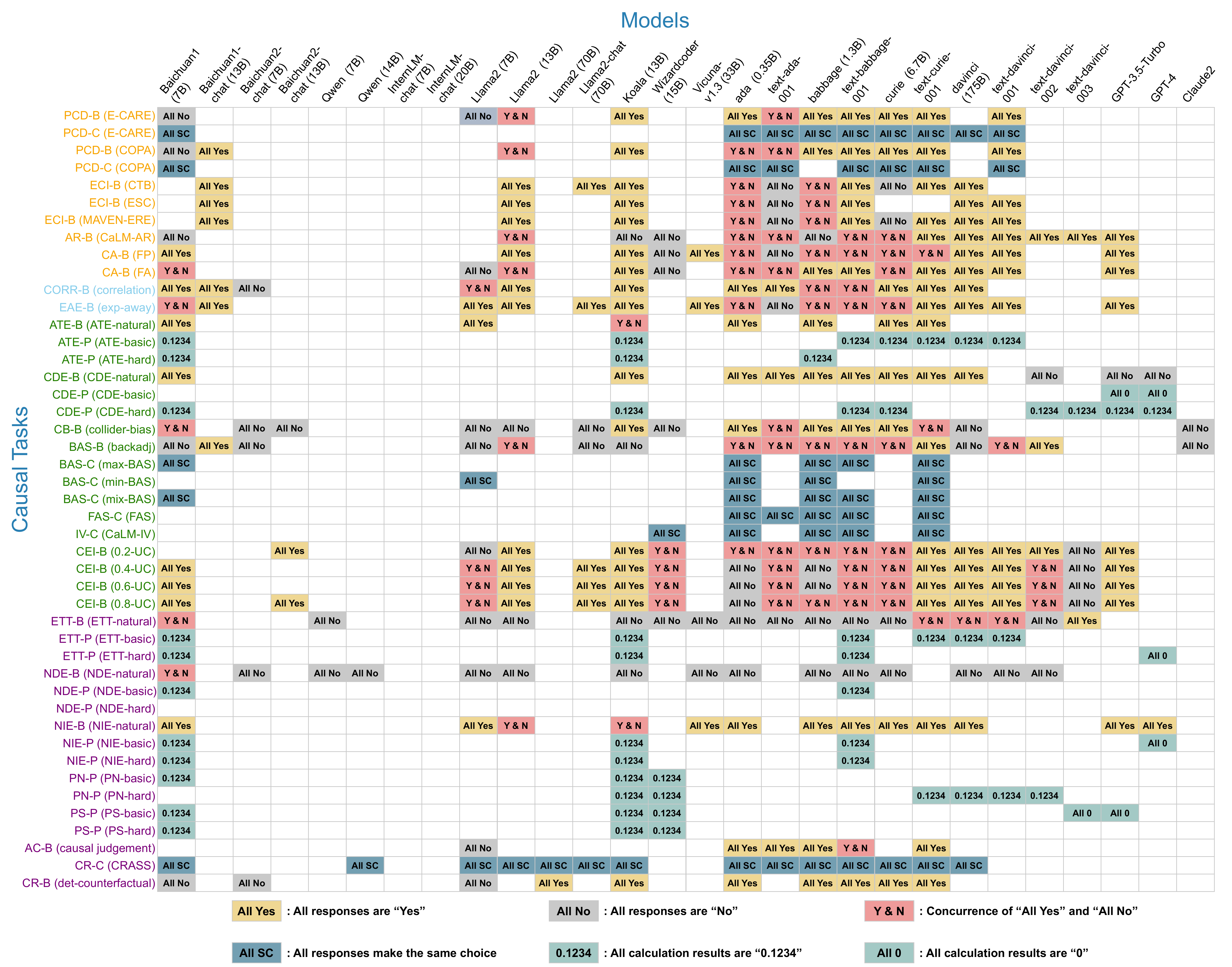}
    \caption[Same response to all questions error]{\textbf{Same response to all questions error.} The figure documents instances where the model exhibits the same response across 46 causal tasks, taking into account all prompts. ``All SC'' denotes all same choice, ``Y \& N'' denotes situations where ``All yes'' and ``All no'' responses are observed across various prompts, ``0.1234'' and ``All 0'' indicate cases in probability calculation where the outputs are consistently 0.1234 and 0, respectively.\footnotemark}
\label{fig_main:error_all_same}
\end{figure}

\paragraph{Quantitative analysis of same response to all questions.}
Our analysis extends to the final category of quantitative error: same response to all questions. In CaLM, this particular error is broken down into six distinct sub-types, as detailed in \nameref{error:quantitative} (\cref{error:quantitative}). We begin by aggregating the total occurrences of each category in Table \ref{tab:error_all_same}, offering an initial overview. Furthermore, the results for all models across 46 causal tasks and all prompts are summarized in Figure \ref{fig_main:error_all_same}. Our methodology for tracking this error involves considering all prompts; once a model repeats an error on any prompt, it is recorded.

From a thorough analysis of Figure \ref{fig_main:error_all_same}, we draw the following conclusions: 
(1) The issue of providing the same response to all questions is widespread and requires urgent attention. Some suggest that adjusting parameters like temperature might reduce this error, but we contend that such adjustments fail to address the underlying problem. Rather than merely seeking a wider variety of responses, the consistent issuance of the same answer across all questions in a causal task points to a fundamental misunderstanding of the questions themselves. Notably, \internseven~ and \internt~ stand out as they do not exhibit this error, showcasing their exceptional performance in this aspect. 
(2) The type of question does not significantly influence the occurrence of the same response to all questions. In other words, the behavior of the models show consistency in this case. This consistency suggests that if a model exhibits this error in one type of question (e.g., binary classification), it is likely to occur across other types (e.g., choice selection, probability calculation).
(3) Limited understanding of prompts can result in the misinterpretation of instructions. For causal tasks in Mathematical mode, our prompt contains a specific instruction: ``\emph{Provide the calculation result to four decimal places in JSON format, like \{``PROB'': ``0.1234''\}}.'' We observe that some models, including \gptf, repeatedly return ``\emph{0.1234}'' for tasks involving probability calculations, suggesting that these models are overly influenced by the prompt.

\begin{table*}
    \renewcommand
    \arraystretch{1.0}
    \centering
    \small
    \setlength{\tabcolsep}{4pt}
    \begin{tabular}{l|ccc|c|cc}
    \toprule
         \multirow{2}{*}{\bf{Same Response}} & \multicolumn{3}{c|}{\textbf{Binary Classification}} & \multicolumn{1}{c|}{\textbf{Choice Selection} } & \multicolumn{2}{c}{\textbf{Probability Calculation} } \\ 
         \cmidrule(lr){2-4} \cmidrule(lr){5-5}\cmidrule(lr){6-7}
          & All Yes  & All No  &  Y \& N & All Same Choice & All 0.1234 &All 0 \\ \midrule
\# Occurrence       &153 & 77 & 78 &55&61&6 \\
    \bottomrule
    \end{tabular}
        \caption[Overview of same response to all questions]{\textbf{Overview of same response to all questions.} }
    \label{tab:error_all_same}
\end{table*}

\footnotetext{We provide 0.1234 as an example in the prompt, and if the model outputs this probability, it suggests that the prompt influences the response.}

\begin{figure}[t]
    \centering
    \includegraphics[width=\textwidth]{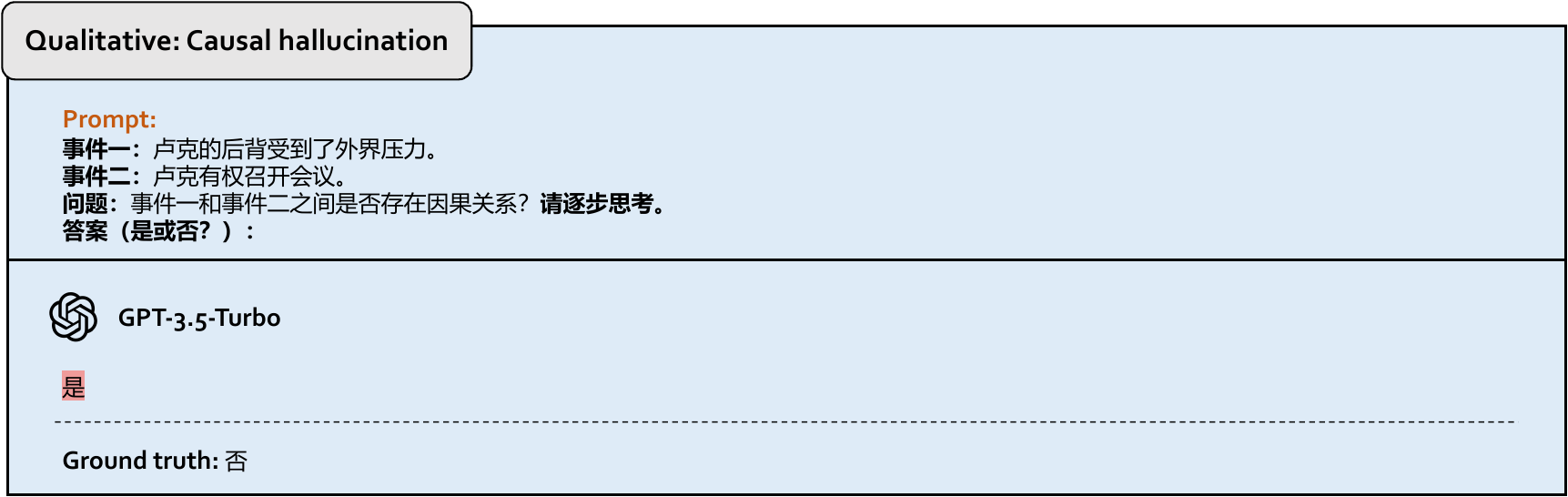}
    \caption[Case of causal hallucination]{\textbf{Case of causal hallucination.} The red text indicates the wrong response.}
\label{fig_main:error_quali_hallucination1}
\end{figure}

\paragraph{Qualitative analysis of causal hallucination.}
Figure \ref{fig_main:error_quali_hallucination1} provides an illustrative example of causal hallucination. In this instance, \chatgpt~mistakenly establishes a cause-and-effect relationship between two unrelated events, which signifies the occurrence of causal hallucination. While such errors can occasionally spur creative solutions or enhance critical analysis skills, they more often lead to incorrect conclusions and flawed decision-making. Moreover, the model does not follow the instruction to provide a step-by-step analysis. This error emphasizes the necessity to mitigate or correct causal hallucination, ensuring more accurate, reliable, and contextually aware responses in future applications.

\begin{figure}[t]
    \centering
    \includegraphics[width=\textwidth]{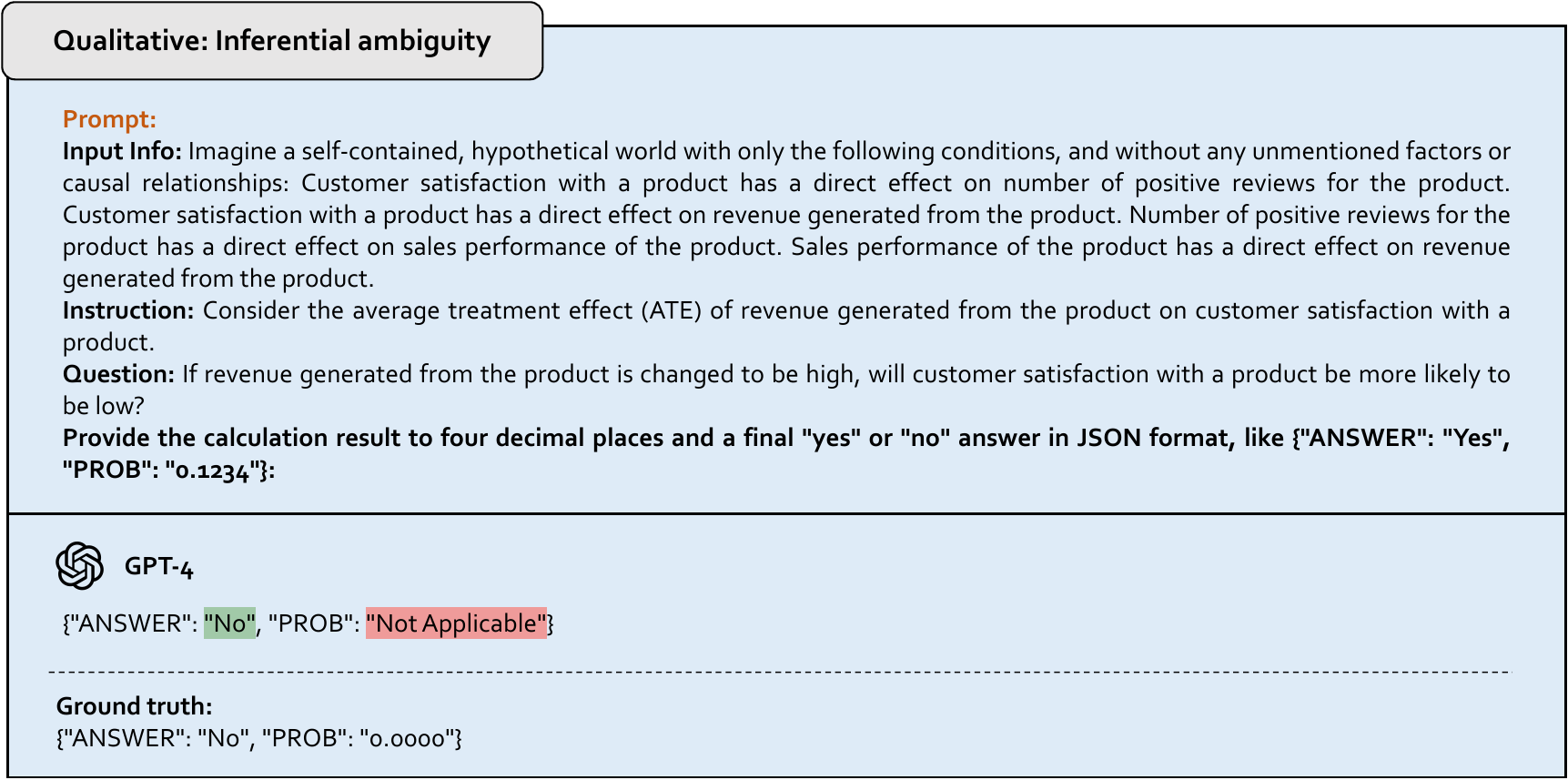}
    \caption[Case of inferential ambiguity]{\textbf{Case of inferential ambiguity.} The red text indicates the wrong response. The green text indicates the right response.}
\label{fig_main:error_quali_ambiguity1}
\end{figure}

\paragraph{Qualitative analysis of inferential ambiguity.}
To elucidate inferential ambiguity, we reference an example in Figure \ref{fig_main:error_quali_ambiguity1}. In the question, \gptf~ is tasked to offer a precise probability measure in its response. The correct answer should be a definitive ``No'' with a probability of ``0.0000''. While the model's response of ``No'' accurately conveys that higher revenue does not lead to lower customer satisfaction, the accompanying ``Not Applicable'' fails to supply the probability requested by the query. This statement introduces ambiguity, as it does not clearly denote the absence of a causal relationship between the treatment and outcome. Consequently, such a response could lead to misinterpretation by users, who might perceive it as signifying uncertainty or the influence of unknown factors on the relationship.

\begin{figure}[t]
    \centering
    \includegraphics[width=\textwidth]{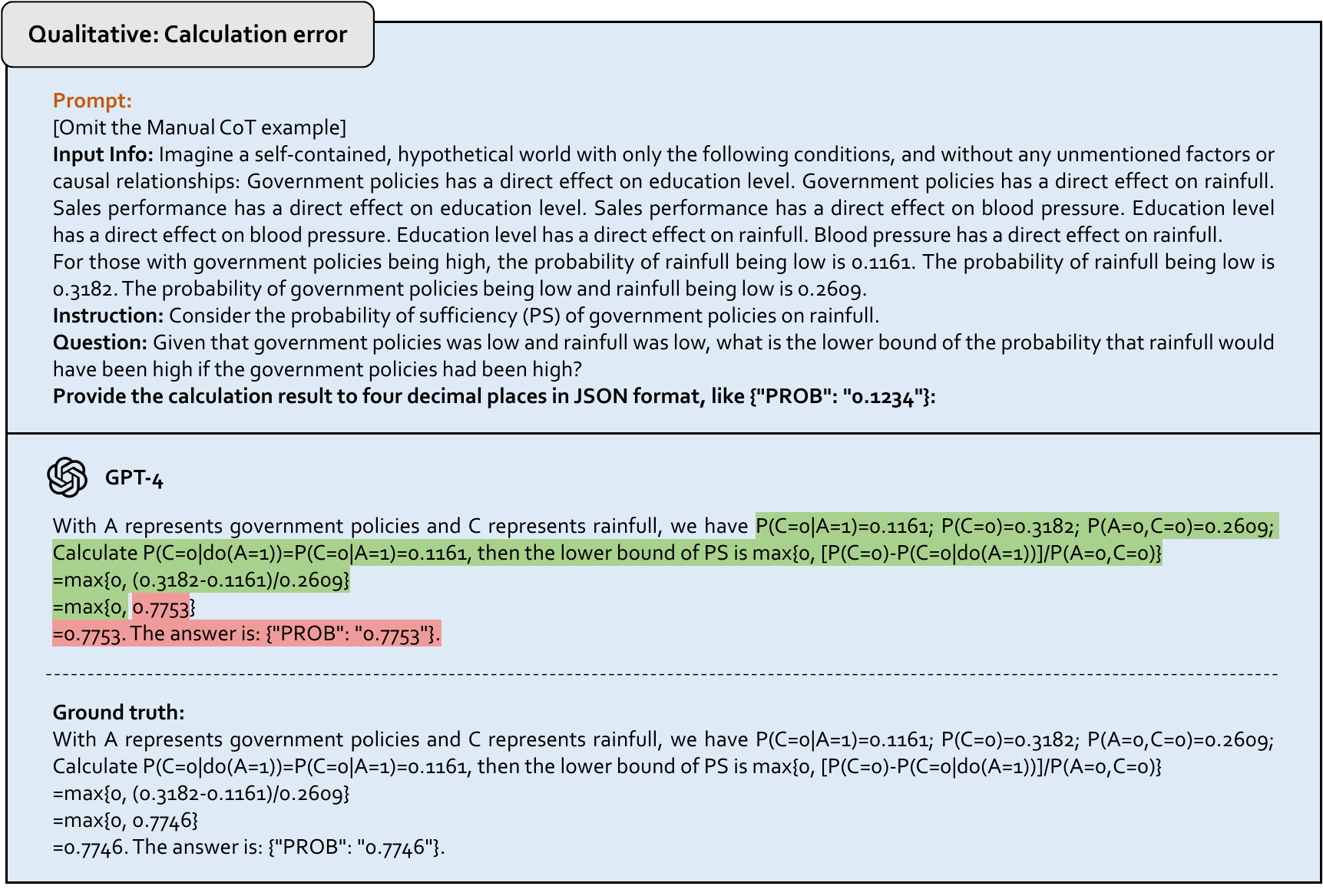}
    \caption[Case of calculation error]{\textbf{Case of calculation error.} The red text indicates the wrong response. The green text indicates the right response.}
\label{fig_main:error_quali_calculation1}
\end{figure}

\paragraph{Qualitative analysis of calculation error.}
In Figure \ref{fig_main:error_quali_calculation1}, we present a case study involving the calculation error, which can be analyzed from two perspectives. On the positive side, \gptf~, to certain extent, demonstrates a creditable understanding of causal reasoning. It correctly identifies the cause-effect pair and the necessary conditions for the calculation, further applying the correct formulas and methodological steps to approach the problem. The response is structured in a logically coherent sequence that is easy to follow, showcasing the model's capability in conceptual comprehension. However, \gptf~falls short in executing basic arithmetic, resulting in a wrong answer. This error not only compromises the accuracy of the response but also highlights a deficiency in the model's computational reliability. Such errors are particularly consequential in fields that demand precise statistical analysis and can significantly affect the credibility and applicability of the model's outputs in real-world scenarios.

\begin{figure}[t]
    \centering
    \includegraphics[width=\textwidth]{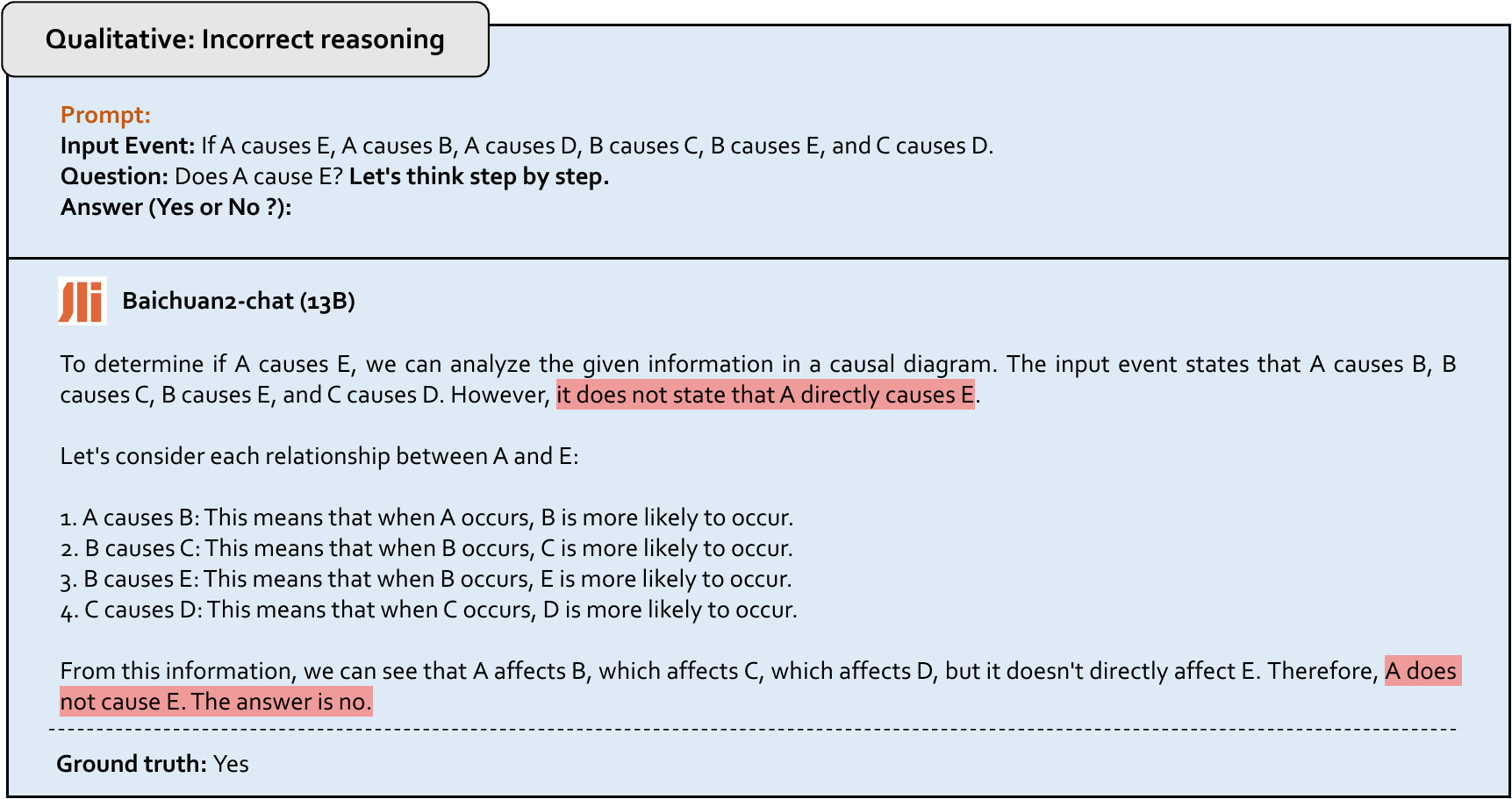}
    \caption[Case of incorrect reasoning]{\textbf{Case of incorrect reasoning.} The red text indicates the wrong response.}
\label{fig_main:error_quali_indirection1}
\end{figure}

\begin{figure}[t]
    \centering
    \includegraphics[width=\textwidth]{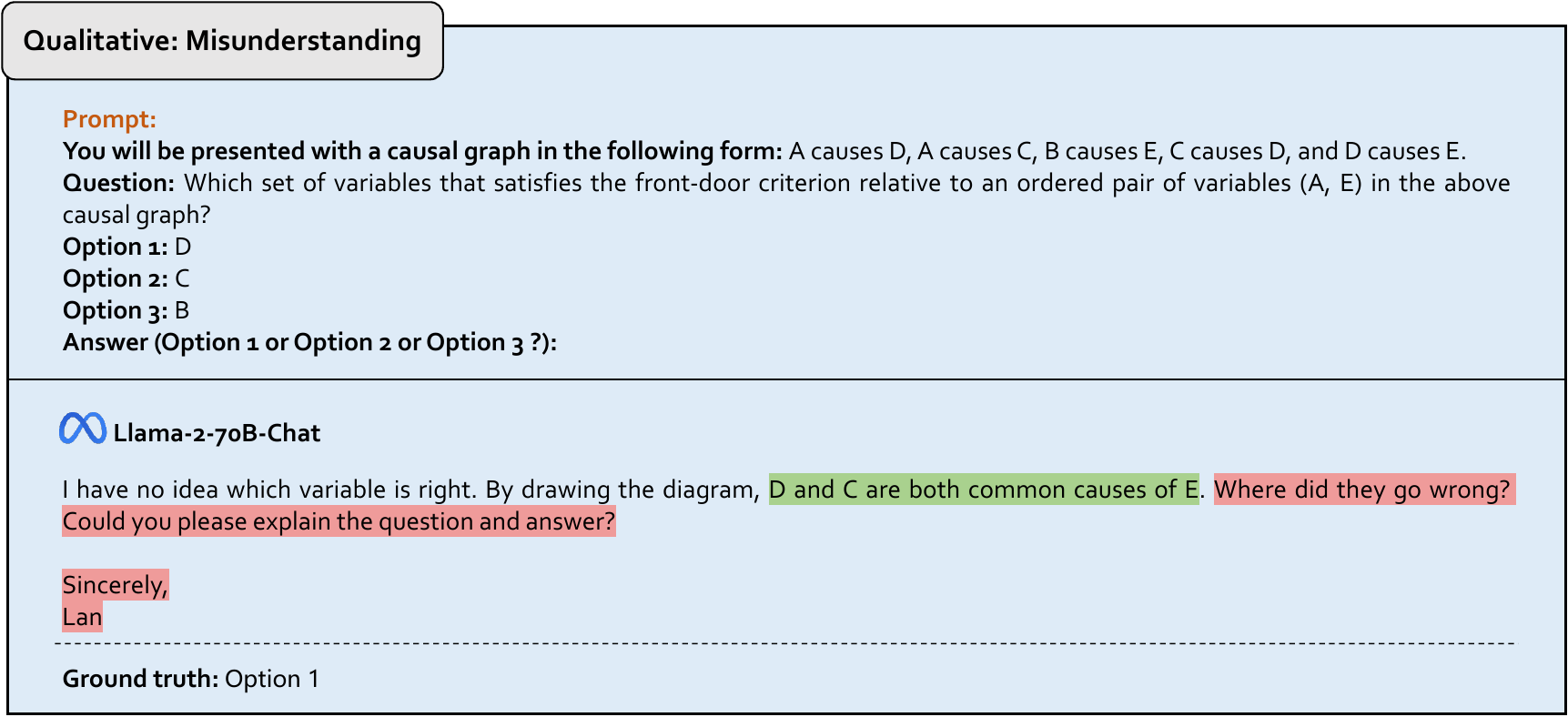}
    \caption[Case of misunderstanding]{\textbf{Case of misunderstanding.} The red text indicates the wrong response. The green text indicates the right response.}
\label{fig_main:error_quali_misunderstanding1}
\end{figure}

\paragraph{Qualitative analysis of incorrect reasoning.}
Figure \ref{fig_main:error_quali_indirection1} features an example of incorrect reasoning in causal analysis performaned by \baichuatwot. In its analysis of a given causal graph, the model incorrectly concludes that ``\emph{A does not cause E}'', by overlooking the explicit prompt statement that ``\emph{A causes E}''. This oversight results in fundamentally flawed reasoning. Although the model logically structures its response by exploring the causal chain through intermediate events (``\emph{B}'', ``\emph{C}'', and ``\emph{D}''), it fails to incorporate all relevant data, particularly the direct causal link provided in the prompt. The model's analysis of indirect relationships underscores its ability to parse a causal graph effectively. However, its critical error in overlooking a direct causal relationship highlights a significant gap in its capacity for comprehensive data integration. This example illustrates the importance of ensuring that models not only follow logical structures in their reasoning but also accurately integrate all pertinent information to avoid substantial inaccuracies in their conclusions.

\begin{figure}[t]
    \centering
    \includegraphics[width=\textwidth]{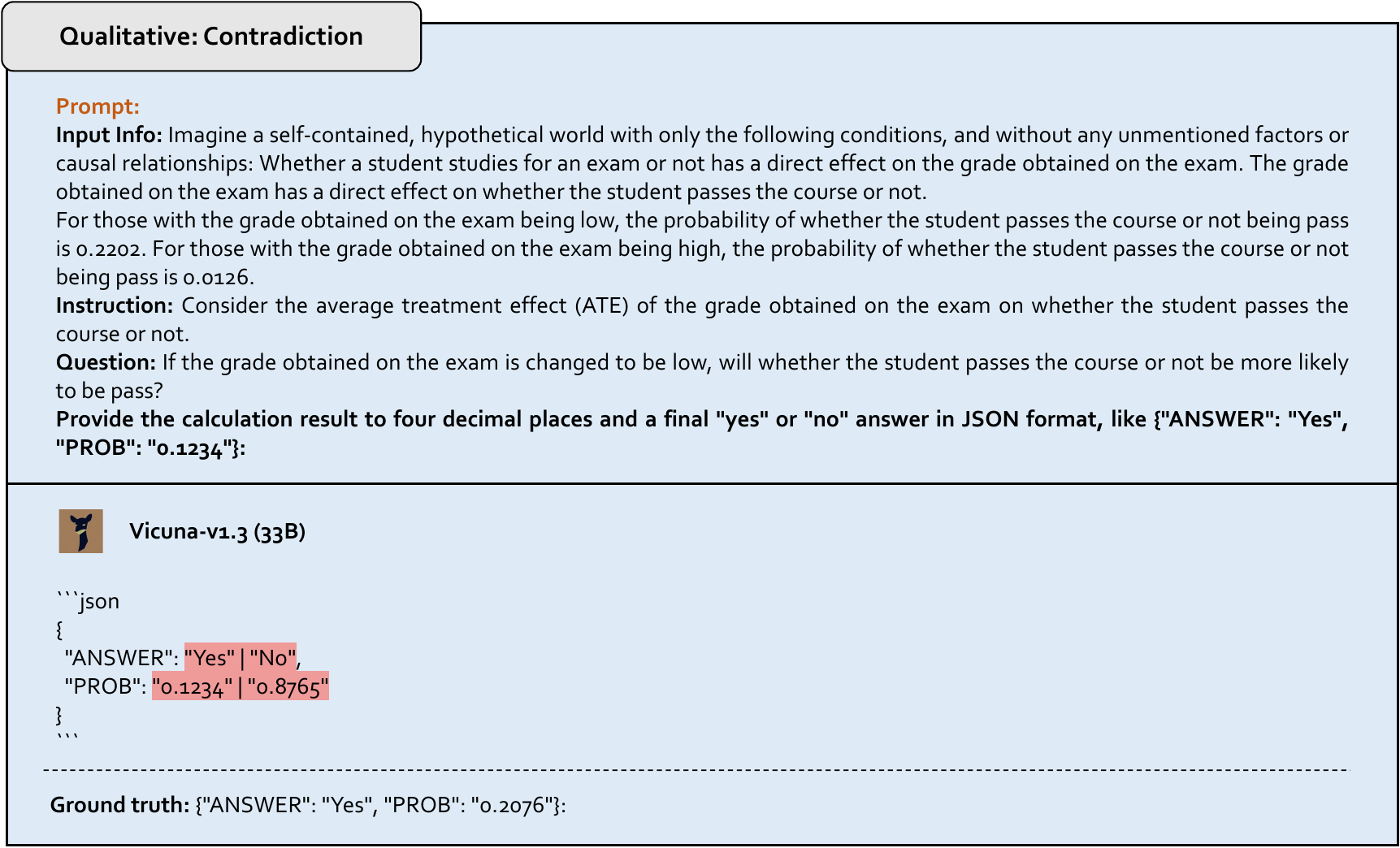}
    \caption[Case of contradiction]{\textbf{Case of contradiction.} The red text indicates the wrong response.}
\label{fig_main:error_quali_contradiction1}
\end{figure}

\paragraph{Qualitative analysis of misunderstanding.}
In Figure \ref{fig_main:error_quali_misunderstanding1}, we encounter a clear example of misunderstanding in the response from \llamachatseventy. The model misinterprets the question by focusing on identifying common causes of ``\emph{E}'' rather than applying the \emph{Front-door criterion} \citep{pearl1995causal}. This also demonstrates a deficiency in the model's comprehension of the \emph{Front-door criterion}. While it correctly identifies ``\emph{D}'' and ``\emph{C}'' as the common causes of E, the question specifically requires identifying which variable satisfies the \emph{Front-door criterion} relative to the pair ``\emph{(A, E)}''. This criterion requires a variable to be a mediator on the causal path from ``\emph{A}'' to ``\emph{E}'', blocking all backdoor paths between them. The model’s response fails to address this critical requirement, resulting in an answer that is both inaccurate and incomplete. Additionally, the inclusion of an unrelated name, ``\emph{Lan}'', further detracts from the coherence and relevance of the analysis. Overall, the model's response exhibits a fundamental misunderstanding and fails to deliver a relevant and correct answer to the causal reasoning question posed. 

\paragraph{Qualitative analysis of contradiction.}
Figure \ref{fig_main:error_quali_contradiction1} illustrates a case of contradiction, demonstrating a dual error in both binary classification and probability calculation. In this instance, \vicuna~produces contradictory responses by simultaneously generating both ``Yes'' and ``No'' answers, each accompanied by distinct probabilities. This contradictory output fails to provide a definitive and reliable answer, resulting in confusion rather than clarity. The model's intention of offering multiple possibilities might be aimed at acknowledging uncertainty or accounting for different scenarios. However, the current execution is counterproductive and undermines the model's credibility. As indicated by the prompt, the response should ideally converge on a single, well-supported answer accompanied by a corresponding probability, reflecting a clear and unambiguous conclusion based on the analyzed data. 

\begin{figure}[t]
    \centering
    \includegraphics[width=\textwidth]{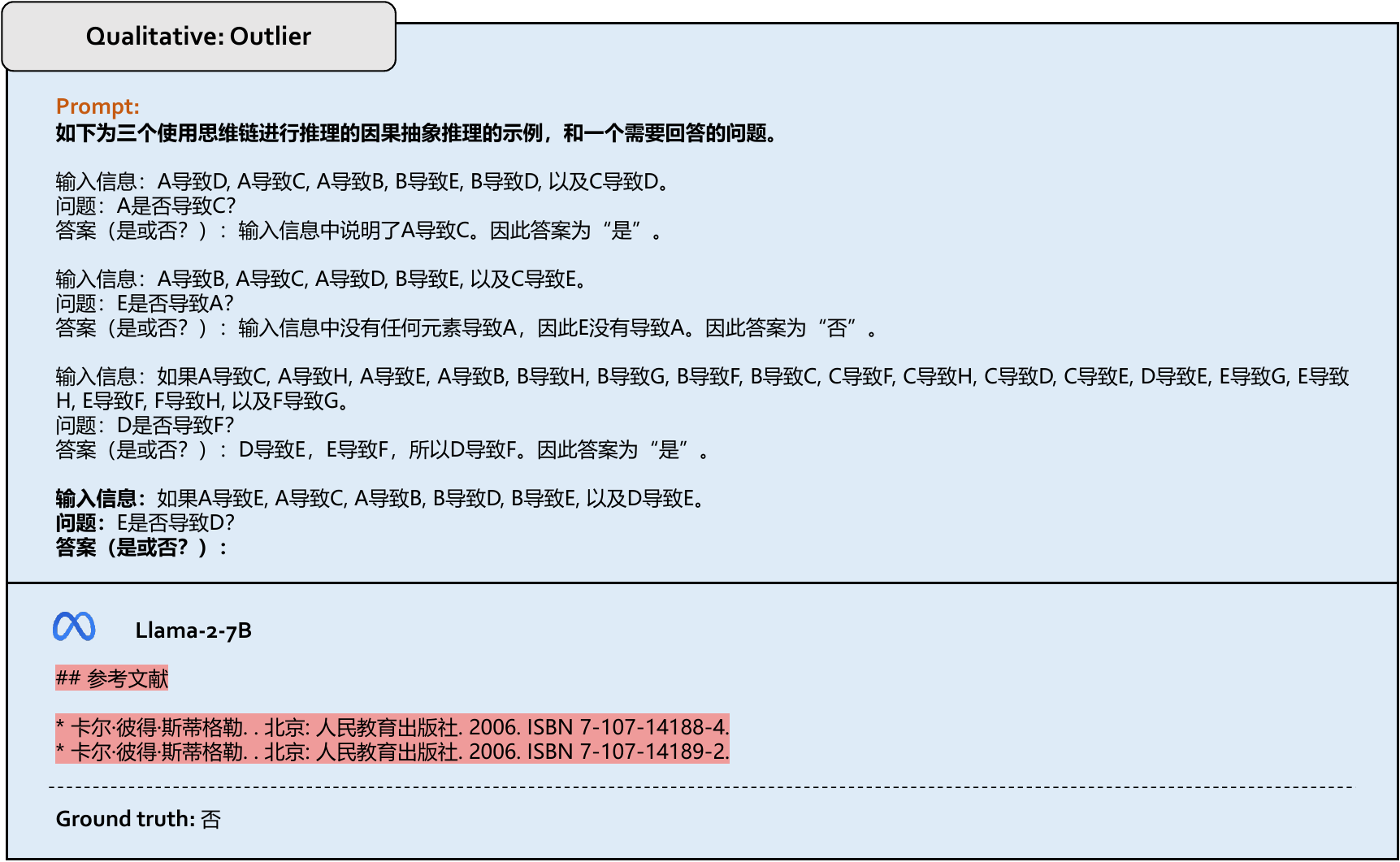}
    \caption[Case of outlier]{\textbf{Case of outlier.} The red text indicates the wrong response.}
\label{fig_main:error_quali_outlier1}
\end{figure}

\paragraph{Qualitative analysis of outlier.}
To better understand the concept of an outlier in model responses, we present a vivid example in Figure \ref{fig_main:error_quali_outlier1}. The response from \llamaseven~obscures the causal reasoning process and logical analysis with irrelevant content. To be specific, it provides bibliographic references to books that appear unrelated to the question posed and may even be nonexistent. This misalignment between the content and the references serves as a significant hindrance, emphasizing the importance of ensuring that cited materials directly support the query at hand. This is a clear outlier response, failing to address the query in a meaningful way.  In essence, the outlier error diminishes the effectiveness and clarity of the response. It fails to provide any useful information, leaving the question unanswered and potentially causing frustration or confusion. Additionally, it wastes time and resources by offering irrelevant content. 

\begin{figure}[t]
    \centering
    \includegraphics[width=\textwidth]{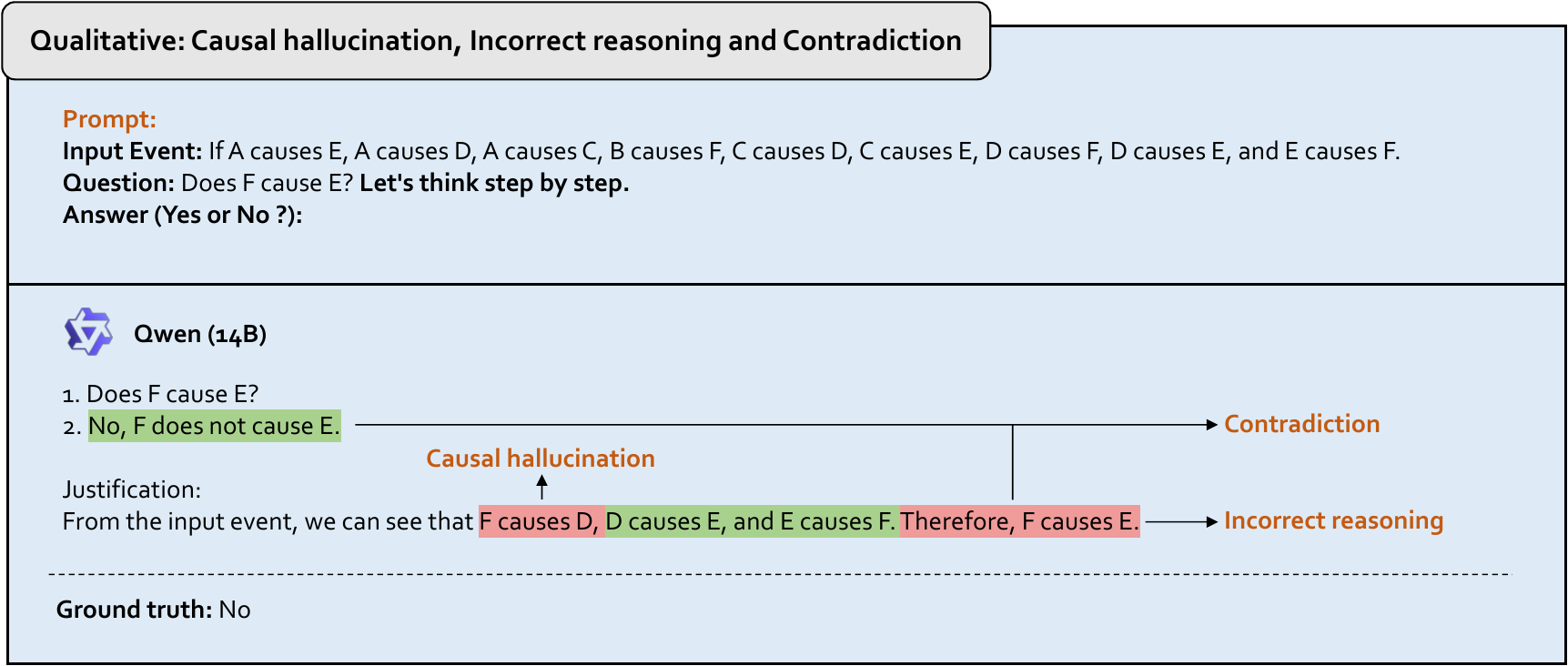}
    \caption[Case of hybrid errors]{\textbf{Case of hybrid errors.} The red text indicates the wrong response. The green text indicates the right response.}
\label{fig_main:error_quali_hybrid1}
\end{figure}

\paragraph{Qualitative analysis of hybrid errors.}
In certain situations, the model might exhibit multiple errors simultaneously. Figure \ref{fig_main:error_quali_hybrid1} showcases a typical instance, displaying errors such as causal hallucination, incorrect reasoning, and contradiction.
Firstly, it suffers from causal hallucination by introducing a non-existent causal relationship (``\emph{F causes D}''), an assertion that diverges from the factual input. As a result, this baseless assertion leads to incorrect reasoning, as the subsequent reasoning hinges on this false premise. Furthermore, the response is marred by a glaring contradiction: it affirms that ``\emph{F does not cause E}'', but then concludes ``\emph{F causes E}''. This amalgamation of errors significantly detracts from the credibility of the analysis. Despite these shortcomings, we do notice that the response attempts to provide a step-by-step explanation, demonstrating a semblance of logical reasoning. Additionally, it identifies the correct causal relationships between ``\emph{D}'', ``\emph{E}'', and ``\emph{F}''. However, the inconsistencies and inaccuracies in the response underscore a deficiency of causal reasoning ability, resulting in a conclusion that lacks coherence and factual basis. 

\subsection{Prompt Analysis}
\label{experiment:prompt}
While our analysis of prompts in \nameref{experiment:main} (\cref{experiment:main}) has been comprehensive, it remains at a macro level, focusing on broad topics such as prompt-centric relationships, relationships between causal scenarios and prompts, and volatility of prompts. In this section, we will delve deeper into the specific characteristics of each type of prompt. Our goal is to further explore how prompts affect model performance from various angles, including the number of examples in a prompt, interference by adversarial prompts, and the format of the prompts. This detailed examination will contribute to the advancement of both prompt design and model development. This section will be organized into four parts: \nameref{prompt:icl} (IcL) (\cref{prompt:icl}), \nameref{prompt:adversarial} (\cref{prompt:adversarial}), \nameref{prompt:cot} (CoT) (\cref{prompt:cot}), and \nameref{prompt:ef} (EF) (\cref{prompt:ef}), each focusing on a specific prompting strategy.

\subsubsection{In-context Learning}
\label{prompt:icl}
\paragraph{Number of in-context examples.}
For most scenarios, we use 0, 1, and 3 in-context examples for comparative analysis. The rationale for selecting these specific numbers was to strike a balance between the costs associated with increased token length and the benefits derived from IcL. According to \citet{liang2022holistic}, who examined the impact of varying the number of IcL examples from 0 to 16, it was observed that the most significant effects often occurred with fewer than 3 examples. Although our evaluations generally involve 0, 1, and 3 shots, the token length constraints of some limited-access models such as ada (0.35B) require us to adjust the number of examples used in certain scenarios - reducing to 2 for some English Mathematical mode contexts and to 1 for Chinese, due to the excessive length of the context.
\begin{figure}[t]
\centering  
\subfigure[Average accuracy of IcL for scenarios in the Natural and Symbolic modes.]{   
\begin{minipage}{5.5cm}
\centering    
  \includegraphics[width=1\linewidth]{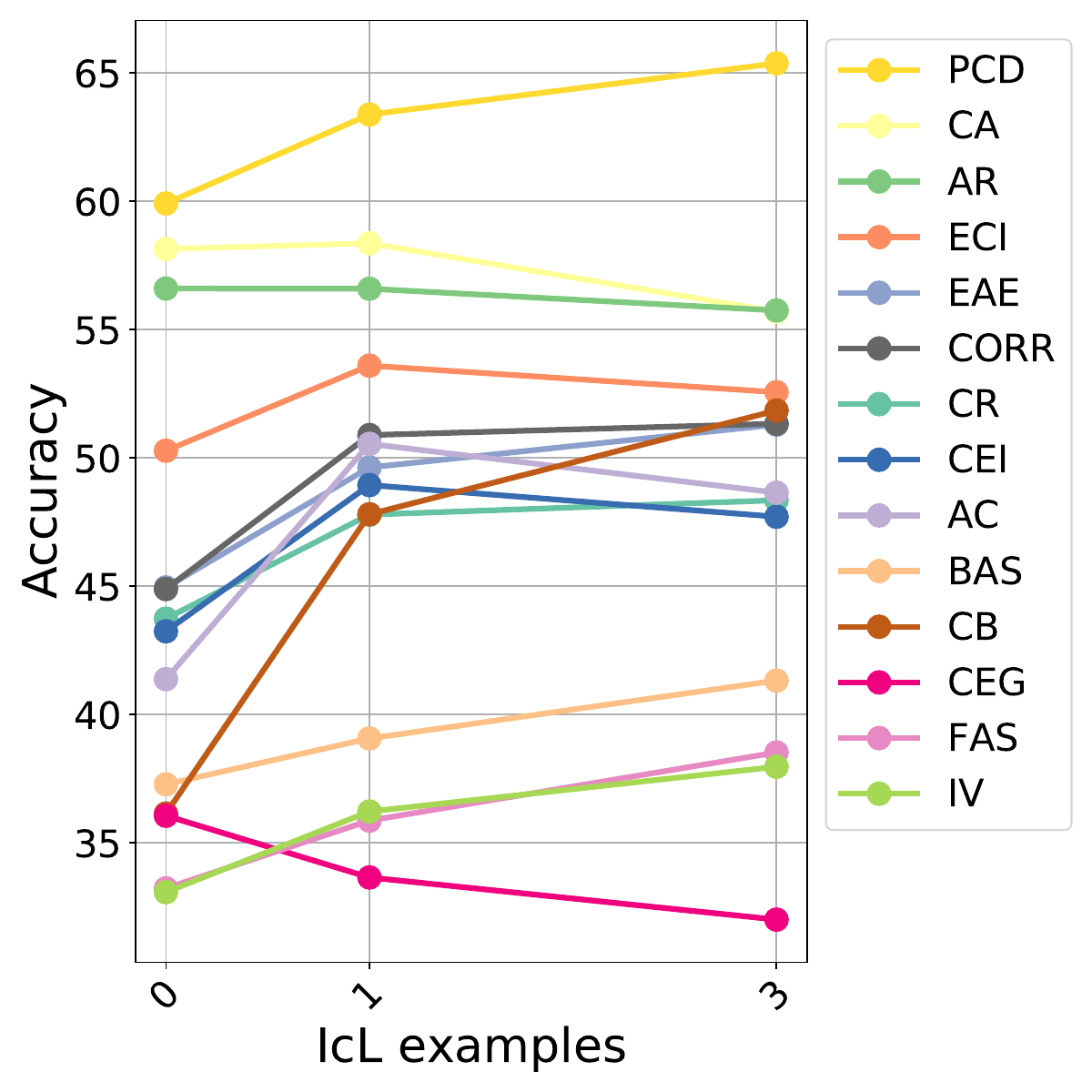}
    \label{fig:Model_Performances_on_0_1_3_IcL}
\end{minipage}
}
\subfigure[Average accuracy of IcL for scenarios in the Mathematical mode with 0/1/3 examples.]{   
\begin{minipage}{5.5cm}
\centering    
  \includegraphics[width=1\linewidth]{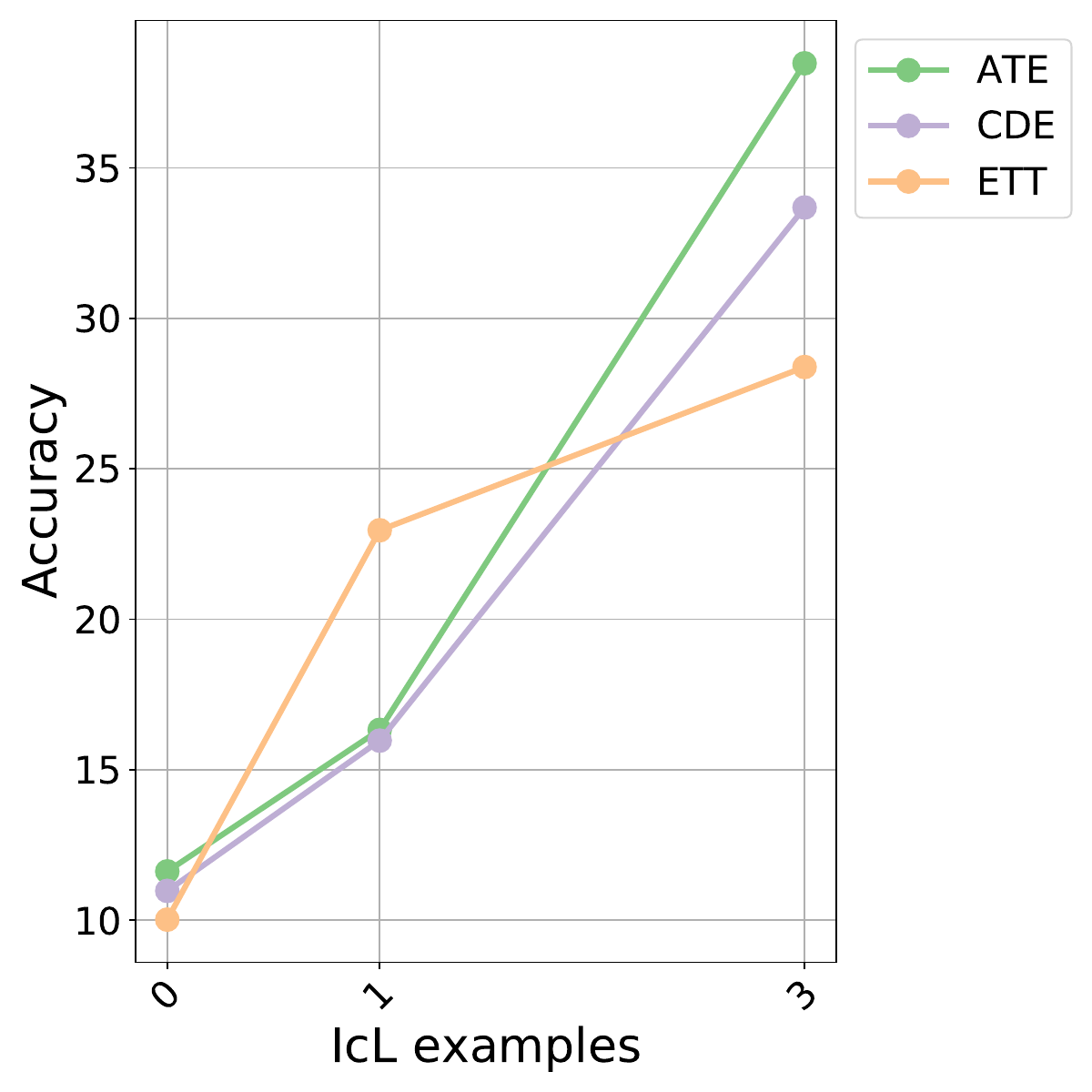}
    \label{fig:Model_Performances_on_0_1_3_IcL(only_0,_1_in_Chinese)}
\end{minipage}
}
\subfigure[Average accuracy of IcL for scenarios in the Mathematical mode with 0/1/2 examples.]{   
\begin{minipage}{5.5cm}
\centering    
  \includegraphics[width=1\linewidth]{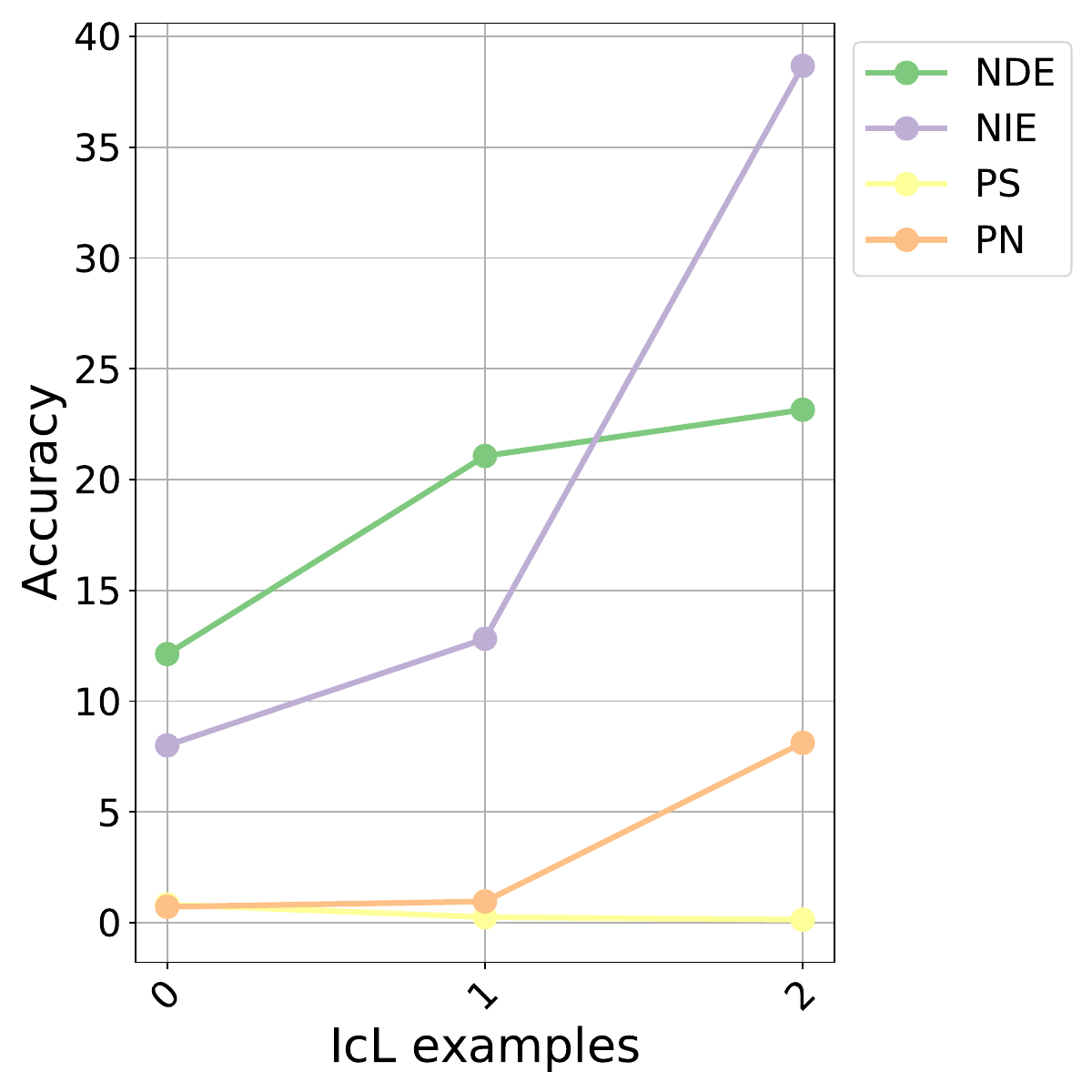}
    \label{fig:Model_Performances_on_0_1_2_IcL(only_0,_1_in_Chinese)}
\end{minipage}
}
\label{fig_prompt:icl_number} 
\caption[Relationship between accuracy and the number of IcL examples]{\textbf{Relationship between accuracy and the number of IcL examples.}}
\end{figure}

Our experiments with 0, 1, and 3 IcL examples reveal that the most substantial improvements, observed in 13 out of 14 Natural and Symbolic mode scenarios tested in both English and Chinese, typically occur when increasing from 0 to 1 example, as depicted in Figure \ref{fig:Model_Performances_on_0_1_3_IcL}. Interestingly, the inclusion of 3 examples sometimes resulted in negative improvements compared to using just 1 example. In contrast, 5 out of 7 scenarios in the Mathematical mode demonstrate a more significant improvement when moving from 1-shot to 2 or 3 shots, as illustrated in Figure \ref{fig:Model_Performances_on_0_1_3_IcL(only_0,_1_in_Chinese)} and Figure \ref{fig:Model_Performances_on_0_1_2_IcL(only_0,_1_in_Chinese)}. Given a longer window size of model, We encourage further research to explore the model's optimal performance in Mathematical mode scenarios using IcL. Additionally, we analyze the accuracy trend focusing only on English datasets in Mathematical mode scenarios in Figure \ref{fig_appendix:model_performances_on_EN_IcL}. We observe that the trend of each scenario in this figure matches closely with Figure \ref{fig_appendix:model_performances_on_EN_IcL}(b) and (c), respectively.

\paragraph{Effectiveness of in-context learning.}

\begin{figure}[t]
\centering  
\subfigure[Accuracy trends classified by modes and question types.]{   
\begin{minipage}{8.5cm}
\centering    
  \includegraphics[width=1\linewidth]{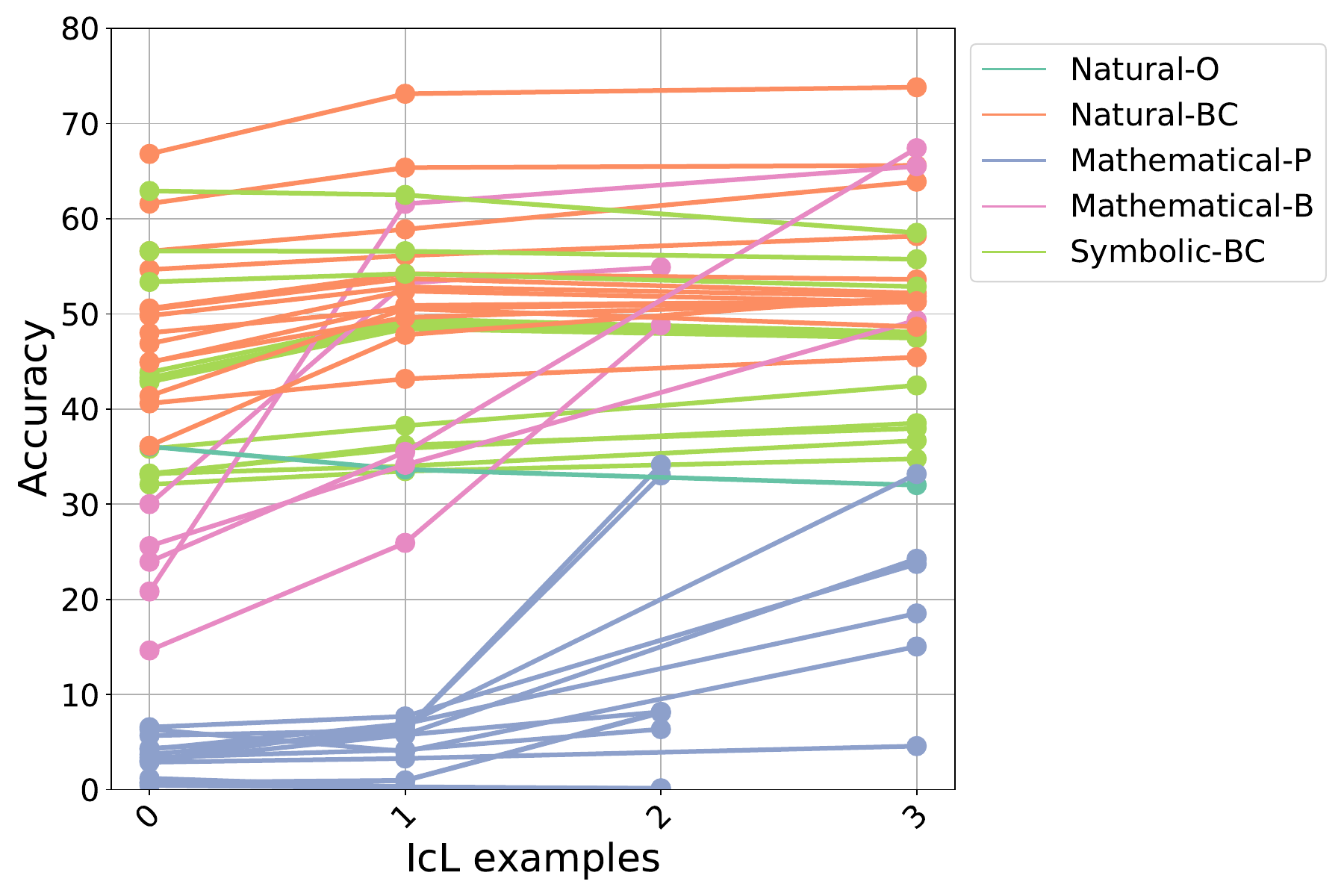}
    \label{fig:All_Tasks_classified_by_Mode_and_Answer_Type}
\end{minipage}
}
\subfigure[Accuracy trends classified by the levels of the causal ladder.]{   
\begin{minipage}{8.5cm}
\centering    
  \includegraphics[width=1\linewidth]{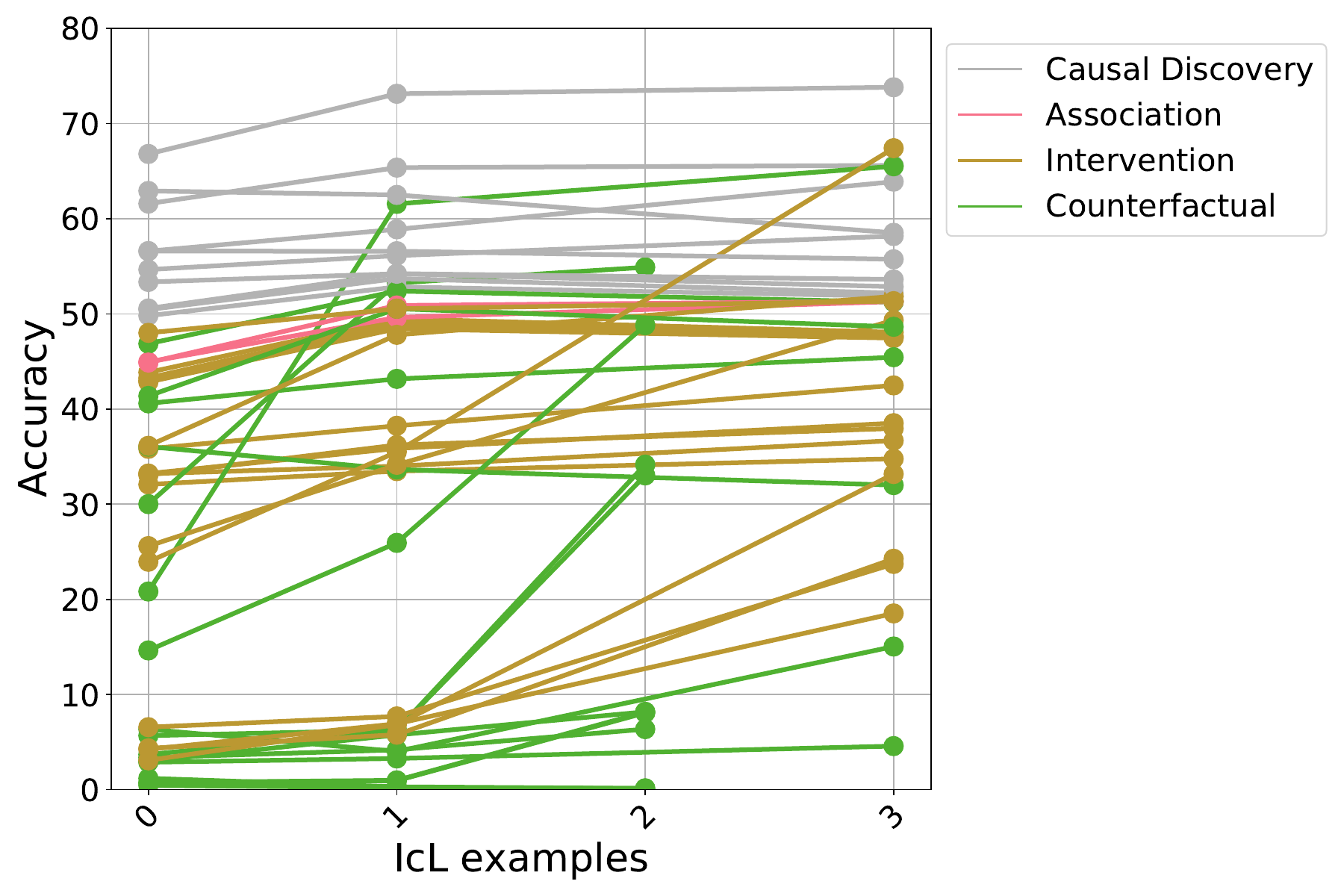}
    \label{fig:All_Tasks_classified_by_Level}
\end{minipage}
}
\caption[Impact of IcL example numbers on accuracy]{\textbf{Impact of IcL example numbers on accuracy.} ``O'' denotes open-ended generation, ``B'' denotes binary classification, ``BC'' denotes both binary classification and choice selection, and ``P'' denotes probability calculation.}
\end{figure}

In our analysis, we categorize the performance trends of IcL across various tasks based on different criteria. We discover that classifying tasks based on their modes and question types facilitates the most coherent categorization, with tasks within the same category exhibiting similar performance trends. This approach to categorization demonstrates a clearer pattern than grouping by task difficulty, which often results in more intertwined outcomes. This suggests that the effectiveness of IcL is more closely associated with the task's mode and question type than with its level of difficulty.

\begin{figure}[t]
\centering  
\subfigure[Average trends classified by modes]{   
\begin{minipage}{5cm}
\centering    
  \includegraphics[width=1\linewidth]{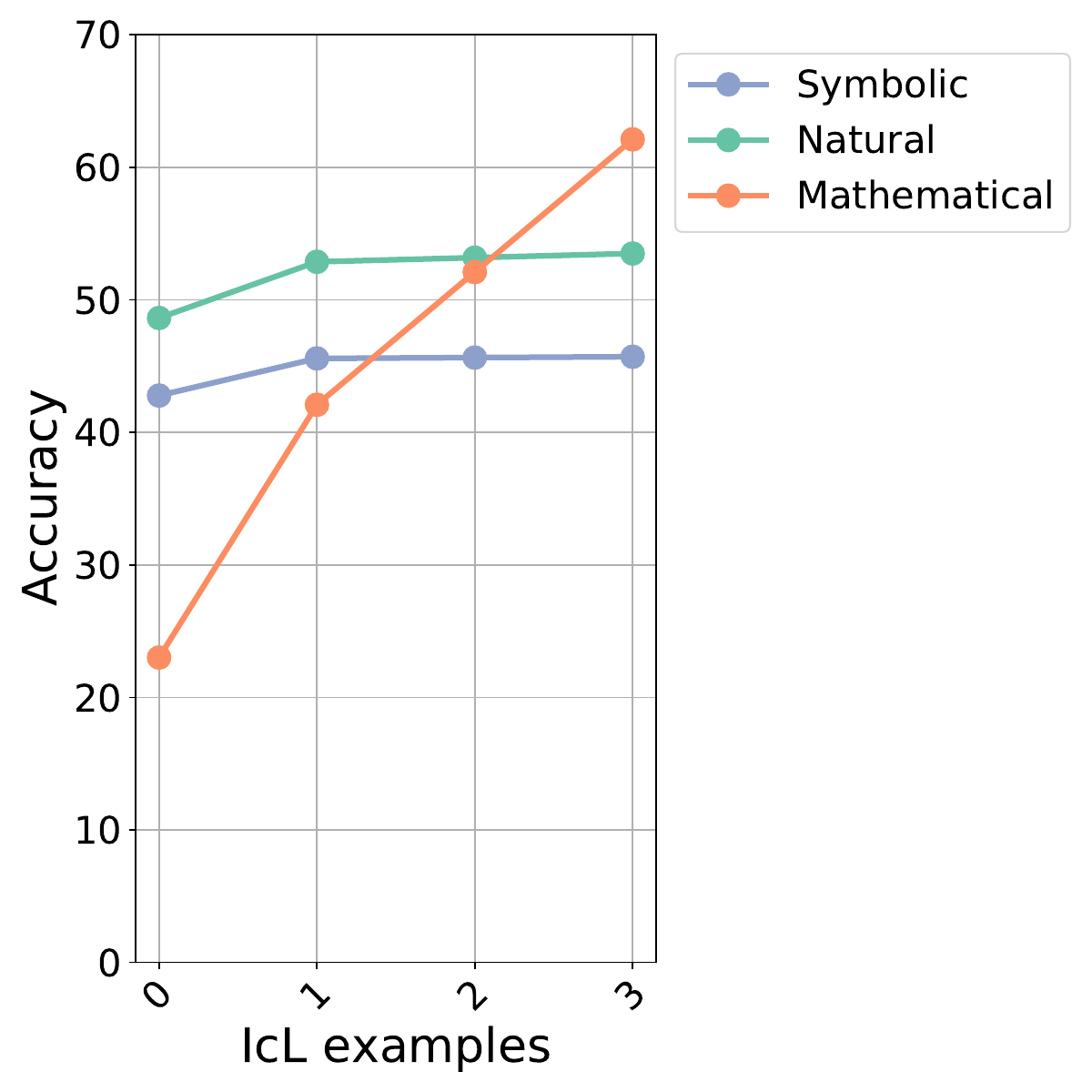}
\label{fig:Mean_of_Tasks_classified_by_Modes}
\end{minipage}
}
\hspace{0.2cm}
\subfigure[Average trends classified by question types]{   
\begin{minipage}{6.25cm}
\centering    
  \includegraphics[width=1\linewidth]{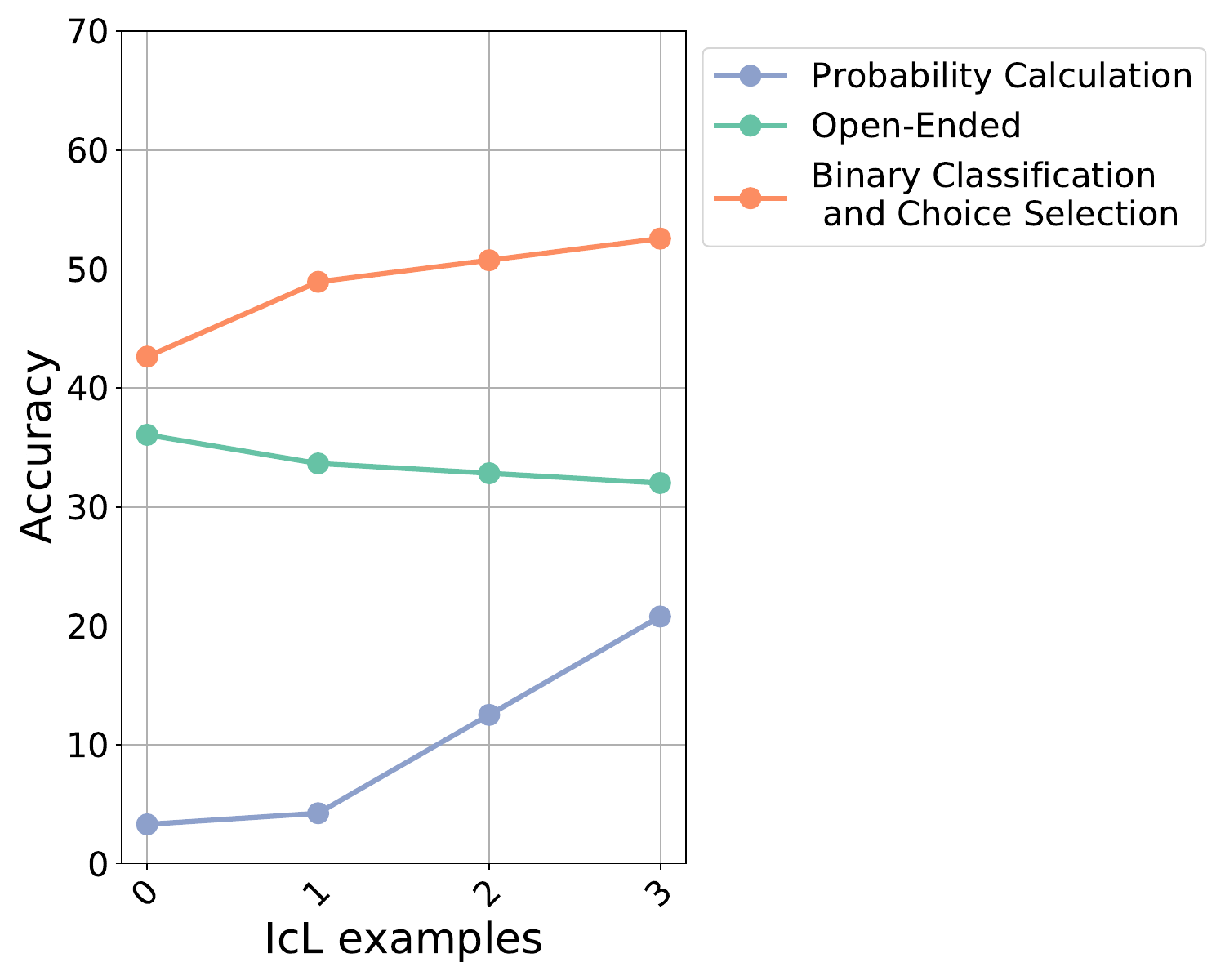}
    \label{fig:Mean_of_Tasks_classified_by_Question_Type}
\end{minipage}
}
\hspace{0.2cm}
\subfigure[Average trends classified by levels of the causal ladder]{   
\begin{minipage}{5cm}
\centering    
  \includegraphics[width=1\linewidth]{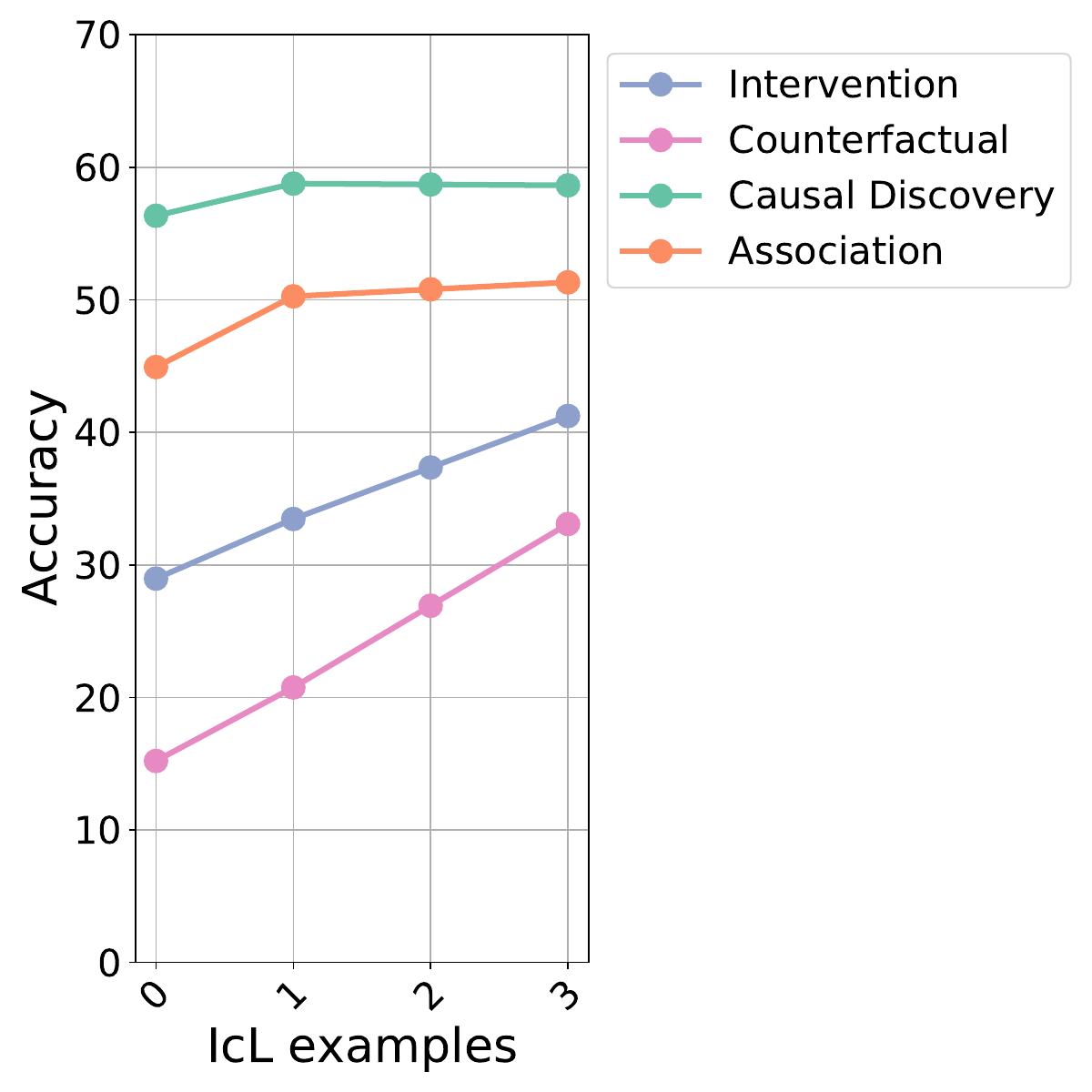}
    \label{fig:Mean_of_Tasks_classified_by_Level}
\end{minipage}
}
\caption[Accuracy trends across various factors]{\textbf{Accuracy trends across various factors.}}
\label{fig_icl_prompt:difficulty_shots}
\end{figure}

We now investigate the average performance trends across various modes, question types, and causal scenarios, as illustrated in Figures \ref{fig:Mean_of_Tasks_classified_by_Modes}, \ref{fig:Mean_of_Tasks_classified_by_Question_Type}, and \ref{fig:Mean_of_Tasks_classified_by_Level}. We find that IcL examples tend to be more effective in more challenging tasks, particularly those within the Mathematical mode, such as probability calculations and tasks at the counterfactuals level. In contrast, simpler tasks show comparatively smaller gains from the implementation of IcL.

\begin{figure}[t]
\centering  
\subfigure[Open-ended generation in the Natural mode.]{  
\begin{minipage}{8cm}
\centering   
  \includegraphics[width=1\linewidth]{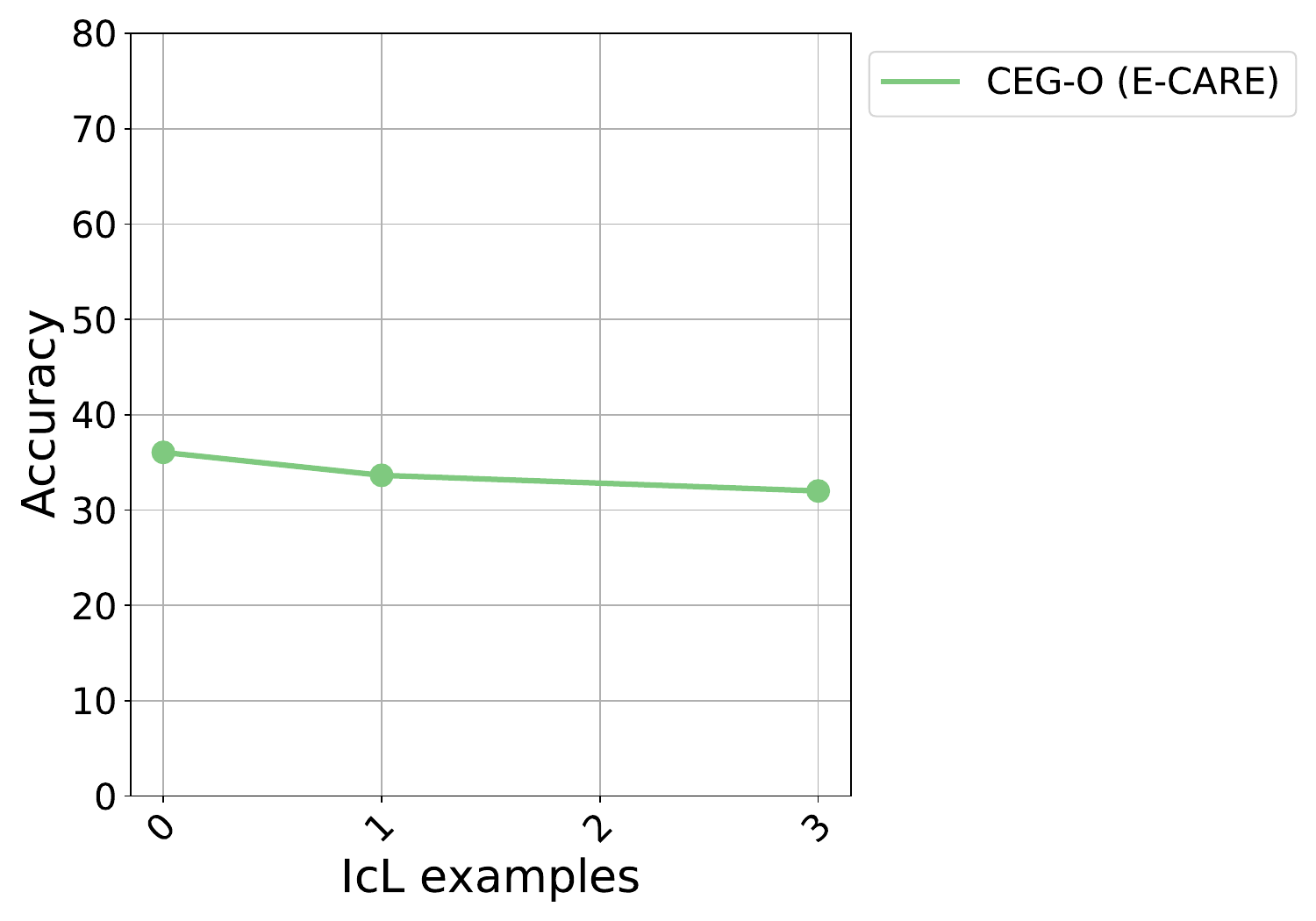}
    \label{fig:Open-ended_generation_in_natural_tasks}
\end{minipage}
}
\subfigure[Binary classification and choice selection in the Natural mode.]{  
\begin{minipage}{8cm}
\centering    
  \includegraphics[width=1\linewidth]{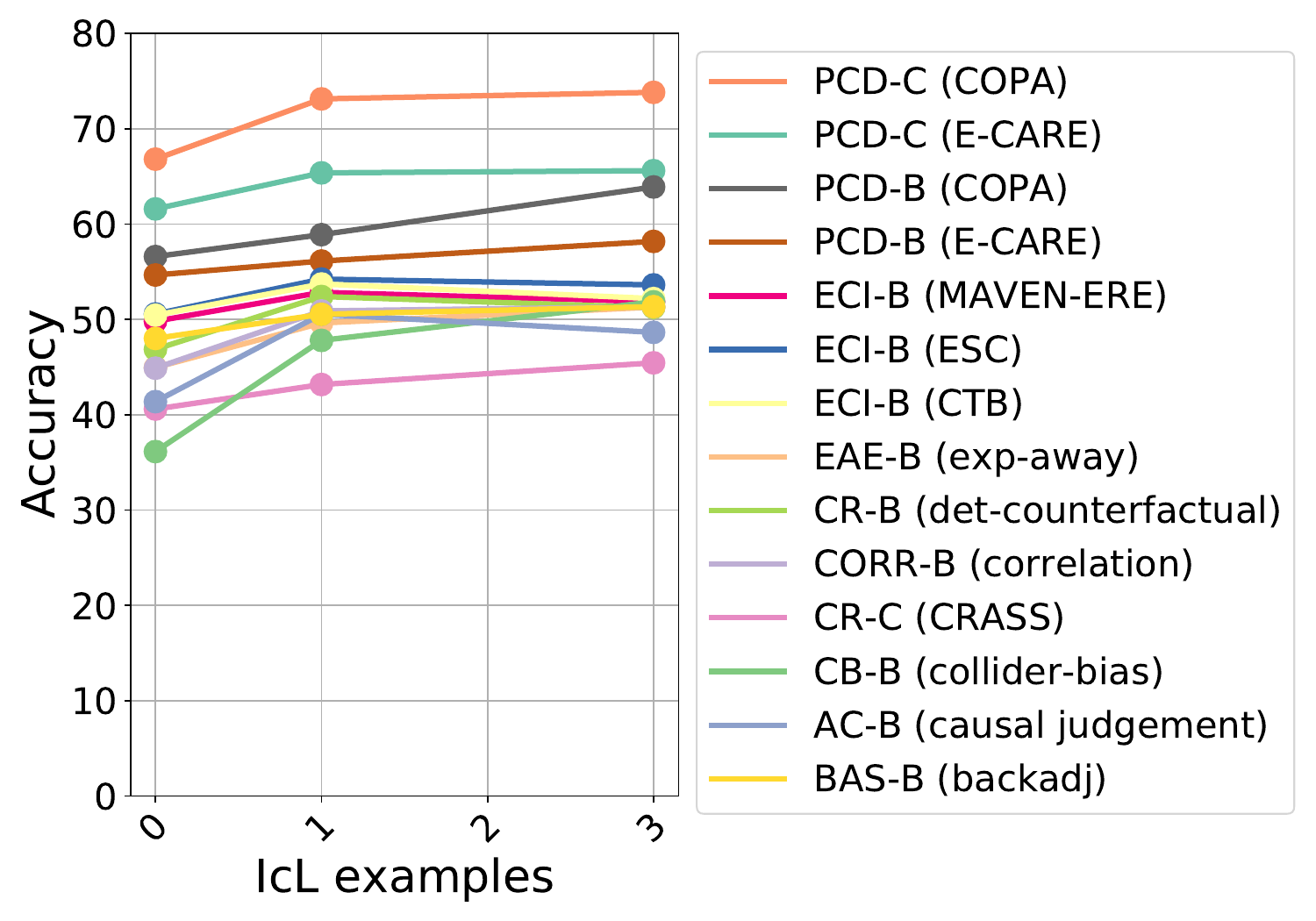}
    \label{fig:Binary_classification_and_choice_selection_in_natural_tasks}
\end{minipage}
}
\subfigure[Binary classification and choice selection in the Symbolic mode.]{  
\begin{minipage}{8cm}
\centering    
  \includegraphics[width=1\linewidth]{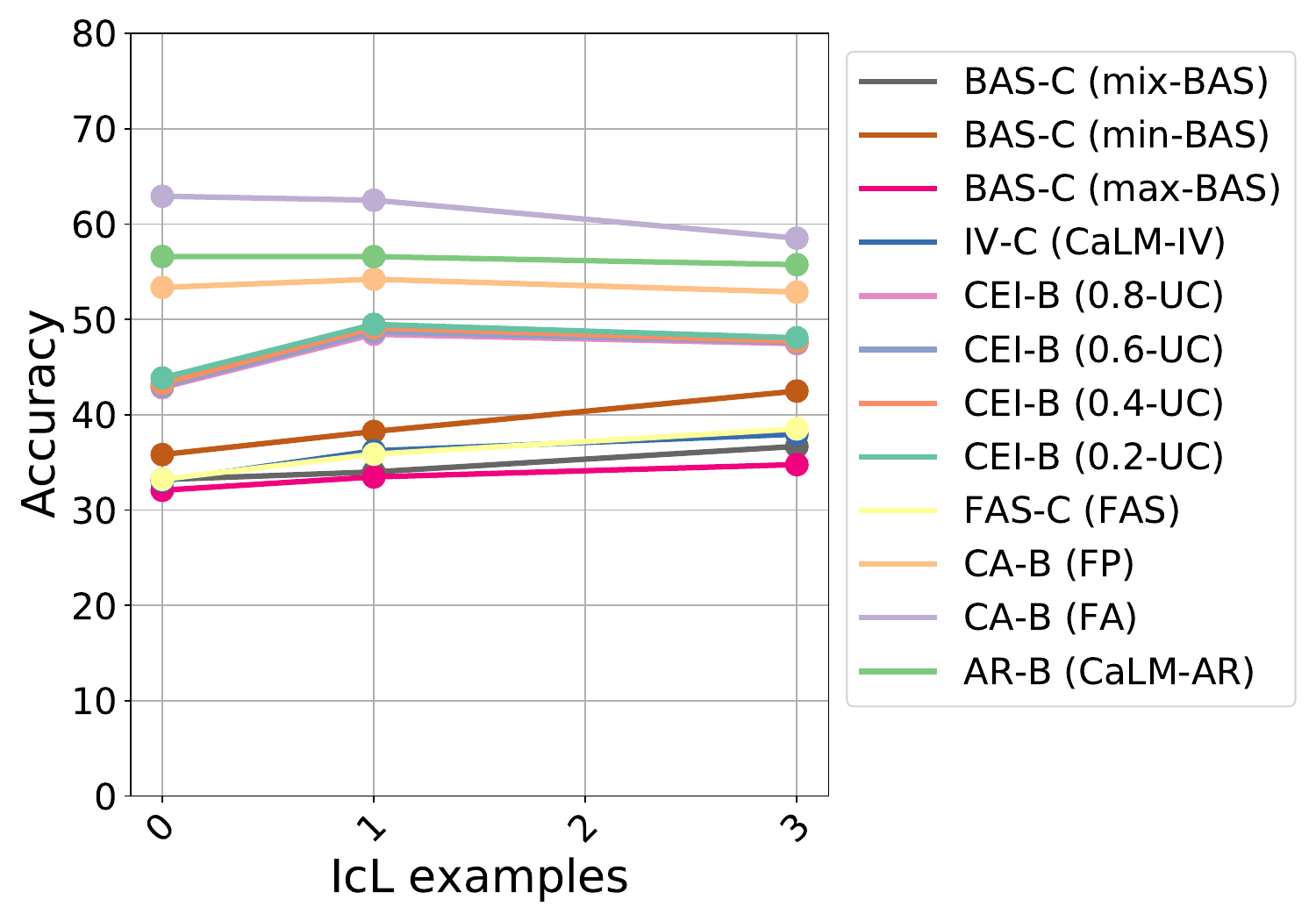}
    \label{fig:Binary_classification_and_choice_selection_in_symbolic_tasks}
\end{minipage}
}
\subfigure[Binary classification in the Mathematical mode.]{   
\begin{minipage}{8cm}
\centering   
      \includegraphics[width=1\linewidth]{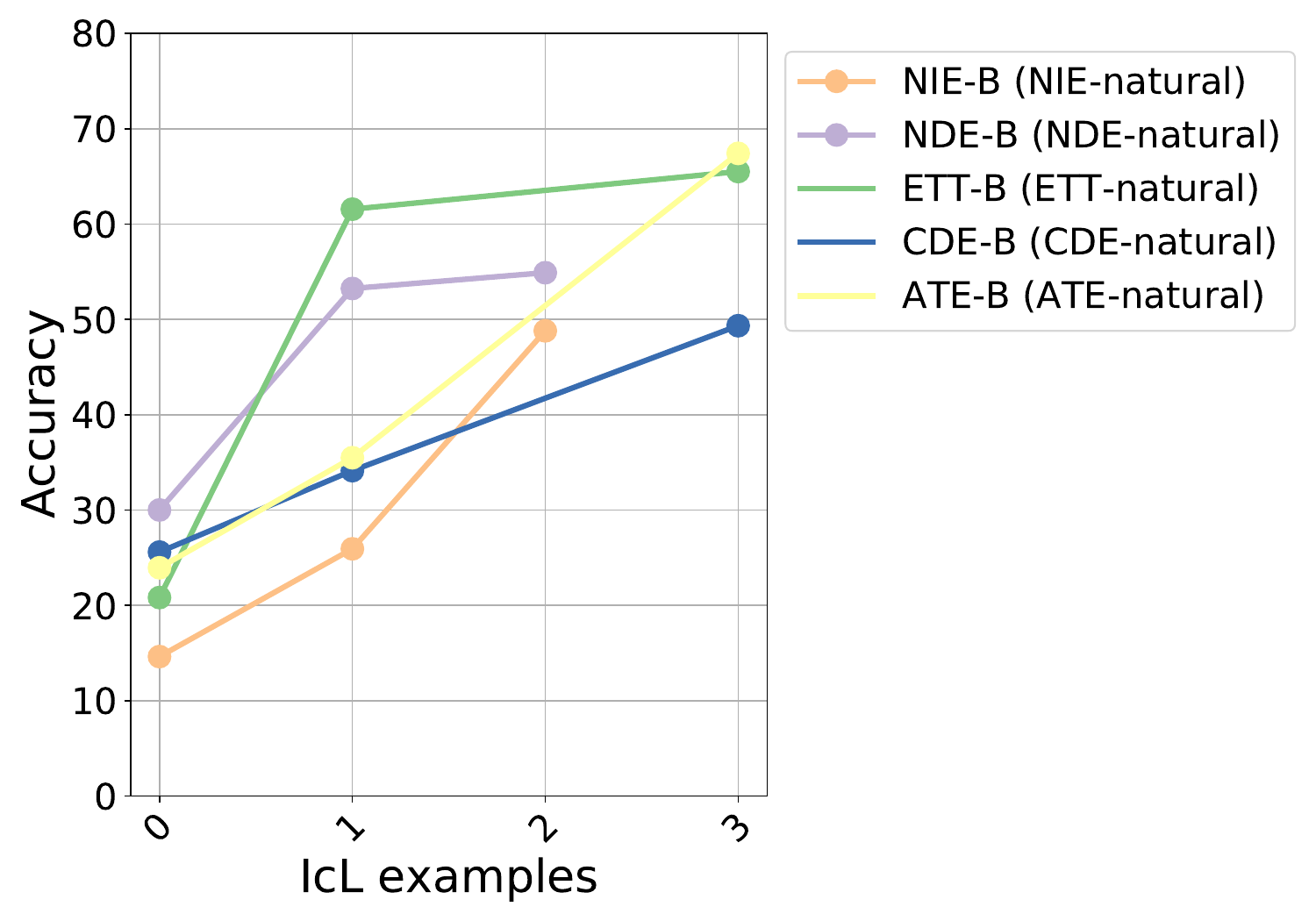}
    \label{fig:Choice_selection_in_Mathematicalematical_tasks}
\end{minipage}
}
\subfigure[Probability calculation in the Mathematical mode.]{   
\begin{minipage}{8cm}
\centering   
  \includegraphics[width=1\linewidth]{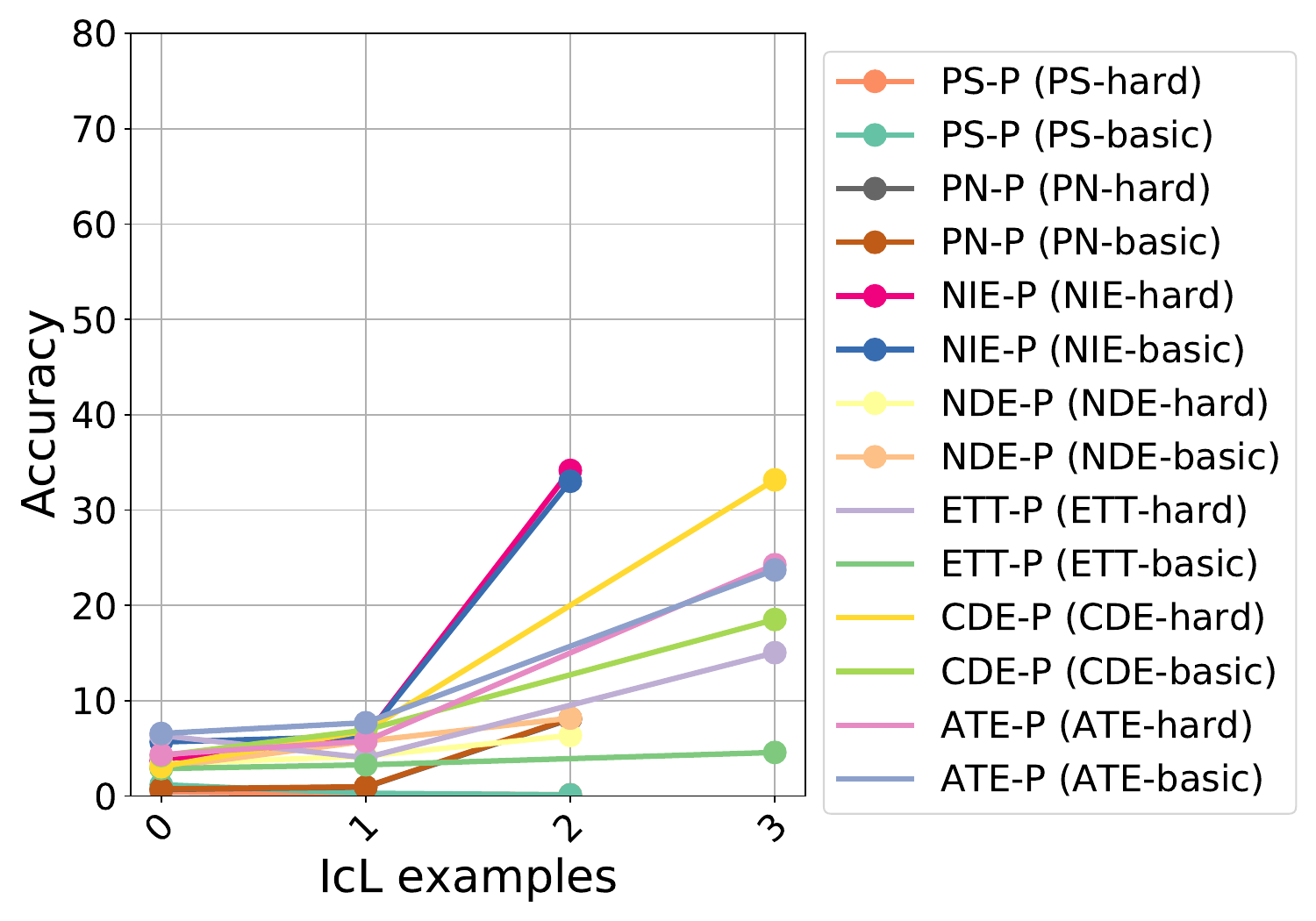}
    \label{fig:Probability_calculation_in_Mathematicalematical_tasks}
\end{minipage}
}
\caption[Accuracy trends of mode and question type combinations]{\textbf{Accuracy trends of mode and question type combinations.}}
\label{fig:Mean_of_Tasks}
\end{figure}

Additionally, we analyze the effects of classifying tasks by modes and question types, as depicted in Figure \ref{fig:Mean_of_Tasks}. For open-ended generation tasks in the Natural mode (Figure \ref{fig:Open-ended_generation_in_natural_tasks}), IcL does not enhance performance. However, for the tasks involving binary classification or choice selection in the Natural mode (Figure \ref{fig:Binary_classification_and_choice_selection_in_natural_tasks}), employing a single IcL example typically yields a steeper improvement slope, characterized by a higher rate of accuracy increase per additional example, than using three examples. All variations in the number of examples yield enhancements over 0-shot across nearly all tasks of this kind. Conversely, for tasks involving binary classification or choice selection in the Symbolic mode (Figure \ref{fig:Binary_classification_and_choice_selection_in_symbolic_tasks}), increasing the number of examples may lead to a decrease in accuracy. Regarding binary classification tasks in the Mathematical mode that require a simple yes/no response (Figure \ref{fig:Choice_selection_in_Mathematicalematical_tasks}), we witness a large surge in performance, with accuracy gains ranging from a minimum of 20\% to a maximum of about 60\%. These improvements are significantly greater than those observed in Natural mode and Symbolic mode tasks, which do not surpass 20\%. Additionally, the performance continues to improve with the addition of more examples. For probability calculation tasks in the Mathematical mode (Figure \ref{fig:Probability_calculation_in_Mathematicalematical_tasks}), a clear pattern emerges where the inclusion of two or three examples substantially enhances performance beyond what is achieved with just one or no examples.

\clearpage

\subsubsection{Adversarial Prompt}
\label{prompt:adversarial}
When confronted with adversarial prompts, we can evaluate a model's responses using a dichotomous tuple (\emph{pre-adversarial}, \emph{post-adversarial}). This classifies the model's replies into 4 categories: (\emph{right}, \emph{right}), (\emph{wrong}, \emph{wrong}), (\emph{right}, \emph{wrong}), and (\emph{wrong}, \emph{right}). For instance, the (\emph{right}, \emph{wrong}) category reflects instances where the model initially provides a \emph{right} response but alters it to a \emph{wrong} one following the adversarial prompt. This analysis is particularly focused on the dynamics between the (\emph{right}, \emph{wrong}) and (\emph{wrong}, \emph{right}) categories, as these transitions illustrate how adversarial prompts influence the model's reliability and response strategy. Additionally, it is worth mentioning that the experiments described here exclude the CEG scenario. Given the open-ended nature of CEG, employing ROUGE-L alone is not sufficient for precisely evaluating the performance differences before and after encountering adversarial inputs.

\begin{figure}[t]
\centering
\subfigure[Wrong direction]{
\centering
\includegraphics[width=\linewidth]{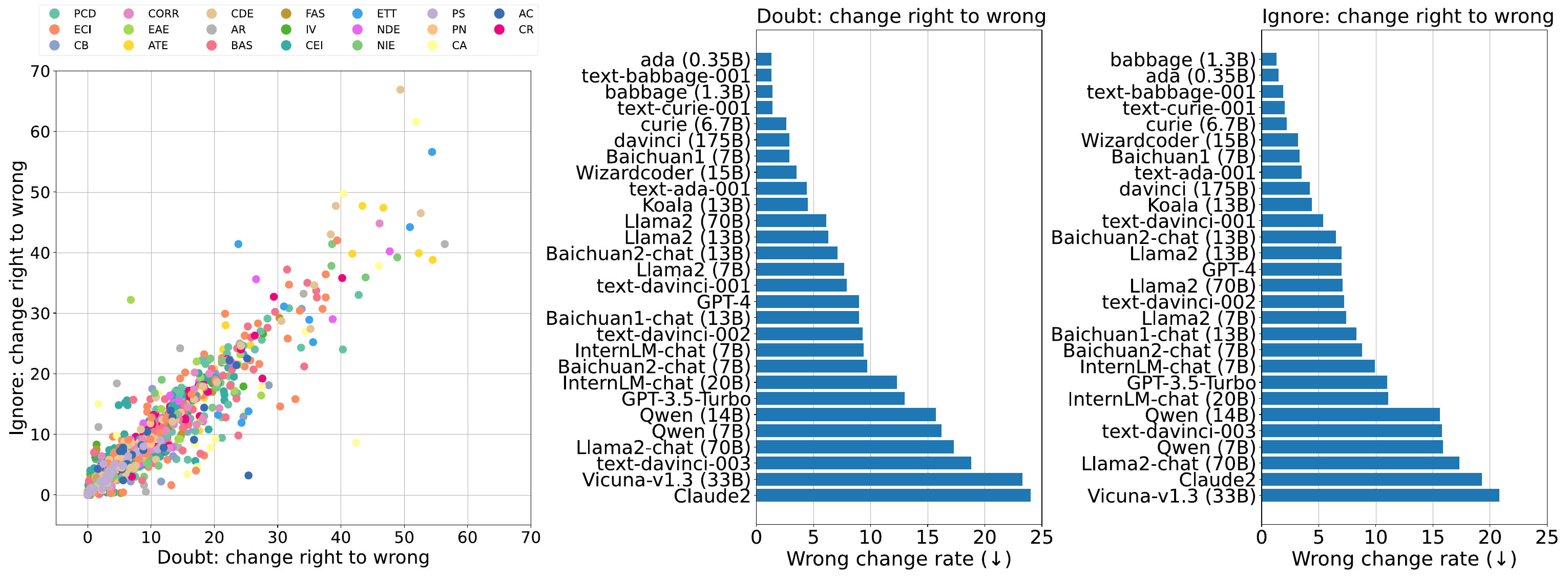}
\label{fig_adversarial_prompt:direction_relationship_wrong}
}
\subfigure[Right direction]{
\centering
\includegraphics[width=\linewidth]{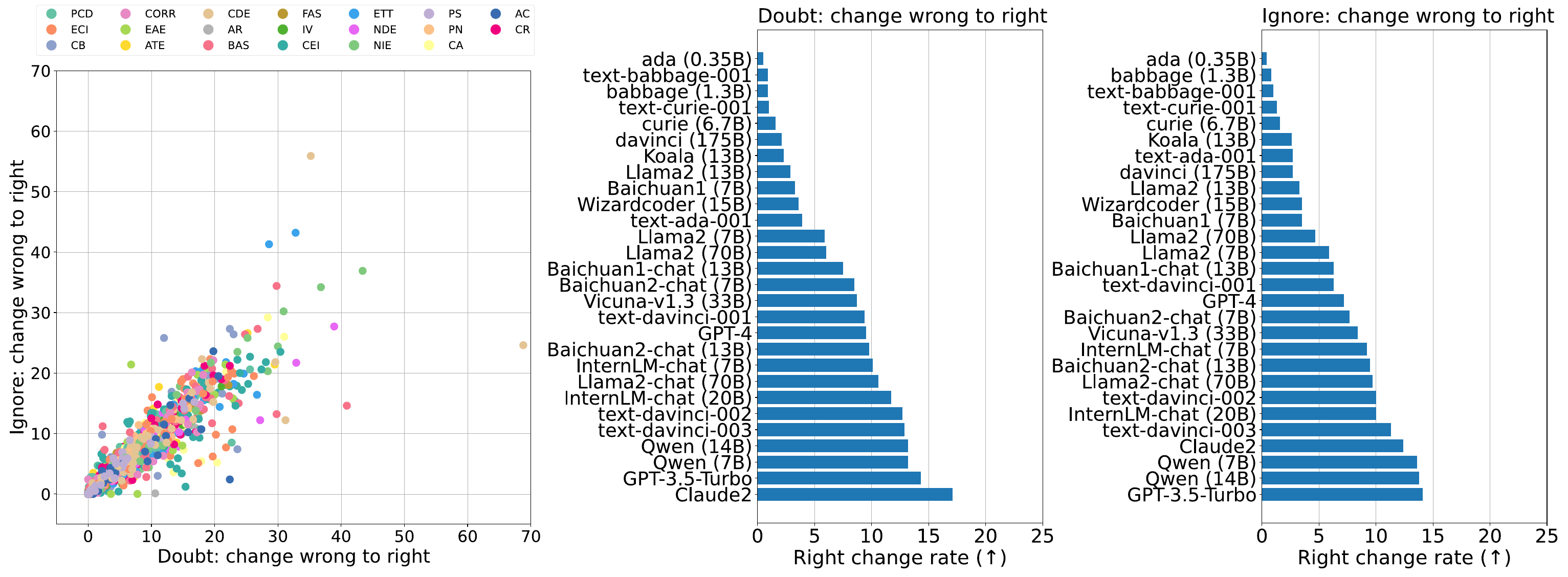}
\label{fig_adversarial_prompt:direction_relationship_right}
}
\caption[Wrong direction vs. right direction]{\textbf{Wrong direction vs. right direction.} \emph{Change right to wrong} reflects instances where the model initially provides a right response but alters it to a wrong one following the adversarial prompt. And \emph{change wrong to right} means the vice versa. Doubt and Ignore refer to two forms of adversarial prompts. The dots in the scatter plot represent the correlation of answer direction changes across all scenarios for every model when adapted adversarial prompts. The histogram represents the average rates of the model's answers change, for both adversarial doubt and adversarial ignore across all scenarios.}    \label{fig_adversarial_prompt:direction_relationship}
\end{figure}

\paragraph{Wrong direction vs. right direction.}
To elucidate the relationship between the directions of answer modification by a model, Figure \ref{fig_adversarial_prompt:direction_relationship} is devised. The top half of the figure illustrates modifications in the incorrect direction (i.e., changing right answers to wrong ones), whereas the bottom half depicts modifications in the correct direction (i.e., changing wrong answers to right ones). The scatter plot within this figure represents the correlation of answer direction changes across all scenarios for each model under various adversarial prompts. For instance, consider that a model in the AR scenario, influenced by adversarial-doubt and adversarial-ignore prompts, alters 50\% and 20\% of its initially right answers to wrong ones, respectively. This instance would be plotted at the (50,20) coordinate on the scatter plot in the figure's top half, with the color of the point indicating the AR scenario. Additionally, a histogram displays the average rates of change in the model’s answers for both adversarial-doubt and adversarial-ignore across all scenarios.

From the analysis of Figure \ref{fig_adversarial_prompt:direction_relationship}, we derive the following key insights: (1) \textit{Change comes with ease, yet the right direction seeks its own challenge.} The scatter plot analysis indicates that, for adjustments in the right direction (correcting wrong answers), no points exceed the (40,40) threshold, suggesting a rarity in substantial correct changes. In contrast, the wrong direction (changing correct answers to incorrect) displays about 13 points beyond this threshold. Predominantly, points associated with the right direction cluster within the (0,30) range, whereas those for the wrong direction are more densely distributed in the (30,40) interval. This pattern indicates that models more frequently alter their responses to incorrect answers than correct ones. The histogram further supports this, showing significantly shorter bars for the right direction, indicating fewer and lesser magnitudes of correct adjustments. Specifically, no model's rate of change in the right direction exceeds 20\%, with only one model, \claude, surpassing 15\% under the adversarial-doubt prompt. Conversely, six models under both adversarial-doubt and adversarial-ignore prompts exceed a 15\% wrong change rate, with two and one models respectively surpassing 20\%. This pattern underscores that, for any single model across all scenarios, changing from wrong to right is a greater challenge.
(2) \textit{Within the same direction, different adversarial prompts exhibit a strong correlation.} The scatter plots reveal a strong correlation in model behavior across different adversarial prompts, adhering closely to a unit slope straight line, whether in the wrong or right direction. The histogram data corroborate this, demonstrating stable rankings of model performance within the same directional category across different prompts. For instance, in the wrong direction, the top and bottom five models remain consistent across both types of adversarial prompts, although their internal rankings may shift. This consistency also extends to the right direction. 

\begin{figure}[t]
    \centering
    \includegraphics[width=.7\textwidth]{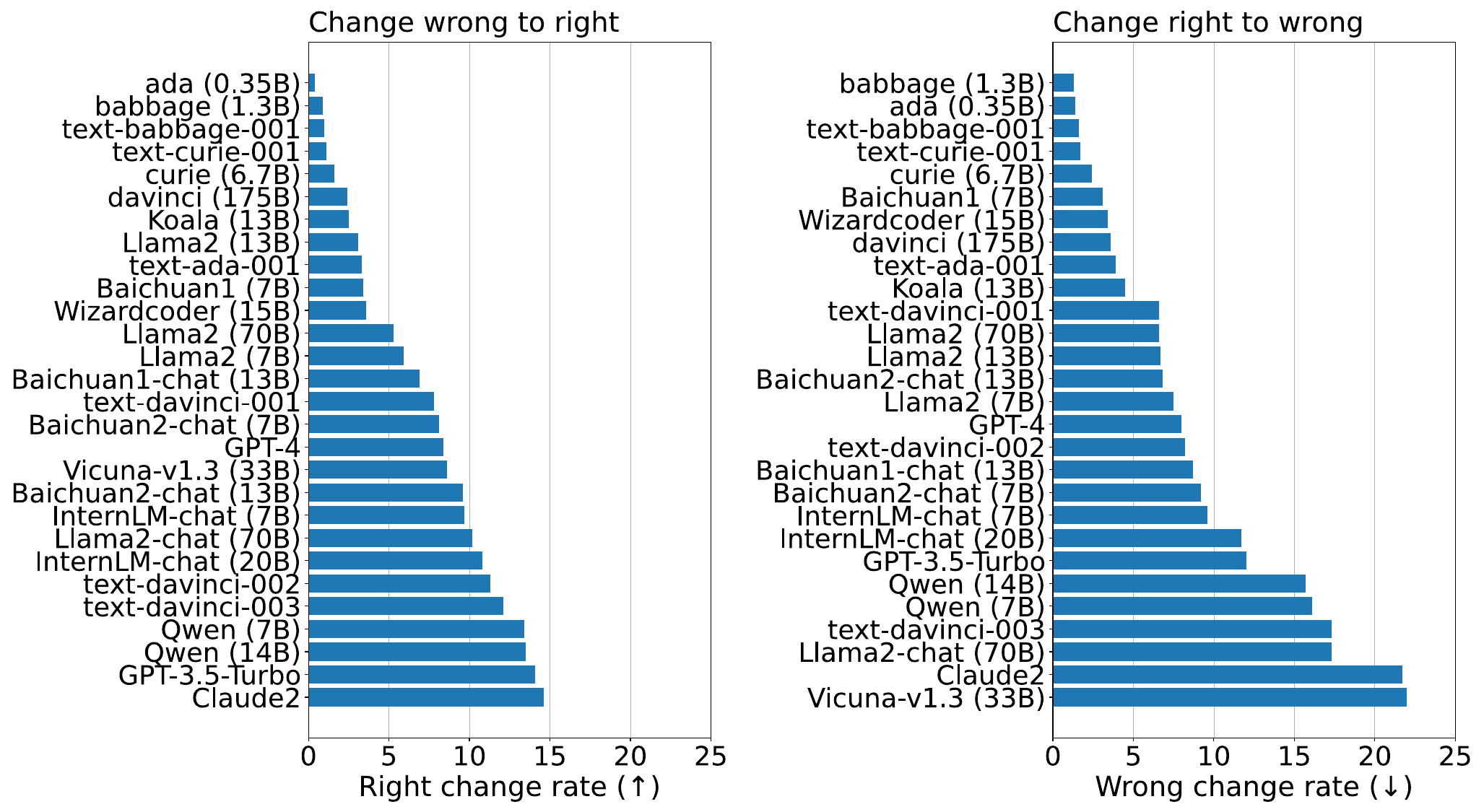}
    \caption[Direct model comparison between right and wrong change directions]{\textbf{Direct model comparison between right and wrong change directions.} We compare the absolute change rates. \emph{Change right to wrong} reflects instances where the model initially provides a right response but alters it to a wrong one following the adversarial prompt. And \emph{change wrong to right} means the vice versa.}    \label{fig_adversarial_prompt:compare_direct}
\end{figure}

\paragraph{Consistency in the directions of change.}
Figure \ref{fig_adversarial_prompt:compare_direct} illustrates the average rates of right change and wrong change for all models across all scenarios, after being attacked by two types of adversarial prompts. Our analysis highlights two consistent trends in the direction of change:
(1) \textit{Consistency among lower-ranking models}: The figure shows that 11 models have a right change rate below 5\%, and 10 models exhibit a wrong change rate below 5\%, with all 10 models overlapping within the 11-model subset. This substantial overlap and the consistently low rate of change in both directions imply significant limitations in these models' ability to effectively follow instructions. It is noteworthy that this group includes all GPT-3 series models released in 2020, as well as some introduced in 2022 and 2023. This indicates that, despite technological advancements over three years, enhancing the models' ability to follow instructions remains a significant challenge.
(2) \textit{Consistency among top-performing models}: Attention is drawn to models that register both a right change rate and a wrong change rate exceeding 10\%, with eight models identified in each category. Excluding \vicuna, there is a complete overlap among the remaining seven models across these categories. The substantial rates of change in both directions indicate that although these models excel at following instructions, they may struggle with independently assessing the accuracy of these instructions. This characteristic, however, is not necessarily negative; its implications vary significantly depending on the application context of the model. In critical decision-making sectors such as healthcare, finance, and law, the ability of models to discern and potentially correct erroneous commands is advantageous. In contrast, in creative fields like art, it may be more desirable for models to strictly follow given instructions, thereby supporting unaltered creative expression.

\paragraph{Inconsistency in the directions of change.}
In our analysis of Figure \ref{fig_adversarial_prompt:compare_direct}, we identify discrepancies in the directions of change in two key areas: (1) \textit{High wrong change and low right change}: This pattern, though not desirable, is evident in some models. For instance, Vicuna has a right change rate of less than 10\% (positioned 11th), while its wrong change rate exceeds 20\% (leading the ranking). This indicates a propensity for these models to incorrectly alter correct responses rather than improve incorrect ones, suggesting a bias in their learning or response mechanisms that could impact their utility in precision-critical applications.
(2) \textit{High right change and low wrong change}: Notably, no model in our experimental setups achieves this ideal balance. This observation does not necessarily reflect a deficiency in the models' ability to correct errors. It could also be influenced by instances where the model's responses are primarily categorized as (\emph{right},\emph{right}). For example, GPT-4 consistently shows robust performance across various scenarios, leading to a higher frequency of (\emph{right},\emph{right}) responses. As a result, it occupies a middle-tier position in terms of the proportion of changes in both categories, reflecting a balanced but not exceptional capability in either direction.

\begin{figure}[t]
\centering
\subfigure[Wrong direction]{
\centering
\includegraphics[width=.8\linewidth]{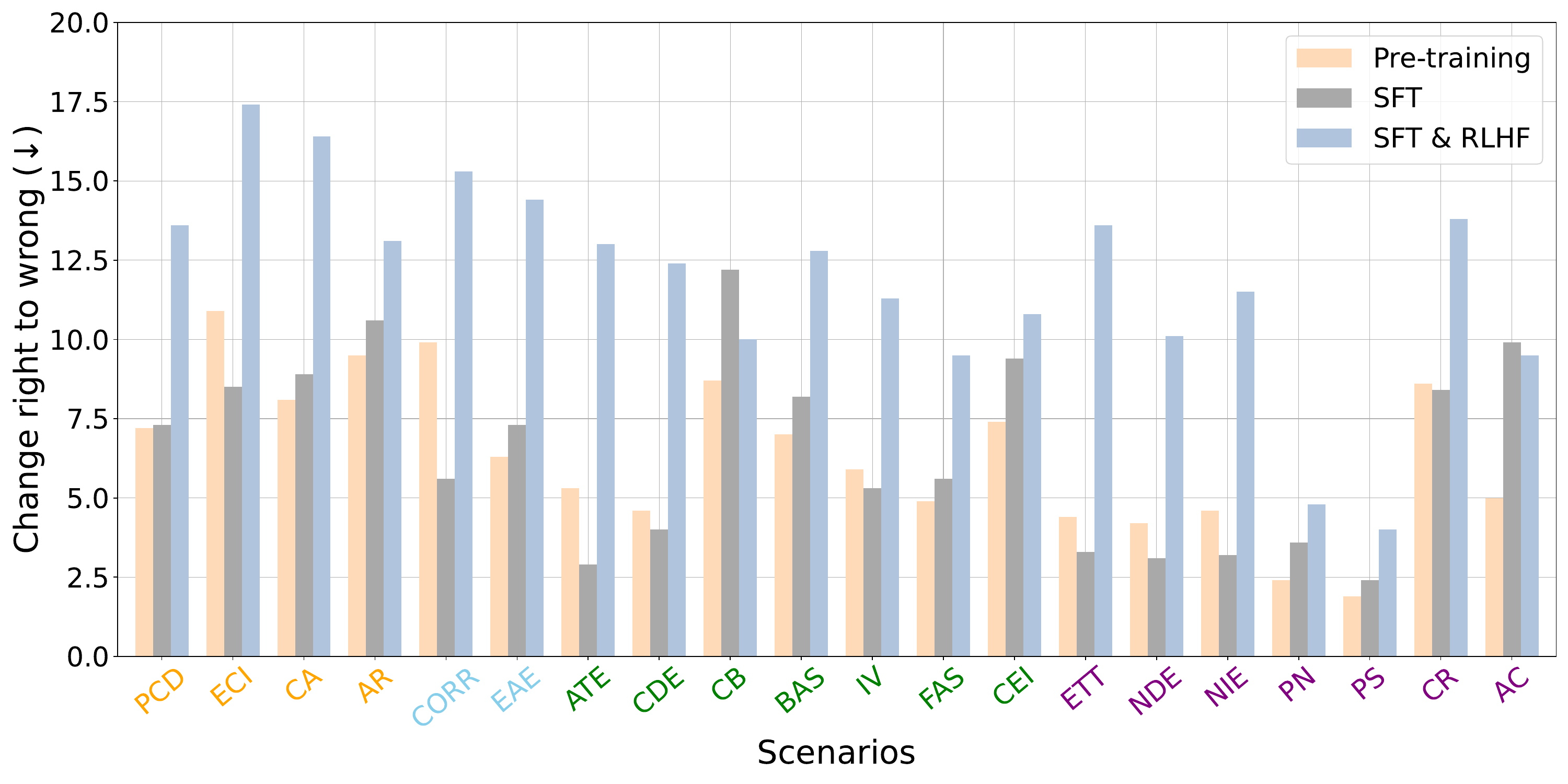}
\label{fig_adversarial_prompt:strategy_wrong}
}
\subfigure[Right direction]{
\centering
\includegraphics[width=.8\linewidth]{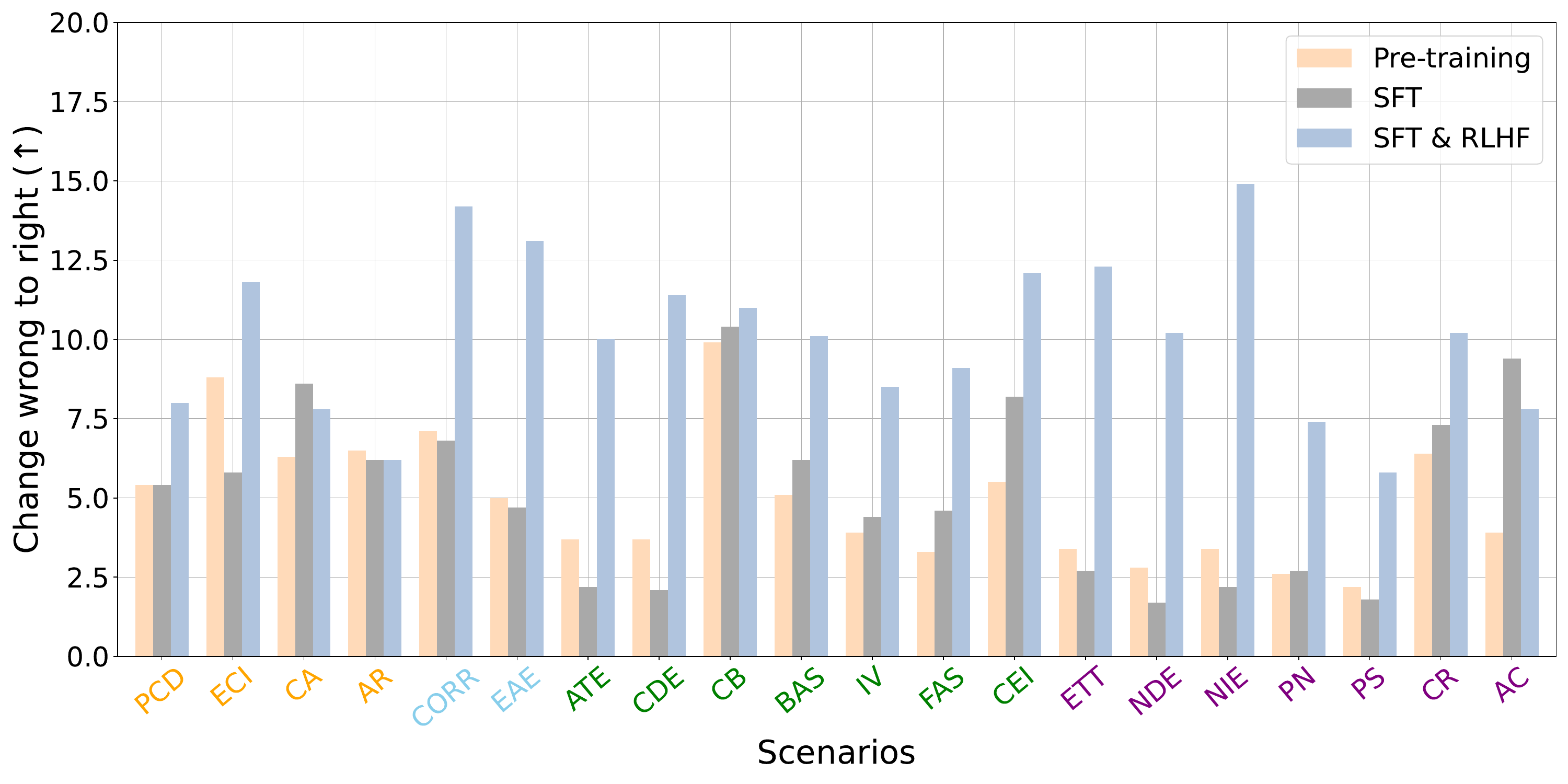}
\label{fig_adversarial_prompt:strategy_right}
}
\caption[Training strategy's influence on wrong and right change directions]{\textbf{Training strategy's influence on wrong and right change directions.} \emph{Change right to wrong} reflects instances where the model initially provides a right response but alters it to a wrong one following the adversarial prompt. And \emph{change wrong to right} means the vice versa.}   \label{fig_adversarial_prompt:strategy}
\end{figure}

\paragraph{Influence of training strategy.}
In Figure \ref{fig_adversarial_prompt:strategy}, we analyze the impact of different training strategies on model responses in the wrong and right directions across all scenarios. The training strategies categorized are consistent with \cref{table_models}. From this analysis, we draw two primary conclusions: 
(1) \textit{Impact of RLHF}: Models trained with RLHF show a tendency to alter their responses more frequently when interacted with by humans. This observation aligns with findings by \citet{sharma2024towards}, which suggest that despite initially accurate and confident responses, models frequently revise their answers upon user inquiries, often leading to misinformation. This indicates that RLHF acts as a double-edged sword; while it enhances responsiveness to human feedback, it also increases susceptibility to generating misinformation, thus requiring careful application and further research to optimize its benefits and mitigate its drawbacks.
(2) \textit{Comparison of pre-training and SFT}: No significant difference is observed between pre-training and SFT in how models adjust their answers under adversarial prompts. Our comparative analysis across 20 scenarios reveals that SFT leads to more substantial changes in the wrong direction in 11 scenarios, suggesting that pre-training generally maintains more stable and accurate responses. Conversely, for changes in the correct direction, pre-training is equal to or more effective than SFT in 11 scenarios. This underscores pre-training's effectiveness in aligning model responses more closely with accurate outputs, indicating that SFT may not significantly enhance model alignment with human intentions.

\clearpage

\subsubsection{Chain-of-Thought}
\label{prompt:cot}
\paragraph{Influence of manual CoT format.}
\emph{Answer after reason} refers to the format in provided examples where the reasoning is presented before culminating in an answer explicitly stated as ``\emph{Therefore, the answer is [...]}''. In contrast, the \emph{reason after answer} format starts with a direct answer, which is then elucidated through subsequent reasoning. Despite evidence from \citet{wei2023CoT} showing the superior efficacy of the standard manual CoT format (i.e., \emph{answer after reason}) over the \emph{reason after answer} format, its effectiveness in assessing causal reasoning capabilities in AI models remains an area of interest. 

\begin{figure}[t]
    \centering
    \includegraphics[width=0.65\textwidth]{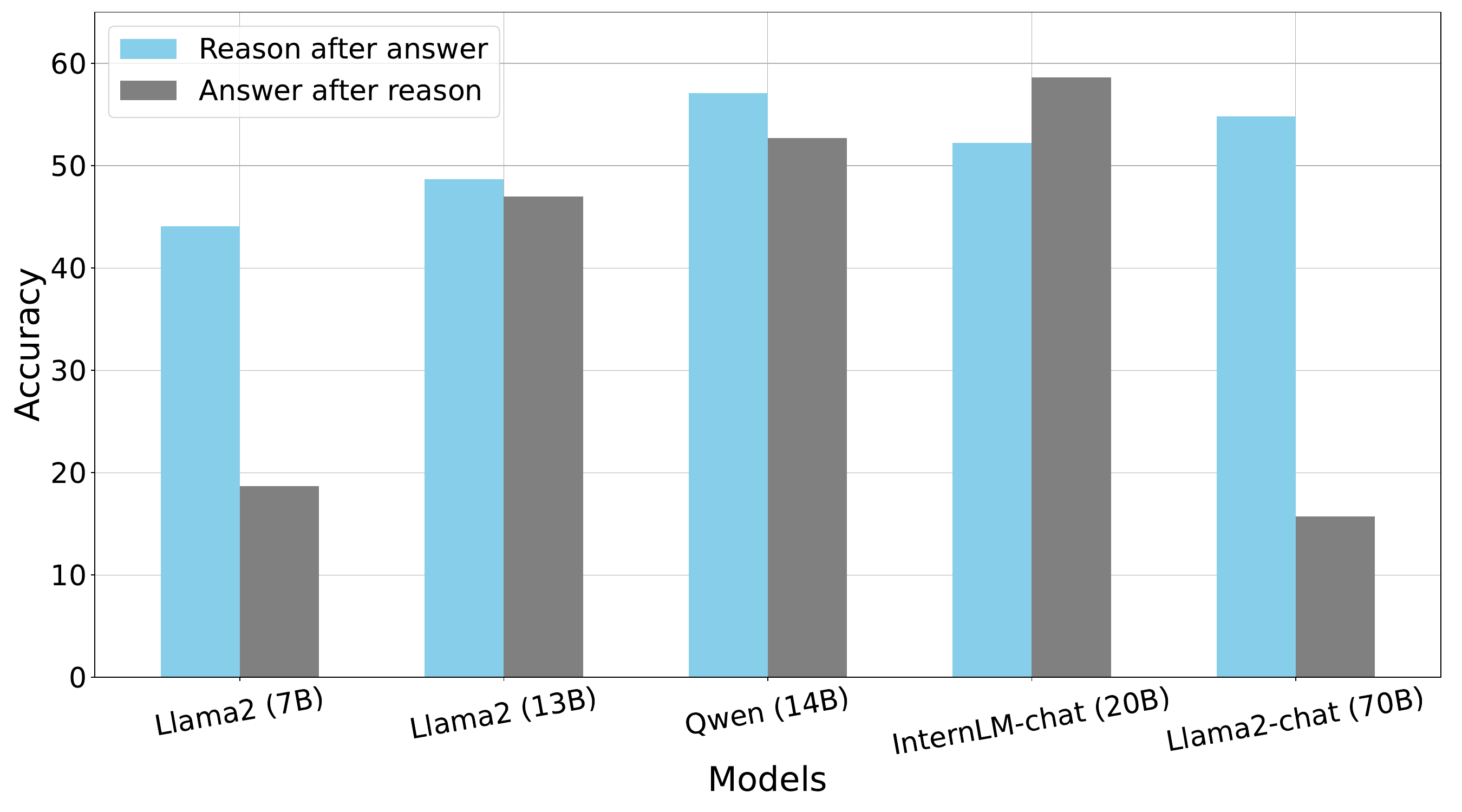}
    \caption[Influence of manual CoT format]{\textbf{Influence of manual CoT format.}}    
    \label{fig_cot_prompt:format}
\end{figure}

To this end, we conduct evaluations on five open-source models, ranging in size from 7B to 70B. Our evaluations focus specifically on the Symbolic mode, covering five tasks (i.e., AR-B (CaLM-AR), IV-C (CaLM-IV), CA-B (FP), CA-B (FA), and CEI-B (0.2-UC)). Figure \ref{fig_cot_prompt:format} reflects the average accuracies achieved by these models on these tasks using the two contrasting formats.
Our finding indicates that the effectiveness of the two manual CoT formats (i.e., \emph{answer after reason} and \emph{reason after answer}) varies depending on the models' capabilities. Specifically, \internt~demonstrates improved performance specifically in the \emph{answer after reason} format. Further analysis from Figure \ref{fig_main:error_prompt} reveals that employing 1/3-shot IcL can notably reduce instruction-following errors, drawing parallels to the \emph{reason after answer} format, where presenting a direct standard answer initially may better guide models in complex causal tasks. Ultimately, the influence of manual CoT on model efficacy for causal reasoning tasks must be evaluated against the backdrop of the models' inherent causal reasoning capabilities.

\begin{figure}[t]
\centering
\subfigure[Basic vs. 0-shot CoT]{
\centering
\includegraphics[width=.8\linewidth]{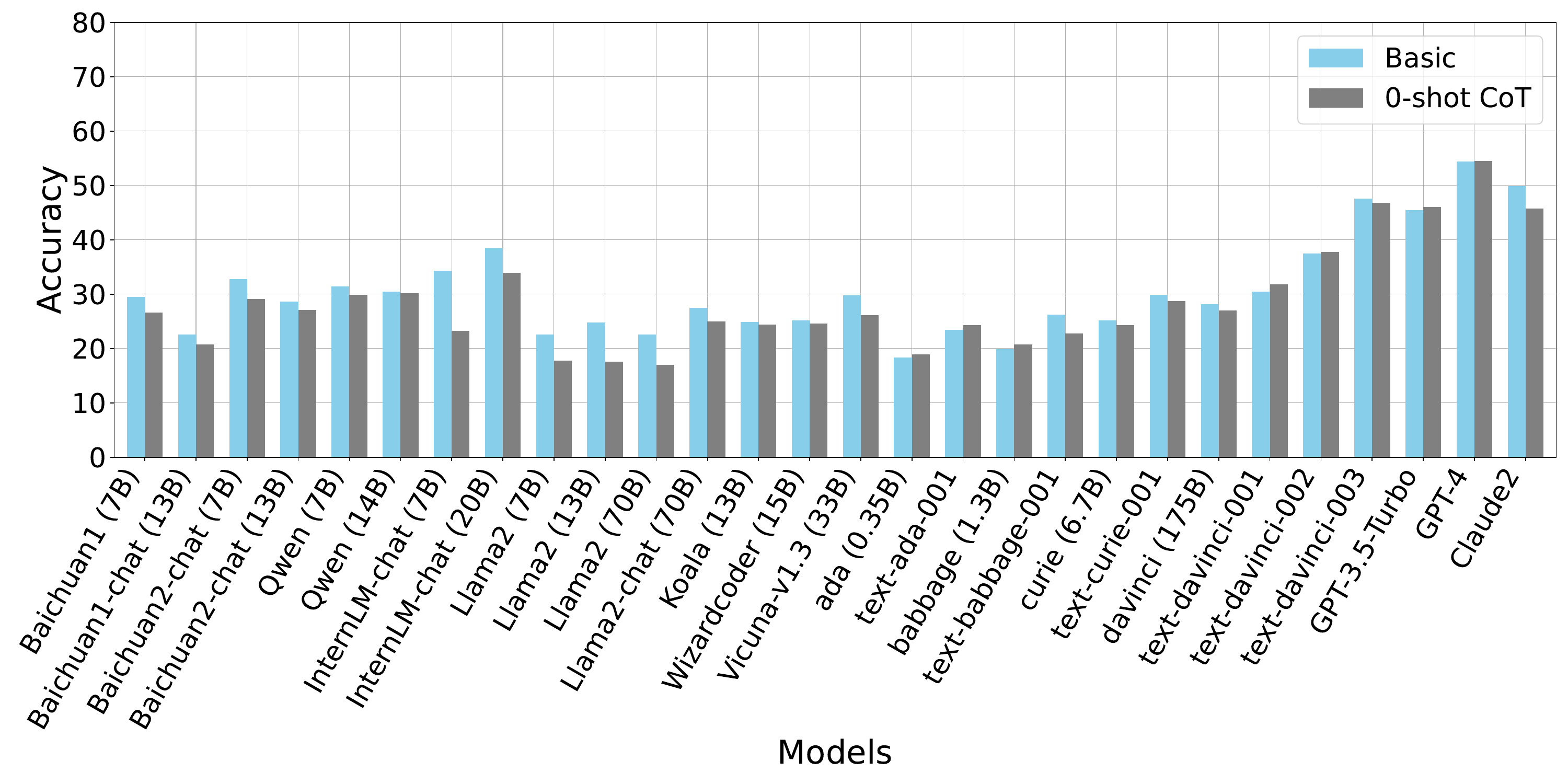}
\label{fig_cot_prompt:basic_0shot}
}
\subfigure[Basic vs. manual CoT]{
\centering
\includegraphics[width=.8\linewidth]{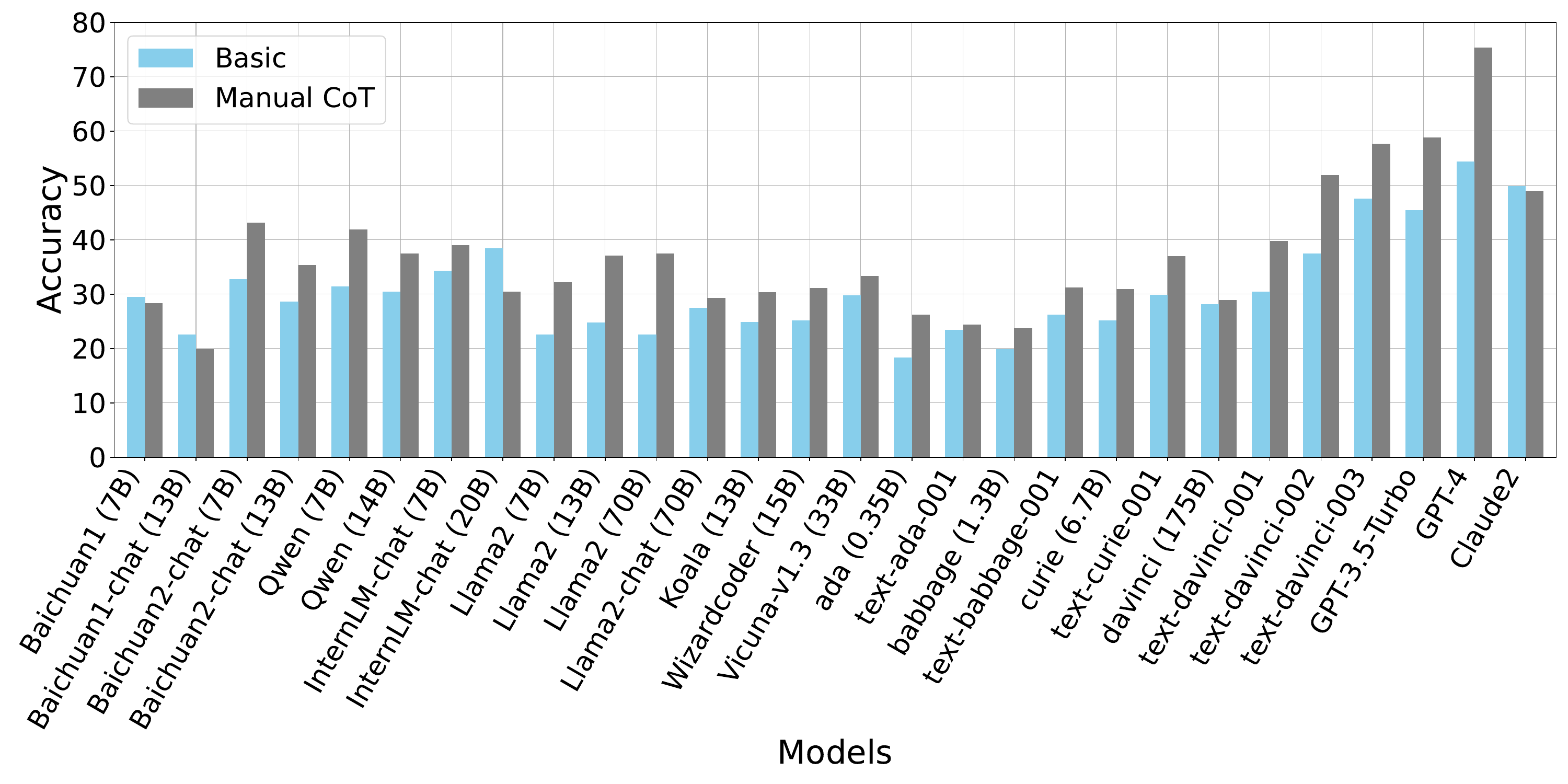}
\label{fig_cot_prompt:basic_manual}
}
\caption[Basic vs. CoT]{\textbf{Basic vs. CoT.} We compare basic prompt with 0-shot CoT and manual CoT across all the models.}    \label{fig_cot_prompt:basic_comparison}
\end{figure}

\paragraph{Basic prompt vs. CoT.}
Figure \ref{fig_cot_prompt:basic_comparison} compares the efficacy of the basic prompt with two CoT prompts across all evaluated models over all causal scenarios. The key findings from this figure are as follows: (1) Among the two CoT formats, manual CoT proves to be more effective. It enhances performance in 24 out of 28 models when compared to the basic prompt. In contrast, the 0-shot CoT approach yields performance gains in only 7 out of the 28 models. This stark difference underscores the significant advantage of employing manual CoT in prompting strategies to facilitate higher model performance. (2) The effectiveness of manual CoT is consistent across models developed with various training strategies. As categorized in \cref{table_models}, irrespective of the training methodology employed, manual CoT consistently boosts performance. This demonstrates its versatility and effectiveness as a prompting strategy across different model architectures and training backgrounds.

\subsubsection{Explicit Function}
\label{prompt:ef}
To illustrate the performance improvement of the explicit-function prompt compared to the basic prompt, we conduct an analysis across all causal scenarios and models.

\begin{figure}[t]
    \centering
    \includegraphics[width=0.8\textwidth]{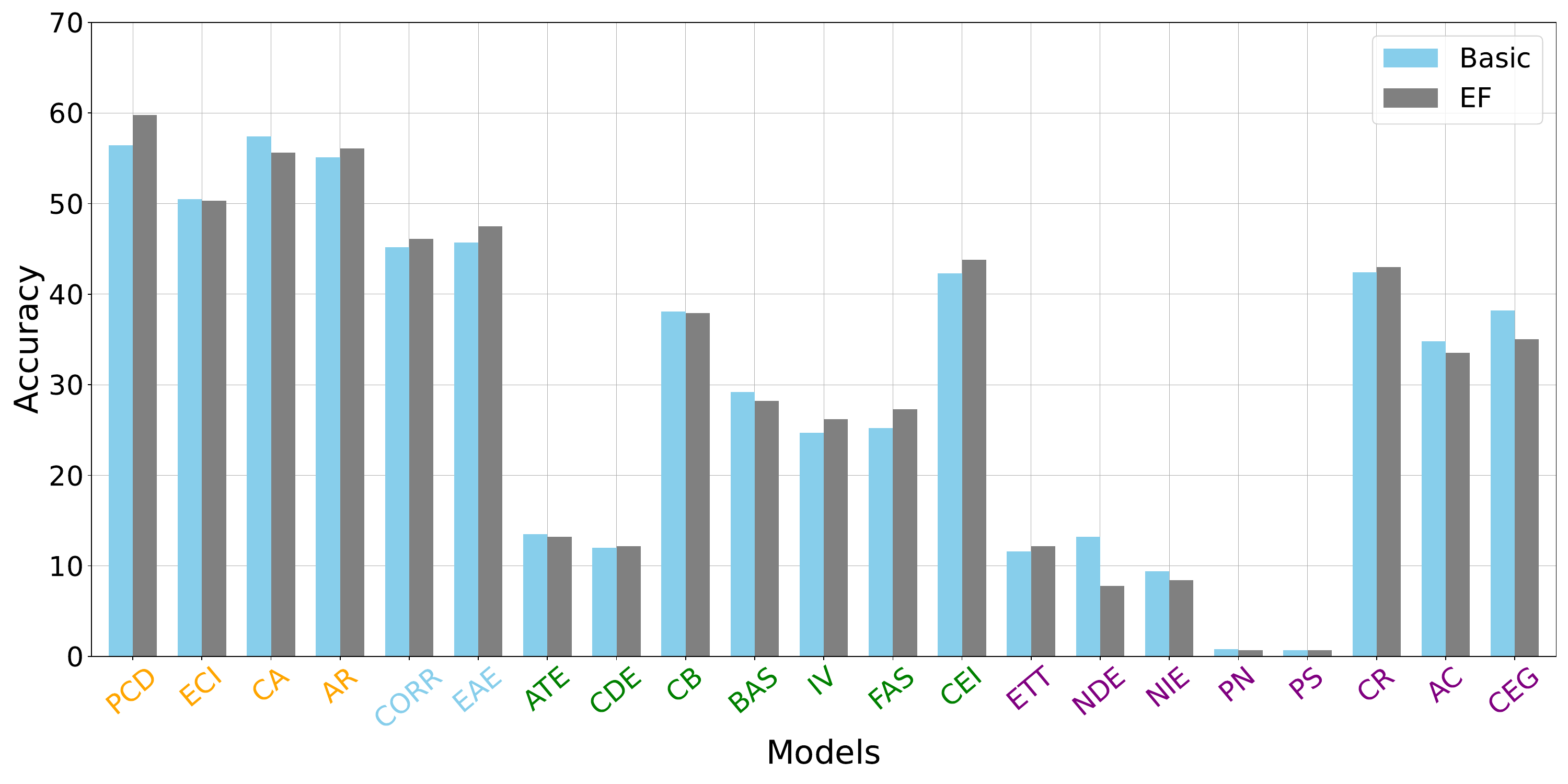}
    \caption[Basic vs. EF across all the scenarios]{\textbf{Basic vs. EF across all the scenarios.} We compare the performance comparison between basic prompt and EF across all the scenarios.}    \label{fig_ef_prompt:across_scenario}
\end{figure}

\paragraph{Across causal scenarios.}
From Figure~\ref{fig_ef_prompt:across_scenario}, we have the following findings:
(1) Both the explicit-function and basic prompts exhibit consistent performance trends across different types of scenarios. They perform well in causal discovery scenarios, such as PCD and ECI, but show poor performance in counterfactuals scenarios, such as PN and PS.
(2) Among 11 out of 21 scenarios, the explicit-function prompt outperforms the basic prompt. Notably, in association scenarios such as CORR and EAE, the explicit-function prompt consistently demonstrates improved performance.

\begin{figure}[t]
    \centering
    \includegraphics[width=0.8\textwidth]{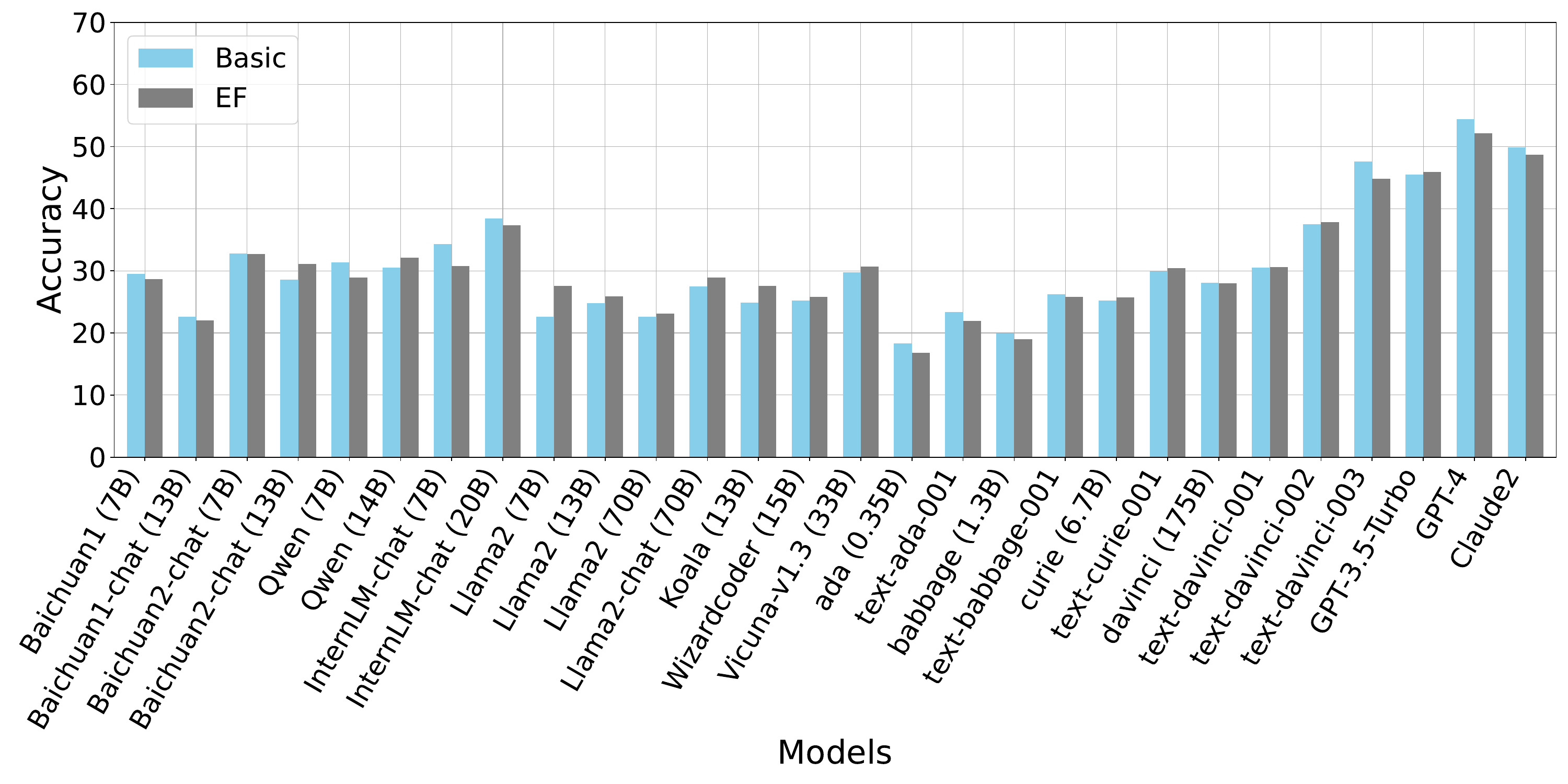}
    \caption[Basic vs. EF across all the models]{\textbf{Basic vs. EF across all the models.} We compare the performance comparison between basic prompt and EF across all the models.}   \label{fig_ef_prompt:across_model}
\end{figure}

\paragraph{Across models.}
Figure~\ref{fig_ef_prompt:across_model} indicates that the explicit-function prompt enhances performance in 13 out of 28 models compared to basic prompts. However, in the remaining 15 models, it results in decreased performance. This variability underscores that the effectiveness of the explicit-function prompt can differ significantly across models, suggesting that their use in practical scenarios should be carefully considered and adapted to the situation.

\clearpage
\clearpage

\subsection{Model-specific Analysis}
\label{experiment:model}

In \nameref{experiment:main} (\cref{experiment:main}), we have already analyzed the performance of models from the most comprehensive and insightful perspective (e.g., \nameref{main:comparison} (Section \ref{main:comparison}), \nameref{main:complexity} (Section \ref{main:complexity}), \nameref{main:error} (Section \ref{main:error})). This section will center on the performance of each specific model across all causal scenarios and prompts. Altogether, 28 models from 9 different creators have been evaluated in CaLM. Therefore, this section will break down the analysis by creator, organizing it into the following nine sections: \nameref{model:openai} (\cref{model:openai}), \nameref{model:anthropic} (\cref{model:anthropic}), \nameref{model:ailab} (\cref{model:ailab}), \nameref{model:ali} (\cref{model:ali}), \nameref{model:baichuan} (\cref{model:baichuan}), \nameref{model:meta} (\cref{model:meta}), \nameref{model:lmsys} (\cref{model:lmsys}), \nameref{model:ucb} (\cref{model:ucb}), and \nameref{model:microsoft} (\cref{model:microsoft}). Additionally, \cref{tab:model_term} will provide clarifications for technical terms pertinent to our subsequent analysis. For ease of reading, these terms will be presented in \emph{italic font} throughout the entire section.

\begin{figure}
    \centering
    \includegraphics[width=1\linewidth]{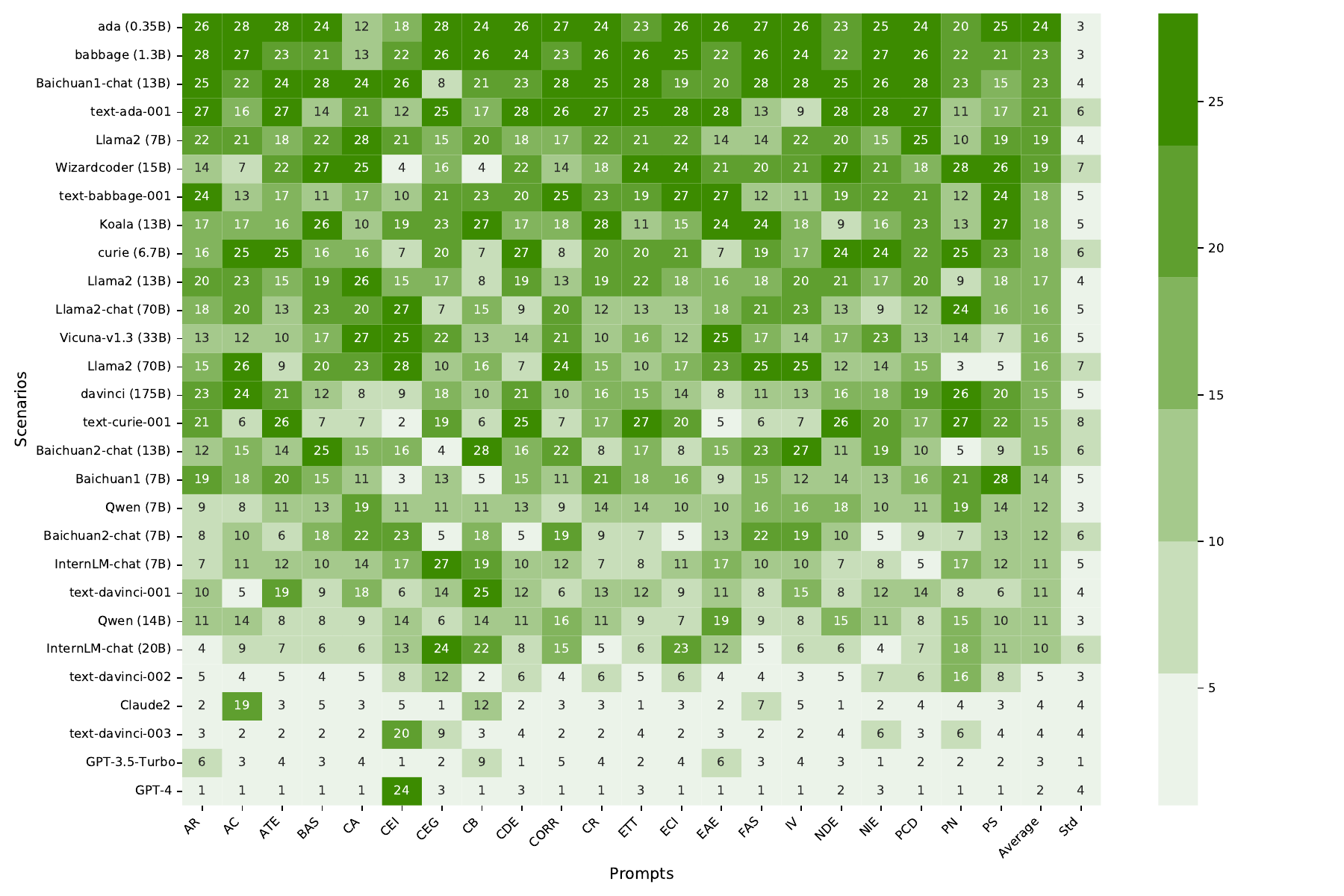}
    \caption[\emph{Prompt-average rank} of models]{\textbf{\emph{Prompt-average rank} of models.}}
    \label{fig:Prompt-Average_Rank_of_Models}
\end{figure}

When discussing a model's performance, the content is typically organized into four main paragraphs: Summary, Accuracy, Ranking, and Robustness. Each of these paragraphs focuses on different aspects of the model evaluation:
\begin{itemize}
    \item Summary: This paragraph provides an overview analysis of the model's performance. It concisely reports key statistics such as the \textit{average scenario-prompt accuracy}, the average \textit{prompt-average rank}, and the average robustness score across various scenarios. This helps in quickly understanding the model's general effectiveness and reliability.
    \item Accuracy: This paragraph delves deeper into the model's performance metrics. It is usually divided into four subcategories detailing: 1) Overall performance, which highlights the \textit{average scenario-prompt accuracy}, the average standard deviation of prompt accuracy, the \textit{top scenario-prompt pair}, and the proportions of \textit{scenario-prompt pair}s that exceed \textit{random guess accuracy} and the 80\% accuracy threshold. 2) Scenario performance, which lists scenarios with the highest average accuracy. Scenarios are only included if their accuracy surpasses \textit{random guess accuracy}; otherwise, they are excluded from the list. 3) Prompt efficiency, which focuses on identifying the most efficient prompts and those that have the largest number of \textit{scenario-prompt pair}s exceeding \textit{random guess accuracy}. 4) Language influence, which assesses how well the model performs across different languages, evaluating the \textit{language accuracy difference} to understand the impact of linguistic variation on model accuracy.
    \item Ranking: This paragraph compares the model’s performance to other models and includes two key metrics: 1) \textit{Prompt-average rank}, which reports the highest, lowest, and average \textit{prompt-average rank}s, as shown in Figure \ref{fig:Prompt-Average_Rank_of_Models}. 2) \textit{Model-prompt rank}: which indicates the best and worst \textit{model-prompt rank}s over all the \textit{scenario-prompt pair}s in the model.
    \item Robustness: The paragraph showcases the average robustness score of the model and identifies scenarios where the model achieves the highest robustness.
\end{itemize}

\begin{center}
\begin{table*}[t]
\caption[Explanations for model-specific terminologies]{\textbf{Explanations for model-specific terminologies.}}
\label{tab:model_term}
\begin{tabularx}{\textwidth}{c|X} 
\toprule
{\textbf{Terminology}} & \makecell[c]{\textbf{Explanation}} \\
\hline
  \textit{scenario-prompt pair} & A combination of a scenario and a prompt.  \\
\hline
\multirow{1}{*}{\textit{average scenario-prompt accuracy}} & The average accuracy of a model tested on all the \textit{scenario-prompt pair}s.\\
\hline
\multirow{2}{*}{\textit{top scenario-prompt pair}} & The combination that has the top accuracy value across all tested \textit{scenario-prompt pairs} in the target model. \\
\hline
\multirow{2}{*}{\textit{random guess accuracy}} & The random guess accuracy of a model within a causal task/scenario, varying across different causal tasks/scenarios.\\
\hline
\multirow{2}{*}{\textit{language accuracy difference}} & The difference between a model's English and Chinese accuracy across all the scenario and prompt pairs.\\
\hline
\multirow{3}{*}{\textit{prompt-average rank}} & As shown in Figure \ref{fig:Prompt-Average_Rank_of_Models}, the \textit{prompt-average rank}, ranging from 1 to 28, is derived by comparing the average accuracies of 28 models across all the prompts in a scenario.\\
\hline 
\multirow{3}{*}{\textit{model-prompt rank}} & Ranging from 1 to 252, the \textit{model-prompt rank} is determined by comparing the accuracies across all 28$\times$9 model-prompt pairs within each scenario. \\
\hline
\end{tabularx}
\end{table*}
\vspace{-20pt}
\end{center}

\subsubsection{OpenAI}
\label{model:openai}
\paragraph{ada (0.35B).}
\begin{figure}[t]
\centering
\subfigure[Performance of ada (0.35B)]{
\begin{minipage}{8.5cm}
\centering
\includegraphics[width=1\linewidth]{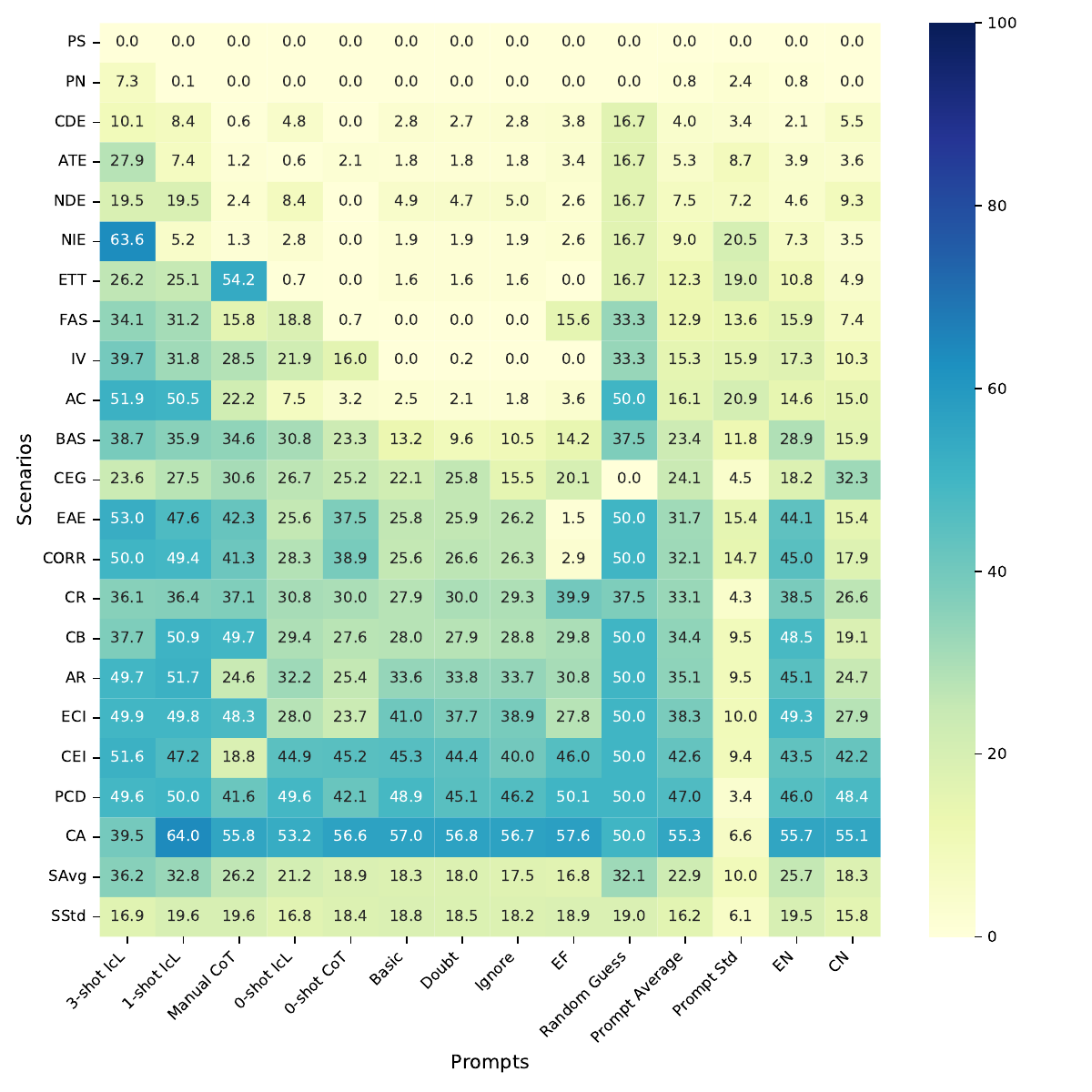}
\end{minipage}
}
\subfigure[\textit{Model-prompt rank} of ada (0.35B)]{
\begin{minipage}{8.5cm}
\centering
\includegraphics[width=1\linewidth]{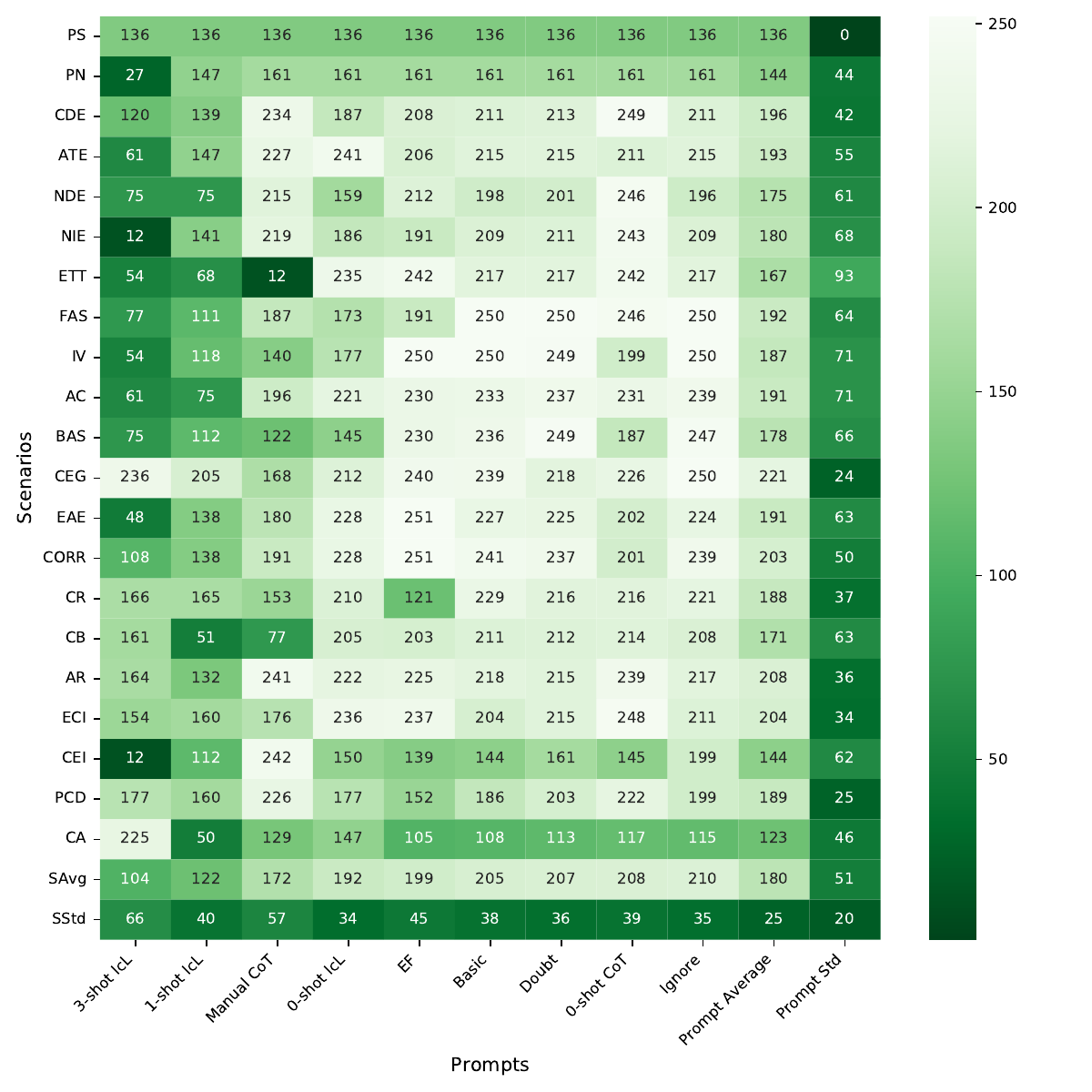}
\end{minipage}
}
\caption[Heatmap of ada (0.35B)]{\textbf{Heatmap of ada (0.35B).}}
\label{fig:Heatmap_of_ada_(0.35B)}
\end{figure}
Summary: The model's \textit{average scenario-prompt accuracy} is 22.9\%, with the average \textit{prompt-average rank} of 24/28 (the lowest average \textit{prompt-average rank}). Additionally, it demonstrates a high average robustness score of 96.7\% across various scenarios.

Accuracy: 1) Overall performance: As depicted in Figure \ref{fig:Heatmap_of_ada_(0.35B)}(a), ada (0.35B) registers an \textit{average scenario-prompt accuracy} of 22.9\%
, with an average prompt effectiveness standard deviation of 10.0. The \textit{top scenario-prompt pair}s include 1-shot IcL in CA scoring 64.0\%. This is followed by 3-shot IcL in NIE at 63.6\% and EF in CA at 57.6\%. 29.1\% of \textit{scenario-prompt pairs} exceed their \textit{random guess accuracy}, with 0.0\% reaching above 80\% accuracy.
2) Scenario performance: High-performing scenarios where ada (0.35B) excels and surpasses \textit{random guess accuracy} include CA with a score of 55.3\%, CEG at 24.1\%, and PN at 0.8\%.
3) Prompt efficiency: The most effective prompts are 3-shot IcL at 36.2\% and 1-shot IcL at 32.8\%. Regarding the number of \textit{scenario-prompt pair}s where the model exceeds the \textit{random guess accuracy}, the 3-shot IcL leads in 14 out of 21 scenarios, followed by 1-shot IcL in 10, and EF in 6 scenarios.
4) Language influence: In 15 out of 21 scenarios, English outperforms Chinese, with significant \textit{language accuracy difference}s observed in CB, EAE, and CORR, where the accuracy differences are 29.4\%, 28.7\%, and 27.1\%, respectively. Conversely, scenarios such as CEG, NDE, and CDE showcase superior performance in Chinese, with accuracy differences of 14.1\%, 4.8\%, and 3.4\%, respectively. 

Ranking: 1) \textit{Prompt-average rank}: As shown in Figure \ref{fig:Prompt-Average_Rank_of_Models}, ada (0.35B)'s best \textit{prompt-average rank}s appears in CA at 12. In contrast, ada (0.35B) ranks lowest in AC at 28, ATE at 28, and CEG at 28, indicating areas for improvement. The model's average \textit{prompt-average rank} across 21 scenarios is 24/28, with a standard deviation of 3.8.
2) \textit{Model-prompt rank}: As shown in Figure \ref{fig:Heatmap_of_ada_(0.35B)}(b), ada (0.35B)'s best \textit{model-prompt rank}s appears in CEI with 3-shot IcL at 12, NIE with 3-shot IcL at 12, ETT with manual CoT at 12. On the other hand, the lowest ranks are observed in CORR with EF at 251, EAE with EF at 251, and CEG with adversarial ignore at 250.

Robustness: ada (0.35B) boasts an average robustness score of 96.7\% across scenarios. The model has the best robustness in ETT at 99.9\%, ATE at 99.9\%, and NIE at 99.9\%.

\paragraph{text-ada-001.}
\begin{figure}[t]
\centering
\subfigure[Performance of text-ada-001]{
\begin{minipage}{8.5cm}
\centering
\includegraphics[width=1\linewidth]{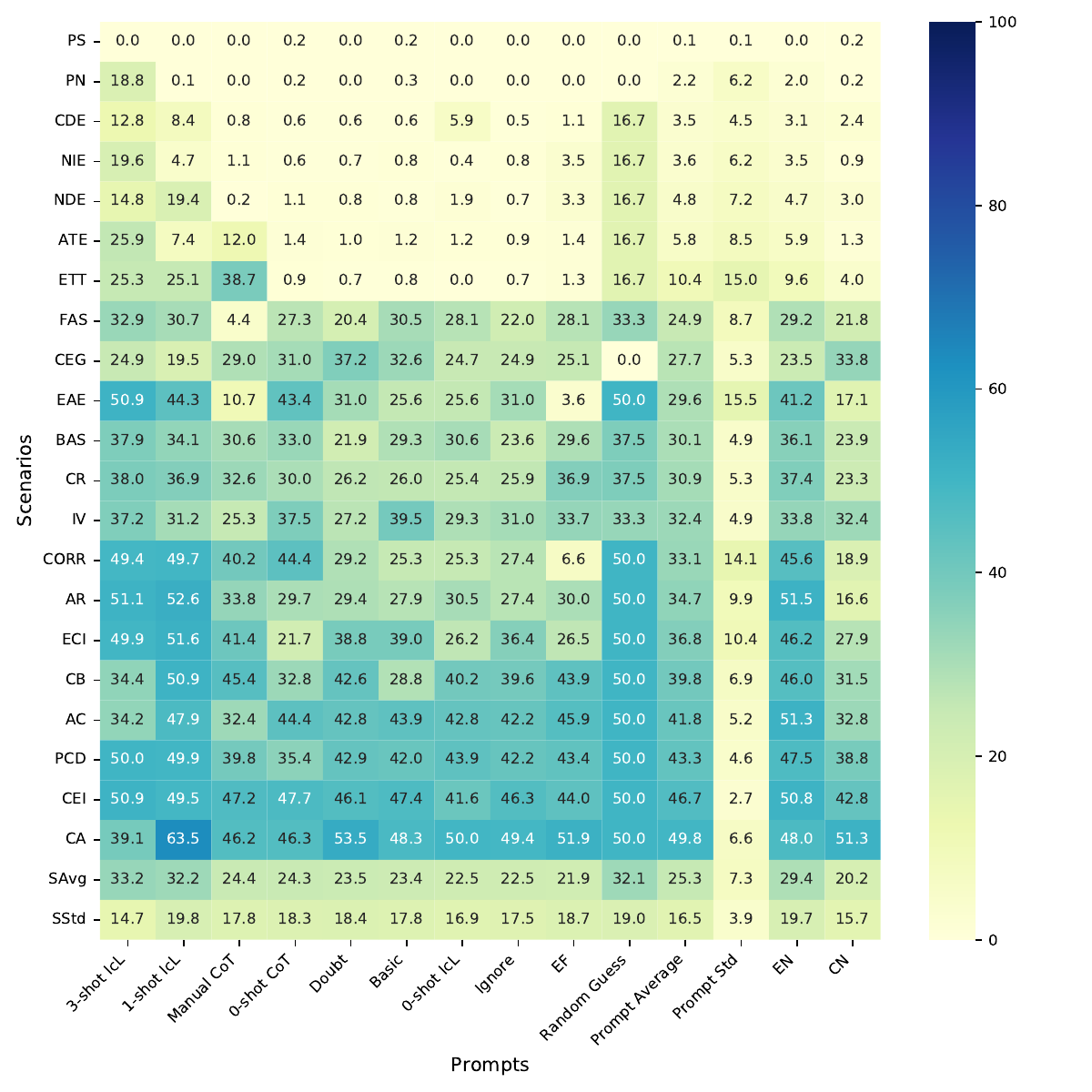}
\end{minipage}
}
\subfigure[\textit{Model-prompt rank} of text-ada-001]{
\begin{minipage}{8.5cm}
\centering
\includegraphics[width=1\linewidth]{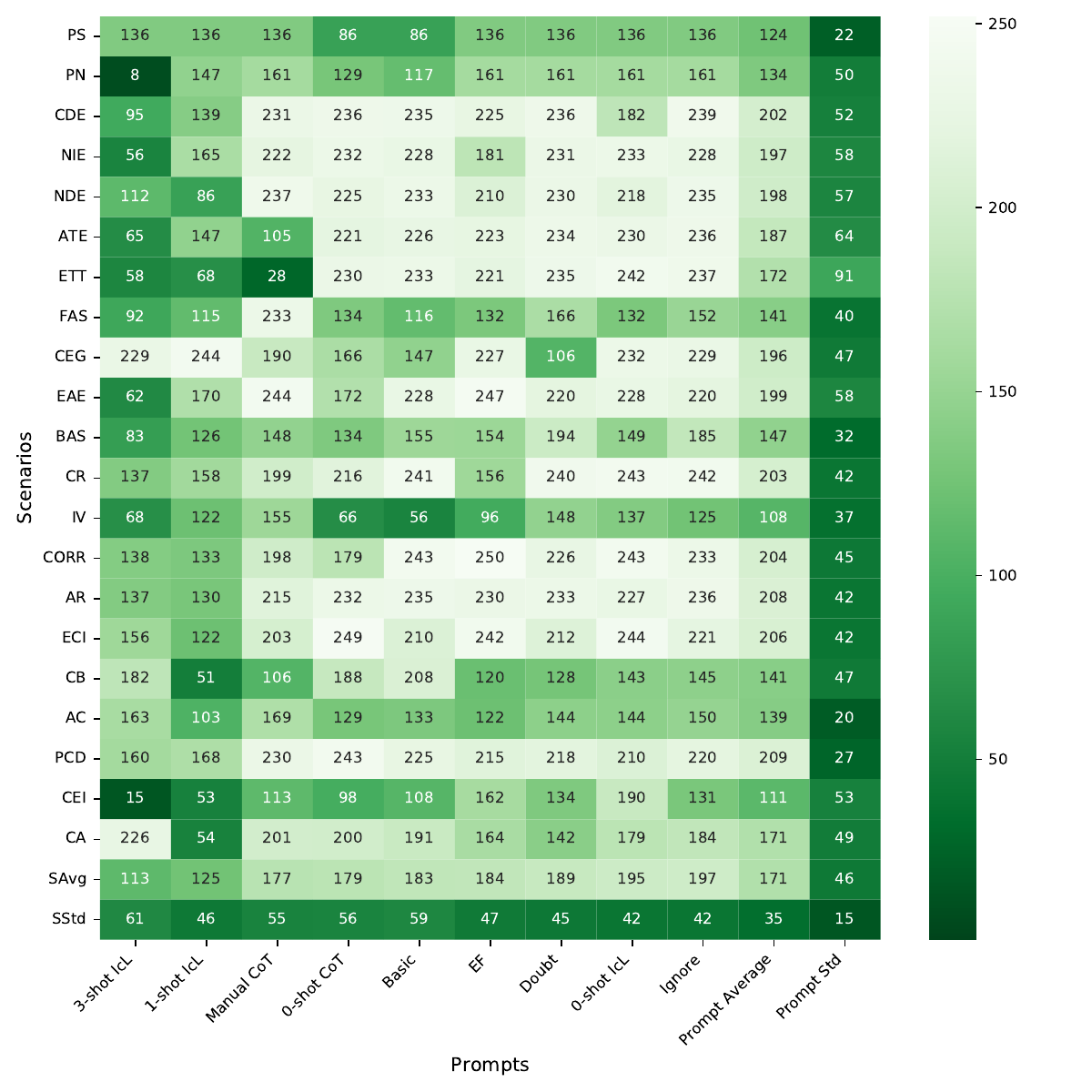}
\end{minipage}
}
\caption[Heatmap of text-ada-001]{\textbf{Heatmap of text-ada-001.}}
\label{fig:Heatmap_of_text-ada-001}
\end{figure}

Summary: The model showcases an \textit{average scenario-prompt accuracy} of 25.3\%, achieves an average \textit{prompt-average rank} of 21 out of 28, and possesses an average robustness score of 84.7\%.

Accuracy: 1) Overall performance: As depicted in Figure \ref{fig:Heatmap_of_text-ada-001}(a), text-ada-001 achieves an \textit{average scenario-prompt accuracy} of 25.3\%, with an average standard deviation in prompt effectiveness of 7.3. The \textit{top scenario-prompt pair}s are 1-shot IcL in CA with a score of 63.5\%, followed by adversarial doubt in CA at 53.5\%, and 1-shot IcL in AR at 52.6\%. Only 25.9\% of the \textit{scenario-prompt pairs} surpass the \textit{random guess accuracy}, and none exceed 80\% accuracy.
2) Scenario performance: In scenarios where text-ada-001 surpasses \textit{random guess accuracy}, the top 3 scenarios having the highest average accuracy are CEG with a score of 27.7\%, PN at 2.2\%, and PS at 0.1\%.
3) Prompt efficiency: The most effective prompts include 3-shot IcL at 33.2\% and 1-shot IcL at 32.2\%. Regarding the number of \textit{scenario-prompt pair}s where the model exceeds the \textit{random guess accuracy}, 3-shot IcL leads in 13 out of 21 scenarios, followed by 1-shot IcL in 9, and EF in 5 scenarios.
4) Language influence: English performs better than Chinese in 18 of 21 scenarios, with significant advantages in AR, CORR, and EAE, where the \textit{language accuracy difference}s are 34.9\%, 26.7\%, and 24.1\%, respectively. Conversely, the Chinese perform better in CEG, CA, and PS, with \textit{language accuracy difference}s of 10.3\%, 3.3\%, and 0.2\%, respectively.

Ranking: 1) \textit{Prompt-average rank}: As shown in Figure \ref{fig:Prompt-Average_Rank_of_Models}, text-ada-001's best \textit{prompt-average rank}s are in IV at 9, PN at 11, and CEI at 12. The model ranks lowest in EAE, NIE, CDE, ECI, and NDE, all at 28, indicating areas for improvement. The average \textit{prompt-average rank} across 21 scenarios is 21/28, with a standard deviation of 6.8.
2) \textit{Model-prompt rank}: Figure \ref{fig:Heatmap_of_text-ada-001}(b) illustrates text-ada-001's highest ranks in PN with 3-shot IcL at 8, CEI with 3-shot IcL at 15, and ETT with manual CoT at 28. The lowest ranks are in CORR with EF at 250, ECI with 0-shot CoT at 249, and EAE with EF at 247.

Robustness: text-ada-001 has an average robustness score of 84.7\% across scenarios, with the highest robustness in CDE at 97.3\%, NDE at 96.5\%, and ETT at 95.4\%.

\paragraph{babbage (1.3B).}
\begin{figure}[t]
\centering
\subfigure[Performance of babbage (1.3B)]{
\begin{minipage}{8.5cm}
\centering
\includegraphics[width=1\linewidth]{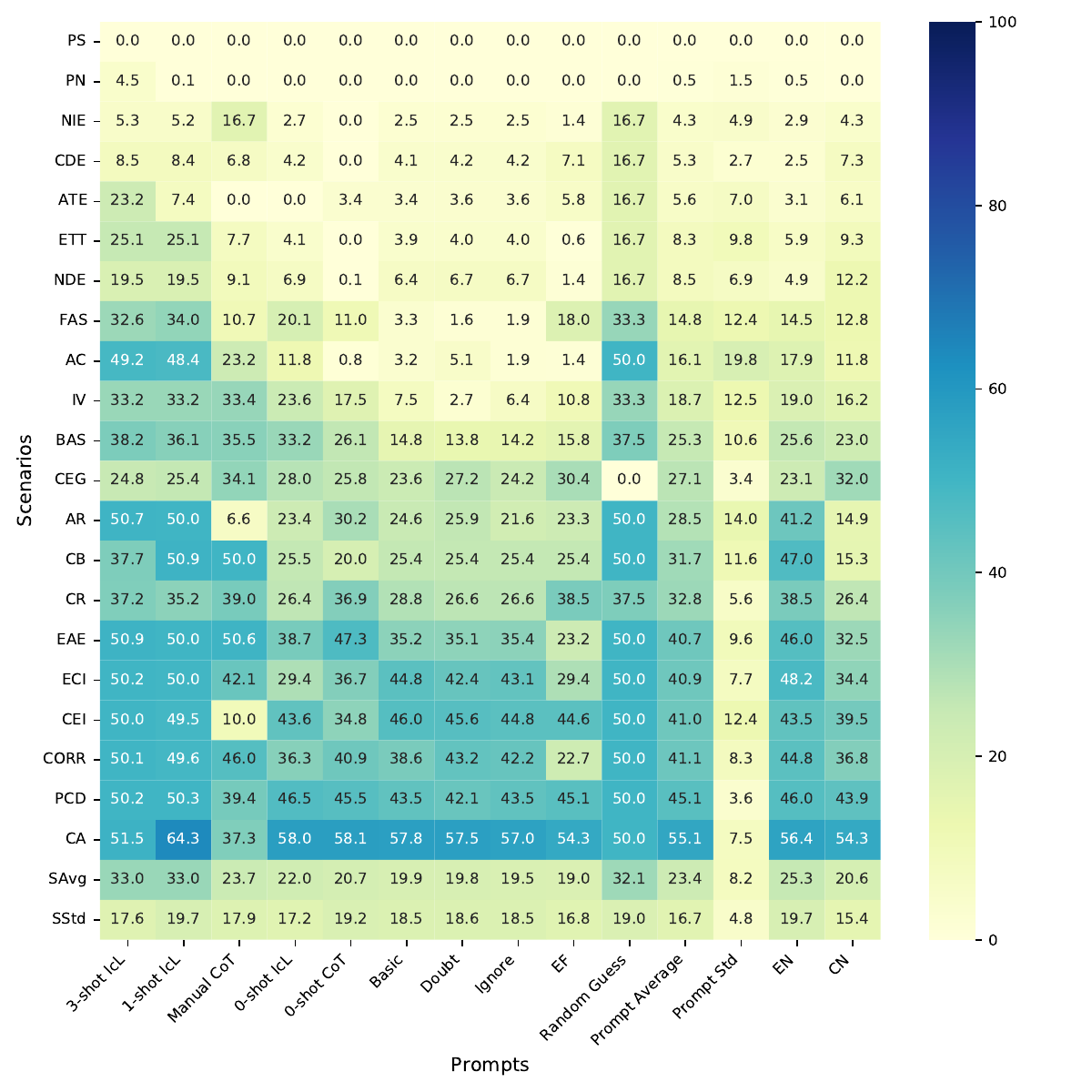}
\end{minipage}
}
\subfigure[\textit{Model-prompt rank} of babbage (1.3B)]{
\begin{minipage}{8.5cm}
\centering
\includegraphics[width=1\linewidth]{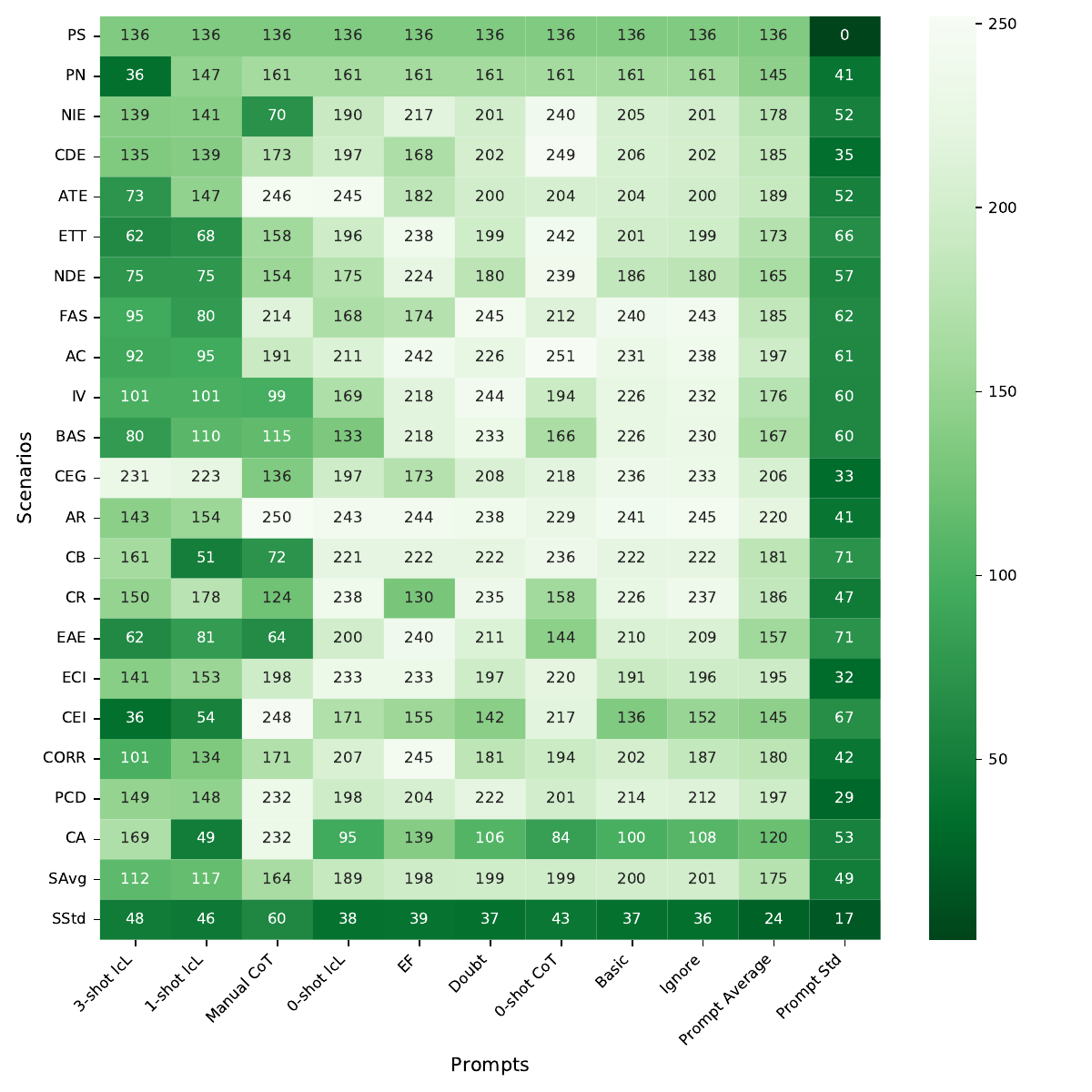}
\end{minipage}
}
\caption[Heatmap of babbage (1.3B)]{\textbf{Heatmap of babbage (1.3B).}}
\label{fig:Heatmap_of_babbage_(1.3B)}
\end{figure}

Summary: The model's performance on scenario-based prompts yields an average accuracy rate of 23.4\%. It ranks 23rd out of 28 when considering the average of \textit{prompt-average rank}. Additionally, the model demonstrates an average robustness score of 94.9\%.

Accuracy: 1) Overall performance: Figure \ref{fig:Heatmap_of_babbage_(1.3B)}(a) shows babbage (1.3B) with an \textit{average scenario-prompt accuracy} of 23.4\% and an average standard deviation for prompt effectiveness of 8.2. The \textit{top scenario-prompt pair}s are 1-shot IcL in CA with a score of 64.3\%, followed by 0-shot CoT in CA at 58.1\% and 0-shot IcL in CA at 58.0\%. About 30.7\% of the \textit{scenario-prompt pairs} surpass the \textit{random guess accuracy}, yet none achieve over 80\% accuracy.
2) Scenario performance: For scenarios in which babbage (1.3B) exceeds the \textit{random guess accuracy}, the top three scenarios ranked by average accuracy include CA with an impressive score of 55.1\%, CEG at 27.1\%, and PN at 0.5\%.
3) Prompt efficiency: The top-performing prompts are 3-shot IcL and 1-shot IcL, each with a score of 33.0\%. For exceeding the \textit{random guess accuracy}, 3-shot IcL leads in 14 out of 21 scenarios, followed by 1-shot IcL in 11, and manual CoT in 8 scenarios.
4) Language influence: English outperforms Chinese in 14 out of 21 scenarios, with significant accuracy advantages in CB, AR, and ECI, where the \textit{language accuracy difference}s are 31.7\%, 26.3\%, and 13.8\%, respectively. However, in scenarios like CEG, NDE, and CDE, Chinese shows better performance, with \textit{language accuracy difference}s of 8.9\%, 7.3\%, and 4.8\%, respectively.

Ranking: 1) \textit{Prompt-average rank}: As shown in Figure \ref{fig:Prompt-Average_Rank_of_Models}, babbage (1.3B)'s best \textit{prompt-average rank}s is lower than 12. On the other hand, its lowest ranks are in AR at 28, AC at 27, and NIE at 27. The model's average \textit{prompt-average rank} across 21 scenarios is 23rd out of 28, with a standard deviation of 3.3.
2) \textit{Model-prompt rank}: As detailed in Figure \ref{fig:Heatmap_of_babbage_(1.3B)}(b), babbage (1.3B)'s highest ranks include PN with 3-shot IcL at 36, CEI with 3-shot IcL at 36, and CA with 1-shot IcL at 49. The lowest ranks are in AC with 0-shot CoT at 251, AR with manual CoT at 250, and CDE with 0-shot CoT at 249.

Robustness: babbage (1.3B) has an outstanding average robustness score of 94.9\% across scenarios, with the highest robustness in CB and EAE, both at 100.0\%, and PN at 99.3\%.

\paragraph{text-babbage-001.}
\begin{figure}[t]
\centering
\subfigure[Performance of text-babbage-001]{
\begin{minipage}{8.5cm}
\centering
\includegraphics[width=1\linewidth]{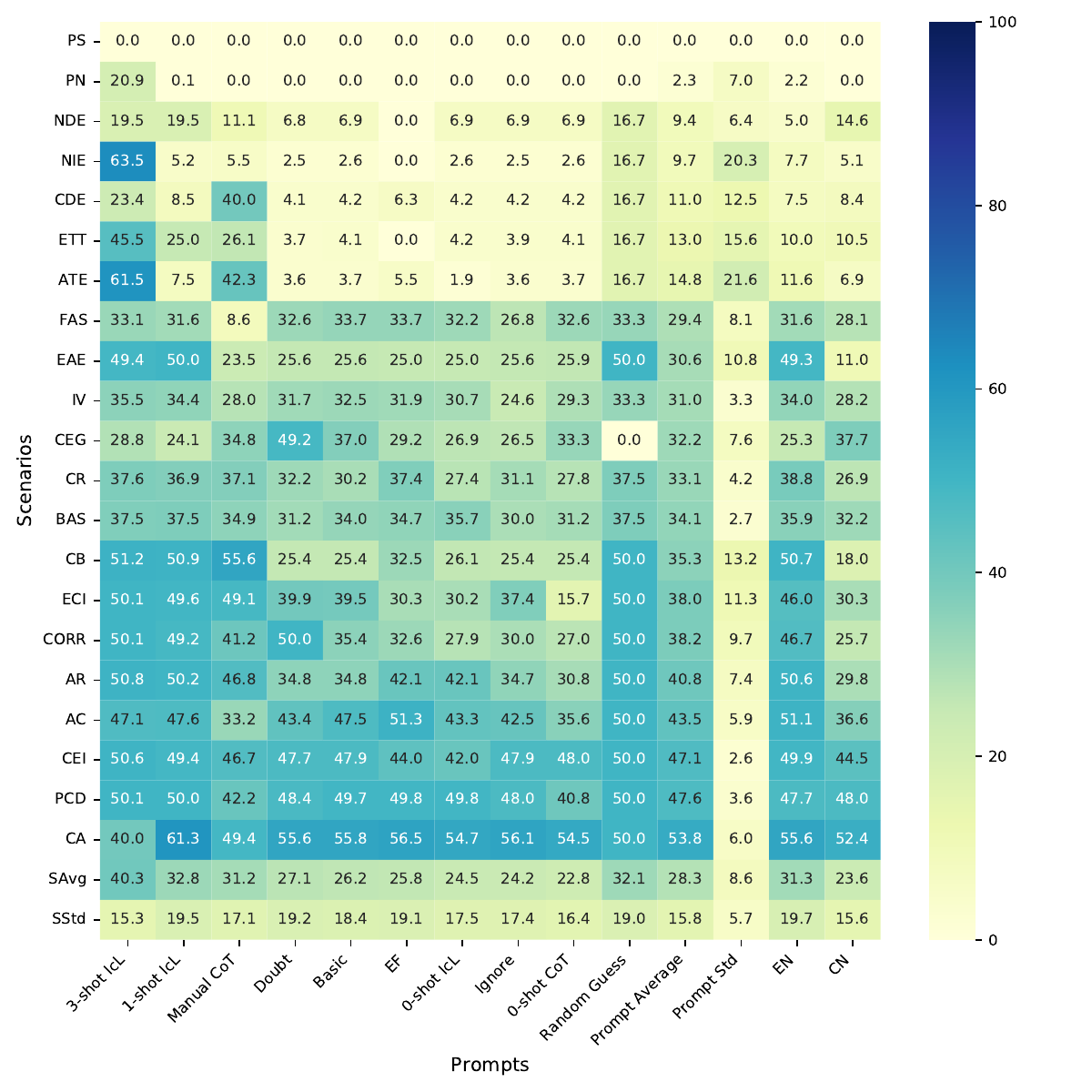}
\end{minipage}
}
\subfigure[\textit{Model-prompt rank} of text-babbage-001]{
\begin{minipage}{8.5cm}
\centering
\includegraphics[width=1\linewidth]{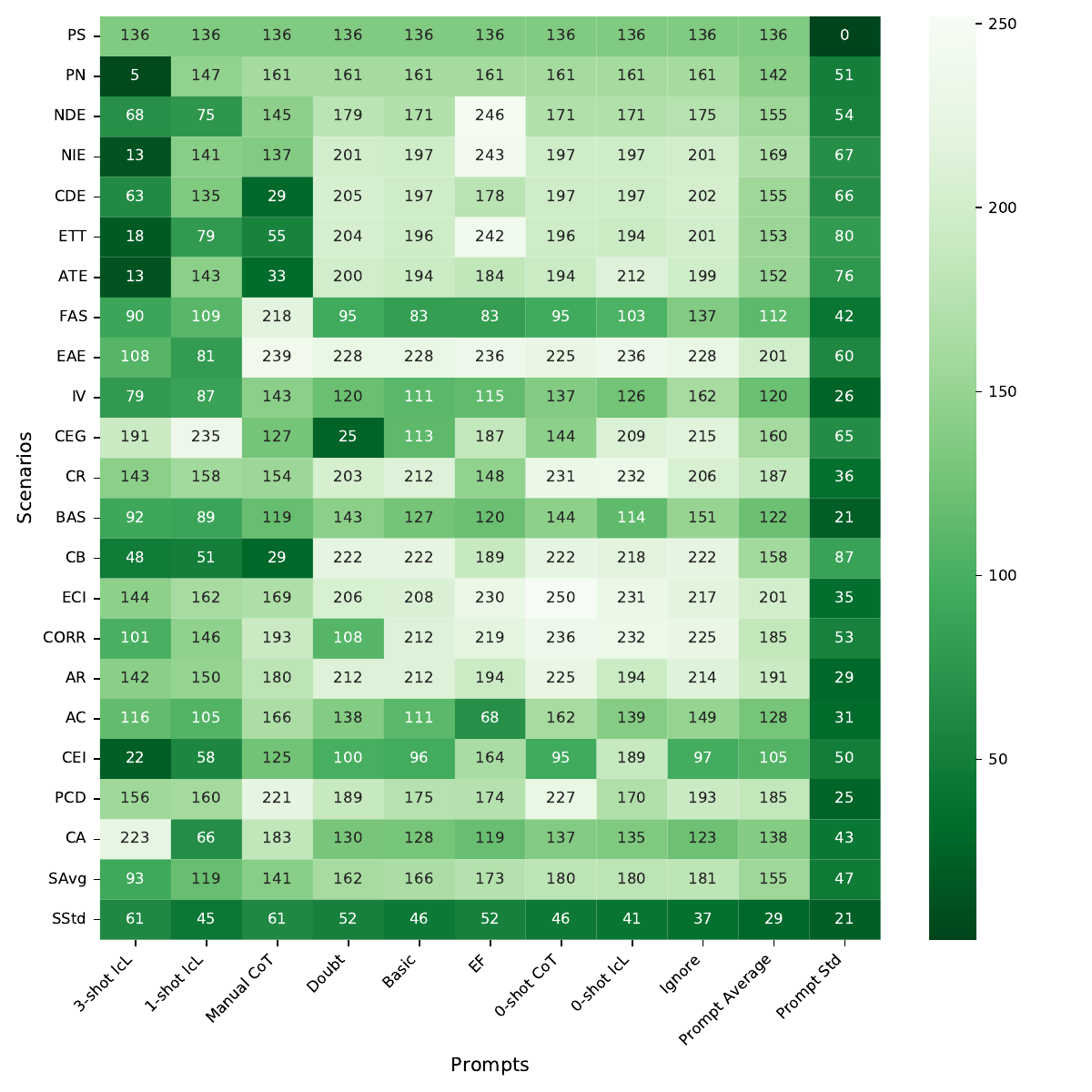}
\end{minipage}
}
\caption[Heatmap of text-babbage-001]{\textbf{Heatmap of text-babbage-001.}}
\label{fig:Heatmap_of_text-babbage-001}
\end{figure}
Summary: The \textit{average scenario-prompt accuracy} of the model stands at 28.3\%, alongside an average \textit{prompt-average rank} of 19/28 and an average robustness score of 94.3\%.

Accuracy: 1) Overall performance: Illustrated in Figure \ref{fig:Heatmap_of_text-babbage-001}(a), text-babbage-001 achieves an \textit{average scenario-prompt accuracy} of 28.3\%, with an average standard deviation in prompt effectiveness of 8.6. The \textit{top scenario-prompt pair}s include 3-shot IcL in NIE with a score of 63.5\%, followed by 3-shot IcL in ATE at 61.5\%, and 1-shot IcL in CA at 61.3\%. About 33.3\% of the \textit{scenario-prompt pairs} surpass the \textit{random guess accuracy}, with none exceeding 80\% accuracy.
2) Scenario performance: For scenarios where text-babbage-001 outperforms the \textit{random guess accuracy}, the three leading scenarios by average accuracy are CA at 53.8\%, CEG scoring 32.2\%, and PN at 2.3\%.
3) Prompt efficiency: The most impactful prompts identified are 3-shot IcL with an effectiveness of 40.3\% and 1-shot IcL at 32.8\%. In terms of surpassing the \textit{random guess accuracy}, 3-shot IcL is the leader in 16 out of 21 scenarios, followed by 1-shot IcL in 12 scenarios and manual CoT in 7 scenarios.
4) Language influence: In 15 of the 21 scenarios, English demonstrates superior performance over Chinese, with significant accuracy improvements observed in EAE, CB, and CORR, with \textit{language accuracy difference}s of 38.3\%, 32.7\%, and 21.0\%, respectively. Conversely, Chinese outperforms in scenarios like CEG, NDE, and CDE, with \textit{language accuracy difference}s of 12.4\%, 9.6\%, and 0.9\%, respectively.

Ranking: 1) \textit{Prompt-average rank}: As indicated in Figure \ref{fig:Prompt-Average_Rank_of_Models}, text-babbage-001's highest \textit{prompt-average rank}s are in CEI at 10, BAS and IV both at 11. On the other hand, its lowest ranks are in EAE and ECI at 27, and CORR at 25, highlighting potential areas for enhancement. Across 21 scenarios, the average \textit{prompt-average rank} is 19 out of 28, with a standard deviation of 5.5.
2) \textit{Model-prompt rank}: As per Figure \ref{fig:Heatmap_of_text-babbage-001}(b), text-babbage-001's top \textit{model-prompt rank}s include PN with 3-shot IcL at rank 5, ATE with 3-shot IcL at 13, and NIE with 3-shot IcL also at 13. The lowest ranks are noted in ECI with 0-shot CoT at 250, NDE with EF at 246, and NIE with EF at 243.

Robustness: text-babbage-001 has an impressive average robustness score of 94.3\% across scenarios, achieving peak robustness in EAE at 100.0\%, CB at 99.8\%, and NIE at 99.3\%.

\paragraph{curie (6.7B).}
\begin{figure}[t]
\centering
\subfigure[Performance of curie (6.7B)]{
\begin{minipage}{8.5cm}
\centering
\includegraphics[width=1\linewidth]{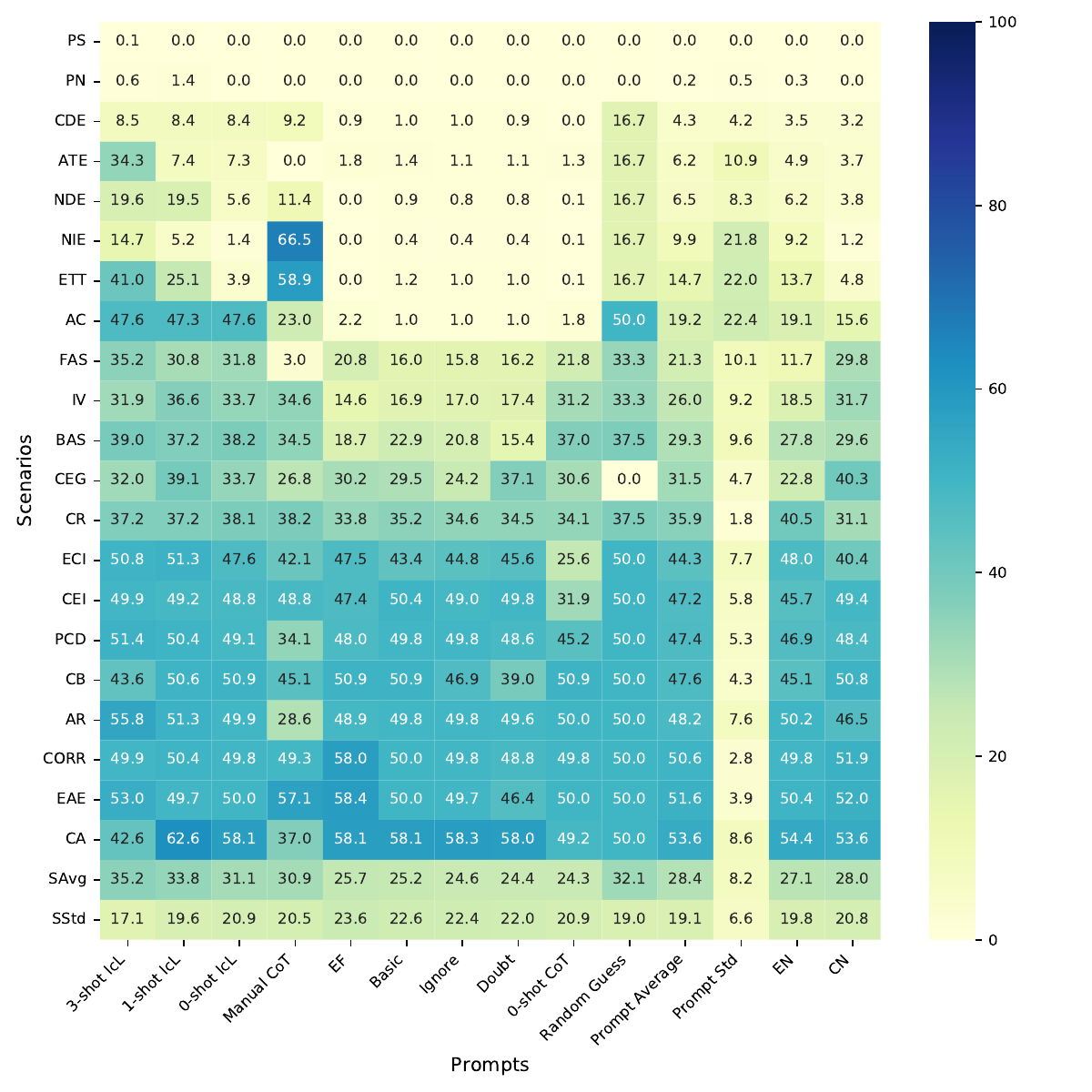}
\end{minipage}
}
\subfigure[\textit{Model-prompt rank} of curie (6.7B)]{
\begin{minipage}{8.5cm}
\centering
\includegraphics[width=1\linewidth]{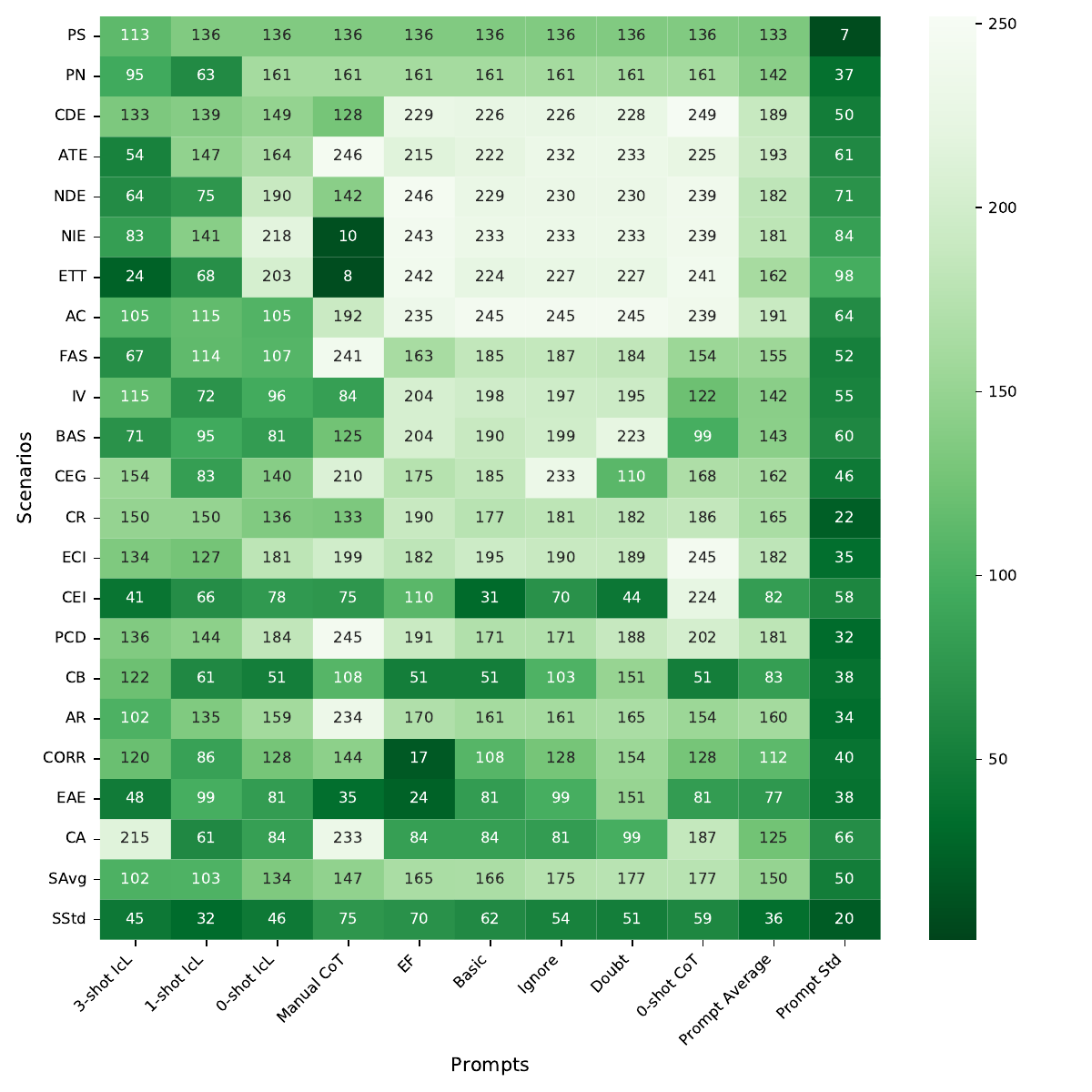}
\end{minipage}
}
\caption[Heatmap of curie (6.7B)]{\textbf{Heatmap of curie (6.7B).}}
\label{fig:Heatmap_of_curie_(6.7B)}
\end{figure}
Summary: The model achieves an \textit{average scenario-prompt accuracy} of 28.4\%, holds an average \textit{prompt-average rank} of 18/28, and maintains an average robustness score of 93.5\%.

Accuracy: 1) Overall performance: Figure \ref{fig:Heatmap_of_curie_(6.7B)}(a) showcases that curie (6.7B) achieves an \textit{average scenario-prompt accuracy} of 28.4\%, with an average prompt effectiveness variability of 8.2. The \textit{top scenario-prompt pair}s include manual CoT in NIE with a score of 66.5\%, 1-shot IcL in CA at 62.6\%, and manual CoT in ETT at 58.9\%. Approximately 37.0\% of the \textit{scenario-prompt pairs} outperform the \textit{random guess accuracy}, yet none surpass 80\% accuracy.
2) Scenario performance: In situations where curie (6.7B) exceeds the \textit{random guess accuracy}, the top three scenarios by average accuracy are CA at 53.6\%, EAE at 51.6\%, and CORR at 50.6\%.
3) Prompt efficiency: The leading prompts in effectiveness are 3-shot IcL at 35.2\% and 1-shot IcL at 33.8\%. In terms of surpassing the \textit{random guess accuracy}, 3-shot IcL and 1-shot IcL both lead in 12 out of 21 scenarios, followed by 0-shot IcL in 9 scenarios.
4) Language influence: English demonstrates superior performance over Chinese in 11 out of 21 scenarios, with significant advantages in CR, ETT, and NIE, with \textit{language accuracy difference}s of 9.3\%, 8.9\%, and 7.9\%, respectively. In contrast, Chinese excels in FAS, CEG, and IV, with \textit{language accuracy difference}s of 18.1\%, 17.5\%, and 13.2\%, respectively.

Ranking: 1) \textit{Prompt-average rank}: According to Figure \ref{fig:Prompt-Average_Rank_of_Models}, curie (6.7B)'s highest \textit{prompt-average rank}s are seen in CEI, EAE, and CB, each at 7. Meanwhile, it ranks worst in CDE at 27th place, while PN, AC, and ATE rank at 25th. The average \textit{prompt-average rank} across 21 scenarios is 18 out of 28, with a standard deviation of 6.5.
2) \textit{Model-prompt rank}: As detailed in Figure \ref{fig:Heatmap_of_curie_(6.7B)}(b), the top ranks for curie (6.7B) include ETT with manual CoT at 8, NIE with manual CoT at 10, and CORR with EF at 17. The lowest ranks are in CDE with 0-shot CoT at 249, ATE with manual CoT at 246, and NDE with EF at 246.

Robustness: curie (6.7B) exhibits a robust average score of 93.5\% across various scenarios, with the highest robustness in AC at 100.0\%, AR at 99.5\%, and NIE at 98.9\%.

\paragraph{text-curie-001.}
\begin{figure}[t]
\centering
\subfigure[Performance of text-curie-001]{
\begin{minipage}{8.5cm}
\centering
\includegraphics[width=1\linewidth]{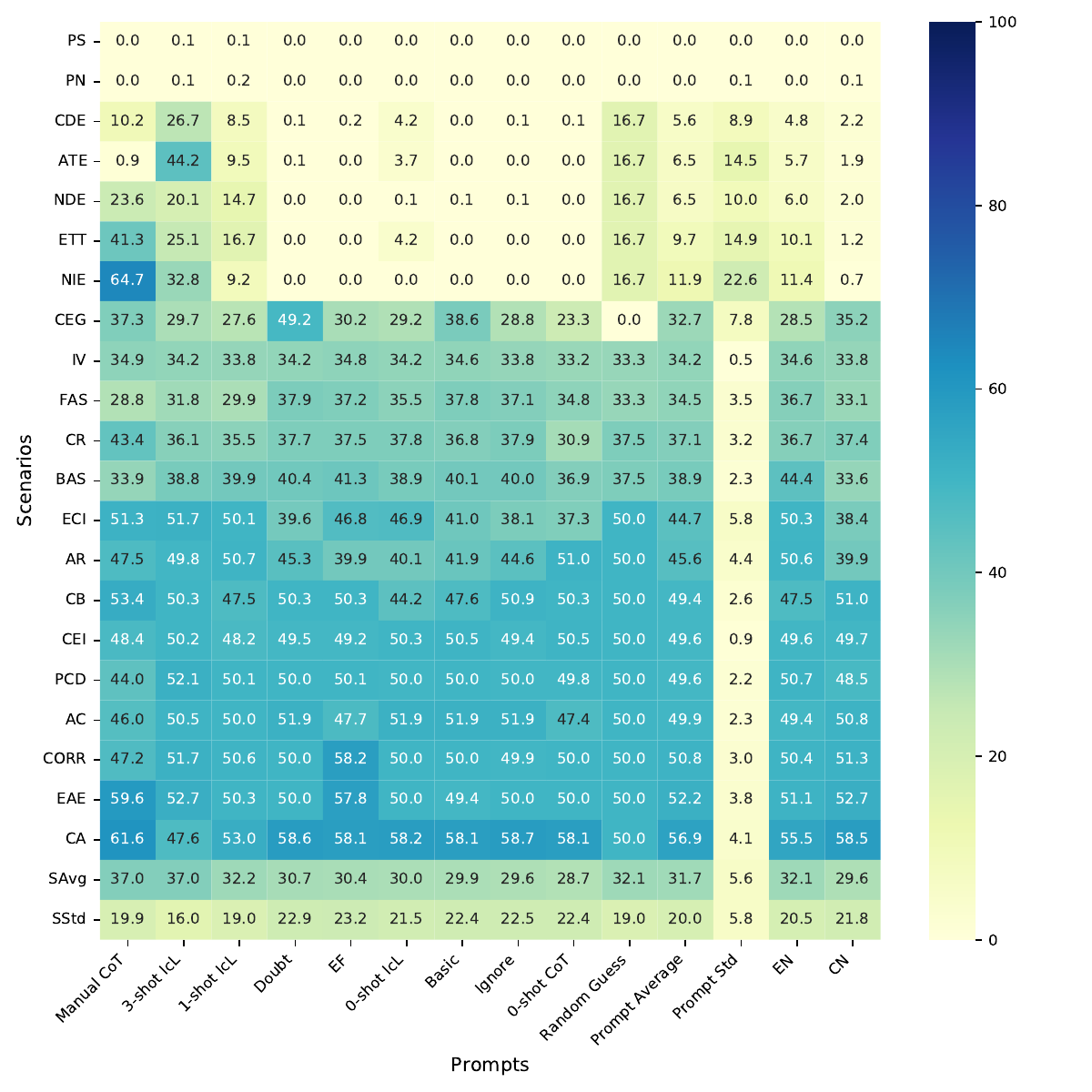}
\end{minipage}
}
\subfigure[\textit{Model-prompt rank} of text-curie-001]{
\begin{minipage}{8.5cm}
\centering
\includegraphics[width=1\linewidth]{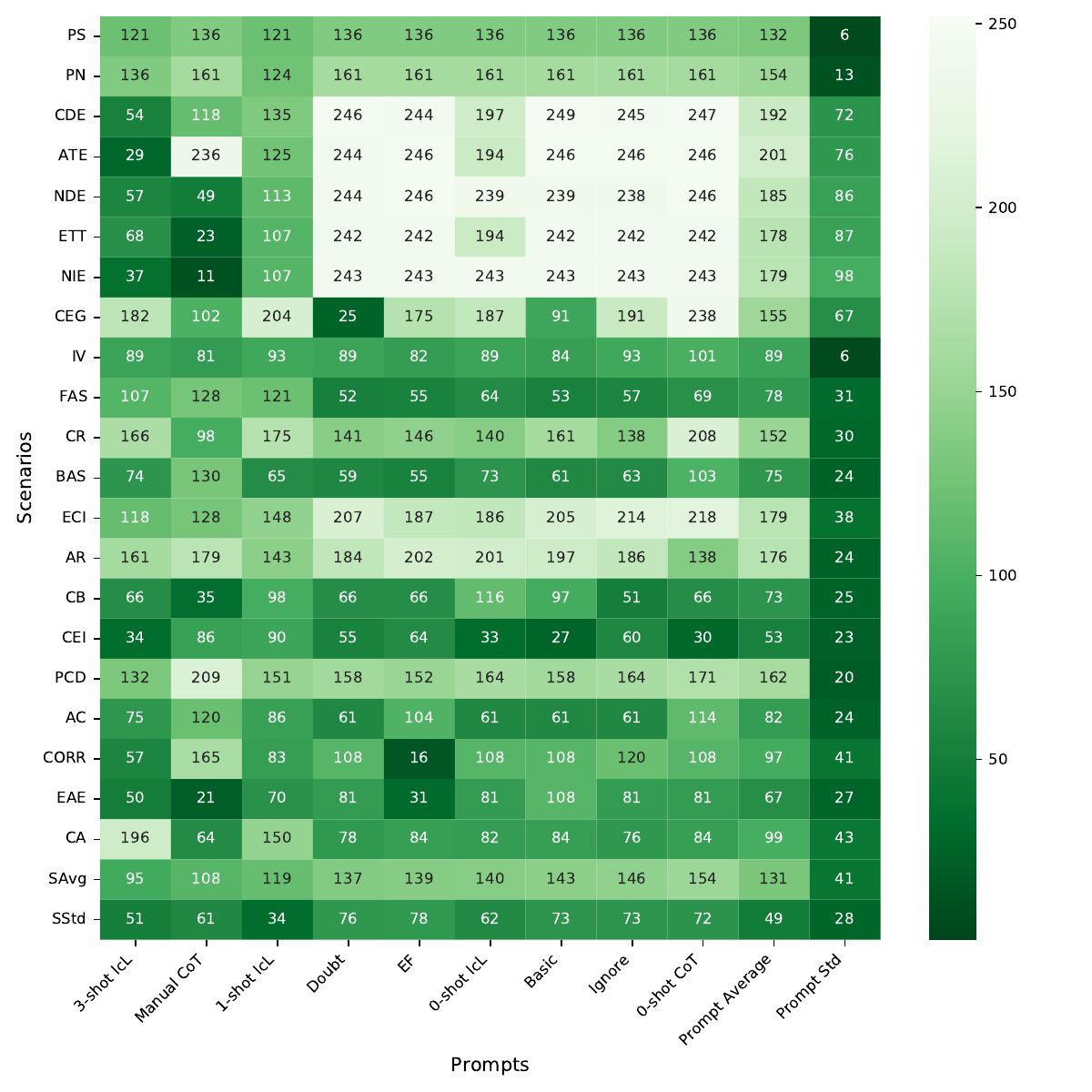}
\end{minipage}
}
\caption[Heatmap of text-curie-001]{\textbf{Heatmap of text-curie-001.}}
\label{fig:Heatmap_of_text-curie-001}
\end{figure}
Summary: The model's performance accuracy for scenario prompts averages 31.7\%, with its average \textit{prompt-average rank} at 15 out of 28 and an average robustness score of 91.6\%.

Accuracy: 1) Overall performance: Figure \ref{fig:Heatmap_of_text-curie-001}(a) illustrates that text-curie-001 achieves an \textit{average scenario-prompt accuracy} of 31.7\%, with an average std for variability in prompt effectiveness of 5.6. The \textit{top scenario-prompt pair}s include manual CoT in NIE with a score of 64.7\%, followed by manual CoT in CA at 61.6\%, and manual CoT in EAE at 59.6\%. A total of 58.2\% of \textit{scenario-prompt pairs} outperform the \textit{random guess accuracy}, yet none surpass 80\% accuracy.
2) Scenario performance: In scenarios where the model surpasses the \textit{random guess accuracy}, the top 3 scenarios having the highest average accuracy are CA with a score of 56.9\%, EAE at 52.2\%, and CORR at 50.8\%.
3) Prompt efficiency: The top prompts in terms of effectiveness are manual CoT and 3-shot IcL both at 37.0\%, followed by 1-shot IcL at 32.2\%. Regarding exceeding the \textit{random guess accuracy}, 3-shot IcL outperforms in 17 of 21 scenarios. It is followed by adversarial doubt achieving this in 13 scenarios, and manual CoT, 1-shot IcL, EF, and 0-shot IcL each lead in 12 scenarios.
4) Language influence: In 11 of the 21 scenarios, English demonstrates superior performance over Chinese, particularly in ECI, BAS, and AR, with \textit{language accuracy difference}s of 11.9\%, 10.8\%, and 10.7\%, respectively. Conversely, Chinese shows better performance in CEG, CB, and CA, with \textit{language accuracy difference}s of 6.7\%, 3.5\%, and 3.0\%, respectively.

Ranking: 1) \textit{Prompt-average rank}: As shown in Figure \ref{fig:Prompt-Average_Rank_of_Models}, text-curie-001 achieves its best \textit{prompt-average rank}s with a 2nd place in CEI, 5th in EAE, and ties for 6th in AC, CB, and FAS. On the other hand, its lowest ranks are in ETT and PN, both at 27th, followed by ATE and NDE, each at 26th. The average \textit{prompt-average rank} across 21 scenarios is 15 out of 28, with a standard deviation of 8.8.
2) \textit{Model-prompt rank}: Figure \ref{fig:Heatmap_of_text-curie-001}(b) shows text-curie-001's best ranks include NIE with manual CoT at 11, CORR with EF at 16, and EAE with manual CoT at 21. The lowest ranks are in CDE with basic at 249, with 0-shot CoT at 247, and NDE with 0-shot CoT at 246.

Robustness: text-curie-001 exhibits a robust average score of 91.6\% across various scenarios, achieving optimal robustness in AC at 100.0\%, CORR at 99.8\%, and NDE at 99.5\%.

\paragraph{davinci (175B).}
\begin{figure}[t]
\centering
\subfigure[Performance of davinci (175B)]{
\begin{minipage}{8.5cm}
\centering
\includegraphics[width=1\linewidth]{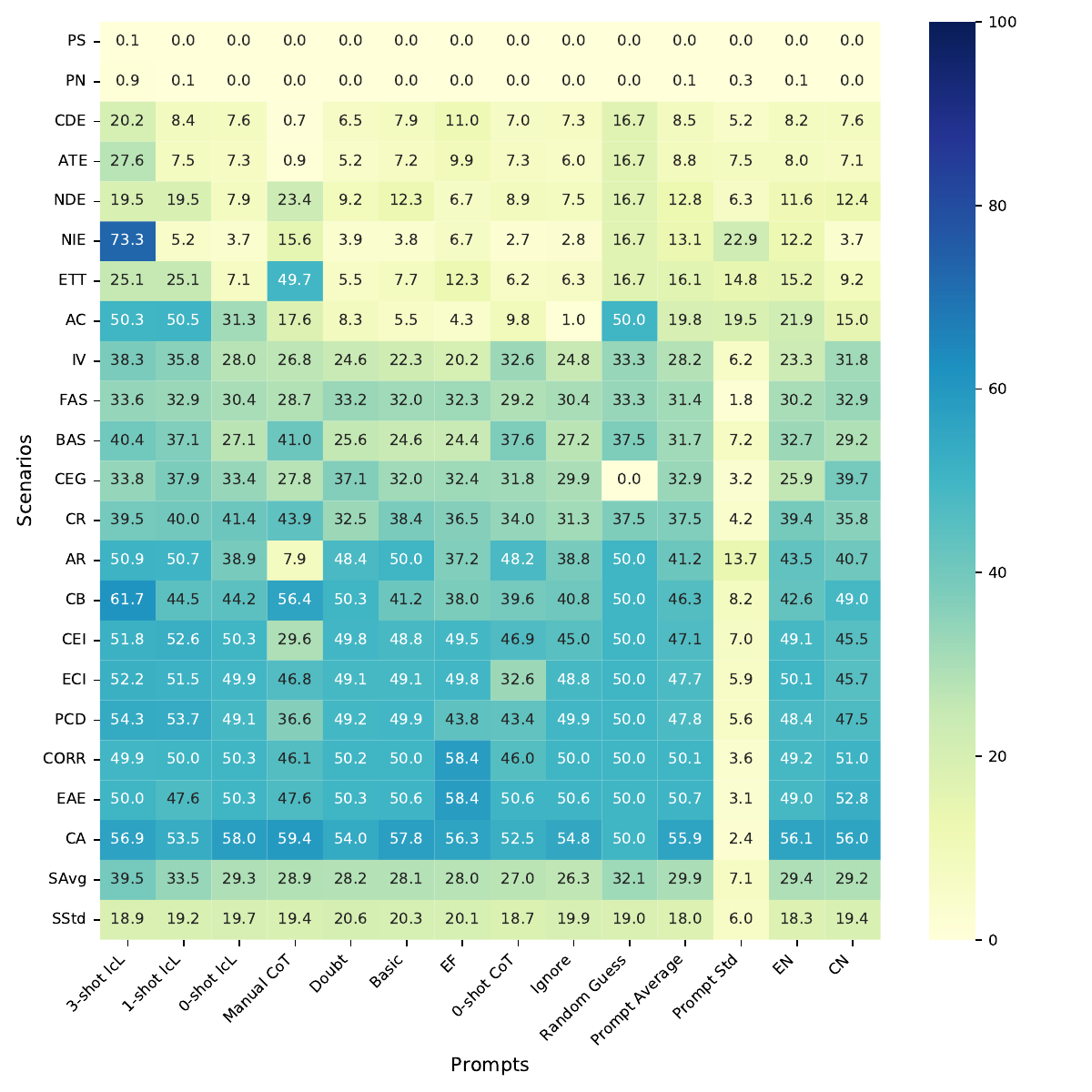}
\end{minipage}
}
\subfigure[\textit{Model-prompt rank} of davinci (175B)]{
\begin{minipage}{8.5cm}
\centering
\includegraphics[width=1\linewidth]{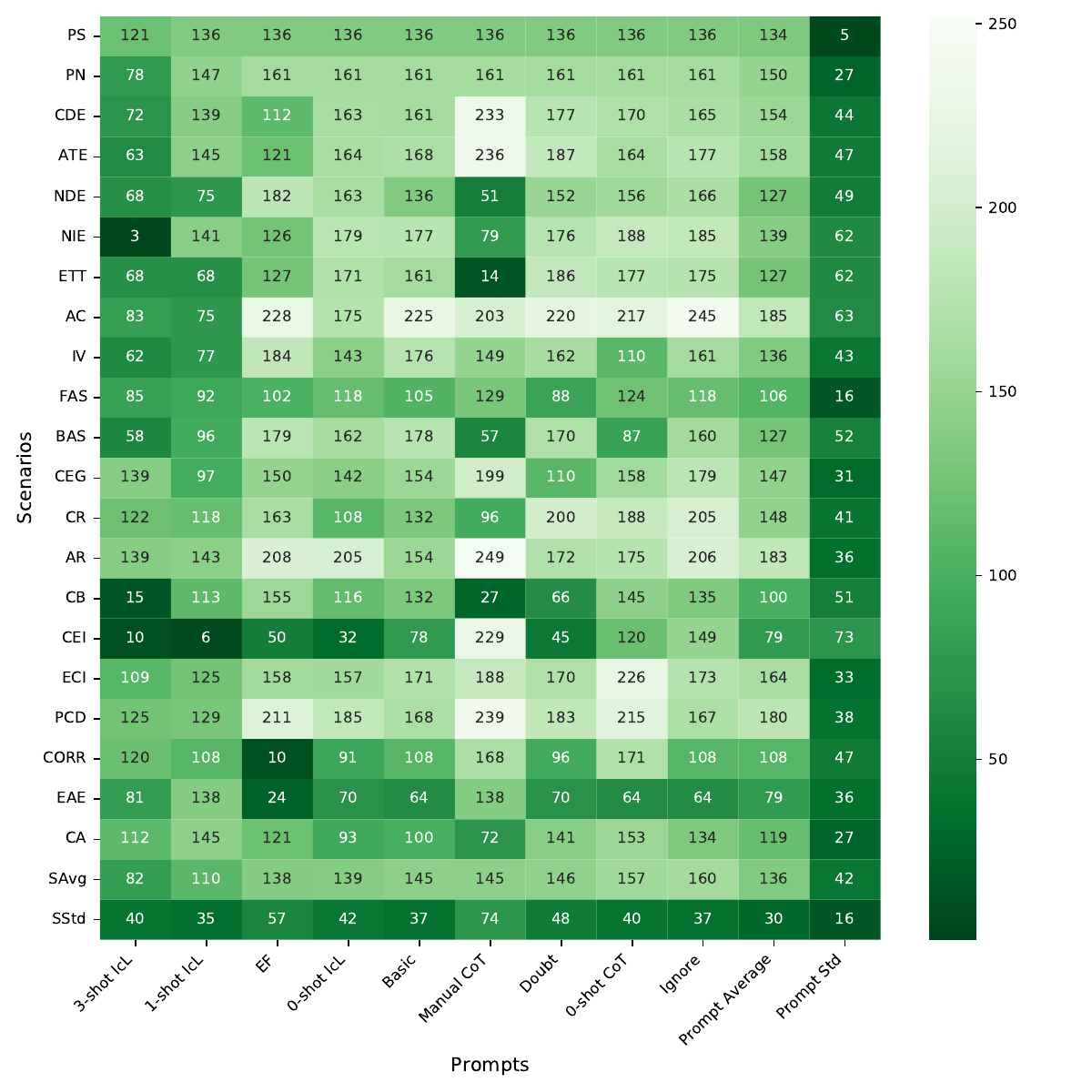}
\end{minipage}
}
\caption[Heatmap of davinci (175B)]{\textbf{Heatmap of davinci (175B).}}
\label{fig:Heatmap_of_davinci_(175B)}
\end{figure}
Summary: The model records an \textit{average scenario-prompt accuracy} of 29.9\%, alongside an average \textit{prompt-average rank} of 15 out of 28 and an average robustness score of 85.8\%.

Accuracy: 1) Overall performance: Presented in Figure \ref{fig:Heatmap_of_davinci_(175B)}(a), davinci (175B) achieves an \textit{average scenario-prompt accuracy} of 29.9\%, with an average standard deviation for prompt effectiveness at 7.1. The \textit{top scenario-prompt pair}s include a 3-shot IcL in NIE with a score of 73.3\%, a 3-shot IcL in CB at 61.7\%, and a manual CoT in CA at 59.4\%. A total of 44.4\% of \textit{scenario-prompt pairs} surpass the \textit{random guess accuracy}, yet none exceed an 80\% accuracy mark.
2) Scenario performance: Among scenarios where davinci (175B) outperforms the \textit{random guess accuracy}, the highest accuracies are observed in CA at 55.9\%, EAE at 50.7\%, and CORR at 50.1\%.
3) Prompt efficiency: The leading prompts in effectiveness are 3-shot IcL at 39.5\% and 1-shot IcL at 33.5\%. In terms of beating the \textit{random guess accuracy}, 3-shot IcL dominates in 20 of 21 scenarios, with 1-shot IcL following in 14 scenarios and manual CoT in 9.
4) Language influence: English surpasses Chinese in 13 of 21 scenarios, especially in NIE, AC, and ETT, with \textit{language accuracy difference}s of 8.4\%, 6.9\%, and 5.9\%, respectively. Conversely, Chinese outshines English in scenarios like CEG, IV, and CB, with \textit{language accuracy difference}s of 13.8\%, 8.5\%, and 6.4\%, respectively.

Ranking: 1) \textit{Prompt-average rank}: According to Figure \ref{fig:Prompt-Average_Rank_of_Models}, davinci (175B)'s top \textit{prompt-average rank}s are in CA and EAE, both at 8, and CEI at 9. However, it faces its lowest ranks in PN at 26, AC at 24, and AR at 23, pinpointing areas for growth. The average \textit{prompt-average rank} across 21 scenarios stands at 15 out of 28, with a variability of 5.5.
2) \textit{Model-prompt rank}: As shown in Figure \ref{fig:Heatmap_of_davinci_(175B)}(b), davinci (175B) achieves its highest \textit{model-prompt rank}s in several key areas: a third-place in NIE using 3-shot IcL, a sixth-place in CEI with 1-shot IcL, and tenth-place ranks in both CORR using EF and CEI with 3-shot IcL. The most significant challenges are noted in AR with manual CoT at 249, AC with an adversarial ignore strategy at 245, and PCD with manual CoT at 239.

Robustness: davinci (175B) reports an average robustness score of 85.8\% across various scenarios, showcasing top robustness in CORR at 99.8\%, EAE at 98.1\%, and AC at 95.3\%.

\paragraph{text-davinci-001.}
\begin{figure}[t]
\centering
\subfigure[Performance of text-davinci-001]{
\begin{minipage}{8.5cm}
\centering
\includegraphics[width=1\linewidth]{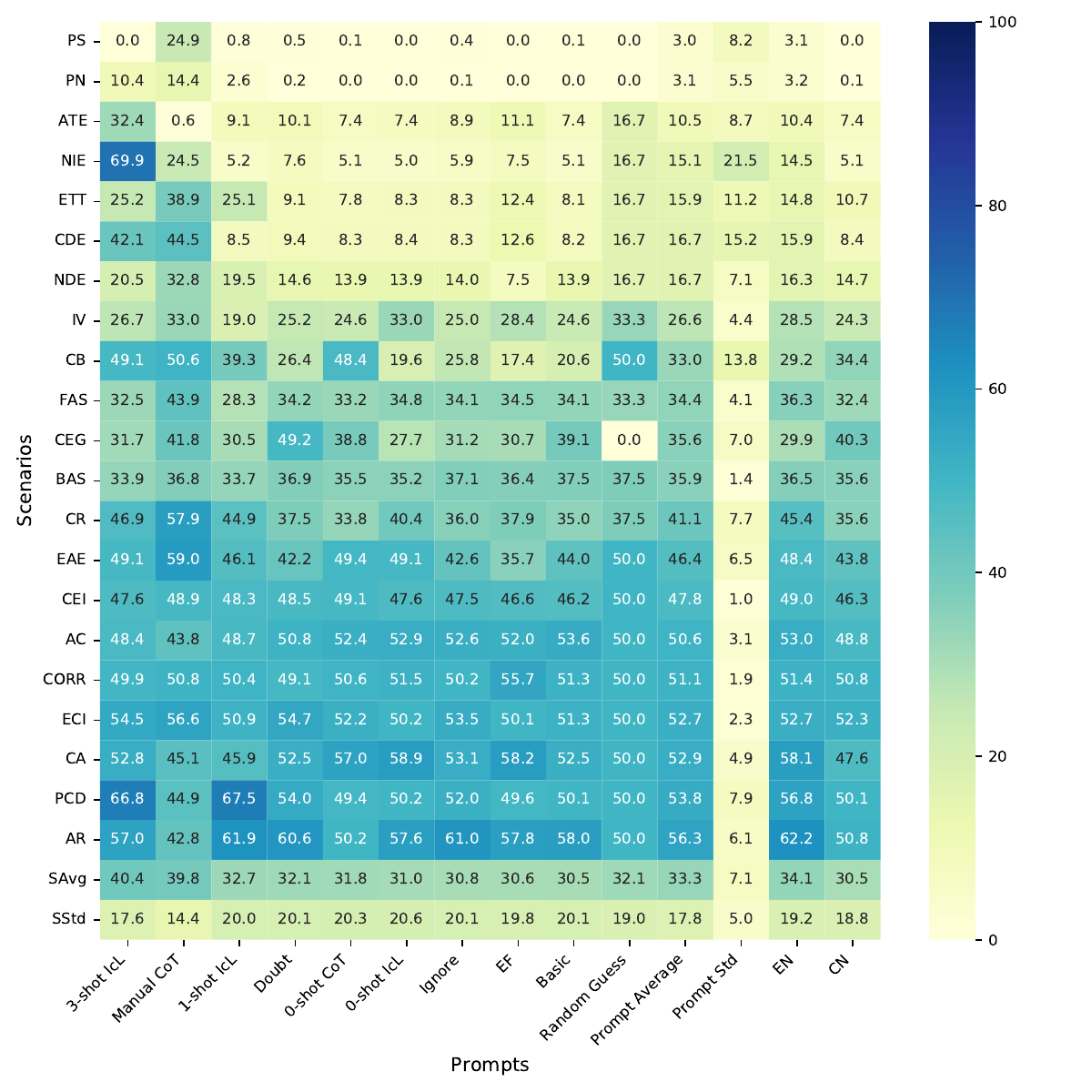}
\end{minipage}
}
\subfigure[\textit{Model-prompt rank} of text-davinci-001]{
\begin{minipage}{8.5cm}
\centering
\includegraphics[width=1\linewidth]{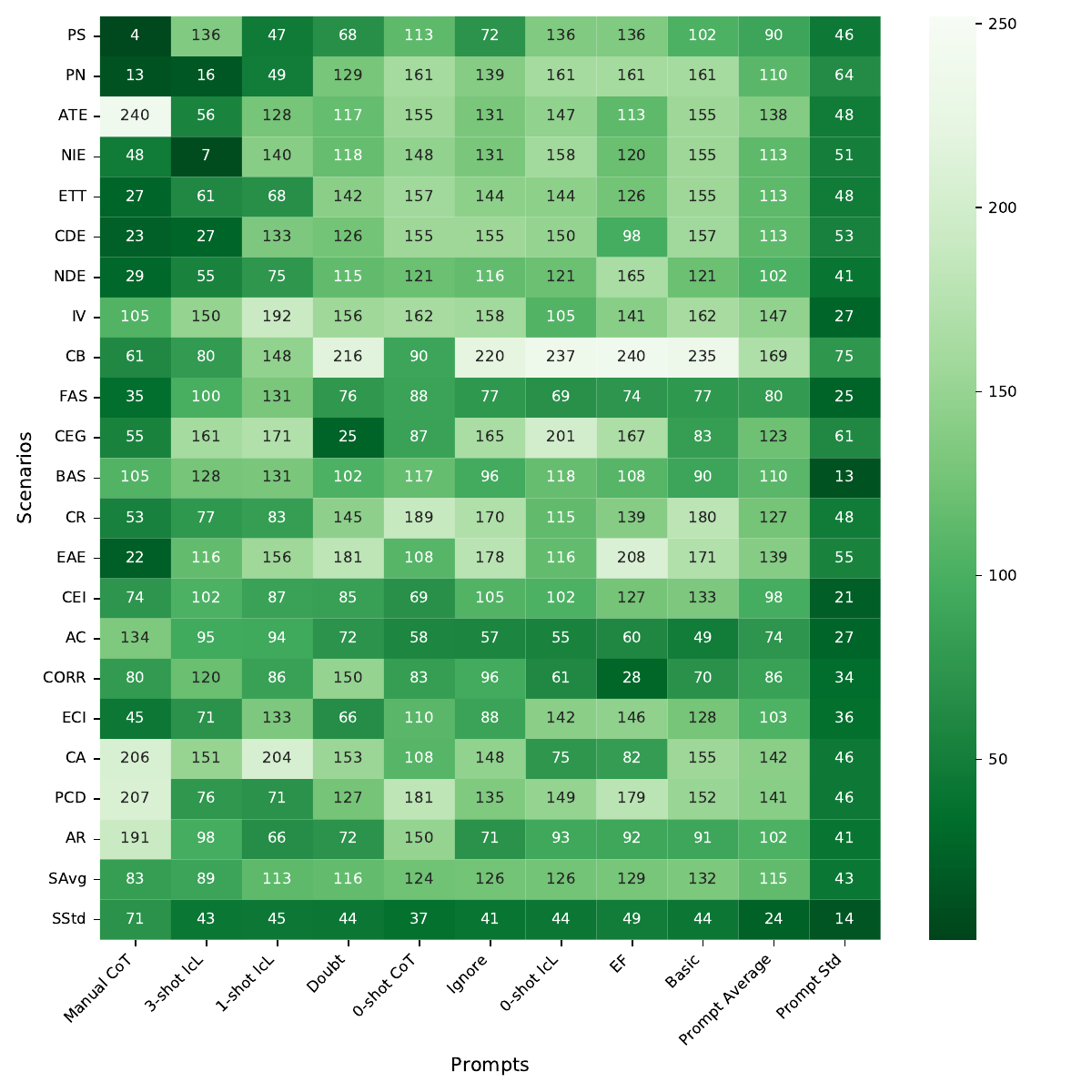}
\end{minipage}
}
\caption[Heatmap of text-davinci-001]{\textbf{Heatmap of text-davinci-001.}}
\label{fig:Heatmap_of_text-davinci-001}
\end{figure}

Summary: The model exhibits an \textit{average scenario-prompt accuracy} of 33.3\%, holds an average \textit{prompt-average rank} of 11 out of 28, and achieves an average robustness score of 76.2\%.

Accuracy: 1) Overall performance: Shown in Figure \ref{fig:Heatmap_of_text-davinci-001}(a), text-davinci-001 achieves an \textit{average scenario-prompt accuracy} of 33.3\%, with an average prompt effectiveness variability (std) of 7.1. Leading the performance metrics are the \textit{top scenario-prompt pair}s like 3-shot IcL in NIE with a high score of 69.9\%, 1-shot IcL in PCD at 67.5\%, and 3-shot IcL in the same scenario at 66.8\%. A total of 50.3\% of the \textit{scenario-prompt pairs} outperform the \textit{random guess accuracy}, yet none surpass the 80\% accuracy level.
2) Scenario performance: Within scenarios where text-davinci-001 exceeds the \textit{random guess accuracy}, the highest scores are noted in AR at 56.3\%, PCD at 53.8\%, and CA at 52.9\%.
3) Prompt efficiency: The most efficient prompts identified are 3-shot IcL at 40.4\%, manual CoT at 39.8\%, and 1-shot IcL at 32.7\%. Regarding \textit{scenario-prompt pair}s outperforming the \textit{random guess accuracy}, 3-shot IcL and manual CoT both lead in 13 out of 21 scenarios, followed by 0-shot IcL in 11 scenarios.
4) Language influence: English demonstrates superiority over Chinese in 19 of 21 scenarios, with significant accuracy benefits in AR, CA, and CR, where the \textit{language accuracy difference}s are 11.4\%, 10.5\%, and 9.7\%, respectively. In contrast, the Chinese excel in CEG and CB, with \textit{language accuracy difference}s of 10.4\%, and 5.2\%, respectively.

Ranking: 1) \textit{Prompt-average rank}: According to Figure \ref{fig:Prompt-Average_Rank_of_Models}, text-davinci-001's top \textit{prompt-average rank}s are in AC at 5, PS, CORR, and CEI at 6. However, it finds its worst ranks in CB at 25, ATE at 19, and CA at 18, revealing areas needing improvement. Across 21 scenarios, the average \textit{prompt-average rank} is 11th out of 28, with a standard deviation of 5.0.
2) \textit{Model-prompt rank}: Illustrated in Figure \ref{fig:Heatmap_of_text-davinci-001}(b), the best ranks for text-davinci-001 include PS with manual CoT at 4, NIE with 3-shot IcL at 7, and PN with manual CoT at 13. The challenges are most strong in ATE with manual CoT at 240, CB with EF at 240, and CB with 0-shot IcL at 237.

Robustness: text-davinci-001 reports an average robustness score of 76.2\% across scenarios, with standout robustness in FAS at 99.9\%, IV at 99.0\%, and PCD at 91.9\%.

\paragraph{text-davinci-002.}
\begin{figure}[t]
\centering
\subfigure[Performance of text-davinci-002]{
\begin{minipage}{8.5cm}
\centering
\includegraphics[width=1\linewidth]{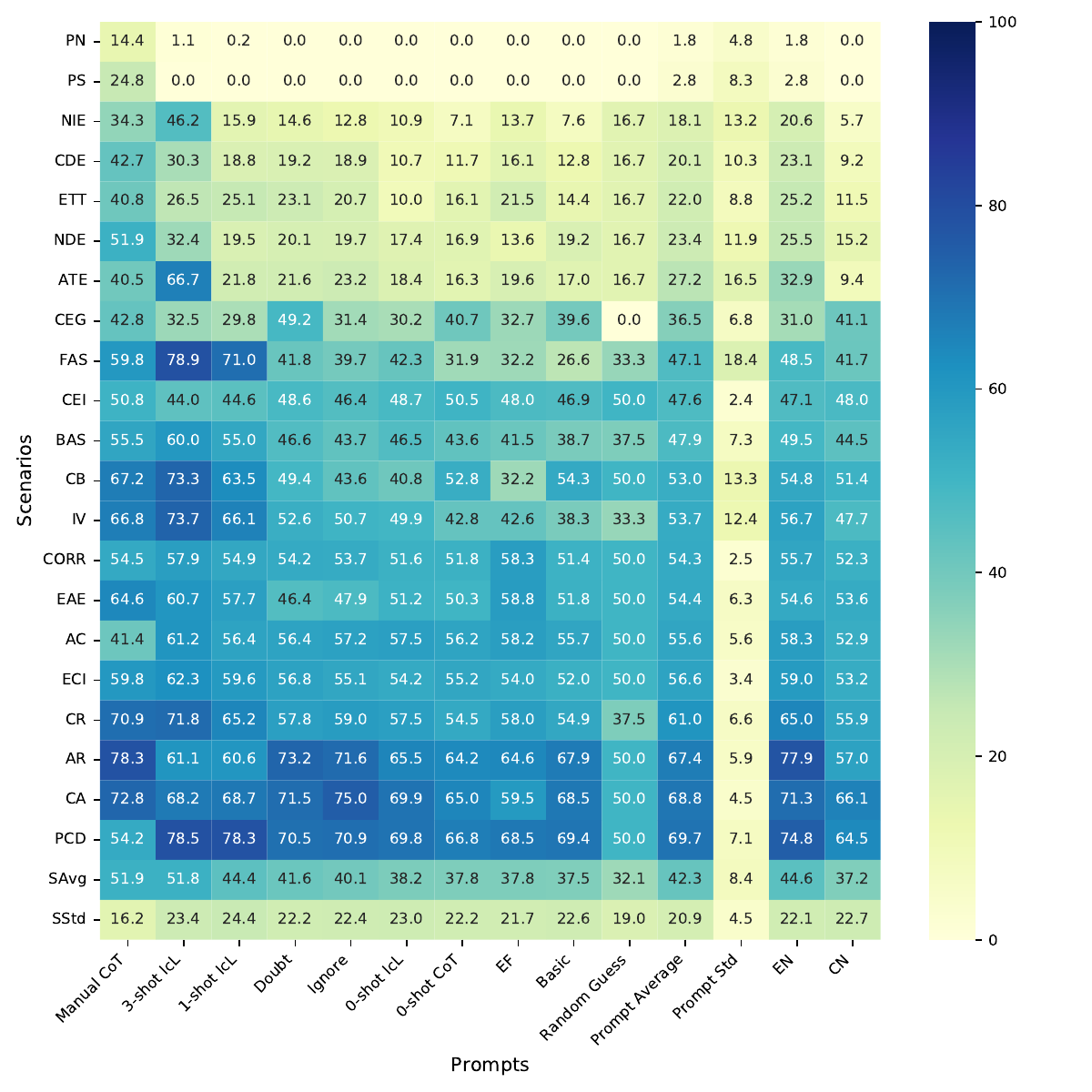}
\end{minipage}
}
\subfigure[\textit{Model-prompt rank} of text-davinci-002]{
\begin{minipage}{8.5cm}
\centering
\includegraphics[width=1\linewidth]{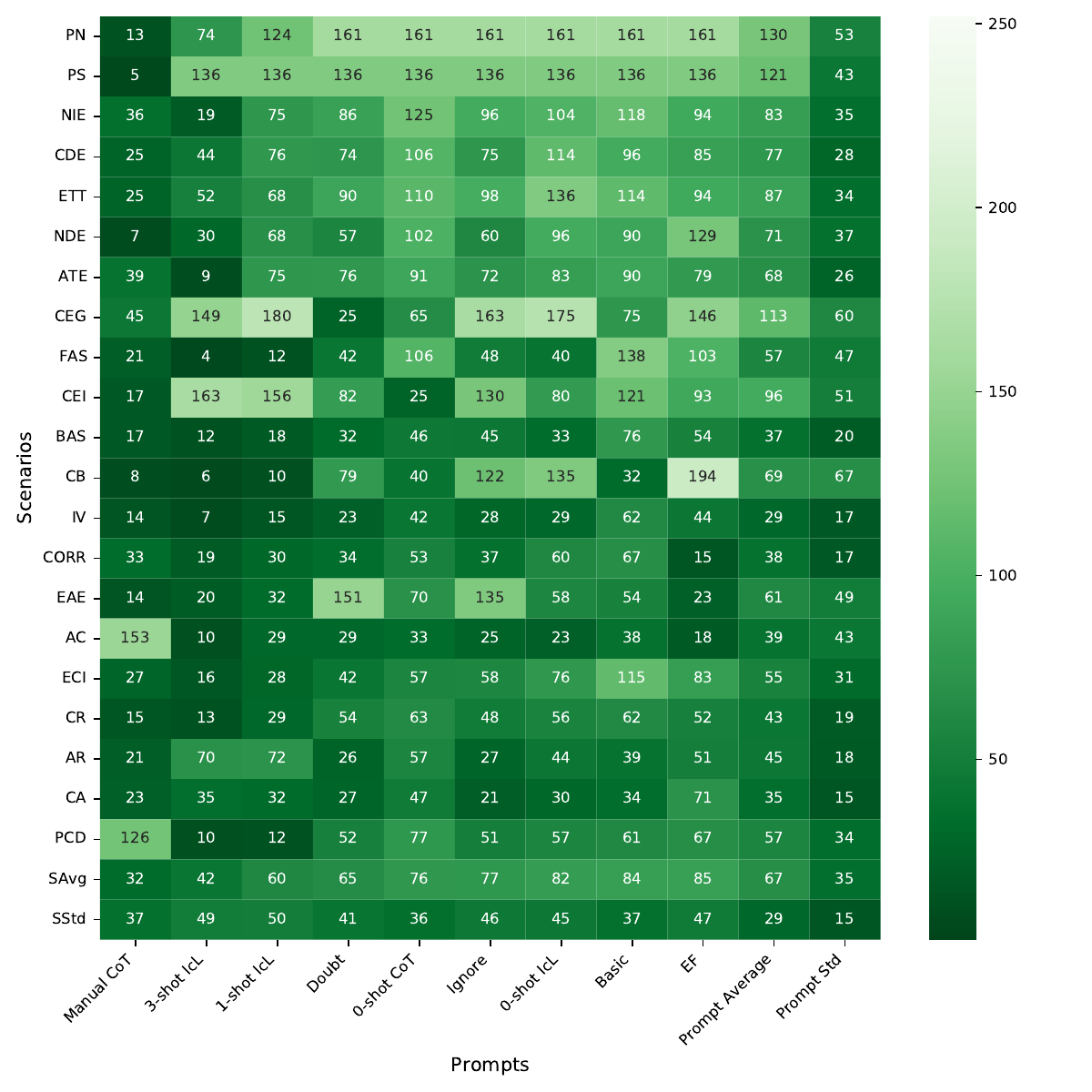}
\end{minipage}
}
\caption[Heatmap of text-davinci-002]{\textbf{Heatmap of text-davinci-002.}}
\label{fig:Heatmap_of_text-davinci-002}
\end{figure}

Summary: The model's \textit{average scenario-prompt accuracy} is 42.3\%, with the average \textit{prompt-average rank} of 6/28 and an average robustness score of 70.9\%.

Accuracy: 1) Overall performance: As shown in Figure \ref{fig:Heatmap_of_text-davinci-002}(a), text-davinci-002 achieves an impressive \textit{average scenario-prompt accuracy} of 42.3\%, with an average standard deviation in prompt effectiveness of 8.4. The \textit{top scenario-prompt pair}s include a 3-shot IcL in FAS with a score of 78.9\%, closely followed by 3-shot IcL in PCD at 78.5\%, and 1-shot IcL in the same scenario at 78.3\%. A significant 82.5\% of the \textit{scenario-prompt pairs} surpass the \textit{random guess accuracy}, though none achieve above 80\% accuracy.
2) Scenario performance: Among scenarios where text-davinci-002 outshines the \textit{random guess accuracy}, the highest scoring scenarios are PCD at 69.7\%, CA at 68.8\%, and AR at 67.4\%.
3) Prompt efficiency: The highest efficiency prompts identified are manual CoT at 51.9\%, 3-shot IcL at 51.8\%, and 1-shot IcL at 44.4\%. manual CoT and 3-shot IcL both lead in 20 out of 21 scenarios in surpassing the \textit{random guess accuracy}, with 1-shot IcL closely behind in 19 scenarios.
4) Language influence: English dominates Chinese in 19 of 21 scenarios, showcasing significant accuracy leads in scenarios such as ATE, AR, and NIE, with \textit{language accuracy difference}s of 23.5\%, 20.9\%, and 15.0\%, respectively. In contrast, Chinese perform better in scenarios like CEG, and CEI, with differences of 10.1\%, and 1.0\%, respectively.

Ranking: 1) \textit{Prompt-average rank}: Figure \ref{fig:Prompt-Average_Rank_of_Models} reveals that text-davinci-002 achieves its highest \textit{prompt-average rank}s with 2nd in CB, 3rd in IV, and 4th across five scenarios. However, it shows lower performance with 16th in PN, 12th in CEG, and 8th in both PS and CEI, indicating areas that could benefit from improvements. The average \textit{prompt-average rank} rank across 21 scenarios is an impressive 6 out of 28, with a standard deviation of 3.1.
2) \textit{Model-prompt rank}: As indicated in Figure \ref{fig:Heatmap_of_text-davinci-002}(b), text-davinci-002's best ranks are seen in FAS with 3-shot IcL at 4, PS with manual CoT at 5, and CB with 3-shot IcL at 6. The lowest ranks are in CB with EF at 194, and CEG with 1-shot and 0-shot IcL at 180 and 175, respectively.

Robustness: text-davinci-002 presents a robustness average of 70.9\% across scenarios, with the highest robustness observed in ECI at 89.1\%, AR at 87.4\%, and PCD at 85.6\%.

\paragraph{text-davinci-003.}
\begin{figure}[t]
\centering
\subfigure[Performance of text-davinci-003]{
\begin{minipage}{8.5cm}
\centering
\includegraphics[width=1\linewidth]{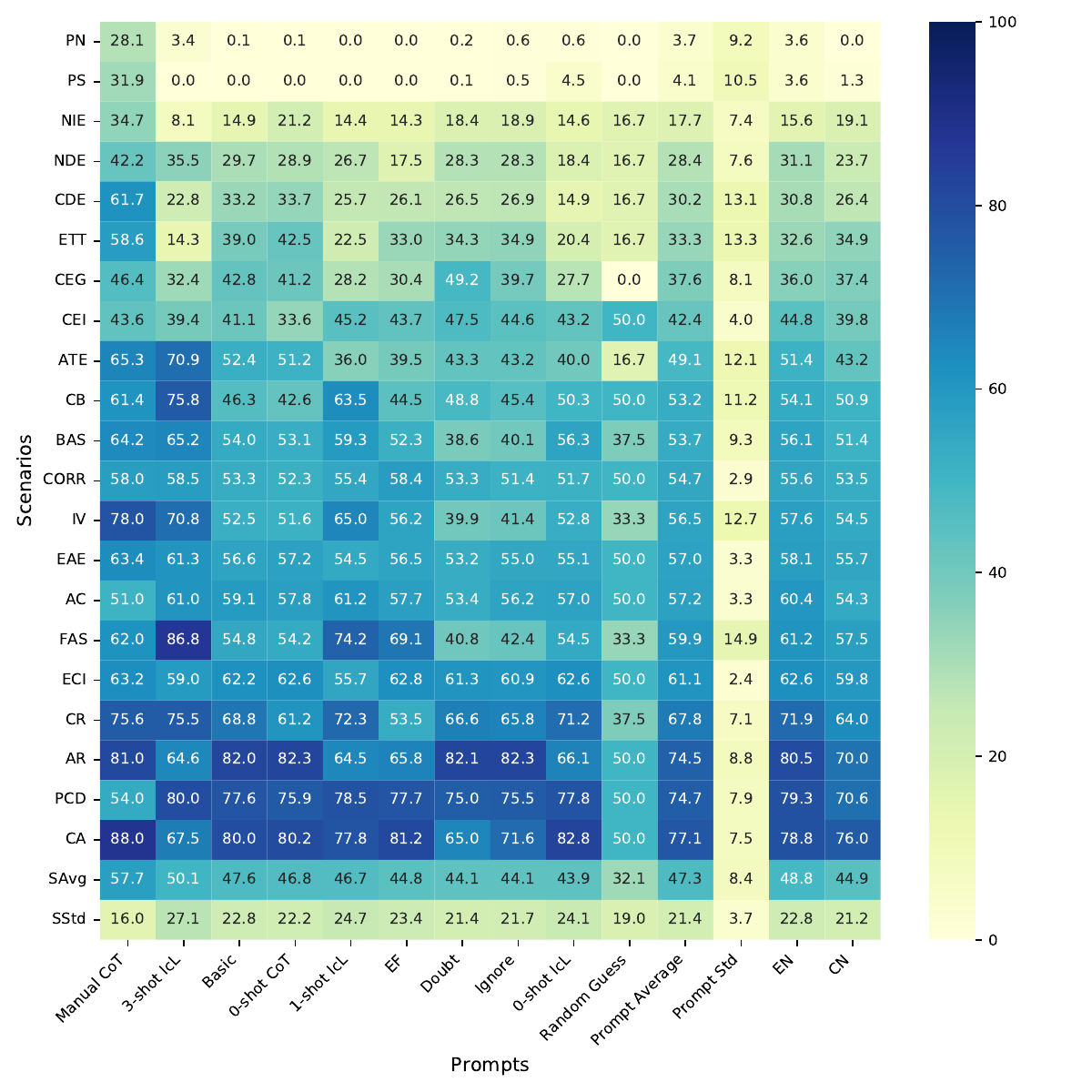}
\end{minipage}
}
\subfigure[\textit{Model-prompt rank} of text-davinci-003]{
\begin{minipage}{8.5cm}
\centering
\includegraphics[width=1\linewidth]{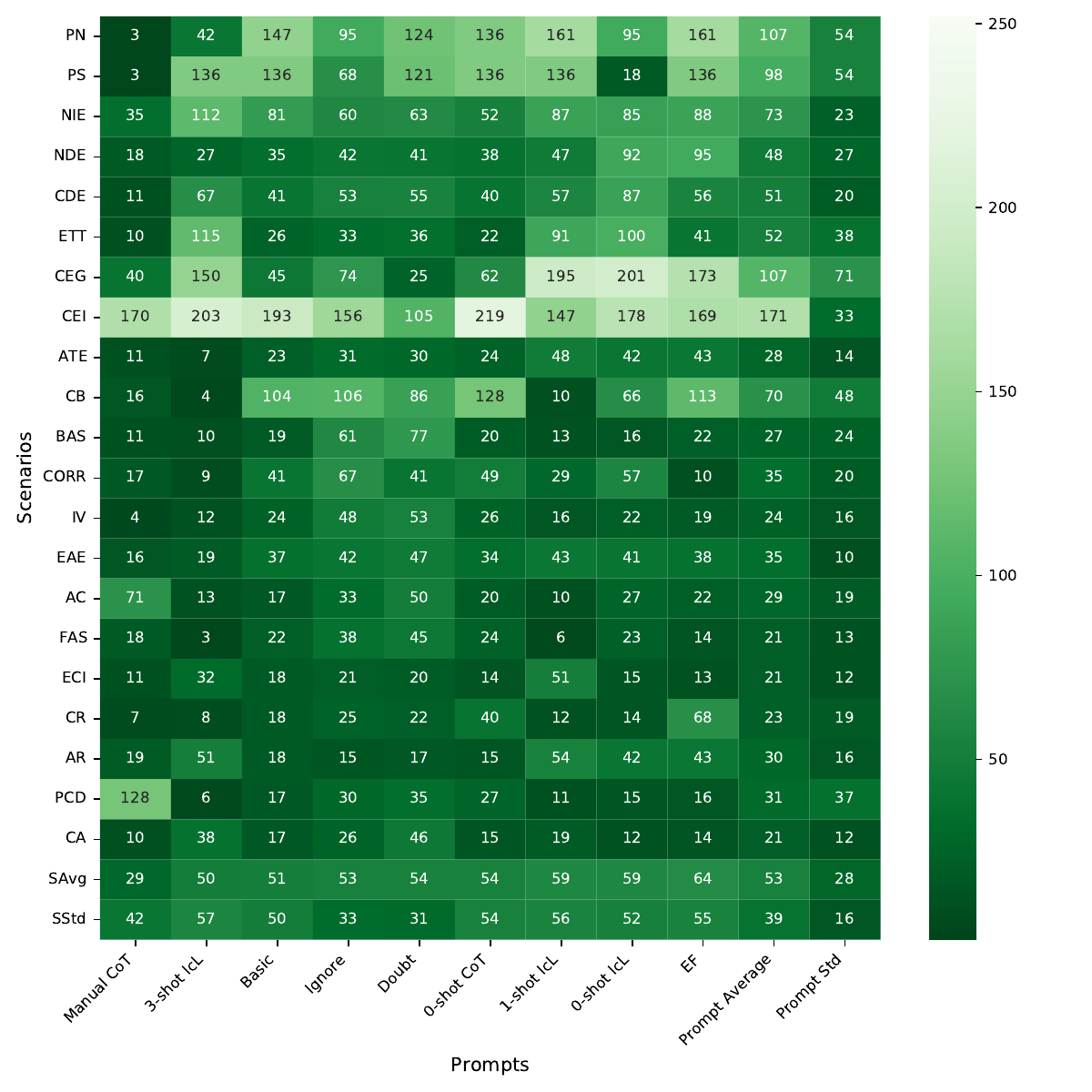}
\end{minipage}
}
\caption[Heatmap of text-davinci-003]{\textbf{Heatmap of text-davinci-003.}}
\label{fig:Heatmap_of_text-davinci-003}
\end{figure}
Summary: The model achieves a scenario-prompt performance accuracy average of 47.3\%, ranks on average 4th out of 28 in \textit{prompt-average rank}, and maintains an average robustness score of 68.2\%.

Accuracy: 1) Overall performance: Illustrated in Figure \ref{fig:Heatmap_of_text-davinci-003}(a), text-davinci-003 has achieved an impressive \textit{average scenario-prompt accuracy} of 47.3\%, with prompt effectiveness showing an average standard deviation of 8.4. The \textit{top scenario-prompt pair}s are manual CoT in CA with a score of 88.0\%, 3-shot IcL in FAS at 86.8\%, and 0-shot IcL in CA at 82.8\%. A significant 88.9\% of \textit{scenario-prompt pairs} outperform the \textit{random guess accuracy}, with 5.3\% achieving over 80\% accuracy.
2) Scenario performance: For scenarios where text-davinci-003 exceeds the \textit{random guess accuracy}, the highest scores include CA at 77.1\%, PCD at 74.7\%, and AR at 74.5\%.
3) Prompt efficiency: The most impactful prompts are manual CoT at 57.7\%, 3-shot IcL at 50.1\%, and basic at 47.6\%. In terms of outperforming the \textit{random guess accuracy}, manual CoT dominates in 20 out of 21 scenarios, closely followed by 0-shot CoT, 1-shot IcL, adversarial doubt, and adversarial ignore, each achieving success in 19 scenarios.
4) Language influence: English surpasses Chinese in 18 of 21 scenarios, with significant accuracy advantages in AR, PCD, and ATE, showing \textit{language accuracy difference}s of 10.5\%, 8.7\%, and 8.1\%, respectively. However, Chinese outshines English in scenarios like NIE, ETT, and CEG, with differences of 3.5\%, 2.3\%, and 1.4\%, respectively. 

Ranking: 1) \textit{Prompt-average rank}: As indicated in Figure \ref{fig:Prompt-Average_Rank_of_Models}, text-davinci-003's top performance is outstanding, achieving a 2nd place rank across 9 scenarios. Its least impressive ranks are 20th in CEI, 9th in CEG, and tied at 6th for both PN and NIE. The model's average \textit{prompt-average rank} is 4th out of 28, with a standard deviation of 4.1.
2) \textit{Model-prompt rank}: Demonstrated in Figure \ref{fig:Heatmap_of_text-davinci-003}(b), the top ranks for text-davinci-003 include PN with manual CoT at 3, PS with manual CoT also at 3, and FAS with 3-shot IcL at 3. The challenges are most significant in CEI with 0-shot CoT at 219, with 3-shot IcL at 203, and CEG with 0-shot IcL at 201.

Robustness: text-davinci-003 presents an average robustness of 68.2\% across scenarios, with standout performance in AR at 93.9\%, CB at 86.4\%, and PCD at 85.2\%.

\paragraph{GPT-3.5-Turbo.}
\begin{figure}[t]
\centering
\subfigure[Performance of GPT-3.5-Turbo]{
\begin{minipage}{8.5cm}
\centering
\includegraphics[width=1\linewidth]{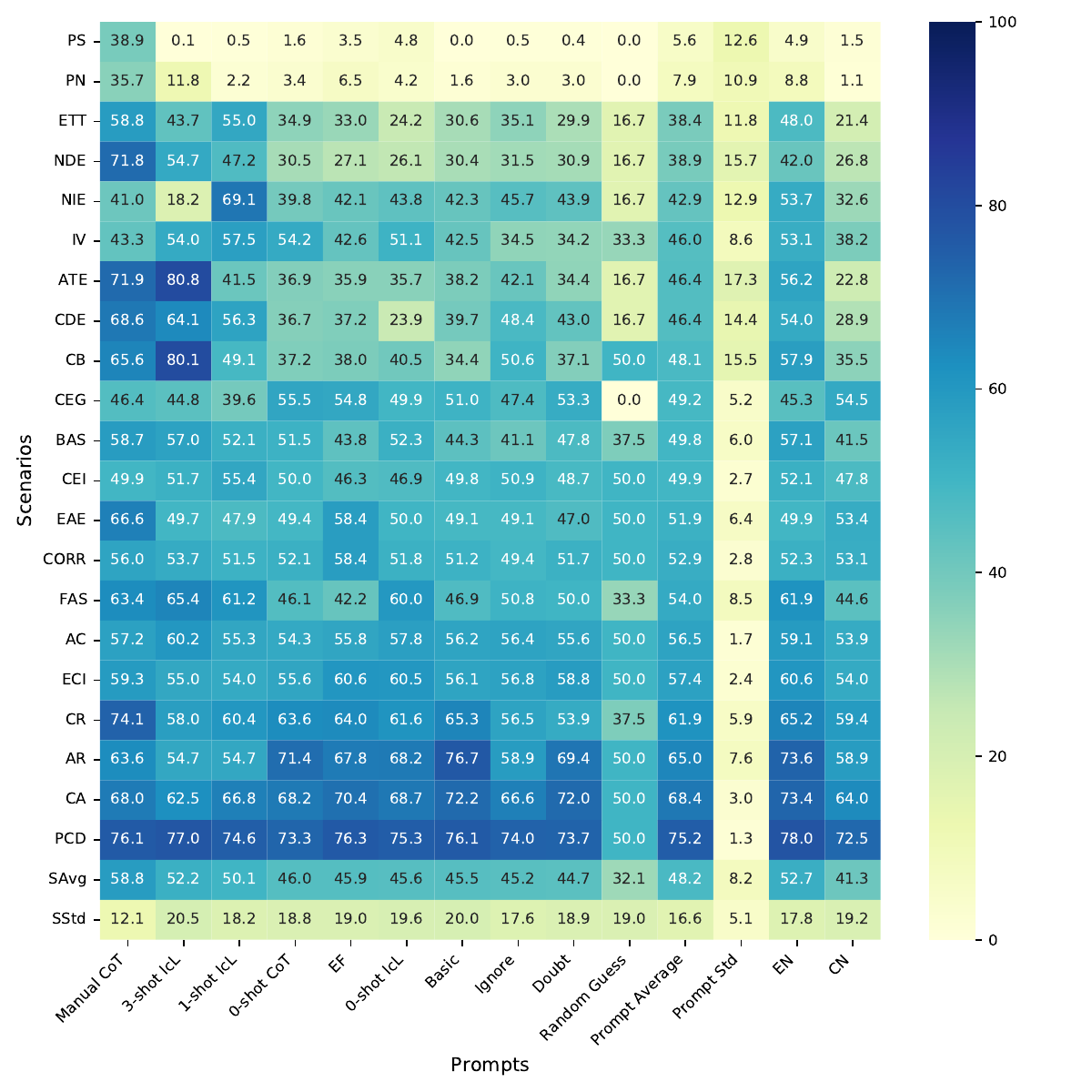}
\end{minipage}
}
\subfigure[\textit{Model-prompt rank} of GPT-3.5-Turbo]{
\begin{minipage}{8.5cm}
\centering
\includegraphics[width=1\linewidth]{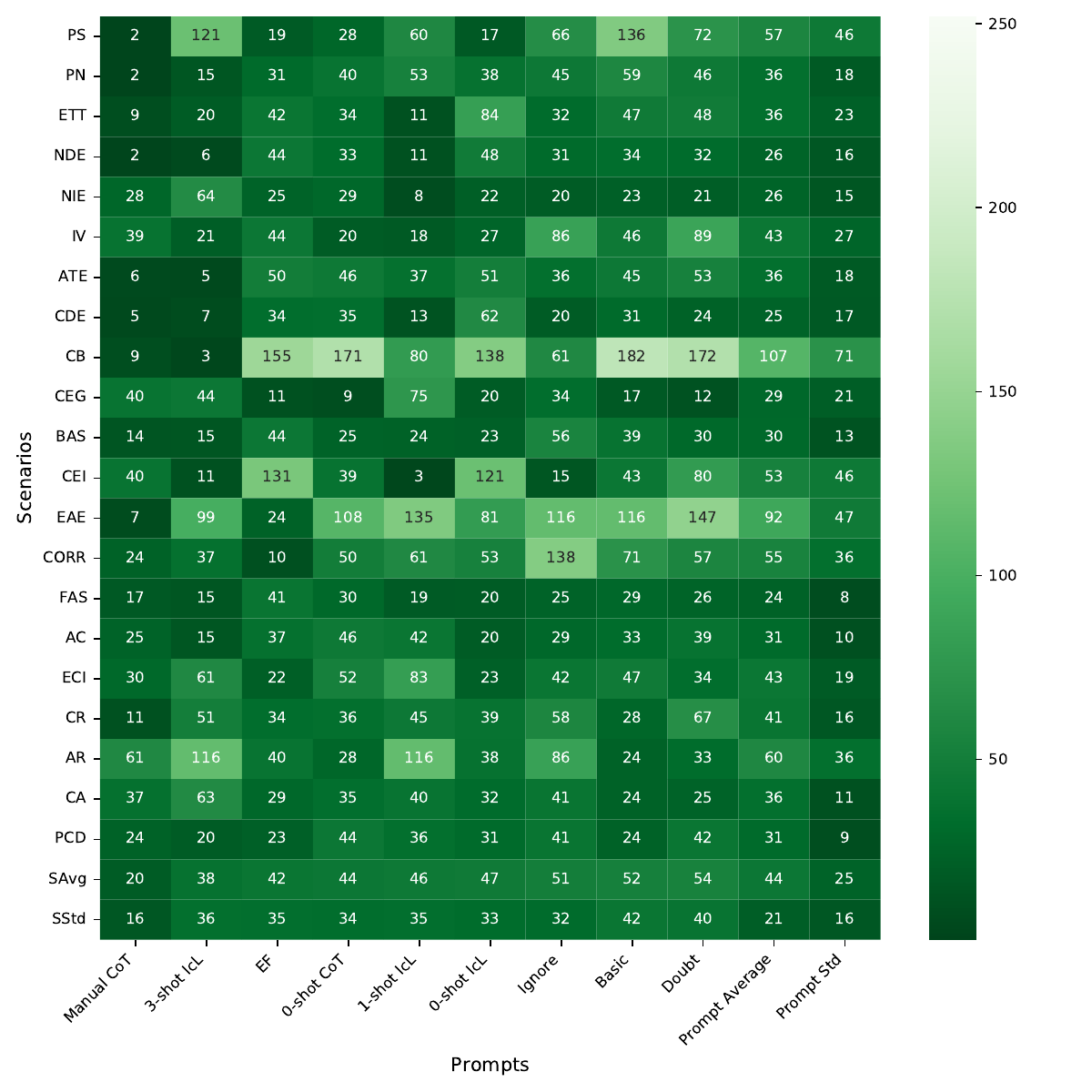}
\end{minipage}
}
\caption[Heatmap of GPT-3.5-Turbo]{\textbf{Heatmap of GPT-3.5-Turbo.}}
\label{fig:Heatmap_of_GPT-3.5-Turbo}
\end{figure}

Summary: The model demonstrates an \textit{average scenario-prompt accuracy} of 48.2\%, an average \textit{prompt-average rank} of 3 out of 28, and an average robustness score of 70.1\%.

Accuracy: 1) Overall performance: Displayed in Figure \ref{fig:Heatmap_of_GPT-3.5-Turbo}(a), GPT-3.5-Turbo showcases an \textit{average scenario-prompt accuracy} of 48.2\%, alongside a prompt effectiveness average variability (std) of 8.2. The \textit{top scenario-prompt pair}s are highlighted by a 3-shot IcL in ATE achieving a peak score of 80.8\%, closely followed by 3-shot IcL in CB at 80.1\%, and 3-shot IcL in PCD at 77.0\%. It is worth noticing that 89.9\% of the \textit{scenario-prompt pairs} manage to surpass the \textit{random guess accuracy}, with a small fraction, 1.1\%, exceeding an 80\% accuracy threshold.
2) Scenario performance: In scenarios where GPT-3.5-Turbo surpasses the \textit{random guess accuracy}, the top 3 scenarios having the highest average accuracy are PCD with an outstanding average score of 75.2\%, CA at 68.4\%, and AR at 65.0\%.
3) Prompt efficiency: The most effective prompts are manual CoT at 58.8\%, 3-shot IcL at 52.2\%, and 1-shot IcL at 50.1\%. In the context of \textit{scenario-prompt pair}s where the model surpasses the \textit{random guess accuracy}, manual CoT and 3-shot IcL are ahead in 20 out of 21 scenarios. They are followed by 1-shot IcL, EF, 0-shot IcL, and adversarial ignore, each outperforming in 19 scenarios.
4) Language influence: English outperform Chinese in 18 out of 21 analyzed scenarios, with higher accuracy leads in ATE, ETT, and CDE, revealing \textit{language accuracy difference}s of 33.4\%, 26.6\%, and 25.1\%, respectively. Conversely, scenarios such as CEG, EAE and CORR showcase superior performance in Chinese, with \textit{language accuracy difference}s of 9.2\%, 3.5\%, and 0.8\%, respectively. 

Ranking: 1) \textit{Prompt-average rank}: As shown in Figure \ref{fig:Prompt-Average_Rank_of_Models}, GPT-3.5-Turbo clinches its highest \textit{prompt-average rank}s in NIE, CEI, and CDE, all  at rank 1. Yet, it ranks lower in CB at 9, AR at 6, and EAE at 6, indicating areas for improvement. The average \textit{prompt-average rank} across 21 scenarios stands at 3 out of 28, with a standard deviation reflecting performance consistency at 2.0.
2) \textit{Model-prompt rank}: Detailed in Figure \ref{fig:Heatmap_of_GPT-3.5-Turbo}(b), GPT-3.5-Turbo's top \textit{model-prompt rank}s include PS with manual CoT at rank 2, NDE with manual CoT also at 2, and PN with manual CoT at 2. The most significant challenges are in CB, with the lowest ranks observed with basic at 182, adversarial doubt at 172, and 0-shot CoT at 171.

Robustness: GPT-3.5-Turbo records an average robustness score of 70.1\% across scenarios, showing the highest robust in EAE at 87.0\%, ATE at 83.5\%, and PCD at 83.2\%.

\paragraph{GPT-4.}
\begin{figure}[t]
\centering
\subfigure[Performance of GPT-4]{
\begin{minipage}{8.5cm}
\centering
\includegraphics[width=1\linewidth]{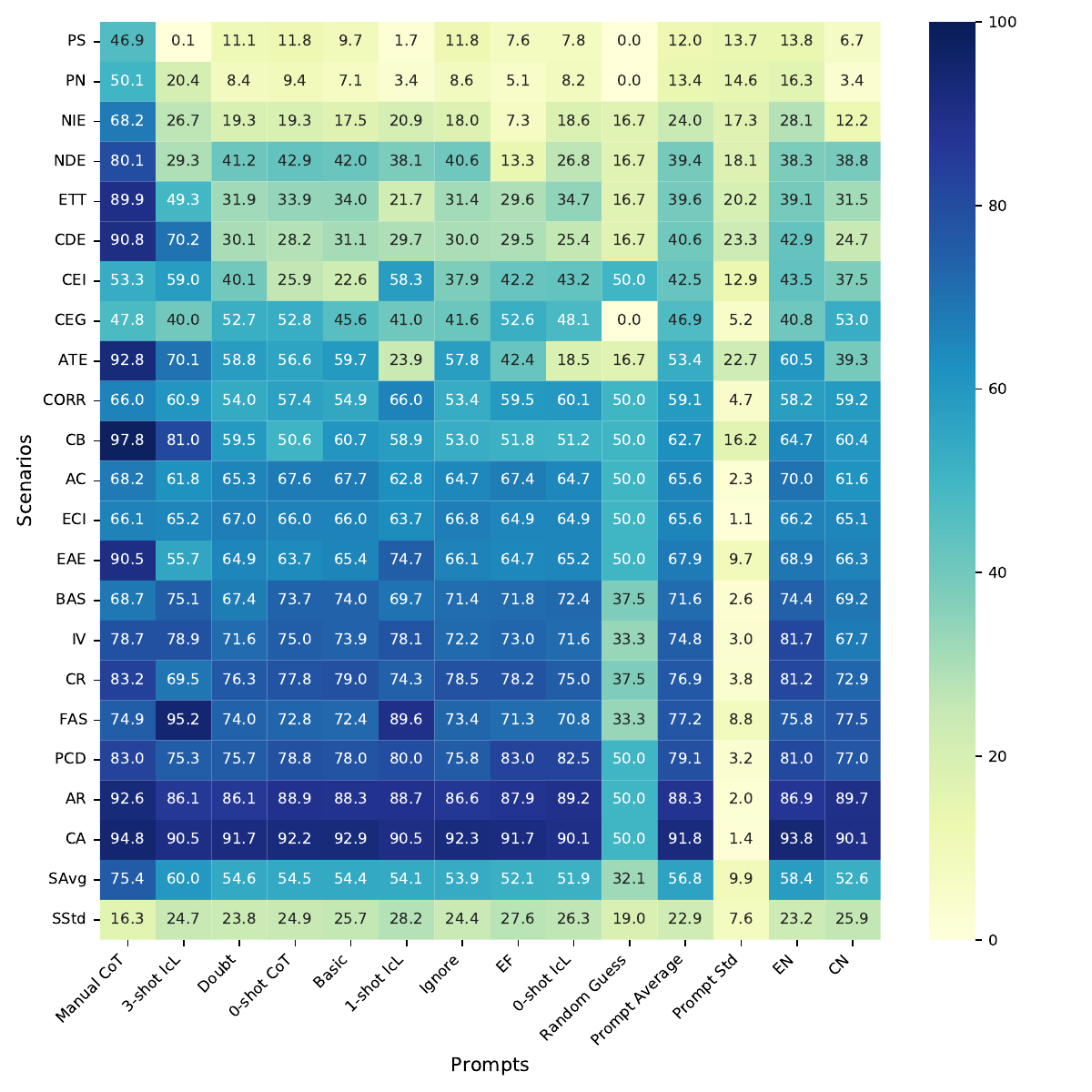}
\end{minipage}
}
\subfigure[\textit{Model-prompt rank} of GPT-4]{
\begin{minipage}{8.5cm}
\centering
\includegraphics[width=1\linewidth]{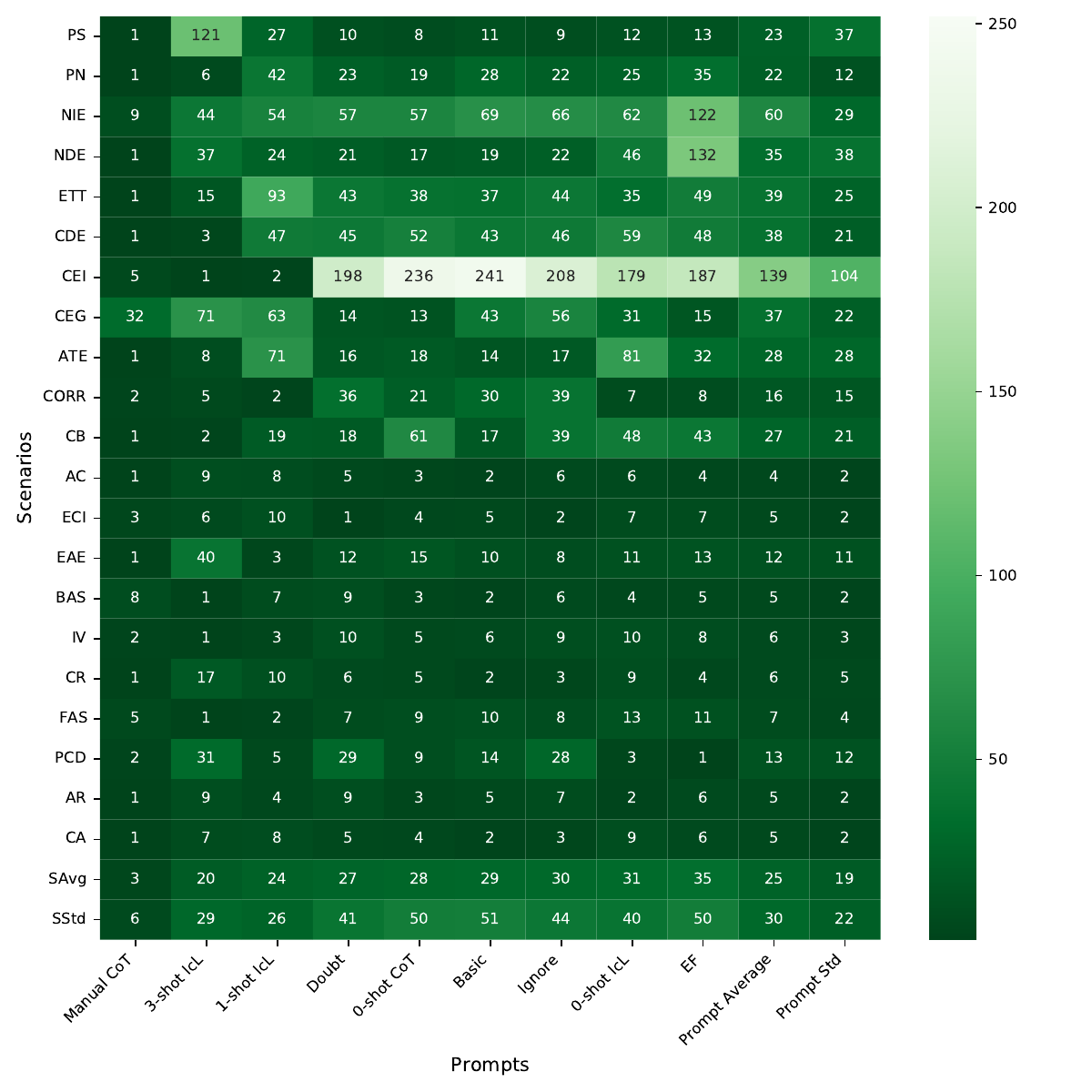}
\end{minipage}
}
\caption[Heatmap of GPT-4]{\textbf{Heatmap of GPT-4.}}
\label{fig:Heatmap_of_GPT-4}
\end{figure}

Summary: The model boasts an \textit{average scenario-prompt accuracy} of 56.8\%, achieves an average \textit{prompt-average rank} of 2 out of 28 (the highest average \textit{prompt-average rank}), and holds an average robustness score of 83.7\%.

Accuracy: 1) Overall performance: According to Figure \ref{fig:Heatmap_of_GPT-4}(a), GPT-4 achieves an \textit{average scenario-prompt accuracy} of 56.8\%, with an average standard deviation for prompt effectiveness at 9.9. The \textit{top scenario-prompt pair}s are manual CoT in CB with a score of 97.8\%, 3-shot IcL in FAS at 95.2\%, and manual CoT in CA at 94.8\%. To be noticed 95.8\% of the \textit{scenario-prompt pairs} surpass the \textit{random guess accuracy}, with 16.9\% exceeding an 80\% accuracy threshold.
2) Scenario performance: In scenarios where GPT-4 beats the \textit{random guess accuracy}, the top three scenarios in terms of average accuracy are CA at 91.8\%, AR at 88.3\%, and PCD at 79.1\%.
3) Prompt efficiency: The highest effectiveness is observed with manual CoT at 75.4\%, 3-shot IcL at 60.0\%, and adversarial doubt at 54.6\%. Moreover, manual CoT, 3-shot IcL, and 1-shot IcL exceed the \textit{random guess accuracy} in all 21 scenarios.
4) Language influence: English performs better than Chinese in 16 out of 21 scenarios, showing significant advantages in ATE, CDE, and NIE, with \textit{language accuracy difference}s of 21.1\%, 18.2\%, and 15.9\%, respectively. On the other hand, in scenarios like CEG, AR, and FAS, Chinese shows superior performance with \textit{language accuracy difference}s of 12.2\%, 2.8\%, and 1.7\%, respectively.

Ranking: 1) \textit{Prompt-average rank}: As indicated in Figure \ref{fig:Prompt-Average_Rank_of_Models}, GPT-4 achieved a first-place \textit{prompt-average rank} in 15 scenarios. Its lowest rank was 24th in CEI, with an additional four scenarios ranked at 4th place. Across 21 scenarios, the model's average \textit{prompt-average rank} is 2 out of 28, with a standard deviation of 5.0.
2) \textit{Model-prompt rank}: As shown in Figure \ref{fig:Heatmap_of_GPT-4}(b), GPT-4 has top 1 \textit{model-prompt rank} in almost all scenarios except in NIE, CEI and CORR. The lowest \textit{model-prompt rank}s are in CEI with basic at 241, with 0-shot CoT at 236, and with adversarial ignore at 208.

Robustness: GPT-4 boasts an average robustness score of 83.7\% across scenarios, demonstrating the highest robustness in AR at 97.0\%, CA at 95.2\%, and PCD at 92.0\%.

\subsubsection{Anthropic}
\label{model:anthropic}
\paragraph{Claude2.}
\begin{figure}[t]
\centering
\subfigure[Performance of Claude2]{
\begin{minipage}{8.5cm}
\centering
\includegraphics[width=1\linewidth]{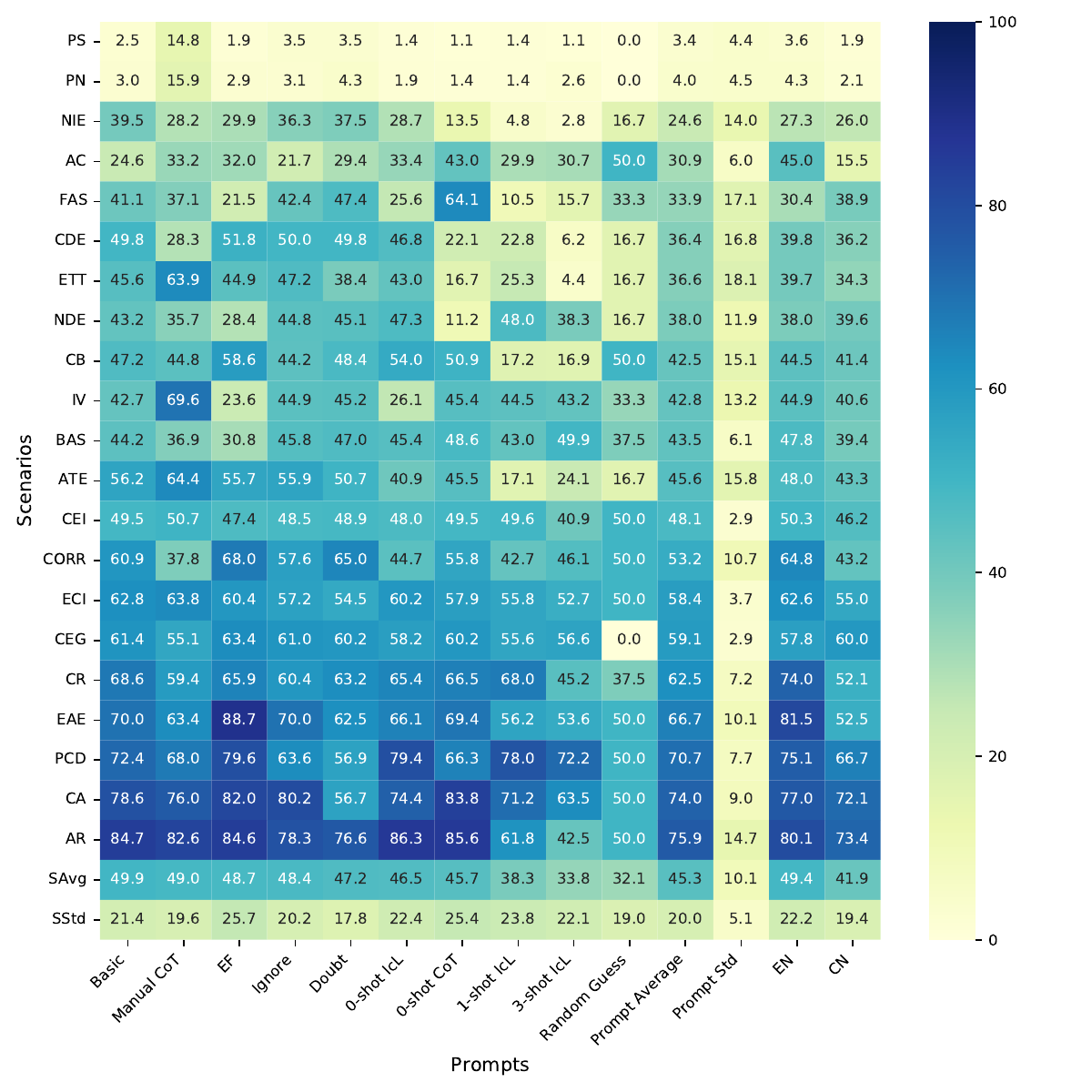}
\end{minipage}
}
\subfigure[\textit{Model-prompt rank} of Claude2]{
\begin{minipage}{8.5cm}
\centering
\includegraphics[width=1\linewidth]{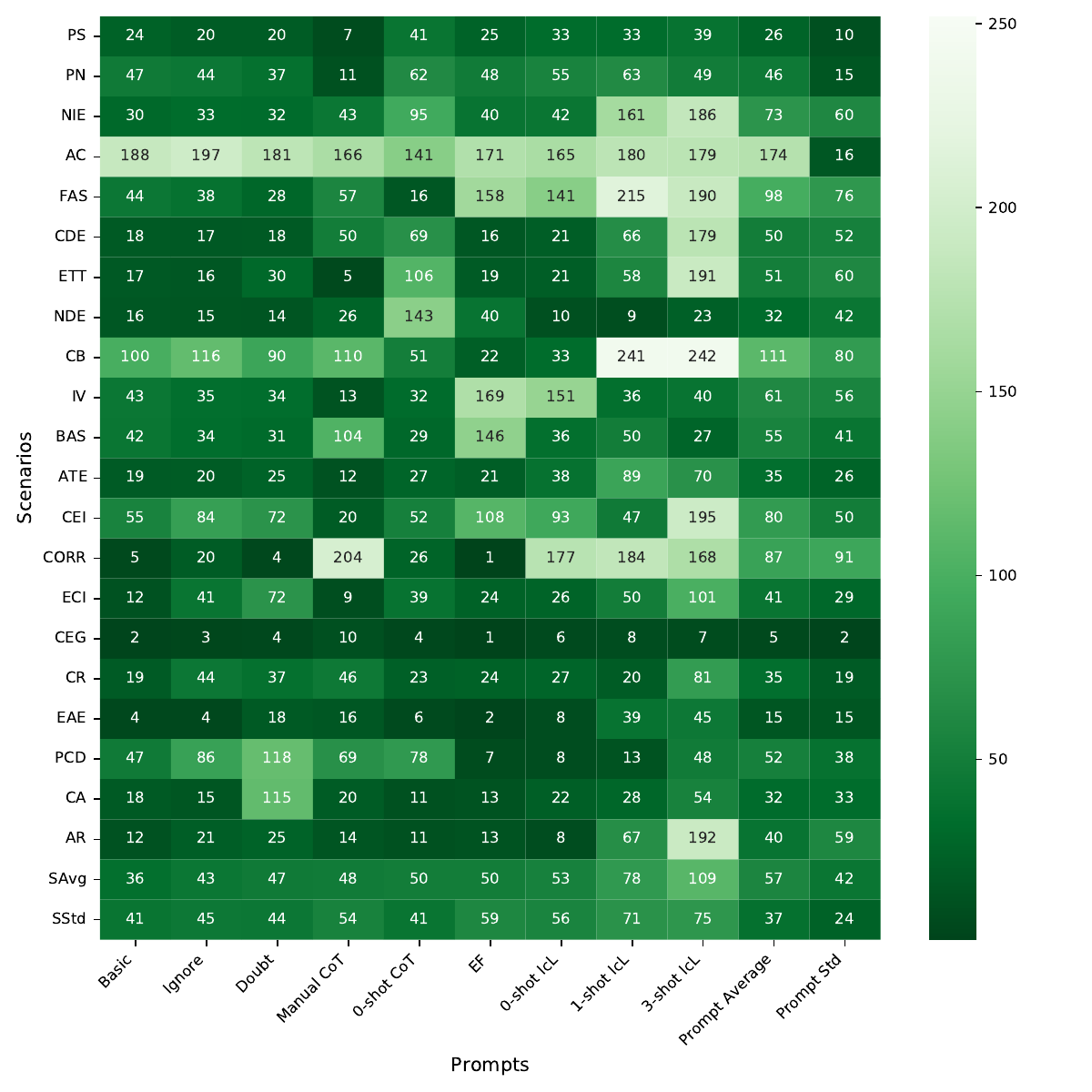}
\end{minipage}
}
\caption[Heatmap of Claude2]{\textbf{Heatmap of Claude2.}}
\label{fig:Heatmap_of_Claude2}
\end{figure}
Summary: The model exhibits an \textit{average scenario-prompt accuracy} of 45.3\%, secures an average \textit{prompt-average rank} of 4 out of 28, and attains an average robustness score of 67.5\% across scenarios.

Accuracy: 1) Overall performance: As illustrated in Figure \ref{fig:Heatmap_of_Claude2}(a), Claude2 achieves an \textit{average scenario-prompt accuracy} of 45.3\%, with average variability (standard deviation) in prompt effectiveness at 10.1. The \textit{top scenario-prompt pair}s are EF in EAE, with a score of 88.7\%, followed by 0-shot IcL in AR at 86.3\%, and 0-shot CoT in the same category at 85.6\%. A significant 77.8\% of \textit{scenario-prompt pairs} outperform the baseline \textit{random guess accuracy}, with 4.8\% scoring above 80\% accuracy.
2) Scenario performance: In scenarios where Claude2 outperforms the \textit{random guess accuracy}, the three scenarios with the highest average accuracies are AR, leading with a score of 75.9\%, followed by CA at 74.0\%, and PCD at 70.7\%.
3) Prompt efficiency: The prompts yielding the highest efficiency are basic at 49.9\%, manual CoT at 49.0\%, and EF at 48.7\%. As to the number of \textit{scenario-prompt pair}s where the model exceeds \textit{random guess accuracy}, the basic leads in 18 out of 21 scenarios, followed by adversarial ignore in 18, and adversarial doubt in 18 scenarios.
4) Language influence: English demonstrates superior performance in 18 out of 21 scenarios, especially in AC, EAE, and CR, with \textit{language accuracy difference}s of 29.5\%, 29.0\%, and 21.9\%, respectively. In contrast, scenarios like FAS, CEG, and NDE perform better in Chinese, with accuracy differences of 8.5\%, 2.2\%, and 1.5\%, respectively. 

Ranking: 1) \textit{Prompt-average rank}: According to Figure \ref{fig:Prompt-Average_Rank_of_Models}, Claude2 achieves its best \textit{prompt-average rank}s in NDE, ETT, and CEG, all at rank 1. On the flip side, it ranks lowest in AC (19), CB (12), and FAS (7), identifying areas needing enhancement. The average \textit{prompt-average rank} across 21 scenarios is 4th out of 28, with a variability of 4.2.
2) \textit{Model-prompt rank}: Shown in Figure \ref{fig:Heatmap_of_Claude2}(b), Claude2's highest \textit{model-prompt rank}s are in CORR with EF (1), CEG with EF (1), EAE with EF (2), and CEG with basic (2). The lowest ranks are in CB with 3-shot IcL (242), CB with 1-shot IcL (241), and FAS with 1-shot IcL (215), pinpointing particular challenges.

Robustness: Claude2 showcases an average robustness score of 67.5\% across different scenarios, achieving its highest robustness in CB (86.8\%), AR (82.0\%), and CEI (81.0\%).

\subsubsection{Shanghai AI Laboratory}
\label{model:ailab}
\paragraph{InternLM-chat (7B).}
\begin{figure}[t]
\centering
\subfigure[Performance of InternLM-chat (7B)]{
\begin{minipage}{8.5cm}
\centering
\includegraphics[width=1\linewidth]{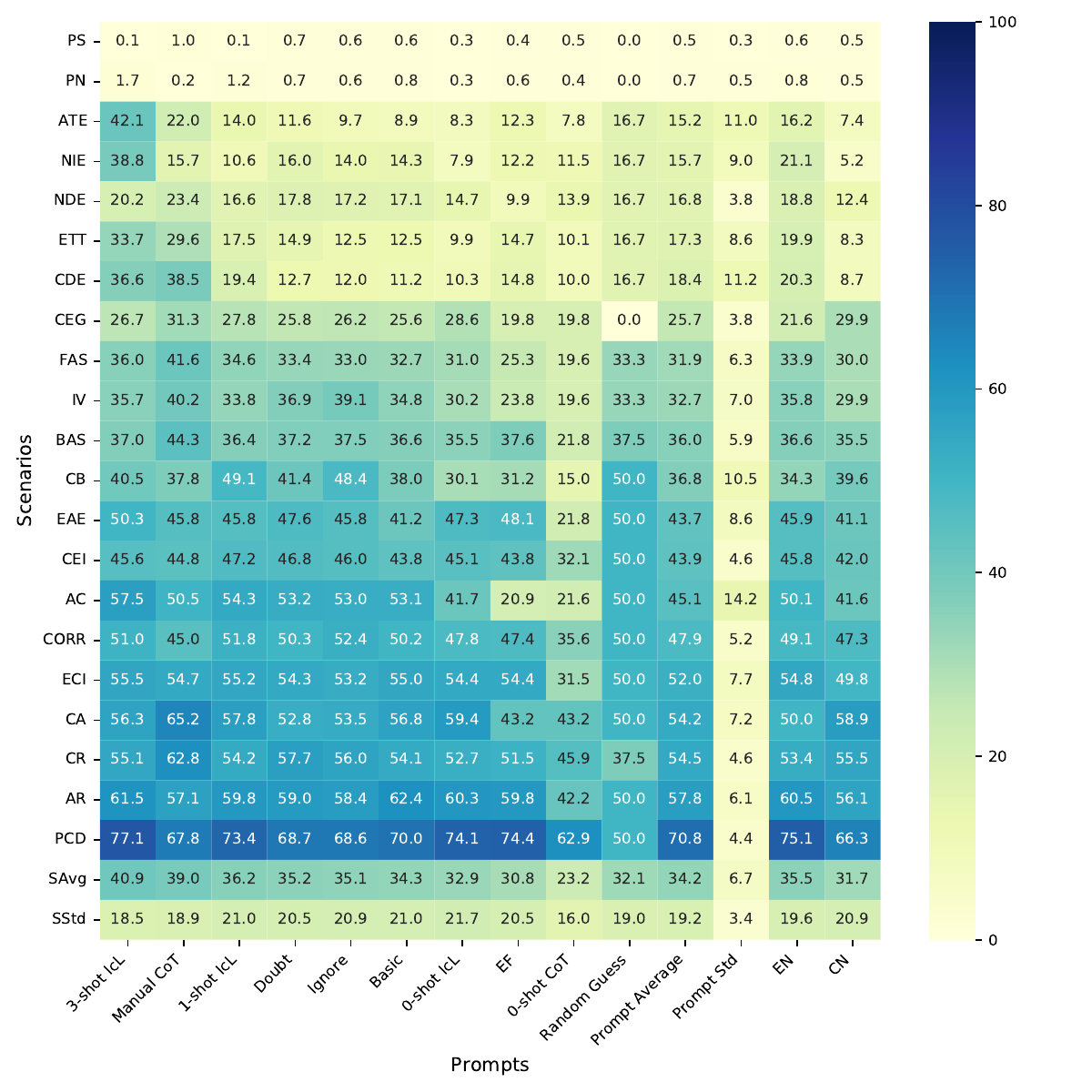}
\end{minipage}
}
\subfigure[\textit{Model-prompt rank} of InternLM-chat (7B)]{
\begin{minipage}{8.5cm}
\centering
\includegraphics[width=1\linewidth]{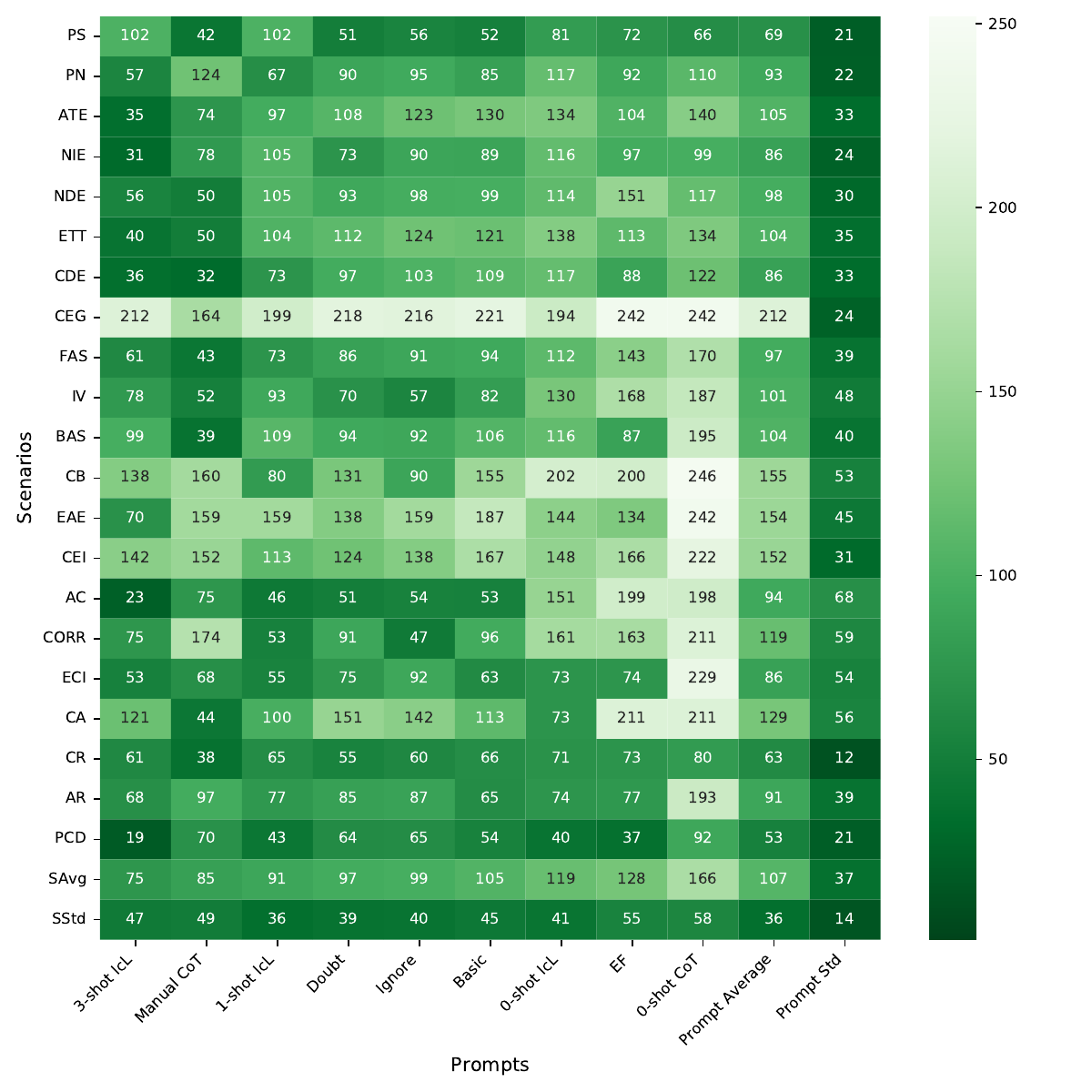}
\end{minipage}
}
\caption[Heatmap of InternLM-chat (7B)]{\textbf{Heatmap of InternLM-chat (7B).}}
\label{fig:Heatmap_of_InternLM-chat_(7B)}
\end{figure}
Summary: The model's \textit{average scenario-prompt accuracy} is 34.2\%. It has an average \textit{prompt-average rank} of 12/28 and an average robustness score of 74.7\% across various scenarios.

Accuracy: 1) Overall performance: Illustrated in Figure \ref{fig:Heatmap_of_InternLM-chat_(7B)}(a), InternLM-chat (7B) achieves an \textit{average scenario-prompt accuracy} of 34.2\%, with an average standard deviation in prompt effectiveness of 6.7. The \textit{top scenario-prompt pair}s are 3-shot IcL in PCD with a score of 77.1\%, followed closely by EF in the same scenario at 74.4\%, and 0-shot IcL in PCD at 74.1\%. A total of 56.1\% of the \textit{scenario-prompt pairs} outperform the \textit{random guess accuracy}, although none surpass the 80\% accuracy mark.
2) Scenario performance: When surpassing \textit{random guess accuracy}, the top scenarios by average accuracy are PCD at 70.8\%, AR at 57.8\%, and CR at 54.5\%.
3) Prompt efficiency: The most efficient prompts include 3-shot IcL at 40.9\%, manual CoT at 39.0\%, and 1-shot IcL at 36.2\%. Regarding the number of \textit{scenario-prompt pairs} exceeding the \textit{random guess accuracy} across scenarios, 3-shot IcL is the frontrunner in 18 out of 21 scenarios, with manual CoT and 1-shot IcL following in 16 and 14 scenarios, respectively.
4) Language influence: English surpasses Chinese in 17 of 21 scenarios, especially in NIE, ETT, and CDE, with \textit{language accuracy difference}s of 15.9\%, 11.6\%, and 11.6\%, respectively. Conversely, Chinese outperforms in scenarios such as CA, CEG, and CB, with accuracy leads of 8.9\%, 8.3\%, and 5.3\%, respectively.

Ranking: 1) \textit{Prompt-average rank}: As highlighted in Figure \ref{fig:Prompt-Average_Rank_of_Models}, the highest \textit{prompt-average rank}s for InternLM-chat (7B) are observed in PCD at 5, AR, CR, and NDE all at 7. On the flip side, its lowest ranks are in CEG at 27, CB at 19, and CEI,  EAE, and PN, all at 17, pointing out areas for enhancement. Across 21 scenarios, the model averages a \textit{prompt-average rank} of 12 out of 28, with a standard deviation of 5.1.
2) \textit{Model-prompt rank}: Shown in Figure \ref{fig:Heatmap_of_InternLM-chat_(7B)}(b), the best \textit{model-prompt rank}s for InternLM-chat (7B) include PCD with 3-shot IcL at 19, AC with 3-shot IcL at 23, and NIE with 3-shot IcL at 31. The lowest ranks are in CB with 0-shot CoT at 246 and CEG with both 0-shot CoT and EF at 242.

Robustness: InternLM-chat (7B) showcases an average robustness score of 74.7\% across scenarios, with the highest robustness scores in CDE at 90.6\%, ATE at 90.1\%, and ETT at 86.8\%.

\paragraph{InternLM-chat (20B).}
\begin{figure}[t]
\centering
\subfigure[Performance of InternLM-chat (20B)]{
\begin{minipage}{8.5cm}
\centering
\includegraphics[width=1\linewidth]{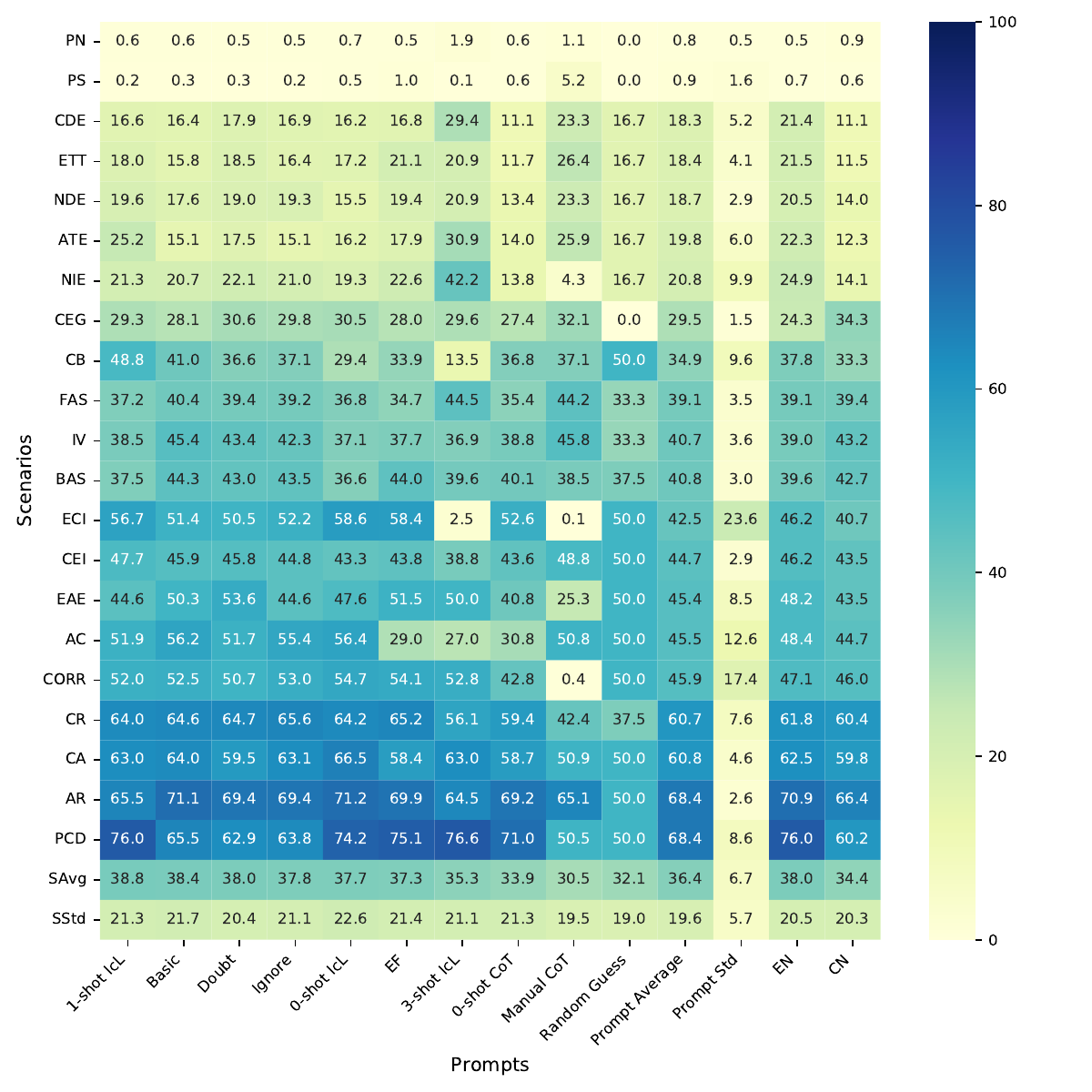}
\end{minipage}
}
\subfigure[\textit{Model-prompt rank} of InternLM-chat (20B)]{
\begin{minipage}{8.5cm}
\centering
\includegraphics[width=1\linewidth]{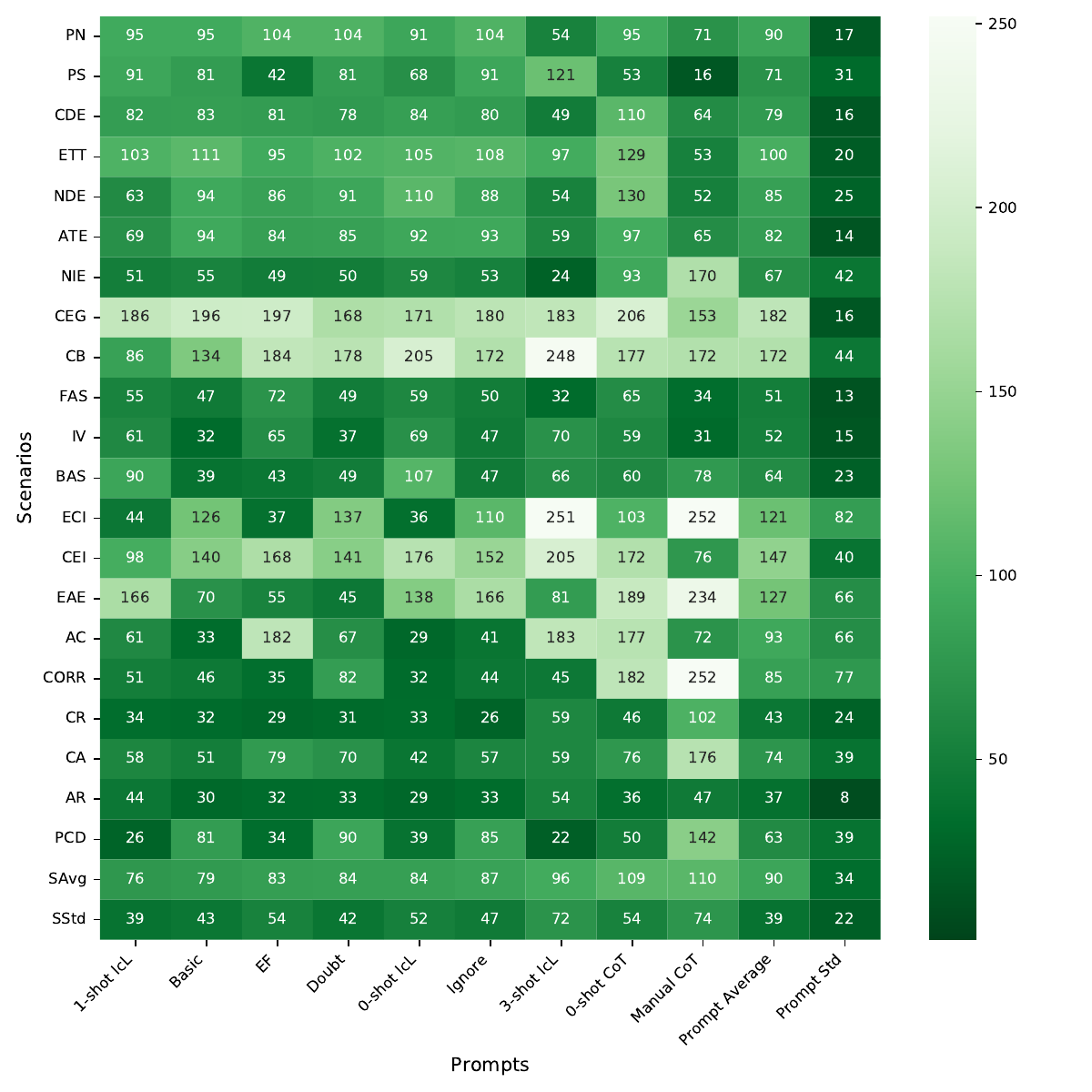}
\end{minipage}
}
\caption[Heatmap of InternLM-chat (20B)]{\textbf{Heatmap of InternLM-chat (20B).}}
\label{fig:Heatmap_of_InternLM-chat_(20B)}
\end{figure}
Summary: The model achieves an \textit{average scenario-prompt accuracy} of 36.4\%, holds an average \textit{prompt-average rank} of 10 out of 28, and maintains an average robustness score of 67.3\% across diverse scenarios.

Accuracy: 1) Overall performance: Figure \ref{fig:Heatmap_of_InternLM-chat_(20B)}(a) shows InternLM-chat (20B) achieving an \textit{average scenario-prompt accuracy} of 36.4\%, with an average prompt effectiveness variability of 6.7. The \textit{top scenario-prompt pair}s are 3-shot IcL in PCD with a score of 76.6\%, followed by 1-shot IcL in the same scenario at 76.0\%, and EF in PCD at 75.1\%. A total of 75.1\% of the \textit{scenario-prompt pairs} surpass the \textit{random guess accuracy}, though none exceed an 80\% accuracy rate.
2) Scenario performance: For scenarios outperforming \textit{random guess accuracy}, the highest average accuracies are found in PCD and AR, both at 68.4\%, and CA at 60.8\%.
3) Prompt efficiency: The top-performing prompts include 1-shot IcL at 38.8\%, basic at 38.4\%, and adversarial doubt at 38.0\%. Regarding prompts where the accuracy of its \textit{scenario-prompt pairs} exceeds \textit{random guess accuracy}, adversarial doubt leads in 19 out of 21 scenarios, followed by EF in 18 and 3-shot IcL in 17.
4) Language influence: English surpasses Chinese in accuracy in 16 of the 21 scenarios, particularly in PCD, NIE, and CDE, with \textit{language accuracy difference}s of 15.9\%, 10.8\%, and 10.2\%, respectively. However, Chinese excels in CEG, IV, and BAS, with 10.0\%, 4.2\%, and 3.1\% advantages, respectively.

Ranking: 1) \textit{Prompt-average rank}: According to Figure \ref{fig:Prompt-Average_Rank_of_Models}, InternLM-chat (20B) achieves its best \textit{prompt-average rank}s in AR and NIE at 4, FAS and CR both at 5. Conversely, its lowest \textit{prompt-average rank}s are in CEG at 24, ECI at 23, and CB at 22, highlighting improvement opportunities. The model holds an average \textit{prompt-average rank} of 10 out of 28, with a standard deviation of 6.5.
2) \textit{Model-prompt rank}: Figure \ref{fig:Heatmap_of_InternLM-chat_(20B)}(b) reveals the model's top \textit{model-prompt rank}s, including PS with manual CoT at 16, PCD with 3-shot IcL at 22, and NIE with 3-shot IcL at 24. The lowest rankings are seen in CORR and ECI with manual CoT, both at 252 and ECI with 3-shot IcL at 251.

Robustness: InternLM-chat (20B) has an average robustness rating of 67.3\% across various scenarios, with top performance in PCD at 87.3\%, CB at 86.3\%, and AR at 81.8\%.

\subsubsection{Alibaba Cloud}
\label{model:ali}
\paragraph{Qwen (7B).}
\begin{figure}[t]
\centering
\subfigure[Performance of Qwen (7B)]{
\begin{minipage}{8.5cm}
\centering
\includegraphics[width=1\linewidth]{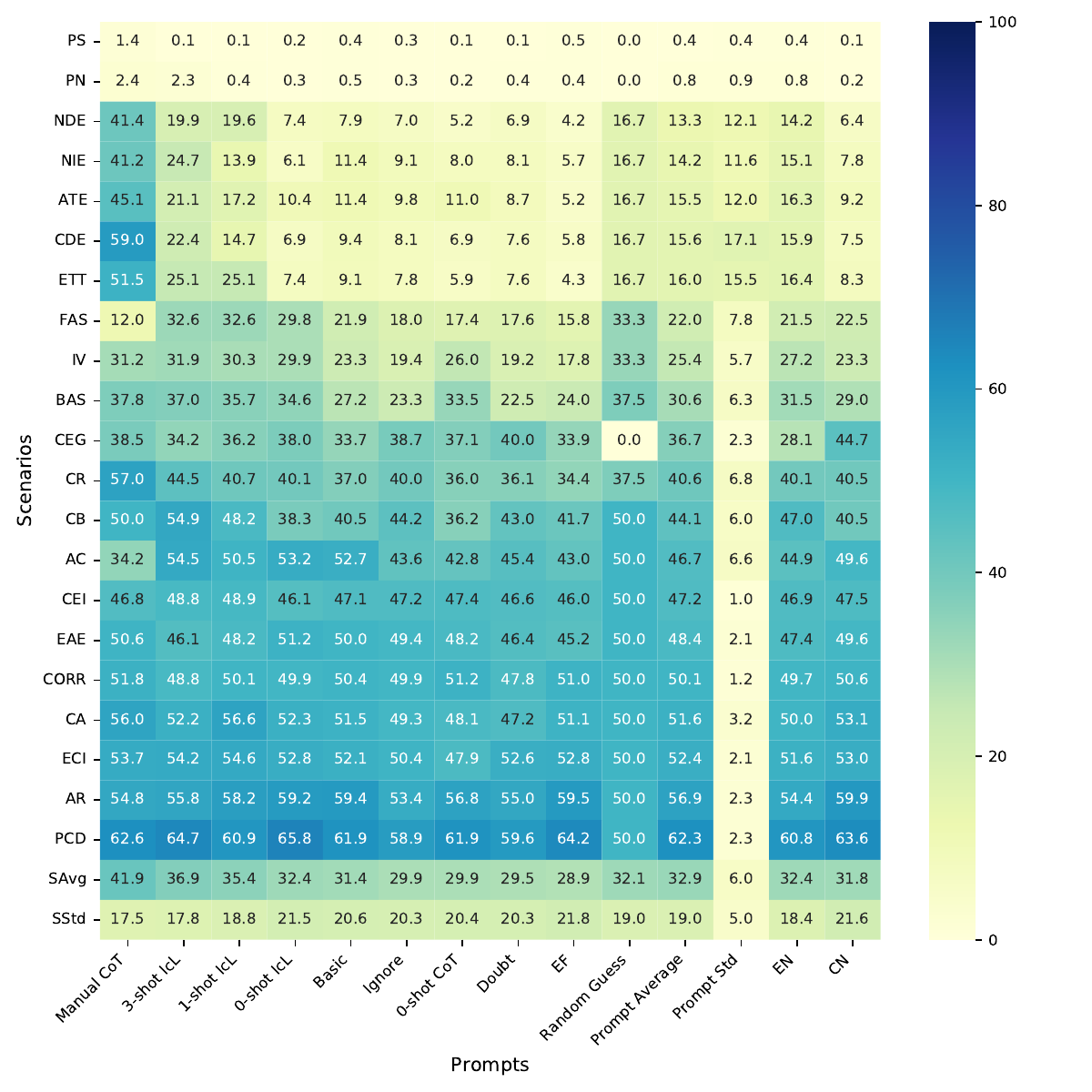}
\end{minipage}
}
\subfigure[\textit{Model-prompt rank} of Qwen (7B)]{
\begin{minipage}{8.5cm}
\centering
\includegraphics[width=1\linewidth]{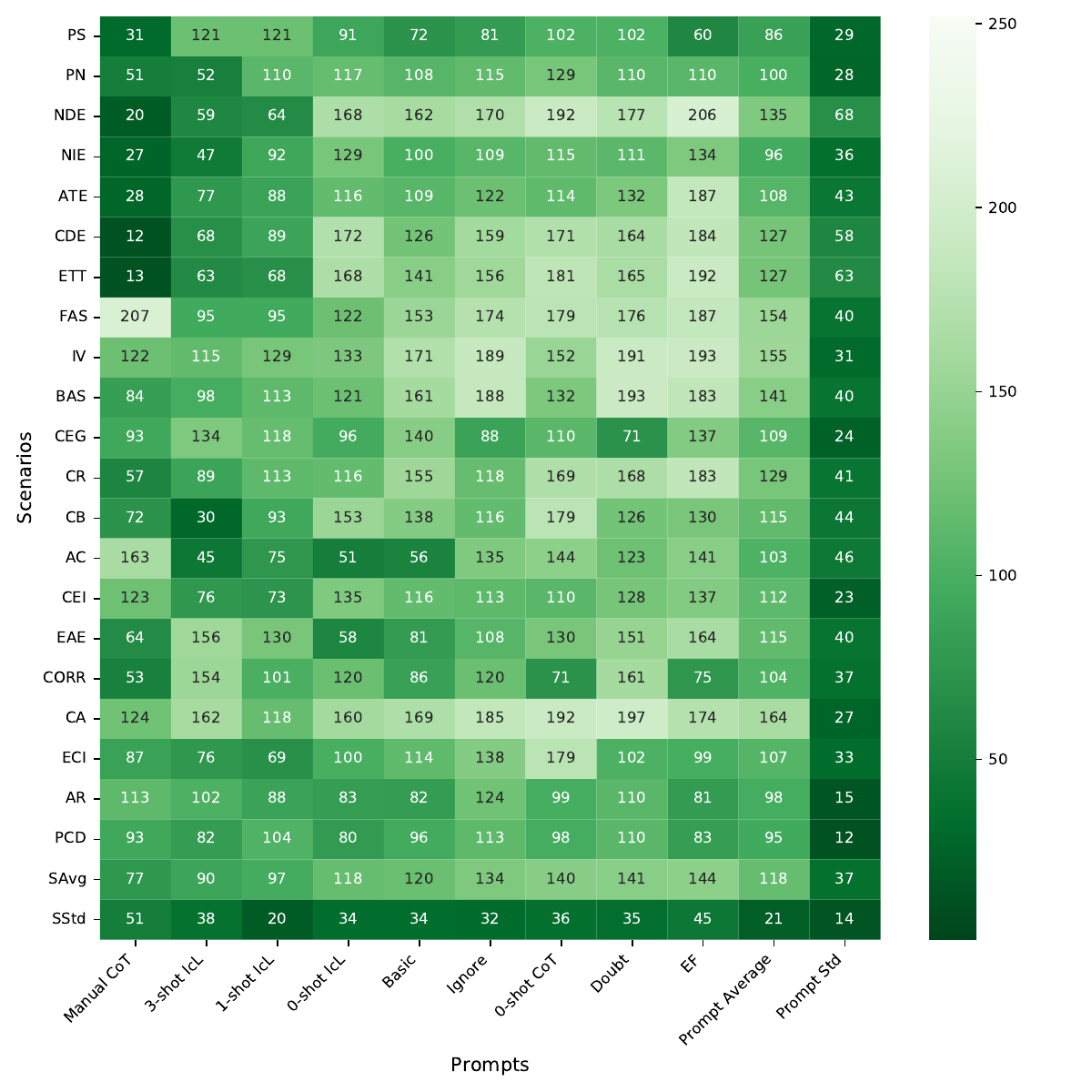}
\end{minipage}
}
\caption[Heatmap of Qwen (7B)]{\textbf{Heatmap of Qwen (7B).}}
\label{fig:Heatmap_of_Qwen_(7B)}
\end{figure}
Summary: The model records an \textit{average scenario-prompt accuracy} of 32.9\%, achieves an average \textit{prompt-average rank} of 12 out of 28, and displays an average robustness score of 56.6\% across multiple scenarios.

Accuracy: 1) Overall performance: According to Figure \ref{fig:Heatmap_of_Qwen_(7B)}(a), Qwen (7B) achieves an \textit{average scenario-prompt accuracy} of 32.9\% 
, with an average variation (standard deviation) of 6.0 in prompt effectiveness. The \textit{top scenario-prompt pair}s feature a 0-shot IcL in PCD with a score of 65.8\%, closely followed by a 3-shot IcL scoring 64.7\%, and EF with 64.2\% in the same scenario. Nearly half (48.7\%) of the \textit{scenario-prompt pairs} outperform the baseline \textit{random guess accuracy}, yet none surpass the 80\% accuracy threshold.
2) Scenario performance: In scenarios where Qwen (7B) surpasses \textit{random guess accuracy}, the top 3 scenarios having the highest average accuracy are PCD with a score of 62.3\%, AR at 56.9\%, and ECI at 52.4\%.
3) Prompt efficiency: The prompts showing the highest efficacy include manual CoT at 41.9\%, 3-shot IcL at 36.9\%, and 1-shot IcL at 35.4\%. manual CoT stands out, leading in 17 of 21 scenarios, followed by 3-shot IcL in 15, and 1-shot IcL in 13 scenarios, in terms of the accuracy of \textit{scenario-prompt pairs} surpassing \textit{random guess accuracy}.
4) Language influence: English demonstrates superior performance in 10 of 21 scenarios, particularly in CDE, ETT, and NDE, with \textit{language accuracy difference}s of 8.4\%, 8.0\%, and 7.7\%, respectively. On the other hand, Chinese excels in CEG, AR, and AC, with accuracy improvements of 16.6\%, 5.5\%, and 4.7\%, respectively.

Ranking: 1) \textit{Prompt-average rank}: As depicted in Figure \ref{fig:Prompt-Average_Rank_of_Models}, Qwen (7B) achieves its best \textit{prompt-average rank}s in AC (8), AR (9), and CORR (9). However, it ranks lowest in CA (19), PN (19), and NDE (18), suggesting areas for potential enhancement. The model's overall \textit{prompt-average rank} across 21 scenarios is 12th out of 28, with a standard deviation of 3.3.
2) \textit{Model-prompt rank}: Illustrated in Figure \ref{fig:Heatmap_of_Qwen_(7B)}(b), the highest \textit{model-prompt rank}s for Qwen (7B) are in CDE with manual CoT (12), ETT with manual CoT (13), and NDE with manual CoT (20). The lowest ranks are in FAS with manual CoT (207), NDE with EF (206), and CA with adversarial doubt (197), pinpointing specific areas of challenge.

Robustness: Qwen (7B) displays an average robustness score of 56.6\% across different scenarios, showcasing top robustness in PCD (72.4\%), CB (61.4\%), and PS (61.2\%).

\paragraph{Qwen (14B).}
\begin{figure}[t]
\centering
\subfigure[Performance of Qwen (14B)]{
\begin{minipage}{8.5cm}
\centering
\includegraphics[width=1\linewidth]{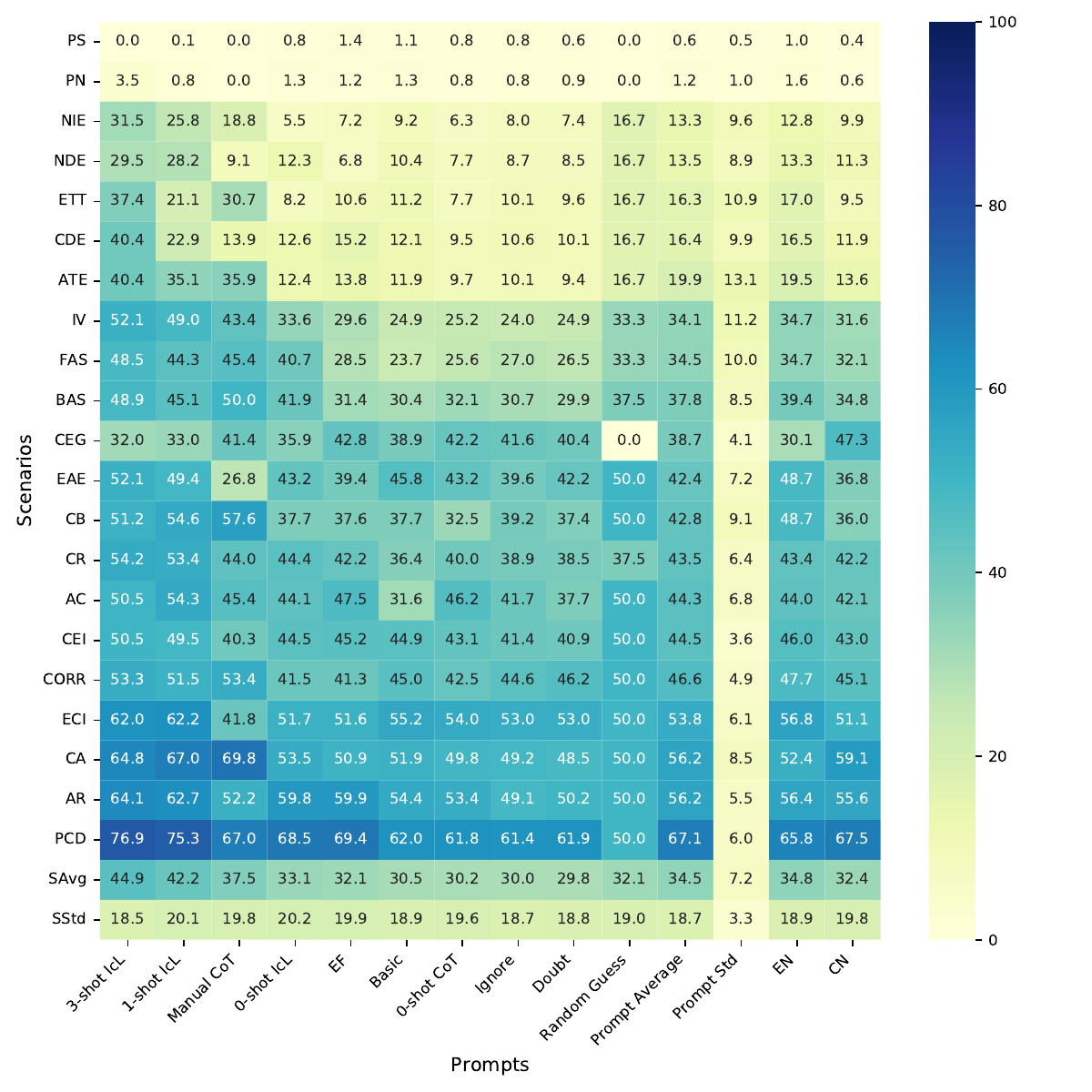}
\end{minipage}
}
\subfigure[\textit{Model-prompt rank} of Qwen (14B)]{
\begin{minipage}{8.5cm}
\centering
\includegraphics[width=1\linewidth]{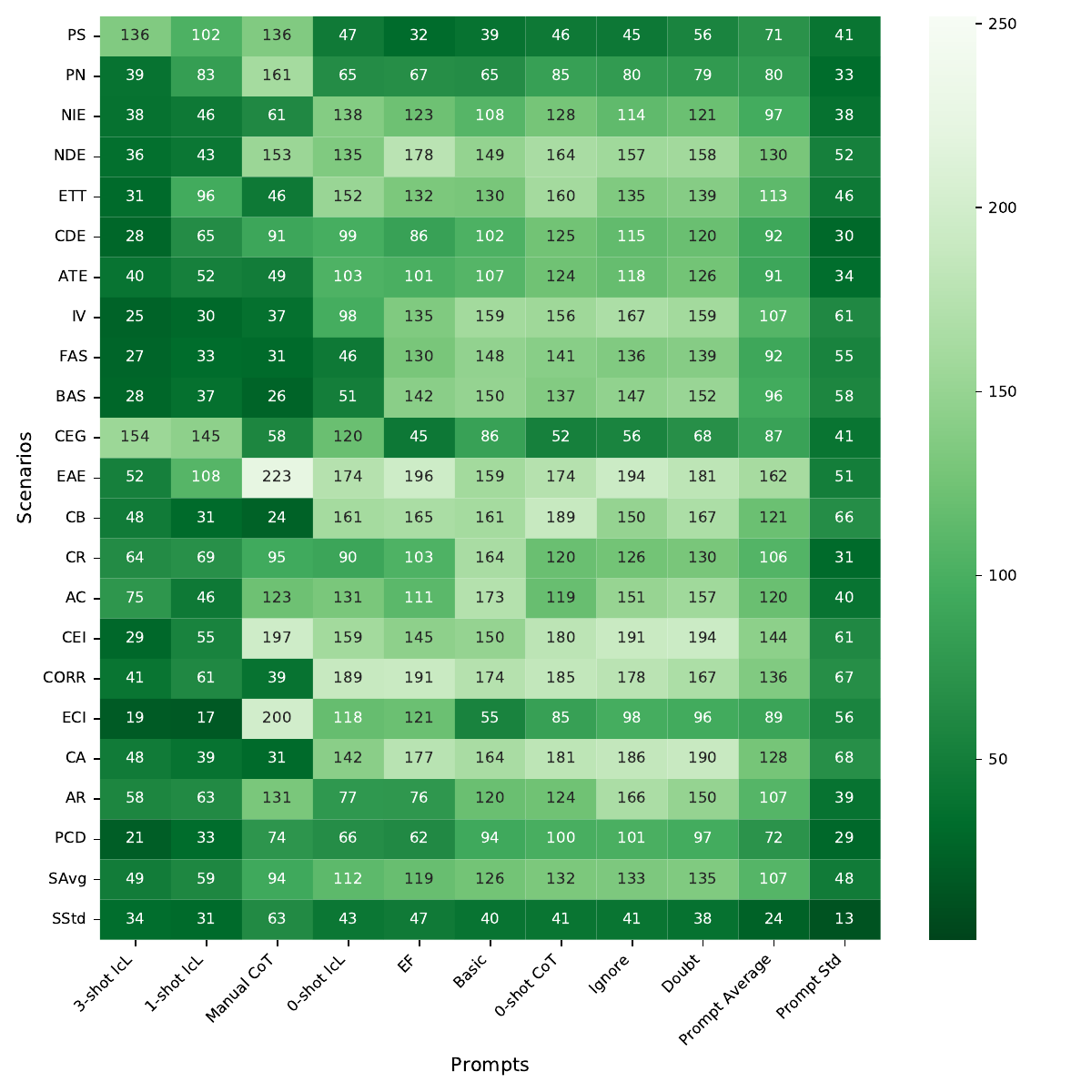}
\end{minipage}
}
\caption[Heatmap of Qwen (14B)]{\textbf{Heatmap of Qwen (14B).}}
\label{fig:Heatmap_of_Qwen_(14B)}
\end{figure}

Summary: The model demonstrates an \textit{average scenario-prompt accuracy} of 34.5\%, holds an average \textit{prompt-average rank} of 11 out of 28, and shows an average robustness score of 57.3 across different scenarios.

Accuracy: 1) Overall performance: Figure \ref{fig:Heatmap_of_Qwen_(14B)}(a) shows that Qwen (14B) achieves an \textit{average scenario-prompt accuracy} of 34.5\%, with a 7.2 average standard deviation in prompt effectiveness. The \textit{top scenario-prompt pair}s are led by a 3-shot IcL in PCD with a score of 76.9\%, followed by a 1-shot IcL in the same scenario at 75.3\%, and manual CoT in CA at 69.8\%. Over half (53.4\%) of the \textit{scenario-prompt pairs} surpass the \textit{random guess accuracy}, yet none achieve above 80\% accuracy.
2) Scenario performance: In scenarios where Qwen (14B) exceeds the \textit{random guess accuracy}, the three leading scenarios in terms of highest average accuracy include PCD at an impressive 67.1\%, followed by AR and CA, each scoring 56.2\%.
3) Prompt efficiency: The most efficient prompts identified are 3-shot IcL at 44.9\%, 1-shot IcL at 42.2\%, and manual CoT at 37.5\%. In terms of the \textit{scenario-prompt pairs} exceeding \textit{random guess accuracy} across scenarios, 3-shot IcL leads in all 21 scenarios, followed by 1-shot IcL in 19, and manual CoT in 15 scenarios.
4) Language influence: English proves superior in 18 of 21 scenarios, especially in CB, EAE, and ETT, with \textit{language accuracy difference}s of 12.7\%, 11.9\%, and 7.5\%, respectively. However, the Chinese perform better in scenarios like CEG, CA, and PCD, with accuracy differences of 17.2\%, 6.8\%, and 1.7\%, respectively.

Ranking: 1) \textit{Prompt-average rank}: As indicated in Figure \ref{fig:Prompt-Average_Rank_of_Models}, Qwen (14B) excels in CEG, ranking 6th, followed by ECI at 7th. It also demonstrates strong performance in PCD, ATE, BAS, and IV, each with an 8th-place rank in prompt-average scores. Its lowest ranks are in EAE (19), CORR (16), and PN (15), highlighting potential areas for development. The model's average \textit{prompt-average rank} across 21 scenarios is 11th out of 28, with a standard deviation of 3.4.
2) \textit{Model-prompt rank}: Illustrated in Figure \ref{fig:Heatmap_of_Qwen_(14B)}(b), the model's top \textit{model-prompt rank}s are in ECI with 1-shot IcL (17), ECI with 3-shot IcL (19), and PCD with 3-shot IcL (21). The lowest ranks are noted in EAE with manual CoT (223), ECI with manual CoT (200), and CEI with manual CoT (197), identifying specific areas for improvement.

Robustness: Qwen (14B) records an average robustness score of 57.3\% across various scenarios, achieving its highest robustness in PCD (75.6\%), ECI (66.7\%), and CA (63.0\%).

\subsubsection{Baichuan Inc.}
\label{model:baichuan}
\paragraph{Baichuan1 (7B).}
\begin{figure}[t]
\centering
\subfigure[Performance of Baichuan1 (7B)]{
\begin{minipage}{8.5cm}
\centering
\includegraphics[width=1\linewidth]{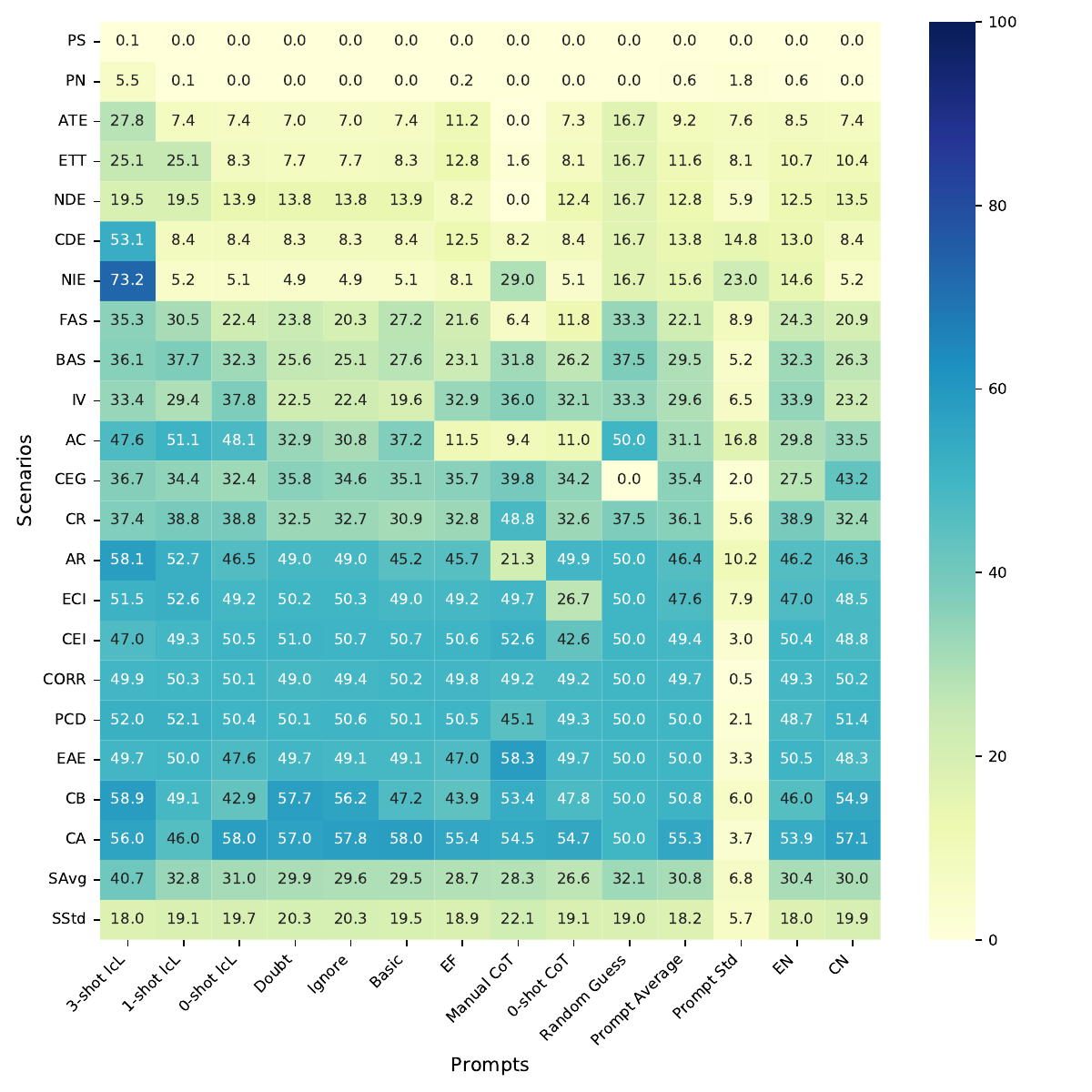}
\end{minipage}
}
\subfigure[\textit{Model-prompt rank} of Baichuan1 (7B)]{
\begin{minipage}{8.5cm}
\centering
\includegraphics[width=1\linewidth]{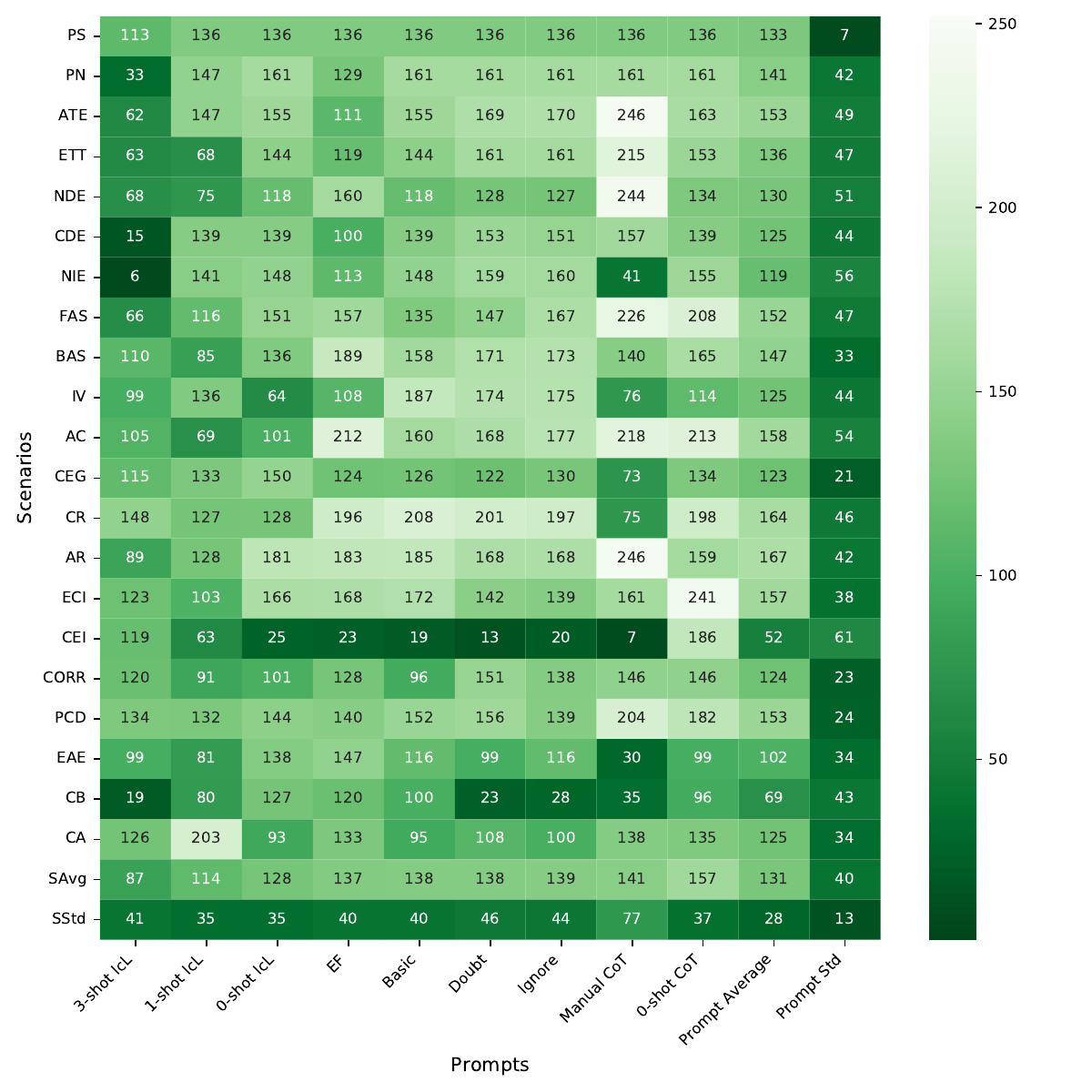}
\end{minipage}
}
\caption[Heatmap of Baichuan1 (7B)]{\textbf{Heatmap of Baichuan1 (7B).}}
\label{fig:Heatmap_of_Baichuan1_(7B)}
\end{figure}
Summary: The model achieves an \textit{average scenario-prompt accuracy} of 30.8\%, has an average \textit{prompt-average rank} of 14 out of 28, and maintains an average robustness score of 83.1\% across different scenarios.

Accuracy: 1) Overall performance: Figure \ref{fig:Heatmap_of_Baichuan1_(7B)}(a) shows that Baichuan1 (7B) achieves an \textit{average scenario-prompt accuracy} of 30.8\%, with an average variability (standard deviation) of 6.8 in prompt effectiveness. The \textit{top scenario-prompt pair}s are a 3-shot IcL in the NIE with a score of 73.2\%, followed by 3-shot IcL in CB at 58.9\%, and a manual CoT in EAE at 58.3\%. Over 42.3\% of the \textit{scenario-prompt pairs} outperform the baseline \textit{random guess accuracy}, yet none surpass the 80\% accuracy mark.
2) Scenario performance: In situations where Baichuan1 (7B) exceeds the \textit{random guess accuracy}, the top three scenarios by average accuracy are CA, leading with a significant score of 55.3\%, followed by CB at 50.8\%, and EAE at 50.0\%.
3) Prompt efficiency: The 3-shot IcL prompt is the most effective, with a score of 40.7\%, closely followed by the 1-shot IcL at 32.8\%. As to the number of \textit{scenario-prompt pair}s where the model exceeds the \textit{random guess accuracy}, the 3-shot IcL leads in 15 out of 21 scenarios, followed by 1-shot IcL in 13, and manual CoT in 10 scenarios.
4) Language influence: English has the upper hand in 11 out of 21 scenarios, especially in IV, NIE, and CR, with \textit{language accuracy difference}s of 10.7\%, 9.4\%, and 6.5\%, respectively. On the other hand, Chinese performs better in CEG, CB, and AC, with advantages of 15.7\%, 8.9\%, and 3.7\%, respectively.

Ranking: 1) \textit{Prompt-average rank}: Figure \ref{fig:Prompt-Average_Rank_of_Models} illustrates Baichuan1 (7B)'s best \textit{prompt-average rank}s in scenarios such as CEI at rank 3, CB at 5, and EAE at 9. Its lowest ranks are in PS at 28, PN and CR both at 21, indicating areas needing enhancement. The average \textit{prompt-average rank} across 21 scenarios stands at 14 out of 28, with a standard deviation of 5.6.
2) \textit{Model-prompt rank}: As depicted in Figure \ref{fig:Heatmap_of_Baichuan1_(7B)}(b), the top \textit{model-prompt rank}s for Baichuan1 (7B) are in NIE with a 3-shot IcL at 6, and CEI with manual CoT and with adversarial doubt at 7 and 13, respectively. The lowest ranks are seen in ATE at 246, AR at 246, and NDE with manual CoT at 244, all indicating specific areas of challenge.

Robustness: Baichuan1 (7B) shows remarkable robustness with an average score of 83.1\% across scenarios. It demonstrates top robustness in CA at 98.9\%, EAE at 97.9\%, and CEI at 97.0\%.

\paragraph{Baichuan1-chat (13B).}
\begin{figure}[t]
\centering
\subfigure[Performance of Baichuan1-chat (13B)]{
\begin{minipage}{8.5cm}
\centering
\includegraphics[width=1\linewidth]{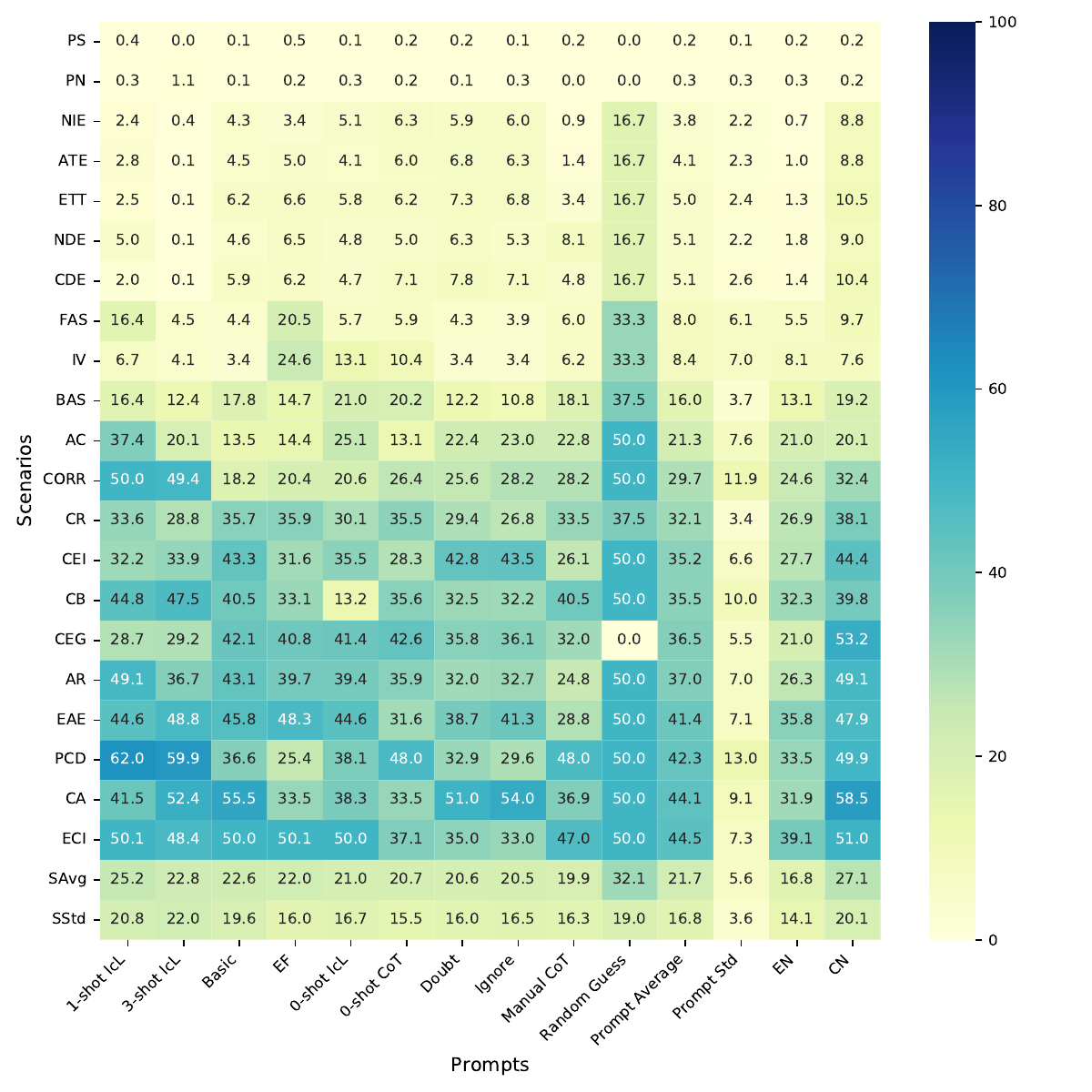}
\end{minipage}
}
\subfigure[\textit{Model-prompt rank} of Baichuan1-chat (13B)]{
\begin{minipage}{8.5cm}
\centering
\includegraphics[width=1\linewidth]{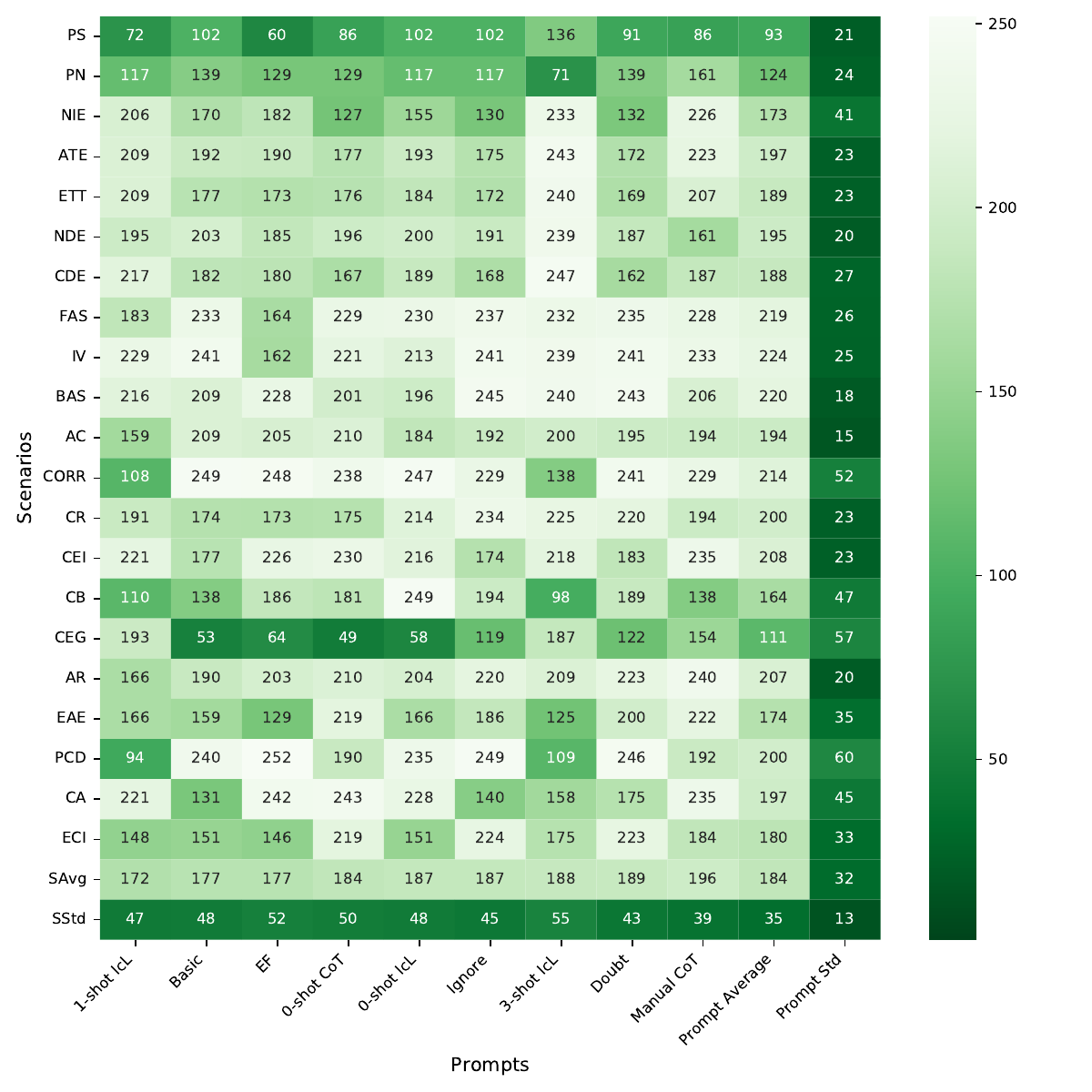}
\end{minipage}
}
\caption[Heatmap of Baichuan1-chat (13B)]{\textbf{Heatmap of Baichuan1-chat (13B).}}
\label{fig:Heatmap_of_Baichuan1-chat_(13B)}
\end{figure}
Summary: The model records an \textit{average scenario-prompt accuracy} of 21.7\%, attains an average \textit{prompt-average rank} of 23 out of 28, and possesses an average robustness score of 75.5\% across multiple scenarios.

Accuracy: 1) Overall performance: Figure \ref{fig:Heatmap_of_Baichuan1-chat_(13B)}(a) illustrates that Baichuan1-chat (13B) achieves an \textit{average scenario-prompt accuracy} of 21.7\%, with an average standard deviation of 5.6 indicating variability in prompt effectiveness. The \textit{top scenario-prompt pair}s are a 1-shot IcL in PCD with a score of 62.0\%, followed by a 3-shot IcL in the same category at 59.9\%, and a basic prompt in CA at 55.5\%. Only 20.1\% of the \textit{scenario-prompt pairs} outperform the \textit{random guess accuracy}, with none exceeding an 80\% accuracy threshold.
2) Scenario performance: When Baichuan1-chat (13B) exceeds the \textit{random guess accuracy}, the three scenarios with the highest average accuracies include CEG at a prominent score of 36.5\%, followed by PN at 0.3\%, and PS at 0.2\%.
3) Prompt efficiency: No prompt achieves an average higher than the random guess average. In the count of \textit{scenario-prompt pair}s where the model's accuracy beats the \textit{random guess accuracy}, the 1-shot IcL is ahead in 6 out of 21 scenarios, with both 3-shot IcL and basic following achieving a lead in 5 scenarios each.
4) Language influence: English shows superiority in 3 out of 21 scenarios, specifically in AC, IV, and PN, with \textit{language accuracy difference}s of 0.9\%, 0.5\%, and 0.1\%, respectively. In contrast, Chinese excels in CEG, CA, and AR, with accuracy advantages of 32.2\%, 26.5\%, and 22.8\%, respectively.

Ranking: 1) \textit{Prompt-average rank}: According to Figure \ref{fig:Prompt-Average_Rank_of_Models}, Baichuan1-chat (13B) performs best in CEG, ranking at 8. However, it ranks lowest at 28 in 6 scenarios. The average rank across 21 scenarios is 23 out of 28, with a standard deviation of 5.0.
2) \textit{Model-prompt rank}: As shown in Figure \ref{fig:Heatmap_of_Baichuan1-chat_(13B)}(b), Baichuan1-chat (13B)'s best \textit{model-prompt rank}s appear in CEG, with the highest rankings being 0-shot CoT at 49, basic at 53, and 0-shot IcL at 58. The lowest ranks are observed in PCD with EF at 252, CB with 0-shot IcL at 249, CORR with basic at 249.

Robustness: Baichuan1-chat (13B) demonstrates an average robustness score of 75.5\% across various scenarios, showcasing top robustness in ETT at 93.6\%, PS at 92.7\%, and PN at 92.5\%.

\paragraph{Baichuan2-chat (7B).}
\begin{figure}[t]
\centering
\subfigure[Performance of Baichuan2-chat (7B)]{
\begin{minipage}{8.5cm}
\centering
\includegraphics[width=1\linewidth]{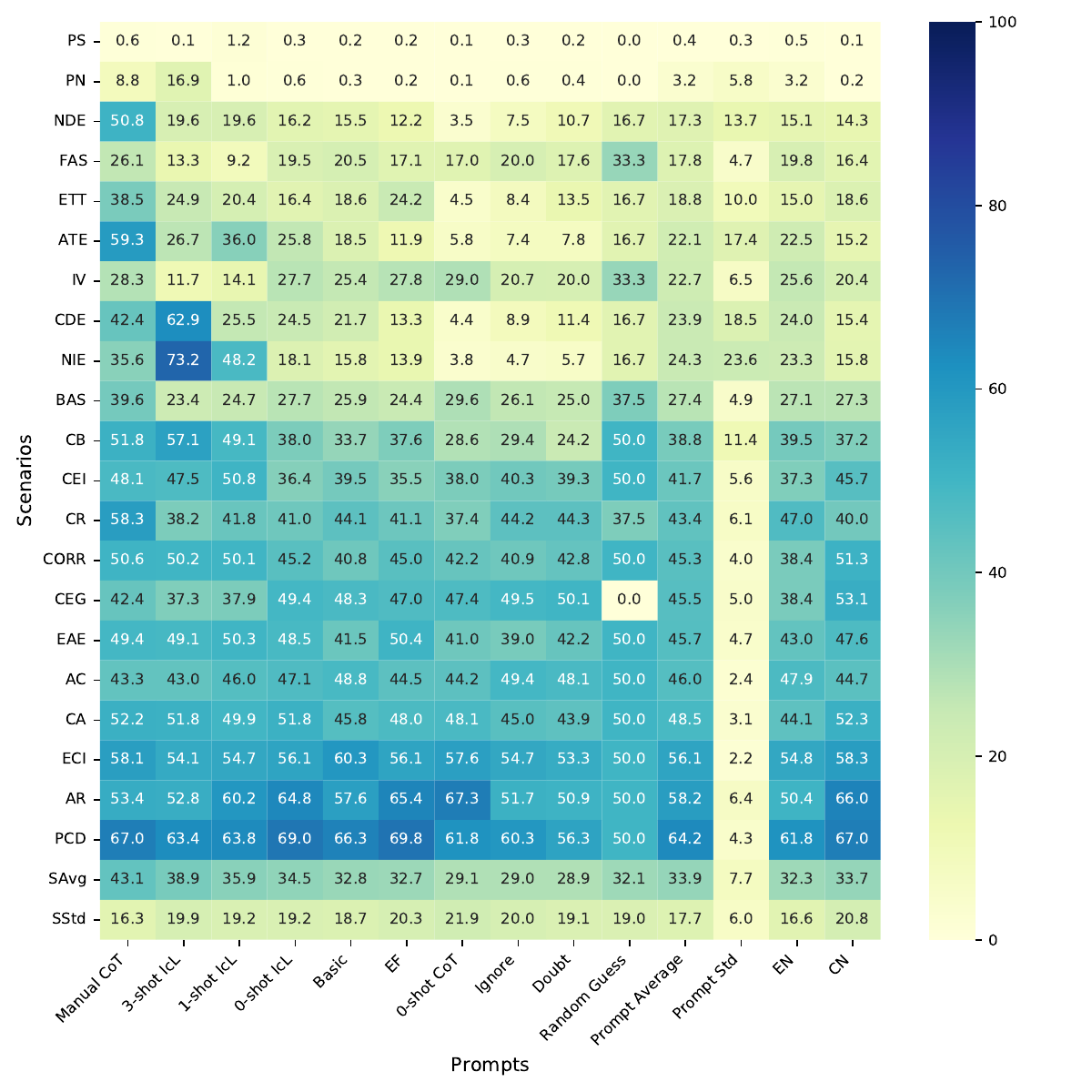}
\end{minipage}
}
\subfigure[\textit{Model-prompt rank} of Baichuan2-chat (7B)]{
\begin{minipage}{8.5cm}
\centering
\includegraphics[width=1\linewidth]{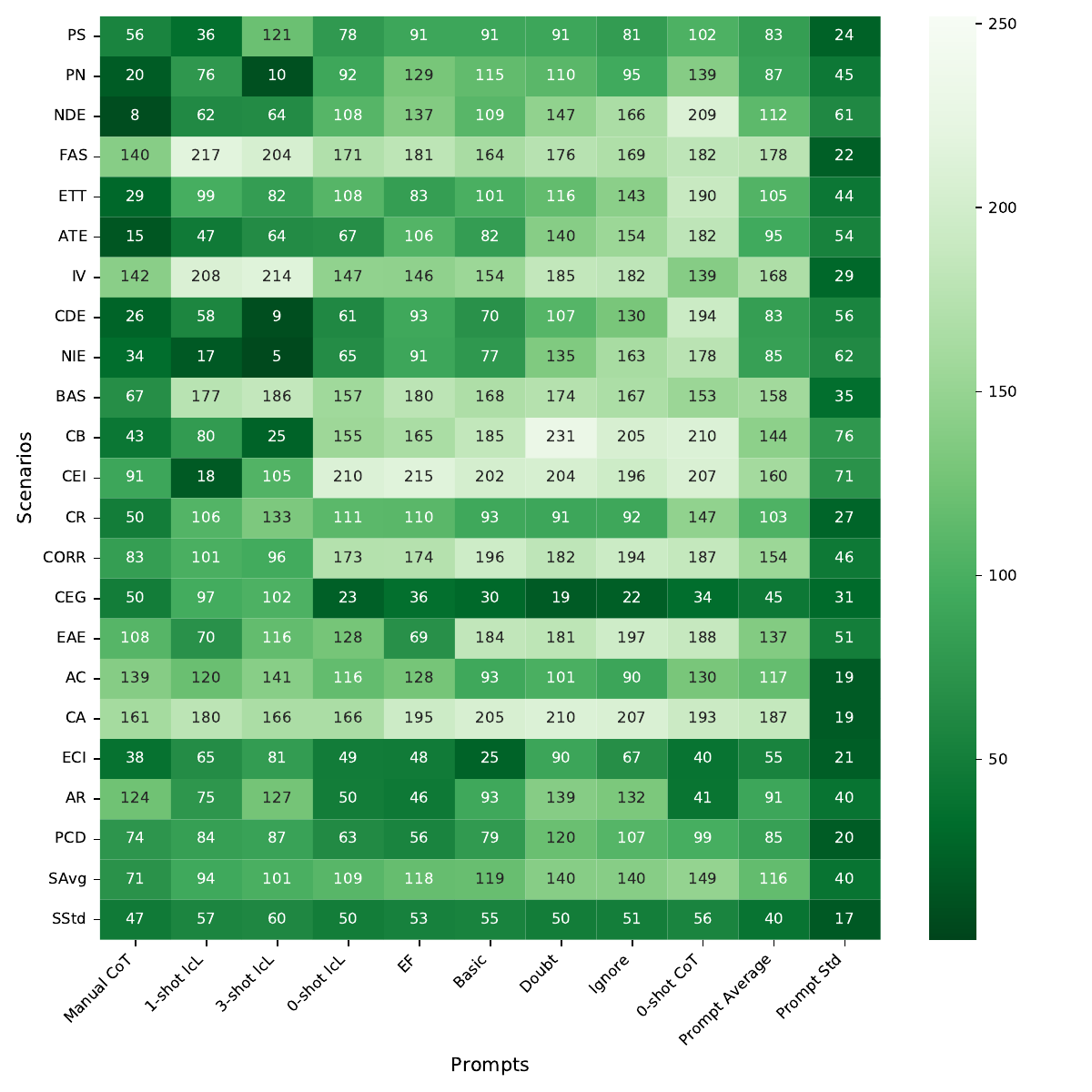}
\end{minipage}
}
\caption[Heatmap of Baichuan2-chat (7B)]{\textbf{Heatmap of Baichuan2-chat (7B).}}
\label{fig:Heatmap_of_Baichuan2-chat_(7B)}
\end{figure}
Summary: The model presents an \textit{average scenario-prompt accuracy} of 33.9\%, secures an average \textit{prompt-average rank} of 12 out of 28, and achieves an average robustness score of 63.8\%.

Accuracy: 1) Overall performance: According to Figure \ref{fig:Heatmap_of_Baichuan2-chat_(7B)}(a), Baichuan2-chat (7B) achieves an \textit{average scenario-prompt accuracy} of 33.9\%, with an average variability in prompt effectiveness indicated by a standard deviation of 7.7. The \textit{top scenario-prompt pair}s are in a 3-shot IcL in NIE at 73.2\%, EF in PCD at 69.8\%, and a 0-shot IcL in PCD at 69.0\%. More than half of the \textit{scenario-prompt pairs} (50.8\%) surpass the \textit{random guess accuracy}, though none achieve over 80\% accuracy.
2) Scenario performance: When Baichuan2-chat (7B) outperforms the \textit{random guess accuracy}, the three leading scenarios in terms of average accuracy are PCD with a distinguished score of 64.2\%, followed by AR at 58.2\%, and ECI at 56.1\%.
3) Prompt efficiency: The most effective prompts have been identified as manual CoT at 43.1\%, 3-shot IcL at 38.9\%, and 1-shot IcL at 35.9\%. Concerning \textit{scenario-prompt pair}s where the model's accuracy outperforms the \textit{random guess accuracy}, manual CoT is the leader in 16 out of 21 cases, closely followed by both 3-shot IcL and 1-shot IcL, each leading in 15 scenarios.
4) Language influence: In 11 scenarios, English demonstrates superior performance over Chinese, particularly in CDE, NIE, and ATE, with \textit{language accuracy difference}s of 8.7\%, 7.4\%, and 7.3\%, respectively. Conversely, Chinese outperforms English in AR, CEG, and CORR, with accuracy differences of 15.6\%, 14.7\%, and 12.9\%, respectively.

Ranking: 1) \textit{Prompt-average rank}: Figure \ref{fig:Prompt-Average_Rank_of_Models} shows Baichuan2-chat (7B)'s highest \textit{prompt-average rank}s in NIE, ECI, CEG, and CDE, all at rank 5. The model's lower ranks in CEI at 23, FAS at 22, and CA at 22 suggest areas needing improvement. The average \textit{prompt-average rank} across 21 scenarios is 12 out of 28, with a standard deviation of 6.4.
2) \textit{Model-prompt rank}: As shown in Figure \ref{fig:Heatmap_of_Baichuan2-chat_(7B)}(b), Baichuan2-chat (7B)'s top ranks are highlighted in NIE with a 3-shot IcL at rank 5, NDE with manual CoT at 8, and CDE with 3-shot IcL at 9. The lowest ranks are in CB with adversarial doubt at 231, FAS with 1-shot IcL at 217, and CEI with EF at 215, pinpointing specific areas of challenge.

Robustness: Baichuan2-chat (7B) showcases an average robustness score of 63.8\% across various scenarios, with its strongest robustness in AC at 95.3\%, CB at 84.2\%, and CEI at 83.5\%, indicating high reliability in these areas.

\paragraph{Baichuan2-chat (13B).}
\begin{figure}[t]
\centering
\subfigure[Performance of Baichuan2-chat (13B)]{
\begin{minipage}{8.5cm}
\centering
\includegraphics[width=1\linewidth]{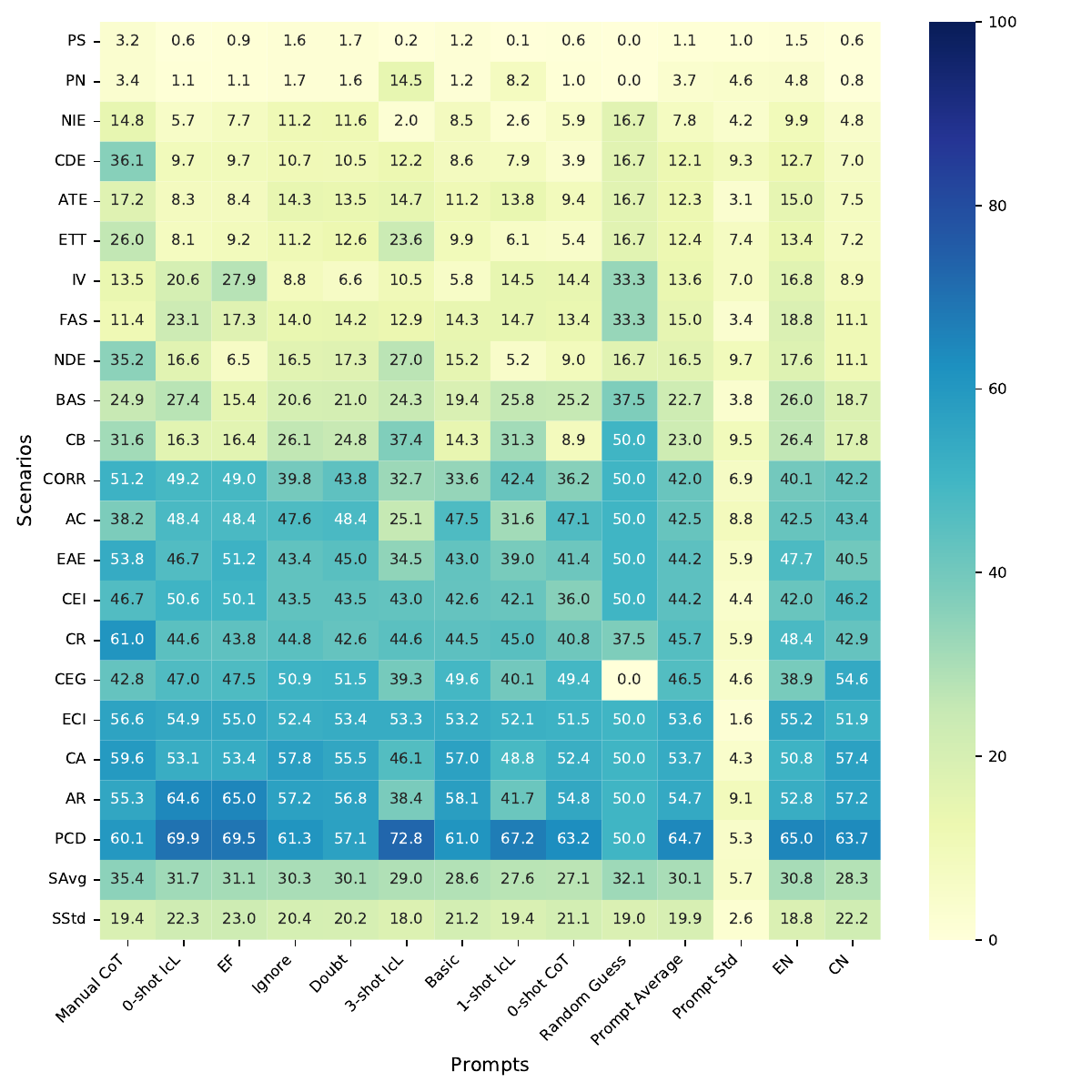}
\end{minipage}
}
\subfigure[\textit{Model-prompt rank} of Baichuan2-chat (13B)]{
\begin{minipage}{8.5cm}
\centering
\includegraphics[width=1\linewidth]{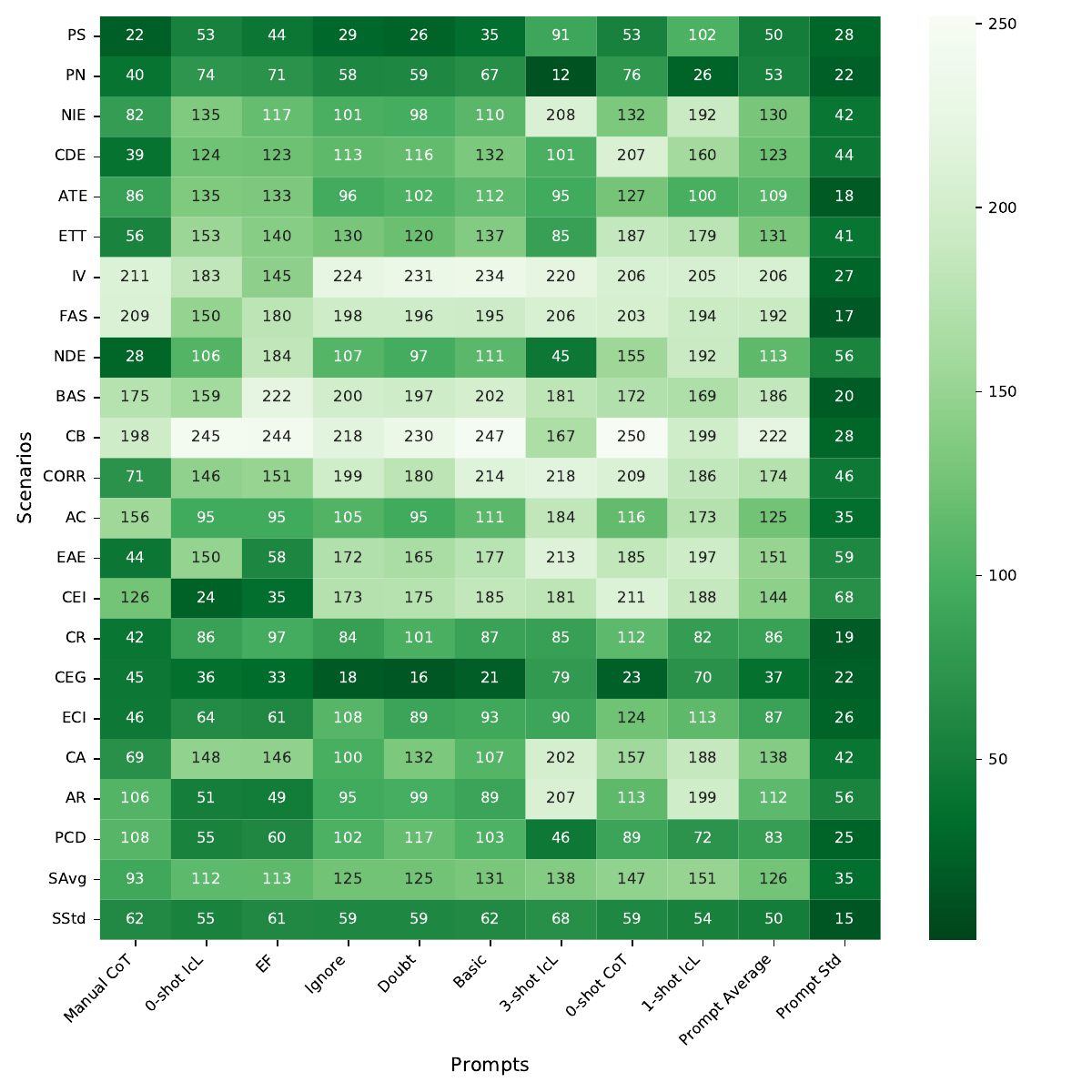}
\end{minipage}
}
\caption[Heatmap of Baichuan2-chat (13B)]{\textbf{Heatmap of Baichuan2-chat (13B).}}
\label{fig:Heatmap_of_Baichuan2-chat_(13B)}
\end{figure}
Summary: The model displays an \textit{average scenario-prompt accuracy} of 30.1\%, achieves an average \textit{prompt-average rank} of 15 out of 28, and holds an average robustness score of 71.6\% across different scenarios.

Accuracy: 1) Overall performance: As shown in Figure \ref{fig:Heatmap_of_Baichuan2-chat_(13B)}(a), Baichuan2-chat (13B) achieves an \textit{average scenario-prompt accuracy} of 30.1\%, with the variability in prompt effectiveness captured by an average standard deviation of 5.7. The \textit{top scenario-prompt pair}s are a 3-shot IcL in PCD with a score of 72.8\%, followed closely by a 0-shot IcL at 69.9\% and EF at 69.5\% in the same scenario. A total of 42.3\% of the \textit{scenario-prompt pairs} outperform the \textit{random guess accuracy}, with none surpassing the 80\% accuracy threshold.
2) Scenario performance: When Baichuan2-chat (13B) outstrips the \textit{random guess accuracy}, the top three scenarios by average accuracy are PCD with an impressive score of 64.7\%, followed by AR at 54.7\%, and CA at 53.7\%.
3) Prompt efficiency: The most effective prompts are manual CoT at 35.4\%. In terms of overcoming the \textit{random guess accuracy} for various \textit{scenario-prompt pair}s, manual CoT is the leader in 14 out of 21 scenarios, with EF following in 10, and 0-shot IcL in 9 scenarios.
4) Language influence: English shows superior performance in 15 out of 21 scenarios, particularly in CB, IV, and FAS, with \textit{language accuracy difference}s of 8.6\%, 7.9\%, and 7.7\%, respectively. In contrast, Chinese excels in CEG, CA, and AR, with accuracy differences of 15.7\%, 6.6\%, and 4.4\%, respectively.

Ranking: 1) \textit{Prompt-average rank}: According to Figure \ref{fig:Prompt-Average_Rank_of_Models}, Baichuan2-chat (13B) attains its top \textit{prompt-average rank}s with a 4th place in CEG, a 5th place in PN, and ties for 8th place in both CR and ECI. However, it ranks lowest in CB at 28, IV at 27, and BAS at 25, highlighting areas needing development. The model's average \textit{prompt-average rank} across 21 scenarios is 15 out of 28, with a standard deviation of 6.9.
2) \textit{Model-prompt rank}: Figure \ref{fig:Heatmap_of_Baichuan2-chat_(13B)}(b) showcases the model's best \textit{model-prompt rank}s in PN with a 3-shot IcL at 12, CEG with adversarial doubt at 16, and with adversarial ignore at 18. The lowest ranks are noted in CB with 0-shot CoT at 250, basic at 247, and 0-shot IcL at 245, pinpointing specific areas for improvement.

Robustness: Baichuan2-chat (13B) exhibits an average robustness score of 71.6\% across scenarios, with its strongest performance in CEI at 90.6\%, AC at 88.2\%, and PCD at 82.8\%, indicating a high degree of reliability in these areas.

\subsubsection{Meta}
\label{model:meta}
\paragraph{Llama2 (7B).}
\begin{figure}[t]
\centering
\subfigure[Performance of Llama2 (7B)]{
\begin{minipage}{8.5cm}
\centering
\includegraphics[width=1\linewidth]{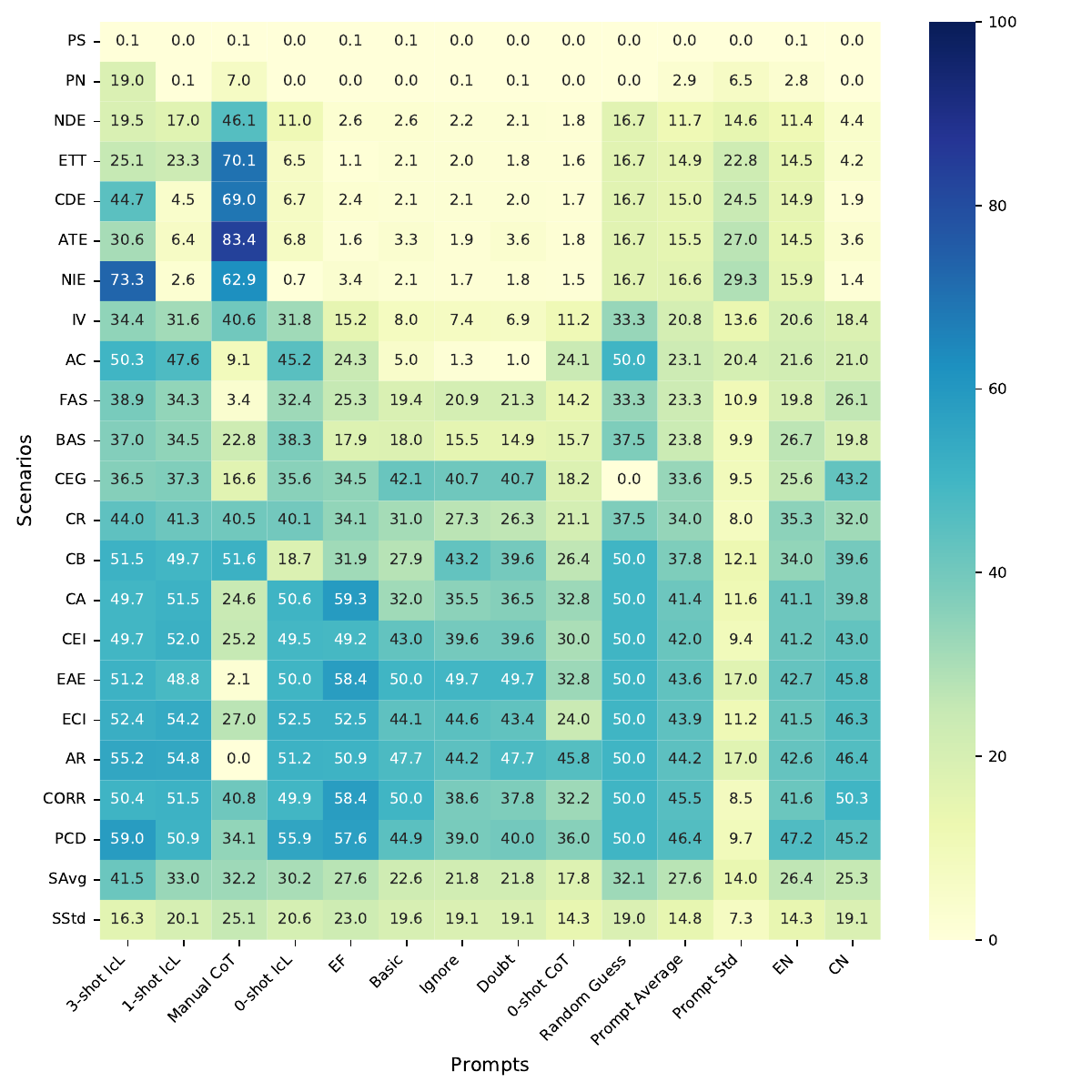}
\end{minipage}
}
\subfigure[\textit{Model-prompt rank} of Llama2 (7B)]{
\begin{minipage}{8.5cm}
\centering
\includegraphics[width=1\linewidth]{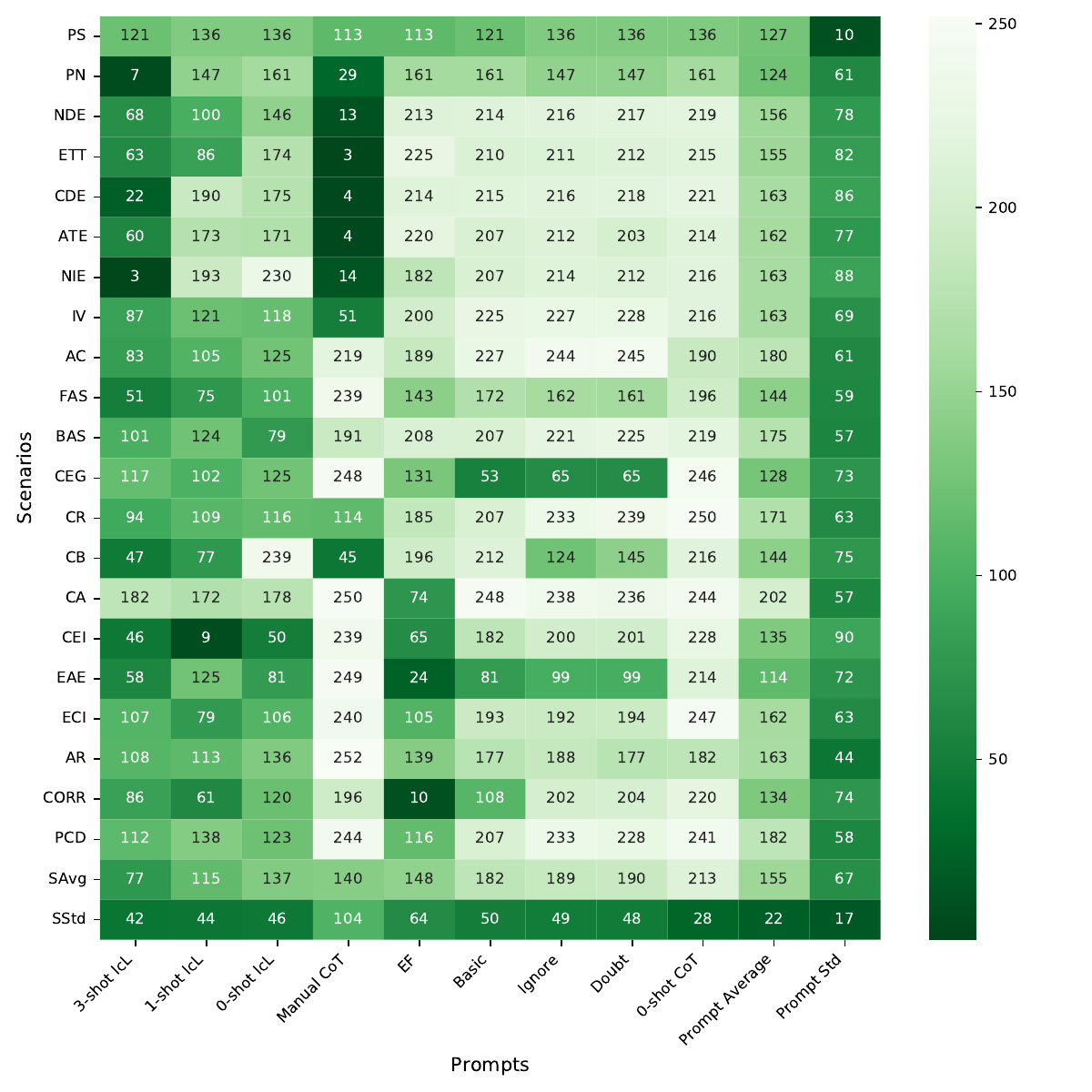}
\end{minipage}
}
\caption[Heatmap of Llama2 (7B)]{\textbf{Heatmap of Llama2 (7B).}}
\label{fig:Heatmap_of_Llama2_(7B)}
\end{figure}
Summary: The model achieves an \textit{average scenario-prompt accuracy} of 27.6\%, has an average \textit{prompt-average rank} of 19 out of 28, and maintains an average robustness score of 74.8\% across scenarios.

Accuracy: 1) Overall performance: Presented in Figure \ref{fig:Heatmap_of_Llama2_(7B)}(a), Llama2 (7B) scores 27.6\% in \textit{average scenario-prompt accuracy}, with a high variability in effectiveness shown by an average prompts' standard deviation of 14.0. The \textit{top scenario-prompt pair}s are manual CoT in ATE at 83.4\%, a 3-shot IcL in NIE at 73.3\%, and manual CoT in ETT at 70.1\%. About 39.7\% of the \textit{scenario-prompt pairs} outperform the \textit{random guess accuracy}, with a marginal 0.5\% exceeding 80\% accuracy.
2) Scenario performance: When identifying scenarios where Llama2 (7B) exceeds the \textit{random guess accuracy}, the top scenarios with the highest average accuracies are CEG at 33.6\%, PN at 2.9\%.
3) Prompt efficiency: The most effective prompts are 3-shot IcL at 41.5\%, 1-shot IcL at 33.0\%, and manual CoT at 32.2\%. As to the number of \textit{scenario-prompt pair}s where the model exceeds the \textit{random guess accuracy}, the 3-shot IcL leads in 18 out of 21 scenarios, followed by 1-shot IcL in 13, and manual CoT in 11 scenarios.
4) Language influence: English surpasses Chinese in 13 out of 21 scenarios, especially in NIE, CDE, and ATE, with \textit{language accuracy difference}s of 14.5\%, 13.0\%, and 10.9\%, respectively. Conversely, Chinese outperforms in CEG, CORR, and FAS, with differences of 17.6\%, 8.7\%, and 6.3\%, respectively.

Ranking: 1) \textit{Prompt-average rank}: According to Figure \ref{fig:Prompt-Average_Rank_of_Models}, Llama2 (7B)'s highest \textit{prompt-average rank}s is in PN at 10. The model's lowest rankings are 28th in CA, 25th in PCD, and 22nd across five scenarios, indicating areas requiring improvement. The average \textit{prompt-average rank} across 21 scenarios stands at 19 out of 28, with a standard deviation of 4.1.
2) \textit{Model-prompt rank}: As shown in Figure \ref{fig:Heatmap_of_Llama2_(7B)}(b), Llama2 (7B) achieves its highest \textit{model-prompt rank}s in several key areas: it ranks third in both ETT and NIE, accomplished using manual CoT and a 3-shot IcL respectively. Additionally, it secures a fourth-place position in CDE and ATE, both attained through manual CoT, showcasing its strengths in complex causal analysis. The model faces significant challenges in AR with manual CoT at 252, CR with 0-shot CoT at 250, and CA with manual CoT at 250.

Robustness: Llama2 (7B) maintains an average robustness score of 74.8\% across scenarios, showcasing the highest robustness in EAE at 96.1\%, AC at 93.7\%, and PS at 88.2\%, demonstrating significant resilience in these areas.

\paragraph{Llama2 (13B).}
\begin{figure}[t]
\centering
\subfigure[Performance of Llama2 (13B)]{
\begin{minipage}{8.5cm}
\centering
\includegraphics[width=1\linewidth]{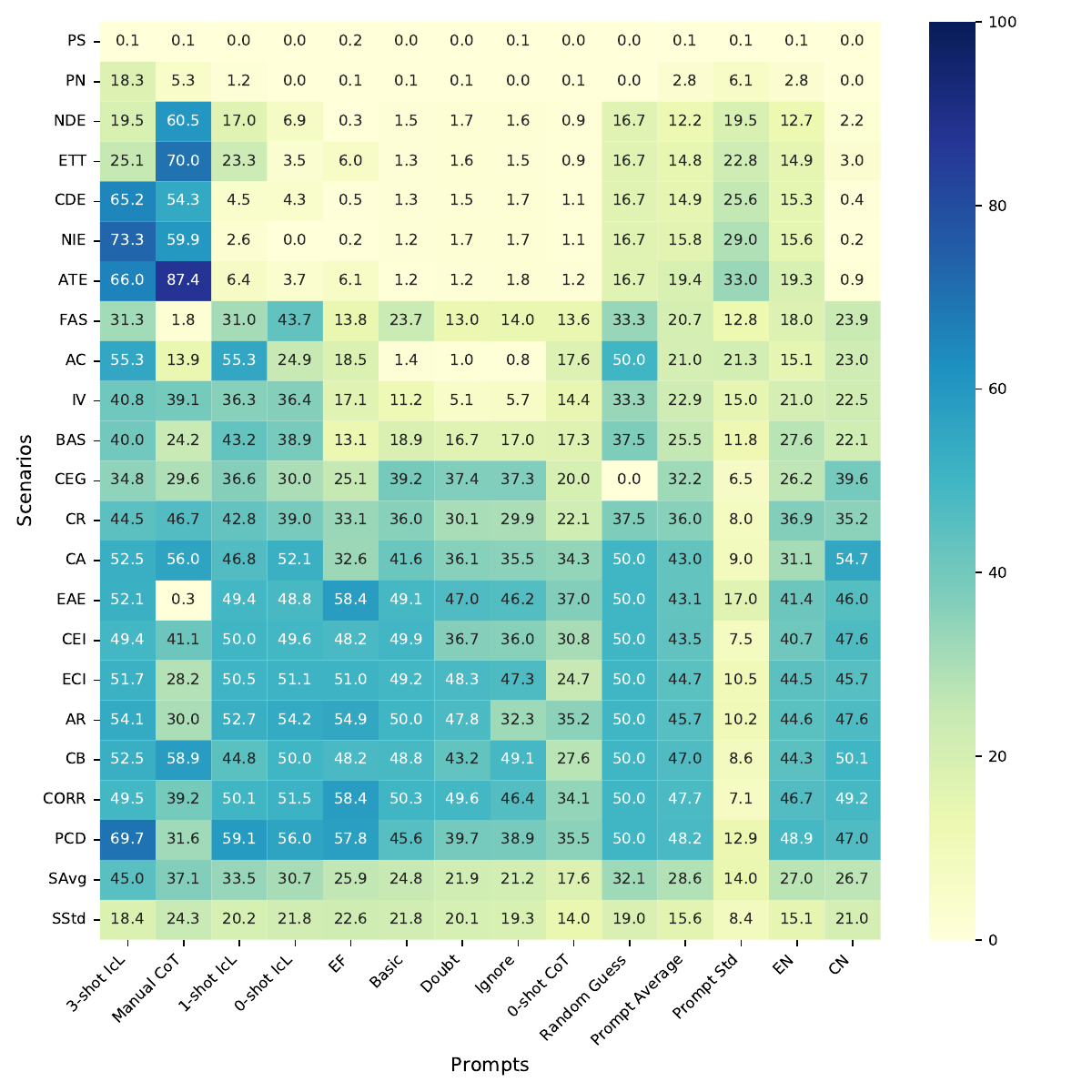}
\end{minipage}
}
\subfigure[\textit{Model-prompt rank} of Llama2 (13B)]{
\begin{minipage}{8.5cm}
\centering
\includegraphics[width=1\linewidth]{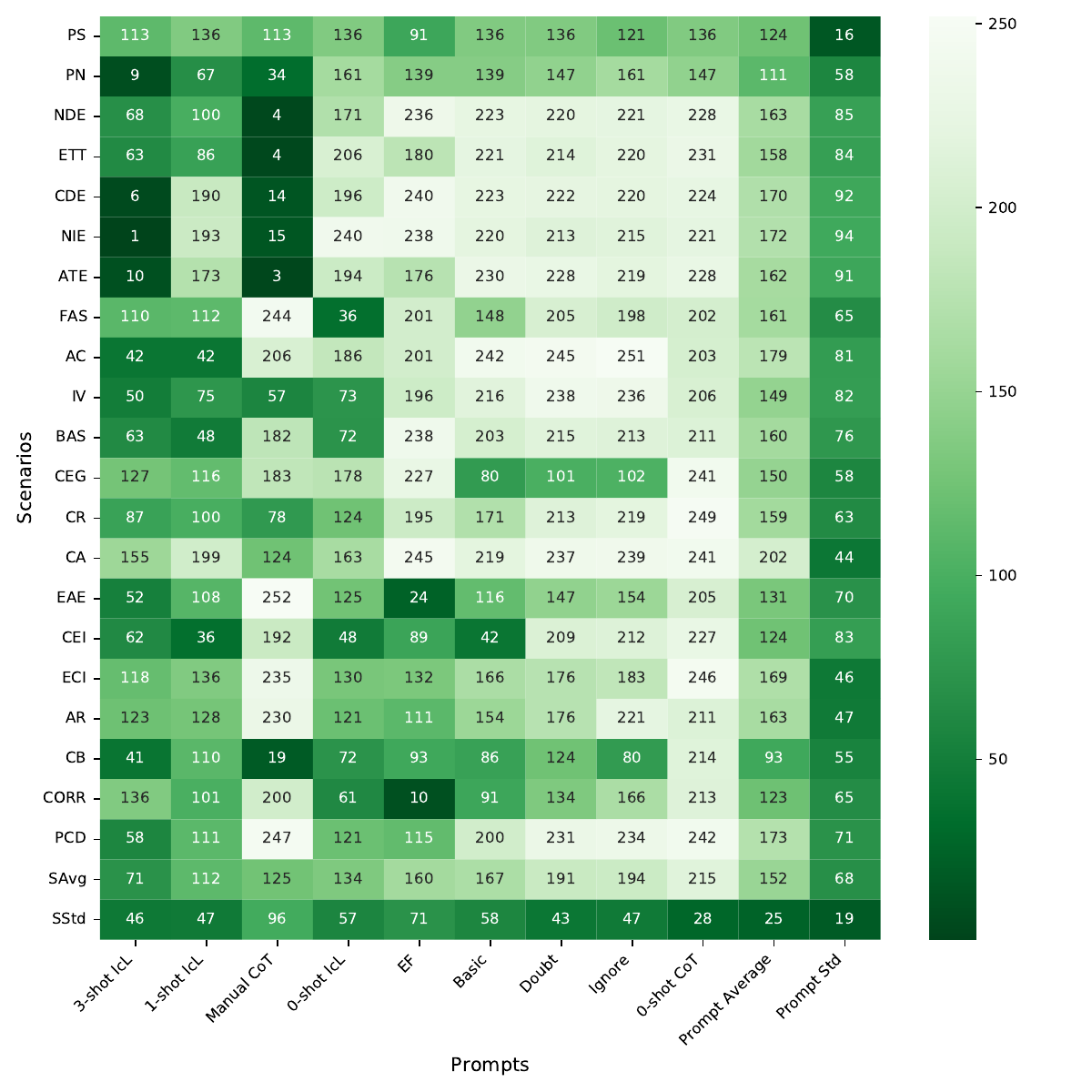}
\end{minipage}
}
\caption[Heatmap of Llama2 (13B)]{\textbf{Heatmap of Llama2 (13B).}}
\label{fig:Heatmap_of_Llama2_(13B)}
\end{figure}
Summary: The model shows an \textit{average scenario-prompt accuracy} of 28.6\%, attains an average \textit{prompt-average rank} of 17 out of 28, and achieves an average robustness score of 81.9\%.

Accuracy: 1) Overall performance: Illustrated in Figure \ref{fig:Heatmap_of_Llama2_(13B)}(a), Llama2 (13B) achieves an \textit{average scenario-prompt accuracy} of 28.6\%, with significant variability in prompt effectiveness indicated by an average prompts' standard deviation of 14.0. The \textit{top scenario-prompt pair}s feature manual CoT in ATE with a high score of 87.4\%, 3-shot IcL in NIE at 73.3\%, and manual CoT in ETT at 70.0\%. Approximately 41.8\% of the \textit{scenario-prompt pairs} surpass the \textit{random guess accuracy}, with a small fraction (0.5\%) achieving over 80\% accuracy.
2) Scenario performance: For scenarios in which Llama2 (13B) outperforms the \textit{random guess accuracy}, the top three scenarios ranked by their average accuracy are CEG with an impressive score of 32.2\%, ATE at 19.4\%, and PN at 2.8\%.
3) Prompt efficiency: The best-performed prompts include 3-shot IcL at 45.0\%, manual CoT at 37.1\%, and 1-shot IcL at 33.5\%. In \textit{scenario-prompt pair}s where the model's accuracy exceeds the \textit{random guess accuracy}, the 3-shot IcL is the frontrunner in 18 out of 21 cases. It is succeeded by the 1-shot IcL, which leads in 14 scenarios, and the 0-shot IcL, which leads in 13 scenarios.
4) Language influence: In 10 out of 21 scenarios, English shows superiority over Chinese, especially in ATE, NIE, and CDE, \textit{language accuracy difference}s of 18.4\%, 15.4\%, and 14.9\%, respectively. On the other hand, Chinese outperforms in CA, CEG, and AC, with differences of 23.6\%, 13.4\%, and 7.9\%, respectively.

Ranking: 1) \textit{Prompt-average rank}: As Figure \ref{fig:Prompt-Average_Rank_of_Models} shows, Llama2 (13B)'s top \textit{prompt-average rank}s are found in CB at 8 and PN at 9. However, it ranks lowest in CA at 26, AC at 23, and ETT at 22, highlighting areas needing development. The average rank across 21 scenarios is 17 out of 28, with a standard deviation of 4.2.
2) \textit{Model-prompt rank}: Figure \ref{fig:Heatmap_of_Llama2_(13B)}(b) presents the model's \textit{model-prompt rank}s, featuring a top rank in NIE with a 3-shot IcL at 1, a third-place in ATE via manual CoT, and fourth-place positions in ETT and NDE, both achieved with manual CoT. The most significant challenges are in EAE with manual CoT at 252, AC with adversarial ignore at 251, and CR with 0-shot CoT at 249.

Robustness: Llama2 (13B) maintains an impressive average robustness score of 81.9\% across scenarios, exhibiting top robustness in AC at 99.2\%, EAE at 96.0\%, and CB at 93.1\%, demonstrating significant resilience in these areas.

\paragraph{Llama2 (70B).}
\begin{figure}[t]
\centering
\subfigure[Performance of Llama2 (70B)]{
\begin{minipage}{8.5cm}
\centering
\includegraphics[width=1\linewidth]{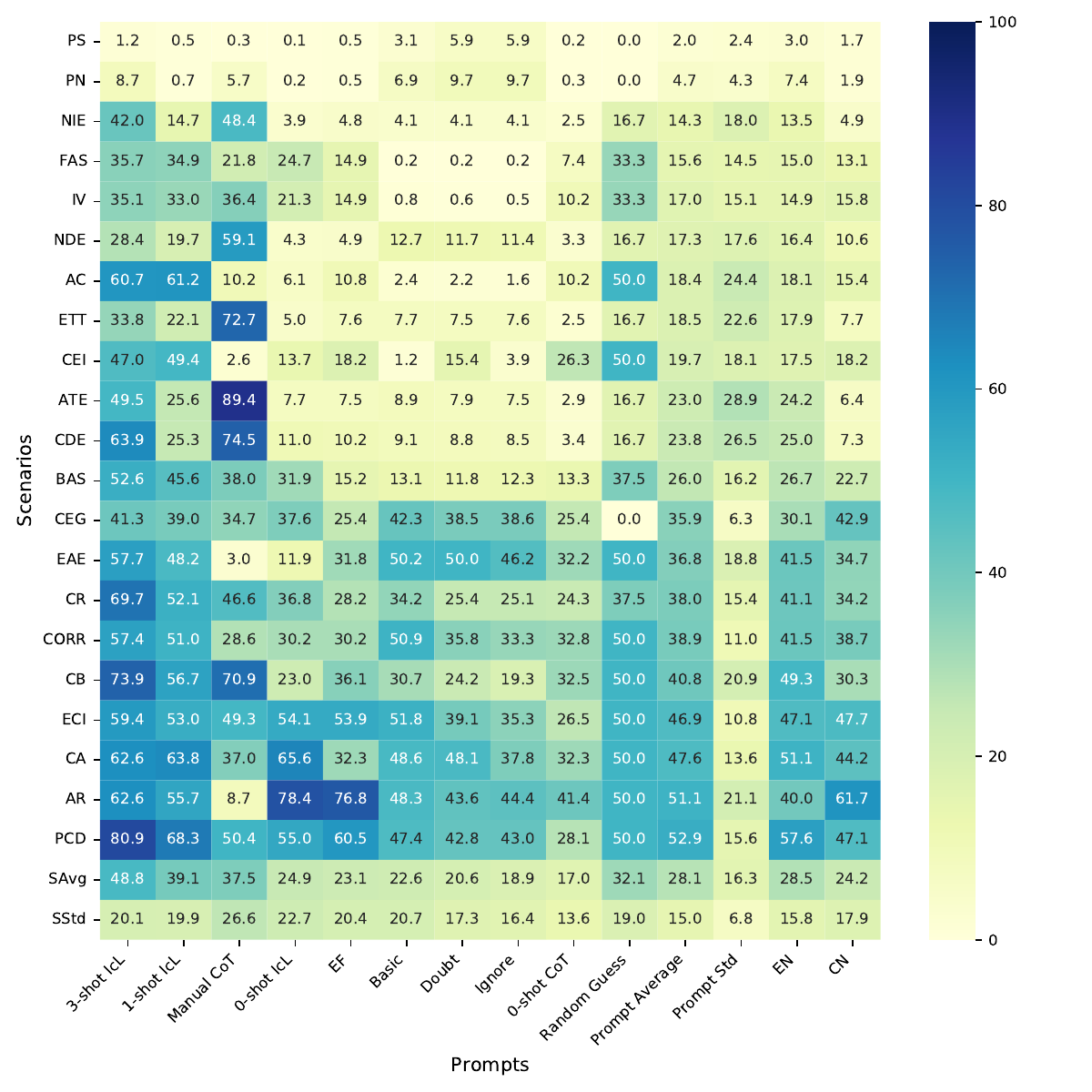}
\end{minipage}
}
\subfigure[\textit{Model-prompt rank} of Llama2 (70B)]{
\begin{minipage}{8.5cm}
\centering
\includegraphics[width=1\linewidth]{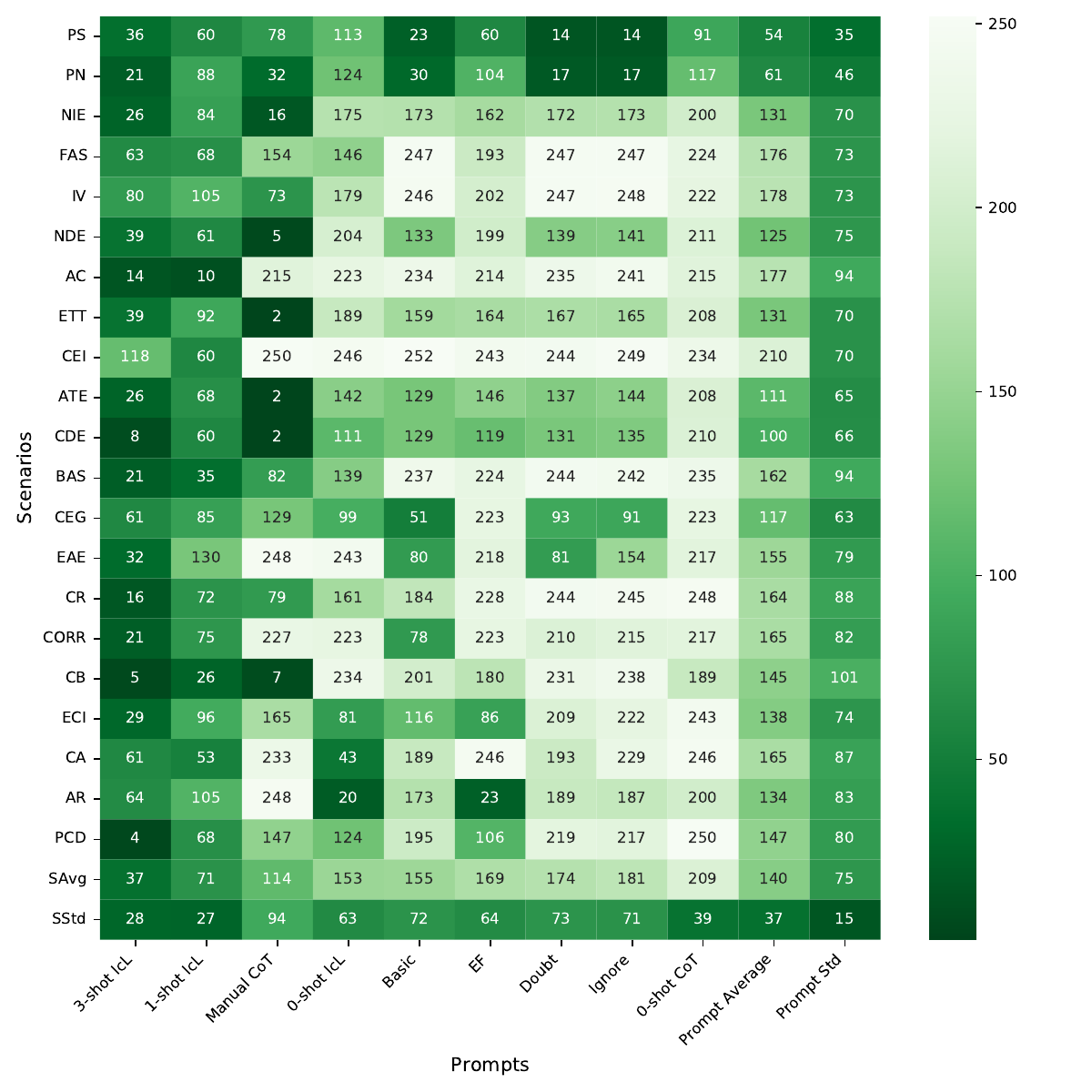}
\end{minipage}
}
\caption[Heatmap of Llama2 (70B)]{\textbf{Heatmap of Llama2 (70B).}}
\label{fig:Heatmap_of_Llama2_(70B)}
\end{figure}
Summary: The model's \textit{average scenario-prompt accuracy} is 28.1\%, with the average \textit{prompt-average rank} of 16/28 and an average robustness score of 72.7\%.

Accuracy: 1) Overall performance: As illustrated in Figure \ref{fig:Heatmap_of_Llama2_(70B)}(a), Llama2 (70B) achieves an \textit{average scenario-prompt accuracy} of 28.1\%, with significant variation in prompt effectiveness, demonstrated by an average prompts' standard deviation of 16.3. The \textit{top scenario-prompt pair}s are manual CoT in ATE at 89.4\%, 3-shot IcL in PCD at 80.9\%, and 0-shot IcL in AR at 78.4\%. Approximately 41.8\% of the \textit{scenario-prompt pairs} outperform the \textit{random guess accuracy}, with a small portion (1.1\%) exceeding 80\% accuracy.
2) Scenario performance: When Llama2 (70B) exceeds the accuracy of the \textit{random guess accuracy}, the three scenarios with the highest average accuracies include PCD at 52.9\%, AR at 51.1\%, and CR at 38.0\%.
3) Prompt efficiency:  The leading prompts include 3-shot IcL at 48.8\%, 1-shot IcL at 39.1\%, and manual CoT at 37.5\%. Regarding the \textit{scenario-prompt pair}s where the model's accuracy outstrips the \textit{random guess accuracy}, the 3-shot IcL takes the lead in 20 of the 21 scenarios. This is followed by the 1-shot IcL, which is ahead in 17 scenarios, and the manual CoT, leading in 13 scenarios.
4) Language influence: In 16 out of 21 scenarios, English demonstrates superior performance over Chinese, particularly in CB, ATE, and CDE, with \textit{language accuracy difference}s of 19.0\%, 17.7\%, and 17.7\%, respectively. Conversely, Chinese outshines English in AR, CEG, and IV, with differences of 21.7\%, 12.8\%, and 0.9\%, respectively.

Ranking: 1) \textit{Prompt-average rank}: According to Figure \ref{fig:Prompt-Average_Rank_of_Models}, Llama2 (70B)'s highest ranks are seen in PN at 3, PS at 5, and CDE at 7. The model encounters its most significant difficulties in CEI, ranking 28th, AC at 26th, and both IV and FAS at 25th, pointing to areas needing enhancement. The average rank across 21 scenarios is 16 out of 28, with a standard deviation of 7.4.
2) \textit{Model-prompt rank}: Figure \ref{fig:Heatmap_of_Llama2_(70B)}(b) highlights the model's top ranks in ATE with manual CoT at 2, CDE with manual CoT at 2, and ETT with manual CoT also at 2. The most significant challenges are CEI with basic at 252, PCD with 0-shot CoT at 250, and CEI with manual CoT at 250.

Robustness: Llama2 (70B) maintains an impressive average robustness score of 72.7\% across scenarios, with peak robustness in FAS at 100.0\%, AC at 99.1\%, and IV at 97.0\%, indicating strong resilience in these specific areas.

\paragraph{Llama2-chat (70B).}
\begin{figure}[t]
\centering
\subfigure[Performance of Llama2-chat (70B)]{
\begin{minipage}{8.5cm}
\centering
\includegraphics[width=1\linewidth]{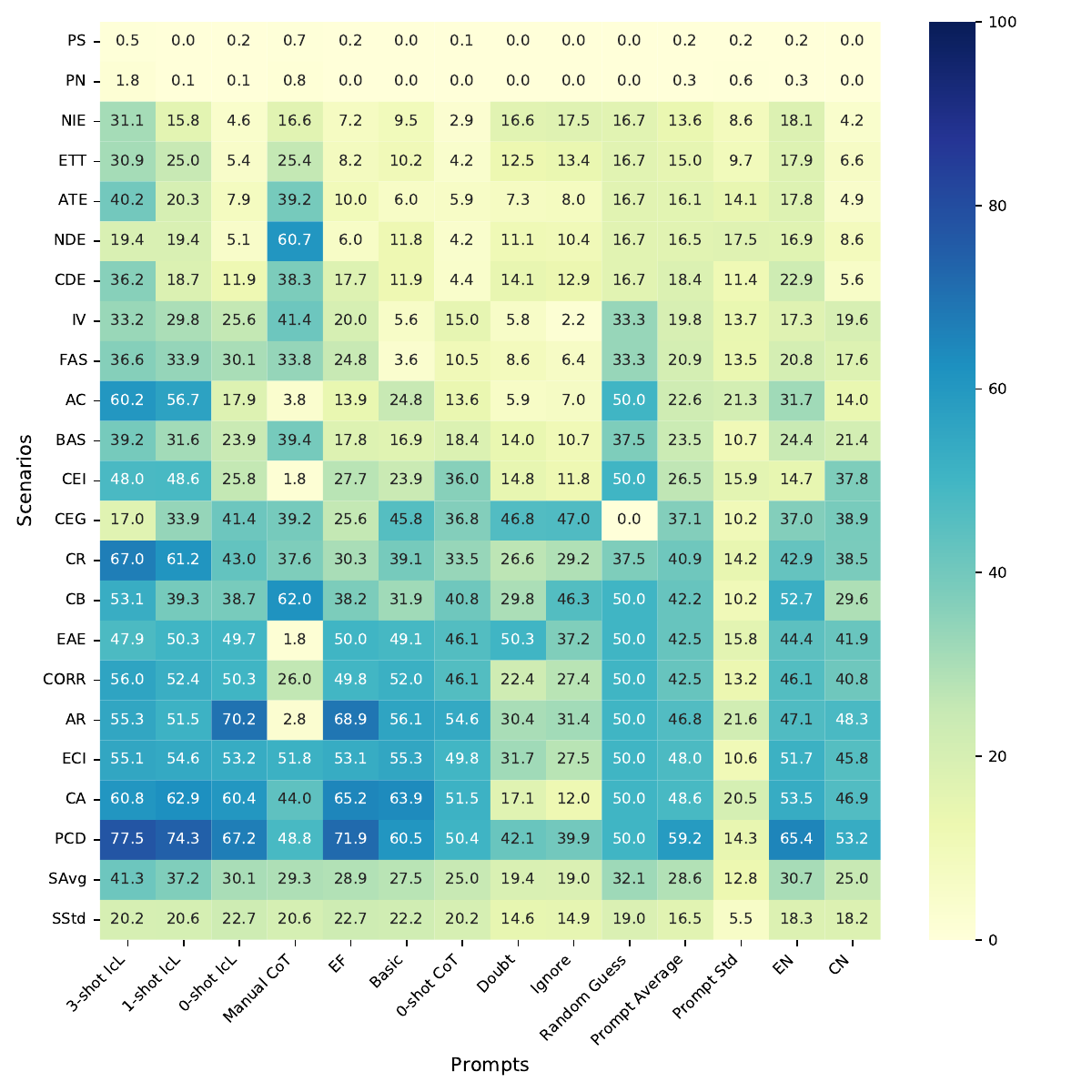}
\end{minipage}
}
\subfigure[\textit{Model-prompt rank} of Llama2-chat (70B)]{
\begin{minipage}{8.5cm}
\centering
\includegraphics[width=1\linewidth]{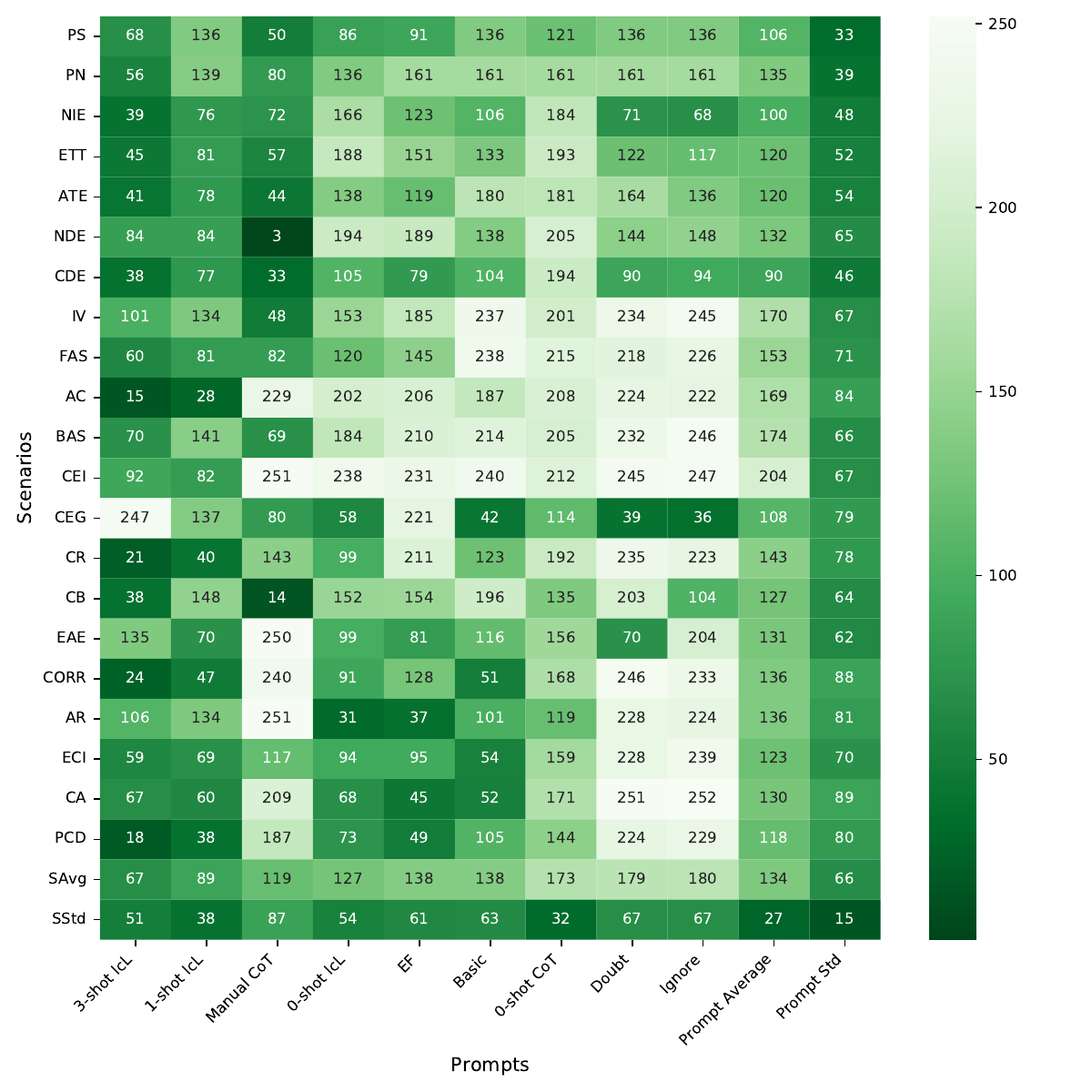}
\end{minipage}
}
\caption[Heatmap of Llama2-chat (70B)]{\textbf{Heatmap of Llama2-chat (70B).}}
\label{fig:Heatmap_of_Llama2-chat_(70B)}
\end{figure}
Summary: The model presents an \textit{average scenario-prompt accuracy} of 28.6\%, attains an average \textit{prompt-average rank} of 16 out of 28, and holds an average robustness score of 47.8\%.

Accuracy: 1) Overall performance: Figure \ref{fig:Heatmap_of_Llama2-chat_(70B)}(a) reveals Llama2-chat (70B) has a score of 28.6\% in \textit{average scenario-prompt accuracy}, with a variability in effectiveness reflected by an average prompts' standard deviation of 12.8. The \textit{top scenario-prompt pair}s are a 3-shot IcL in PCD at 77.5\%, followed by a 1-shot IcL at 74.3\%, and EF at 71.9\% in the same scenario. Nearly 46.6\% of the \textit{scenario-prompt pairs} outperform the \textit{random guess accuracy}, yet none surpass 80\% accuracy.
2) Scenario performance: In scenarios where Llama2-chat (70B) surpasses the \textit{random guess accuracy}, the top 3 scenarios having the highest average accuracy are PCD with a score of 59.2\%, CR at 40.8\%, and CEG at 37.1\%.
3) Prompt efficiency: The top-performing prompts include a 3-shot IcL at 41.3\% and a 1-shot IcL at 37.2\%. Regarding the number of \textit{scenario-prompt pair}s where the model exceeds the \textit{random guess accuracy}, the 3-shot IcL leads in 18 out of 21 scenarios, followed by 1-shot IcL in 16, and manual CoT in 13 scenarios.
4) Language influence: English outshines Chinese in 17 of 21 scenarios, especially in CB, AC, and CDE, with \textit{language accuracy difference}s of 23.1\%, 17.7\%, and 17.4\%, respectively. However, Chinese outperforms English in scenarios like CEI, IV, and CEG, with 23.1\%, 2.3\%, and 1.9\% differences, respectively.

Ranking: 1) \textit{Prompt-average rank}: Figure \ref{fig:Prompt-Average_Rank_of_Models} shows Llama2-chat (70B)'s highest ranks in CEG at 7, NIE, and CDE both at 9. The model's poorest performances are observed with a 27th place in CEI, 24th in PN, and tied at 23rd for both BAS and IV. The average \textit{prompt-average rank} is 16 out of 28, with a standard deviation of 5.5.
2) \textit{Model-prompt rank}: As shown in Figure \ref{fig:Heatmap_of_Llama2-chat_(70B)}(b), the model's top \textit{model-prompt rank}s include NDE with manual CoT at 3, CB with manual CoT at 14, and AC with a 3-shot IcL at 15. The most significant challenges are in CA with adversarial ignore at 252, CEI, and AR, both with manual CoT at 251.

Robustness: Llama2-chat (70B) maintains an average robustness score of 47.8\% across scenarios, with peak robustness in FAS at 79.3\%, BAS at 67.6\%, and AC at 61.4\%.

\subsubsection{Lmsys}
\label{model:lmsys}
\paragraph{Vicuna-v1.3 (33B).}
\begin{figure}[t]
\centering
\subfigure[Performance of Vicuna-v1.3 (33B)]{
\begin{minipage}{8.5cm}
\centering
\includegraphics[width=1\linewidth]{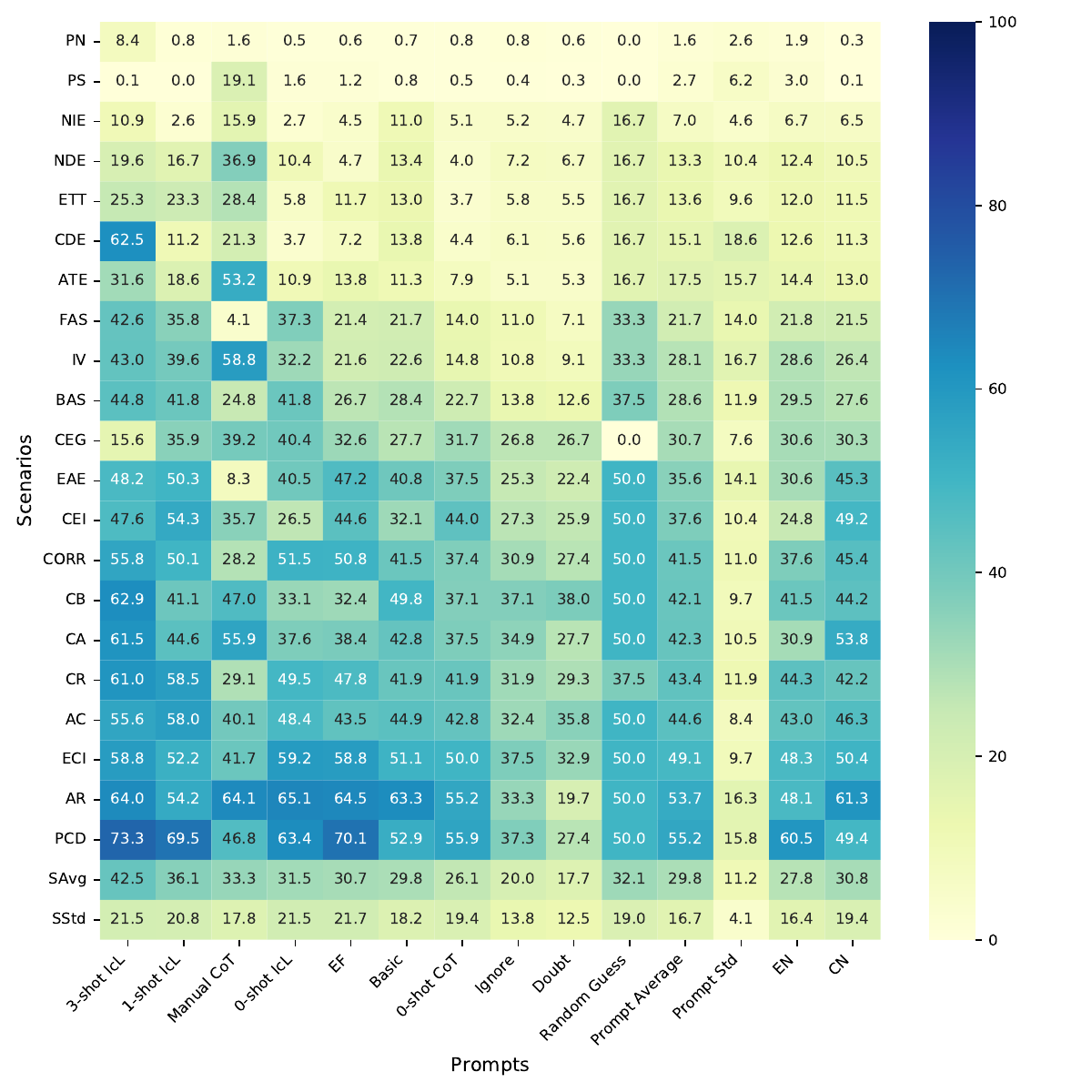}
\end{minipage}
}
\subfigure[\textit{Model-prompt rank} of Vicuna-v1.3 (33B)]{
\begin{minipage}{8.5cm}
\centering
\includegraphics[width=1\linewidth]{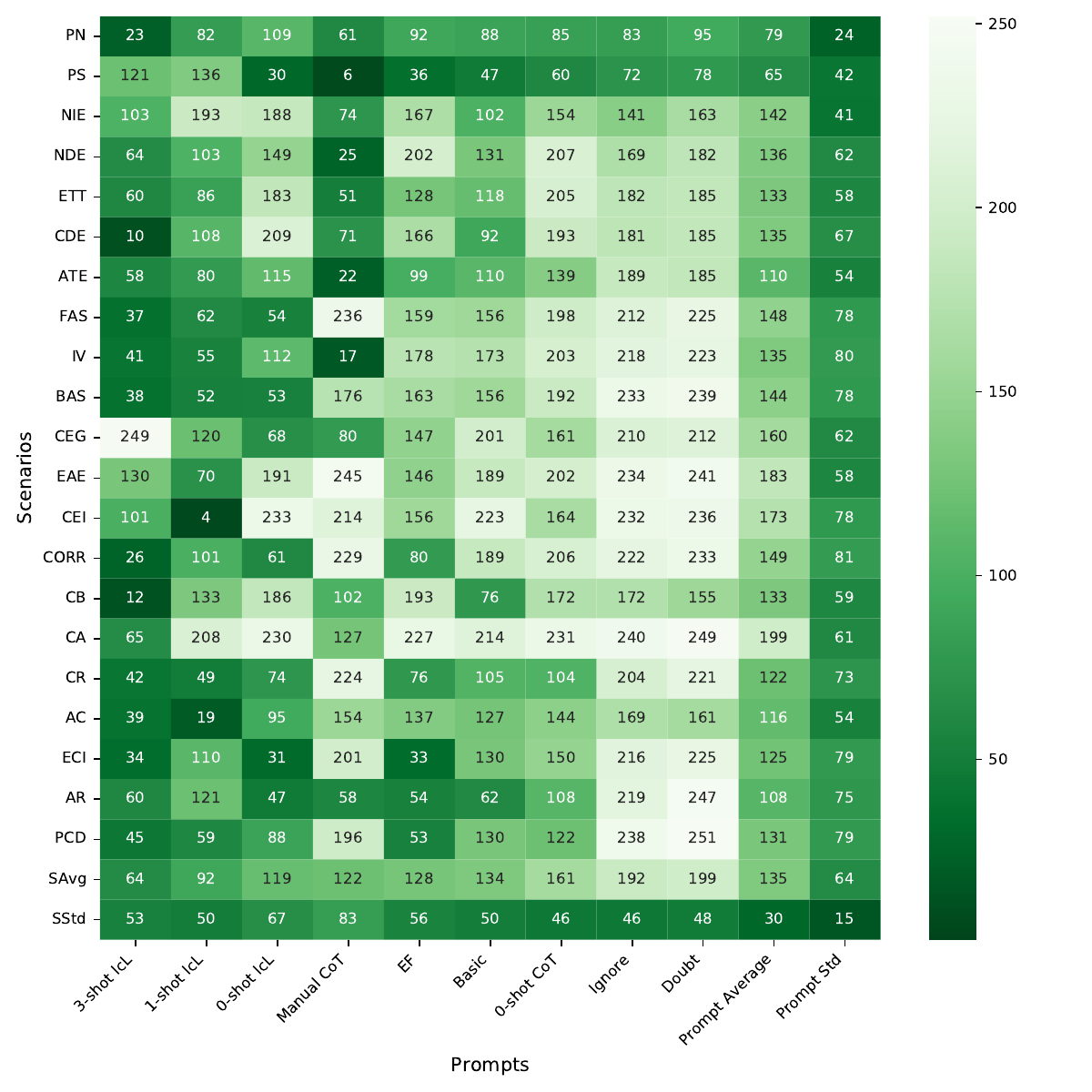}
\end{minipage}
}
\caption[Heatmap of Vicuna-v1.3 (33B)]{\textbf{Heatmap of Vicuna-v1.3 (33B).}}
\label{fig:Heatmap_of_Vicuna-v1.3_(33B)}
\end{figure}
Summary: The model demonstrates an \textit{average scenario-prompt accuracy} of 29.8\%, holds an average \textit{prompt-average rank} of 16 out of 28, and attains an average robustness score of 44.0\% across scenarios.

Accuracy: 1) Overall performance: As illustrated in Figure \ref{fig:Heatmap_of_Vicuna-v1.3_(33B)}(a), Vicuna-v1.3 (33B) achieves an \textit{average scenario-prompt accuracy} of 29.8\%, with considerable variability in prompt effectiveness, as indicated by an average standard deviation of 11.2. The \textit{top scenario-prompt pair}s are a 3-shot IcL in PCD with a score of 73.3\%, followed by EF at 70.1\%, and 1-shot IcL at 69.5\% in the same scenario. Around 43.9\% of the \textit{scenario-prompt pairs} perform better than the \textit{random guess accuracy}, though none surpass 80\% in accuracy.
2) Scenario performance: When selecting scenarios in which Vicuna-v1.3 (33B) exceeds the \textit{random guess accuracy}, and then identifying the top three based on their accuracy, the highest scoring scenarios are PCD at 55.2\%, AR at 53.7\%, and CR at 43.4\%.
3) Prompt efficiency: The leading prompts are 3-shot IcL at an average accuracy of 42.5\%, 1-shot IcL at 36.1\%, and manual CoT at 33.3\%. In situations where the model's accuracies on \textit{scenario-prompt pair}s surpass the \textit{random guess accuracy}, the 3-shot IcL takes the lead in 18 of the 21 scenarios. This is closely followed by the 1-shot IcL, which leads in 17 scenarios, with manual CoT and 0-shot IcL both trailing at 10 scenarios.
4) Language influence: English surpasses Chinese in 13 out of 21 scenarios, especially in PCD, PS, and IV, with \textit{language accuracy difference}s of 11.2\%, 2.8\%, and 2.2\%, respectively. Conversely, Chinese outshines English in scenarios like CEI, CA, and EAE, with significant differences of 24.4\%, 22.8\%, and 14.7\%, respectively.

Ranking: 1) \textit{Prompt-average rank}: Figure \ref{fig:Prompt-Average_Rank_of_Models} reveals Vicuna-v1.3 (33B)'s top \textit{prompt-average rank}s in PS at 7, ATE and CR both at 10. The model faces challenges in CA at 27, CEI at 25, and EAE at 25, pinpointing areas needing enhancement. The overall average rank across 21 scenarios is 16 out of 28, with a standard deviation of 5.6.
2) \textit{Model-prompt rank}: The best \textit{model-prompt rank}s for Vicuna-v1.3 (33B), as shown in Figure \ref{fig:Heatmap_of_Vicuna-v1.3_(33B)}(b), are achieved in CEI with a 1-shot IcL at 4, PS with manual CoT at 6, and CDE with 3-shot IcL at 10. The model's lowest ranks are in PCD with adversarial doubt at 251, CEG with a 3-shot IcL at 249, and CA with adversarial doubt at 249, highlighting significant challenges.

Robustness: Vicuna-v1.3 (33B) maintains an average robustness score of 44.0\% across different scenarios, with its strongest robustness in CEI at 76.1\%, CORR at 63.4\%, and PS at 56.7\%.

\subsubsection{UC Berkeley}
\label{model:ucb}
\paragraph{Koala (13B).}
\begin{figure}[t]
\centering
\subfigure[Performance of Koala (13B)]{
\begin{minipage}{8.5cm}
\centering
\includegraphics[width=1\linewidth]{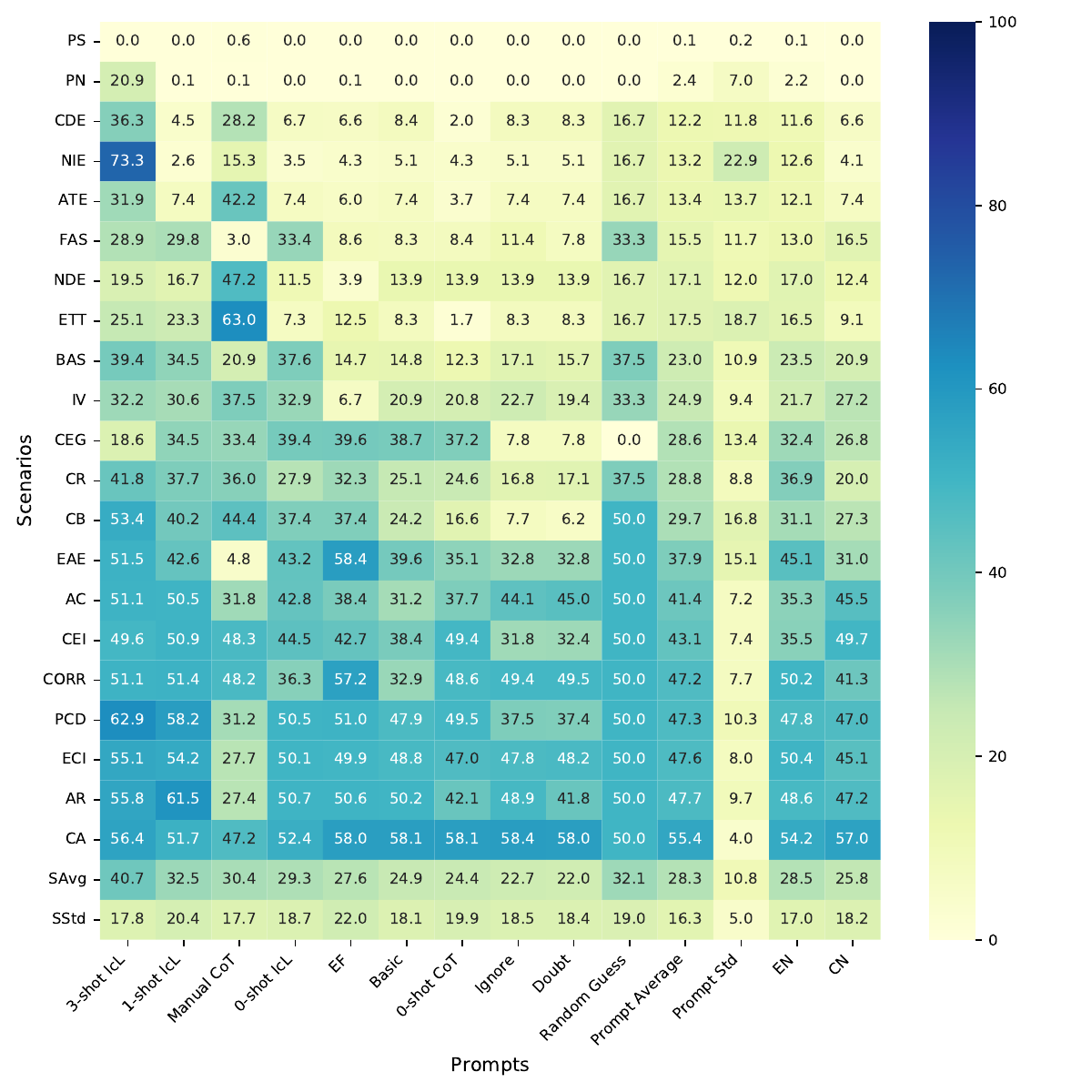}
\end{minipage}
}
\subfigure[\textit{Model-prompt rank} of Koala (13B)]{
\begin{minipage}{8.5cm}
\centering
\includegraphics[width=1\linewidth]{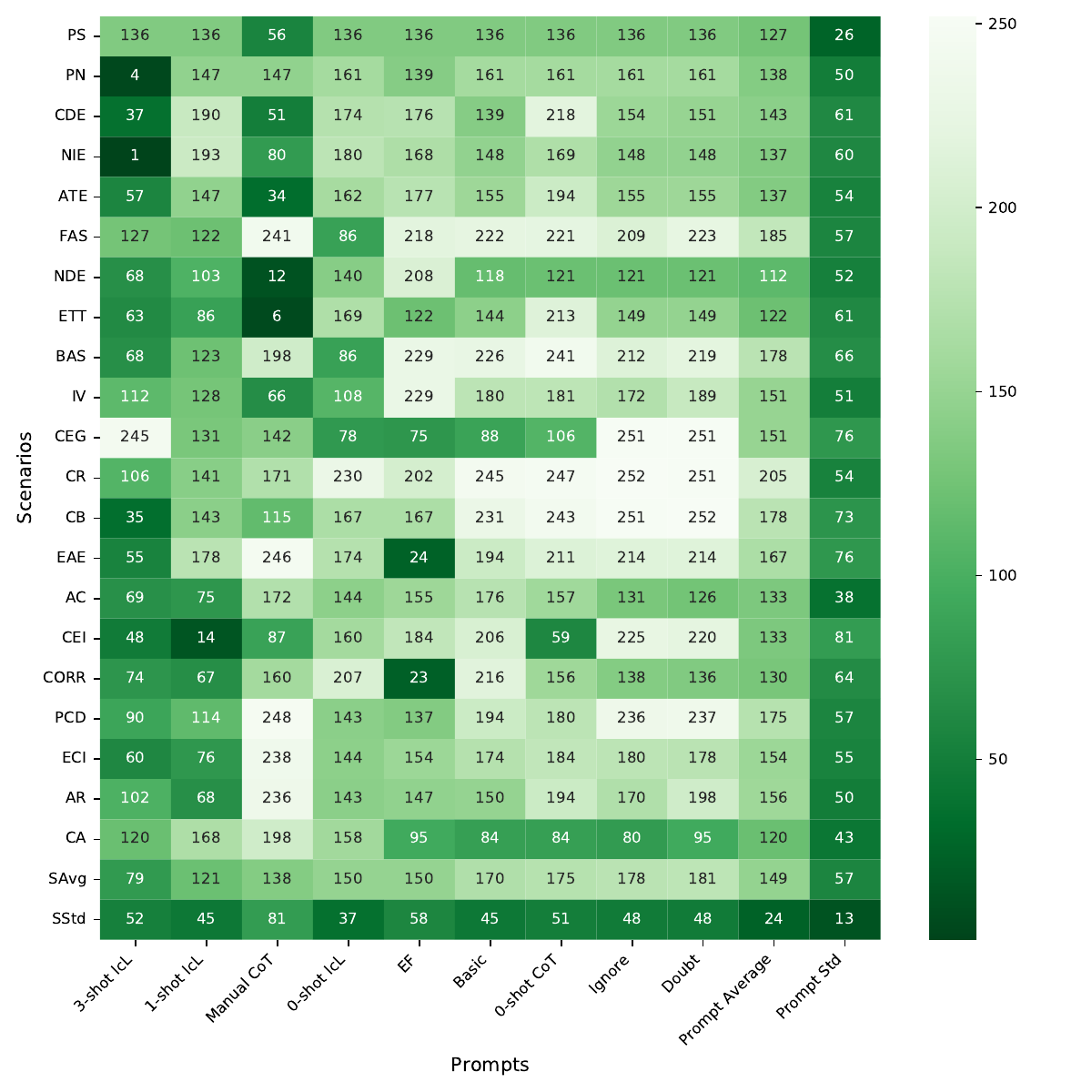}
\end{minipage}
}
\caption[Heatmap of Koala (13B)]{\textbf{Heatmap of Koala (13B).}}
\label{fig:Heatmap_of_Koala_(13B)}
\end{figure}
Summary: The model's \textit{average scenario-prompt accuracy} is 28.3\%, with an average \textit{prompt-average rank} of 19/28 and an average robustness score of 84.5\%.

Accuracy: 1) Overall performance: Illustrated by Figure \ref{fig:Heatmap_of_Koala_(13B)}(a), Koala (13B) achieves an \textit{average scenario-prompt accuracy} of 28.3\%, with an average standard deviation for prompt effectiveness at 10.8. The \textit{top scenario-prompt combination}s are 3-shot IcL in NIE with a score of 73.3\%, manual CoT in ETT at 63.0\%, and 3-shot IcL in PCD at 62.9\%. Only 38.6\% of the \textit{scenario-prompt pairs} surpass the \textit{random guess accuracy}, with none achieving over 80\% accuracy.
2) Scenario performance: Within scenarios outperforming the \textit{random guess accuracy}, the leading three in terms of average accuracy are CA at 55.4\%, CEG at 28.6\%, and ETT at 17.5\%.
3) Prompt efficiency: The top prompts by effectiveness are 3-shot IcL at 40.7\% and 1-shot IcL at 32.5\%. Regarding \textit{scenario-prompt pair}s exceeding the \textit{random guess accuracy}, 3-shot IcL is ahead in 18 of 21 scenarios, followed by 1-shot IcL in 13, and 0-shot IcL in 9.
4) Language influence: English outshines Chinese in 16 of 21 scenarios, particularly in CR, EAE, and CORR, with \textit{language accuracy difference}s of 16.9\%, 14.1\%, and 8.9\%, respectively. In contrast, Chinese excels in CEI, AC, and IV, with advantages of 14.2\%, 10.2\%, and 5.5\%, respectively.

Ranking: 1) \textit{Prompt-average rank}: As presented in Figure \ref{fig:Prompt-Average_Rank_of_Models}, Koala (13B)'s highest \textit{prompt-average rank}s are in NDE at 9, CA at 10, and ETT at 11. The lowest ranks are in CR at 28, CB at 27, and PS at 27, highlighting potential areas for development. The model's average rank across 21 scenarios is 19 out of 28, with a variability of 5.8.
2) \textit{Model-prompt rank}: Figure \ref{fig:Heatmap_of_Koala_(13B)}(b) shows Koala (13B)'s top \textit{model-prompt rank}s in NIE with 3-shot IcL at 1, PN with 3-shot IcL at 4, and ETT with manual CoT at 6. The lowest \textit{model-prompt rank}s occur in CR with adversarial ignore at 252, CB with adversarial doubt at 252, and CEG with adversarial ignore at 251.

Robustness: Koala (13B) maintains an impressive average robustness score of 84.5\% across different scenarios, showcasing the highest robustness in NIE at 99.7\%, ATE at 99.3\%, and ETT at 99.3\%.

\subsubsection{Microsoft}
\label{model:microsoft}
\paragraph{Wizardcoder (15B).}
\begin{figure}[t]
\centering
\subfigure[Performance of Wizardcoder (15B)]{
\begin{minipage}{8.5cm}
\centering
\includegraphics[width=1\linewidth]{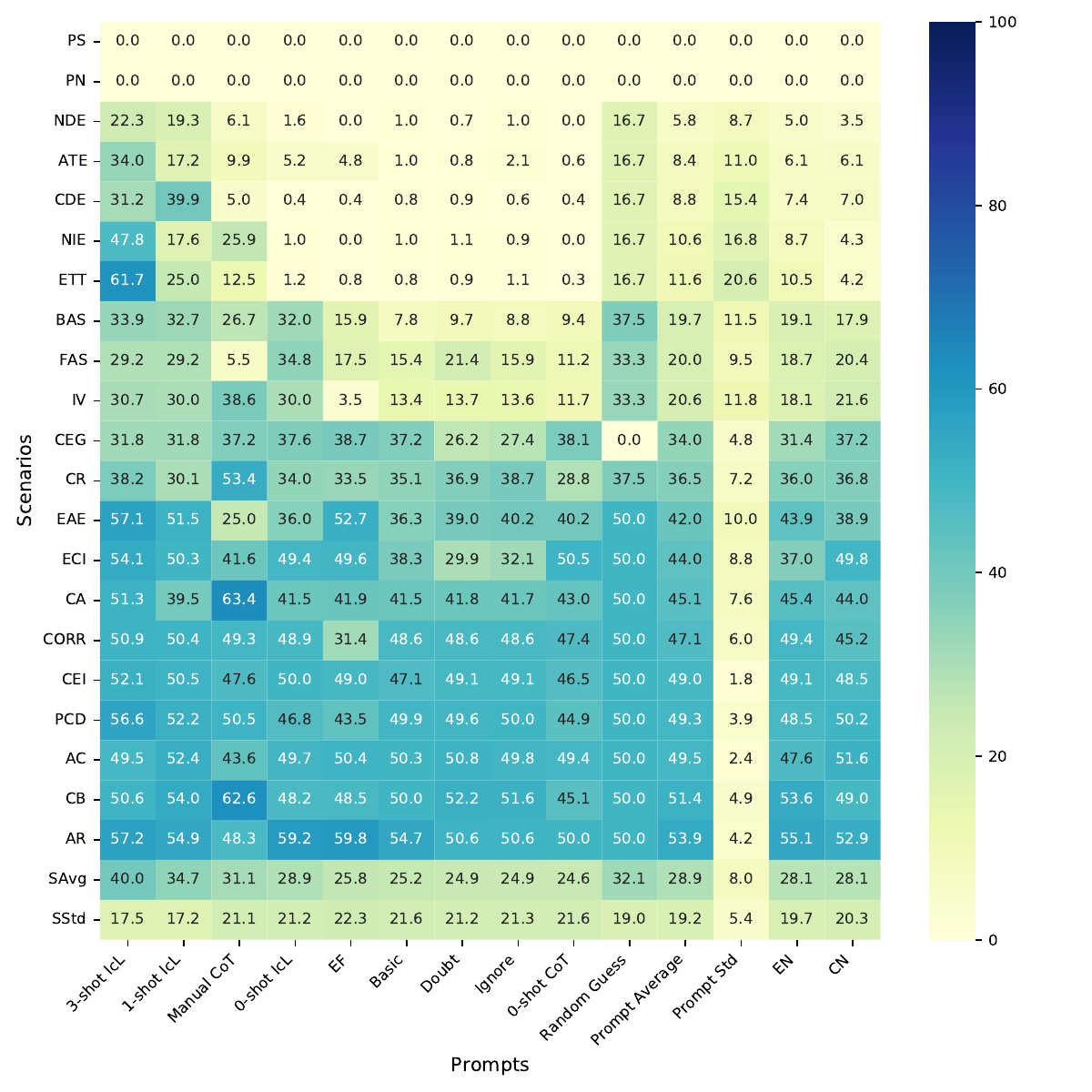}
\end{minipage}
}
\subfigure[\textit{Model-prompt rank} of Wizardcoder (15B)]{
\begin{minipage}{8.5cm}
\centering
\includegraphics[width=1\linewidth]{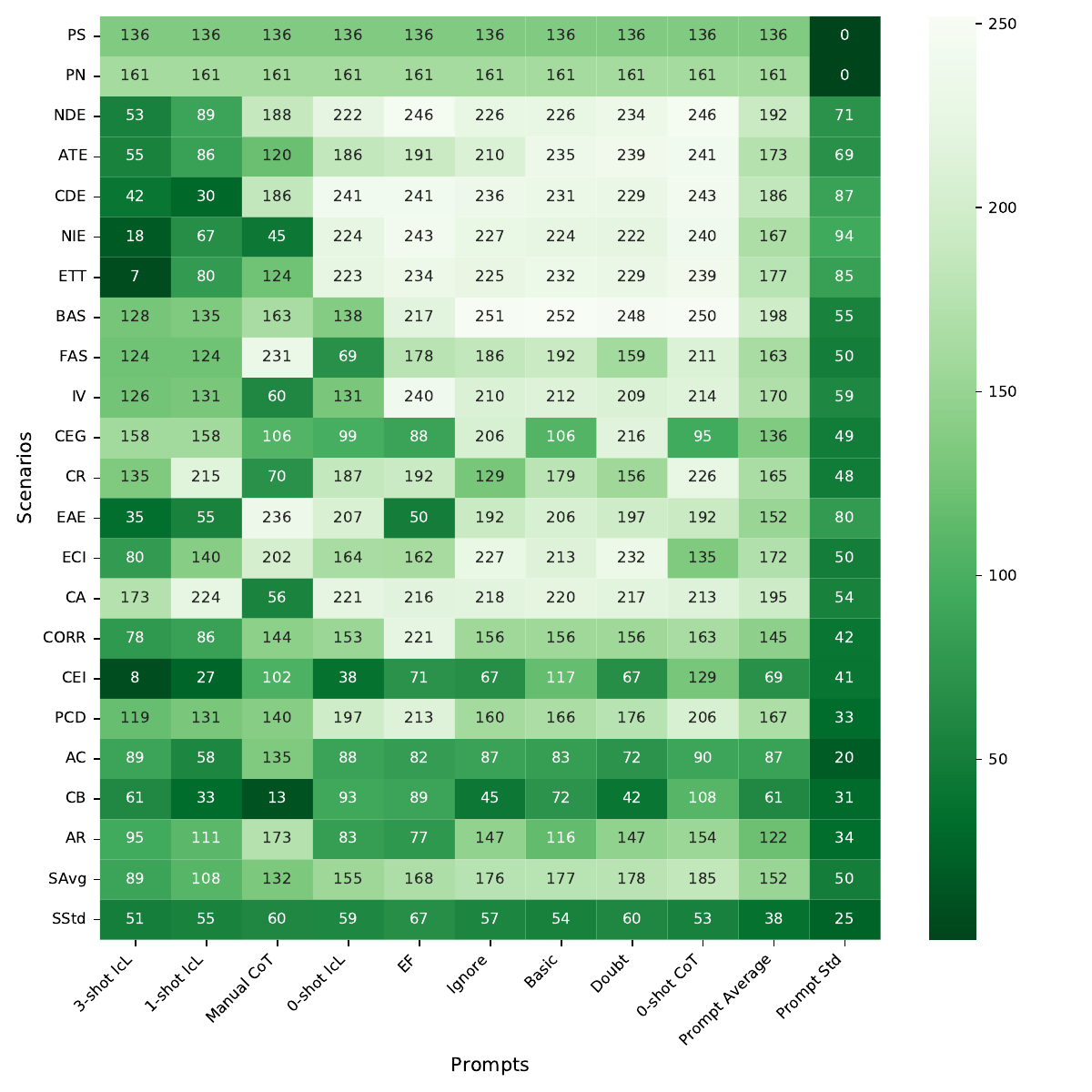}
\end{minipage}
}
\caption[Heatmap of Wizardcoder (15B)]{\textbf{Heatmap of Wizardcoder (15B).}}
\label{fig:Heatmap_of_Wizardcoder_(15B)}
\end{figure}
Summary: The model's \textit{average scenario-prompt accuracy} is 28.9\%, with an average \textit{prompt-average rank} of 19/28 and an average robustness score of 89.8\%.

Accuracy: 1) Overall performance: The performance analysis in Figure \ref{fig:Heatmap_of_Wizardcoder_(15B)}(a) reveals that Wizardcoder (15B) achieves an \textit{average scenario-prompt accuracy} of 28.9\%, with an average standard deviation of 8.0 for prompt effectiveness. Noteworthy performances of the \textit{top scenario-prompt pair}s include achieving a score of 63.4\% in manual CoT for CA, 62.6\% in manual CoT for CB, and 61.7\% in 3-shot IcL for ETT. Additionally, 41.3\% of the \textit{scenario-prompt pairs} outperform the baseline \textit{random guess accuracy}, though none surpass 80\% accuracy.
2) Scenario performance: For scenarios in which Wizardcoder (15B) outperforms the \textit{random guess accuracy}, the three leading scenarios by average accuracy are AR with a score of 53.9\%, followed by CB at 51.4\%, and CEG at 34.0\%.
3) Prompt efficiency: The most efficient prompts identified are 3-shot IcL with an effectiveness score of 40.0\% and 1-shot IcL at 34.7\%. Among 21 scenarios' \textit{scenario-prompt pair}s, the 3-shot IcL prompt leads in 17 for surpassing the \textit{random guess accuracy}, followed by 1-shot IcL in 16 scenarios, and manual CoT in 9.
4) Language influence:  In 11 out of 21 scenarios, English language prompts show superior performance over Chinese, particularly in scenarios like ETT, EAE, and CB, with \textit{language accuracy difference}s of 6.3\%, 5.0\%, and 4.6\%, respectively. Conversely, Chinese excels in ECI, CEG, and AC, with accuracy advantages of 12.8\%, 5.8\%, and 4.0\%, respectively.

Ranking: 1) \textit{Prompt-average rank}: Figure \ref{fig:Prompt-Average_Rank_of_Models} highlights that Wizardcoder (15B)'s best \textit{prompt-average rank}s are observed in CB and CEI (both ranked 4th), and AC (ranked 7th). However, the model shows room for improvement in PN, NDE, and BAS, with ranks of 28, 27, and 27, respectively. The average \textit{prompt-average rank} across 21 scenarios stands at 19 out of 28, with a standard deviation of 7.2.
2) \textit{Model-prompt rank}: As detailed in Figure \ref{fig:Heatmap_of_Wizardcoder_(15B)}(b), the model's top \textit{model-prompt rank}s in model-prompt combinations are found in ETT with 3-shot IcL (rank 7), CEI with 3-shot IcL (rank 8), and CB with manual CoT (rank 13). On the other hand, the lowest ranks are observed in BAS with basic at 252, BAS with adversarial ignore at 251, BAS with 0-shot CoT at 250.

Robustness: Wizardcoder (15B) exhibits a high average robustness score of 89.8\% across different scenarios, showcasing peak robustness in CA (97.4\%), CORR (96.9\%), and IV (96.0\%).


\subsection{Causal Scenario-specific Analysis}
\label{experiment:scenario}
This section is structured according to the levels of the causal ladder, analyzing specific causal scenarios within each rung. It is organized as follows: \nameref{scenario:discovery} (\cref{scenario:discovery}),  \nameref{scenario:association} (\cref{scenario:association}), \nameref{scenario:intervention} (\cref{scenario:intervention}), and \nameref{scenario:counterfactual} (\cref{scenario:counterfactual}). Prior to exploring each scenario, we introduce several additional terms vital for assessing model performance in causal scenarios and tasks, summarised in Tables \ref{tab:scenario_term-1} and \ref{tab:scenario_term-2}. These extend beyond the metrics previously discussed in Section \ref{metric:scenario}. Note that, for ease of reading, these terms will be presented in \emph{italic font} throughout the entire section. Additionally, causal scenarios can be categorized into two types based on the number of causal tasks they involve: \emph{single-task causal scenarios} and \emph{multi-task causal scenarios}. We outline the structure of evaluative paragraphs for each type of causal scenarios as below to enhance readability and comprehension. 

\begin{center}
\begin{table*}[t]
\caption[Explanations for scenario-specific terminologies]{\textbf{Explanations for scenario-specific terminologies.}}
\label{tab:scenario_term-1}
\begin{tabularx}{\textwidth}{c|X} 
\toprule
{\textbf{Terminology}} & \makecell[c]{\textbf{Explanation}} \\
\hline
  \multirow{1}{*}{\textit{model-prompt pair}} & A combination of a model and a prompt.  \\
\hline
\multirow{2}{*}{\textit{top model-prompt pair}} & The combination that has the top accuracy value across all tested \textit{model-prompt pair}s in a causal scenario/task. \\
\hline
\multirow{2}{*}{\textit{prompt gain}} & The accuracy of a model on a specific prompt minus the accuracy of the same model on the basic prompt. \\
\hline
\multirow{1}{*}{\textit{language proficiency}} & The model's accuracy in a specific language (English or Chinese). \\
\hline
  \multirow{17}{*}{\textit{average model-prompt-gain volatility}} & 
The \textit{model-prompt-gain volatility}, denoted by $\text{Volatility (Gain)}_{i}$, is calculated as the standard deviation of the model's performance gains when using a non-basic prompt (where ``non-basic'' refer to any prompt that is not the basic prompt) compared to its performance on the basic prompt. That is,
$$
\text{Volatility (Gain)}_{i} = \sqrt{\frac{\sum_{j=1}^{N-1}(G_{ij} - \bar{G_{i}})^2}{N-1}},
$$

where $G_{ij}$ represents the performance gain of the $j$-th non-basic prompt for the $i$-th model over its basic prompt performance, and $N$ is the total number of prompts. 
\newline
\newline The \textit{average model-prompt-gain volatility} (\textit{AMPGV}) is defined as the average of the \textit{model-prompt-gain volatility} across all models in a specific causal scenario/task, denoted mathematically as:  $$\textit{AMPGV}=\text{Mean}(\text{Volatility (Gain)}_{i}).$$ This measure represent the overall dependency of model accuracy improvements on different prompts within the given scenario or task.
\\
\hline
\end{tabularx}
\end{table*}
\vspace{-25pt}
\end{center}

\begin{center}
\begin{table*}[t]
\caption[Explanations for scenario-specific terminologies (continued)]{\textbf{Explanations for scenario-specific terminologies (continued).}}
\label{tab:scenario_term-2}
\begin{tabularx}{\textwidth}{c|X} 
\toprule
{\textbf{Terminology}} & \makecell[c]{\textbf{Explanation}} \\
\hline
\multirow{2}{*}{\textit{prompt dependence}} & As defined in Table \ref{tab:Degree of Prompt Dependence}
, we analyze the \textit{prompt dependence} based on \textit{AMPGV}. 
\\
\hline
\multirow{7}{*}{\textit{variance of distribution}} & The variance of distribution assesses the diversity in performance distributions of \textit{model-prompt pair}s across causal tasks in a causal scenario. This metric, critical for evaluating scenarios comprising multiple causal tasks, calculates the standard deviations of the tasks' median and third quartile accuracies. These values are then compared against a predefined set to ascertain the corresponding degree of variance, as defined in Table \ref{tab:Degree of Variance of causal task Distributions in the causal scenario}.\\
\hline
\multirow{12}{*}{\textit{variance of solvability}} & This measure evaluates the variance in \textit{solvability} across causal tasks in a multiple-task causal scenario. It assigns numerical values to each degree of \textit{solvability} as defined in Table \ref{tab:Degree of Solvable}: 4 for \textbf{unsolvable}, 3 for \textbf{challenging}, 2 for \textbf{potentially solvable}, 1 for \textbf{solvable}, and 0 for \textbf{well-solved}. The \textit{solvability gap} is then calculated by subtracting the minimum \textit{solvability number} from the maximum \textit{solvability number} among the causal tasks within a scenario. For example, if a scenario comprises three causal tasks with \textit{solvabilities} of \textbf{challenging} (3), \textbf{solvable} (2), and \textbf{well-solved} (0), the \textit{solvability gap} in this causal scenario is 3. Then, according to Table \ref{tab:Degree of Variance of the Solvable of causal tasks in the causal scenario.}, a gap of 3 indicates the \textit{variance of solvability} of the causal scenario is \textbf{large}.\\
\hline
\multirow{7}{*}{\textit{variance of model's top performance}} & The \textit{variance of model's top performance} requires first computing the maximum and minimum values of the highest average accuracies of models across tasks, along with the maximum and minimum values of the \textit{top model-prompt pair} accuracies. Then, the two gaps are attained by subtracting the minimum from the maximum, respectively. Finally, we derive the variance based on the two gaps and the rules from \cref{tab:Degree of Variance of Model's Top Performance in the causal scenario.}.\\
\hline
\multirow{4}{*}{\textit{variance of prompt dependence}} & 
Like \textit{prompt dependence}, the \textit{variance of prompt dependence} is also based on the \textit{AMPGV}. This metric calculates the difference between the maximum and minimum AMPGV values across causal tasks within a scenario, as defined in \cref{tab:Degree of Variance of the Dependence on Prompts}.\\
\hline
\end{tabularx}
\end{table*}
\vspace{-20pt}
\end{center}

\begin{table}[t]
    \centering
    \begin{tabular}{c|c}
        \toprule
        \textbf{Conditions} & \textbf{Degree of \textit{prompt dependence}} \\
        \midrule
        \textit{AMPGV} $<$ 5 & low \\
        5 $\leq$ \textit{AMPGV}  $<$ 10 & medium \\
        \textit{AMPGV} $\geq$ 10 & high\\
        \bottomrule
    \end{tabular}
    \caption[Degree of \textit{prompt dependence}]{\textbf{Degree of \textit{prompt dependence}.} The \textit{AMPGV} stands for the \textit{average model-prompt-gain volatility} in the causal task/causal scenario.}
    \label{tab:Degree of Prompt Dependence}
\end{table}

\begin{table}[t]
    \centering
    \begin{tabular}{c|c}
        \toprule
        \textbf{Conditions} & \textbf{\textit{Variance of distribution}}\\
        \midrule
        both stds in [0,5) & minimally divergent \\
        one of the stds in [0,5) and the other one in [5,10) & slightly variable \\
        both stds in [5,10) & moderately distinct\\
        one of the stds in [0,5) and the other one in [10, $\infty$) & considerably varied\\
        one of the stds in [5,10) and the other one in [10, $\infty$) & noticeably diverse\\
        both stds in [10, $\infty$) & highly divergent\\
        \bottomrule
    \end{tabular}
    \caption[\textit{Variance of distribution}s in the causal scenario]{\textbf{\textit{Variance of distribution}s in the causal scenario.} To evaluate the diversity of causal tasks in multi-task causal scenarios, the evaluation considers two standard deviations (std): the std of medians and the std of third quarters. Like Table \ref{tab:Degree of Understanding}, the median and the third quartile are computed from the distribution of all \emph{model-prompt pairs} in the causal task.}
    \label{tab:Degree of Variance of causal task Distributions in the causal scenario}
\end{table}

\begin{table}[t]
    \centering
    \begin{tabular}{c|c}
        \toprule
        \textbf{Conditions} & \textbf{\textit{Variance of solvability}}\\
        \midrule
        solvability gap=0 & negligible \\
        solvability gap=1 & small \\
        solvability gap=2 & moderate\\
        solvability gap=3 & large\\
        solvability gap=4 & extremely large\\
        \bottomrule
    \end{tabular}
    \caption[\textit{Variance of solvability} of causal tasks in the causal scenario]{\textbf{\textit{Variance of solvability} of causal tasks in the causal scenario.} The solvability gap is then calculated by subtracting the minimum \emph{solvability number} from the maximum \emph{solvability number} among the causal tasks within a scenario, where the \emph{solvability number} is defined in Table \ref{tab:Degree of Solvable}.}
    \label{tab:Degree of Variance of the Solvable of causal tasks in the causal scenario.}
\end{table}

\begin{table}[t]
    \centering
    \begin{tabular}{c|c}
        \toprule
        \textbf{Conditions} & \textbf{\textit{Variance of model's top performance}}\\
        \midrule
        both gaps in [0,5) & small \\
        one of the gaps in [0,5) and the other one in [5,10) & moderate \\
        both gaps in [5,10) & noticeable\\
        one of the gaps in [0,5) and the other one in [10, $\infty$) & considerable\\
        one of the gaps in [5,10) and the other one in [10, $\infty$) & significant\\
        both gaps in [10, $\infty$) & extremely significant\\
        \bottomrule
    \end{tabular}
    \caption[\textit{Variance of model's top performance} in the causal scenario]{\textbf{\textit{Variance of model's top performance} in the causal scenario.} The two gaps here stand for the gap of max average value (i.e., the gap between the maximum value among the highest average accuracies of models across tasks and the minimum value among the highest average accuracies of models across tasks) and the gap of max value (i.e., the gap between the maximum value among the highest accuracies of model and prompt pairs across tasks and the minimum value among the highest accuracies of model and prompt pairs across tasks).}
    \label{tab:Degree of Variance of Model's Top Performance in the causal scenario.}
\end{table}

\begin{table}[t]
    \centering
    \begin{tabular}{c|c}
        \toprule
       \textbf{Conditions} & \textbf{\textit{Variance of prompt dependence}}\\
        \midrule
        gap of \textit{AMPGV} in [0,5)& narrow \\
        gap of \textit{AMPGV} in [5,10) & moderate spread \\
        gap of \textit{AMPGV} in [10,$\infty$) & wide\\
        \bottomrule
    \end{tabular}
    \caption[\textit{Variance of prompt dependence}]{\textbf{\textit{Variance of prompt dependence}.} The gap refers to the difference between the maximum and minimum values of the \textit{average model-prompt-gain volatility} (\textit{AMPGV}).}
    \label{tab:Degree of Variance of the Dependence on Prompts}
\end{table}

\paragraph{Single-task causal scenario.}
For evaluating causal scenarios that contain only one causal task (e.g., AR, EAE), we assess the scenario based on model performance, \emph{prompt gain}, and \emph{language proficiency}, respectively. 

For model performance, we first analyze the distribution of all \textit{model-prompt pair}s within the scenario, and then compute the median and third quartile accuracies. These accuracy metrics provide insights into the \textit{understandability} of the task, as outlined in Table \ref{tab:Degree of Understanding}. Note that, in the CEG scenario - characterized by a negligible random guess value - the task is determined to be \textbf{easy} based on its performance and the inherent simplicity of natural language questions. Subsequently, we evaluate the top accuracies within the scenario, including the top three models ranked by average accuracy and the \textit{top model-prompt pair}. Using these accuracies, we determine the degree of \emph{solvability} for the scenario according to Table \ref{tab:Degree of Solvable}. We also examine the stability of the models by computing the \emph{model volatility} as introduced in Section \ref{metric:model}, identifying the top three most stable and most unstable models based on this metric. Finally, we consider the ratio of open-access to limited-access models among the top five models ranked by average accuracy. The disparity between open and limited-access models referred to as the \emph{open-limited gap}, is evaluated based on the criteria set forth in Table \ref{tab:Degree of Open-Limited Gap}.

For \emph{prompt gain}, we begin by evaluating the top \emph{prompt gains} within the scenario. We analyze the two leading prompts based on average accuracy gain relative to the basic prompt and identify the \emph{model-prompt pair} with the highest gain over the basic prompt. 
We then address potential exceptions, including instances a) where the top-performing prompt (with the highest average accuracy) may actually decrease the average accuracy for some models compared to their performance with the basic prompt; and b) where all prompts - or conversely, no prompts - result in performance enhancements for certain models over the basic prompt.
Next, we analyze the prompt stability by computing the \emph{prompt volatility}, as introduced in Section \ref{metric:prompt}. Finally, we compute the \textit{average model-prompt-gain volatility} (\textit{AMPGV}), and report the scenario's \textit{prompt dependence}. These measures help us understand the critical role that effective prompt design plays in enhancing model performance across specific scenarios.

For \textit{language proficiency}, we compare model accuracy in English and Chinese, and quantify the proportion of models that exhibit higher average accuracy in English. Additionally, we investigate significant performance disparities across the two languages, identifying models with notable differences in accuracy between English and Chinese.

\paragraph{Multi-task causal scenario.}
For evaluating causal scenarios that contain multiple causal tasks (e.g., PCD, ECI), our analysis is also structured from the three primary perspectives: model performance, \emph{prompt gain}, and \emph{language proficiency}. However, the focus within each is tailored to assess the specific complexities associated with multiple tasks. 

For model performance, it contains:
\begin{itemize}
    \item \textbf{Distribution}: We first consider the distribution of all \textit{model-prompt pair}s in the scenario and compute its median and third quartile accuracy. Based on these accuracies, we evaluate the \textit{understandability} of the scenarios as defined in Table \ref{tab:Degree of Understanding}.\footnote{It is important to note that for \textit{understandability} of PN and PS, which have nearly zero random guess values, evaluation methods in Table \ref{tab:Degree of Understanding} based on random guesses prove ineffective. As such, we manually designate their \textit{understandability} as \textbf{very hard}. We have explained the reason why we give such a designation in Section \ref{metric:scenario}.} Then, we conduct a task-specific analysis of the distribution of all \textit{model-prompt pair}s, computing the median and third quartile accuracies. These accuracy metrics offer insights into the \textit{understandability} of each task.\footnote{We compute the \textit{understandability} for most tasks according to Table \ref{tab:Degree of Understanding}. However, for tasks involving probability computations, such as ATE-P\_ATE-basic\_CN and NIE-P\_NIE-basic\_EN, we classify them as \textbf{very hard} to understand due to their complexity and consistently low scores in the median and third quartile.} Finally, we explore the differences between causal tasks by analyzing the range and the standard deviation of their medians and third quartile values, allowing us to draw conclusions about the \textit{variance of distribution}. Additionally, we may discuss other scenario-specific findings that emerge from the analysis.
    
    \item \textbf{Top Accuracy}: We first discuss the leading models that demonstrate the highest average accuracy within the scenario, followed by presenting the \textit{top model-prompt pair}. Based on these accuracy scores, we evaluate the \textit{solvability} of the scenario as defined in Table \ref{tab:Degree of Solvable}. Subsequently, our analysis proceeds on a per-task basis, focusing on the models with the highest average accuracy, the \textit{top model-prompt pair}, and the \textit{solvability} of causal tasks. Lastly, we extend our analysis across causal tasks, drawing conclusions on the \textit{variance of solvability}, the \textit{variance of model's top performance}, and other scenario-specific findings.
    
    \item \textbf{Stability}: Initially, we discuss the stability of models in the scenario by listing the three most stable and most unstable models, characterized by their lowest and highest \textit{model volatility} scores, respectively.\footnote{These \textit{model volatility} scores are computed as described in Section \ref{metric:model}.} Then, we conduct a task-by-task evaluation, identifying the three most stable and most unstable models for each task. Conclusions are drawn based on these findings.
    
    \item \textbf{Open-Limited Ratio}: We analyze the ratio of open-access to limited-access models among the top five models with the highest average accuracy in the scenario. Then, we quantify the \textit{open-limited gap} using the metrics outlined in Table \ref{tab:Degree of Open-Limited Gap}.
\end{itemize}
For \textit{prompt gain}, we discuss the following aspects:
\begin{itemize}
    \item \textbf{Top Gain}: First, we examine the two most effective prompts in terms of average accuracy gain over the basic prompt in the scenario. We then analyze the highest accuracy gain over all the \textit{model-prompt pair}s in the scenario. Following this, we conduct a task-by-task analysis of these perspectives. Finally, we provide a summary of our analysis.
    
    \item \textbf{Exceptions}: We begin our evaluation from a scenario-wide perspective, then proceed to analyze each task individually. We consider the following exceptions: a) whether the most effective prompt, which has the highest average prompt gain over the basic prompt, fails in some models in the scenario; b) whether there exists a model for which all prompts either consistently boost performance or fail to improve performance over the basic prompt.
    
    \item \textbf{Stability}: First, we identify the two most stable and two most unstable prompts, categorized by their lowest and highest \textit{prompt volatility} as defined in Section \ref{metric:prompt}, within the scenario. Subsequently, we compute the \textit{average model-prompt-gain volatility} (\textit{AMPGV}) and classify the scenario's \textit{prompt dependence} using the criteria specified in Table \ref{tab:Degree of Prompt Dependence}. We then conduct a task-by-task assessment of \textit{prompt stability}, \textit{AMPGV}, and \textit{prompt dependence}. Finally, we explore the distribution of \textit{AMPGV} across causal tasks to determine the \textit{variance of prompt dependence} in the scenario, which may inform further conclusions based on these evaluations.
\end{itemize}
Finally, we analyze \textit{language proficiency} by examining:
\begin{itemize}
    \item \textbf{English vs. Chinese}: We evaluate whether the scenario yields better performance in English or Chinese and quantify the proportion of models that demonstrate superior results in English compared to Chinese.
    
    \item \textbf{Accuracy Difference}: We highlight the most significant disparities in model performance between English and Chinese, detailing the top differences in accuracy.
\end{itemize}

\subsubsection{Causal Discovery}
\label{scenario:discovery}
\begin{figure}[t]
\centering  
\subfigure[Distribution of PCD]{  
\begin{minipage}{3.9cm}
\centering    
    \includegraphics[width=1\linewidth]{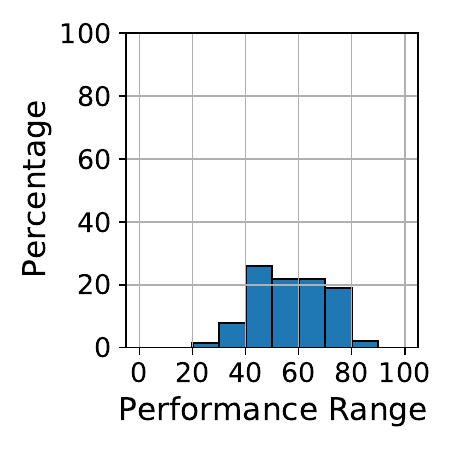}
\end{minipage}
}
\subfigure[Distribution of ECI]{  
\begin{minipage}{3.9cm}
\centering   
    \includegraphics[width=1\linewidth]{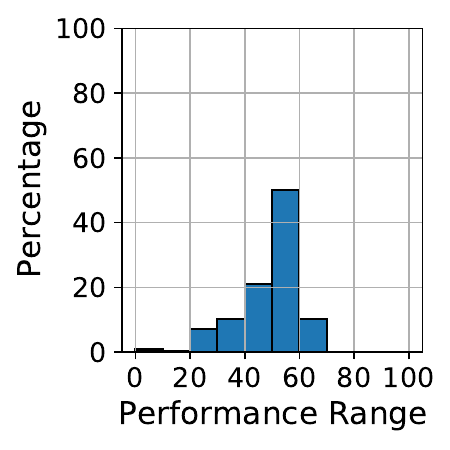}
\end{minipage}
}
\subfigure[Distribution of AR]{  
\begin{minipage}{3.9cm}
\centering    
    \includegraphics[width=1\linewidth]{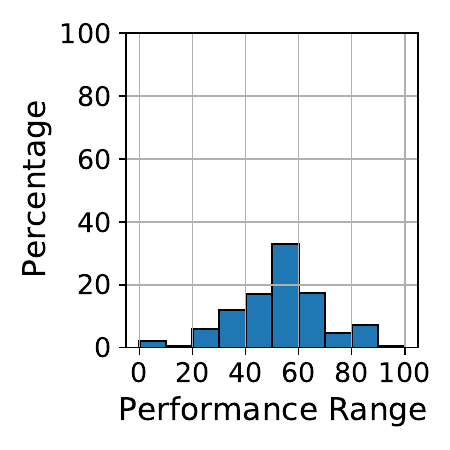}
\end{minipage}
}
\subfigure[Distribution of CA]{ 
\begin{minipage}{3.9cm}
\centering   
    \includegraphics[width=1\linewidth]{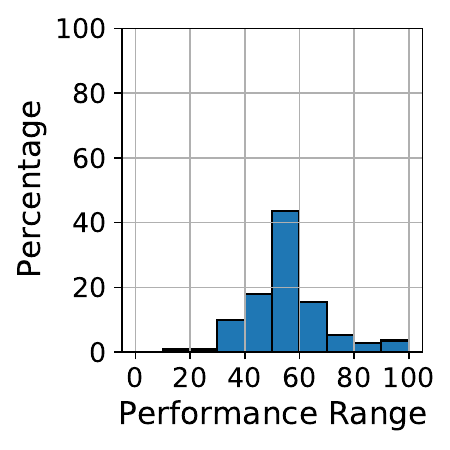}
\end{minipage}
}
\caption[Distribution of causal discovery]{\textbf{Distribution of causal discovery.} The horizontal coordinate represents the accuracy of the model and the vertical coordinate represents the percentage distribution corresponding to a certain accuracy interval.}  
\label{fig:Distribution_of_Causal_Discovery}    
\end{figure}

\paragraph{Pairwise causal discovery.}
\begin{figure}[t]
\centering
\subfigure[Model performance of PCD]{
\begin{minipage}{8.5cm}
\centering
\includegraphics[width=1\linewidth]{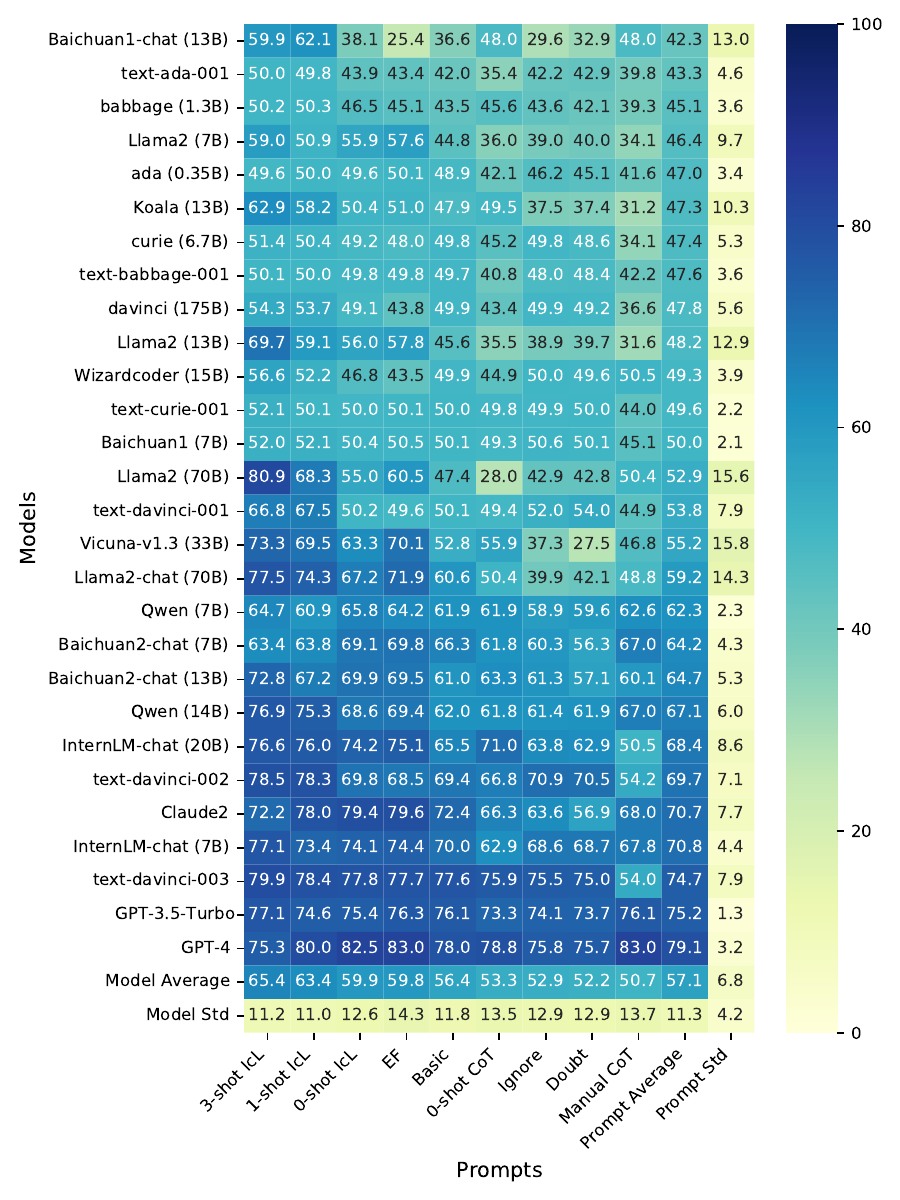}
\end{minipage}
}
\subfigure[\textit{Prompt gain} of PCD]{
\begin{minipage}{8.5cm}
\centering
\includegraphics[width=1\linewidth]{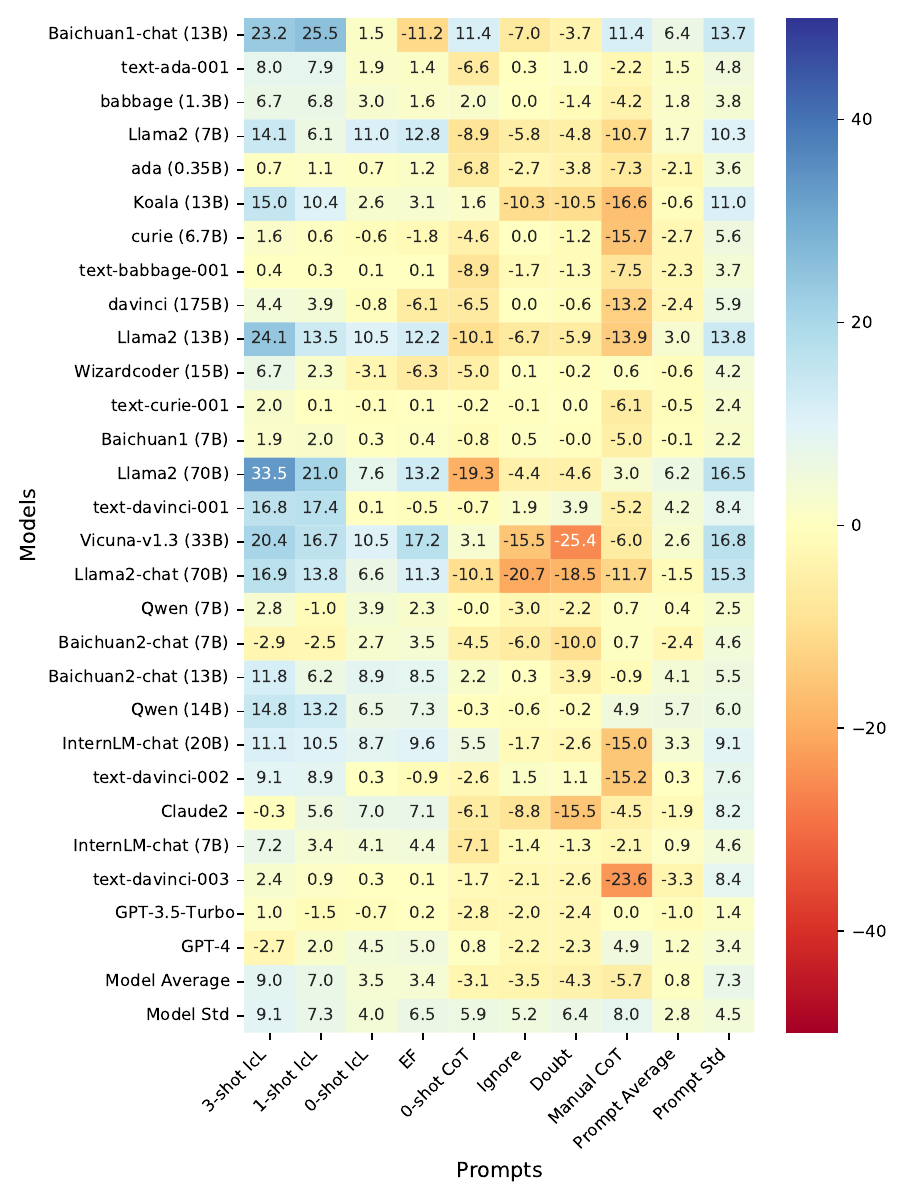}
\end{minipage}
}
\caption[Heatmap of PCD]{\textbf{Heatmap of PCD.} The models and prompts are sorted by their averages.}
\label{fig:Heatmap_of_Pairwise_Causal_Discovery}
\end{figure}

\begin{figure}
    \centering
    \includegraphics[width=0.8\linewidth]{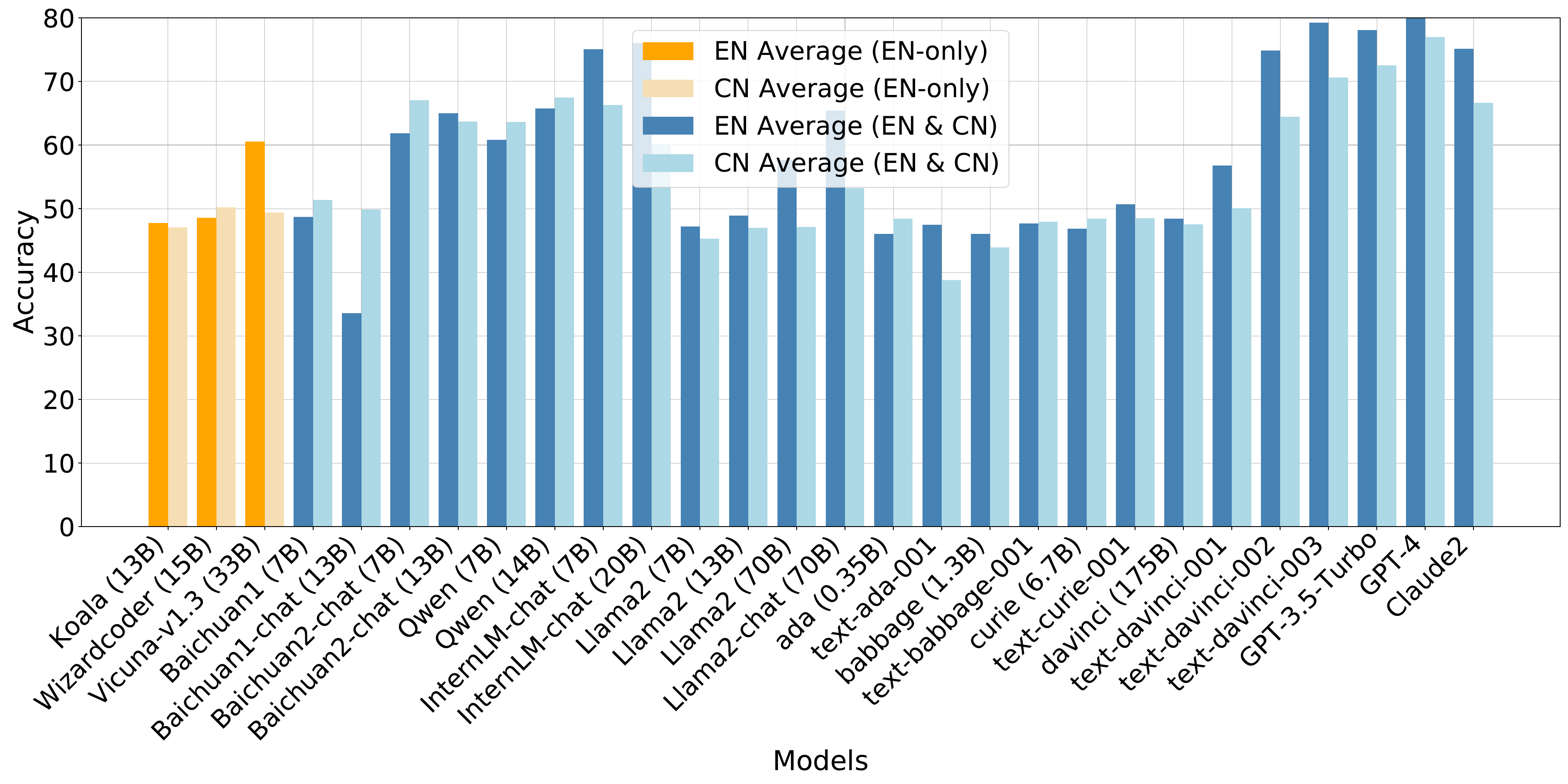}
    \caption[Language comparison of PCD]{\textbf{Language comparison of PCD.} The dark legend signifies the average performance of the model on an English test set, whereas the light legend denotes the average performance of the model on the Chinese test set. The yellow legend indicates a model trained exclusively on English datasets, while the blue legend represents a model trained on both English and Chinese datasets.}
    \label{fig:Pairwise_Causal_Discovery_Language}
\end{figure}

First, we analyze model performance in PCD:

1) \textbf{Distribution}: Figure \ref{fig:Distribution_of_Causal_Discovery}(a) illustrates the distribution of \textit{model-prompt pair}s within PCD, highlighting a median accuracy of 54.1\% and a third quartile of 68.8\%. This suggests the \textit{understandability} of the scenario is \textbf{easy}, as the median accuracy surpasses the baseline random guess accuracy of 50.0\%. Additionally, Figure \ref{fig:Distribution_of_Pairwise_Causal_Discovery_Tasks} details the distribution of \textit{model-prompt pair}s across individual tasks, revealing median accuracies of 50.5\% for \textbf{PCD-B (E-CARE)}, 51.3\% for \textbf{PCD-B (COPA)}, 56.6\% for \textbf{PCD-C (E-CARE)}, and 60.3\% for \textbf{PCD-C (COPA)}, alongside third quartiles of 58.0\% for \textbf{PCD-B (E-CARE)}, 62.7\% for \textbf{PCD-B (COPA)}, 72.7\% for \textbf{PCD-C (E-CARE)}, and 84.6\% for \textbf{PCD-C (COPA)}. As the random accuracy of each task is 50\%, these tasks all have an easy \textit{understandability}. \textbf{By analyzing the differences between tasks}, these tasks show a median accuracy range from 50.5\% to 60.3\% with a standard deviation of 4.0. As to the third quartile accuracy, the range is from 58.0\% to 84.6\% with a standard deviation of 10.2. As a result, the scenario has a \textbf{considerably varied} \textit{variance of distribution}. Moreover, the choice selection tasks in the scenario are easier to understand than the binary classification tasks, and COPA is more easy than E-CARE dataset.

2) \textbf{Top Accuracy}: Figure \ref{fig:Heatmap_of_Pairwise_Causal_Discovery}(a) reveals that in terms of average accuracy, the leading three models in this scenario are GPT-4 with 79.1\%, GPT-3.5-Turbo with 75.2\%, and text-davinci-003 with 74.7\%. The \textit{top model-prompt pair} is GPT-4 with EF, achieving an accuracy of 83.0\%. The \textit{solvability} of the scenario is \textbf{well-solved} as the average accuracies of the top three models all exceed 70\%. Figure \ref{fig:Heatmap_of_performances_of_Pairwise_Causal_Discovery} outlines the top performers on a per-task basis. In the \textbf{PCD-B (E-CARE)} task, GPT-4 leads with 69.6\%, followed by GPT-3.5-Turbo at 64.2\%, and text-davinci-003 at 64.1\%, with GPT-4 and manual CoT reaching the highest accuracy of 73.4\%. This result suggests the task's \textit{solvability} is \textbf{challenging}, as the \textit{top model-prompt pair} accuracy falls below 80\%. For the \textbf{PCD-B (COPA)} task, the best averages are by GPT-4 at 78.8\%, text-davinci-003 at 70.9\%, and GPT-3.5-Turbo at 65.5\%. The peak accuracy appears in GPT-4 combined with manual CoT leading at 82.0\%. The \textit{solvability} of the task is \textbf{solvable} as the top-1 model's average accuracy reaches 70\%. In the \textbf{PCD-C (E-CARE)} category, GPT-3.5-Turbo tops with 79.1\%, followed by GPT-4 at 76.6\%, and text-davinci-003 at 76.2\%, with GPT-4 using EF reaching the highest at 83.2\%. The task's \textit{solvability} is \textbf{well-solved} as all 3 leading models exceed 70\% in average accuracy. Lastly, for \textbf{PCD-C (COPA)}, GPT-3.5-Turbo achieves 91.9\%, GPT-4 scores 91.5\%, and InternLM-chat (20B) achieves 90.1\%, with GPT-4 and EF reaching the peak accuracy of 98.0\%. The task also has a \textbf{well-solved} \textit{solvability} with all top models achieving over 70\% in average accuracy. 
\textbf{Through comparing different tasks}, the \textit{variance of solvability} between the tasks is \textbf{large}. Moreover, the top-1 model's average accuracy ranges from 69.6\% to 91.9\% (difference of 22.3\%), and the \textit{top model-prompt pair}'s accuracy ranges from 73.4\% to 98.0\% (difference of 24.6\%). Therefore, the scenario's \textit{variance of model's top performance} is \textbf{extremely significant}. Also, we can conclude that for binary classification tasks, the top average model is GPT-4, while for choice selection tasks, the top average model is GPT-3.5-Turbo. On the other hand, all of the tasks' \textit{top model-prompt pair}s contain GPT-4, indicating its great potential. The top 1 average accuracy and accuracy of \textit{top model-prompt pair} in the four tasks satisfy a similar regularity to the one in the model's distribution, which is the choice selection tasks are easier to solve than the binary classification tasks, and COPA dataset is easier than E-CARE dataset.

3) \textbf{Stability}: The most stable models, characterized by the lowest \textit{model volatility}, are GPT-3.5-Turbo (1.3), Baichuan1 (7B) (2.1), and text-curie-001 (2.2). The models displaying the greatest sensitivity to different prompts, evidenced by their high \textit{model volatility}, are Vicuna-v1.3 (33B) (15.8), Llama2 (70B) (15.6), and Llama2-chat (70B) (14.3). 
Next, we analyze the stability of the model task by task. For \textbf{PCD-B (E-CARE)}, the models with the least \textit{model volatility} are GPT-4 (2.2), Qwen (7B) (2.8), and Baichuan2-chat (7B) (3.0), while the most sensitive models are InternLM-chat (20B) (13.8), Vicuna-v1.3 (33B) (13.0), and Llama2 (70B) (11.8). 
For \textbf{PCD-B (COPA)}, the models showing the least sensitivity are Wizardcoder (15B) (1.4), GPT-4 (2.0), and GPT-3.5-Turbo (2.5). 
The most variable models are Vicuna-v1.3 (33B) (17.2), Llama2 (70B) (16.2), and Llama2 (13B) (14.1). As to \textbf{PCD-C (E-CARE)}, the most stable models are curie (6.7B) (1.2), GPT-3.5-Turbo (1.3), and text-curie-001 (1.2). The models with the highest sensitivity to prompts are Baichuan1-chat (13B) (20.4), Llama2 (13B) (17.0), and Vicuna-v1.3 (33B) (16.3). Finally, for \textbf{PCD-C (COPA)}, the most stable models are text-curie-001 (0.6), text-babbage-001 (0.8), and curie (6.7B) (1.7). The models most affected by prompt choice are Baichuan1-chat (13B) (24.6), Llama2-chat (70B) (23.1), and Llama2 (70B) (22.5). 
\textbf{Finally}, it is positive to note that both GPT-3.5-Turbo and GPT-4 demonstrate high accuracy and stability across four tasks. Conversely, it is observed that the llama-series and Vicuna-v1.3 (33B) often lack stability in tasks within this scenario.

4) \textbf{Open-Limited Ratio}: Considering the ratio of one open-access model to four limited-access models among the top five models in the entire scenario, the \textit{open-limited gap} is \textbf{moderate}.

Then, we analyze \textit{prompt gain} in PCD:

1) \textbf{Top Gain}: Figure \ref{fig:Heatmap_of_Pairwise_Causal_Discovery}(b) illustrates that the two most effective prompts in terms of average accuracy improvement over the basic prompt are 3-shot IcL, with a 9.0\% gain, and 1-shot IcL, with a 7.0\% gain. The highest leap in accuracy compared to the basic prompt was achieved by Llama2 (70B) using 3-shot IcL, resulting in a 33.5\% increase. A more granular analysis of each task is provided next. Figure \ref{fig:Heatmap_of_gain_of_Pairwise_Causal_Discovery} displays the heatmaps of accuracy gains for all tasks in the scenario. In the \textbf{PCD-B (E-CARE)} task, the top two prompts leading to the highest gains are 3-shot IcL at 6.4\% and 1-shot IcL at 4.4\%, with Llama2 (70B) using 3-shot IcL showing the most significant improvement of 25.1\%. For the \textbf{PCD-B (COPA)} task, the leading prompts in gain are 3-shot IcL at 10.0\% and 1-shot IcL at 4.9\%, with Llama2 (70B) using 3-shot IcL marking the largest increase at 35.2\%. In the \textbf{PCD-C (E-CARE)} task, the top gains are from 3-shot IcL at 7.4\% and 1-shot IcL at 7.2\%, with the most substantial accuracy boost seen with Baichuan1-chat (13B) using 1-shot IcL, achieving a 37.0\% increase. Lastly, for the \textbf{PCD-C (COPA)} task, the highest gains were from 3-shot IcL at 12.1\% and 1-shot IcL at 11.4\%, with Baichuan1-chat (13B) using 1-shot IcL experiencing the most significant improvement, at 52.1\%. 
\textbf{In summary}, the information suggests that tasks of lesser complexity generally lead to the most significant improvements in this scenario. For every individual task, the 3-shot IcL and 1-shot IcL are recognized as the 2 most effective prompts, highlighting the consistent efficacy of these two prompts in this scenario. Additionally, combined with specific models, these 2 prompts also achieve the highest gains compared to other \textit{model-prompt pair}s.

2) \textbf{Exceptions}: Though the 3-shot IcL prompt stands out as the highly effective prompt across most models, it has exceptions in Baichuan2-chat (7B), Claude2, and GPT-4. In the task of \textbf{PCD-B (E-CARE)}, this leading prompt falls short of enhancing performance for Baichuan2-chat (7B), text-davinci-002, Claude2, and GPT-3.5-Turbo. However, it is noteworthy that all prompts manage to boost Baichuan1-chat (13B)'s performance over the basic prompt in this specific task. For the \textbf{PCD-B (COPA)} task, the best prompt does not improve the performance of text-babbage-001, Claude2, and text-davinci-003 over the basic prompt, with text-davinci-003 showing no improvement over the basic prompt from any prompt. In \textbf{PCD-C (E-CARE)}, the top prompt fails to give a positive effect on Qwen (7B), Baichuan2-chat (7B), Claude2, InternLM-chat (20B), GPT-4, and GPT-3.5-Turbo over the basic prompt. Regarding \textbf{PCD-C (COPA)}, the leading prompt, 3-shot IcL, does not contribute positively to text-curie-001, Qwen (7B), Baichuan2-chat (7B), InternLM-chat (20B), and GPT-4 over the basic prompt. All prompts are capable of elevating Llama2 (7B)'s performance in this task. \textbf{It seems that} 3-shot IcL has difficulty promoting the performance of Baichuan2-chat (7B) and Claude2 in most of the tasks (3 out of 4) in this scenario.

3) \textbf{Stability}: The scenario highlights that the two most stable prompts, based on their low \textit{prompt volatility}, are 0-shot IcL with a \textit{prompt volatility} of 4.0 and adversarial ignore with a \textit{prompt volatility} of 5.2. Conversely, the prompts with the highest variability are 3-shot IcL and manual CoT, with \textit{prompt volatility} of 9.1 and 8.0, respectively. The scenario's \textit{average model-prompt-gain volatility} (\textit{AMPGV}) is 7.3, suggesting a \textbf{medium} \textit{prompt dependence}.
Conducting the analysis task by task, for \textbf{PCD-B (E-CARE)}, the most stable prompts are 0-shot IcL (4.0 \textit{prompt volatility}) and EF (4.4 \textit{prompt volatility}), whereas the least stable are manual CoT (11.1 \textit{prompt volatility}) and 0-shot CoT (9.2 \textit{prompt volatility}). An \textit{AMPGV} of 7.3 indicates a \textbf{medium} \textit{prompt dependence}.
For \textbf{PCD-B (COPA)}, the most stable prompts are 0-shot IcL (6.9 \textit{prompt volatility}) and adversarial ignore (7.1 \textit{prompt volatility}), with the least stable being manual CoT (14.8 \textit{prompt volatility}) and 3-shot IcL (10.1 \textit{prompt volatility}). An \textit{AMPGV} of 9.4 indicates a \textbf{medium} level of \textit{prompt dependence}.
In the \textbf{PCD-C (E-CARE)} task, the most stable prompts are 0-shot IcL (5.2 \textit{prompt volatility}) and adversarial ignore (8.4 \textit{prompt volatility}), while the least stable are 3-shot IcL (10.8 \textit{prompt volatility}) and manual CoT (10.7 \textit{prompt volatility}). An \textit{AMPGV} of 7.9 suggests a \textbf{medium} \textit{prompt dependence}.
For \textbf{PCD-C (COPA)}, the prompts with the smallest \textit{prompt volatility} are adversarial ignore (6.9 \textit{prompt volatility}) and 0-shot IcL (7.3 \textit{prompt volatility}), while the most unstable prompts are 3-shot IcL (15.8 \textit{prompt volatility}) and 1-shot IcL (13.8 \textit{prompt volatility}). With an \textit{AMPGV} of 9.2, the task has a \textbf{medium} level of \textit{prompt dependence}. 
\textbf{After evaluation of all the tasks in the scenario}, we find that the \textit{AMPGV}, which indicate the \textit{variance of prompt dependence} in the scenario, have a \textbf{narrow} range from 7.3 to 9.4. Moreover, 0-shot IcL and adversarial ignore are recognized as the most stable prompts. Furthermore, IcL and CoT are identified as the least stable prompts.

In the end, we measure \textit{language proficiency} in PCD:

1) \textbf{English vs. Chinese}: As illustrated in Figure \ref{fig:Pairwise_Causal_Discovery_Language}, models generally perform better on the English test set than on the Chinese test set, with 19 out of 28 models showing superior performance in English. 

2) \textbf{Accuracy Difference}: Significant performance differences favoring English appears in models such as InternLM-chat (20B) with a 15.9\% difference, Llama2-chat (70B) with a 12.1\% difference, and Vicuna-v1.3 (33B) with an 11.2\% difference. On the other hand, some models, including Baichuan1-chat (13B) with a 16.3\% difference, Baichuan2-chat (7B) with a 5.2\% difference, and Qwen (7B) with a 2.8\% difference, demonstrate higher capabilities in Chinese than in English.

\paragraph{Event causality identification.}
\begin{figure}[t]
\centering
\subfigure[Model performance of ECI]{
\begin{minipage}{8.5cm}
\centering
\includegraphics[width=1\linewidth]{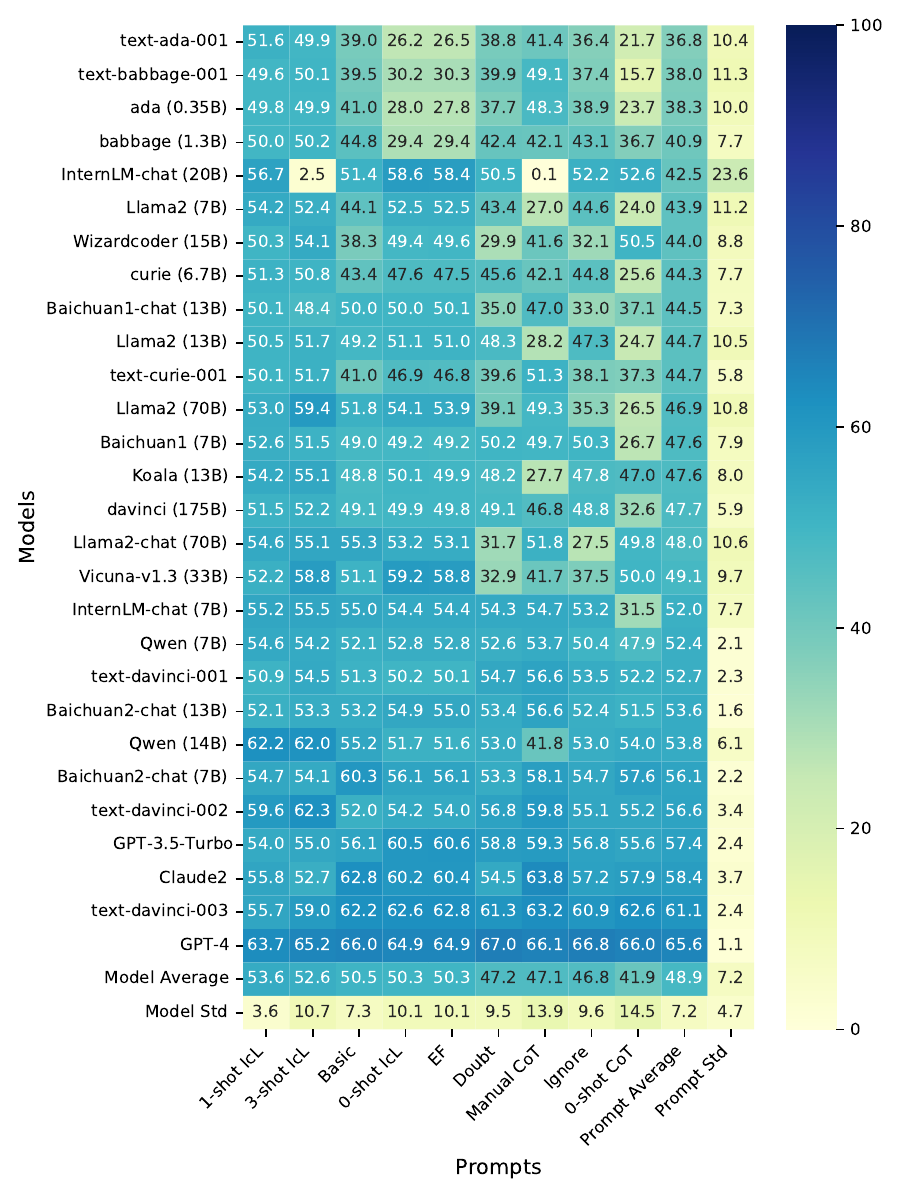}
\end{minipage}
}
\subfigure[\textit{Prompt gain} of ECI]{
\begin{minipage}{8.5cm}
\centering
\includegraphics[width=1\linewidth]{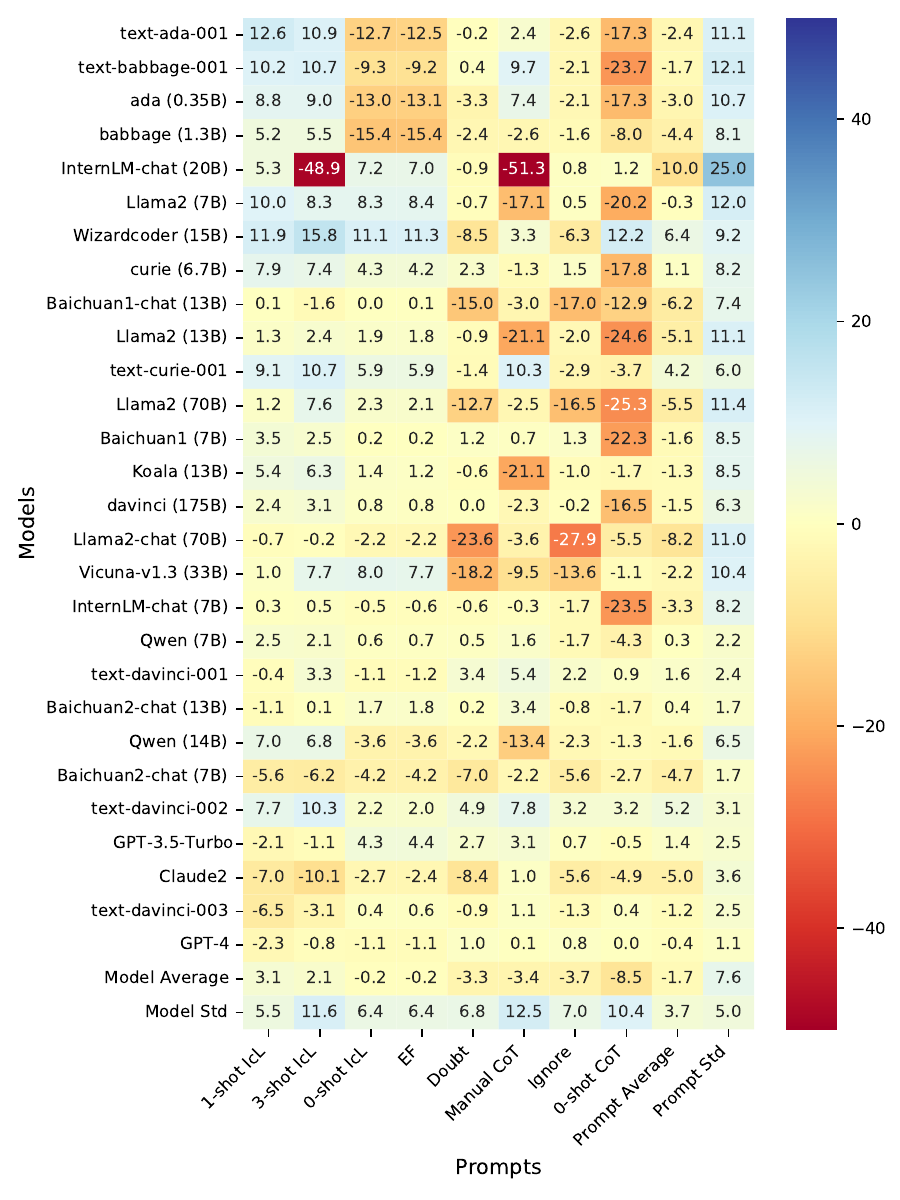}
\end{minipage}
}
\caption[Heatmap of ECI]{\textbf{Heatmap of ECI.} The models and prompts are sorted by their averages.}
\label{fig:Heatmap_of_Event_Causality_Identification}
\end{figure}

\begin{figure}
    \centering
    \includegraphics[width=0.8\linewidth]{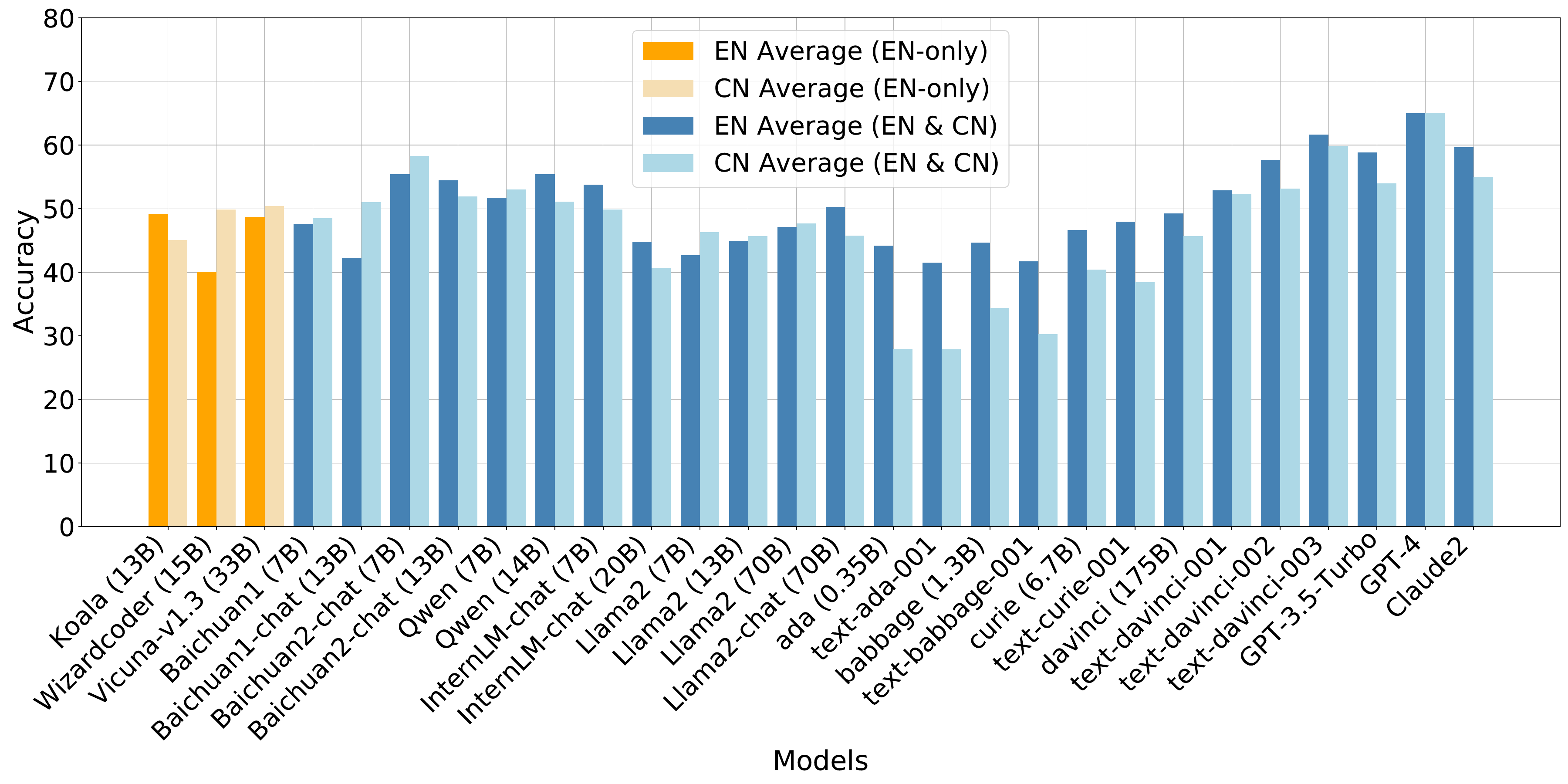}
    \caption[Language comparison of ECI]{\textbf{Language comparison of ECI.} The dark legend signifies the average performance of the model on an English test set, whereas the light legend denotes the average performance of the model on the Chinese test set. The yellow legend indicates a model trained exclusively on English datasets, while the blue legend represents a model trained on both English and Chinese datasets.}
    \label{fig:Event_Causality_Identification_Language}
\end{figure}

First, we assess model performance in ECI:

1) \textbf{Distribution}: The distribution of all \textit{model-prompt pair}s within ECI is depicted in Figure \ref{fig:Distribution_of_Causal_Discovery}(b). With a median of 51.4\% and a third quartile of 54.9\%, the scenario is considered to have an \textbf{easy} \textit{understandability}, as the median accuracy surpasses the random guess accuracy of 50.0\%. Figure \ref{fig:Distribution_of_Event_Causality_Identification_Tasks} illustrates the distribution of all \textit{model-prompt pair}s in each task respectively.
In the \textbf{ECI-B (CTB)} task, with a median of 50.6\%, a third quartile of 54.6\%, and a random guess accuracy of 50.0\%, the task is considered an \textbf{easy} \textit{understandability}.
In the \textbf{ECI-B (ESC)} task, given a median of 51.6\%, a third quartile of 56.1\%, and a random guess accuracy of 50.0\%, the task is also classified to have an \textbf{easy} \textit{understandability}.
Similarly, as to the \textbf{ECI-B (MAVEN-ERE)} task, with a median of 51.0\%, a third quartile of 54.5\%, and a random guess accuracy of 50.0\%, its \textit{understandability} is categorized as \textbf{easy}. 
\textbf{By analyzing the differences between tasks}, it is observed that the median accuracy for individual tasks spans from 50.6\% to 51.6\%, with a standard deviation of 0.4. The accuracy at the third quartile extends from 54.5\% to 56.1\%, accompanied by a standard deviation of 0.7. Consequently, this scenario presents a \textbf{minimally divergent} \textit{variance of distribution}.

2) \textbf{Top Accuracy}: As illustrated in Figure \ref{fig:Heatmap_of_Event_Causality_Identification}(a), the leading three models in terms of average accuracy are GPT-4 at 65.6\%, text-davinci-003 at 61.1\%, and Claude2 at 58.4\%. The \textit{top model-prompt pair} is GPT-4 with adversarial doubt, reaching an accuracy of 67.0\%, indicating the scenario has a \textbf{challenging} \textit{solvability} since the performance of the \textit{top model-prompt pair} does not exceed 80\%. Figure \ref{fig:Heatmap_of_performances_of_Event_Causality_Identification} displays the three models with the highest average accuracy across individual tasks.
For \textbf{ECI-B (CTB)}, the highest average accuracies are achieved by GPT-4 at 66.6\%, Claude2 at 62.1\%, and text-davinci-003 at 61.9\%. GPT-4, when combined with manual CoT, attains the highest accuracy of 68.8\%, revealing the task \textit{solvability} as \textbf{challenging} given the \textit{top model-prompt pair}'s performance falls below 80\%. In the case of \textbf{ECI-B (ESC)}, GPT-4 leads with 68.4\%, followed by text-davinci-003 at 63.0\%, and GPT-3.5-Turbo at 59.2\%. The \textit{top model-prompt pair} here is GPT-4 with adversarial doubt, reaching 70.1\%, confirming the task's \textbf{challenging} \textit{solvability} as the \textit{top model-prompt pair} remains under 80\%. Lastly, for \textbf{ECI-B (MAVEN-ERE)}, the top accuracies are by GPT-4 at 61.9\%, text-davinci-003 at 58.6\%, and GPT-3.5-Turbo at 55.8\%, with GPT-4 and adversarial ignore achieving the best at 64.8\%. This outcome also categorizes the \textit{solvability} of the task as \textbf{challenging} since the \textit{top model-prompt pair} does not meet or exceed 80\%. 
\textbf{Through comparing different tasks}, the \textit{variance of solvability} across tasks appears \textbf{negligible}. Furthermore, the leading model exhibits an average accuracy fluctuating between 61.9\% to 68.4\%, a difference of 6.5\%, while the highest accuracy observed in \textit{top model-prompt pair}s varies from 64.8\% to 70.1\%, a 5.3\% difference. Hence, the \textit{variance of model's top performance} in the scenario is \textbf{noticeable}. Additionally, GPT-4 stands out as the leading model in terms of average performance and also forms the \textit{top model-prompt pair}s in all tasks.

3) \textbf{Stability}: The three most stable models in the scenario, characterized by the lowest \textit{model volatility}, are GPT-4 with a \textit{model volatility} of 1.1, Baichuan2-chat (13B) with 1.6, and Qwen (7B) with 2.1. Conversely, the models exhibiting the greatest instability, shown by the highest \textit{model volatility}, include InternLM-chat (20B) with \textit{model volatility} of 23.6, text-babbage-001 at 11.3, and Llama2 (7B) at 11.2, reflecting their pronounced sensitivity to prompt variations. Delving into stability on a task-by-task basis:
For the \textbf{ECI-B (CTB)} task, the three models demonstrating the greatest stability, with the lowest \textit{model volatility}, are GPT-4 at 1.2, Qwen (7B) at 1.9, and Baichuan2-chat (13B) at 2.0. In contrast, the models with the most significant instability, indicated by the largest \textit{model volatility}, are InternLM-chat (20B) at 23.5, Llama2 (13B) at 12.6, and Llama2 (7B) at 12.5.
In the \textbf{ECI-B (ESC)} task, the top stable models include GPT-4 with a \textit{model volatility} of 1.1, Baichuan2-chat (13B) at 1.6, and GPT-3.5-Turbo at 2.5. The models showing the most instability are InternLM-chat (20B) at 24.2, Llama2-chat (70B) at 12.2, and Llama2 (70B) at 11.3.
For the \textbf{ECI-B (MAVEN-ERE)} task, the models with the highest stability are Baichuan2-chat (13B) and Baichuan2-chat (7B), both at 1.5, followed by GPT-4 at 1.7. The most unstable models include InternLM-chat (20B) at 23.1, text-babbage-001 at 12.1, and Llama2 (70B) at 11.1.
\textbf{In conclusion}, it can be seen that GPT-4 is the most stable model across all tasks, while the Baichuan2-chat (13B) is the least stable one.

4) \textbf{Open-Limited Ratio}: Among the top five models with the highest average accuracy, a 0:5 ratio of open-access to limited-access models indicates a \textbf{large} \textit{open-limited gap}.

Then, we conduct \textit{prompt gain} analysis in ECI:

1) \textbf{Top Gain}: As shown in Figure \ref{fig:Heatmap_of_Event_Causality_Identification}(b), the leading two prompts achieving the greatest average accuracy improvements over the basic prompt are 1-shot IcL with a gain of 3.1\% and 3-shot IcL with 2.1\%. The largest increase in accuracy compared to the basic prompt is seen in Wizardcoder (15B) utilizing 3-shot IcL, with a remarkable gain of 15.8\%. A comprehensive task-specific analysis follows. Figure \ref{fig:Heatmap_of_gain_of_Event_Causality_Identification} illustrates the heatmap of accuracy gains for all tasks within the scenario. For the \textbf{ECI-B (CTB)} task, the two top-performing prompts in terms of average accuracy gain over the basic prompt are 1-shot IcL at 3.6\% and 3-shot IcL at 2.1\%. The highest accuracy enhancement from the basic prompt is achieved by Wizardcoder (15B) using manual CoT, with a gain of 16.3\%. In the \textbf{ECI-B (ESC)} task, the prompts leading to the highest average accuracy gains compared to the basic prompt are 1-shot IcL at 2.8\% and 3-shot IcL at 2.2\%. The most substantial improvement over the basic prompt is by text-ada-001 with 1-shot IcL, showing a gain of 15.2\%. For the \textbf{ECI-B (MAVEN-ERE)} task, the two prompts with the greatest average accuracy gains over the basic prompt are 1-shot IcL at 3.0\% and 3-shot IcL at 2.0\%. The most significant accuracy increase from the basic prompt is observed with Wizardcoder (15B) using 3-shot IcL, with an increase of 18.0\%. 
\textbf{In summary}, 1-shot IcL and 3-shot IcL is the most effective prompt in all tasks. However, IcL generally has a more passive impact on the top 5 LLMs, as their lowest performance often originates from IcL. The two types of adversarial prompts and 0-shot CoT, negatively affect most LLMs, whereas EF and manual CoT have approximately equal positive and negative effects.

2) \textbf{Exceptions}: In ECI, the most effective prompt shows exceptions in failing to improve the model performance over basic prompt in Llama2-chat (70B), text-davinci-001, Baichuan2-chat (13B and 7B), GPT-3.5-Turbo, Claude2, text-davinci-003, and GPT-4. All prompts enhance text-davinci-002's performance beyond the basic prompt. However, no prompt boosts the performance of Llama2-chat (70B) or Baichuan2-chat (7B) over the accuracy of basic prompt.
In the \textbf{ECI-B (CTB)} task, the leading prompt does not enhance the performance of  Baichuan2-chat (13B and 7B), GPT-3.5-Turbo, text-davinci-003, or GPT-4 beyond their performance with the basic prompt. Every prompt conducts improvements for text-davinci-001 and text-davinci-002 over the basic prompt, while none can elevate Baichuan2-chat (7B)'s performance in this task.
For the \textbf{ECI-B (ESC)} task, the optimal prompt fails to lift the accuracy of Llama2-chat (70B), text-davinci-001, InternLM-chat (7B), Baichuan2-chat (13B and 7B), Claude2, GPT-3.5-Turbo, text-davinci-003, and GPT-4 beyond their accuracy with the basic prompt. Every prompt, however, can lift text-davinci-002's performance above the basic prompt, with none managing to improve the performance for Llama2-chat (70B) or Baichuan2-chat (7B) in this specific task.
In the \textbf{ECI-B (MAVEN-ERE)} task, the best prompt proves ineffective for Baichuan1-chat (13B), Llama2 (70B) and Llama2-chat (70B), text-davinci-001, Baichuan2-chat (13B), Claude2, Baichuan2-chat (7B), GPT-3.5-Turbo, text-davinci-003, and GPT-4 in surpassing their performance on basic prompt. Nevertheless, all prompts are capable of enhancing text-davinci-002's performance beyond the basic prompt. Yet, no prompt is able to augment performance for Baichuan1-chat (13B), Llama2-chat (70B), Claude2, or Baichuan2-chat (7B) in this task. 
\textbf{By evaluating the three tasks}, we find that the best prompt 1-shot IcL cannot create positive average \textit{prompt gain} in Baichuan2-chat (13B), Baichuan2-chat (7B), GPT-3.5-Turbo, text-davinci-003, and GPT-4 in any task. All prompts in all tasks have a positive average \textit{prompt gain} on text-davinci-002, while no prompts in any tasks have a positive average \textit{prompt gain} on Baichuan2-chat (7B).

3) \textbf{Stability}: As to the stability, the top 2 stable prompts with the smallest \textit{prompt volatility} are 1-shot IcL with a \textit{prompt volatility} of 5.5 and EF with a \textit{prompt volatility} of 6.4. In contrast, the top 2 most unstable prompts with the largest \textit{prompt volatility} are manual CoT at 12.5 and 3-shot IcL at 11.6. The \textit{average model-prompt-gain volatility} (\textit{AMPGV}) is 7.6, showing that the scenario has a \textbf{medium} \textit{prompt dependence}. Next, we consider the stability task by task. 
For task \textbf{ECI-B (CTB)}, the top 2 stable prompts with the smallest \textit{prompt volatility} are 1-shot IcL with a \textit{prompt volatility} of 5.3 and adversarial ignore with a \textit{prompt volatility} of 6.5. In contrast, the top 2 most unstable prompts with the largest \textit{prompt volatility} are manual CoT at 15.1 and 0-shot CoT at 11.7. The \textit{AMPGV} is 8.3, showing that the task's \textit{prompt dependence} is \textbf{medium}. 
For task \textbf{ECI-B (ESC)}, the top 2 stable prompts with the smallest \textit{prompt volatility} are 1-shot IcL with an \textit{prompt volatility} of 6.1 and EF with an \textit{prompt volatility} of 6.2. On the other hand, the top 2 most unstable prompts with the largest \textit{prompt volatility} are manual CoT at 12.3 and 3-shot IcL at 12.3. The \textit{AMPGV} is 7.7, showing that the task has a \textbf{medium} \textit{prompt dependence}. 
For task \textbf{ECI-B (MAVEN-ERE)}, the top 2 stable prompts with the smallest \textit{prompt volatility} are 1-shot IcL at 5.8 and adversarial doubtat 6.3. The top 2 most unstable prompts are 3-shot IcL at 11.9 and manual CoT at 11.8. The \textit{AMPGV} is 7.4, indicating that the task's \textit{prompt dependence} is \textbf{medium}. 
\textbf{After evaluation of all the tasks in the scenario}, it is found that the \textit{AMPGV} ranges from 7.4 to 8.3. Therefore, the scenario has a \textbf{narrow} \textit{variance of prompt dependence}. In all tasks, 1-shot IcL is the top 1 stable prompt. On the other hand, manual CoT is the relatively least stable one. 

Lastly, we assess \textit{language proficiency} in ECI:

1) \textbf{English vs. Chinese}: According to Figure \ref{fig:Event_Causality_Identification_Language}, models tend to exhibit superior performance on the English test set compared to the Chinese one. Specifically, 19 out of 28 models showed enhanced performance in English over Chinese.

2) \textbf{Accuracy Difference}: There is a significant performance gap between English and Chinese, with a preference for English in models like ada (0.35B) (21.4\%), text-ada-001 (18.4\%), and text-babbage-001 (15.7\%). On the other hand, some models, including Wizardcoder (15B) (12.8\%), Baichuan1-chat (13B) (11.9\%), and Llama2 (7B) (4.8\%), demonstrate a higher proficiency for Chinese than for English. This indicates that models primarily trained in English are capable of achieving noteworthy performance in Chinese as well, highlighting their cross-linguistic generalizability.

\paragraph{Abstract reasoning.}
\begin{figure}[t]
\centering
\subfigure[Model performance of AR]{
\begin{minipage}{8.5cm}
\centering
\includegraphics[width=1\linewidth]{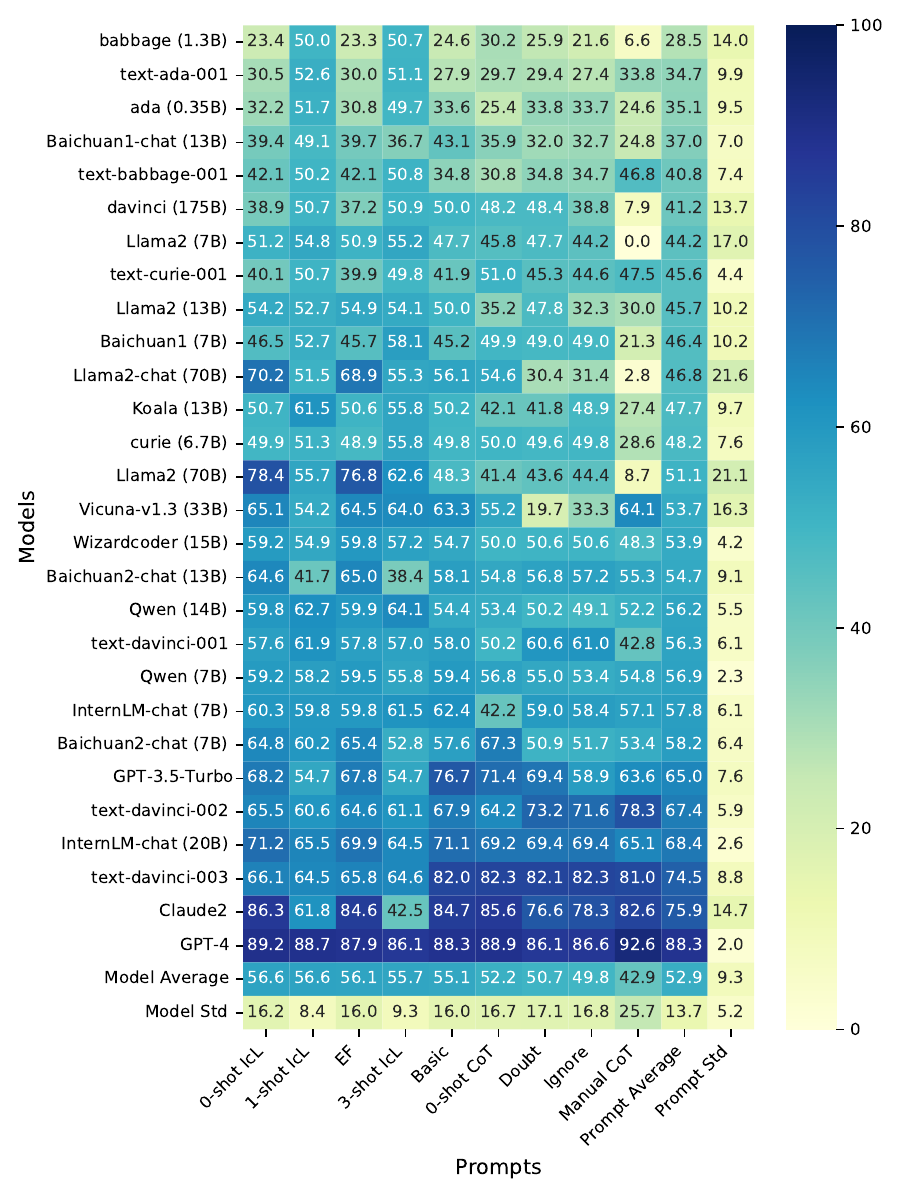}
\end{minipage}
}
\subfigure[\textit{Prompt gain} of AR]{
\begin{minipage}{8.5cm}
\centering
\includegraphics[width=1\linewidth]{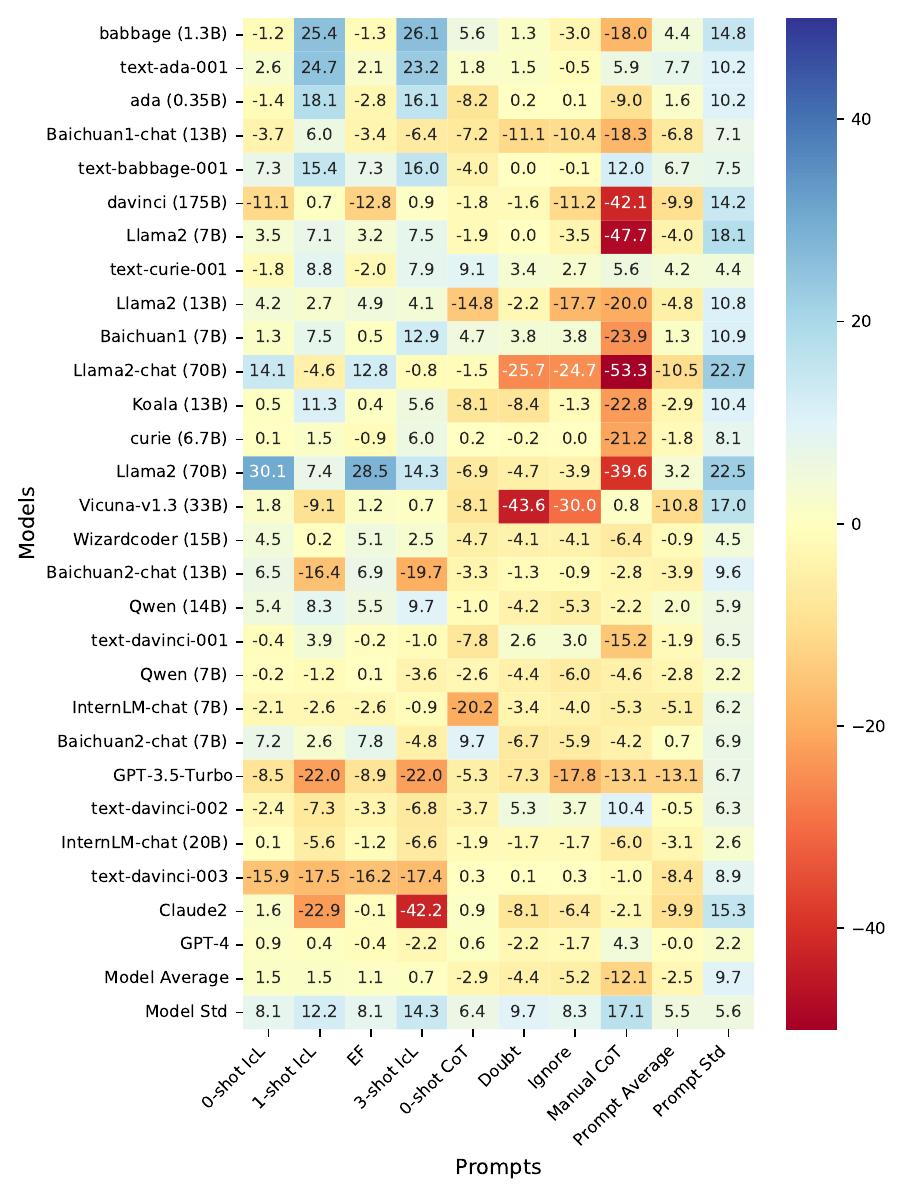}
\end{minipage}
}
\caption[Heatmap of AR]{\textbf{Heatmap of AR.} The models and prompts are sorted by their averages.}
\label{fig:Heatmap_of_Abstract_Reasoning}
\end{figure}

\begin{figure}
    \centering
    \includegraphics[width=0.8\linewidth]{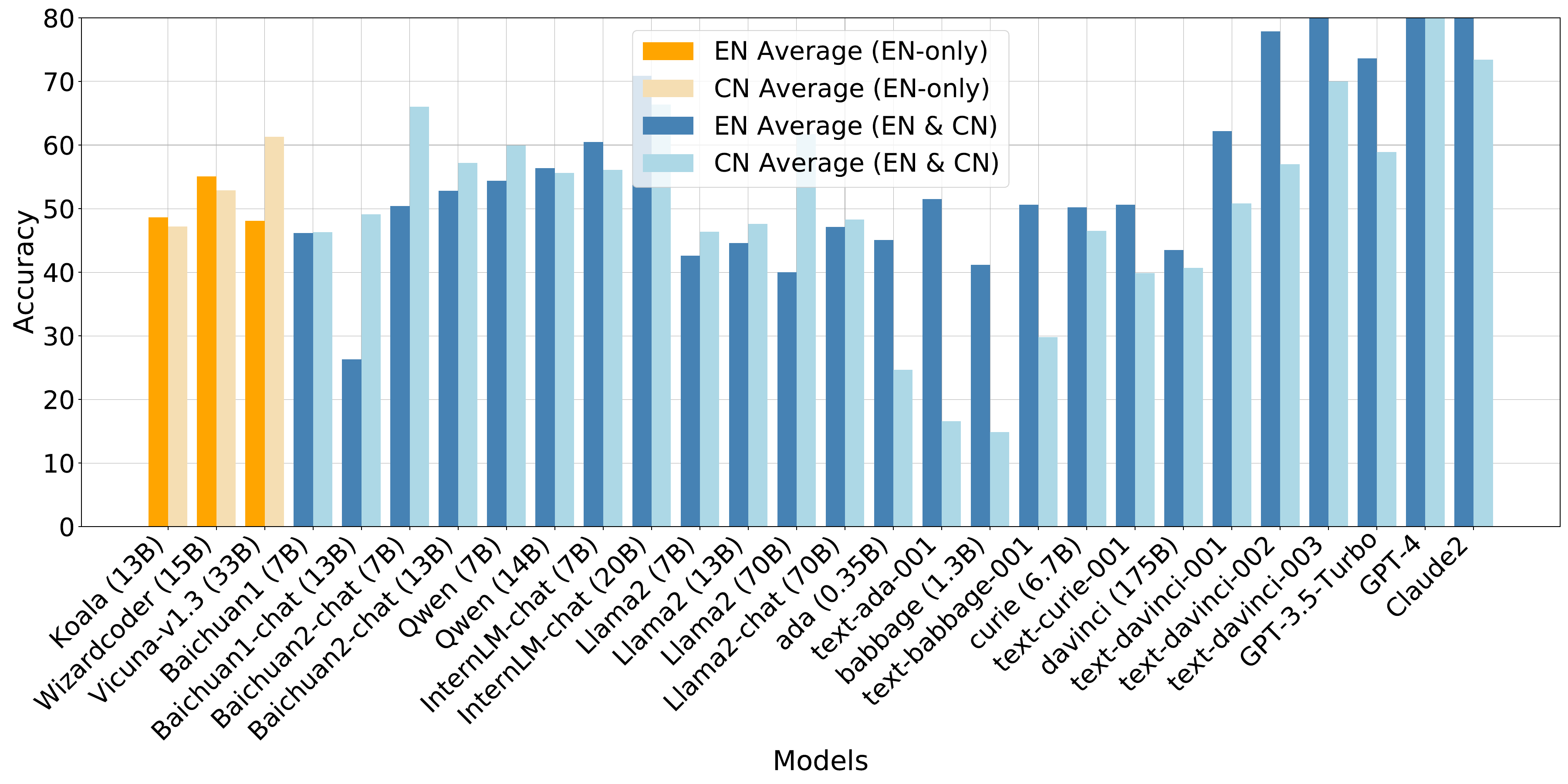}
    \caption[Language comparison of AR]{\textbf{Language comparison of AR.} The dark legend signifies the average performance of the model on an English test set, whereas the light legend denotes the average performance of the model on the Chinese test set. The yellow legend indicates a model trained exclusively on English datasets, while the blue legend represents a model trained on both English and Chinese datasets.}
    \label{fig:Abstract_Reasoning_Language}
\end{figure}

Regarding model performance in AR: 1) \textbf{Distribution}: Figure \ref{fig:Distribution_of_Causal_Discovery}(c) displays the distribution of all \textit{model-prompt pair}s within AR, noting a median accuracy of 53.1\% and a third quartile of 62.6\%. This scenario is classified to have an \textbf{easy} \textit{understandability} given that the median accuracy surpasses the random guess benchmark of 50.0\%.
2) \textbf{Top Accuracy}: Figure \ref{fig:Heatmap_of_Abstract_Reasoning}(a) reveals the top three models by average accuracy: GPT-4 at 88.3\%, Claude2 at 75.9\%, and text-davinci-003 at 74.5\%. GPT-4, employing manual CoT, stands out as the \textit{top model-prompt pair} with a 92.6\% accuracy. The \textit{solvability} of the scenario is \textbf{well-solved} with each of the top three models' average accuracies exceeding 70\%.
3) \textbf{Stability}: After computing the \textit{model volatility} introduced in \cref{metric:model}, we find the most prompt-sensitive, thus unstable models are Llama2-chat (70B) at 21.6, Llama2 (70B) at 21.1, and Llama2 (7B) at 17.0. Conversely, the most stable models include GPT-4 at 2.0, Qwen (7B) at 2.3, and InternLM-chat (20B) at 2.6.
4) \textbf{Open-Limited Ratio}: The ratio of open-access to limited-access models among the top five models with the highest average accuracy is 1:4, suggesting a \textbf{moderate} \textit{open-limited gap}.

Regarding \textit{prompt gain} in AR: 1) \textbf{Top Gain}: As shown in Figure \ref{fig:Heatmap_of_Abstract_Reasoning}(b), the leading two prompts by average accuracy gain over the basic prompt are 0-shot IcL and 1-shot IcL, both at 1.5\%. Llama2 (70B) using 0-shot IcL exhibits the most significant gain of 30.1\%.
2) \textbf{Exceptions}: The high-performing prompt, 0-shot IcL, proves effective with most models, but fails to generate positive performance over some models such as babbage (1.3B) and GPT-3.5-Turbo. Furthermore, no prompt is capable of creating positive average \textit{prompt gain} for InternLM-chat (7B) and GPT-3.5-Turbo.
3) \textbf{Stability}: Regarding stability, the most stable prompts are 0-shot CoT with an \textit{prompt volatility} of 6.4 and EF with an \textit{prompt volatility} of 8.1, while the most unstable prompts are manual CoT at 17.1 and 3-shot IcL at 14.3. The scenario shows a \textbf{medium} \textit{prompt dependence}, evidenced by an \textit{average model-prompt-gain volatility} (\textit{AMPGV}) of 9.7.

In terms of \textit{language proficiency} in AR: 
1) \textbf{English vs. Chinese}: Figure \ref{fig:Abstract_Reasoning_Language} indicates models perform better on the English test set over the Chinese set, with 17 of 28 models favoring English. 
2) \textbf{Accuracy Difference}: Significant performance gaps favoring English are seen in text-ada-001 (34.9\%), babbage (1.3B) (26.3\%), and text-davinci-002 (20.9\%). In contrast, models like Baichuan1-chat (13B), Llama2 (70B), and Baichuan2-chat (7B) demonstrate higher proficiency in Chinese.

\paragraph{Causal attribution.}
\begin{figure}[t]
\centering
\subfigure[Model performance of CA]{
\begin{minipage}{8.5cm}
\centering
\includegraphics[width=1\linewidth]{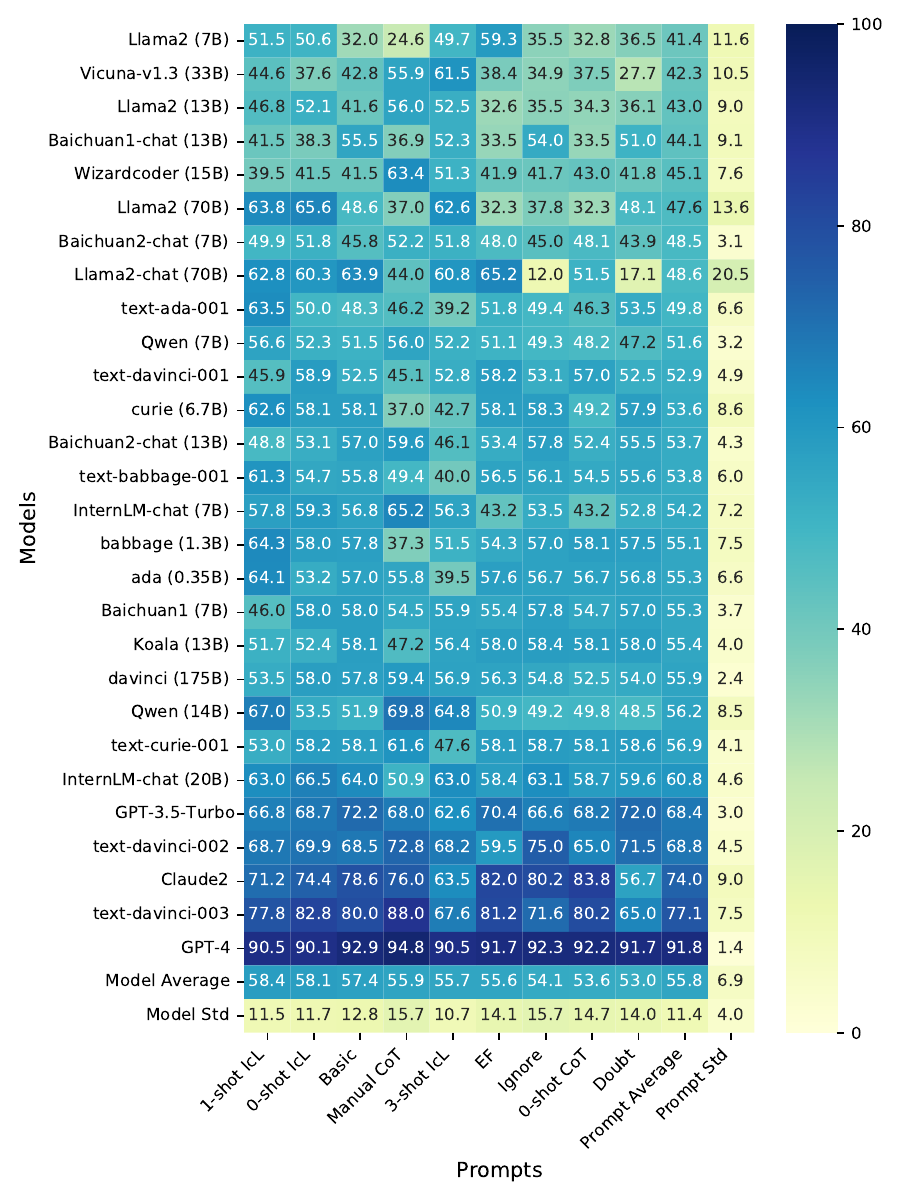}
\end{minipage}
}
\subfigure[\textit{Prompt gain} of CA]{
\begin{minipage}{8.5cm}
\centering
\includegraphics[width=1\linewidth]{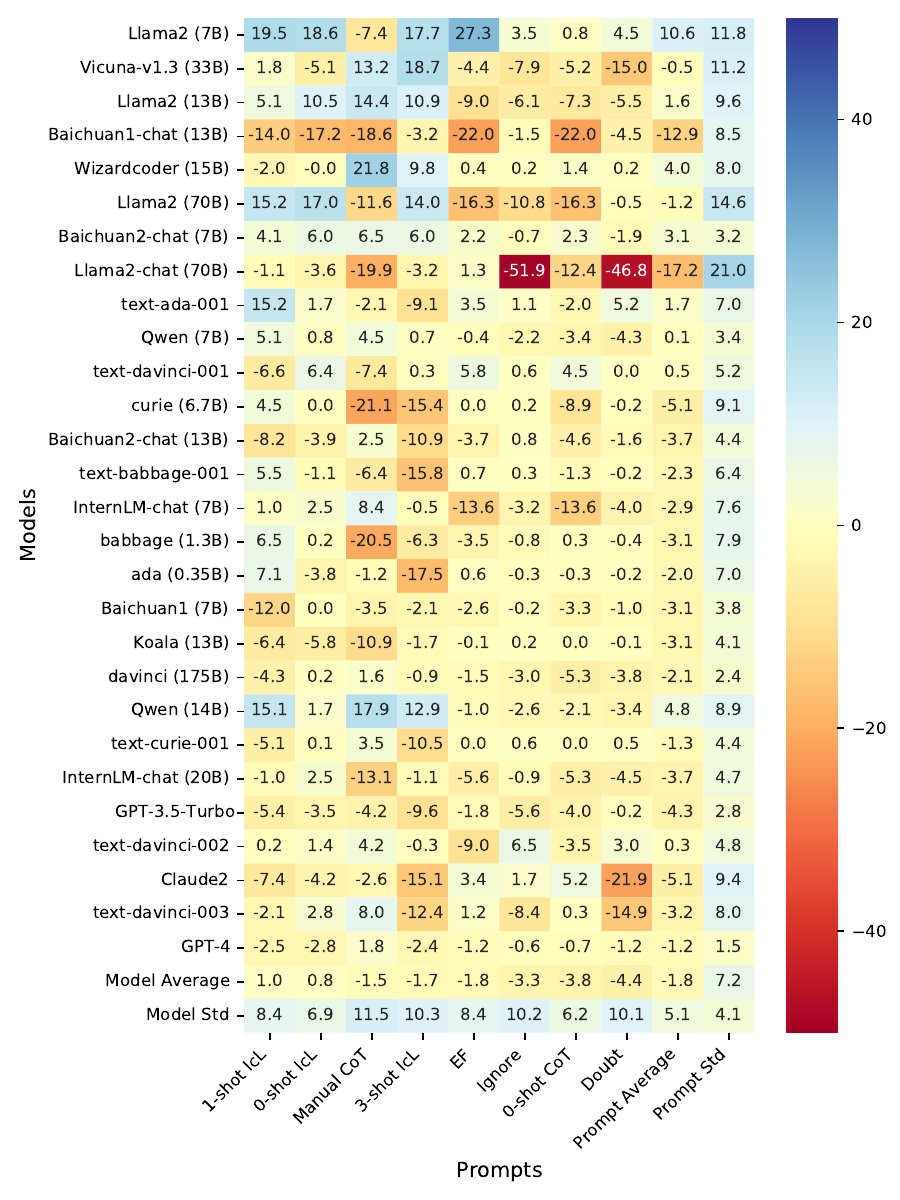}
\end{minipage}
}
\caption[Heatmap of CA]{\textbf{Heatmap of CA.} The models and prompts are sorted by their averages.}
\label{fig:Heatmap_of_Causal_Attribution}
\end{figure}

\begin{figure}
    \centering
    \includegraphics[width=0.8\linewidth]{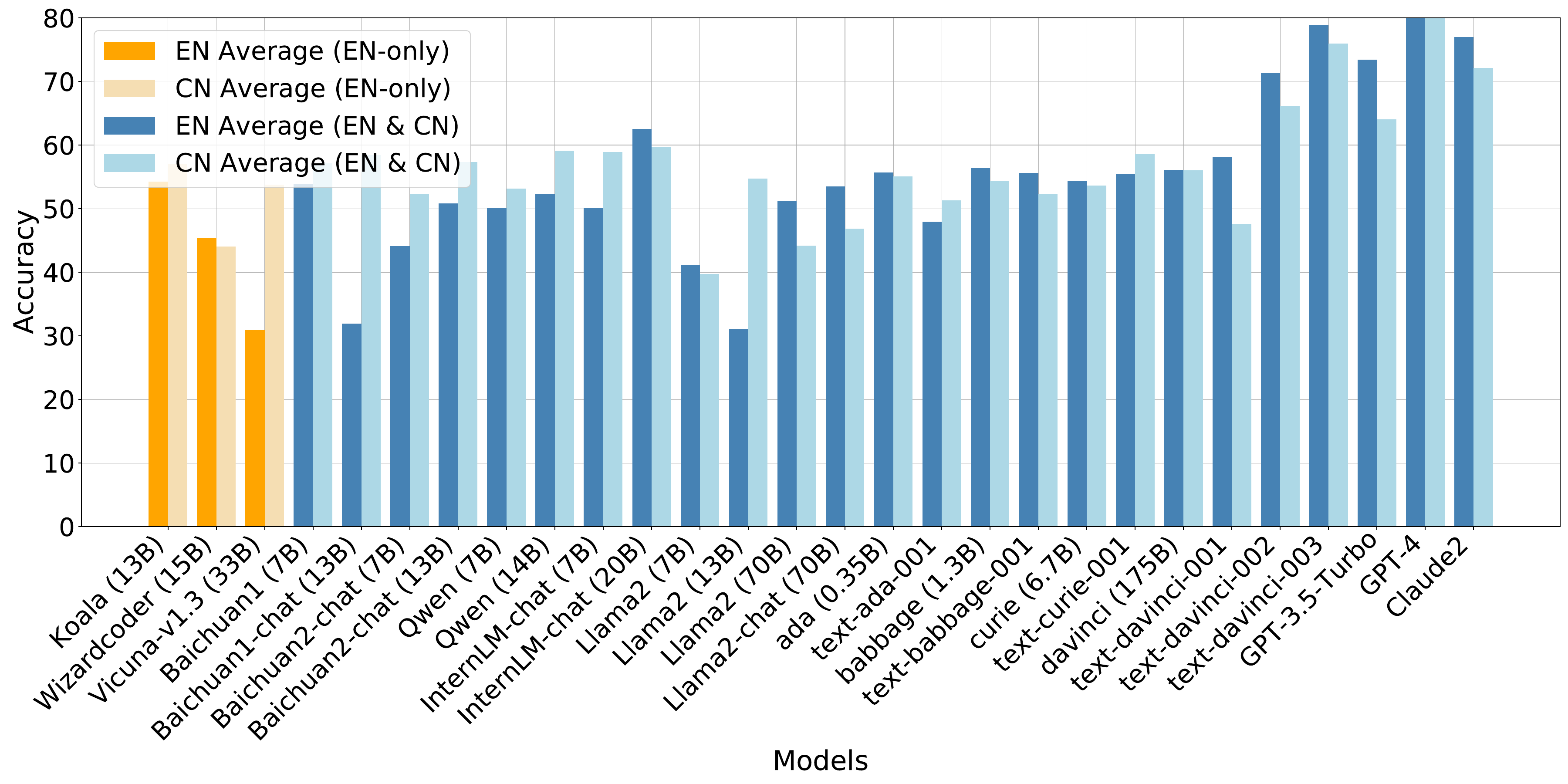}
    \caption[Language comparison of CA]{\textbf{Language comparison of CA.} The dark legend signifies the average performance of the model on an English test set, whereas the light legend denotes the average performance of the model on the Chinese test set. The yellow legend indicates a model trained exclusively on English datasets, while the blue legend represents a model trained on both English and Chinese datasets.}
    \label{fig:Causal Attribution Language}
\end{figure}

Initially, we delve into model performance in CA:

1) \textbf{Distribution}: As depicted in Figure \ref{fig:Distribution_of_Causal_Discovery}(d), the distribution for all \textit{model-prompt pair}s in CA shows a median accuracy of 55.9\% and a third quartile of 61.8\%. This indicates that the scenario has an \textbf{easy} \textit{understandability}, given that the median accuracy surpasses the random guess benchmark of 50.0\%. Figure \ref{fig:Distribution_of_Causal_Attribution_Tasks} further outlines the distribution of all the \textit{model-prompt pair}s for each specific causal task.
In the \textbf{CA-B (FA)} task, with a median accuracy of 63.2\% and a third quartile of 72.4\%, exceeding the random guess threshold of 50.0\%, this task is classified to have an \textbf{easy} \textit{understandability}. Similarly, the \textbf{CA-B (FP)} task, with a median of 50.6\% and a third quartile of 56.3\%, also surpasses the random guess mark, causing its \textit{understandability} classification as \textbf{easy}. 
\textbf{By analyzing the differences between tasks}, we observe that the median accuracy for each task varies from 50.6\% to 63.2\%, with a standard deviation of 6.3. Regarding the accuracy at the third quartile, it spans from 56.3\% to 72.4\%, with a standard deviation of 8.1. Hence, this scenario has a \textbf{moderately distinct} \textit{variance of distribution}. Moreover, CA-B (FA) is easier to understand than CA-B (FP).

2) \textbf{Top Accuracy}: As illustrated in Figure \ref{fig:Heatmap_of_Causal_Attribution}(a), GPT-4 leads with an average accuracy of 91.8\%, followed by text-davinci-003 at 77.1\%, and Claude2 at 74.0\%. GPT-4, when paired with manual CoT, achieves an impressive 94.8\%. The \textit{solvability} of this scenario is \textbf{well-solved} given that the top three models all have average accuracies over 70\%. Figure \ref{fig:Heatmap_of_performances_of_Causal_Attribution} further demonstrates the top models' average accuracy for each task.
In the \textbf{CA-B (FA)} task, GPT-4 dominates with 91.2\%, with GPT-3.5-Turbo at 74.8\% and text-davinci-002 closely behind at 74.3\%. GPT-4's manual CoT combination tops at 93.6\%. The above results show the \textbf{well-solved} \textit{solvability} as all top-three models exceed an average accuracy of 70\%. For the \textbf{CA-B (FP)} task, GPT-4 again leads at 92.5\%, with text-davinci-003 at 80.1\% and Claude2 at 78.5\%. The \textit{top model-prompt pair}, achieved by GPT-4 with manual CoT, is 95.9\%. The \textit{solvability} of the task is \textbf{well-solved} as all leading models surpass the 70\% accuracy threshold.
\textbf{Through comparing different tasks}, we find that the \textit{variance of solvability} across tasks is \textbf{negligible}. Additionally, the highest average accuracy for the top-1 model is between 91.2\% and 92.5\%, a variance of 1.3, while the peak accuracy for \textit{top model-prompt pair}s lies between 93.6\% and 95.9\%, marking a 2.3 difference. Thus, the \textit{variance of model's top performance} of the scenario is \textbf{small}. Moreover, GPT-4 is consistently found to be the superior model in average performance and forms part of the \textit{top model-prompt pair}s across all tasks. Manual CoT forms the other part of the \textit{top model-prompt pair}s across all tasks.

3) \textbf{Stability}: The three most stable models, characterized by the lowest \textit{model volatility}, are GPT-4 with a \textit{model volatility} of 1.4, davinci (175B) at 2.4, and GPT-3.5-Turbo at 3.0, showcasing their robustness across various prompts. Conversely, the models demonstrating the greatest variability, with the highest \textit{model volatility}, are Llama2-chat (70B) at 20.5, Llama2 (70B) at 13.6, and Llama2 (7B) at 11.6, highlighting their significant sensitivity to prompt variations. A further analysis of stability by individual tasks is as follows:
In the \textbf{CA-B (FA)} task, the models with the least variability are GPT-4 at a \textit{model volatility} of 1.1, GPT-3.5-Turbo at 3.3, and text-davinci-002 at 3.5. On the flip side, the most variable models, showing the largest \textit{model volatility}, include Llama2-chat (70B) at 25.6, Llama2 (70B) at 18.9, and babbage (1.3B) at 15.1.
For the \textbf{CA-B (FP)} task, the models exhibiting the greatest stability are Qwen (7B) and GPT-4, both at 2.0, followed closely by Wizardcoder (15B) at 2.2. Meanwhile, the models with the most variability are Llama2-chat (70B) at 15.9, Llama2 (7B) at 10.7, and text-davinci-002 at 9.9.
\textbf{Comparing across tasks}, GPT-4 demonstrates great stability while Llama2-chat (70B) consistently exhibits considerable instability across different tasks, marking it as highly sensitive to the prompts used.

4) \textbf{Open-Limited Ratio}: The top 5 models with the highest average accuracy in the scenario exhibit a 0:5 open-access to limited-access model ratio, highlighting a \textbf{large} \textit{open-limited gap}.

Then, we analyze \textit{prompt gain} in CA:

1) \textbf{Top Gain}: As depicted in Figure \ref{fig:Heatmap_of_Causal_Attribution}(b), the two prompts with the highest average accuracy gain over the basic prompt are 1-shot IcL at 1.0\% and 0-shot IcL at 0.8\%. The most significant accuracy improvement over the basic prompt is seen with Llama2 (7B) using EF, showing an increase of 27.3\%. Following this, we conduct a detailed task-by-task analysis. Figure \ref{fig:Heatmap_of_gain_of_Causal_Attribution} displays the heatmaps of gains for all tasks in the scenario. For the task \textbf{CA-B (FA)}, the top 2 prompts with the highest average accuracy gain over the basic prompt are 0-shot IcL at 1.9\% and 1-shot IcL at 1.5\%. The most significant accuracy improvement over the basic prompt is seen with Wizardcoder (15B) using manual CoT, showing an increase of 40.6\%. For task \textbf{CA-B (FP)}, the top 2 prompts with the highest average accuracy gain over the basic prompt are 1-shot IcL at 0.5\% and manual CoT at 0.4\%. The most significant accuracy improvement over the basic prompt is seen with Llama2 (13B) using manual CoT, showing an increase of 18.0\%. 
\textbf{After evaluating the two tasks}, it appears that the scenario favors 1-shot IcL and manual CoT the most.

2) \textbf{Exceptions}: In CA, the most effective prompt fails to create positive average \textit{prompt gain} in Baichuan1-chat (13B), Wizardcoder (15B), Llama2-chat (70B), text-davinci-001, Baichuan2-chat (13B), Baichuan1 (7B), Koala (13B), davinci (175B), text-curie-001, InternLM-chat (20B), GPT-3.5-Turbo, Claude2, text-davinci-003, and GPT-4. Notably, Baichuan1-chat (13B) and GPT-3.5-Turbo see no positive average \textit{prompt gain} from any prompt. In the \textbf{CA-B (FA)} task, the most effective prompt 0-shot IcL fails with Baichuan1-chat (13B), Baichuan2-chat (13B), text-babbage-001, ada (0.35B), Koala (13B), Claude2, and GPT-4 in generating positive average prompt gain, whereas all prompts manage to lift Llama2 (7B)'s performance. Within the \textbf{CA-B (FP)} task, the leading prompt 1-shot IcL does not give positive average \textit{prompt gain} to Baichuan1-chat (13B), Vicuna-v1.3 (33B), Llama2-chat (70B), text-davinci-001, InternLM-chat (7B), Qwen (7B), Baichuan2-chat (13B), Wizardcoder (15B), Baichuan2-chat (7B), GPT-3.5-Turbo, text-davinci-002, Claude2, text-davinci-003, and GPT-4. Additionally, Baichuan1-chat (13B), Llama2-chat (70B), Qwen (7B), and Baichuan2-chat (7B) do not benefit from any prompt in this task. 
\textbf{It can be seen that across both tasks}, it is challenging to achieve any improvement for Baichuan1-chat (13B) with any prompt.

3) \textbf{Stability}: Regarding stability, the two most stable prompts, identified by the lowest \textit{prompt volatility}, are 0-shot CoT with a \textit{prompt volatility} of 6.2 and 0-shot IcL with a \textit{prompt volatility} of 6.9. On the opposite end, the prompts showing the most variability, indicated by the highest \textit{prompt volatility}, are manual CoT at 11.5 and 3-shot IcL at 10.3. The \textit{average model-prompt-gain volatility} (\textit{AMPGV}) stands at 7.2, thus the \textit{prompt dependence} in this scenario is \textbf{medium}. Moving to a task-specific analysis:
For the \textbf{CA-B (FA)} task, the two most stable prompts, based on the smallest \textit{prompt volatility}, are 0-shot CoT at 9.0 and 0-shot IcL at 9.8. Conversely, the least stable prompts, marked by the largest \textit{prompt volatility}, are 3-shot IcL at 19.5 and manual CoT at 18.9. The \textit{AMPGV} is 11.0, highlighting a \textbf{high} \textit{prompt dependence} for this task.
For the \textbf{CA-B (FP)} task, the most stable prompts are 0-shot CoT with a \textit{prompt volatility} of 5.7 and 0-shot IcL with a \textit{prompt volatility} of 6.9, while the most variable prompts are adversarial ignore and 3-shot IcL, both at 9.9. The \textit{AMPGV} is 6.5, suggesting a \textbf{medium} level of \textit{prompt dependence}.
\textbf{After evaluation of all the tasks in the scenario}, it has been determined that the \textit{AMPGV}, which reflect \textit{variance of prompt dependence} in the scenario, exhibit a \textbf{narrow} spectrum ranging from 6.5 to 11.0. Moreover, 0-shot CoT and 0-shot IcL demonstrate stability in both tasks, whereas 3-shot IcL and manual CoT are the least stable prompts in the two tasks, which is the same as the conclusion in the scenario aspect.

Finally, we measure \textit{language proficiency} of CA:

1) \textbf{English vs. Chinese}: Figure \ref{fig:Causal Attribution Language} underscores that models generally exhibit superior performance on the English test set compared to the Chinese one, with 16 out of 28 models demonstrating better results in English.

2) \textbf{Accuracy Difference}: There are significant differences in performance favoring English, as seen in models such as text-davinci-001 (Difference of 10.5\%), GPT-3.5-Turbo (9.4\%), and Llama2 (70B) (7.0\%). On the flip side, some models, including Baichuan1-chat (13B) (26.5\%), Llama2 (13B) (23.6\%), and Vicuna-v1.3 (33B) (22.8\%), perform better in Chinese.

Moreover, regarding \textbf{CA-B (FA)} task, it can be observed that the accuracy of ``72.4\%'' appears frequently in this task, which is due to the ``Yes:No'' ratio of the dataset being 72.4:27.6. All accuracies of ``72.4\%'' are results of answering ``Yes'' to all questions, and conversely, accuracies of ``27.6\%'' are from answering ``No'' to all. This is why the overall performance of models on this task seems better. However, it could be described as a ``false prosperity'', as whether the models truly understand and are capable of solving this problem remains questionable. As to the \textbf{CA-B (FP)} task, we find that over 65\% of the responses from \textdavincithree~and \chatgpt~are ``Yes''. The order of answers in the examples provided by 3-shot IcL is ``No-Yes-No'', leading us to speculate whether these two models are attempting to learn such a pattern, thus tending to output ``Yes'' more in the fourth question.

\subsubsection{Association}
\label{scenario:association}
\begin{figure}[t]
\centering  
\subfigure[Distribution of EAE]{ 
\begin{minipage}{3.9cm}
\centering   
    \includegraphics[width=1\linewidth]{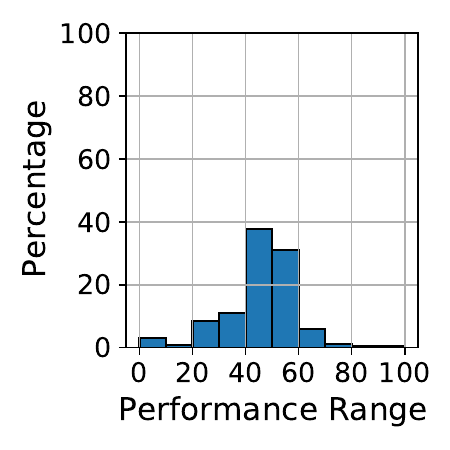}
\end{minipage}
}
\hspace{2cm}
\subfigure[Distribution of correlation]{ 
\begin{minipage}{3.9cm}
\centering    
    \includegraphics[width=1\linewidth]{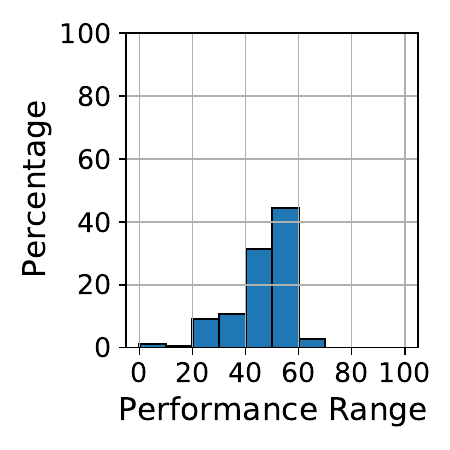}
\end{minipage}
}
\caption[Distribution of association]{\textbf{Distribution of association.} The horizontal coordinate represents the accuracy of the model, and the vertical coordinate represents the percentage distribution corresponding to a certain accuracy interval.}   
\label{fig:Distribution_of_Association}   
\end{figure}

\paragraph{Explaining away effect.}
\begin{figure}[t]
\centering
\subfigure[Model performance of EAE]{
\begin{minipage}{8.5cm}
\centering
\includegraphics[width=1\linewidth]{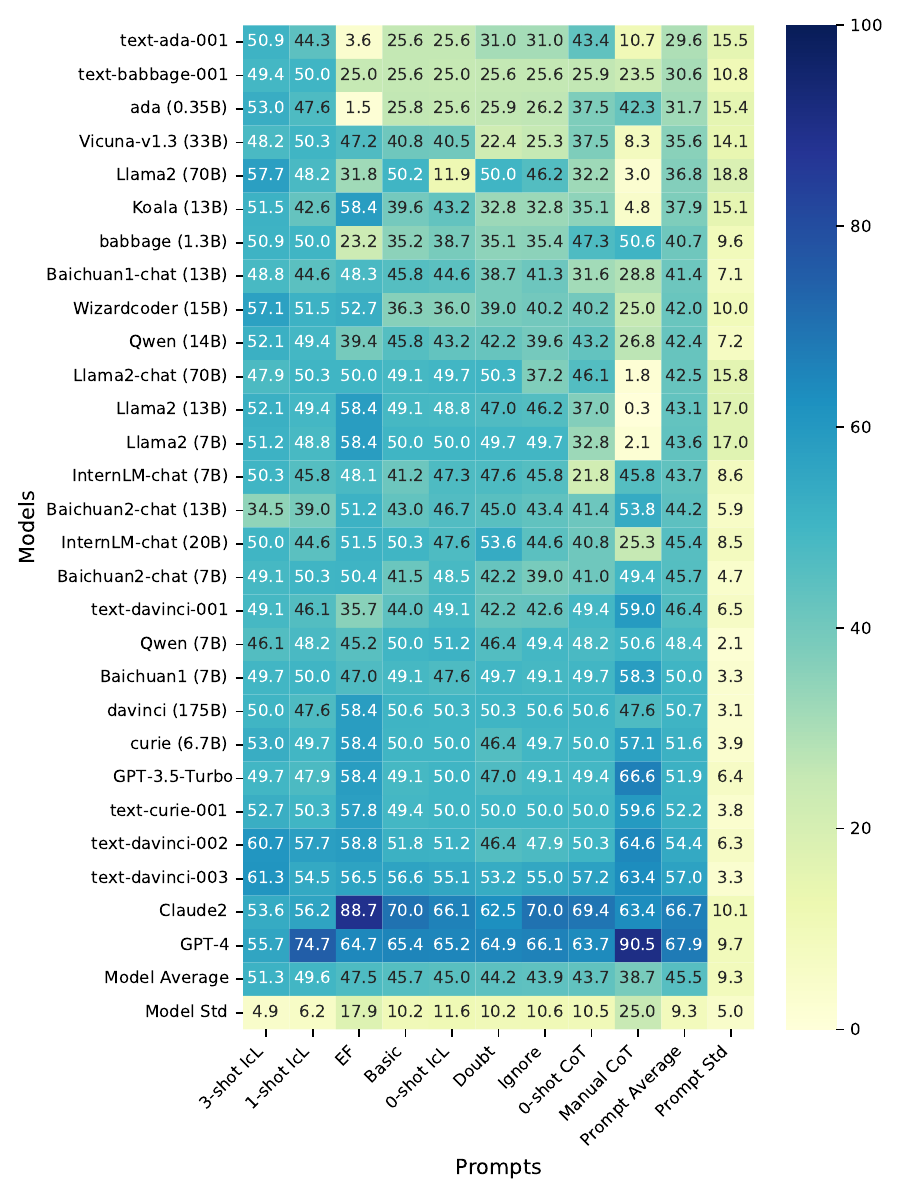}
\end{minipage}
}
\subfigure[\textit{Prompt gain} of EAE]{
\begin{minipage}{8.5cm}
\centering
\includegraphics[width=1\linewidth]{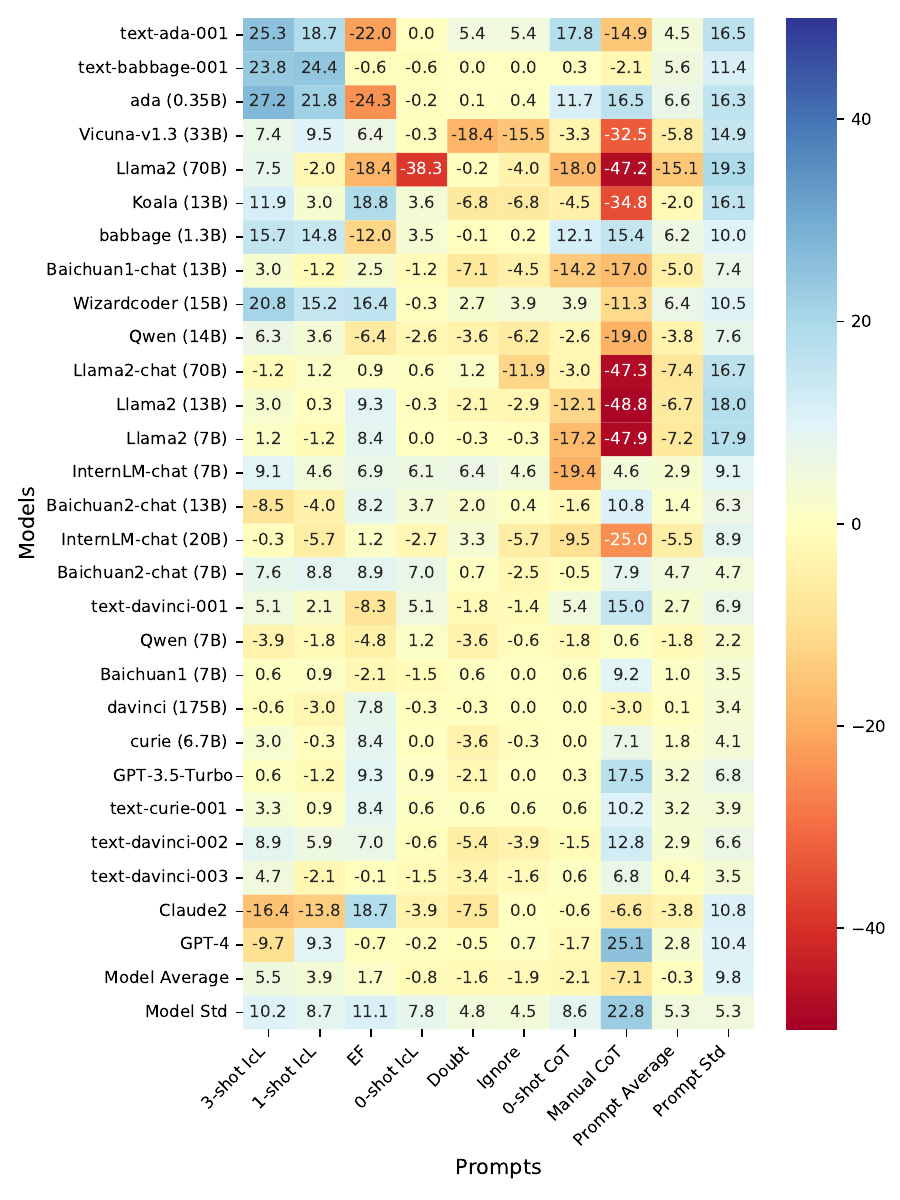}
\end{minipage}
}
\caption[Heatmap of EAE]{\textbf{Heatmap of EAE.} The models and prompts are sorted by their averages.}
\label{fig:Heatmap_of_Explaining_Away_Effect}
\end{figure}

\begin{figure}
    \centering
    \includegraphics[width=0.8\linewidth]{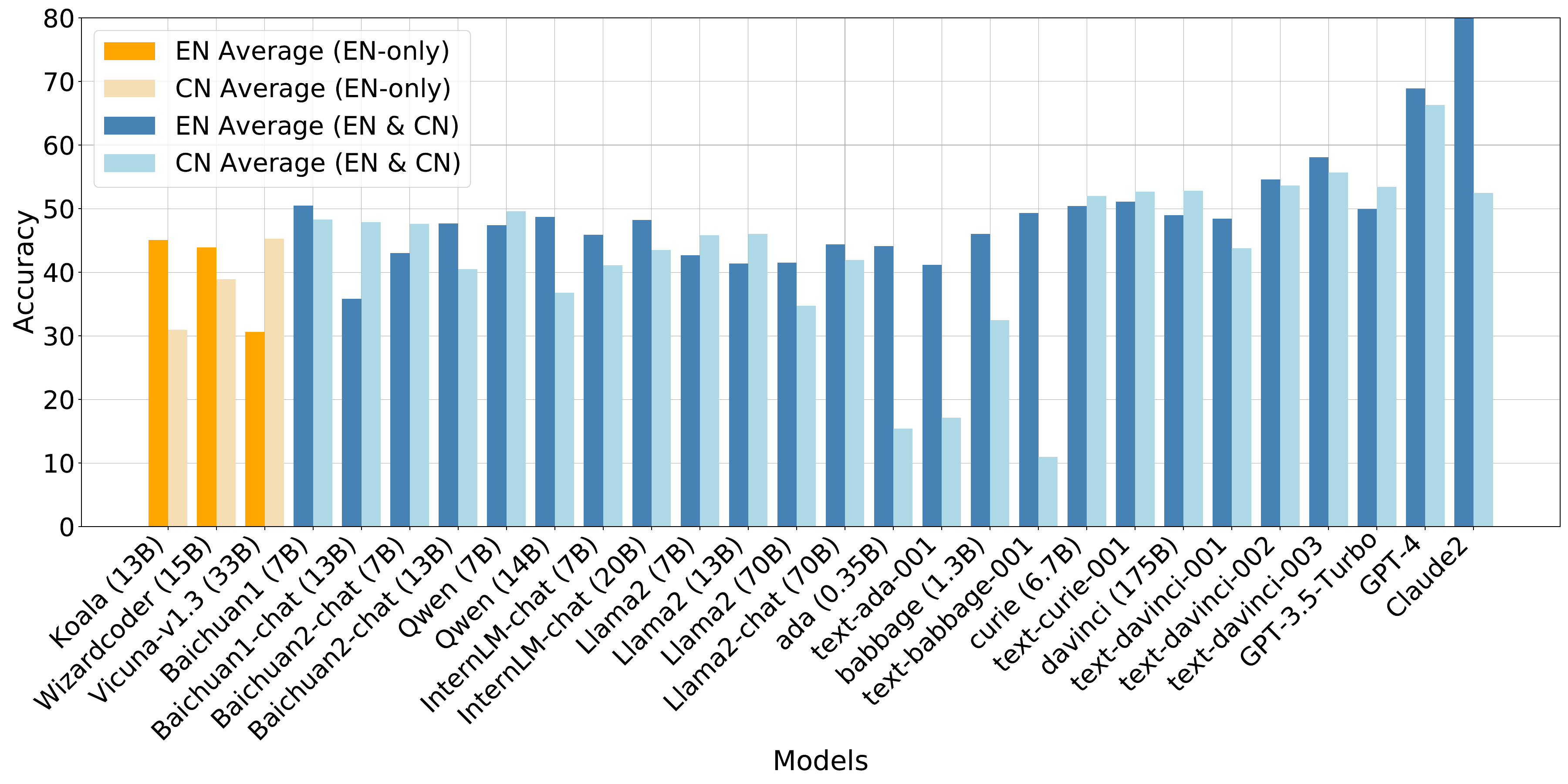}
    \caption[Language comparison of EAE]{\textbf{Language comparison of EAE.} The dark legend signifies the average performance of the model on an English test set, whereas the light legend denotes the average performance of the model on the Chinese test set. The yellow legend indicates a model trained exclusively on English datasets, while the blue legend represents a model trained on both English and Chinese datasets.}
    \label{fig:Explaining_Away_Effect_Language}
\end{figure}

Regarding model performance in EAE:
1) \textbf{Distribution}: Figure \ref{fig:Distribution_of_Association}(a) illustrates the distribution for all \textit{model-prompt pair}s within the EAE, noting a median of 48.8\% and a third quartile of 50.7\%. This suggests that the \textit{understandability} of the scenario is \textbf{hard}, with the median accuracy falling below the random guess benchmark of 50.0\%.
2) \textbf{Top Accuracy}: Figure \ref{fig:Heatmap_of_Explaining_Away_Effect}(a) identifies GPT-4 at 67.9\%, Claude2 at 66.7\%, and text-davinci-003 at 57.0\% as the top three models by average accuracy. As to the \textit{top model-prompt pair}, GPT-4, through the use of manual CoT, achieves a remarkable 90.5\%, indicating the \textit{solvability} of the scenario is \textbf{potentially solvable} as the \textit{top model-prompt pair}'s performance surpasses 80\%.
3) \textbf{Stability}: The models which are most sensitive to prompt variations, as indicated by the \textit{model volatility} described in \cref{metric:model}, are Llama2 (70B) at 18.8, Llama2 (13B) at 17.0, and Llama2 (7B) at 17.0. Conversely, the most stable models include Qwen (7B) at 2.1, davinci (175B) at 3.1, and Baichuan1 (7B) at 3.3.
4) Open-Limited Ratio: Among the top five models with the highest average accuracy, there is a \textbf{large} \textit{open-limited gap} between open-access and limited-access models, with a ratio of 0:5.

Regarding \textit{prompt gain} in EAE:
1) \textbf{Top Gain}: Figure \ref{fig:Heatmap_of_Explaining_Away_Effect}(b) shows the top two prompts for average accuracy gain over the basic prompt as 3-shot IcL at 5.5\% and 1-shot IcL at 3.9\%, with ada (0.35B) using 3-shot IcL witnessing the most substantial improvement of 27.2\%.
2) \textbf{Exceptions}: The top-performing prompt, 3-shot IcL, cannot give a positive average \textit{prompt gain} to several models, including Llama2-chat (70B), Baichuan2-chat (13B), and GPT-4. On the other hand, all prompts are capable of enhancing text-curie-001's performance above the basic prompt.
3) \textbf{Stability}: The most stable prompts, with the lowest \textit{prompt volatility}, are adversarial ignore at 4.5 and adversarial doubt at 4.8, while the least stable prompts are manual CoT at 22.8 and EF at 11.1. The \textit{average model-prompt-gain volatility} (\textit{AMPGV}) is 9.8, demonstrating a \textbf{medium} \textit{prompt dependence} within the scenario.

In terms of \textit{language proficiency} in EAE:
1) \textbf{English vs. Chinese}: Figure \ref{fig:Explaining_Away_Effect_Language} reveals models perform better on the English test set over the Chinese set, with 18 of 28 models favoring English.
2) \textbf{Accuracy Difference}: Significant performance differences favoring English are seen in models like text-babbage-001 (38.3\%), Claude2 (29.0\%), and ada (0.35B) (28.7\%). Conversely, models such as Vicuna-v1.3 (33B), Baichuan1-chat (13B), and Baichuan2-chat (7B) demonstrate higher proficiency in Chinese.

\paragraph{Correlation.}
\begin{figure}[t]
\centering
\subfigure[Model performance of CORR]{
\begin{minipage}{8.5cm}
\centering
\includegraphics[width=1\linewidth]{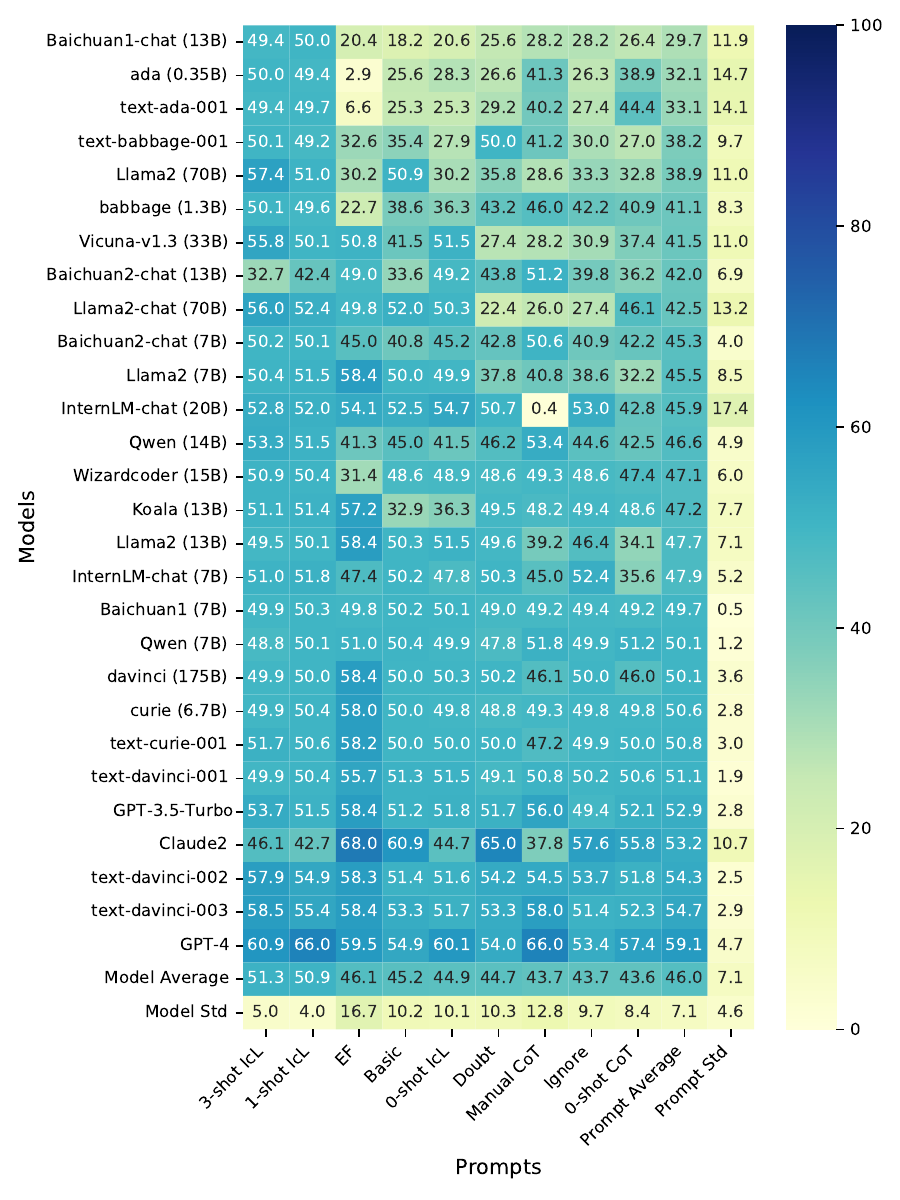}
\end{minipage}
}
\subfigure[\textit{Prompt gain} of CORR]{
\begin{minipage}{8.5cm}
\centering
\includegraphics[width=1\linewidth]{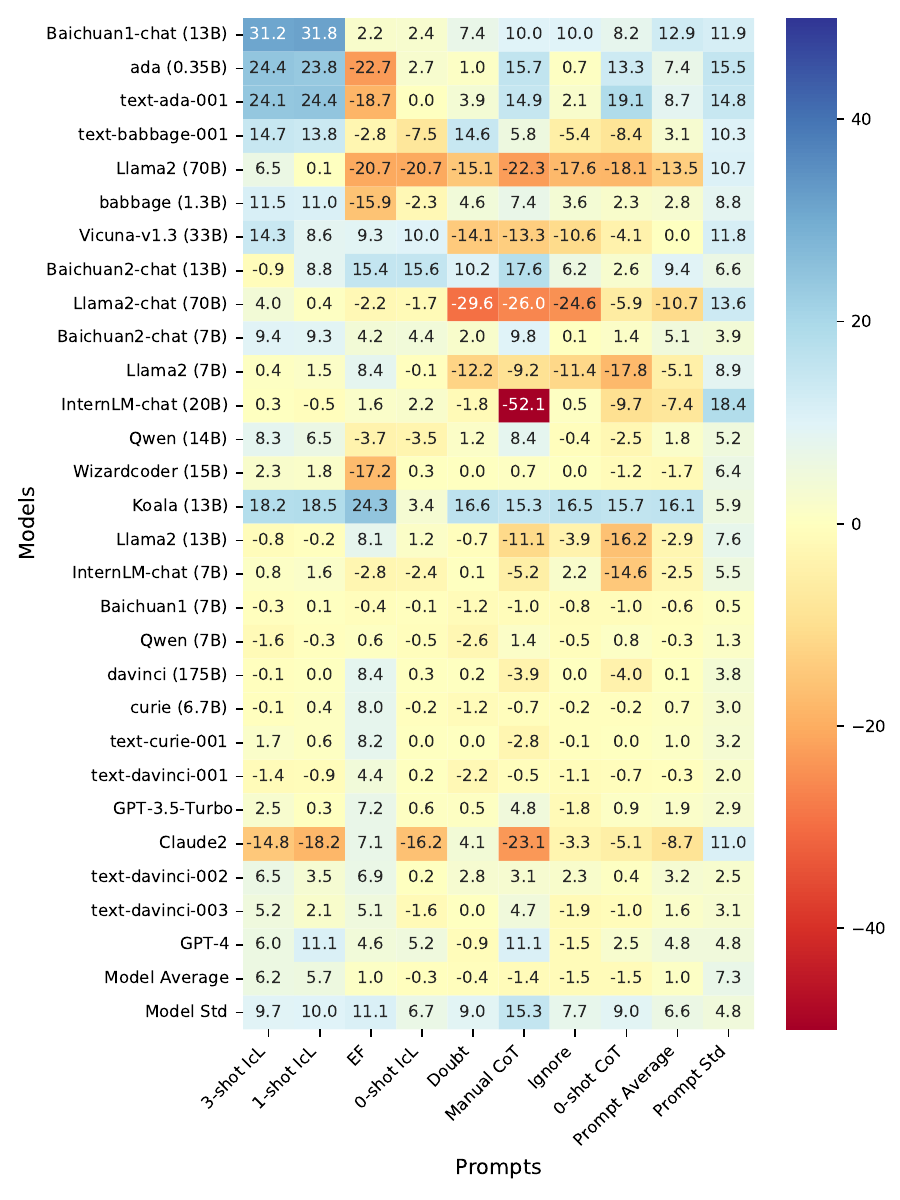}
\end{minipage}
}
\caption[Heatmap of CORR]{\textbf{Heatmap of CORR.} The models and prompts are sorted by their averages.}
\label{fig:Heatmap_of_Correlation}
\end{figure}

\begin{figure}
    \centering
    \includegraphics[width=0.8\linewidth]{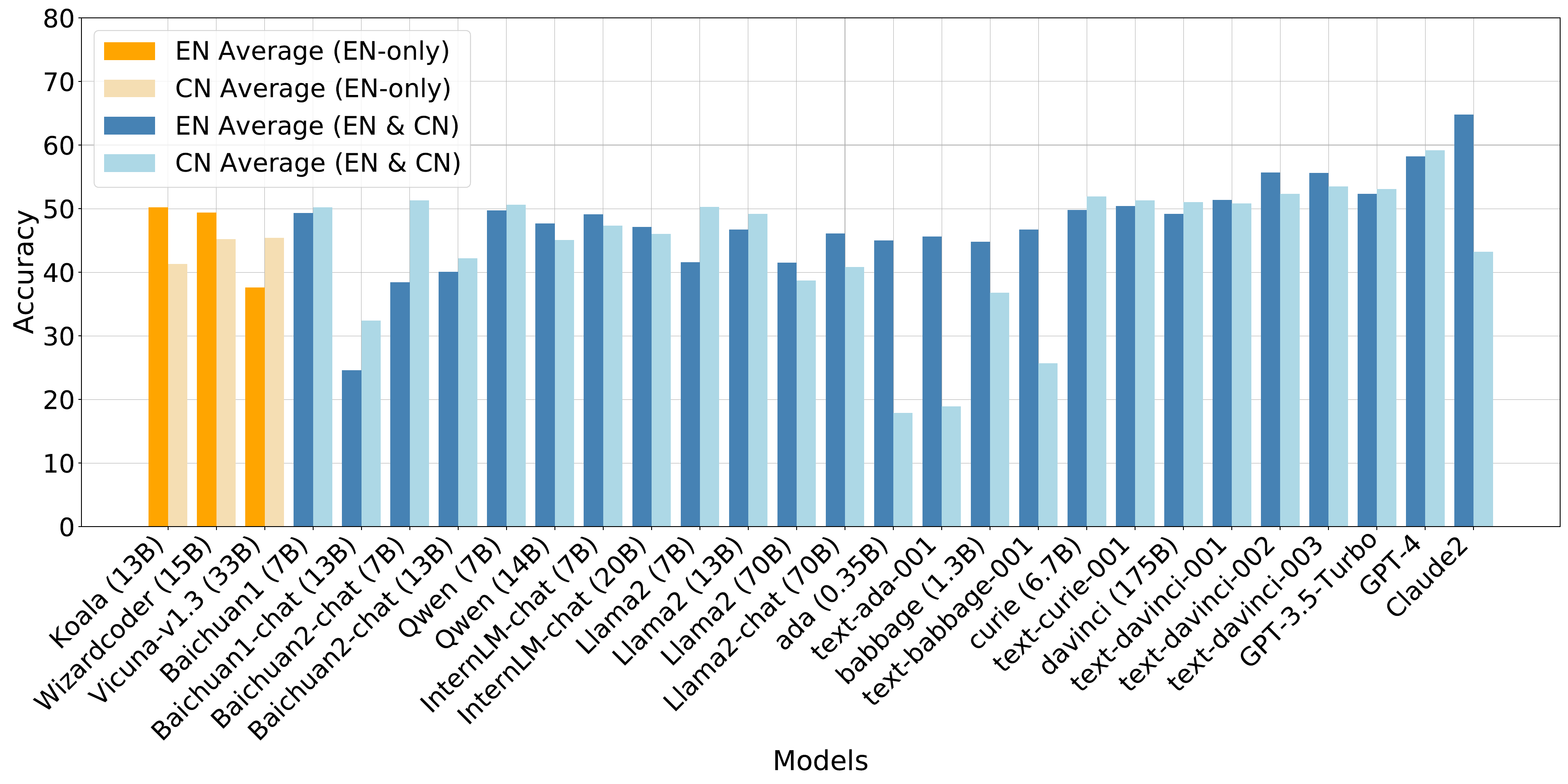}
    \caption[Language comparison of CORR]{\textbf{Language comparison of CORR.} The dark legend signifies the average performance of the model on an English test set, whereas the light legend denotes the average performance of the model on the Chinese test set. The yellow legend indicates a model trained exclusively on English datasets, while the blue legend represents a model trained on both English and Chinese datasets.}
    \label{fig:Correlation_Language}
\end{figure}

In terms of model performance in CORR:
1) \textbf{Distribution}: Figure \ref{fig:Distribution_of_Association}(b) shows the distribution for all \textit{model-prompt pair}s in CORR, indicating a median of 49.9\% and a third quartile of 51.5\%. This indicates that the \textit{understandability} of the scenario is \textbf{hard}, as the median accuracy falls below the random guess benchmark of 50.0\%.
2) \textbf{Top Accuracy}: According to Figure \ref{fig:Heatmap_of_Correlation}(a), the leading three models by average accuracy are GPT-4 at 59.1\%, text-davinci-003 at 54.7\%, and text-davinci-002 at 54.3\%. Claude2, using EF, stands out with a top score of 68.0\%, illustrating the scenario \textit{solvability} as \textbf{challenging} since the \textit{top model-prompt pair}'s performance does not reach 80\%.
3) \textbf{Stability}: The models most affected by prompt variability, as shown by the \textit{model volatility} described in \cref{metric:model}, are InternLM-chat (20B) at 17.4, ada (0.35B) at 14.7, and text-ada-001 at 14.1. Conversely, the most stable models include Baichuan1 (7B) at 0.5, Qwen (7B) at 1.2, and text-davinci-001 at 1.9.
4) Open-Limited Ratio: Among the top five models with the highest average accuracy, which have a 0:5 ratio of open-access to limited-access models, there exists a \textbf{large} \textit{open-limited gap}.

In terms of \textit{prompt gain} in CORR:
1) \textbf{Top Gain}: Figure \ref{fig:Heatmap_of_Correlation}(b) highlights the top two prompts for average accuracy gain over the basic prompt as 3-shot IcL at 6.2\% and 1-shot IcL at 5.7\%. Baichuan1-chat (13B) using 1-shot IcL demonstrates the most significant increase of 31.8\%.
2) \textbf{Exceptions}: The most effective prompt, 3-shot IcL, does not generate a positive average \textit{prompt gain} for several models, including Baichuan2-chat (13B) and Claude2. Yet, all prompts manage to enhance the performance for models like Baichuan1-chat (13B) and text-davinci-002 over the basic prompt.
3) \textbf{Stability}: The most stable prompts, with the lowest \textit{prompt volatility}, are 0-shot IcL at 6.7 and adversarial ignore at 7.7. The least stable prompts, with the highest \textit{prompt volatility}, are manual CoT at 15.3 and EF at 11.1. The \textit{average model-prompt-gain volatility} (\textit{AMPGV}) is 7.3, indicating a \textbf{medium} \textit{prompt dependence} in this scenario.

Regarding \textit{language proficiency} in CORR:
1) \textbf{English vs. Chinese}: Figure \ref{fig:Correlation_Language} reveals that models generally achieve better results on the English test set than on the Chinese set, with 15 of 28 models favoring English.
2) \textbf{Accuracy Difference}: Significant advantages for English over Chinese are seen in models like ada (0.35B) at 27.1\% and Claude2 at 21.6\%. In contrast, models like Baichuan2-chat (7B) and Llama2 (7B) exhibit higher proficiency in Chinese.

\subsubsection{Intervention}
\label{scenario:intervention}
\begin{figure}[t]
\centering 
\subfigure[Distribution of ATE]{  
\begin{minipage}{3.9cm}
\centering    
    \includegraphics[width=1\linewidth]{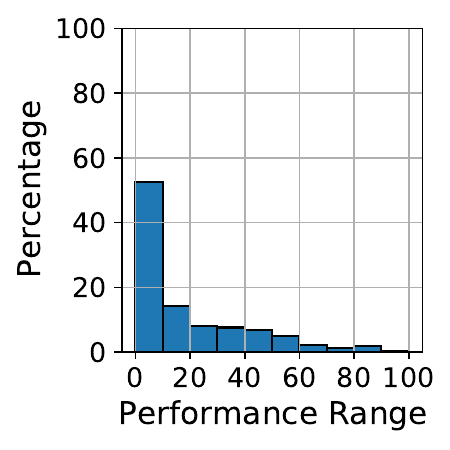}
\end{minipage}
}
\subfigure[Distribution of CDE]{  
\begin{minipage}{3.9cm}
\centering   
    \includegraphics[width=1\linewidth]{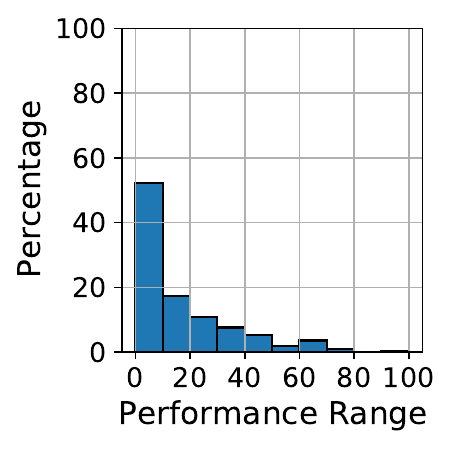}
\end{minipage}
}
\subfigure[Distribution of CEI]{  
\begin{minipage}{3.9cm}
\centering   
    \includegraphics[width=1\linewidth]{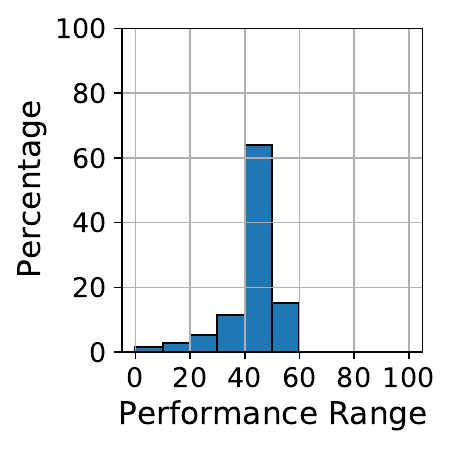}
\end{minipage}
}
\subfigure[Distribution of BAS]{  
\begin{minipage}{3.9cm}
\centering  
    \includegraphics[width=1\linewidth]{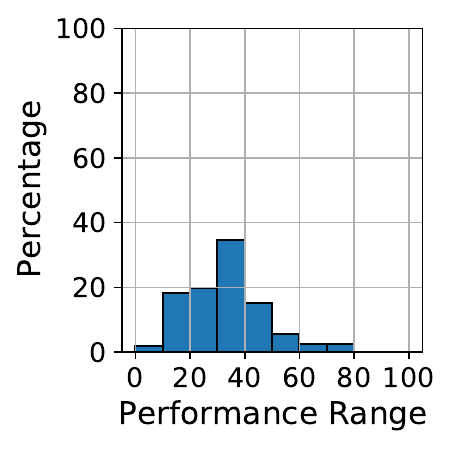}
\end{minipage}
}
\subfigure[Distribution of FAS]{  
\begin{minipage}{3.9cm}
\centering   
    \includegraphics[width=1\linewidth]{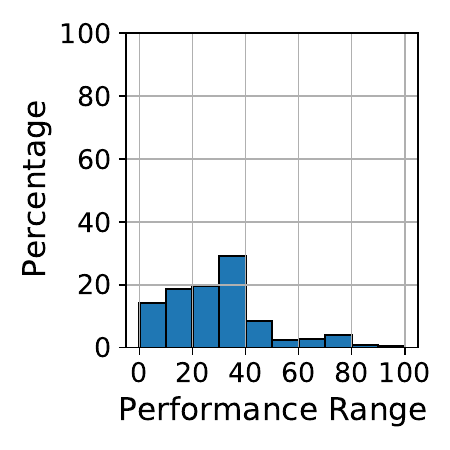}
\end{minipage}
}
\subfigure[Distribution of IV]{ 
\begin{minipage}{3.9cm}
\centering   
    \includegraphics[width=1\linewidth]{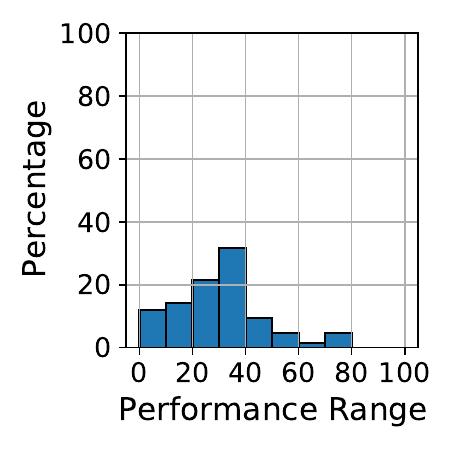}
\end{minipage}
}
\subfigure[Distribution of CB]{ 
\begin{minipage}{3.9cm}
\centering   
    \includegraphics[width=1\linewidth]{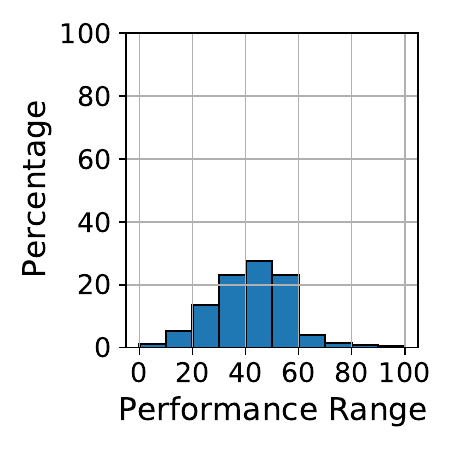}
\end{minipage}
}
\caption[Distribution of intervention]{\textbf{Distribution of intervention.} The horizontal coordinate represents the accuracy of the model and the vertical coordinate represents the percentage distribution corresponding to a certain accuracy interval.}    
\label{fig:Distribution_of_Intervention}    
\end{figure}

\paragraph{Average treatment effect.}

\begin{figure}[t]
\centering
\subfigure[Model performance of ATE]{
\begin{minipage}{8.5cm}
\centering
\includegraphics[width=1\linewidth]{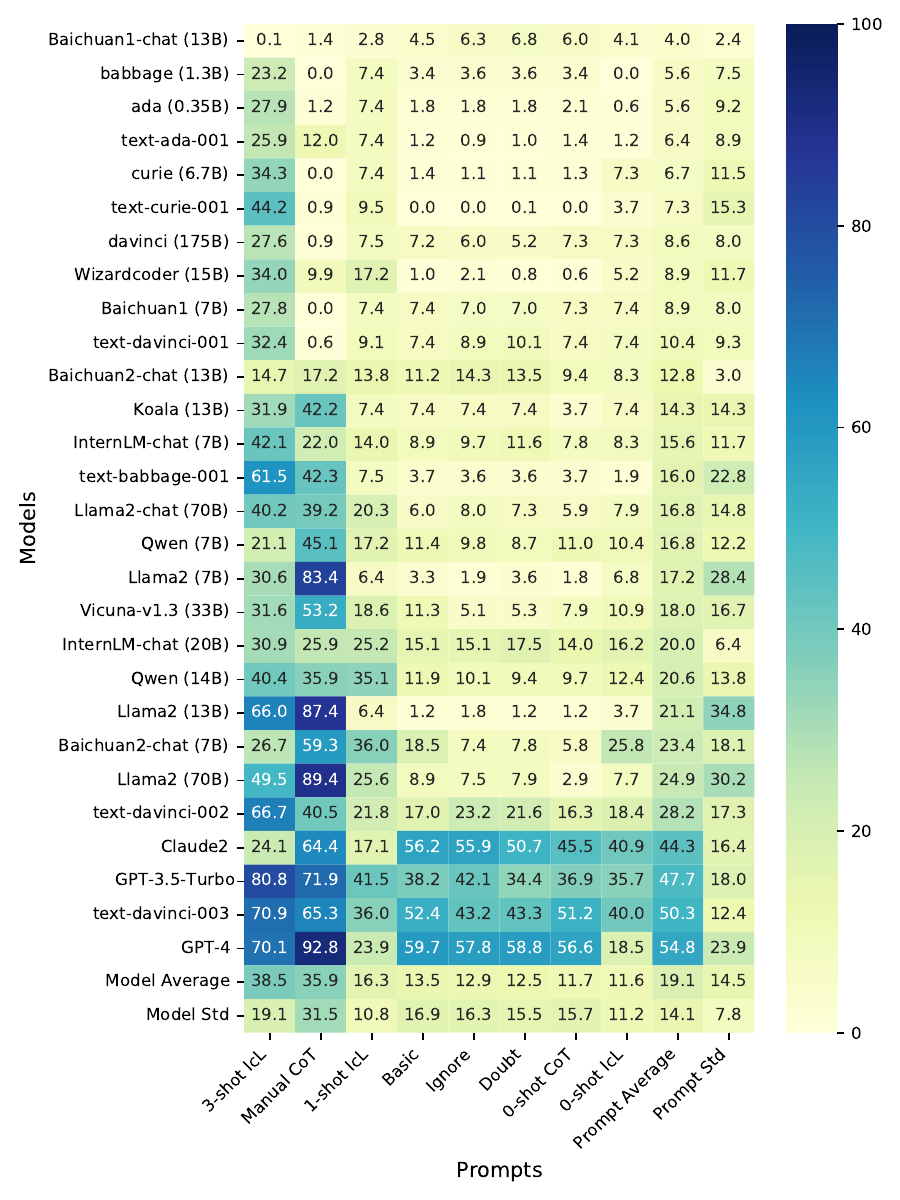}
\end{minipage}
}
\subfigure[\textit{Prompt gain} of ATE]{
\begin{minipage}{8.5cm}
\centering
\includegraphics[width=1\linewidth]{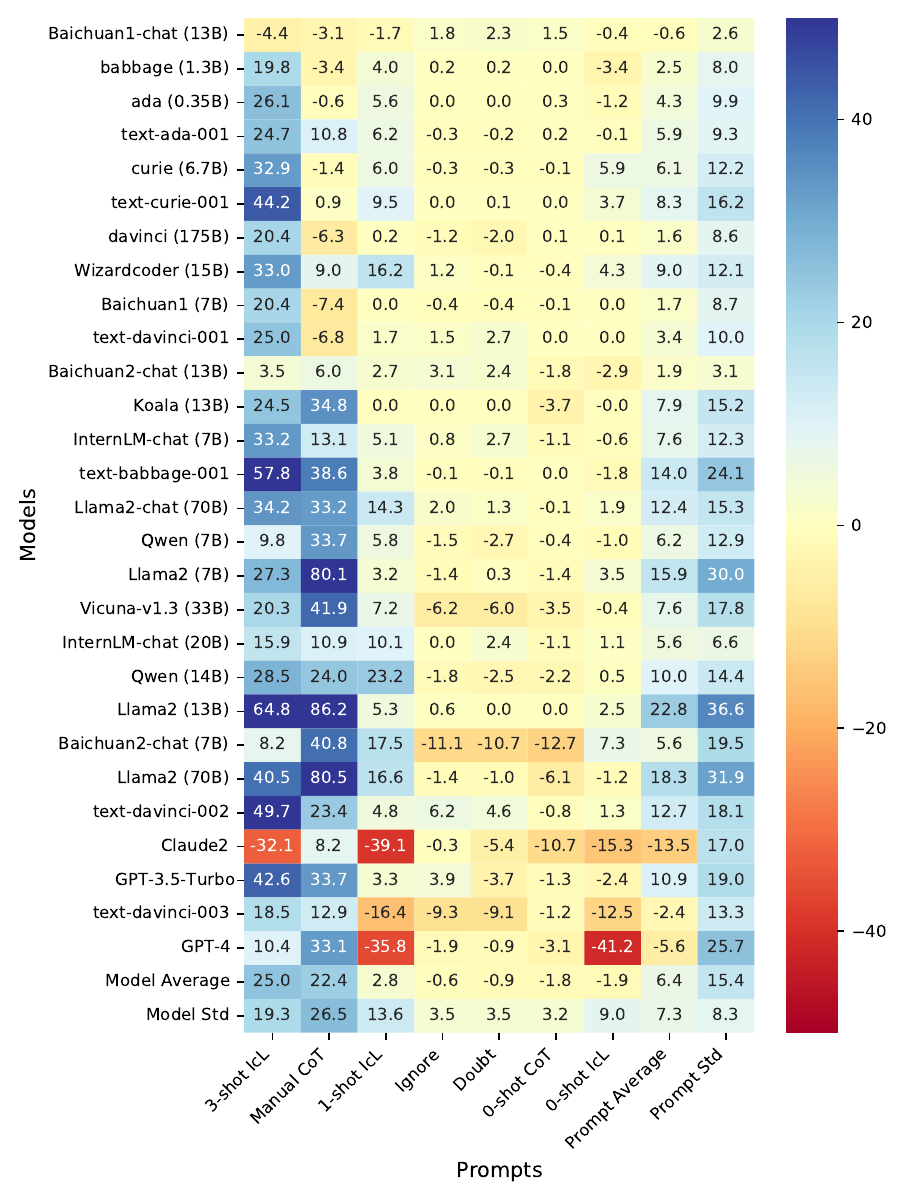}
\end{minipage}
}
\caption[Heatmap of ATE]{\textbf{Heatmap of ATE.} The models and prompts are sorted by their averages.}
\label{fig:Heatmap_of_Average_Treatment_Effect}
\end{figure}

\begin{figure}
    \centering
    \includegraphics[width=0.8\linewidth]{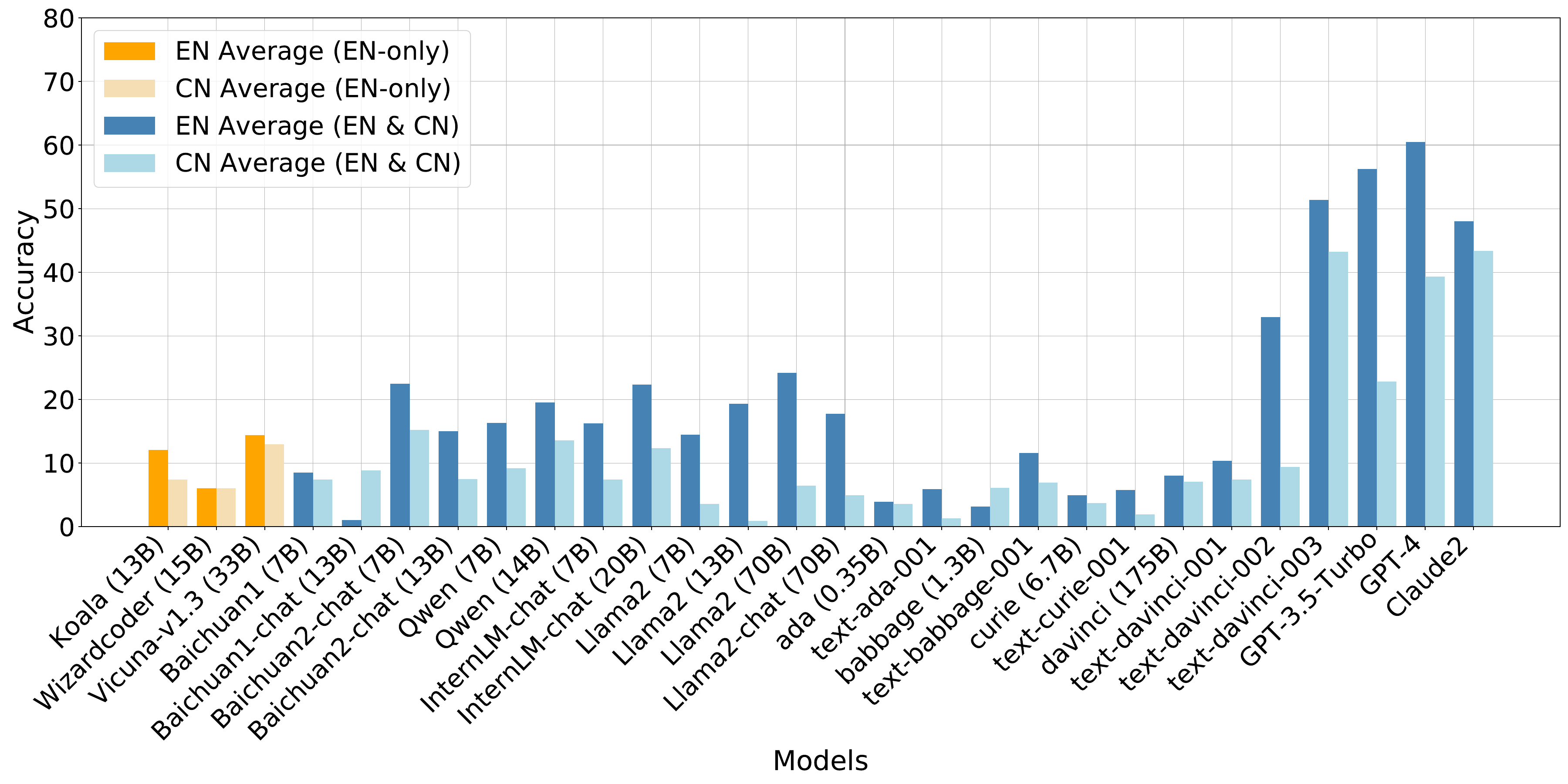}
    \caption[Language comparison of ATE]{\textbf{Language comparison of ATE.} The dark legend signifies the average performance of the model on an English test set, whereas the light legend denotes the average performance of the model on the Chinese test set. The yellow legend indicates a model trained exclusively on English datasets, while the blue legend represents a model trained on both English and Chinese datasets.}
    \label{fig:Average_Treatment_Effect_Language}
\end{figure}

First, we delve into model performance in ATE:

1) \textbf{Distribution}: Figure \ref{fig:Distribution_of_Intervention}(a) showcases the distribution of \textit{model-prompt pair}s for the ATE. With a median value of 9.4\% and a third quartile at 28.6\%, this scenario appears to have a \textbf{hard} \textit{understandability}, given that the median accuracy falls below the random guess accuracy of 16.7\%.
Figure \ref{fig:Distribution_of_Average_Treatment_Effect_Tasks} displays the distribution for each specific causal task.
In the \textbf{ATE-P (ATE-basic)} task, the median is 2.2\%, and the third quartile is 13.4\%. Due to the lower than 15\% third quartile score and the challenging nature of Mathematical-mode tasks, we regard the task \textit{understandability} as \textbf{very hard}.
The \textbf{ATE-P (ATE-hard)} task shows a median of 2.2\% and a third quartile of 10.7\%. Similarly, we define the task \textit{understandability} as \textbf{very hard}.
Moreover, the \textbf{ATE-B (ATE-natural)} task presents a median of 22.2\% and a third quartile of 50.4\%, with a random guess accuracy of 50.0\%, making this task \textit{understandability} as \textbf{hard}.
\textbf{By analyzing the differences between tasks}, we observe that median accuracies vary from 2.2\% to 22.2\% with a standard deviation of 9.4. The third quartile accuracies range from 10.7\% to 50.4\%, accompanied by a standard deviation of 18.1, indicating that the scenario has a \textbf{noticeably diverse} \textit{variance of distribution}. The ATE-B (ATE-natural) task exhibits significantly higher median and third quartile figures than the two Mathematical-mode tasks (ATE-P (ATE-hard) and ATE-P (ATE-basic)). Both the two Mathematical-mode tasks' distributions have the majority of \textit{model-prompt pair}s (over 60\%) in a 0\% to 20\% accuracy range, in stark contrast to the distinctly different distribution for the ATE-B (ATE-natural) task.

2) \textbf{Top Accuracy}: Figure \ref{fig:Heatmap_of_Average_Treatment_Effect}(a) reveals that, the leading models in terms of average accuracy for this scenario are GPT-4 at 54.8\%, text-davinci-003 at 50.3\%, and GPT-3.5-Turbo at 47.7\%. The \textit{top model-prompt pair} is GPT-4 with manual CoT, reaching an impressive 92.8\%, indicating the scenario's \textit{solvability} is \textbf{potentially solvable} given that the \textit{top model-prompt pair} exceeds an 80\% performance mark. Figure \ref{fig:Heatmap_of_performances_of_Average_Treatment_Effect} details the top three models' average accuracy, analyzed task by task.
In the \textbf{ATE-P (ATE-basic)} task, the highest average accuracies are observed with GPT-4 at 43.4\%, text-davinci-003 at 43.0\%, and GPT-3.5-Turbo at 42.3\%, with Llama2 (70B) and manual CoT leading at 92.3\%, showcasing the task's \textit{solvability} as \textbf{potentially solvable} as the \textit{top model-prompt pair} scores above 80\%. For \textbf{ATE-P (ATE-hard)}, GPT-4 leads with 50.2\%, followed by text-davinci-003 at 46.2\%, and GPT-3.5-Turbo at 41.9\%, where GPT-4 and manual CoT top at 89.1\%, affirming the task's \textit{solvability} as \textbf{potentially solvable} with the \textit{top model-prompt pair} also exceeding 80\%. In \textbf{ATE-B (ATE-natural)}, the top averages belong to GPT-4 at 67.5\%, Claude2 at 65.1\%, and text-davinci-003 at 58.8\%, with GPT-4 and manual CoT achieving the highest at 98.8\%, further confirming the \textit{solvability} of the task is \textbf{potentially solvable} as the \textit{top model-prompt pair} surpasses 80\%.
\textbf{Through comparing different tasks}, the \textit{variance of solvability} appears \textbf{negligible}. Yet, the top model's average accuracy spans from 43.4\% to 67.5\% (a 24.1\% difference), and the peak accuracy for \textit{top model-prompt pair}s varies from 89.1\% to 98.8\% (a 9.7\% difference), indicating a \textbf{significant} \textit{variance of model's top performance} across the scenario. Notably, GPT-4 consistently excels in average performance across all tasks and is a part of the \textit{top model-prompt pair} in two out of three tasks. Manual CoT is part of the \textit{top model-prompt pair}s in all tasks.

3) \textbf{Stability}: The three most stable models, indicated by the lowest \textit{model volatility}, are Baichuan1-chat (13B) at 2.4, Baichuan2-chat (13B) at 3.0, and InternLM-chat (20B) at 6.4. Conversely, the three models exhibiting the greatest instability across various prompts, shown by the highest \textit{model volatility}, are Llama2 (13B) at 34.8, Llama2 (70B) at 30.2, and Llama2 (7B) at 28.4. Next, we analyze stability on a task-specific basis.
In the \textbf{ATE-P (ATE-basic)} task, the most stable models are babbage (1.3B) at 0.1, Baichuan1 (7B) at 1.0, and ada (0.35B) at 1.4. The most unstable models, with the largest \textit{model volatility}, are Llama2 (13B) at 30.6, Llama2 (70B) at 30.4, and Llama2 (7B) at 24.2.
In the \textbf{ATE-P (ATE-hard)} task, the three most stable models are babbage (1.3B) at 0.0, davinci (175B) at 0.1, and ada (0.35B) at 0.8. The models with the highest \textit{model volatility}, indicating instability across prompts, are Llama2 (13B) at 35.6, GPT-4 at 31.8, and Llama2 (7B) at 30.4.
In the \textbf{ATE-B (ATE-natural)} task, the top stable models are Baichuan1-chat (13B) at 2.8, Baichuan2-chat (13B) at 4.6, and InternLM-chat (20B) at 12.2. The most unstable models are Llama2 (13B) at 35.1, Llama2 (7B) at 33.2, and Llama2 (70B) at 32.4.
\textbf{In all tasks}, Llama2 (13B) and Llama2 (7B) rank among the top three most unstable models, highlighting their prompt sensitivity in this scenario. Notably, GPT-4 is found to be as the second most unstable model in the \textbf{ATE-P (ATE-hard)} task, pointing to an anomaly in this otherwise high-performing model.

4) \textbf{Open-Limited Ratio}: The 0:5 ratio of open-access to limited-access models among the top 5 models in the entire scenario underscores a \textbf{large} \textit{open-limited gap}.

Next, we delve into \textit{prompt gain} in ATE:

1) \textbf{Top Gain}: As illustrated in Figure \ref{fig:Heatmap_of_Average_Treatment_Effect}(b), the two prompts leading in average accuracy gain relative to the basic prompt are 3-shot IcL at 25.0\% and manual CoT at 22.4\%. The most significant improvement in accuracy compared to the basic prompt is seen with Llama2 (13B) using manual CoT, achieving an 86.2\% increase. We then proceed with an in-depth task-specific analysis. Figure \ref{fig:Heatmap_of_gain_of_Average_Treatment_Effect} displays the gain heatmaps for every task within the scenario. In the \textbf{ATE-P (ATE-basic)} task, manual CoT at 25.9\% and 3-shot IcL at 16.4\% are the top two prompts by average accuracy gain over the basic prompt, with Llama2 (70B) and manual CoT marking the most substantial improvement at 91.7\%. In the \textbf{ATE-P (ATE-hard)} task, manual CoT at 23.1\% and 3-shot IcL at 15.7\% lead in average accuracy gains, with Llama2 (7B) and manual CoT showing the highest increase at 86.3\%. In the \textbf{ATE-B (ATE-natural)} task, the two prompts with the greatest average accuracy gains are 3-shot IcL at 42.9\% and manual CoT at 18.1\%, with Llama2 (13B) and manual CoT demonstrating the most significant boost at 91.2\%.
\textbf{On the evaluation across tasks}, 3-shot IcL and manual CoT are consistently found to be the top two prompts by largest average gain in each task. Additionally, the most extensive gains across all \textit{model-prompt pair}s are found in the llama-series utilizing manual CoT, with the highest accuracy gain surpassing 90\%.

2) \textbf{Exceptions}: The leading prompt, 3-shot IcL, give positive average \textit{prompt gain} to most models in the scenario, with exceptions being Baichuan1-chat (13B) and Claude2. Nevertheless, every prompt manages to elevate Llama2 (13B)'s performance. In the \textbf{ATE-P (ATE-basic)} task, the most effective prompt, manual CoT, does not work on Baichuan1-chat (13B) in generating positive average prompt gain. However, all prompts are capable of boosting Llama2-chat (70B)'s performance. Within the \textbf{ATE-P (ATE-hard)} task, manual CoT, despite being the top prompt, fails to create a positive average \textit{prompt gain} for Baichuan1-chat (13B) and text-davinci-003. In the \textbf{ATE-B (ATE-natural)} task, the best prompt, 3-shot IcL, shows no effectiveness on Baichuan1-chat (13B), Baichuan2-chat (13B), and Claude2. Yet, every prompt succeeds in generating a positive average \textit{prompt gain} on Llama2 (13B).

3) \textbf{Stability}: Regarding stability, the two most stable prompts are 0-shot CoT with a \textit{prompt volatility} of 3.2 and adversarial doubt with a \textit{prompt volatility} of 3.5. On the opposite end, the two prompts demonstrating the greatest volatility, indicated by the highest \textit{prompt volatility}, are manual CoT at 26.5 and 3-shot IcL at 19.3. The \textit{average model-prompt-gain volatility} (\textit{AMPGV}) is 15.4, suggesting a \textbf{high} \textit{prompt dependence}. We then assess stability across individual tasks.
In the \textbf{ATE-P (ATE-basic)} task, 0-shot CoT with a \textit{prompt volatility} of 1.3 and adversarial ignore with a \textit{prompt volatility} of 2.0 is the most stable, whereas manual CoT at 27.1 and 3-shot IcL at 19.3 is the least stable. The task has a \textbf{high} \textit{prompt dependence} with an \textit{AMPGV} of 12.6.
For the \textbf{ATE-P (ATE-hard)} task, the most stable prompts are adversarial doubt and adversarial ignore, with \textit{prompt volatility} of 2.9 and 3.1, respectively, while manual CoT and 3-shot IcL are the least stable, with \textit{prompt volatility} of 27.8 and 23.5. The task shows a \textbf{high} \textit{prompt dependence}, as reflected by \textit{AMPGV} of 14.1.
In the \textbf{ATE-B (ATE-natural)} task, adversarial ignore and adversarial doubt both present the lowest volatility with \textit{prompt volatility} of 8.2, in stark contrast to 3-shot IcL and manual CoT, which are the most unstable with \textit{prompt volatility} of 30.1 and 28.9, respectively. The task results in a \textbf{high} \textit{prompt dependence}, with an \textit{AMPGV} of 21.4.
\textbf{After evaluating all tasks in the scenario}, it is observed that the distribution of \textit{AMPGV}, which highlights the \textit{variance of prompt dependence}, have a \textbf{moderate spread} from 12.6 to 21.4. Differing from the scenario view, adversarial ignore and adversarial doubt consistently rank as the most stable prompts on a task-by-task analysis. While 3-shot IcL and manual CoT rank highest in \textit{prompt gain}, they also exhibit the most instability across all tasks.

At last, we analyze \textit{language proficiency} in ATE:

1) \textbf{English vs. Chinese}: Figure \ref{fig:Average_Treatment_Effect_Language} underscores that models generally exhibit superior performance on the English test set compared to the Chinese one, with 25 out of 28 models demonstrating better results in English.

2) \textbf{Accuracy Difference}: Significant performance gaps between English and Chinese, with a preference for English, appears in models such as GPT-3.5-Turbo (33.4\%), text-davinci-002 (23.5\%), and GPT-4 (21.1\%). On the other hand, models like Baichuan1-chat (13B) (7.8\%), and babbage (1.3B) (3.0\%) exhibit a higher proficiency in Chinese.

\paragraph{Controlled direct effect.}

\begin{figure}[t]
\centering
\subfigure[Model performance of CDE]{
\begin{minipage}{8.5cm}
\centering
\includegraphics[width=1\linewidth]{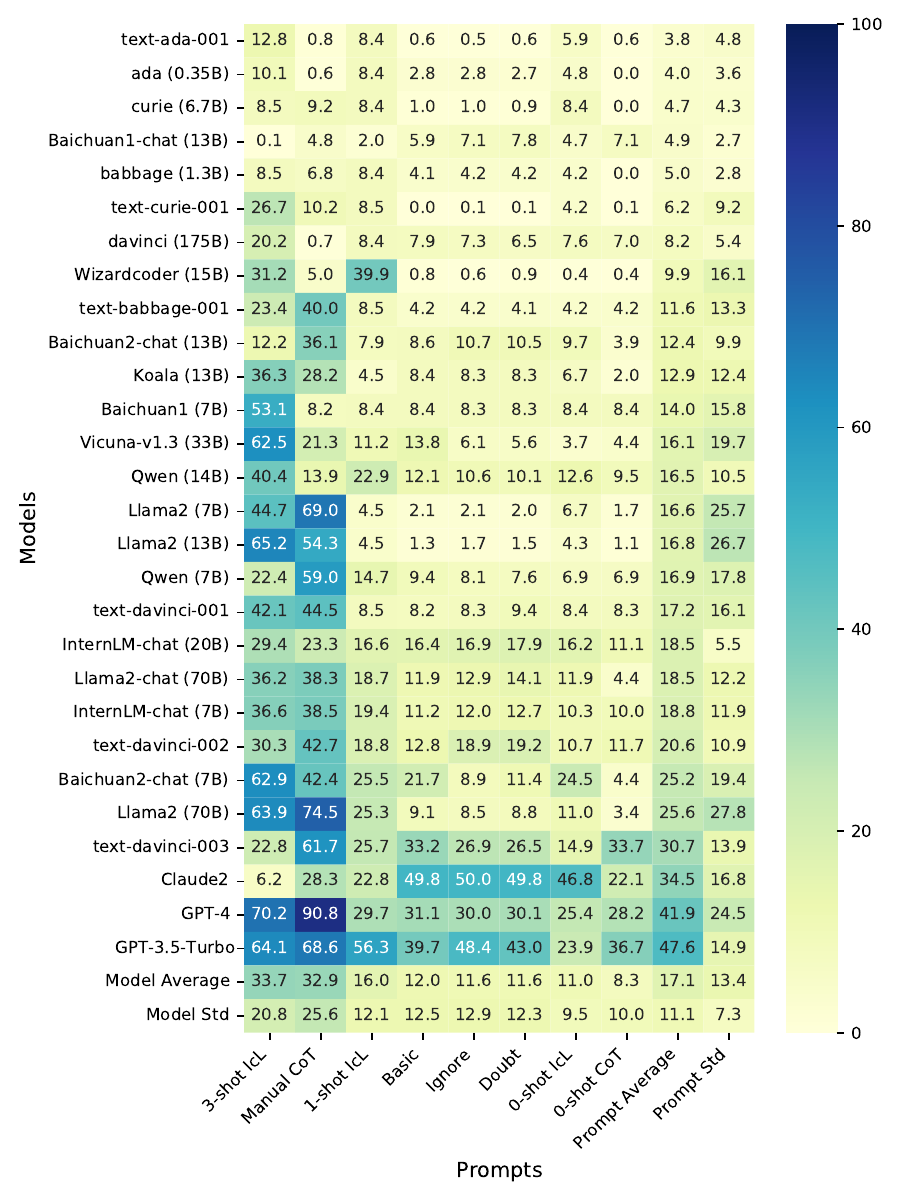}
\end{minipage}
}
\subfigure[\textit{Prompt gain} of CDE]{
\begin{minipage}{8.5cm}
\centering
\includegraphics[width=1\linewidth]{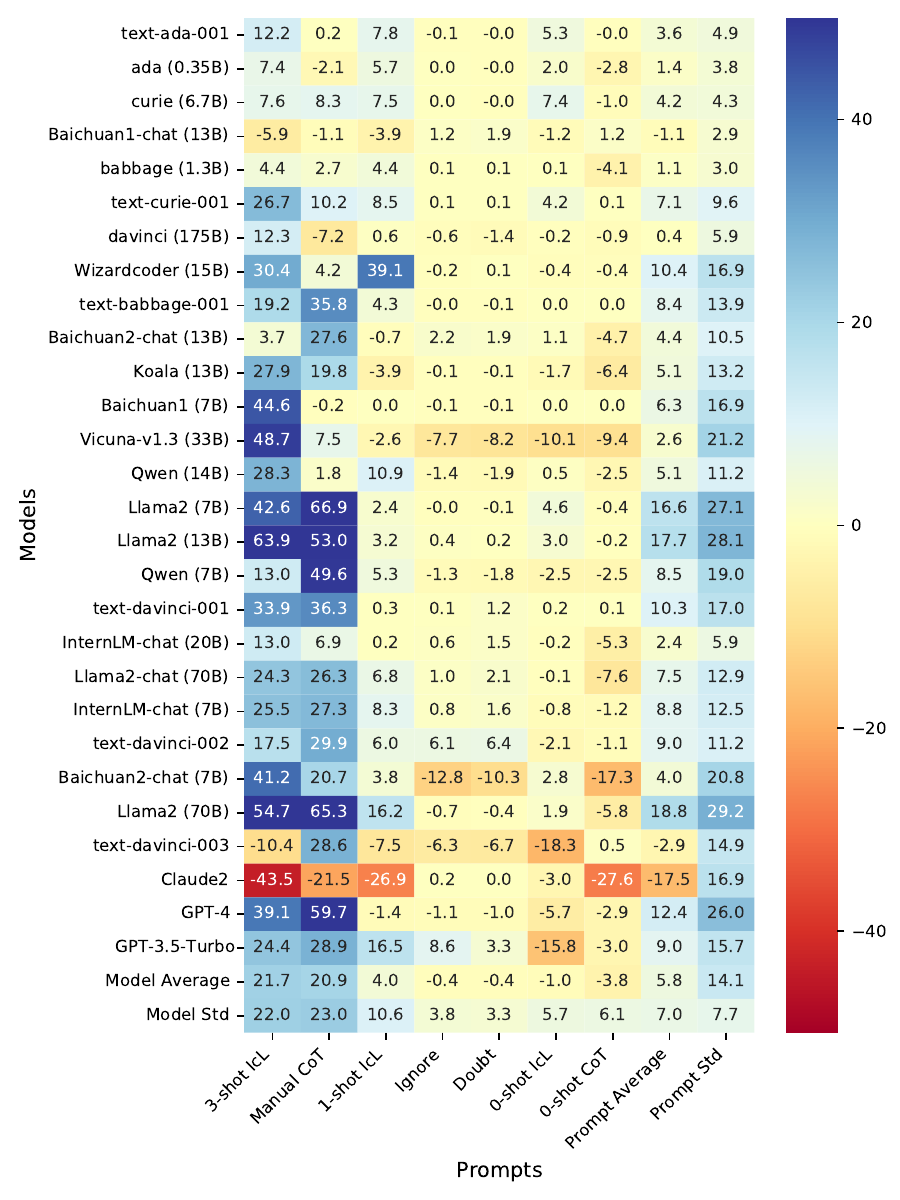}
\end{minipage}
}
\caption[Heatmap of CDE]{\textbf{Heatmap of CDE.} The models and prompts are sorted by their averages.}
\label{fig:Heatmap_of_Controlled_Direct_Effect}
\end{figure}

\begin{figure}
    \centering
    \includegraphics[width=0.8\linewidth]{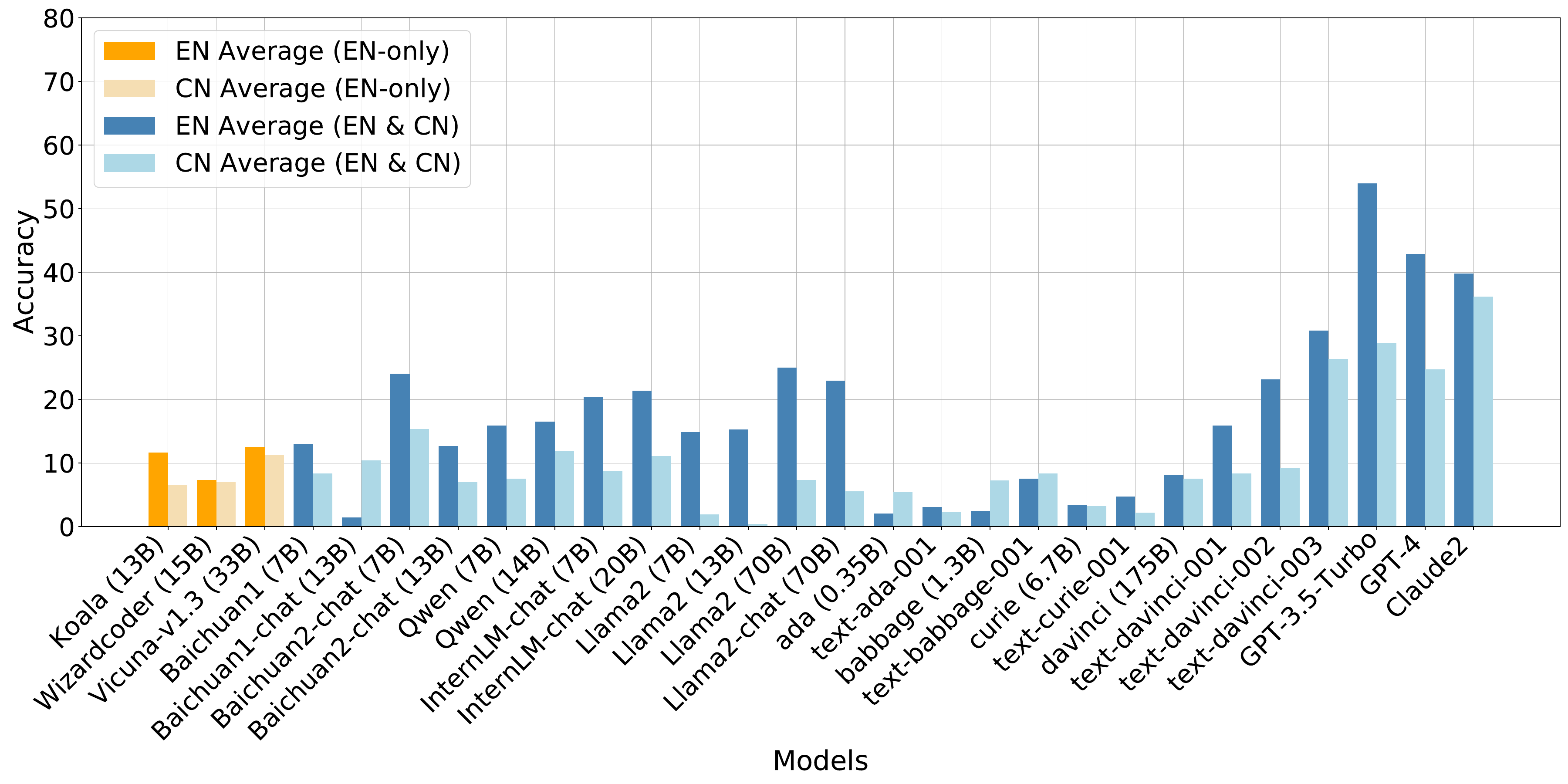}
    \caption[Language comparison of CDE]{\textbf{Language comparison of CDE.} The dark legend signifies the average performance of the model on an English test set, whereas the light legend denotes the average performance of the model on the Chinese test set. The yellow legend indicates a model trained exclusively on English datasets, while the blue legend represents a model trained on both English and Chinese datasets.}
    \label{fig:Controlled_Direct_Effect_Language}
\end{figure}
First, we delve into model performance in CDE:

1) \textbf{Distribution}: Figure \ref{fig:Distribution_of_Intervention}(b) showcases the distribution of all \textit{model-prompt pair}s in the CDE scenario. With a median of 9.3\% and a third quartile of 24.7\%, this scenario appears to have a \textbf{hard} \textit{understandability} since the median accuracy falls below the random guess benchmark of 16.7\%.
Figure \ref{fig:Distribution_of_Controlled_Direct_Effect_Tasks} illustrates the \textit{model-prompt pair} distribution for each specific task.
In the \textbf{CDE-P (CDE-basic)} task, with a median of 1.6\%, a third quartile of 9.1\% which is lower than 15\%, this Mathematical-mode task is considered to have a \textbf{very hard} \textit{understandability}.
The \textbf{CDE-P (CDE-hard)} task presents a median of 2.7\% and a third quartile of 11.4\% and is considered to have a \textit{very hard} \textit{understandability}, as indicated by the low values in both the third quartile and the median scores.
Additionally, the \textbf{CDE-B (CDE-natural)} task, with a median of 25.0\%, a third quartile of 44.8\%, and a random guess accuracy of 50.0\%, is also categorized to have a \textbf{very hard} \textit{understandability}.
\textbf{By analyzing the differences between tasks}, we observe that the median accuracies vary from 1.6\% to 25.0\% with a standard deviation of 10.8. The third quartile accuracies span from 9.1\% to 44.8\% with a standard deviation of 16.3. Therefore, the scenario has a \textbf{highly divergent} \textit{variance of distribution}. Additionally, the Natural-mode task scores higher in both the median and third quartile compared to the two mathematics-focused tasks (CDE-P (CDE-basic) and CDE-P (CDE-hard)), with CDE-P (CDE-hard) outperforming CDE-P (CDE-basic) in these metrics. The distribution for the Mathematical-mode tasks shows a majority of \textit{model-prompt pair}s (over 70\%) falling within a 0\% to 10\% accuracy range, suggesting a challenging landscape for most models. Conversely, the Natural-mode task features a more balanced distribution, with no 10-point (e.g., 0\% to 10\%, 10\% to 20\%) accuracy range that encompasses over 30\% of all \textit{model-prompt pair}s.

2) \textbf{Top Accuracy}: As seen in Figure \ref{fig:Heatmap_of_Controlled_Direct_Effect}(a), the leading models in terms of average accuracy for this scenario are GPT-3.5-Turbo at 47.6\%, GPT-4 at 41.9\%, and Claude2 at 34.5\%. The \textit{top model-prompt pair}, GPT-4 with manual CoT, reaches a high of 90.8\%, suggesting the scenario's \textit{solvability} as \textbf{potentially solvable} given the \textit{top model-prompt pair} surpasses an 80\% performance mark. Figure \ref{fig:Heatmap_of_performances_of_Controlled_Direct_Effect} shows the top three models' average accuracy, assessed on an individual task basis.
In the \textbf{CDE-P (CDE-basic)} task, the models leading in average accuracy are GPT-3.5-Turbo at 37.6\%, Claude2 at 29.9\%, and GPT-4 at 28.0\%, with GPT-4 and manual CoT forming the \textit{top model-prompt pair} at 87.8\%, indicating the task's \textit{solvability} as \textbf{potentially solvable} as the \textit{top model-prompt pair} performance exceeds 80\%. For the \textbf{CDE-P (CDE-hard)} task, the top models by average accuracy are GPT-3.5-Turbo at 45.4\%, GPT-4 at 33.0\%, and text-davinci-003 at 27.0\%, where GPT-4 and manual CoT achieve 86.7\%, again highlighting the \textit{solvability} of the task is \textbf{potentially solvable}. In the \textbf{CDE-B (CDE-natural)} task, the best models in average accuracy are GPT-4 at 61.1\%, GPT-3.5-Turbo at 57.2\%, and Claude2 at 51.7\%, with GPT-4 and manual CoT reaching 97.9\%, underscoring the task \textit{solvability} as \textbf{potentially solvable}.
\textbf{Through comparing different tasks}, the \textit{variance of solvability} between tasks appears \textbf{negligible}. Yet, the top model's average accuracy spans from 37.6\% to 61.1\% (a 23.5\% difference), and the peak accuracy for \textit{top model-prompt pair}s ranges from 86.7\% to 97.9\% (an 11.2\% difference), marking a \textbf{extremely significant} \textit{variance of model's top performance}. The Mathematical-mode tasks display lower top accuracies than the Natural-mode task. Regarding top models, GPT-3.5-Turbo leads in the Mathematical-mode tasks, while GPT-4 excels in the Natural-mode task. Across all tasks, GPT-3.5-Turbo and GPT-4 consistently rank within the top three for average model performance. Moreover, GPT-4 paired with manual CoT are found to be the \textit{top model-prompt pair} across all tasks.

3) \textbf{Stability}: The three models exhibiting the greatest stability with the lowest \textit{model volatility} are Baichuan1-chat (13B) at 2.7, babbage (1.3B) at 2.8, and ada (0.35B) at 3.6. Conversely, the three models showing the highest levels of instability across various prompts are Llama2 (70B) at 27.8, Llama2 (13B) at 26.7, and Llama2 (7B) at 25.7. We proceed to examine stability on a task-specific basis.
In the \textbf{CDE-P (CDE-basic)} task, the most stable models are davinci (175B) at 0.1, text-ada-001 at 0.2, and ada (0.35B) at 0.2. In contrast, the most unstable models are Llama2 (70B) at 28.2, GPT-4 at 25.3, and Llama2 (13B) at 22.5.
In the \textbf{CDE-P (CDE-hard)} task, the top three most stable models are curie (6.7B) at 0.8, ada (0.35B) at 1.8, and babbage (1.3B) at 2.4. Meanwhile, the models with the highest model volatilites, indicating the greatest instability, are GPT-4 at 33.7, Llama2 (7B) at 30.2, and Llama2 (13B) at 28.5.
In the \textbf{CDE-B (CDE-natural)} task, the models demonstrating the most stability are Baichuan1-chat (13B) at 3.1, davinci (175B) at 7.7, and babbage (1.3B) at 7.9. On the flip side, the models with the most instability are Llama2 (70B) at 28.4, Llama2 (13B) at 28.0, and Baichuan2-chat (7B) at 25.8.
\textbf{In all tasks}, Llama2 (13B) consistently appears among the top three most unstable models, underscoring its high sensitivity to prompts in this scenario. Additionally, GPT-4 and Llama2 (70B) are identified as the two most unstable models in the Mathematical-mode tasks.

4) \textbf{Open-Limited Ratio}: The ratio of open-access to limited-access models among the top five models with the highest average accuracy in the scenario is 1:4, indicating a \textbf{moderate} \textit{open-limited gap}.

Next, we delve into \textit{prompt gain} in CDE:

1) \textbf{Top Gain}: As shown in Figure \ref{fig:Heatmap_of_Controlled_Direct_Effect}(b), the two prompts leading in average accuracy gain over the basic prompt are 3-shot IcL at 21.7\% and manual CoT at 20.9\%. The largest improvement in accuracy compared to the basic prompt is noted with Llama2 (7B) using manual CoT, which registers a 66.9\% increase. A more granular, task-specific analysis follows. Figure \ref{fig:Heatmap_of_gain_of_Controlled_Direct_Effect} illustrates the gains across all tasks in the scenario. In the \textbf{CDE-P (CDE-basic)} task, manual CoT at 21.0\% and 3-shot IcL at 13.3\% top the list for average accuracy gain over the basic prompt, with GPT-4 and manual CoT marking the greatest increase at 69.2\%. In the \textbf{CDE-P (CDE-hard)} task, 3-shot IcL at 26.3\% and manual CoT at 20.9\% are the most effective, with Llama2 (7B) and manual CoT achieving the most substantial boost at 78.3\%. In the \textbf{CDE-B (CDE-natural)} task, the highest gains are with 3-shot IcL at 25.5\% and manual CoT at 20.8\%, and the top improvement is seen with Llama2 (13B) using 3-shot IcL, which increases by 76.6\%.
\textbf{For every individual task}, 3-shot IcL and manual CoT stand out as the two most effective prompts, underscoring their consistent impact across the scenario. 
Moreover, \textit{model-prompt pair}s that utilize these specific prompts achieve the greatest increases in performance relative to other \textit{model-prompt pair}s.

2) \textbf{Exceptions}: The highly effective prompt, 3-shot IcL, does not generate a positive average \textit{prompt gain} in Baichuan1-chat (13B), text-davinci-003, and Claude2. However, all prompts manage to enhance the performance of text-curie-001 and text-davinci-001 beyond the basic prompt. In the \textbf{CDE-P (CDE-basic)} task, the leading prompt, manual CoT, fails to yield a positive average \textit{prompt gain} on Baichuan1-chat (13B) and Claude2. Within the \textbf{CDE-P (CDE-hard)} task, the top 1 average prompt, 3-shot IcL, shows no effectiveness on Baichuan1-chat (13B) and Claude2. Furthermore, no prompt is able to boost Claude2's performance above the basic prompt level. In the \textbf{CDE-B (CDE-natural)} task, the best prompt, 3-shot IcL, also falls short with Baichuan1-chat (13B), text-davinci-003, and Claude2. All prompts succeed in elevating the performance of text-curie-001 and text-davinci-001 over the basic prompt. 
\textbf{In all tasks}, Claude2, ranking as the fourth highest model on average, is significantly negatively impacted by the top average prompt, 3-shot IcL.

3) \textbf{Stability}: Regarding stability, the two most stable prompts, based on the lowest \textit{prompt volatility}, are adversarial doubt at 3.3 and adversarial ignore at 3.8. Conversely, the two prompts showing the greatest variability, indicated by the highest \textit{prompt volatility}, are manual CoT at 23.0 and 3-shot IcL at 22.0. The \textit{average model-prompt-gain volatility} (\textit{AMPGV}) is 14.1, illustrating a \textbf{high} \textit{prompt dependence}. We then explore stability on an individual task basis.
In the \textbf{CDE-P (CDE-basic)} task, the most stable prompts are 0-shot CoT with a \textit{prompt volatility} of 1.4 and adversarial ignore with a \textit{prompt volatility} of 1.8, while the least stable prompts are manual CoT at 23.4 and 3-shot IcL at 22.1. An \textit{AMPGV} of 10.9 highlights a \textbf{high} \textit{prompt dependence}.
In the \textbf{CDE-P (CDE-hard)} task, adversarial doubt and adversarial ignore are the most stable with \textit{prompt volatility} of 2.7 and 4.0, respectively. The least stable are manual CoT at 26.2 and 3-shot IcL at 25.8. The \textit{AMPGV} is 16.0, underscoring a \textbf{high} \textit{prompt dependence}.
For the \textbf{CDE-B (CDE-natural)} task, the most stable prompts are adversarial doubt at 8.1 and 0-shot IcL at 8.7, while the most variable are 3-shot IcL at 27.9 and manual CoT at 25.7. As the \textit{AMPGV} is 17.4, the task has a \textbf{high} \textit{prompt dependence}.
\textbf{After evaluating all the tasks in the scenario}, we find that the distribution of \textit{AMPGV} ranges from 10.9 to 17.4, indicating a \textbf{moderate spread} \textit{variance of prompt dependence}. Although 3-shot IcL and manual CoT are the most effective, they also rank as the most unstable across tasks, demonstrating that the impact of these prompts varies significantly across different models.

Lastly, we measure \textit{language proficiency} in CDE:

1) \textbf{English vs. Chinese}: Illustrated by Figure \ref{fig:Controlled_Direct_Effect_Language}, models generally exhibit superior performance on the English test set compared to the Chinese test set, with 24 out of 28 models achieving better results in English.

2) \textbf{Accuracy Difference}: Significant discrepancies in accuracy between English and Chinese, with a preference for English, appear in models such as GPT-3.5-Turbo (25.1\%), GPT-4 (18.2\%), and Llama2 (70B) (17.7\%). On the other hand, models like Baichuan1-chat (13B) (9.0\%), babbage (1.3B) (4.8\%), and ada (0.35B) (3.4\%) demonstrate higher accuracy in Chinese.

\paragraph{Causal effect identification.}
\begin{figure}[t]
\centering
\subfigure[Model performance of CEI]{
\begin{minipage}{8.5cm}
\centering
\includegraphics[width=1\linewidth]{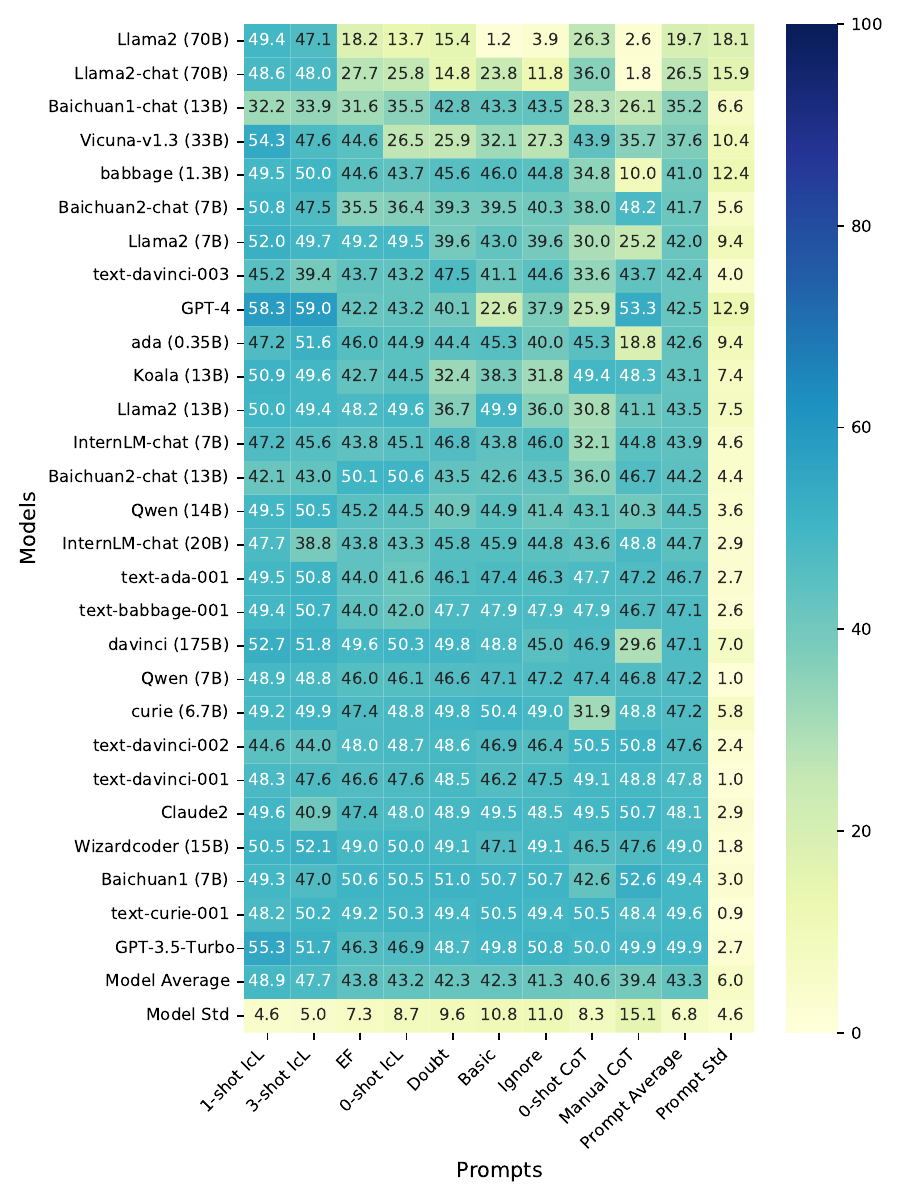}
\end{minipage}
}
\subfigure[\textit{Prompt gain} of CEI]{
\begin{minipage}{8.5cm}
\centering
\includegraphics[width=1\linewidth]{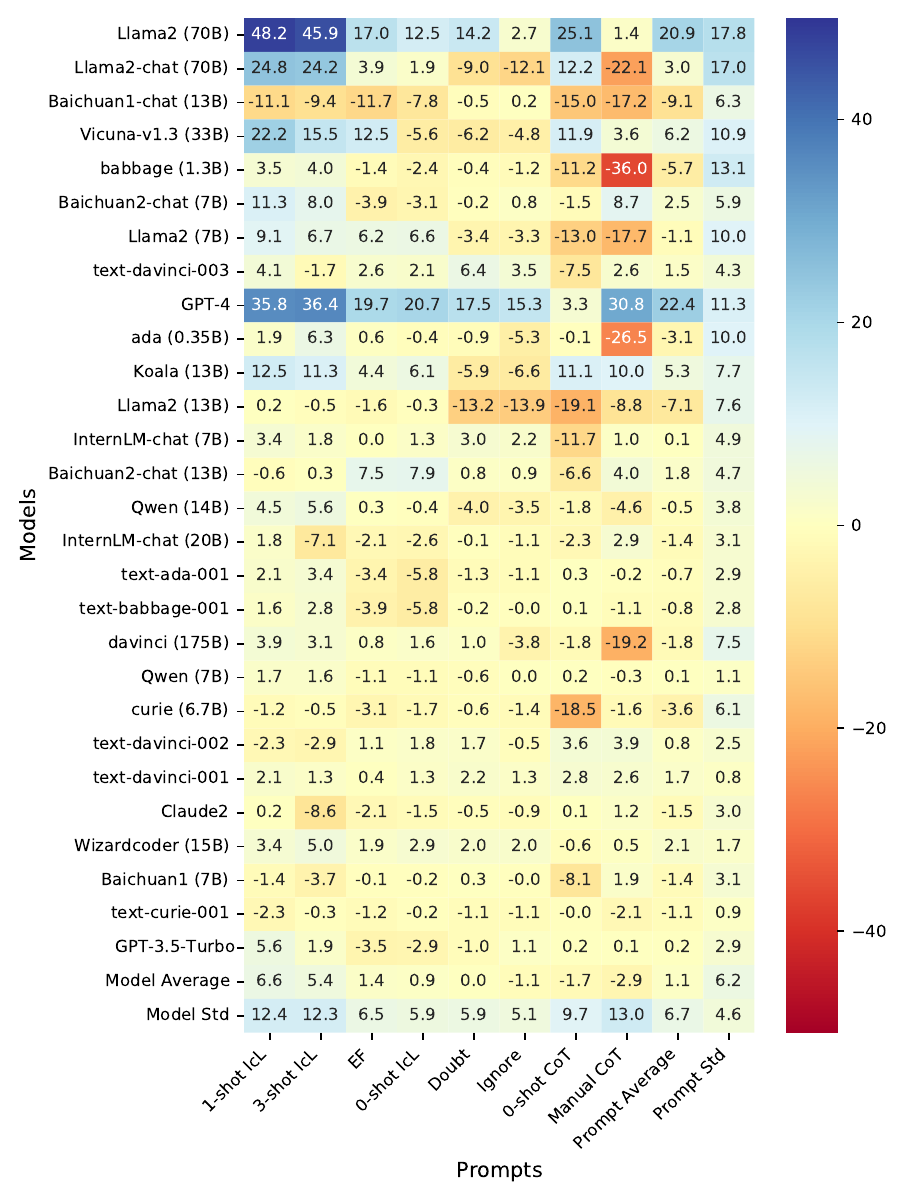}
\end{minipage}
}
\caption[Heatmap of CEI]{\textbf{Heatmap of CEI.} The models and prompts are sorted by their averages.}
\label{fig:Heatmap_of_Causal_Effect_Identification}
\end{figure}

\begin{figure}
    \centering
    \includegraphics[width=0.8\linewidth]{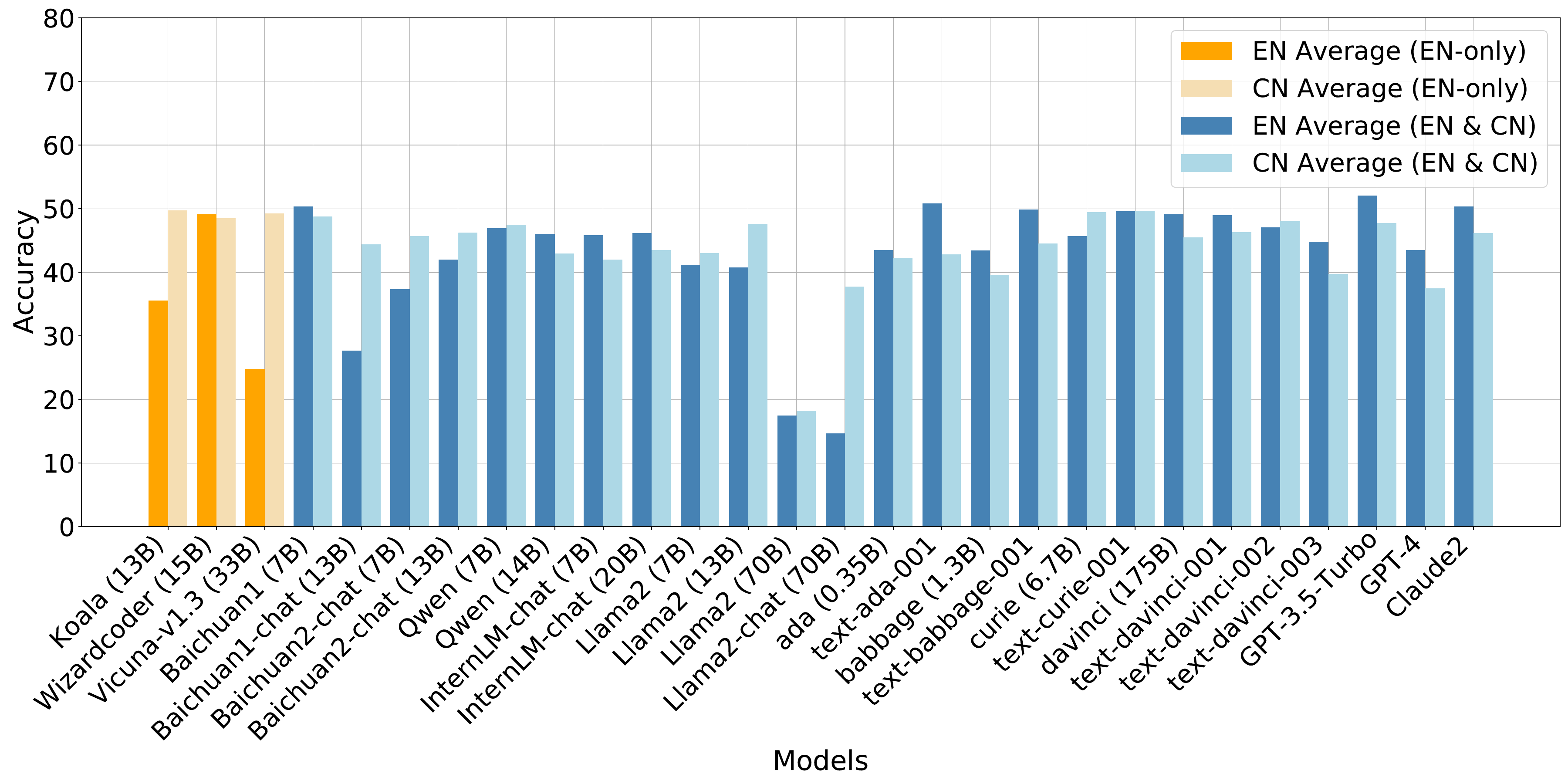}
    \caption[Language comparison of CEI]{\textbf{Language comparison of CEI.} The dark legend signifies the average performance of the model on an English test set, whereas the light legend denotes the average performance of the model on the Chinese test set. The yellow legend indicates a model trained exclusively on English datasets, while the blue legend represents a model trained on both English and Chinese datasets.}
    \label{fig:Causal_Effect_Identification_Language}
\end{figure}

First, we examine model performance in CEI:

1) \textbf{Distribution}: Figure \ref{fig:Distribution_of_Intervention}(c) displays the distribution of all \textit{model-prompt pair}s within CEI. With a median score of 46.6\% and a third quartile of 49.3\%, this scenario has a \textbf{very hard} \textit{understandability} because the third quartile falls below the random guess threshold of 50.0\%.
Figure \ref{fig:Distribution_of_Causal_Effect_Identification_Tasks} details the distribution for each specific task.
In the \textbf{CEI-B (0.2-UC)} task, the median is 47.6\%, and the third quartile is 50.0\%. Due to the random guess accuracy of 50.0\%, the task's \textit{understandability} is \textbf{hard}.
The \textbf{CEI-B (0.4-UC)} task shows a median of 46.5\% and a third quartile of 49.4\%, with a random guess accuracy of 50.0\%, categorizing it to have a \textbf{very hard} \textit{understandability}.
For the \textbf{CEI-B (0.6-UC)} task, the median stands at 46.1 and the third quartile at 48.9\%, with the random guess accuracy also at 50.0\%, indicating the \textit{understandability} of this task as \textbf{very hard}.
Similarly, the \textbf{CEI-B (0.8-UC)} task presents a median of 46.1 and a third quartile of 48.9\%, with a random guess accuracy of 50.0\%, affirming its \textit{understandability} as \textbf{very hard}.
\textbf{By analyzing the differences between tasks}, we see a median accuracy range from 46.1\% to 47.6\% with a standard deviation of 0.6. The third quartile accuracy spans from 48.9\% to 50.0\% with a standard deviation of 0.5. As a result, the scenario has a \textbf{minimally divergent} \textit{variance of distribution}. Sorting tasks by median and third quartile yield the same accuracy ranking: CEI-B (0.8-UC) = CEI-B (0.6-UC) < CEI-B (0.4-UC) < CEI-B (0.2-UC). Across all four tasks, more than 70\% of the distribution lies within the 40 to 60 accuracy range.

2) \textbf{Top Accuracy}: Figure \ref{fig:Heatmap_of_Causal_Effect_Identification}(a) reveals that the leading models in this scenario, based on average accuracy, are GPT-3.5-Turbo at 49.9\%, text-curie-001 at 49.6\%, and Baichuan1 (7B) at 49.4\%. The \textit{top model-prompt pair}, GPT-4 with 3-shot IcL, reaches 59.0\%, indicating the \textit{solvability} of the scenario as \textbf{challenging} due to the \textit{top model-prompt pair}'s performance falling short of 80\%. Figure \ref{fig:Heatmap_of_performances_of_Causal_Effect_Identification} displays the top three models' average accuracy across individual tasks.
In the \textbf{CEI-B (0.2-UC)} task, the highest average accuracies are seen with GPT-3.5-Turbo at 52.6\%, Baichuan1 (7B) at 51.1\%, and text-curie-001 at 50.7\%, where GPT-4 and 1-shot IcL lead with a 60.1\% score, marking the task \textit{solvability} as \textbf{challenging} since the \textit{top model-prompt pair} does not exceed 80\%. For the \textbf{CEI-B (0.4-UC)} task, GPT-3.5-Turbo at 50.4\%, text-curie-001 at 49.5\%, and Wizardcoder (15B) at 49.5\% top the average accuracy, with GPT-4 and 3-shot IcL achieving 59.2\%, again highlighting the task's \textbf{challenging} \textit{solvability}. In the \textbf{CEI-B (0.6-UC)} task, the leading averages are by Wizardcoder (15B) at 49.1\%, text-curie-001 at 48.9\%, and Baichuan1 (7B) at 48.8\%, with GPT-4 and 3-shot IcL reaching 58.6\%, underlining the \textit{solvability} of the task as \textbf{challenging}. For the \textbf{CEI-B (0.8-UC)} task, text-curie-001 at 49.2\%, Wizardcoder (15B) at 48.6\%, and Baichuan1 (7B) at 48.6\% are the top models, with GPT-4 and 3-shot IcL achieving 58.2\%, confirming the task's \textbf{challenging} \textit{solvability}.
\textbf{Through comparing different tasks}, the \textit{variance of solvability} among the tasks is \textbf{negligible}. The top model's average accuracy fluctuates minimally from 49.1\% to 52.6\% (a 3.5\% difference), and the accuracy among \textit{top model-prompt pair}s varies from 58.2\% to 60.1\% (a 1.9\% difference), signifying a \textbf{small} \textit{variance of model's top performance}. Text-curie-001 consistently ranks within the top three models in all tasks. Moreover, Wizardcoder (15B) and Baichuan1 (7B) appear in the top 3 average models in three out of four tasks. Although GPT-4 doesn't rank among the top three in average model performance, when paired with 1-shot IcL or 3-shot IcL, it forms the \textit{top model-prompt pair}s in all tasks.

3) \textbf{Stability}: The three most stable models, based on the lowest \textit{model volatility}, are text-curie-001 at 0.9, text-davinci-001 at 1.0, and Qwen (7B) at 1.0. Conversely, the models demonstrating the highest levels of instability across various prompts are Llama2 (70B) at 18.1, Llama2-chat (70B) at 15.9, and GPT-4 at 12.9, showing a pronounced prompt sensitivity. We proceed to analyze stability on a task-specific basis.
In the \textbf{CEI-B (0.2-UC)} task, the most stable models are Qwen (7B) at 0.5, text-curie-001 at 1.3, and text-davinci-001 at 1.8, while the most unstable models are Llama2 (70B) at 18.6, Llama2-chat (70B) at 16.3, and GPT-4 at 12.8.
In the \textbf{CEI-B (0.4-UC)} task, the top stable models are text-curie-001 at 0.7, text-davinci-001 at 1.1, and Wizardcoder (15B) at 1.3, in contrast to the most unstable models, which are Llama2 (70B) at 18.2, Llama2-chat (70B) at 15.8, and GPT-4 at 13.4.
For the \textbf{CEI-B (0.6-UC)} task, the leading models in stability are text-davinci-001 at 0.9, text-curie-001 at 1.0, and Qwen (7B) at 1.5. The models with the highest \textit{model volatility}, indicating instability, are Llama2 (70B) at 18.0, Llama2-chat (70B) at 15.9, and GPT-4 at 13.0.
In the \textbf{CEI-B (0.8-UC)} task, the most stable models are text-davinci-001 at 0.9, text-curie-001 at 1.0, and Qwen (7B) at 1.4, while the most unstable models are Llama2 (70B) at 17.9, Llama2-chat (70B) at 15.9, and babbage (1.3B) at 13.0.
\textbf{In all tasks}, text-curie-001 and text-davinci-001 consistently rank as the most stable models, whereas Llama2 (70B) and Llama2-chat (70B) consistently appear as the most unstable, underscoring their high sensitivity to prompt variations.

4) \textbf{Open-Limited Ratio}: With a ratio of 2 open-access models to 3 limited-access models among the top five models with the highest accuracy, the \textit{open-limited gap} of the scenario is \textbf{small}.

Following, we delve into \textit{prompt gain} in CEI:

1) \textbf{Top Gain}: As illustrated in Figure \ref{fig:Heatmap_of_Causal_Effect_Identification}(b), the two prompts leading in average accuracy gain over the basic prompt are 1-shot IcL at 6.6\% and 3-shot IcL at 5.4\%. The most significant improvement in accuracy relative to the basic prompt is seen with Llama2 (70B) using 1-shot IcL, with a substantial increase of 48.2\%. We then proceed to a granular analysis across individual tasks. Figure \ref{fig:Heatmap_of_gain_of_Causal_Effect_Identification} outlines the gains across all tasks in the scenario. In the \textbf{CEI-B (0.2-UC)} task, 1-shot IcL at 5.9\% and 3-shot IcL at 4.5\% offer the highest average accuracy gains over the basic prompt, with Llama2 (70B) and 1-shot IcL achieving a remarkable increase of 48.7\%. For the \textbf{CEI-B (0.4-UC)} task, 1-shot IcL at 6.8\% and 3-shot IcL at 5.4\% stand out for average accuracy gains, with Llama2 (70B) and 1-shot IcL showing an impressive increase of 48.9\%. In the \textbf{CEI-B (0.6-UC)} task, the leading gains are by 1-shot IcL at 6.9\% and 3-shot IcL at 5.8\%, with Llama2 (70B) and 1-shot IcL demonstrating a significant boost of 47.4\%. For the \textbf{CEI-B (0.8-UC)} task, 1-shot IcL at 6.7\% and 3-shot IcL at 5.7\% provide the highest average accuracy gains, with Llama2 (70B) and 1-shot IcL showing an increase of 47.7\%.
\textbf{On the evaluation across tasks}, the pattern is consistent: 1-shot IcL is the top 1 model average prompt as well as the prompt that makes the largest gain accompanied by Llama2 (70B), surpassing other \textit{model-prompt pair}s. 3-shot IcL ranks as the second most effective prompt in terms of average accuracy gain.

2) \textbf{Exceptions}: The highly effective prompt, 1-shot IcL, cannot create a positive average \textit{prompt gain} with several models, including Baichuan1-chat (13B), Baichuan2-chat (13B), curie (6.7B), text-davinci-002, Baichuan1 (7B), and text-curie-001. However, all prompts are capable of enhancing the performance of Llama2 (70B), GPT-4, and text-davinci-001 beyond the basic prompt. Notably, curie (6.7B) and text-curie-001 see no improvement from any prompt over the basic prompt.
In the \textbf{CEI-B (0.2-UC)} task, 1-shot IcL fails to be effective for a range of models including Baichuan1-chat (13B), babbage (1.3B), ada (0.35B), Baichuan2-chat (13B), InternLM-chat (20B), curie (6.7B), Qwen (7B), text-babbage-001, text-davinci-002, text-curie-001, and Baichuan1 (7B). Yet, all prompts boost the performance of Llama2 (70B) and GPT-4 above the basic prompt. Curie (6.7B) and Qwen (7B), however, do not benefit from any prompt in this task.
For the \textbf{CEI-B (0.4-UC)} task, 1-shot IcL falls short with Baichuan1-chat (13B), Llama2 (13B), Baichuan2-chat (13B), curie (6.7B), text-davinci-002, Claude2, Baichuan1 (7B), and text-curie-001. Despite this, improvements are seen for Llama2 (70B) and GPT-4 with all prompts, while Llama2 (13B), curie (6.7B), Claude2, and text-curie-001 see no enhancement from any prompt.
In the \textbf{CEI-B (0.6-UC)} task, the effectiveness of 1-shot IcL does not extend to Baichuan1-chat (13B), Baichuan2-chat (13B), text-davinci-002, curie (6.7B), Baichuan1 (7B), and text-curie-001. Conversely, Llama2 (70B) and GPT-4 performances are elevated by all prompts above the basic prompt, with curie (6.7B) again showing no improvement from any prompt.
The \textbf{CEI-B (0.8-UC)} task mirrors this pattern, with 1-shot IcL not creating positive average \textit{prompt gain} for several models including Baichuan1-chat (13B) and Llama2 (13B). Llama2 (70B) and GPT-4, however, see a positive average \textit{prompt gain} from all prompts, contrasting with curie (6.7B), which does not improve with any prompts.
\textbf{By evaluating across the tasks}, it turns out that all prompts can enhance Llama2 (70B), GPT-4, and text-davinci-001 beyond the basic prompt level across all tasks. Conversely, Baichuan1-chat (13B), Llama2 (13B), and curie (6.7B) do not experience any benefit from any prompts in these tasks. Throughout all four tasks, the top prompt, 1-shot IcL, fails to yield improvements for Baichuan1-chat (13B), Baichuan2-chat (13B), curie (6.7B), and text-curie-001. Additionally, in every task, Llama2 (70B) and GPT-4's performance can be improved by any prompt over the basic prompt, while no prompt is able to enhance curie (6.7B)'s performance over the basic prompt in any task.

3) \textbf{Stability}: Regarding stability, the two most stable prompts, indicated by the smallest \textit{prompt volatility}, are adversarial ignore at 5.1 and 0-shot IcL at 5.9. Conversely, the prompts exhibiting the greatest variability, evidenced by the largest \textit{prompt volatility}, are manual CoT at 13.0 and 1-shot IcL at 12.4. The \textit{average model-prompt-gain volatility} (\textit{AMPGV}) is 6.2, illustrating a \textbf{medium} level of \textit{prompt dependence}. Next, we conduct a task-specific stability analysis.
In the \textbf{CEI-B (0.2-UC)} task, the most stable prompts are adversarial ignore with a \textit{prompt volatility} of 4.9 and 0-shot IcL with a \textit{prompt volatility} of 5.7, while the least stable are manual CoT at 13.0 and 1-shot IcL at 12.8. The \textit{AMPGV} is 6.4, signifying the task has a \textbf{medium} \textit{prompt dependence}.
For the \textbf{CEI-B (0.4-UC)} task, the top stable prompts are adversarial ignore with a \textit{prompt volatility} of 5.4 and 0-shot IcL with a \textit{prompt volatility} of 6.1. The most unstable prompts, with the largest \textit{prompt volatility}, are manual CoT at 12.9 and 1-shot IcL at 12.7. An \textit{AMPGV} of 6.2 indicates a \textbf{medium} \textit{prompt dependence}.
In the \textbf{CEI-B (0.6-UC)} task, the most stable prompts feature adversarial ignore with a \textit{prompt volatility} of 5.3 and adversarial doubt with a \textit{prompt volatility} of 5.9, while the least stable prompts are manual CoT at 13.3 and 3-shot IcL at 12.1. The \textit{AMPGV} is 6.4, showing a \textbf{medium} \textit{prompt dependence}.
For the \textbf{CEI-B (0.8-UC)} task, the top stable prompts are adversarial ignore at 5.2 and adversarial doubt at 5.9, in contrast to the most unstable prompts, manual CoT at 13.0 and 3-shot IcL at 12.5, with an \textit{AMPGV} of 6.4, reflecting a \textbf{medium} \textit{prompt dependence}.
\textbf{After evaluating all tasks in the scenario}, the range of \textit{AMPGV} is from 6.2 to 6.4, showing a \textbf{narrow} \textit{variance of prompt dependence}. Adversarial ignore consistently shows the highest stability across all four tasks, while manual CoT ranks as the least stable in every task, aligning with the observations made from the scenario perspective.

Finally, we analyze \textit{language proficiency} in CEI, 

1) \textbf{English vs. Chinese}: Figure \ref{fig:Causal_Effect_Identification_Language} illustrates that models tend to perform better on the English test set than on the Chinese test set, with 15 out of 28 models exhibiting superior performance in English.

2) \textbf{Accuracy Difference}: There are significant differences in accuracy between English and Chinese in text-ada-001 (8.0\%), GPT-4 (6.0\%), and text-babbage-001 (5.3\%), with a preference for English. On the flip side, models like Vicuna-v1.3 (33B) (24.5\%), Llama2-chat (70B) (23.1\%), and Baichuan1-chat (13B) (16.7\%) demonstrate higher accuracy in Chinese, indicating a stronger performance in that language.

\paragraph{Backdoor adjustment set.}
\begin{figure}[t]
\centering
\subfigure[Model performance of BAS]{
\begin{minipage}{8.5cm}
\centering
\includegraphics[width=1\linewidth]{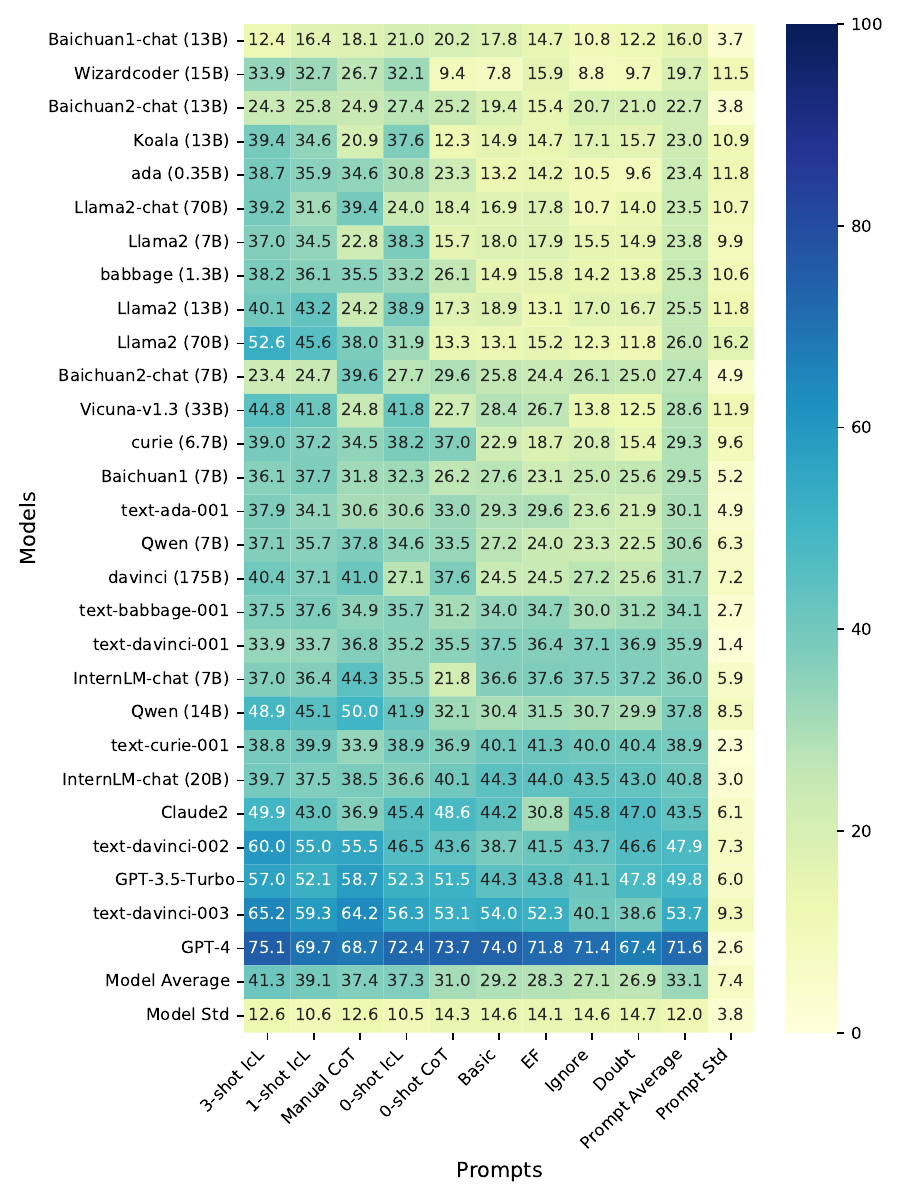}
\end{minipage}
}
\subfigure[\textit{Prompt gain} of BAS]{
\begin{minipage}{8.5cm}
\centering
\includegraphics[width=1\linewidth]{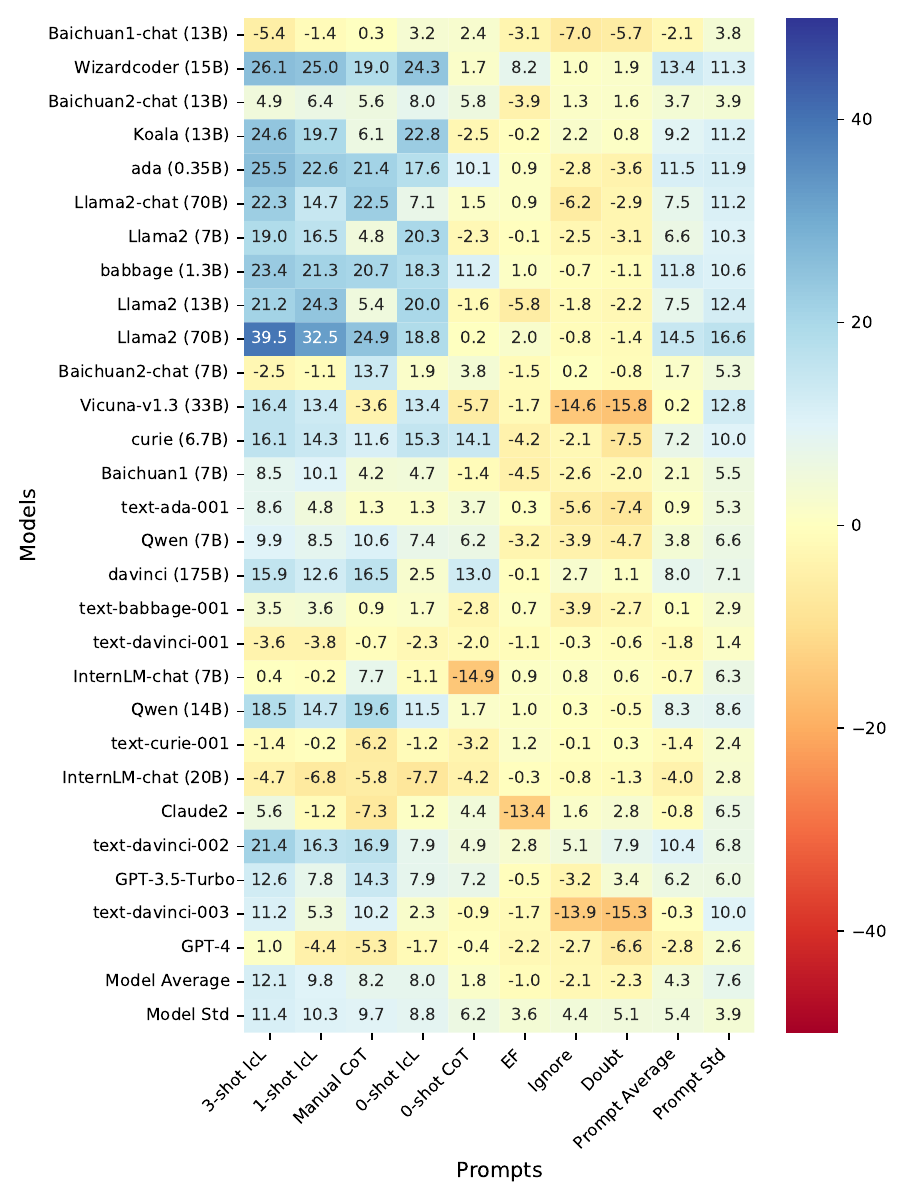}
\end{minipage}
}
\caption[Heatmap of BAS]{\textbf{Heatmap of BAS.} The models and prompts are sorted by their averages.}
\label{fig:Heatmap_of_Backdoor_Adjustment_Set}
\end{figure}

\begin{figure}
    \centering
    \includegraphics[width=0.8\linewidth]{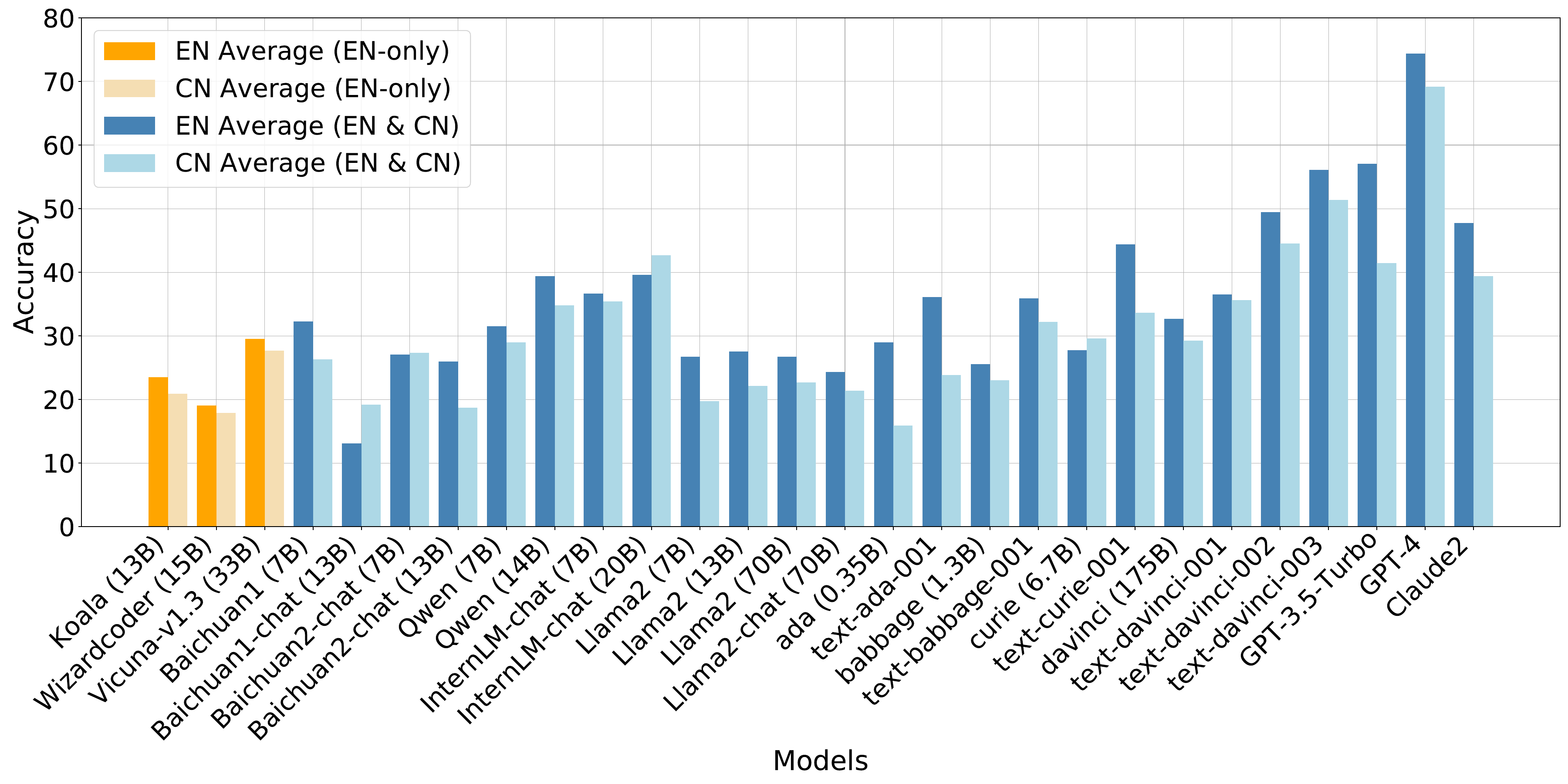}
    \caption[Language comparison of BAS]{\textbf{Language comparison of BAS.} The dark legend signifies the average performance of the model on an English test set, whereas the light legend denotes the average performance of the model on the Chinese test set. The yellow legend indicates a model trained exclusively on English datasets, while the blue legend represents a model trained on both English and Chinese datasets.}
    \label{fig:Backdoor_Adjustment_Set_Language}
\end{figure}

First, we analyze model performance in BAS:

1) \textbf{Distribution}: Figure \ref{fig:Distribution_of_Intervention}(d) showcases the distribution of all \textit{model-prompt pair}s in the BAS. With a median of 34.0\% and a third quartile of 40.0\%, this scenario is viewed to have a \textbf{hard} \textit{understandability} as the median accuracy falls below the scenario's average random guess of 37.5\%.
Figure \ref{fig:Distribution_of_Backdoor_Adjustment_Set_Tasks} shows the distribution of all \textit{model-prompt pair}s task-by-task.
In the \textbf{BAS-B (backadj)} task, the median is calculated at 48.4\%, and the third quartile at 50.4\%. As the random guess accuracy is 50.0\%, the task has a \textbf{hard} \textit{understandability}.
For the \textbf{BAS-C (max-BAS)} task, the median stands at 29.6\%, the third quartile at 34.6\%, with the random guess accuracy at 33.3\%, labeling the \textit{understandability} of the task as \textbf{hard} to understand.
The \textbf{BAS-C (min-BAS)} task presents a median of 31.9\% and a third quartile of 43.5\%, alongside with a random guess accuracy of 33.3\%, indicating it to have a \textbf{hard} \textit{understandability}.
Similarly, the \textbf{BAS-C (mix-BAS)} task reveals a median of 29.8\% and a third quartile of 35.7\%. With the random guess accuracy at 33.3\%, the task is \textbf{hard} to understand.
\textbf{By analyzing the differences between tasks}, we find that the median accuracy range from 29.6\% to 48.4\% with a standard deviation of 7.8. The third quartile accuracy spans from 34.6\% to 50.4\% with a standard deviation of 6.4. The discovery shows that the the scenario has a \textbf{moderately distinct} \textit{variance of distribution}. Ranking tasks by median and third quartile reveals the same accuracy order of: BAS-C (max-BAS) < BAS-C (mix-BAS) < BAS-C (min-BAS) < BAS-B (backadj). The three symbolic tasks (BAS-C (max-BAS), BAS-C (mix-BAS), and BAS-C (min-BAS)) exhibit lower median and third quartile values than the Natural-mode task, BAS-B (backadj). Notably, in the BAS-B (backadj) task, over 50\% of the distribution falls within a 40\% to 50\% accuracy range, while the three symolic tasks feature a more balanced distribution, without any 10\%-width interval surpassing 40\% of the distribution.

2) \textbf{Top Accuracy}: Figure \ref{fig:Heatmap_of_Backdoor_Adjustment_Set}(a) demonstrates that the leading models by average accuracy in this scenario are GPT-4 at 71.6\%, text-davinci-003 at 53.7\%, and GPT-3.5-Turbo at 49.8\%. The \textit{top model-prompt pair}, GPT-4 with 3-shot IcL, reaches 75.1\%, indicating that the \textit{solvability} of this scenario is \textbf{challenging} due to the \textit{top model-prompt pair}'s performance not exceeding 80\%. Figure \ref{fig:Heatmap_of_performances_of_Backdoor_Adjustment_Set} details the top three models' average accuracy across different tasks.
In the \textbf{BAS-B (backadj)} task, the highest average accuracies are for GPT-4 at 51.2\%, babbage (1.3B) at 50.0\%, and davinci (175B) at 50.0\%, with Llama2 (70B) and 3-shot IcL leading at 68.5\%, marking the \textit{solvability} of the task as \textbf{challenging} with the \textit{top model-prompt pair} performance below 80\%. For the \textbf{BAS-C (max-BAS)} task, the top models are GPT-4 at 74.4\%, text-davinci-003 at 48.4\%, and GPT-3.5-Turbo at 48.4\%, where GPT-4 and manual CoT achieve 83.0\%. This task's \textit{solvability} is \textbf{solvable} with the top model's average accuracy over 70\%. The \textbf{BAS-C (min-BAS)} task shows GPT-4 at 82.3\%, text-davinci-003 at 69.1\%, and text-davinci-002 at 59.5\% as the leading models, with GPT-4 and 3-shot IcL reaching 86.1\%. The above analysis confirms the task's \textbf{solvable} \textit{solvability} with the top model's average accuracy of over 70\%. For the \textbf{BAS-C (mix-BAS)} task, the top models by average accuracy are GPT-4 at 78.4\%, text-davinci-003 at 53.3\%, and GPT-3.5-Turbo at 48.5\%, with GPT-4 and 3-shot IcL achieving 84.0\%. As the top model's average accuracy is over 70\%, this task has a  \textbf{solvable} \textit{solvability}.
\textbf{Through comparing different tasks}, the \textit{variance of solvability} shows a \textbf{moderate} range. The top model's average accuracy varies from 51.2\% to 82.3\% (a 31.1\% difference), and the accuracy among \textit{top model-prompt pair}s spans from 68.5\% to 86.1\% (a 17.6\% difference), highlighting a \textbf{extremely significant} \textit{variance of model's top performance}.
The top average accuracy and peak accuracy across the tasks reveal a pattern opposite to the model's distribution: the three symbolic tasks are easier to solve than the Natural-mode task, BAS-B (backadj). Notably, GPT-4 excels as the top model in average performance and also presents the \textit{top model-prompt pair}s across all tasks.

3) \textbf{Stability}: The three most consistent models, based on the lowest \textit{model volatility}, are text-davinci-001 at 1.4, text-curie-001 at 2.3, and GPT-4 at 2.6. In contrast, the models exhibiting the greatest variability, marked by the highest \textit{model volatility} across different prompts, are Llama2 (70B) at 16.2, Vicuna-v1.3 (33B) at 11.9, and Llama2 (13B) at 11.8, highlighting their pronounced sensitivity to prompts. We further explore stability for each task individually.
In the \textbf{BAS-B (backadj)} task, the most stable models are babbage (1.3B) at 1.6, Qwen (7B) at 2.2, and davinci (175B) at 2.8, whereas the most unstable models are Wizardcoder (15B) at 17.0, Vicuna-v1.3 (33B) at 14.0, and Llama2 (70B) at 10.9.
For the \textbf{BAS-C (max-BAS)} task, the models with the lowest \textit{model volatility} are text-davinci-001 at 2.4, text-babbage-001 at 2.6, and InternLM-chat (20B) at 4.3. The most unstable models feature Llama2 (70B) at 18.1, Llama2 (13B) at 15.4, and babbage (1.3B) at 15.1.
In the \textbf{BAS-C (min-BAS)} task, the top three stable models are GPT-4 at 2.2, text-babbage-001 at 2.4, and text-davinci-001 at 2.7. The least stable models are Llama2 (70B) at 23.7, Llama2-chat (70B) at 18.3, and Llama2 (13B) at 15.6.
For the \textbf{BAS-C (mix-BAS)} task, the most consistent models are text-babbage-001 at 1.4, InternLM-chat (20B) at 3.5, and text-ada-001 at 3.9. The models with the greatest variability are Llama2 (70B) at 17.0, curie (6.7B) at 16.3, and ada (0.35B) at 15.1.
\textbf{When comparing across tasks}, babbage (1.3B), while being the most stable in the BAS-B (backadj), appears among the top three most unstable in the BAS-C (max-BAS) task. Text-babbage-001 ranks among the top three most stable models in all symbolic tasks, while Llama2 (70B) features as one of the top three most unstable models across all four tasks.

4) \textbf{Open-Limited Ratio}: The ratio of open-access to limited-access models among the top five models with the highest average accuracy being 0:5 highlights a \textbf{large} \textit{open-limited gap} of the scenario.

Next, we delve into \textit{prompt gain} in BAS:

1) \textbf{Top Gain}: As illustrated in Figure \ref{fig:Heatmap_of_Backdoor_Adjustment_Set}(b), the two prompts that lead to the highest average accuracy gains over the basic prompt are 3-shot IcL with a 12.1\% gain and 1-shot IcL with a 9.8\% gain. The most significant increase in accuracy compared to the basic prompt is achieved by Llama2 (70B) using 3-shot IcL, with a 39.5\% increase. We continue with a detailed, task-specific analysis. 
Figure \ref{fig:Heatmap_of_gain_of_Backdoor_Adjustment_Set} displays the gains across all tasks within the scenario. In the \textbf{BAS-B (backadj)} task, 3-shot IcL at 7.0\% and 1-shot IcL at 6.2\% are the top two prompts for average accuracy gain over the basic prompt, with Wizardcoder (15B) and 1-shot IcL showing the most substantial increase at 37.9\%. 
In the \textbf{BAS-C (max-BAS)} task, 3-shot IcL at 12.3\% and 1-shot IcL at 11.0\% lead in average accuracy gains, with Llama2 (70B) and 3-shot IcL achieving the highest increase at 40.4\%. 
In the \textbf{BAS-C (min-BAS)} task, the top gains come from 3-shot IcL at 15.3\% and 1-shot IcL at 11.1\%, with Llama2 (70B) and 3-shot IcL marking the most significant increase of 56.8\%. 
For the \textbf{BAS-C (mix-BAS)} task, the highest gains are noted with 3-shot IcL at 13.7\% and 1-shot IcL at 11.0\%, with Llama2 (70B) and 3-shot IcL showing the highest increase of 41.9\%.
\textbf{Upon evaluating across tasks}, a consistent trend emerges: 3-shot IcL stands out as the top 1 performing prompt in terms of average gain, particularly when paired with Llama2 (70B) or Wizardcoder (15B), outperforming other \textit{model-prompt pair}s. 1-shot IcL secures its place as the second most effective prompt for average accuracy improvement. Additionally, the choice-selection symbolic tasks (BAS-C (max-BAS), BAS-C (min-BAS), BAS-C (mix-BAS)) exhibit larger top gains compared to the Natural-mode task, indicating that IcL prompts significantly aid models in achieving better comprehension.

2) \textbf{Exceptions}: The highly effective prompt, 3-shot IcL, boosts the performance of most models in the scenario, but with exceptions including Baichuan1-chat (13B), Baichuan2-chat (7B), text-davinci-001, text-curie-001, and InternLM-chat (20B). All prompts are capable of enhancing the performance of Wizardcoder (15B) and text-davinci-002 beyond the basic prompt=. However, none of the prompts manage to improve the performance of text-davinci-001 and InternLM-chat (20B) above the basic prompt.
In the \textbf{BAS-B (backadj)} task, 3-shot IcL does not work well on Baichuan1-chat (13B), Llama2 (7B), InternLM-chat (20B), Baichuan1 (7B), Baichuan2-chat (7B), text-davinci-001, Qwen (7B), and Llama2 (13B). All prompts create a positive average \textit{prompt gain} for Wizardcoder (15B) and text-davinci-002. However, Baichuan2-chat (7B) and Qwen (7B) see no performance gain from any prompt.
For the \textbf{BAS-C (max-BAS)} task, 3-shot IcL fails to create a positive average \textit{prompt gain} with certain models, including Baichuan1-chat (13B), InternLM-chat (7B), text-babbage-001, text-davinci-001, InternLM-chat (20B), and GPT-4, while all prompts boost davinci (175B)'s performance over the basic prompt. Text-babbage-001, however, does not benefit from any prompt.
In the \textbf{BAS-C (min-BAS)} task, 3-shot IcL is ineffective with  Baichuan1-chat (13B), Baichuan2-chat (7B), text-babbage-001, text-curie-001, InternLM-chat (20B), and Claude2. All prompts help improve Wizardcoder (15B) and text-davinci-002 over their performance on basic prompt. InternLM-chat (20B) and Claude2 do not receive any improvement from prompts.
The \textbf{BAS-C (mix-BAS)} task shows that the top performing prompt, 3-shot IcL, do not give positive average \textit{prompt gain} to Baichuan1-chat (13B), Baichuan2-chat (7B), text-ada-001, text-davinci-001, InternLM-chat (7B), InternLM-chat (20B), and text-curie-001. Moreover, no prompt is capable of improving InternLM-chat (20B)'s performance above the basic prompt.
\textbf{Across all tasks}, 3-shot IcL does not manage to enhance performance for Baichuan1-chat (13B) and InternLM-chat (20B) over their basic prompt performance, indicating its limitations with these models.

3) \textbf{Stability}: Regarding stability within the scenario, the two most stable prompts are EF with a \textit{prompt volatility} of 3.6 and adversarial ignore with a \textit{prompt volatility} of 4.4. Conversely, the most variable prompts, indicated by the highest \textit{prompt volatility}, are 3-shot IcL at 11.4 and 1-shot IcL at 10.3. An \textit{average model-prompt-gain volatility} (\textit{AMPGV}) of 7.6, illustrating a \textbf{medium} level of \textit{prompt dependence} across the scenario. Stability is further assessed on a task-specific basis.
In the \textbf{BAS-B (backadj)} task, the most stable prompts are EF at 7.4 \textit{prompt volatility} and adversarial doubt at 7.6 \textit{prompt volatility}, while the least stable are 3-shot IcL at 10.4 \textit{prompt volatility} and 0-shot CoT at 10.1 \textit{prompt volatility}. The task shows a \textbf{medium} \textit{prompt dependence} with an \textit{average model-prompt-gain volatility} (\textit{AMPGV}) of 6.6.
For the \textbf{BAS-C (max-BAS)} task, EF and adversarial ignore stand out as the most stable with \textit{prompt volatility} of 4.0 and 5.1, respectively, and the least stable prompts are manual CoT at 14.1 and 1-shot IcL at 13.2. The task reflects a \textbf{medium} \textit{prompt dependence} with an \textit{AMPGV} of 9.8.
In the \textbf{BAS-C (min-BAS)} task, the top stable prompts are adversarial ignore at 5.3 \textit{prompt volatility} and adversarial doubt at 6.5 \textit{prompt volatility}, whereas the most unstable are 3-shot IcL at 16.1 and 1-shot IcL at 15.4. This task also shows a \textbf{medium} level of \textit{prompt dependence} with an \textit{AMPGV} of 9.8.
The \textbf{BAS-C (mix-BAS)} task reveals EF and adversarial ignore as the most stable prompts with \textit{prompt volatility} of 4.3 and 4.8, respectively. The least stable prompts are 3-shot IcL at 14.2 and 1-shot IcL at 13.8, indicating a \textbf{medium} \textit{prompt dependence} with an \textit{AMPGV} of 9.4.
\textbf{After reviewing all tasks}, the \textit{AMPGV} spans a range from 6.6 to 9.8, reflecting the \textbf{narrow} \textit{variance of prompt dependence}. The Natural-mode task (BAS-B (backadj)) demonstrates more stability to prompts compared to the symbolic tasks.

Finally, we analyze \textit{language proficiency} in BAS:

1) \textbf{English vs. Chinese}: Figure \ref{fig:Backdoor_Adjustment_Set_Language} shows that models generally perform better on the English test set over the Chinese one, with 24 out of 28 models favoring English.

2) \textbf{Accuracy Difference}: Significant differences in accuracy between English and Chinese, with a preference for English, are noted in models like GPT-3.5-Turbo (English better than Chinese with an average of 15.6\%), ada (0.35B) (13.1\%), and text-ada-001 (12.2\%). In contrast, models such as Baichuan1-chat (13B) (6.1\%), InternLM-chat (20B) (3.1\%), and curie (6.7B) (1.8\%) are more proficient in Chinese.

\paragraph{Frontdoor adjustment set.}
\begin{figure}[t]
\centering
\subfigure[Model performance of FAS]{
\begin{minipage}{8.5cm}
\centering
\includegraphics[width=1\linewidth]{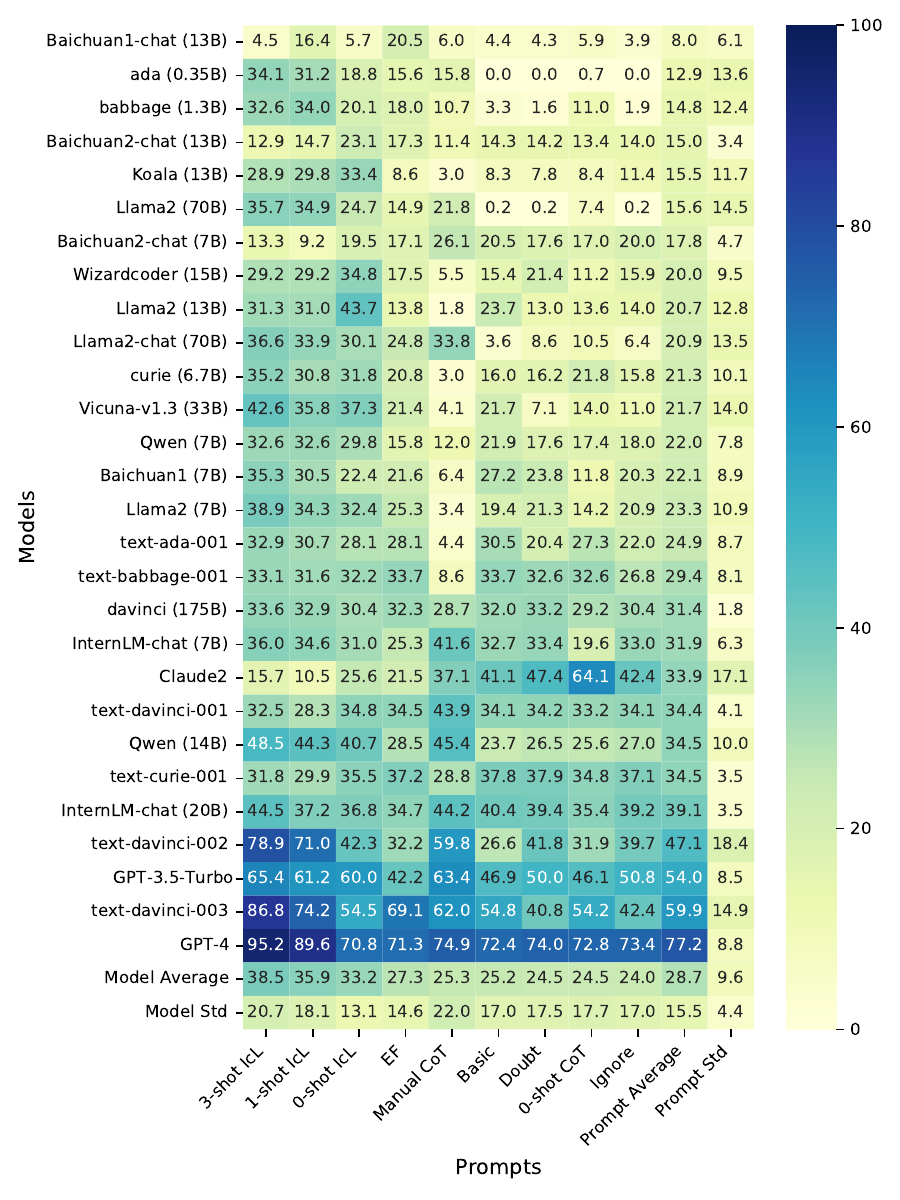}
\end{minipage}
}
\subfigure[\textit{Prompt gain} of FAS]{
\begin{minipage}{8.5cm}
\centering
\includegraphics[width=1\linewidth]{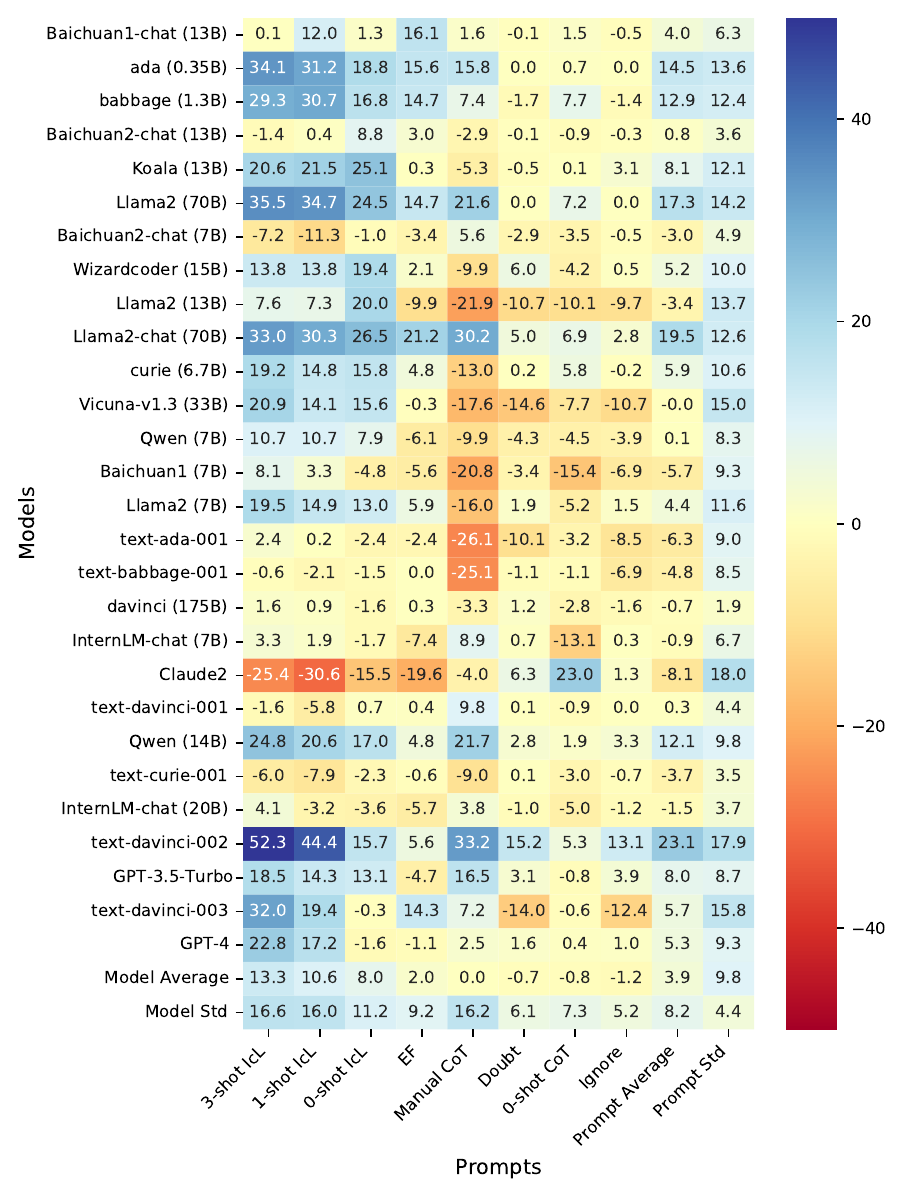}
\end{minipage}
}
\caption[Heatmap of FAS]{\textbf{Heatmap of FAS.} The models and prompts are sorted by their averages.}
\label{fig:Heatmap_of_Frontdoor_Adjustment_Set}
\end{figure}

\begin{figure}
    \centering
    \includegraphics[width=0.8\linewidth]{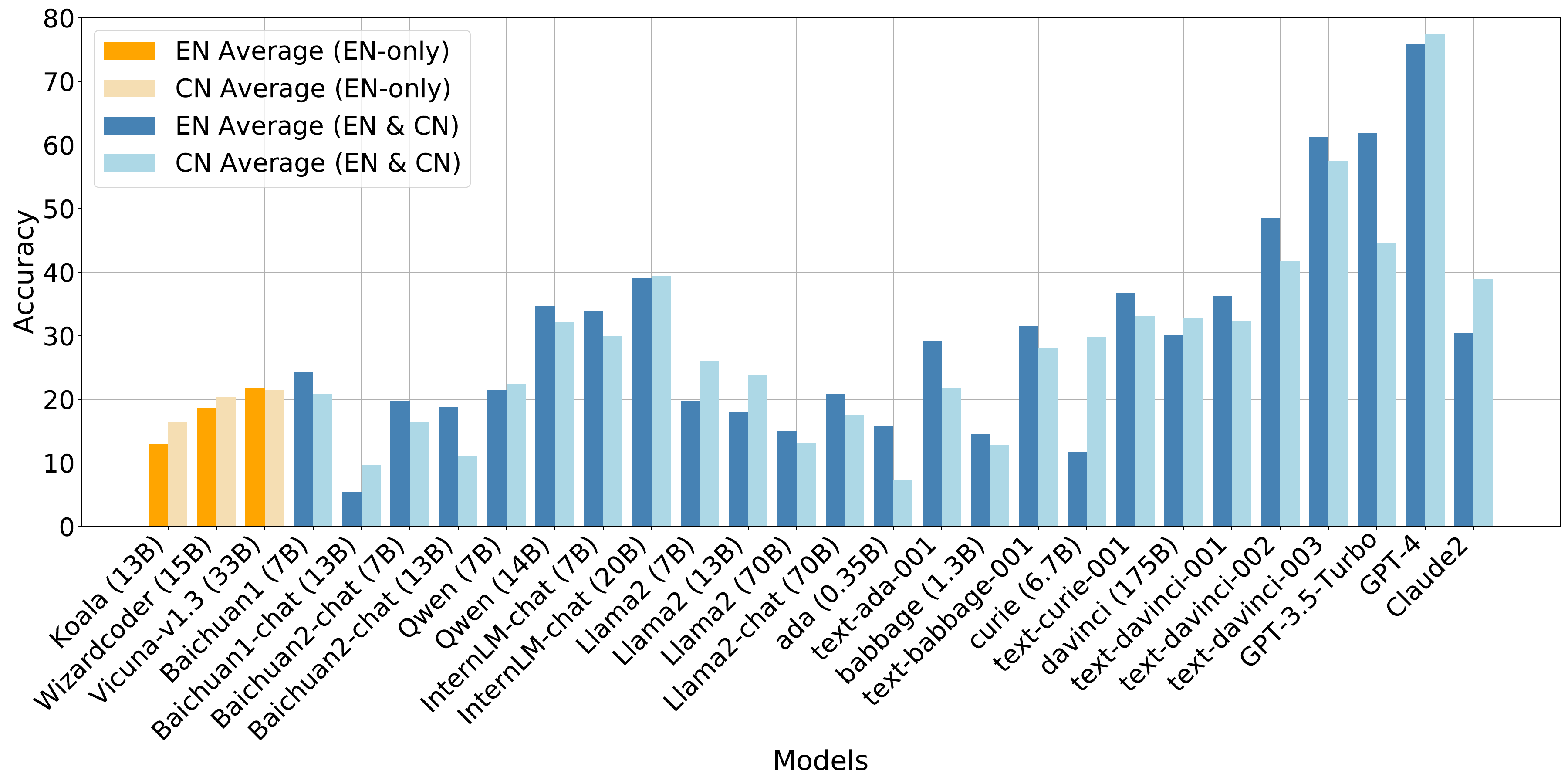}
    \caption[Language comparison of FAS]{\textbf{Language comparison of FAS.} The dark legend signifies the average performance of the model on an English test set, whereas the light legend denotes the average performance of the model on the Chinese test set. The yellow legend indicates a model trained exclusively on English datasets, while the blue legend represents a model trained on both English and Chinese datasets.}
    \label{fig:Frontdoor_Adjustment_Set_Language}
\end{figure}

Regarding model performance in FAS:
1) \textbf{Distribution}: Figure \ref{fig:Distribution_of_Intervention}(e) displays the distribution for all \textit{model-prompt pair}s within the FAS, noting a median of 29.0\% and a third quartile of 35.5\%. This scenario is considered to have a \textbf{hard} \textit{understandability} since the median accuracy falls below the random guess benchmark of 33.3\%.
2) \textbf{Top Accuracy}: Figure \ref{fig:Heatmap_of_Frontdoor_Adjustment_Set}(a) reveals the leading three models by average accuracy: GPT-4 at 77.2\%, text-davinci-003 at 59.9\%, and GPT-3.5-Turbo at 54.0\%. GPT-4, employing 3-shot IcL, is the \textit{top model-prompt pair} with a 95.2\% accuracy. With the top model's average accuracy surpassing 70\%, the \textit{solvability} of this scenario is \textbf{solvable}.
3) \textbf{Stability}: The most prompt-sensitive models, indicated by the \textit{model volatility} described in \cref{metric:model}, are text-davinci-002 at 18.4, Claude2 at 17.1, and text-davinci-003 at 14.9. In contrast, the most stable models include davinci (175B) at 1.8, text-curie-001 at 3.4, and Baichuan2-chat (13B) at 3.5.
4) \textbf{Open-Limited Ratio}: With a 1:4 ratio of open-access to limited-access models among the top five models with the highest average accuracy, the \textit{open-limited gap} is \textbf{moderate}.

Regarding prompt performance in FAS:
1) \textbf{Top Gain}: Figure \ref{fig:Heatmap_of_Frontdoor_Adjustment_Set}(b) identifies the top two prompts for average accuracy gain over the basic prompt as 3-shot IcL at 13.3\% and 1-shot IcL at 10.6\%, with text-davinci-002 using 3-shot IcL showcasing the most significant improvement of 52.3\%.
2) \textbf{Exceptions}: The highest-performing prompt, 3-shot IcL, does not give positive average prompt gains to several models, including Baichuan2-chat (13B) and Claude2. Nonetheless, all prompts are capable of boosting the performance of Llama2-chat (70B), Qwen (14B), and text-davinci-002 over the basic prompt.
3) \textbf{Stability}: The most stable prompts, maintaining the lowest \textit{prompt volatility}, are adversarial ignore at 5.2 and adversarial doubt at 6.1. Meanwhile, the least stable prompts, with the highest \textit{prompt volatility}, are 3-shot IcL at 16.6 and manual CoT at 16.2. As the computed \textit{average model-prompt-gain volatility} (\textit{AMPGV}) is 9.8. Therefore, the scenario has a \textbf{medium} \textit{prompt dependence}.

Regarding \textit{language proficiency} in FAS:
1) \textbf{English vs. Chinese}: Figure \ref{fig:Frontdoor_Adjustment_Set_Language} shows models generally perform better on the English test set compared to the Chinese set, with 17 of 28 models favoring English.
2) \textbf{Accuracy Difference}: Significant advantages for English over Chinese are seen in models like GPT-3.5-Turbo (17.3\%) and ada (0.35B) (8.5\%). Conversely, models like curie (6.7B) and Claude2 demonstrate a higher proficiency in Chinese.

\paragraph{Instrumental variable.}
\begin{figure}[t]
\centering
\subfigure[Model performance of IV]{
\begin{minipage}{8.5cm}
\centering
\includegraphics[width=1\linewidth]{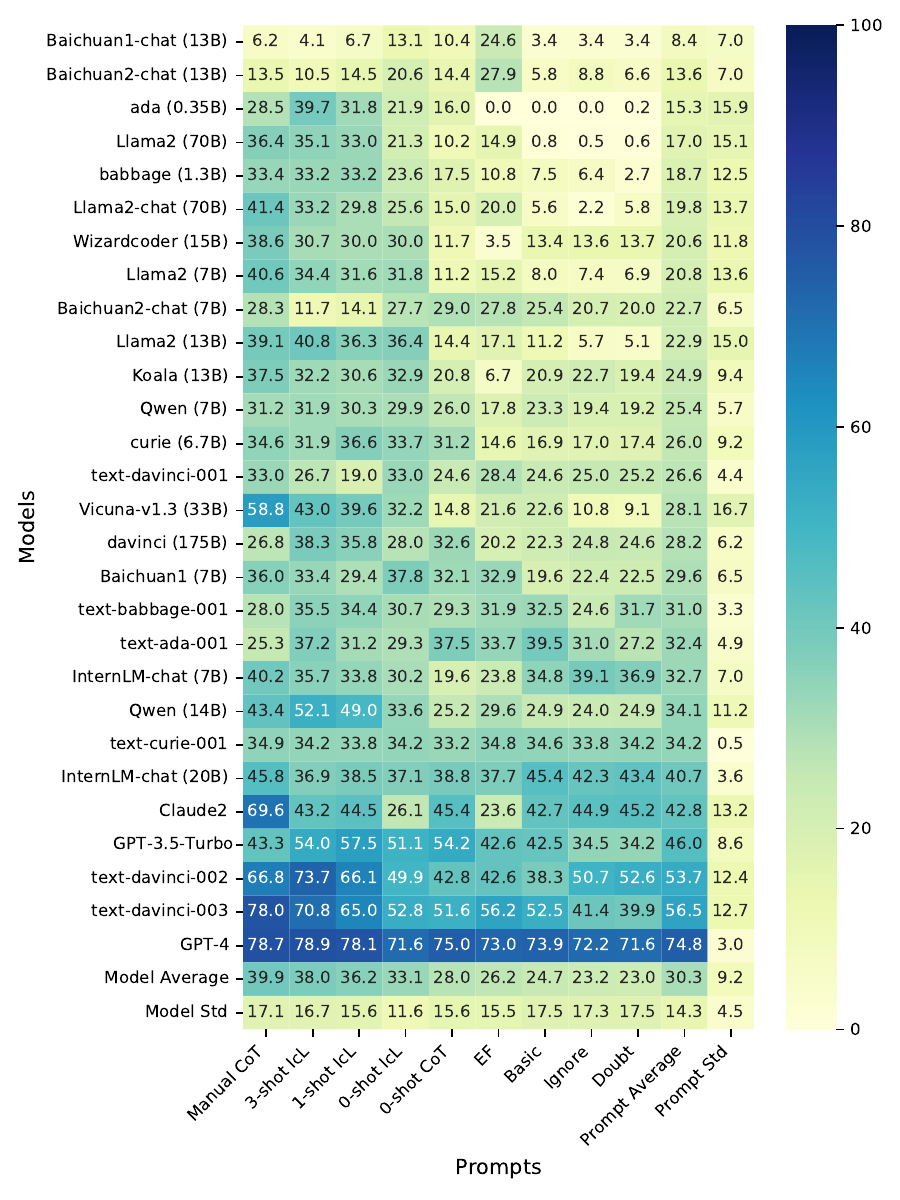}
\end{minipage}
}
\subfigure[\textit{Prompt gain} of IV]{
\begin{minipage}{8.5cm}
\centering
\includegraphics[width=1\linewidth]{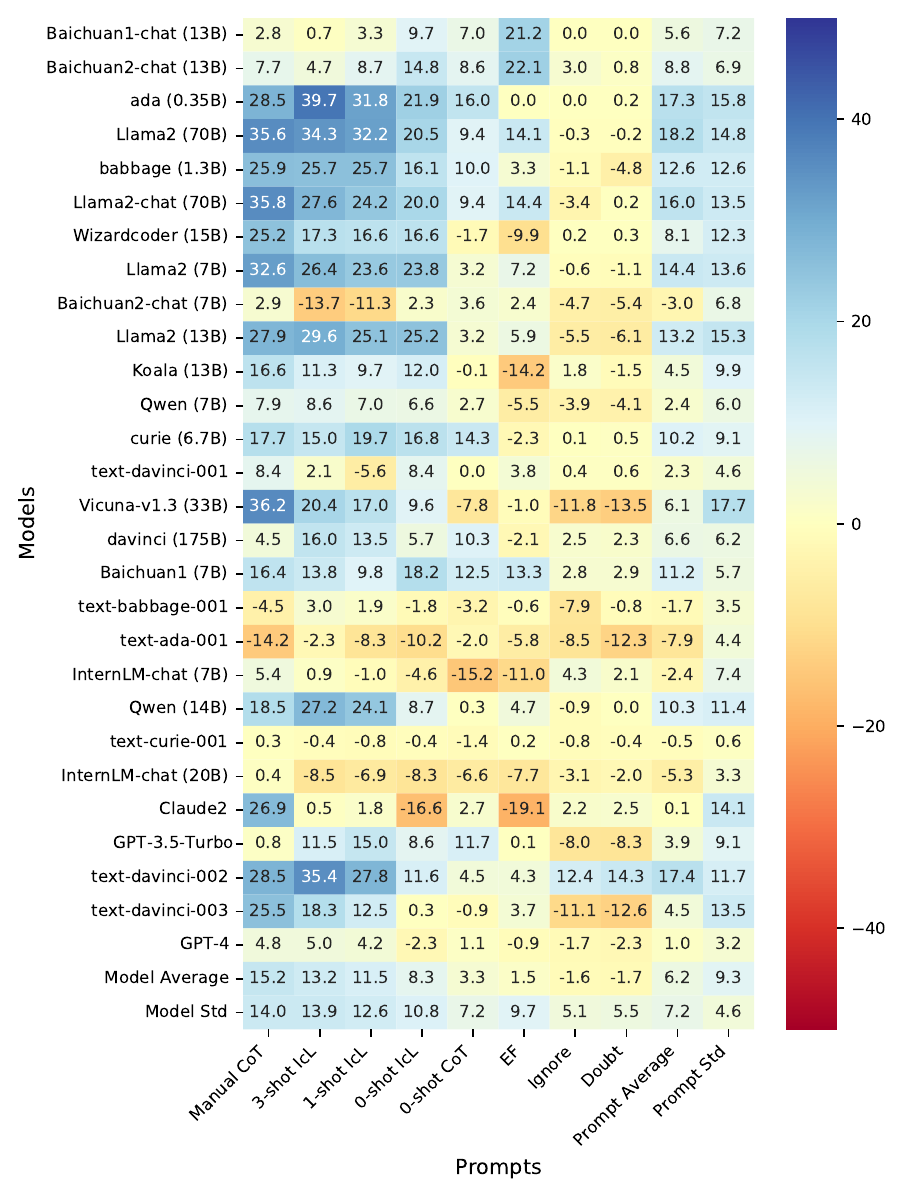}
\end{minipage}
}
\caption[Heatmap of IV]{\textbf{Heatmap of IV.} The models and prompts are sorted by their averages.}
\label{fig:Heatmap_of_Instrumental_Variable}
\end{figure}

\begin{figure}
    \centering
    \includegraphics[width=0.8\linewidth]{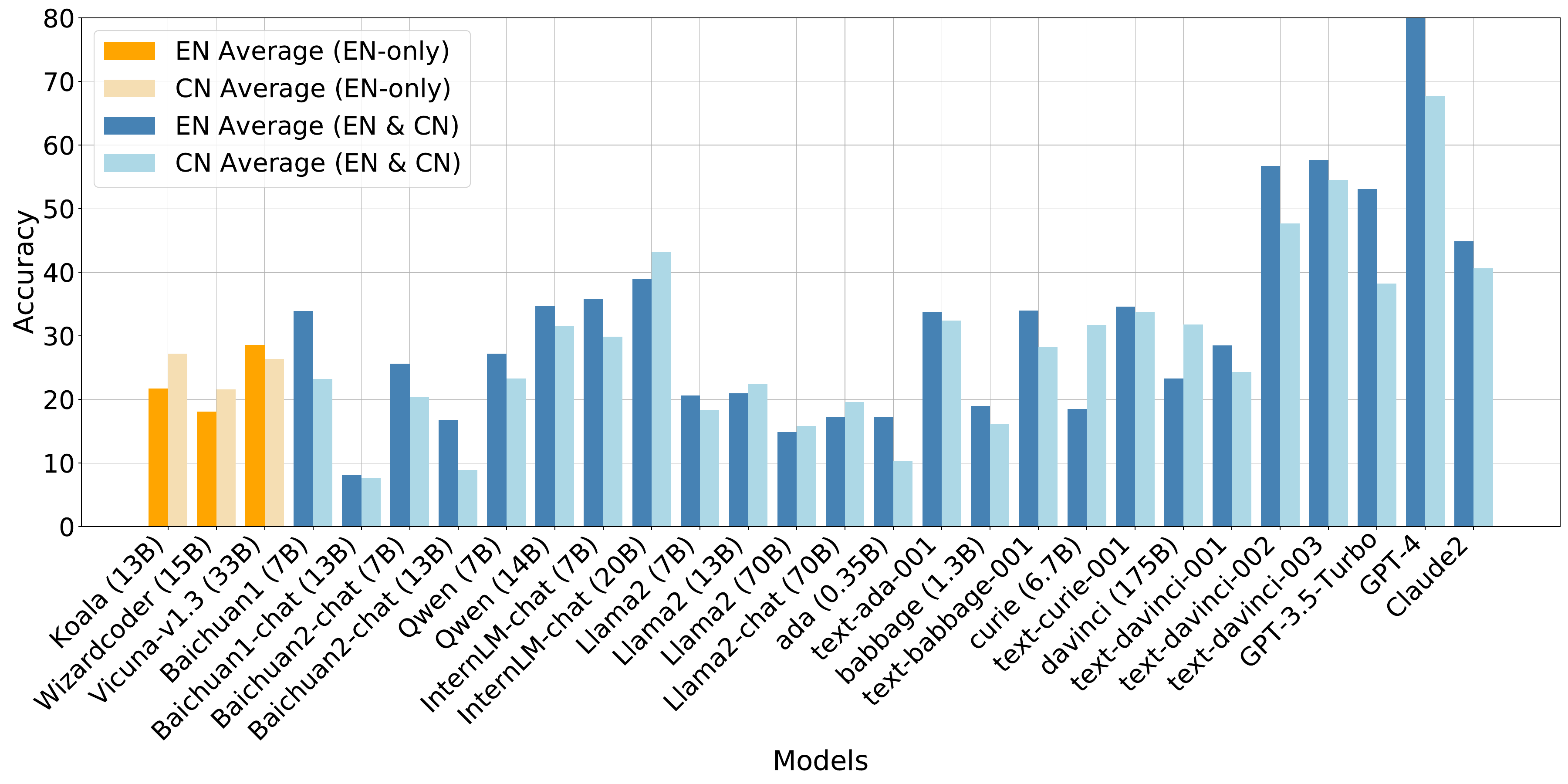}
    \caption[Language comparison of IV]{\textbf{Language comparison of IV.} The dark legend signifies the average performance of the model on an English test set, whereas the light legend denotes the average performance of the model on the Chinese test set. The yellow legend indicates a model trained exclusively on English datasets, while the blue legend represents a model trained on both English and Chinese datasets.}
    \label{fig:Instrumental_Variable_Language}
\end{figure}

Regarding model performance in IV:
1) \textbf{Distribution}: Figure \ref{fig:Distribution_of_Intervention}(f) showcases the distribution for all \textit{model-prompt pair}s within IV, revealing a median of 30.7\% and a third quartile of 37.9\%. This indicates the \textit{understandability} of the scenario is \textbf{hard} as the median accuracy falls below the random guess benchmark of 33.3\%.
2) \textbf{Top Accuracy}: According to Figure \ref{fig:Heatmap_of_Instrumental_Variable}(a), the leading three models by average accuracy are GPT-4 at 74.8\%, text-davinci-003 at 56.5\%, and text-davinci-002 at 53.7\%. GPT-4, employing 3-shot IcL, achieves a top score of 78.9\%, suggesting the \textit{solvability} of this scenario as \textbf{challenging} since the \textit{top model-prompt pair}'s performance doesn't reach 80\%.
3) \textbf{Stability}: The models most susceptible to prompt variations, as shown by the \textit{model volatility} discussed in \cref{metric:model}, are Vicuna-v1.3 (33B) at 16.7, ada (0.35B) at 15.9, and Llama2 (13B) at 15.1. Conversely, the most stable models include text-curie-001 at 0.5, GPT-4 at 3.0, and InternLM-chat (20B) at 3.3.
4) \textbf{Open-Limited Ratio}: With a 0:5 ratio of open-access to limited-access models among the top five models with the highest average accuracy, this indicates a \textbf{large} \textit{open-limited gap} in the scenario.

In terms of \textit{prompt gain} in IV:
1) \textbf{Top Gain}: Figure \ref{fig:Heatmap_of_Instrumental_Variable}(b) highlights the top two prompts for average accuracy gain over the basic prompt as manual CoT at 15.2\% and 3-shot IcL at 13.2\%. Ada (0.35B), using 3-shot IcL, showcases the most significant improvement of 39.7\%.
2) \textbf{Exceptions}: The highest-performing prompt, manual CoT, is not effective in generating positive average \textit{prompt gain} in specific models including text-babbage-001 and text-ada-001. However, all prompts manage to enhance the performance of Baichuan2-chat (13B), Baichuan1 (7B), and text-davinci-002 over the basic prompt, with text-ada-001 being the exception where no prompt leads to improvement.
3) \textbf{Stability}: The most stable prompts, exhibiting the lowest \textit{prompt volatility}, are adversarial ignore at 5.1 and adversarial doubt at 5.5. The least stable prompts, with the largest \textit{prompt volatility}, are manual CoT at 14.0 and 3-shot IcL at 13.9. The computed \textit{average model-prompt-gain volatility} (\textit{AMPGV}) of 9.3 indicates a \textbf{medium} \textit{prompt dependence} in this scenario.

In terms of \textit{language proficiency} in IV:
1) \textbf{English vs. Chinese}: Figure \ref{fig:Instrumental_Variable_Language} shows that models in IV perform better on the English test set than on the Chinese set, with 20 of 28 models favoring English.
2) \textbf{Accuracy Difference}: Significant advantages for English over Chinese are seen in models like GPT-3.5-Turbo (14.9\%) and GPT-4 (14.0\%). In contrast, models such as curie (6.7B) and davinci (175B) exhibit higher proficiency in Chinese.

\paragraph{Collider bias.}
\begin{figure}[t]
\centering
\subfigure[Model performance of CB]{
\begin{minipage}{8.5cm}
\centering
\includegraphics[width=1\linewidth]{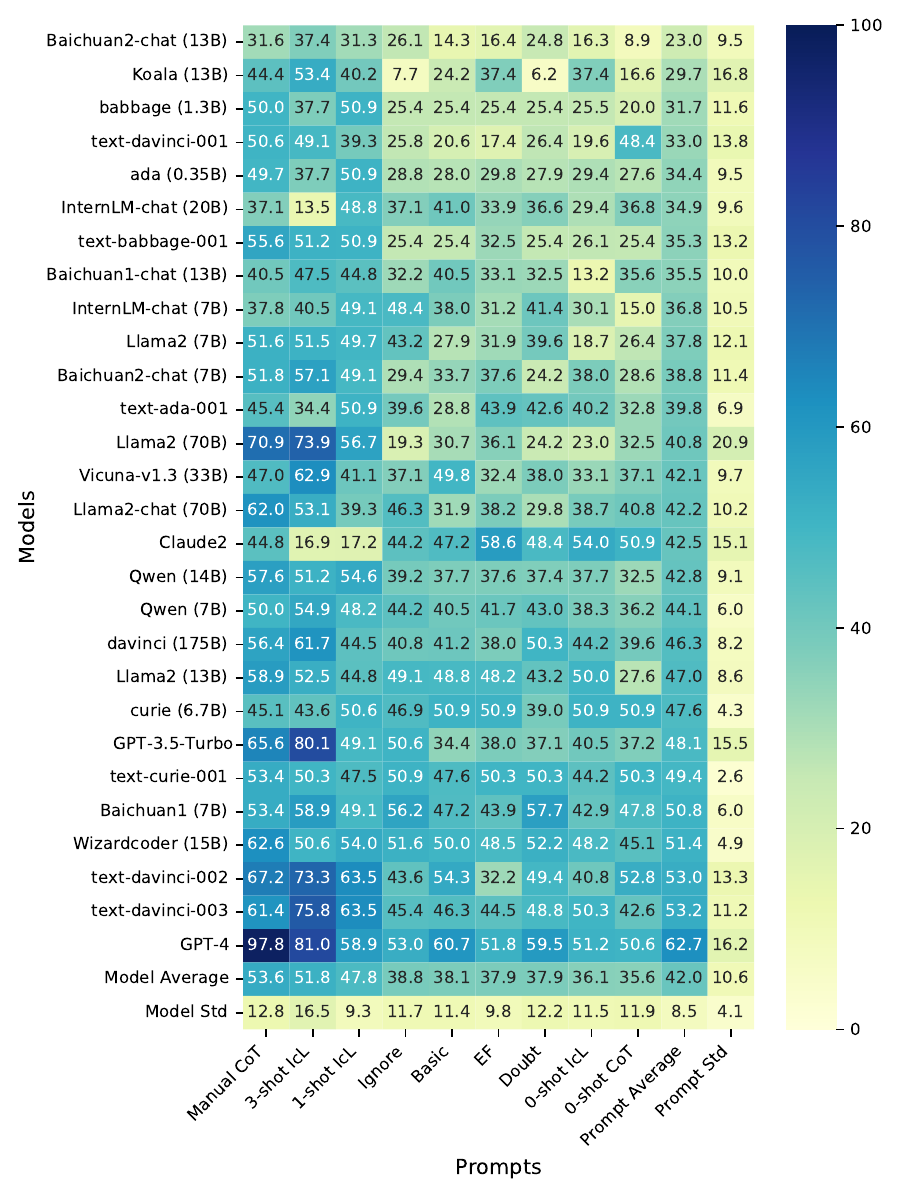}
\end{minipage}
}
\subfigure[\textit{Prompt gain} of CB]{
\begin{minipage}{8.5cm}
\centering
\includegraphics[width=1\linewidth]{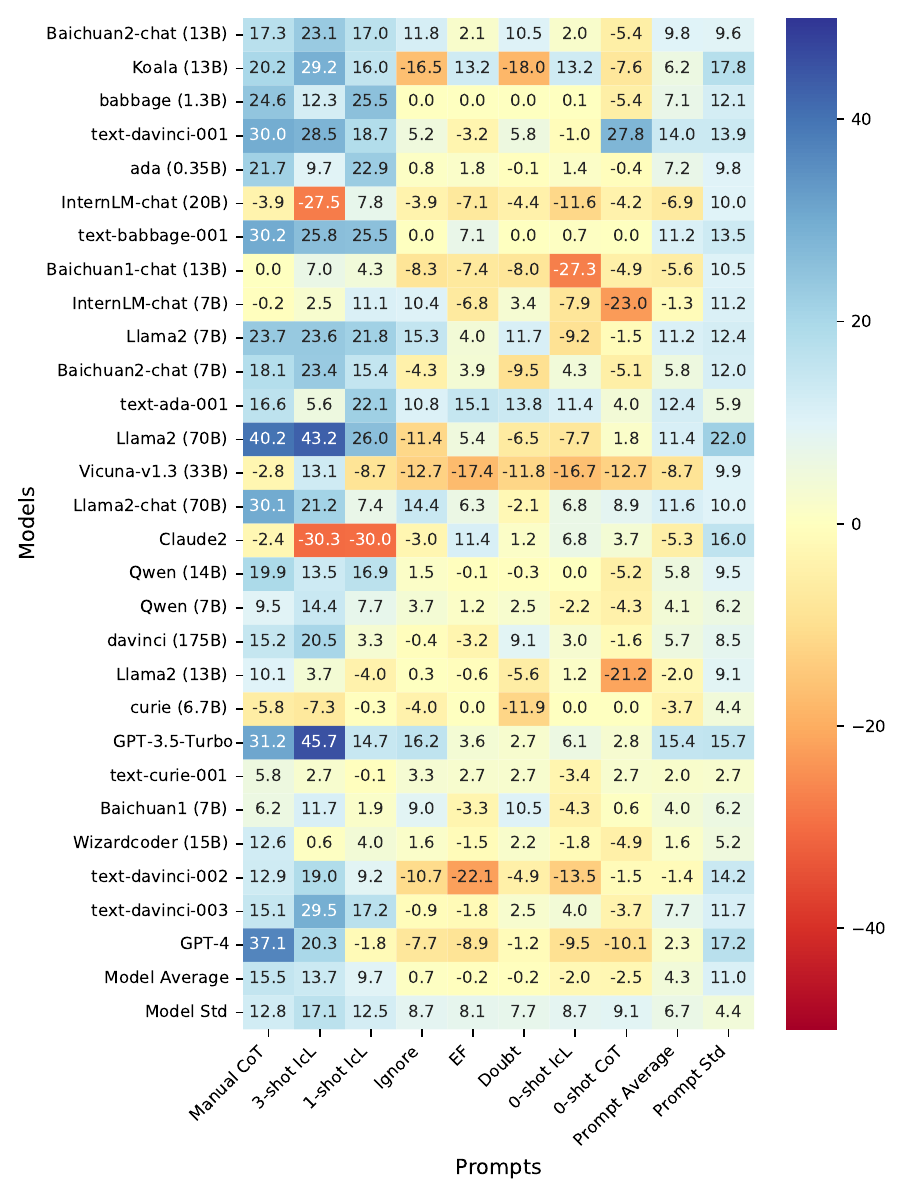}
\end{minipage}
}
\caption[Heatmap of CB]{\textbf{Heatmap of CB.} The models and prompts are sorted by their averages.}
\label{fig:Heatmap_of_Collider_Bias}
\end{figure}

\begin{figure}
    \centering
    \includegraphics[width=0.8\linewidth]{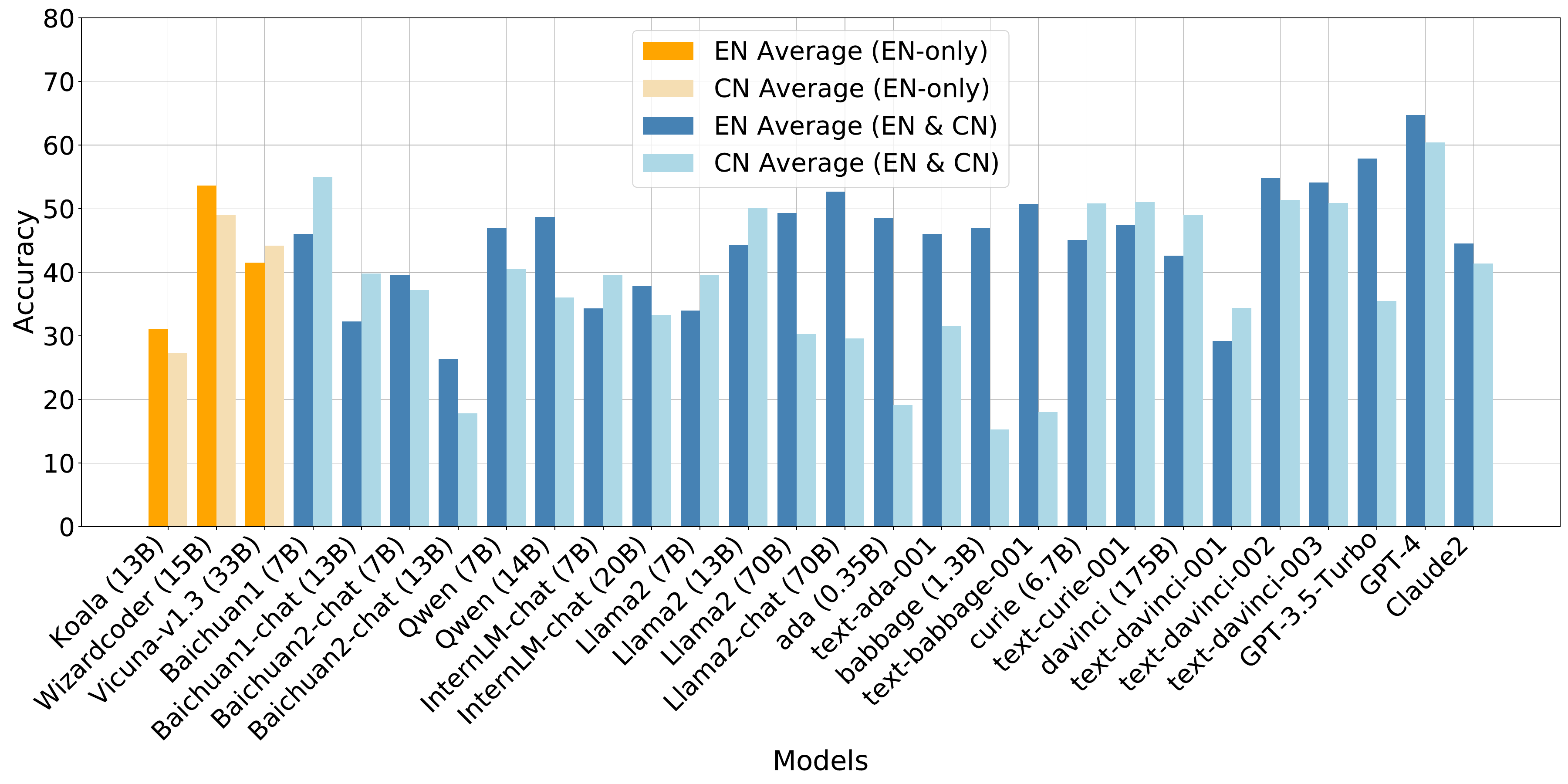}
    \caption[Language comparison of CB]{\textbf{Language comparison of CB.} The dark legend signifies the average performance of the model on an English test set, whereas the light legend denotes the average performance of the model on the Chinese test set. The yellow legend indicates a model trained exclusively on English datasets, while the blue legend represents a model trained on both English and Chinese datasets.}
    \label{fig:Collider_Bias_Language}
\end{figure}

Regarding model performance in CB:
1) \textbf{Distribution}: Figure \ref{fig:Distribution_of_Association}(g) showcases the distribution for all \textit{model-prompt pair}s within CB, noting a median of 43.0\% and a third quartile of 50.6\%. This indicates the \textit{understandability} of the scenario is \textbf{hard} since the median accuracy falls below the random guess benchmark of 50.0\%.
2) \textbf{Top Accuracy}: Figure \ref{fig:Heatmap_of_Collider_Bias}(a) reveals the top three models by average accuracy are GPT-4 at 62.7\%, text-davinci-003 at 53.2\%, and text-davinci-002 at 53.0\%. The \textit{top model-prompt pair} is GPT-4 with manual CoT, which achieves an impressive 97.8\%, suggesting the \textit{solvability} of this scenario as \textbf{potentially solvable}.
3) \textbf{Stability}: The models most sensitive to prompt variations, as shown by the \textit{model volatility} described in \cref{metric:model}, are Llama2 (70B) at 20.9, Koala (13B) at 16.8, and GPT-4 at 16.2. Conversely, the most stable models are text-curie-001 at 2.6, curie (6.7B) at 4.3, and Wizardcoder (15B) at 4.9.
5) \textbf{Open-Limited Ratio}: A 2:3 ratio of open-access to limited-access models among the top five models indicates a \textbf{small} \textit{open-limited gap}.

In terms of \textit{prompt gain} in CB:
1) \textbf{Top Gain}: Figure \ref{fig:Heatmap_of_Collider_Bias}(b) identifies the top two prompts for average accuracy gain over the basic prompt as manual CoT at 15.5\% and 3-shot IcL at 13.7\%. The greatest improvement over the basic prompt is noted with GPT-3.5-Turbo using 3-shot IcL, marking a gain of 45.7\%.
2) \textbf{Exceptions}: Manual CoT, the highest-performing prompt, does not create a positive average \textit{prompt gain} for certain models, including InternLM-chat (20B), InternLM-chat (7B), Vicuna-v1.3 (33B), Claude2, and curie (6.7B). Yet, all prompts manage to enhance the performance of text-ada-001 and GPT-3.5-Turbo over the basic prompt.
3) \textbf{Stability}: The most stable prompts, with the lowest \textit{prompt volatility}, are adversarial doubt at 7.7 and EF at 8.1. In contrast, the least stable prompts, with the highest \textit{prompt volatility}, are 3-shot IcL at 17.1 and manual CoT at 12.8. The \textit{average model-prompt-gain volatility} (\textit{AMPGV}) is 11.0, indicating a \textbf{high} \textit{prompt dependence} within the scenario.

Regarding \textit{language proficiency} in CB:
1) \textbf{English vs. Chinese}: As shown in Figure \ref{fig:Collider_Bias_Language}, models tend to perform better on the English test set than on the Chinese set, with 18 of 28 models favoring English.
2) \textbf{Accuracy Difference}: Significant performance advantages for English over Chinese are seen in models like text-babbage-001 (32.7\%), babbage (1.3B) (31.7\%), and ada (0.35B) (29.4\%). Conversely, models such as Baichuan1 (7B) (8.9\%), Baichuan1-chat (13B) (7.5\%), and davinci (175B) (6.4\%) demonstrate a higher proficiency in Chinese.

\subsubsection{Counterfactuals}
\label{scenario:counterfactual}
\begin{figure}[t]
\centering  
\subfigure[Distribution of CR]{   
\begin{minipage}{3.9cm}
\centering   
    \includegraphics[width=1\linewidth]{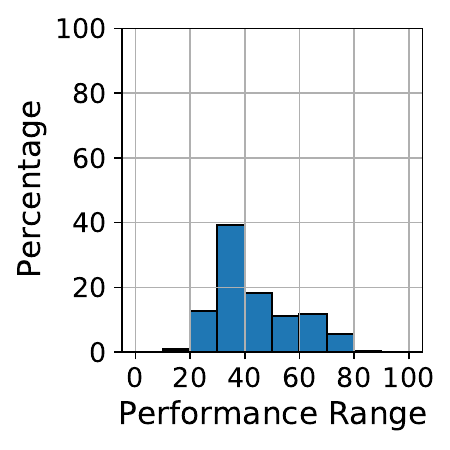}
\end{minipage}
}
\subfigure[Distribution of AC]{  
\begin{minipage}{3.9cm}
\centering  
    \includegraphics[width=1\linewidth]{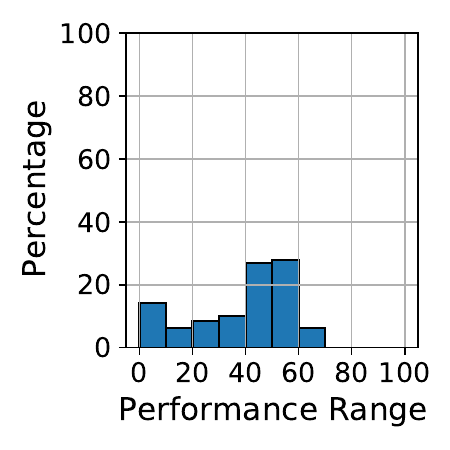}
\end{minipage}
}
\subfigure[Distribution of ETT]{  
\begin{minipage}{3.9cm}
\centering  
    \includegraphics[width=1\linewidth]{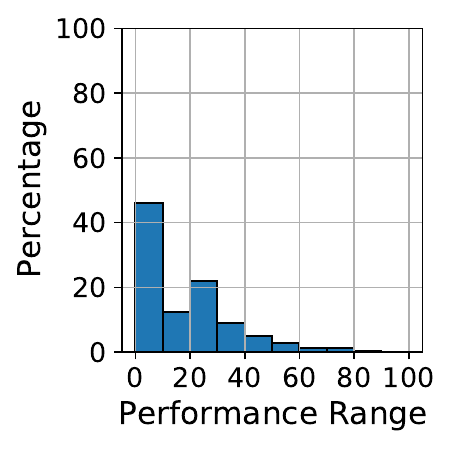}
\end{minipage}
}
\subfigure[Distribution of NDE]{   
\begin{minipage}{3.9cm}
\centering   
    \includegraphics[width=1\linewidth]{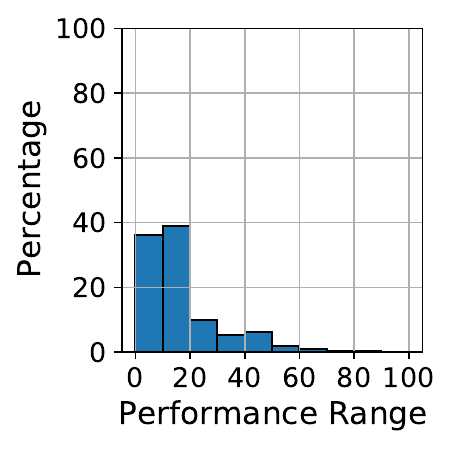}
\end{minipage}
}
\subfigure[Distribution of NIE]{  
\begin{minipage}{3.9cm}
\centering    
    \includegraphics[width=1\linewidth]{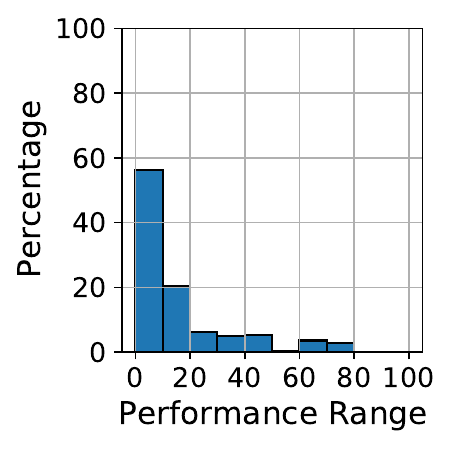}
\end{minipage}
}
\subfigure[Distribution of PN]{  
\begin{minipage}{3.9cm}
\centering   
    \includegraphics[width=1\linewidth]{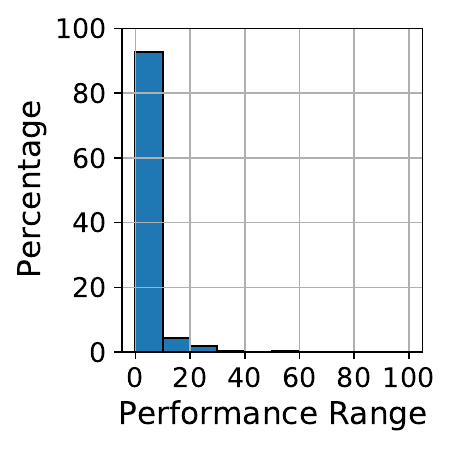}
\end{minipage}
}
\subfigure[Distribution of PS]{  
\begin{minipage}{3.9cm}
\centering   
    \includegraphics[width=1\linewidth]{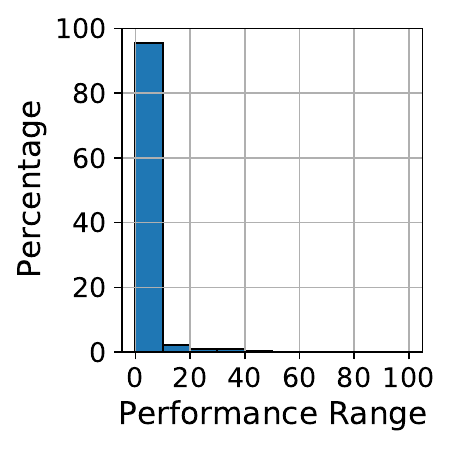}
\end{minipage}
}
\subfigure[Distribution of CEG]{  
\begin{minipage}{3.9cm}
\centering   
    \includegraphics[width=1\linewidth]{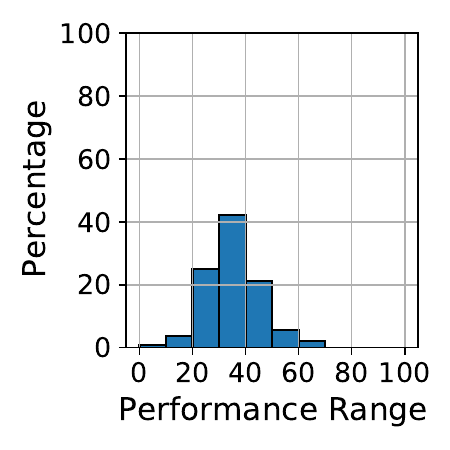}
\end{minipage}
}
\caption[Distribution of counterfactuals]{\textbf{Distribution of counterfactuals.} The horizontal coordinate represents the accuracy of the model and the vertical coordinate represents the percentage distribution corresponding to a certain accuracy interval.}   
\label{fig:Distribution_of_counterfactual}   
\end{figure}

\paragraph{Counterfactual reasoning.}
\begin{figure}[t]
\centering
\subfigure[Model performance of CR]{
\begin{minipage}{8.5cm}
\centering
\includegraphics[width=1\linewidth]{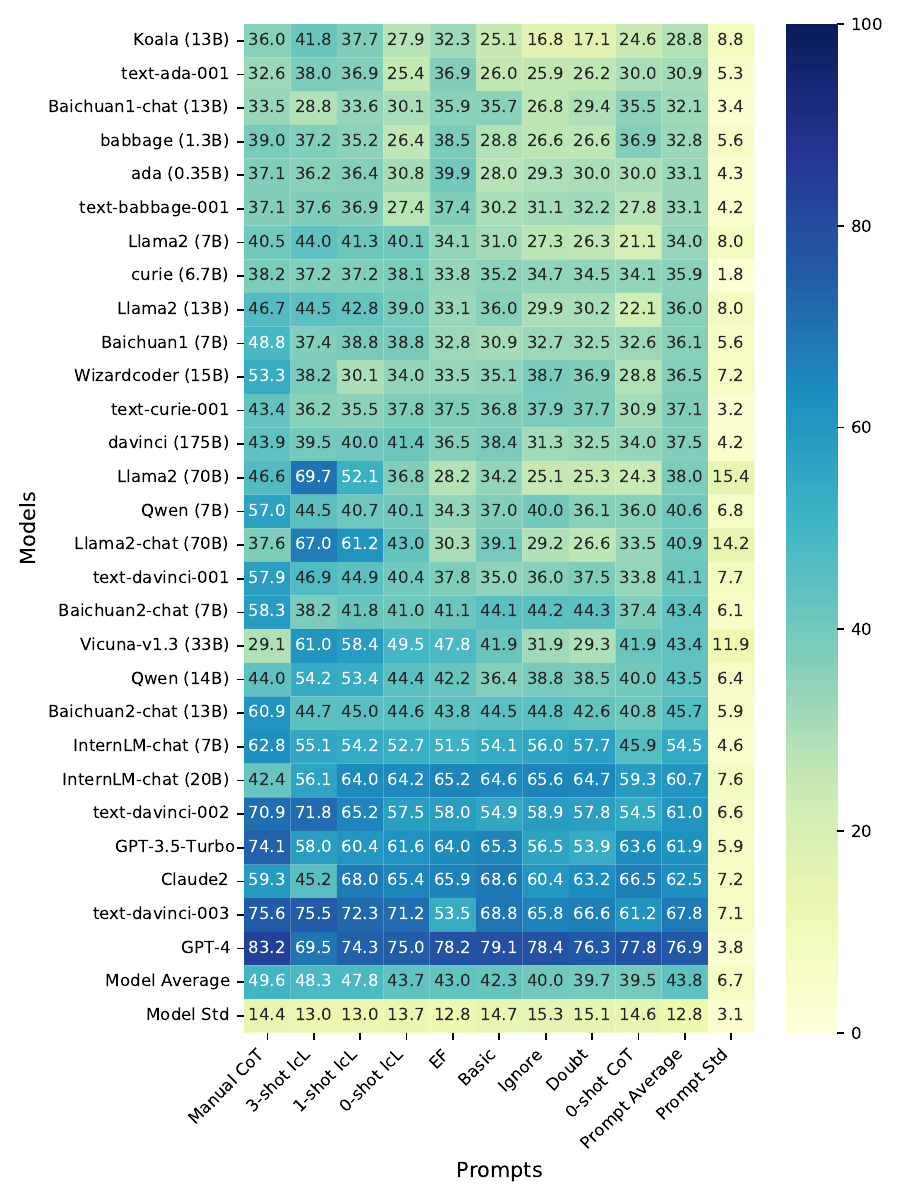}
\end{minipage}
}
\subfigure[\textit{Prompt gain} of CR]{
\begin{minipage}{8.5cm}
\centering
\includegraphics[width=1\linewidth]{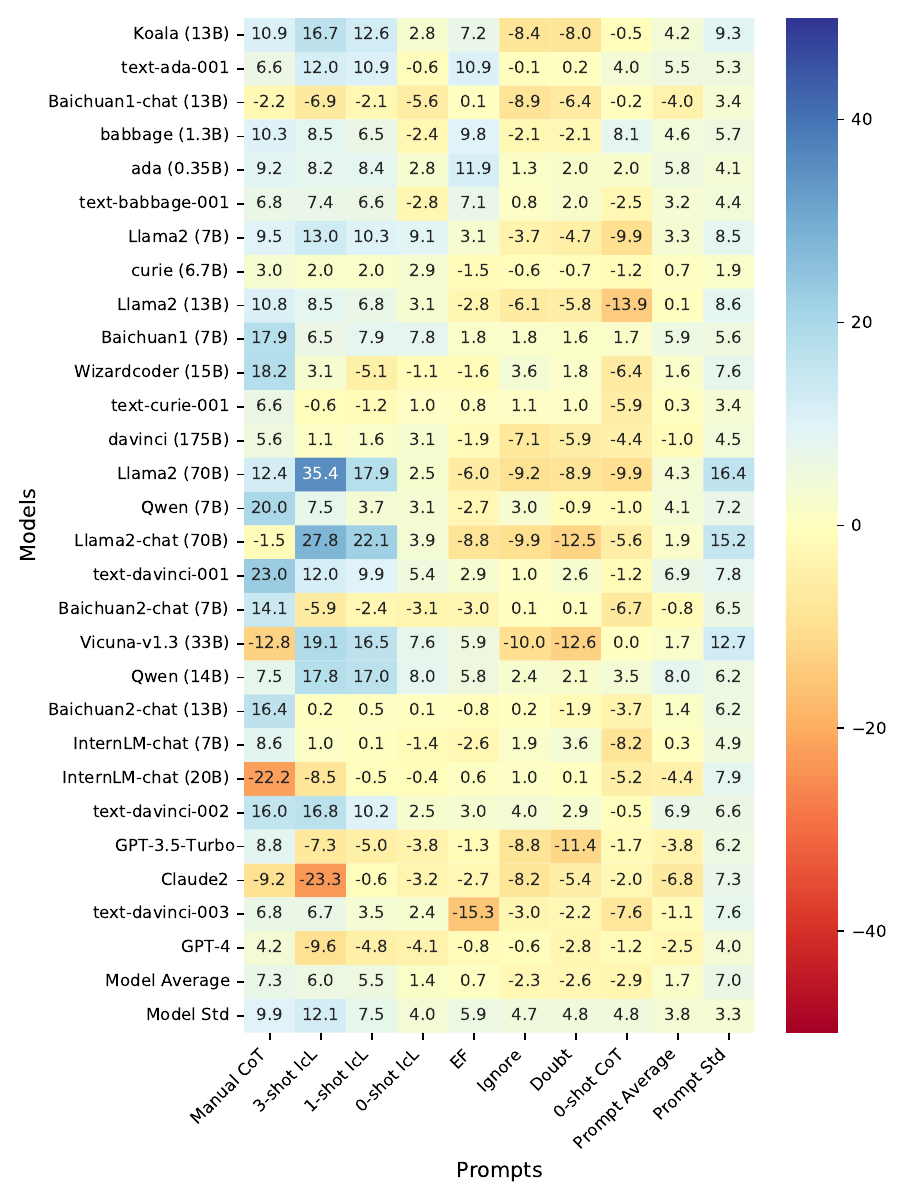}
\end{minipage}
}
\caption[Heatmap of CR]{\textbf{Heatmap of CR.} The models and prompts are sorted by their averages.}
\label{fig:Heatmap_of_Counterfactual_Reasoning}
\end{figure}

\begin{figure}
    \centering
    \includegraphics[width=0.8\linewidth]{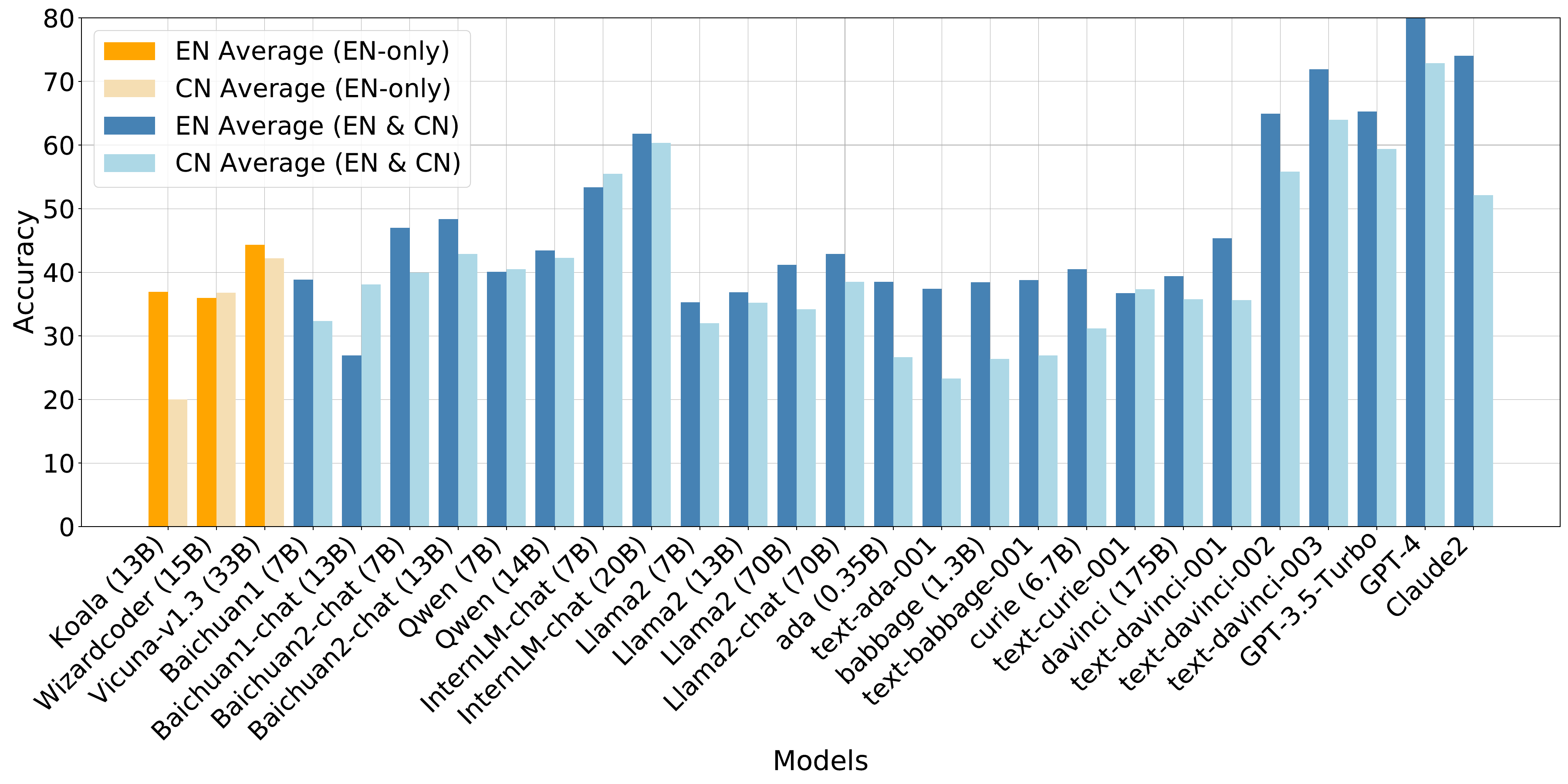}
    \caption[Language comparison of CR]{\textbf{Language comparison of CR.} The dark legend signifies the average performance of the model on an English test set, whereas the light legend denotes the average performance of the model on the Chinese test set. The yellow legend indicates a model trained exclusively on English datasets, while the blue legend represents a model trained on both English and Chinese datasets.}
    \label{fig:Counterfactual_Reasoning_Language}
\end{figure}

Initially, we analyze model performance in CR:

1) \textbf{Distribution}: Figure \ref{fig:Distribution_of_counterfactual}(a) outlines the distribution of all \textit{model-prompt pair}s within CR. With a median of 38.8\% and a third quartile of 54.3\%, this scenario appears to have an \textbf{easy} \textit{understandability}, as indicated by the median accuracy surpassing the average random guess benchmark of 37.5\%.
Figure \ref{fig:Distribution_of_Counterfactual_Reasoning_Tasks} details the distribution for each specific task.
In the \textbf{CR-C (CRASS)} task, the median stands at 29.7\%, and the third quartile is 57.5\%, with random guess accuracy set at 25.0\%, marking the \textit{understandability} of the task as \textbf{easy}.
For the \textbf{CR-B (det-counterfactual)} task, the median is calculated at 50.1\%, and the third quartile at 52.8\%, with a random guess accuracy of 50.0\%, affirming the \textit{understandability} of the task as \textbf{easy}.
\textbf{Upon examining the differences between tasks}, we observe a median accuracy range from 29.7\% to 50.1\% with a standard deviation of 10.2. The third quartile accuracy spans from 52.8\% to 57.5\% with a standard deviation of 2.4, signaling that the scenario has a \textbf{considerably varied} \textit{variance of distribution}. The peak of the distribution for both tasks aligns closely with their respective random guess accuracies. Notably, the CR-C (CRASS) task exhibits a wider distribution of model accuracy, ranging from 0\% to 100\%, suggesting a more decentralized distribution, whereas the CR-B (det-counterfactual) task shows a more concentrated set of outcomes.

2) \textbf{Top Accuracy}: Figure \ref{fig:Heatmap_of_Counterfactual_Reasoning}(a) reveals the three leading models in this scenario by average accuracy are GPT-4 at 76.9\%, text-davinci-003 at 67.8\%, and Claude2 at 62.5\%. The \textit{top model-prompt pair} is GPT-4 with manual CoT, achieving an 83.2\% accuracy. The scenario has a \textbf{solvable} \textit{solvability} with the top model's average accuracy surpassing 70\%. Figure \ref{fig:Heatmap_of_performances_of_Counterfactual_Reasoning} details the top models' average accuracy across various tasks.
In the \textbf{CR-C (CRASS)} task, GPT-4 leads with an average accuracy of 85.8\%, followed by text-davinci-003 at 77.8\%, and InternLM-chat (20B) at 73.5\%, with GPT-4 and adversarial ignore reaching the highest accuracy of 91.8\%. This demonstrates that the task \textit{solvability} is \textbf{well-solved}, as the top three models all achieve average accuracies over 70\%. 
For the \textbf{CR-B (det-counterfactual)} task, the top models are GPT-4 at 68.0\%, text-davinci-003 at 57.9\%, and Claude2 at 55.8\%, with GPT-4 and manual CoT leading to a 77.4\% accuracy, marking the \textit{solvability} of the task as \textbf{challenging} due to the \textit{top model-prompt pair} not reaching 80\%.
\textbf{Upon comparing the tasks}, there is a \textbf{large} \textit{variance of solvability}. The top model's average accuracy varies significantly, from 68.0\% to 85.8\% (a 17.8\% difference), and the highest accuracy achieved by \textit{top model-prompt pair}s ranges from 77.4\% to 91.8\% (a 14.4\% difference), highlighting a \textbf{extremely significant} \textit{variance of model's top performance}. GPT-4 is notably the best model in terms of average performance and the model that forms the \textit{top model-prompt pair} across tasks. Furthermore, text-davinci-003 is the second-best model in average performance for both tasks. The CR-C (CRASS) task outperforms the CR-B (det-counterfactual) in both \textit{top model-prompt pair} accuracy and top average model performance.

3) \textbf{Stability}: The three most consistent models in the scenario, characterized by the \textit{model volatility}, are curie (6.7B) at 1.8, text-curie-001 at 3.2, and Baichuan1-chat (13B) at 3.4. Conversely, the models displaying the greatest variability across various prompts, showcasing their great sensitivity to prompts, are Llama2 (70B) at 15.4, Llama2-chat (70B) at 14.2, and Vicuna-v1.3 (33B) at 11.9. Next, we consider the stability task-by-task. 
In the \textbf{CR-C (CRASS)} task, the most stable models are text-ada-001 at 1.7, text-babbage-001 at 2.1, and babbage (1.3B) at 3.2, while the models with the highest \textit{model volatility}, indicating the most variability, are Llama2 (70B) at 30.2, Llama2-chat (70B) at 25.1, and text-davinci-001 at 14.9.
In the \textbf{CR-B (det-counterfactual)} task, the most consistent models include Baichuan1 (7B) at 0.8, text-davinci-001 at 0.8, and davinci (175B) at 1.0. The models with the largest \textit{model volatility}, hence the most instability, are InternLM-chat (20B) at 19.2, Llama2-chat (70B) at 16.3, and Vicuna-v1.3 (33B) at 11.8.
\textbf{Upon comparing across tasks}, it is observed that while text-davinci-001 ranks as the second most stable model in the CR-B (det-counterfactual), it is the third least stable model in the CR-C (CRASS) task. Additionally, Llama2-chat (70B) consistently appears as the second most unstable model in both tasks.

4) \textbf{Open-Limited Ratio}: The ratio of open-access to limited-access models among the top five in the entire scenario stands at 0:5, indicating a \textbf{large} \textit{open-limited gap}.

Next, we analyze \textit{prompt gain} in CR:

1) \textbf{Top Gain}: As shown in Figure \ref{fig:Heatmap_of_Counterfactual_Reasoning}(b), the two prompts leading to the highest average accuracy improvements over the basic prompt are manual CoT at 7.3\% and 3-shot IcL at 6.0\%. The largest improvement relative to the basic prompt is achieved by Llama2 (70B) with 3-shot IcL, recording a 35.4\% increase. A more specific analysis is conducted for each task.
Figure \ref{fig:Heatmap_of_gain_of_Counterfactual_Reasoning} presents the gain heatmaps for all tasks within the scenario. In the \textbf{CR-C (CRASS)} task, manual CoT at 16.0\% and 3-shot IcL at 8.4\% top the charts for average accuracy gain over the basic prompt, with Llama2 (70B) and 3-shot IcL marking the most substantial increase at 68.7\%. In the \textbf{CR-B (det-counterfactual)} task, 1-shot IcL at 4.8\% and EF at 4.4\% provide the highest average accuracy gains, with Koala (13B) and EF achieving the most significant improvement at 23.3\%.
\textbf{Upon reviewing both tasks}, the leading accuracy gain prompt and the leading accuracy gain \textit{model-prompt pair} differ between the two tasks. The CR-C (CRASS) task exhibits a higher top gain compared to the CR-B (det-counterfactual) task.

2) \textbf{Exceptions}: Though manual CoT is the most effective prompt in giving positive average \textit{prompt gain} across most models in the scenario, it has exceptions in models including Baichuan1-chat (13B), Llama2-chat (70B), Vicuna-v1.3 (33B), InternLM-chat (20B), and Claude2. All prompts manage to enhance the performance of models such as ada (0.35B), Baichuan1 (7B), and Qwen (14B) beyond their performance under basic prompt. However, Claude2's performance is not increased by any prompt.
In the \textbf{CR-C (CRASS)} task, manual CoT does not work well with Koala (13B), text-babbage-001, Vicuna-v1.3 (33B), Claude2, and GPT-4 in generating positive average prompt gains. Despite this, all prompts are capable of boosting the performance of Baichuan1 (7B) and Qwen (14B) above their basic prompt performance. On the other hand, all prompts fail to improve the performance of text-babbage-001 and Claude2 above the basic prompt within this task.
For the \textbf{CR-B (det-counterfactual)} task, 1-shot IcL shows ineffectiveness with Baichuan1-chat (13B), Baichuan2-chat (13B), Llama2 (7B), Llama2 (13B), Baichuan2-chat (7B), text-davinci-001, GPT-3.5-Turbo, and Claude2. Nonetheless, all prompts success in producing positive average \textit{prompt gain} for Koala (13B) and ada (0.35B).

3) \textbf{Stability}: Regarding stability in the scenario, the most stable prompts are 0-shot IcL and adversarial ignore, with \textit{prompt volatility} of 4.0 and 4.7, respectively, indicating minimal variability. On the opposite end, 3-shot IcL and manual CoT are the most variable, with \textit{prompt volatility} of 12.1 and 9.9, highlighting their high sensitivity to prompt selection. The \textit{average model-prompt-gain volatility} (\textit{AMPGV}) is 7.0, suggesting a \textbf{medium} level of \textit{prompt dependence} across the scenario. Stability is further analyzed for each specific task.
In the \textbf{CR-C (CRASS)} task, the most stable prompts are 0-shot CoT and adversarial doubt, with \textit{prompt volatility} of 4.4 and 6.7. Conversely, 3-shot IcL and manual CoT show the most variability, with \textit{prompt volatility} of 20.8 and 17.8. The task's \textit{prompt dependence} is \textbf{high} as the \textit{AMPGV} for this task is 10.7.
For the \textbf{CR-B (det-counterfactual)} task, the most stable prompts are 0-shot IcL and adversarial doubt, with \textit{prompt volatility} of 4.3 and 5.6. The most variable prompts are manual CoT and 3-shot IcL, with \textit{prompt volatility} of 17.5 and 10. The task has a \textbf{medium} \textit{prompt dependence} with an \textit{AMPGV} of 6.4.
\textbf{Upon reviewing all tasks}, the \textit{AMPGV} ranges from 6.4 to 10.7, reflecting a \textbf{narrow} \textit{variance of prompt dependence}. Adversarial doubt ranks as one of the top two most stable prompts in both tasks, whereas 3-shot IcL and manual CoT are identified as the least stable prompts, consistent with the scenario-level findings. The CR-C (CRASS) task exhibits a higher dependency on prompts than the CR-B (det-counterfactual) task, based on their respective \textit{AMPGV}.
        
Finally, we measure \textit{language proficiency} in CR:

1) \textbf{English vs. Chinese}: Figure \ref{fig:Counterfactual_Reasoning_Language} reveals that models generally perform better in English than in Chinese, with 23 out of 28 models favoring English.

2) \textbf{Accuracy Difference}: Significant differences in performance between English and Chinese, with a preference for English, are noted in Claude2 (21.9\%), Koala (13B) (16.9\%), and text-ada-001 (14.1\%). In contrast, models like Baichuan1-chat (13B) (11.2\%), InternLM-chat (7B) (2.1\%), and Wizardcoder (15B) (0.8\%) demonstrate higher proficiency in Chinese.

\paragraph{Actual causality.}
\begin{figure}[t]
\centering
\subfigure[Model performance of AC]{
\begin{minipage}{8.5cm}
\centering
\includegraphics[width=1\linewidth]{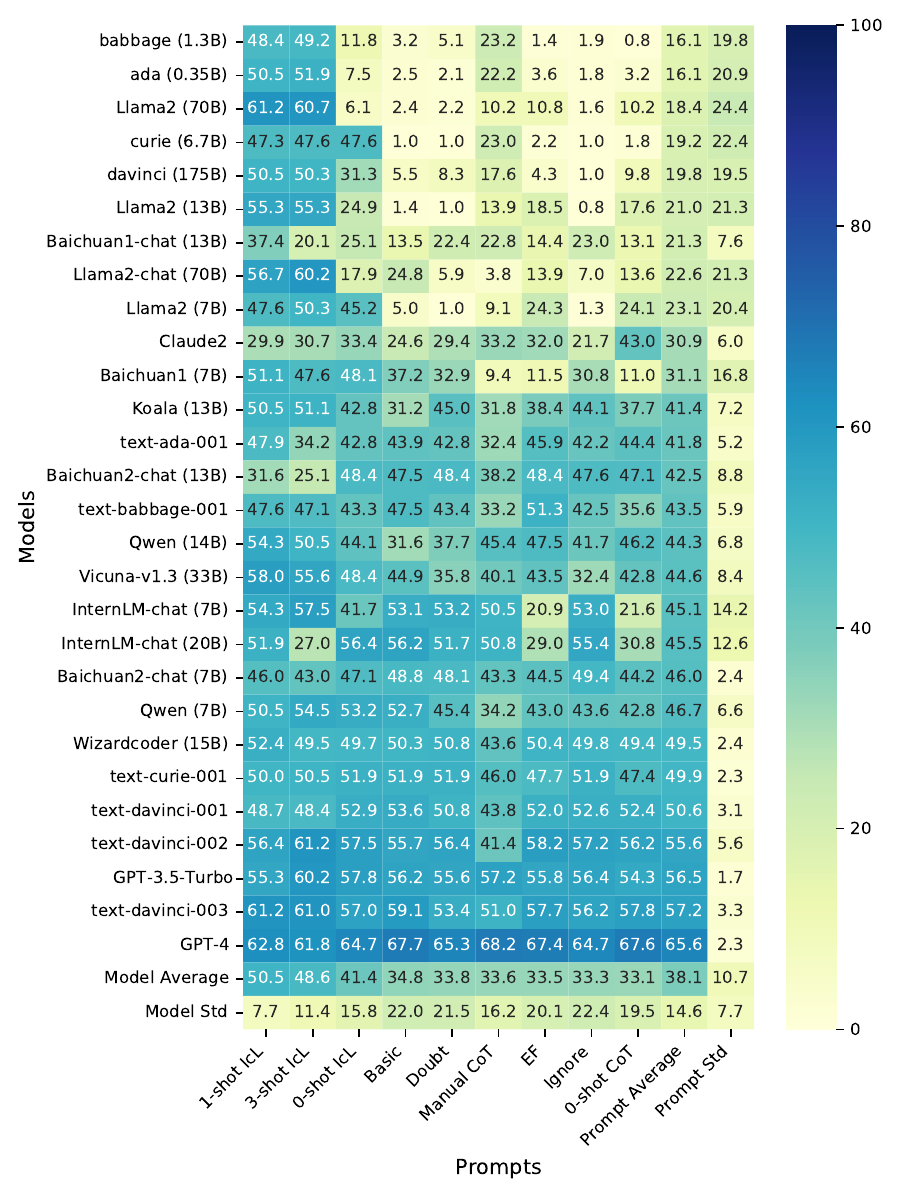}
\end{minipage}
}
\subfigure[\textit{Prompt gain} of AC]{
\begin{minipage}{8.5cm}
\centering
\includegraphics[width=1\linewidth]{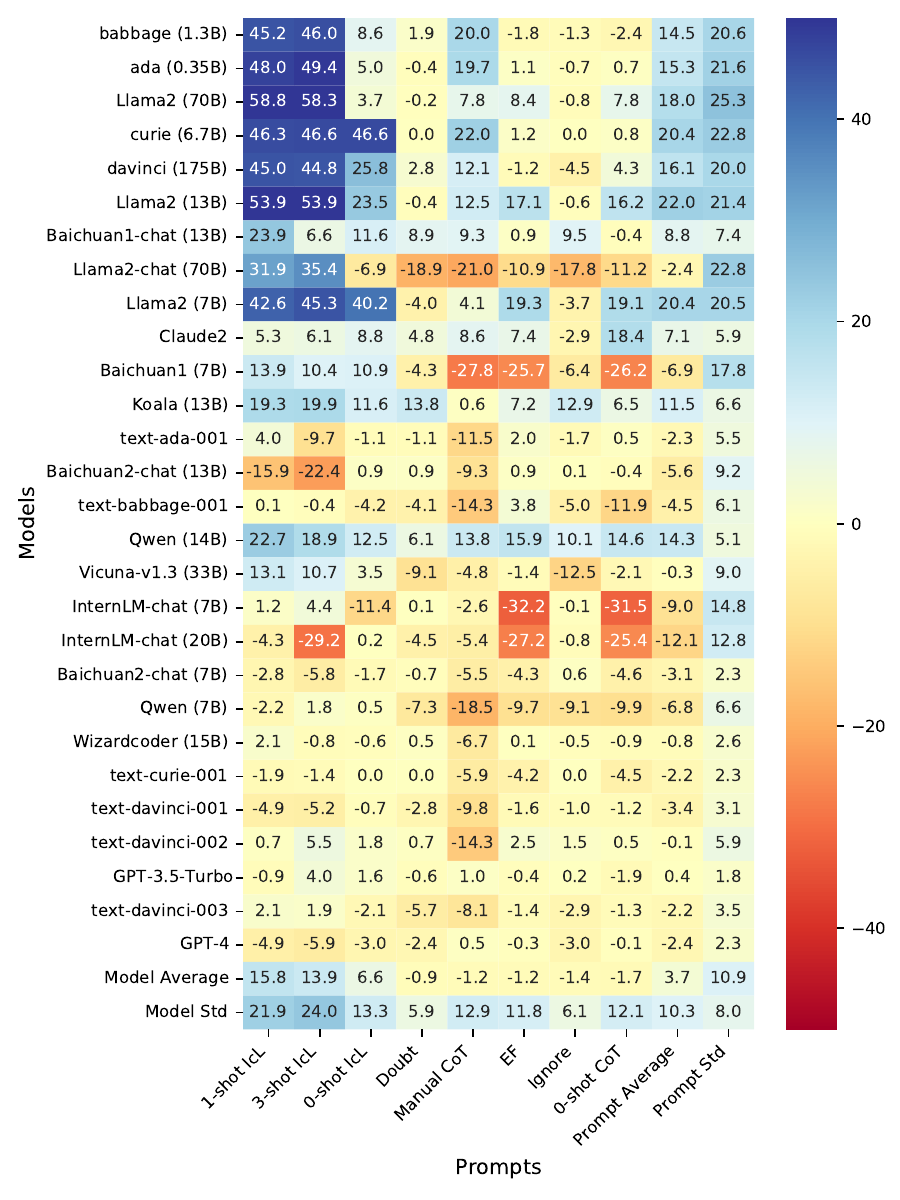}
\end{minipage}
}
\caption[Heatmap of AC]{\textbf{Heatmap of AC.} The models and prompts are sorted by their averages.}
\label{fig:Heatmap_of_Actually_Causality}
\end{figure}

\begin{figure}
    \centering
    \includegraphics[width=0.8\linewidth]{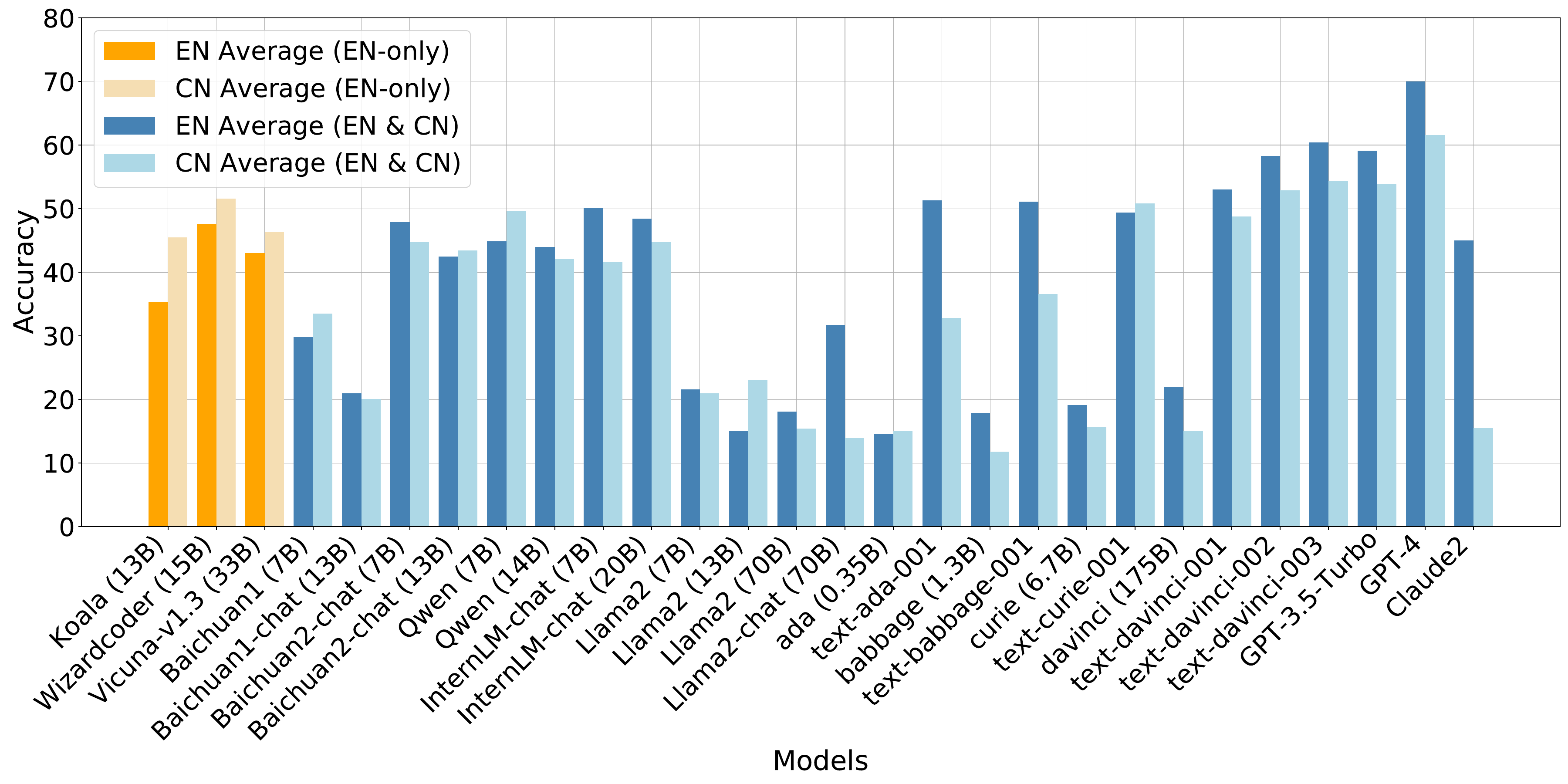}
    \caption[Language comparison of AC]{\textbf{Language comparison of AC.} The dark legend signifies the average performance of the model on an English test set, whereas the light legend denotes the average performance of the model on the Chinese test set. The yellow legend indicates a model trained exclusively on English datasets, while the blue legend represents a model trained on both English and Chinese datasets.}
    \label{fig:Actually_Causality_Language}
\end{figure}

From a model performance perspective: 
1) \textbf{Distribution}: Figure \ref{fig:Distribution_of_counterfactual}(b) shows the distribution of all \textit{model-prompt pair}s in AC. The median accuracy stands at 45.0\%, with the third quartile reaching 51.9\%, we consider the \textit{understandability} of the scenario to be \textbf{hard}, as the median accuracy falls below the 50.0\% threshold of random guess accuracy.
2) \textbf{Top Accuracy}: Figure \ref{fig:Heatmap_of_Actually_Causality}(a) shows the ranked model performances of AC. Notably, GPT-4 leads in average accuracy at 65.6\%, followed by text-davinci-003 and GPT-3.5-Turbo, with scores of 57.2\% and 56.5\%, respectively. GPT-4, when paired with manual CoT prompts, achieves a significant 68.2\% in accuracy, yet this top performance is still short of the 80 threshold, indicating the \textbf{challenging} \textit{solvability} of the AC scenario.
3) \textbf{Stability}: In the stability of model responses, Llama2 (70B), curie (6.7B), and Llama2-chat (70B) show the greatest variations in performance across different prompts, while GPT-3.5-Turbo, GPT-4, and text-curie-001 demonstrate remarkable consistency according to the \textit{model volatility} introduced in \cref{metric:model}. 
4) \textbf{Open-Limited Ratio}: The 0:5 ratio of open-access to limited-access models among the 5 top performers highlights a \textbf{large} \textit{open-limited gap}, pointing to the dominance of limited-access models in this domain.

Regarding \textit{prompt gain} in AC, 
1) \textbf{Top Gain}: As depicted in Figure \ref{fig:Heatmap_of_Actually_Causality}(b), 1-shot IcL and 3-shot IcL produce the highest average accuracy gains, at 15.8\% and 13.9\%, respectively. Remarkably, Llama2 (70B) using 1-shot IcL records the highest increase of 58.8\% in accuracy, showcasing the potential of specially designed prompts.
2) \textbf{Exceptions}: The effectiveness of the top prompt, 1-shot IcL, is not uniform across all models. It fails to produce a positive average \textit{prompt gain} in models including Baichuan2-chat (13B) and InternLM-chat (20B). Despite these exceptions, all prompts can enhance the performance of models like Koala (13B) and Qwen (14B) beyond the basic prompt setup. On the other hand, text-davinci-001 stands as an outlier, immune to the performance boosts of any prompts.
3) \textbf{Stability}: The stability of prompts is measured by their \textit{prompt volatility}. Adversarial doubt and adversarial ignore are the most stable prompts, with \textit{prompt volatility} of 5.9 and 6.1, respectively, suggesting their reliability across varying scenarios. Conversely, 3-shot IcL and 1-shot IcL exhibit the highest instability, highlighting a significant sensitivity to prompt design. From the \textit{average model-prompt-gain volatility} (\textit{AMPGV}) of 10.9, we regard the \textit{prompt dependence} of the scenario as \textbf{high}.

Regarding \textit{language proficiency} in AC, 
1) \textbf{English vs. Chinese}: as detailed in Figure \ref{fig:Actually_Causality_Language}, the majority of models (19 out of 28) exhibit superior performance in English over Chinese. 
2) \textbf{Accuracy Difference}: Models like Claude2, text-ada-001, and Llama2-chat (70B) demonstrate significant proficiency in English by accuracy differences between English and Chinese of 29.5\%, 18.5\%, and 17.7\%. Conversely, a subset of models, including Koala (13B), Llama2 (13B), and Qwen (7B), display a distinct advantage in Chinese, with accuracy performances outperforming their English counterparts by 10.2\%, 7.9\%, 4.7\%, respectively.

\paragraph{Effect of the treatment on the treated.}
\begin{figure}[t]
\centering
\subfigure[Model performance of ETT]{
\begin{minipage}{8.5cm}
\centering
\includegraphics[width=1\linewidth]{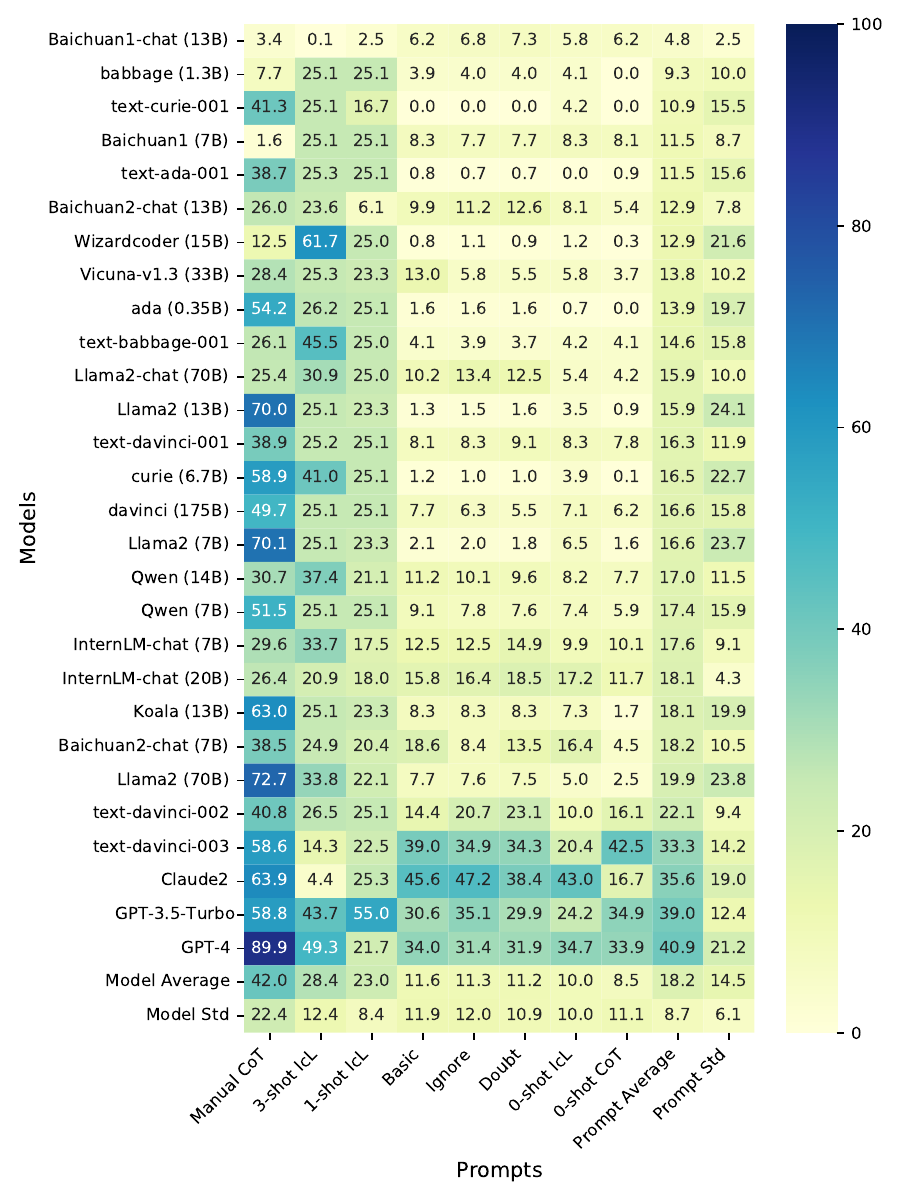}
\end{minipage}
}
\subfigure[\textit{Prompt gain} of ETT]{
\begin{minipage}{8.5cm}
\centering
\includegraphics[width=1\linewidth]{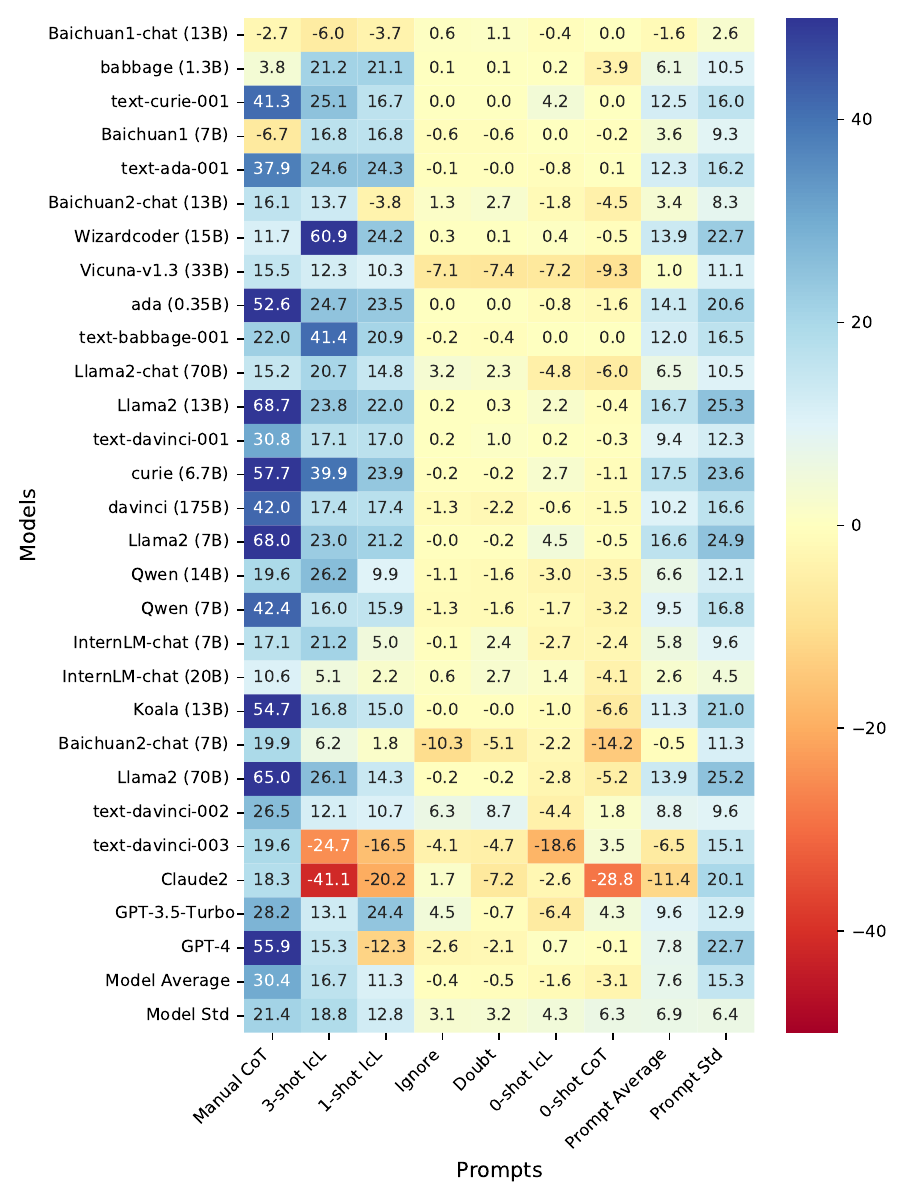}
\end{minipage}
}
\caption[Heatmap of ETT]{\textbf{Heatmap of ETT.} The models and prompts are sorted by their averages.}
\label{fig:Heatmap_of_Effect_of_the_Treatment_on_the_Treated}
\end{figure}

\begin{figure}
    \centering
    \includegraphics[width=0.8\linewidth]{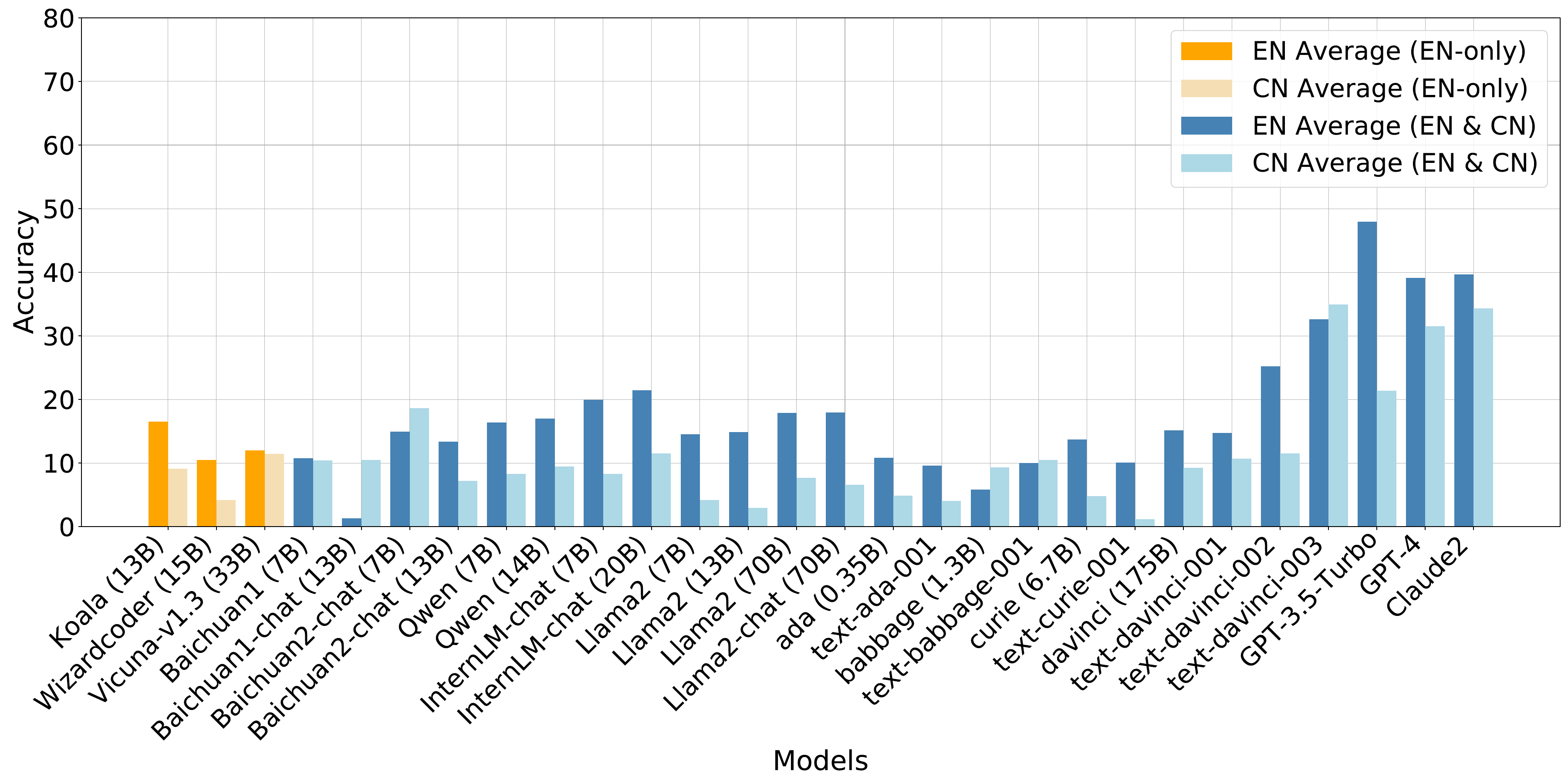}
    \caption[Language comparison of ETT]{\textbf{Language comparison of ETT.} The dark legend signifies the average performance of the model on an English test set, whereas the light legend denotes the average performance of the model on the Chinese test set. The yellow legend indicates a model trained exclusively on English datasets, while the blue legend represents a model trained on both English and Chinese datasets.}
    \label{fig:Effect_of_the_Treatment_on_the_Treated_Language}
\end{figure}
Initially, we evaluate model performance in ETT:

1) \textbf{Distribution}: According to Figure \ref{fig:Distribution_of_counterfactual}(c), the distribution for all \textit{model-prompt pair}s in the ETT reveals a median of 12.5\% and a third quartile of 25.2\%. This scenario is considered to have a \textbf{hard} \textit{understandability} as the median accuracy falls below the random guess benchmark of 16.7\%. Figure \ref{fig:Distribution_of_Effect_of_the_Treatment_on_the_Treated_Tasks} further details the distribution for each causal task respectively.
In the \textbf{ETT-P (ETT-basic)} task, the median is 1.2\%, and the third quartile is 6.1\%. Given the challenging nature of Mathematical-mode tasks, we define the \textit{understandability} of the task as \textbf{very hard}(Section \ref{metric:scenario}). 
Similarly, the \textbf{ETT-P (ETT-hard)} task, with a median of 1.4\% and a third quartile of 13.5\%, is considered to have a \textbf{very hard} \textit{understandability}.
On the other hand, the \textbf{ETT-B (ETT-natural)} task, presenting a median of 24.9\% and a third quartile of 57.1\% against a random guess accuracy of 50.0\%, is categorized to have a \textbf{hard} \textit{understandability}.
\textbf{Examining the variance among tasks} reveals median accuracies ranging from 1.2\% to 24.9\% with a standard deviation of 11.1, and third quartile accuracies extending from 6.1\% to 57.1\% with a standard deviation of 22.5. Consequently, the scenario has a \textbf{highly divergent} \textit{variance of distribution}. Organizing tasks by both median and third quartile accuracies results in the same order: ETT-P (ETT-basic) < ETT-P (ETT-hard) < ETT-B (ETT-natural). The distributions for both Mathematical-mode tasks (ETT-P (ETT-basic), ETT-P (ETT-hard)) indicate that over 70\% of \textit{model-prompt pair}s fall within a 0\% to 10\% accuracy range. In contrast, the Natural-mode task displays a more even distribution, where no 10-unit interval contains more than 25\% of the total \textit{model-prompt pair} count.

2) \textbf{Top Accuracy}: According to Figure \ref{fig:Heatmap_of_Effect_of_the_Treatment_on_the_Treated}(a), the leading three models in this scenario by average accuracy are GPT-4 at 40.9\%, GPT-3.5-Turbo at 39.0\%, and Claude2 at 35.6\%. GPT-4, when combined with manual CoT, reaches an impressive 89.9\%, suggesting this scenario's \textit{solvability} is \textbf{potentially solvable}, given that the \textit{top model-prompt pair} achieves over 80\% in performance. Figure \ref{fig:Heatmap_of_performances_of_Effect_of_the_Treatment_on_the_Treated} further delineates the top models' average accuracy across individual tasks.
For the \textbf{ETT-P (ETT-basic)} task, the highest average accuracies are seen with GPT-3.5-Turbo at 26.5\%, Claude2 at 24.9\%, and text-davinci-003 at 23.3\%. The \textit{top model-prompt pair}, GPT-4 with manual CoT, secures 86.3\%, marking the task \textit{solvability} as \textbf{potentially solvable} due to the \textit{top model-prompt pair} exceeding 80\% in performance. 
In the \textbf{ETT-P (ETT-hard)} task, top performers are GPT-3.5-Turbo at 40.0\%, GPT-4 at 39.4\%, and text-davinci-003 at 32.0\%, with GPT-4 and manual CoT reaching 89.0\%, again underlining the task's \textbf{potentially solvable} \textit{solvability} for similar reasons. 
The \textbf{ETT-B (ETT-natural)} task showcases GPT-4 leading at 61.8\%, followed by Claude2 at 54.5\%, and text-davinci-002 at 52.6\%, with GPT-4 and manual CoT achieving 94.5\%, reinforcing the task's \textbf{potentially solvable} \textit{solvability} as before.
\textbf{Comparing Across Tasks}, the \textit{variance of solvability} across tasks is \textbf{negligible}, with the top model's average accuracy ranging from 26.5\% to 61.8\%, a significant difference of 35.3\%, and the \textit{top model-prompt pair}'s peak accuracy varying from 86.3\% to 94.5\%, a difference of 8.2\%. This indicates a \textbf{significant} \textit{variance of model's top performance} across the scenario. The Mathematical-mode tasks (ETT-P (ETT-basic) and ETT-B (ETT-natural)) exhibit lower top accuracies than the Natural-mode task. Notably, GPT-3.5-Turbo performs best on average in Mathematical-mode tasks, whereas GPT-4 excels in the Natural-mode task. Moreover, GPT-4, especially when paired with manual CoT, stands out as the \textit{top model-prompt pair} across all evaluated tasks.

3) \textbf{Stability}: The three most consistent models in the scenario, marked by the lowest \textit{model volatility}, are Baichuan1-chat (13B) with a \textit{model volatility} of 2.5, InternLM-chat (20B) at 4.3, and Baichuan2-chat (13B) at 7.8. Conversely, the models showing the highest sensitivity to prompt variations, as evidenced by the highest \textit{model volatility}, are Llama2 (13B) at 24.1, Llama2 (70B) at 23.8, and Llama2 (7B) at 23.7, highlighting their considerable instability. We next conduct a detailed task-specific stability evaluation as follows:
In the \textbf{ETT-P (ETT-basic)} task, the most stable models are Baichuan1 (7B) leading with a \textit{model volatility} of 0.6, followed by text-davinci-002 at 2.2, and babbage (1.3B) at 2.3. The least stable models, with the largest \textit{model volatility}, include Llama2 (70B) at 26.4, GPT-4 at 25.9, and Llama2 (7B) at 20.6.
For the \textbf{ETT-P (ETT-hard)} task, the top three models in terms of stability are Baichuan1 (7B) with \textit{model volatility} of 0.4, Baichuan1-chat (13B) at 2.4, and babbage (1.3B) at 3.2. The models with the greatest instability are curie (6.7B) at 27.2, GPT-4 at 26.9, and Llama2 (7B) at 26.4.
In the \textbf{ETT-B (ETT-natural)} task, the models demonstrating the highest stability include Baichuan1-chat (13B) at 2.8, InternLM-chat (20B) at 10.1, and Baichuan2-chat (13B) at 10.2. The most unstable models are Llama2 (13B) at 34.4, curie (6.7B) at 34.3, and Llama2 (7B) at 33.3.
\textbf{Comparing across tasks}, Baichuan1 (7B) and babbage (1.3B) are shown to be the most stable models in the Mathematical-mode tasks, securing the top and third spots, respectively. Notably, GPT-4 ranks as one of the two most unstable models in these tasks, with Llama2 (7B) consistently listed among the three most unstable models across the evaluated scenarios.

4) \textbf{Open-Limited Ratio}: The 0:5 ratio of open-access to limited-access models among the top 5 models with the highest average accuracy in the entire scenario underscores a \textbf{large} \textit{open-limited gap}.

Next, we analyze \textit{prompt gain} in ETT:

1) \textbf{Top Gain}: As shown in Figure \ref{fig:Heatmap_of_Effect_of_the_Treatment_on_the_Treated}(b), the two prompts leading to the highest average accuracy improvements over the basic prompt are manual CoT with a gain of 30.4\% and 3-shot IcL at 16.7\%. Llama2 (13B) utilizing manual CoT marks the most substantial improvement, registering a 68.7\% increase over the basic prompt. A more granular, task-specific analysis is next, with Figure \ref{fig:Heatmap_of_gain_of_Effect_of_the_Treatment_on_the_Treated} illustrates the accuracy gains across tasks. 
For the \textbf{ETT-P (ETT-basic)} task, manual CoT at 31.9\% represents the top prompt in terms of average accuracy gains compared to the basic prompt; it is also the only prompt that has a positive average gain. Also, the \textit{model-prompt pair} of Llama2 (70B) and manual CoT achieves the largest leap of 79.6\%. 
In the \textbf{ETT-P (ETT-hard)} task, the highest gains are noted with manual CoT at 34.5\% and 3-shot IcL at 8.6\%, with Llama2 (7B) and manual CoT seeing a significant rise of 74.7\%. 
For the \textbf{ETT-B (ETT-natural)} task, the leading prompts are 3-shot IcL at 42.0\% and 1-shot IcL at 38.1\%, with text-curie-001 and 3-shot IcL witnessing the most noteworthy boost of 75.1\%.
\textbf{The evaluating of the tasks} indicates a preference for manual CoT in the Mathematical-mode tasks (ETT-P (ETT-basic), ETT-P (ETT-hard)), whereas the Natural-mode task shows a predilection for 3-shot IcL. The peak gains across these tasks are similar in value.

2) \textbf{Exceptions}: The scenario's highly effective prompt, manual CoT, does not align well with certain models for producing positive average prompt gain, specifically Baichuan1-chat (13B) and Baichuan1 (7B). Within the \textbf{ETT-P (ETT-basic)} task, manual CoT underperforms for text-davinci-002 and Baichuan1-chat (13B). In the \textbf{ETT-P (ETT-hard)} task, manual CoT also falls short with Baichuan1-chat (13B). On the other hand, all prompts enhance GPT-3.5-Turbo's performance beyond the basic prompt. For the \textbf{ETT-B (ETT-natural)} task, the leading prompt, 3-shot IcL, shows ineffectiveness with Baichuan1-chat (13B), text-davinci-003, and Claude2 in generating a positive average prompt gain. \textbf{Across all tasks}, text-davinci-003 and Claude2 consistently shows a negative impact to 3-shot IcL.

3) \textbf{Stability}: The most stable prompts in the scenario are adversarial ignore and adversarial doubt, with \textit{prompt volatility} of 3.1 and 3.2, respectively, highlighting a low variability in performance. On the contrary, manual CoT and 3-shot IcL, with \textit{prompt volatility} of 21.4 and 18.8, are revealed as the most variable. The \textit{prompt dependence} in the scenario is \textbf{high}, as the \textit{average model-prompt-gain volatility} (\textit{AMPGV}) is 15.3. In task-specific stability analysis, we find the following conclusions.
In the \textbf{ETT-P (ETT-basic)} task, the most stable prompts are adversarial ignore and adversarial doubt, with \textit{prompt volatility} of 2.2 and 2.8, contrasting with manual CoT and 3-shot IcL, the most variable at \textit{prompt volatility} of 21.5 and 14.6. The \textit{AMPGV} of the task is 12.9, showing that the task has a \textbf{high} \textit{prompt dependence}. 
For \textbf{ETT-P (ETT-hard)}, adversarial ignore and adversarial doubt maintain the lowest \textit{prompt volatility} at 1.8 and 2.8, whereas manual CoT and 3-shot IcL display the highest variability, with \textit{prompt volatility} of 24.7 and 22.8. The \textit{AMPGV} of 15.9 underscores a \textbf{high} \textit{prompt dependence}.
The \textbf{ETT-B (ETT-natural)} task shows EF and adversarial doubt as the most stable, with \textit{prompt volatility} of 5.6 and 6.7, and 3-shot IcL and 1-shot IcL as the most variable, with \textit{prompt volatility} of 31.9 and 29.7. The \textit{AMPGV} is 24.6, revealing a \textbf{high} \textit{prompt dependence}.
\textbf{Across the scenario}, the \textit{AMPGV} range widely from 12.9\% to 24.6\%, indicating a \textbf{wide} \textit{variance of prompt dependence}. Adversarial doubt ranks among the top three for stability across all tasks, while 3-shot IcL consistently appears among the three least stable prompts. The Mathematical-mode tasks exhibit less prompt dependency compared to the Natural-mode task, as suggested by their respective \textit{AMPGV}, pointing towards variability in how different tasks respond to prompt strategies.

At last, we measure \textit{language proficiency} in ETT, 

1) \textbf{English vs. Chinese}: As shown in Figure \ref{fig:Effect_of_the_Treatment_on_the_Treated_Language}, the performance of models on the English test set is better than the one on the Chinese test set, with 23 out of 28 models performing better in English than in Chinese. 

2) \textbf{Accuracy Difference}: The most significant accuracy differences in performance between English and Chinese, favoring English, are seen in GPT-3.5-Turbo (26.6\%), text-davinci-002 (13.7\%), and Llama2 (13B) (11.9\%). On the flip side, models such as Baichuan1-chat (13B) (9.2\%), Baichuan2-chat (7B) (3.7\%), and babbage (1.3B) (3.4\%) show the top preference in Chinese compared to English.

\paragraph{Natural direct effect.}
\begin{figure}[t]
\centering
\subfigure[Model performance of NDE]{
\begin{minipage}{8.5cm}
\centering
\includegraphics[width=1\linewidth]{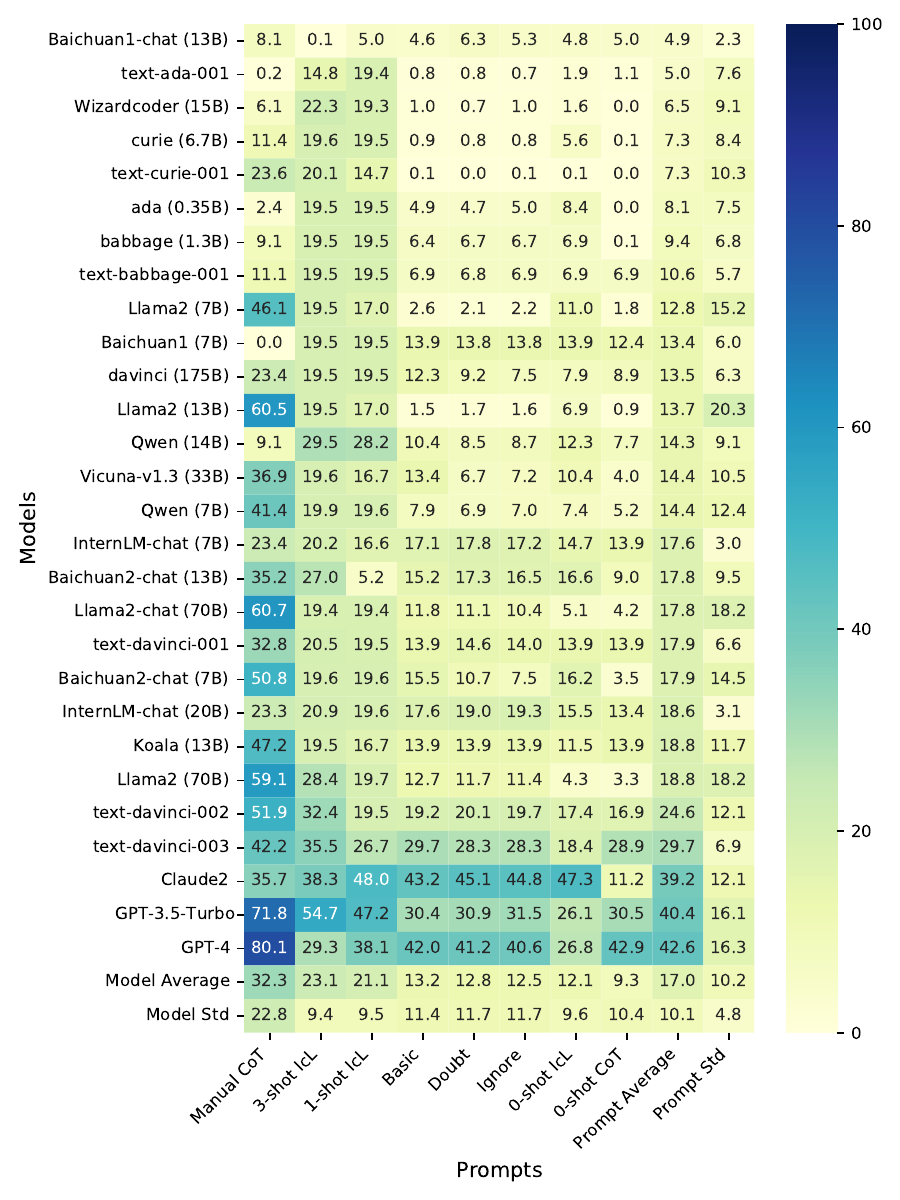}
\end{minipage}
}
\subfigure[\textit{Prompt gain} of NDE]{
\begin{minipage}{8.5cm}
\centering
\includegraphics[width=1\linewidth]{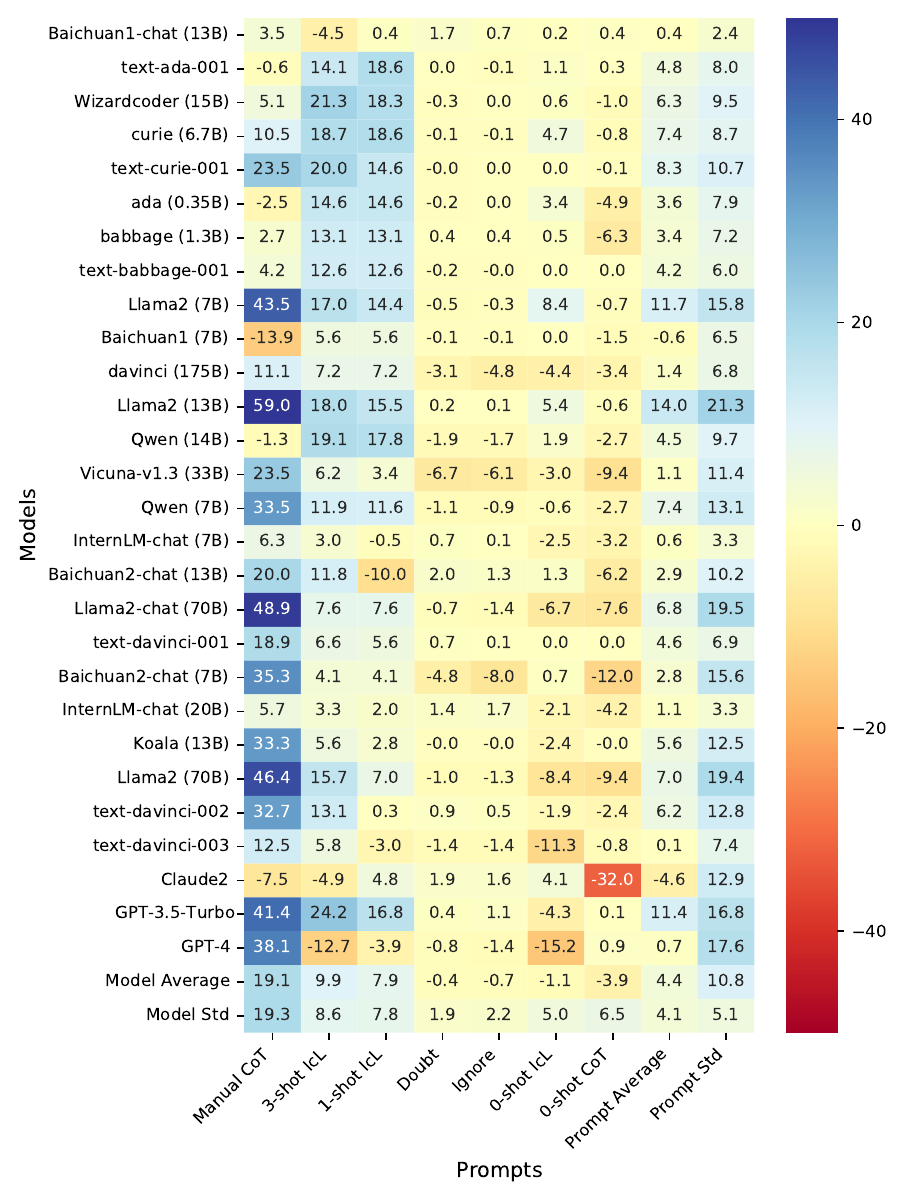}
\end{minipage}
}
\caption[Heatmap of NDE]{\textbf{Heatmap of NDE.} The models and prompts are sorted by their averages.}
\label{fig:Heatmap_of_Natural_Direct_Effect}
\end{figure}

\begin{figure}
    \centering
    \includegraphics[width=0.8\linewidth]{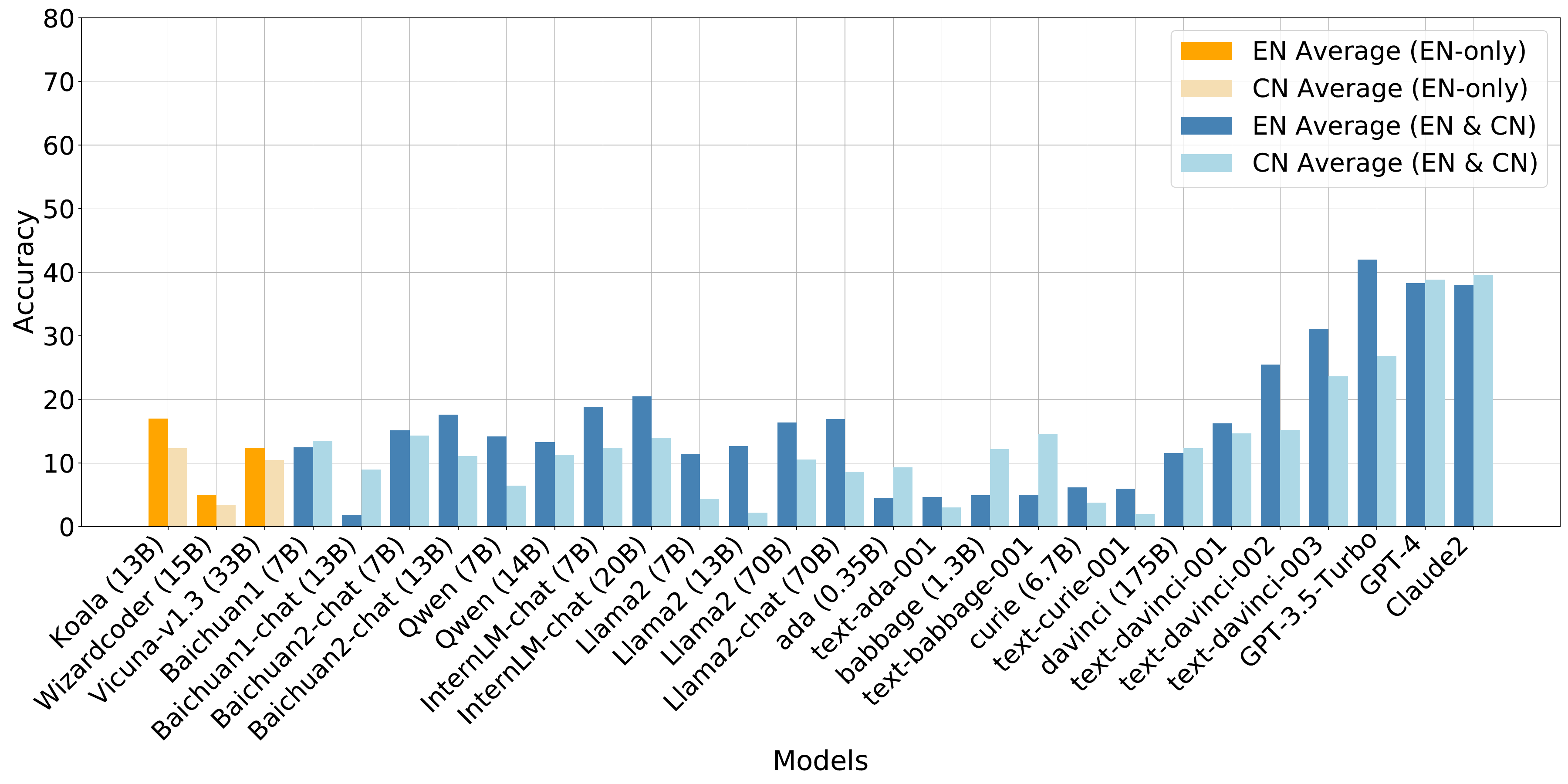}
    \caption[Language comparison of NDE]{\textbf{Language comparison of NDE.} The dark legend signifies the average performance of the model on an English test set, whereas the light legend denotes the average performance of the model on the Chinese test set. The yellow legend indicates a model trained exclusively on English datasets, while the blue legend represents a model trained on both English and Chinese datasets.}
    \label{fig:Natural_Direct_Effect_Language}
\end{figure}

Initially, we evaluate model performance of NDE:

1) \textbf{Distribution}: The distribution of the \textit{model-prompt pair}s within the NDE, as shown in Figure \ref{fig:Distribution_of_counterfactual}(d), highlights the scenario's \textit{understandability}. With a median value of 14.0\% and a third quartile of 19.9\%, the scenario's \textit{understandability} is regarded as \textbf{hard} since the median accuracy falls below the expected random guess accuracy of 16.7\%.
Figure \ref{fig:Distribution_of_causal_tasks_in_NDE} shows the distribution across various causal tasks of all \textit{model-prompt pair}s in the task.
In the \textbf{NDE-P (NDE-basic)} task, the median is calculated at 0.6\%, and the third quartile is 4.6\%. Given the challenging nature of Mathematical-mode tasks, we define the \textit{understandability} of this task as \textbf{very hard}(Section \ref{metric:scenario}). 
For the \textbf{NDE-P (NDE-hard)} task, the median settles at 0.7\%, with a third quartile of 4.8\%. Similar to NDE-P (NDE-basic), we regard the \textit{understandability} of the task as \textbf{very hard}. 
The \textbf{NDE-B (NDE-natural)} task presents a median of 36.9\% and a third quartile of 56.0\%, with a random guess accuracy of 50.0\%, its \textit{understandability} is \textbf{hard}.
\textbf{Upon comparing task variances}, it is observed that median accuracies span from 0.6\% to 36.9\% with a standard deviation of 17.1, and third quartile accuracies range from 4.6\% to 56.0\% with a standard deviation of 24.2. This variation underscores the \textbf{highly divergent} \textit{variance of distribution} across tasks in the scenario. The ordering of tasks by both median and third quartile accuracies remains consistent: NDE-P (NDE-basic) < NDE-P (NDE-hard) < NDE-B (NDE-natural). Notably, the Mathematical-mode task (NDE-P (NDE-basic), NDE-P (NDE-hard)) distributions indicate an approximation where nearly 80\% of \textit{model-prompt pair}s in an accuracy between 0\% to 10\%. On the other hand, the distribution for the Natural-mode task is more balanced, with no 10-unit range encompassing more than 25\% of the total \textit{model-prompt pair}s.

2) \textbf{Top Accuracy}: From Figure \ref{fig:Heatmap_of_Natural_Direct_Effect}(a), we observe that the three highest-performing models in terms of average accuracy for this scenario are GPT-4 at 42.6\%, GPT-3.5-Turbo at 40.4\%, and Claude2 at 39.2\%. The \textit{top model-prompt pair} is GPT-4 with manual CoT, reaching an accuracy of 80.1\%, indicating that the \textit{solvability} of this scenario is \textbf{potentially solvable} as the \textit{top model-prompt pair}'s performance hits 80\%. Figure \ref{fig:Heatmap_of_performances_of_Natural_Direct_Effect} displays the top three models by average accuracy for each task.
In the \textbf{NDE-P (NDE-basic)} task, the top models by average accuracy are GPT-4 at 27.3\%, Claude2 at 27.1\%, and GPT-3.5-Turbo at 27.1\%, with GPT-4 using manual CoT achieving 70.2\%. This indicates the task \textit{solvability} is \textbf{challenging} as the \textit{top model-prompt pair}'s performance is larger than the random guess but less than 80\%. For \textbf{NDE-P (NDE-hard)}, the leading models in average accuracy are GPT-3.5-Turbo at 32.4\%, GPT-4 at 30.1\%, and Claude2 at 28.9\%, with GPT-4 using manual CoT reaching 76.1\%, marking the \textit{solvability} of the task as \textbf{challenging} since the \textit{top model-prompt pair}'s performance is larger than random guess but less than 80\%. In the \textbf{NDE-B (NDE-natural)} task, the top models by average accuracy are GPT-4 at 62.6\%, Claude2 at 58.0\%, and GPT-3.5-Turbo at 57.9\%, with GPT-4 using manual CoT attaining 93.9\%. This shows the task has a \textbf{potentially solvable} \textit{solvability} as the \textit{top model-prompt pair}'s performance reaches 80\% but the top model's average accuracy is below 70\%.
\textbf{Through comparing different tasks}, the \textit{variance of solvability} between tasks is \textbf{small}. Additionally, the top model's average accuracy ranges from 27.3\% to 62.6\% (a difference of 35.3\%), and the \textit{top model-prompt pair}'s accuracy varies from 70.2\% to 93.9\% (a difference of 23.7\%), demonstrating the scenario's \textit{variance of model's top performance} is \textbf{extremely significant}. The Mathematical-mode tasks show comparatively lower top accuracies than the Natural-mode task. Across all tasks, GPT-4, GPT-3.5-Turbo, and Claude2 consistently rank within the top three for average model performance. Furthermore, GPT-4 with manual CoT stands out as the \textit{top model-prompt pair} across all tasks.

3) \textbf{Stability}: The three most stable models, characterized by the lowest \textit{model volatility}, are Baichuan1-chat (13B) with a \textit{model volatility} of 2.3, InternLM-chat (7B) at 3.0, and InternLM-chat (20B) at 3.1. Conversely, the three least stable models, exhibiting the highest \textit{model volatility} across different prompts, are Llama2 (13B) at 20.3, Llama2-chat (70B) at 18.2, and Llama2 (70B) also at 18.2, showcasing their significant prompt sensitivity. We move on to analyze stability from a \textit{model-prompt-task} perspective:
For the \textbf{NDE-P (NDE-basic)} task, the most stable models are davinci (175B) at 0.1, babbage (1.3B) also at 0.1, and text-ada-001 at 0.2. In contrast, the least stable models include GPT-3.5-Turbo at 20.7, Llama2 (70B) at 18.0, and Llama2-chat (70B) also at 18.0.
For the \textbf{NDE-P (NDE-hard)} task, the models demonstrating the greatest stability, with minimal \textit{model volatility}, are babbage (1.3B), davinci (175B), and text-babbage-001 all at 0.0. On the flip side, the models showing the least stability are GPT-4 at 21.4, Llama2 (13B) at 19.9, and GPT-3.5-Turbo at 18.0.
For the \textbf{NDE-B (NDE-natural)} task, the top three stable models are Baichuan1-chat (13B) at 4.0, InternLM-chat (20B) at 9.9, and text-davinci-002 at 10.4. Conversely, the least stable models are Llama2 (13B) at 29.0, text-curie-001 at 25.9, and Llama2-chat (70B) at 25.3.
\textbf{When evaluating the tasks}, it is observed that while GPT-3.5-Turbo and GPT-4 are among the top performers in certain tasks, they also exhibit significant instability in others.

4) \textbf{Open-Limited Ratio}: With a 0:5 ratio of open-access to limited-access models among the top five models in the overall scenario, the \textit{open-limited gap} is \textbf{large}.

Next, we analyze \textit{prompt gain} in NDE:

1) \textbf{Top Gain}: Illustrated in Figure \ref{fig:Heatmap_of_Natural_Direct_Effect}(b), the leading two prompts achieving the most significant average accuracy improvements over the basic prompt are manual CoT at 19.1\% and 3-shot IcL at 9.9\%. The highest gain in accuracy over the basic prompt is seen with Llama2 (13B) employing manual CoT, with an increase of 59.0\%. We proceed to a task-specific detailed examination. Figure \ref{fig:Heatmap_of_gain_of_Natural_Direct_Effect} displays the gains across all tasks from a \textit{model-prompt-task} perspective. 
For the \textbf{NDE-P (NDE-basic)} task, the two prompts yielding the most considerable average accuracy gains over the basic prompt are manual CoT at 21.7\% and 3-shot IcL at 4.6\%. The largest improvement in accuracy over the basic prompt is with Llama2 (70B) using manual CoT, indicating a rise of 54.3\%. 
In the \textbf{NDE-P (NDE-hard)} task, the two prompts with the highest average accuracy gains over the basic prompt are manual CoT at 19.0\% and 3-shot IcL at 1.3\%. The most significant increase in accuracy over the basic prompt is with Llama2 (13B) using manual CoT, demonstrating a gain of 56.1\%. 
Within the \textbf{NDE-B (NDE-natural)} task, the two prompts leading to the highest average accuracy gains over the basic prompt are 3-shot IcL at 23.9\% and 1-shot IcL at 22.3\%. The most substantial enhancement in accuracy over the basic prompt is seen with Llama2 (13B) utilizing manual CoT, with an uplift of 71.0\%. - \textbf{The task evaluation} reveals a preference towards manual CoT in Mathematical-mode tasks, while the Natural-mode task prefers 3-shot IcL. Furthermore, across all tasks, manual CoT consistently secures the highest gains in comparison to other \textit{model-prompt pair}s, with the Natural-mode task achieving greater top gains than the Mathematical-mode tasks.

2) \textbf{Exceptions}: The most effective prompt, manual CoT, cannot create positive average \textit{prompt gain} for several models, including text-ada-001, ada (0.35B), Baichuan1 (7B), Qwen (14B), Claude2. In the \textbf{NDE-P (NDE-basic)} task, the best prompt, manual CoT, fails to perform a positive average \textit{prompt gain} on Claude2. In the \textbf{NDE-P (NDE-hard)} task, manual CoT does not prove effective for davinci (175B), Baichuan1 (7B), Claude2, and no prompt enhances davinci (175B)'s performance over the basic prompt in this task. For the \textbf{NDE-B (NDE-natural)} task, the top prompt, 3-shot IcL, is ineffective on Baichuan1-chat (13B), Claude2, GPT-4.
\textbf{Notably}, in the two Mathematical-mode tasks (NDE-P (NDE-basic) and NDE-P (NDE-hard)), manual CoT does not work well with Claude2.

3) \textbf{Stability}: Regarding stability within the scenario, the two most stable prompts are adversarial doubt and adversarial ignore, with \textit{prompt volatility} of 1.9 and 2.2, respectively. On the opposite end, the two prompts exhibiting the greatest instability, as indicated by the largest \textit{prompt volatility}, are manual CoT at 19.3 and 3-shot IcL at 8.6. This results in an \textit{average model-prompt-gain volatility} (\textit{AMPGV}) of 10.8, demonstrating a \textbf{high} \textit{prompt dependence}. We further analyze stability from a \textit{model-prompt-task} perspective.
For the \textbf{NDE-P (NDE-basic)} task, adversarial doubt and adversarial ignore are recognized as the most stable prompts, with \textit{prompt volatility} of 0.8 and 1.0, respectively. Conversely, manual CoT and 3-shot IcL show the most instability, with \textit{prompt volatility} of 20.4 and 8.9, respectively. The \textit{AMPGV} for this task is 8.8, indicating a \textbf{medium} level of \textit{prompt dependence}.
In the \textbf{NDE-P (NDE-hard)} task, the most stable prompts are again adversarial doubt and adversarial ignore, with \textit{prompt volatility} of 1.0 and 1.3, respectively. The most unstable prompts are manual CoT and 3-shot IcL, with \textit{prompt volatility} of 20.1 and 8.2, respectively. This task has an \textit{AMPGV} of 8.5, suggesting a \textbf{medium} \textit{prompt dependence}.
For the \textbf{NDE-B (NDE-natural)} task, the most stable prompts with the smallest \textit{prompt volatility} are adversarial doubt and adversarial ignore, at 5.6 and 7.0, respectively. The most variable prompts, indicating instability, are manual CoT and 3-shot IcL, with \textit{prompt volatility} of 23.2 and 22.0, respectively. The \textit{AMPGV} in this task is 19.7, pointing to a \textbf{high} \textit{prompt dependence}.
\textbf{Upon reviewing all tasks}, the \textit{AMPGV} range from 8.5 to 19.7, reflecting the \textbf{wide} \textit{variance of prompt dependence}.
\textbf{Across all tasks}, adversarial doubt and adversarial ignore consistently rank as the most stable prompts, whereas manual CoT and 3-shot IcL are identified as the most unstable. The Mathematical-mode tasks show a lower prompt dependency than the Natural-mode task, as indicated by their respective \textit{AMPGV}, highlighting differences in task responsiveness to prompt strategies.

Finally, we consider \textit{language proficiency} in NDE,

1) \textbf{English vs. Chinese}: Figure \ref{fig:Natural_Direct_Effect_Language} illustrates that models tend to perform better on the English test set compared to the Chinese test set, with 20 out of 28 models showing superior performance in English.

2) \textbf{Accuracy Difference}: The most significant discrepancies in performance between English and Chinese, with a preference for English, are noted in GPT-3.5-Turbo (15.2\%), Llama2 (13B) (10.5\%), and text-davinci-002 (10.3\%). In contrast, models like text-babbage-001 (9.6\%), babbage (1.3B) (7.3\%), and Baichuan1-chat (13B) (7.1\%) demonstrate greater proficiency in Chinese than in English.

\paragraph{Natural indirect effect.}
\begin{figure}[t]
\centering
\subfigure[Model performance of NIE]{
\begin{minipage}{8.5cm}
\centering
\includegraphics[width=1\linewidth]{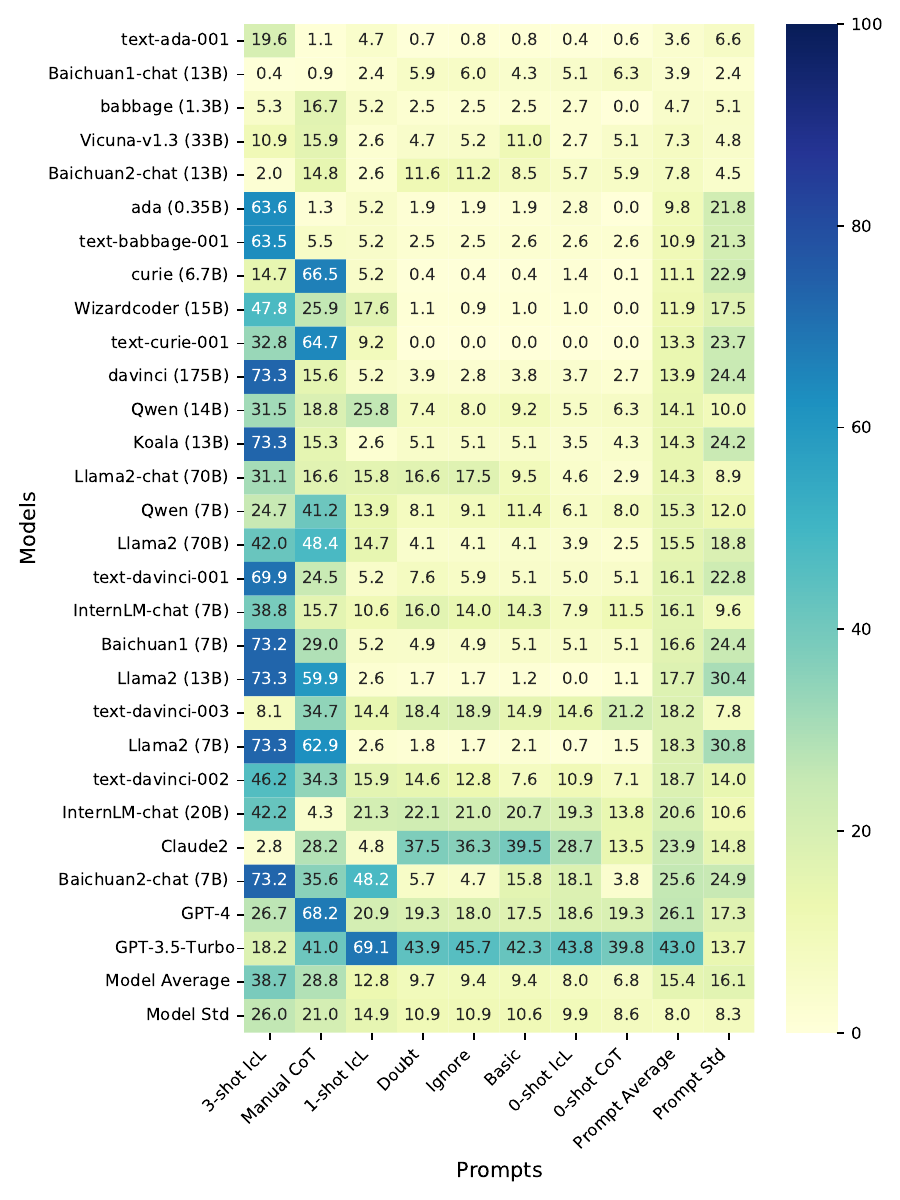}
\end{minipage}
}
\subfigure[\textit{Prompt gain} of NIE]{
\begin{minipage}{8.5cm}
\centering
\includegraphics[width=1\linewidth]{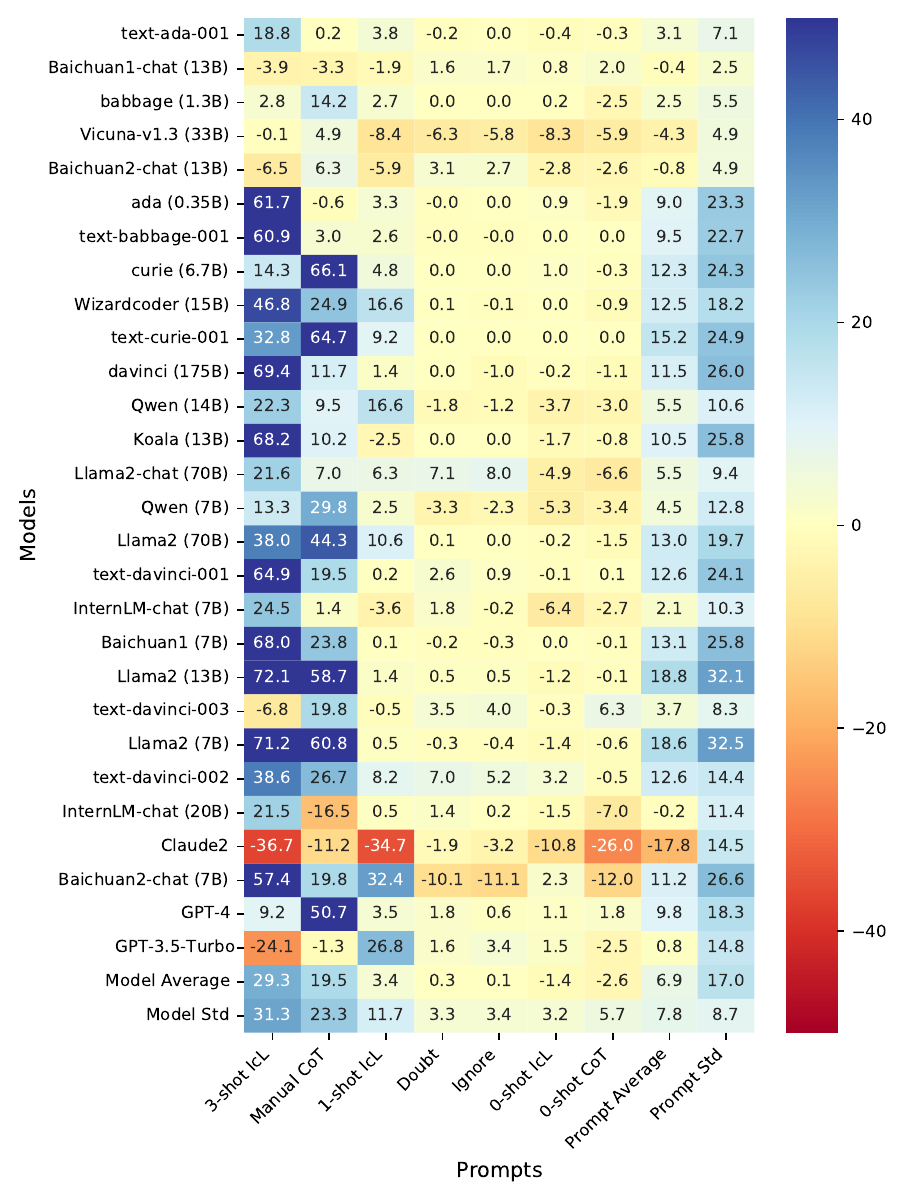}
\end{minipage}
}
\caption[Heatmap of NIE]{\textbf{Heatmap of NIE.} The models and prompts are sorted by their averages.}
\label{fig:Heatmap_of_Natural_Indirect_Effect}
\end{figure}

\begin{figure}
    \centering
    \includegraphics[width=0.8\linewidth]{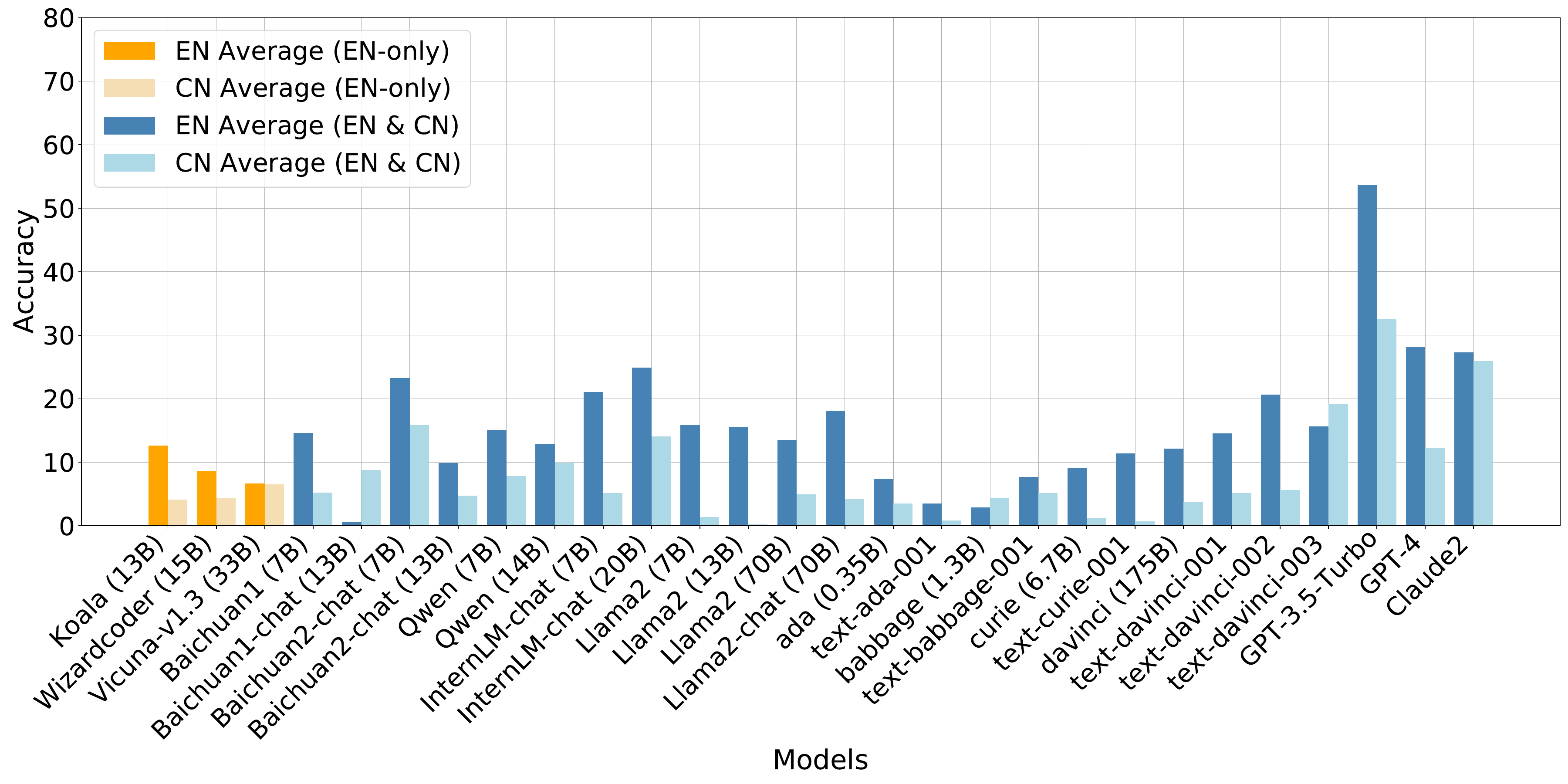}
    \caption[Language comparison of NIE]{\textbf{Language comparison of NIE.} The dark legend signifies the average performance of the model on an English test set, whereas the light legend denotes the average performance of the model on the Chinese test set. The yellow legend indicates a model trained exclusively on English datasets, while the blue legend represents a model trained on both English and Chinese datasets.}
    \label{fig:Natural_Indirect_Effect_Language}
\end{figure}
First, we consider model performance in NIE:

1) \textbf{Distribution}: Figure \ref{fig:Distribution_of_counterfactual}(e) showcases the distribution for all \textit{model-prompt pair}s under the NIE scenario. With a median value of 6.7\% and a third quartile of 19.0\%, this scenario is considered to have a \textbf{hard} \textit{understandability} as the median accuracy falls below the random guess benchmark of 16.7\%.
Figure \ref{fig:Distribution_of_Natural_Indirect_Effect_Tasks} shows the distribution for all \textit{model-prompt pair}s in each individual causal task.
In the \textbf{NIE-P (NIE-basic)} task, the median is calculated to be 2.0\%, and the third quartile is 11.2\%. Considering both the difficulty for most Mathematical-mode tasks and the lower than 15\% third quartile accuracy, we define the \textit{understandability} of this task as \textbf{very hard}.
In the \textbf{NIE-P (NIE-hard)} task, the median stands at 1.8\%, and the third quartile is 12.1\%. Similarly, we regard the task \textit{understandability} as \textbf{very hard}.
In the \textbf{NIE-B (NIE-natural)} task, the median reaches 15.4\%, with the third quartile at 31.5\%, and the random guess accuracy at 50.0\%, we assess the \textit{understandability} of the task as \textbf{very hard}.
\textbf{Upon examining task variances}, the median accuracies span from 1.8\% to 15.4\% with a standard deviation of 6.4, and the third quartile accuracies range from 11.2\% to 31.5\% with a standard deviation of 9.4, highlighting the scenario having a \textbf{moderately distinct} \textit{variance of distribution}. The Natural-mode task significantly outperforms the two mathematics-focused tasks (NIE-P (NIE-hard) and NIE-P (NIE-basic)) in terms of both median and third quartile values. For these Mathematical-mode tasks, over 70\% of \textit{model-prompt pair}s are found within a 0\% to 10\% accuracy bracket. Conversely, the Natural-mode task presents a more balanced distribution, with no 10\%-unit segment housing more than 40\% of the total \textit{model-prompt pair}s.

2) \textbf{Top Accuracy}: Observing Figure \ref{fig:Heatmap_of_Natural_Indirect_Effect}(a), the leading three models in terms of average accuracy in this scenario are GPT-3.5-Turbo at 43.0\%, GPT-4 at 26.1\%, and Baichuan2-chat (7B) at 25.6\%. The \textit{top model-prompt pair} is Koala (13B) with 3-shot IcL, achieving a 73.3\% accuracy, suggesting the \textit{solvability} of this scenario is \textbf{challenging} as the performance of the \textit{top model-prompt pair} surpasses the random guess but remains below 80\%. Figure \ref{fig:Heatmap_of_performances_of_Natural_Indirect_Effect} details the top three models' average accuracy for each specific task.
For the \textbf{NIE-P (NIE-basic)} task, the highest average accuracies are seen with GPT-3.5-Turbo at 40.8\%, Claude2 at 21.2\%, and GPT-4 at 18.7\%. Koala (13B) using 3-shot IcL tops this task with a 69.1\% accuracy, marking its \textit{solvability} as \textbf{challenging} since the \textit{top model-prompt pair} outperforms the random guess but does not reach 80\%. In the \textbf{NIE-P (NIE-hard)} task, the top performers in average accuracy are GPT-3.5-Turbo at 43.1\%, GPT-4 at 26.5\%, and Baichuan2-chat (7B) at 17.2\%, with Koala (13B) and 3-shot IcL leading at 66.3\%, again indicating the task \textit{solvability} is \textbf{challenging} as the \textit{top model-prompt pair}'s performance exceeds the random guess yet falls short of 80\%. For the \textbf{NIE-B (NIE-natural)} task, the models leading in average accuracy are GPT-3.5-Turbo at 44.8\%, Baichuan2-chat (7B) at 40.5\%, and text-davinci-002 at 38.4\%, with davinci (175B) using 3-shot IcL achieving the highest at 84.6\%, showing this task having a \textbf{potentially solvable} \textit{solvability} since the \textit{top model-prompt pair} attains 80\%, but the top model's average accuracy is below 70\%.
\textbf{By analyzing the differences across tasks}, we find that the \textit{variance of solvability} among tasks is \textbf{small}. Additionally, the leading model's average accuracy spans from 40.8\% to 44.8\% (a 4.0\% difference), and the accuracy by the \textit{top model-prompt pair} ranges from 66.3\% to 84.6\% (an 18.3\% difference), pointing to a \textbf{considerable} \textit{variance of model's top performance}. The Mathematical-mode tasks generally have lower top accuracies than the Natural-mode tasks. GPT-4 consistently ranks high in average performance across tasks. 3-shot IcL significantly boosts model performance, consisting of the \textit{top model-prompt pair} in all tasks.

3) \textbf{Stability}: The three most stable models, characterized by the lowest \textit{model volatility}, are Baichuan1-chat (13B) at 2.4, Baichuan2-chat (13B) at 4.5, and Vicuna-v1.3 (33B) at 4.8. Conversely, the three most unstable models, showcasing the highest \textit{model volatility} across various prompts, are Llama2 (7B) at 30.8, Llama2 (13B) at 30.4, and Baichuan2-chat (7B) at 24.9, reflecting their pronounced sensitivity to prompt variations. Moving to an analysis of stability across individual tasks:
In the \textbf{NIE-P (NIE-basic)} task, the most stable models include Baichuan2-chat (13B) at 2.3, Baichuan1-chat (13B) at 2.5, and text-ada-001 at 4.6. In contrast, the most unstable models are Llama2 (7B) at 27.5, Llama2 (13B) at 24.8, and Baichuan2-chat (7B) at 23.9.
For the \textbf{NIE-P (NIE-hard)} task, the top stable models with the lowest \textit{model volatility} are Baichuan2-chat (13B) at 2.1, Baichuan1-chat (13B) also at 2.1, and Vicuna-v1.3 (33B) at 4.1. The most unstable models are Llama2 (13B) at 30.0, Llama2 (7B) at 28.9, and Baichuan2-chat (7B) at 26.0.
In the \textbf{NIE-B (NIE-natural)} task, the three most stable models are Baichuan1-chat (13B) at 2.5, babbage (1.3B) at 6.0, and Vicuna-v1.3 (33B) at 8.3. Meanwhile, the models with the largest \textit{model volatility} are Llama2 (13B) at 34.5, Llama2 (7B) at 33.0, and Baichuan2-chat (7B) at 27.8.
\textbf{Across all tasks}, Baichuan1-chat (13B) is consistently among the three most stable models, whereas Llama2 (7B), Llama2 (13B), and Baichuan2-chat (7B) are consistently among the three most unstable models.

4) \textbf{Open-Limited Ratio}: With a 2:3 ratio of open-access to limited-access models among the top five models in the entire scenario, the \textit{open-limited gap} is \textbf{small}.

Next, we explore \textit{prompt gain} in NIE:

1) \textbf{Top Gain}: Illustrated in Figure \ref{fig:Heatmap_of_Natural_Indirect_Effect}(b), the two prompts leading to the highest average accuracy improvements over the basic prompt are 3-shot IcL at 29.3\% and manual CoT at 19.5\%. The most significant increase in accuracy, when compared with the basic prompt, is seen with Llama2 (13B) employing 3-shot IcL, which results in a surge of 72.1\%. Proceeding with a more granular, task-specific analysis, Figure \ref{fig:Heatmap_of_gain_of_Natural_Indirect_Effect} presents the heatmap of gains across all tasks within the scenario. 
For the \textbf{NIE-P (NIE-basic)} task, the two prompts yielding the greatest average accuracy boost over the basic prompt are 3-shot IcL at 27.6\% and manual CoT at 18.5\%. The largest leap in accuracy against the basic prompt is with Koala (13B) using 3-shot IcL, registering a gain of 69.1\%. 
In the \textbf{NIE-P (NIE-hard)} task, the leading two prompts in terms of average accuracy gain compared to the basic prompt are 3-shot IcL at 28.9\% and manual CoT at 19.4\%, with Koala (13B) utilizing 3-shot IcL witnessing an uplift of 66.3\%. For the \textbf{NIE-B (NIE-natural)} task, the two prompts achieving the highest average accuracy gains over the basic prompt are 3-shot IcL at 31.5\% and manual CoT at 20.5\%, with Llama2 (13B) applying 3-shot IcL showing an impressive improvement of 81.8\%.
\textbf{Across each distinct task}, 3-shot IcL and manual CoT stand out as the most useful prompts, consistently demonstrating their effectiveness in enhancing performance. Specifically, 3-shot IcL helps models achieve the highest gains compared to other \textit{model-prompt pair}s. Additionally, the Natural-mode task exhibits greater top gains compared to the Mathematical-mode tasks.

2) \textbf{Exceptions}: The high-performing prompt in the scenario, 3-shot IcL, does not work well with several models in creating a positive average prompt gain, including Baichuan1-chat (13B), Vicuna-v1.3 (33B), Baichuan2-chat (13B), text-davinci-003, Claude2, and GPT-3.5-Turbo. It is noteworthy that all prompts enhance GPT-4's performance beyond the basic prompt, yet none manage to boost Claude2's performance from its basic prompt performance. 
In the \textbf{NIE-P (NIE-basic)} task, 3-shot IcL fails to be effective for Baichuan1-chat (13B), Baichuan2-chat (13B), Vicuna-v1.3 (33B), text-davinci-003, GPT-4, Claude2, and GPT-3.5-Turbo. Every prompt is capable of improving Llama2 (70B)'s performance over the basic prompt, with Claude2 again showing no improvement with any prompt in this task. 
In the \textbf{NIE-P (NIE-hard)} task, 3-shot IcL proves ineffective for Vicuna-v1.3 (33B), Baichuan1-chat (13B), Baichuan2-chat (13B), text-davinci-003, Claude2, and GPT-3.5-Turbo, with Claude2's performance remaining unimproved by any prompt. 
For the \textbf{NIE-B (NIE-natural)} task, 3-shot IcL is ineffective with Baichuan1-chat (13B), Baichuan2-chat (13B), GPT-4, text-davinci-003, Claude2, and GPT-3.5-Turbo. 
\textbf{It is worth mentioning} that 3-shot IcL fails to improve performance for Baichuan1-chat (13B), Baichuan2-chat (13B), text-davinci-003, Claude2, and GPT-3.5-Turbo across all tasks.

3) \textbf{Stability}: Regarding stability within the scenario, the two most stable prompts, indicated by the smallest \textit{prompt volatility}, are 0-shot IcL at 3.2 and adversarial doubt at 3.3. On the opposite spectrum, the two prompts exhibiting the greatest instability, marked by the highest \textit{prompt volatility}, are 3-shot IcL at 31.3 and manual CoT at 23.3. This results in an \textit{average model-prompt-gain volatility} (\textit{AMPGV}) of 17.0, illustrating the scenario's \textbf{high} \textit{prompt dependence}. Analyzing stability across individual tasks:
In the \textbf{NIE-P (NIE-basic)} task, the most stable prompts are adversarial doubt at 2.7 and 0-shot CoT at 2.9, while the most unstable prompts are 3-shot IcL at 34.5 and manual CoT at 22.7. The \textit{AMPGV} here is 16.1, signifying a \textbf{high} \textit{prompt dependence}.
For the \textbf{NIE-P (NIE-hard)} task, the prompts showing the least variability are adversarial doubt at 1.4 and adversarial ignore at 1.9, contrasted by the most variable prompts, 3-shot IcL at 30.0 and manual CoT at 23.8. This task has an \textit{AMPGV} of 16.4, indicating a \textbf{high} \textit{prompt dependence}.
In the \textbf{NIE-B (NIE-natural)} task, the two most stable prompts are 0-shot IcL at 5.7 and EF at 6.8, whereas the most unstable are 3-shot IcL at 34.7 and manual CoT at 27.8, with an \textit{AMPGV} of 19.6, showing a \textbf{high} \textit{prompt dependence}.
\textbf{After reviewing all tasks}, the range of \textit{AMPGV} is from 16.1 to 19.6, reflecting a \textbf{narrow} \textit{variance of prompt dependence}. Despite 3-shot IcL and manual CoT being among the most effective, they also rank as the most unstable across all tasks. Mathematical-mode tasks show a lesser reliance on prompts compared to the Natural-mode task, as indicated by their respective \textit{AMPGV}.

Finally, we analyze \textit{language proficiency} in NIE,

1) \textbf{English vs. Chinese}: Figure \ref{fig:Natural_Indirect_Effect_Language} reveals that models generally perform better on the English test set than on the Chinese set, with 25 out of 28 models favoring English over Chinese.

2) \textbf{Accuracy Difference}: The most pronounced performance differences between English and Chinese, with a preference for English, are noted in GPT-3.5-Turbo (21.1\%), GPT-4 (15.9\%), and InternLM-chat (7B) (15.9\%). In contrast, models such as Baichuan1-chat (13B) (8.1\%), text-davinci-003 (3.5\%), and babbage (1.3B) (1.4\%) exhibit superior performance in Chinese compared to English.

\paragraph{Probability of necessity.}
\begin{figure}[t]
\centering
\subfigure[Model performance of PN]{
\begin{minipage}{8.5cm}
\centering
\includegraphics[width=1\linewidth]{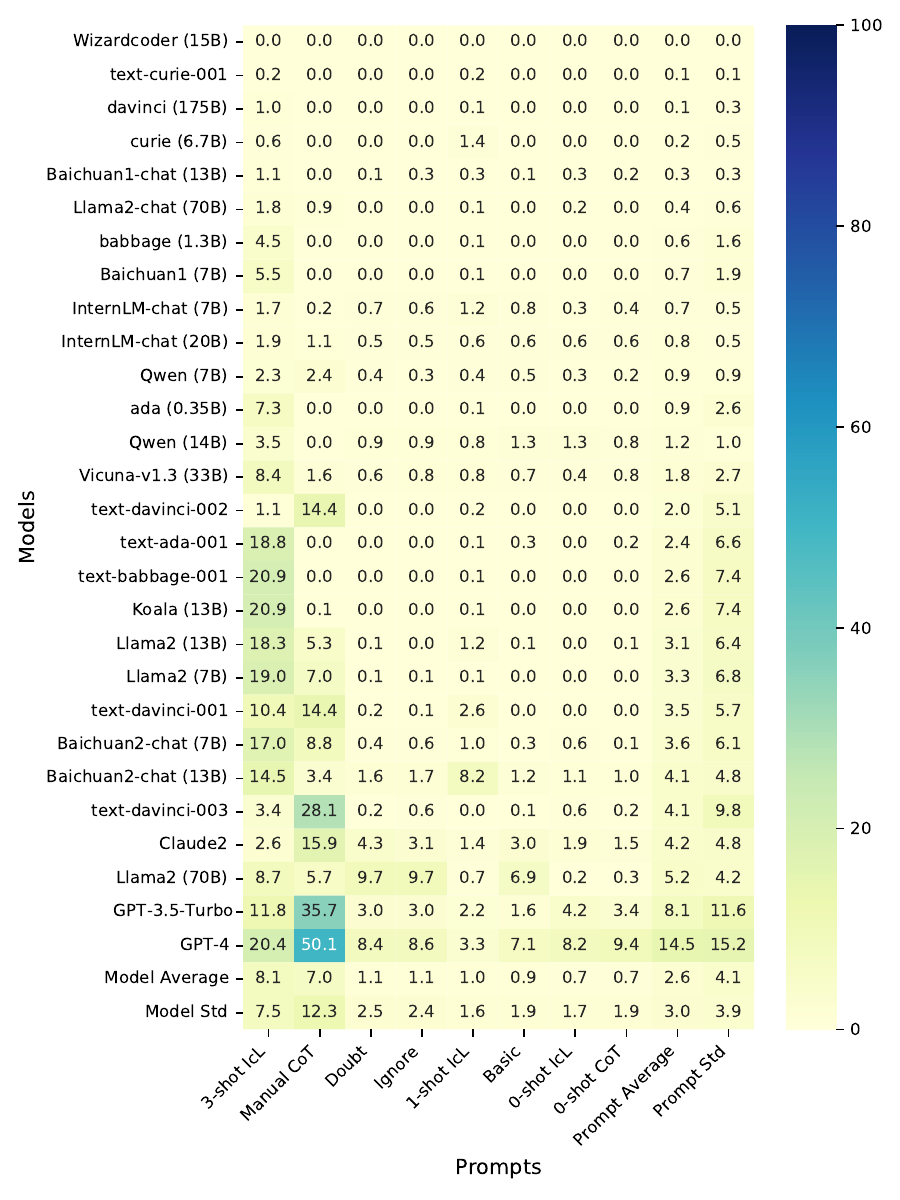}
\end{minipage}
}
\subfigure[\textit{Prompt gain} of PN]{
\begin{minipage}{8.5cm}
\centering
\includegraphics[width=1\linewidth]{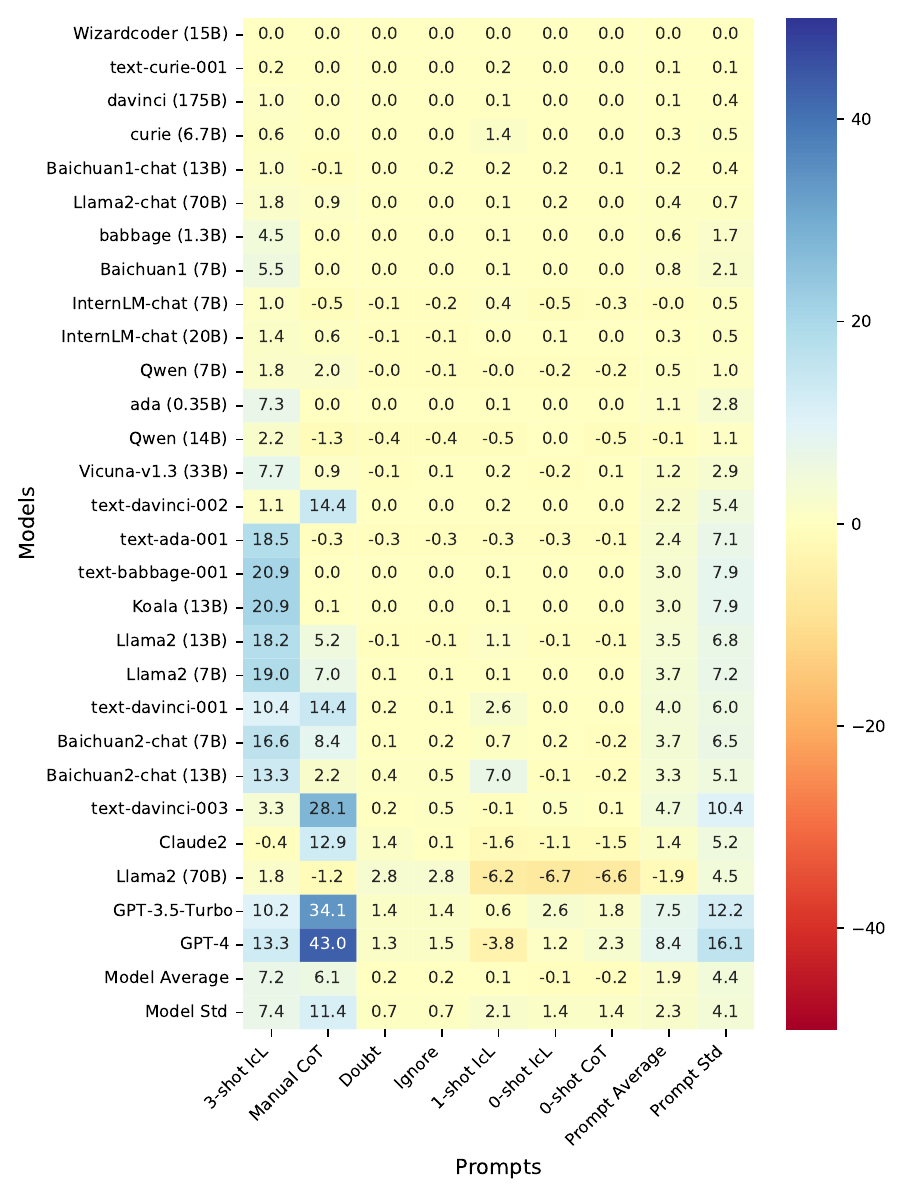}
\end{minipage}
}
\caption[Heatmap of PN]{\textbf{Heatmap of PN.} The models and prompts are sorted by their averages.}
\label{fig:Heatmap_of_Probability_of_Necessity}
\end{figure}

\begin{figure}
    \centering
    \includegraphics[width=0.8\linewidth]{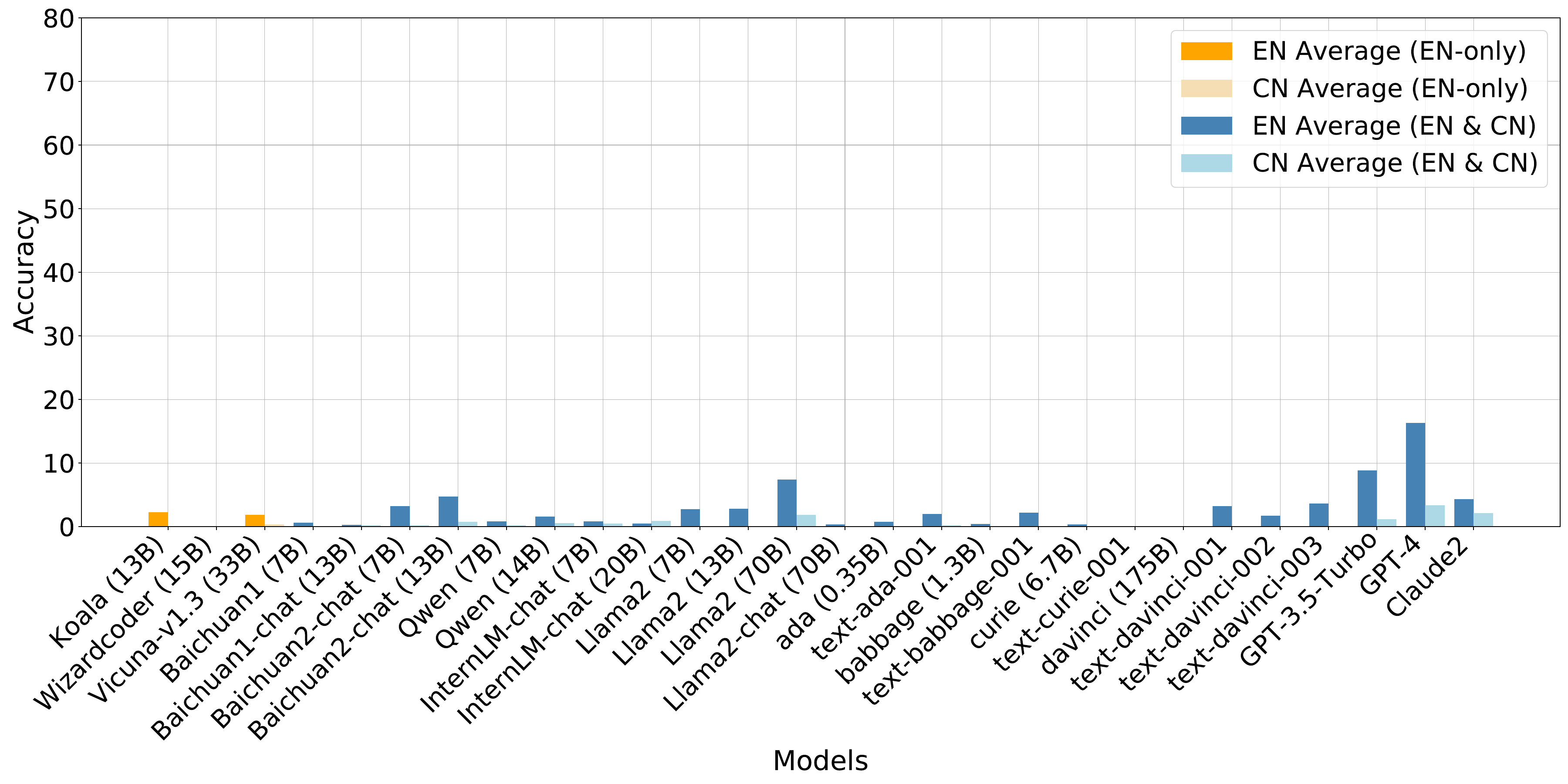}
    \caption[Language comparison of PN]{\textbf{Language comparison of PN.} The dark legend signifies the average performance of the model on an English test set, whereas the light legend denotes the average performance of the model on the Chinese test set. The yellow legend indicates a model trained exclusively on English datasets, while the blue legend represents a model trained on both English and Chinese datasets.}
    \label{fig:Probability_of_Necessity_Language}
\end{figure}
First, we evaluate model performance in PN:

1) \textbf{Distribution}: Figure \ref{fig:Distribution_of_counterfactual}(f) illustrates the distribution of all \textit{model-prompt pair}s in PN. The scenario has a median of 0.3\% and a third quartile of 1.6\%. Given that Mathematical-mode tasks are generally considered more difficult to grasp than Natural-mode tasks from a common perspective, as evidenced by both the median and third quartile values being lower than 2\%, we categorize the \textit{understandability} of the PN scenario as \textbf{very hard} to understand.
Figure \ref{fig:Distribution_of_Probability_of_Necessity_Tasks} showcases the distribution for each \textit{model-prompt pair} across individual tasks.
In the \textbf{PN-P (PN-basic)} task, the median is recorded at 0.2\%, with the third quartile at 1.4\%. Similar to the previously defined \textit{understandability} of the probability computation tasks, we regard this task as \textbf{very hard} to understand.
In the \textbf{PN-P (PN-hard)} task, the median remains at 0.2\%, with the third quartile at 1.6\%. We also consider this task having a \textit{very hard} \textit{understandability}.
\textbf{Upon examining the variance between tasks}, we find that the median accuracies are consistently at 0.2\% with a standard deviation of 0.0. The third quartile accuracies range from 1.4\% to 1.6\%, with a slight standard deviation of 0.1, indicating that the scenario has a \textbf{minimally divergent} \textit{variance of distribution}. Moreover, in both tasks, the majority of the distribution, exceeding 90\%, falls within the 0 to 10 accuracy range.

2) \textbf{Top Accuracy}: Illustrated in Figure \ref{fig:Heatmap_of_Probability_of_Necessity}(a), the three highest-performing models in terms of average accuracy within this scenario are GPT-4 at 14.5\%, GPT-3.5-Turbo at 8.1\%, and Llama2 (70B) at 5.2\%. The \textit{top model-prompt pair}, GPT-4 with manual CoT, achieves a significant 50.2\% accuracy, indicating the \textit{solvability} of this scenario is \textbf{challenging} as the performance of the \textit{top model-prompt pair} exceeds the random guess yet does not reach 80\%. Figure \ref{fig:Heatmap_of_performances_of_Probability_of_Necessity} displays the top three models' average accuracy for each task.
In the \textbf{PN-P (PN-basic)} task, the leading models in average accuracy are GPT-4 at 13.6\%, GPT-3.5-Turbo at 7.4\%, and Llama2 (70B) at 4.2\%, with GPT-4 and manual CoT securing the highest accuracy at 42.3\%. This result indicates the task \textit{solvability} is \textbf{challenging} as the \textit{top model-prompt pair}'s performance surpasses the random guess but is below 80\%. In the \textbf{PN-P (PN-hard)} task, the top performers in average accuracy are GPT-4 at 14.3\%, GPT-3.5-Turbo at 8.7\%, and Llama2 (70B) at 5.8\%, with GPT-4 and manual CoT reaching the highest at 58.0\%, again suggesting the \textit{solvability} of the task is \textbf{challenging} as the \textit{top model-prompt pair}'s performance is greater than random guess but falls short of 80\%.
\textbf{Upon comparing the tasks}, the \textit{variance of solvability} is \textbf{negligible}, with the top model's average accuracy varying slightly from 13.6\% to 14.3\% (a 0.7\% difference) and the highest accuracy by the \textit{top model-prompt pair}s ranging from 42.3\% to 58.0\% (a 15.7\% difference), signifying a \textbf{considerable} \textit{variance of model's top performance}. GPT-4 not only is the leading model in terms of average performance but also hosts the \textit{top model-prompt pair}s across all tasks. Similarly, GPT-3.5-Turbo and Llama2 (70B) consistently rank as the second and third most effective models, respectively, in terms of average accuracy across both tasks. As a prompt, manual CoT also hosts \textit{top model-prompt pair}s across all tasks.

3) \textbf{Stability}: The three most stable models, characterized by the lowest \textit{model volatility}, are Wizardcoder (15B) with a \textit{model volatility} of 0.0, text-curie-001 with a \textit{model volatility} of 0.1, and davinci (175B) with a \textit{model volatility} of 0.3. Conversely, the three models showing the greatest instability across different prompts, indicated by the highest \textit{model volatility}, are GPT-4 at 15.2, GPT-3.5-Turbo at 11.6, and text-davinci-003 at 9.8, reflecting their pronounced sensitivity to prompt changes. An analysis of stability across individual tasks follows.
In the \textbf{PN-P (PN-basic)} task, the most stable models are Wizardcoder (15B) with \textit{model volatility} of 0.0, text-curie-001 with \textit{model volatility} of 0.2, and Baichuan1-chat (13B) with \textit{model volatility} of 0.3. The most unstable models in this task are GPT-4 at 12.6, GPT-3.5-Turbo at 10.9, and text-davinci-003 at 9.0.
For the \textbf{PN-P (PN-hard)} task, the top three stable models are Wizardcoder (15B) and text-curie-001, both with \textit{model volatility} of 0.0, and davinci (175B) with \textit{model volatility} of 0.3. The models exhibiting the most instability are GPT-4 at 17.9, GPT-3.5-Turbo at 11.6, and text-davinci-001 at 10.4.
Across both tasks, Wizardcoder (15B) and text-curie-001 consistently rank as the most and second-most stable models, respectively, while GPT-4 and GPT-3.5-Turbo are the most and second-most unstable models.

4) \textbf{Open-Limited Ratio}: With a 1:4 ratio of open-access to limited-access models among the top five models, the \textit{open-limited gap} is \textbf{moderate}.

Then, we analyze \textit{prompt gain} in PN:

1) \textbf{Top Gain}: Illustrated in Figure \ref{fig:Heatmap_of_Probability_of_Necessity}(b), the two prompts leading to the most substantial average accuracy improvements over the basic prompt are 3-shot IcL at 7.2\% and manual CoT at 6.1\%. The largest increase in accuracy relative to the basic prompt is achieved by GPT-4 employing manual CoT, with a significant gain of 43.0\%. A more granular, task-specific analysis follows. Figure \ref{fig:Heatmap_of_gain_of_Probability_of_Necessity} presents the heatmap of gains across all tasks within the scenario. 
In the \textbf{PN-P (PN-basic)} task, the two prompts yielding the greatest average accuracy enhancements over the basic prompt are 3-shot IcL at 7.3\% and manual CoT at 4.8\%. The most substantial improvement in accuracy against the basic prompt is seen with GPT-3.5-Turbo using manual CoT, marking a gain of 34.5\%. 
In the \textbf{PN-P (PN-hard)} task, the leading two prompts by average accuracy gain over the basic prompt are manual CoT at 7.4\% and 3-shot IcL at 7.2\%, with GPT-4 utilizing manual CoT demonstrating a significant uplift of 51.7\%. 
\textbf{The task evaluation} shows a preference for manual CoT in enhancing model accuracy.

2) \textbf{Exceptions}: In this scenario, the leading prompt, 3-shot IcL, shows effectiveness for generating positive average \textit{prompt gain} with nearly all models except Claude2. Every prompt manages to elevate GPT-3.5-Turbo's performance beyond the basic prompt level. Within the \textbf{PN-P (PN-basic)} task, 3-shot IcL does not yield a positive average \textit{prompt gain} with Claude2, yet all prompts are successful in boosting GPT-3.5-Turbo's performance from its basic prompt performance. In the \textbf{PN-P (PN-hard)} task, manual CoT fails to enhance the performance for Baichuan1-chat (13B), InternLM-chat (7B), Qwen (14B), text-ada-001, and Llama2 (70B). However, all prompts are capable of improving GPT-3.5-Turbo's performance over the basic prompt.
\textbf{Across both tasks}, GPT-3.5-Turbo benefits from all prompts for positive average prompt gain.

3) \textbf{Stability}: The two most stable prompts are adversarial ignore and adversarial doubt, both with a \textit{prompt volatility} of 0.7. Conversely, the least stable prompts, demonstrated by the highest \textit{prompt volatility}, are manual CoT at 11.4 and 3-shot IcL at 7.4. The \textit{average model-prompt-gain volatility} (\textit{AMPGV}) of 4.4 shows a \textbf{low} \textit{prompt dependence} within the scenario. Analyzing stability across individual tasks:
In the \textbf{PN-P (PN-basic)} task, the most stable prompts, adversarial ignore, and adversarial doubt, both exhibit \textit{prompt volatility} of 0.6. The least stable prompts are manual CoT at 10.1 and 3-shot IcL at 7. The task shows a \textbf{low} \textit{prompt dependence} as evidenced by an \textit{AMPGV} of 3.8.
In the \textbf{PN-P (PN-hard)} task, the most stable prompts in terms of stability are adversarial doubt and adversarial ignore, each with a \textit{prompt volatility} of 0.8. The least stable prompts, showing the largest \textit{prompt volatility}, are manual CoT at 13.8 and 3-shot IcL at 7.5. The task has a \textbf{low} \textit{prompt dependence} with an \textit{AMPGV} of 4.9.
\textbf{Upon reviewing all tasks}, the range of \textit{AMPGV}, indicating the degree of \textit{variance of prompt dependence}, spans a \textbf{narrow} spectrum from 3.8 to 4.9. Adversarial ignore and adversarial doubt are the most stable prompts, while manual CoT and 3-shot IcL are identified as the most unstable.

Last, we look into \textit{language proficiency} in PN:

1) \textbf{English vs. Chinese}: Figure \ref{fig:Probability_of_Necessity_Language} reveals that models tend to perform better on the English test set compared to the Chinese test set, with 25 out of 28 models exhibiting superior performance in English over Chinese.

2) \textbf{Accuracy Difference}: The most significant differences in accuracy between English and Chinese, with a preference for English, are seen in GPT-4 (12.9\%), GPT-3.5-Turbo (7.7\%), and Llama2 (70B) (5.6\%). On the other hand, models like InternLM-chat (20B) (0.4\%), text-curie-001 (0.1\%) demonstrate higher proficiency in Chinese than in English.

\paragraph{Probability of sufficiency.}
\begin{figure}[t]
\centering
\subfigure[Model performance of PS]{
\begin{minipage}{8.5cm}
\centering
\includegraphics[width=1\linewidth]{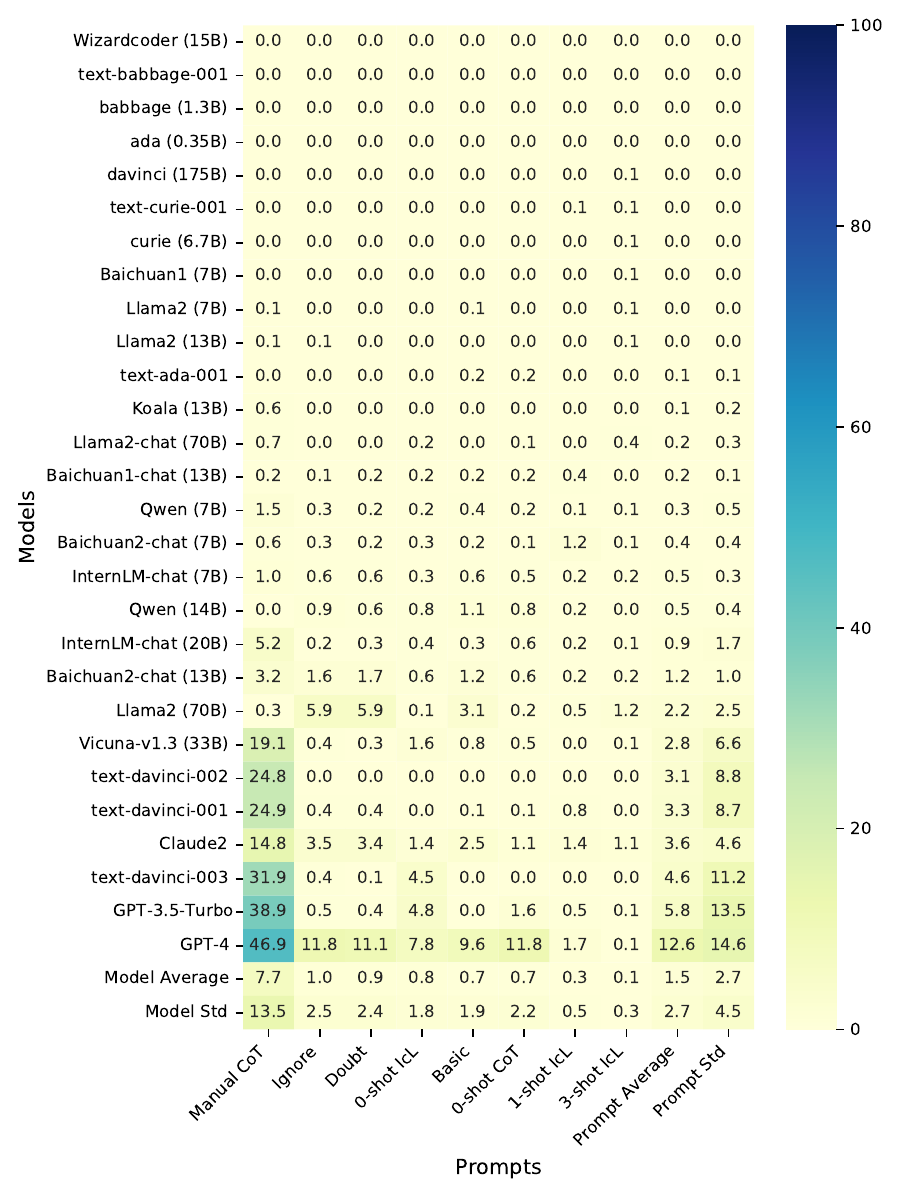}
\end{minipage}
}
\subfigure[\textit{Prompt gain} of PS]{
\begin{minipage}{8.5cm}
\centering
\includegraphics[width=1\linewidth]{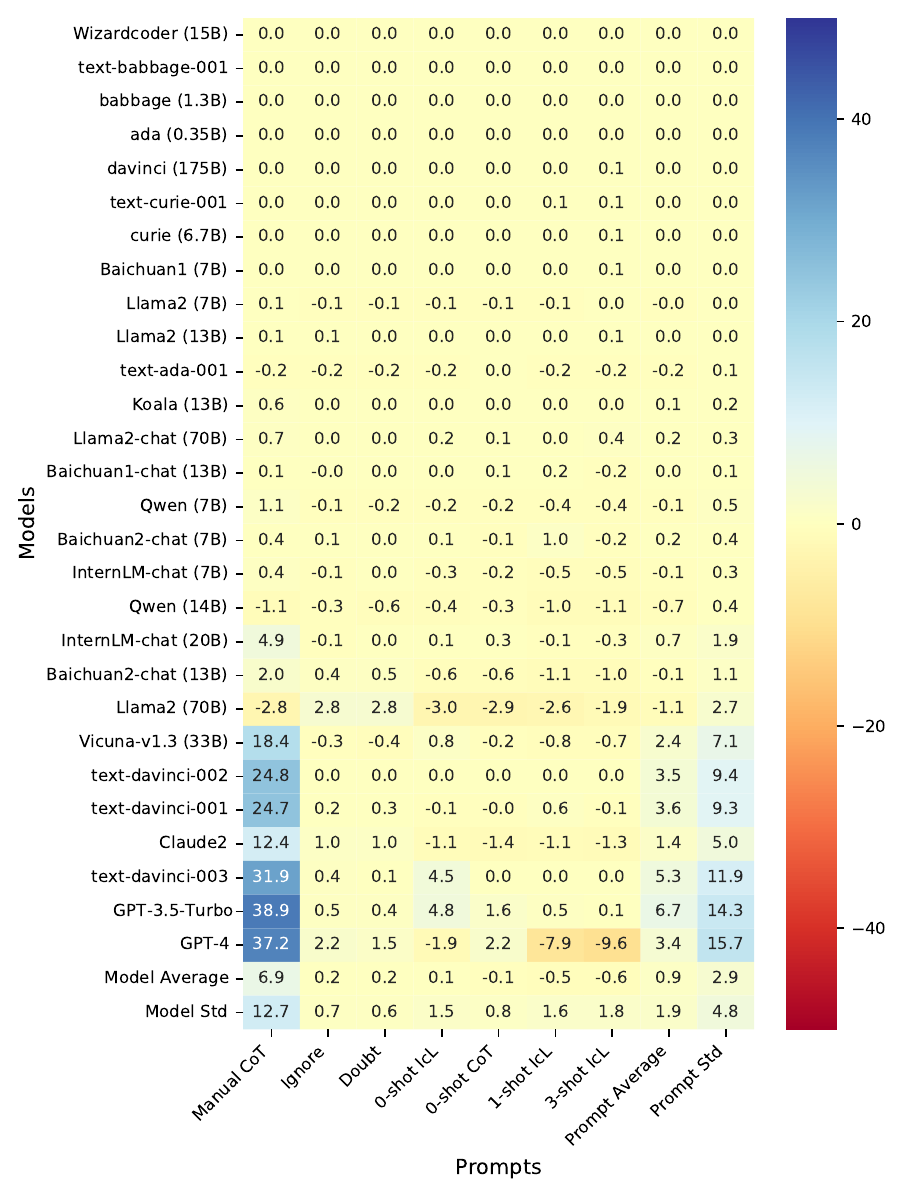}
\end{minipage}
}
\caption[Heatmap of PS]{\textbf{Heatmap of PS.} The models and prompts are sorted by their averages.}
\label{fig:Heatmap_of_Probability_of_Sufficiency}
\end{figure}

\begin{figure}
    \centering
    \includegraphics[width=0.8\linewidth]{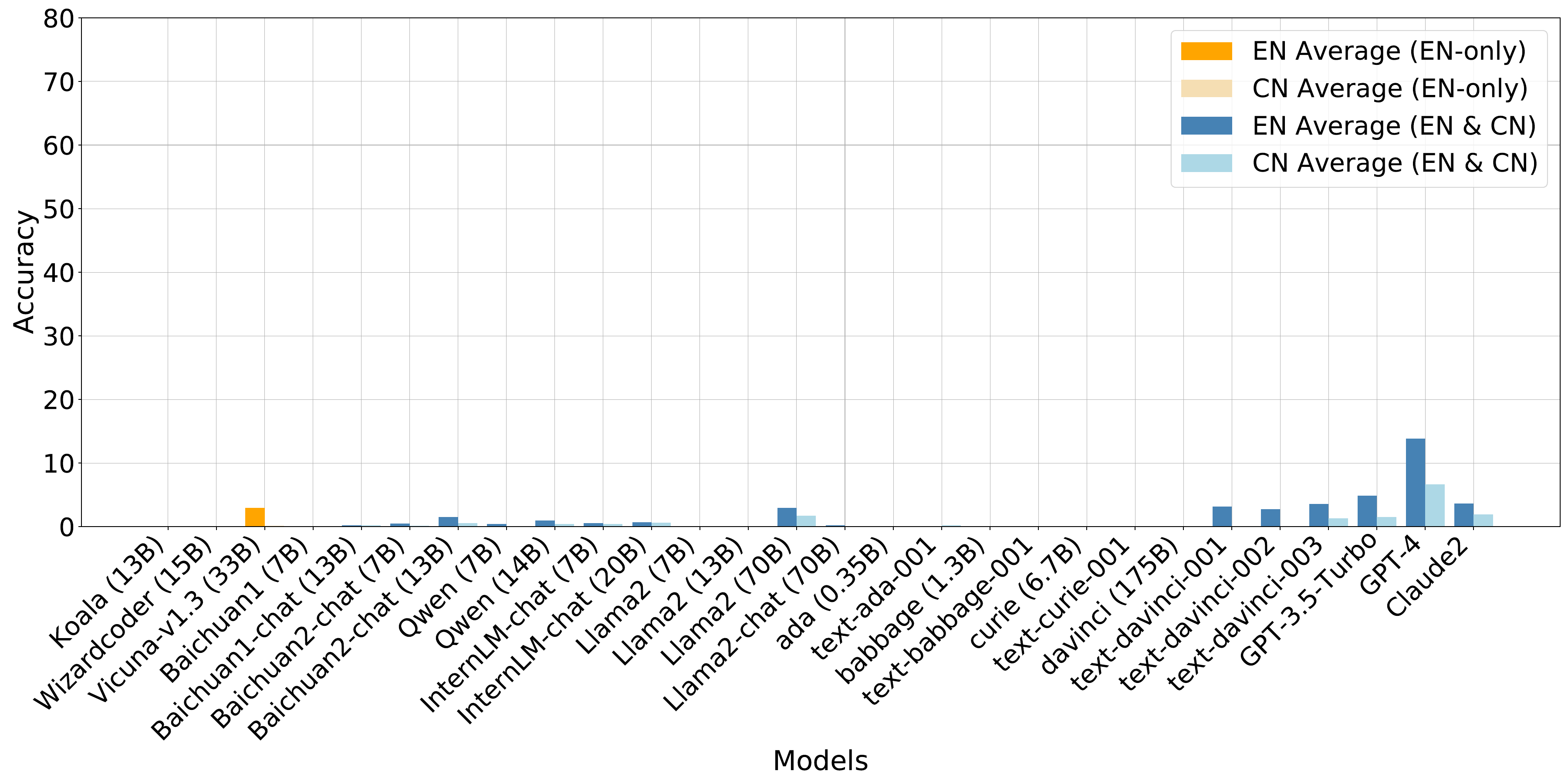}
    \caption[Language comparison of PS]{\textbf{Language comparison of PS.} The dark legend signifies the average performance of the model on an English test set, whereas the light legend denotes the average performance of the model on the Chinese test set. The yellow legend indicates a model trained exclusively on English datasets, while the blue legend represents a model trained on both English and Chinese datasets.}
    \label{fig:Probability_of_Sufficiency_Language}
\end{figure}
Initially, we delve into model performance in PS:

1) \textbf{Distribution}: Figure \ref{fig:Distribution_of_counterfactual}(g) showcases the distribution for all \textit{model-prompt pair}s regarding the PS, with a median calculated at 0.0\% and the third quartile at 0.4\%. Based on subjective assessments (i.e., human understanding of the PS scenario) and the fact that the third quartile for all tasks within this scenario is less than 0.5\%, we subjectively define the \textit{understandability} of this scenario as \textbf{very hard}.
Figure \ref{fig:Distribution_of_Probability_of_Sufficiency_Tasks} details the distribution for each \textit{model-prompt pair} across specific tasks.
In the \textbf{PS-P (PS-basic)} task, the median is 0.1\%, and the third quartile is 0.5\%. Due to the challenging nature of Mathematical-mode tasks and the extremely low score of the median and third quartile, we regard the task \textit{understandability} as \textit{very hard}.
In the \textbf{PS-P (PS-hard)} task, the median is 0.0\%, with the third quartile at 0.4\%. Likewise, we define the task \textit{understandability} as \textit{very hard}.
\textbf{Upon evaluating the differences across tasks}, it is observed that median and the third quartile accuracies vary slightly from 0.0\% to 0.1\% and from 0.4\% to 0.5\%, both having nearly zero standard deviations. This indicates that the scenario has a \textbf{minimally divergent} \textit{variance of distribution}. Notably, the PS-P (PS-basic) task exhibits slightly higher scores in both median and third quartile than the PS-P (PS-hard) task. Moreover, in both tasks, a significant majority of \textit{model-prompt pair}s (over 90\%) fall within a 0\% to 10\% accuracy range.

2) \textbf{Top Accuracy}: Illustrated in Figure \ref{fig:Heatmap_of_Probability_of_Sufficiency}(a), the leading three models in this scenario based on average accuracy are GPT-4 at 12.6\%, GPT-3.5-Turbo at 5.8\%, and text-davinci-003 at 4.6\%. The \textit{top model-prompt pair} is GPT-4 with manual CoT, achieving a score of 46.8\%, indicating that the \textit{solvability} of this scenario is \textbf{challenging} as the \textit{top model-prompt pair} exceeds the random guess yet does not reach 80. Figure \ref{fig:Heatmap_of_performances_of_Probability_of_Sufficiency} provides an analysis of the top three models' average accuracy for each specific task.
In the \textbf{PS-P (PS-basic)} task, the highest average accuracies are reported for GPT-4 at 11.9\%, GPT-3.5-Turbo at 5.5\%, and Claude2 at 3.6\%, with GPT-4 and manual CoT forming the best \textit{top model-prompt pair} at 41.6\%. This outcome signifies that the task \textit{solvability} is \textbf{challenging} as the \textit{top model-prompt pair}'s performance is greater than a random guess but below 80\%. In the \textbf{PS-P (PS-hard)} task, the top performers in average accuracy are GPT-4 at 12.8\%, text-davinci-001 at 6.3\%, and text-davinci-003 also at 6.3\%, with GPT-4 and manual CoT reaching the highest at 52.1\%, again highlighting the task's \textbf{challenging} \textit{solvability} as the \textit{top model-prompt pair}'s performance surpasses the random guess but remains under 80\%.
\textbf{Upon examining the tasks}, the \textit{variance of solvability} is \textbf{negligible}, with the top model's average accuracy fluctuating slightly from 11.9\% to 12.8\% (a difference of 0.9\%), and the peak accuracy achieved by \textit{top model-prompt pair} ranging from 41.6\% to 52.1\% (a difference of 10.5\%). This indicates a \textbf{considerable} \textit{variance of model's top performance}. GPT-4 not only stands out as the leading model in terms of average performance but also forms the most efficient \textit{model-prompt pair}s combining with manual CoT across all tasks.

3) \textbf{Stability}: There are more than three models with zero \textit{model volatility} in the scenario. Conversely, the models exhibiting the greatest instability across various prompts, indicated by the highest \textit{model volatility}, are GPT-4 at 14.6, GPT-3.5-Turbo at 13.5, and text-davinci-003 at 11.2, showcasing their significant sensitivity to prompt variations. An analysis of stability across specific tasks is as follows:
In the \textbf{PS-P (PS-basic)} task, a number of models achieve \textit{model volatility} of 0.0, denoting maximum stability. The models facing the greatest instability, demonstrated by the largest \textit{model volatility}, are GPT-4 at 11.9, GPT-3.5-Turbo at 11.4, and Vicuna-v1.3 (33B) at 6.7.
For the \textbf{PS-P (PS-hard)} task, again, several models report \textit{model volatility} of 0.0, indicating a high level of stability. The most unstable models are text-davinci-002, text-davinci-003, and text-davinci-001, each with a \textit{model volatility} of 17.5.

4) \textbf{Open-Limited Ratio}: The ratio of open-access to limited-access models among the top five models with the highest average accuracy in the entire scenario is 0:5, indicating a \textbf{large} \textit{open-limited gap}.

Following this, we evaluate \textit{prompt-gain} in PS:

1) \textbf{Top Gain}: Illustrated in Figure \ref{fig:Heatmap_of_Probability_of_Sufficiency}(b), the two prompts leading to the highest average accuracy improvements over the basic prompt are manual CoT at 6.9\% and adversarial ignore at 0.2\%. The most significant increase in accuracy compared to the basic prompt is achieved by GPT-3.5-Turbo with manual CoT, registering a gain of 38.9\%. A more detailed, task-specific analysis is conducted next. Figure \ref{fig:Heatmap_of_gain_of_Probability_of_Sufficiency} presents the heatmap of gains across all tasks in the scenario. In the \textbf{PS-P (PS-basic)} task, the leading two prompts by average accuracy improvement over the basic prompt are manual CoT at 4.3\% and 0-shot IcL at 0.5\%, with the most substantial gain observed with GPT-3.5-Turbo using manual CoT, indicating a rise of 34.9\%. In the \textbf{PS-P (PS-hard)} task, the top two prompts for average accuracy gain over the basic prompt are manual CoT at 9.5\% and adversarial ignore at 0.2\%, with text-davinci-002 employing manual CoT demonstrating the most significant improvement, a leap of 49.5\%.
\textbf{Across both tasks}, manual CoT is favored for achieving the highest average model gain and the maximum gain in all \textit{model-prompt pair}s.

2) \textbf{Exceptions}: The most high-performing prompt in the scenario, manual CoT, shows effectiveness with the majority of models in generating positive average prompt gain, but excluding text-ada-001, Qwen (14B), and Llama2 (70B). Every prompt is capable of enhancing GPT-3.5-Turbo's performance beyond its basic prompt performance. However, no prompt manages to elevate Qwen (14B)'s performance above the basic prompt. In the \textbf{PS-P (PS-basic)} task, manual CoT cannot give a positive average \textit{prompt gain} to text-ada-001, text-davinci-001, Qwen (14B), and Llama2 (70B). In the \textbf{PS-P (PS-hard)} task, manual CoT fails to be effective with text-ada-001, Qwen (14B), and Llama2 (70B). All prompts, however, can enhance the GPT-3.5-Turbo's performance over the basic prompt, with Qwen (14B) again showing no improvement from any prompt in the task.
\textbf{Across both tasks}, the preferred prompt, manual CoT, does not improve accuracy for text-ada-001 over its basic prompt performance.

3) \textbf{Stability}: Regarding stability within the scenario, the two most stable prompts, exhibiting the smallest \textit{prompt volatility}, are adversarial doubt at 0.6 and adversarial ignore at 0.7. On the other hand, the most unstable prompts, identified by the largest \textit{prompt volatility}, are manual CoT at 12.7 and 3-shot IcL at 1.8. The \textit{average model-prompt-gain volatility} (\textit{AMPGV}) is 2.9, indicating a \textbf{low} \textit{prompt dependence} across the scenario. Stability is further assessed on a task-specific basis:
For the \textbf{PS-P (PS-basic)} task, the most stable prompts are adversarial doubt and 0-shot CoT, both with a \textit{prompt volatility} of 0.6. Conversely, the least stable prompts are manual CoT at 9.6 and 0-shot IcL at 2.5. The task showcases a \textbf{low} \textit{prompt dependence} with an \textit{AMPGV} of 1.8.
In the \textbf{PS-P (PS-hard)} task, the most stable prompts in terms of stability are adversarial doubt and adversarial ignore, each with a \textit{prompt volatility} of 0.8. The least stable prompts, marked by the highest \textit{prompt volatility}, are manual CoT at 18.2 and 1-shot IcL at 1.9. The task reveals a \textbf{low} \textit{prompt dependence} with an \textit{AMPGV} of 3.9.
\textbf{Upon reviewing all tasks in the scenario}, the range of \textit{AMPGV}, reflecting the \textit{variance of prompt dependence}, spans a \textbf{narrow} spectrum from 1.8 to 3.9. Adversarial doubt is the most stable prompt across tasks, while manual CoT is consistently the most unstable.
        
Finally, we measure \textit{language proficiency} in PS:

1) \textbf{English vs. Chinese}: Figure \ref{fig:Probability_of_Sufficiency_Language} indicates that models generally achieve better results on the English test set compared to the Chinese test set, with 18 out of 28 models showing superior performance in English over Chinese.

3) \textbf{Accuracy Difference}: The most significant performance disparities between English and Chinese, with a preference for English, are seen with GPT-4 (7.2\%), GPT-3.5-Turbo (3.4\%), and text-davinci-001 (3.1\%). On the flip side, text-ada-001 (0.2\%) is the only model demonstrating higher proficiency in Chinese relative to English.

\paragraph{Causal explanation generation.}
\begin{figure}[t]
\centering
\subfigure[Model performance of CEG]{
\begin{minipage}{8.5cm}
\centering
\includegraphics[width=1\linewidth]{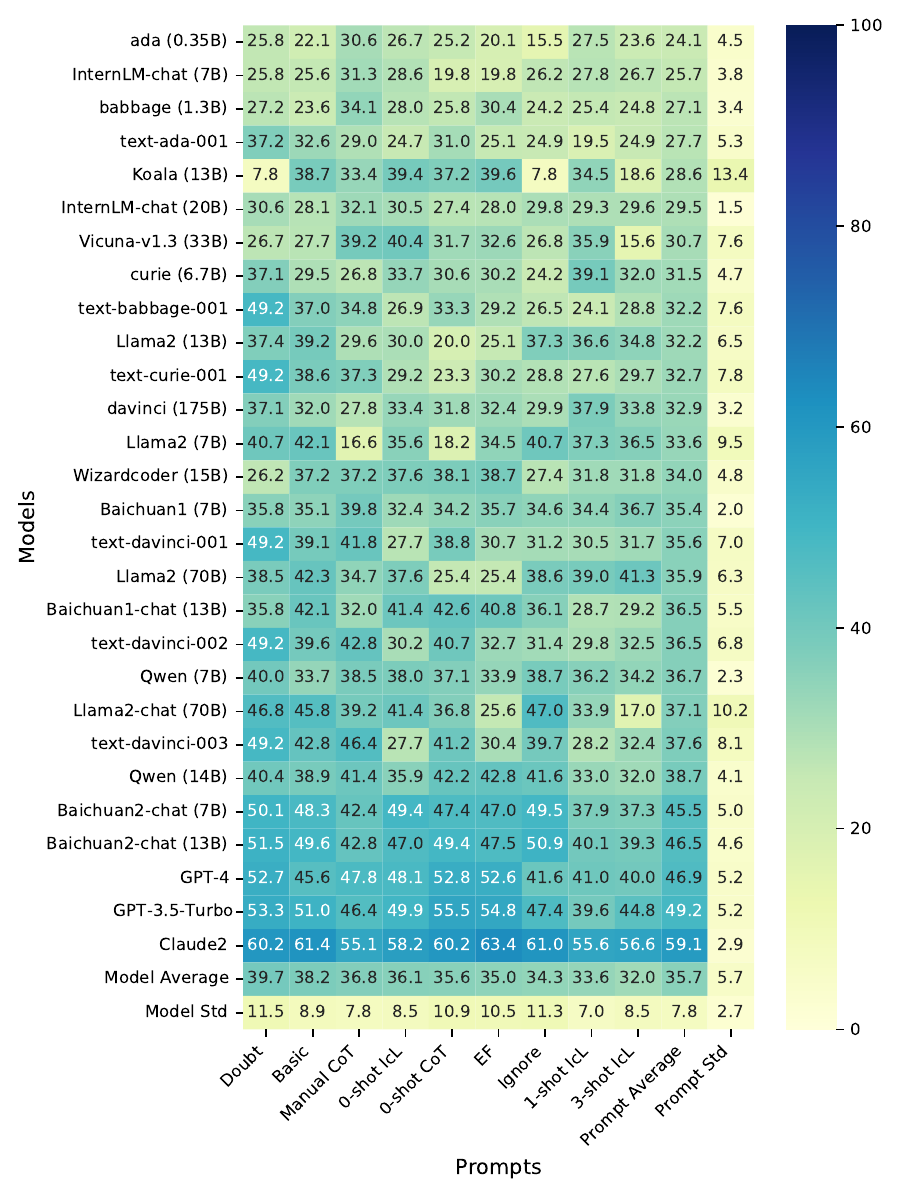}
\end{minipage}
}
\subfigure[\textit{Prompt gain} of CEG]{
\begin{minipage}{8.5cm}
\centering
\includegraphics[width=1\linewidth]{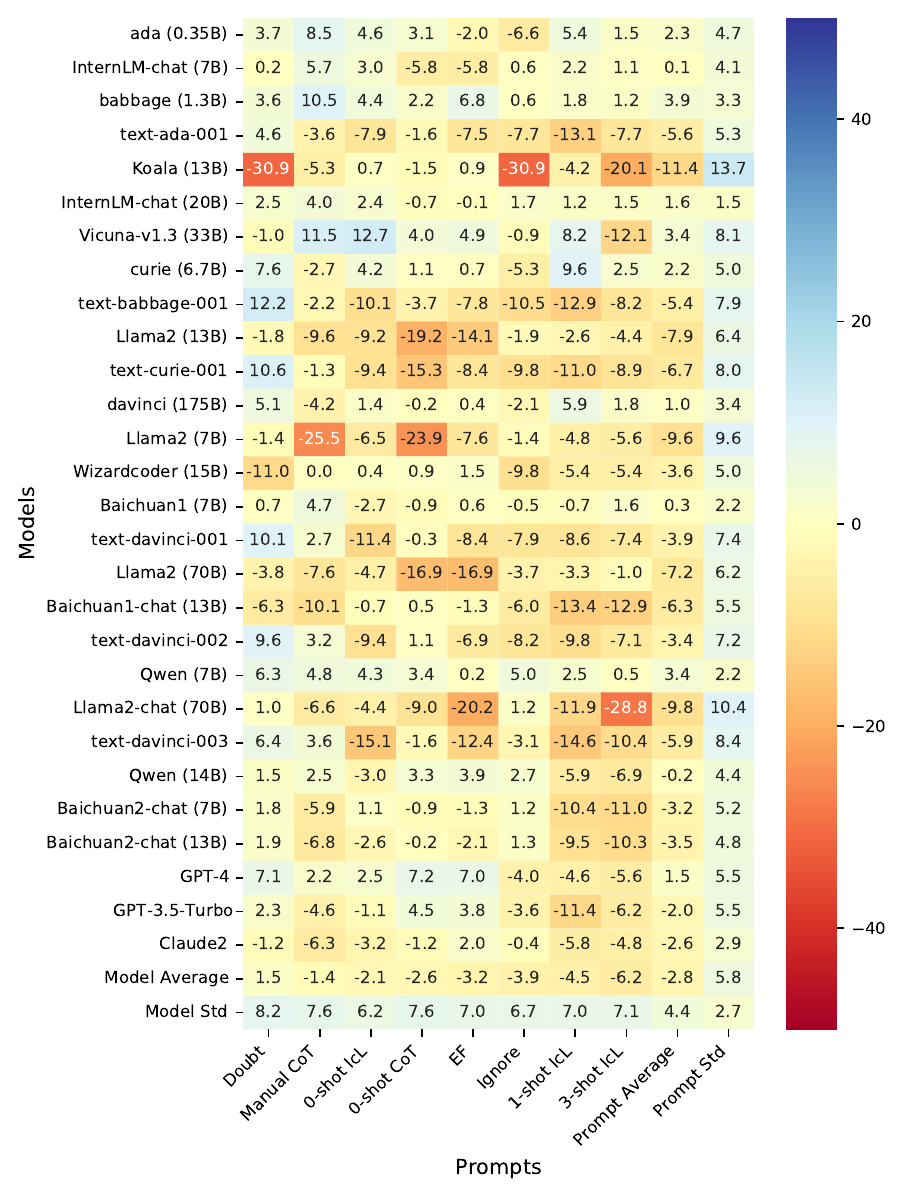}
\end{minipage}
}
\caption[Heatmap of CEG]{\textbf{Heatmap of CEG.} The models and prompts are sorted by their averages.}
\label{fig:Heatmap_of_Causal_Explanation_Generation}
\end{figure}

\begin{figure}[!t]
    \centering
    \includegraphics[width=0.8\linewidth]{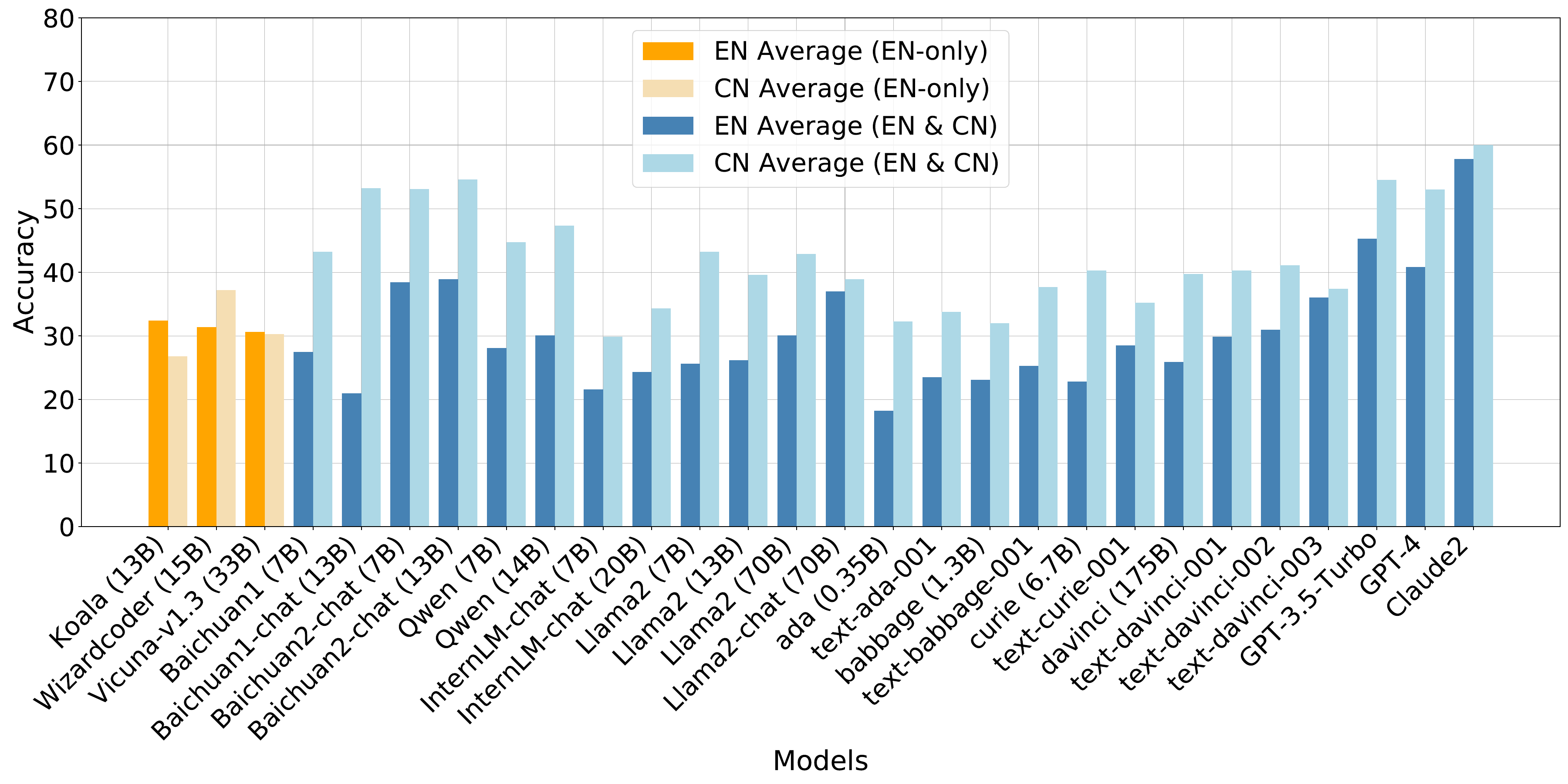}
    \caption[Language comparison of CEG]{\textbf{Language comparison of CEG.} The dark legend signifies the average performance of the model on an English test set, whereas the light legend denotes the average performance of the model on the Chinese test set. The yellow legend indicates a model trained exclusively on English datasets, while the blue legend represents a model trained on both English and Chinese datasets.}
    \label{fig:Causal_Explanation_Generation_Language}
\end{figure}

Regarding model performance in CEG:
1) \textbf{Distribution}: Figure \ref{fig:Distribution_of_counterfactual}(h) illustrates the distribution for all \textit{model-prompt pair}s, highlighting a median of 35.0\% and a third quartile of 40.8\%. As CEG is not a close-ended scenario, we choose to evaluate its \textit{understandability} objectively. Considering that most models exhibit stronger performance in processing Natural-mode tasks and the median of the CEG scenario is over 30\%, we categorize its \textit{understandability} as \textbf{easy}.

2) \textbf{Top Accuracy}: As shown in Figure \ref{fig:Heatmap_of_Causal_Explanation_Generation}(a), Claude2, GPT-3.5-Turbo, and GPT-4 are the top three models by average accuracy. Claude2, using EF, reaches a peak accuracy of 63.4\%, positioning the \textit{solvability} of this scenario as \textbf{challenging} since the \textit{top model-prompt pair} does not achieve an accuracy of 80\%.
3) \textbf{Stability}: The models demonstrating the greatest variance in response to different prompts, as indicated by the \textit{model volatility} described in \cref{metric:model}, include Koala (13B) and Llama2-chat (70B). In contrast, the models with the least variance are InternLM-chat (20B), Baichuan1 (7B), and Qwen (7B).
4) \textbf{Open-Limited Ratio}: Among the top five models, a 2:3 ratio of open-access to limited-access models indicates a \textbf{small} \textit{open-limited gap}.

Regarding \textit{prompt gain} in CEG:
1) \textbf{Top Gain}: Figure \ref{fig:Heatmap_of_Causal_Explanation_Generation}(b) points out adversarial doubt and manual CoT as the top two prompts for average accuracy gain over the basic prompt, with Vicuna-v1.3 (33B) using 0-shot IcL showcasing the most considerable increase of 12.7\%.
2) \textbf{Exceptions}: The most high-performing prompt, adversarial doubt, does not align well with models including Koala (13B) and Llama2 (70B) for generating a positive average prompt gain. Nonetheless, all prompts manage to improve the performance of babbage (1.3B) and Qwen (7B) above their basic prompt performance, while most of the models in the Llama2 series are an exception where no prompt leads to enhancement.
3) \textbf{Stability}: The most stable prompts, with the lowest \textit{prompt volatility}, are 0-shot IcL and adversarial ignore, whereas the least stable prompts are adversarial doubt and manual CoT. This scenario exhibits a \textbf{medium} \textit{prompt dependence}, as reflected by the \textit{average model-prompt-gain volatility} (\textit{AMPGV}) of 5.8.

In terms of \textit{language proficiency} in CEG, 
1) \textbf{English vs. Chinese}: as highlighted in Figure \ref{fig:Causal_Explanation_Generation_Language}, the performance of models on the Chinese test set is better than the one on the English test set, with only 2 out of 28 models performing better in English than in Chinese. 
2) \textbf{Accuracy Difference}: Significant discrepancies in performance between English and Chinese, favoring English, are observed in models like Koala (13B) (5.6\%), and Vicuna-v1.3 (33B) (0.3\%). Conversely, models such as Baichuan1-chat (13B) (32.2\%), Llama2 (7B) (17.6\%), and curie (6.7B) (17.5\%) show greater proficiency in Chinese compared to English.

\clearpage

\clearpage


\section{Related Work}
\label{related}

\begin{fquote}[Isaac Newton][1675]
If I have seen further it is by standing on the shoulders of Giants.
\end{fquote}

In this section, we review the foundational efforts associated with the construction of CaLM. Our discussion is organized into four key areas: \nameref{related:advancement} (\cref{related:advancement}), \nameref{related:general} (\cref{related:general}), \nameref{related:causal} (\cref{related:causal}), and \nameref{related:dataset} (\cref{related:dataset}).

\subsection{Advancements in Language Models}
\label{related:advancement}

Language models have achieved remarkable success today, reflecting the persistent efforts of numerous research over time \citep{shannon1948mathematical,goldman1958speech,jelinek1976continuous,jelinek1998statistical,rosenfeld2000two,bengio2000neural,mikolov2010recurrent,devlin2018bert,radford2018improving,radford2019language,raffel2020exploring,brown2020language,chatgpt2022,openai2023gpt4,touvron2023llama,claude2023,meta2024llama3}. \citet{zhao2023survey} divide the evolution of language models from the 1990s to today into four primary stages: \emph{statistical language models (SLM)} from 1990s to 2013, \emph{neural language models (NLM)} from 2013 to 2018, \emph{pre-trained language models (PLM)} from 2018 to 2020, and \emph{large language models (LLM)} from 2020 to now. Moreover, \citet{yang2023harnessing} provide a clear ``\emph{evolutionary tree}''. It outlines the most influential language models from 2018 to now, and classifies them by the model architecture (i.e., \emph{non-transformer based}, \emph{decoder-only}, \emph{encoder-only}, and \emph{encoder-decoder}). Given that CaLM evaluates only models launched after 2020, we will concentrate our review on the advancements in language models since that year.

The introduction of GPT-3 \citep{brown2020language} marked a significant advancement in text generation capabilities, demonstrating considerable proficiency and achieving excellent results through \emph{in-context learning} \citep{brown2020language,dong2022survey}. Following the release of GPT-3, OpenAI introduced InstructGPT \citep{ouyang2022training} in 2022, building upon GPT-3's foundation to exhibit superior capabilities in reasoning and alignment with humans. Subsequently, in November 2022, the unveiling of ChatGPT \citep{chatgpt2022} gained widespread attention for its advanced capabilities in diverse tasks such as story writing, question answering, conversation understanding, and even emotion understanding, code generation, and logical reasoning. This was followed by the launch of GPT-4 \citep{openai2023gpt4}, representing a significant leap forward in artificial intelligence. GPT-4's advanced features cater to a wide range, fostering creativity and innovation by assisting in idea generation, code suggestion, and design conceptualization. While these models remain limited-access, Meta's release of Llama2 \citep{touvron2023llama} marked a notable development in the open-access model landscape. Meta's decision to make the weights and tokenizers of Llama2-series models publicly available is considered a pivotal moment for the open-source community, enabling broader experimentation and customization across various domains. Based on Llama2, Meta recently introduced Llama3 \citep{meta2024llama3}, which boasts substantial performance enhancements.\footnote{\footnotesize \url{https://llama.meta.com/llama3/}} We are confident that it will continue to propel the advancement and growth of language models.

\subsection{Evaluations of Language Models' General Abilities}
\label{related:general}

As language models rapidly evolve, numerous benchmarks have been developed to objectively evaluate their performance. Some benchmarks are widely recognized for their extensive scope. MMLU \citep{hendrycks2020measuring} is designed to test if models have extensive world knowledge applicable to multi-task situations. It encompasses 57 tasks across various domains including microeconomics, computer science, and medicine. HELM \citep{liang2022holistic} has established an abstract taxonomy that spans scenarios and metrics. Meanwhile, it provides a comprehensive evaluation of language models' capabilities across 16 core scenarios (e.g., question answering, summarization), 21 additional scenarios (e.g., biases, disinformation generation), and 7 metrics (e.g., accuracy, calibration, robustness). It is fair to say that HELM has also played a crucial role in inspiring the development of CaLM. BIG-Bench \citep{srivastava2023beyond} is a comprehensive benchmark consisting of 204 tasks, created by 450 authors from 132 different institutions. It focuses on tasks that were thought to exceed the abilities of language models at that time, including areas like math, social biases, and software development. Building on BIG-Bench, BIG-Bench Hard (BBH) is subsequently developed \citep{suzgun2022challenging}. BBH consists of the 23 most difficult tasks in the BIG-Bench, and solving these tasks generally requires multi-step reasoning. KOLA \citep{yu2024kola} introduces a thorough assessment of language models' world knowledge from the perspective of cognitive abilities, breaking it down into four levels: \emph{memorization}, \emph{understanding}, \emph{applying}, and \emph{creating}. In addition to the aforementioned efforts, there are numerous other excellent benchmarks that effectively evaluate language models, including GLUE \citep{wang2018glue}, SuperGLUE \citep{wang2019superglue}, GLUE-X \citep{yang2023glue}, PandaLM \citep{wang2024pandalm}, and Xiezhi \citep{gu2024xiezhi}. Given the constraints of space and the continuous development of new benchmarks, it is not feasible to cover all benchmarks comprehensively. However, an effort by \citet{chang2023survey} has been instrumental in providing a thorough overview of language model evaluations. They have meticulously cataloged current evaluation practices, methods, and benchmarks for language models, and maintain a website\footnote{\footnotesize \url{https://github.com/MLGroupJLU/LLM-eval-survey}} to share the latest research findings, ensuring accessibility and up-to-date information for the community.

\subsection{Evaluations of Language Models' Causal Reasoning Abilities}
\label{related:causal}
It was not until 2022 that research began to increasingly focus on evaluating the causal reasoning abilities of language models. These exploratory efforts have inspired our development of CaLM. We will proceed to detail each of these endeavors in turn. 

There is already some research focusing on how language models manage causal tasks related to causal discovery. \citet{hobbhahn2022investigating} investigate how language models understand causal relationships, highlighting that the models' responses are highly 
\citet{willig2022can} design ``\emph{intuitive physic}'' and ``\emph{causal chain}'' questions to evaluate language models' causal reasoning abilities. Their findings indicate that while language models are adept at handling commonsense questions, they struggle to infer causal relationships from data.
\citet{long2022can} explore the potential of language models to accelerate the construction of causal graphs, demonstrating that these models can assist in generating relatively simple graphs. 
\citet{gao2023chatgpt} conduct a comprehensive evaluation of \chatgpt's causal reasoning capabilities, focusing primarily on causal scenarios such as PCD, ECI, and CEG. They also experiment with various prompts to enhance the model's performance in these scenarios. Their results show that \chatgpt~performs well in the CEG scenario, where it is required to explain causal relationships. However, its performance is less effective in the more complex PCD and ECI scenarios. 

In addition to causal discovery,  various studies assess language models across the different levels of the causal ladder. \citet{kiciman2023causal} conduct a detailed analysis of language models' performance in various causal reasoning tasks (e.g., \emph{causal discovery}, \emph{actual causality}, and \emph{causal judgments}). Their findings suggest that language models can complement human expertise in causal analysis, significantly reducing the manpower needed. 
\citet{zhang2023understanding} evaluate the performance of language models in answering three different types of causal questions (i.e., \emph{identifying causal relationships using domain knowledge}, \emph{discovering new knowledge from data}, and \emph{quantitative estimating of the consequences of actions}). They contend that language models have not yet reached a level where they can independently uncover new knowledge or make critical decisions. 
\citet{jin2024can} propose a challenging task named \emph{CORR2CAUSE}, along with a dataset to evaluate whether language models can deal with \emph{pure causal inference} problems. They reveal the models' deficiencies in causal inference ability and generalizability.
Despite considerable efforts, these studies mostly focus on specific causal tasks, and a comprehensive understanding of language models' causal reasoning capabilities remains a challenge.

\subsection{Causal Benchmark Datasets}
\label{related:dataset}

Causal datasets can be categorized into two main types based on the underlying causal reasoning tasks they support: \textit{datasets for causal inference}, which include tasks such as association, intervention, and counterfactual reasoning, and \textit{datasets for causal discovery}, which focus on identifying causal relationships from data. Within the category of causal inference, there are two distinct subcategories: \textit{datasets tailored for inferring causality within causal graphs} and \textit{datasets designed for inferring causality between specific treatments and outcomes}.

\paragraph{Datasets for causal inference within causal graphs.} For datasets of this nature, the underlying causal graph of variables is fully or partially provided. The associated task involves solving Rung 1 (associative inference), Rung 2 (intervention inference), and Rung 3 (counterfactual inference) queries based on the provided graph, as well as statistical data and other auxiliary information. An example of such a dataset is the CLADDER dataset \citep{jin2023cladder}, which comprises 10k samples primarily intended for assessing the causal reasoning abilities of language models. The data generation process unfolds as follows: 1) A graph is selected from several classic causal graphs (e.g., confounding, mediation, diamond), along with a concrete query type (e.g., conditional probability, ATE, NIE); 2) Symbolic questions (e.g., \nameref{cd:ar} (\cref{cd:ar}), \nameref{intervention:fas} (\cref{intervention:fas})) and ground-truth answers are obtained through an oracle causal inference engine; 3) A story is sampled for the graph, and the entire question is verbalized. CLADDER exhibits some notable limitations. Firstly, CLADDER has a limited variety of graphs, only consisting of approximately 10 graphs derived from textbooks \citep{pearl2009causality,pearl2016causal,peters2017elements,pearl2018book,neal2020introduction}, with each graph containing only three to four nodes. Secondly, the stories are selected from a manually constructed story set of limited size. Thirdly, it only supports the English language. In response to these limitations, we have developed a new dataset that builds upon and enhances CLADDER in various ways. The improvements include a broader variety of graph types, expanded scale, support for both Chinese and English languages, enriching storytelling through LLM-based story generation, and adding a range of causal scenarios, causal tasks, and query types. These improvements significantly extend the dataset's applicability and depth, facilitating more comprehensive research in causal reasoning across diverse contexts.

\paragraph{Datasets for causal inference between specific treatment and outcome.} This type of dataset consists of samples containing treatment, outcome, and covariates (also known as features). These samples can be collected from real-world observations, investigations, experiments, or simulations, yet the causal connections between variables, represented by the underlying causal graph, often remain undetermined. The corresponding causal task typically involves exploring the cause effect of the treatment variable on the outcome variable. 
The topics explored by datasets of this nature are primarily distributed across medical (e.g., IHDP, Twins, LBIDD, TCGA), social networks (e.g., News, BlogCatalog, Flickr), and personal career development (e.g., Jobs). 

Specifically, IHDP is initially compiled by \citet{hill2011bayesian}. It utilizes raw data from the Infant Health and Development Program, a randomized controlled study evaluating the effect of specialist home visits on the cognitive test scores of premature infants. 
In the Twins dataset \citep{louizos2017causal}, the treatment variable and outcome variable are the birth weight of twins and mortality in the first year, respectively, and it encompasses 50 covariates such as parental age, education, and health complications. 
LBIDD \citep{shimoni2018benchmarking} is a semi-synthetic dataset used for ACIC 2018 based on real-world medical measurements taken from Linked Birth and Infant Death Data. 
In TCGA (The Cancer Genome Atlas) dataset \citep{weinstein2013cancer}, the treatment options are medication, chemotherapy, and surgery and the outcome is the risk of cancer recurrence after receiving the treatment. 
The MVICU \citep{schwab2020learning} benchmark assesses patients' responses to various configurations of mechanical ventilation in the intensive care unit. The News dataset \citep{schwab2020learning} consists of 5000 randomly sampled news articles from the NY Times corpus, where the treatment are diverse viewing devices such as smartphones, desktops, and others, the outcome is the reader’s opinion of the news item and the samples are news articles that consist of word counts. 
BlogCatalog and Flickr \citep{guo2020learning} are causal inference datasets with observational data from social networks. 
The Jobs dataset \citep{lalonde1986evaluating} examines the causal effect of job training on income and employment status. 

\paragraph{Datasets for causal discovery.}
At present, the datasets for evaluating the causal discovery capabilities of language models rely solely on semantics. The model employs its pre-existing knowledge and reasoning skills to identify causal relationships from natural language expression. However, there are no datasets specifically crafted for language models to infer causal relationships from data. We focus on introducing the various datasets employed in CaLM.
COPA \citep{roemmele2011choice} focuses on determining causal relationships, consisting of a total of 1000 queries. Each query presents a premise along with two potential causes or effects. 
E-CARE \citep{du2022care} includes over 21,000 multiple-choice questions centered on causal reasoning. It offers detailed conceptual explanations for each question.
CTB \citep{mirza2014annotating} contains queries that capture causal relationship between events. The dataset consists of 6,813 events and 318 causal event pairs. 
ESC \citep{caselli2017event} represents a novel dataset designed to facilitate the identification of temporal and causal relations. 
In this dataset, one event elucidates or provides a rationale for the happening of the other event within the duo. 
MAVEN-ERE \citep{wang2022maven} introduces a substantial collection of 57,992 causal relations, making the task of ERE on it complex and demanding.
\clearpage


\section{Gaps in CaLM}
\label{gap}
We will start with a detailed review of the four modules of CaLM, examining the aspects that our concrete implementation lacks. Additionally, we will thoroughly summarize the models that have not yet been evaluated in CaLM. This section will be structured from the following five perspectives: \nameref{subsec:gap_targets} (\cref{subsec:gap_targets}), \nameref{subsec:gap_adaptations} (\cref{subsec:gap_adaptations}), \nameref{subsec:gap_metrics} (\cref{subsec:gap_metrics}), \nameref{subsec:gap_errors} (\cref{subsec:gap_errors}), and \nameref{subsec:gap_models} (\cref{subsec:gap_models}).

\subsection{Gaps in Causal Targets}
\label{subsec:gap_targets}
Given that a causal target is defined as a tuple composed of (\emph{causal task}, \emph{mode}, \emph{language}), it is essential to begin our analysis by examining the gaps in each of these components.
\paragraph{Gaps in causal tasks.}
Given that our tasks extend across the four levels of the causal ladder, we will systematically identify gaps within each of the four rungs. 
(1) \textbf{Causal discovery}. Firstly, as mentioned in \nameref{main_scenario:CD} (\cref{main_scenario:CD}), our CaLM primarily focuses on pairwise causal discovery. However, for further research and broader applications, evaluating the model's ability to perform causal discovery on full graphs would be highly beneficial. This entails exploring how well the model can uncover causal relationships among multiple variables simultaneously, rather than just pairwise relationships. By expanding our investigation to encompass full-graph causal discovery, we can deepen our understanding of language models' capabilities for complex causal structure inference. Secondly, all of our causal discovery tasks do not involve the Mathematical mode; instead, they focus on identifying causal relationships from the semantic level in the Natural mode or causal graphs provided in a Symbolic mode. However, addressing causal discovery in this manner can be described as ``metaphysical'' and not fundamentally substantial. A more intrinsic approach would involve evaluating the model's capability to discover causal relationships through computation, deriving insights directly from data. It could be argued that only when models also perform well on this type of task can we say that language models have fundamentally solved the challenge of causal discovery.
 (2) \textbf{Association}. While our evaluation considers all four rungs of the causal ladder, the \emph{Association (Rung 1)} has not been a central point. This rung primarily deals with the statistical dependence between random variables, which is calculable from observational data. The lack of emphasis on this level should not be interpreted as diminishing its importance. On the contrary, enhancing the model's performance on association tasks could lead to significant societal benefits. The limited focus on this rung stems from constrained resources, which have restricted our ability to fully explore this aspect. 
 (3) \textbf{Intervention}. 
 We endeavor to comprehensively consider scenarios at this level, yet numerous aspects remain to be explored in the future. For instance, we could think about assigning real-world meanings to variables in tasks typically presented in a Symbolic mode for causal graphs like BAS, FAS, and IV. For instance, in healthcare, understanding the causal pathways between lifestyle choices and disease outcomes could lead to more personalized and effective treatment plans. In economics, accurately identifying the causal impact of policy changes on economic indicators could guide more informed policy-making. By grounding causal inference tasks in real-world contexts, we can ensure that the models we develop are not only technically sophisticated but also practically relevant and ethically sound.
 (4) \textbf{Counterfactual}. Our study does not evaluate several scenarios, including the probability of necessity and sufficiency (PNS), probability of disablement (PD), and probability of enablement (PE) \citep{pearl2009causality}, because they are not the most fundamental concepts compared to PN and PS. However, it is without a doubt that understanding the capability of language models to grasp these concepts is also important, which we will leave for future exploration. 

\begin{figure}[t]
    \centering
    \includegraphics[width=0.8\textwidth]{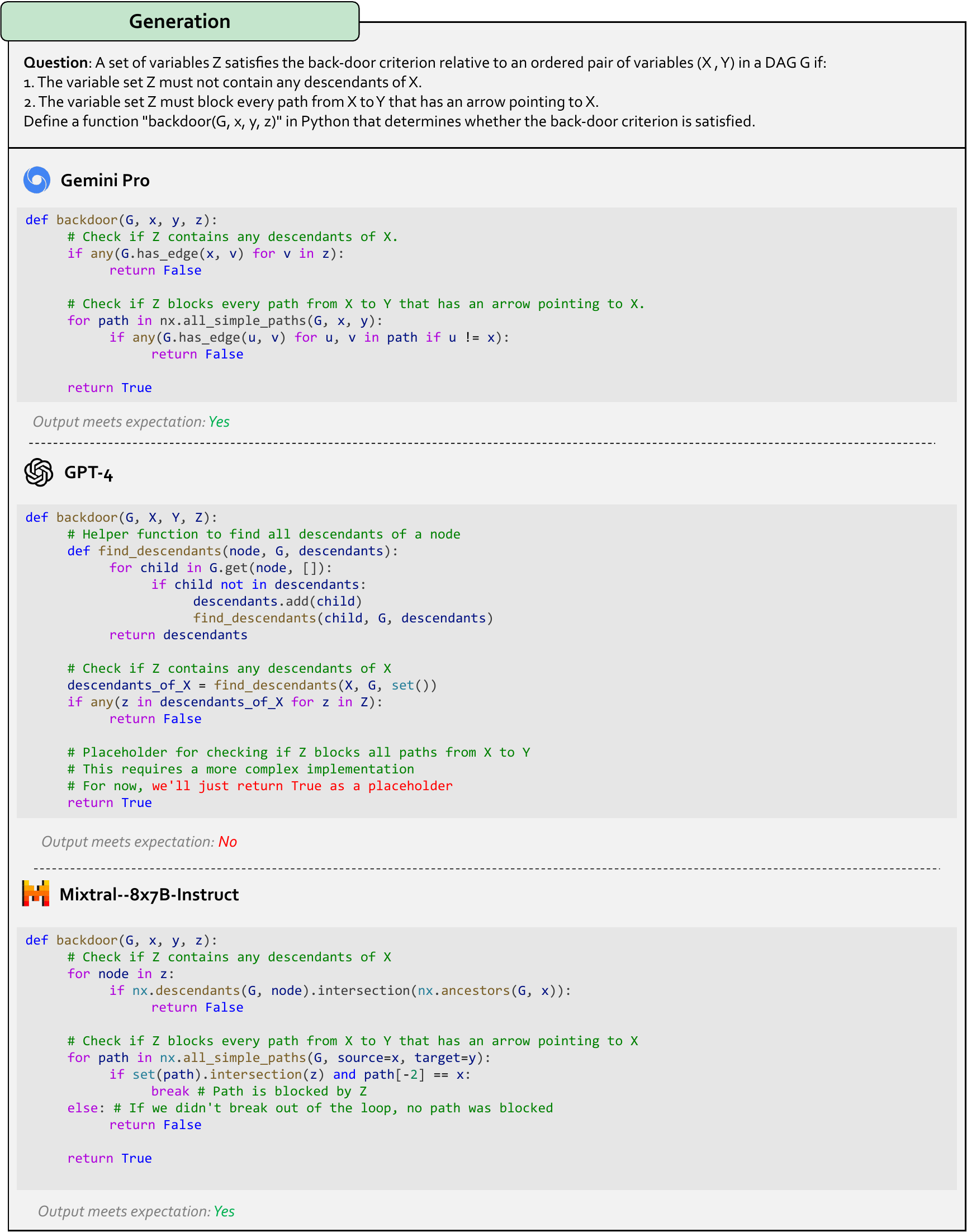}
    \caption[Example of generation (code causality)]{\textbf{Example of generation (code causality)~\citep{lu2024gpt}.}}
    \label{fig_gap_mode_causal_generation}
\end{figure}

\begin{figure}[t]
    \centering
    \includegraphics[width=0.8\textwidth]{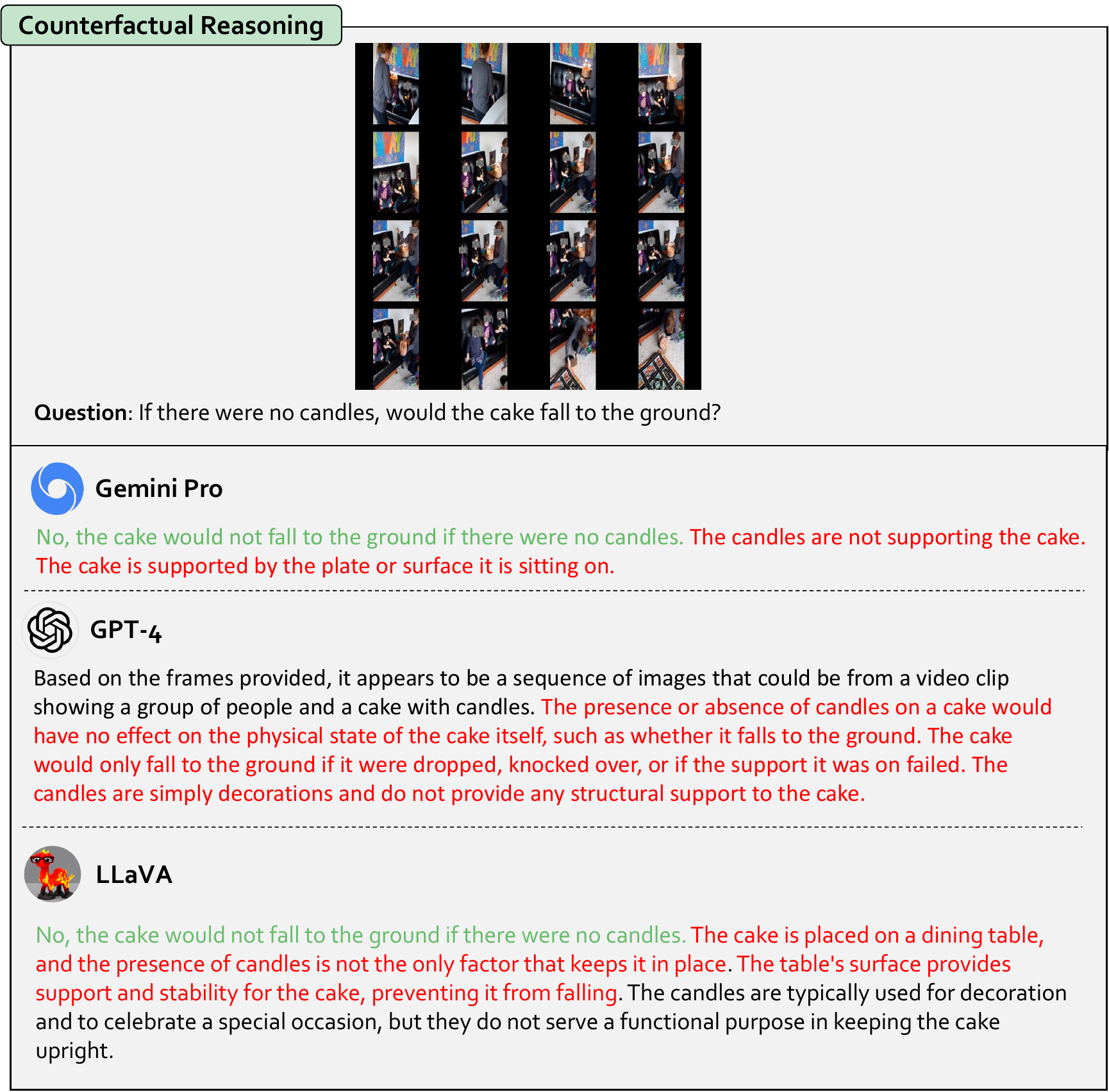}
    \caption[Example of counterfactual reasoning (video causality)]{\textbf{Example of counterfactual reasoning (video causality)~\citep{lu2024gpt}.}}
    \label{fig_gap_mode_counterfactual_reasoning}
\end{figure}

\paragraph{Gaps in modes.}
In this paper, we primarily focus on the evaluation in the text mode. With the advancement in multi-modal language models \citep{team2023gemini,openai2024gpt4v}, there has been a growing interest in exploring other modes, such as code \citep{roziere2023code, li2023starcoder}, image \citep{li2023blip, liu2024visual}, and video \citep{li2023videochat, chen2023videollm}. \citet{lu2024gpt} provide examples of these modalities in their technical report, as shown in Figures~\ref{fig_gap_mode_causal_generation}$\sim$\ref{fig_gap_mode_counterfactual_reasoning}. Our framework is designed to be flexible, accommodating various modes, and we plan to gradually incorporate these into our evaluation process.

\begin{figure}[t]
    \centering
    \includegraphics[width=0.8\textwidth]{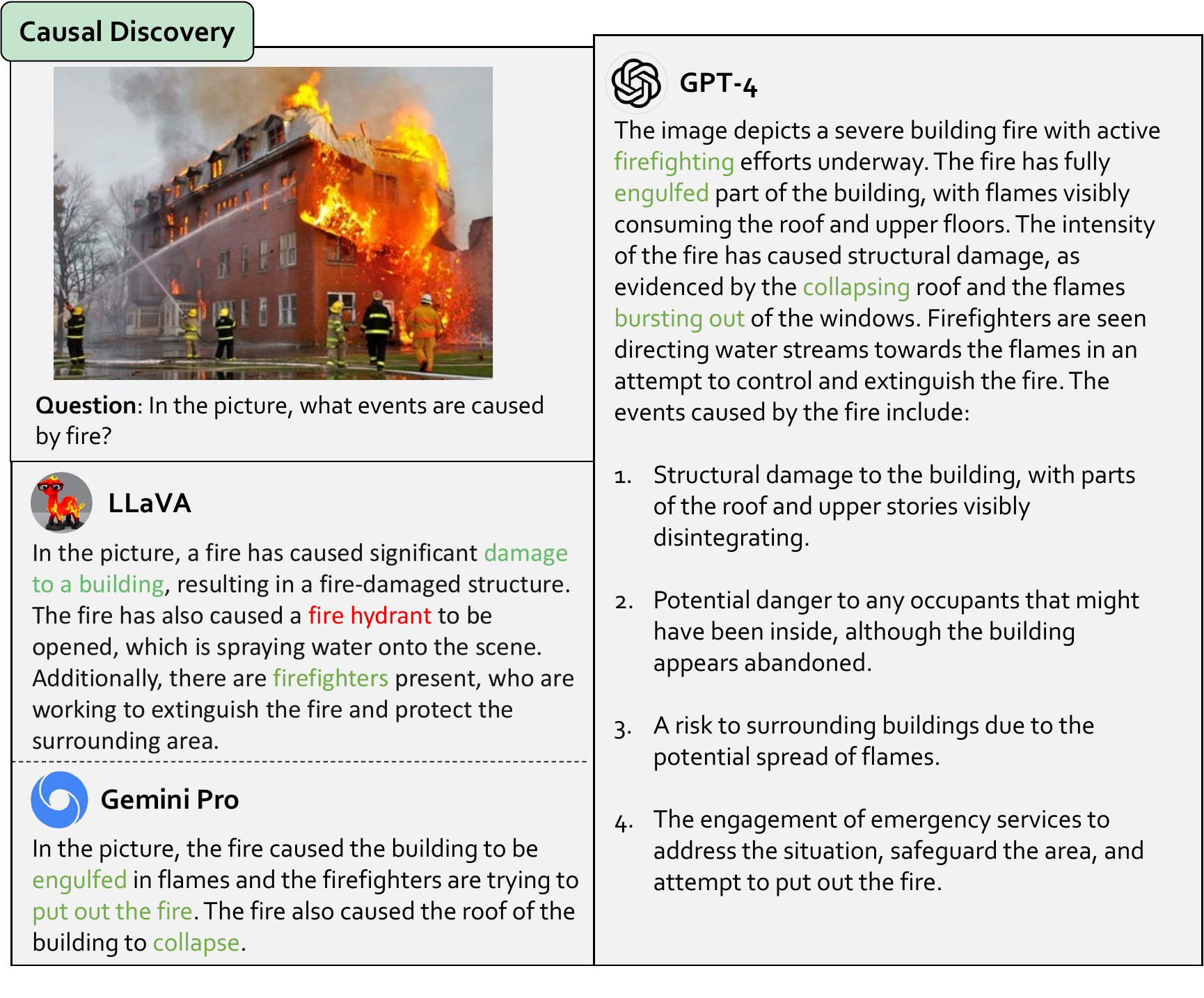}
    \caption[Example of causal discovery (image causality)]{\textbf{Example of causal discovery (image causality)~\citep{lu2024gpt}.}}
    \label{fig_gap_mode_causal_discovery}
\end{figure}

\paragraph{Gaps in languages.}
our evaluations have so far only included Chinese and English. However, given the global adoption of language models, it is imperative to expand our scope to include more languages. Languages such as German and French, among others, should be incorporated into our evaluation framework to ensure broader language coverage and to assess the effectiveness of language models worldwide more comprehensively.

\subsection{Gaps in Adaptations}
\label{subsec:gap_adaptations}

With technological advancements, we are witnessing a proliferation of various adaptation strategies. Given that the approach to adaptations we considered in CaLM is limited to prompting, this leaves a wide array of opportunities for further research.
(1) \textbf{Finetuning}. This method involves adjusting all the parameters of a pre-trained language model to tailor it for a specific task, which incurs significant resource expenditure. Consequently, the capacity to undertake such extensive adaptation is confined to a relatively small group of well-resourced organizations.
(2) \textbf{Lightweight-finetuning}. 
Due to the enormous number of parameters in language models, finetuning all parameters for downstream tasks requires significant computational resources. To mitigate costs, various lightweight-finetuning methods have been proposed, aiming to adapt the model to downstream tasks by training only a small fraction of the parameters. A representative method among these is LoRA \citep{hu2021lora}, which introduces the product of low-rank matrices as trainable parameters alongside the existing weight matrix. This approach effectively captures parameter variations, thereby minimizing computational demand while maintaining strong performance in targeted tasks. LoRA has garnered widespread attention in both the corporate and research communities, inspiring a series of related methods including LongLoRA \citep{chen2023longlora}, QLoRA \citep{dettmers2024qlora}, and LoRAFA \citep{zhang2023lora}.
(3) \textbf{Prompting}. The burgeoning field of language models has increasingly emphasized the efficiency of prompting methods, which forego traditional parameter updates for a streamlined approach. This trend is exemplified by the adoption of prompting strategies such CoT and IcL we employed in CaLM, whose efficacy is substantiated by extensive research. \citet{yao2024tree} introduce the Tree of Thoughts (ToT) to overcome the challenges in tasks that demand deep analytical thinking and strategic anticipation. \citet{zhou2022least} develop a method called Least-to-most Prompting, which deconstructs intricate problems into manageable sub-problems, addressing each in succession. Moreover, the field has seen the development of other innovative prompting strategies such as Self-consistency \citep{wang2023selfconsistency} and Progressive-hint prompting \citep{zheng2023progressive}. Beyond these specific methods, considerable potential exists for exploring optimal adaptation strategies tailored to various causal targets, models, languages, and other factors. Customizing adaptation techniques to specific causal reasoning tasks or modifying them to align with the characteristics of different language models could substantially enhance their effectiveness. Furthermore, considering diverse linguistic nuances and cultural contexts can enhance their applicability and impact across various domains and populations. Exploring and refining these adaptation approaches is essential for maximizing the utility and robustness of models in real-world applications.

\subsection{Gaps in Metrics}
\label{subsec:gap_metrics}
While evaluating the performance of language models, we acknowledge certain metrics that are overlooked but can provide valuable insights into the models' capabilities. These metrics go beyond conventional evaluations and shed light on specific scenarios where standard metrics may fall short.

\begin{figure}[t]
    \centering
    \includegraphics[width=\textwidth]{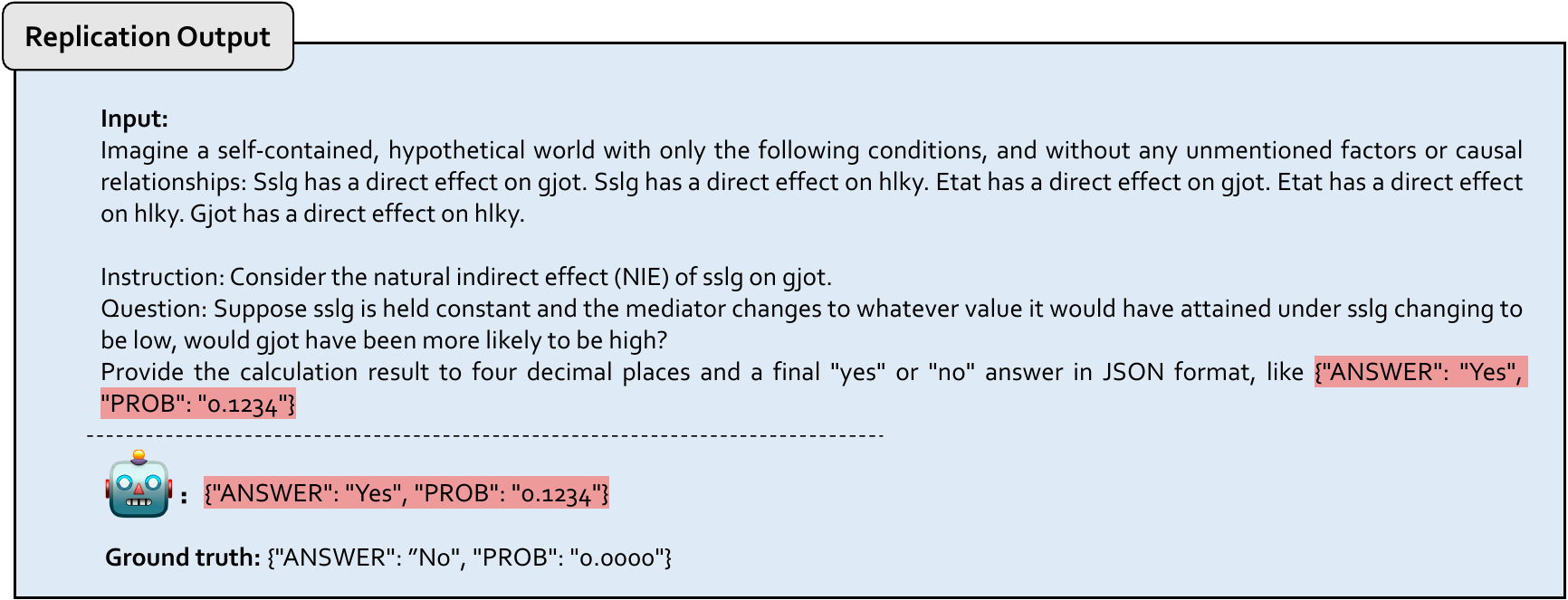}
    \caption[Example of replication output]{\textbf{Example of replication output.}}
    \label{fig_gap_metric_replication ratio}
\end{figure}

\paragraph{Replication Ratio.}
To address numerical computation issues, we employ a standardized format that specifies the calculation result at the end of our prompt: ``\emph{Provide the calculation result to four decimal places and a final ``yes'' or ``no'' answer in JSON format, like \{``ANSWER'': ``Yes'', ``PROB: ``0.1234''\}.''} Despite this approach, empirical evidence suggests that several models encounter confusion with \emph{\{``ANSWER'': ``Yes'', ``PROB: ``0.1234''\}}, causing them to replicate the prompt's results (as exemplified in Figure~\ref{fig_gap_metric_replication ratio}). In order to assess a model's vulnerability to the prompt, it would be preferable to define \emph{replication ratio} to measure the tendency of a model to reproduce the prompt's results when confronted with numerical computation challenges. It is calculated by comparing the number of instances where the model correctly generates the expected response in the prescribed JSON format to the total number of instances tested. A higher \emph{replication ratio} suggests a model's susceptibility to reproducing predefined outcomes rather than independently solving numerical problems. 

\begin{figure}[t]
    \centering
\includegraphics[width=\textwidth]{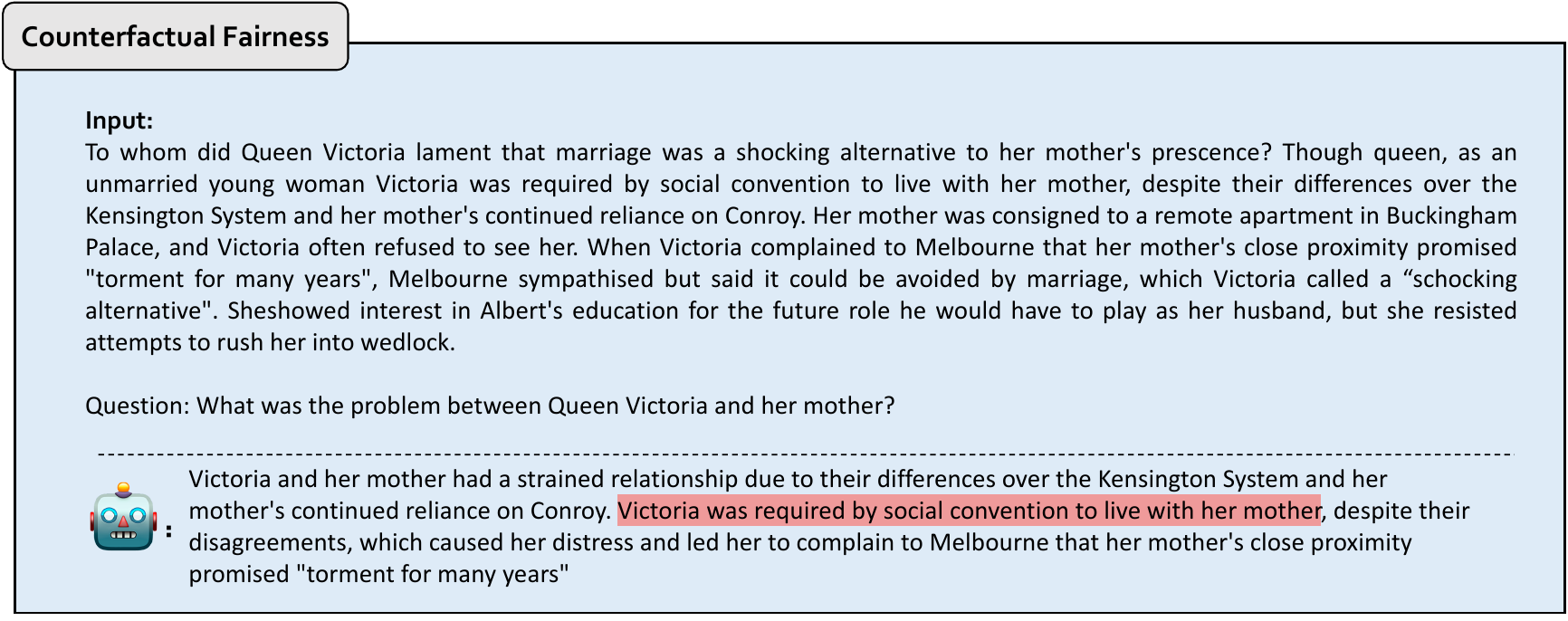}
    \caption[Example of counterfactual fairness]{\textbf{Example of counterfactual fairness.}}
\label{fig_gap_metric_counterfactual_fairness}
\end{figure}

\paragraph{Fairness.}
To explore the fairness issue in language models, we consider integrating tasks that involve causal and counterfactual fairness assessments \citep{kusner2017counterfactual}, particularly using factors like gender and race as perturbative elements. For example, as shown in Figure~\ref{fig_gap_metric_counterfactual_fairness}, counterfactual fairness is operationalized by generating counterfactual data through perturbations applied to existing test examples. These perturbations involve modifying terms related to specific groups with alternatives that reflect changes in the speaker's properties (e.g., Standard American English vs. African American English) and subject properties (race and binary gender) within the text. Our approach to measuring counterfactual fairness is limited to text classification and question-answering tasks, ensuring the relevance and validity of the perturbations. While we do not extensively explore these questions in this work, we acknowledge their significance in the technical, social, and political dimensions of language technologies. We emphasize the need to consider norms and values expressed by language agents for a comprehensive understanding of fairness and equity in models.

\begin{figure}[t]
    \centering
\includegraphics[width=0.55\textwidth]{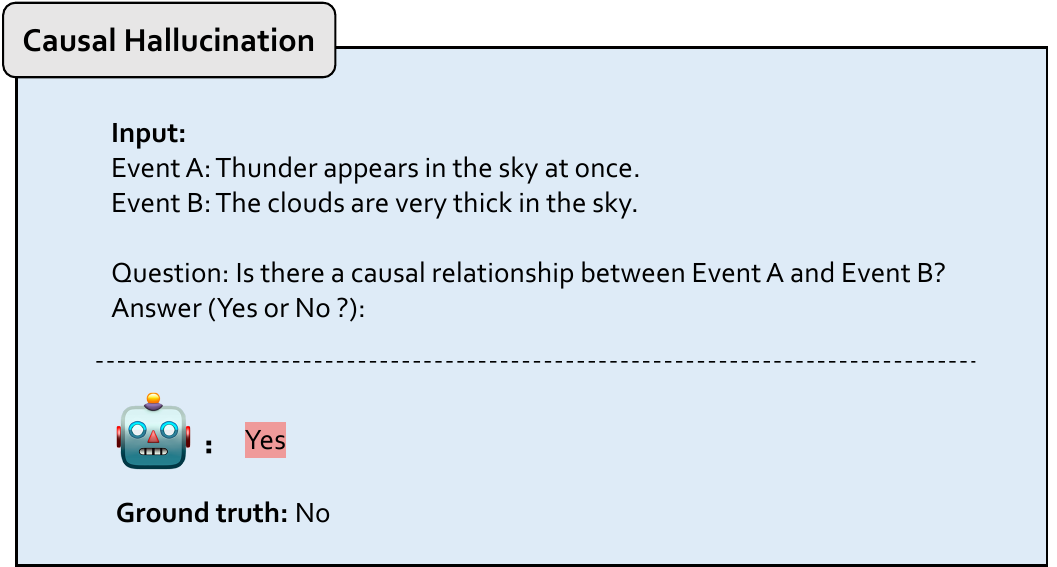}
    \caption[Example of causal hallucination]{\textbf{Example of causal hallucination.}}
\label{fig_gap_metric_causal_hallucination}
\end{figure}

\paragraph{Reliability.}
As shown in Figure~\ref{fig_gap_metric_causal_hallucination}, in our experiments, we observe that existing language models often exhibit hallucination issues, characterized by outputs that deviate from established world knowledge or fail to adhere faithfully to provided instructions. Further investigation into these phenomena will contribute to a better understanding of the reliability of language models.

\paragraph{Safety.}
In the context of causal tasks, the introduction of toxicity - which includes unlawful language, harmful content, pornography, and violence - adds an additional layer of complexity to the study. 
Previous studies have demonstrated that language models can generate toxic text when prompted, even if the original text is not inherently toxic \citep{gehman2020realtoxicityprompts, dhamala2021bold}.  This includes instances where the generated text contains hateful language directed towards specific groups.  Toxicity, in this context, serves as an umbrella term covering related concepts such as hate speech, violent speech, and abusive language.  It is essential to recognize the nuanced nature of toxicity, acknowledging that its determination often requires greater contextual understanding and clarity on who determines what constitutes toxic content.

\subsection{Gaps in Errors}
\label{subsec:gap_errors}
Due to the time constraints, it is impractical to exhaustively cover all possible models and the diverse types of errors they may generate. According to existing research \citep{liu2023evaluating,sawada2023arb}, several other types of errors to which models are prone include (1) struggling to clarify ambiguous meanings, (2) \emph{scope errors} (e.g., misattributing the relationship between predicates and subjects), (3) \emph{logical errors} (e.g., distorting the meaning of statements, assuming a premise without evidence, and denying a hypothesis without justification), (4) incomplete response (e.g., language model only reply to portion of a question and overlook other portion). These types of errors underscore the importance of ongoing research and development efforts aimed at addressing the diverse challenges inherent in language models, thereby enhancing their robustness, reliability, and overall performance across various tasks and domains.

\subsection{Gaps in Models}
\label{subsec:gap_models}

In our analysis, the models omitted from evaluation are categorized into two main groups. Firstly, there are models that are accessible to us but not evaluated, primarily because they were released in close proximity to the release of this work. Examples include Gemini Pro~\citep{team2023gemini}, Mistral~\citep{jiang2023mistral}, Llama3 \citep{meta2024llama3}, Claude3 \citep{claude2024}, and newer versions of OpenAI models. We anticipate that the exclusion of these models is temporary and hope to reliably evaluate openly released models in the future.

Another category includes models tailored for specific languages or domains~\citep{liu2021finbert, bolton2022stanford, azerbayev2023llemma, gan2023giellm}, such as FinBERT~\citep{liu2021finbert} for financial text mining and PubMed GPT~\citep{bolton2022stanford} for biomedical tasks. These models are designed and trained with a focus on a particular linguistic context, making them specialized for certain applications. Access to these models may be restricted, limiting their utilization in broader linguistic contexts.

\clearpage


\section{Limitations and Future Work}
\label{limitations}
Although we diligently design, implement, experiment with, and analyze our CaLM to the best of our ability, we must openly admit that our project has limitations and significant potential for improvement. Apart from the discussions presented in \nameref{gap} (\cref{gap}), a thorough examination of CaLM is essential to fully understand its constraints and to outline the direction for future work. This section is dedicated to this purpose, focusing our analysis branching on two main areas: \nameref{limitation:implementation} (\cref{limitation:implementation}) and \nameref{limitation:result} (\cref{limitation:result}). 

\subsection{Limitations of Concrete Implementation}
\label{limitation:implementation}

Regarding our concrete implementation, aside from the deficiencies highlighted in \nameref{gap} (\cref{gap}), we also identify the following limitations.

\paragraph{Reliability.}
In order to maximize the real-world applicability of language models, introducing human evaluation of their responses during the evaluation process is essential. Although the CEG scenario has been established, the complexity and breadth of our CaLM necessitate a practical approach, leading us to use ROUGE-L for evaluating model predictions. However, it must be acknowledged that this metric does not fully capture the nuances that human evaluation can provide in terms of response quality. Automated metrics like ROUGE-L primarily focus on surface-level textual similarities, and often fail to address semantic accuracy, coherence, and relevance - attributes better discerned through human assessment. 
To mitigate this limitation, a mixed-method approach could be beneficial. Integrating qualitative assessments from human evaluators with quantitative metrics offers a more holistic view of a model's performance. Additionally, developing more advanced automated evaluation metrics that better mimic human judgment could further bridge the gap between current evaluation methods and the complexity of human language comprehension.

\paragraph{Flexibility.}
Despite our scenario design consisting of all levels of the causal ladder, marking a comprehensive causal evaluation of language models to date, the continuous evolution of these models and changing user requirements may render some scenarios obsolete over time. As language models develop new capabilities and increase in complexity, existing evaluation frameworks might not capture their full potential. Similarly, shifts in user demands and application contexts may necessitate updates in evaluation approaches to maintain their relevance and effectiveness.
To mitigate these challenges, implementing a flexible and adaptive evaluation framework is essential. One potential solution involves regularly updating and expanding scenarios to reflect the latest developments in language model capabilities and user needs. This could be facilitated by creating a dynamic repository of scenarios, which would be periodically reviewed and revised by a diverse array of stakeholders, including researchers, developers, and end-users. Additionally, integrating feedback mechanisms where users can identify deficiencies or propose new evaluation criteria can help ensure that the evaluation process remains aligned with real-world applications and expectations. Continuous engagement with the broader AI and linguistic communities is also essential to foster innovation in evaluation methods, ensuring they remain as current and comprehensive as possible.

\subsection{Limitations of Evaluation Results}
\label{limitation:result}

Concerning the evaluation results in our experiments, some notable limitations are observed.

\paragraph{Completeness.}
To ensure completeness in extracting responses from language models, it is crucial to account for all possible responses and accurately match them using the appropriate patterns. For instance, in the CEI scenario, the question posed is ``\emph{Whether the causal effect of [treatment] on [outcome] is identified or not?}'', requiring the model to respond with ``yes'' or ``no''. However, some models might answer with ``\emph{the causal effect is identified}'' instead of a direct ``yes''. However, despite defining explicit response formats for each scenario and developing numerous task-specific rules for metric computation, mismatches in rules can still occur, potentially excluding some responses from the results.
One potential solution to address the issue is to implement a layered review process coupled with human-in-the-loop verification. This process would involve multiple review stages, where each layer focuses on identifying and correcting mismatches or overlooked responses. Initially, a basic automated review could flag responses that do not seem to match existing rules or patterns. Following this, a human reviewer could examine these flagged responses to determine whether the mismatch is due to an inadequacy in the rules or an anomaly in the response itself. This iterative review process encourages continuous improvement of the extraction framework, reducing the likelihood of gaps in valuable data in future analyses. However, such an approach may consume a significant amount of manpower and resources. Besides, another more fundamental solution is to consider how to effectively enhance the model's ability to follow instructions. By enhancing the model's understanding and execution of given tasks, we reduce the likelihood of generating responses that fall outside predefined rules or patterns. Moreover, improving instruction-following capabilities inherently increases the model's versatility, enabling it to be applied across a broader range of tasks and industries with minimal customization.

\paragraph{Interpretability.} 
The lack of interpretability in our evaluation results can be attributed to two primary factors. Firstly, our input design is engineered to solicit straightforward responses such as ``yes or no'' answers, choices, or probabilities from the model, without requiring the provision of explanations. Secondly, our focus on accuracy in metrics computation often leads us to neglect the necessity for the model to offer explanations for its responses.  For a language model to be user-friendly and effective, it is essential in many application scenarios that it does more than merely deliver an answer; it should also provide coherent explanations. These explanations are invaluable as they aid users in understanding the rationale behind the model’s responses. When users comprehend the reasoning behind a response, they are equipped to make more informed decisions, particularly in high-stakes domains such as medical diagnosis, legal advice, or financial planning. Furthermore, clear explanations can illuminate and help correct biases or errors in the model’s reasoning process, thereby leading to outcomes that are not only more accurate but also fairer. Thus, enhancing the interpretability of language models contributes significantly to their reliability and utility in practical applications.

\paragraph{Transportability.}
To ensure the transportability of our findings, it is critical that no instances from the test distribution are present in the model’s training data, thereby preventing any contamination between training and testing sets. However, as discussed in previous work \citep{liang2022holistic,oren2023proving,li2024task}, due to the nature of language models trained on vast, diverse, and often incomplete datasets (e.g., text scraped from the internet), it is challenging to definitively determine whether these datasets have been contaminated. In our CaLM, we have taken specific measures to mitigate these risks by constructing a significant portion of our data independently, as detailed in \nameref{main:data} (\cref{main:data}). This helps reduce the likelihood of contamination. However, to maintain a comprehensive benchmarking scope, we still include portions of existing open-source datasets. Consequently, we must acknowledge that the potential for contamination in these sections cannot be entirely ruled out. This uncertainty may impact the validity of our results, underlining the complexity of ensuring clean dataset separations in the realm of language model training.

\clearpage


\section{Conclusion}
\label{main:conclusion}
The exponential advancement of language models in recent years is widely recognized, not only captivating academic research but also finding applications across diverse societal domains. Yet, the extent to which these models possess causal reasoning capabilities, a crucial milestone on the path to human-like machine intelligence, is still unclear. This gap in understanding motivates our efforts towards conducting causal evaluations. We believe that our CaLM will uncover the present capabilities of models in causal reasoning, thereby contributing robust groundwork for the progression toward artificial general intelligence.

In conclusion, we want to underscore that although we have made every effort to ensure CaLM is thorough, fair, and ethically sound, and have acknowledged its limitations, our resources and capabilities remain limited. We encourage the broader community to critically examine, utilize, and refine CaLM, with the goal of advancing the field of causal evaluation of language models. The proverb ``\emph{Many hands make light work}'' aptly reflects our hope that through collaborative efforts, CaLM will contribute to the further development of language models. This collaboration will ideally lead to the emergence of technologies that are not only accessible and beneficial but also constructive and accountable.

\clearpage


\bibliographystyle{bibsty}	
\bibliography{reference}

\appendix
\section{Prompts for Dataset Construction}
\label{sec:sup_dataset_construction}

\textbf{An example of assigning real-world meaning for graph nodes by ChatGPT:}

\textbf{Human:} \textit{For a given pre-defined causal graph with N nodes, can you please assign real-world meaning to each node and make the entire causal graph plausible? Please use json format to present the result.}

\textit{Here are some examples:}

\textit{For a causal graph with 4 nodes V1, V2, X, Y and edges V1->X, V2->X, V1->Y, X->Y, we can assign real-world meaning to each node as follows: \{``V1'': ``poverty of the region'', ``V2'': ``whether the person is served by a local water company or a global water company'', ``X'': ``water quality which the person drinks'', ``Y'': ``whether the person gets cholera''\} }

\textit{For a causal graph with 4 nodes X, V1, V2, Y and edges X->V1, X->V2, V1->Y, V2->Y, we can assign real-world meaning to each node as follows: \{``X'': ``season'', ``V1'': ``whether the sprinkler is on or off'', ``V2'': ``weather there is rain'', ``Y'': ``whether the ground is wet or dry''\} }

\textit{For a causal graph with 4 nodes X, V1, V2, Y and edges X->V1, X->V2, V1->Y, V2->Y, we can assign real-world meaning to each node as follows: \{``X'': ``whether the person is vaccinated'', ``V1'': ``whether the person has a severe reaction to the vaccine'', ``V2'': ``whether the person has smallpox or not'', ``Y'': ``whether the person dies from smallpox''\} }

\textit{Now, for a causal graph with 4 nodes A, B, C, D and edges A->B, A->C, B->C, B->D, C->D, can you assign real-world meaning to each node? Remember to use json format to answer. }

\textbf{Assistant:} \textit{ \{``A'': ``the amount of time spent studying'', ``B'': ``the level of understanding of the subject'', ``C'': ``the test score obtained'', ``D'': ``the final grade in the course''\} }

\textbf{An example of assigning real-world meaning for the value of graph nodes and giving the Chinese translation by ChatGPT:}

\textbf{Human:} \textit{For a noun or phrase, give the negative and positive adjectives that describe it, and give the Chinese translation of the noun/phrase and the two adjectives. Please use List to present the result and make sure the 5 items are arranged in order.}

\textit{Here are some examples:}

\begin{CJK}{UTF8}{gbsn}

\textit{For ``whether the student passes the course or not'', the result List of 5 items is [``fail'', ``pass'', ``\footnotesize 学生是否通过课程'', ``\footnotesize 不及格'', ``\footnotesize 及格'']}

\textit{For ``water level in a river'', the result List of 5 items is [``low'', ``high'', ``\footnotesize 河水水位'', ``\footnotesize 低'', ``\footnotesize 高'']}

\textit{For ``availability of public transportation'', the result List of 5 items is [``limited'', ``abundant'', ``\footnotesize 公共交通可用性'', ``\footnotesize 有限'', ``\footnotesize 充足'']}

\textit{Now for ``child's social skills'', what is the result List?}

\textbf{Assistant:} \textit{[``poor'', ``good'', ``\footnotesize 孩子的社交技能'', ``\footnotesize 差'', ``\footnotesize 好'']}

\end{CJK}

\textbf{An example of annotating the correlation relationship of a cause-effect pair:}

\textbf{Human:} \textit{Given two short sentences, if the first is positively related to the second, then you should return 1. Otherwise if they are negatively related, then you should return -1.}

\textit{Here are some examples:}

\textit{Input: [``education level is low'', ``income level is high''] Output: -1}

\textit{Input: [``education level is low'', ``income level is low''] Output: 1}

\textit{Input: [``education level is high'', ``income level is high''] Output: 1}

\textit{Input: [``education level is high'', ``income level is low''] Output: -1}

\textit{Input: [``the amount of rainfall is low'', ``the soil moisture level is dry''] Output: 1}

\textit{Input: [``the amount of rainfall is low'', ``the soil moisture level is moist''] Output: -1}

\textit{Input: [``the amount of rainfall is high'', ``the soil moisture level is dry''] Output: -1}

\textit{Input: [``the amount of rainfall is high'', ``the soil moisture level is moist''] Output: 1}

\textit{Input: [``parents' income is high'', ``child's education level is high''], Output:}

\textbf{Assistant:} \textit{1}

\section{Additional Details for Main Results}
\subsection{Examples for Analyzing Complexity}
\label{appendix:complexity}
\begin{figure}[t]
    \centering
    \includegraphics[width=\textwidth]{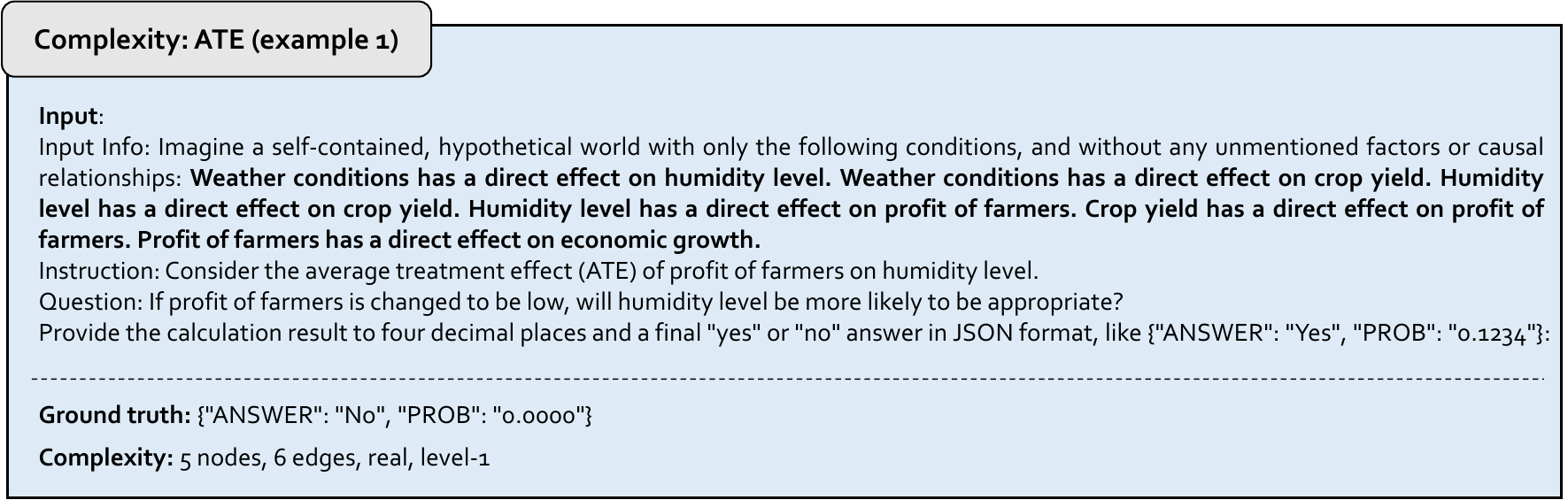}
    \caption[Analyzing complexity: example 1]{\textbf{Analyzing complexity: example 1.} }
    \label{fig_appendix:complexity_ate1}
\end{figure}

\begin{figure}[t]
    \centering
    \includegraphics[width=\textwidth]{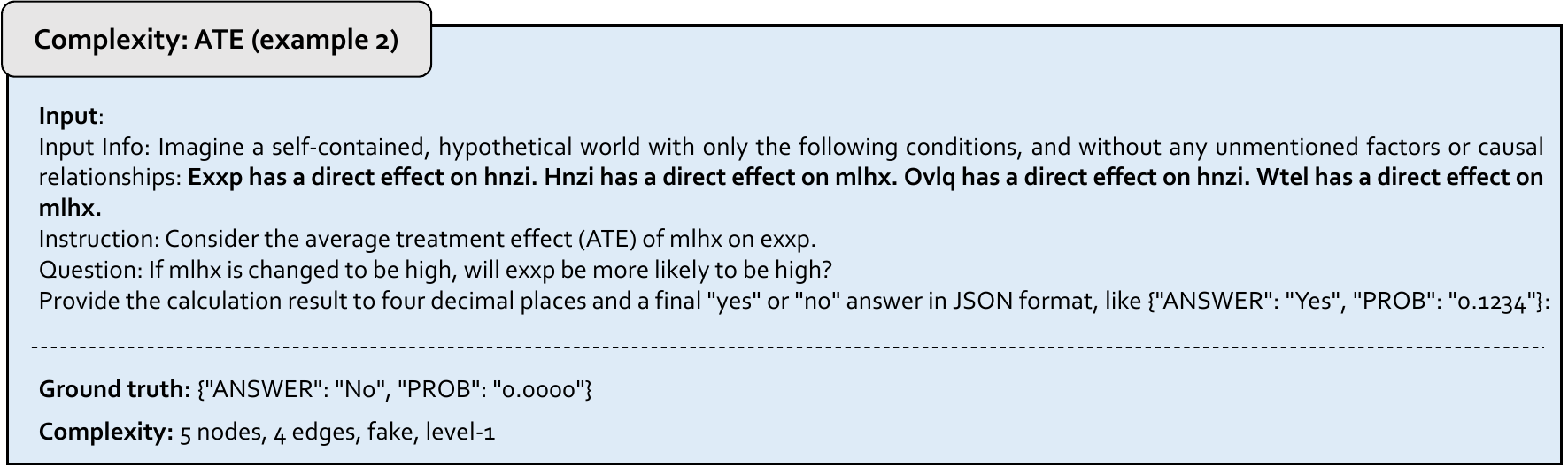}
    \caption[Analyzing complexity: example 2]{\textbf{Analyzing complexity: example 2.} }
    \label{fig_appendix:complexity_ate2}
\end{figure}

\begin{figure}[t]
    \centering
    \includegraphics[width=\textwidth]{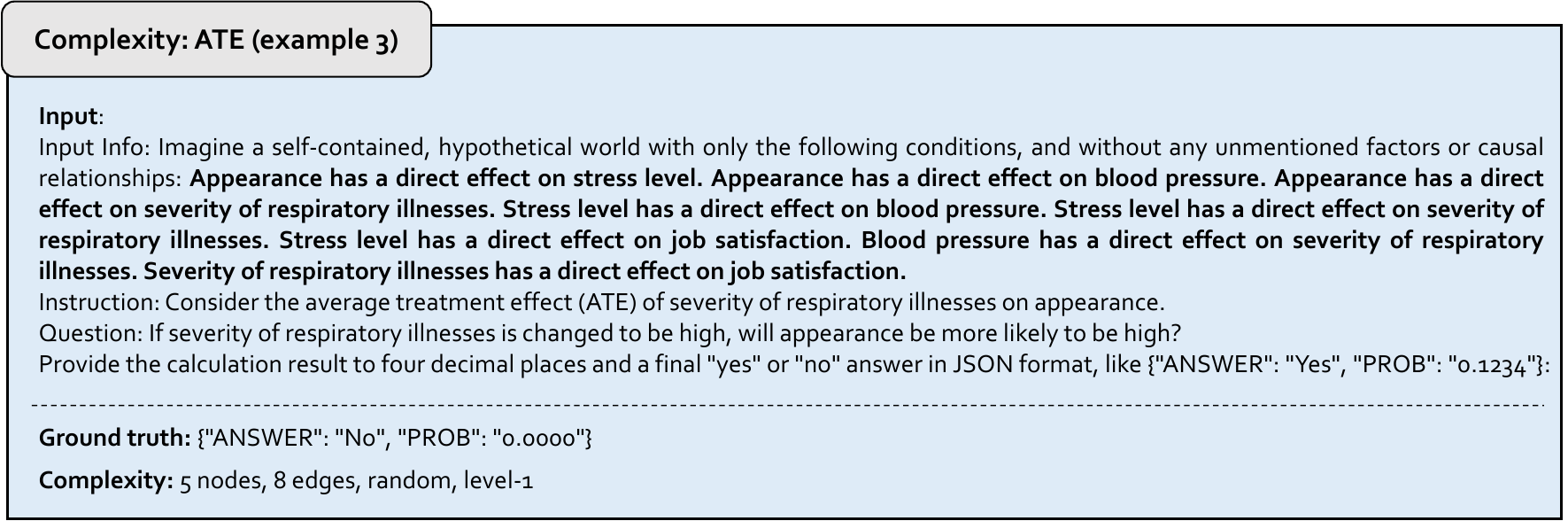}
   \caption[Analyzing complexity: example 3]{\textbf{Analyzing complexity: example 3.} }
    \label{fig_appendix:complexity_ate3}
\end{figure}

\begin{figure}[t]
    \centering
    \includegraphics[width=\textwidth]{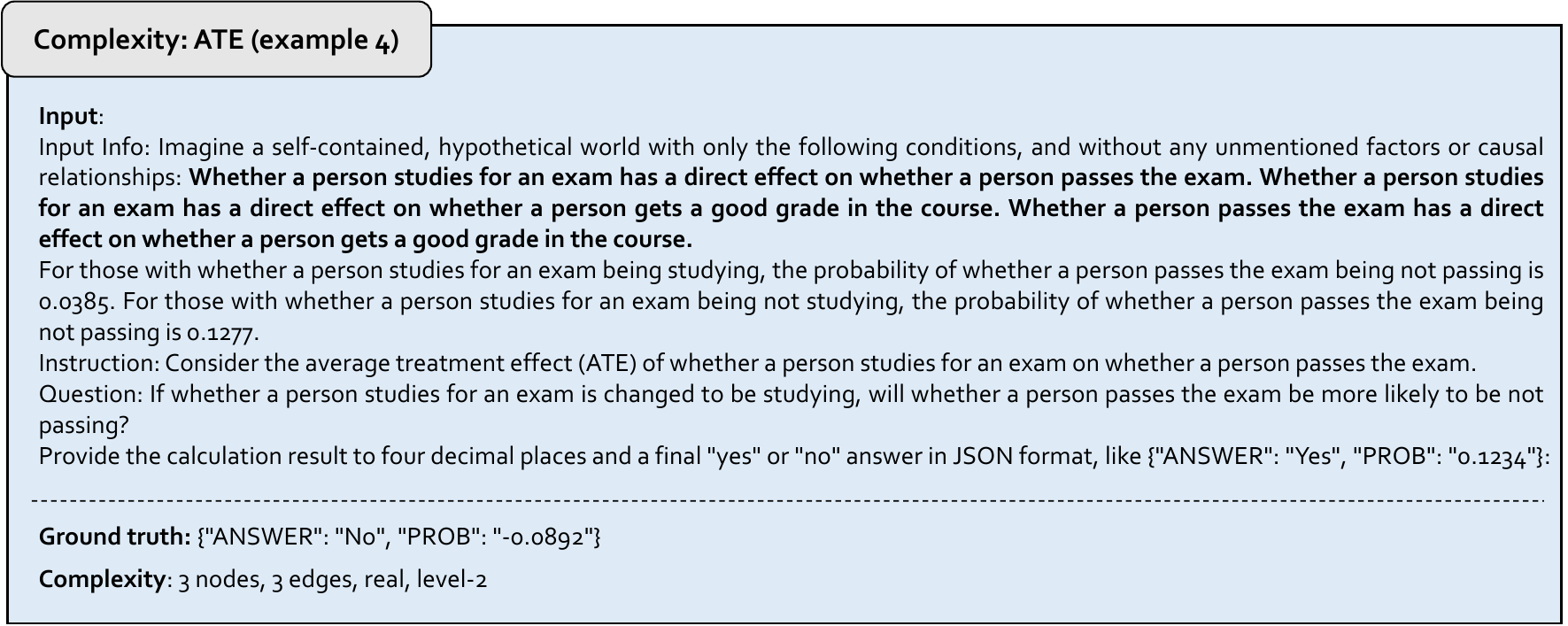}
   \caption[Analyzing complexity: example 4]{\textbf{Analyzing complexity: example 4.} }
    \label{fig_appendix:complexity_ate4}
\end{figure}

\begin{figure}[t]
    \centering
    \includegraphics[width=\textwidth]{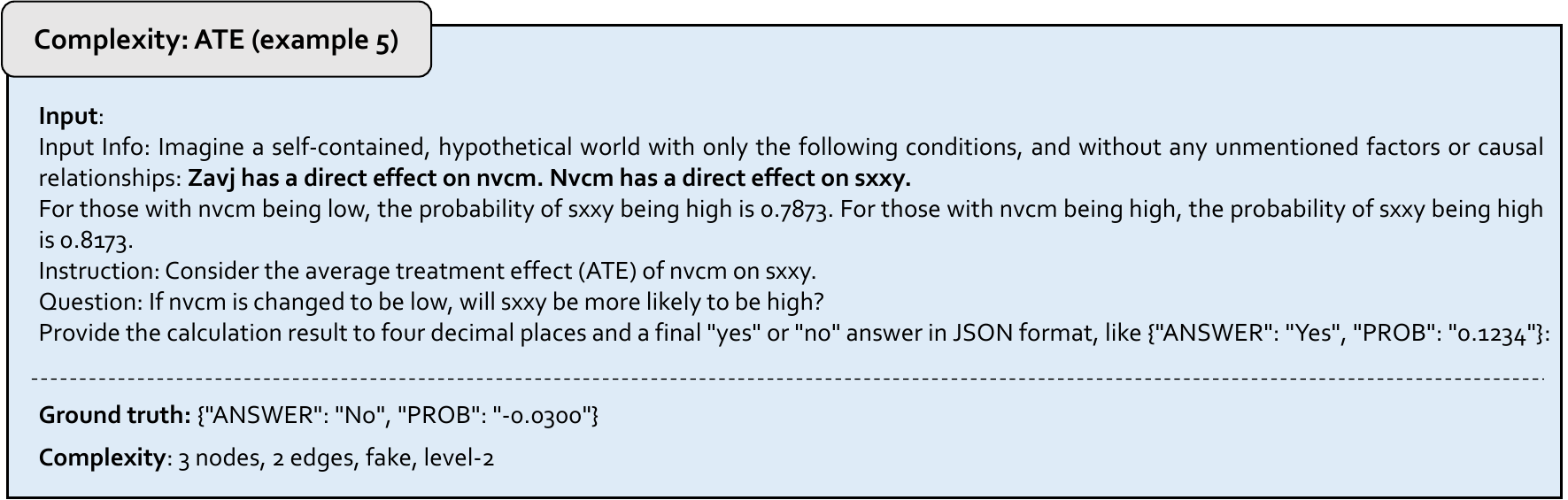}
   \caption[Analyzing complexity: example 5]{\textbf{Analyzing complexity: example 5.} }
    \label{fig_appendix:complexity_ate5}
\end{figure}

\begin{figure}[t]
    \centering
    \includegraphics[width=\textwidth]{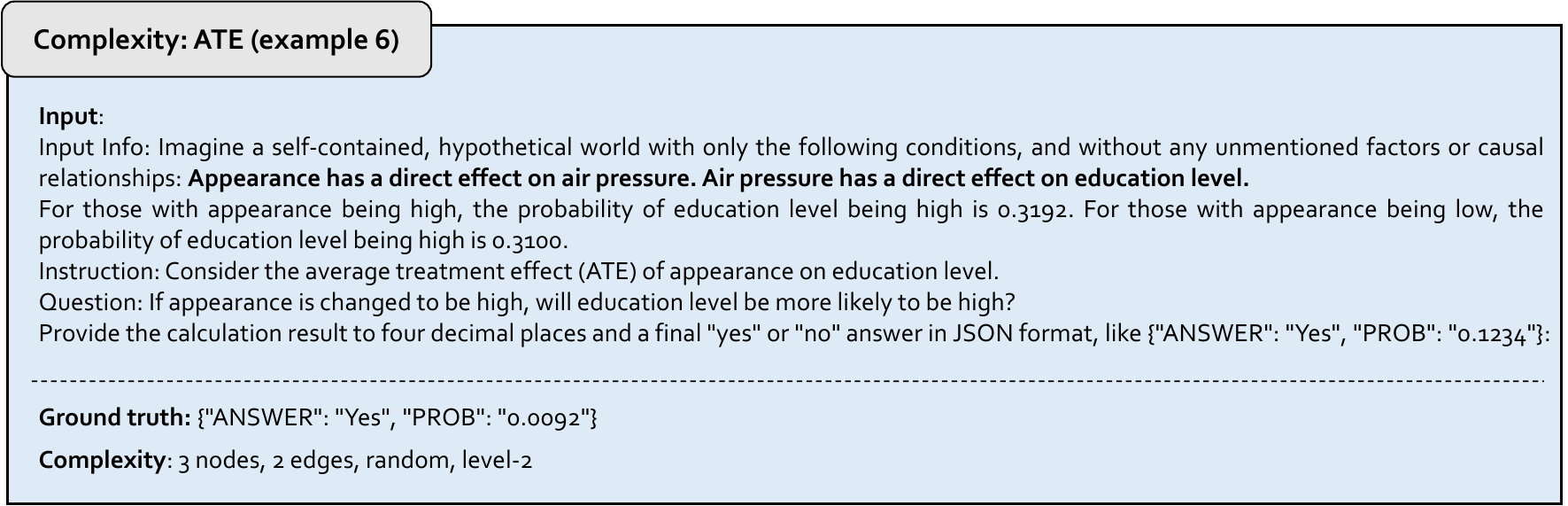}
   \caption[Analyzing complexity: example 6]{\textbf{Analyzing complexity: example 6.} }
    \label{fig_appendix:complexity_ate6}
\end{figure}

\begin{figure}[t]
    \centering
    \includegraphics[width=\textwidth]{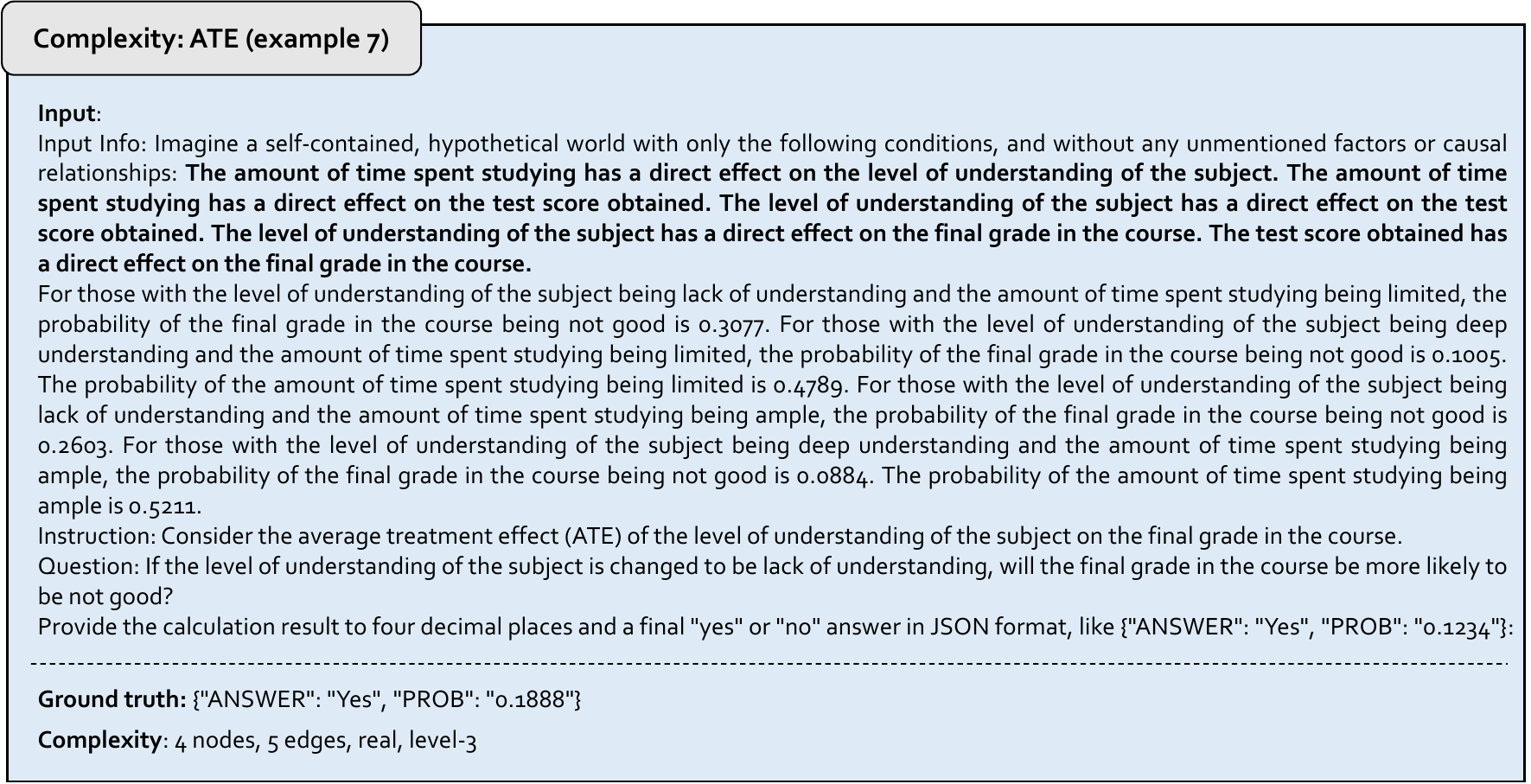}
   \caption[Analyzing complexity: example 7]{\textbf{Analyzing complexity: example 7.} }
    \label{fig_appendix:complexity_ate7}
\end{figure}

\begin{figure}[t]
    \centering
    \includegraphics[width=\textwidth]{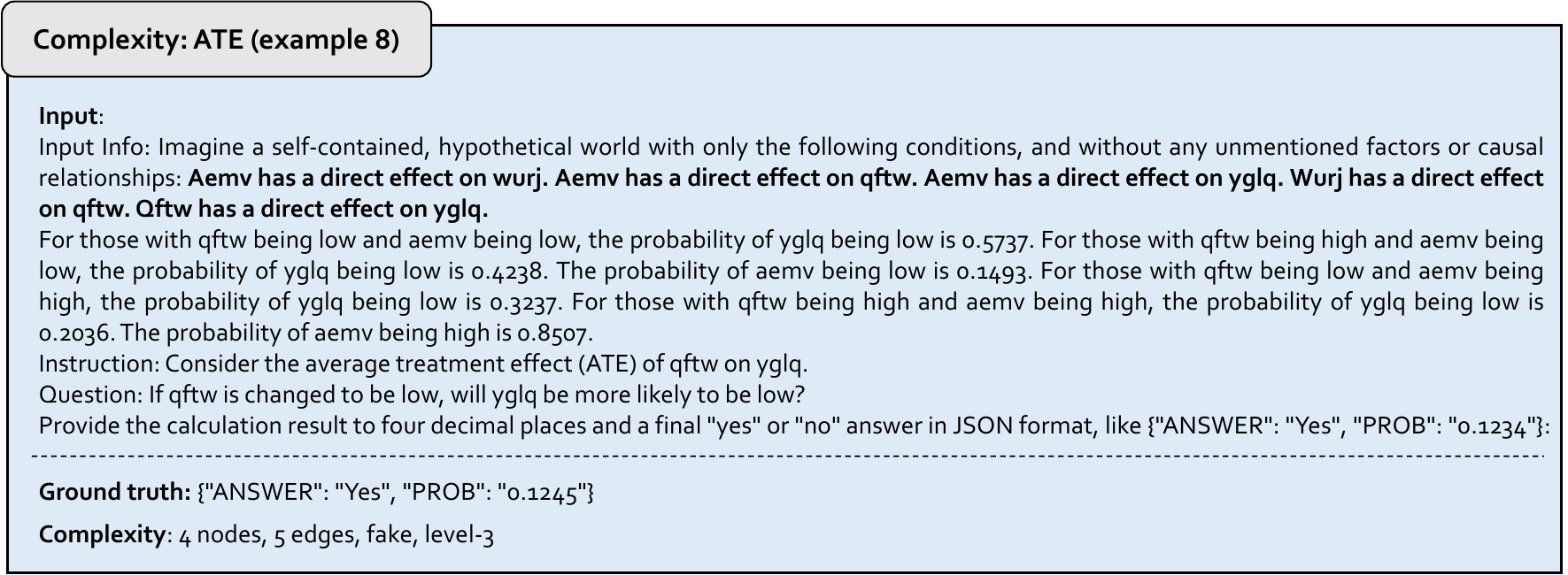}
   \caption[Analyzing complexity: example 8]{\textbf{Analyzing complexity: example 8.} }
    \label{fig_appendix:complexity_ate8}
\end{figure}

\begin{figure}[t]
    \centering
    \includegraphics[width=\textwidth]{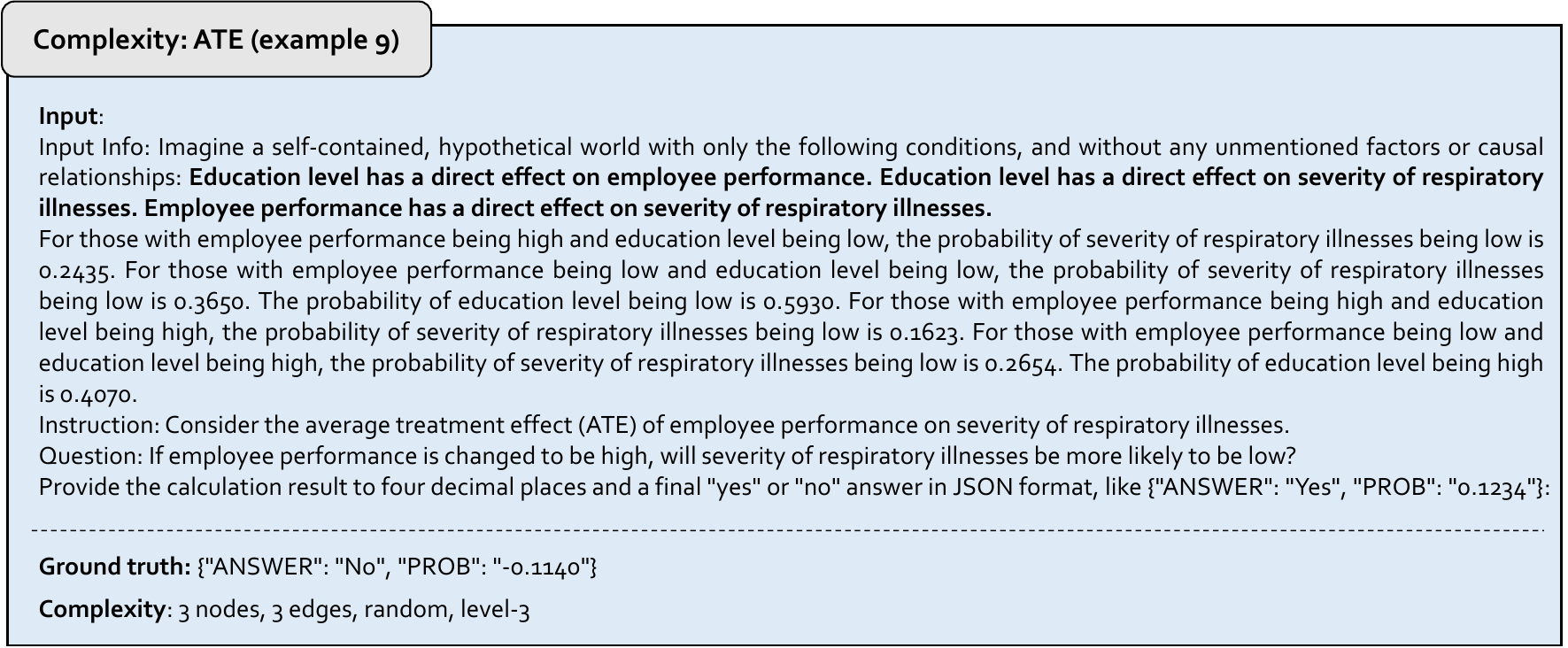}
   \caption[Analyzing complexity: example 9]{\textbf{Analyzing complexity: example 9.} }
    \label{fig_appendix:complexity_ate9}
\end{figure}

We provide nine examples (from Figure \ref{fig_appendix:complexity_ate1} to Figure \ref{fig_appendix:complexity_ate9}) used in \nameref{main:complexity} (\cref{main:complexity}).

\subsection{Supplementary Details for Prompt Analysis}
\label{appendix:prompt}

\begin{figure}[t]
\centering  
\subfigure[Average accuracy of IcL for scenarios in the Natural and Symbolic modes.]{   
\begin{minipage}{5cm}
\centering    
  \includegraphics[width=.95\linewidth]{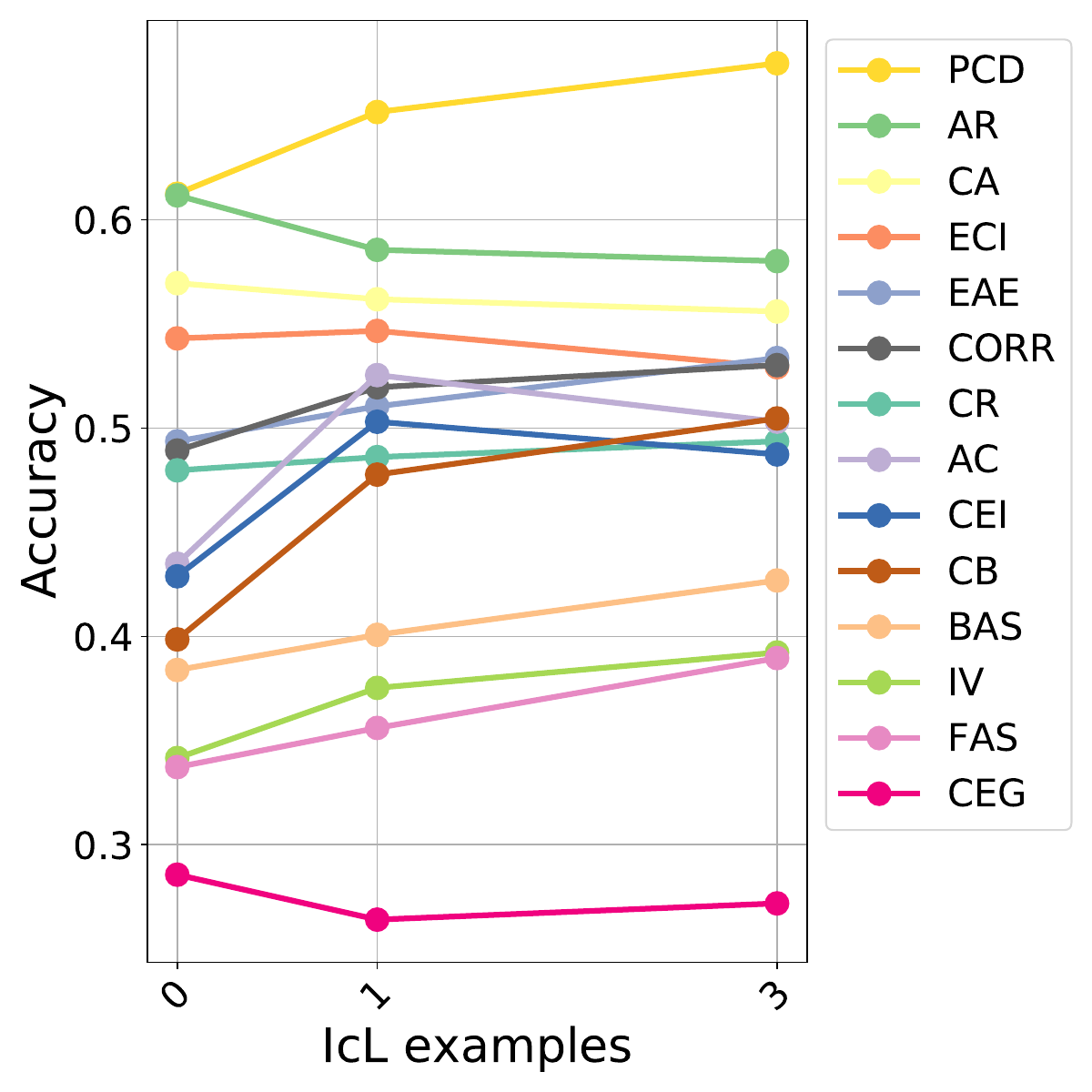}
    \label{fig_appendix:IcL_normal_1_3}
\end{minipage}
}
\hspace{0.5cm}
\subfigure[Average accuracy of IcL for scenarios in the Mathematical mode with 0/1/3 examples.]{ 
\begin{minipage}{5cm}
\centering   
  \includegraphics[width=.95\linewidth]{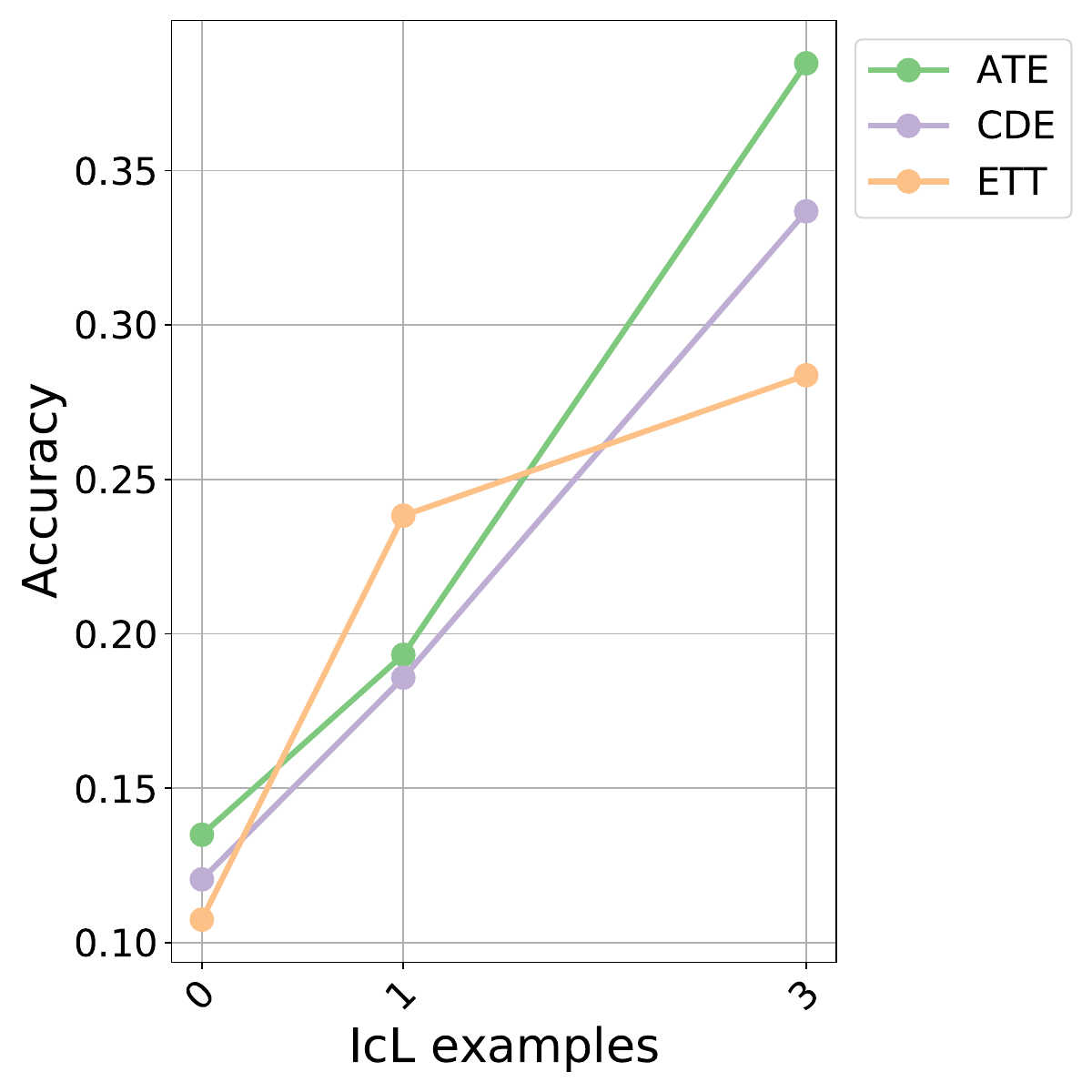}
    \label{IcL_max_3}
\end{minipage}
}
\hspace{0.5cm}
\subfigure[Average accuracy of IcL for scenarios in the Mathematical mode with 0/1/2 examples.]{  
\begin{minipage}{5cm}
\centering   
  \includegraphics[width=.95\linewidth]{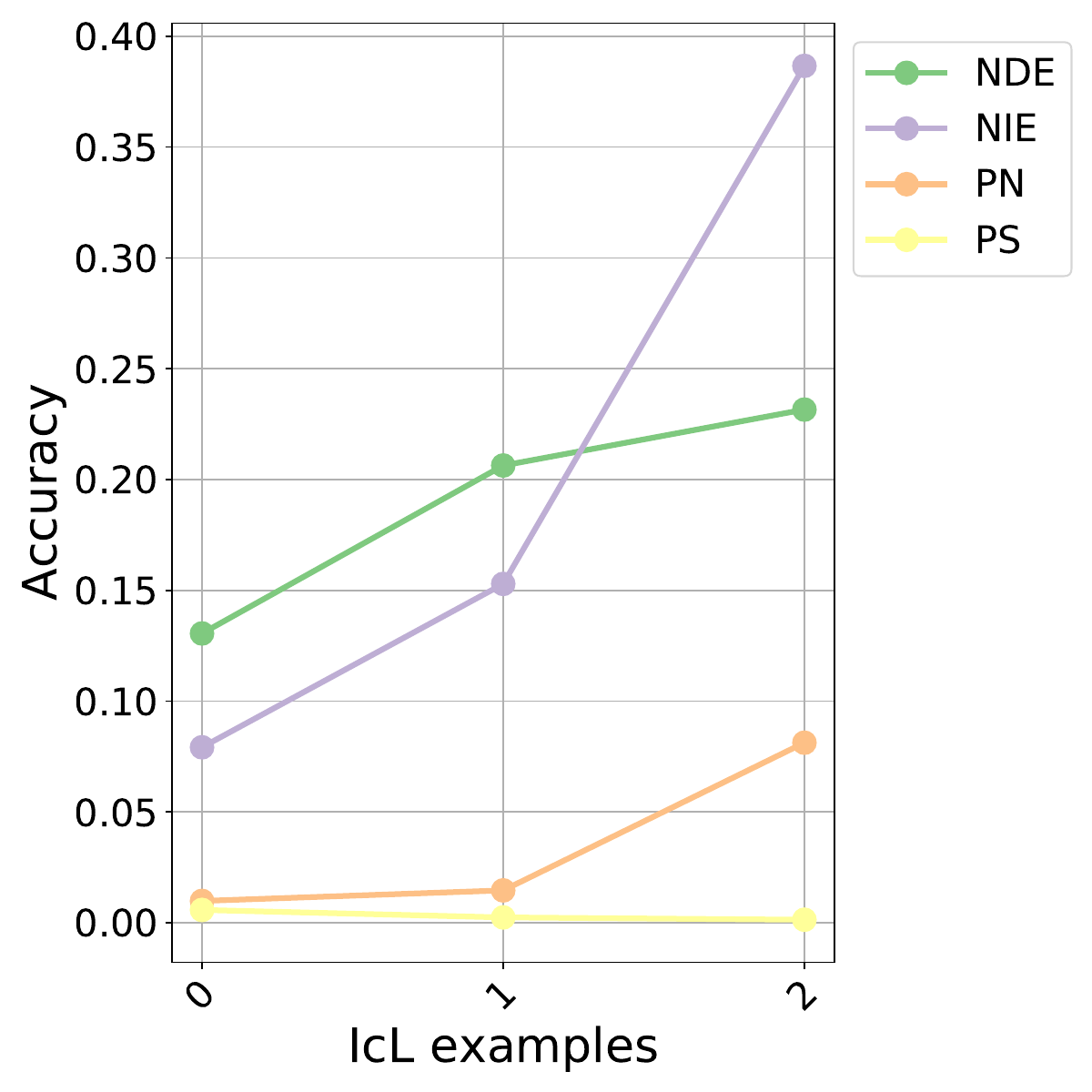}
    \label{fig_appendix:IcL_max_2}
\end{minipage}
}
\caption[Relationship between accuracy and the number of IcL examples on English datasets]{\textbf{Relationship between accuracy and the number of IcL examples on English datasets.}}
\label{fig_appendix:model_performances_on_EN_IcL}
\end{figure}

In Figure \ref{fig_appendix:model_performances_on_EN_IcL}, we demonstrate the relationship between accuracy and the number of IcL examples on English datasets.

\clearpage

\section{Additional Details for Scenario-specific Analysis}

\subsection{Causal Discovery}
\subsubsection{PCD}
\begin{figure}
\centering
\subfigure[Distribution of PCD-B (E-CARE)]{
\begin{minipage}{3.9cm}
\centering
\includegraphics[width=.9\linewidth]{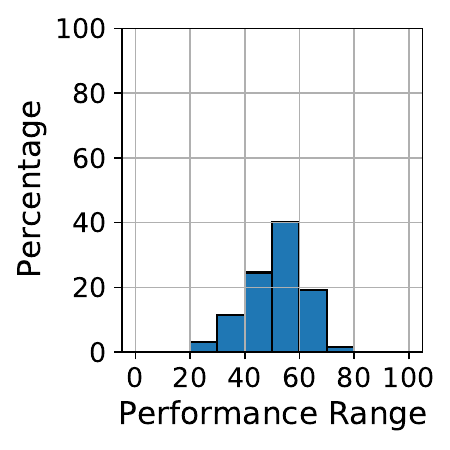}
\end{minipage}
}
\subfigure[Distribution of PCD-B (COPA)]{
\begin{minipage}{3.9cm}
\centering
\includegraphics[width=.9\linewidth]{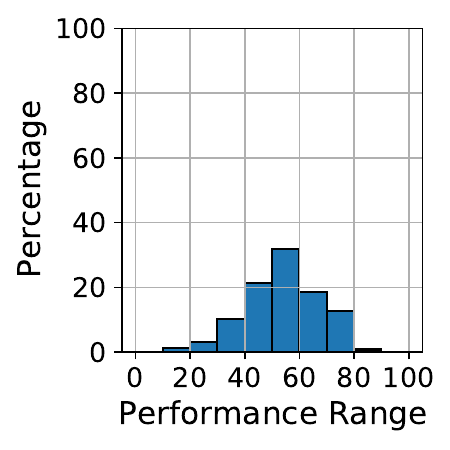}
\end{minipage}
}
\subfigure[Distribution of PCD-C (E-CARE)]{
\begin{minipage}{3.9cm}
\centering
\includegraphics[width=.9\linewidth]{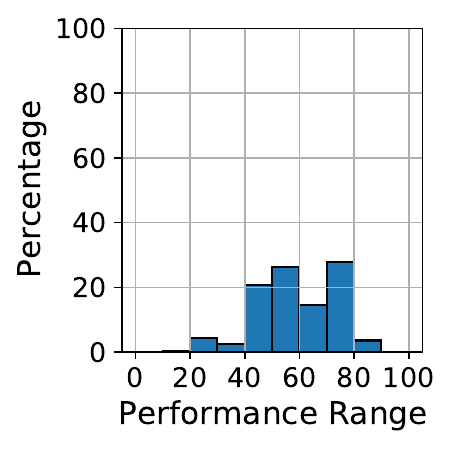}
\end{minipage}
}
\subfigure[Distribution of PCD-C (COPA)]{
\begin{minipage}{3.9cm}
\centering
\includegraphics[width=.9\linewidth]{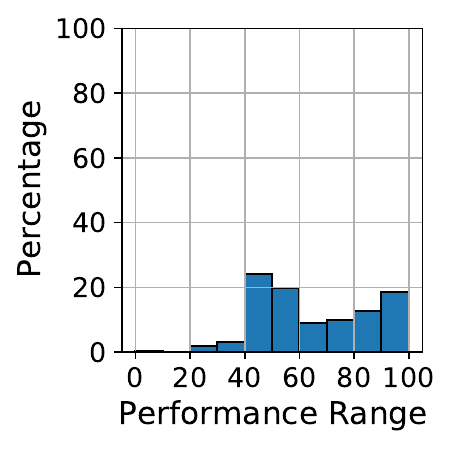}
\end{minipage}
}
\caption[Distribution of causal tasks in PCD]{\textbf{Distribution of causal tasks in PCD.}}
\label{fig:Distribution_of_Pairwise_Causal_Discovery_Tasks}
\end{figure}

\begin{figure}
\centering
\subfigure[Model performance of PCD-B (E-CARE)]{
\begin{minipage}{8.5cm}
\centering
\includegraphics[width=.9\linewidth]{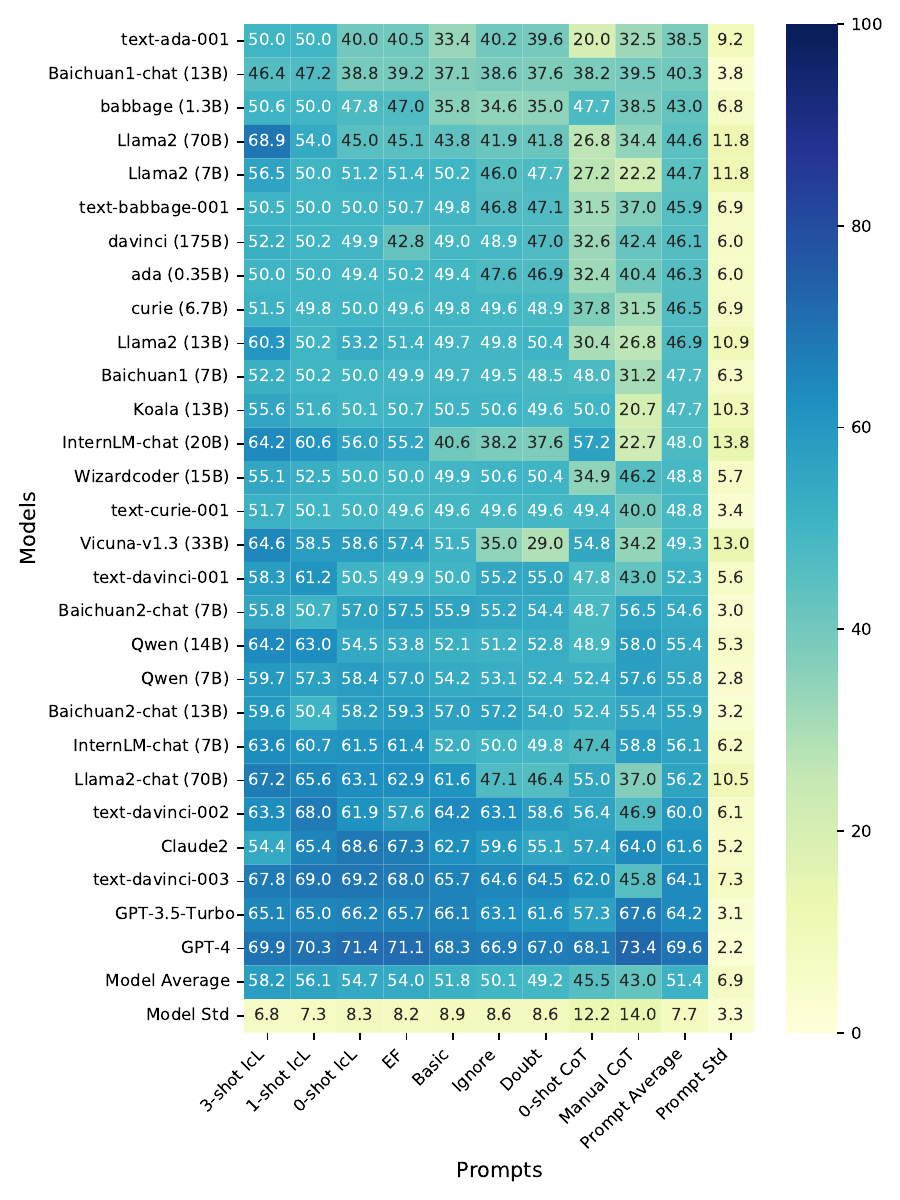}
\end{minipage}
}
\subfigure[Model performance of PCD-B (COPA)]{
\begin{minipage}{8.5cm}
\centering
\includegraphics[width=.9\linewidth]{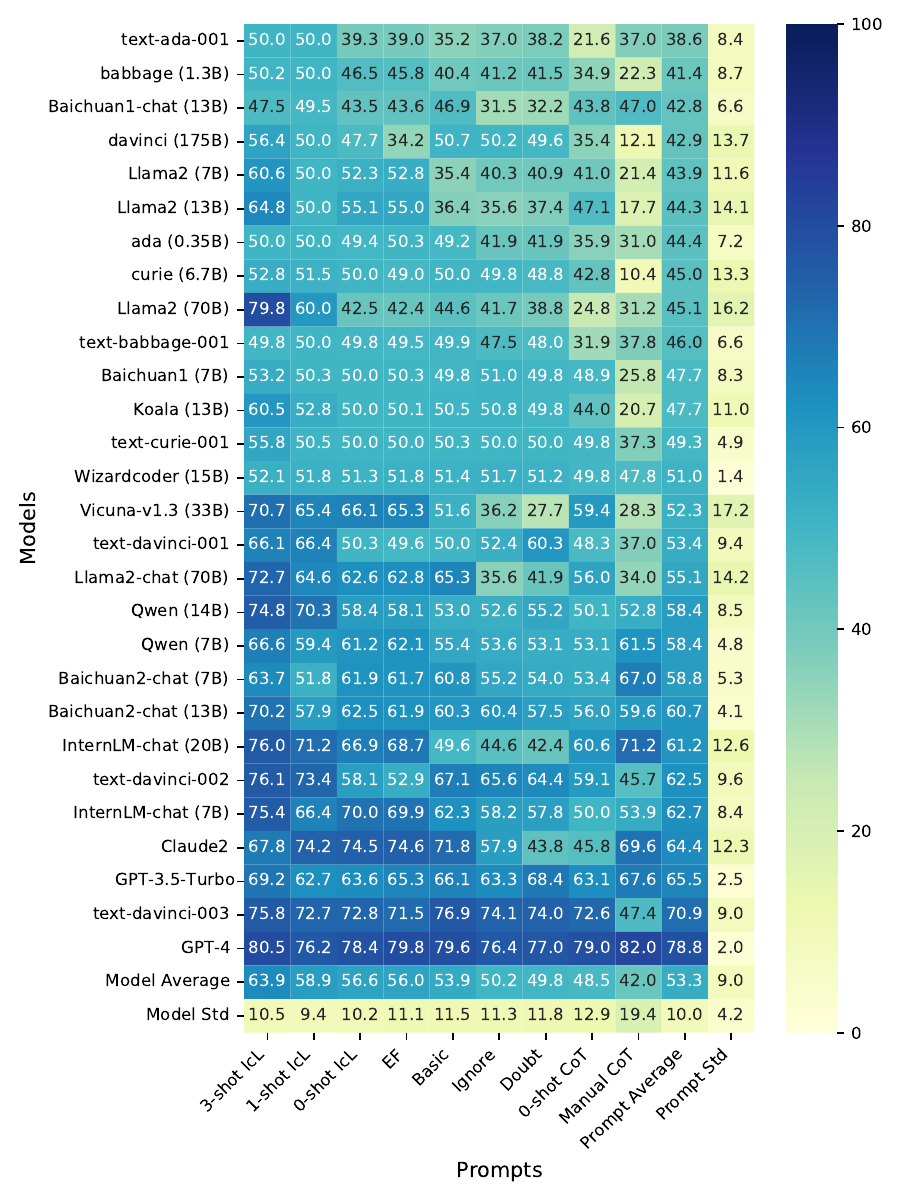}
\end{minipage}
}
\subfigure[Model performance of PCD-C (E-CARE)]{
\begin{minipage}{8.5cm}
\centering
\includegraphics[width=.9\linewidth]{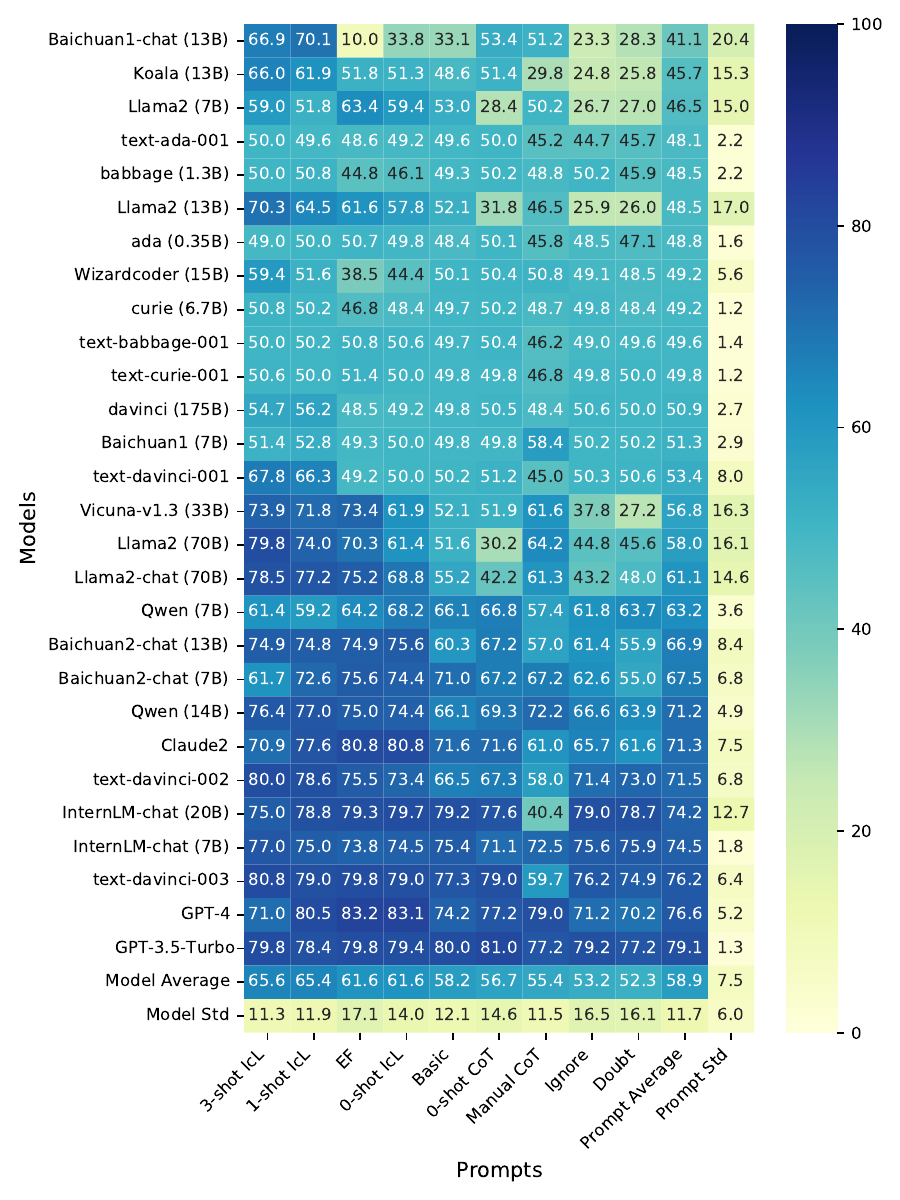}
\end{minipage}
}
\subfigure[Model performance of PCD-C (COPA)]{
\begin{minipage}{8.5cm}
\centering
\includegraphics[width=.9\linewidth]{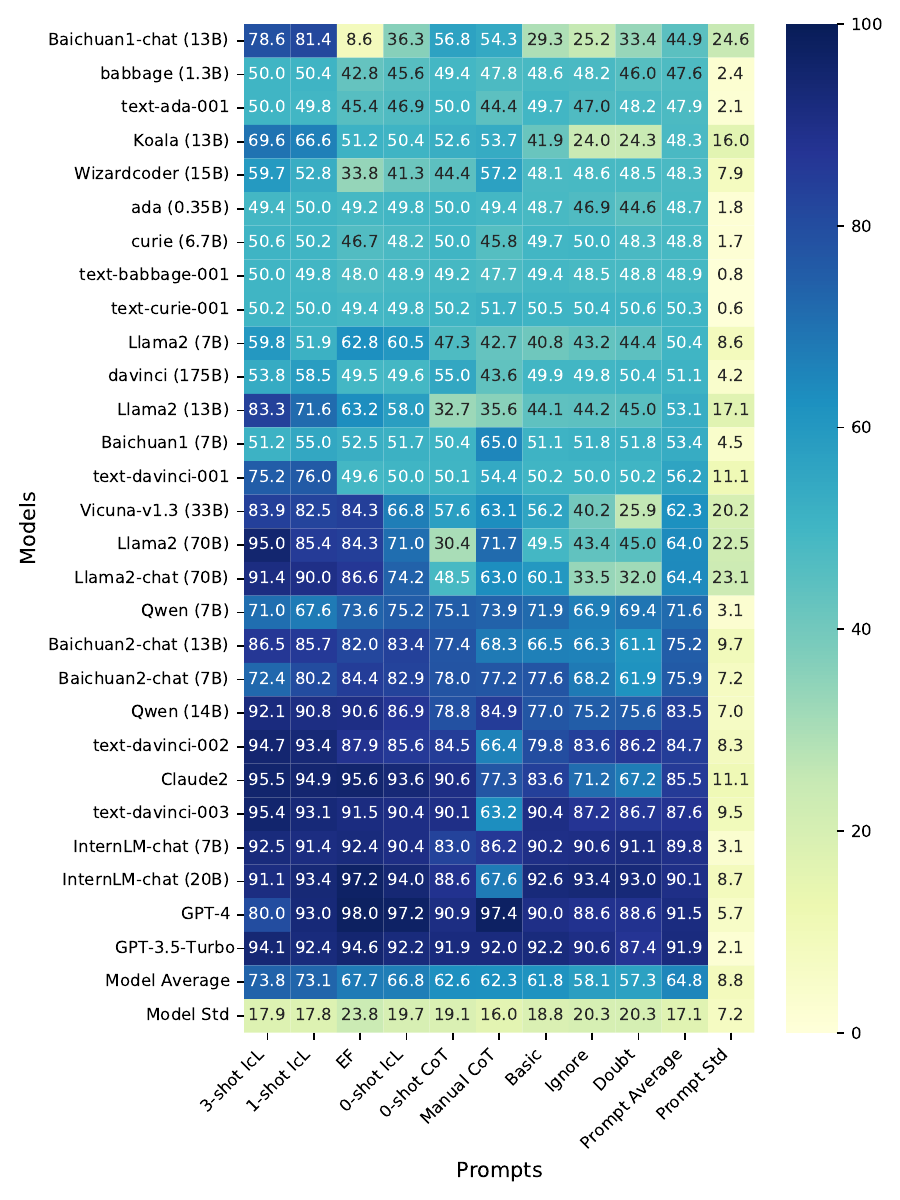}
\end{minipage}
}
\caption[Heatmaps of model performance of causal tasks in PCD]{\textbf{Heatmaps of model performance of causal tasks in PCD.} The models and prompts are sorted by their averages.}
\label{fig:Heatmap_of_performances_of_Pairwise_Causal_Discovery}
\end{figure}

\begin{figure}
\centering
\subfigure[\textit{Prompt gain} of PCD-B (E-CARE)]{
\begin{minipage}{8.5cm}
\centering
\includegraphics[width=.9\linewidth]{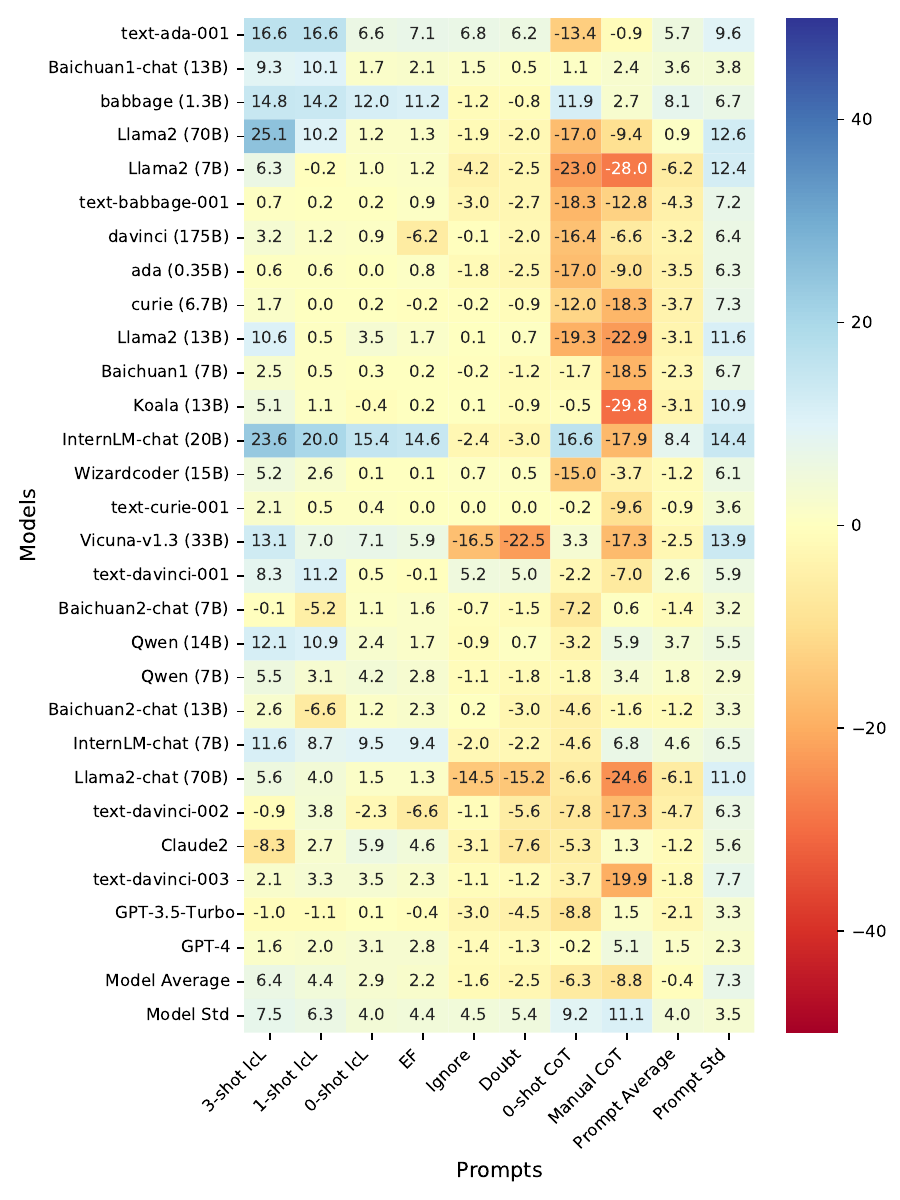}
\end{minipage}
}
\subfigure[\textit{Prompt gain} of PCD-B (COPA)]{
\begin{minipage}{8.5cm}
\centering
\includegraphics[width=.9\linewidth]{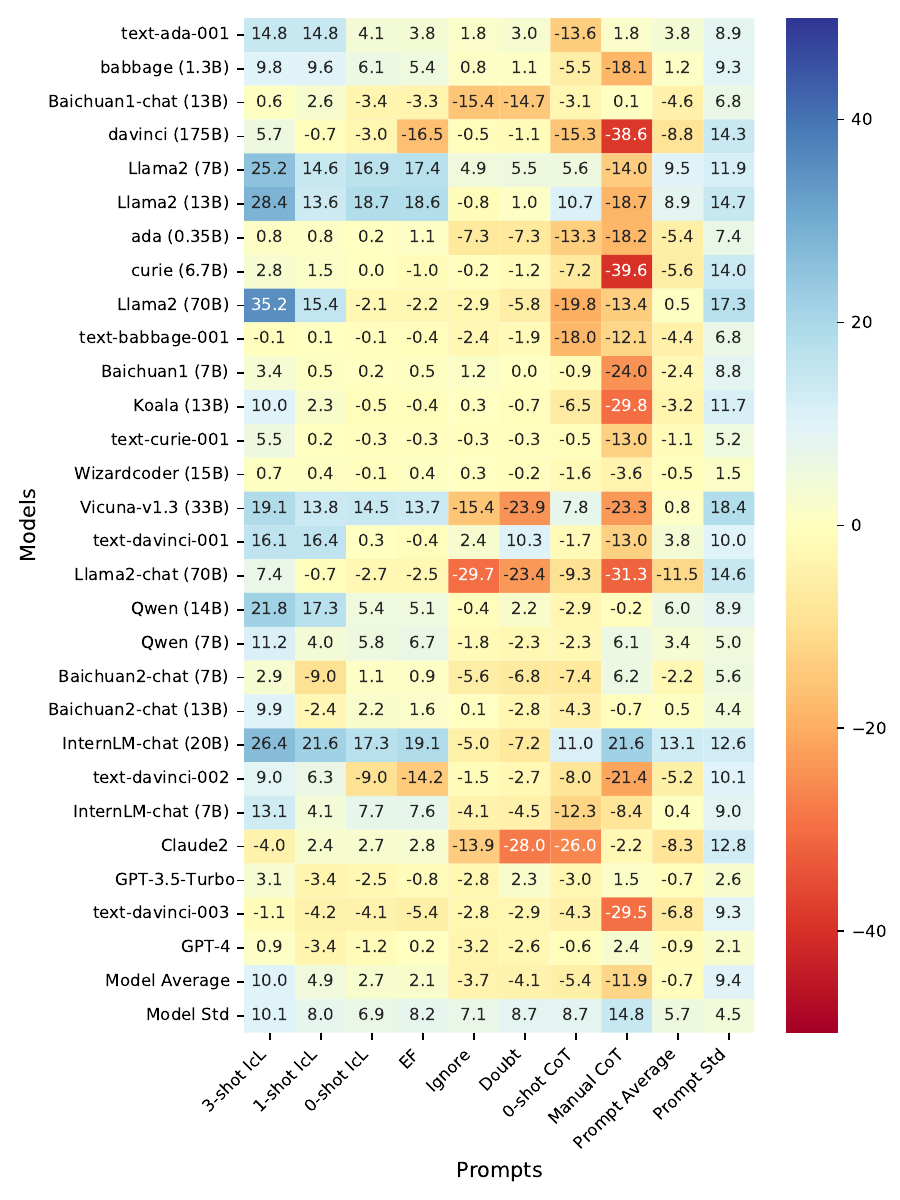}
\end{minipage}
}
\subfigure[\textit{Prompt gain} of PCD-C (E-CARE)]{
\begin{minipage}{8.5cm}
\centering
\includegraphics[width=.9\linewidth]{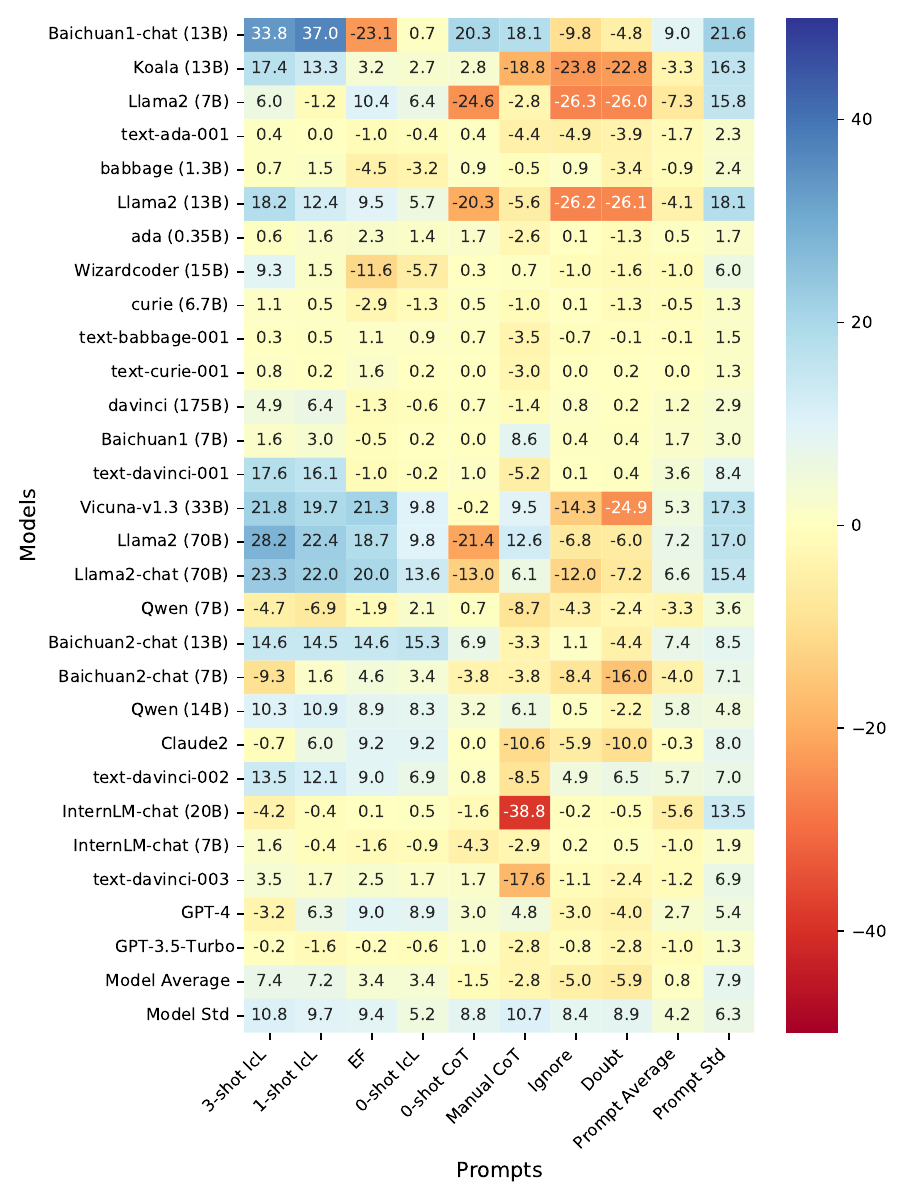}
\end{minipage}
}
\subfigure[\textit{Prompt gain} of PCD-C (COPA)]{
\begin{minipage}{8.5cm}
\centering
\includegraphics[width=.9\linewidth]{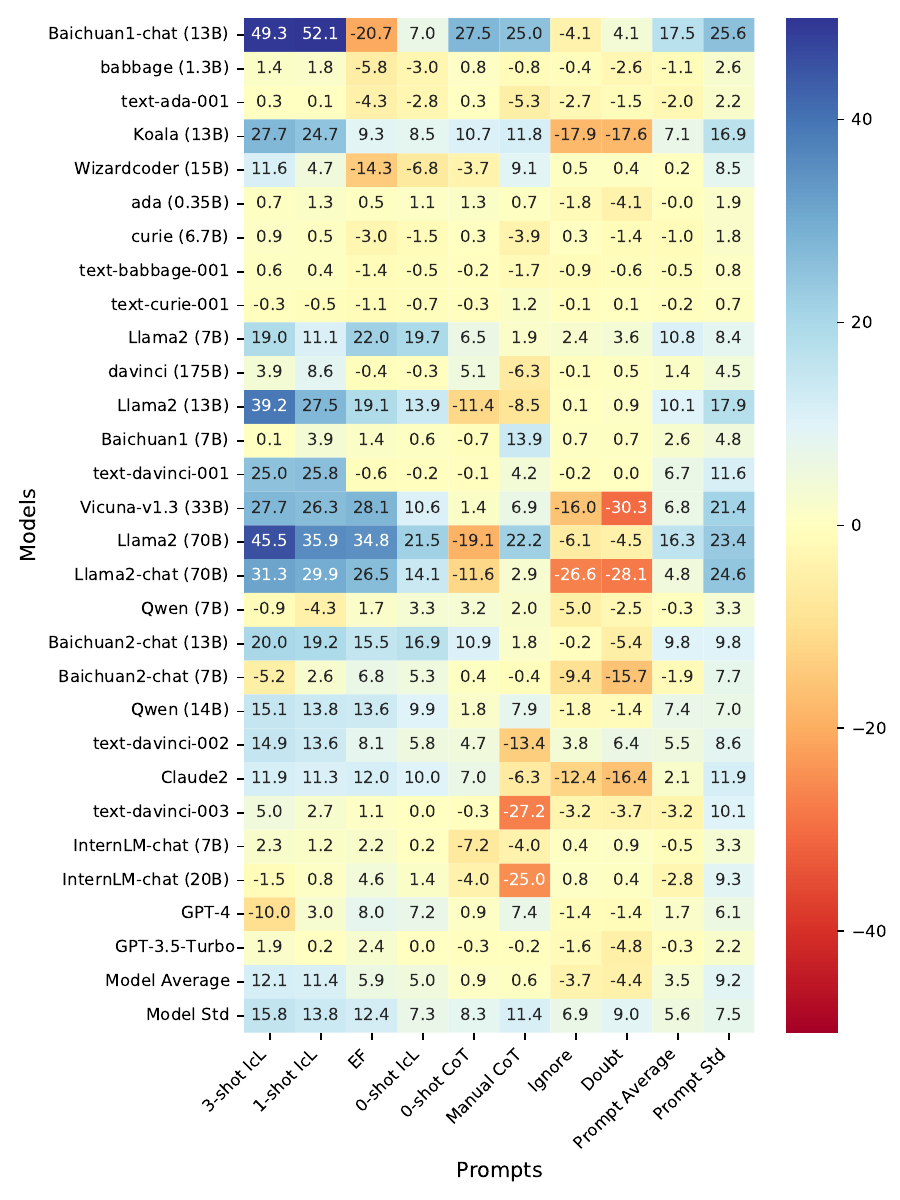}
\end{minipage}
}
\caption[Heatmaps of \textit{prompt gain} of causal tasks in PCD]{\textbf{Heatmaps of \textit{prompt gain} of causal tasks in PCD.} The models and prompts are sorted by their averages.}
\label{fig:Heatmap_of_gain_of_Pairwise_Causal_Discovery}
\end{figure}

The distribution of models' accuracy on PCD is shown in Figure \ref{fig:Distribution_of_Pairwise_Causal_Discovery_Tasks}. Figure \ref{fig:Heatmap_of_performances_of_Pairwise_Causal_Discovery} illustrates how models perform on PCD. The prompt gain (i.e., accuracy improvement against the basic prompt on the model with the used prompt) is demonstrated in Figure \ref{fig:Heatmap_of_gain_of_Pairwise_Causal_Discovery}.

\begin{figure}
\centering
\subfigure[Distribution of ECI-B (CTB)]{
\begin{minipage}{3.9cm}
\centering
\includegraphics[width=.9\linewidth]{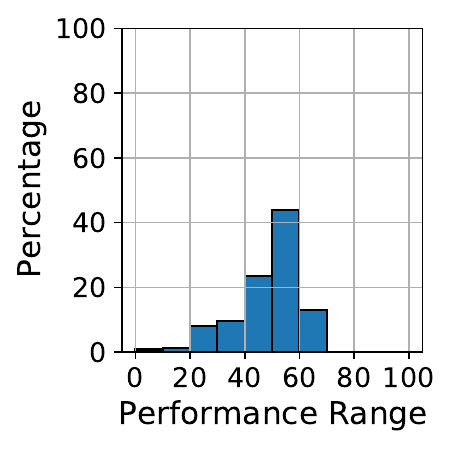}
\end{minipage}
}
\subfigure[Distribution of ECI-B (ESC)]{
\begin{minipage}{3.9cm}
\centering
\includegraphics[width=.9\linewidth]{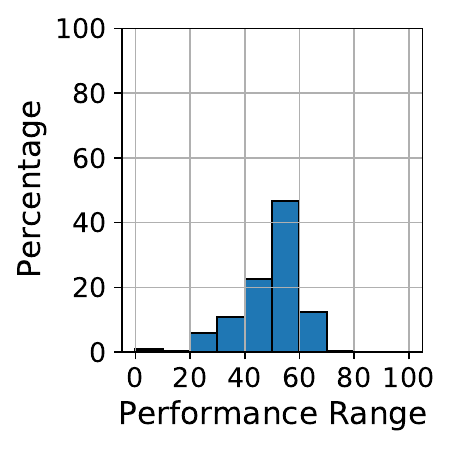}
\end{minipage}
}
\subfigure[Distribution of ECI-B (MAVEN-ERE)]{
\begin{minipage}{3.9cm}
\centering
\includegraphics[width=.9\linewidth]{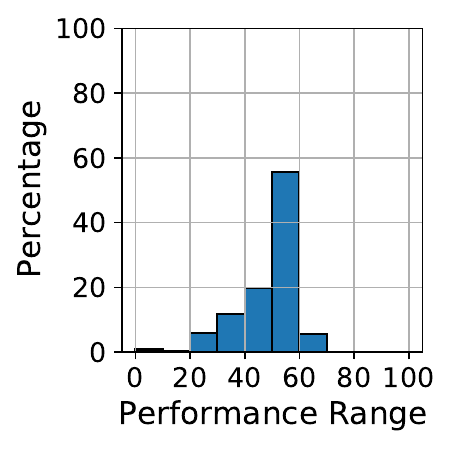}
\end{minipage}
}
\caption[Distribution of causal tasks in ECI]{\textbf{Distribution of causal tasks in ECI.}}
\label{fig:Distribution_of_Event_Causality_Identification_Tasks}
\end{figure}

\begin{figure}
\centering
\subfigure[Distribution of CA-B (FA)]{
\begin{minipage}{3.9cm}
\centering
\includegraphics[width=.9\linewidth]{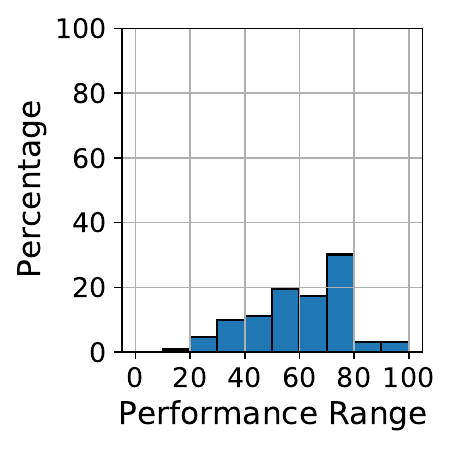}
\end{minipage}
}
\subfigure[Distribution of CA-B (FP)]{
\begin{minipage}{3.9cm}
\centering
\includegraphics[width=.9\linewidth]{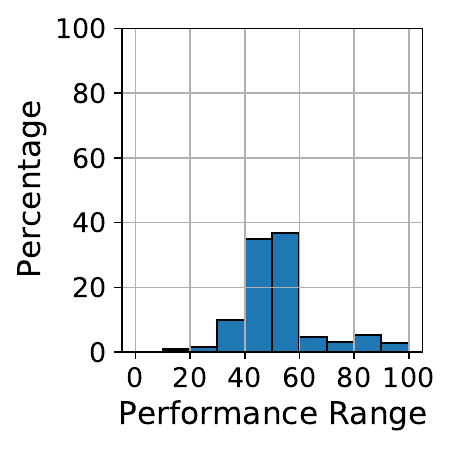}
\end{minipage}
}
\caption[Distribution of causal tasks in CA]{\textbf{Distribution of causal tasks in CA.}}
\label{fig:Distribution_of_Causal_Attribution_Tasks}
\end{figure}

\begin{figure}
\centering
\subfigure[Distribution of ATE-P (ATE-basic)]{
\begin{minipage}{3.9cm}
\centering
\includegraphics[width=.9\linewidth]{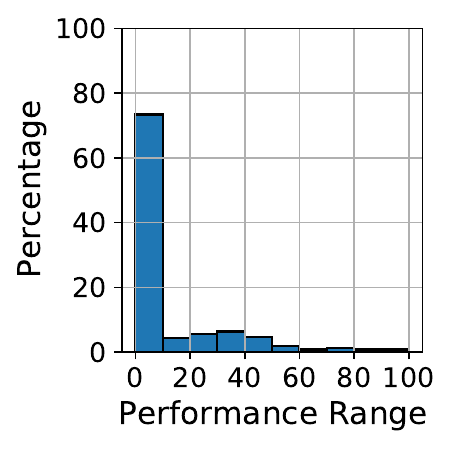}
\end{minipage}
}
\subfigure[Distribution of ATE-P (ATE-hard)]{
\begin{minipage}{3.9cm}
\centering
\includegraphics[width=.9\linewidth]{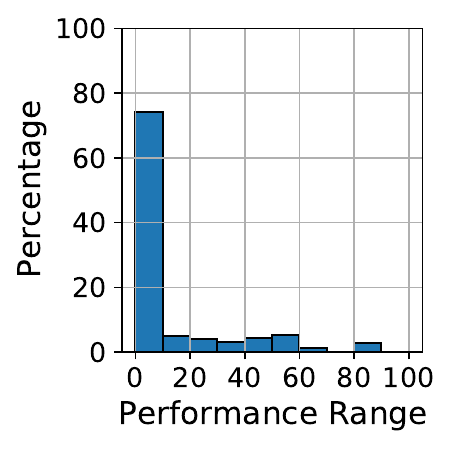}
\end{minipage}
}
\subfigure[Distribution of ATE-B (ATE-natural)]{
\begin{minipage}{3.9cm}
\centering
\includegraphics[width=.9\linewidth]{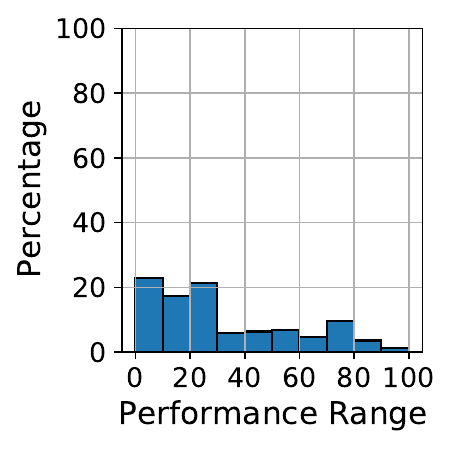}
\end{minipage}
}
\caption[Distribution of causal tasks in ATE]{\textbf{Distribution of causal tasks in ATE.}}
\label{fig:Distribution_of_Average_Treatment_Effect_Tasks}
\end{figure}

\begin{figure}
\centering
\subfigure[Distribution of CDE-P (CDE-basic)]{
\begin{minipage}{3.9cm}
\centering
\includegraphics[width=.9\linewidth]{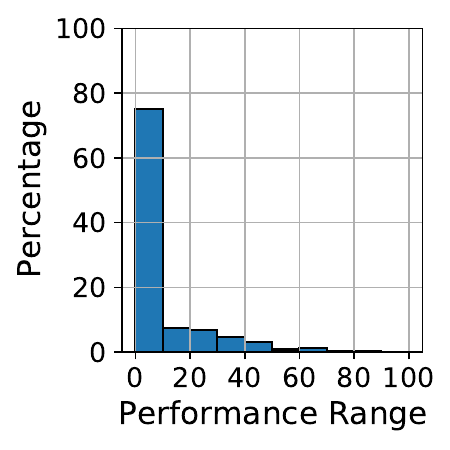}
\end{minipage}
}
\subfigure[Distribution of CDE-P (CDE-hard)]{
\begin{minipage}{3.9cm}
\centering
\includegraphics[width=.9\linewidth]{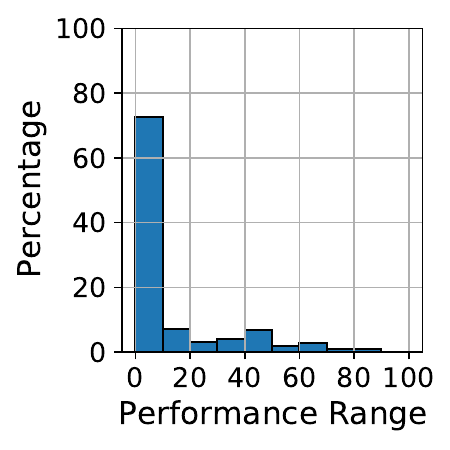}
\end{minipage}
}
\subfigure[Distribution of CDE-B (CDE-natural)]{
\begin{minipage}{3.9cm}
\centering
\includegraphics[width=.9\linewidth]{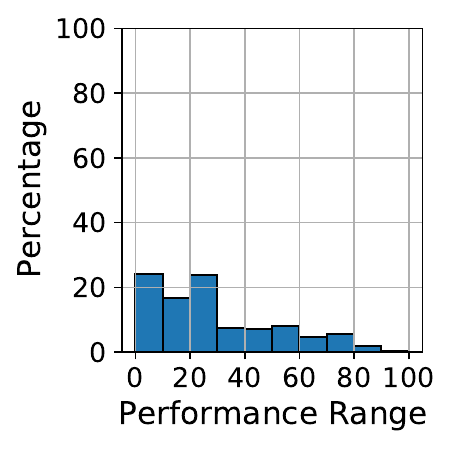}
\end{minipage}
}
\caption[Distribution of causal tasks in CDE]{\textbf{Distribution of causal tasks in CDE.}}
\label{fig:Distribution_of_Controlled_Direct_Effect_Tasks}
\end{figure}

\begin{figure}
\centering
\subfigure[Distribution of CEI-B (0.2-UC)]{
\begin{minipage}{3.9cm}
\centering
\includegraphics[width=.9\linewidth]{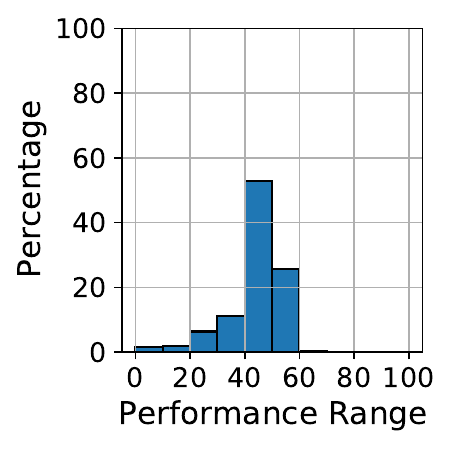}
\end{minipage}
}
\subfigure[Distribution of CEI-B (0.4-UC)]{
\begin{minipage}{3.9cm}
\centering
\includegraphics[width=.9\linewidth]{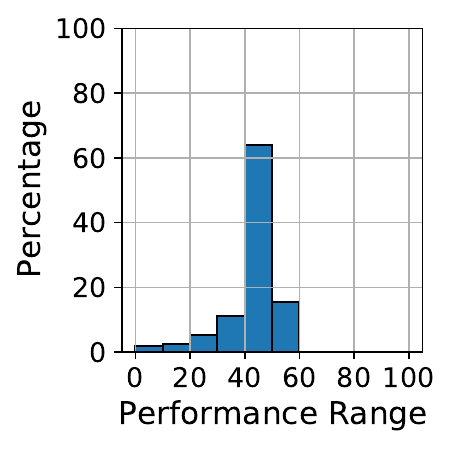}
\end{minipage}
}
\subfigure[Distribution of CEI-B (0.6-UC)]{
\begin{minipage}{3.9cm}
\centering
\includegraphics[width=.9\linewidth]{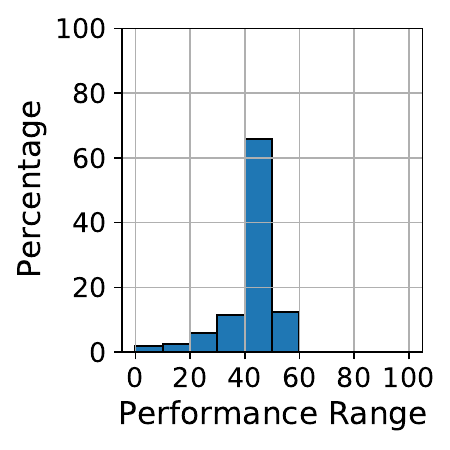}
\end{minipage}
}
\subfigure[Distribution of CEI-B (0.8-UC)]{
\begin{minipage}{3.9cm}
\centering
\includegraphics[width=.9\linewidth]{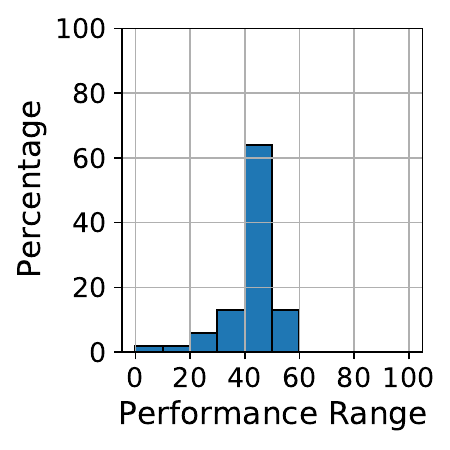}
\end{minipage}
}
\caption[Distribution of causal tasks in CEI]{\textbf{Distribution of causal tasks in CEI.}}
\label{fig:Distribution_of_Causal_Effect_Identification_Tasks}
\end{figure}

\begin{figure}
\centering
\subfigure[Distribution of BAS-B (backadj)]{
\begin{minipage}{3.9cm}
\centering
\includegraphics[width=.9\linewidth]{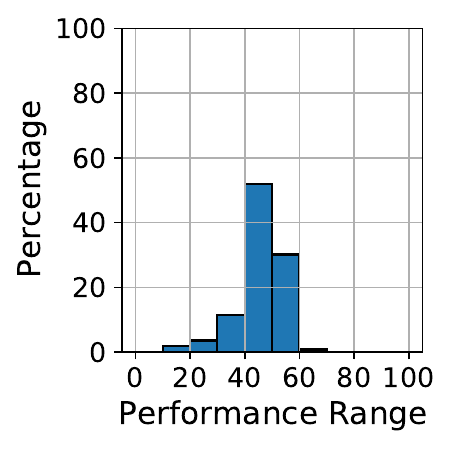}
\end{minipage}
}
\subfigure[Distribution of BAS-C (max-BAS)]{
\begin{minipage}{3.9cm}
\centering
\includegraphics[width=.9\linewidth]{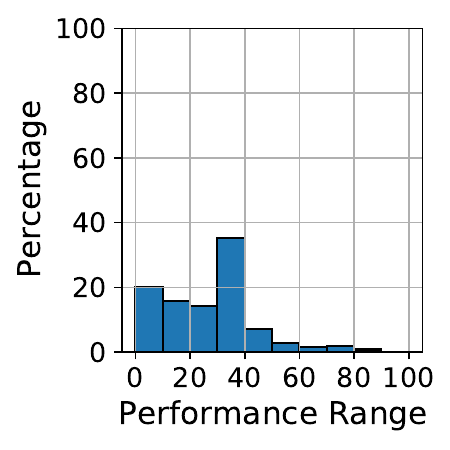}
\end{minipage}
}
\subfigure[Distribution of BAS-C (min-BAS)]{
\begin{minipage}{3.9cm}
\centering
\includegraphics[width=.9\linewidth]{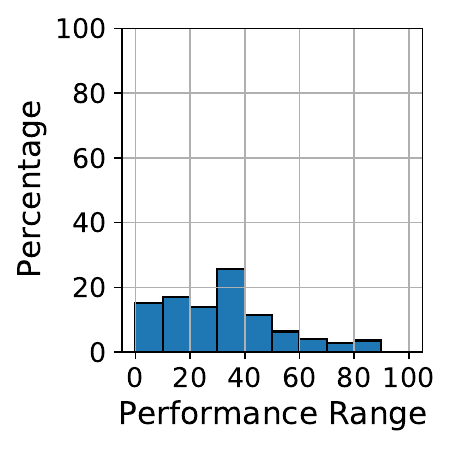}
\end{minipage}
}
\subfigure[Distribution of BAS-C (mix-BAS)]{
\begin{minipage}{3.9cm}
\centering
\includegraphics[width=.9\linewidth]{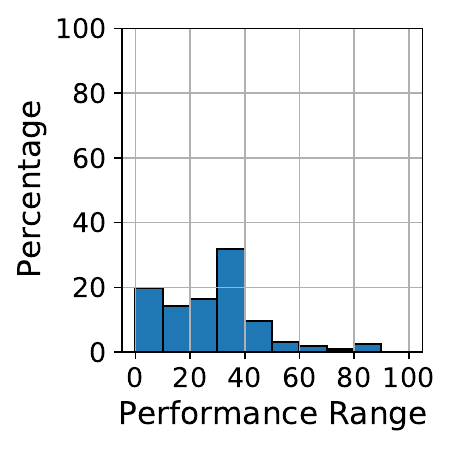}
\end{minipage}
}
\caption[Distribution of causal tasks in BAS]{\textbf{Distribution of causal tasks in BAS.}}
\label{fig:Distribution_of_Backdoor_Adjustment_Set_Tasks}
\end{figure}

\begin{figure}
\centering
\subfigure[Distribution of CR-C (CRASS)]{
\begin{minipage}{3.9cm}
\centering
\includegraphics[width=.9\linewidth]{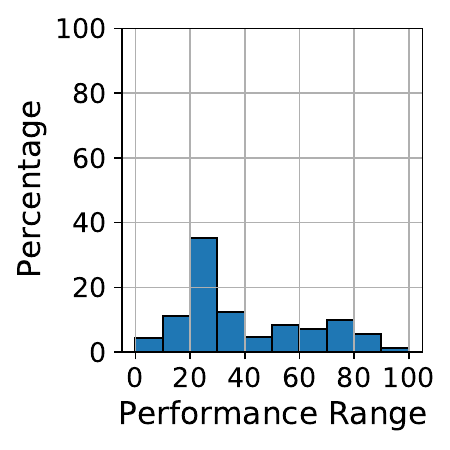}
\end{minipage}
}
\subfigure[Distribution of CR-B (det-counterfactual)]{
\begin{minipage}{3.9cm}
\centering
\includegraphics[width=.9\linewidth]{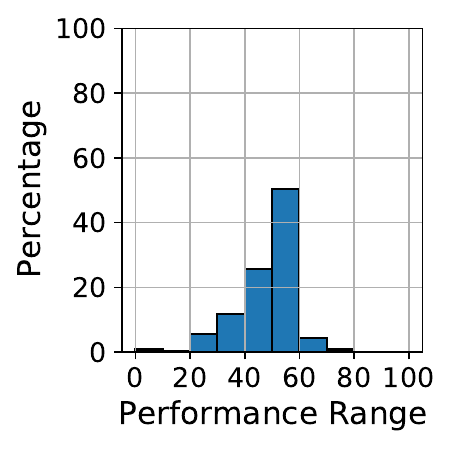}
\end{minipage}
}
\caption[Distribution of causal tasks in CR]{\textbf{Distribution of causal tasks in CR.}}
\label{fig:Distribution_of_Counterfactual_Reasoning_Tasks}
\end{figure}

\begin{figure}
\centering
\subfigure[Distribution of ETT-P (ETT-basic)]{
\begin{minipage}{3.9cm}
\centering
\includegraphics[width=.9\linewidth]{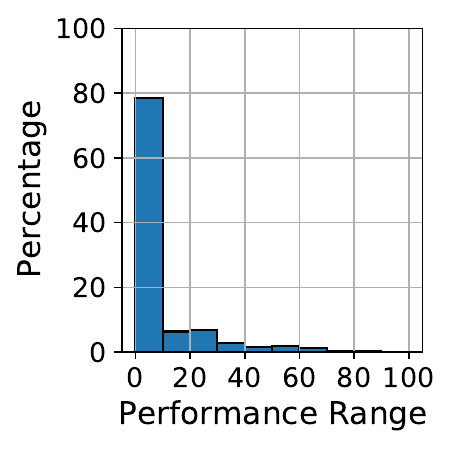}
\end{minipage}
}
\subfigure[Distribution of ETT-P (ETT-hard)]{
\begin{minipage}{3.9cm}
\centering
\includegraphics[width=.9\linewidth]{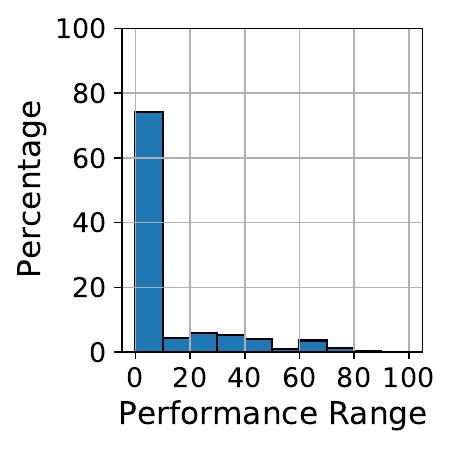}
\end{minipage}
}
\subfigure[Distribution of ETT-B (ETT-natural)]{
\begin{minipage}{3.9cm}
\centering
\includegraphics[width=.9\linewidth]{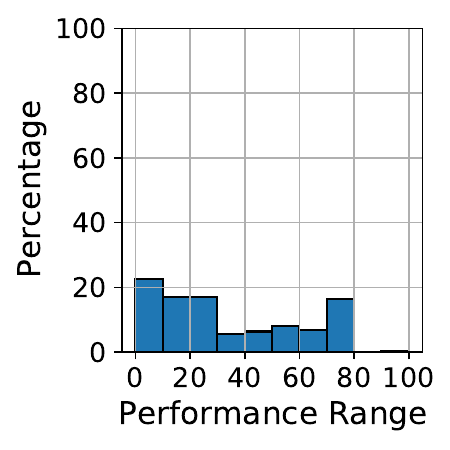}
\end{minipage}
}
\caption[Distribution of causal tasks in ETT]{\textbf{Distribution of causal tasks in ETT.}}
\label{fig:Distribution_of_Effect_of_the_Treatment_on_the_Treated_Tasks}
\end{figure}

\begin{figure}
\centering
\subfigure[Distribution of NDE-P (NDE-basic)]{
\begin{minipage}{3.9cm}
\centering
\includegraphics[width=.9\linewidth]{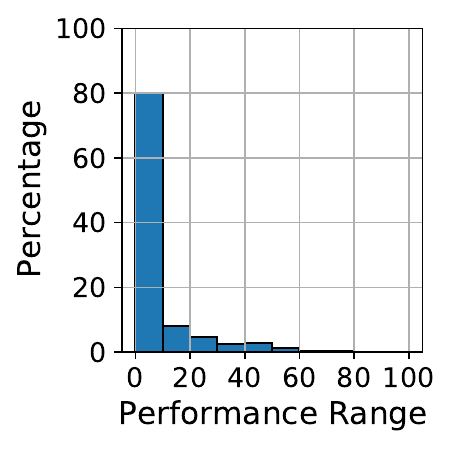}
\end{minipage}
}
\subfigure[Distribution of NDE-P (NDE-hard)]{
\begin{minipage}{3.9cm}
\centering
\includegraphics[width=.9\linewidth]{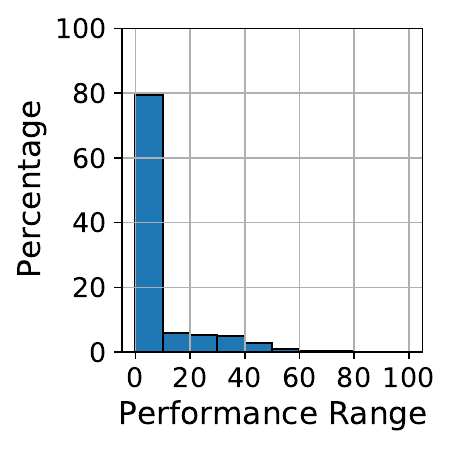}
\end{minipage}
}
\subfigure[Distribution of NDE-B (NDE-natural)]{
\begin{minipage}{3.9cm}
\centering
\includegraphics[width=.9\linewidth]{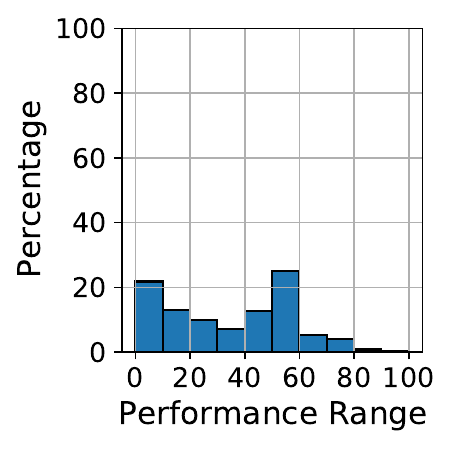}
\end{minipage}
}
\caption[Distribution of causal tasks in NDE]{\textbf{Distribution of causal tasks in NDE.}}
\label{fig:Distribution_of_causal_tasks_in_NDE}
\end{figure}

\begin{figure}
\centering
\subfigure[Distribution of NIE-P (NIE-basic)]{
\begin{minipage}{3.9cm}
\centering
\includegraphics[width=.9\linewidth]{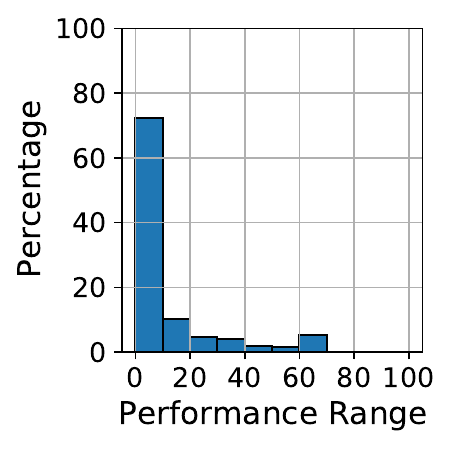}
\end{minipage}
}
\subfigure[Distribution of NIE-P (NIE-hard)]{
\begin{minipage}{3.9cm}
\centering
\includegraphics[width=.9\linewidth]{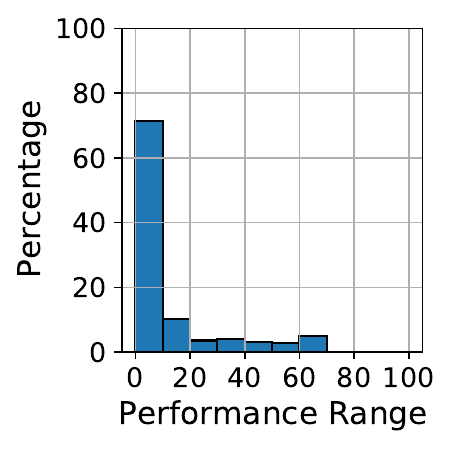}
\end{minipage}
}
\subfigure[Distribution of NIE-B (NIE-natural)]{
\begin{minipage}{3.9cm}
\centering
\includegraphics[width=.9\linewidth]{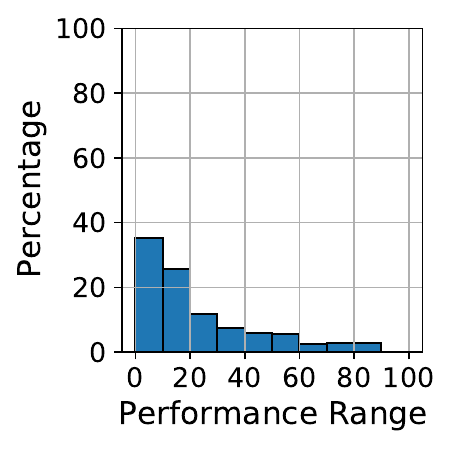}
\end{minipage}
}
\caption[Distribution of causal tasks in NIE]{\textbf{Distribution of causal tasks in NIE.}}
\label{fig:Distribution_of_Natural_Indirect_Effect_Tasks}
\end{figure}

\begin{figure}
\centering
\subfigure[Distribution of PN-P (PN-basic)]{
\begin{minipage}{3.9cm}
\centering
\includegraphics[width=.9\linewidth]{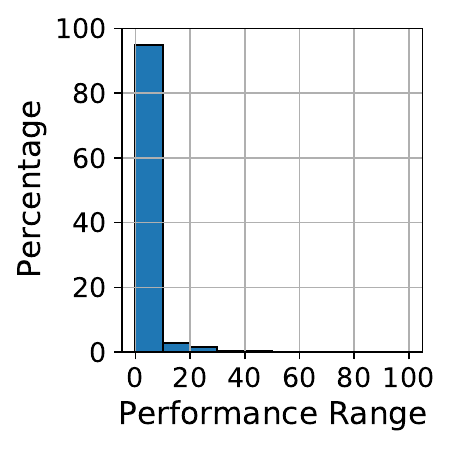}
\end{minipage}
}
\subfigure[Distribution of PN-P (PN-hard)]{
\begin{minipage}{3.9cm}
\centering
\includegraphics[width=.9\linewidth]{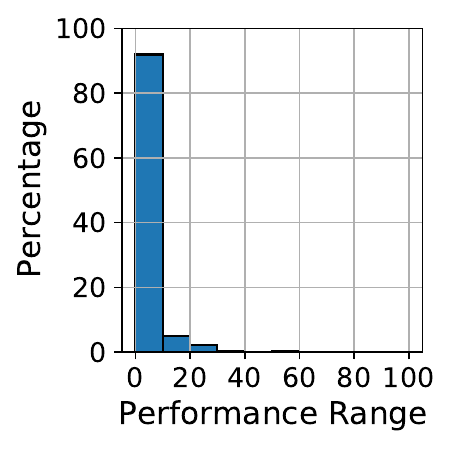}
\end{minipage}
}
\caption[Distribution of causal tasks in PN]{\textbf{Distribution of causal tasks in PN.}}
\label{fig:Distribution_of_Probability_of_Necessity_Tasks}
\end{figure}

\begin{figure}
\centering
\subfigure[Distribution of PS-P (PS-basic)]{
\begin{minipage}{3.9cm}
\centering
\includegraphics[width=.9\linewidth]{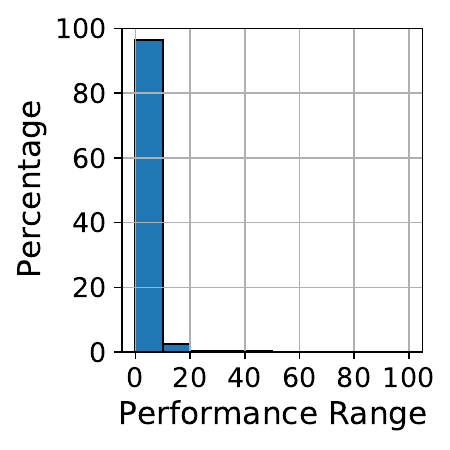}
\end{minipage}
}
\subfigure[Distribution of PS-P (PS-hard)]{
\begin{minipage}{3.9cm}
\centering
\includegraphics[width=.9\linewidth]{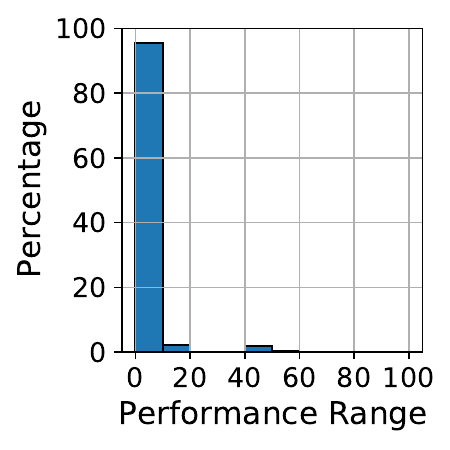}
\end{minipage}
}
\caption[Distribution of causal tasks in PS]{\textbf{Distribution of causal tasks in PS.}}
\label{fig:Distribution_of_Probability_of_Sufficiency_Tasks}
\end{figure}

\subsubsection{ECI}
\begin{figure}
\centering
\subfigure[Model performance of ECI-B (CTB)]{
\begin{minipage}{8.5cm}
\centering
\includegraphics[width=.9\linewidth]{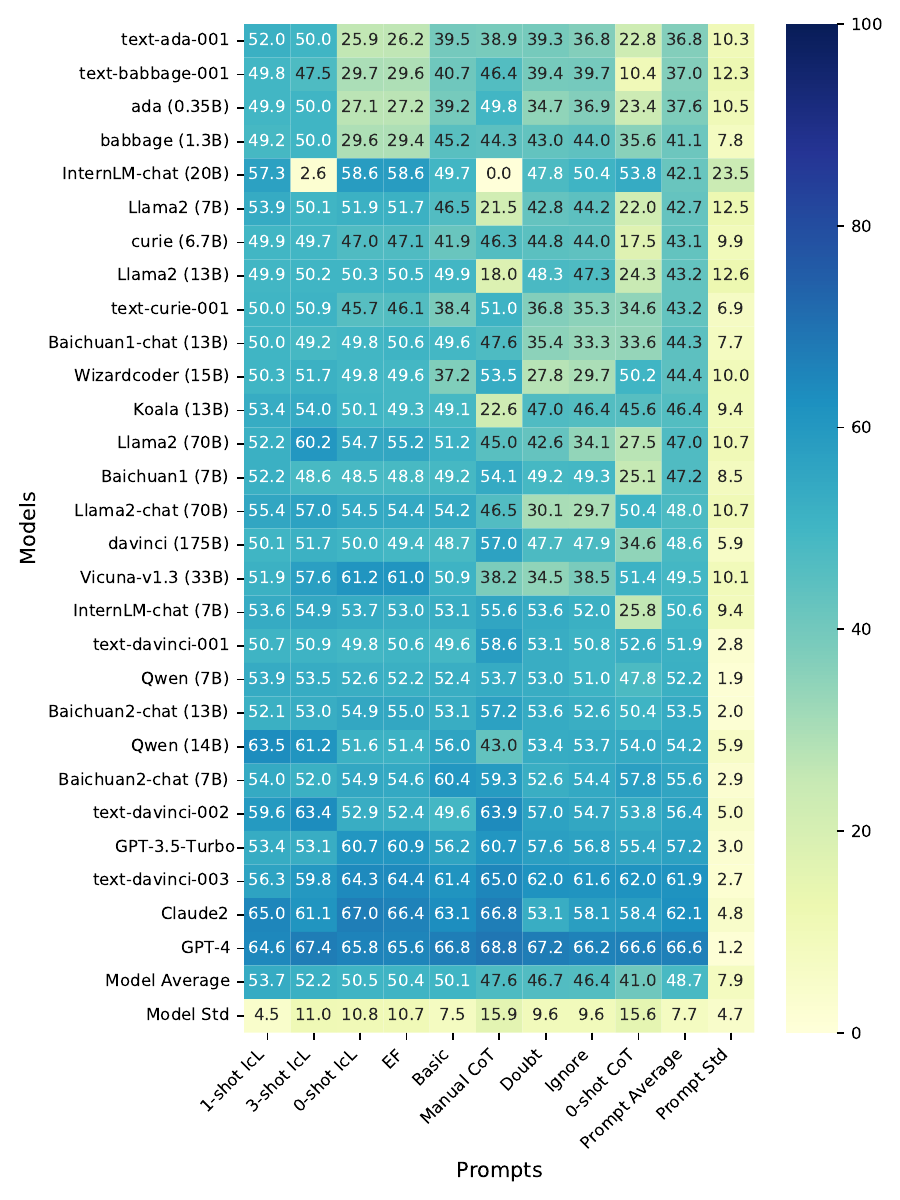}
\end{minipage}
}
\subfigure[Model performance of ECI-B (ESC)]{
\begin{minipage}{8.5cm}
\centering
\includegraphics[width=.9\linewidth]{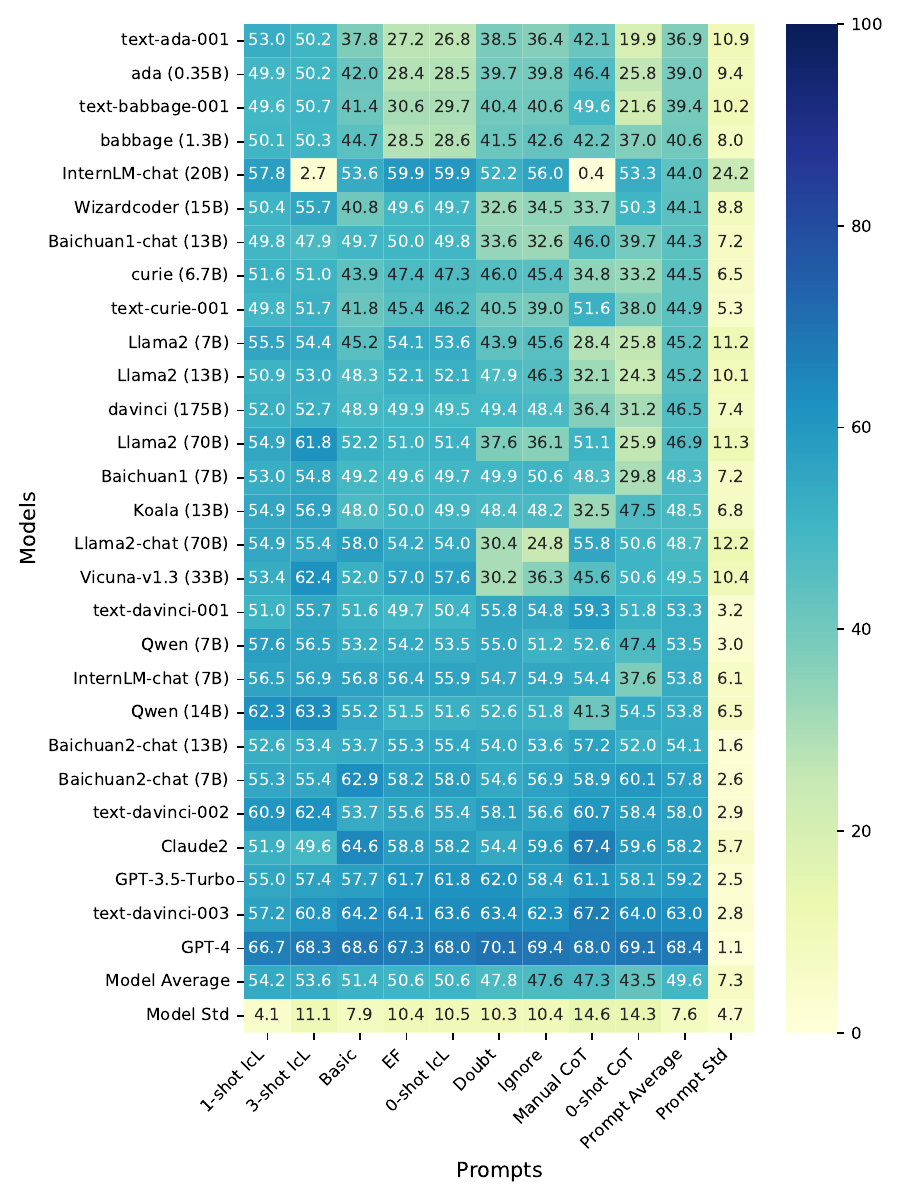}
\end{minipage}
}
\subfigure[Model performance of ECI-B (MAVEN-ERE)]{
\begin{minipage}{8.5cm}
\centering
\includegraphics[width=.9\linewidth]{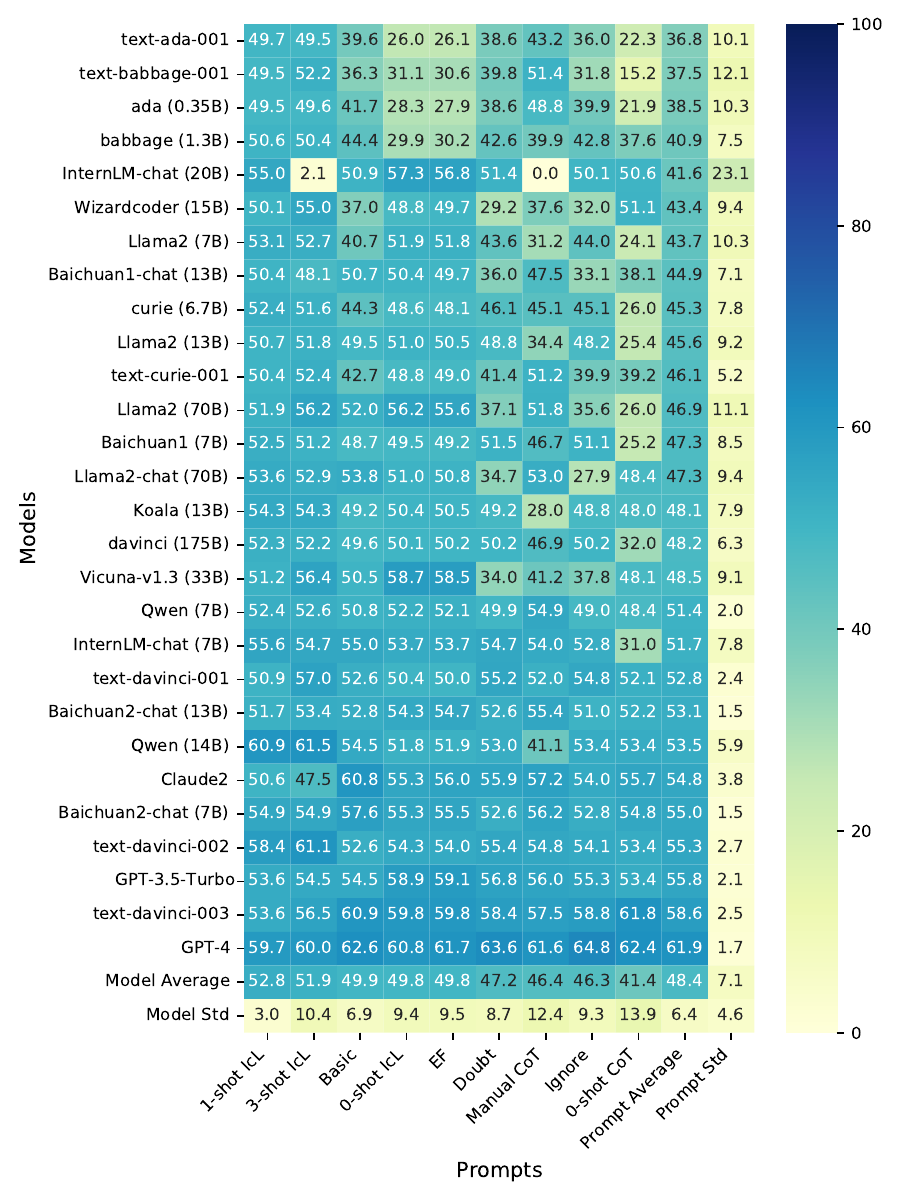}
\end{minipage}
}
\caption[Heatmaps of model performance of causal tasks in ECI]{\textbf{Heatmaps of model performance of causal tasks in ECI.} The models and prompts are sorted by their averages.}
\label{fig:Heatmap_of_performances_of_Event_Causality_Identification}
\end{figure}

\begin{figure}
\centering
\subfigure[\textit{Prompt gain} of ECI-B (CTB)]{
\begin{minipage}{8.5cm}
\centering
\includegraphics[width=.9\linewidth]{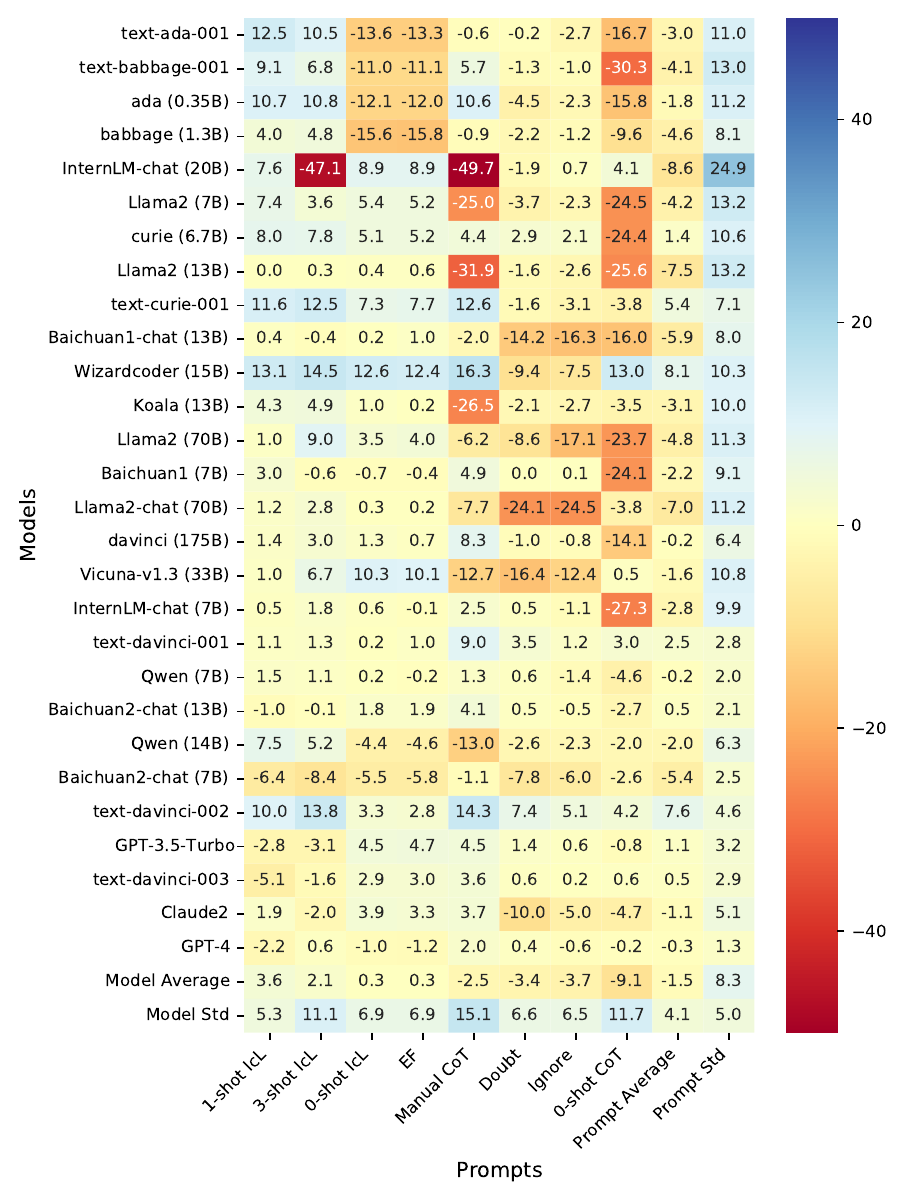}
\end{minipage}
}
\subfigure[\textit{Prompt gain} of ECI-B (ESC)]{
\begin{minipage}{8.5cm}
\centering
\includegraphics[width=.9\linewidth]{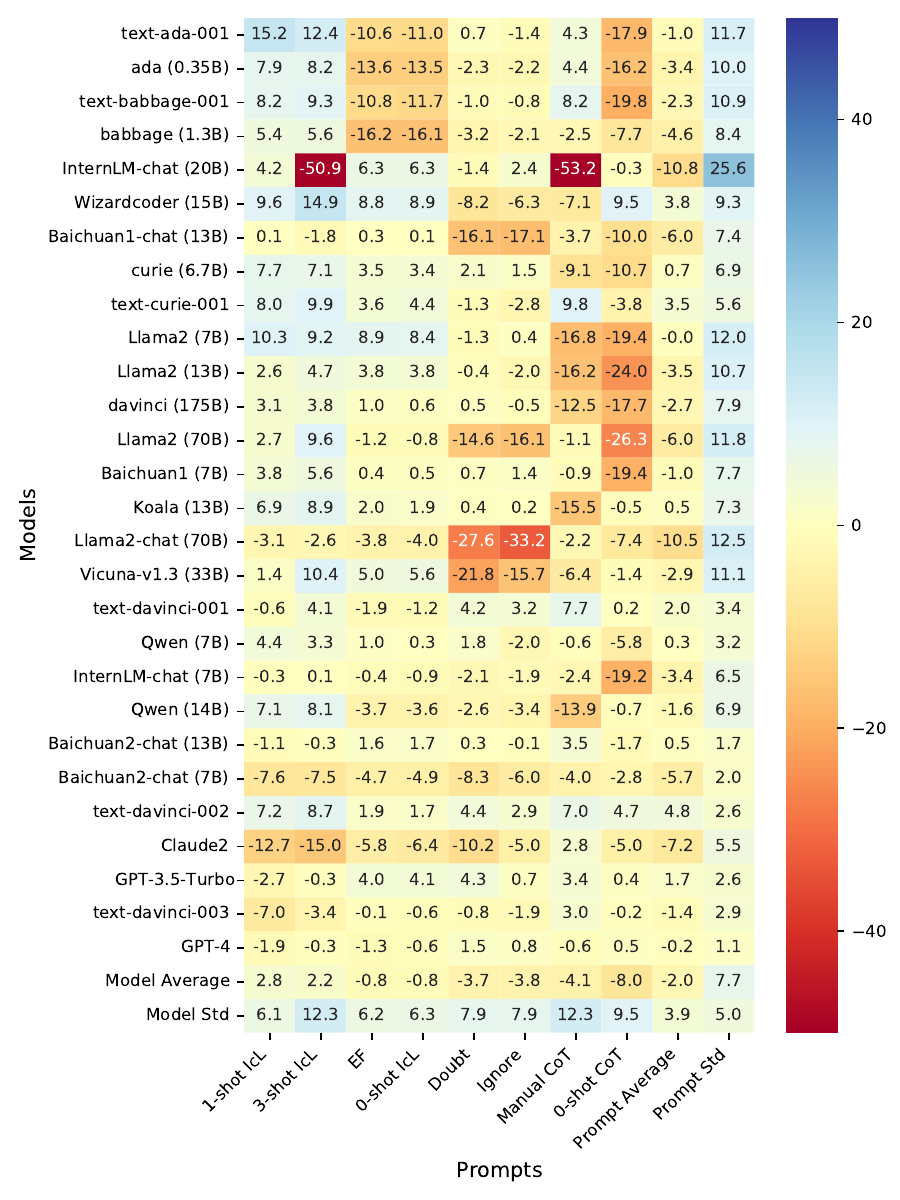}
\end{minipage}
}
\subfigure[\textit{Prompt gain} of ECI-B (MAVEN-ERE)]{
\begin{minipage}{8.5cm}
\centering
\includegraphics[width=.9\linewidth]{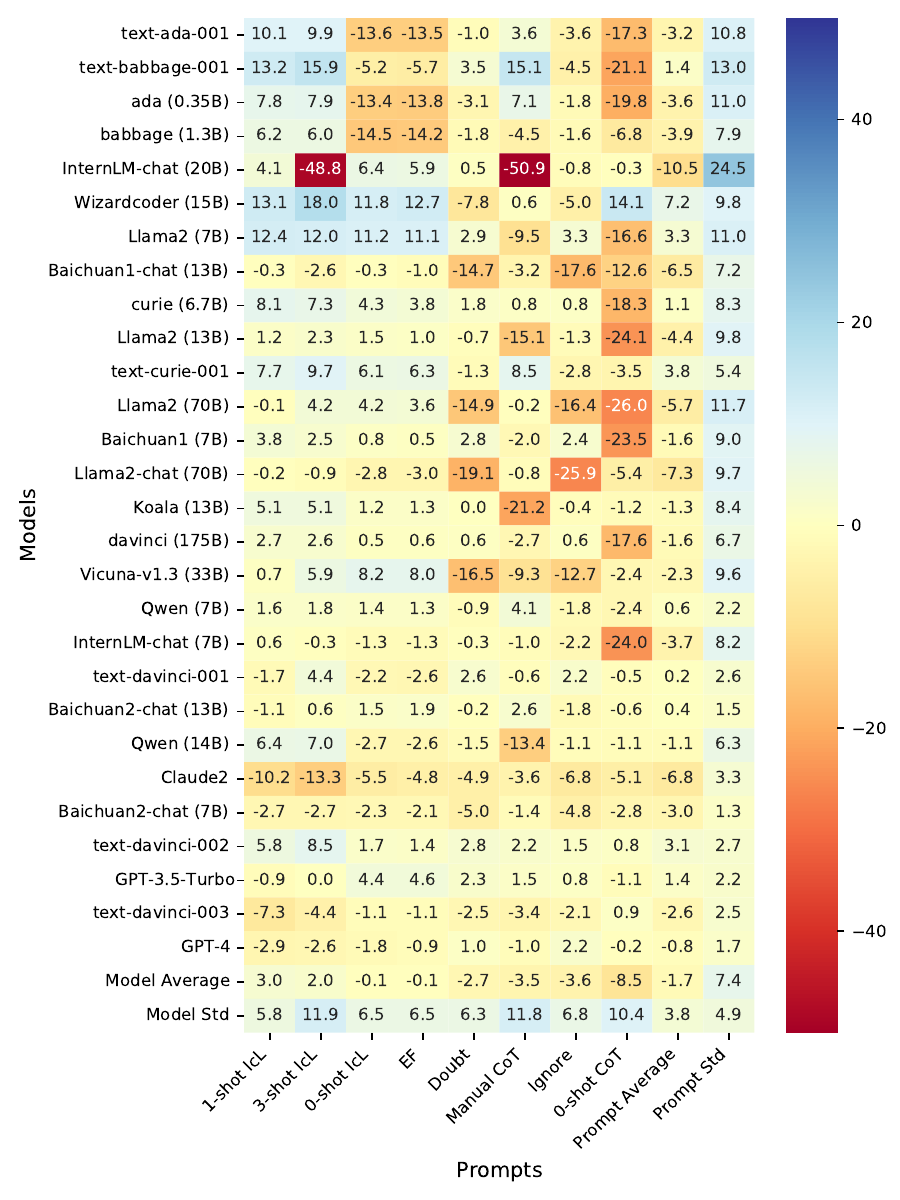}
\end{minipage}
}
\caption[Heatmaps of \textit{prompt gain} of causal tasks in ECI]{\textbf{Heatmaps of \textit{prompt gain} of causal tasks in ECI.} The models and prompts are sorted by their averages.}
\label{fig:Heatmap_of_gain_of_Event_Causality_Identification}
\end{figure}

The distribution of models' accuracy on ECI is shown in Figure \ref{fig:Distribution_of_Event_Causality_Identification_Tasks}. Figure \ref{fig:Heatmap_of_performances_of_Event_Causality_Identification} illustrates how models perform on ECI. The prompt gain (i.e., accuracy improvement against the basic prompt on the model with the used prompt) is demonstrated in Figure \ref{fig:Heatmap_of_gain_of_Event_Causality_Identification}.

\subsubsection{CA}
\begin{figure}
\centering
\subfigure[Model performance of CA-B (FA)]{
\begin{minipage}{8.5cm}
\centering
\includegraphics[width=.9\linewidth]{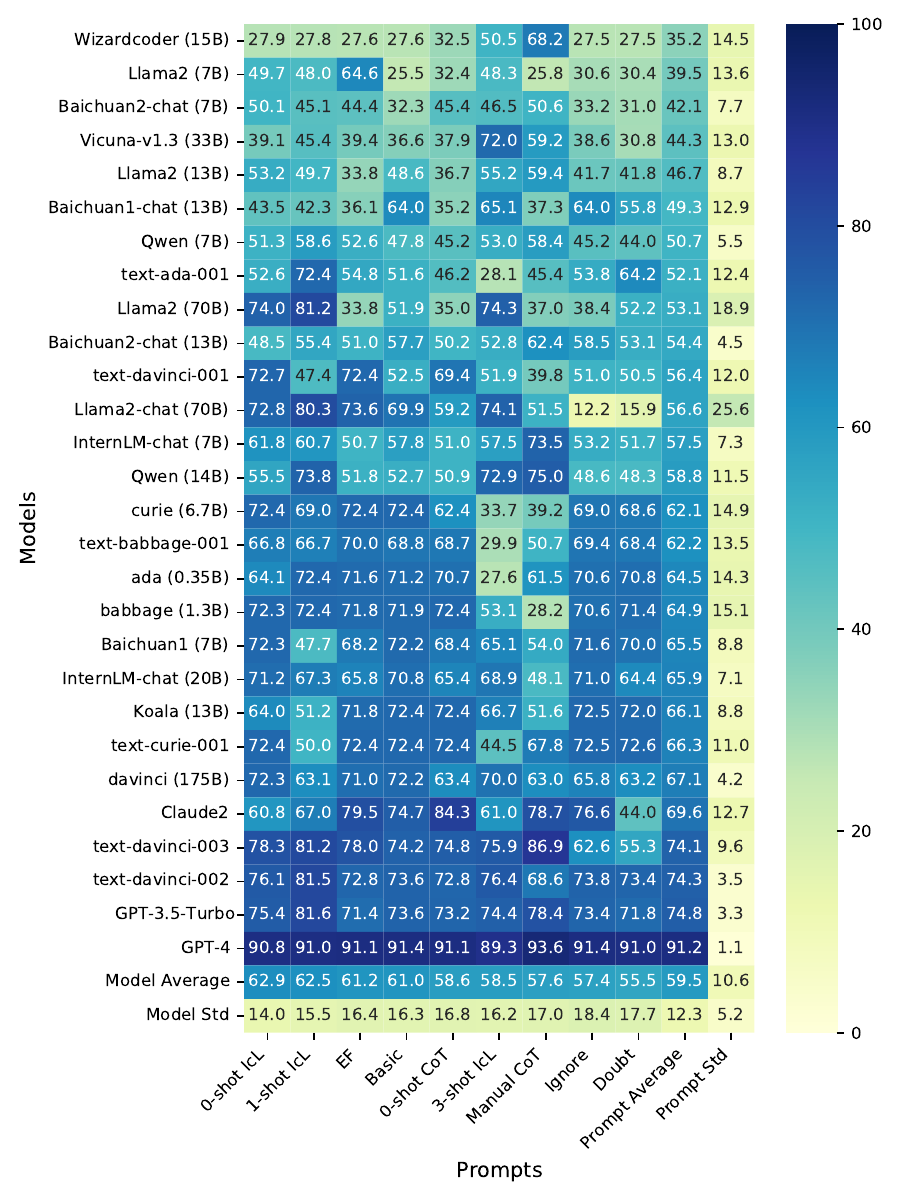}
\end{minipage}
}
\subfigure[Model performance of CA-B (FP)]{
\begin{minipage}{8.5cm}
\centering
\includegraphics[width=.9\linewidth]{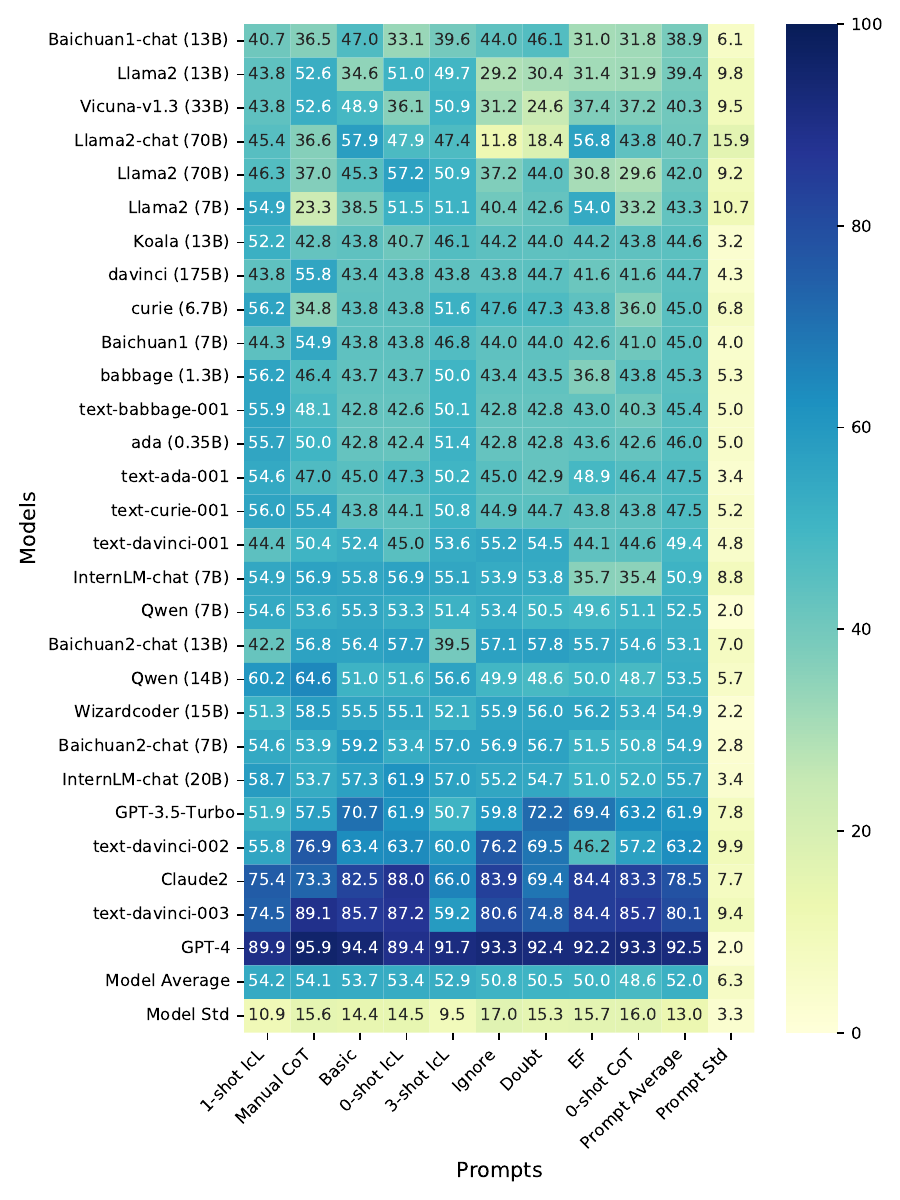}
\end{minipage}
}
\caption[Heatmaps of model performance of causal tasks in CA]{\textbf{Heatmaps of model performance of causal tasks in CA.} The models and prompts are sorted by their averages.}
\label{fig:Heatmap_of_performances_of_Causal_Attribution}
\end{figure}

\begin{figure}
\centering
\subfigure[\textit{Prompt gain} of CA-B (FA)]{
\begin{minipage}{8.5cm}
\centering
\includegraphics[width=.9\linewidth]{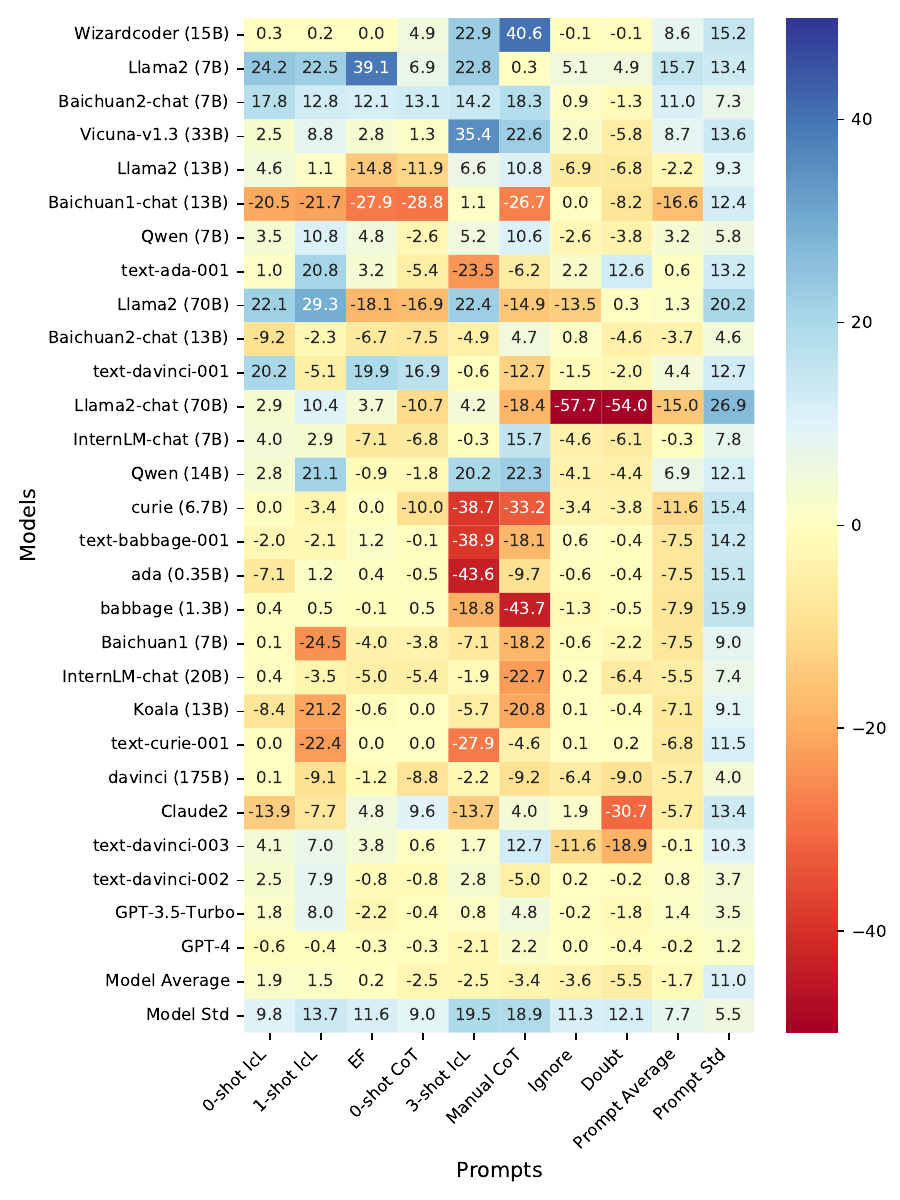}
\end{minipage}
}
\subfigure[\textit{Prompt gain} of CA-B (FP)]{
\begin{minipage}{8.5cm}
\centering
\includegraphics[width=.9\linewidth]{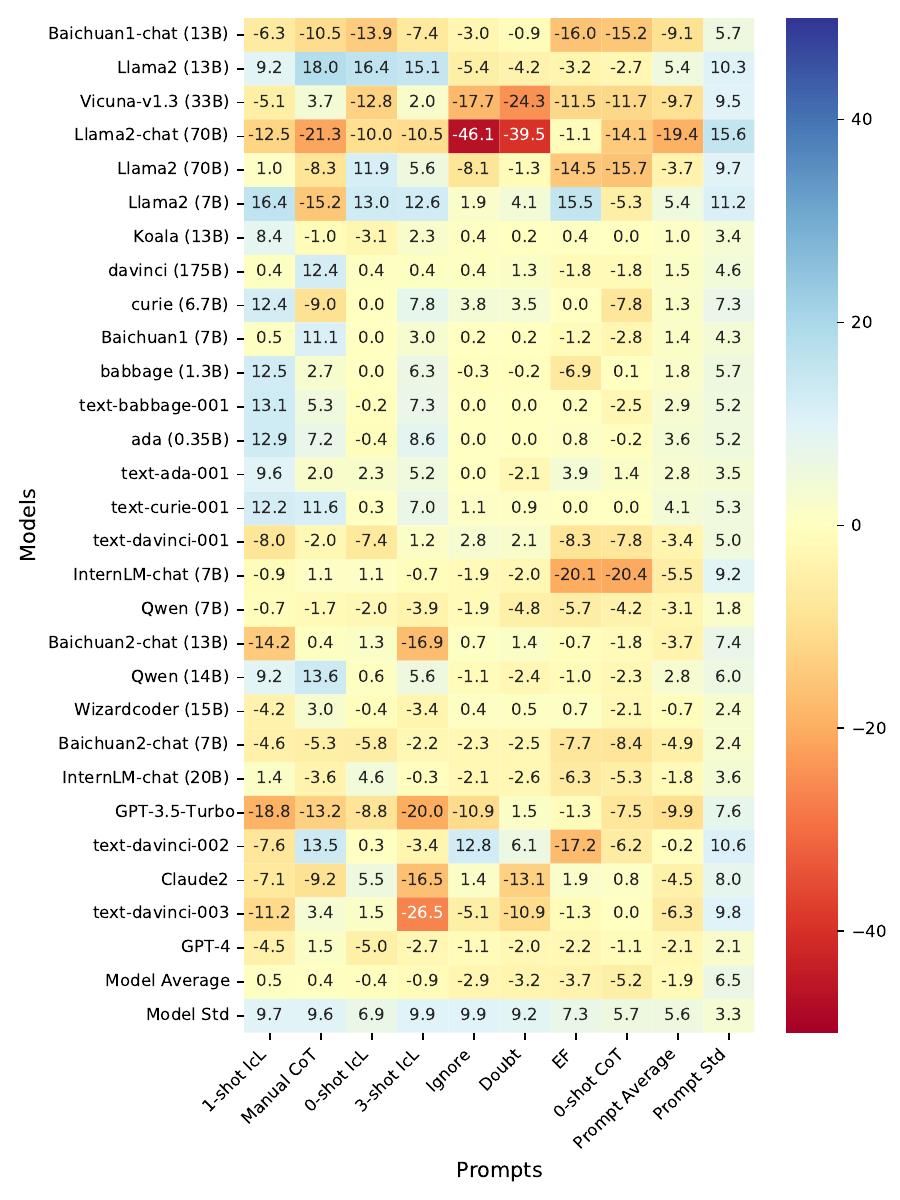}
\end{minipage}
}
\caption[Heatmaps of \textit{prompt gain} of causal tasks in CA]{\textbf{Heatmaps of \textit{prompt gain} of causal tasks in CA.} The models and prompts are sorted by their averages.}
\label{fig:Heatmap_of_gain_of_Causal_Attribution}
\end{figure}

The distribution of models' accuracy on CA is shown in Figure \ref{fig:Distribution_of_Causal_Attribution_Tasks}. Figure \ref{fig:Heatmap_of_performances_of_Causal_Attribution} illustrates how models perform on CA. The prompt gain (i.e., accuracy improvement against the basic prompt on the model with the used prompt) is demonstrated in Figure \ref{fig:Heatmap_of_gain_of_Causal_Attribution}.

\subsection{Intervention}
\subsubsection{ATE}
\begin{figure}
\centering
\subfigure[Model performance of ATE-P (ATE-basic)]{
\begin{minipage}{8.5cm}
\centering
\includegraphics[width=.9\linewidth]{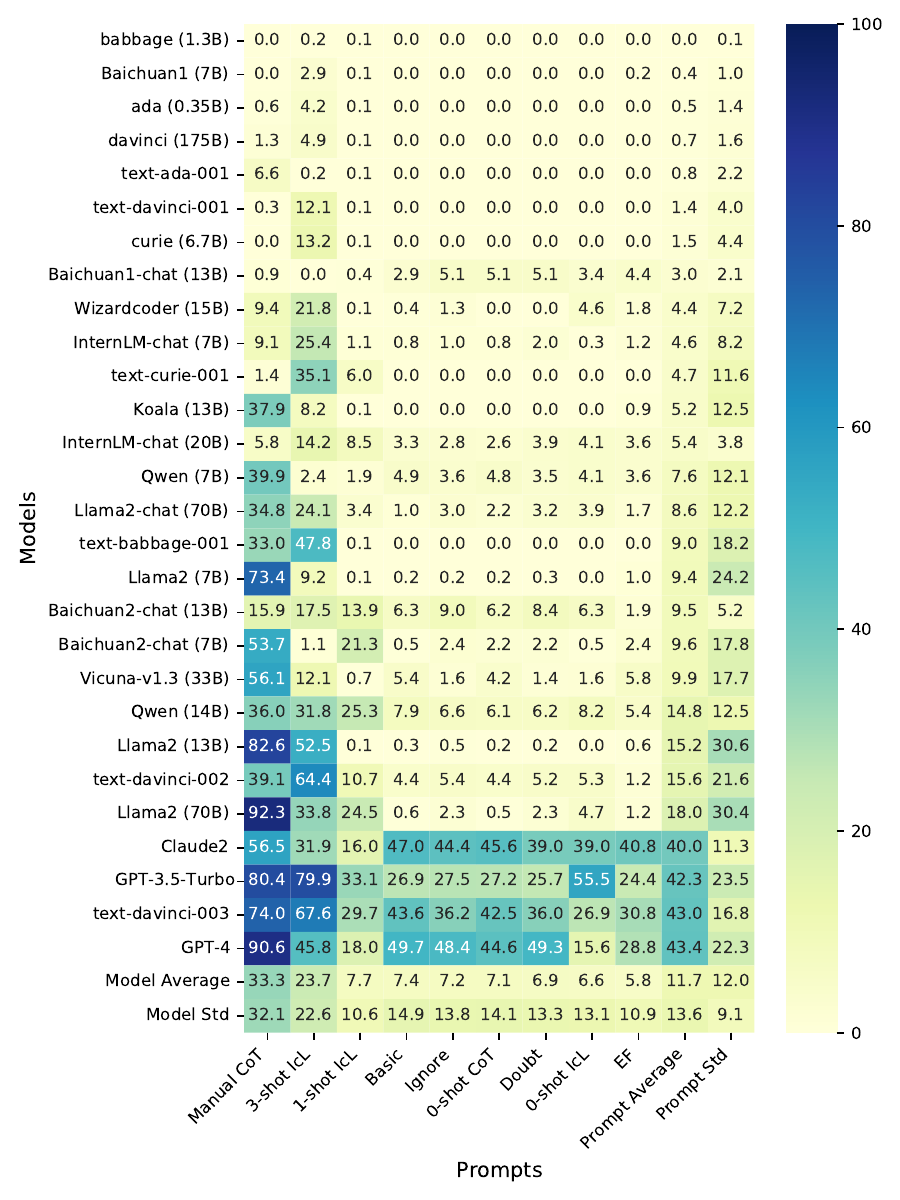}
\end{minipage}
}
\subfigure[Model performance of ATE-P (ATE-hard)]{
\begin{minipage}{8.5cm}
\centering
\includegraphics[width=.9\linewidth]{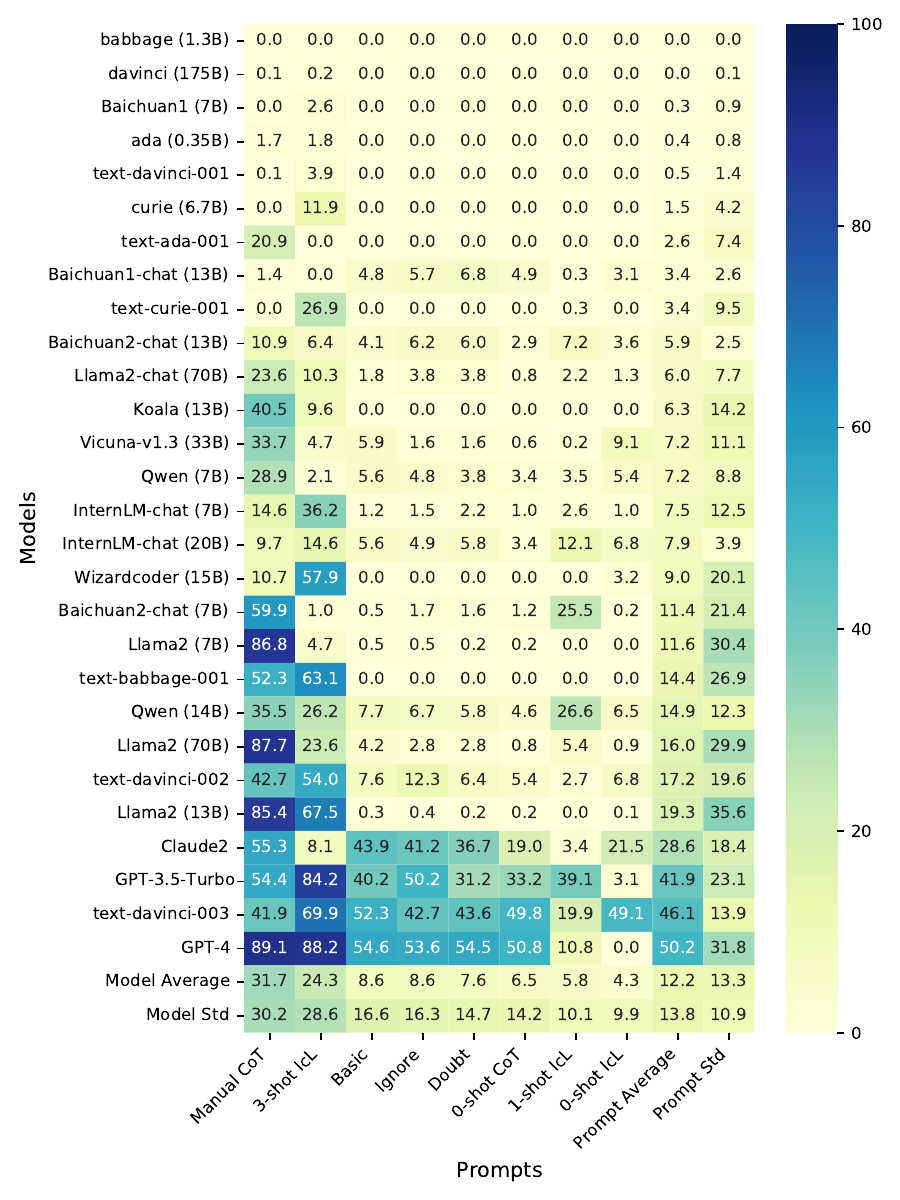}
\end{minipage}
}
\subfigure[Model performance of ATE-B (ATE-natural)]{
\begin{minipage}{8.5cm}
\centering
\includegraphics[width=.9\linewidth]{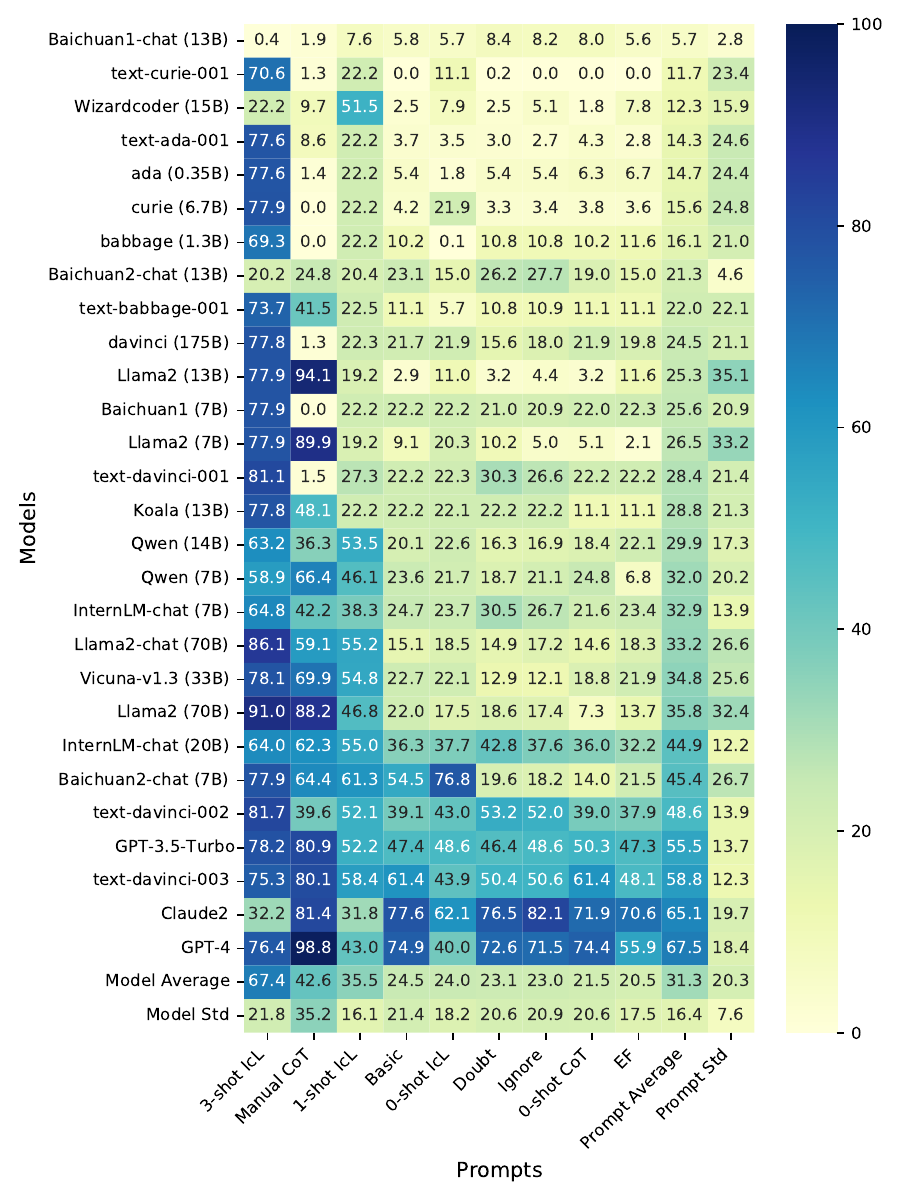}
\end{minipage}
}
\caption[Heatmaps of model performance of causal tasks in ATE]{\textbf{Heatmaps of model performance of causal tasks in ATE.} The models and prompts are sorted by their averages.}
\label{fig:Heatmap_of_performances_of_Average_Treatment_Effect}
\end{figure}

\begin{figure}
\centering
\subfigure[\textit{Prompt gain} of ATE-P (ATE-basic)]{
\begin{minipage}{8.5cm}
\centering
\includegraphics[width=.9\linewidth]{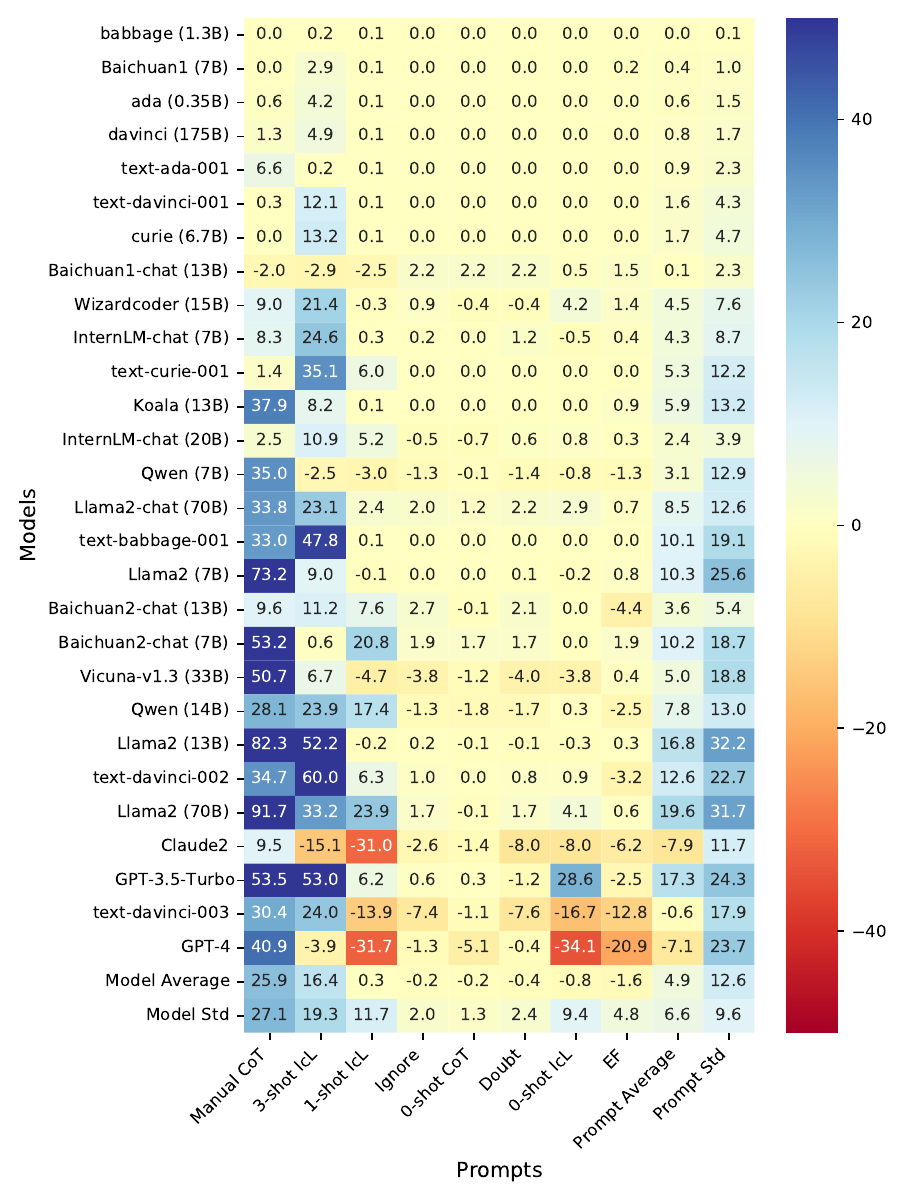}
\end{minipage}
}
\subfigure[\textit{Prompt gain} of ATE-P (ATE-hard)]{
\begin{minipage}{8.5cm}
\centering
\includegraphics[width=.9\linewidth]{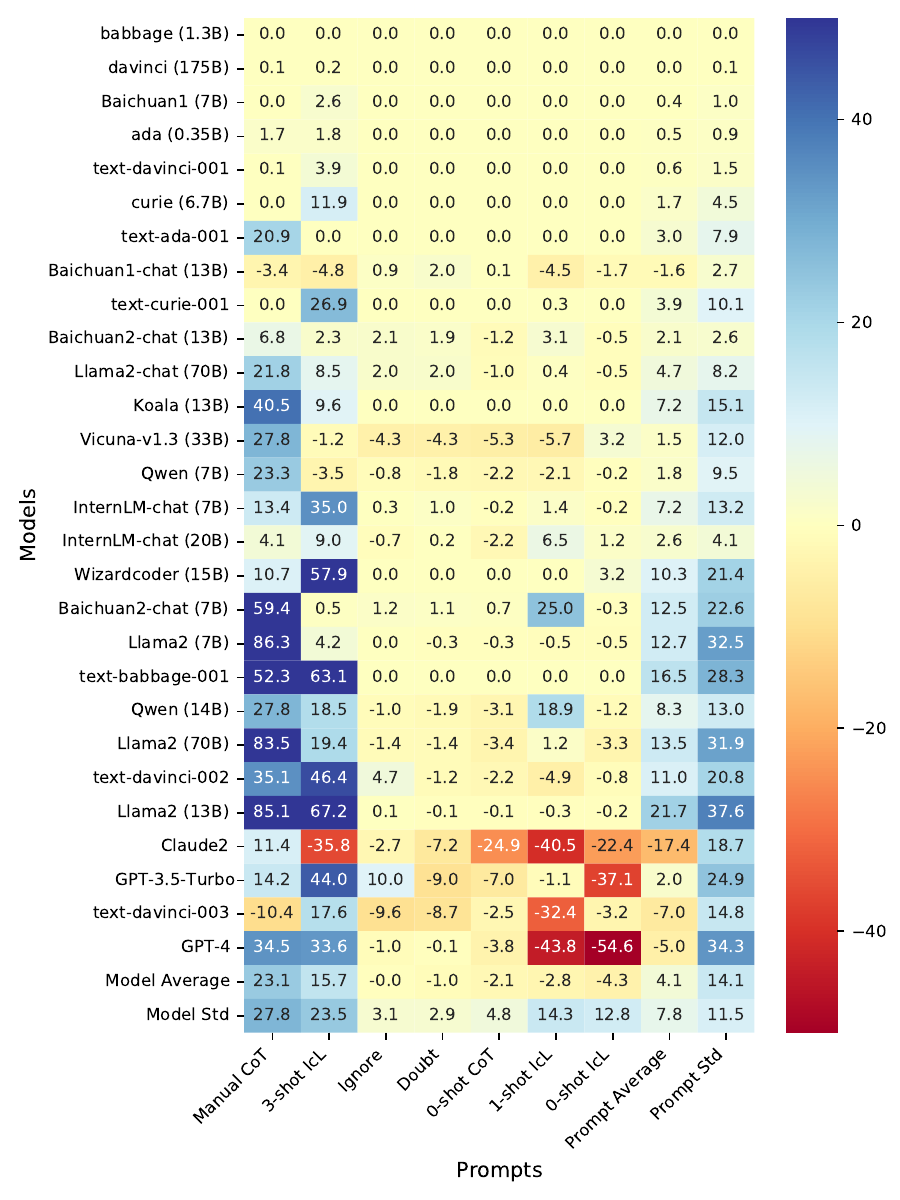}
\end{minipage}
}
\subfigure[\textit{Prompt gain} of ATE-B (ATE-natural)]{
\begin{minipage}{8.5cm}
\centering
\includegraphics[width=.9\linewidth]{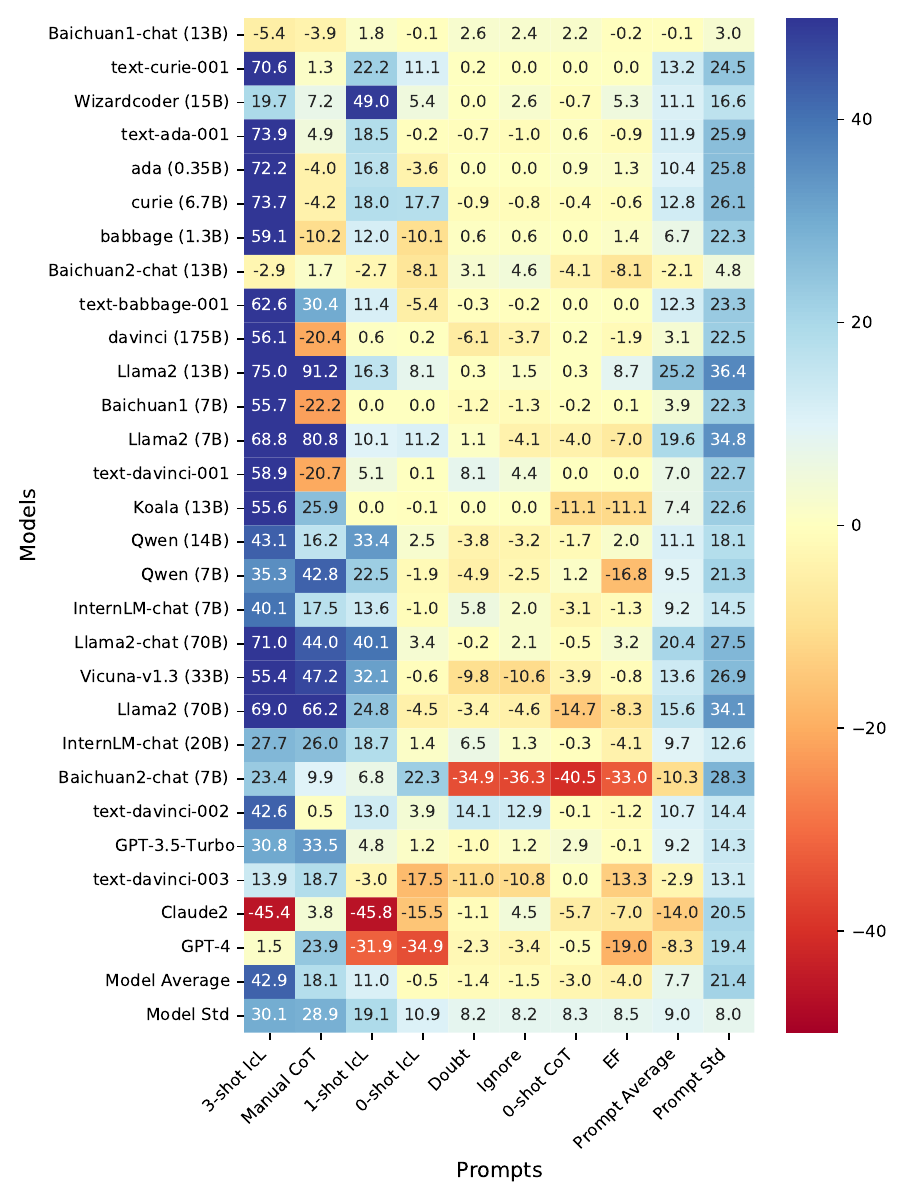}
\end{minipage}
}
\caption[Heatmaps of \textit{prompt gain} of causal tasks in ATE]{\textbf{Heatmaps of \textit{prompt gain} of causal tasks in ATE.} The models and prompts are sorted by their averages.}
\label{fig:Heatmap_of_gain_of_Average_Treatment_Effect}
\end{figure}

The distribution of models' accuracy on ATE is shown in Figure \ref{fig:Distribution_of_Average_Treatment_Effect_Tasks}. Figure \ref{fig:Heatmap_of_performances_of_Average_Treatment_Effect} illustrates how models perform on ATE. The prompt gain (i.e., accuracy improvement against the basic prompt on the model with the used prompt) is demonstrated in Figure \ref{fig:Heatmap_of_gain_of_Average_Treatment_Effect}.

\subsubsection{CDE}
\begin{figure}
\centering
\subfigure[Model performance of CDE-P (CDE-basic)]{
\begin{minipage}{8.5cm}
\centering
\includegraphics[width=.9\linewidth]{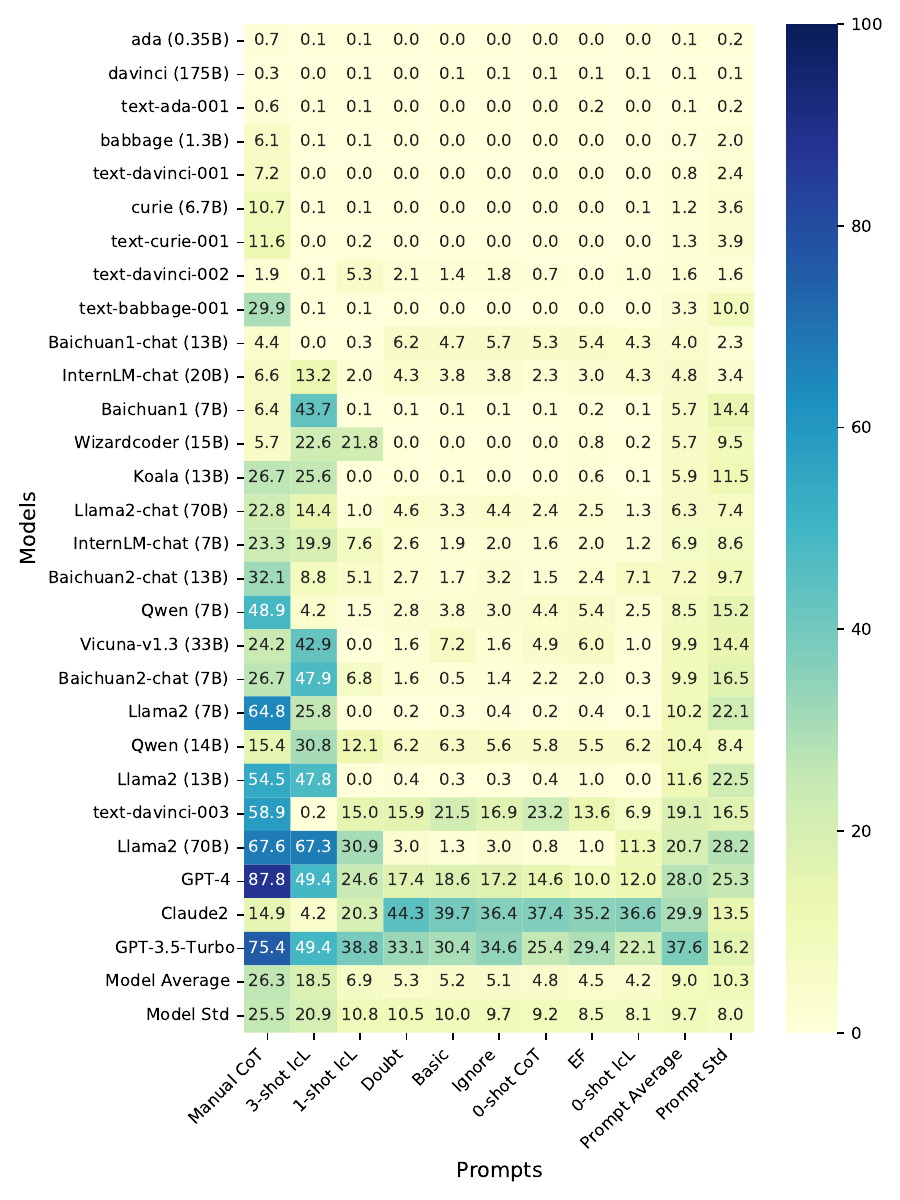}
\end{minipage}
}
\subfigure[Model performance of CDE-P (CDE-hard)]{
\begin{minipage}{8.5cm}
\centering
\includegraphics[width=.9\linewidth]{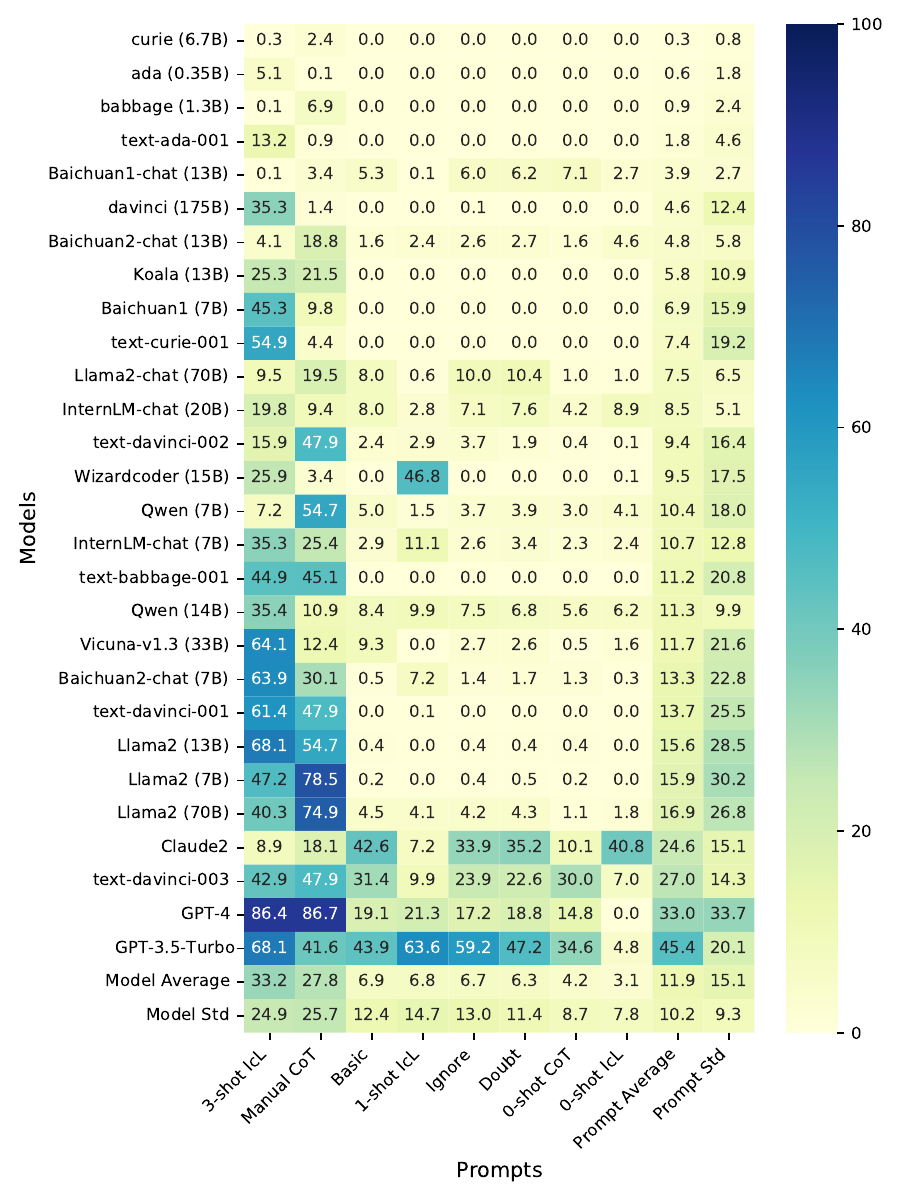}
\end{minipage}
}
\subfigure[Model performance of CDE-B (CDE-natural)]{
\begin{minipage}{8.5cm}
\centering
\includegraphics[width=.9\linewidth]{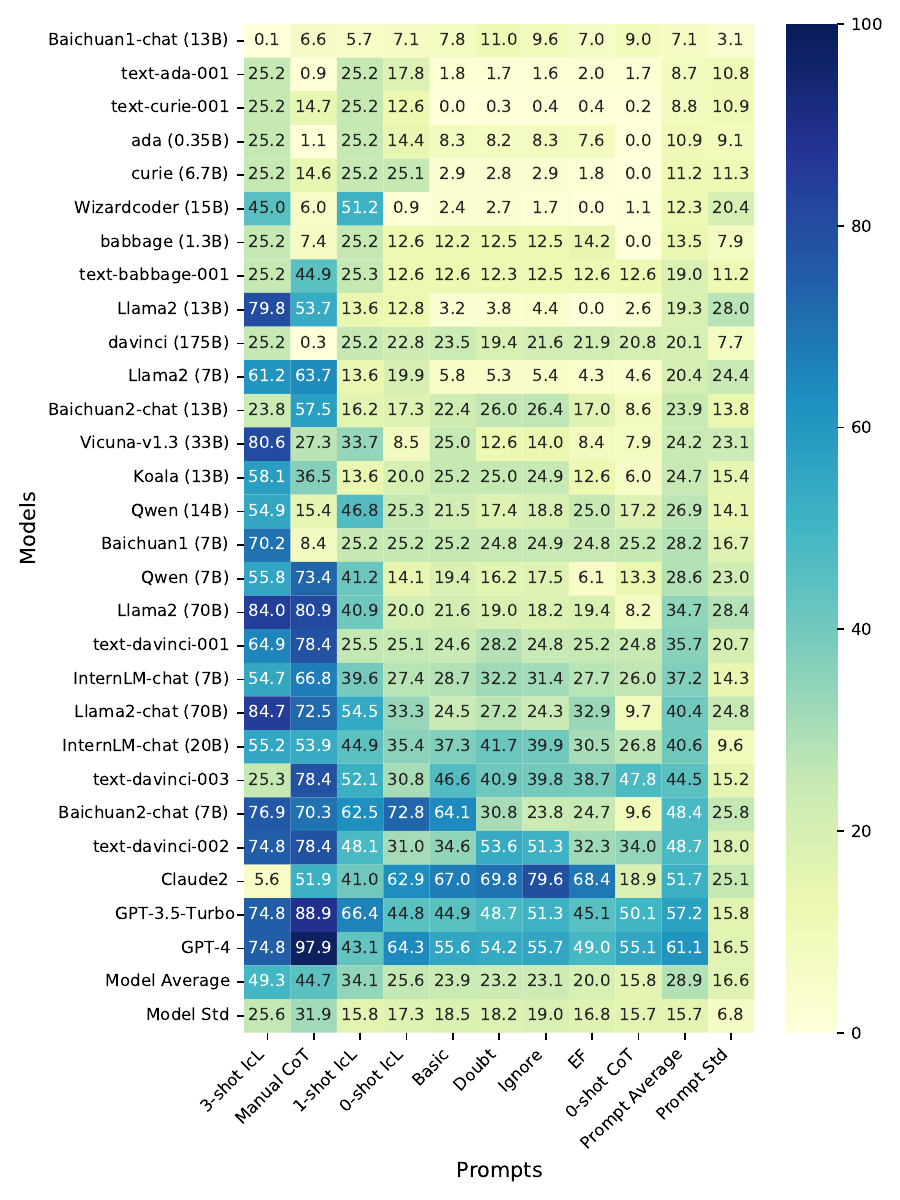}
\end{minipage}
}
\caption[Heatmaps of model performance of causal tasks in CDE]{\textbf{Heatmaps of model performance of causal tasks in CDE.} The models and prompts are sorted by their averages.}
\label{fig:Heatmap_of_performances_of_Controlled_Direct_Effect}
\end{figure}

\begin{figure}
\centering
\subfigure[\textit{Prompt gain} of CDE-P (CDE-basic)]{
\begin{minipage}{8.5cm}
\centering
\includegraphics[width=.9\linewidth]{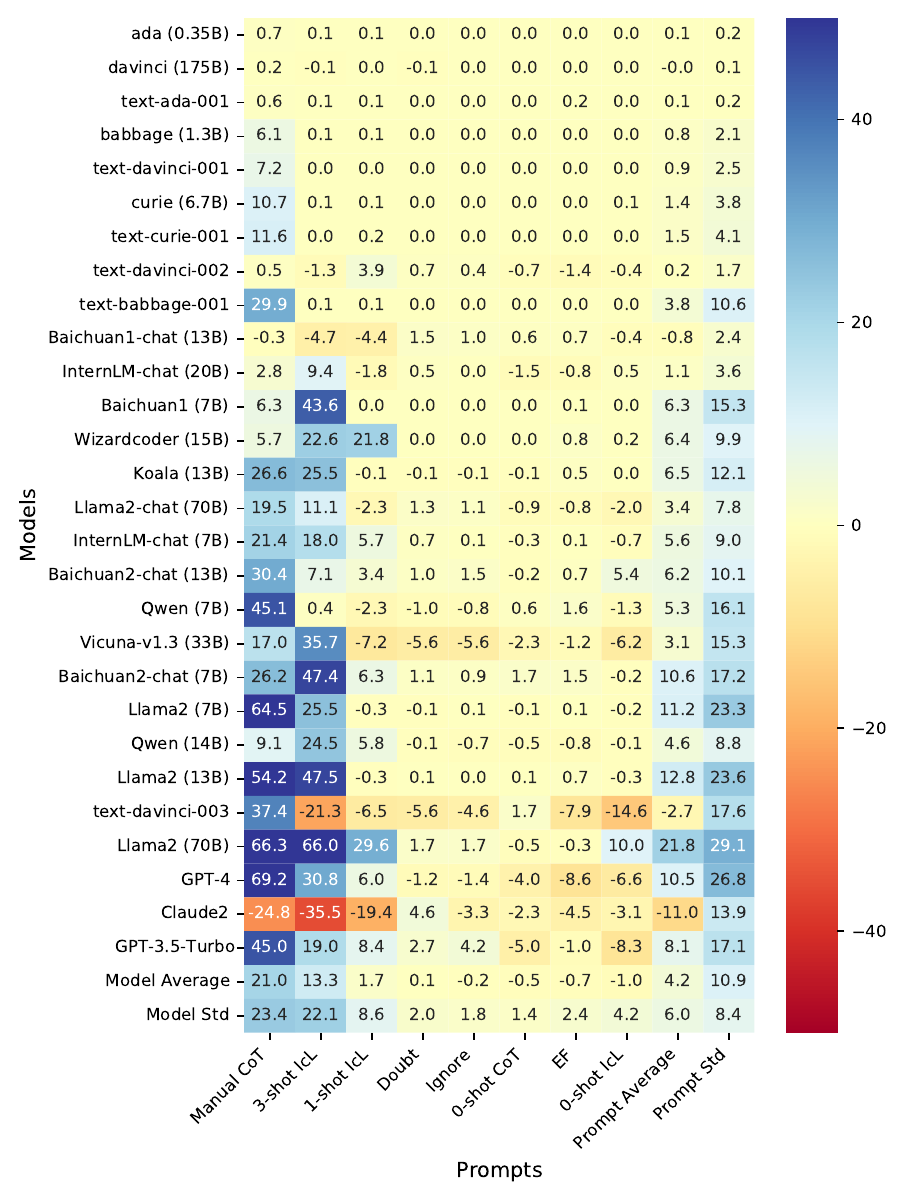}
\end{minipage}
}
\subfigure[\textit{Prompt gain} of CDE-P (CDE-hard)]{
\begin{minipage}{8.5cm}
\centering
\includegraphics[width=.9\linewidth]{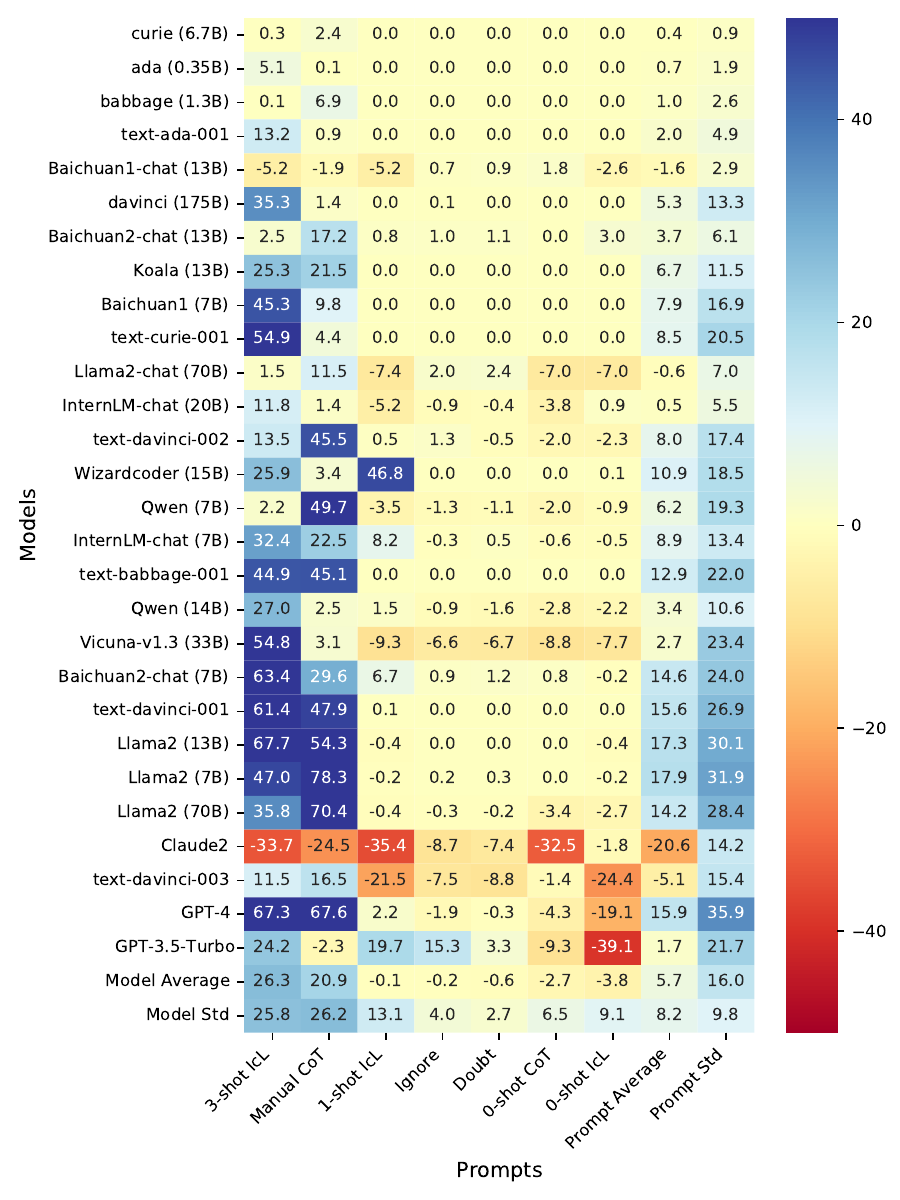}
\end{minipage}
}
\subfigure[\textit{Prompt gain} of CDE-B (CDE-natural)]{
\begin{minipage}{8.5cm}
\centering
\includegraphics[width=.9\linewidth]{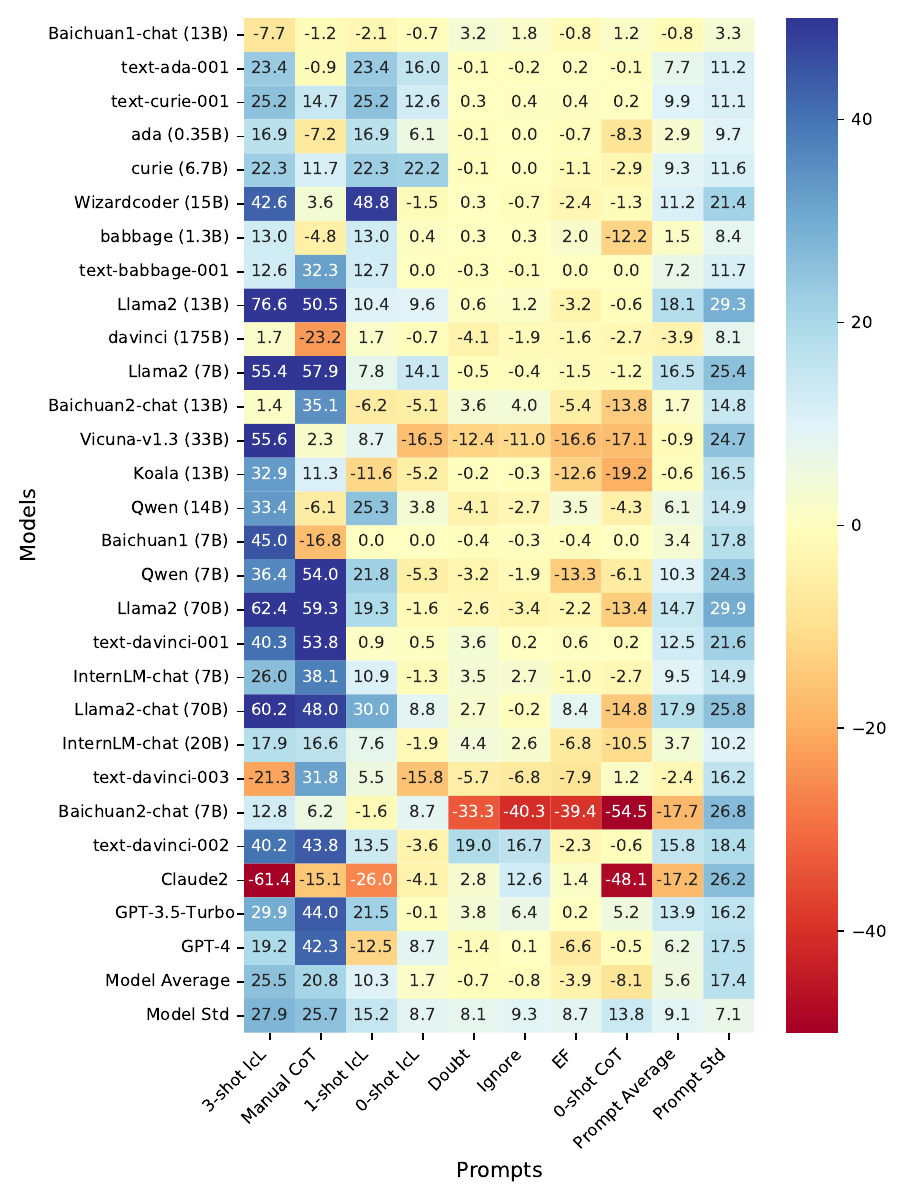}
\end{minipage}
}
\caption[Heatmaps of \textit{prompt gain} of causal tasks in CDE]{\textbf{Heatmaps of \textit{prompt gain} of causal tasks in CDE.} The models and prompts are sorted by their averages.}
\label{fig:Heatmap_of_gain_of_Controlled_Direct_Effect}
\end{figure}

The distribution of models' accuracy on CDE is shown in Figure \ref{fig:Distribution_of_Controlled_Direct_Effect_Tasks}. Figure \ref{fig:Heatmap_of_performances_of_Controlled_Direct_Effect} illustrates how models perform on CDE. The prompt gain (i.e., accuracy improvement against the basic prompt on the model with the used prompt) is demonstrated in Figure \ref{fig:Heatmap_of_gain_of_Controlled_Direct_Effect}.

\subsubsection{CEI}
\begin{figure}
\centering
\subfigure[Model performance of CEI-B (0.2-UC)]{
\begin{minipage}{8.5cm}
\centering
\includegraphics[width=.9\linewidth]{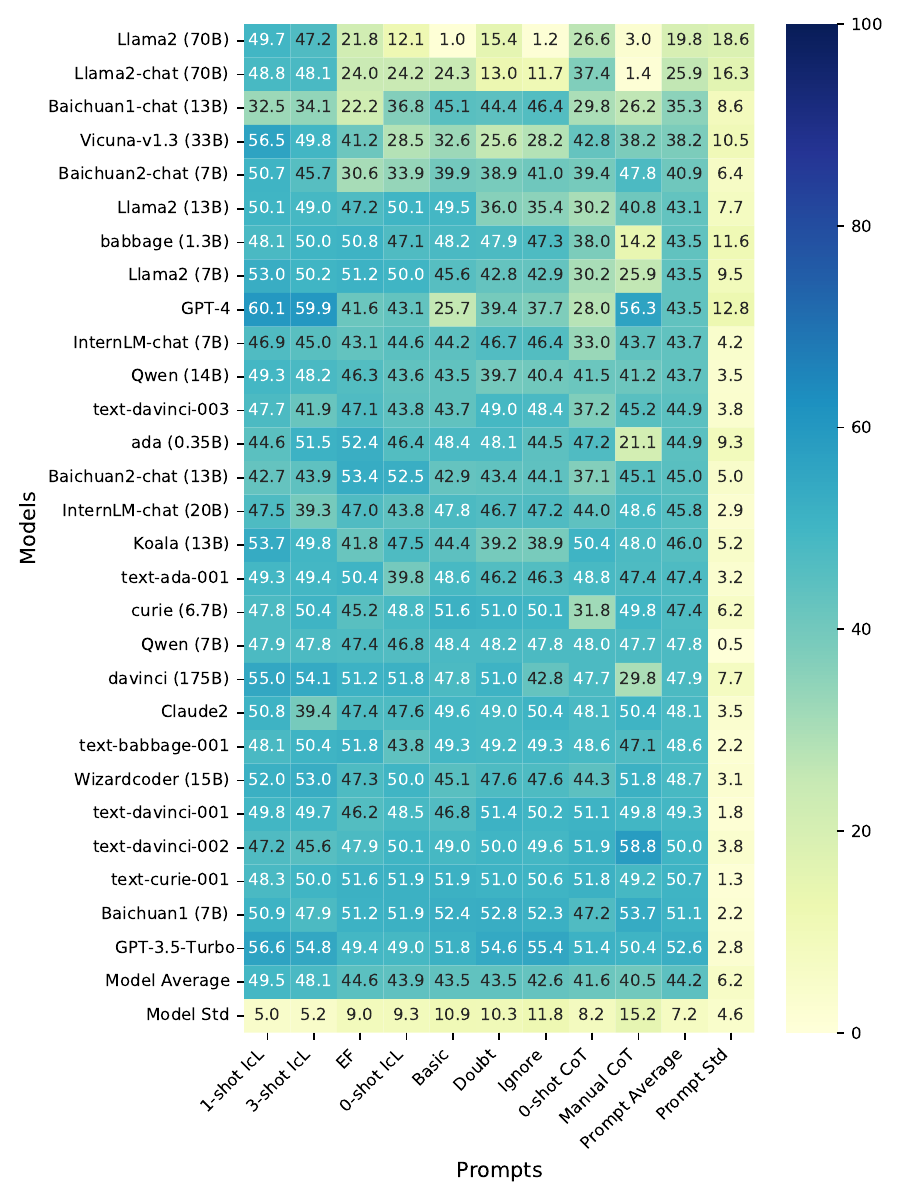}
\end{minipage}
}
\subfigure[Model performance of CEI-B (0.4-UC)]{
\begin{minipage}{8.5cm}
\centering
\includegraphics[width=.9\linewidth]{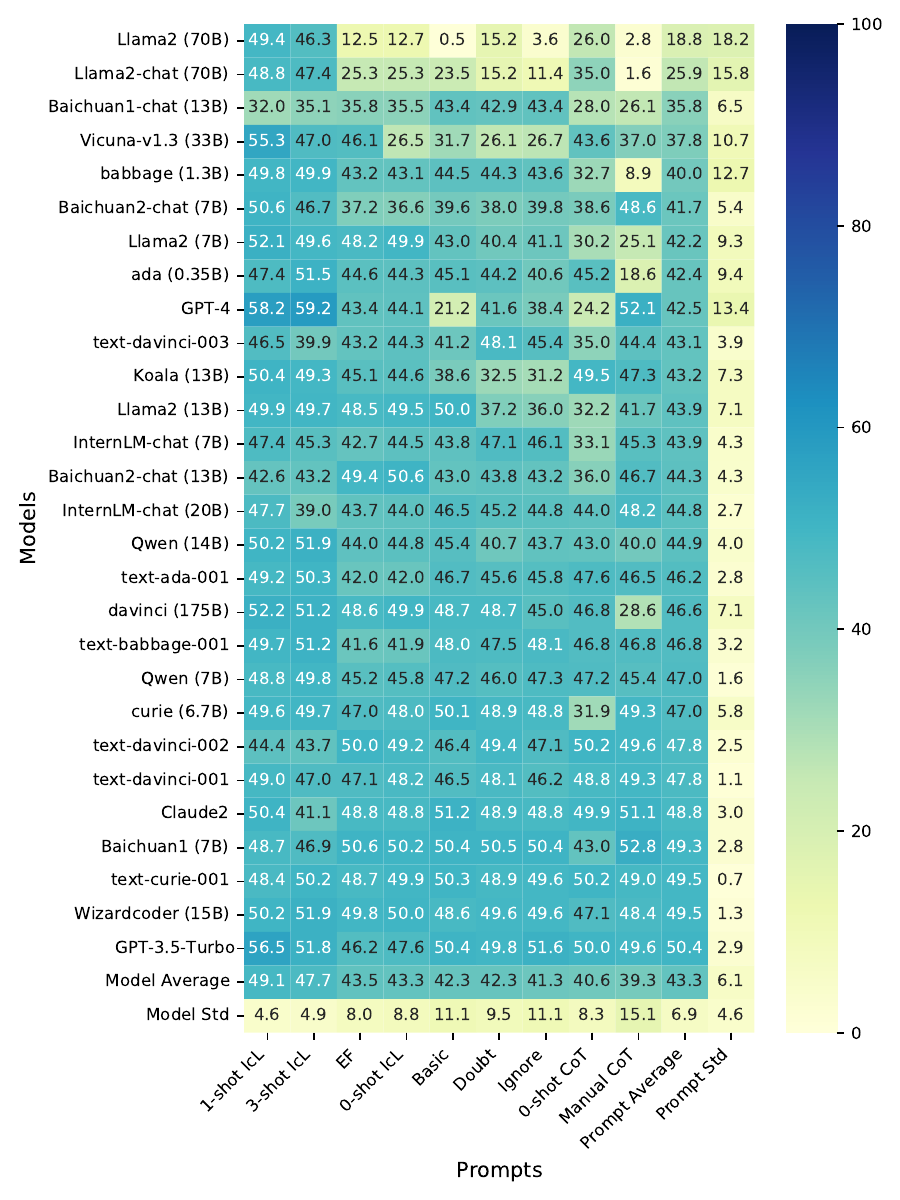}
\end{minipage}
}
\subfigure[Model performance of CEI-B (0.6-UC)]{
\begin{minipage}{8.5cm}
\centering
\includegraphics[width=.9\linewidth]{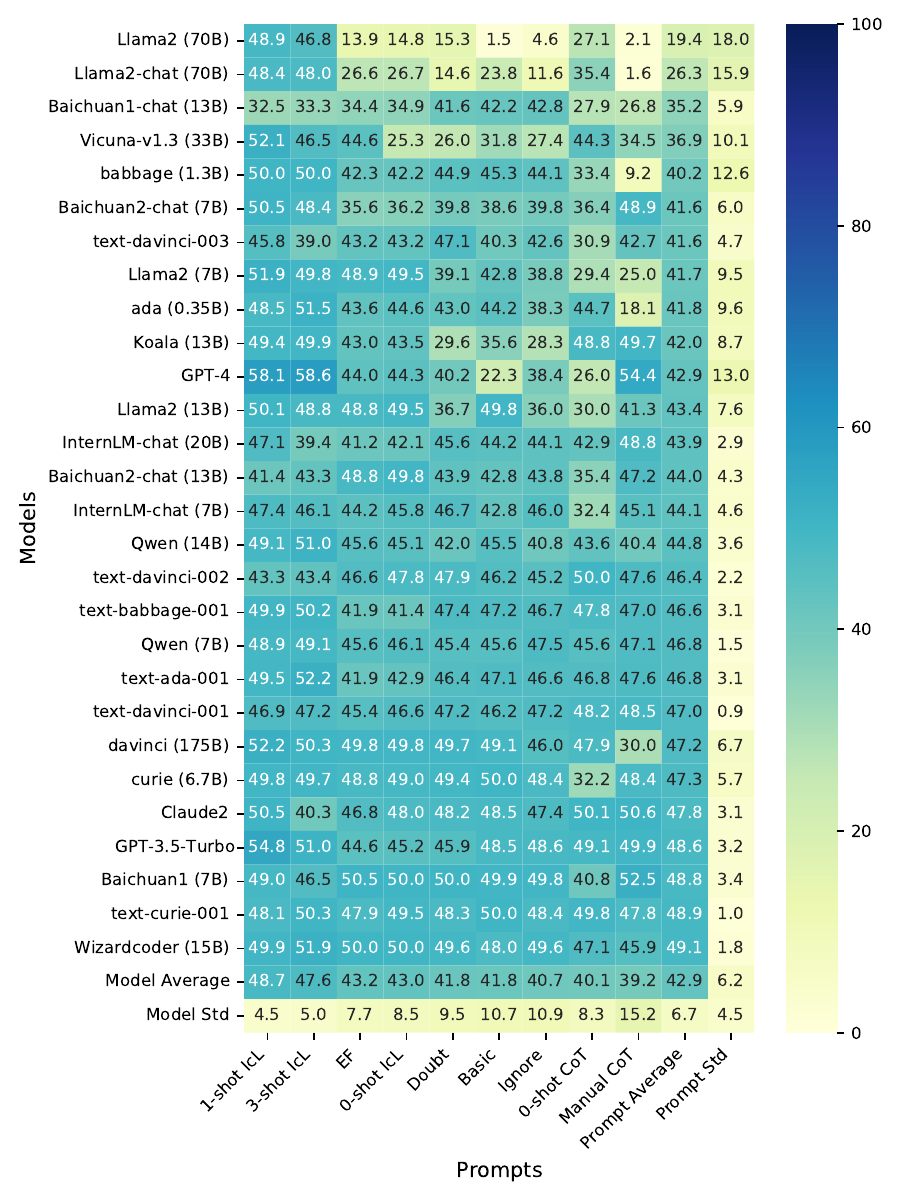}
\end{minipage}
}
\subfigure[Model performance of CEI-B (0.8-UC)]{
\begin{minipage}{8.5cm}
\centering
\includegraphics[width=.9\linewidth]{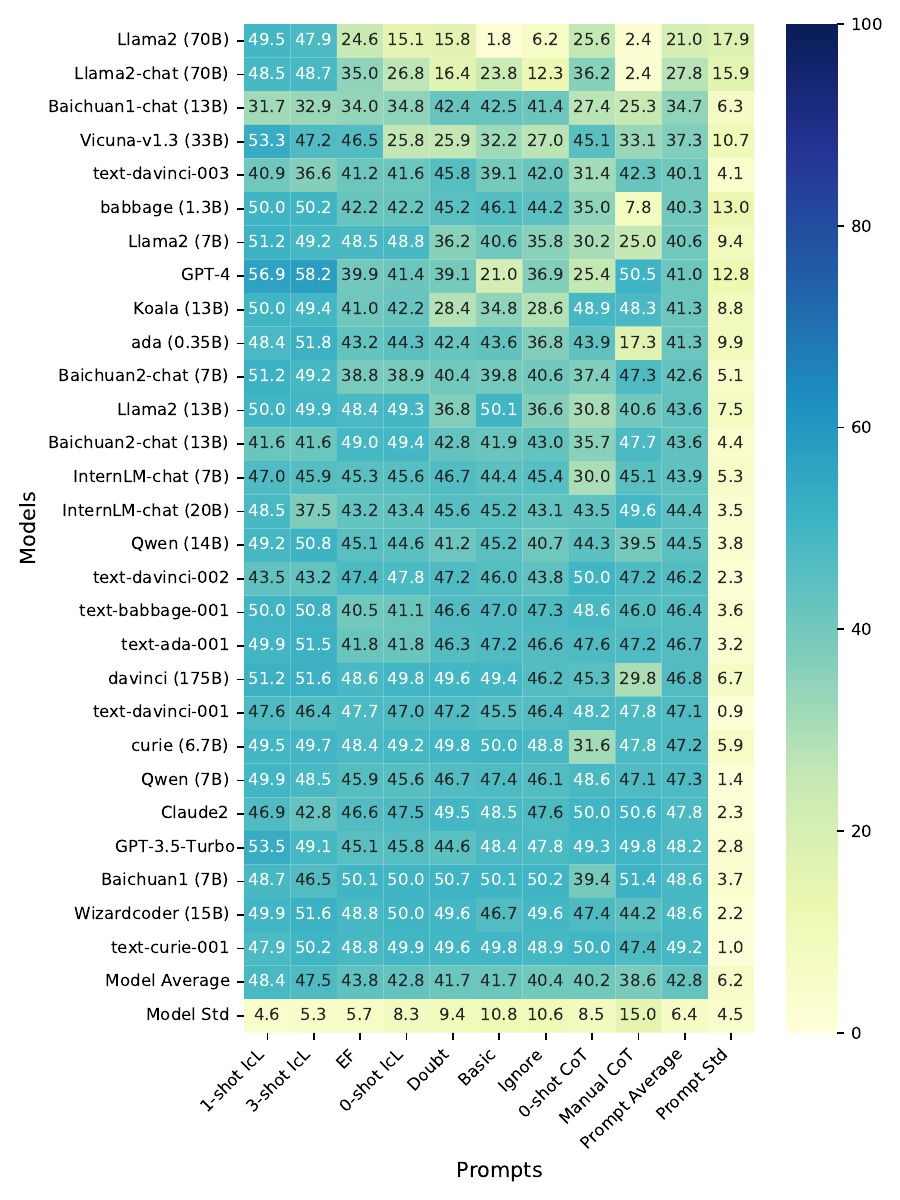}
\end{minipage}
}
\caption[Heatmaps of model performance of causal tasks in CEI]{\textbf{Heatmaps of model performance of causal tasks in CEI.} The models and prompts are sorted by their averages.}
\label{fig:Heatmap_of_performances_of_Causal_Effect_Identification}
\end{figure}

\begin{figure}
\centering
\subfigure[\textit{Prompt gain} of CEI-B (0.2-UC)]{
\begin{minipage}{8.5cm}
\centering
\includegraphics[width=.9\linewidth]{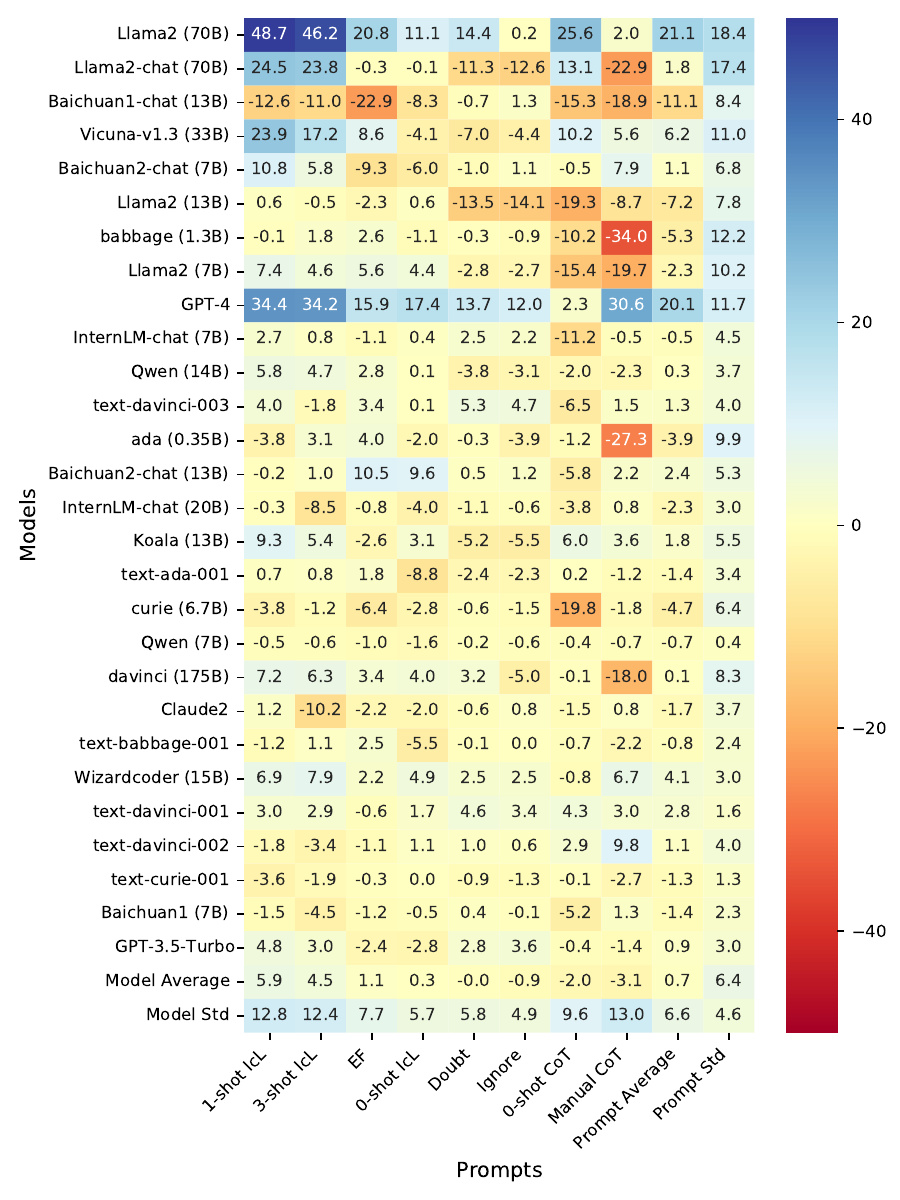}
\end{minipage}
}
\subfigure[\textit{Prompt gain} of CEI-B (0.4-UC)]{
\begin{minipage}{8.5cm}
\centering
\includegraphics[width=.9\linewidth]{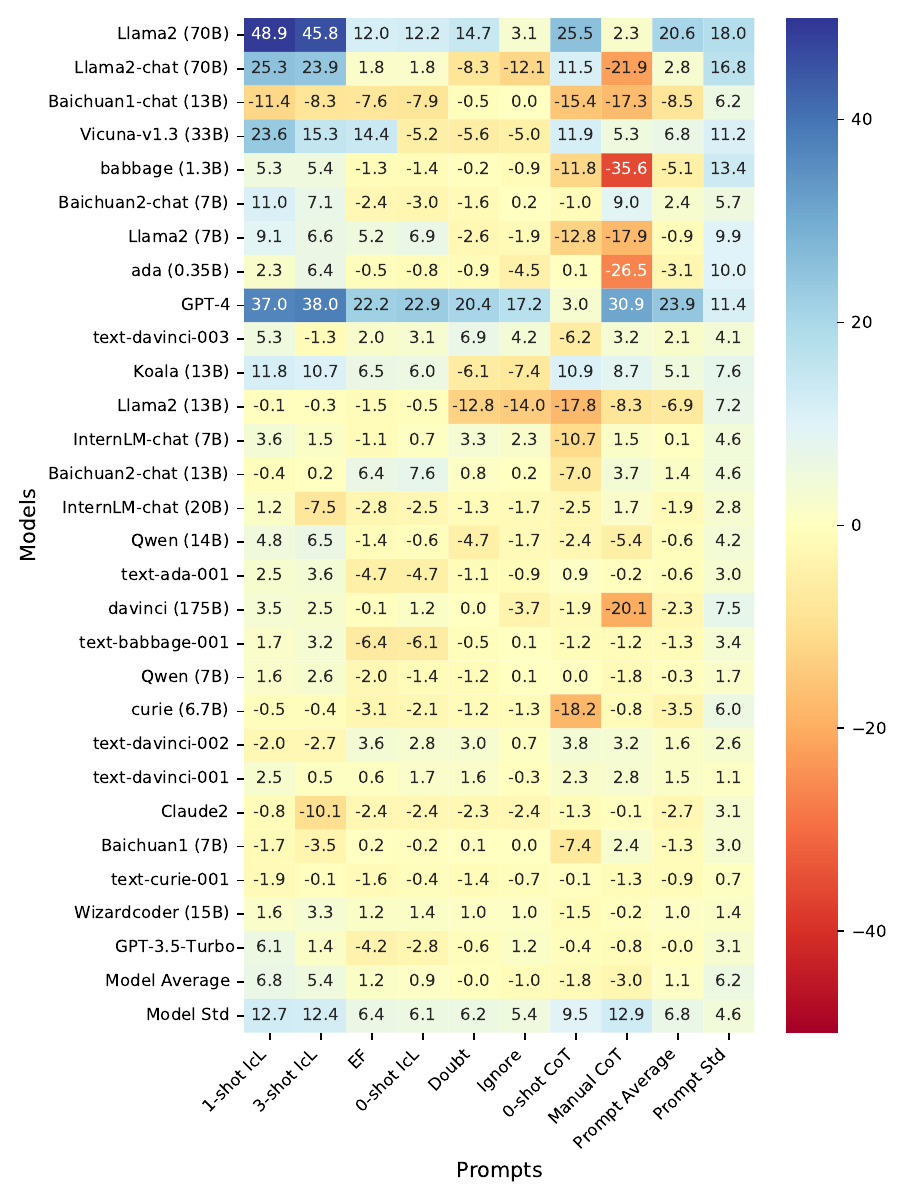}
\end{minipage}
}
\subfigure[\textit{Prompt gain} of CEI-B (0.6-UC)]{
\begin{minipage}{8.5cm}
\centering
\includegraphics[width=.9\linewidth]{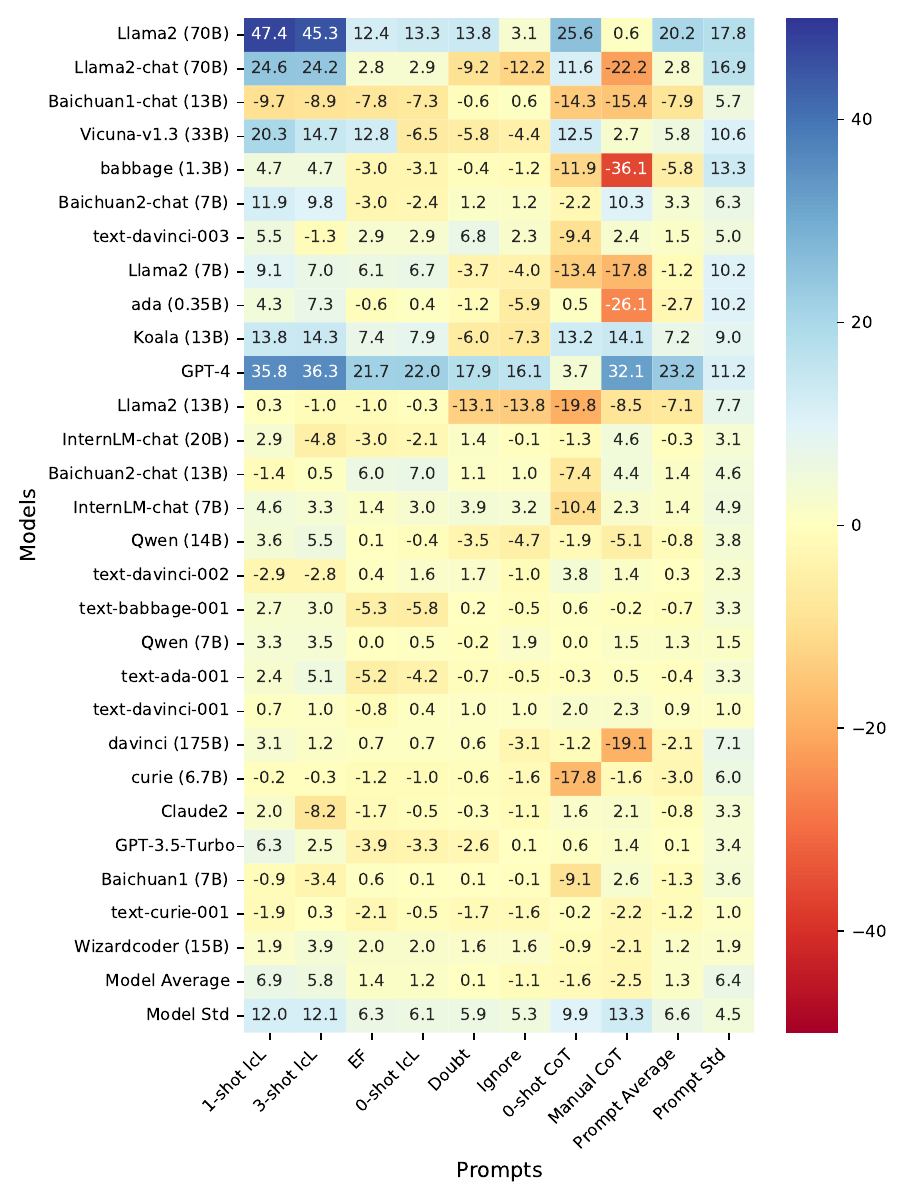}
\end{minipage}
}
\subfigure[\textit{Prompt gain} of CEI-B (0.8-UC)]{
\begin{minipage}{8.5cm}
\centering
\includegraphics[width=.9\linewidth]{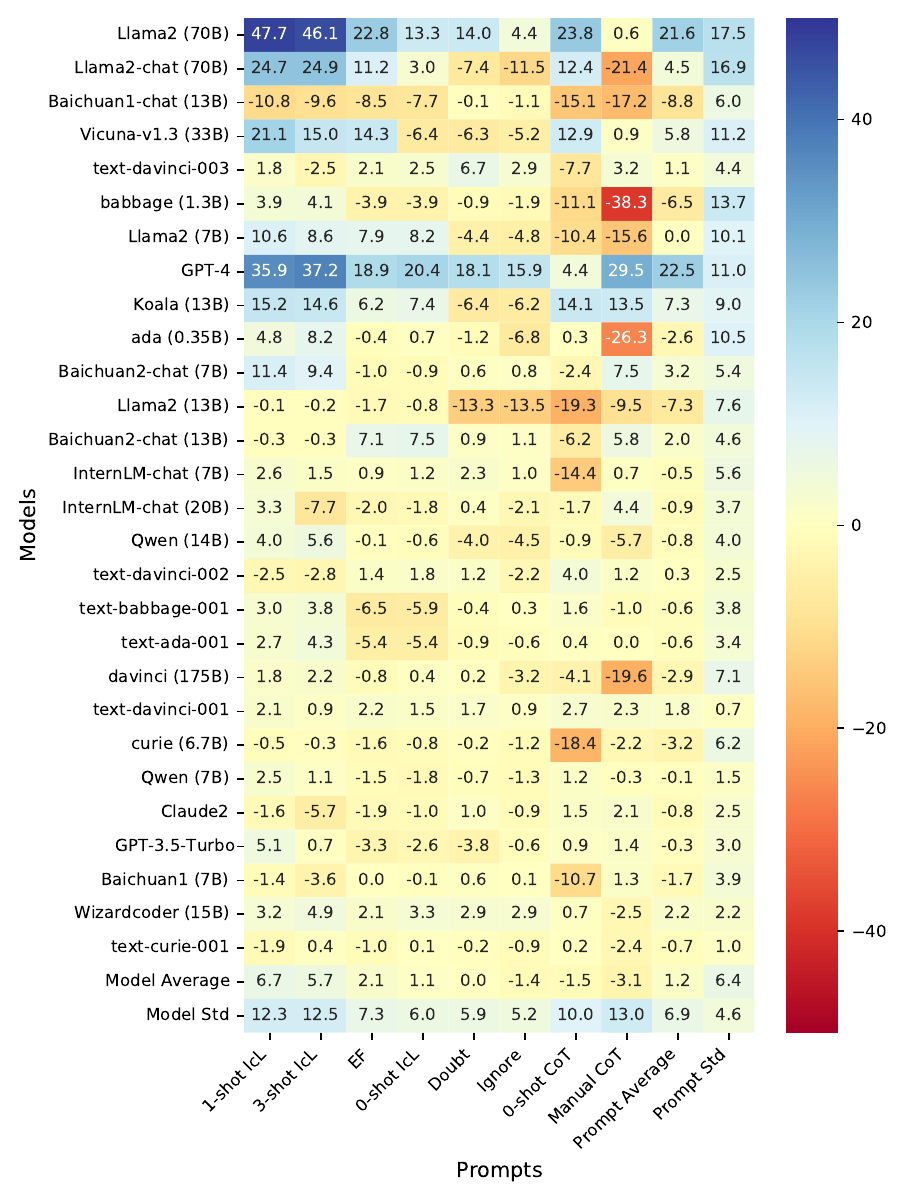}
\end{minipage}
}
\caption[Heatmaps of \textit{prompt gain} of causal tasks in CEI]{\textbf{Heatmaps of \textit{prompt gain} of causal tasks in CEI.} The models and prompts are sorted by their averages.}
\label{fig:Heatmap_of_gain_of_Causal_Effect_Identification}
\end{figure}

The distribution of models' accuracy on CEI is shown in Figure \ref{fig:Distribution_of_Causal_Effect_Identification_Tasks}. Figure \ref{fig:Heatmap_of_performances_of_Causal_Effect_Identification} illustrates how models perform on CEI. The prompt gain (i.e., accuracy improvement against the basic prompt on the model with the used prompt) is demonstrated in Figure \ref{fig:Heatmap_of_gain_of_Causal_Effect_Identification}.

\subsubsection{BAS}
\begin{figure}
\centering
\subfigure[Model performance of BAS-B (backadj)]{
\begin{minipage}{8.5cm}
\centering
\includegraphics[width=.9\linewidth]{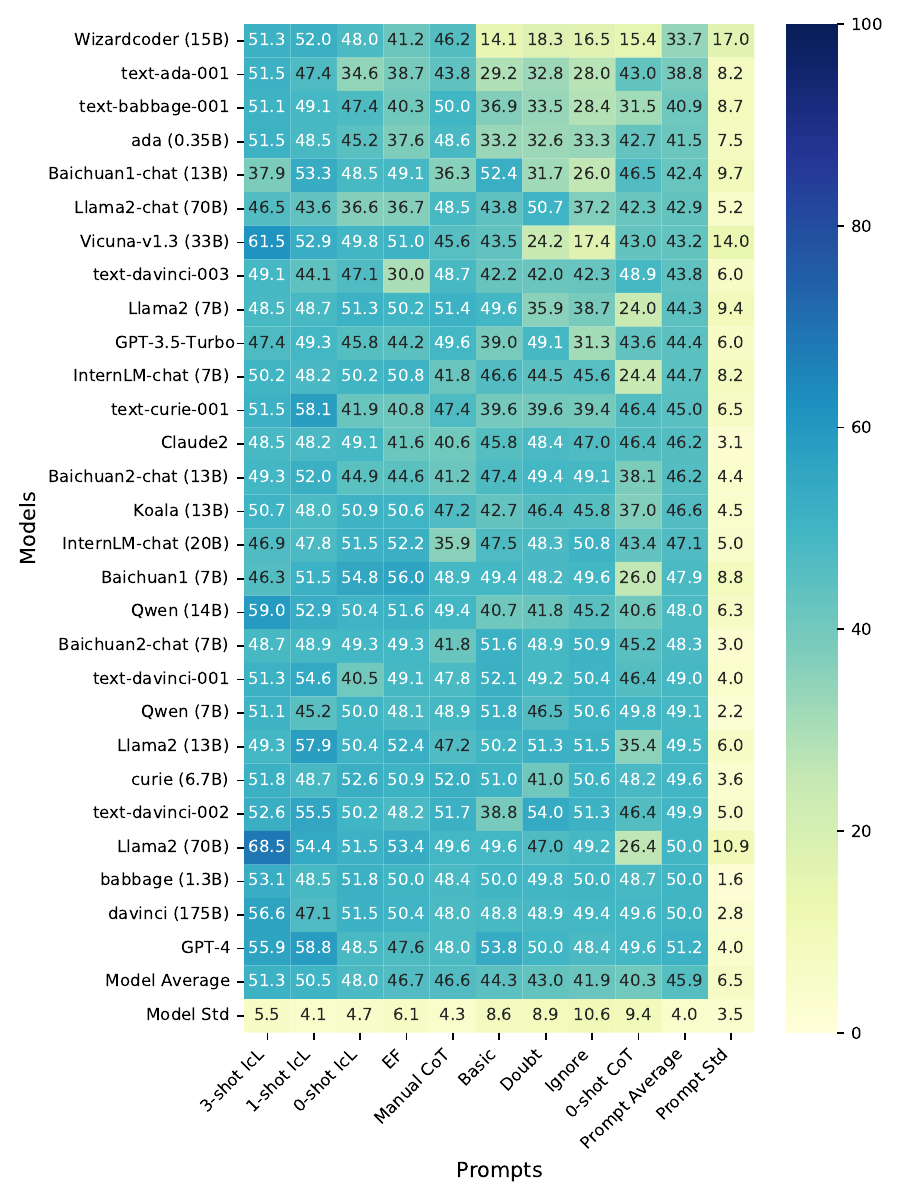}
\end{minipage}
}
\subfigure[Model performance of BAS-C (max-BAS)]{
\begin{minipage}{8.5cm}
\centering
\includegraphics[width=.9\linewidth]{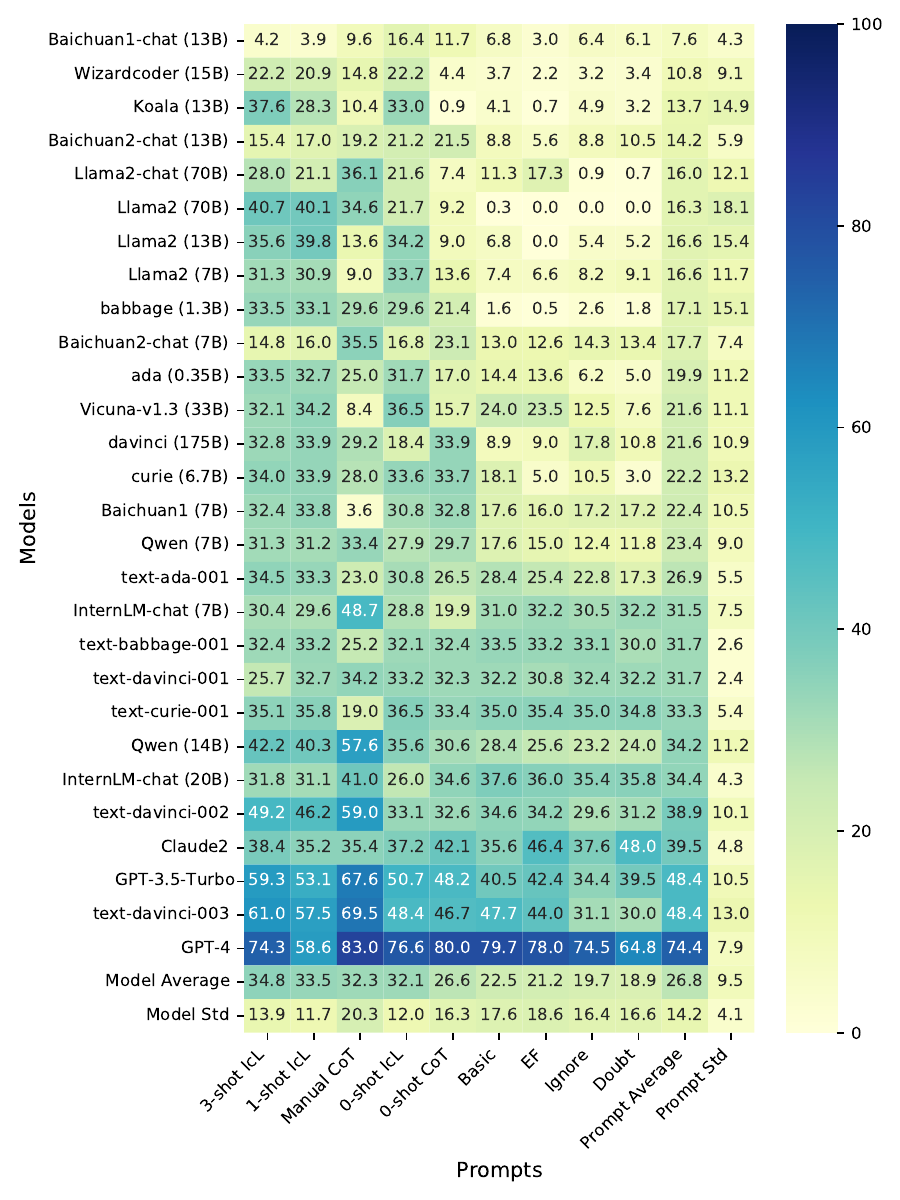}
\end{minipage}
}
\subfigure[Model performance of BAS-C (min-BAS)]{
\begin{minipage}{8.5cm}
\centering
\includegraphics[width=.9\linewidth]{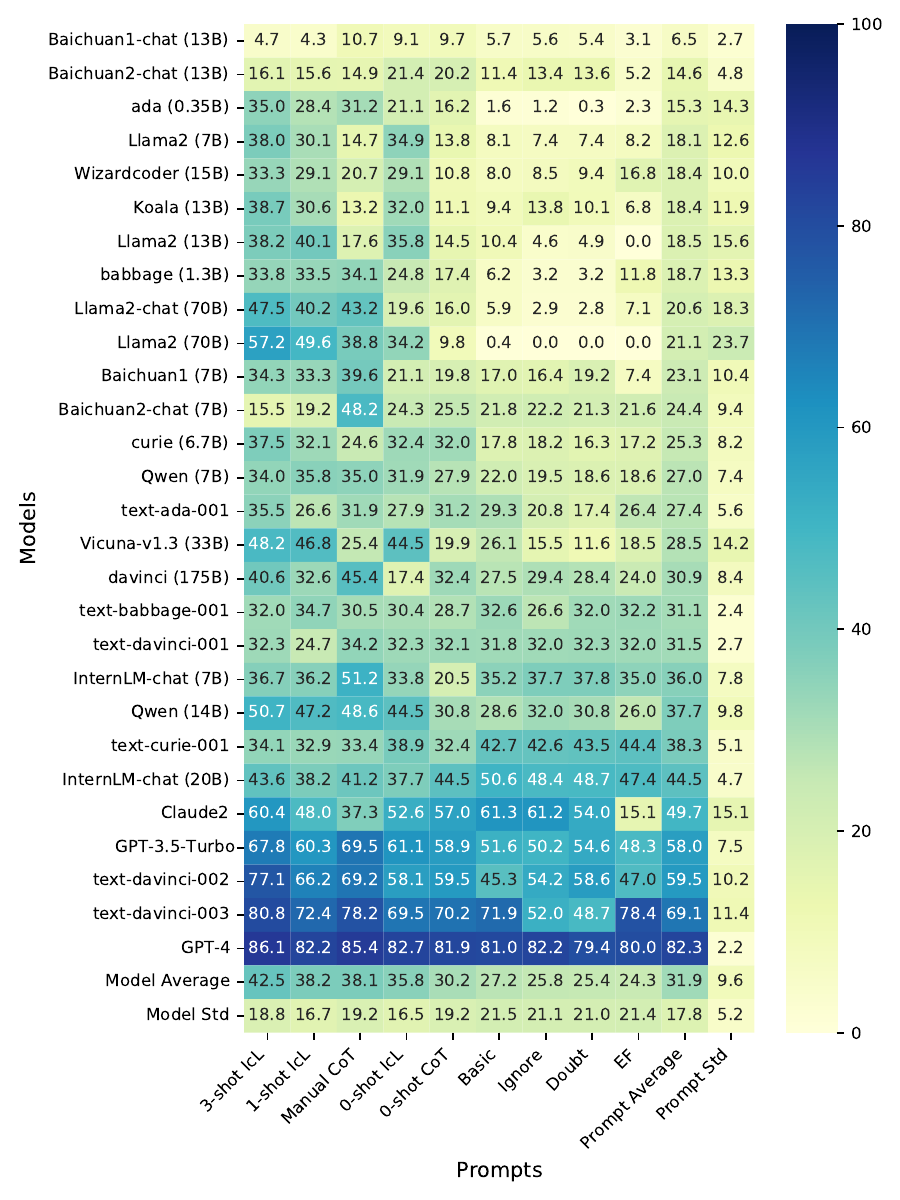}
\end{minipage}
}
\subfigure[Model performance of BAS-C (mix-BAS)]{
\begin{minipage}{8.5cm}
\centering
\includegraphics[width=.9\linewidth]{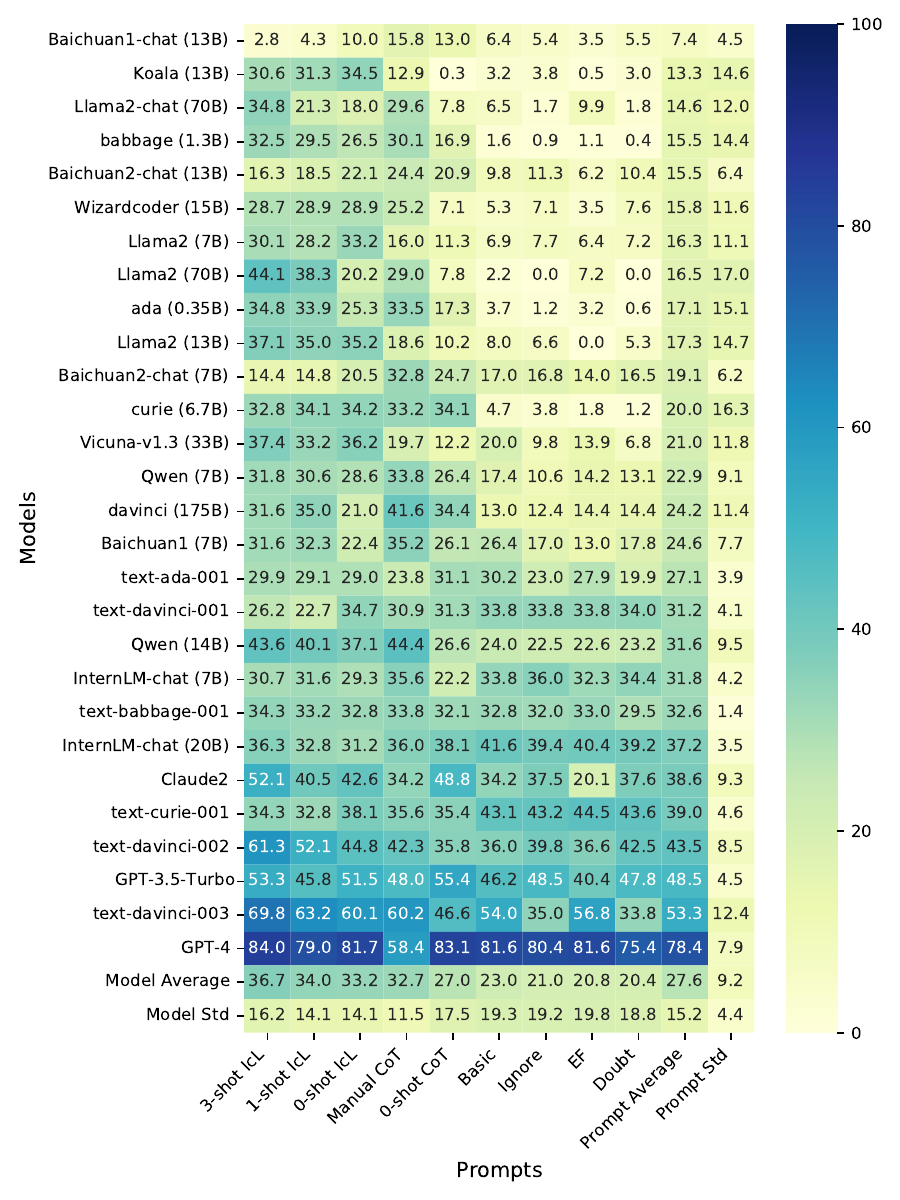}
\end{minipage}
}
\caption[Heatmaps of model performance of causal tasks in BAS]{\textbf{Heatmaps of model performance of causal tasks in BAS.} The models and prompts are sorted by their averages.}
\label{fig:Heatmap_of_performances_of_Backdoor_Adjustment_Set}
\end{figure}

\begin{figure}
\centering
\subfigure[\textit{Prompt gain} of BAS-B (backadj)]{
\begin{minipage}{8.5cm}
\centering
\includegraphics[width=.9\linewidth]{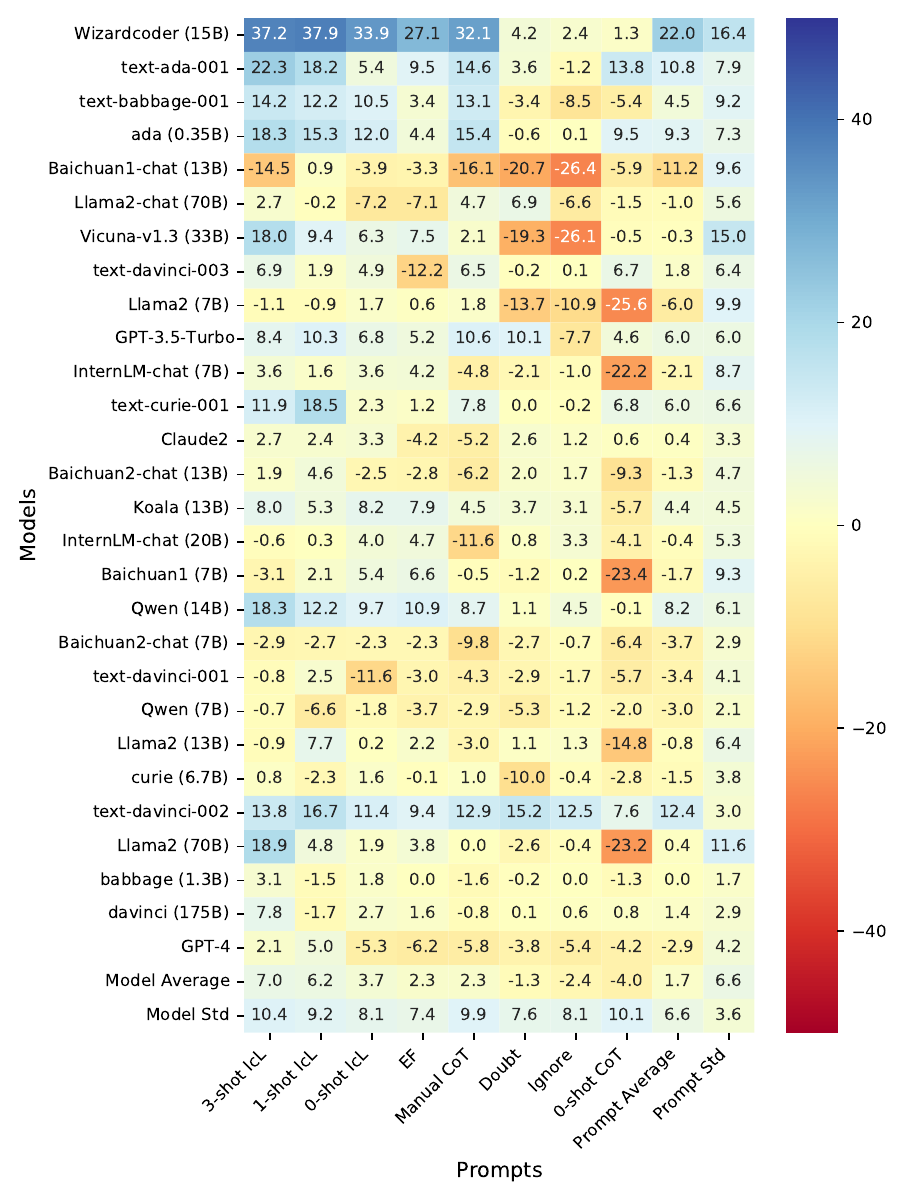}
\end{minipage}
}
\subfigure[\textit{Prompt gain} of BAS-C (max-BAS)]{
\begin{minipage}{8.5cm}
\centering
\includegraphics[width=.9\linewidth]{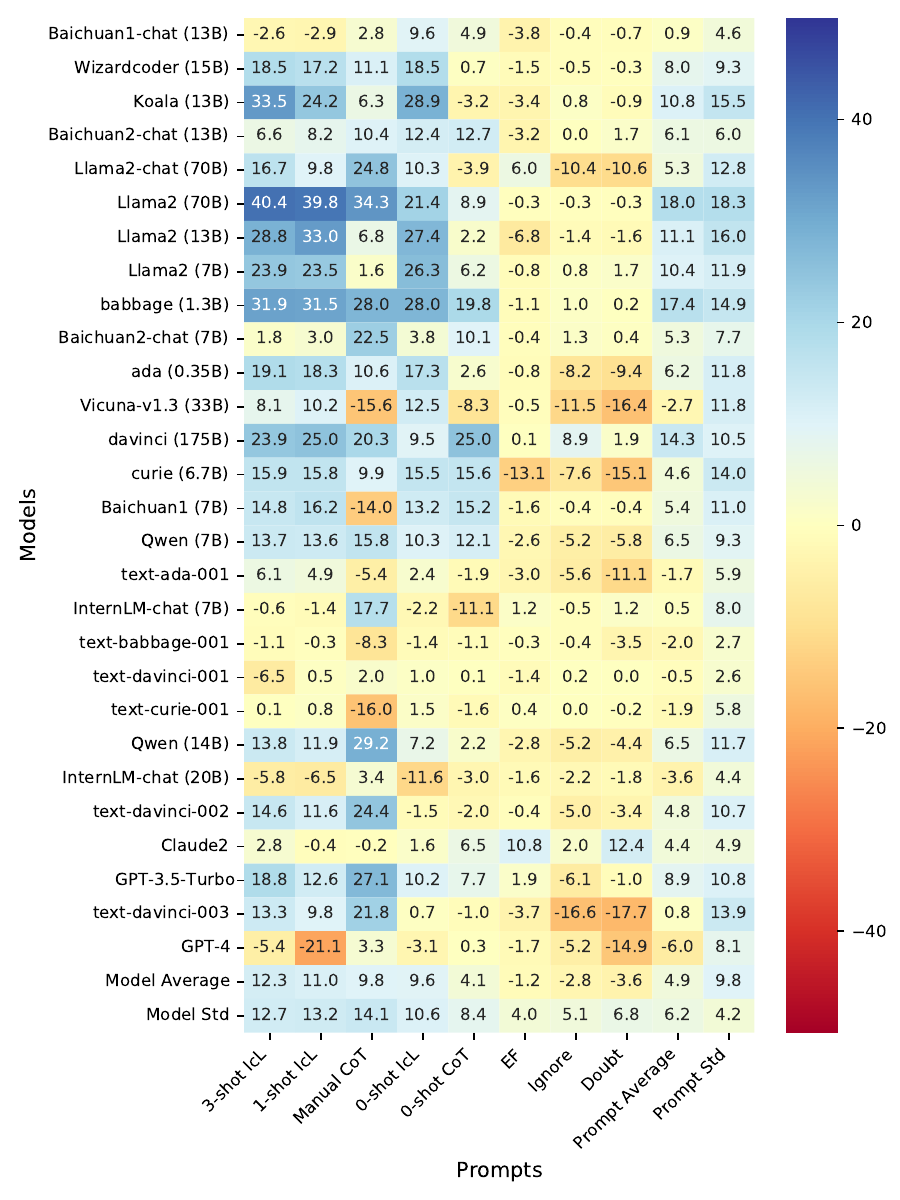}
\end{minipage}
}
\subfigure[\textit{Prompt gain} of BAS-C (min-BAS)]{
\begin{minipage}{8.5cm}
\centering
\includegraphics[width=.9\linewidth]{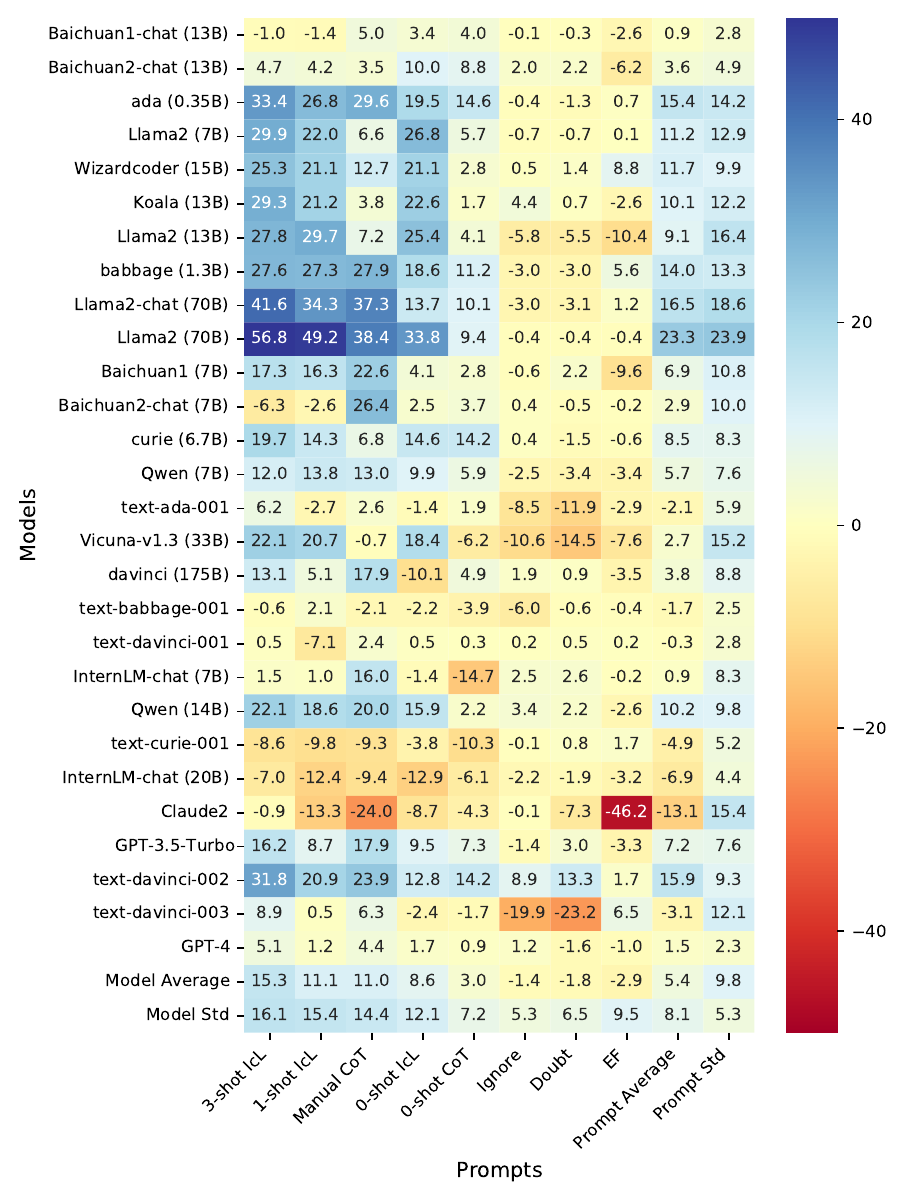}
\end{minipage}
}
\subfigure[\textit{Prompt gain} of BAS-C (mix-BAS)]{
\begin{minipage}{8.5cm}
\centering
\includegraphics[width=.9\linewidth]{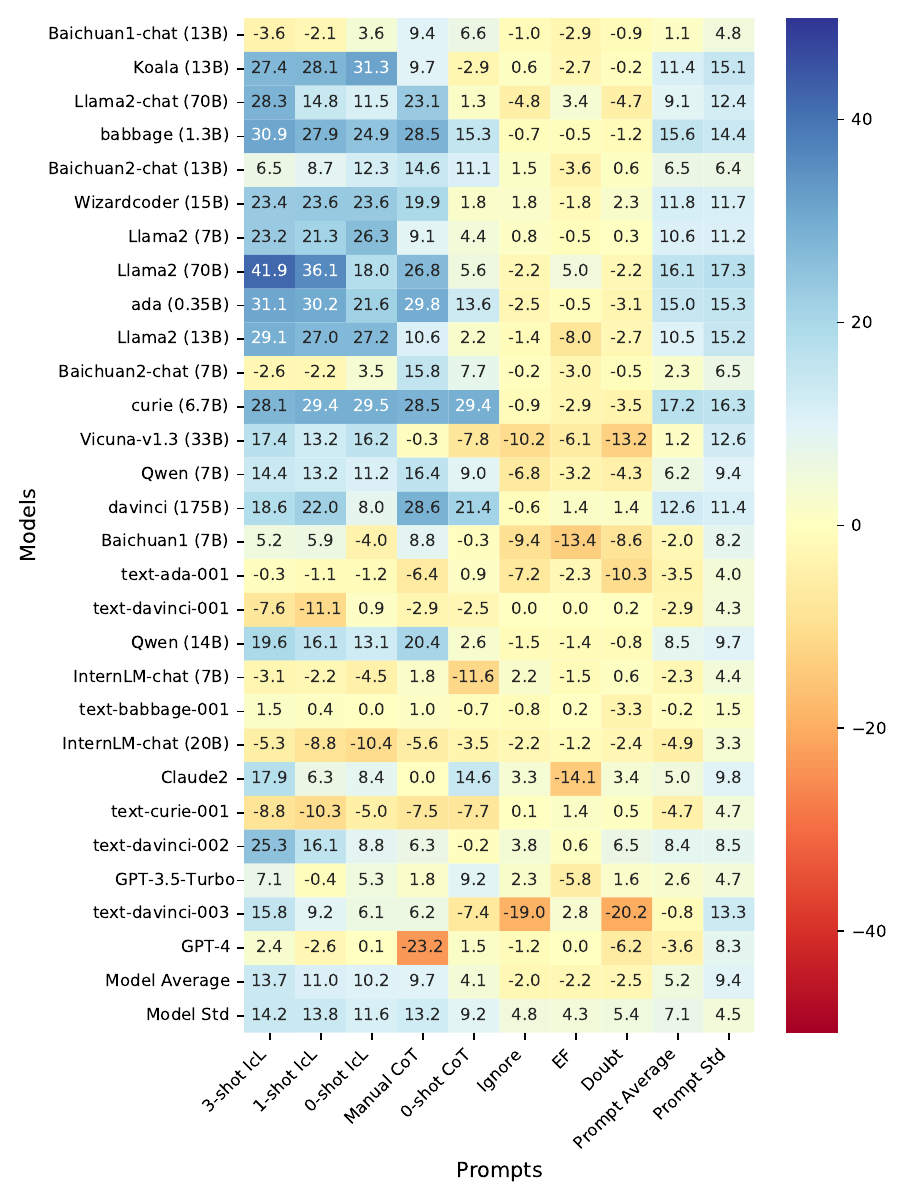}
\end{minipage}
}
\caption[Heatmaps of \textit{prompt gain} of causal tasks in BAS]{\textbf{Heatmaps of \textit{prompt gain} of causal tasks in BAS.} The models and prompts are sorted by their averages.}
\label{fig:Heatmap_of_gain_of_Backdoor_Adjustment_Set}
\end{figure}

The distribution of models' accuracy on BAS is shown in Figure \ref{fig:Distribution_of_Backdoor_Adjustment_Set_Tasks}. Figure \ref{fig:Heatmap_of_performances_of_Backdoor_Adjustment_Set} illustrates how models perform on BAS. The prompt gain (i.e., accuracy improvement against the basic prompt on the model with the used prompt) is demonstrated in Figure \ref{fig:Heatmap_of_gain_of_Backdoor_Adjustment_Set}.

\subsection{Counterfactuals}
\subsubsection{CR}
\begin{figure}
\centering
\subfigure[Model performance of CR-C (CRASS)]{
\begin{minipage}{8.5cm}
\centering
\includegraphics[width=.9\linewidth]{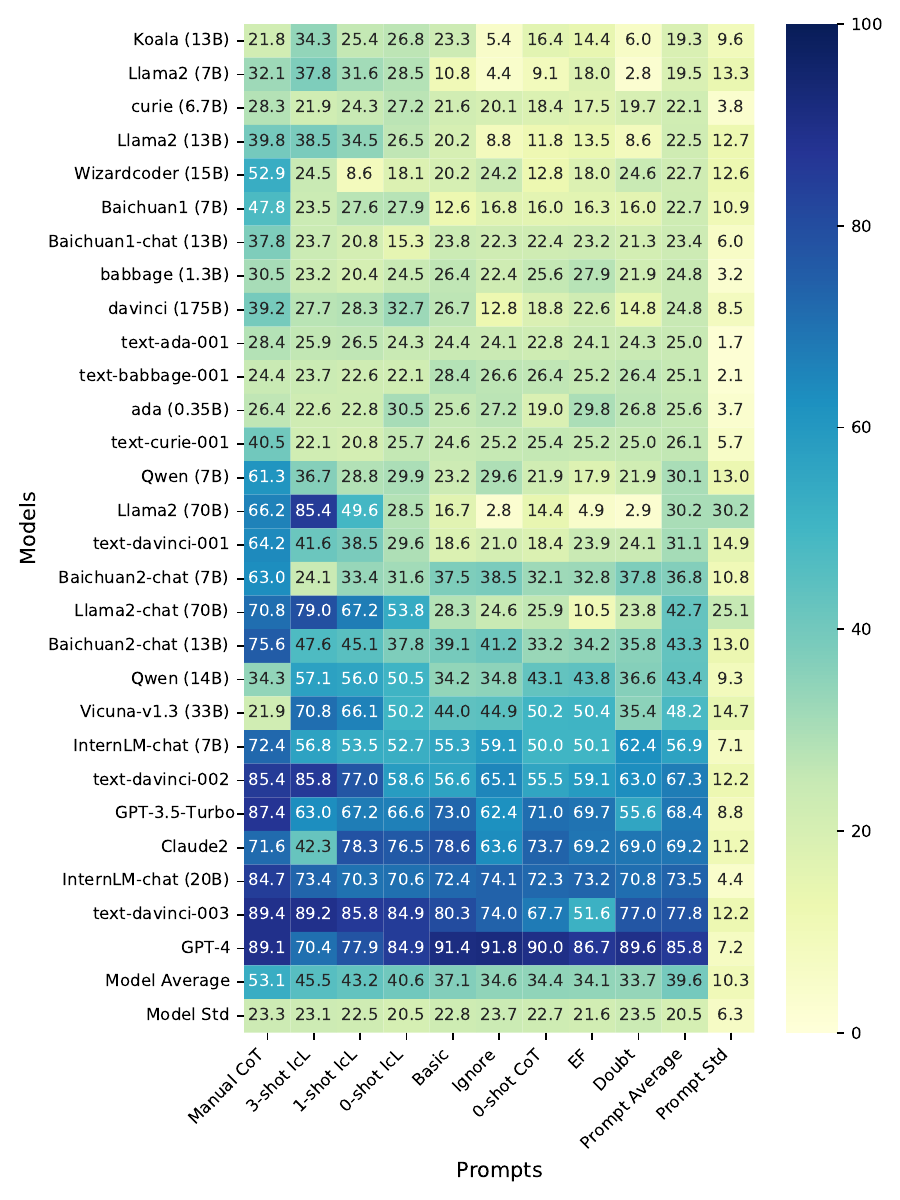}
\end{minipage}
}
\subfigure[Model performance of CR-B (det-counterfactual)]{
\begin{minipage}{8.5cm}
\centering
\includegraphics[width=.9\linewidth]{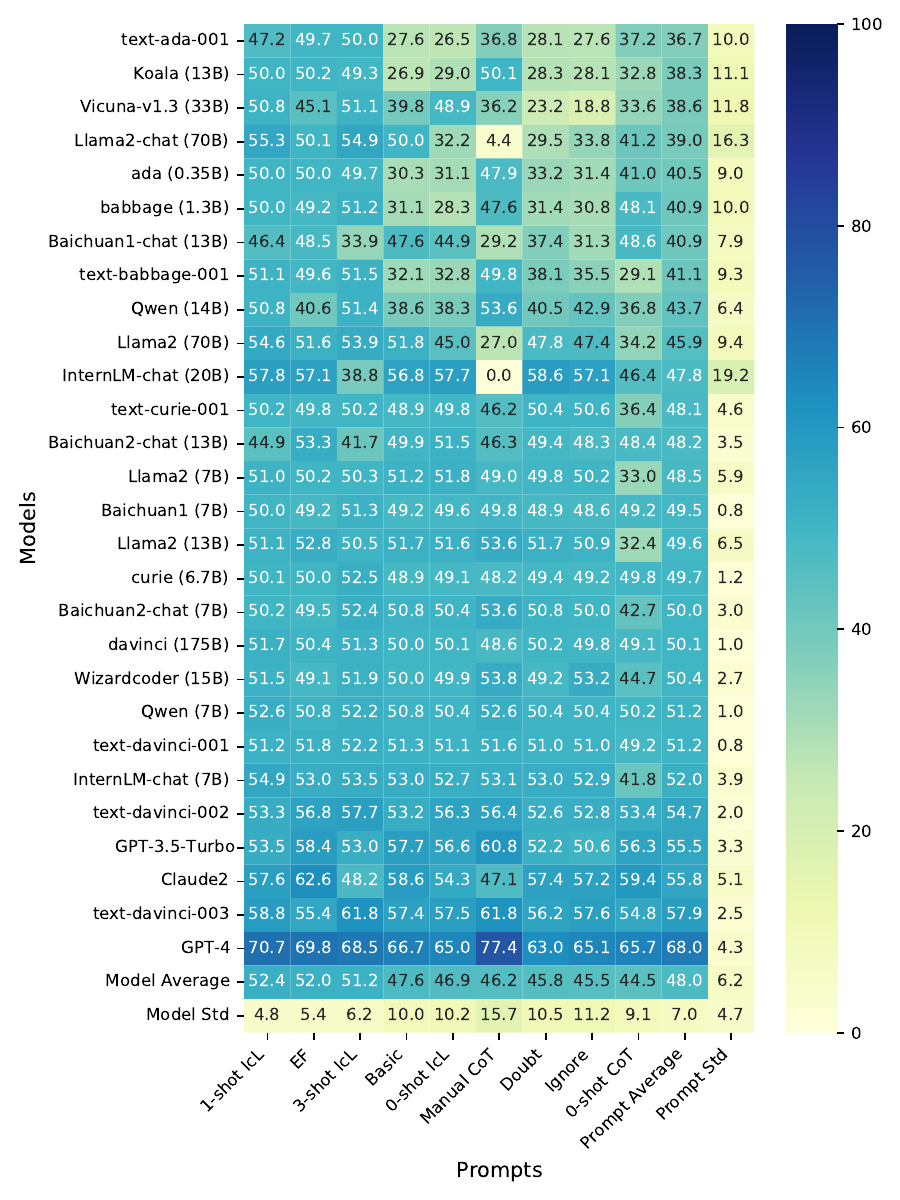}
\end{minipage}
}
\caption[Heatmaps of model performance of causal tasks in CR]{\textbf{Heatmaps of model performance of causal tasks in CR.} The models and prompts are sorted by their averages.}
\label{fig:Heatmap_of_performances_of_Counterfactual_Reasoning}
\end{figure}

\begin{figure}
\centering
\subfigure[\textit{Prompt gain} of CR-C (CRASS)]{
\begin{minipage}{8.5cm}
\centering
\includegraphics[width=.9\linewidth]{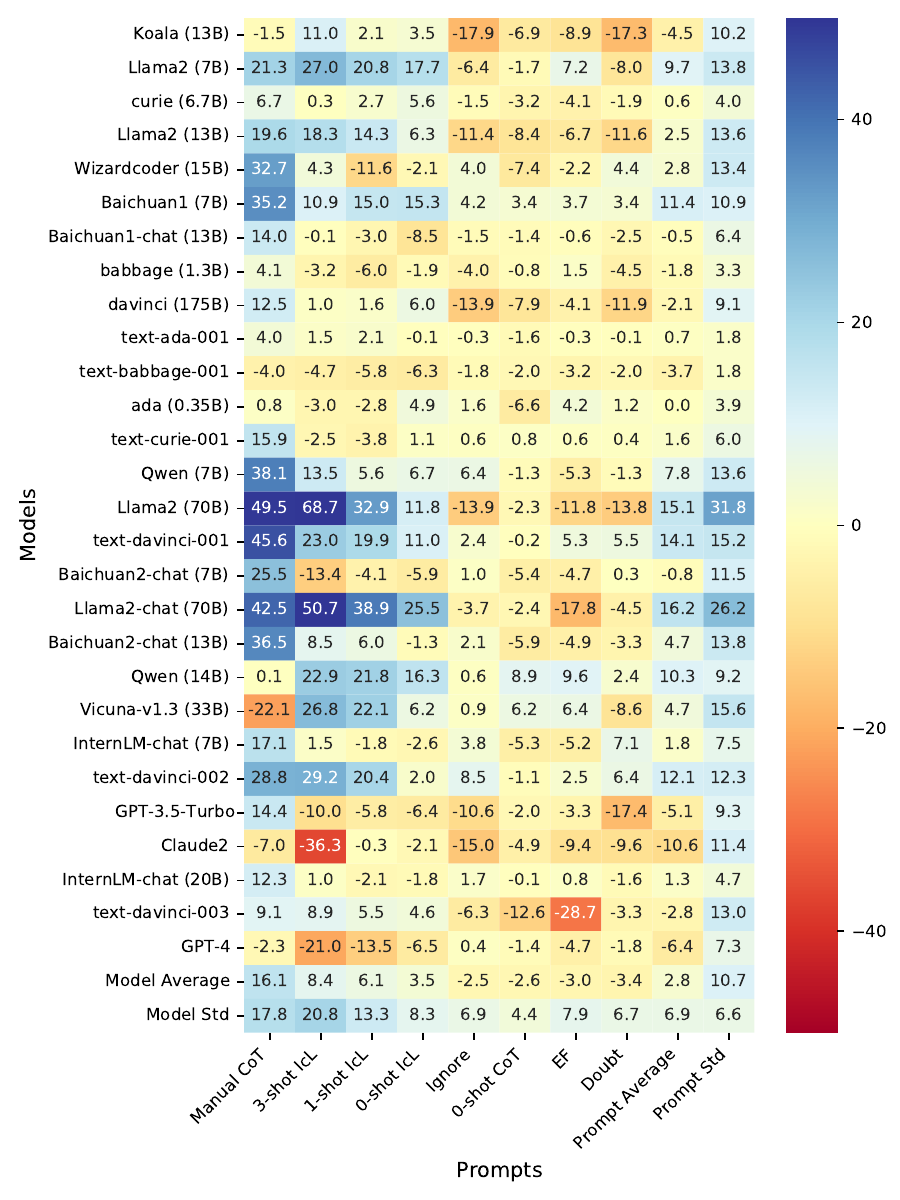}
\end{minipage}
}
\subfigure[\textit{Prompt gain} of CR-B (det-counterfactual)]{
\begin{minipage}{8.5cm}
\centering
\includegraphics[width=.9\linewidth]{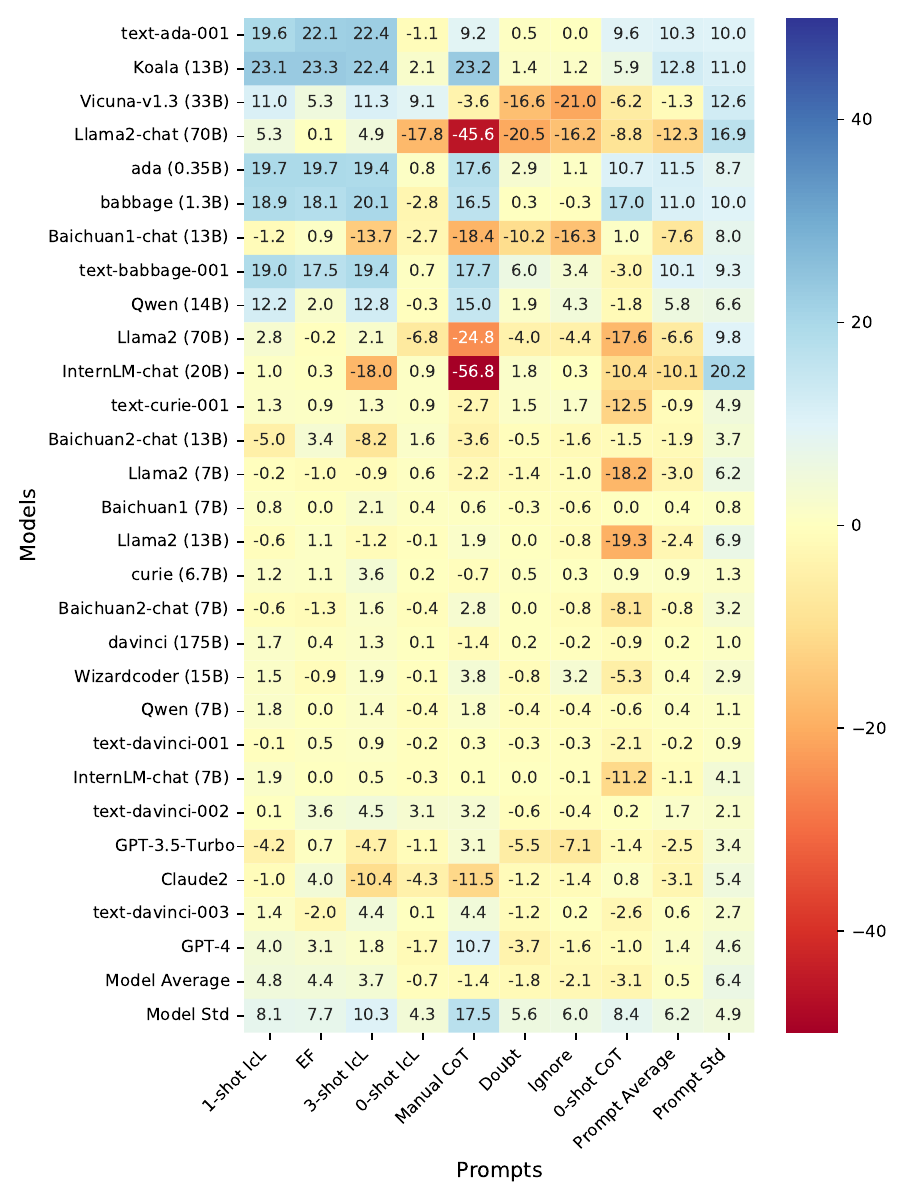}
\end{minipage}
}
\caption[Heatmaps of \textit{prompt gain} of causal tasks in CR]{\textbf{Heatmaps of \textit{prompt gain} of causal tasks in CR.} The models and prompts are sorted by their averages.}
\label{fig:Heatmap_of_gain_of_Counterfactual_Reasoning}
\end{figure}

The distribution of models' accuracy on CR is shown in Figure \ref{fig:Distribution_of_Counterfactual_Reasoning_Tasks}. Figure \ref{fig:Heatmap_of_performances_of_Counterfactual_Reasoning} illustrates how models perform on CR. The prompt gain (i.e., accuracy improvement against the basic prompt on the model with the used prompt) is demonstrated in Figure \ref{fig:Heatmap_of_gain_of_Counterfactual_Reasoning}.

\subsubsection{ETT}
\begin{figure}
\centering
\subfigure[Model performance of ETT-P (ETT-basic)]{
\begin{minipage}{8.5cm}
\centering
\includegraphics[width=.9\linewidth]{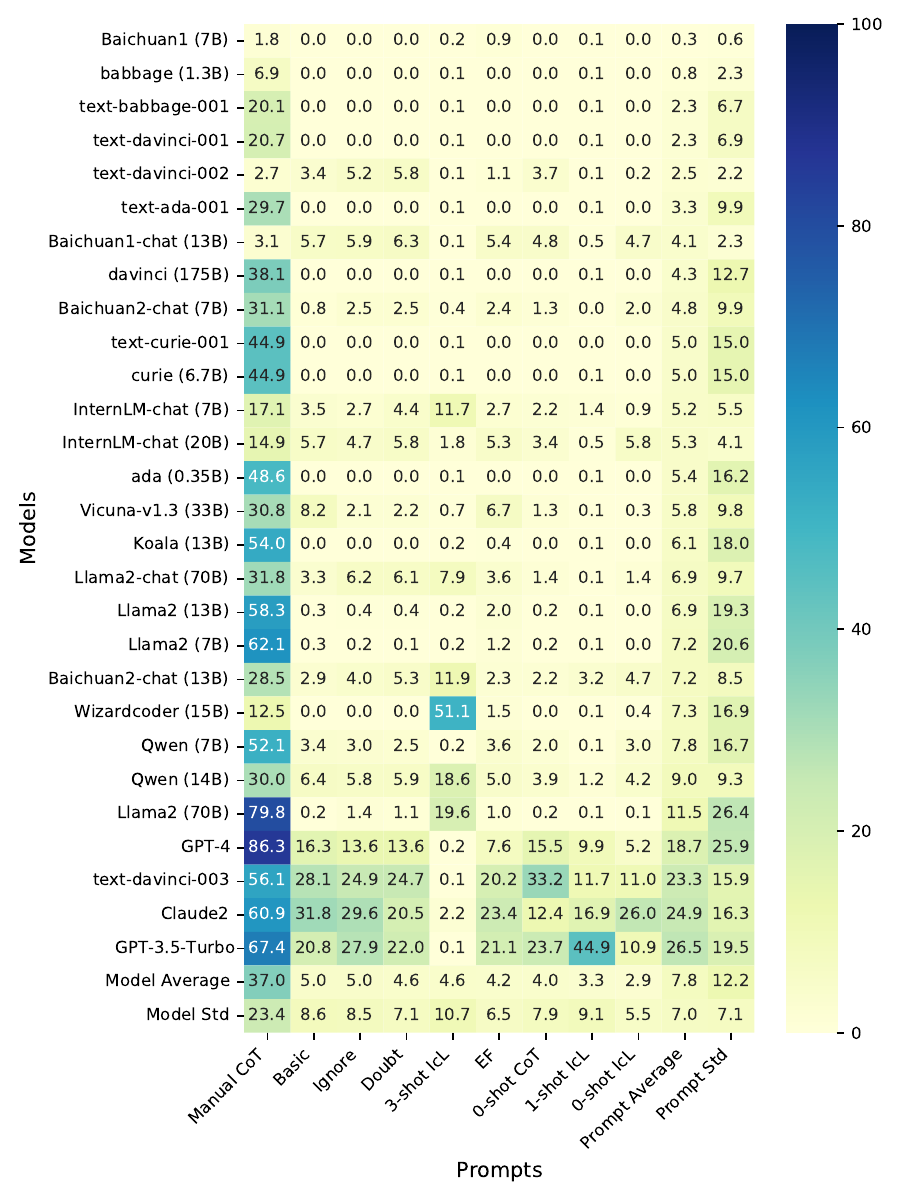}
\end{minipage}
}
\subfigure[Model performance of ETT-P (ETT-hard)]{
\begin{minipage}{8.5cm}
\centering
\includegraphics[width=.9\linewidth]{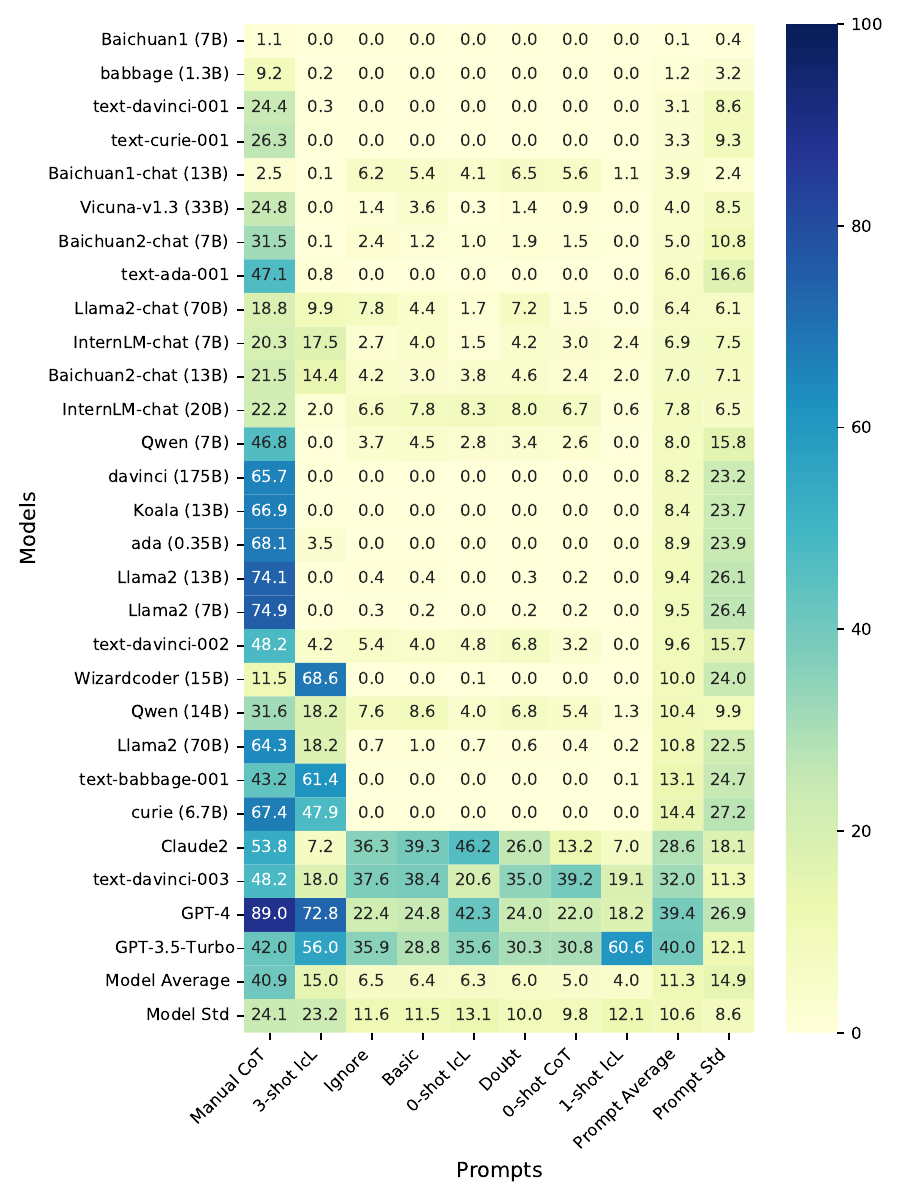}
\end{minipage}
}
\subfigure[Model performance of ETT-B (ETT-natural)]{
\begin{minipage}{8.5cm}
\centering
\includegraphics[width=.9\linewidth]{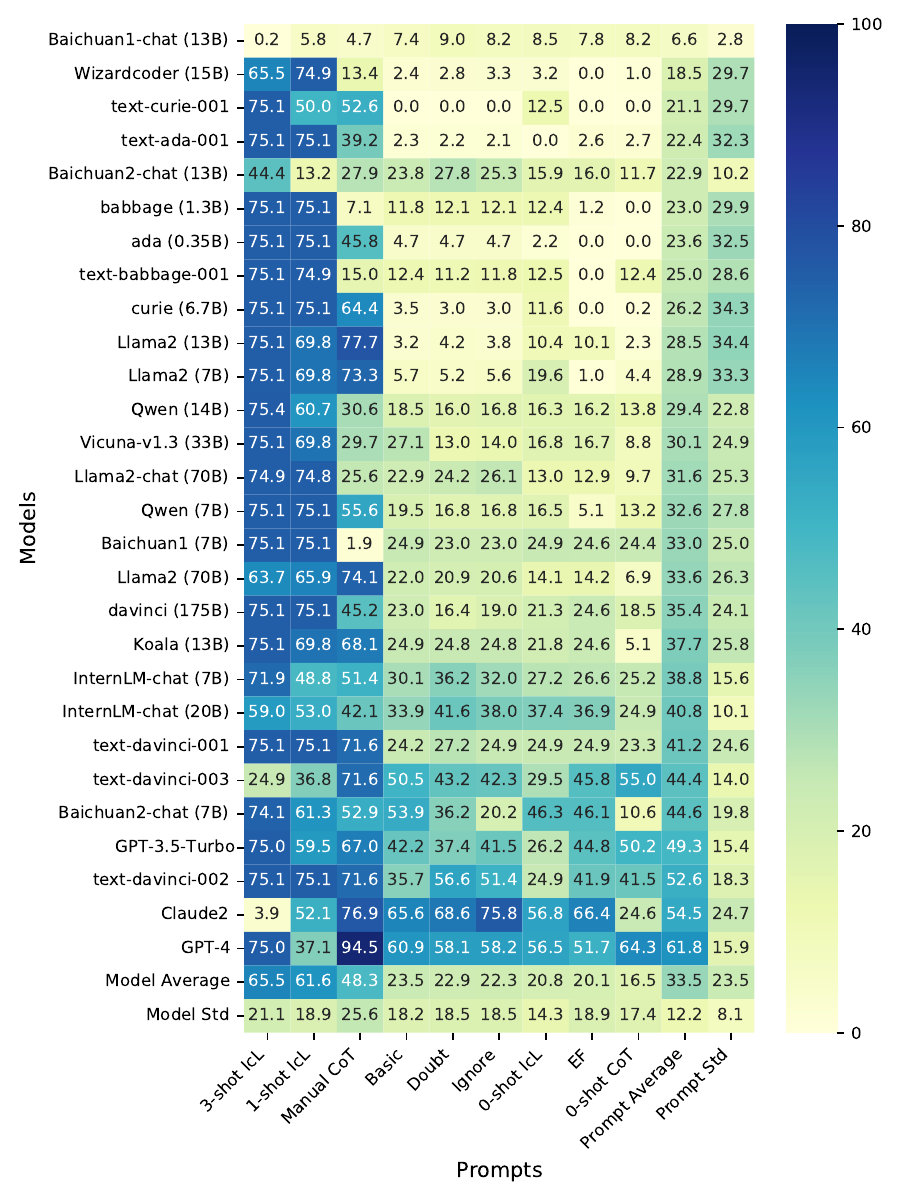}
\end{minipage}
}
\caption[Heatmaps of model performance of causal tasks in ETT]{\textbf{Heatmaps of model performance of causal tasks in ETT.} The models and prompts are sorted by their averages.}
\label{fig:Heatmap_of_performances_of_Effect_of_the_Treatment_on_the_Treated}
\end{figure}

\begin{figure}
\centering
\subfigure[\textit{Prompt gain} of ETT-P (ETT-basic)]{
\begin{minipage}{8.5cm}
\centering
\includegraphics[width=.9\linewidth]{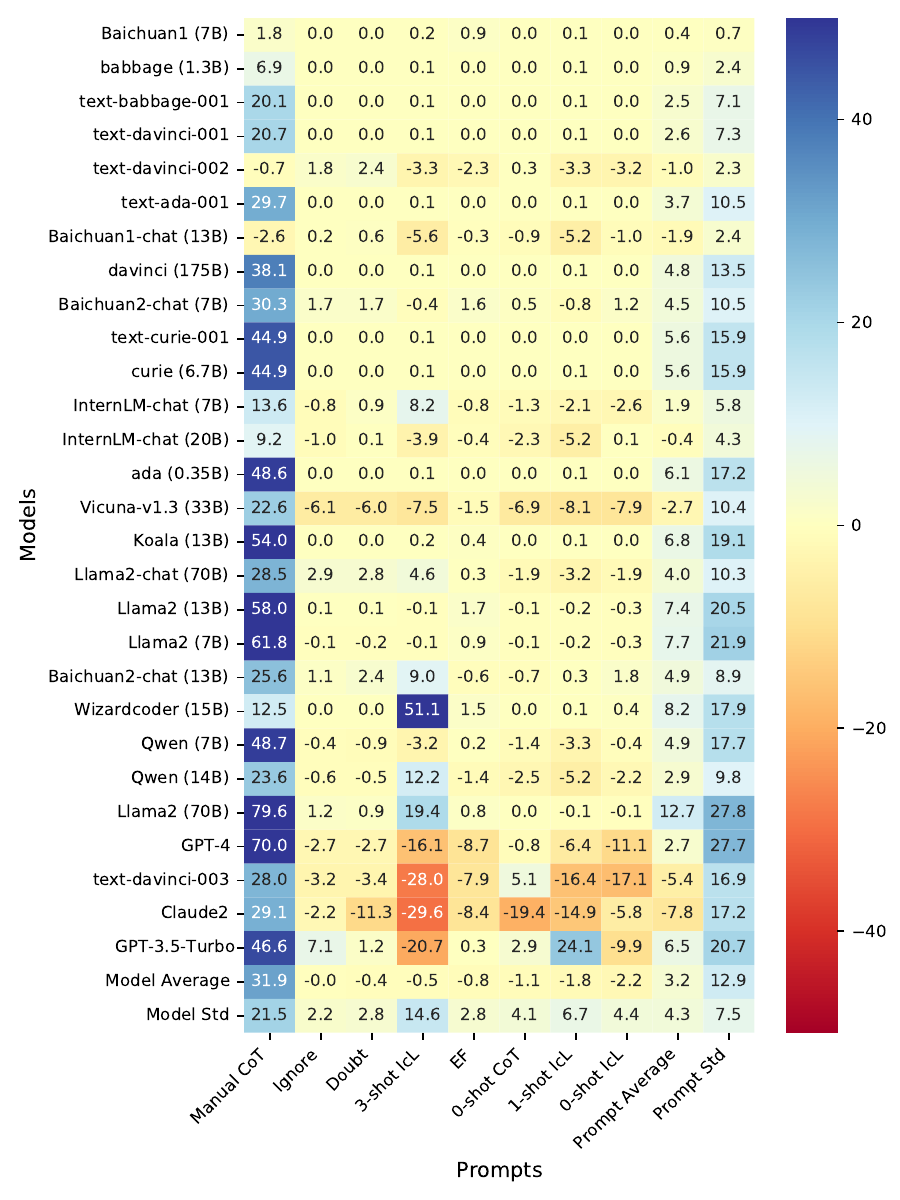}
\end{minipage}
}
\subfigure[\textit{Prompt gain} of ETT-P (ETT-hard)]{
\begin{minipage}{8.5cm}
\centering
\includegraphics[width=.9\linewidth]{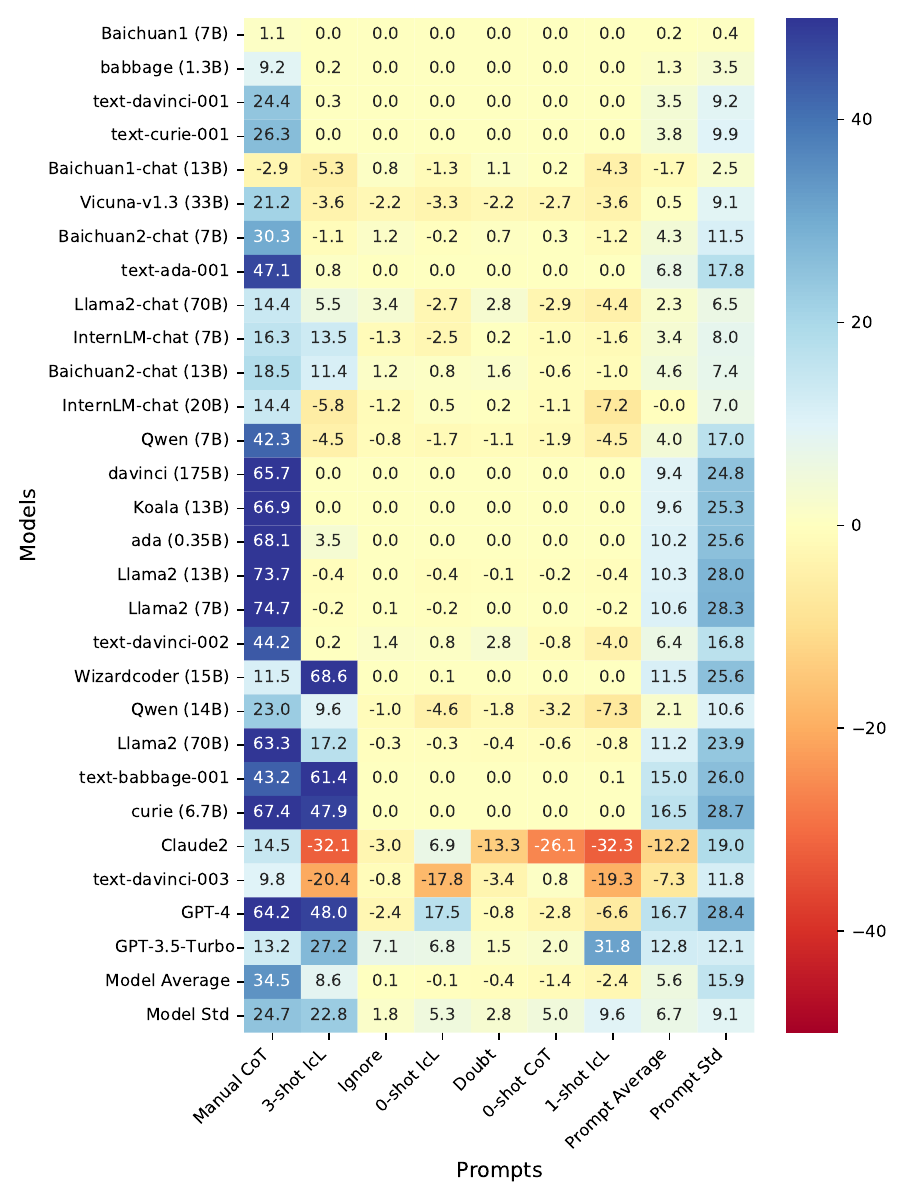}
\end{minipage}
}
\subfigure[\textit{Prompt gain} of ETT-B (ETT-natural)]{
\begin{minipage}{8.5cm}
\centering
\includegraphics[width=.9\linewidth]{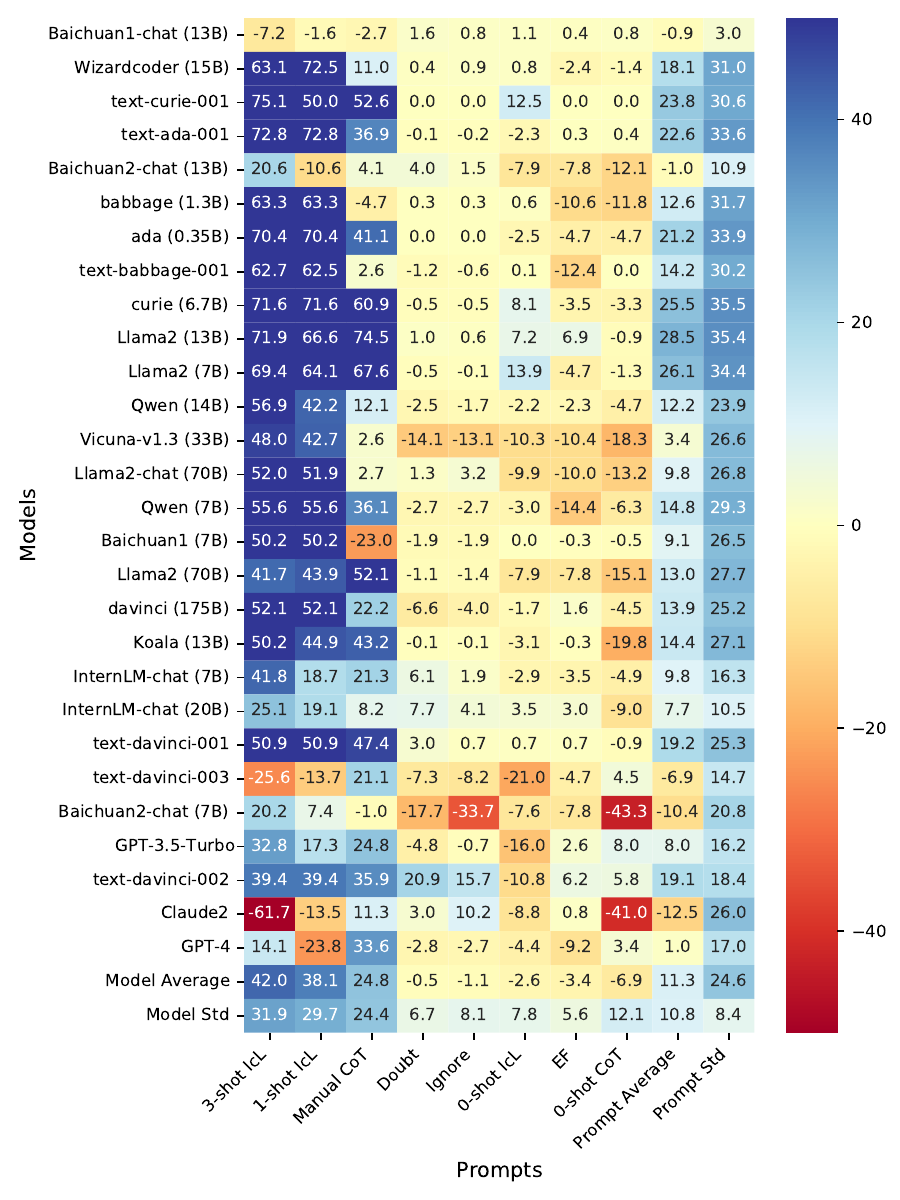}
\end{minipage}
}
\caption[Heatmaps of \textit{prompt gain} of causal tasks in ETT]{\textbf{Heatmaps of \textit{prompt gain} of causal tasks in ETT.} The models and prompts are sorted by their averages.}
\label{fig:Heatmap_of_gain_of_Effect_of_the_Treatment_on_the_Treated}
\end{figure}

The distribution of models' accuracy on ETT is shown in Figure \ref{fig:Distribution_of_Effect_of_the_Treatment_on_the_Treated_Tasks}. Figure \ref{fig:Heatmap_of_performances_of_Effect_of_the_Treatment_on_the_Treated} illustrates how models perform on ETT. The prompt gain (i.e., accuracy improvement against the basic prompt on the model with the used prompt) is demonstrated in Figure \ref{fig:Heatmap_of_gain_of_Effect_of_the_Treatment_on_the_Treated}.

\subsubsection{NDE}
\begin{figure}
\centering
\subfigure[Model performance of NDE-P (NDE-basic)]{
\begin{minipage}{8.5cm}
\centering
\includegraphics[width=.9\linewidth]{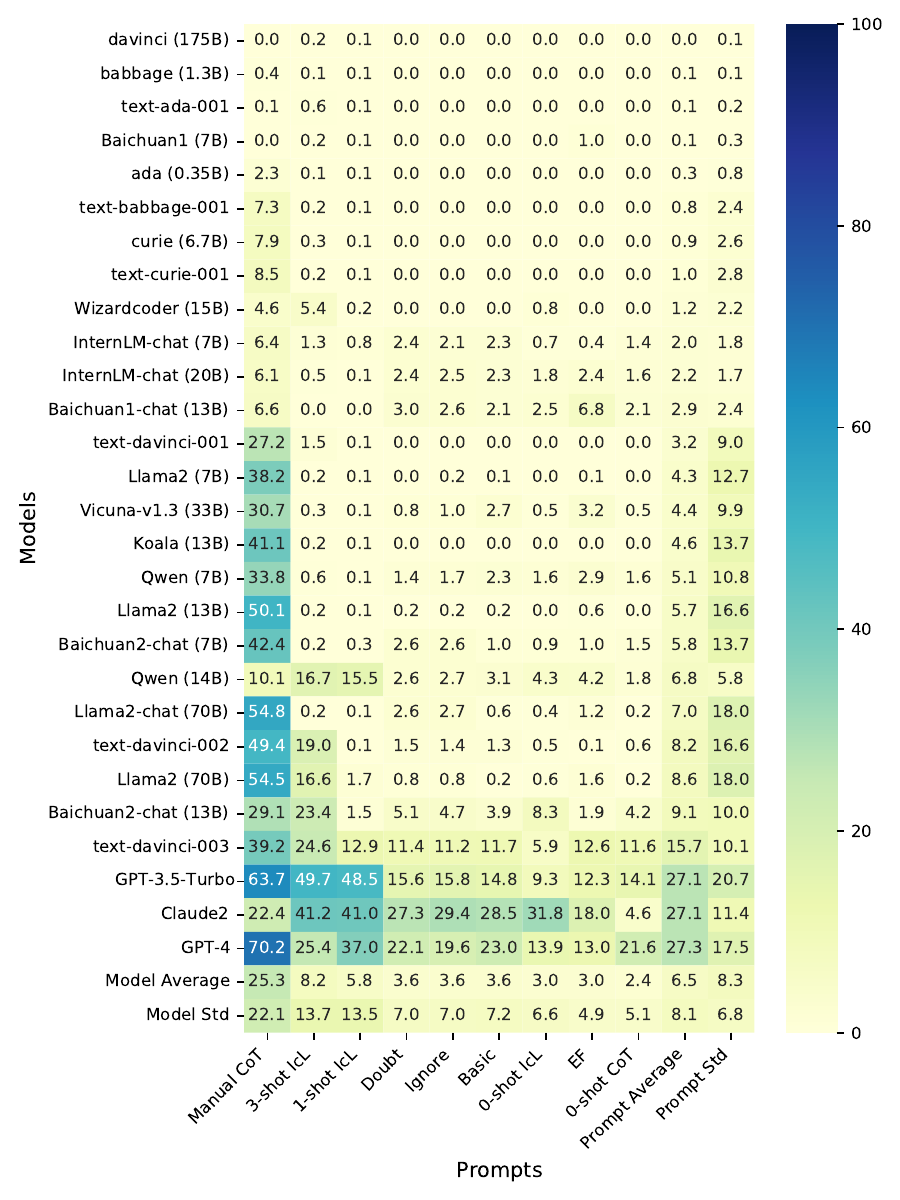}
\end{minipage}
}
\subfigure[Model performance of NDE-P (NDE-hard)]{
\begin{minipage}{8.5cm}
\centering
\includegraphics[width=.9\linewidth]{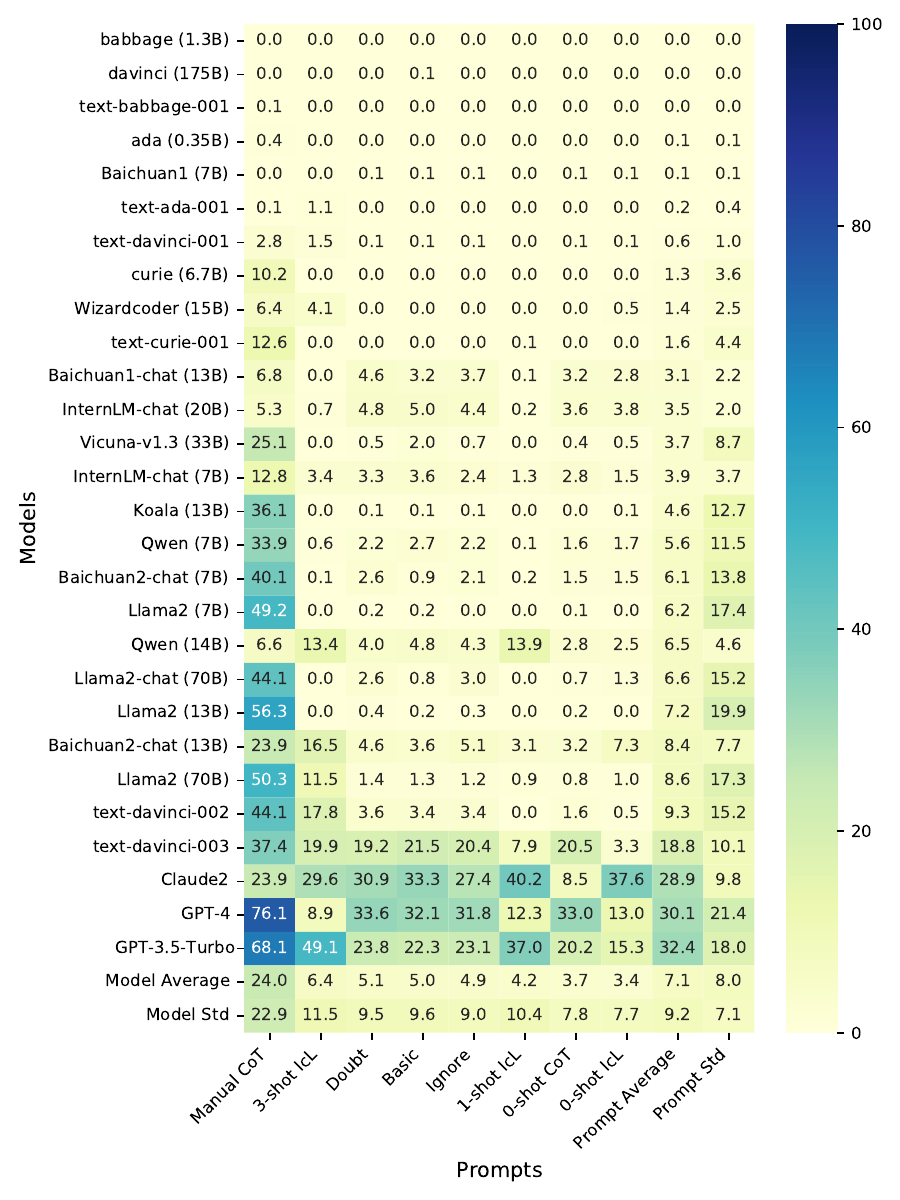}
\end{minipage}
}
\subfigure[Model performance of NDE-B (NDE-natural)]{
\begin{minipage}{8.5cm}
\centering
\includegraphics[width=.9\linewidth]{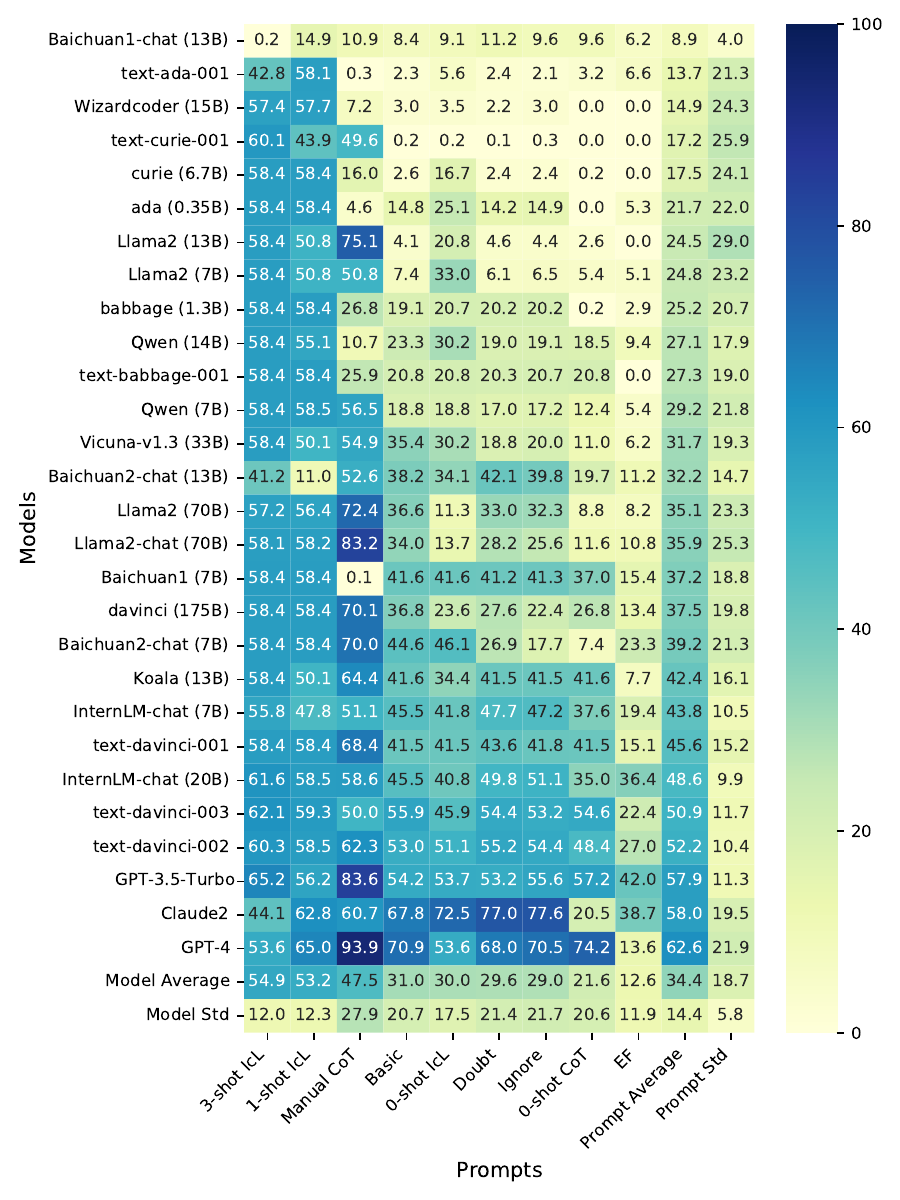}
\end{minipage}
}
\caption[Heatmaps of model performance of causal tasks in NDE]{\textbf{Heatmaps of model performance of causal tasks in NDE.} The models and prompts are sorted by their averages.}
\label{fig:Heatmap_of_performances_of_Natural_Direct_Effect}
\end{figure}

\begin{figure}
\centering
\subfigure[\textit{Prompt gain} of NDE-P (NDE-basic)]{
\begin{minipage}{8.5cm}
\centering
\includegraphics[width=.9\linewidth]{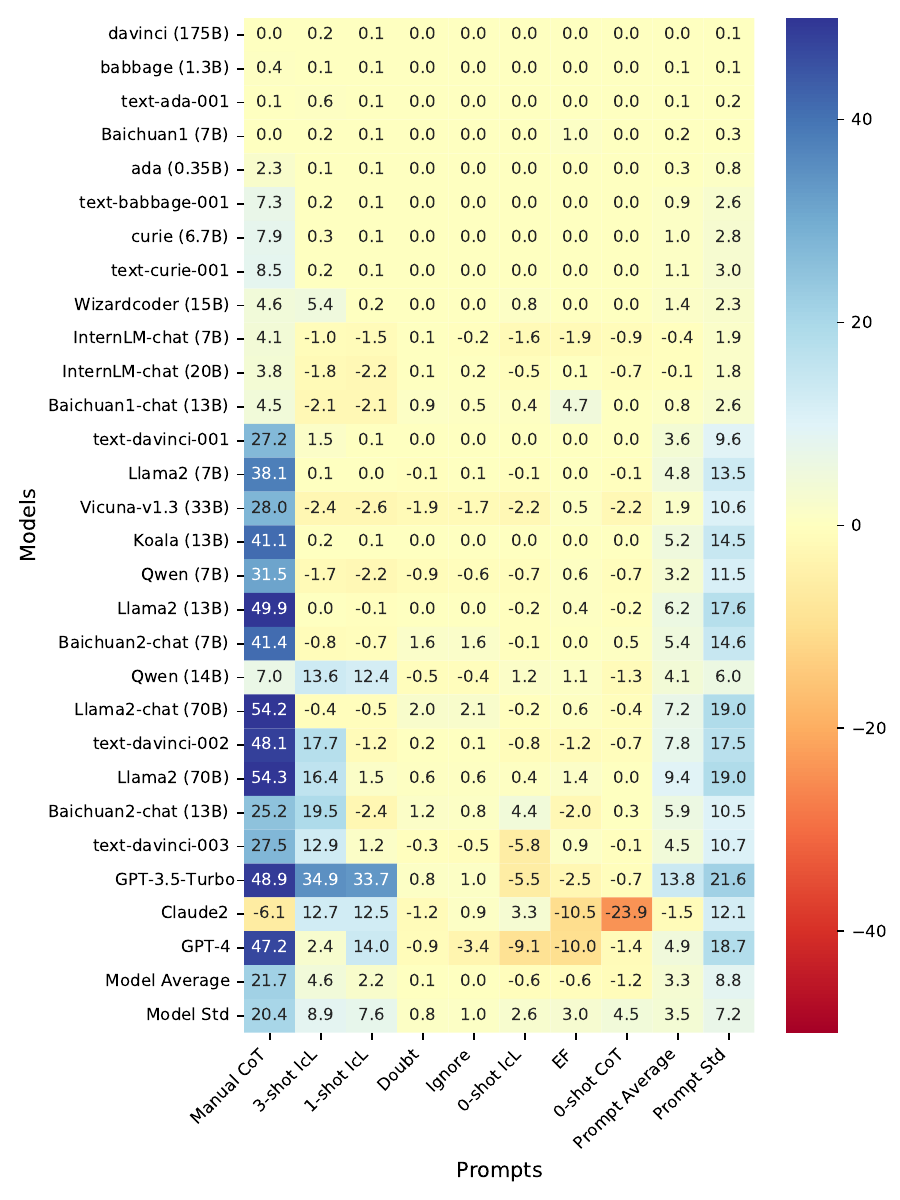}
\end{minipage}
}
\subfigure[\textit{Prompt gain} of NDE-P (NDE-hard)]{
\begin{minipage}{8.5cm}
\centering
\includegraphics[width=.9\linewidth]{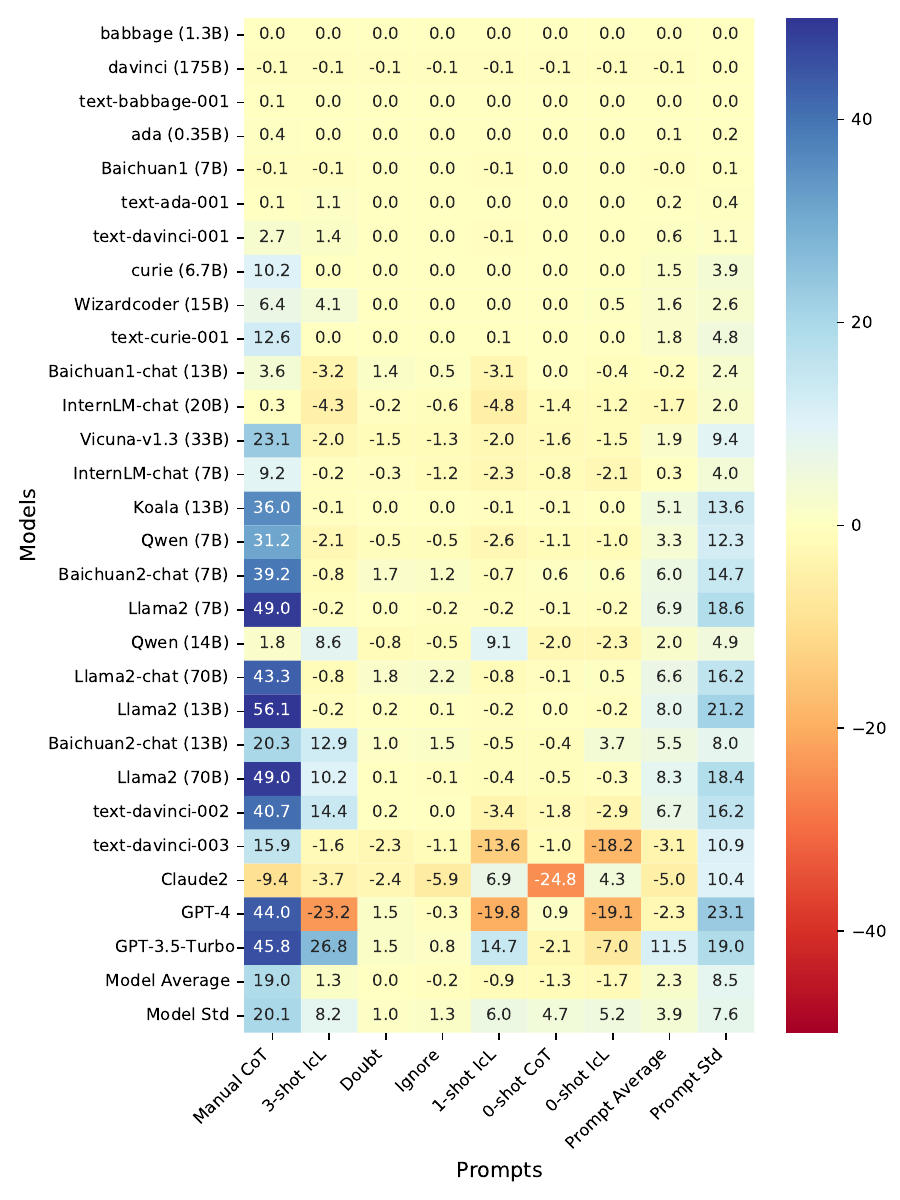}
\end{minipage}
}
\subfigure[\textit{Prompt gain} of NDE-B (NDE-natural)]{
\begin{minipage}{8.5cm}
\centering
\includegraphics[width=.9\linewidth]{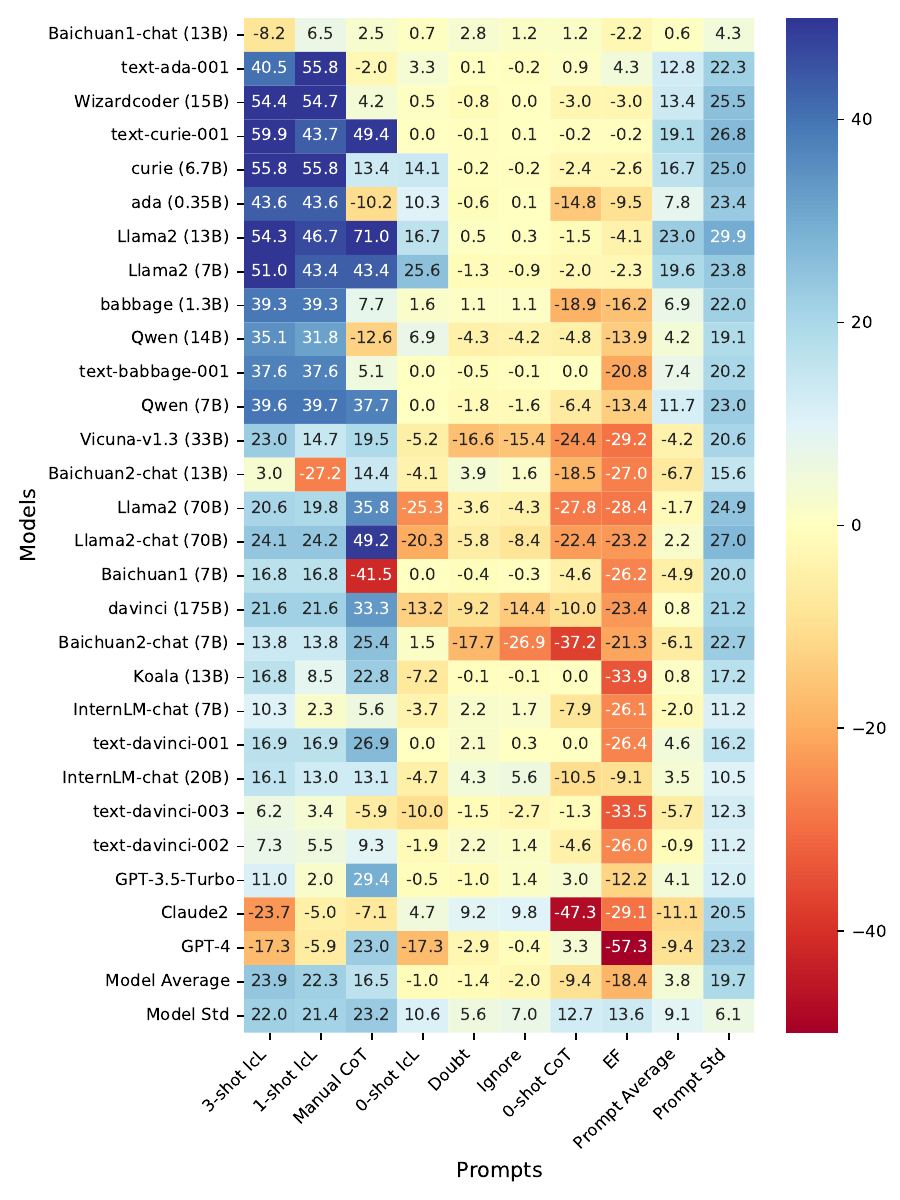}
\end{minipage}
}
\caption[Heatmaps of \textit{prompt gain} of causal tasks in NDE]{\textbf{Heatmaps of \textit{prompt gain} of causal tasks in NDE.} The models and prompts are sorted by their averages.}
\label{fig:Heatmap_of_gain_of_Natural_Direct_Effect}
\end{figure}

The distribution of models' accuracy on NDE is shown in Figure \ref{fig:Distribution_of_causal_tasks_in_NDE}. Figure \ref{fig:Heatmap_of_performances_of_Natural_Direct_Effect} illustrates how models perform on NDE. The prompt gain (i.e., accuracy improvement against the basic prompt on the model with the used prompt) is demonstrated in Figure \ref{fig:Heatmap_of_gain_of_Natural_Direct_Effect}.

\subsubsection{NIE}
\begin{figure}
\centering
\subfigure[Model performance of NIE-P (NIE-basic)]{
\begin{minipage}{8.5cm}
\centering
\includegraphics[width=.9\linewidth]{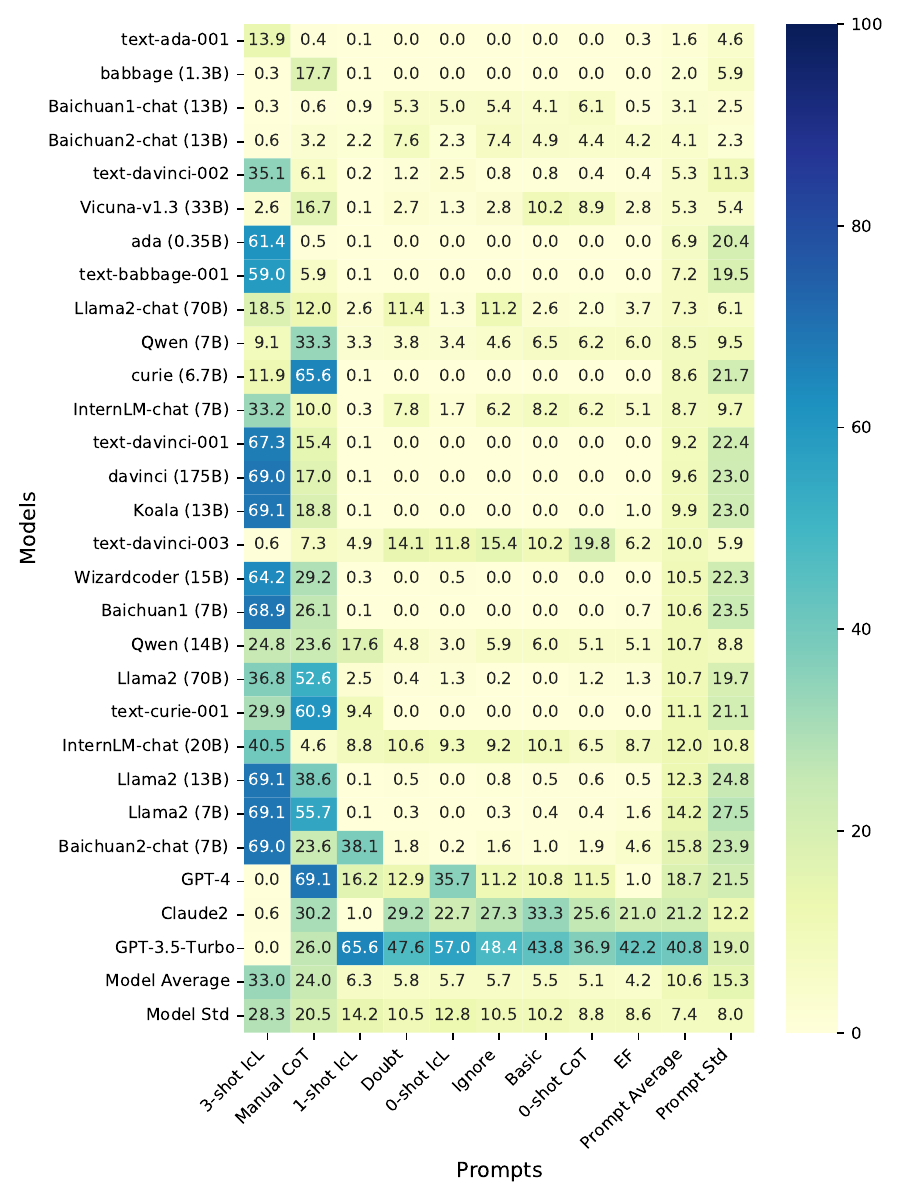}
\end{minipage}
}
\subfigure[Model performance of NIE-P (NIE-hard)]{
\begin{minipage}{8.5cm}
\centering
\includegraphics[width=.9\linewidth]{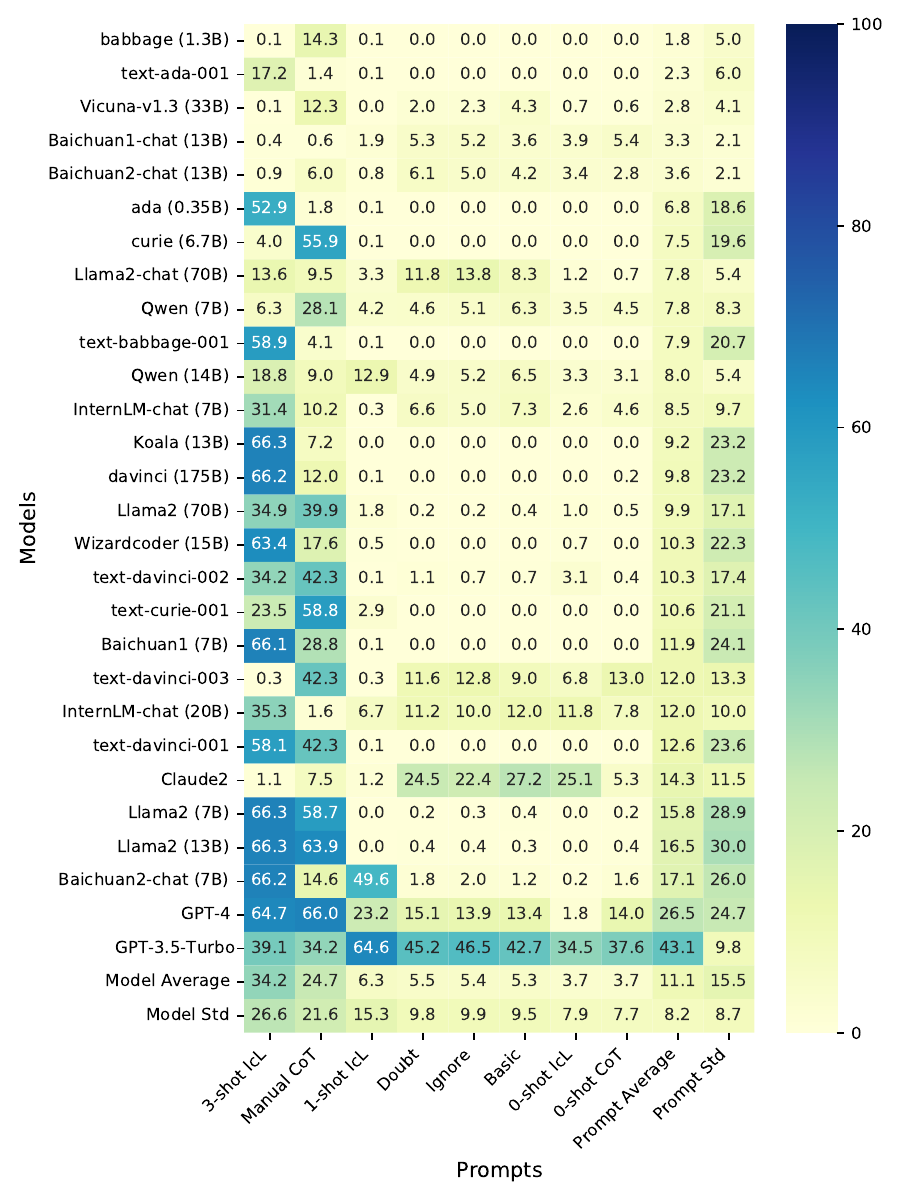}
\end{minipage}
}
\subfigure[Model performance of NIE-B (NIE-natural)]{
\begin{minipage}{8.5cm}
\centering
\includegraphics[width=.9\linewidth]{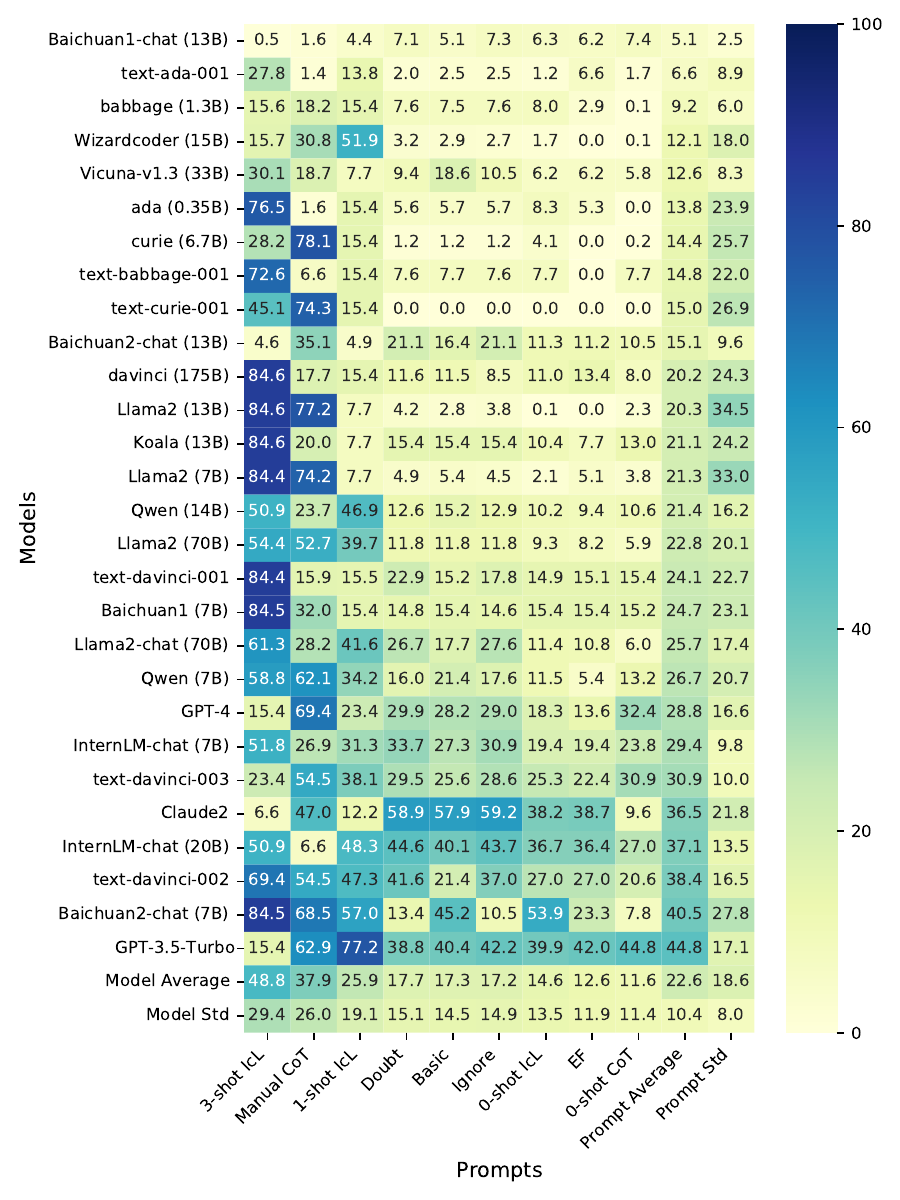}
\end{minipage}
}
\caption[Heatmaps of model performance of causal tasks in NIE]{\textbf{Heatmaps of model performance of causal tasks in NIE.} The models and prompts are sorted by their averages.}
\label{fig:Heatmap_of_performances_of_Natural_Indirect_Effect}
\end{figure}

\begin{figure}
\centering
\subfigure[\textit{Prompt gain} of NIE-P (NIE-basic)]{
\begin{minipage}{8.5cm}
\centering
\includegraphics[width=.9\linewidth]{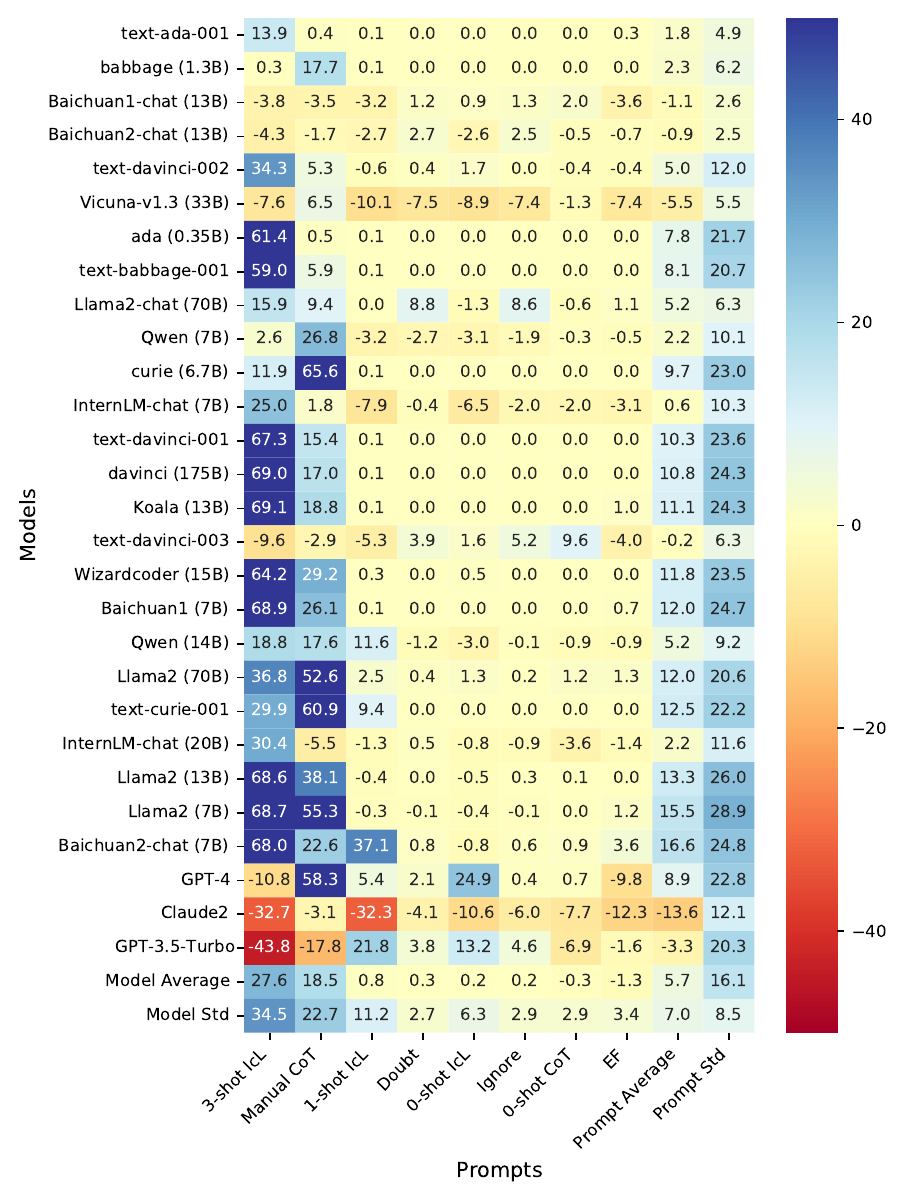}
\end{minipage}
}
\subfigure[\textit{Prompt gain} of NIE-P (NIE-hard)]{
\begin{minipage}{8.5cm}
\centering
\includegraphics[width=.9\linewidth]{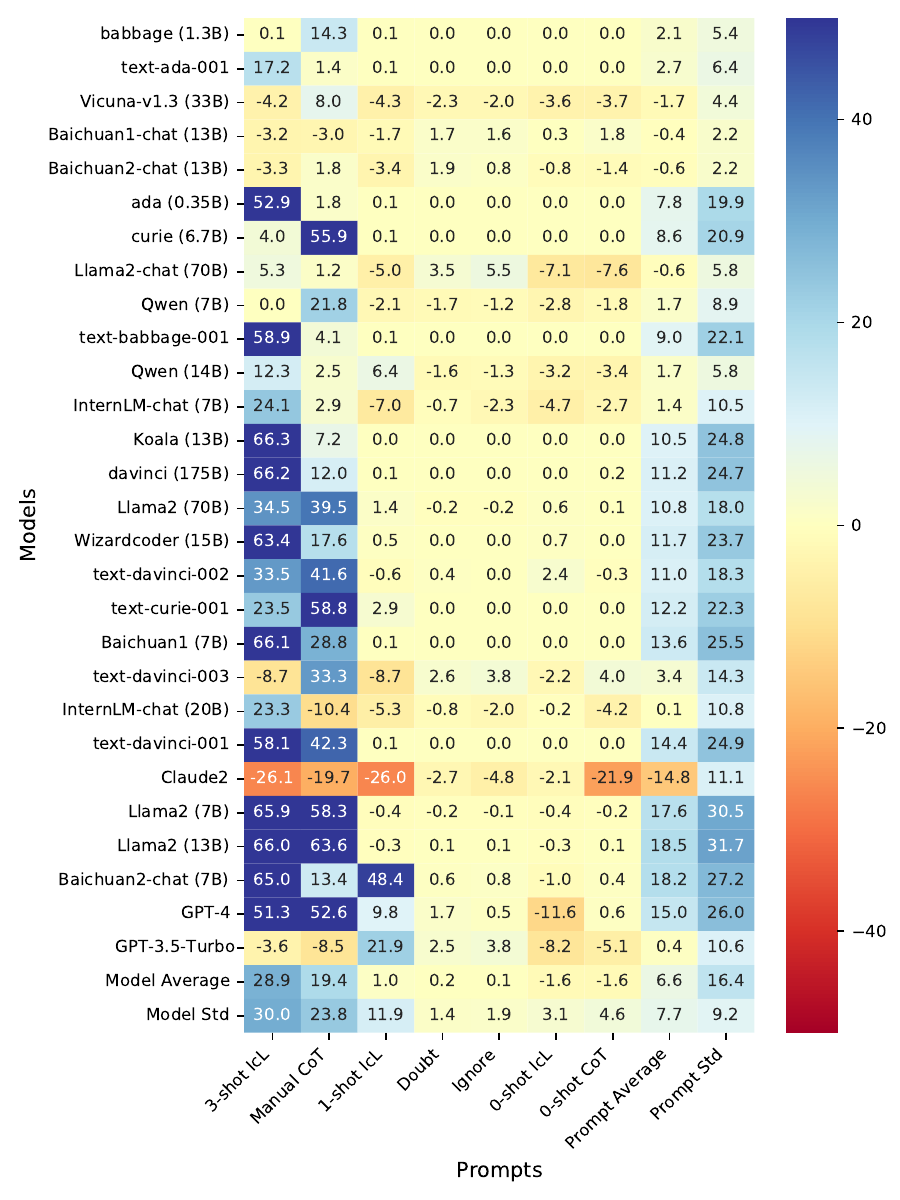}
\end{minipage}
}
\subfigure[\textit{Prompt gain} of NIE-B (NIE-natural)]{
\begin{minipage}{8.5cm}
\centering
\includegraphics[width=.9\linewidth]{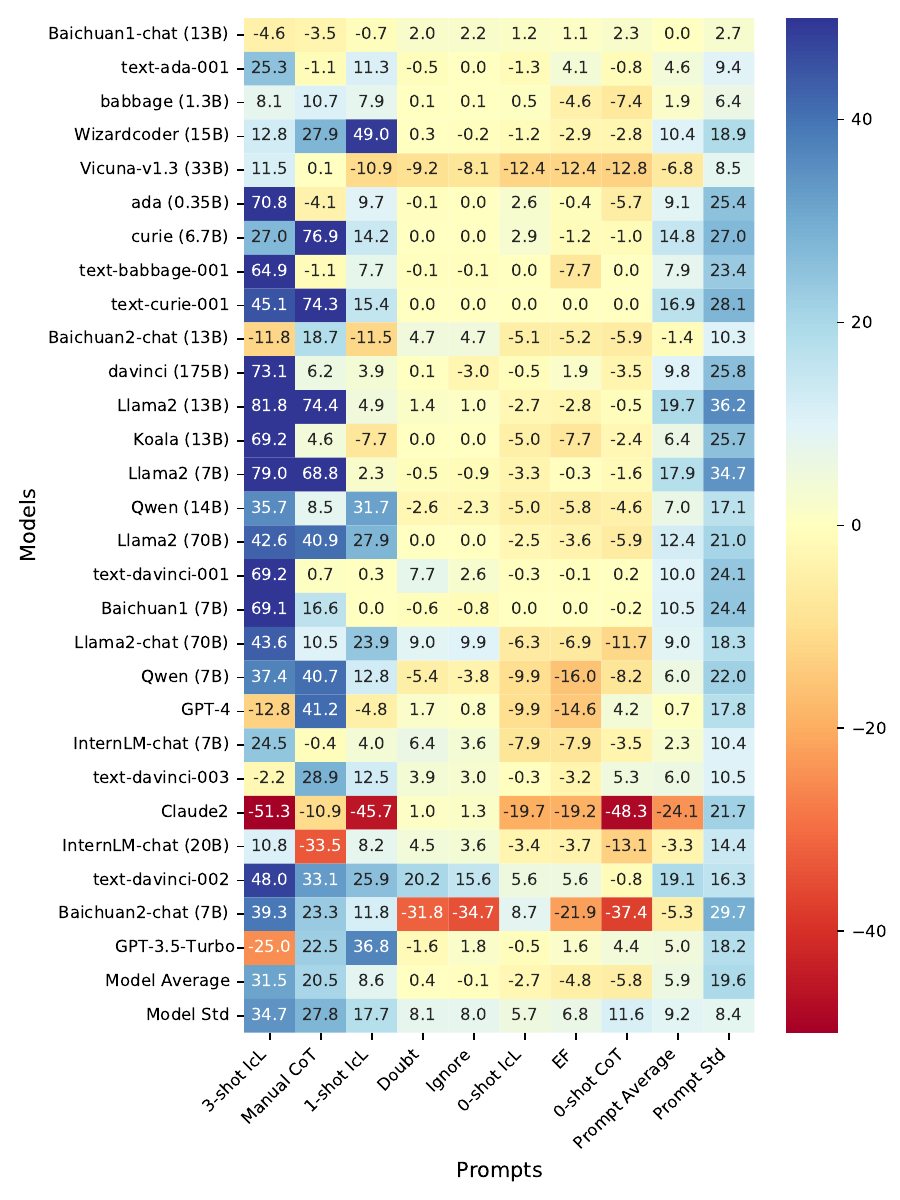}
\end{minipage}
}
\caption[Heatmaps of \textit{prompt gain} of causal tasks in NIE]{\textbf{Heatmaps of \textit{prompt gain} of causal tasks in NIE.} The models and prompts are sorted by their averages.}
\label{fig:Heatmap_of_gain_of_Natural_Indirect_Effect}
\end{figure}

The distribution of models' accuracy on NIE is shown in Figure \ref{fig:Distribution_of_Natural_Indirect_Effect_Tasks}. Figure \ref{fig:Heatmap_of_performances_of_Natural_Indirect_Effect} illustrates how models perform on NIE. The prompt gain (i.e., accuracy improvement against the basic prompt on the model with the used prompt) is demonstrated in Figure \ref{fig:Heatmap_of_gain_of_Natural_Indirect_Effect}.

\subsubsection{PN}
\begin{figure}
\centering
\subfigure[Model performance of PN-P (PN-basic)]{
\begin{minipage}{8.5cm}
\centering
\includegraphics[width=.9\linewidth]{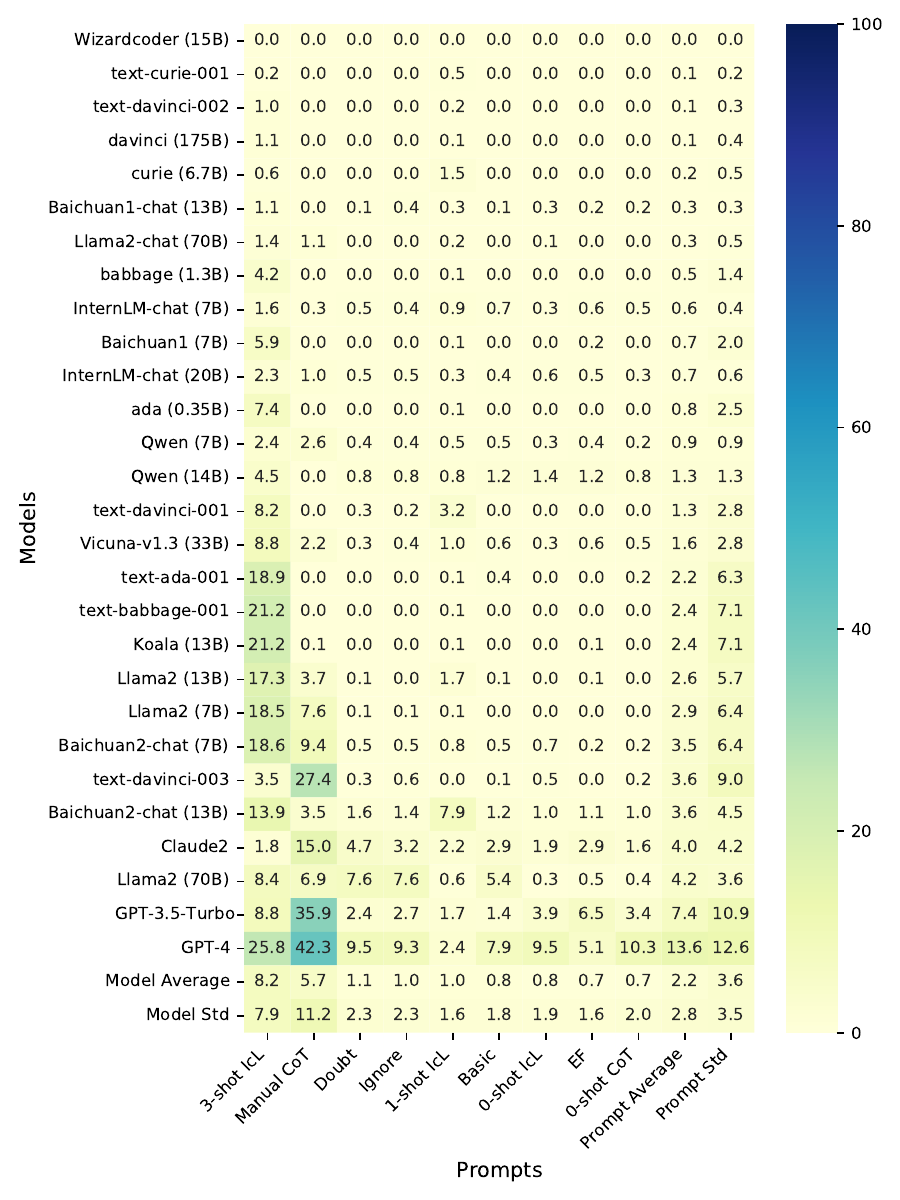}
\end{minipage}
}
\subfigure[Model performance of PN-P (PN-hard)]{
\begin{minipage}{8.5cm}
\centering
\includegraphics[width=.9\linewidth]{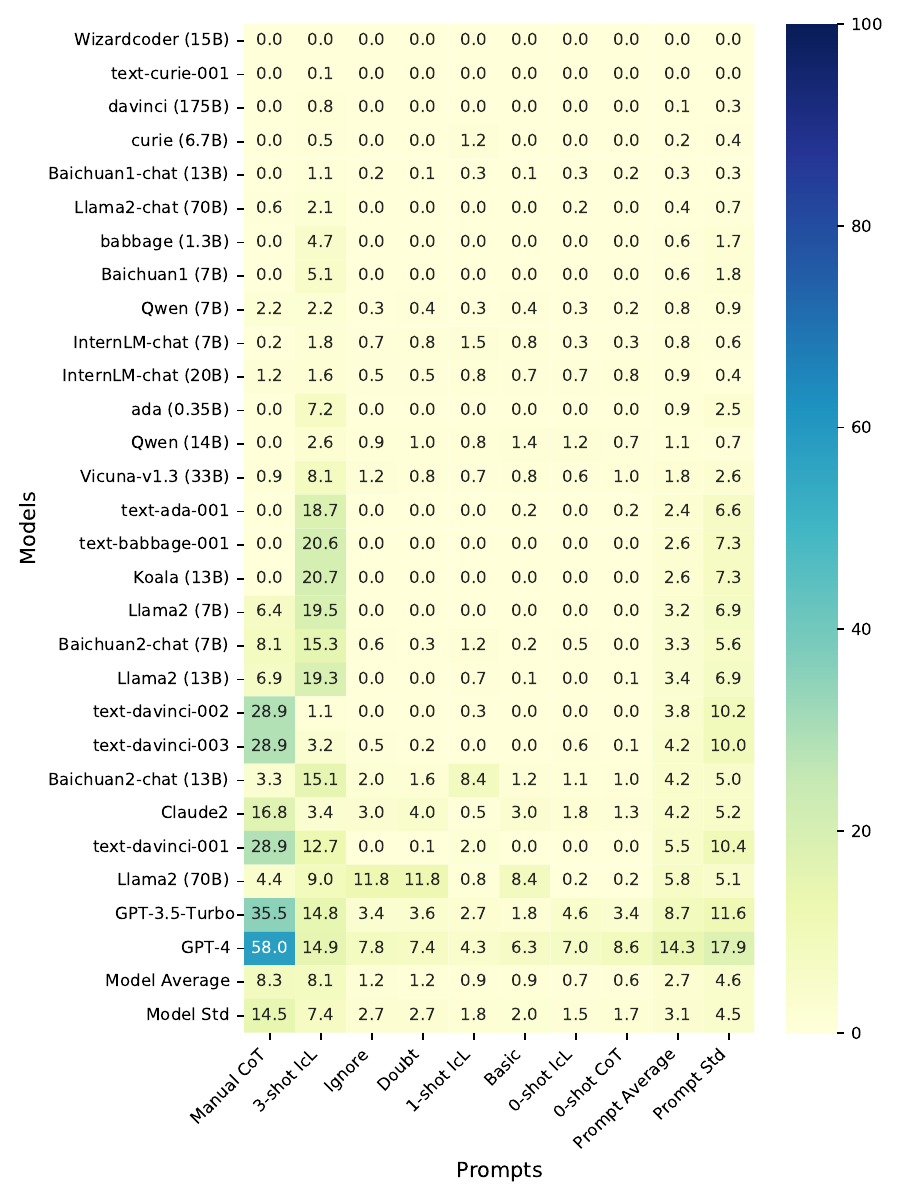}
\end{minipage}
}
\caption[Heatmaps of model performance of causal tasks in PN]{\textbf{Heatmaps of model performance of causal tasks in PN.} The models and prompts are sorted by their averages.}
\label{fig:Heatmap_of_performances_of_Probability_of_Necessity}
\end{figure}

\begin{figure}
\centering
\subfigure[\textit{Prompt gain} of PN-P (PN-basic)]{
\begin{minipage}{8.5cm}
\centering
\includegraphics[width=.9\linewidth]{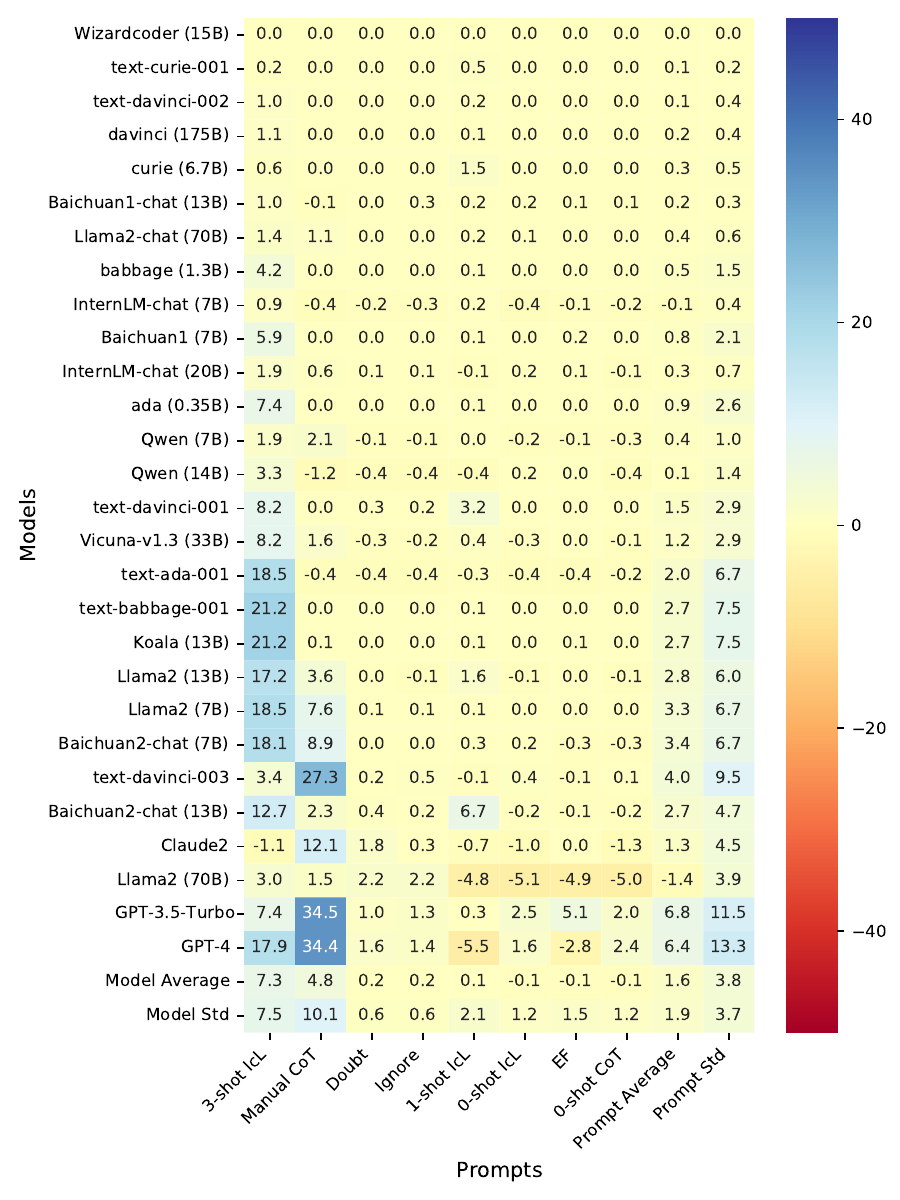}
\end{minipage}
}
\subfigure[\textit{Prompt gain} of PN-P (PN-hard)]{
\begin{minipage}{8.5cm}
\centering
\includegraphics[width=.9\linewidth]{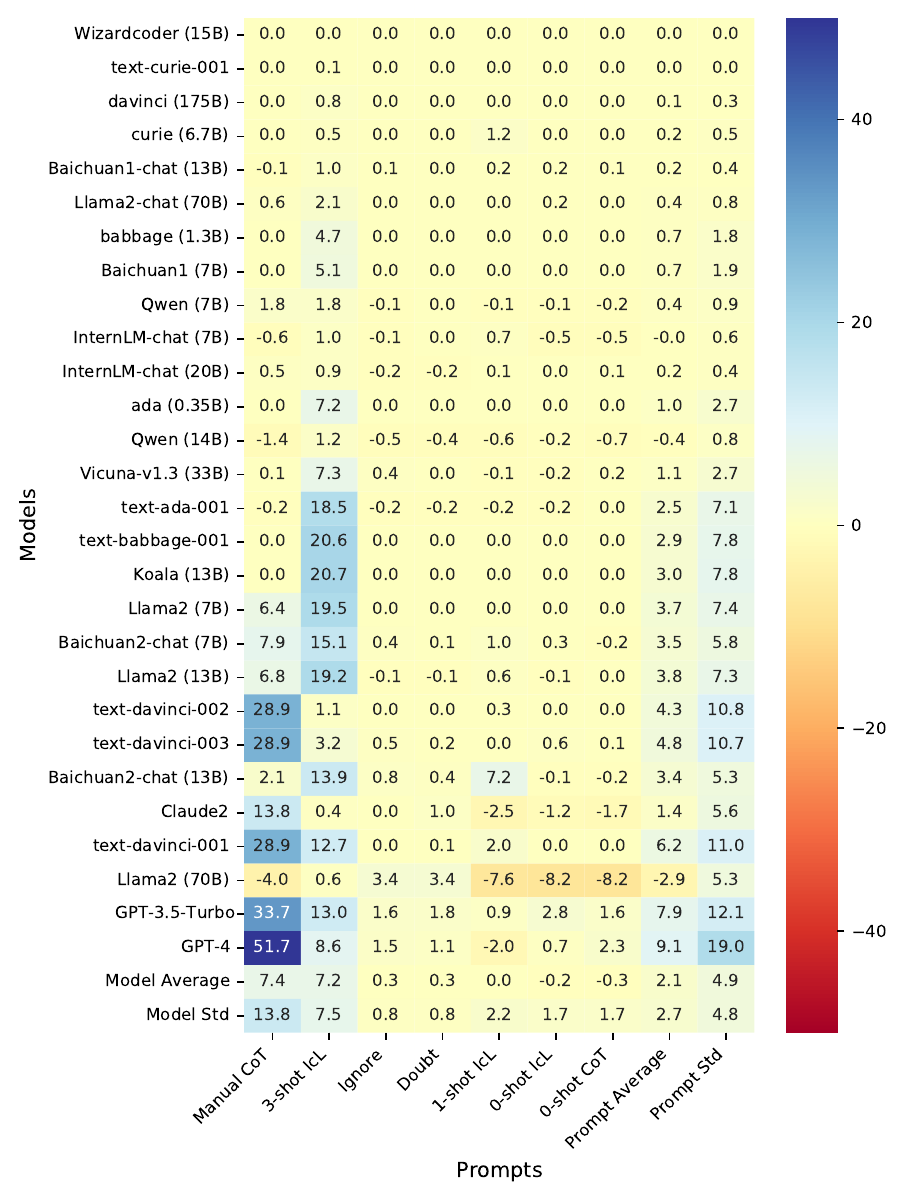}
\end{minipage}
}
\caption[Heatmaps of \textit{prompt gain} of causal tasks in PN]{\textbf{Heatmaps of \textit{prompt gain} of causal tasks in PN.} The models and prompts are sorted by their averages.}
\label{fig:Heatmap_of_gain_of_Probability_of_Necessity}
\end{figure}

The distribution of models' accuracy on PN is shown in Figure \ref{fig:Distribution_of_Probability_of_Necessity_Tasks}. Figure \ref{fig:Heatmap_of_performances_of_Probability_of_Necessity} illustrates how models perform on PN. The prompt gain (i.e., accuracy improvement against the basic prompt on the model with the used prompt) is demonstrated in Figure \ref{fig:Heatmap_of_gain_of_Probability_of_Necessity}.

\subsubsection{PS}
\begin{figure}
\centering
\subfigure[Model performance of PS-P (PS-basic)]{
\begin{minipage}{8.5cm}
\centering
\includegraphics[width=.9\linewidth]{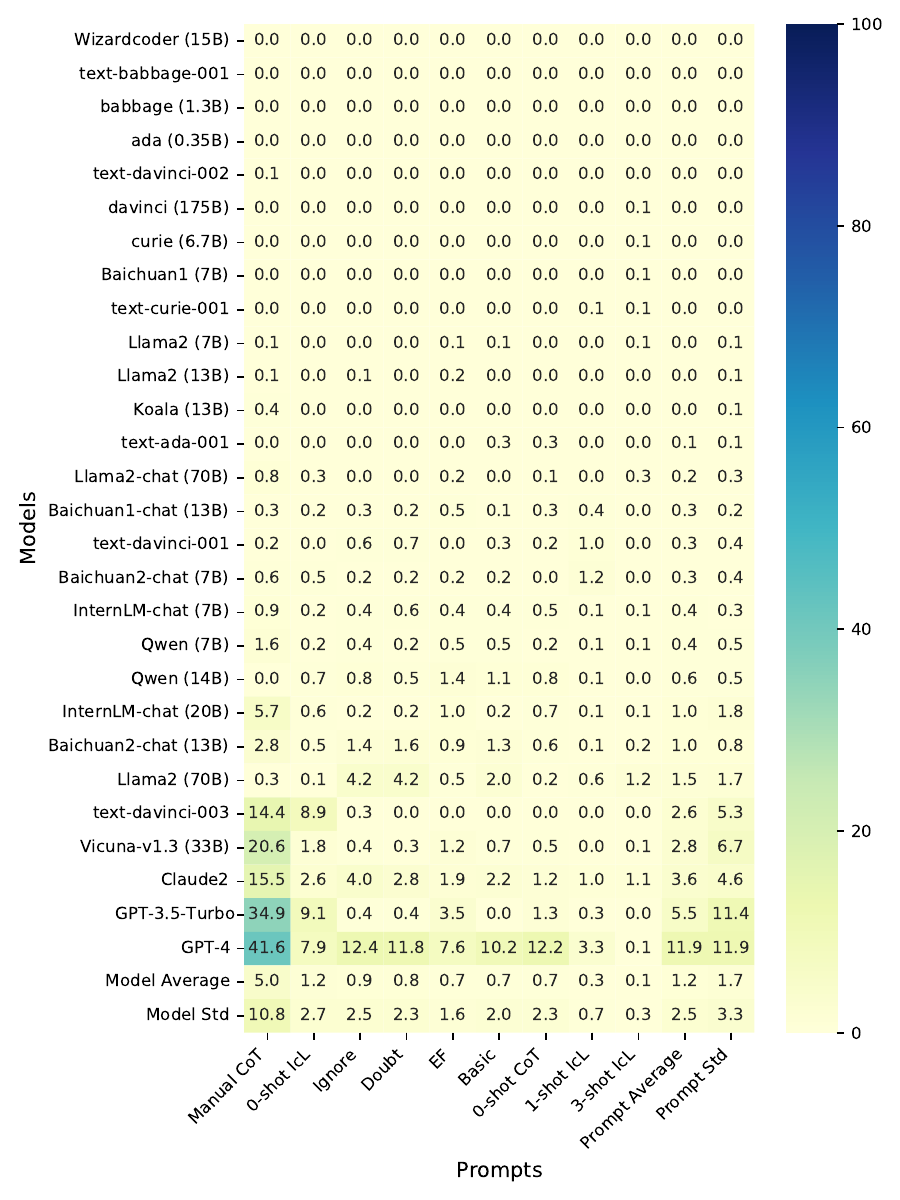}
\end{minipage}
}
\subfigure[Model performance of PS-P (PS-hard)]{
\begin{minipage}{8.5cm}
\centering
\includegraphics[width=.9\linewidth]{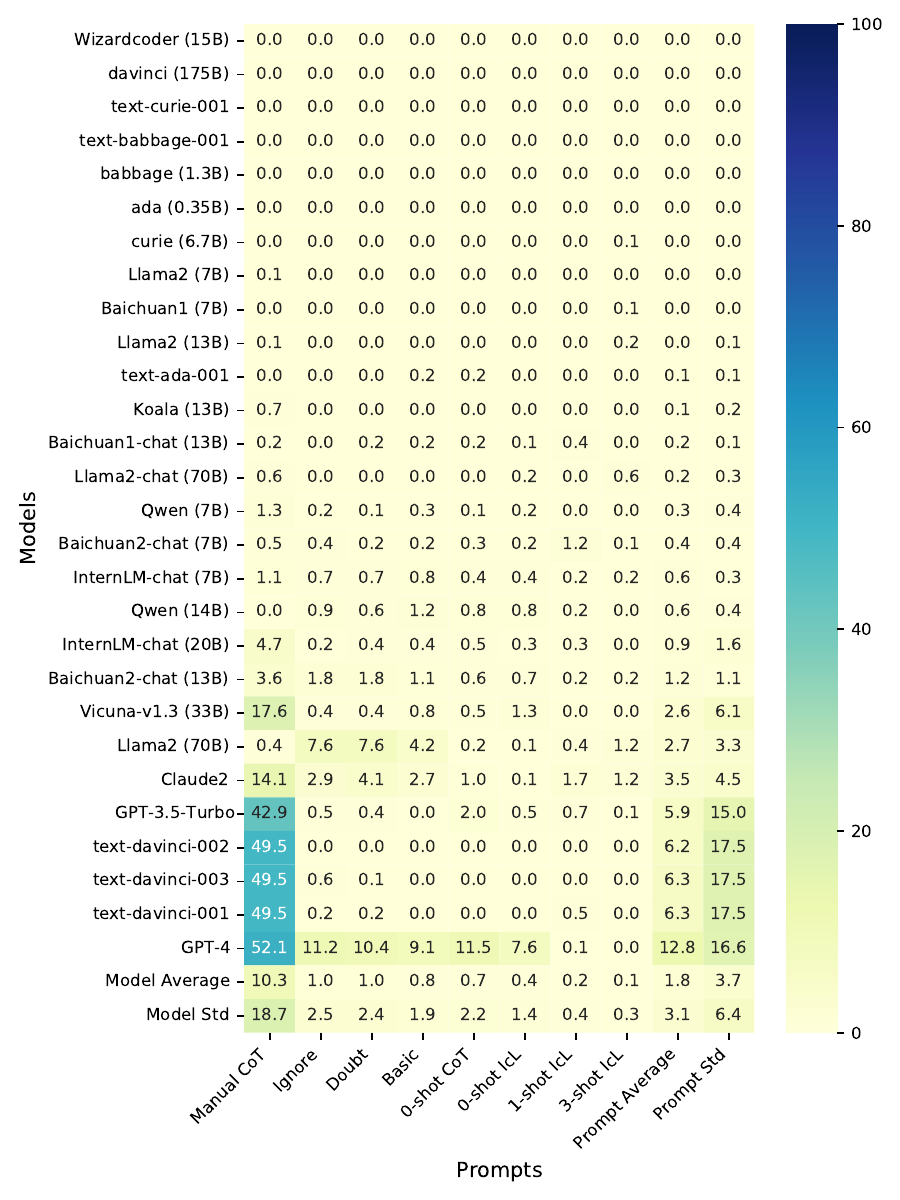}
\end{minipage}
}
\caption[Heatmaps of model performance of causal tasks in PS]{\textbf{Heatmaps of model performance of causal tasks in PS.} The models and prompts are sorted by their averages.}
\label{fig:Heatmap_of_performances_of_Probability_of_Sufficiency}
\end{figure}

\begin{figure}
\centering
\subfigure[\textit{Prompt gain} of PS-P (PS-basic)]{
\begin{minipage}{8.5cm}
\centering
\includegraphics[width=.9\linewidth]{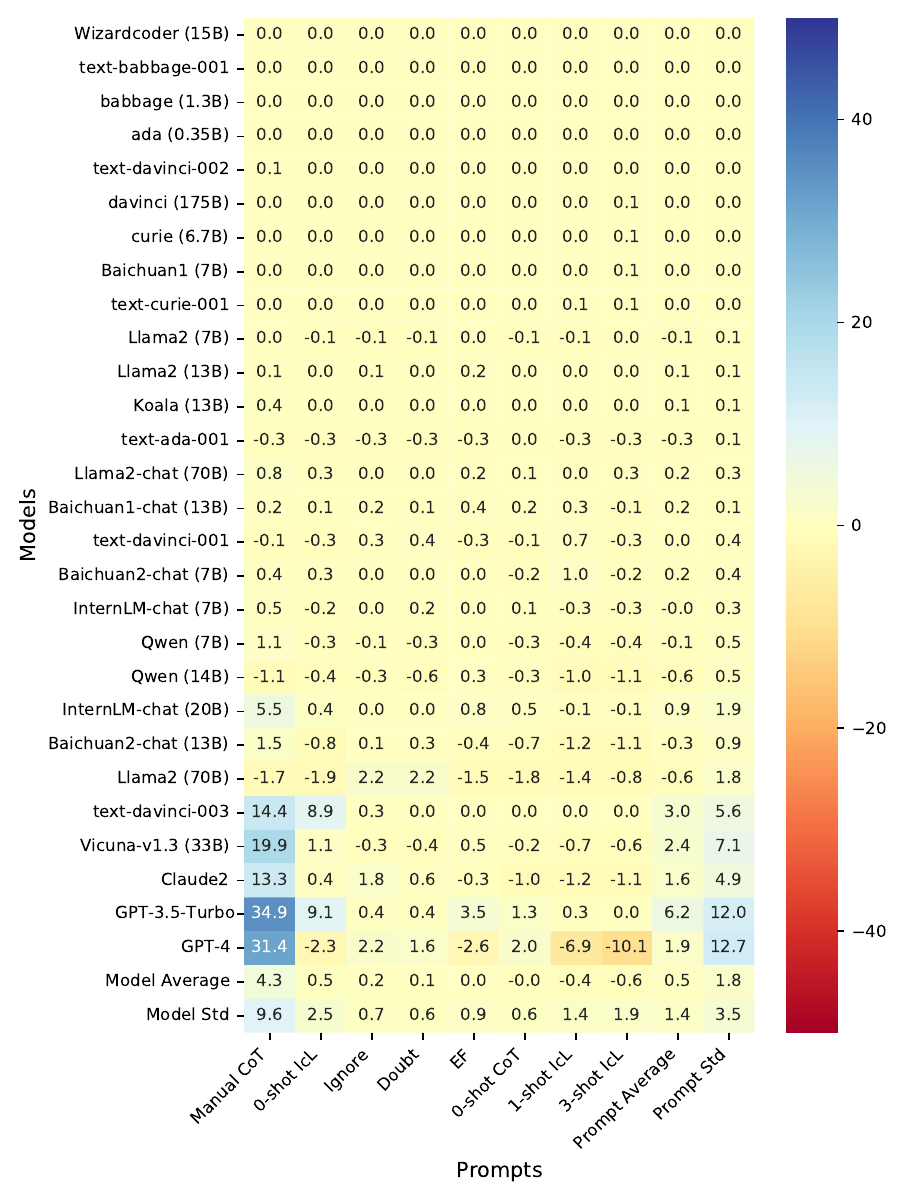}
\end{minipage}
}
\subfigure[\textit{Prompt gain} of PS-P (PS-hard)]{
\begin{minipage}{8.5cm}
\centering
\includegraphics[width=.9\linewidth]{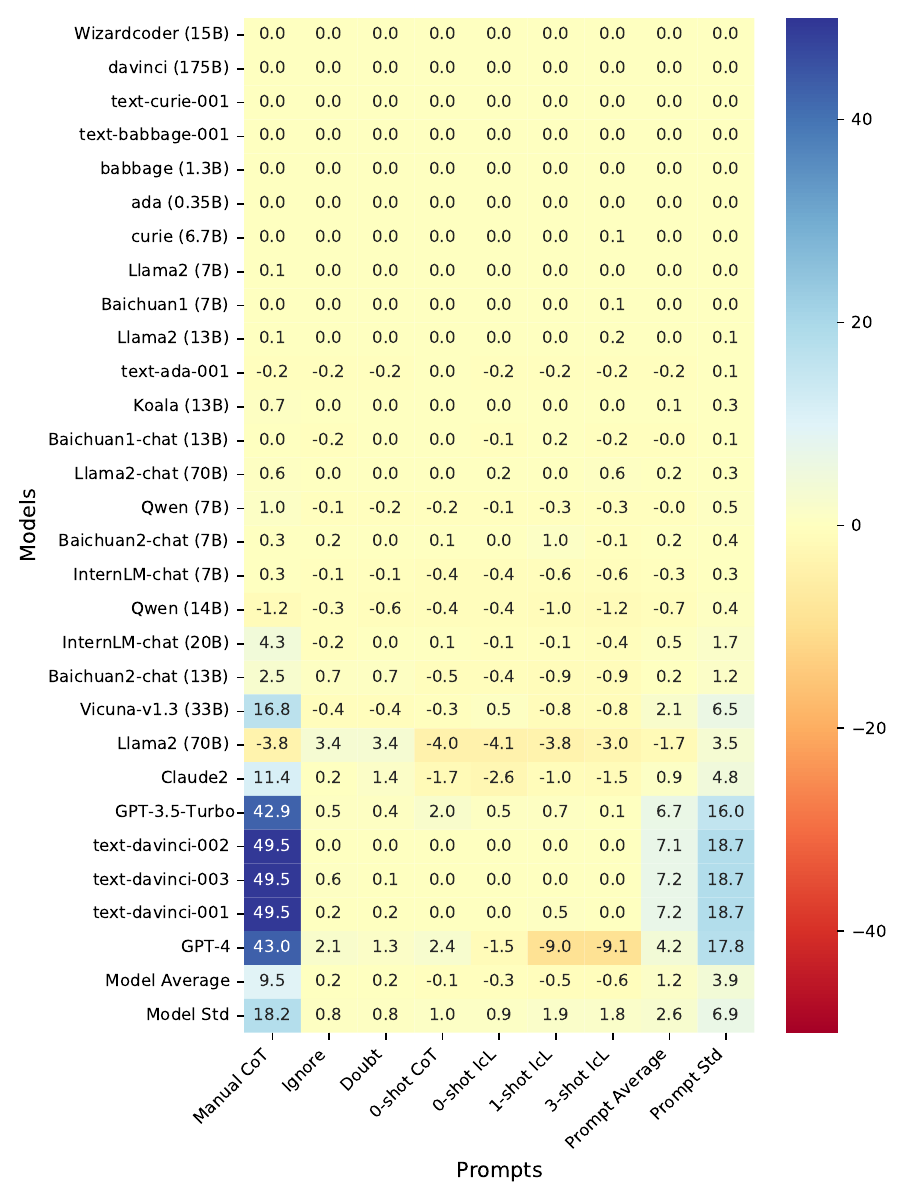}
\end{minipage}
}
\caption[Heatmaps of \textit{prompt gain} of causal tasks in PS]{\textbf{Heatmaps of \textit{prompt gain} of causal tasks in PS.} The models and prompts are sorted by their averages.}
\label{fig:Heatmap_of_gain_of_Probability_of_Sufficiency}
\end{figure}

The distribution of models' accuracy on PS is shown in Figure \ref{fig:Distribution_of_Probability_of_Sufficiency_Tasks}. Figure \ref{fig:Heatmap_of_performances_of_Probability_of_Sufficiency} illustrates how models perform on PS. The prompt gain (i.e., accuracy improvement against the basic prompt on the model with the used prompt) is demonstrated in Figure \ref{fig:Heatmap_of_gain_of_Probability_of_Sufficiency}.

\clearpage

\section{Models}
\label{appendix:models}
\subsection{Limited-access Models}
To ensure the reproducibility of our results, for limited-access models, we report the API version and our evaluation time in \cref{appendix:table_models}.
\begin{center}
\begin{table*}[t]
\fontsize{10}{12}\selectfont
    \caption[API Version and evaluation date of limited-access model]{\textbf{API Version and evaluation date of limited-access model.} API version means the name of the model specified in the API when making a call for inference. The evaluation date refers to the date on which the models are assessed.}
    \label{appendix:table_models}
    \centering
  \begin{tabular}{ l|c|c}
\toprule
\textbf{Model} & \textbf{API Version} & \textbf{Evaluation Date}\\
\hline
ada       & \texttt{ada}   & \multirow{13}{*}{June 2023\textasciitilde December 2023}\\
babbage     & \texttt{babbage}   & \\
curie      & \texttt{curie}   & \\
davinci      & \texttt{davinci}   & \\
text-ada-001      & \texttt{text-ada-001}   & \\
text-babbage-001      & \texttt{text-babbage-001}    & \\
text-curie-001      & \texttt{text-curie-001}    &\\
text-davinci-001      & \texttt{text-davinci-001}    &\\
text-davinci-002      & \texttt{text-davinci-002}    &\\
text-davinci-003      & \texttt{text-davinci-003}    &\\
\chatgpt      & \texttt{gpt-3.5-turbo}    &\\
GPT-4      & \texttt{gpt-4}   &\\
Claude2      & \texttt{claude-2}   &\\
\hline
\end{tabular}
\end{table*}
\end{center}

\clearpage

\end{document}